\documentclass[onecolumn, 12 pt, doublespace, fullpage, a4paper]{report}

\usepackage[utf8]{inputenc} 
\usepackage[T1]{fontenc}    
\usepackage[colorinlistoftodos,bordercolor=orange,backgroundcolor=orange!20,linecolor=orange,textsize=scriptsize]{todonotes}
\usepackage{amsmath, latexsym, amsthm}
\usepackage{mathtools}
\usepackage{url}                    
\usepackage{booktabs}               
\usepackage{amsfonts}               
\usepackage{nicefrac}               
\usepackage{microtype}              
\usepackage{natbib}
\usepackage{comment}
\usepackage{tablefootnote}
\usepackage{graphicx}
\usepackage{cmap}                   
\usepackage{mathtext}
\usepackage{icomma}                 
\usepackage{longtable}              
\usepackage{array}

\usepackage{multirow}               
\usepackage{multicol}               
\usepackage{color}
\usepackage{url}                    
\usepackage{algorithm}
\usepackage{algorithmic}
\usepackage{xspace}
\usepackage{color}
\usepackage{pifont}
\usepackage{float}
\usepackage{subcaption}
\usepackage{mdframed}
\usepackage{enumitem}
\usepackage{latexsym}
\usepackage{etoolbox}\AtBeginEnvironment{algorithmic}{\footnotesize}
\usepackage{tikz}

\usepackage{natbib}
\usepackage[dvipsnames,table]{xcolor}
\usepackage{sectsty}
\usepackage{fancyhdr}
\usepackage[acronym,toc, nogroupskip]{glossaries}
\usepackage{setspace}
\usepackage{lmodern}

\usepackage[lmargin=40mm, rmargin=25mm, vmargin=25mm, headsep=2.5mm]{geometry}

\newcommand{\R}{\mathbb{R}}

\newcounter{savechapter}  
\newcounter{savesection}
\newcounter{savesubsection}
\newcounter{apdxsection}
\newcounter{adjsection}

\renewcommand\appendix{\par
  \setcounter{savechapter}{\value{chapter}}
  \setcounter{savesection}{\value{section}}
  \setcounter{savesubsection}{\value{subsection}}
  \setcounter{apdxsection}{\value{section}}

  \setcounter{adjsection}{0}
  \renewcommand{\thesection}{\Alph{adjsection}\arabic{savechapter}}
  
}

\newcommand\unappendix{\par
  \par
  \setcounter{chapter}{\value{savechapter}} 
  \setcounter{section}{\value{savesection}} 
  \setcounter{subsection}{\value{savesubsection}} 
  \renewcommand{\thesection}{\arabic{chapter}.\arabic{section}} 
}

\def\2{$^2$}			 
\def\3{$^3$}			 
\def\-2{$^{-2}$}		 
\def\-3{$^{-3}$}		 
\def\-1{$^{-1}$}		 

\chapterfont{\fontsize{14}{15}\selectfont}   
\sectionfont{\fontsize{14}{15}\selectfont}
\subsectionfont{\fontsize{14}{15}\selectfont}
\subsubsectionfont{\fontsize{14}{15}\selectfont}

\fancyheadoffset[L]{0.01mm}
\renewcommand{\headrulewidth}{0pt}

\makeatletter
\patchcmd{\@makechapterhead}{50\p@}{20pt}{}{}
\patchcmd{\@makeschapterhead}{50\p@}{20pt}{}{}
\makeatother

\fancypagestyle{plain}{
\fancyhf{} 
\fancyhead[C]{\thepage} 
\renewcommand{\headrulewidth}{0pt}
\renewcommand{\footrulewidth}{0pt}}

\newglossary[slg]{symbols}{syi}{sbl}{List of Symbols}
 
\makeglossaries

\newglossaryentry{symb:Pi}{
name=$\pi$, type=symbols,
description=A mathematical constant whose value is the ratio of any circle's circumference to its diameter,
sort=symbolpi
}

\newglossaryentry{symb:Phi}{
name=$\varphi$, type=symbols,
description=An angle,
sort=symbolphi
}

\newglossaryentry{symb:Lambda}{
name=$\lambda$, type=symbols,
description=Lambda indicates usually an eigenvalue in linear algebra,
sort=symbollambda
}

\newacronym{toc}{ToC}{Table of Contents}
\newacronym{los}{LoS}{List of Symbols}
\newacronym{loa}{LoA}{List of Abbreviations}
\newacronym{phd}{PhD}{Doctoral}
\newacronym{MS}{MS}{Masters}
\newacronym{M$}{MS}{Microsoft}
\newacronym{CD}{CD}{Compact Disc}
\newacronym{kaust}{KAUST}{King Abdullah University of Science and Technology}

\newacronym{AD}{AD}{Active Directory\protect\glsadd{glos:AD}}

\newglossaryentry{glos:AD}{
name=Active Directory,
description={Active Directory is the directory service for Windows-based networks, that allows central organization and administration of any network resource. It allows a single-sign-on concept independent from network topologies or network protocols. As a prerequisite, you need a Windows Server acting as a Domain Controller. This computer stores all necessary data, e.g.~usernames, and corresponding passwords}
}

\newglossaryentry{glos:RespF}{name=response file, description={A file 
that allows unattended software installation}}

\newcommand{\mathsym}[1]{{}}
\newcommand{\unicode}[1]{{}}
\renewcommand{\thechapter}{\arabic{chapter}}
\renewcommand\bibname{\centering BIBLIOGRAPHY}
\newcommand{\orcid}{\includegraphics[width=8pt]{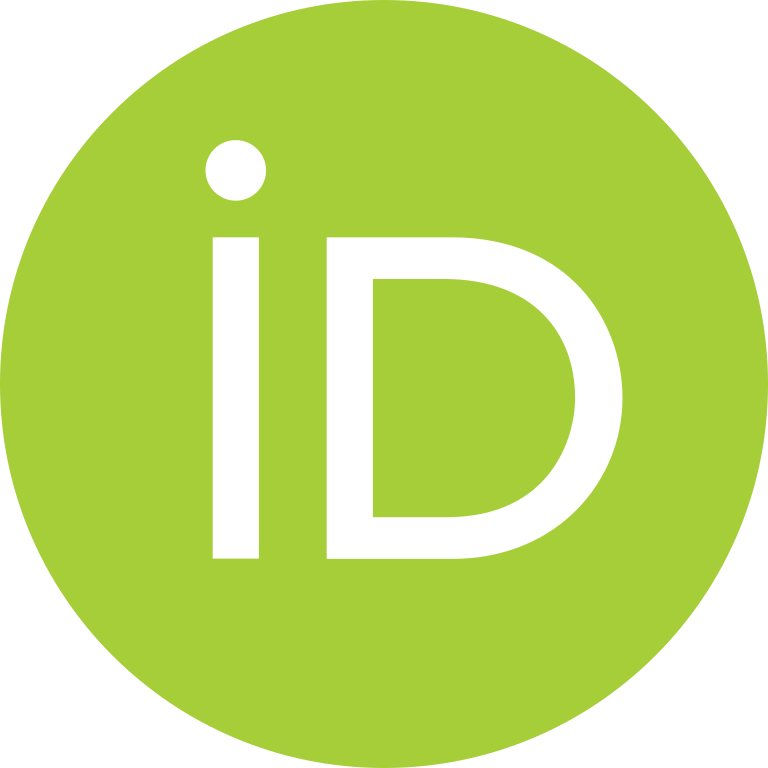}} 

\newcommand{\norm}[1]{\left\| #1 \right\|}
\newcommand{\lin}[1]{\left\langle #1\right\rangle} 

\newcommand{\cD}{\mathcal{D}}

\newcommand{\mI}{\mathbf{I}}

\newcommand{\mQ}{\mathbf{Q}}
\newcommand{\mX}{\mathbf{X}}

\newcommand{\ExpBr}[1]{\mathbb{E}\left[#1\right]}
\newcommand{\ExpBrBig}[1]{\mathbb{E}\Big[#1\Big]}

\newcommand{\cL}{\mathcal{L}}
\newcommand{\cZ}{\mathcal{Z}}

\newcommand{\cJ}{\mathcal{J}}

\newcommand{\sumin}{\sum_{i=1}^n}

\newcommand{\sumjn}{\sum_{j=1}^n}

\newcommand{\avein}{\frac{1}{n}\sum_{i=1}^n}

\newcommand{\RR}{\mathbb{R}}
\newcommand{\cC}{\mathcal{C}}
\newcommand{\cO}{\mathcal{O}}

\newcommand{\bA}{\mathbf{A}}

\newcommand{\bM}{\mathbf{M}}

\newcommand{\del}[1]{}

\newcommand{\inp}[2]{\left\langle #1 ,  #2 \right\rangle}

\newcommand{\eqdef}{\stackrel{\text{def}}{=}}

\newcommand{\Prob}{\mathbf{Prob}}
\newcommand{\E}{\mathbf{E}}
\newcommand{\Exp}[1]{\mathbb{E}\left[#1\right]}

\newcommand{\RD}{\mathbb{R}^d}
\newcommand{\RS}{\mathbb{R}}

\newcommand{\ZS}{\mathbb{Z}}

\def\lfk{\left\lfloor}
\def\rfk{\right\rfloor}

\definecolor{linenumbercolor}{rgb}{0.98, 0.81, 0.69}

\definecolor{classcolor}{RGB}{0,0,255}

\definecolor{apicolor}{RGB}{255,0,0}

\definecolor{linenumbercolor}{rgb}{0.1, 0.1, 0.1}

\newcommand{\cmark}{{\color{PineGreen}\ding{51}}}%
\newcommand{\xmark}{{\color{BrickRed}\ding{55}}}%

\definecolor{modelcolor}{RGB}{108,57,0}
\definecolor{datacolor}{RGB}{108,57,0}
\definecolor{abrcolor}{RGB}{108,57,0}
\definecolor{algcolor}{RGB}{108,57,0}
\definecolor{aescolor}{RGB}{0,0,255}
\definecolor{algnamewithaescolor}{RGB}{0,0,255}
\definecolor{encryptcolor}{RGB}{108,57,0}
\definecolor{attackcolor}{RGB}{108,57,0}
\definecolor{compcolor}{RGB}{108,57,0}
\definecolor{libcolor}{RGB}{108,57,0}

\newcommand{\modelname}[1]{{\color{modelcolor}\sf \small {#1}}} 
\newcommand{\dataname}[1]{{\color{datacolor}\sf \small {#1}}}  
\newcommand{\abr}[1]{{\color{abrcolor}\sf \small {#1}}}       
\newcommand{\algname}[1]{{\color{algcolor}\sf \small {#1}}}   
\newcommand{\aesname}[1]{{\color{aescolor}\sf \small {#1}}}   
\newcommand{\algnamewithaes}[1]{{\color{algnamewithaescolor}\sf \small {#1}}} 
\newcommand{\ecryptname}[1]{{\color{encryptcolor}\sf \small {#1}}} 
\newcommand{\attackname}[1]{{\color{attackcolor}\sf \small {#1}}} 
\newcommand{\compname}[1]{{\color{compcolor}\sf \small {#1}}}   
\newcommand{\libname}[1]{{\sf \color{libcolor} \small {#1}}}    

\newcommand{\algnamesmall}[1]{{\color{algcolor}\sf \scriptsize {#1}}}

\newcommand{\cI}{{\cal I}}

\newcommand{\myred}[1]{{\color{red} #1}}

\newcommand{\myblue}[1]{{\color{blue} #1}}

\newcommand{\dotprod}[2]{\left\langle #1,#2\right\rangle} 
\newcommand{\lp}{\left(}
\newcommand{\rp}{\right)}
\DeclareMathOperator*{\argmin}{argmin}

\usepackage[colorinlistoftodos,bordercolor=orange,backgroundcolor=orange!20,linecolor=orange,textsize=scriptsize]{todonotes}

\newcommand{\abdulmajeed}[1]{\todo[inline]{\textbf{Abdulmajeed: } #1}}
\newcommand{\fahad}[1]{\todo[inline]{\textbf{Fahad: } #1}}
 
\newcommand{\kostya}[1]{\todo[inline]{\textbf{Kostya: } #1}}

\newcommand{\pr}[1]{{\color{red} {\bf Peter:} #1}}
\newcommand{\ks}[1]{{\color{blue} {\bf Kostya:} #1}}
\newtheorem{assumption}{Assumption}
\newtheorem{lemma}{Lemma}

\newtheorem{theorem}{Theorem}

\newtheorem{example}{Example}

\theoremstyle{plain}

\newtheorem{remark}[theorem]{Remark}

\theoremstyle{definition}

\newtheorem{definition}[theorem]{Definition} 

\usepackage{bm}

\newcommand{\serveropt}{\textsc{ServerOpt}\xspace}
\newcommand{\clientopt}{\textsc{ClientOpt}\xspace}

\newcommand{\ServerGlobalState}{\textsc{ServerGlobalState}\xspace}
\newcommand{\ServerGradient}{\textsc{ServerGradient}\xspace}
\newcommand{\ClientState}{\textsc{ClientState}\xspace}
\newcommand{\LocalGradient}{\textsc{LocalGradient}\xspace}
\newcommand{\InitializeServerState}{\textsc{InitializeServerState}\xspace}

\newcommand{\fl}{\libname{FL\_PyTorch}\,}

\newcommand{\obj}{F}

\newcommand{\numClients}{\ensuremath{M}}

\newcommand{\localStep}{\tau}

\newcommand{\data}{\ensuremath{\mathcal{D}}}
\newcommand{\clientDist}{\ensuremath{\mathcal{P}}}

\newcommand{\activeClients}{\mathcal{S}}
\newcommand{\sgrad}{g}

\newcommand{\localChange}{\Delta}

\newcommand{\lr}{\eta}
\newcommand{\slr}{\lr_{s}}

\newcommand{\vxflp}{\bm{x}}

\newcommand{\LAM}{{\color{blue}L_{\rm AM}}}
\newcommand{\LQM}{{\color{red}L_{\rm QM}}}
\newcommand{\Lvar}{L_{\rm var}}

\newcommand{\LAMsq}{{\color{blue}L_{\rm AM}^2}}
\newcommand{\LQMsq}{{\color{red}L_{\rm QM}^2}}

\newcommand{\sqnorm}[1]{\left\| #1 \right\|^{2}}
\newcommand{\rb}[1]{\left(#1\right)}

\usepackage{xspace}
\newcommand{\algnametiny}[1]{{\color{ForestGreen}\tiny\sf#1}\xspace}

\usepackage[flushleft]{threeparttable}
\usepackage{apptools}
\usepackage{thmtools}
\usepackage{siunitx}

\usepackage{amsmath}
\usepackage{amsthm}
\usepackage{amssymb}
\usepackage{amsfonts}
\usepackage{mathtools}

\newcommand{\samplefunc}{{\bf S}}
\newcommand{\samplingname}[1]{{\textit{#1}}}
\newcommand{\normsq}[1]{\left\| #1 \right\|^2}
\newcommand{\inner}[2]{\left< #1 , #2 \right>}

\newcommand{\ExpSub}[2]{{\rm E}_{#1}\left[#2\right]}

\newcommand{\mA}{\mathbf{A}}

\newcommand{\smartparagraph}[1]{\vspace{2pt} \noindent {\bf #1}}
\newcommand{\myNum}[1]{(\emph{#1})}

\usepackage{amsthm}

\usepackage{layout}

\usepackage{amsfonts}
\usepackage{amsmath}
\usepackage{amsthm}
\usepackage{amssymb} 

\usepackage{mdframed} 
\usepackage{thmtools}

\newtheorem*{theorem*}{Theorem}
\newtheorem*{proposition*}{Proposition}
\newtheorem*{lemma*}{Lemma}

\usepackage{amsmath}
\usepackage{mathtools}
\usepackage{amsthm}
\usepackage{algorithm}
\usepackage{algorithmic}
\usepackage{csquotes}
\usepackage{longtable}
\usepackage[flushleft]{threeparttable}
\usepackage{colortbl}
\usepackage{xcolor}
\usepackage{makecell}
\usepackage{nicefrac}
\usepackage{enumitem}
\usepackage{microtype}
\usepackage{graphicx}
\usepackage{makecell}
\usepackage{nicefrac}
\usepackage{enumitem}
\usepackage{graphicx}

\usepackage{csquotes}
\usepackage{longtable}
\usepackage[flushleft]{threeparttable}
\usepackage{colortbl}
\usepackage{xcolor}

\usepackage{makecell}
\usepackage{nicefrac}
\usepackage{enumitem}
\usepackage{microtype}
\usepackage{graphicx}

\usepackage{xcolor}

\newcommand{\HS}{L_{*}}      
\newcommand{\HF}{L_{\rm F}}  
\newcommand{\HM}{L_{\infty}} 
\newcommand{\clr}[1]{#1}
\newcommand{\clrshort}[1]{#1}

\newcommand{\mH}{\mathbf{H}}
\newcommand{\mS}{\mathbf{S}}

\definecolor{bgcolorwe}{rgb}{0.8,1,0.8}
\definecolor{myblue}{rgb}{0.1, 0.3, 0.7}

\definecolor{abscolor}{rgb}{0.501,0.521,0.533}

\usepackage{csquotes}
\usepackage{longtable}
\usepackage[flushleft]{threeparttable}
\usepackage{colortbl}
\usepackage{xcolor}
\usepackage{makecell}
\usepackage{nicefrac}
\usepackage{enumitem}
\usepackage{microtype}
\usepackage{graphicx}
\usepackage{multicol}
\usepackage{listings}

\lstdefinestyle{mystyle}{
    backgroundcolor=\color{white},   
    commentstyle=\color{darkgray},      
    keywordstyle=\color{magenta},    
    numberstyle=\tiny\color{gray},   
    stringstyle=\color{purple},      
    basicstyle=\ttfamily\footnotesize, 
    breakatwhitespace=false,         
    breaklines=true,                 
    captionpos=t,                    
    numbers=left,                    
    numbersep=5pt,                   
    showstringspaces=false           
}
\usepackage{hyperref}               
\definecolor{mydarkblue}{rgb}{0,0.0,1.0}
\hypersetup{ %
	colorlinks=true,
	linkcolor=mydarkblue,
	citecolor=mydarkblue,
	filecolor=mydarkblue,
	urlcolor=mydarkblue,
}

\begin{document}


\thispagestyle{empty}
\addvspace{5mm}  


\begin{center}
\begin{doublespace}
{\textbf{{\large Optimization Methods and Software \\for Federated Learning}}}
\end{doublespace}

{Dissertation by}\\
{Konstantin Burlachenko} 


{ In Partial Fulfillment of the Requirements}\\[12pt]
{ For the Degree of}\\[12pt]
{Doctor of Philosophy} \vfill
{King Abdullah University of Science and Technology }\\
{Thuwal, Kingdom of Saudi Arabia}
\vfill

\begin{onehalfspace}
{\copyright March, 2025}\\
Konstantin Burlachenko\\               
All rights reserved\\

\orcid{} \small \href{https://orcid.org/0000-0001-5986-0855}{https://orcid.org/0000-0001-5986-0855}\\
\end{onehalfspace}

\end{center}
\newpage


%
\chaptertitlefont{\fontsize{14}{15}\selectfont\centering}

\begin{center}

\end{center}

\begin{center}
{{\bf\fontsize{14pt}{14.5pt}\selectfont \uppercase{ABSTRACT}}}
\end{center}

\singlespacing
\addcontentsline{toc}{chapter}{Abstract}

\begin{center}
	\begin{doublespace}
{\fontsize{14pt}{14.5pt}\selectfont {Optimization Methods and Software \\for Federated Learning}}\\
		{\fontsize{14pt}{14.5pt}\selectfont {Konstantin Burlachenko}}\\
	\end{doublespace}
\end{center}
Federated Learning (FL) is a novel, multidisciplinary Machine Learning paradigm where multiple clients, such as mobile devices, collaborate to solve machine learning problems. Initially introduced in \citet{konevcny2016afederated, FEDLEARN, mcmahan17fedavg}, FL has gained further attention through its inclusion in the National AI Research and Development Strategic Plan (2023 Update) of the United States \citep{national2019national}. The FL training process is inherently decentralized and often takes place in less controlled settings compared to data centers, posing unique challenges distinct from those in fully controlled environments.

In this thesis, we identify five key challenges in Federated Learning and propose novel approaches to address them. These challenges arise from the heterogeneity of data and devices, communication issues, and privacy concerns for clients in FL training. Moreover, even well-established theoretical advances in FL require diverse forms of practical implementation to enhance their real-world applicability.

Our contributions advance FL algorithms and systems, bridging theoretical advancements and practical implementations. More broadly, our work serves as a guide for researchers navigating the complexities of translating theoretical methods into efficient real-world implementations and software. Additionally, it offers insights into the reverse process of adapting practical implementation aspects back into theoretical algorithm design. This reverse process is particularly intriguing, as the practical perspective compels us to examine the underlying mechanics and flexibilities of algorithms more deeply, often uncovering new dimensions of the algorithms under study.



\begin{center}

\end{center}
\begin{center}

{\bf\fontsize{14pt}{14.5pt}\selectfont \uppercase{Acknowledgements}}\\\vspace{1cm}
\end{center}

\addcontentsline{toc}{chapter}{Acknowledgements} 

Above all, I would like to express my deepest gratitude to my advisor, Peter Richt\'{a}rik, for his unwavering guidance throughout my PhD. Our discussions, spanning various forms and research projects, have shaped my research interests and helped me navigate the path toward becoming a researcher who can seamlessly combine mathematical rigor with aspects of real-world computation and communication systems. His mentorship has resulted in many papers that I am immensely proud of. Most importantly, Peter Richt\'{a}rik has consistently encouraged me to engage with diverse scientific communities at KAUST and beyond, fostering collaborations with researchers worldwide.

I am also profoundly grateful to my coauthors for their contributions in making this thesis possible. My achievement would not have been possible without their support. Therefore, I would like to thank, in particular, those with whom I have worked closely: Abdulmajeed Alrowithi, Abdurakhmon Sadiev, Ahmed Khaled, Alexander Tyurin, Aritra Dutta, Eduard Gorbunov, El Houcine Bergou, Elnur Gasanov, Fahad Ali Albalawi, Grigory Malinovsky, Haoyu Zhao, Igor Sokolov, Ivan Ilin, Kai Yi, Laurent Condat, Lukang Sun, Samuel Horv\'{a}th, and Zhize Li.

Beyond my direct coauthors, I have also been fortunate to engage in discussions with many individuals who have shaped my research interests: Avetik Karagulyan, Artavazd Maranjyan, Dmitry Kovalev, Egor Shulgin, Mher Safaryan, Filip Hanzely, Kaja Gruntkowska, Konstantin Mishchenko, 
Oleg Ovcharenko, Rustem Islamov, Steven Diamond, Slavom\'{i}r Hanzely, and Yury Demidovich.

Special thanks to Elnur Gasanov, with whom I have had the pleasure of collaborating on several projects. Together, we have supported each other through challenging moments.

I also want to thank the professors and teaching assistants whose courses have contributed to my academic growth and developing a constructively critical mindset. In the long run, this has enabled me to re-examine established ideas from new perspectives and generate new ones.

In addition, I would like to thank the following individuals for considering the potential integration of our research into industrial processes: Samy Bengio (Apple), Filip Granqvist (Apple), Eeshan Dhekane (Apple), Mona Chitnis (Apple), Holger R. Roth (NVIDIA), Hao-Jun Michael Shi (Meta), Maxim Naumov (Meta), Timour Paltashev (AMD), David Pugh (SDAIA-KAUST AI), Fawaz S. Al Qahtani (SDAIA), and Mona Alshahrani (ARAMCO).

Finally, I would not have been able to begin my PhD journey without the support of my family.

\begin{onehalfspacing}

\tableofcontents
\cleardoublepage

\printglossary[type=\acronymtype,style=long3col, title=\centerline{LIST OF ABBREVIATIONS}, toctitle=List of Abbreviations, nonumberlist=true] 

\printglossary[type=symbols,style=long3col, title=\centerline{LIST OF SYMBOLS}, toctitle=List of Symbols, nonumberlist=true]

\cleardoublepage
\phantomsection
\addcontentsline{toc}{chapter}{\listfigurename} 
\renewcommand*\listfigurename{LIST OF FIGURES}
\listoffigures

\cleardoublepage
\phantomsection
\addcontentsline{toc}{chapter}{\listtablename}  
\renewcommand*\listtablename{LIST OF TABLES}
\listoftables

\end{onehalfspacing}


\chaptertitlefont{\fontsize{14}{15}\selectfont}  

\chapter{Introduction}
\label{chapter1}

\paragraph{Artificial Intelligence and Machine Learning.}

Over the past few decades, advancements in Artificial Intelligence (AI) have significantly transformed various systems and improved many aspects of our lives. The mathematical techniques underpinning AI have deep roots in other disciplines: Principal Component Analysis (PCA) stems from singular value decomposition in linear algebra \citep{moore1920reciprocal}, Linear Regression dates back to Gauss \citep{gauss1995theory}, Logistic Regression emerged from statistics \citep{berkson1944application}, Classical Feed-Forward Neural Networks were inspired by psychological theories \citep{rosenblatt1958perceptron}, and Decision Trees originated in statistics \citep{belson1959matching}. While these contributions are deserving of recognition, AI has evolved into a unifying field that integrates and connects these underlying mathematical models. Before AI's rise, these methods were often studied in isolation within their respective disciplines, limiting their rapid dissemination. Today, AI is recognized as an interdisciplinary field that encompasses a broad range of mathematical concepts, including Unsupervised Learning, State-Based Models, Reinforcement Learning and Adaptive Control, Game Theory, Logic-Based Models, Bayesian Networks, and widely-used Machine Learning Supervised Models. AI has become ubiquitous in scenarios where models must be tuned based on data, rather than relying solely on first principles. While first principles offer universal laws, deriving these principles outside physics in the relatively short term can be challenging. It is increasingly evident that, alongside established engineering fields such as Chemical, Electrical, Computer, and Mechanical Engineering, we are witnessing the emergence of a new discipline: AI Engineering. This field is fundamentally shaped by advances in Machine Learning (ML), where the complexity of modeling and explicit algorithm construction is increasingly shifted to data itself. The vision for this phenomenon was also articulated by Michael I. Jordan during his research visit to King Abdullah University of Science and Technology (KAUST) in April 2024\footnote{\href{https://cemse.kaust.edu.sa/events/by-type/lecture/2024/04/24/distribution-free-risk-control}{Michael I. Jordan, KAUST, 24 April 2024: On Distribution-Free Risk Control.}}.

One prominent example of ML in action is its role in search engines, which help users navigate the vast expanse of the Internet. The Internet itself is one of the most complex systems ever built by humankind, as noted by \citet{kurose2005computer}. Companies like Google, Baidu, Yandex, and Microsoft provide web services that enable users to search for documents across the internet. From a user's perspective, these systems take a search query and user information as input. Then they compute a relevance score for each document, and then rank and present the results based on relevance. However, the interaction does not end once users receive the search results. By clicking on links, refining queries, or spending time on specific pages, users implicitly contribute to the training of future ML models. These interactions serve as feedback, generating continuous data that aids in refining the underlying training algorithms, even though these processes remain hidden from the user. This phenomenon illustrates a form of implicit training data generation. Another example of data generation occurs in simulation-based training, where a simulator produces training data. This approach is useful when the forward problem describing the system's behavior under certain conditions is well-understood and can be simulated, while the inverse problem may be complex and becomes the focus of the learning process.
\paragraph{User data.} A significant source of data comes from user-generated content, whether from companies or individuals. However, using such data \textit{naively} for training purposes raises serious privacy concerns \citep{singel2010netflix,kosinski2013private,rocher2019estimating,carlini2019secret}. The challenge arises from the potential inclusion of sensitive or proprietary information in the training data, which could be inadvertently revealed during the training process. Furthermore, once a model is trained, it is crucial to ensure that private or confidential data is not exposed. Beyond privacy concerns, the data locality paradigm is becoming a crucial element in modern Machine Learning~\citep{qiu2021first}. While data locality has been extensively explored in fields such as Computer Architecture, Data Mining, and Data-Intensive Computing, its relevance to ML has only recently obtained attention. The form of data locality we focus on can be motivated by the following question: 
\begin{center}
\textit{"If data is generated and stored on the devices that produce it, is transferring this data to a centralized location truly necessary?"}
\end{center}
While certain data is critical and must not be lost, machine learning applications typically aim to construct approximate predictive models. In this context, the strict requirements of traditional Database Management Systems, which ensure data consistency, can often be relaxed in the case of ML.

Next, we highlight a key limitation of traditional classical ML: its dependence on a single task and a single data distribution. In real-world scenarios, a model trained for one task may not transfer effectively to another, or a model that performs well in one environment may not be suitable in a different one. Therefore, to navigate in a space with varying data distributions, some form of ML model personalization is crucial.

\paragraph{Federated Learning.} To address a more holistic view of training, computation, communication, model personalization, and privacy issues, a new paradigm known as Federated Learning (FL) has recently emerged. FL is a subfield of Machine Learning (ML) that seeks to provide solutions to challenges such as distributed training, heterogeneity in clients' computing and communication capabilities, personalization, and various communication issues during training. FL enables model training without data ever leaving users' devices, ensuring that quantities communicated during training can be protected by multiple privacy-preserving mechanisms. FL allows clients to collaboratively train a shared model while performing computations on their own devices, ensuring that raw data never leaves these devices.

Federated Learning (FL) was proposed in the works of \citet{konevcny2016afederated, FEDLEARN, mcmahan17fedavg, kairouz2019advances}. Recently, it was incorporated into the 2023 update of the National AI Research and Development Strategic Plan of the United States \citep{national2019national}. FL involves distributed training that operates as follows. In each training round, \textit{participating devices} download or update information in a way that allows them to continue training according to the prescribed training algorithm. Next, each selected device performs training using its local data, resulting in updated quantities of the model state. Subsequently, participants submit these updates to the system as prescribed by the training algorithm, and the process repeats.

\begin{figure}[t]
	\centering
	\includegraphics[width=0.85\textwidth]{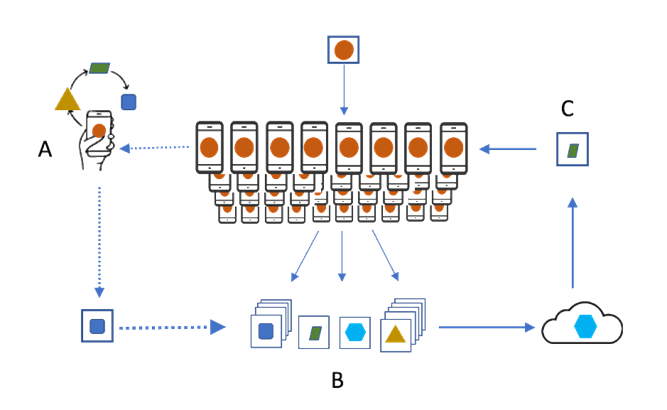}
	\caption{The lifecycle of a Federated Learning training in the cross-device setting. Source of Image: Google Search.}
	\label{ch1:fig:fl_google}
\end{figure}

\citet{kairouz2019advances} identified two main practical settings in FL: \textit{cross-silo} and \textit{cross-device}. In the \textit{cross-silo} setting, several organizations share the common objective of training a model based on the union of their underlying data. However, they do not wish to share their data directly due to privacy concerns. Typically, this setting assumes that the number of organizations involved is relatively small. In the \textit{cross-device} setting, the clients are computational devices with varying computation and storage capacities. Their goal is to collaborate in training a prediction model that will be deployed on these devices. A visualization of the cross-device FL lifecycle is provided in Figure~\ref{ch1:fig:fl_google}.

\paragraph{Federated Learning applications.} Federated Learning (FL) has gained significant academic attention due to its integration of diverse scientific and engineering subfields, and necessarily of addressing complex challenges. On the practical side, several real-world applications have already adopted this paradigm. Notable examples include Apple's vocal classifier for Siri and the iOS 13 QuickType keyboard~\citep{apple19wwdc}, Google Gboard keyboard for next-word prediction, and Android Messages~\citep{hard18gboard}. The growth of FL has also driven the emergence of companies such as \href{https://xayn.com}{Xayn}, \href{https://www.owkin.com}{Owkin}, \href{https://edgedelta.com}{Edge Delta}, \href{bittensor.com}{Bittensor}, \href{http://flower.dev}{Flower}, and others. FL is a particularly advantageous paradigm when the underlying datasets are distributed across numerous clients, either naturally or due to restrictions on centralized data collection. 

Another promising area for FL is its application within the Internet of Things (IoT), where computational devices face limitations in computational power, storage, and energy capacity. The rise of specialized IoT devices is driven by their lower costs and improved energy efficiency. These devices can play a crucial role in modern cities and find applications in areas such as waste management, street lighting, automated parking, sleep monitoring, manufacturing, security, vehicle connectivity, baby monitoring, and food safety~\citep{kotha2018iot}.

\section{Problem Formulation}
\label{ch1:sec:main-problem-formulation}

From the perspective of supervised ML, our objective is to select a function from a parameterized function class $\mathcal{F}_x$ indexed by $x \in \mathbb{R}^d$. 

In the context of Federated Learning (FL), this is achieved by solving the following optimization problem, either exactly or approximately:

\begin{equation}
	\label{ch1:eq:main}
	\min \limits_{x\in \RS^d} \left\{f(x)\eqdef \dfrac{1}{n}\sum \limits_{i=1}^n f_i(x) \right\}, 
\end{equation}

Here, $n\in \mathbb{N}$ represents the number of clients or devices, and $x\in \mathbb{R}^d$ denotes the $d$ parameters or weights of a model $\hat{F}(\cdot;x) \in \mathcal{X}\to \mathcal{Y}$ that need to be learned across all $n$ clients. The function $f_i:\mathbb{R}^d \to \mathbb{R}$ provides the score criteria for using the model $\hat{F}(\cdot;x)$ on the $i$-th client's data. In the context of {FL}, the functions $f_i$ are typically represented as:

\begin{equation}\label{ch1:eq:fi_for_erm} 
	f_i(x) = \dfrac{w_i}{n_{i}} \sum \limits_{j=1}^{n_{i}} \left(\mathcal{L}_{ij}(b_{ij}, \hat{F}(a_{ij};x)) + R_{i}(x) \right),
\end{equation}

In Equation~\eqref{ch1:eq:fi_for_erm}, $n_i \in \mathbb{N}$ represents the number of data points at client $i$, and $(a_{ij}, b_{ij}) \in \mathcal{X} \times \mathcal{Y}$ are the input-output pairs at client $i$. The function $\mathcal{L}_{ij}(y_{\mathrm{real}}, y_{\mathrm{pred}}): \mathcal{Y} \times \mathcal{Y} \to \mathbb{R}$ is a loss function that scores the prediction, while $R_i: \mathbb{R}^d \to \mathbb{R}$ is a regularization function applied to the parameter $x$ at client $i$. The weight $w_i \in \mathbb{R}$ encodes the relative importance of client $i$ and can be used to adjust the training objective. For example:
\begin{enumerate}
	\item $w_i \eqdef \dfrac{n_i \cdot n}{\sum_{i=1}^{n} n_i}$ reflects a scenario where all data points are equally important.	
	\item $w_i \eqdef 1$ corresponds to a case where all devices are equally important.
\end{enumerate}

The presented form represents the most common approach to Federated Learning, where a single global model \( x \in \mathbb{R}^d \) is learned from decentralized data stored across \( n \) clients. An extension of this model, where each client has its own model \( x_i \in \mathbb{R}^d \) for \( i \in [n] \), is also possible and will be considered in Chapter~\ref{chapter6}.

After the model is trained and the solution $x^*$ has been obtained, it can be used for inference in the form $\hat{y} = \hat{F}(a; x^*)$.

Therefore, in Federated Learning (FL), we aim to learn either a single global model $x$ or individual models $x_i \in \mathbb{R}^d$ for each client $i \in [n]$, under the constraint that each client’s data is kept and processed locally. The learning objective must be achieved using only intermediate client updates. This setup differs significantly from traditional distributed learning due to privacy guarantees, the inability to control client participation, possible asymmetries in the computing capabilities of participating agents, and variations in data distribution across clients. 

As a result, FL presents unique scientific challenges that drive advancements in privacy, security, information theory, large-scale Machine Learning, and Distributed Optimization. In addition to these scientific challenges arising from FL's multidisciplinary nature, it is important to highlight the hidden practical challenges that influence both the adoption of implementable methods and the development of new ones.

\section{Challenges}
\label{ch1:sec:challenges}

As previously discussed, Federated Learning presents specific scientific and practical challenges, which we now briefly describe.

\subsection{Data distribution heterogeneity}
\label{ch1:sec:s-data-distribution}

In FL, although clients share a common goal, their underlying data distributions can vary significantly. Each client may hold data that is not only heterogeneous but also non-independent and identically distributed (non-IID). In supervised ML, this means that both class labels and feature distributions can differ across clients. Such variations arise from differences in data collection processes, sensor types, sensor noise, geographical regions, and other factors. As a result, models trained in FL must be robust to these variations. Enforcing a homogeneous data distribution across all clients is not feasible, as data exchange between clients is prohibited. This inherent heterogeneity poses challenges for model convergence. These issues can be mitigated in various ways, one of which is through appropriate FL personalization techniques \citep{gasanov2021flix, Hanzely2020}.

\subsection{Device heterogeneity}
\label{ch1:sec:s-computation}

The cross-device setting of Federated Learning (FL) is designed to operate across a diverse range of devices, including the Internet of Things (IoT) and mobile devices, which often have limited computing power, memory capacity, and energy resources. Limited memory in edge devices poses a challenge during inference \citep{laskaridis2024melting} and becomes an even greater issue during on-device training. Expanding memory in IoT devices is typically infeasible due to their integrated System-on-Chip (SoC) design. These systems are highly optimized for specific tasks and generally do not allow for components to be upgraded over time. This system heterogeneity introduces diverse challenges due to computational asymmetries, which are less common in data center settings. For example, the participation of specific edge devices in Federated Learning becomes completely infeasible if peak memory consumption during training exceeds the available memory on the client device.

The memory aspect also impacts energy consumption. Prior works by \citet{chen2018understanding} and \citet{horowitz20141} have shown that energy consumption is primarily influenced by data flow and memory hierarchy, rather than the utilization of functional units for arithmetic. One approach to addressing computational heterogeneity is asynchronous training \citep{maranjyan2024mindflayer, maranjyan2025ringmaster}.

\subsection{Communication bottleneck}
\label{ch1:sec:s-communication}

In modern data centers, ML training computations are typically executed on traditional data center clusters. The most common network topology in such environments is the hierarchical fat-tree topology. The name comes from the fact that the communicated links higher in the hierarchical tree structure have greater bandwidth (fatter) than those lower down. This topology is widely recognized as an effective choice for data centers and high-performance computing environments\footnote{\href{https://www.cs.cornell.edu/courses/cs5413/2014fa/lectures/08-fattree.pdf}{High-Performance Systems and Networking, Cornell University: Data Center Network Topologies}}.

In FL, the underlying communication network is often unreliable, and, unlike in data centers, a specific communication topology cannot be strictly enforced. For instance, the work of \citet{wang2023topoopt}, which suggests optimizing network topology as an optimization variable for specific ML tasks, is applicable in data centers where the topology is controllable. However, such approaches are challenging to adapt to FL due to its decentralized nature and the variability of network conditions. In FL, clients are typically connected via an uncontrollable and less reliable communication network, such as the Internet. The transferred messages can experience arbitrarily long latency delays, and if a subset of clients shares a common (bottleneck) link, the effective bandwidth is divided among them, reducing throughput.

Since FL has no control over the construction of the communication network, communication between clients and the server infrastructure is often slower than the actual computation \citep{huang2013depth, van2009multi}. Furthermore, training algorithms that conceptually require aggregation across billions of clients simultaneously face technical challenges at multiple levels, including operating systems, network protocol implementations, and hardware constraints. 

To mitigate the communication bottleneck, various strategies can be employed, such as increasing the computational workload per client, applying communication compression, and enabling partial client participation \citep{gorbunov2021marina, malinovsky2023federated}.

\subsection{Privacy and security guarantees}
\label{ch1:sec:s-privacy}

Unfortunately, the partial derivative of the function $f_i(x)|_{x=x_0}$ from Equation~\eqref{ch1:eq:fi_for_erm} is also a function of the data which is suppressed in the $f_i$ notation. Revealing information about one partial derivative may expose some information about $D_i$. With a sufficiently large number of partial derivatives, it is possible to reveal enough information to reconstruct the client’s dataset $D_i$. 

Such attacks are not merely theoretical, they can be practically executed \citep{cohen2018linear, kasiviswanathan2013power}. To address this issue, the training process should be protected. There is no universal agreement on which mechanism is the best, as each mechanism has its limitations. Nevertheless, various mechanisms can be employed in FL to ensure \textit{privacy} at different stages of the training process. The notion of \textit{privacy} refers to the guarantee that the protected party can participate in an algorithm without revealing sensitive information. Each of the following mechanisms, in the context of FL, targets a specific aspect of protection (such as input, output, execution, or communication channel) and is designed for a particular adversary model. Notable examples include the following:

\begin{enumerate}
	\item \textit{Differential Privacy (DP).} Differential Privacy protects the output of an algorithm to ensure that it does not disclose sensitive information about the input data \citep{dwork2006our}.
	
	\item \textit{Secure Multi-Party Computation (MPC).} Secure Multi-Party Computation ensures that the input data remains protected from exposure to other parties during computation \citep{zhao2019secure}.
	
	\item \textit{Trusted Execution Environments (TEE).} Trusted Execution Environments safeguards the execution environment from unauthorized interventions that could disrupt training. It emphasizes self-isolation and is more robust than traditional operating system practices \citep{sabt2015trusted}.
	
	\item \textit{Homomorphic Encryption (HE).} HE hides both the plaintext input and output from the executor, who only sees the encrypted versions. HE allows any device with knowledge of the public key to perform computations on encrypted data and return encrypted results. The result can only be decrypted with the private key \citep{rivest1978data,gentry2009fully}.
	
\end{enumerate}

\subsection{Practical applicability beyond experimental evidence}
\label{ch1:sec:s-brige-th-and-pr}

We define practical applicability as any activity aimed at bridging fundamental research with the real world. In health studies, \citet{booth2004research} defines 
\begin{center}
	\textit{applicability as a relationship that describes how research results are likely to impact practice,}
\end{center}
which closely aligns with our definition. 

Traditionally, when a theory is approaching readiness for practical application, its relevance is demonstrated through experiments in research papers. However, we would like to highlight that there are other formats as well. 

\paragraph{1. Simulators.}

A simulator is a system that mimics the behavior of the real world without necessarily replicating the exact hardware or environment. Simulators are valuable when constructing physical environments is costly or impossible due to scale. In Networks, network simulators are complex software programs that replicate the behavior of real networks. Notable examples include \libname{ns-2} and \libname{ns-3} \citep{issariyakul2009introduction, riley2010ns}. In Digital Design, simulators are tools used for modeling, algorithmic verification, functional verification, synthesis verification, and timing verification of electrical circuits. They also help identify errors in designs, provide graphical feedback on underlying physical processes, and sometimes offer ways to \textit{prove} the correctness of a circuit. A notable example of such a simulator is \libname{Vivado Design Suite} \citep{feist2012vivado}.

A complementary approach to simulators is the construction of \textit{miniaturized physical systems} or \textit{testbeds} for end-to-end analysis. In the context of Federated Learning, the recent work exploring this is \libname{CoLExT}~\citep{bovzivc2024testbed}.

\paragraph{2. Ready-to-Use implementations.} 

The final aspect of practical applicability lies in implementing theoretical methods in software or hardware. As Stephen Boyd noted in his interview \citep{interviewboyd}, the absence of publicly available implementations in {Control Theory} has significantly hindered its widespread adoption, despite its origins in the mid-20th century. Successful implementation of scientific methods often emerges at the intersection of industrial and academic research. Notable examples of implementations born in academia include \libname{CVXGEN}\citep{mattingley2012cvxgen}, which played a crucial role in SpaceX's Falcon 9 landing, and \libname{Apache Spark}\citep{zaharia2016apache}, widely used for data-intensive computations.

In the realm of FL, middleware systems such as \libname{Flower}~\citep{beutel2020flower} are making serious long-term strides in this direction.

\paragraph{3. Estimation of unknown constants.} 

Another key aspect of practical applicability is the \textit{estimation of unknown constants}, which may be required during the runtime of an algorithm. If eliminating this dependency at runtime is theoretically complex or remains an open research question, an alternative approach is to develop a mechanism for \textit{approximately estimating} these constants or providing rigorous bounds. However, when applying the proposed algorithm in practice or comparing it with other algorithms, the cost of the \textit{estimation process} should ultimately be accounted for in the final evaluation.

\paragraph{4. Correlation of theoretical and real cost models.} 

Theoretical cost models may not be entirely accurate and may require adjustments to align with real-world conditions. In complex situations, practical implementation and theoretical models intertwine in iterative processes, leading to continuous refinement of both if aiming for perfection. Sometimes the actual cost models present a mixture of theoretical cost models and actual measurements \citep{jia2019taso}.

\paragraph{5. Simplicity in system design.} 

Simplicity is sometimes associated with poor implementation or design. However, in this context, we refer to the inherent properties of the algorithms and designs themselves. Simplicity is particularly important in Federated Learning because: (i) it reduces the complexity of real-world implementations; (ii) facilitates interdisciplinary collaboration; (iii) models and designs that prioritize simplicity are often easier to analyze, debug, create, and computationally optimize; (iv) reduced complexity ultimately supports real-world deployment. 

In the realm of Convex Optimization, these lessons have been observed by \citet{mattingley2012cvxgen} and further emphasized by the authors in various public talks \citep{interviewboyd}.

\subsection{Thesis Focus}
\label{thesis:sec:focus}

\begin{table}[!t]
	{
		\footnotesize
		\begin{center}
			\begin{tabular}{|c|c|c|c|c|c|c|c|}
				\hline
				\multirow{3}{*}{\bf Ch} & & \multirow{3}{*}{\bf Solution and Reference} & {\bf Data} &  {\bf Device} & {\bf Com-} & {\bf Pri-} & {\bf Soft-}  \\
				  & {\bf Over-} &  & {\bf Heter.} & {\bf Heter.} & {\bf mun.}  & {\bf vacy}  &  {\bf ware$^\dagger$}\\
				\cline{4-8}
				  & {\bf view} &  &  {\ref{ch1:sec:s-data-distribution}} & {\ref{ch1:sec:s-computation}} & \ref{ch1:sec:s-communication} & \ref{ch1:sec:s-privacy} & \ref{ch1:sec:s-brige-th-and-pr}{$^\ast$} \\
				\hline
				\hline
				\ref{chapter2} & \ref{ch1:sec:overview-2}  &
				\makecell[c]{
				 {\tt FL\_PyTorch} \\
				 {\scriptsize\citep{burlachenko2021fl_pytorch}}
				}
				 & \xmark & \xmark & \xmark & \xmark &\cmark      \\ \hline
				\ref{chapter3} & \ref{ch1:sec:overview-3}  & 
				\makecell[c]{
				{\tt EF21-W} \\
				{\scriptsize\citep{richtarik2024error}}
				}
			    & \cmark & \xmark & \cmark & \xmark & \xmark     \\ \hline
				\ref{chapter4} & \ref{ch1:sec:overview-4}  & 
				\makecell[c]{
				{\tt DCGD/PERMK/AES} \\
				{\scriptsize\citep{burlachenko2023federated}}
				}
			  	& \xmark & \cmark & \cmark & \cmark & \xmark     \\ \hline
				\ref{chapter5} & \ref{ch1:sec:overview-5}  & 
				\makecell[c]{
				{\tt PAGE Extensions} \\
				{\scriptsize\citep{tyurin2022sharper}}
				}
			   	& \xmark & \cmark & \xmark & \xmark & \xmark     \\ \hline
				\ref{chapter6} & \ref{ch1:sec:overview-6}  & 
				\makecell[c]{
				{\tt Compressed L2GD} \\
				{\scriptsize\citep{houcine2022personalized}}
				}
				& \cmark & \xmark & \cmark & \xmark & \xmark     \\ \hline
				\ref{chapter7} & \ref{ch1:sec:overview-7}  & 
				\makecell[c]{
				{\tt Unlocking FedNL} \\
				{\scriptsize\citep{burlachenko2024unlocking}}
				}
				& \xmark & \cmark & \cmark & \xmark & \cmark     \\ \hline
				\ref{chapter8} & \ref{ch1:sec:overview-8}  & 
				\makecell[c]{
				{\tt BurTorch} \\
				{\scriptsize\citep{burlachenko2025burtorch}}
				}
			    & \xmark & \cmark & \xmark & \xmark & \cmark     \\ \hline
			\end{tabular}
		\end{center}
		\caption{
A summary of representative contributions proposed in each chapter, along with the explicitly addressed challenges. The columns are: \textbf{Ch} = Chapter, \textbf{Overview} = Overview section, \textbf{Solution} = Main algorithm or implementation, \textbf{Data Heter.} = Data distribution heterogeneity, \textbf{Device Heter.} = Device heterogeneity, \textbf{Commun.} = Communication bottleneck, \textbf{Privacy} = Privacy and Security, \textbf{Software} = Software practically applicable beyond experiments. \textbf{Clarifications:} $\dagger$: All underlying papers from the presented thesis contain experimental evidence. In the context of this table, \textbf{Software} refers to advanced practical applicability beyond experimental prototypes; {$^\ast$}: References to specific challenge described earlier.
		}
		\label{ch1:tbl:algorithms}
	}
\end{table}

\begin{figure}[t]
	\centering
	\includegraphics[width=0.90\textwidth]{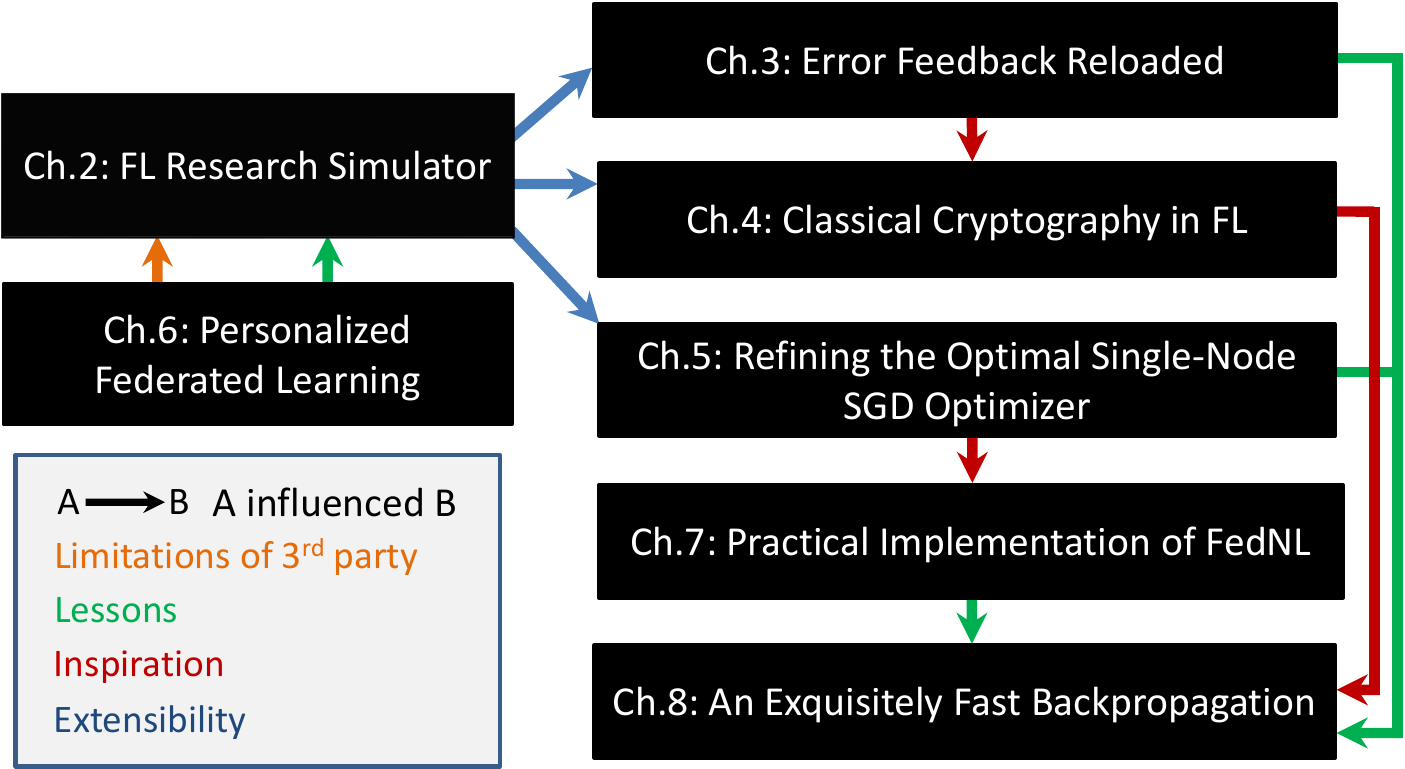}
	\caption{The interconnections between various aspects of Federated Learning explored in this thesis and the influences of each component on the others.}
	\label{ch1:fig:thesis-focus}
\end{figure}

This thesis consists of \textit{seven research papers}, each targeting a specific aspect of the five challenges outlined in Section~\ref{ch1:sec:challenges}. These papers present novel solutions and contribute to advancing the field of Federated Learning in various ways.

Table~\ref{ch1:tbl:algorithms} summarizes the scope of each chapter by highlighting a representative algorithm and the specific challenges addressed by each chapter. Figure~\ref{ch1:fig:thesis-focus} visualizes the relationships between the works presented in this thesis, illustrating how they interconnect and influence one another.

\section{Scope of this Chapter}

This chapter aims to introduce the specific aspects of Federated Learning (FL) addressed in this thesis, setting the stage for detailed discussions in subsequent chapters. A more comprehensive overview of FL will be provided later, as needed. For a broader understanding of the current research landscape in FL, we recommend the survey papers by \citet{kairouz2019advances} and \citet{ding2022federated}.

\section{Chapters Organization}
\label{ch1:sec:organization}

This introductory chapter precedes seven chapters, each dedicated to a distinct paper. Some of these papers have been published, while others are currently undergoing peer review. Each chapter addresses a critical challenge in FL, as outlined in Section~\ref{ch1:sec:challenges}, and proposes a novel, and at times surprising, solutions.

Each chapter is structured as follows. First, we identify a specific problem and review the relevant literature. Next, we provide an overview of our contributions and present the proposed solution, followed by experimental evaluations. Each chapter concludes with a conclusion section. Finally, each chapter includes appendices specific to that chapter, containing additional background, detailed derivations for theoretically inclined papers, and a broader set of experiments, which are designed to be optional for reading. Each chapter's appendix contains a reproducibility statement at the end. The chapters are self-contained and do not include cross-references, except for the back reference to Table \ref{ch1:tbl:algorithms} and Section~\ref{ch1:sec:organization} to facilitate navigation. Chapter~\ref{chapater-conclusion} provides concluding remarks and suggestions for future research. 

The following sections provide a concise overview of each main chapter.

\subsection{Chapter 2: FL research simulator}
\label{ch1:sec:overview-2}

Federated Learning (FL) has emerged as a promising technique for edge devices to collaboratively learn a shared machine learning model while keeping training data locally on the device, thereby removing the need to store and access the full data in the cloud. However, FL is difficult to implement, test, and deploy in practice considering heterogeneity in common edge device settings, making it fundamentally hard for researchers to efficiently prototype and test their optimization algorithms. In this chapter, we aim to alleviate this problem by introducing \fl: a suite of open-source software written in Python that builds on top of one of the most popular Deep Learning (DL) framework \libname{PyTorch} \citep{paszke2019pytorch}. We built \fl as a research simulator for FL to enable fast development, prototyping, and experimenting with new and existing FL optimization algorithms. Our system supports abstractions that provide researchers with a sufficient level of flexibility to experiment with existing and novel approaches to advance the state-of-the-art.  In this chapter, we introduce a new (since 2021) and, to the best of our knowledge, still (as of 2025) unique theoretical research simulator designed with the needs of scientists in mind. Chapter~\ref{chapter2} primarily addresses Challenge~\ref{ch1:sec:s-brige-th-and-pr} (Software) and is based on:
\begin{quote}
\citet{burlachenko2021fl_pytorch}: Konstantin Burlachenko, Samuel Horv{\'a}th, Peter  Richt{\'a}rik. Fl\_PyTorch: optimization research simulator for Federated Learning. In \textit{Proceedings of the 2nd ACM International Workshop on Distributed Machine Learning, 2021}.
\end{quote}

\subsection{Chapter 3: Error feedback reloaded}
\label{ch1:sec:overview-3}

Error Feedback (\algname{EF}) is a highly popular and immensely effective mechanism for fixing convergence issues that arise in distributed training methods (such as distributed \algname{GD} or \algname{SGD}) when these are enhanced with greedy communication compression techniques such as \compname{TopK}. While \algname{EF} was proposed almost a decade ago by \citet{Seide2014}, and despite concentrated effort by the community to advance the theoretical understanding of this mechanism, there is still a lot to explore. In this work, we study a modern form of error feedback called \algname{EF21} by \citet{EF21} which offers the best-known theoretical guarantees, under the weakest assumptions, and also works well in practice. In particular, while the theoretical communication complexity of \algname{EF21} depends on the {\em quadratic mean} of certain smoothness parameters, we improve this dependence to their {\em arithmetic mean}, which is always smaller and can be substantially smaller. In this chapter, we introduce a theoretical refinement of the currently state-of-the-art first-order distributed non-convex optimization method with contractive compressors. We take the reader on a journey of our discovery process, which starts pretty intricate. Chapter~\ref{chapter3} primarily addresses Challenges~\ref{ch1:sec:s-data-distribution} (Data heterogeneity) and \ref{ch1:sec:s-communication} (Communication). It is based on:
\begin{quote}
\citet{richtarik2024error}: Peter Richt{\'a}rik, Elnur Gasanov, Konstantin Burlachenko. Error Feedback Reloaded: From Quadratic to Arithmetic Mean of Smoothness Constants. In \textit{Proceedings of the 12th International Conference on Learning Representations, 2024}.
\end{quote}
	
\subsection{Chapter 4: Classical cryptography in FL}
\label{ch1:sec:overview-4}

Traditional AI methodologies necessitate centralized data collection, which becomes impractical when facing problems with network communication, data privacy, or storage capacity. Federated Learning ({FL}) offers a paradigm that empowers distributed AI model training without collecting raw data. There are different choices for providing privacy during FL training. One of the popular methodologies is employing Homomorphic Encryption (\abr{HE}) - a breakthrough in privacy-preserving computation from Cryptography. However, these methods have a price in the form of extra computation and memory footprint. To resolve these issues, we propose an innovative framework that synergizes permutation-based compressors with Classical Cryptography, even though employing Classical Cryptography was assumed to be impossible in the past \citep{kaissis2020secure,lauter2022private,jain2023revisiting,DBLP:journals/cybersec/PanCWYLW24}. The combination of a special class of permuted correlated compressors, proposed by \citet{szlendak2021permutation}, with classical block ciphers, \aesname{AES} \citep{daemen2001reijndael}, leads to a significant reduction in memory usage compared to the current best-practice solution such as \algname{CKKS}~\citep{cheon2017homomorphic}. Chapter~\ref{chapter4} addresses Challenges~\ref{ch1:sec:s-computation} (Device heterogeneity), \ref{ch1:sec:s-communication} (Communication), \ref{ch1:sec:s-privacy} (Privacy). It is based on:
\begin{quote}
\citet{burlachenko2023federated}: Konstantin Burlachenko, Abdulmajeed Alrowithi, Fahad Ali Albalawi, Peter Richt{\'a}rik. Federated Learning is Better with Non-Homomorphic Encryption. In \textit{Proceedings of the 4th International Workshop on Distributed Machine Learning, 2023}.
\end{quote}	
	
\subsection{Chapter 5: Refining the optimal single-node optimizer}
\label{ch1:sec:overview-5}
  
This work revisits the classical problem of finding an approximately stationary point of the average of $n$ smooth and possibly non-convex functions. The optimal complexity of stochastic first-order methods in terms of the number of gradient computations of individual functions, attained by the optimal \algname{SGD} methods \algname{SPIDER} \citep{fang2018spider} and \algname{PAGE} \citep{li2021page}. However, i) the big-$\mathcal{O}$ notation hides crucial dependencies on the smoothness constants associated with the functions, and ii) the rates and theory in these methods assume simplistic sampling mechanisms that do not offer any flexibility. In this work, we remedy the situation. Finally, our theoretical findings are supported with carefully designed experiments. Chapter~\ref{chapter5} primarily addresses Challenge~\ref{ch1:sec:s-computation} (Device heterogeneity) and is based on:
\begin{quote}
\citet{tyurin2022sharper}:  Alexander Tyurin, Lukang Sun, Konstantin Burlachenko, Peter Richt{\'a}rik. Sharper rates and flexible framework for nonconvex \algname{SGD} with client and data sampling. \textit{Transactions on Machine Learning Research, 2023}.
\end{quote}
	
\subsection{Chapter 6: Personalized FL with compression}
\label{ch1:sec:overview-6}

In contrast to training traditional machine learning (ML) models in data centers, Federated
Learning (FL) trains ML models over local datasets on resource-constrained heterogeneous
edge devices. Existing FL algorithms aim to learn a single global model for all participating devices, which may not be helpful to all devices participating in the training due to the heterogeneity of the data across the devices. Recently, \citet{Hanzely2020} proposed a new formulation for training personalized FL models aimed at balancing the trade-off between the traditional global model and the local models that could be trained by individual devices using their private data only. They derived a new algorithm, called loopless local gradient descent (\algname{L2GD}), to solve it and showed that this algorithm leads to improved communication complexity guarantees in regimes when more personalization is required. In this work, we equip their \algname{L2GD} algorithm with a bidirectional compression mechanism to further reduce the communication bottleneck between the local devices and the server. Unlike other compression-based algorithms used in the FL setting, our compressed \algname{L2GD} algorithm operates on a probabilistic communication protocol, where communication does not happen on a fixed schedule. Chapter~\ref{chapter6} primarily addresses Challenges~\ref{ch1:sec:s-data-distribution} (Data heterogeneity) and~\ref{ch1:sec:s-communication} (Communication). It is based on:
\begin{quote}
\citet{houcine2022personalized}: El Houcine Bergou, Konstantin Burlachenko, Aritra Dutta, Peter Richt{\'a}rik. Personalized Federated Learning with Communication Compression. \textit{Transactions on Machine Learning Research, 2023}.
\end{quote}

\subsection{Chapter 7: Practical implementation of FedNL}
\label{ch1:sec:overview-7}

Federated Learning ({FL}) is an emerging paradigm that enables intelligent agents to collaboratively train ML models in a distributed manner, eliminating the need for sharing their local data. The recent work \citet{safaryan2021fednl} introduces a family of Federated Newton Learn (\algname{FedNL}) algorithms, marking a significant step towards applying second-order methods to {FL} and large-scale optimization. However, the reference \algname{FedNL} prototype exhibits three serious practical drawbacks: (i) It requires $4.8$ hours to launch a single experiment in a server-grade workstation; (ii) The prototype only simulates multi-node setting; (iii) Prototype integration into resource-constrained applications is challenging. To bridge the gap between theory and practice, we present a self-contained implementation of \algname{FedNL}, \algname{FedNL-LS},  \algname{FedNL-PP} for single-node and multi-node settings. Our work resolves the aforementioned issues and reduces the wall clock time by $\times 1000$. With this \algname{FedNL} outperforms alternatives for training \modelname{logistic regression} in a single-node -- \libname{CVXPY} \citep{diamond2016cvxpy}, and in a multi-node -- \libname{Apache Spark} \citep{meng2016mllib}, \libname{Ray/Scikit-Learn} \citep{moritz2018ray}. Finally, we propose two practical-oriented compressors for \algname{FedNL} - adaptive \compname{TopLEK} and cache-aware \compname{RandSeqK}, which fulfill the theory of \algname{FedNL}. Chapter~\ref{chapter7} primarily addresses Challenges~\ref{ch1:sec:s-computation} (Device heterogeneity), \ref{ch1:sec:s-communication} (Communication), \ref{ch1:sec:s-brige-th-and-pr} (Software) and is based on:
\begin{quote}
\citet{burlachenko2024unlocking}: Konstantin Burlachenko, Peter Richt{\'a}rik. Unlocking FedNL: Self-Contained Compute-Optimized Implementation. \textit{arXiv: 2410.08760, 2024}.
\end{quote}

\subsection{Chapter 8: An exquisitely fast backpropagation}
\label{ch1:sec:overview-8}

In this chapter, we introduce \libname{BurTorch}, a compact high-performance framework designed to optimize Deep Learning (DL) training on single-node workstations through an exceptionally efficient CPU-based backpropagation \citep{rumelhart1986learning, linnainmaa1970representation} implementation. Although modern DL frameworks rely on compiler-like optimizations internally, \libname{BurTorch} takes a different path. It adopts a minimalist design and demonstrates that, in these circumstances, classical compiled programming languages can play a significant role in DL research. By eliminating the overhead of large frameworks and making efficient implementation choices, \libname{BurTorch} achieves orders-of-magnitude improvements in performance and memory efficiency when computing $\nabla f(x)$ on a CPU. \libname{BurTorch} features a compact codebase designed to achieve two key goals simultaneously. First, it provides a user experience similar to script-based programming environments. Second, it dramatically minimizes runtime overheads. In large DL frameworks, the primary source of memory overhead for relatively small computation graphs $f(x)$ is due to feature-heavy implementations. We benchmarked \libname{BurTorch} against widely used DL frameworks in their execution modes: \libname{JAX} \citep{jax2018github}, \libname{PyTorch} \citep{paszke2019pytorch}, \libname{TensorFlow} \citep{abadi2016tensorflow}; and several standalone libraries: \libname{Autograd} \citep{maclaurin2015autograd}, \libname{Micrograd} \citep{karpathy2020micrograd}, \libname{Apple MLX} \citep{mlx2023}. For small compute graphs, \libname{BurTorch} outperforms best-practice solutions by up to $\times 2000$ in runtime and reduces memory consumption by up to $\times 3500$. For a miniaturized \modelname{GPT-3} model \citep{gpt}, \libname{BurTorch} achieves up to a $\times 20$ speedup and reduces memory up to $\times 80$ compared to \libname{PyTorch}. Chapter~\ref{chapter8} addresses Challenges \ref{ch1:sec:s-computation} (Device heterogeneity) and \ref{ch1:sec:s-brige-th-and-pr} (Software) and is based on:

\begin{quote}
Konstantin Burlachenko, Peter Richt{\'a}rik. BurTorch: Revisiting Training from First Principles by Coupling Autodiff, Math Optimization, and Systems. \textit{arXiv: 2503.13795, 2025.}
\end{quote}



\subsection{Excluded papers}
\label{thesis:sec:excluded-papers}

\begin{table}
	\centering
	\small
	\caption{Papers excluded from this thesis.}
	\label{ch1:tab:excluded}
	\begin{tabular}{|c|c|c|c|}
		\hline
		\textbf{\#} & \textbf{Title and Reference} & \makecell[c]{\textbf{FL}\\\textbf{Paper}} & \makecell[c]{\textbf{LLM}\\\textbf{Paper}}
		\\
		\hline
		\hline
		1 & \begin{tabular}[c]{@{}c@{}}
			MARINA: Faster Non-Convex Distributed Learning \\ with Compression \citep{gorbunov2021marina}.\\ \textit{ICML 2021}. 
		\end{tabular}                       & \cmark & \xmark  \\ \hline
		2 & \begin{tabular}[c]{@{}c@{}} Don't Compress Gradients in Random Reshuffling:\\Compress Gradient Differences \citep{DBLP:conf/nips/SadievMG00BR24, sadiev2022federated}.\\
		\textit{NeurIPS 2024}.
		\end{tabular}                       & \cmark & \xmark  \\ \hline		
		3 & \begin{tabular}[c]{@{}c@{}}Federated Learning with Regularized Client Participation \\
		\citep{malinovsky2023federated}.\\
		\textit{arXiv: 2302.03662, 2023.}
		\end{tabular}                          & \cmark & \xmark  \\ \hline		
		4 & \begin{tabular}[c]{@{}c@{}}Error Feedback Shines when 
		Features are Rare \\ \citep{richtarik2023error}.\\
		\textit{arXiv: 2305.15264, 2023.}
		\end{tabular}                                           & \cmark & \xmark  \\ \hline		
		5 & \begin{tabular}[c]{@{}c@{}}PV-Tuning: Beyond Straight-Through Estimation\\for Extreme LLM Compression\\ \citep{malinovskii2024pv}.\\
		\textit{NeurIPS 2024}.
		\end{tabular}                               & \xmark & \cmark  \\ \hline		
		6 & \begin{tabular}[c]{@{}c@{}}Faster Rates for Compressed Federated Learning\\ with Client-Variance Reduction \citep{zhao2024faster}\\
		\textit{SIAM Journal on Mathematics of Data Science 2024}.
		\end{tabular}                                     & \cmark & \xmark  \\ \hline
		
	\end{tabular}	
\end{table}

During my PhD, I have co-authored six additional papers that are not included in this thesis. A summary of these excluded papers is provided in Table~\ref{ch1:tab:excluded}. The list is as follows:

\begin{enumerate}
\item \citet{gorbunov2021marina} introduced \algname{MARINA} algorithm, which held the best-known worst-case convergence guarantees for a significant period of time. This work established a state-of-the-art theoretical framework for distributed non-convex training using unbiased compressors, such as randomly selecting $k$ out of $d$ coordinates for transmission, for quantities like $\nabla f_i(x)$. A notable refinement of \algname{MARINA} was later presented by \citet{szlendak2021permutation}, which used correlated and permutation compressors, \compname{PermK}.

\item \citet{DBLP:conf/nips/SadievMG00BR24, sadiev2022federated} analyzed \algname{Quantized SGD} \citep{khirirat2018distributed} and \algname{DIANA} \citep{mishchenko2024distributed} in the context of random reshuffling (also known as iterated mini-batch) within practical machine learning settings.

\item \citet{malinovsky2023federated} explored double random reshuffling of clients organized into groups (cohorts), where these groups were connected in a round-robin fashion. This operational regime helps reduce aggregation complexity in Federated Learning systems.

\item \citet{richtarik2023error} examined a specific regime of \algname{EF21} \citep{EF21}, highlighting improvements that arise when the training objective components $f_i$ in Equation~\eqref{ch1:eq:main} exhibit a particular regularity structure.

\item \citet{malinovskii2024pv} \footnote{Vladimir Malinovskii (from the paper Table~\ref{ch1:tab:excluded} (5)) and Grigory Malinovsky (from the paper Table~\ref{ch1:tab:excluded} (3)) are two distinct researchers with similar surnames.} investigated training quantization techniques for Large Language Models (LLMs) through the lens of optimization. The proposed mechanism generalizes and improves upon existing fine-tuning strategies while providing convergence guarantees in restricted cases. On the practical side, it outperformed prior techniques.

\item \citet{zhao2024faster} addressed a key limitation of \algname{MARINA}, which requires periodic use of uncompressed information, and \algname{PP-MARINA}, which requires periodic Full Participation (FP). This work introduced \algname{COFIG} and \algname{FRECON}, which eliminate these requirements, thereby improving practical applicability.

\end{enumerate}

The complete list of publications authored during the PhD, including their inclusion or exclusion status for this thesis and the venues where they have been published or their current status, is presented in Appendix~\ref{thesis:app:all-papers}.

\section{Basic Facts and Notation}
\label{ch1:basic-facts-and-notation}

\subsection{Notation}

Before proceeding with the main results, let us first elaborate on the most common notation used in the thesis and provide some theoretical background.

In addition to the notation introduced in Section~\ref{ch1:sec:main-problem-formulation}, we denote by $x^\star$ an optimal solution of~\eqref{ch1:eq:main} and let $f^\star\eqdef f(x^\star)$. We always assume that our objective is lower-bounded, i.e., $f^\star > - \infty$. We assume that we can compute the gradient of $f_i(x)$ or $f_{ij}(x)$, denoted by $\nabla f_i(x)$ or $\nabla f_{ij}(x)$, respectively. In some scenarios, we cannot directly access the local gradient $\nabla f_i(x)$, and we only see its stochastic estimator $g_i(x)$, which we commonly assume to be unbiased with respect to the current iterate $x$, i.e., $\mathbb{E}[g_i(x) | x] = \nabla f_i(x)$. 

In most cases, when considering a regularization term (e.g., the $\ell_2$ penalty), we implicitly assume it is included in $f_i(x)$.

We use $\mathbb{R}^d$ to denote the Euclidean space of $d$-dimensional real vectors. We use $\lin{ x,y } \eqdef \sum_{i=1}^d x_i y_i$ to denote the standard inner product of two vectors $x, y\in\R^d$, where $x_i$ corresponds to the $i$-th component of $x$ in the standard basis of $\R^d$. 

We denote the $\ell_2$-norm in $\R^d$ as $\norm{x} \eqdef\sqrt{\lin{ x, x }}$. We denote $\ell_p$-norms as $\|x\|_p \eqdef (\sum_{i=1}^d|x_i|^p)^{\nicefrac{1}{p}}$ for $p\in(1,\infty)$. When discussing convergence guarantees, we often use big-$\cO$ notation for ease of presentation, denoted by $\cO(\cdot)$.

The $[n]$ denotes the set of integers $\{1,2,\dots,n\}$ for any positive integer $n$.

\subsection{Smoothness and convexity}

Throughout the thesis, during complexity analysis, we assume the global loss $f(x)$, the local loss $f_i(x)$, and its elements $f_{ij}(x)$ satisfy certain properties. Two of the most basic properties are defined next. 

\begin{definition}[Convexity]
	\label{ch1:ass:1_optimal}
	Differentiable function $h:\R^d\to\R$ is $\mu$-(strongly) convex with $\mu \geq 0$ if
	\begin{equation}
		\label{ch1:eq:def_strongly_convex}
		h(y) \geq h(x) + \dotprod{\nabla h(x)}{y - x} + \dfrac{\mu}{2}\norm{y - x}^2,\quad\forall x,y \in \R^d.
	\end{equation}
	We say $h$ is convex if it satisfies \eqref{ch1:eq:def_strongly_convex} with $\mu = 0$.
\end{definition}

\begin{definition}[Smoothness]
	\label{ch1:ass:smooth}
	Differentiable function $h:\R^d\to\R$ is $L$-smooth if
	\begin{equation*}
		\label{ch1:eq:smooth}
		\norm{\nabla h(x) - \nabla h(y)}\leq L\norm{x - y},\quad\forall x,y \in \R^d.
	\end{equation*}
\end{definition}

The above are standard definitions of convexity and smoothness.

\subsection{Communication compression}
\label{ch1:sec:quant_and_comp}

One way to alleviate the communication bottleneck in FL is to compress messages before communication. We start by defining unbiased and general compression operators as commonly defined in literature~\cite {beznosikov2020biased, Stich-EF-NIPS2018}.

\begin{definition}[Unbiased Compression Operator]
	\label{ch1:def:omegaquant} A randomized mapping $\cC\colon \R^d \to \R^d$  is an {\em unbiased compression operator (unbiased compressor)}  if there exists $\omega \geq 0$ such that
	\begin{equation}
		\label{ch1:eq:omega_quant}
		\Exp{\cC(x)}=x, \qquad \Exp{\norm{\cC(x)}^2} \leq (\omega + 1) \norm{x}^2, \qquad \forall x \in \R^d.
	\end{equation}
\end{definition}

\begin{remark}
	As $\Exp{ \norm{X-\Exp{X}}^2} = \Exp{\norm{X}^2} - \norm{\Exp{X}}^2$ for any random vector $X$, condition~\eqref{ch1:eq:omega_quant} implies 
	\begin{align}
		\Exp{\norm{\cC(x) - x}^2} \leq \omega \norm{x}^2\,, \qquad \forall x \in \R^d. \label{ch1:def:omega}
	\end{align}
\end{remark}

\begin{definition}[Contractive Compression Operator]
	\label{ch1:def:contractive} A (possibly) randomized mapping $\cC\colon \R^d \to \R^d$  is a {\em contractive compression operator} if there exists $\alpha \in (0,1]$ such that
	\begin{equation}
		\label{ch1:eq:quant} 
		\Exp{\norm{\cC(x) - x}^2} \leq \lp 1 - \alpha \rp \norm{x}^2, \qquad \forall x \in \R^d.
	\end{equation}
\end{definition}

Sparsification is a particularly popular technique for compressing vectors. Below, we include an example that leads to an unbiased compression operator and a contractive compression operator.

\begin{example}[Random sparsifiction (\compname{RandK})]
	The random sparsification operator $\cC:\R^d\to\R^d$ with sparsity parameter $k \in \{1,2, \hdots, d\}$ is defined by
	$$\cC(x) \eqdef \dfrac{d}{k} \cdot \sum_{i=1}^{d} \left(\xi_{i} \cdot x_i\right) e_i$$ 
	where $\xi \sim_{\rm u.a.r.} \{y \in  \{0,1\}^d \colon \norm{y}_0 = k\}$ is a random vector with $k$ non-zero elements (i.e.,  $\norm{y}_0 = k$) equal to one sampled uniformly at random (u.a.r.). The $\{e_i\},i \in [d]$ is unit-vectors which constitutes the standard basis of $\RD$. The \compname{RandK} is unbiased compression operator (Definition~\ref{ch1:def:omegaquant}) with $\omega = \dfrac{d}{k}-1$.
\end{example}

\begin{example}[Greedy (aka \compname{TopK}) sparsifiction]
	The \compname{TopK} sparsification operator is defined as
	\begin{equation*}
		\cC(x) \eqdef \sum \limits_{i=d-k+1}^d x_{(i)} e_{(i)},
	\end{equation*}
	where $\{e_i\},i \in [d]$ is unit-vectors which constitutes the standard basis of $\RD$. The $x_{(j)}$ is the $j$-th largest coordinate in magnitude, i.e., $\lvert{x_{(1)}}\rvert \leq \lvert{x_{(2)}}\rvert \leq \cdots \leq \lvert{x_{(d)}}\rvert$, and $k \in \{1,2, \hdots, d\}$ is a sparsity parameter. The $e_{(j)}$ is the unit vector which corresponds to the $x_{(j)}$ is the $j$-th largest coordinate in magnitude. The \compname{TopK} is contractive compression operator (Definition~\ref{ch1:def:contractive}) with $\alpha = \dfrac{k}{d}$.	
\end{example}

\chapter{FL\_PyTorch: A Federated Learning Research Simulator}
\label{chapter2}

The goals and summaries of this chapter are outlined in Table \ref{ch1:tbl:algorithms} and Section~\ref{ch1:sec:overview-2}.

\section{Introduction}
Over the past few years, the continual development of DL capabilities has revolutionized the way we interact with everyday devices. Much of this success depends on the availability of large-scale training infrastructures such as large GPU clusters and the ever-increasing demand for vast amounts of training data.
In contrary to this, users and providers are becoming increasingly aware of their privacy leakage because of this centralized data collection, leading to the creation of various privacy-preserving initiatives by industry service providers~\citep{apple} or government regulators~\citep{gdpr}.

Recently, a viable solution that has the potential to address the aforementioned issue is Federated Learning (FL). FL a term was initially proposed by \citet{mcmahan17fedavg} as an approach to solving learning tasks by a loose federation of mobile devices. However, the underlying concept of training models without data being available in a single location is applicable beyond the originally considered scenario and it turns out to be useful in other practical use cases. For example, learning from institutional data silos such as hospitals or banks which cannot share data due to confidentiality or legal constraints, or applications in edge devices~\citep{yang2019federated, horvath2021fjord}.

The main goal of FL is to provide strong privacy protection, which is obtained by storing data locally rather than transferring it to the central storage. To solve the underlying machine learning objective, each client provides focused updates intended for immediate aggregation. Stronger privacy properties are possible when FL is combined with other technologies such as differential privacy~\citep{dwork2008differential} and secure multiparty computation (SMPC) protocols such as secure aggregation~\citep{bell2020secagg}. 

Recently, Federated Learning has seen increasing attention not only in academia but also in the industry, which already uses FL in its deployed systems. For instance,
Google uses it in the Gboard mobile keyboard for applications including next word prediction~\citep{hard18gboard}, emoji suggestion~\citep{gboard19emoji} or “Hey Google” Assistant~\citep{googleassistant2021}. Apple uses FL for applications like the QuickType keyboard and the vocal classifier for “Hey Siri”~\citep{apple19wwdc}. In the finance space, FL is used to detect money laundering~\citep{webank2020} or financial fraud~\citep{intel2020}. In the medical space, Federated Learning is used for drug discovery~\citep{melloddy2020}, for predicting COVID-19 patients’ oxygen needs~\citep{nvidia2020} or medical images analysis~\citep{owkin2020}.

\iftrue
	
	\begin{algorithm}[t]
		\caption{Generalized Federated Averaging.}
		\label{ch2:algo:generalized_fedavg}
		\begin{algorithmic}[1]
			\STATE \textbf{Input:} Initial model $\vxflp^{(0)}$, \textsc{ClientOpt}, \textsc{ServerOpt}
			\STATE {\bfseries Initialize:} Initialize server state $H^{0}=\InitializeServerState()$
			\FOR{$t = 0$ to $T-1$}
			\STATE Sample a subset $\activeClients^{(t)}$ of available clients
			\STATE Generate state: $s^{(t)} = \ClientState(H^{(t)})$
			\STATE Broadcast $(\vxflp^{(t)},s^{(t)})$ to workers
			\FORALL{client $i \in \activeClients^{(t)}$ \textbf{in parallel}}
			\STATE Initialize local model $\vxflp_i^{(t,0)}=\vxflp^{(t)}$
			\FOR{$k = 0$ to $\localStep_i -1$}
			\STATE Compute local stochastic gradient $\sgrad_i =\LocalGradient(\vxflp_i^{(t,k)}, s_t)$
			\STATE Perform local update $\vxflp_i^{(t,k+1)} = \textsc{ClientOpt}(\vxflp_i^{(t,k)}, \sgrad_i, {k}, t)$
			\ENDFOR
			\STATE Compute local model changes $\localChange_i^{(t)} = \vxflp_i^{(t,\localStep_i)} - \vxflp_i^{(t,0)}$
			\STATE Create local state update: $U_i^{(t)} = \textsc{LocalState}(\vxflp^{(t)}, \vxflp_i^{(t,\localStep_i)})$
			\STATE Send $(\localChange_i^{(t)},U_i^{(t)})$ to Master
			\ENDFOR
			\STATE Obtain $(\localChange_i^{(t)},U_i^{(t)})$ for all $i \in \activeClients^{(t)}$
			\STATE Compute $G^{(t)} = \ServerGradient(\{\localChange_i^{(t)},U_i^{(t)}\}_{i \in S^{(t)}}, H^{(t)})$
			\STATE Update global model $\vxflp^{(t+1)} = \textsc{ServerOpt}(\vxflp^{(t)}, G^{(t)},\slr,t)$
			\STATE Update: $H^{(t+1)}=\ServerGlobalState(\{\localChange_i^{(t)},U_i^{(t)}\}_{i \in S^{(t)}}, H^{(t)})$
			\ENDFOR
		\end{algorithmic}
	\end{algorithm}

\fi

To enable research in Federated Learning, several frameworks have been proposed including \libname{LEAF} \citep{caldas2018leaf}, \libname{FedML} \citep{he2020fedml}, \libname{Flower} \citep{beutel2020flower}, (\libname{PySyft} \citep{ryffel2018generic}, \libname{TensorFlow-Federated} (TFF) \citep{TFF2019}, \libname{FATE} \citep{yang2019federated}, \libname{Clara} \citep{ClaraTraining}, \libname{PaddleFL} \citep{ma2019paddlepaddle}, \libname{Open FL} \citep{OpenFLFramework}. These frameworks are mainly built with a focus on being deployed on real-world systems while also providing users with the ability to run experiments with the same code. This desired property often comes with the price that the entry bar for researchers to extend or experiment with these frameworks is limited in the sense that they either need to have extensive experience with distributed systems or require assistance from experts on the given framework.

\subsection{FL Objective and FedAVG}
\label{ch2:sec:gen_objective}
In this section, we introduce the FL objective in its general form
\begin{align}
	\obj(\vxflp) = \E_{i \sim \clientDist}[ \obj_i(\vxflp)] \quad \text{where} \ \obj_i(\vxflp) = \E_{\xi \in \data_i}[f_i(\vxflp, \xi)] + R(x). \label{ch2:eqn:global_obj}
\end{align}

The global objective $\obj$ is an expectation over local objectives $\obj_i$ over the randomness inherited from the client distribution $\clientDist$, and the local objectives $\{\obj_i\}$ have the form of an expectation over the local datasets $\{\data_i\}$. The $\{\obj_i\}$ has an additional regularization term that is useful for incorporating prior knowledge of parameters to find $x$.

In the case of the finite number of clients and local data points, both global objectives $\obj$ and local losses $\{\obj_i\}$ can be written as simple weighted averages in the empirical risk minimization form (ERM). The most common algorithm to solve \eqref{ch2:eqn:global_obj} is federated averaging \algname{FedAVG}~\citep{mcmahan17fedavg}. This algorithm divides the training process into communication rounds. At the beginning of the $t$-th round ($t \geq 0$), the server broadcasts the current global model $\vxflp^{(t)}$ to a random subset of clients $\activeClients^{(t)}$ (often uniformly sampled without replacement in simulations). Then, each sampled client performs $\localStep_i$ local \algname{SGD} updates on its own local dataset and sends its local model update $\Delta_i^{(t)}=\vxflp_i^{(t,\localStep_i)}-\vxflp^{(t)}$ to the server. Finally, the server uses the aggregated model updates to obtain the new global model as follows:
\begin{align}
	\vxflp^{(t+1)} 
	= \vxflp^{(t)} + \dfrac{\sum_{i \in \activeClients^{(t)}} p_i \Delta_i^{(t)}}{\sum_{i \in \activeClients^{(t)}} p_i}. \label{ch2:eqn:upadte_fedavg}
\end{align}
where $p_i$ is the relative weight of client $i$. The above procedure will repeat until the algorithm converges. In the \emph{cross-silo} setting where all clients participate in the training at every round, we have $\activeClients^{(t)}=\{1,2,\dots,\numClients\}$. 

\subsection{Contributions}

In our work, we decided to take one step back and focus on the construction of a framework that, although not aimed for being deployed to edge devices as the primary goal, can provide a useful simulation environment for researchers with the following goals:
\begin{enumerate}
	\item \textbf{{Low entry bar / Simplicity.}} We aim our tool to be as simple as possible for the user while providing all necessary functionalities. 
	
	\item \textbf{{Extensibility.}} It is easy to bring your own algorithm or dataset or extend existing ones. We aim to achieve this by providing universal abstractions with a sufficient level of flexibility to experiment with existing and novel approaches to advance the state-of-the-art.
	
	\item \textbf{{Hardware utilization.}} Cross-device FL experiments are usually of a much smaller scale compared to the centralized setting. This is mainly because of limited device capabilities. Running such experiments on GPU can lead to the under-utilization of available hardware. We aim to resolve this via the ability to parallelize clients' computation.
	
	\item \textbf{{Easy debugging.}} Debugging multi-process or multi-thread systems is hard. We only require the user to provide a single thread implementation which is automatically adjusted to multi-GPU and multi-node setup as a separate step attainable from the configuration process.
	
\end{enumerate}

To the best of our knowledge, there is no such tool that could provide a sufficient level of freedom and is simple to use, and therefore, we design \fl to achieve the aforementioned goal.

\fl is an optimization research simulator for FL implemented in Python based on \libname{PyTorch}~\citep{paszke2019pytorch}. \fl is a simple-to-use tool with its own Graphical User Interface (GUI) implemented in \libname{PyQt5}~\citep{pyqt_docu}. During the simulation process, the selected local CPUs/GPUs are accessed in a parallel way. In addition, there is a possibility to use remote CPUs/GPUs. Remote devices are required to have a \abr{TCP} transport layer for communicating with the master. Regarding supported devices, we target server and desktop stations running on Linux, macOS, or Windows OS for efficient simulations.

For the chapter organization, we introduced the general FL minimization objective in Section~\ref{ch2:sec:gen_objective}. Subsequently, we provide a deep dive into our \fl system in Section~\ref{ch2:sec:fl_pytorch} and, finally, we demonstrate \fl capabilities by concluding several experiments on multiple FL baselines in Section~\ref{ch2:sec:exp}. 

\section{FL Optimization Simulator}
\label{ch2:sec:fl_pytorch}
\fl is a system that has been built using Python programming language, and it is based on the DL framework \libname{PyTorch}. We have made the code repository publicly available at {\href{https://github.com/burlachenkok/flpytorch}{https://github.com/burlachenkok/flpytorch}}.

Its backbone consists of a general form of \algname{FedAVG} displayed in Algorithm~\ref{ch2:algo:generalized_fedavg} partially inspired by Algorithm 1 in~\citet{reddi2020adaptive}. Our proposed general form preserves the standard structure of federated optimization where in each round $t$, subset $\activeClients^{(t)}$ of all available clients is selected. As a next step, the master generates its state $s_t$ which is broadcasted together with the current copy of the global solution $\vxflp^{(t)}$ to the selected subset of clients $\activeClients^{(t)}$. Afterwards, each participating client initializes its local solution $\vxflp_i^{(t,0)}$ to be a copy of the global solution $\vxflp^{(t)}$. Each of these clients then performs $\tau_i$ steps using the local optimizer $\clientopt$ with gradient estimated by the $\LocalGradient$ function. After this step, each client computes the model and state update, which are then sent back to the master, which estimates the global update direction using the $\ServerGradient$ function. This estimate is used in $\serveropt$ that updates the global solution to its new value $\vxflp^{(t+1)}$. Lastly, the server's global state is updated. In Section~\ref{ch2:sec:algorithms}, we show that this general scheme captures all standard algorithms, thus our scheme is sufficient and we believe that it gives researchers a sufficient level of freedom to develop and experiment with each component of our general scheme to push both practical and theoretical FL state-of-the-art. 

Our current implementation is fully defined and allows us to configure up to 51 parameters, which can be specified in one of the following ways: (i) through our user-friendly GUI tool, (ii) directly via the command line, as discussed in the next subsection, or (iii) by converting the originally configured parameters from the GUI into Command Line Interface format.

These parameters can be grouped into 4 categories based on their function: 
\begin{itemize}
	\item \textit{Server Optimizer.} The number of communication rounds $T$; the number of clients sampled per round; the global initial local learning rate; the global learning rate schedule; the global optimizer and its parameters, such as momentum; and the name of the algorithm to be executed as the backbone, on top of which modifications are applied.
	
	\item \textit{Local Optimizer.} The number of local epochs or local iterations $\tau_i$; batch size for data loading; local optimizer with its parameters.
	
	\item \textit{Model and Data.} The model's and dataset's names.
	
	\item \textit{System Setup.} Directory to store run metadata, target compute devices, usage of remote computing devices, logging level, number of workers for loading dataset, random seed, thread pool sizes, experiments grouping, user-defined comment, enable usage of NVIDIA Ampere GPU Tensor Cores, optional cleanup of \libname{PyTorch} GPU cache at the end of each round, device for store client state.
	
\end{itemize}

\subsection{Front end}

As we mentioned previously, \fl supports two modes for running federated optimization--Graphical User Interface (GUI) and Command Line Interface (CLI).

\subsubsection{Graphical user interface (GUI)}
We implemented our Graphical User Interface using \libname{PyQt5} GUI framework~\citep{pyqt_docu}. This GUI framework supports all the standard desktop operating systems such as  Windows, Linux, or macOS. In addition, the GUI part of \fl has built-in VNC server support. If the target machine does not have a native Windows manager system, one can use this mode and connect to the GUI part via VNC Viewer software.

\begin{figure*}[ht!]
	\centering
	\begin{subfigure}[ht]{0.7\textwidth}
		\includegraphics[width=\textwidth]{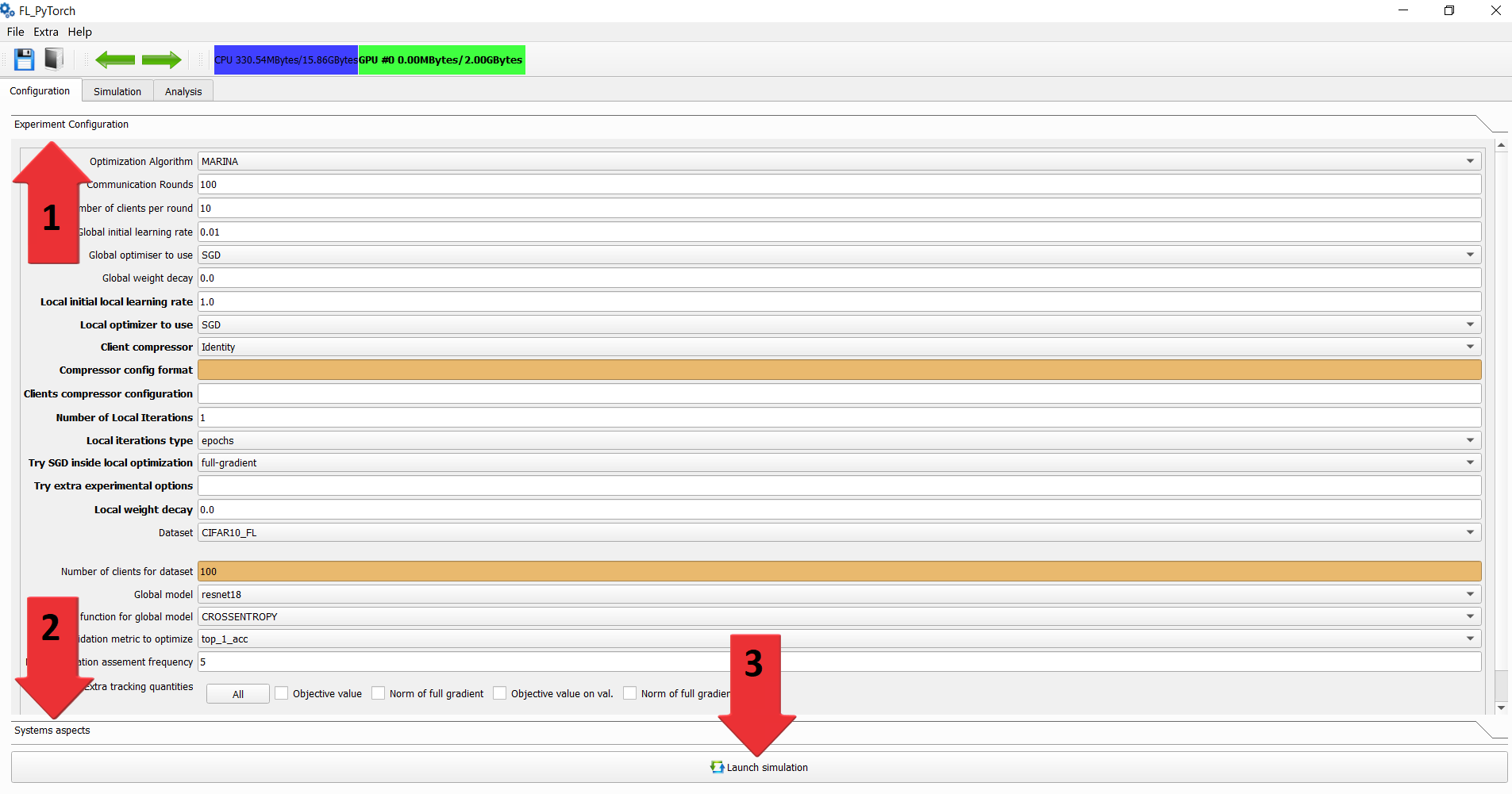} 
		\caption{The main tab of the GUI}
	\end{subfigure}
	
	\begin{subfigure}[ht]{0.7\textwidth}
		\includegraphics[width=\textwidth]{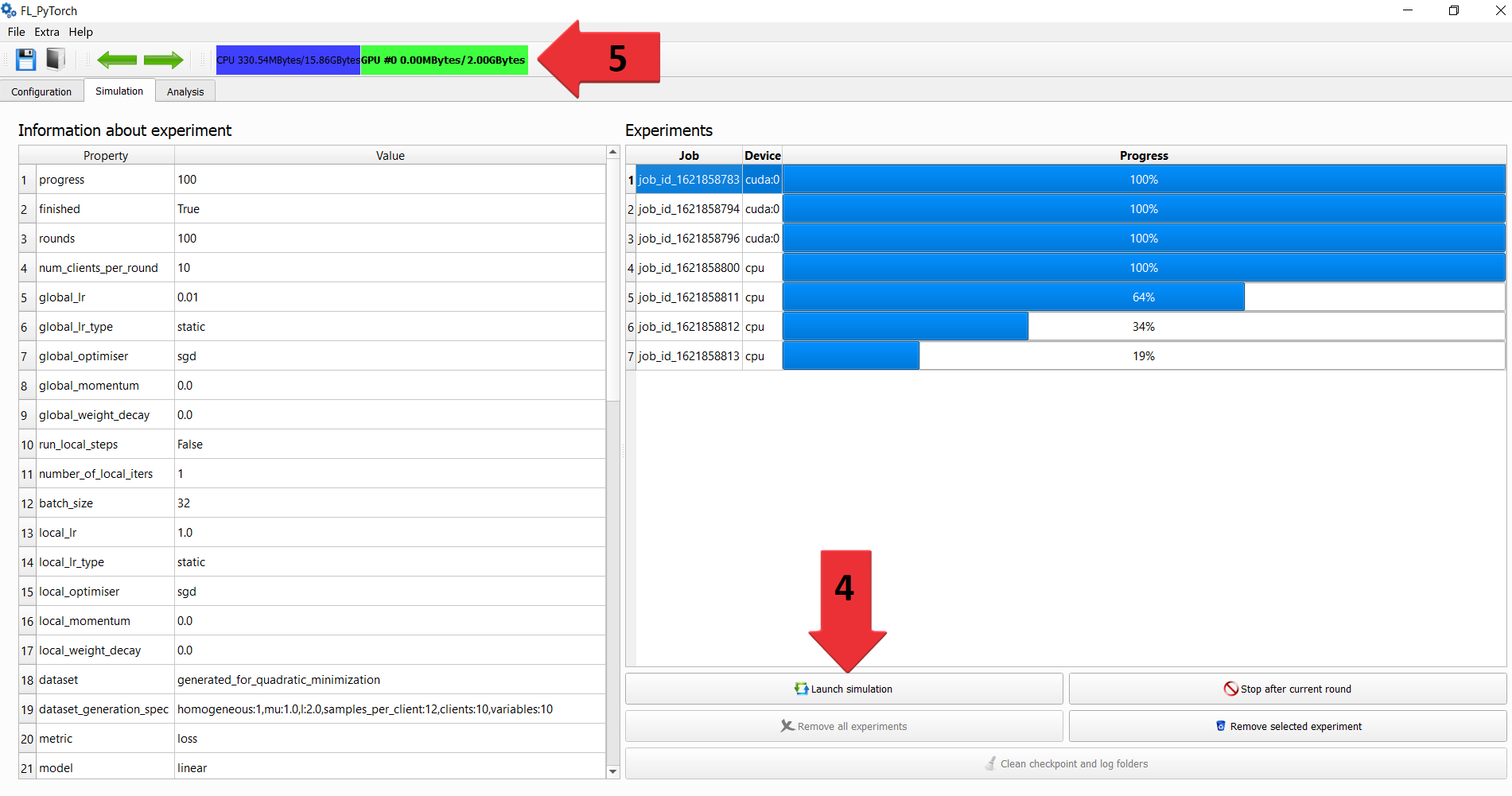} 
		\caption{The progress tab}
	\end{subfigure}
	
	\begin{subfigure}[ht]{0.7\textwidth}
		\includegraphics[width=\textwidth]{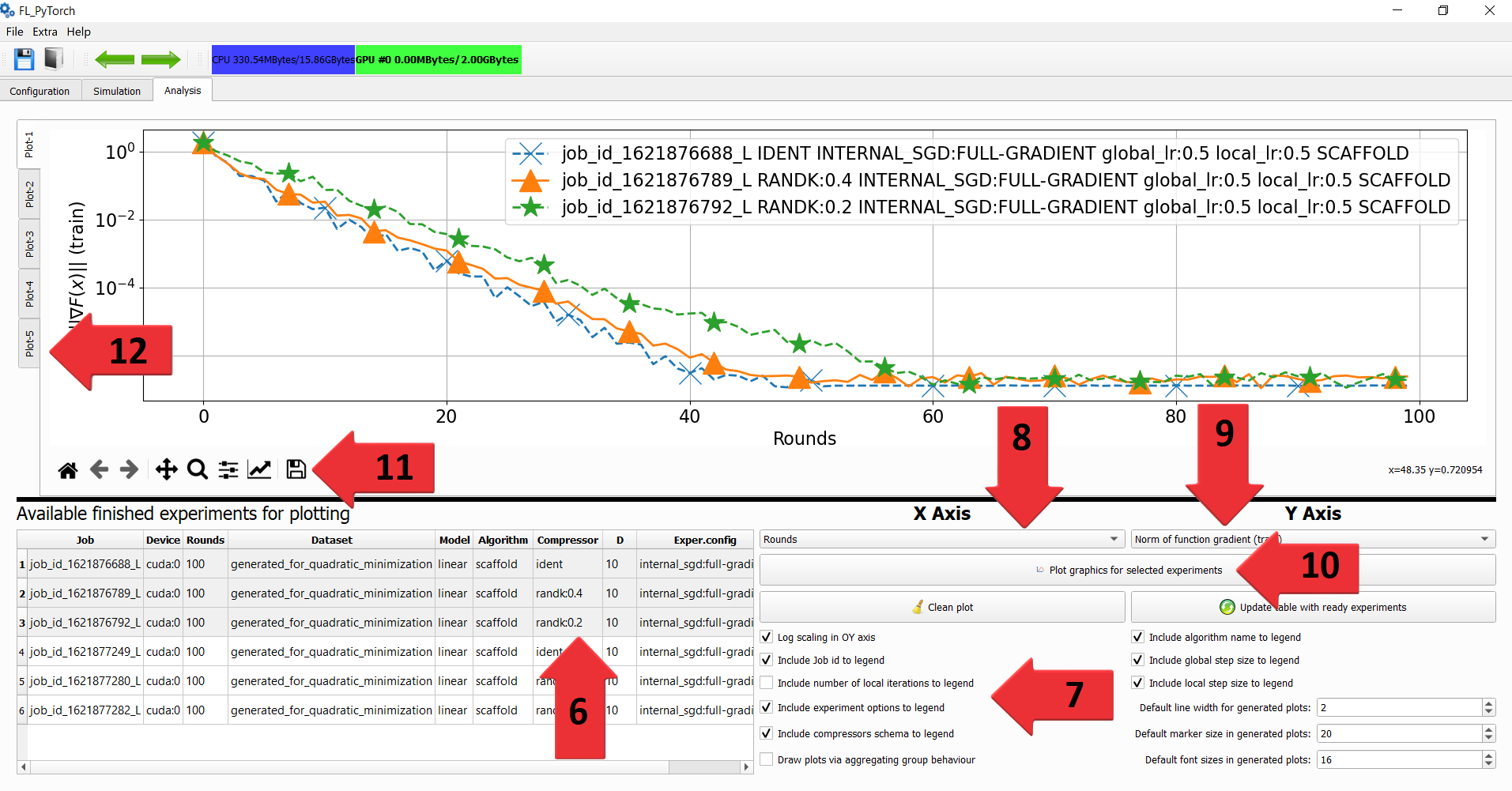}
		\caption{The Analysis tab}
	\end{subfigure}
	
	\begin{subfigure}[ht]{0.55\textwidth}
		\includegraphics[width=\textwidth]{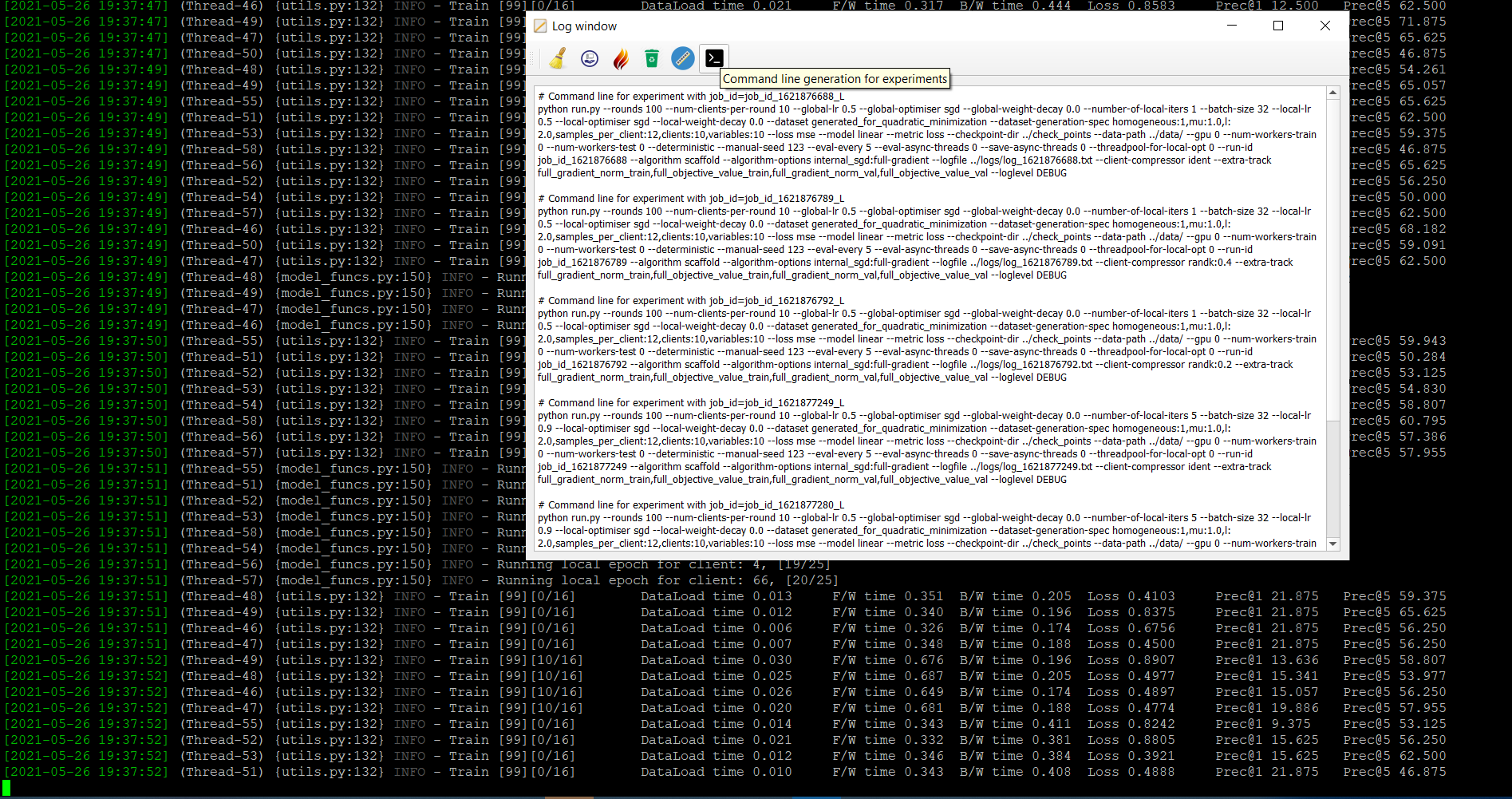} 
		\caption{Console output log}
	\end{subfigure}
	
	\caption{Graphical User Interface (GUI) of the \fl simulator.}
	\label{ch2:fig:fl_gui_simulation_gui}
\end{figure*}

To get a better picture of our GUI, we display its main menu provided in Figure~\ref{ch2:fig:fl_gui_simulation_gui} (a, b, c). The components of our GUI help users to build their run setup with the steps depicted as numbered red arrows: 1.) experiment configuration,  2.) system setup configuration, 3.) thread pool setup, 4.) button to launch experiments,  5.) navigation pane, 6.) system monitoring, 7.) plotting setup -- experiments selection, 8.) plotting setup -- format 8.) plotting setup -- x-axis, 9.) plotting setup -- y-axis, 10.)   plotting setup -- generate plots, 11.) plotting setup --  save option, 12.)  plotting setup -- clean output option.


%
%
%

To obtain information regarding the currently used system and environment, the user can select GUI Help$\to$About.

Some extra configurable displayed information from the GUI includes a number of currently executing experiments, showing the available size of physical DRAM and swap memory in OS.

\subsubsection{Command line interface (CLI)}
Another alternative option to pass the run arguments is directly through the console via the Command Line Interface (CLI). 
We also assist in seeing the available arguments' options that can be accessed by launching \texttt{./run.sh} script provided in the supplementary materials with the \texttt{--help} option. 

To facilitate an efficient switch between GUI and CLI, we provide an option to retrieve arguments passed through the GUI tool in command line format, as shown in Figure~\ref{ch2:fig:fl_gui_simulation_gui}. This feature allows the generation of a command line for both (i) completed experiments and (ii) experiments that are currently configured in the GUI but have not yet been launched. The CLI interface is used also to instantiate remote workers in remote machines, to provide extra remote compute resources for the simulation process. 

The simulator can be used from the CLI or GUI, allowing users to monitor experiment progress in \libname{WandB} online plotting tool \citep{wandb}, due to its integration with the \fl simulator. To perform early stopping, users should send a POSIX SIGTERM or SIGINT signals~\citep{kerrisk2010linux}. In this case, the experiment will terminate, but simulation results will be properly saved. While \libname{WandB} is useful for monitoring experiments, professional plots often require more advanced customization. To address this, the GUI component of our system supports generating specific plots using \libname{Matplotlib}~\citep{barrett2005matplotlib}, with the ability to customize them directly within the GUI, as shown in Figure~\ref{ch2:fig:fl_gui_simulation_gui} (c).

\subsection{Optimization algorithms}
\label{ch2:sec:algorithms}
As discussed previously, our general level of abstraction introduced in Algorithm~\ref{ch2:algo:generalized_fedavg} allows for a sufficient level of freedom to implement standard and also more exotic FL optimization algorithms. In the current version, we implemented following state-of-the-arts methods considered in the literature: Distributed Compressed Gradient Descent (\algname{DCGD})~\citep{alistarh2017qsgd,khirirat2018distributed,horvath2019natural} \algname{FedAVG}~\citep{mcmahan17fedavg}, \algname{SCAFFOLD}~\citep{karimireddy2020scaffold}, \algname{FedProx}~\citep{li2018federated}, \algname{DIANA}~\citep{mishchenko2024distributed}, \algname{MARINA}~\citep{gorbunov2021marina}, \algname{PP-MARINA}~\citep{gorbunov2021marina}, \algname{EF21}~\citep{richtarik2021ef21}, \algname{EF21-PP}~\citep{fatkhullin2021ef21}, \algname{COFIG} and \algname{FRECON}~\citep{zhao2021faster}.

Some optimization algorithms require a client to store in its internal state-specific information about algorithm-specific shifts that have a notion about the gradient estimator. The analysis is very often irrelevant to the specific strategy of initializing shifts, but in practice, different initialization policies may have consequences in convergence speed. We provide two policies with the initializing shift by zero or by full gradient at $x^{(0)}$.

As an example, we show how Algorithm~\ref{ch2:algo:generalized_fedavg} is adjusted to \algname{FedAVG} and \algname{SCAFFOLD}. For \algname{FedAVG}, both global states are empty dictionaries. Locally, we run $\tau_i$ iterations, which is usually set to be a constant $T$ for the theoretical and size of the local dataset for the experiments. $\LocalGradient$ returns the unbiased stochastic gradient at $\vxflp^{(t,k)}$ and $\clientopt$ is Stochastic Gradient Descent (\algname{SGD}) with given step size $\eta_l$. Similarly to the global state, local step update is none. $\ServerGradient$ is a (weighted) average of local updates and $\serveropt$ is \algname{SGD} with fixed step size $1$. For \algname{SCAFFOLD}, there are $2$ changes compared to \algname{FedAVG}. Firstly, the global state $s^{(t)}$ is a non-empty vector of the same dimension as $\vxflp^{(t)}$ initialized as zero, and each $\clientopt$ has its own local  $s_i^{(t)}$, which is set to be a stochastic local gradient at $\vxflp^{(t)}$. $\LocalGradient$ then returns the unbiased stochastic gradient at $\vxflp^{(t,k)}$ shifted by $s_i^{(t)} - s_i^{(t)}$. Secondly, the local model update is sent to the master together with the local state update, which is set to the difference between current and previous round $U_i^{(t)} = s_i^{(t)} - s_i^{(t-1)}$. The $\ServerGlobalState$ is then updated using the average of $U_i^{(t)}$'s.

\subsection{Supported compressors}
In FL (especially in the cross-device setting), clients may experience severe network latency and bandwidth limitations. Therefore, practical FL algorithms generally use a communication reduction mechanism to speed up the training procedure. Three common methods to reduce the communication cost are:
\begin{enumerate}
	\item Reduce the communication frequency by allowing local updates.
	\item Reduce communication traffic by limiting the participating clients per round.
	\item Reduce communication volume by compressing messages.
\end{enumerate} 
We naturally support the first and the second option, and we also added support for the compression. \fl allows compressing messages both from the server to local clients and vice-versa. We support several unbiased and biased compressors. The supported unbiased compressor are: Identical compressor (no compression), \compname{Bernoulli} or \compname{Lazy} compressor (update is communicated with probability $p$ rescaled for the update to be unbiased), \compname{RandK} (only random $K$ coordinates are preserved uniformly at random rescaled for the update to be unbiased), \compname{Natural compressor}~\citep{horvath2019natural}, \compname{Standard dithering}~\citep{alistarh2017qsgd}, \compname{Natural Dithering}~\citep{horvath2019natural}, \compname{Terngrad}~\citep{DBLP:conf/nips/WenXYWWCL17} and \algname{QSGD compressor}~\citep{alistarh2017qsgd}. The supported biased compressor are: \compname{TopK}~\citep{beznosikov2020biased} and \compname{RankK}~\citep{safaryan2021fednl}. For \compname{TopK} and \compname{RankK} compressors, parameter $K$ can be specified as a percentage in $[0,100]$ from $d$ or as an absolute number greater than $1$.

Also, we provide means to construct new compressors via function composition and probabilistic switching from existing ones.


\begin{figure*}[t]
	\centering
	\begin{subfigure}[ht]{0.65\textwidth}
		\includegraphics[width=\textwidth]{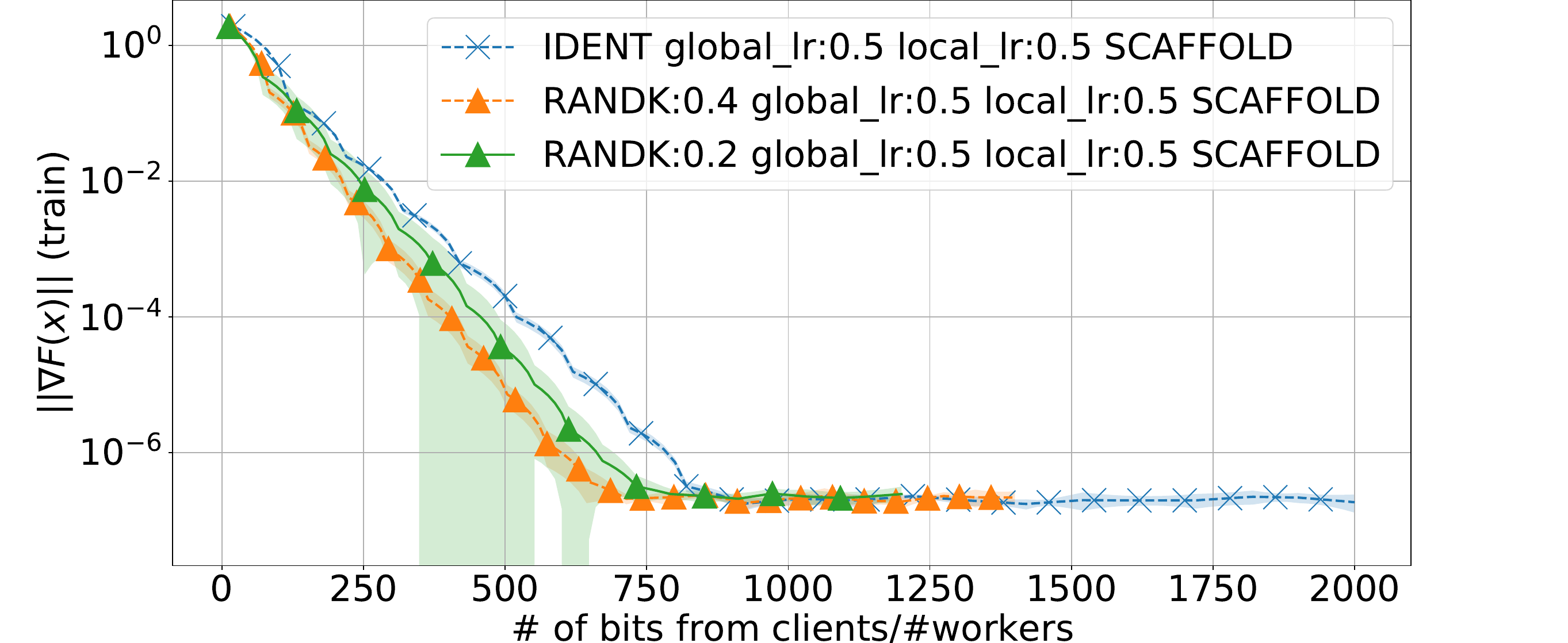} 
		\caption{One local iteration}
	\end{subfigure}
	
	\begin{subfigure}[ht]{0.65\textwidth}
		\includegraphics[width=\textwidth]{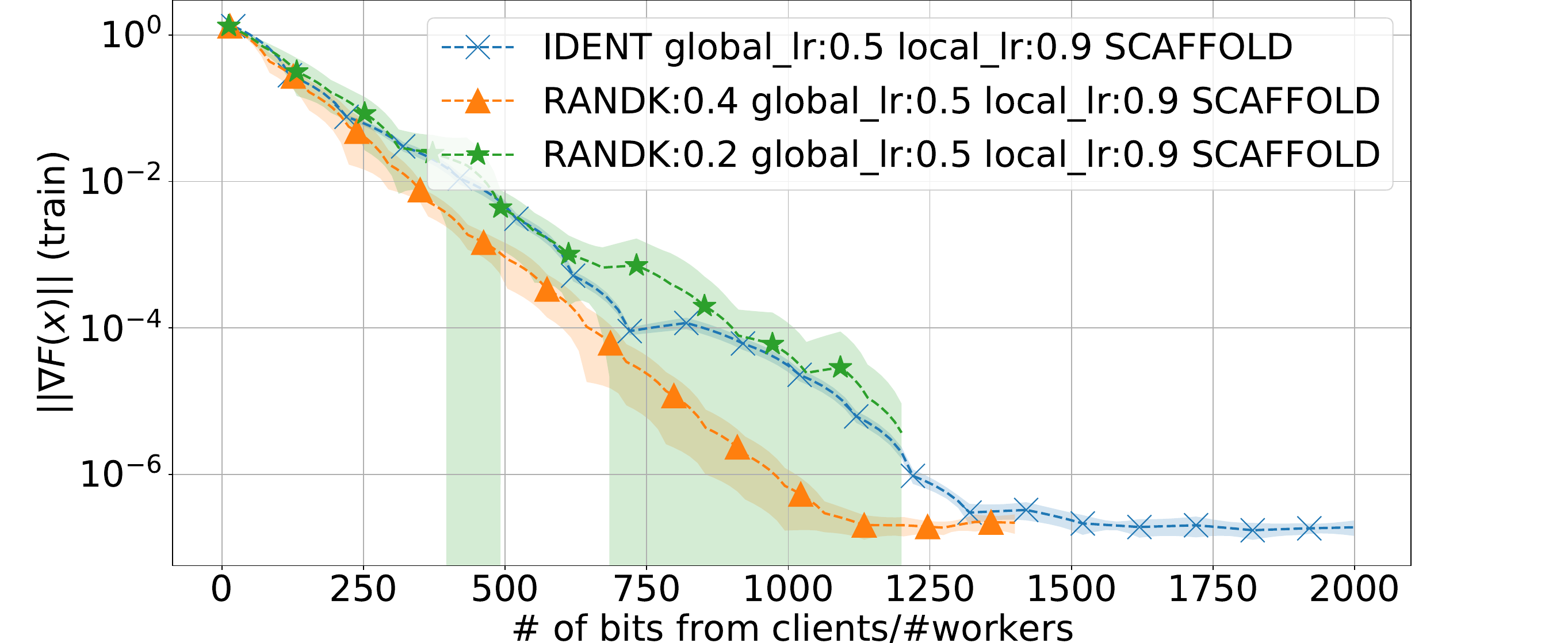}
		\caption{Five local iterations}
	\end{subfigure}

	
	
	\caption{Function gradient diminishing for 1 and 5 local iterations of \algname{SCAFFOLD} for \modelname{quadratic} minimization. Mean and variance across 10 realizations.}
	\label{ch2:fig:scaffold_experiment_convex}
\end{figure*}

\subsection{Supported models (or pattern) structures}

Current \fl's implementation allows users to experiment with the following list of image classification models: 
\modelname{ResNets}~\citep{resnet} (18, 34, 50, 101, 152), \modelname{VGGs}~\citep{vgg} (11, 13, 16, 19), \modelname{WideResNets}~\citep{wideresnet} (28\_2, 28\_4, 28\_8), and \modelname{Logistic Regression}. 

Besides DL models, we provide support for simple quadratic loss function to enable users to explore algorithms in this simplistic regime or debug their implementation. 
We synthetically generate our objective such that local loss is of the form
\begin{equation}
	\label{ch2:eq:synthetic_quadratic}
	\min_{x \in \RD} \dfrac{1}{n_i} \sum_{i=1}^{n_i} \| a_i^\top x - b_i\|^2.
\end{equation}
Users can specify: dimensionality $d$, strong convexity parameter $\mu$, and $L$ smoothness constant for \eqref{ch2:eq:synthetic_quadratic} and the number of samples per client. In addition, we consider two settings. In the first case, all clients optimize the same objective in an independent and identically distributed (i.i.d.) setting. In the second case, each client has a different objective to optimize, in other words, a non-i.i.d. setting.

\subsection{Tracking metrics and experiments analysis}
An important feature to evaluate experiments is the ability to track all important quantities during the run of the algorithms. \fl allows users to track the number of communication rounds, loss, accuracy, the norm of computed gradient, the norm of the full objective gradient, function values, number of gradient oracle calls, used GPU memory, number of bits which would be sent from workers to master in an actual real-world system, number of clients per round, wall-clock execution time. Furthermore, we provide a dedicated visualization tab in our GUI tool which allows the user to load and visualize their experiments interactively in several plots. User can observe such quantities during experiment execution via two-dimensional plots.

Each experiment is associated with a wealth of information. We provide tools to configure the view of experimental results by selecting relevant attributes, sorting experiments by specific attributes in the analysis tab, and manually reordering experiments. All of this can be accomplished through the interface shown in Figure~\ref{ch2:fig:fl_gui_simulation_gui} (c). Additionally, experiments with the same group name can be grouped together, with error bars plotted based on the statistics of the selected set of grouped experiments.

\subsection{The internals of the system}
To achieve our goal of making \fl computationally efficient on GPUs, we analyzed various aspects of \libname{PyTorch} infrastructure, including its position within NVIDIA's computation stack, its initialization process, forward and backward computation, and the speed of typical buses in a local node setup. This analysis led us to the system design, which asynchronously exploits available hardware. 

The Algorithm~\ref{ch2:algo:generalized_fedavg} has several independent parts, and instead of sequential execution of these parts in a simulated environment, independent parts can be partitioned across independent thread-pools of workers. In our implementation, each worker within a thread pool is a separate CPU thread that lives within a Python interpreter process launched in the operating system (OS) if the system's resources allow it. Since \libname{PyTorch} and therefore \fl allow launching of the computations in a GPUs, for providing separate CPU threads ability to work independently, each thread has its separate GPU CUDA stream to submit its computation work for GPU independently. By design, each worker with the same purpose is assigned to its task-specific thread pool with a thread-safe tasks queue. There are three specific types of tasks -- deferred execution, process request for finish execution, and wait on the completion of all submitted work.

Finally, we emphasize that all of this happens "behind the curtain", with user interaction—including the implementation of new methods—remaining agnostic to the implementation details presented in Appendix~\ref{ch2:app:fl_vis}.

\subsection{Bringing custom algorithm / model / data}
\label{ch2:sec:custom_alg}

As described in Algorithm~\ref{ch2:algo:generalized_fedavg}, each algorithm supported by \fl requires implementation of \InitializeServerState,  \ClientState, \LocalGradient, \textsc{ClientOpt}, \textsc{LocalState}, \textsc{ServerOpt}, \ServerGradient, \ServerGlobalState. These are standard centralized-like \libname{PyTorch} functions that need to be provided to \fl to run simulations. 

One example of adding a new algorithm might be \algname{SCAFFOLD} assuming that we are given implementation of \algname{FedAVG} in the required Algorithm~\ref{ch2:algo:generalized_fedavg} format. As described in Section~\ref{ch2:sec:algorithms}, one needs to set \ClientState to return global shift as proposed by \citet{karimireddy2020scaffold} and update \LocalGradient to account for both global and local shifts. One further needs to define \textsc{LocalState} to return a local shift update, see \citep{karimireddy2020scaffold}  for the detailed formula.  The last modification is in \ServerGlobalState, which updates the global shift using the local shift updates. 

For the datasets, users are required to provide them in a format compatible with \libname{PyTorch}'s \libname{DataLoader}, and their models must inherit from \libname{PyTorch}'s \libname{nn.Module}\footnote{Detailed manuals are provided in the project repository: \href{https://github.com/burlachenkok/flpytorch}{github.com/burlachenkok/flpytorch}}.

\section{Experiments}
\label{ch2:sec:exp}
In this section, we provide several experiments to showcase potential use cases of \fl. In general, \fl can serve as a simulator for researchers to test and compare different algorithms, extend them, and analyze each algorithmic component in a plug-and-play fashion as described in Section~\ref{ch2:sec:fl_pytorch} and Algorithm~\ref{ch2:algo:generalized_fedavg}. 

For instance, one might be interested in how to reduce the communication burden of a given algorithm. In the literature, three approaches have been proposed. For the first approach, one performs local updates to reduce the frequency of the global model updates, thus also reducing communication frequency. With this strategy,  communication cost per client model update can be effectively reduced by a number of local steps. However, while often in practice this leads to improvement, it is not always guaranteed, as it was theoretically shown that there is a trade-off between communication reduction and convergence~\citep{charles2021convergence}. Another possibility is to employ compression techniques that reduce the bits transmitted between the clients and the server. This compression operator should be selected carefully as it comes with an extra variance and it might affect the convergence. The third option is to employ partial participation, for which in each round only a sampled subset of clients participates. There is also a trade-off for this strategy as although having only a few clients in each round has a positive effect on reducing communication, it might negatively affect the quality of the obtained solution due to inexact updates based only on sub-sampling. \fl supports experimentation with all three approaches. Below, we provide three examples to illustrate this.


\begin{figure*}[t]
	\centering
	\begin{subfigure}[ht]{0.49\textwidth}
		\includegraphics[width=\textwidth]{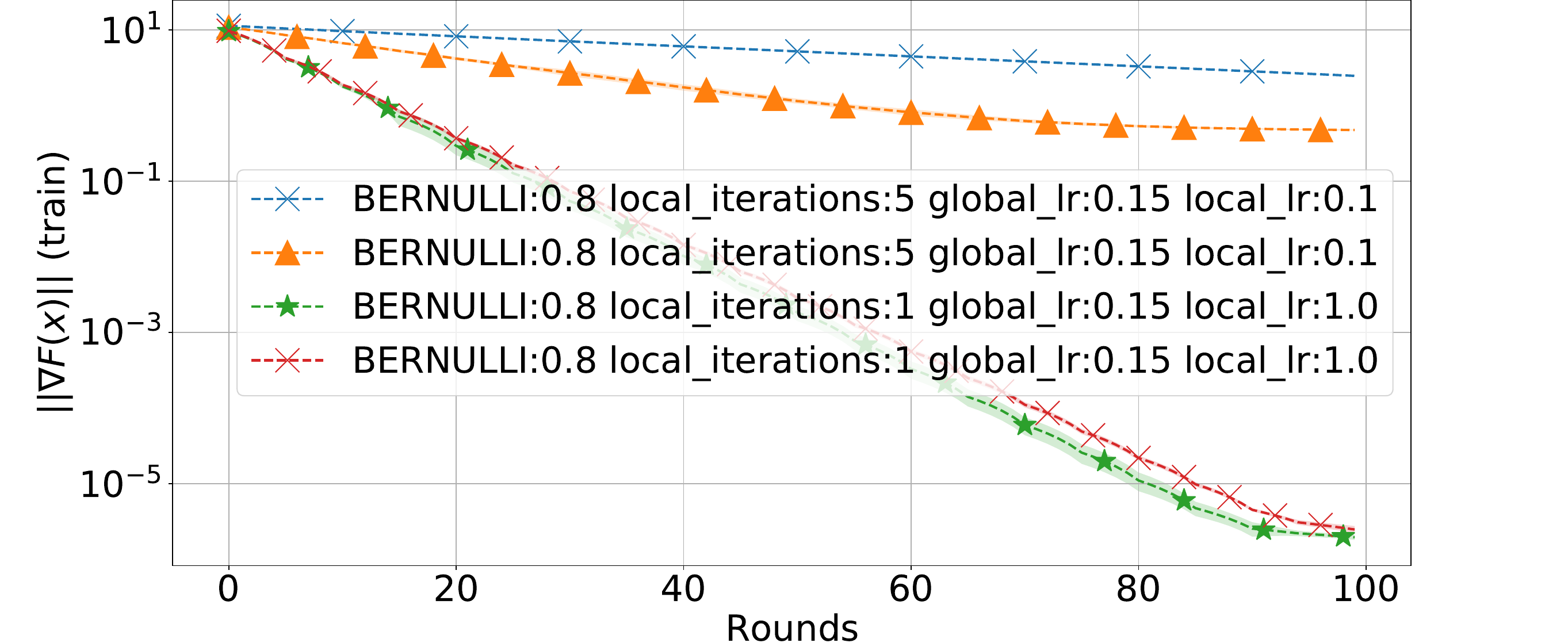} 
		\caption{Convergence}
	\end{subfigure}	
	\begin{subfigure}[ht]{0.49\textwidth}
		\includegraphics[width=\textwidth]{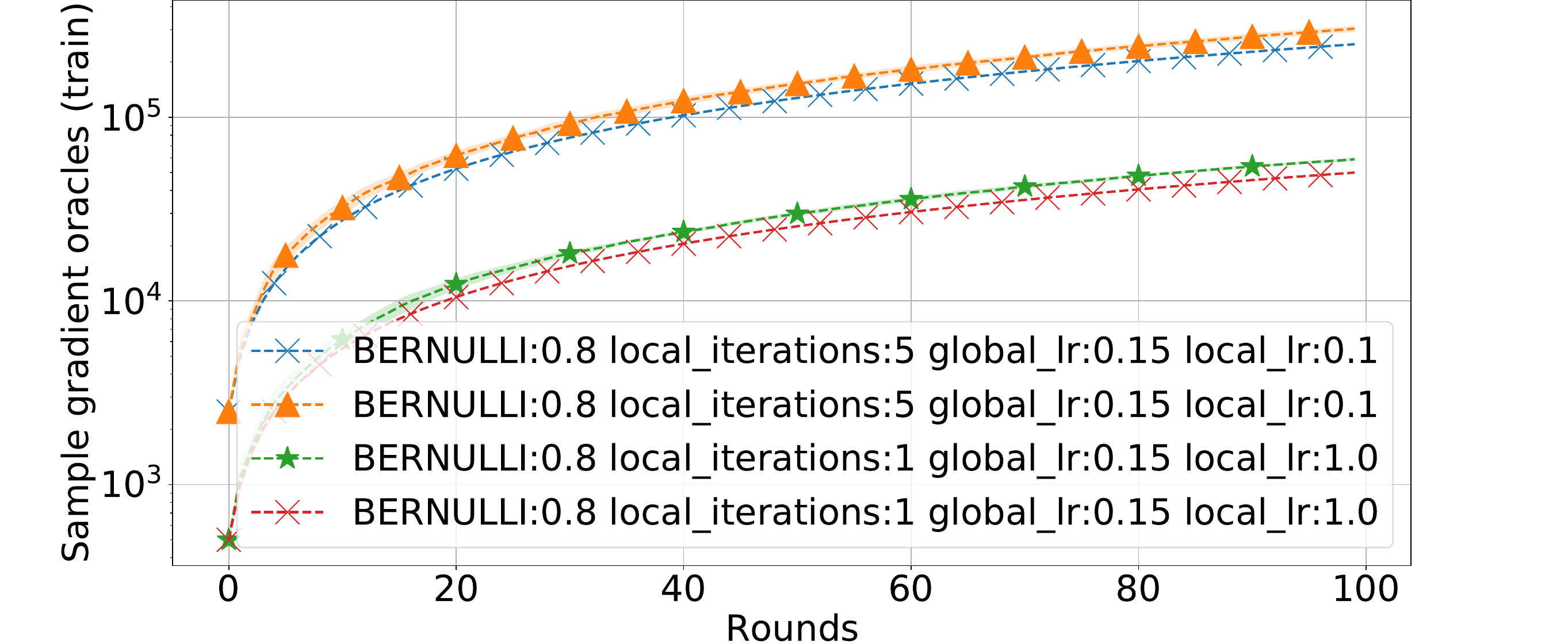}
		\caption{Gradient oracles}
	\end{subfigure}
	
	\caption{Experiments with \algname{MARINA} and \algname{DIANA} algorithms with making local steps for \modelname{quadratic} minimization. Mean and variance have been estimated across 10 realizations.}
	\label{ch2:fig:marina_diana_2_local_iteration_convergence}	
\end{figure*}


\begin{figure*}[t]
	\centering
	\begin{subfigure}[ht]{0.49\textwidth}
		\includegraphics[width=\textwidth]{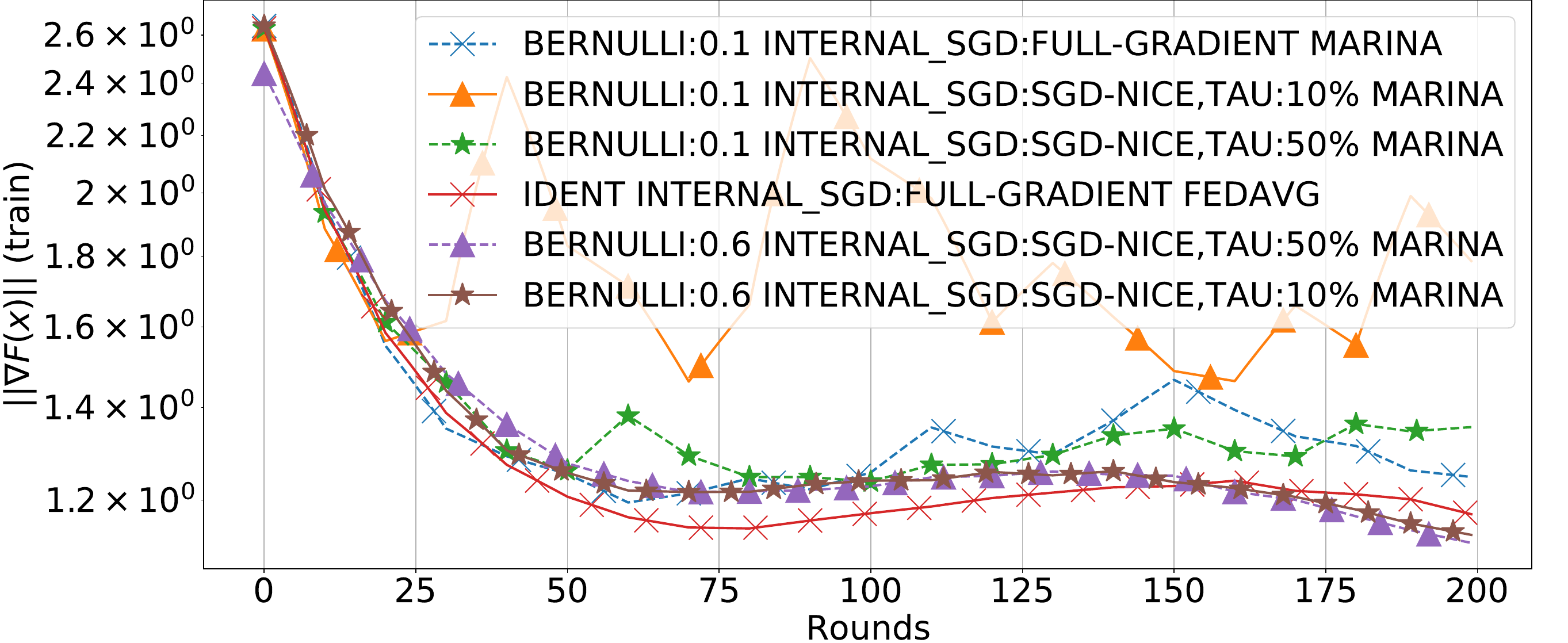} 
		\caption{Convergence}
	\end{subfigure}	
	\begin{subfigure}[ht]{0.49\textwidth}
		\includegraphics[width=\textwidth]{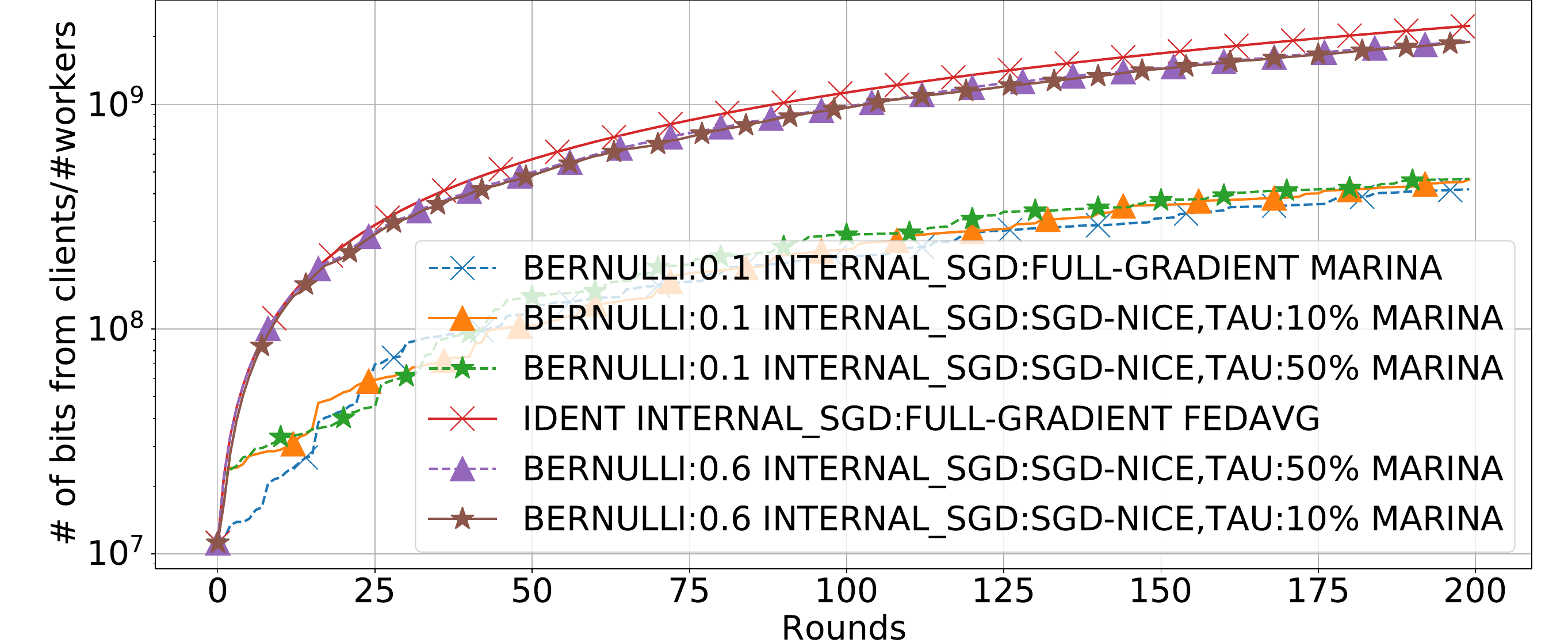}
		\caption{Client-master communication}
	\end{subfigure}
	
	\caption{Experiments with \algname{MARINA} and \algname{DIANA} algorithms for training \modelname{ResNet-18} on \dataname{CIFAR-10} dataset.}
	\label{ch2:fig:neural_nets}	
\end{figure*}

\subsection{{SCAFFOLD} with compression}
In the first example, we used our framework to analyze whether \algname{SCAFFOLD} can work with compression, specifically if both $(\localChange_i^{(t)}, U_i^{(t)})$ can be compressed without a significant decrease in performance 

This experiment was carried out on a synthetically generated \modelname{quadratic} minimization Problem~\eqref{ch2:eq:synthetic_quadratic}. We set the dimensionality of the problem to be $20$. Both features $a_i$'s and responses $b_i$ are generated using a uniform distribution on the interval $[0, 1]$.  After this step, we update the data matrix such that the objective is L-smooth with $L = 2$ and strongly convex with $\mu=1$ using singular value decomposition (SVD). We generate $10$ clients and consider a full participation scenario. The number of communications rounds is chosen to be $100$. We set the global learning rate to $0.5$ and the local learning rate to $1.0$. For the compressor, we choose \compname{RandK}, with 3 values of $K$: $100\%$ (no compression), $40\%$ and $20\%$. We provide the results of this experiment in Figure~\ref{ch2:fig:scaffold_experiment_convex}.  One can observe that dropping $60\%$ coordinates at random has minimal effect on the convergence with respect to iterates.  Dropping $80\%$ brings a visible slow-down in per-iterate convergence, but it is comparable with respect to the number of communicated bits. 

This experiment demonstrates that it may be worthwhile to consider an extension of \algname{SCAFFOLD} by adding compression.

\subsection{Benefits of local updates}

For the second experiment, we adopt a setup similar to that of the previous experiment. We set the smoothness constant $L = 5$ and the strong convexity parameter $\mu = 1$. As compressors, we select either the \compname{Lazy} or \compname{Bernoulli} compressor with $p = 0.8$. This compressor is described in Equation~\eqref{ch2:eq:bernoulli_compressor}.

\begin{equation}
	\label{ch2:eq:bernoulli_compressor}
	C(x)=\begin{cases}
		\dfrac{x}{p}, &\text{with probability } p\\
		0,&\text{with probability} 1 - p.
	\end{cases}
\end{equation}
We run \algname{MARINA} and \algname{DIANA} algorithms with full gradient estimation for $100$ rounds with $10$ clients and full participation. These methods were not analyzed in the setting with several local iterations that motivated us to consider these methods for this experiment.

Looking into Figure~\ref{ch2:fig:marina_diana_2_local_iteration_convergence}, we can observe that both \algname{MARINA} and \algname{DIANA} do not benefit from extra local updates. We also experimented with different problem condition numbers ($\nicefrac{L}{\mu}$), but it seems that local iterations do not speed up the convergence in the strongly convex regime. 

This might suggest that a naive combination of these algorithms with local updates will not lead to theoretical benefits.

\subsection{Stochastic updates and compression}
For the third example, we run \algname{MARINA} and choose the model to be \modelname{ResNet-18}, and for the dataset, federated version of \dataname{CIFAR-10} split uniformly at random (u.a.r.) among $100$ clients. We run $200$ communications rounds and in each round, we randomly select $25$ clients. The local learning rate is $1.0$ and the global learning rate is $0.001$. We perform one local iteration by specifying $\tau_i = 1$ in Algorithm~\ref{ch2:algo:generalized_fedavg}.

We investigate the effect of the combined effect of compression and stochastic gradient estimation since both of these approaches introduce extra variance that might negatively impact the quality of the obtained solution. For gradient estimation, we employ \algname{SGD-NICE} sampling strategy that estimates gradient via selecting uniformly at random a subset of samples of fixed size. $TAU$ in the experiments reflects the relative subset size with respect to the total size of the local dataset. The compressor operator is \compname{Bernoulli} compression.  

Looking into Figure~\ref{ch2:fig:neural_nets}, one can see that having \compname{Lazy} compression with $p=0.1$ combined with \algname{SGD-NICE} with $TAU=10\%$ hurts the convergence. On the other hand,  having a less aggressive compressor or using a bigger sample size for \algname{SGD-NICE} does not deteriorate the performance.

\section{Conclusions}

In this work, we introduce \fl, an efficient FL simulator built on \libname{PyTorch}, designed to empower FL researchers to experiment with optimization algorithms and advance the current state of the art. \fl is a user-friendly tool that supports SOTA FL algorithms and commonly used image classification datasets. Also, \fl leverages multiple levels of parallelism for efficient execution while remaining easy to extend using standard \libname{PyTorch} abstractions. Its basic version requires no programming, with all pre-implemented components accessible via its GUI, enabling users to set up, run, monitor, and evaluate various FL methods. \fl requires a single implementation of an algorithm, with its runtime and computation planning largely decoupled from the algorithm's design.



\clearpage
\appendix

\part*{Appendices to Chapter \ref{chapter2}}
\label{ch2:app:toc_1}
\newpage

\phantomsection
\addcontentsline{toc}{chapter}{Appendices to Chapter 2}


\addtocounter{adjsection}{1}
\section{FL\_PyTorch: Template Methods of Algorithm \ref{ch2:algo:generalized_fedavg}}
\label{ch2:app:fl_skeleton}

Here we present an overview of templated methods under researcher responsibilities for Algorithm \ref{ch2:algo:generalized_fedavg}. For subtle technical details please familiarize yourself with the provided readme, tutorial, automatically generated code documentation in the code repository, and well-documented code.

\begin{table}[h!]
	\caption{Brief description of template methods of Algorithm \ref{ch2:algo:generalized_fedavg}.}
	\label{ch2:tbl:generalized_fedavg_steps}
	\centering
	\bgroup
	\definecolor{headcolor}{RGB}{47,79,79}
	\def\arraystretch{1.5}%
	\begin{tabular}{|p{0.33\textwidth}|p{0.61\textwidth}|}
		\hline
		{\textcolor{black}{\textbf{Template Method}}} &
		{\textcolor{black}{\textbf{Description}}} \\
		\hline
		\hline
		\InitializeServerState &  This method should return a dictionary that initializes the server state. The method obtains a dimension of the problem, and the constructed and initialized model. \\
		\hline
		\ClientState &  
		By our design client state is stateless. The client state is instantiated at the beginning of each round for each of the selected clients. User should reconstruct the client state based on the initialized or updated server state.
		\\
		\hline
		\LocalGradient &  
		This method should evaluate the local gradient that is optimization algorithm-specific.
		\\
		\hline
		\textsc{ClientOpt} & 
		Local classical optimizer provided out of the box by \libname{PyTorch} used in client for local steps.\\
		\hline
		\textsc{LocalState} & The default implementation of this step is presented in some sense in \textit{local\_training} method. This step happens automatically, and in rare cases, there is a need for customization of this step.\\
		\hline
		\textsc{ServerOpt} & Classical optimizer provided out of the box by \libname{PyTorch} used in sever for global steps.\\
		\hline
		\ServerGradient &  
		Server gradient is the method that estimates the direction of the global server model update, which should return a flat vector with $d$ elements.
		\\
		\hline
		\ServerGlobalState &  
		This logic is dedicated to the global server state update. This method as input obtains collected and ready-to-use information about clients' responses and clients in that communication round, the previous global model, and a new model with updated parameters model.
		\\
		\hline
	\end{tabular}
	\egroup
\end{table}

\newpage
\addtocounter{adjsection}{1}
\section{FL\_PyTorch: Internal Mechanisms}
\label{ch2:app:fl_vis}
In this section, we would like to depict \fl main components schematically.

\begin{figure}[ht]
	\centering
	\includegraphics[width=0.70\textwidth, keepaspectratio=true]
	{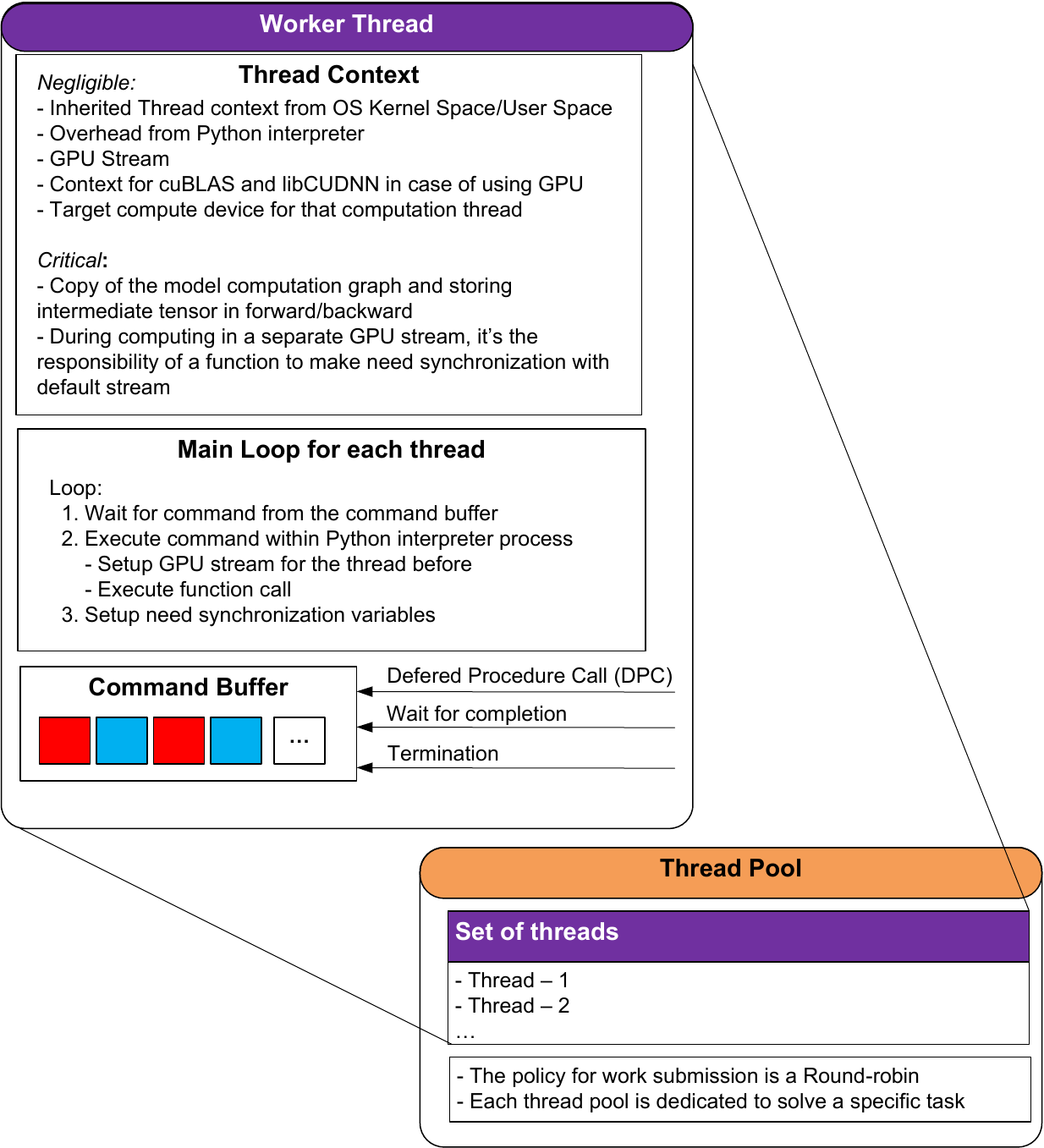}
	\caption{A single worker thread structure and its role in a thread pool.}
	\label{ch2:fig:fl_worker_th}
\end{figure}

\begin{figure}[ht]
	\centering
	\includegraphics[width=0.45\textwidth, keepaspectratio=true]
	{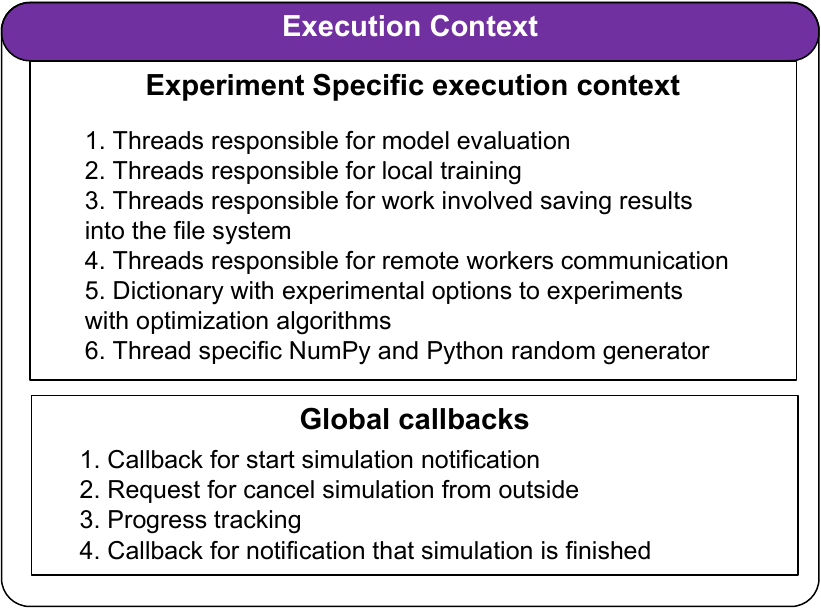}
	\caption{\fl execution context for a single experiment. The GUI can handle several experiments at the same time.}
	\label{ch2:fig:fl_exec_ctx}
\end{figure}

\begin{figure}[t]
	\centering
	\includegraphics[width=1.0\textwidth, keepaspectratio=true]
	{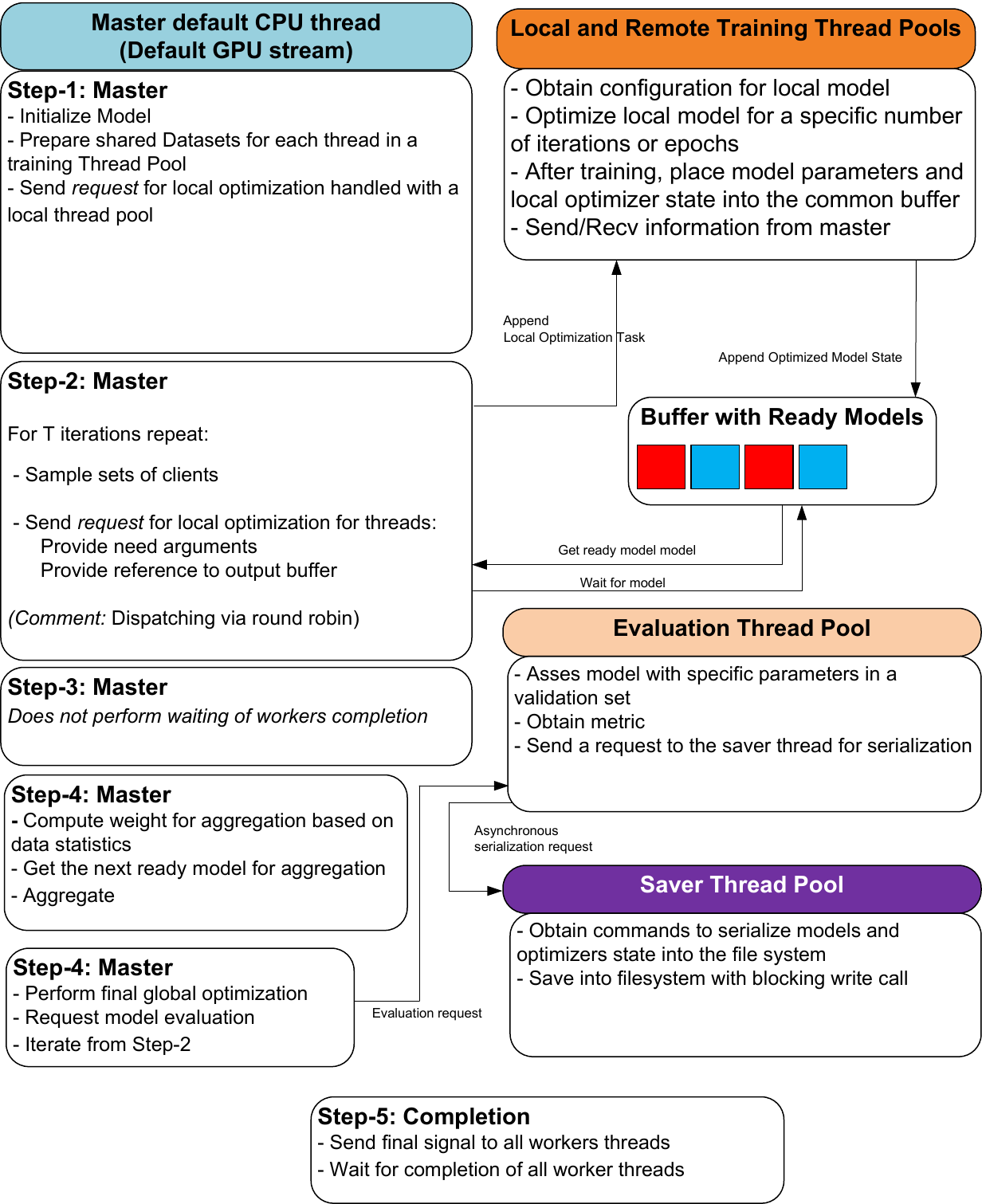}
	\caption{Communication between different threads during Algorithm \ref{ch2:algo:generalized_fedavg} execution.}
	\label{ch2:fig:fl_components}
\end{figure}

\clearpage
\addtocounter{adjsection}{1}
\section{Functional Comparison to Related Frameworks}
\label{ch2:app:fl_vs_others}

Next, we compare \fl with \libname{Flower} and \libname{FedML.ai}. It's important to note that while these well-recognized frameworks are designed for FL experiments and offer a variety of capabilities, to the best of our knowledge, they were not specifically intended as extensive research simulators (as of 2021)\footnote{We have connected with the authors of \libname{Flower} to suggest integrating this work into \libname{Flower}, but it was not straightforward due to their different focus.} The results of the comparison, including built-in support, are presented in Table~\ref{ch2:tbl:cmp}.

\begin{table}[h!]
	\footnotesize
	\caption{Comparison of \fl, \libname{FedML}, and \libname{Flower} in 2021.}
	\label{ch2:tbl:cmp}
	\centering
	\bgroup
	\definecolor{headcolor}{RGB}{47,79,79}
	\def\arraystretch{1.5}%
	\begin{tabular}{|p{0.02\textwidth}|p{0.35\textwidth}|p{0.15\textwidth}|p{0.15\textwidth}|p{0.15\textwidth}|}
		\hline
		\textcolor{black}{\textbf{\#}} & 
		\textcolor{black}{\textbf{Built-in Comparison Criteria}} & \textcolor{black}{\fl} & \textcolor{black}{\textbf{\libname{FedML.ai}}} & \textcolor{black}{\textbf{\libname{Flower.dev}}} \\
		\hline
		\hline
		1 & Home Page & \href{https://github.com/burlachenkok/flpytorch}{flpytorch} & \href{https://fedml.ai}{fedml.ai} & \href{https://flower.dev}{flower.dev} \\
		\hline
		2 & Stochastic Compressors & Yes & No & No \\
		\hline
		3 & Support for Local Steps & Yes & Yes & Yes \\
		\hline
		4 & Plot Creation (with Internet Connection) & Yes & Yes & Yes \\
		\hline
		5 & Plot Creation (without Internet Connection) & Yes & No & No \\
		\hline
		6 & Highly Customizable Plots & Easy & Harder & Harder \\
		\hline
		7 & Serialization of Results from Numerical Experiments & Yes & No & No \\
		\hline
		8 & Support for Standalone and Distributed Mode & Yes & Yes & Yes \\
		\hline
		9 & Communication Protocol in Multi-node Setup & {TCP/IP} & {MPI}, {gRPC} & {gRPC} \\
		\hline
		10 & Synthetically Controlled Optimization Problems & Yes & No & No \\
		\hline
		11 & Number of Supported Models and Datasets & Modest & High & High \\
		\hline
		12 & Client Parallelization on a Single GPU & Yes & No & No \\
		\hline
		13 & Distributed Debugging & Easy\tablefootnote{Computations can be reduced to a single-thread system setup} & Harder & Harder \\
		\hline
		14 & Parallelization Across Several GPUs (Standalone) & Yes & No & No \\
		\hline
		15 & GUI Interface & Yes & No & No \\
		\hline
		16 & Console Interface for Launching Experiments & Yes & Yes & Yes \\
		\hline
		17 & Support for Theoretical Step Sizes During Training & Yes & No & No \\
		\hline
		18 & Built-in Mechanism to Load/Save Experiments & Yes & No & No \\
		\hline
	\end{tabular}
	\egroup
	\\
\end{table}

\clearpage
\addtocounter{adjsection}{1}
\section{FL\_PyTorch as a GPU-Accelerated Application}
\label{ch3:app:nv_compute_ecosystem}

Machine learning applications today predominantly rely on GPUs or CPUs. Recently, there has been growing interest in using FPGA devices. Unlike CPUs and GPUs with fixed instruction sets, FPGAs consist of programmable logic blocks that can be dynamically interconnected to create custom hardware circuits for specialized tasks. FPGAs offer an interesting option and unique capabilities for tighter algorithms and computational hardware integration.

Historically, between 2003 and 2010, in the absence of general-purpose GPU computing support, researchers relied on graphics APIs to encode computations within the framework of rasterized image processing. Since 2010, direct programming interfaces for GPU computing have become widely available,\footnote{Some GPU Compute API: AMD ROCm, CUDA, OpenCL, Apple Metal, Vulkan Compute.} enabling explicit access to GPU acceleration. For Deep Learning training, NVIDIA GPUs remain the dominant choice, though other vendors are beginning to catch up\footnote{Example is the \href{https://rocm.docs.amd.com/en/latest}{AMD ROCm} platform which is integrated into TensorFlow, PyTorch, JAX.}. Low-level programming of NVIDIA GPUs is done through CUDA\footnote{Compute Unified Device Architecture (CUDA): \href{https://developer.nvidia.com/cuda-toolkit}{https://developer.nvidia.com/cuda-toolkit}}. CUDA is not just a programming language but a comprehensive platform and ecosystem that extends several programming languages, both syntactically and semantically, to facilitate efficient GPU programming.

\begin{figure}[t]
	\centering
	\includegraphics[width=1.0\textwidth, keepaspectratio=true]
	{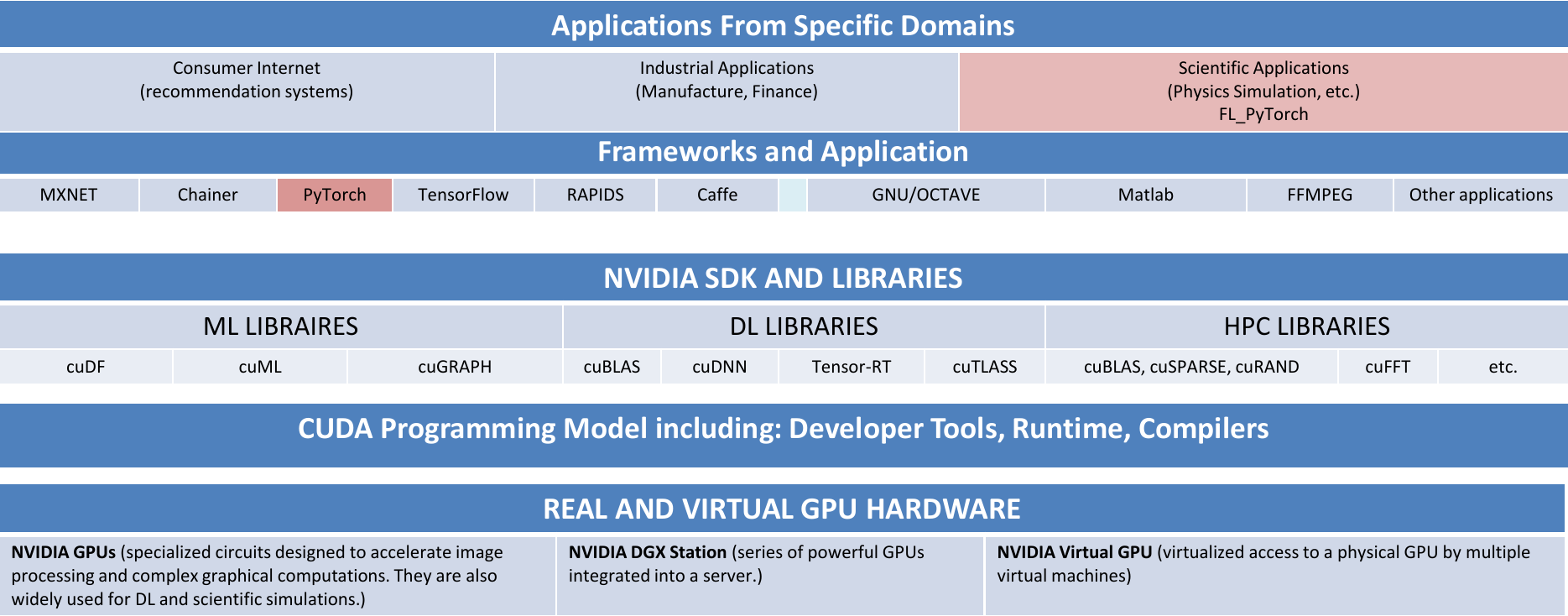}
	\caption{NVIDIA compute ecosystem and the place of \fl in it if using GPU compute.}
	\label{ch2:fig:flpytorch-in-comp-gpu-stack}
\end{figure}

Because \fl uses \texttt{GPU}s indirectly through \libname{PyTorch}, it can be worthwhile to observe the larger picture of how \fl fits into the overall computational stack once configured to use an {NVIDIA GPU}. We illustrate this in Figure~\ref{ch2:fig:flpytorch-in-comp-gpu-stack}.

\addtocounter{adjsection}{1}
\section{Reproducibility}

The \fl is distributed under the Apache License, Version 2.0 (January 2004) and is publicly available at the following link:
\begin{center}
	\href{https://github.com/burlachenkok/flpytorch}{https://github.com/burlachenkok/flpytorch}
\end{center}

\unappendix

\chapter{EF21-W: Error Feedback Reloaded}
\label{chapter3}

The goals and summaries of this chapter are outlined in Table \ref{ch1:tbl:algorithms} and Section~\ref{ch1:sec:overview-3}.

\section{Introduction}

Due to their ability to harness the computational capabilities of modern devices and their capacity to extract value from the enormous data generated by organizations, individuals, and various digital devices and sensors, Machine Learning (ML) methods~\citep{bishop2016pattern, shalev2014understanding} have become indispensable in numerous practical applications~\citep{krizhevsky2012, lin-etal-2022-truthfulqa, transformer, onay2018review, cardio_vas_DL, gavriluct2009malware, sun2017deep}. 

The necessity to handle large datasets has driven application entities to store and process their data in powerful computing centers~\citep{YANG2019278, dean2012large, verbraeken2020survey} via {\em distributed} training algorithms. Besides this industry-standard centralized approach, decentralized forms of distributed learning are becoming increasingly popular. For example, Federated Learning (FL) facilitates a collaborative learning process in which various clients, such as hospitals or owners of edge devices, collectively train a model on their devices while retaining their data locally, without uploading it to a centralized location \citep{FEDLEARN, konevcny2016afederated, mcmahan17fedavg, FL_overview, kairouz2019advances, FieldGuide2021}. 

Distributed training problems are typically formulated as optimization problems of the form
\begin{equation}\label{ch3:eq:main_problem}
	\min\limits_{x \in \RR^d} \left\{ f(x) \eqdef \dfrac{1}{n}\sum \limits_{i=1}^n f_i(x) \right\},
\end{equation}
where $n$ is the number of clients/workers/nodes, vector $x\in \RR^d$ represents the $d$ trainable parameters, and  $f_i(x)$ is the loss of the model parameterized by $x$ on the training data stored on client $i \in [n]\eqdef \{1,\dots,n\}$. One of the key issues in distributed training in general, and FL in particular, is the \textit{communication bottleneck}~\citep{FEDLEARN,kairouz2019advances}. The overall efficiency of a distributed algorithm for solving \eqref{ch3:eq:main_problem} can be characterized by multiplying the number of communication rounds needed to find a solution of acceptable accuracy by the cost of each communication round:
\begin{eqnarray}
	\label{ch3:eq:comm_burden_formula}
	\mathsf{Communication\ Complexity} & =  \mathsf{\#\ Communication\ Rounds} \notag \\
	&\times \mathsf{Cost\ per\ Round}.
\end{eqnarray}
This simple formula clarifies the rationale behind two orthogonal approaches to alleviating the communication bottleneck. i) The first approach aims to minimize the first factor in~\eqref{ch3:eq:comm_burden_formula}. This is done by carefully deciding on {\em what work should be done} on the clients in each communication round in order for it  to reduce the total number of communication rounds needed and includes methods based on local training~\citep{stich2018local,lin2018don, mishchenko2022proxskip, condat2023tamuna, li2019convergence} and  momentum~\citep{nesterov_accelerated,NesterovBook,AccMethodsBook}. Methods in this class  communicate dense $d$-dimensional vectors. ii) The second approach aims to minimize the second factor in~\eqref{ch3:eq:comm_burden_formula}. Methods in this category {\em compress the information} (typically $d$-dimensional vectors) transmitted between the clients and the server~\citep{alistarh2017qsgd, DCGD, signsgd, safaryan2021}.

\subsection{Communication compression}
Vector compression can be achieved through the application of a compression operator. Below, we outline two primary classes of these operators: unbiased (with conically bounded  variance) and contractive.

\begin{definition}[Compressors]
	A randomized mapping $\cC: \RR^d \to \RR^d$ is called i) an {\em unbiased compressor} if for some $\omega > 0$ it satisfies 
	\begin{equation}\label{ch3:eq:unbiased}
		\ExpBr{\cC(x)} = x, \quad \ExpBr{\|\cC(x) - x \|^2} \leq \omega \|x\|^2, \quad 
		\forall x\in\RR^d,\end{equation}
	and ii) a {\em contractive compressor} if for some $\alpha \in (0, 1]$ it satisfies 
	\begin{equation}\label{ch3:eq:compressor_contraction}		
		\ExpBr{\|\cC(x) - x \|^2} \leq (1 - \alpha) \|x\|^2, \quad 
		\forall x\in\RR^d.
	\end{equation}
\end{definition}
It is well known that whenever a compressor $\cC(x)$ satisfies \eqref{ch3:eq:unbiased}, then the scaled compressor $\cC(x)/(\omega+1)$ satisfies \eqref{ch3:eq:compressor_contraction} with $\alpha=(\omega+1)^{-1}$. In this sense, the class of contractive compressors includes all unbiased compressors as well. However, it is also strictly larger. For example, the \compname{TopK} compressor, which retains the $K$ largest elements in the absolute value of the vector it is applied to and replaces the rest by zeros, and happens to be very powerful in practice~\citep{Alistarh-EF-NIPS2018}, satisfies \eqref{ch3:eq:compressor_contraction} with $\alpha=
\nicefrac{K}{d}$, but does not satisfy \eqref{ch3:eq:unbiased}. From now on, we write $\mathbb{C}(\alpha)$ to denote the class of compressors satisfying \eqref{ch3:eq:compressor_contraction}. 

It will be convenient to define the following functions of the contraction parameter 
\begin{equation}
	\label{ch3:eq:xi}
	\begin{split}
		\theta &= \theta(\alpha) \eqdef 1 - \sqrt{1 - \alpha}, \\
		\beta &= \beta(\alpha) \eqdef \dfrac{1 - \alpha}{1 - \sqrt{1 - \alpha}}, \\
		\xi &= \xi(\alpha) \eqdef \sqrt{\dfrac{\beta(\alpha)}{\theta(\alpha)}} = \dfrac{1 + \sqrt{1 - \alpha}}{\alpha} - 1.
	\end{split}
\end{equation}

Note that $$0\leq \xi(\alpha) <\dfrac{2}{\alpha}-1.$$
The behavior of distributed algorithms  utilizing unbiased compressors for solving \eqref{ch3:eq:main_problem} is relatively well-understood from a theoretical standpoint~\citep{DCGD, mishchenko2024distributed, ADIANA, gorbunov2021marina, DASHA}. By now, the community possesses a robust theoretical understanding of the advantages such methods can offer and the mechanisms behind their efficacy~\citep{sigma_k,khaled2023,DASHA}.  However, it is well known that the class of contractive compressors includes some practically very powerful operators, such as the greedy sparsifier \compname{TopK}~\citep{Stich-EF-NIPS2018,Alistarh-EF-NIPS2018} and the low-rank approximator \compname{RankK}~\citep{PowerSGD, safaryan2021fednl}, which are biased, and hence their behavior is not explainable by the above developments. These compressors have demonstrated surprisingly effective performance in practice~\citep{Seide2014,Alistarh-EF-NIPS2018}, even when compared to the best results we can get with unbiased compressors~\citep{szlendak2021permutation}, and are indispensable on difficult tasks such as the fine-tuning of foundation models in a geographically distributed manner over slow networks~\citep{cocktailsgd}. 

However, our theoretical understanding of algorithms based on contractive compressors in general, and these powerful biased compressors in particular, is very weak. Indeed, while the SOTA theory involving unbiased compressors offers significant and often several-degrees-of-magnitude improvements  over the baseline methods that do not use compression~\citep{mishchenko2024distributed, DIANA2, ADIANA,sigma_k,gorbunov2021marina, DASHA}, the best theory we currently have for methods that can provably work with contractive compressors, i.e., the theory behind the error feedback method called \algname{EF21}  developed by \citet{EF21} (see Algorithm~\ref{ch3:alg:EF21})  and its variants~\citep{fatkhullin2021ef21, EF-BV,EF21M}, merely matches the communication complexity of the underlying methods that do not use any compression~\citep{szlendak2021permutation}. 

To the best of our knowledge, the only exception is the recent work of \citet{richtarik2023error}, which demonstrates that in the {\em rare features} regime, the \algname{EF21} method \citep{EF21} can be substantially improved in theory for this restricted class of optimization problems. However, \citet{richtarik2023error} do not establish any improvements over the current best theoretical results for vanilla \algname{EF21} \citep{EF21} in the general smooth non-convex regime outlined in Section~\ref{ch3:sec:assumptions}, which is the focus of our investigation.

\begin{algorithm}[t]
	\begin{algorithmic}[1]
		\STATE {\bfseries Input:} initial model $x^0 \in \RR^d$; initial gradient estimates $g_1^0, g_2^0, \dots,g_n^0 \in \R^d$ stored at the server and the clients; step size $\gamma>0$; number of iterations $T > 0$
		\STATE {\bfseries Initialize:} $g^0 = \avein g_i^0 $ on the server
		\FOR{$t = 0, 1, 2, \dots, T - 1 $}
		\STATE Server computes $x^{t+1} = x^t - \gamma g^t$ and  broadcasts  $x^{t+1}$ to all $n$ clients
		\FOR{$i = 1, \dots, n$ {\bf on the clients in parallel}} 
		\STATE Compute $u_i^t=\cC_i^t (\nabla f_i(x^{t+1}) - g_i^t)$ and update $g_i^{t+1} = g_i^t +u_i^t$ \label{ch3:alg_line:g_update_step}
		\STATE Send the compressed message $u_i^{t}$ to the server
		\ENDFOR 
		\STATE Server updates $g_i^{t+1} = g_i^t +u_i^t$ for all $i\in [n]$, and computes $g^{t+1} = \avein g_i^{t+1}$
		\ENDFOR
		\STATE {\bfseries Output:} Point $\hat{x}^T$ chosen from the set $\{x^0, \dots, x^{T-1}\}$ uniformly at random
	\end{algorithmic}
	\caption{\algname{EF21}: Error Feedback 2021 from \citet{EF21} \textbf{[Existing]}.}	
	\label{ch3:alg:EF21}
\end{algorithm}

\subsection{Assumptions}
\label{ch3:sec:assumptions}

We adopt the same very weak assumptions as those used by \citet{EF21} in their analysis of \algname{EF21}.

\begin{assumption}\label{ch3:as:smooth}
	The function $f$ is $L$-smooth, i.e., there exists $L>0$ such that
	\begin{equation}\label{ch3:eq:smoothness_def}
		\norm{\nabla f(x) - \nabla f(y)} \leq L \norm{x - y}, \quad \forall x, y \in \R^d.
	\end{equation}
\end{assumption}

\begin{assumption}\label{ch3:as:L_i} 
	The functions $f_i$ are $L_i$-smooth,  i.e., for all $i\in [n]$ there exists $L_i>0$ such that
	\begin{equation} \label{ch3:eq:L_i}
		\norm{\nabla f_i(x) - \nabla f_i(y)} \leq L_i \norm{x - y}, \quad \forall x, y \in \R^d.
	\end{equation}
\end{assumption}

Note  that if \eqref{ch3:eq:L_i} holds,  then \eqref{ch3:eq:smoothness_def} holds, and $$L\leq \LAM \eqdef {\color{blue}\dfrac{1}{n}\sum_{i=1}^n L_i}.$$
So, Assumption~\ref{ch3:as:smooth} does {\em not} further limit the class of functions already covered by Assumption~\ref{ch3:as:L_i}. Indeed, it merely provides a new parameter $L$ better characterizing the smoothness of $f$ than the estimate $\LAM$ obtainable from Assumption~\ref{ch3:as:L_i} could. 

Since our goal in \eqref{ch3:eq:main_problem} is to minimize $f$, the below assumption is necessary for the problem to be meaningful.

\begin{assumption}\label{ch3:as:lower_bound}
	There exists $f^\ast \in \RR$ such that $\inf f \geq f^\ast$.
\end{assumption}

\subsection{Contributions}

\begin{enumerate}
\item  In our work, we  improve the current SOTA theoretical communication complexity guarantees for distributed algorithms that  work with contractive compressors in general, and  empirically powerful biased compressors such as \compname{TopK} and \compname{RankK} in particular~\citep{EF21,fatkhullin2021ef21}. In particular, under Assumptions~\ref{ch3:as:smooth}, \ref{ch3:as:L_i}, \ref{ch3:as:lower_bound}, the best known guarantees were obtained by \citet{EF21} for the \algname{EF21} method: to find a (random) vector $\hat{x}^T$ satisfying $$\ExpBr{\norm{\nabla f(\hat{x}^T)}^2} \leq \varepsilon,$$
Algorithm~\ref{ch3:alg:EF21} requires 
$$
T=\cO \left({\left(L + \LQM \xi (\alpha)\right)}{\varepsilon}^{-1}\right)
$$ 
iterations, where
$\color{red} \LQM \eqdef \sqrt{\dfrac{1}{n} \sum_{i=1}^n L_i^2}$ is the {\color{red}Quadratic Mean} of the smoothness constants $L_1,\dots,L_n$. Our main finding is an improvement of this result to \begin{equation}\label{ch3:eq:98y98fd}T=\cO\left({\left( L + \LAM \xi (\alpha)\right)}{\varepsilon}^{-1} \right),\end{equation}
where $\color{blue} \LAM \eqdef \dfrac{1}{n} \sum_{i=1}^n L_i$ is the {\color{blue}Arithmetic Mean} of the smoothness constants $L_1,\dots,L_n$. 

\item We obtain this improvement in {\em three different ways:} 
\begin{itemize}
	\item [i)] by {\em client cloning} (see Sections~\ref{ch3:sec:clone1} and \ref{ch3:sec:clone2} and Theorem~\ref{ch3:thm:clone}), 
	\item [ii)] by proposing a new {\em smootness-weighted} variant of \algname{EF21} which we call \algname{EF21-W} (see Section~\ref{ch3:sec:weights} and Theorem~\ref{ch3:thm:EF21-W}), 
	\item [iii)] and by a new {\em smoothness-weighted} analysis of classical \algname{EF21} (see Section~\ref{ch3:sec:weighted-analysis} and Theorem~\ref{ch3:thm:ef21_new_result}).
\end{itemize}

\item We obtain refined linear convergence results in cases under the Polyak-Łojasiewicz condition. 

\item Further, our analysis technique extends to many variants of \algname{EF21}, including \algname{EF21-SGD} which uses stochastic gradients instead of gradients (Section~\ref{ch3:sec:EF21-W-SGD}), and \algname{EF21-PP} which enables partial participation of clients (Section~\ref{ch3:sec:EF21-W-PP}).
\item Our analysis also improves upon the results of \citet{richtarik2023error} who study \algname{EF21} in the {\em rare features} regime (Section~\ref{ch3:sec:RF}). 

\item Finally, we validate our theory with suitable computational experiments (Sections~\ref{ch3:sec:experiments-main}, \ref{ch3:app:exp-additional-details-main-part} and \ref{ch3:app:exp-additional-experiments}).

\end{enumerate}

\section{EF21 Reloaded: Our Discovery Story} \label{ch3:sec:reloaded}

We now take the reader along on a ride of our discovery process.

\subsection{Step 1: Cloning the client with the worse smoothness constant}
\label{ch3:sec:clone1}

The starting point of our journey is  a simple observation described in the following example.
\begin{example} Let $n=4$ and $$f(x) = \dfrac14 (f_1(x) + f_2(x) + f_3(x) + f_4(x)).$$ Assume the smoothness constants $L_1, L_2, L_3$ of $f_1, f_2, f_3$ are equal to 1, and $L_4$ is equal to $100$. In this case, \algname{EF21} needs to run for
	$$
	T_1 \eqdef \cO\left({\left(L + \LQM \xi(\alpha)\right)}{\varepsilon}^{-1}\right) = \cO\left({\left(L +  \sqrt{2501.5}\xi(\alpha)\right)}{\varepsilon}^{-1}\right)
	$$ 
	iterations. Now, envision the existence of an additional machine capable of downloading the data from the fourth ``problematic'' machine. By rescaling local loss functions, we maintain the overall loss function as: 
	\begin{eqnarray*}
		f(x) &=& \dfrac14 (f_1(x) + f_2(x) + f_3(x) + f_4(x)) \\
			 &=&  \dfrac15 \left( \dfrac54 f_1(x) + \dfrac54 f_2(x) + \dfrac54 f_3(x) + \dfrac58 f_4(x) +\dfrac58 f_4(x) \right) \eqdef \tilde{f}(x).
	\end{eqnarray*}
	
	Rescaling of the functions modifies the smoothness constants to $\hat{L}_i = \dfrac54 L_i$ for $i = 1, 2, 3$, and $\hat{L}_i = \dfrac58 L_4$ for $i=4, 5$. \algname{EF21}, launched on this setting of five nodes, requires 
	$$
	T_2 \eqdef \cO\left({\left(L + {\color{red}\tilde{L}_{\rm QM}} \xi(\alpha)\right)}{\varepsilon}^{-1} \right) \approx \cO\left({\left(L + \sqrt{1564} \xi (\alpha)\right)}{\varepsilon}^{-1}\right)
	$$
	iterations, where ${\color{red}\tilde{L}_{\rm QM}}$ is the quadratic mean of the new smoothness constants $\hat{L}_1,\dots,\hat{L}_5$.
\end{example}

This simple observation highlights that the addition of just one more client significantly enhances the convergence rate. Indeed, \algname{EF21} requires approximately $\nicefrac{\xi(\alpha)}{\varepsilon} \left(\sqrt{2501.5} - \sqrt{1564}\right) \approx 10\,\nicefrac{\xi(\alpha)}{\varepsilon}$ fewer iterations. We will generalize this client cloning idea in the next section.

\subsection{Step 2: Generalizing the cloning idea} 
\label{ch3:sec:clone2}

We will now take the above motivating example further, allowing each client $i$ to be cloned arbitrarily many ($N_i$) times. Let us see where this gets us. For each $i\in [n]$, let $N_i$ denote  a positive integer. We define $N\eqdef \sum_{i=1}^n N_i$ (the total number of clients after cloning), and observe that $f$ can be equivalently written as
\begin{eqnarray}
	\label{ch3:eq:cloning}
	f(x)  &\overset{\eqref{ch3:eq:main_problem}}{=}& \dfrac{1}{n}\sum \limits_{i=1}^n f_i(x) \notag \\
	&=&  \dfrac{1}{n}\sum \limits_{i=1}^n  \sum\limits_{j=1}^{N_i} \dfrac{1}{N_i} f_i(x) \notag \\
	&=& \dfrac{1}{N}\sum \limits_{i=1}^n  \sum\limits_{j=1}^{N_i}\dfrac{N}{n N_i} f_i(x) = \dfrac{1}{N}\sum \limits_{i=1}^n  \sum\limits_{j=1}^{N_i} f_{ij}(x),
\end{eqnarray}
where 
$ f_{ij}(x) \eqdef \dfrac{N}{n N_i} f_i(x)$ for all $ i\in [n]$ and $ j\in [N_i].$
Notice that we scaled the functions as before and that $f_{ij}$ is $L_{ij}$-smooth, where $L_{ij} \eqdef \dfrac{N}{n N_i} L_i$. 

{\bf Analysis of the convergence rate.} The performance of \algname{EF21}, when applied to the Problem~\eqref{ch3:eq:cloning} involving $N$ clients, depends on the quadratic mean of the new smoothness constants:
\begin{eqnarray} 
	\label{ch3:eq:M-solve}
	M(N_1,\dots,N_n) &\eqdef& \sqrt{\dfrac{1}{N} \sum\limits_{i=1}^n \sum\limits_{j=1}^{N_i} L_{ij}^2} \notag = \sqrt{\dfrac{1}{N} \sum\limits_{i=1}^n {N_i} \cdot \left(\dfrac{N}{n N_i} L_i\right)^2} \notag \\
	&=& \sqrt{\sum\limits_{i=1}^n \dfrac{N}{n^2 N_i} L_i^2} = \dfrac{1}{n}\sqrt{\sum\limits_{i=1}^n \dfrac{L_i^2}{N_i/N} }.
\end{eqnarray}
Note that if $N_i=1$ for all $i\in [n]$, then $M(1,\dots,1)=\LQM$.

{\bf Optimal choice of cloning frequencies.} Our goal is to find integer values $N_1\in \mathbb{N},\dots,N_n \in \mathbb{N} $ minimizing the function $M(N_1,\dots,M_n)$ defined in \eqref{ch3:eq:M-solve}. While we do not have a closed-form formula for the global minimizer, we can explicitly find a  solution that is at most $\sqrt{2}$ times worse than the optimal one in terms of the objective value. In particular, if we let $N^\star_i =  \left \lceil  \nicefrac{L_i}{\LAM} \right \rceil$ for all $i\in [n]$, then 
\[\LAM \leq  \min\limits_{N_1 \in \mathbb{N},\dots, N_n\in \mathbb{N}} M(N_1,\dots,N_n) \leq   M(N^\star_1,\dots,N^\star_n) \leq \sqrt{2} \LAM,\]
and moreover, $n \leq N^\star \eqdef \sum_i N^\star_i \leq 2n$. That is, we need at most double the number of clients in our client cloning construction.  See Lemma~\ref{ch3:lem:sandwitch} in the Appendix for details. 

By applying \algname{EF21} theory from \citep{EF21} to Problem~\eqref{ch3:eq:cloning} involving $N^\star$ clients, we obtain the advertised improvement from $\LQM$ to $\LAM$.

\begin{theorem}[\textbf{Convergence of \algname{EF21} applied to Problem~\eqref{ch3:eq:cloning} with $N^\star$ machines}] \label{ch3:thm:clone}  Consider Algorithm~\ref{ch3:alg:EF21} (\algname{EF21}) applied to the ``cloning reformulation'' \eqref{ch3:eq:cloning} of the distributed optimization Problem~\eqref{ch3:eq:main_problem}, where $N^\star_i = \left \lceil \nicefrac{L_i}{\LAM} \right \rceil$ for all $i \in [n]$.  Let Assumptions~\ref{ch3:as:smooth}, \ref{ch3:as:L_i}, \ref{ch3:as:lower_bound} hold, assume that
	$\cC_{ij}^t \in \mathbb{C}(\alpha)$ for all $i\in [n]$, $j\in [N_i]$ and $t\geq 0$,  set
	\begin{equation*}
		G^t \eqdef \dfrac{1}{N}\sum \limits_{i=1}^N \sum\limits_{j=1}^{N_i} \norm{g_{ij}^t - \nabla f_{ij}(x^t)}^2,
	\end{equation*}
	and let the step size satisfy
	$$
	0 < \gamma \leq \dfrac{1}{L + \sqrt{2}\LAM \xi (\alpha)}.
	$$
	If for $T \geq 1$ we define $\hat{x}^T$ as an element of the set $\{x^0, x^1, \dots, x^{T-1}\}$ chosen uniformly at random, then
	\begin{equation*}
		\ExpBr{\norm{\nabla f(\hat{x}^T)}^2} \leq \dfrac{2 (f(x^0) - f^\ast) }{\gamma T} + \dfrac{G^0}{\theta(\alpha) T}.
	\end{equation*}
\end{theorem}

When we choose the largest allowed step size and $g_{ij}^0 = \nabla f_{ij}(x^0)$ for all $i,j$, this leads to the complexity \eqref{ch3:eq:98y98fd}; that is, by cloning client machines, we can replace $\color{red}L_{\rm QM}$ in the standard rate with $\sqrt{2} \color{blue} L_{\rm AM}$. A similar result can be obtained even if we do not ignore the integrality constraint, but we do not include it for brevity reasons. However, it is important to note that the cloning approach has several straightforward shortcomings, which we will address in the next section.\footnote{	In our work, we address an optimization problem of the form 
	$
	\min_{\substack{w_j \geq 0 \ \forall j\in[n]; \sum_{i=1}^{n} w_i = 1}} \sum_{i=1}^{n} \dfrac{a_i^2}{w_i},
	$
	where \(a_i\) represent certain constants. This formulation bears a resemblance to the meta-problem in the importance sampling strategy discussed in \citep{so_importance_sampling}. Despite the apparent similarities in the abstract formulation, our approach and the one in the referenced work diverge significantly in both motivation and implementation. While \citet{so_importance_sampling} applies importance sampling to reduce the variance of a stochastic gradient estimator by adjusting sampling probabilities, our method involves adjusting client cloning weights without sampling. Furthermore, our gradient estimator is biased, unlike the unbiased estimator in the referenced paper, and we aim to minimize the quadratic mean of the smoothness constants, which is inherently different from the objectives in \citet{so_importance_sampling}. Although both approaches can be expressed through a similar mathematical framework, they are employed in vastly different contexts, and any parallelism may be coincidental rather than indicative of a direct connection.}

\subsection{Step 3: From client cloning to update weighting} 
\label{ch3:sec:weights}

It is evident that employing client cloning improves the convergence rate. Nevertheless, there are obvious drawbacks associated with this approach. Firstly, it necessitates a larger number of computational devices, rendering its implementation less appealing from a resource allocation perspective. Secondly, the utilization of \algname{EF21} with cloned machines results in a departure from the principles of Federated Learning, as it inherently compromises user privacy -- transferring data from one device to another is prohibited in FL.

However, a  simpler approach to implementing the cloning idea emerges when we assume the compressors used to be {\em deterministic}. To illustrate this, let us initially examine how we would typically implement \algname{EF21} with cloned machines:
\begin{align} 
	&x^{t+1} = x^t - \gamma \dfrac{1}{N}\sum\limits_{i=1}^n \sum\limits_{j=1}^{N_i} g_{ij}^t, \label{ch3:eq:AlgStep1}  \\
	&g_{ij}^{t+1} = g_{ij}^t + \cC_{ij}^t ( \nabla f_{ij}(x^{t+1}) - g_{ij}^t), \quad i \in [n],  \quad j \in [N_i]. \label{ch3:eq:AlgStep2}
\end{align}

We will now rewrite the same method in a different way. Assume we choose $g_{ij}^0 = g_i^0$ for all $j\in[N_i]$. We show by induction that $g_{ij}^t$ is the same for all $j \in [N_i]$. We have just seen that this holds for $t=0$. Assume this holds for some $t$. Then since $ \nabla f_{ij}(x^{t+1}) = \dfrac{N}{n N_i} \nabla f_i(x^{t+1})$ for all $j\in[N_i]$ combined with the induction hypothesis, \eqref{ch3:eq:AlgStep2} and the determinism of $\cC_{ij}^t$, we see that $g_{ij}^{t+1}$ is the same for all $j \in [N_i]$. Let us define $g_{i}^t \equiv g_{ij}^t$ for all $t$. This is a valid definition since we have shown that $g_{ij}^t$ does not depend on $j$. Because of all of the above, iterations \eqref{ch3:eq:AlgStep1}, \eqref{ch3:eq:AlgStep2} can be equivalently written in the form
\begin{align}
	&x^{t+1} = x^t - \gamma \sum\limits_{i=1}^n {\color{ForestGreen}w_i}  g_{i}^t, \label{ch3:eq:AlgStep1X} \\
	&g_{i}^{t+1} = g_{i}^t + \cC_i^t\left( \dfrac{1}{n {\color{ForestGreen}w_i}} \nabla f_i(x^{t+1}) - g_{i}^t \right), \quad i\in [n], \label{ch3:eq:AlgStep2X} 
\end{align}
where ${\color{ForestGreen}w_i} = \dfrac{L_i}{\sum_j L_j}$. 

This transformation effectively enables us to operate the method on the original $n$ clients, eliminating the need for $N$ clients! This refinement has led to the creation of a new algorithm that outperforms \algname{EF21} in terms of convergence rate, which we call \algname{EF21-W} (Algorithm~\ref{ch3:alg:EF21-W}).

\begin{algorithm}[!t]
	\begin{algorithmic}[1]
		\STATE {\bfseries Input:} initial model parameters $x^0 \in \RR^d$; initial gradient estimates $g_1^0, g_2^0, \dots,g_n^0 \in \R^d$ stored at the server and the clients; weights ${\color{ForestGreen}w_i} = \nicefrac{L_i}{\sum_j L_j}$; step size $\gamma>0$; number of iterations $T > 0$
		\STATE {\bfseries Initialize:} $g^0 = \sum_{i=1}^n {\color{ForestGreen}w_i} g_i^0 $ on the server
		\FOR{$t = 0, 1, 2, \dots, T - 1 $}
		\STATE Server computes $x^{t+1} = x^t - \gamma g^t$		 and broadcasts  $x^{t+1}$ to all $n$ clients
		\FOR{$i = 1, \dots, n$ {\bf on the clients in parallel}} 
		\STATE Compute $u_i^t=\cC_i^t (\dfrac{1}{n {\color{ForestGreen}w_i}}\nabla f_i(x^{t+1}) - g_i^t)$ and update $g_i^{t+1} = g_i^t +u_i^t$ \label{ch3:alg_line:g_update_step}
		\STATE Send the compressed message $u_i^{t}$ to the server
		\ENDFOR 
		\STATE Server updates $g_i^{t+1} = g_i^t +u_i^t$ for all $i\in [n]$, and computes $g^{t+1} = \sum_{i=1}^n {\color{ForestGreen}w_i} g_i^{t+1}$
		\ENDFOR
		\STATE {\bfseries Output:} Point $\hat{x}^T$ chosen from the set $\{x^0, \dots, x^{T-1}\}$ uniformly at random
	\end{algorithmic}
	\caption{\algname{EF21-W}: Weighted Error Feedback 2021.}
	\label{ch3:alg:EF21-W}
\end{algorithm}
While we relied on assuming that the compressors are \textit{deterministic} in order to motivate the transition from $N$ to $n$ clients, it turns out that \algname{EF21-W} converges without the need to invoke this assumption. \begin{theorem}[\textbf{Theory for \algname{EF21-W}}]\label{ch3:thm:EF21-W}
	Consider Algorithm~\ref{ch3:alg:EF21-W} (\algname{EF21-W}) applied to the distributed optimization Problem~\eqref{ch3:eq:main_problem}.
	Let Assumptions~\ref{ch3:as:smooth}, \ref{ch3:as:L_i}, \ref{ch3:as:lower_bound} hold, assume that $\cC_i^t \in \mathbb{C}(\alpha)$ for all $i\in [n]$ and $t\geq 0$, set $$G^t \eqdef \sum\limits_{i=1}^n {\color{ForestGreen}w_i}  \norm{g_i^t - \dfrac{1}{n\color{ForestGreen}w_i} \nabla f_i(x^t)}^2,$$ where ${\color{ForestGreen}w_i} = \dfrac{L_i}{\sum_j L_j}$ for all $i\in [n]$, and let the step size satisfy
	$$
	0< \gamma \leq \dfrac{1}{L +  \LAM  \xi(\alpha)}.
	$$
	If for $T>1$ we define $\hat{x}^T$ as an element of the set  $\{x^0, x^1, \dots, x^{T-1}\}$ chosen uniformly at random, then
	\begin{equation}
		\ExpBr{\norm{ \nabla f(\hat{x}^T)  }^2} \leq \dfrac{2(f(x^0) - f^\ast)}{\gamma T} + \dfrac{G^0}{\theta(\alpha) T}.
	\end{equation}
\end{theorem}
\subsection{Step 4: From weights in the algorithm to weights in the analysis}
\label{ch3:sec:weighted-analysis}

In the preceding section, we introduced a novel algorithm: \algname{EF21-W}. While it bears some resemblance to the vanilla \algname{EF21} algorithm~\citep{EF21} (we recover it for uniform weights), the reliance on  particular non-uniform weights  enables it to achieve a faster convergence rate. However, this is not the end of the story as another insight reveals yet another surprise.

Let us consider the scenario when the compressors in Algorithm~\ref{ch3:alg:EF21-W} are  {\em positively homogeneous}\footnote{A compressor $\cC:\R^d\to \R^d$ is positively homogeneous if $\cC(tx) = t\cC(x)$ for all $t>0$ and $x\in\RR^d$.}. Introducing the new variable $h_i^t = n {\color{ForestGreen}w_i} g_i^t$, we can reformulate the gradient update in Algorithm~\ref{ch3:alg:EF21-W} to \begin{align*}
		h_i^{t+1} & = {n{\color{ForestGreen}w_i}} g_i^{t+1} \overset{\eqref{ch3:eq:AlgStep2X}}{=} {n{\color{ForestGreen}w_i}} \left[g_{i}^t + \cC_i^t\left( \dfrac{\nabla f_i(x^{t+1})}{n {\color{ForestGreen}w_i}}  - g_{i}^t \right) \right]  =  h_i^t + \cC_i^t (\nabla f_i(x^t) - h_i^t),
\end{align*}
indicating that $h_i^t$ adheres to the update rule of the vanilla \algname{EF21} method! Furthermore, the iterates $x^t$  also follow the same rule as \algname{EF21}:
\[
x^{t+1} \overset{\eqref{ch3:eq:AlgStep1X}}{=} x^t - \gamma \sum \limits_{i=1}^n{\color{ForestGreen}w_i} g^t =  x^t - \gamma \sum \limits_{i=1}^n {\color{ForestGreen}w_i} \dfrac{1}{n{\color{ForestGreen}w_i}} h_i^t = x^t - \gamma \dfrac{1}{n}\sum \limits_{i=1}^n h_i^t. 
\]

\begin{center}
	\textit{So, what does this mean?}
\end{center}

One interpretation suggests that for positively homogeneous contractive compressors, the vanilla \algname{EF21} algorithm is equivalent to \algname{EF21-W}, and hence inherits its faster convergence rate that depends on $\LAM$ rather than on $\LQM$. However, it turns out that we can establish the same result without having to resort to positive homogeneity altogether. For example, the "natural compression" quantizer, which rounds the components of an input vector from 
$\RD$ elementwise to one of the two nearest powers of two is not positively homogeneous \citep{horvath2019natural}.

\begin{theorem}[{\bf New theory for \algname{EF21}}]\label{ch3:thm:ef21_new_result}
	Consider Algorithm~\ref{ch3:alg:EF21} (\algname{EF21}) applied to the distributed optimization Problem~\eqref{ch3:eq:main_problem}. Let Assumptions~\ref{ch3:as:smooth}, \ref{ch3:as:L_i}, \ref{ch3:as:lower_bound} hold, assume that $\cC_i^t \in \mathbb{C}(\alpha)$ for all $i\in [n]$ and $t\geq 0$, set $$ G^t \eqdef \dfrac{1}{n}\sum \limits_{i=1}^n \dfrac{1}{n {\color{ForestGreen}w_i}} \norm{g_i^t - \nabla f_i(x^t)}^2,$$ where ${\color{ForestGreen}w_i} = \dfrac{L_i}{\sum_j L_j}$ for all $i\in [n]$, and let the step size satisfy
	$$
	0< \gamma \leq \dfrac{1}{L + \LAM \xi(\alpha) }.
	$$
	If for $T>1$ we define $\hat{x}^T$ as an element of the set $\{x^0, x^1, \dots, x^{T-1}\}$ chosen uniformly at random, then
	\begin{equation}
		\ExpBr{\| \nabla f(\hat{x}^T)  \|^2} \leq \dfrac{2(f(x^0) - f^\ast)}{\gamma T} + \dfrac{G^0}{\theta(\alpha) T}.
	\end{equation}
\end{theorem}
This last result effectively pushes the weights from the algorithm in \algname{EF21-W} to the proof, which enabled us to show that the original \algname{EF21} method also enjoys the same improvement: from $\LQM$ to $\LAM$.

\section{Experiments}
\label{ch3:sec:experiments-main}

\subsection{Non-convex {logistic regression} on benchmark datasets}
\label{ch3:sec:experiments-main-real}

\begin{center}	
	\begin{figure*}[t!]
		\centering
		\captionsetup[sub]{font=normalsize,labelfont={}}	
		\captionsetup[subfigure]{labelformat=empty}
		\newcommand{\myVar}{0.49\textwidth}
		\newcommand{\myVarN}{1.0\textwidth}
		
		\begin{subfigure}[ht]{\myVar}
			\includegraphics[width=\myVarN]{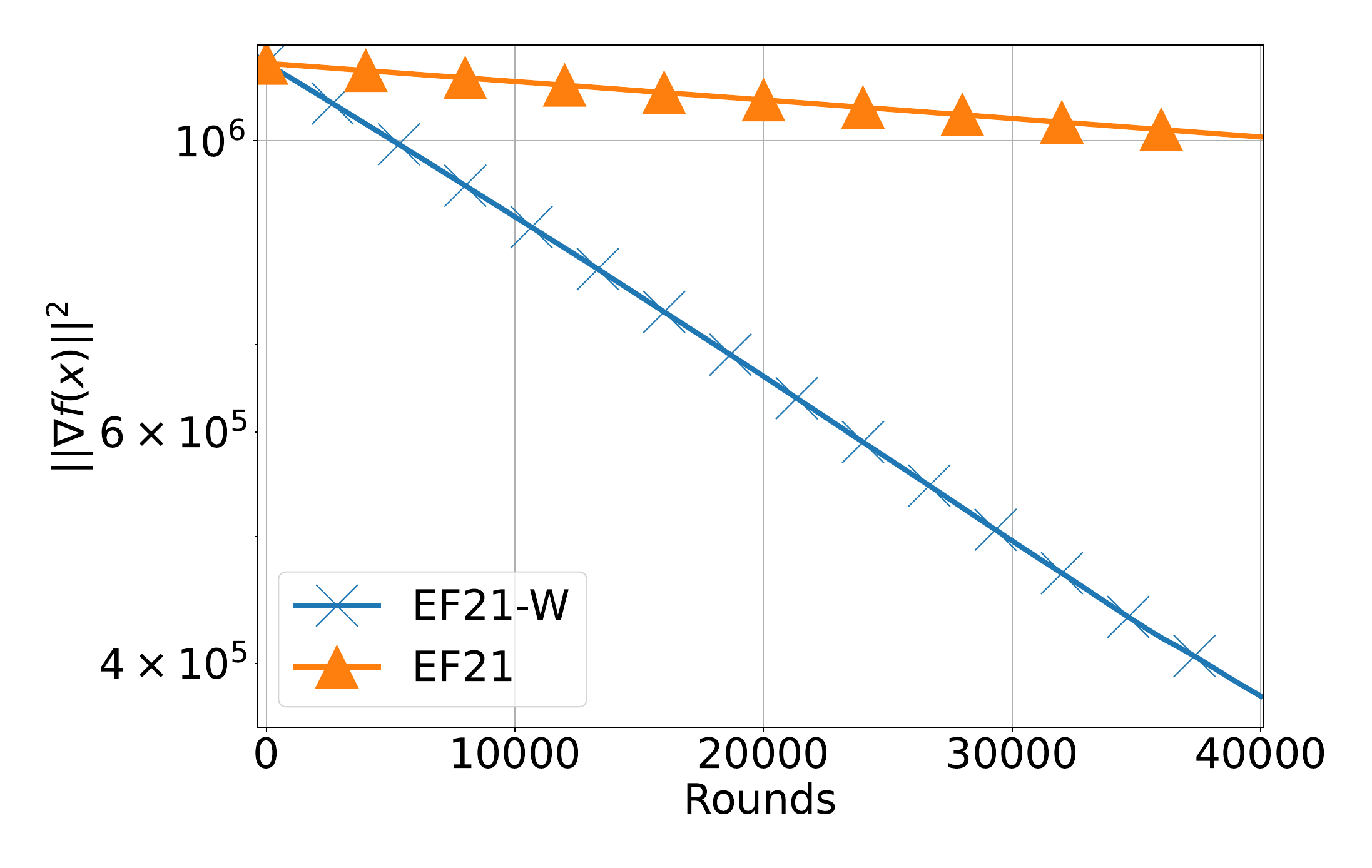} 
			\caption{ { (a) \dataname{AUSTRALIAN}, $\Lvar \approx 10^{16}$ } }
		\end{subfigure}		
		\begin{subfigure}[ht]{\myVar}
			\includegraphics[width=\myVarN]{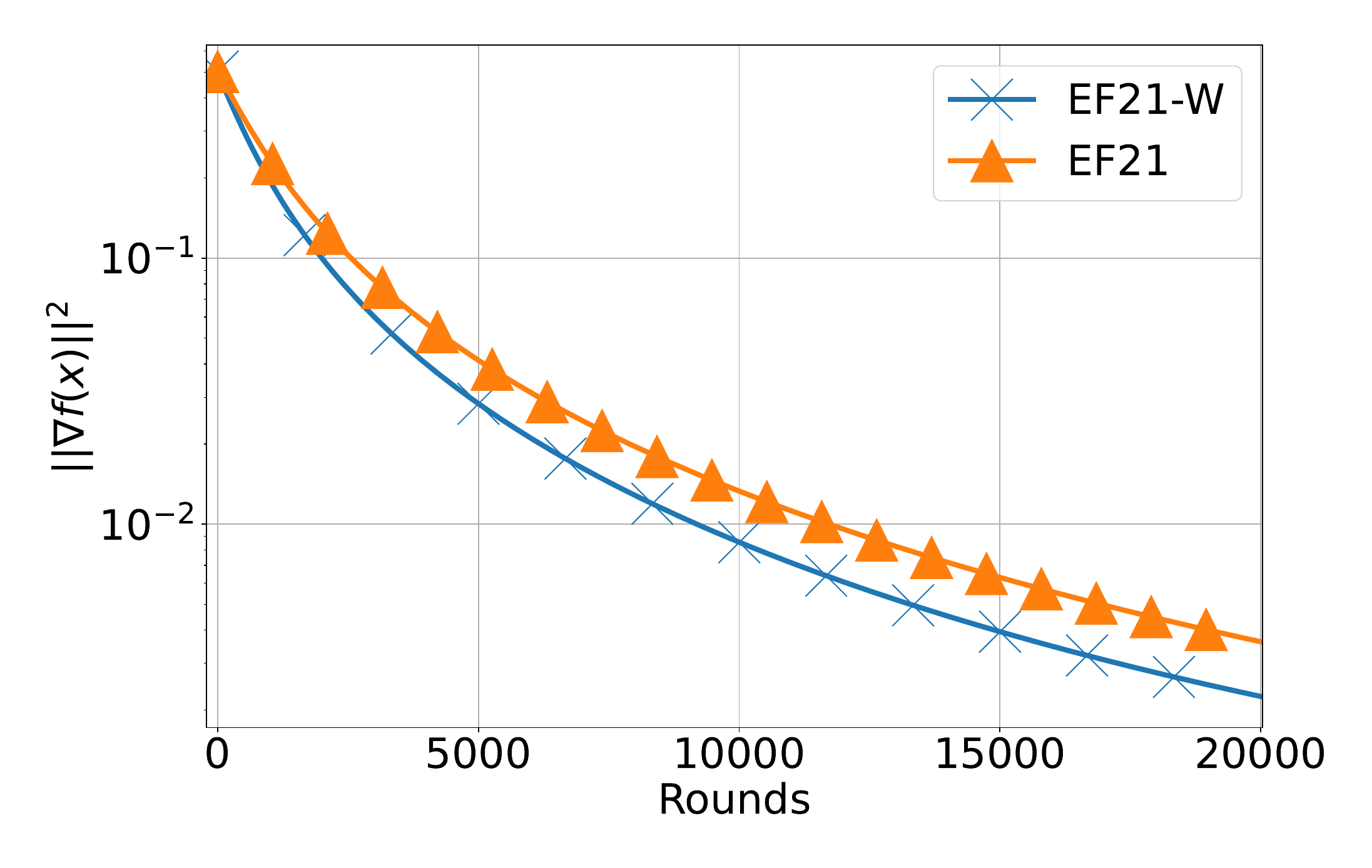} \caption{ {(b) \dataname{W1A}, $\Lvar\approx 3.28$ } }
		\end{subfigure}
		
		\begin{subfigure}[ht]{\myVar}
			\includegraphics[width=\myVarN]{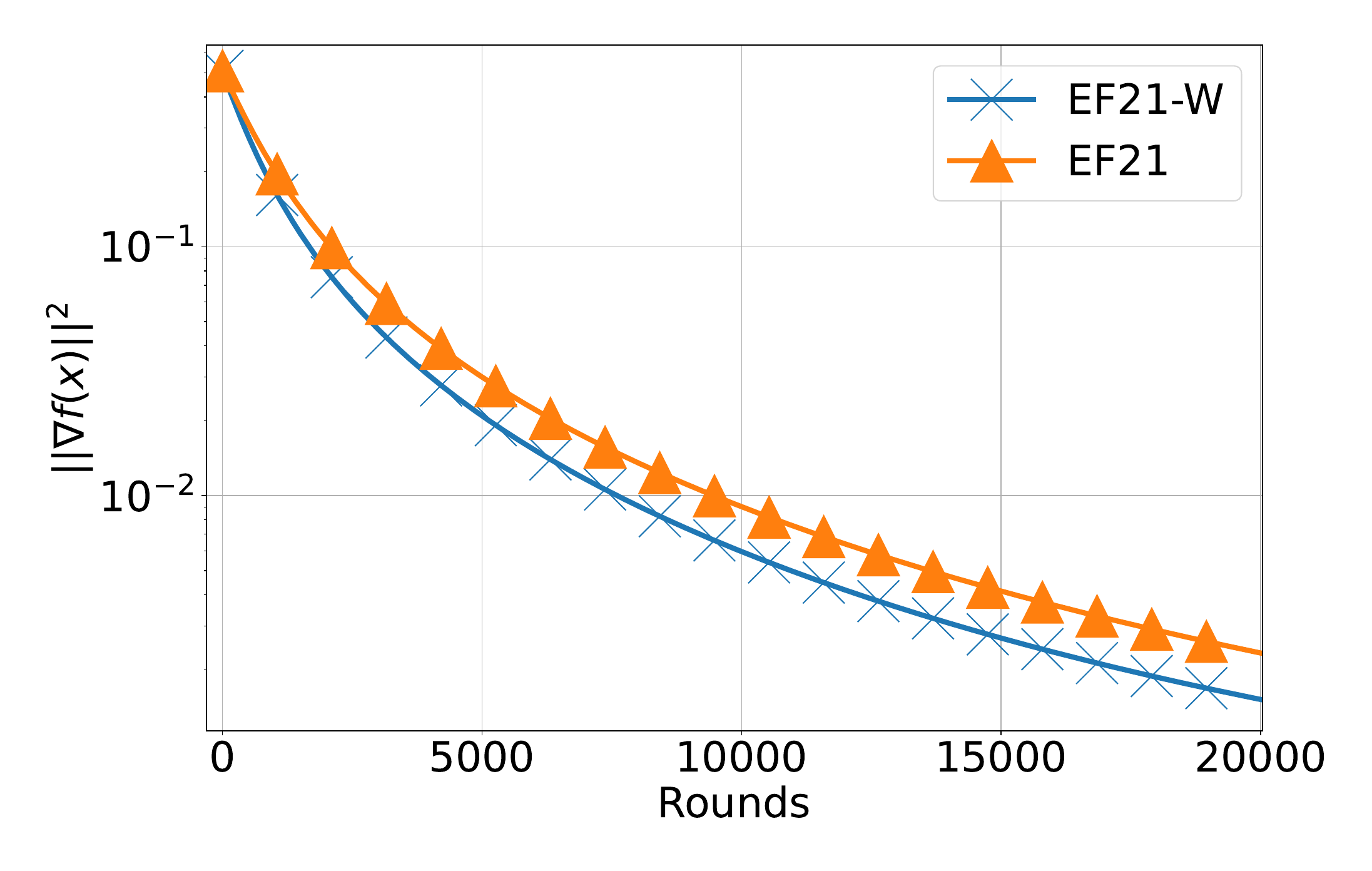} 
			\caption{ {(c) \dataname{W2A}, $\Lvar\approx2.04$ } }
		\end{subfigure}		
		\begin{subfigure}[ht]{\myVar}
			\includegraphics[width=\myVarN]{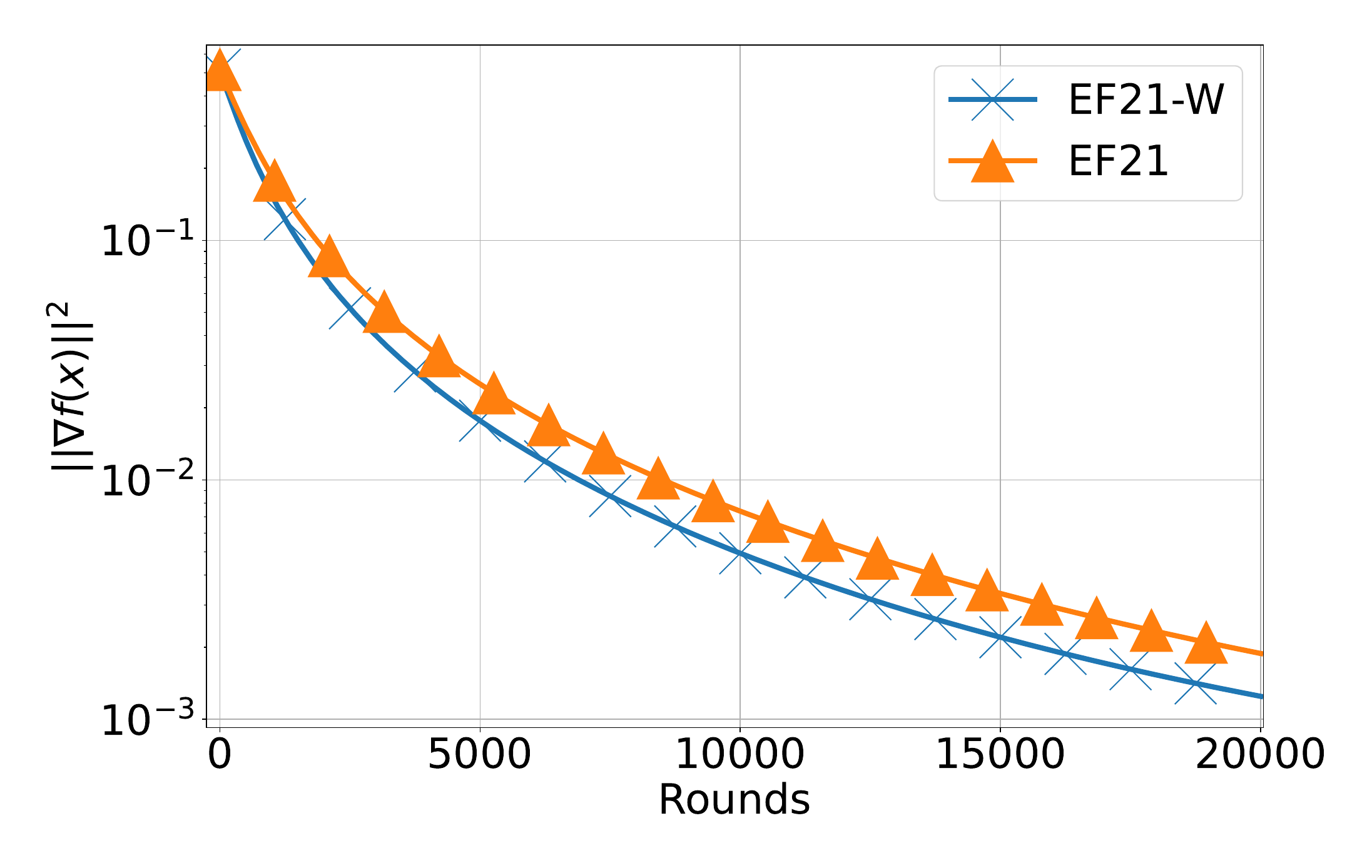} 
			\caption{ {(d) \dataname{W3A}, $\Lvar\approx1.58$ } }
		\end{subfigure}
		
		\begin{subfigure}[ht]{\myVar}
			\includegraphics[width=\myVarN]{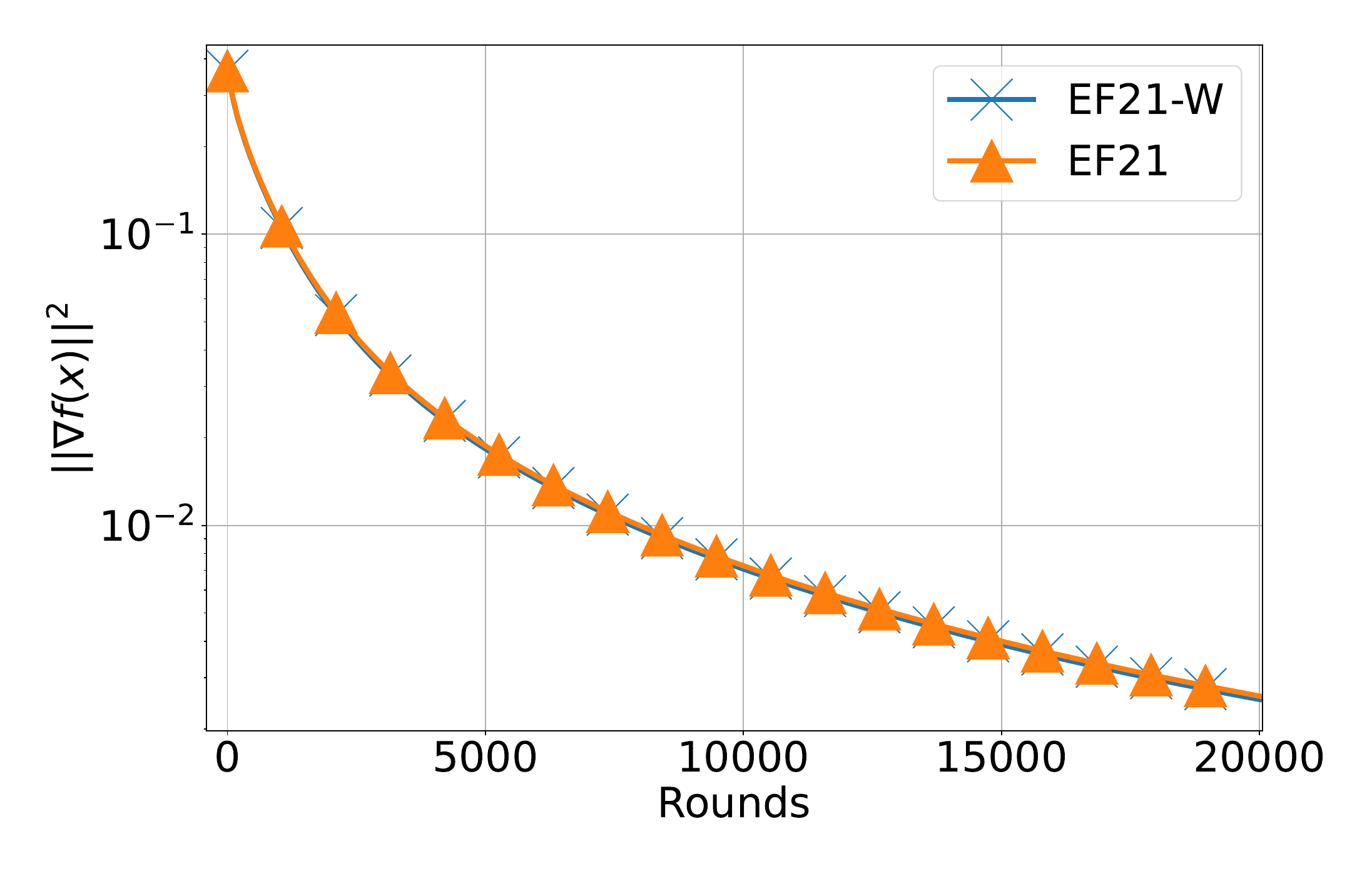} \caption{ {(e) \dataname{MUSHROOMS}, $\Lvar\approx 5 \cdot 10^{-1}$ } }
		\end{subfigure}
		\begin{subfigure}[ht]{\myVar}
			\includegraphics[width=\myVarN]{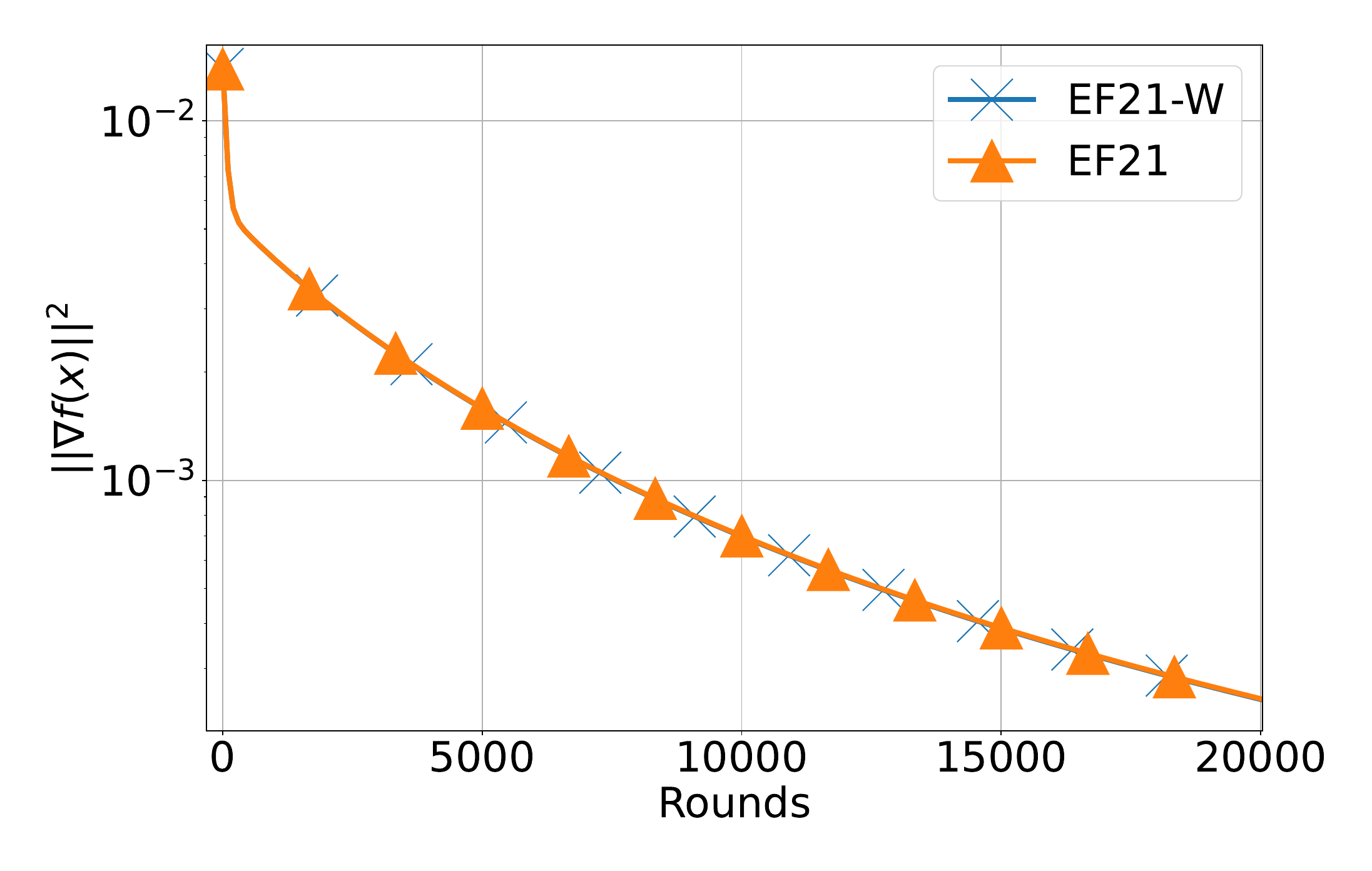} \caption{ {(f) \dataname{PHISHING}, $\Lvar=9 \cdot 10^{-4}$ } }
		\end{subfigure}
		
		\caption{{Comparison of  \algname{EF21} versus our new \algname{EF21-W} with the \compname{Top1} compressor on the non-convex \modelname{logistic regression} problem. The number of clients $n$ is $1,000$. The step size for~\algname{EF21} is set according to~\citep{EF21}, and the step size for~\algname{EF21-W} is set according to Theorem~\ref{ch3:thm:EF21-W}. The coefficient $\lambda$ for (b)--(f) is set to $0.001$, and for (a) is set to $1,000$ for numerical stability. We let $\Lvar \eqdef \LQMsq - \LAMsq = {\color{red}\avein L_i^2} - {\color{blue}\left(\avein L_i \right)^2}$.}}
		\label{ch3:fig:real-ef21-vc-ncvx}
	\end{figure*}
\end{center}

In our first experiment, we employed a \modelname{logistic regression} model with a non-convex regularizer, i.e., 
\begin{equation*}
	f_i(x) \eqdef \dfrac{1}{n_i} \sum \limits_{j=1}^{n_i} \log \left(1+\exp({-y_{ij} \cdot a_{ij}^{\top} x})\right) + \lambda  \sum\limits_{j=1}^{d} \dfrac{x_j^2}{x_j^2 + 1},
\end{equation*}
where $(a_{ij},  y_{ij}) \in \mathbb{R}^{d} \times \{-1,1\}$ represents the $j$-th data point out from a set of $n_i$ data points stored at client $i$, and $\lambda>0$ denotes a regularization coefficient. We utilized six datasets from \dataname{LIBSVM} \citep{chang2011libsvm}.  The dataset shuffling strategy, detailed in Appendix~\ref{ch3:app:dataset-shuffling-for-libsvm}, was employed to emulate heterogeneous data distribution. Each client was assigned the same number of data points. Figure~\ref{ch3:fig:real-ef21-vc-ncvx} provides a comparison between \algname{EF21} employing the original step size~\citep{EF21} and \algname{EF21-W} with the better step size. The initial gradient estimators were chosen as $g_i^0 = \nabla f_i(x^0)$ for all $ i \in [n]$. As evidenced empirically, the \algname{EF21-W} algorithm emerges as a practical choice when utilized in situations characterized by high variance in smoothness constants. As evident from the plots, the algorithm employing the new step size exhibits superior performance compared to its predecessor.

\begin{center}
	\begin{figure*}[t!]
		\centering
		\captionsetup[sub]{font=normalsize,labelfont={}}	
		\captionsetup[subfigure]{labelformat=empty}
		\newcommand{\sfwidth}{0.49\textwidth}
		\newcommand{\figwidth}{1.0\textwidth}
		
		\begin{subfigure}[ht]{\sfwidth}			\includegraphics[width=\figwidth]{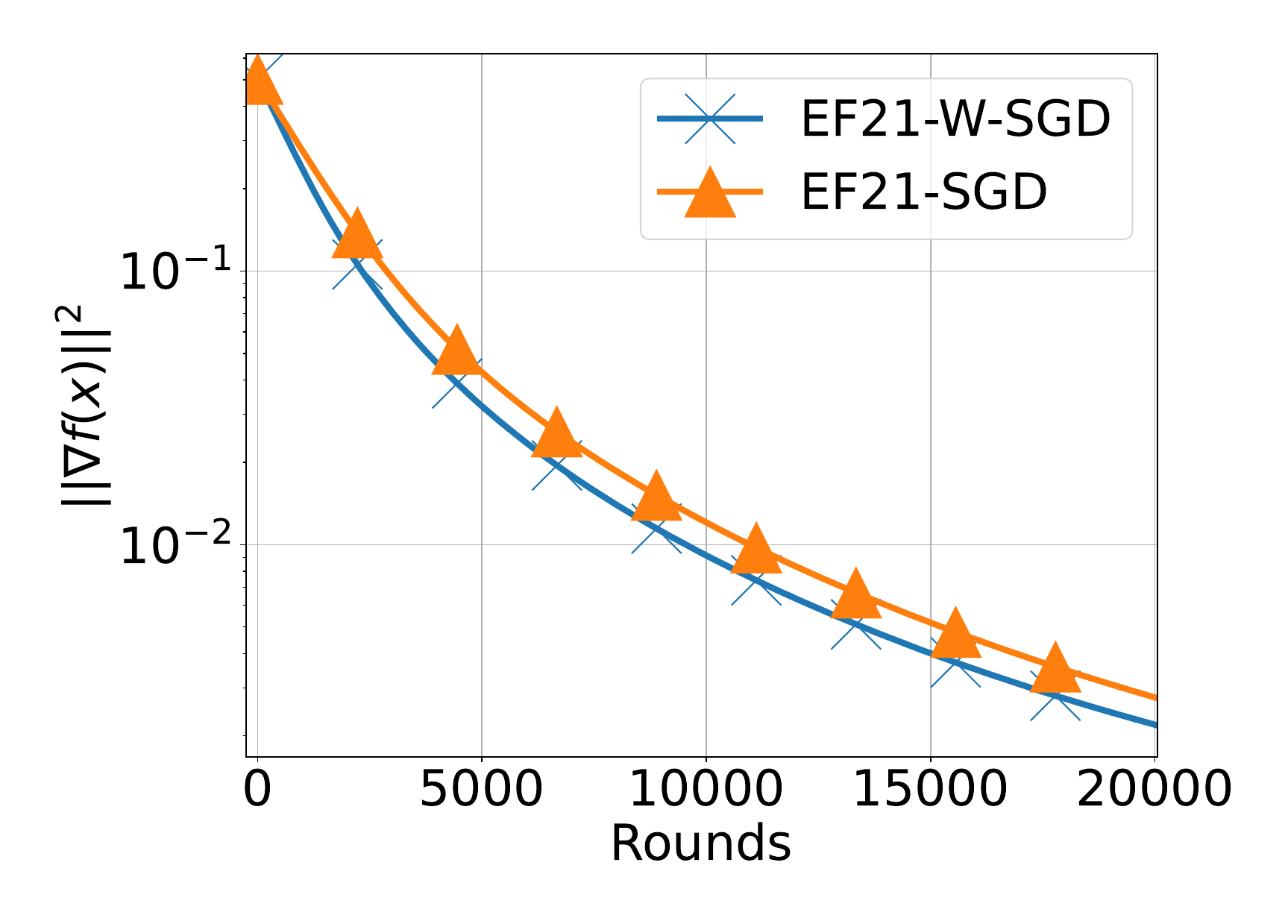} \caption{ {(a) \dataname{W1A}, \algname{SGD} } }
		\end{subfigure}
		\begin{subfigure}[ht]{\sfwidth}
			\includegraphics[width=\figwidth]{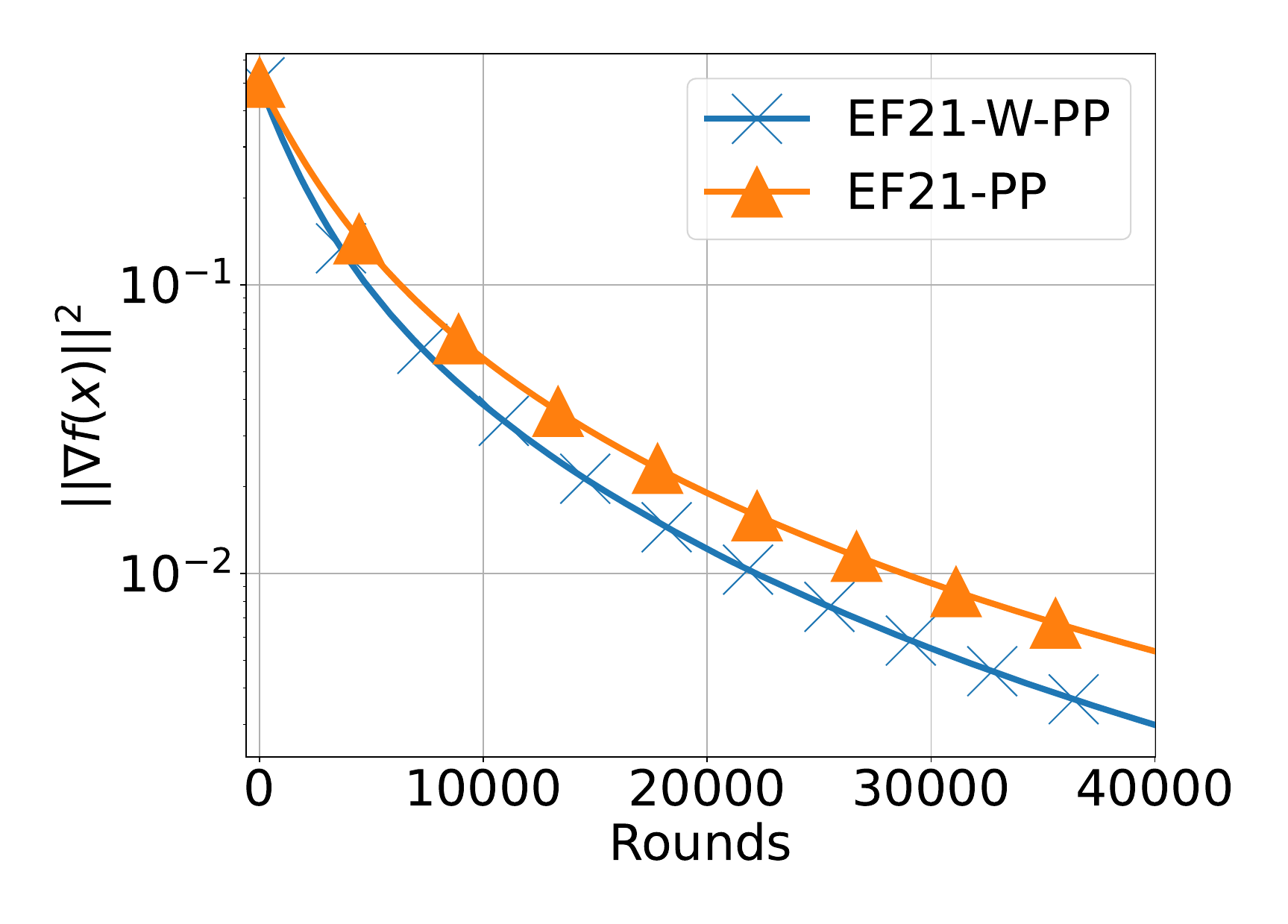} \caption{ {(b) \dataname{W1A}, \algname{PP} } }
		\end{subfigure}
		
		\begin{subfigure}[ht]{\sfwidth}
			\includegraphics[width=\figwidth]{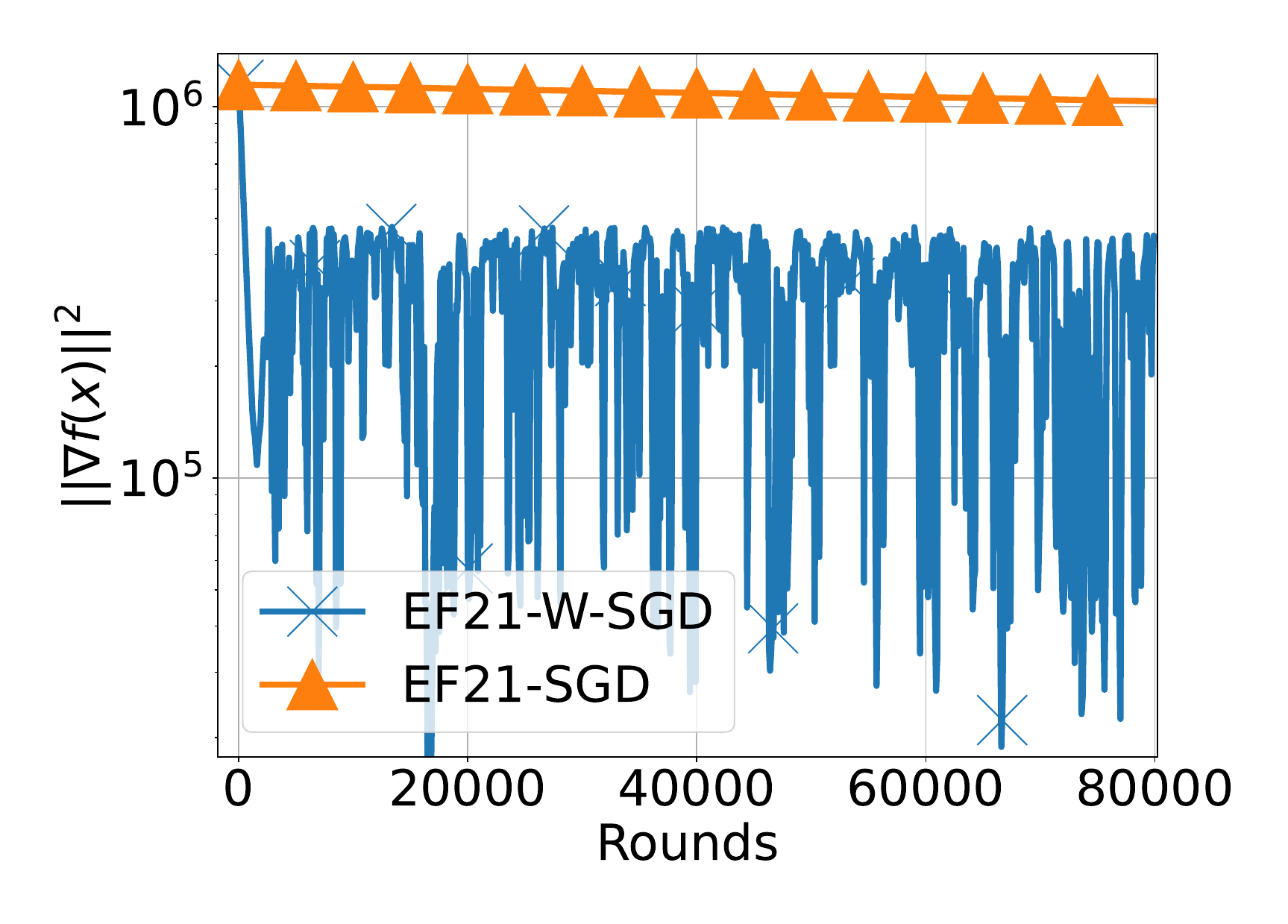} \caption{ { (c) \dataname{AUSTRALIAN}, \algname{SGD} } }
		\end{subfigure}
		\begin{subfigure}[ht]{\sfwidth}
			\includegraphics[width=\figwidth]{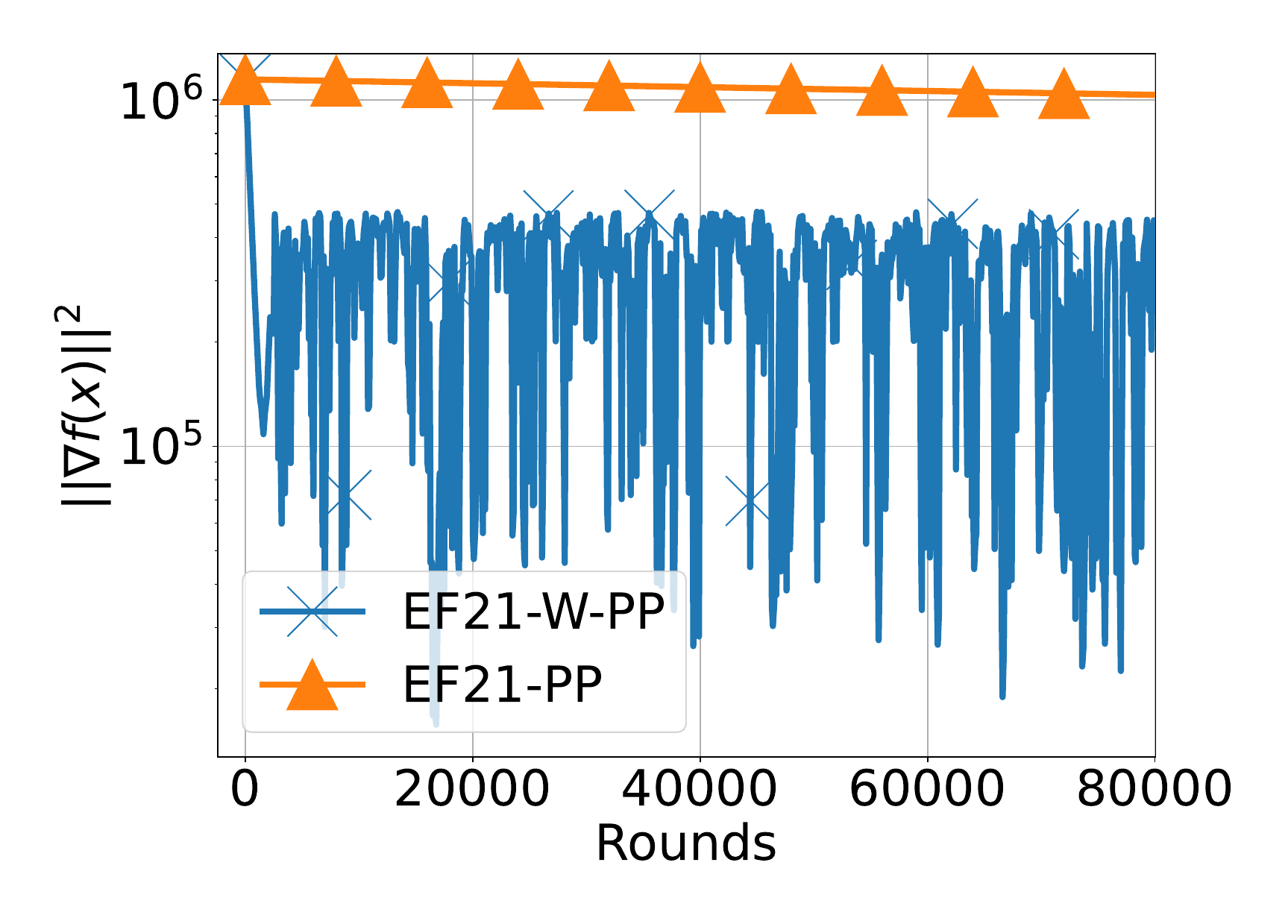} \caption{ { (d) \dataname{AUSTRALIAN}, \algname{PP} } }
		\end{subfigure}
		
		\caption{{Comparison of \algname{EF21-W} with partial partial participation (\algname{EF21-W-PP}) or stochastic gradients (\algname{EF21-W-SGD}) versus \algname{EF21}  with partial partial participation (\algname{EF21-PP}) or stochastic gradients (\algname{EF21-SGD})~\citep{fatkhullin2021ef21}. The \compname{Top1} compressor was employed in all experiments. The number of clients $n=1,000$. All step sizes are theoretical. The coefficient $\lambda$ was set to $0.001$ for (a, b) and to $1,000$ for (c, d).}}
		\label{ch3:fig:ext-real-ef21-vc-ncvx}
	\end{figure*}
\end{center}

Next, we conducted a comparative analysis of the performance of \algname{EF21-W-PP} and \algname{EF21-W-SGD}, as elucidated in the appendix, compared to  their non-weighted counterparts. In the \algname{EF21-PP}/\algname{EF21-W-PP} algorithms, each client participated independently in each round with probability $p_i=0.5$. Moreover, in the case of \algname{EF21-SGD}/\algname{EF21-W-SGD} algorithms, a single data point was stochastically sampled from a uniform distribution at each client during each iteration of the algorithm. As observed in Figure~\ref{ch3:fig:ext-real-ef21-vc-ncvx}, the algorithms employing the new learning rates demonstrate faster convergence. Notably, Figure~\ref{ch3:fig:ext-real-ef21-vc-ncvx}~(c) depicts more pronounced oscillations with updated step sizes, as the new analysis permits larger step sizes, which can induce oscillations in stochastic methods.

\subsection{Non-convex linear model on synthetic datasets}
\label{ch3:sec:experiments-main-syb-ncvx}
\begin{center}	
	\begin{figure*}[t!]
		\centering
		\captionsetup[sub]{font=normalsize,labelfont={}}	
		\captionsetup[subfigure]{labelformat=empty}
		
		\begin{subfigure}[ht]{0.65\textwidth}
			\includegraphics[width=\textwidth]{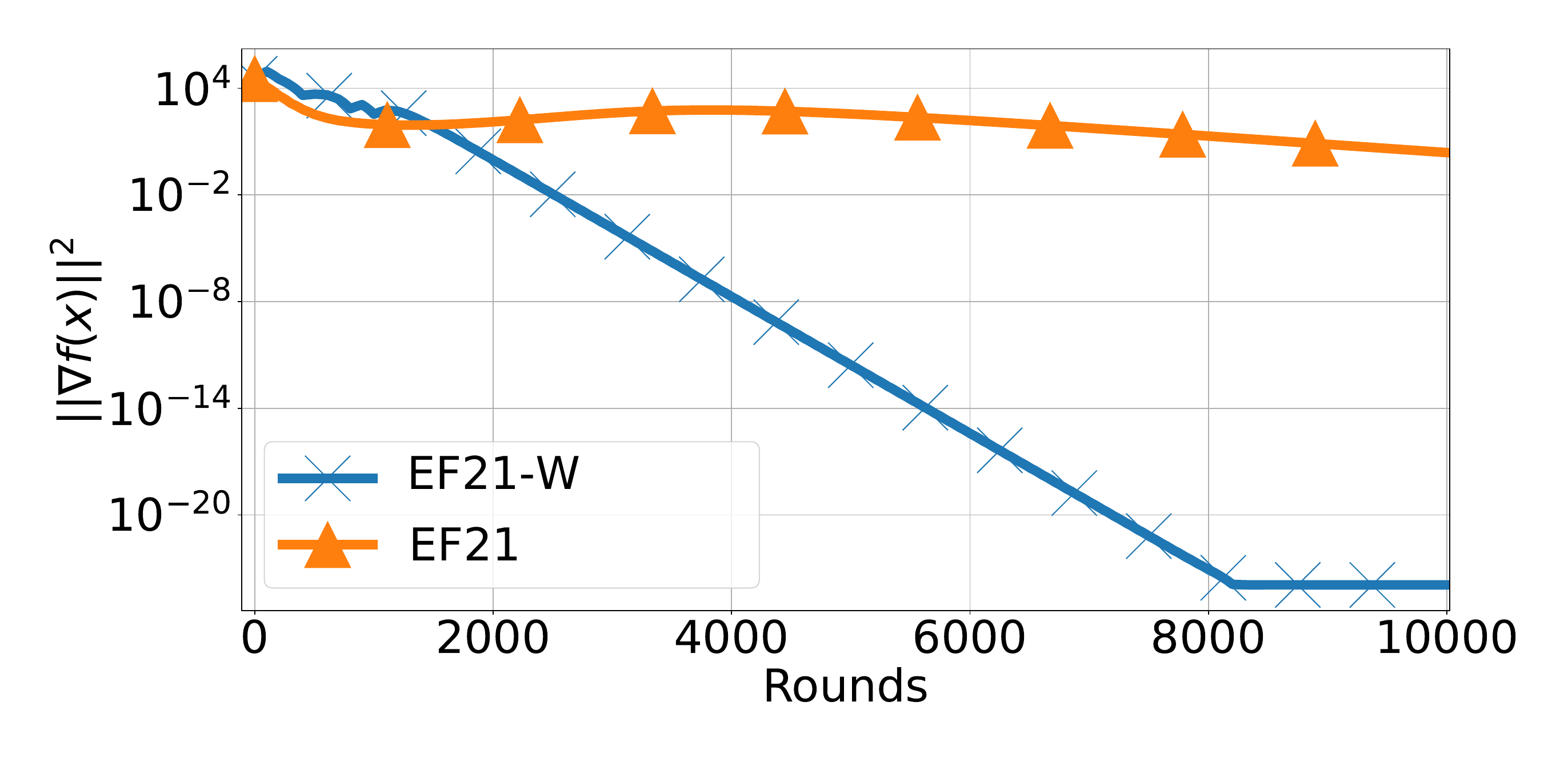} \caption{{(a) $\Lvar  \approx 4.4 \cdot 10^6$} \quad {$\LQM  \approx 2126, \LAM \approx 252$}}
		\end{subfigure}
		
		\begin{subfigure}[ht]{0.65\textwidth}
			\includegraphics[width=\textwidth]{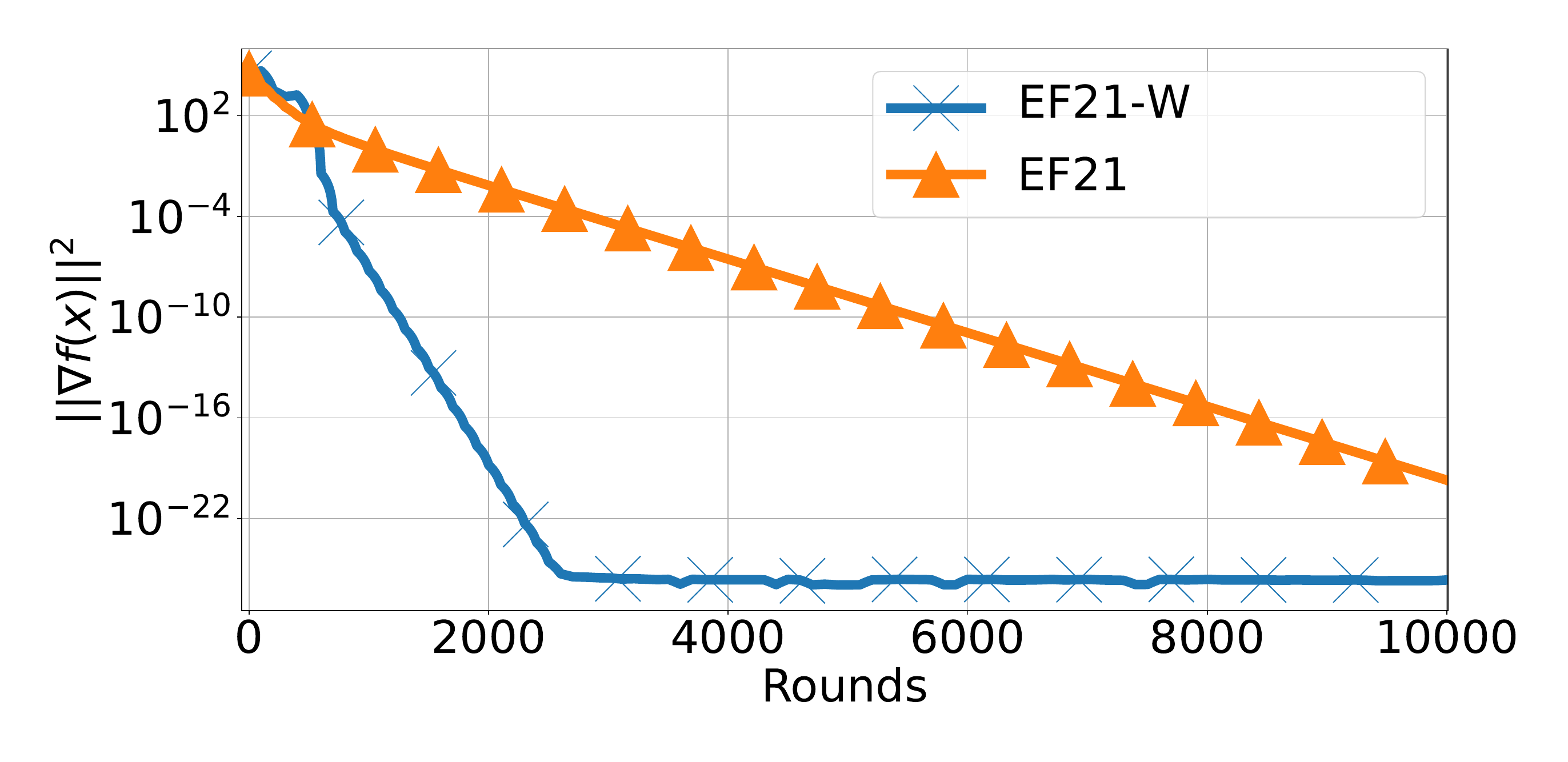} \caption{{(b) $\Lvar \approx 1.9 \cdot 10^6$} \quad { $\LQM  \approx 1431, \LAM \approx 263$
			}} 
		\end{subfigure}
		
		\begin{subfigure}[ht]{0.65\textwidth}
			\includegraphics[width=\textwidth]{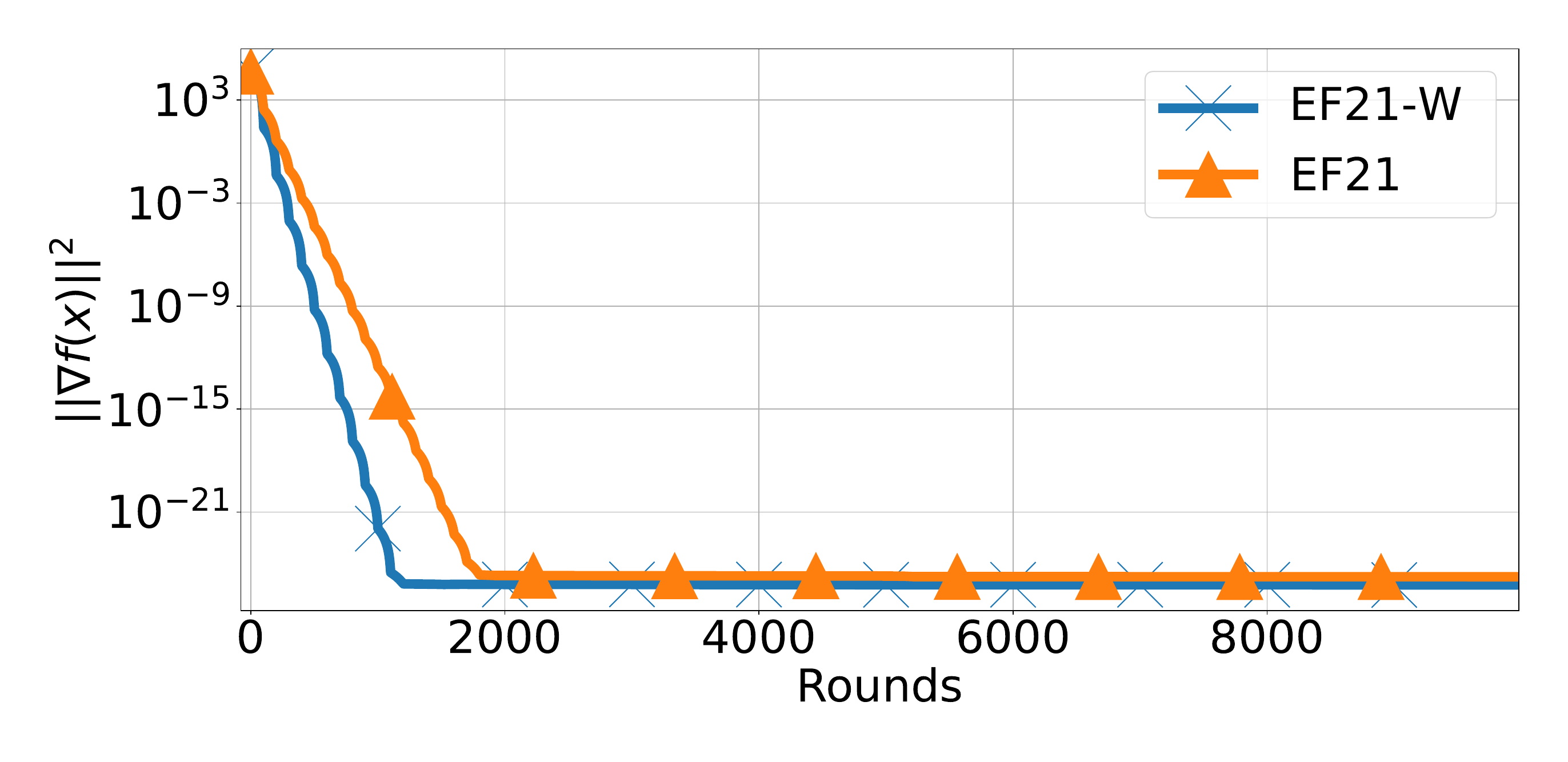} \caption{{(c) $\Lvar \approx 1.0 \cdot 10^5$} \quad { $\LQM \approx 433, \LAM\approx 280$
			}}
		\end{subfigure}
		
		\begin{subfigure}[ht]{0.65\textwidth}
			\includegraphics[width=\textwidth]{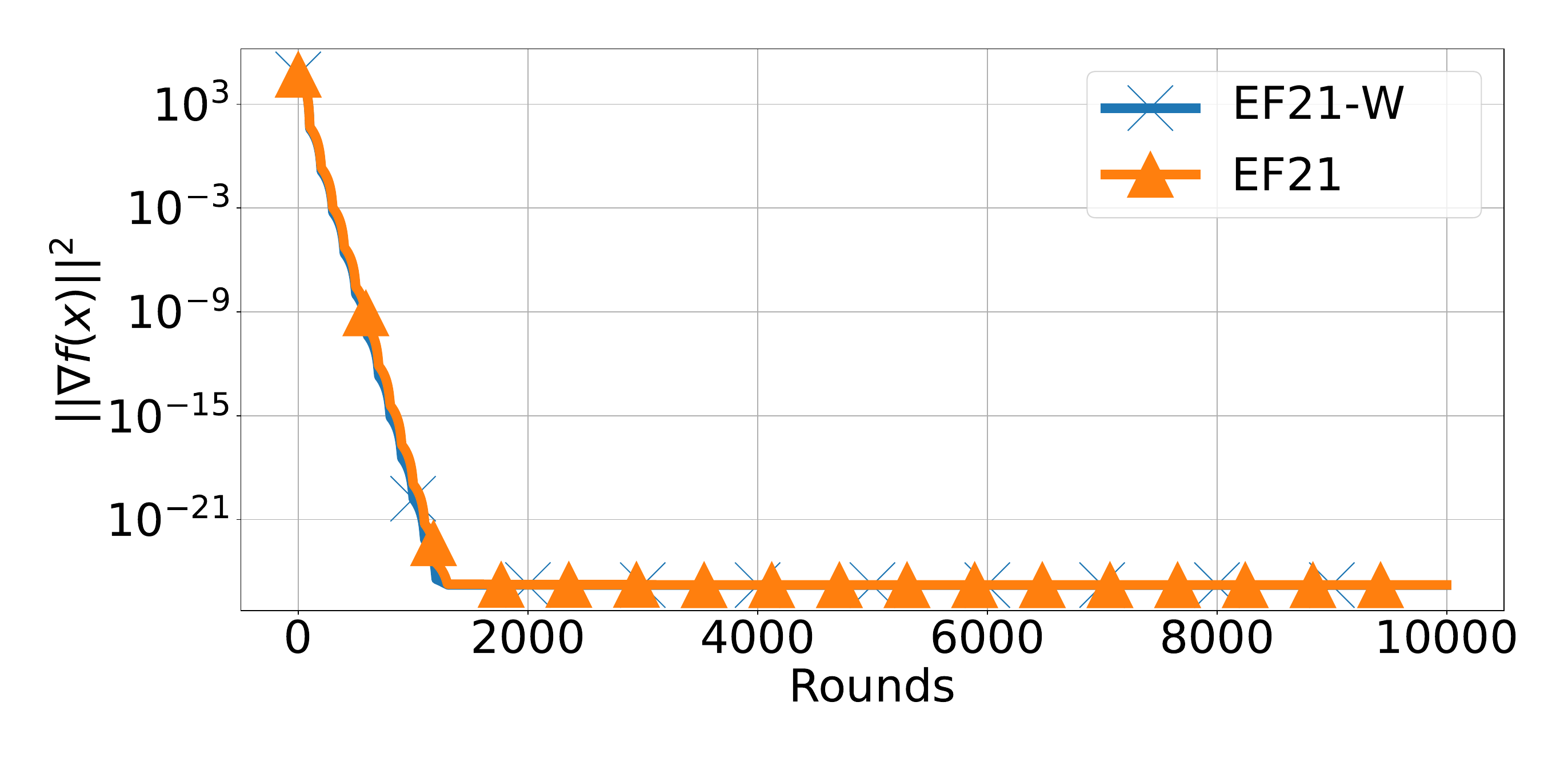} \caption{{(d) $\Lvar \approx 5.4 \cdot 10^3$} \quad { $\LQM  \approx 294, \LAM \approx 285$
			}} 
		\end{subfigure}
		
		\caption{ {Comparison of~\algname{EF21} and \algname{EF21-W} with the~\compname{Top1} compressor on the non-convex linear problem. The number of clients $n$ is $2,000$. The coefficient $\lambda$ has been set to $100$. The step size for~\algname{EF21} is set according to~\citep{EF21}, and the step size for~\algname{EF21-W} is set according to Theorem~\ref{ch3:thm:EF21-W}. In all cases, the smoothness constant $L$ equals $50$}.}
		\label{ch3:fig:syn-ef21-vc-noncvx}
	\end{figure*}
\end{center}

In our second set of experiments, we trained a linear regression model with a non-convex regularizer. The function $f_i$ for the linear regression problem is defined as follows:
$$f_i(x) \eqdef \dfrac{1}{n_i} \norm{\mA_i x - {b_i}}^2 + \lambda  \sum_{j=1}^{d} \dfrac{x_j^2}{x_j^2 + 1}.
$$
Here, $\mA_i \in \RR^{n_i \times d}$ and $b_i \in \RR^{n_i}$ represent the feature matrix and labels stored on client $i$ encompassing $n_i$ data points.   The data employed in four experiments, as illustrated in Figure~\ref{ch3:fig:syn-ef21-vc-noncvx}, was generated in such a manner that the smoothness constant $L$ remained fixed, while $L_i$ varied so that the difference between two crucial to analysis terms $\LQM$ and $\LAM$ changed from a relatively large value to negligible. As evident from Figure~\ref{ch3:fig:syn-ef21-vc-noncvx}, the performance of \algname{EF21-W} consistently matches or surpasses that of the original \algname{EF21}, particularly in scenarios characterized by significant variations in the smoothness constants. For additional details and supplementary experiments, we refer the reader to Sections~\ref{ch3:app:exp-additional-details-main-part} and~\ref{ch3:app:exp-additional-experiments}.


\section{Conclusions}

In this work, we present significant improvements to the \algname{EF21} algorithm and its extensions, which are distributed optimization algorithms that utilize contractive compressors, achieving state-of-the-art theoretical communication complexity guarantees. We demonstrate both theoretically and empirically that our improvements are fundamental. These enhancements apply not only to \algname{EF21}, but also to its variants addressing stochastic gradients (\algname{EF21-W-SGD}, Section~\ref{ch3:sec:EF21-W-SGD}), partial client participation (\algname{EF21-W-PP}, Section~\ref{ch3:sec:EF21-W-PP}), and \algname{EF21} in the {\em rare features} regime (Section~\ref{ch3:sec:RF}). We share our discovery process with the reader, hoping it will be helpful to researchers encountering challenges at any stage of their research journey. We believe that the empirical success of \algname{EF21} stems from a complex underlying factor and that further discoveries in this area are needed.

\clearpage
\appendix

\part*{Appendices to Chapter \ref{chapter3}}
\label{ch3:app:toc_1}
\newpage

\phantomsection
\addcontentsline{toc}{chapter}{Appendices to Chapter 3}

\addtocounter{adjsection}{1}
\section{Basic Results and Lemmas}
In this section, we offer a few results that serve as essential prerequisites for establishing the main findings.

\subsection{Optimal client cloning frequencies}

\begin{lemma}[Optimal weights] \label{ch3:lem:ow}
	Let $a_i > 0$ for $i \in [n]$. Then
	\begin{equation}\label{ch3:eq:0998fddff-89y9fd}
		\min\limits_{ \substack{w_1 > 0, \dots, w_n>0 \\ \sumin w_i = 1}} \sumin \dfrac{a_i^2}{w_i} =\left(\sumin a_i\right)^2,
	\end{equation}
	which is achieved when $w^\ast_i = \dfrac{a_i}{\sum_j a_j}$. This means that
	\begin{equation}\label{ch3:eq:0998fddff}
		\min\limits_{ \substack{w_1 > 0, \dots, w_n>0 \\ \sumin w_i = 1}} \dfrac{1}{n}\sqrt{\sumin \dfrac{a_i^2}{w_i}} =\dfrac{1}{n}\sumin a_i.
	\end{equation}
\end{lemma}

We now show that the cloning frequencies given by $N^\star_i =  \left \lceil  \dfrac{L_i}{\LAM} \right \rceil$ form a $\sqrt{2}$-approximation for the optimization problem of finding the optimal integer client frequencies. 

\begin{lemma}[$\sqrt{2}$-approximation]\label{ch3:lem:sandwitch} If we let $N^\star_i =  \left \lceil  \dfrac{L_i}{\LAM} \right \rceil$ for all $i\in [n]$, then 
	\[ \LAM \leq  \min_{N_1 \in \mathbb{N},\dots, N_n\in \mathbb{N}} M(N_1,\dots,N_n) \leq   M(N^\star_1,\dots,N^\star_n) \leq \sqrt{2} \LAM.\]
\end{lemma}
\begin{proof}  Recall that
	\[M(N_1,\dots,N_n) \eqdef  \dfrac{1}{n}\sqrt{\sumin \dfrac{L_i^2}{N_i/N}} .\]
	The first inequality in the claim follows by relaxing the integrality constraints, which gives us the bound
	\[\min\limits_{ \substack{w_1 > 0, \dots, w_n>0 \\ \sumin w_i = 1}} \dfrac{1}{n}\sqrt{\sumin \dfrac{L_i^2}{w_i}}   \leq  \min_{N_1 \in \mathbb{N},\dots, N_n\in \mathbb{N}} \dfrac{1}{n}\sqrt{\sumin \dfrac{L_i^2}{N_i/N}}  , \] 
	and subsequently applying Lemma~\ref{ch3:lem:ow}.
	
	Next, we argue that the quantity $N^\star \eqdef \sum_i N^\star_i$ is at most $2n$. Indeed,
	\begin{equation}\label{ch3:eq:98y08fd=-=f-d} N^\star = \sum_{i=1}^n N^\star_i = \sum_{i=1}^n \left \lceil  \dfrac{L_i}{\LAM} \right \rceil \leq \sum_{i=1}^n \left( \dfrac{L_i}{\LAM} + 1\right) = 2n.\end{equation}
	We will now use this to bound $M(N^\star_1,\dots,N^\star_n)$ from above:
	\begin{eqnarray*}
		M(N^\star_1,\dots,N^\star_n) = \dfrac{1}{n}\sqrt{\sumin \dfrac{L_i^2}{N^\star_i/N^\star}} 
		\overset{\eqref{ch3:eq:98y08fd=-=f-d}}{\le}  \dfrac{\sqrt{2}}{\sqrt{n}}\sqrt{\sumin \dfrac{L_i^2}{N^\star_i}}  =  \dfrac{\sqrt{2}}{\sqrt{n}}\sqrt{\sumin \dfrac{\dfrac{L_i}{\LAM} }{N^\star_i}L_i \LAM} .
	\end{eqnarray*}
	Since $ \dfrac{\dfrac{L_i}{\LAM} }{N^\star_i} \leq 1$ for all $i\in [n]$, as stated in Lemma~\ref{ch3:lem:sandwitch}, the proof proceeds as follows: 
	\begin{eqnarray*}
		M(N^\star_1,\dots,N^\star_n) &\leq& \dfrac{\sqrt{2}}{\sqrt{n}}\sqrt{\sumin L_i \LAM}  =  \dfrac{\sqrt{2}}{\sqrt{n}} \sqrt{\LAM}\sqrt{\sumin L_i } \\
		& = & \sqrt{2} \sqrt{\LAM}\sqrt{\dfrac{1}{n}\sumin L_i } = \sqrt{2} \LAM.
	\end{eqnarray*}
\end{proof}

\subsection{Descent lemma}

\begin{lemma}[\citet{li2021page}]\label{ch3:lm:descent_lemma}
	Let Assumption~\ref{ch3:as:smooth} hold and $x^{t+1} = x^t - \gamma g^t$, where $g^t \in \RR^d$ is any vector, and $\gamma > 0$ is any scalar. Then, we have
	\begin{eqnarray}
		\label{ch3:eq:descent_lemma}
		f(x^{t+1}) &\leq& f(x^t) - \dfrac{\gamma}{2} \|\nabla f(x^t)\|^2 \notag \\
		&-& \left(\dfrac{1}{2\gamma} - \dfrac{L}{2}\right) \|x^{t+1} - x^t\|^2 + \dfrac{\gamma}{2} \|g^t - \nabla f(x^t)\|^2.
	\end{eqnarray}
\end{lemma}

\subsection{Young's inequality}

\begin{lemma}[Young's inequality]
	For any $a, b \in \RR^d$ and any positive scalar $s > 0$ it holds that
	\begin{equation}\label{ch3:eq:young}
		\|a + b\|^2 \leq (1 + s) \|a\|^2 + (1 + s^{-1})\|b\|^2.
	\end{equation}
\end{lemma}

\subsection{$2$-Suboptimal but simple step size rule}

\begin{lemma}[Lemma 5,~\citet{EF21}]\label{ch3:lm:stepsize_bound}
	Let $a, b > 0$. If $$0 < \gamma \leq \dfrac{1}{\sqrt{a} + b},$$ then $$a \gamma^2 + b \gamma \leq 1.$$
	Moreover, the bound is tight up to the factor of 2 since $$\dfrac{1}{\sqrt{a} + b} \leq \min\left\{\dfrac{1}{\sqrt{a}}, \dfrac{1}{b}\right\} \leq \dfrac{2}{\sqrt{a} + b}.$$
\end{lemma}

\subsection{Optimal coefficient in Young's inequality}

\begin{lemma}[Lemma 3, \citet{EF21}]\label{ch3:lm:best_theta_beta_choice}
	Let $0 < \alpha \leq 1$ and for $s> 0$, let $\theta(\alpha,s) \eqdef 1- (1- \alpha )(1+s)$ and $\beta(\alpha,s)\eqdef (1- \alpha ) \left(1+ s^{-1} \right)$. Then, the solution of the optimization problem
	\begin{equation}
		\min\limits_{s} \left\{\dfrac{\beta(\alpha,s)}{\theta(\alpha,s)} : 0 < s < \dfrac{\alpha}{1 - \alpha} \right\}
	\end{equation}
	is given by $$s^\ast = \dfrac{1}{\sqrt{1 - \alpha}} - 1.$$ 
	Furthermore, $$\theta(\alpha,s^\ast) = 1 - \sqrt{1-\alpha},$$
	and 
	$$\beta(\alpha,s^\ast) = \dfrac{1-\alpha}{1 - \sqrt{1 - \alpha}}.$$
\end{lemma}

\clearpage
\addtocounter{adjsection}{1}
\section{Cloning Reformulation for Polyak-\L ojasiewicz Functions}

For completeness, we also provide a series of convergence results under Polyak-\L ojasiewicz condition. We commence our exposition with the subsequent definition.

\begin{assumption}[Polyak-\L ojasiewicz]\label{ch3:as:PL}
	There exists a positive scalar $\mu > 0$ such that for all points $x \in \RR^d$, the following inequality is satisfied: 
	\begin{equation}\label{ch3:eq:PL}
		f(x) - f(x^\ast) \leq \dfrac{1}{2\mu}\|\nabla f(x)\|^2,
	\end{equation} where $x^\ast \eqdef \argmin f(x)$.
\end{assumption}

\begin{theorem}
	Let Assumptions~\ref{ch3:as:smooth},~\ref{ch3:as:L_i},~and~\ref{ch3:as:PL} hold. Assume that $\cC_i^t \in \mathbb{C}(\alpha)$ for all $i \in [n]$ and $t \geq 0$. Consider Algorithm~\ref{ch3:alg:EF21} (\algname{EF21}) applied to the ``cloning'' reformulation~\eqref{ch3:eq:cloning} of the distributed optimization Problem~\eqref{ch3:eq:main_problem}, where $N_i^\ast = \lceil \dfrac{L_i}{\LAM} \rceil$ for all $i \in [n]$. Let the step size be set as
	$$
	0 \leq \gamma \leq \min\left\{ \left(L + \sqrt{2} \LAM \sqrt{\dfrac{2\beta}{\theta}} \right)^{-1}, \dfrac{\theta}{2\mu}\right\},
	$$
	where $\theta = 1 - \sqrt{1 - \alpha}$ and $\beta = \dfrac{1 - \alpha}{1 - \sqrt{1 - \alpha}}$.
	Let $$\Psi^t \eqdef f(x^t) - f(x^\ast) + \dfrac{\gamma}{\theta} G^t .$$ Then, for any $T \geq 0$, we have
	$$
	\ExpBr{\Psi^T} \leq (1 - \gamma \mu)^T \Psi^0.
	$$
\end{theorem}
\begin{proof}
	This theorem is a corollary of Theorem 2 in~\citet{EF21} and Lemma~\ref{ch3:lem:sandwitch}.
\end{proof}

\clearpage
\addtocounter{adjsection}{1}
\section{Proof of Theorem~\ref{ch3:thm:EF21-W} (Theory for EF21-W)}

In this section, we present a proof for Theorem~\ref{ch3:thm:EF21-W}. To start this proof, we establish a corresponding contraction Lemma. We define the following quantities:
\begin{equation}\label{ch3:eq:weighted_def_grad_distortion}
	G_i^{t} \eqdef \sqnorm{ g_i^t - \dfrac{\nabla f_i(x^{t})}{n w_i}}; \qquad G^t \eqdef \sumin w_i G_i^t,
\end{equation}
where the weights $w_i$ are defined as specified in Algorithm~\ref{ch3:alg:EF21-W}, that is, 
\begin{equation}\label{ch3:eq:weight_definition}
	w_i = \dfrac{L_i}{\sum_{j=1}^n L_j}.
\end{equation}

\subsection{A lemma}

With these definitions in place, we are now prepared to proceed to the Lemma.

\begin{lemma} Let $\cC_i^t\in \mathbb{C}(\alpha)$ for all $i\in [n]$ and $t\geq 0$. Let $W^t \eqdef \{g_1^t, g_2^t, \dots, g_n^t, x^t, x^{t+1}\}$. Then, for iterates of Algorithm~\ref{ch3:alg:EF21-W} we have
	\begin{equation}\label{ch3:eq:weighted_ind_grad_dist_evolution}
		\ExpBr{ G_i^{t+1} \;|\; W^t} \leq (1-\theta(\alpha,s))   G_i^t + \beta(\alpha,s) \dfrac{1}{n^2 w_i^2} \sqnorm{\nabla f_i(x^{t+1}) - \nabla f_i(x^t)},
	\end{equation}
	and 
	\begin{equation}\label{ch3:eq:weighted_full_grad_dist_evolution}
		\ExpBr{G^{t+1} } {\leq} \rb{1 - \theta(\alpha,s)} \ExpBr{G^t} + \beta(\alpha,s) \LAMsq \ExpBr{ \sqnorm{x^{t+1} - x^t}},
	\end{equation}
	where $s > 0$ is an arbitrary positive scalar, and
	\begin{equation}\label{ch3:eq:theta-beta-def} \theta(\alpha,s) \eqdef 1- (1- \alpha )(1+s), \qquad \text{and} \qquad \beta(\alpha,s)\eqdef (1- \alpha ) \left(1+ s^{-1} \right). \end{equation}
\end{lemma}

\begin{proof}
	
	The proof is straightforward and bears resemblance to a similar proof found in a prior work \citep{EF21}.
	\begin{eqnarray*}
		\ExpBr{ G_i^{t+1} \;|\; W^t} & \overset{\eqref{ch3:eq:weighted_def_grad_distortion}}{=} & \ExpBr{  \sqnorm{g_i^{t+1} 
				- \dfrac{\nabla f_i(x^{t+1})}{n w_i}}  \;|\; W^t}	 \\
		&= & \ExpBr{  \sqnorm{g_i^t + \cC_i^t \left( \dfrac{\nabla f_i(x^{t+1})}{n w_i} - g_i^t \right) 
				- \dfrac{\nabla f_i(x^{t+1})}{n w_i}}  \;|\; W^t}	 \\
		&\overset{\eqref{ch3:eq:compressor_contraction}}{\leq} &  (1-\alpha) 
		\sqnorm{\dfrac{\nabla f_i(x^{t+1})}{n w_i} - g_i^t} \\
		& = & (1-\alpha) 
		\sqnorm{\dfrac{\nabla f_i(x^{t+1})}{n w_i} - g_i^t + \dfrac{\nabla f_i(x^{t})}{n w_i} - \dfrac{\nabla f_i(x^{t})}{n w_i}} \\
		&\overset{\eqref{ch3:eq:young}}{\leq}& (1-\alpha) (1+ s) \sqnorm{ \dfrac{\nabla f_i(x^{t})}{n w_i} - g_i^t}\\
		&& \qquad  + (1-\alpha)  \left(1+s^{-1}\right) \dfrac{1}{n^2 w_i^2} 
		\sqnorm{\nabla f_i(x^{t+1}) - \nabla f_i(x^t)}, 
	\end{eqnarray*}
	with the final inequality holding for any positive scalar $s > 0$. Consequently, we have successfully established the first part of the Lemma.
	
	By employing~\eqref{ch3:eq:weighted_def_grad_distortion} and the preceding inequality, we can derive the subsequent bound for the conditional expectation of $G^{t+1}$:
	\begin{eqnarray}
		\ExpBr{G^{t+1} \mid W^t } &\overset{\eqref{ch3:eq:weighted_def_grad_distortion}}{=} & \ExpBr{\sumin w_i G_i^{t+1} \mid W^t} \notag \\
		&\overset{\eqref{ch3:eq:weighted_def_grad_distortion}}{=} & \sumin w_i \ExpBr{\sqnorm{ g_i^{t+1} - \dfrac{\nabla f_i(x^{t+1})}{n w_i}} \mid W^t}  \notag \\
		&\overset{\eqref{ch3:eq:weighted_ind_grad_dist_evolution}}{\leq} & \rb{1 - \theta(\alpha,s)} \sumin w_i 
		\sqnorm{g_i^t - \dfrac{\nabla f_i(x^{t})}{n w_i} } \notag \\
		&& \quad+ \beta(\alpha,s) \sumin \dfrac{w_i}{w_i^2 n^2} \sqnorm {\nabla f_i(x^{t+1}) - \nabla f_i(x^t)}  .\label{ch3:eq:weighted_aux_1}
	\end{eqnarray}
	Applying Assumption~\ref{ch3:as:L_i} and~\eqref{ch3:eq:weight_definition}, we further proceed to:
	\begin{eqnarray}
		\ExpBr{G^{t+1} \mid W^t } &\overset{\eqref{ch3:eq:weighted_aux_1}}{\leq} &  \rb{1 - \theta(\alpha,s)} \sumin w_i 
		\sqnorm{g_i^t - \dfrac{\nabla f_i(x^{t})}{n w_i} } \notag \\
		& & \qquad	+ \beta(\alpha,s) \sumin \dfrac{w_i}{w_i^2 n^2} \sqnorm {\nabla f_i(x^{t+1}) - \nabla f_i(x^t)}  \notag \\
		&\overset{\eqref{ch3:eq:weighted_def_grad_distortion}}{=}  & \rb{1 - \theta(\alpha,s)}G^t+ \beta(\alpha,s) \sumin \dfrac{1}{w_i n^2} \sqnorm {\nabla f_i(x^{t+1}) - \nabla f_i(x^t)}   \notag \\
		&\overset{\eqref{ch3:eq:L_i}}{\leq} &  \rb{1 - \theta(\alpha,s)}G^t
		+ \beta(\alpha,s) \rb{\sumin \dfrac{L_i^2}{w_i n^2} }\sqnorm{x^{t+1} - x^t}   \notag \\
		&\overset{\eqref{ch3:eq:weight_definition}}{=} & \rb{1 - \theta(\alpha,s)}G^t
		+ \beta(\alpha,s) \rb{\sumin \dfrac{L_i^2}{\dfrac{L_i}{\left(\sumjn L_j\right)} n^2}}\sqnorm{x^{t+1} - x^t}   \notag \\
		&=& \rb{1 - \theta(\alpha,s)}G^t
		+ \beta(\alpha,s) \rb{\sumin \dfrac{L_i \sumjn L_j}{n^2}}\sqnorm{x^{t+1} - x^t}  \notag  \\
		&= & \rb{1 - \theta(\alpha,s)}G^t
		+ \beta(\alpha,s) \LAMsq \sqnorm{x^{t+1} - x^t} .\label{ch3:eq:weighted_aux_2}
	\end{eqnarray}
	
	Using the tower property,  we get 
	\begin{eqnarray*}
		\ExpBr{G^{t+1}} = \ExpBr{\ExpBr{G^{t+1} \mid W^t}} &\overset{\eqref{ch3:eq:weighted_aux_2}}{\leq}& \rb{1 - \theta(\alpha,s)}\ExpBr{G^t} \notag \\ 
		& & \qquad	+ \beta(\alpha,s) \LAMsq \ExpBr{ \sqnorm{x^{t+1} - x^t}},
	\end{eqnarray*}
	and this finalizes the proof.
\end{proof}

\subsection{Main result}

We are now prepared to establish the proof for Theorem~\ref{ch3:thm:EF21-W}.
\begin{proof} Note that, according to~\eqref{ch3:eq:AlgStep1X}, the gradient estimate for Algorithm~\ref{ch3:alg:EF21-W} gets the following form:
	\begin{equation}\label{ch3:eq:weighted_grad_estimate_def}
		g^t=\sum\limits_{i=1}^{n} w_i g_i^t.
	\end{equation}
	Using Lemma~\ref{ch3:lm:descent_lemma} and Jensen's inequality applied to the function $x\mapsto \sqnorm{x}$ (since $\sumin w_i = 1$), we obtain the following bound:
	\begin{eqnarray}
		f(x^{t+1}) &\overset{\eqref{ch3:eq:descent_lemma}}{\leq} & 
		f(x^{t})-\dfrac{\gamma}{2}\sqnorm{\nabla f(x^{t})}-\left(\dfrac{1}{2 \gamma}-\dfrac{L}{2}\right)\sqnorm{x^{t+1}-x^{t}} \notag \\
		& & \qquad +\dfrac{\gamma}{2}\sqnorm{
			g^t - \sumin \dfrac{\nabla f_i(x^{t})}{n} } \notag \\
		& \overset{\eqref{ch3:eq:weighted_grad_estimate_def}}{=} & 
		f(x^{t})-\dfrac{\gamma}{2}\sqnorm{\nabla f(x^{t})}-\left(\dfrac{1}{2 \gamma}-\dfrac{L} 
		{2}\right)\sqnorm{x^{t+1}-x^{t}} \notag \\
		& & \qquad +\dfrac{\gamma}{2}\sqnorm{
			\sumin w_i \left(g_i^t-  \dfrac{\nabla f_i(x^{t})}{n w_i} \right) }   \notag \\
		& \leq & 
		f(x^{t})-\dfrac{\gamma}{2}\sqnorm{\nabla f(x^{t})}-\left(\dfrac{1}{2 \gamma}-\dfrac{L}       
		{2}\right)\sqnorm{x^{t+1}-x^{t}} \notag \\
		& & \qquad +\dfrac{\gamma}{2}
		\sumin w_i \sqnorm{ g_i^t-  \dfrac{\nabla f_i(x^{t})}{n w_i}  }  \notag \\
		& \overset{\eqref{ch3:eq:weighted_def_grad_distortion}}{=}  & f(x^{t})-\dfrac{\gamma}{2}\sqnorm{\nabla f(x^{t})}-\left(\dfrac{1}{2 \gamma}-\dfrac{L}       
		{2}\right)\sqnorm{x^{t+1}-x^{t}} \notag \\
		&& \qquad +\dfrac{\gamma}{2} G^t.\label{ch3:eq:weighted_aux_3}
	\end{eqnarray}
	
	Subtracting $f^\ast$ from both sides and taking expectation, we get
	\begin{align}
		\ExpBr{f(x^{t+1})-f^\ast} & \leq  \ExpBr{f(x^{t})-f^\ast}
		-\dfrac{\gamma}{2} \ExpBr{\sqnorm{\nabla f(x^{t})}}  \notag \\
		& \qquad -\left(\dfrac{1}{2 \gamma}
		-\dfrac{L}{2}\right) \ExpBr{\sqnorm{x^{t+1}-x^{t}}}+ \dfrac{\gamma}{2}\ExpBr{G^t}.
		\label{ch3:eq:weighted_function_descent}
	\end{align}
	
	Let
	\begin{align*}
		\delta^{t} &\eqdef \mathbb{E}\left[f(x^{t}) - f^\ast\right], \\
		s^{t} &\eqdef \mathbb{E}\left[G^t\right], \\
		r^{t} & \eqdef\ExpBr{\sqnorm{x^{t+1}-x^{t}}}.
	\end{align*}
	The last Equation~\eqref{ch3:eq:weighted_function_descent} can be re-expressed using the new notation. 

	Subsequently, by adding \eqref{ch3:eq:weighted_full_grad_dist_evolution} with a $\dfrac{\gamma}{2 \theta(\alpha,s)}$ multiplier, we obtain 
	\begin{eqnarray*}
		\delta^{t+1}+\dfrac{\gamma}{2 \theta(\alpha,s)} s^{t+1} &\overset{\eqref{ch3:eq:weighted_function_descent}}{\leq} & \delta^{t}-\dfrac{\gamma}{2}\sqnorm{\nabla f(x^{t})}-\left(\dfrac{1}{2 \gamma}-\dfrac{L}{2}\right) r^{t}+\dfrac{\gamma}{2} s^{t} \notag \\
		& & \qquad + \dfrac{\gamma}{2 \theta(\alpha,s)} s^{t+1}\\
		&\overset{\eqref{ch3:eq:weighted_full_grad_dist_evolution}}{\leq}& \delta^{t}-\dfrac{\gamma}{2}\sqnorm{\nabla f(x^{t})}-          \left(\dfrac{1}{2 \gamma}-\dfrac{L}{2}\right) r^{t}+\dfrac{\gamma}{2} s^{t} \notag \\ 
		& & \qquad	+\dfrac{\gamma}{2 \theta(\alpha,s)}\left(\beta(\alpha,s) \LAMsq r^t + (1 - \theta(\alpha,s)) s^{t}\right) \\
		&=&\delta^{t}+\dfrac{\gamma}{2\theta(\alpha,s)} s^{t}-\dfrac{\gamma}{2}\sqnorm{\nabla f(x^{t})} \notag \\
		& & \qquad -\left(\dfrac{1}{2\gamma} -\dfrac{L}{2} - \dfrac{\gamma}{2\theta(\alpha,s)}\beta(\alpha,s) \LAMsq \right) r^{t} \\
		& \leq& \delta^{t}+\dfrac{\gamma}{2\theta(\alpha,s)} s^{t} -\dfrac{\gamma}{2}\sqnorm{\nabla f(x^{t})}.
	\end{eqnarray*}
	The last inequality is a result of the bound $$\gamma^2\dfrac{\beta(\alpha,s) \LAMsq}{\theta(\alpha,s)} + L\gamma \leq 1,$$ which is satisfied for the step size $$\gamma \leq \dfrac{1}{L + \LAM \xi(\alpha,s)},$$ 
	where $ \xi(\alpha,s)\eqdef \sqrt{\dfrac{\beta(\alpha,s)}{\theta(\alpha,s)}}$.
	Maximizing the step size bound over the choice of $s$ using Lemma~\ref{ch3:lm:best_theta_beta_choice}, we obtain the final step size.
	By summing up inequalities for $t =0, \ldots, T-1,$ we get
	$$
	0 \leq \delta^{T}+\dfrac{\gamma}{2 \theta} s^{T} \leq \delta^{0}+\dfrac{\gamma}{2 \theta} s^{0}-\dfrac{\gamma}       {2} \sum_{t=0}^{T-1} \ExpBr{\sqnorm{\nabla f(x^{t})}}.
	$$
	Multiplying both sides by $\dfrac{2}{\gamma T}$, after rearranging we get
	$$
	\sum_{t=0}^{T-1} \dfrac{1}{T} \ExpBr{\sqnorm{\nabla f (x^{t})}} \leq \dfrac{2 \delta^{0}}{\gamma T} + \dfrac{s^0}      {\theta T}.
	$$
	It remains to notice that the left-hand side can be interpreted as $$\ExpBr{ \sqnorm{\nabla f(\hat{x}^{T})} },$$
	where $\hat{x}^{T}$ is chosen from $\{ x^{0}, x^{1}, \ldots, x^{T-1} \}$ uniformly at random.
\end{proof}

\subsection{Main result for Polyak-\L ojasiewicz functions}

The main result is presented next.

\begin{theorem}
	Let Assumptions~\ref{ch3:as:smooth},~\ref{ch3:as:L_i}, and~\ref{ch3:as:PL} hold. Assume that $\cC_i^t \in \mathbb{C}(\alpha)$ for all $i\in [n]$ and $t\geq 0$. Let the step size in Algorithm~\ref{ch3:alg:EF21-W} be set as 
	$$
	0 < \gamma \leq \min\left\{ \dfrac{1}{L + \sqrt{2} \LAM  \xi(\alpha)}, \dfrac{\theta(\alpha)}{2\mu}\right\}.
	$$
	Let $$\Psi^t \eqdef f(x^t) - f(x^\ast) + \dfrac{\gamma}{\theta} G^t.$$ Then, for any $T>0$ the following inequality holds:
	\begin{equation}
		\ExpBr{\Psi^T} \leq (1 - \gamma \mu)^T \Psi^0.
	\end{equation}
\end{theorem}

\begin{proof}
	We proceed as in the previous proof, starting from the descent Lemma~\ref{ch3:lm:descent_lemma} with the same vector but using the PL inequality~\eqref{ch3:eq:PL} and            subtracting $f(x^\star)$ from both sides:
	\begin{eqnarray}
		\ExpBr{f(x^{t+1})- f(x^\star)}  &\overset{\eqref{ch3:eq:descent_lemma}}{\leq} & \ExpBr{ f(x^{t})- f(x^\star) }-\dfrac{\gamma}{2}\sqnorm{\nabla f(x^{t})} \notag \\
		&& \qquad -\left(\dfrac{1}{2 \gamma}-\dfrac{L}{2}\right)\sqnorm{x^{t+1}-x^{t}}+\dfrac{\gamma}{2} G^t \notag \\ 
		& \overset{\eqref{ch3:eq:PL}}{\leq}  & (1-\gamma \mu) \ExpBr{ f(x^{t})- f(x^\star) } \notag \\
		&& \qquad - \left(\dfrac{1}{2 \gamma}-\dfrac{L}{2}\right)\sqnorm{x^{t+1}-x^{t}}+\dfrac{\gamma}{2} G^t. \label{ch3:eq:weighted_aux_4}
	\end{eqnarray}
	
	Let
	\begin{align*}
		\delta^{t} &\eqdef \ExpBr{f(x^{t})-f(x^\star)}, \\
		s^{t} &\eqdef \ExpBr{G^t }, \\
		r^{t} &\eqdef \ExpBr{\sqnorm{x^{t+1}-x^{t}}}.
	\end{align*}
	Thus, after adding $\dfrac{\gamma}{\theta} s^{t+1}$ to both sides of inequality~\eqref{ch3:eq:weighted_aux_4}, it can be rewritten as follows:
	\begin{eqnarray*}
		\delta^{t+1}+\dfrac{\gamma}{ \theta} s^{t+1} &\overset{\eqref{ch3:eq:weighted_aux_4}}{\leq}& (1-\gamma \mu)\delta^{t} -\left(\dfrac{1}{2 \gamma} - \dfrac{L}{2}\right) r^{t} + \dfrac{\gamma}{2} s^{t} \notag \\
		&& \qquad + \dfrac{\gamma}{ \theta} s^{t+1} \\
		& \overset{\eqref{ch3:eq:weighted_full_grad_dist_evolution}}{\leq}  & (1-\gamma \mu)\delta^{t} -\left(\dfrac{1}{2 \gamma} - \dfrac{L}{2}\right) r^{t} + \dfrac{\gamma}{2} s^{t} \notag \\
		&& \qquad + \dfrac{\gamma}{ \theta}\left( (1-\theta) s^t + \beta \left(\avein L_i \right)^2 r^t       \right) \\
		&=&  (1-\gamma \mu) \delta^{t}  + \dfrac{\gamma}{\theta} \left(1-\dfrac{\theta}{2}\right) s^t  -\left(\dfrac{1}{2 \gamma} - \dfrac{L}{2} - \dfrac{\beta \LAMsq \gamma}{\theta}\right) r^{t},
	\end{eqnarray*}
	where $\theta$ and $\beta$ are set as in Lemma~\ref{ch3:lm:best_theta_beta_choice}.
	Note that our extra assumption on the step size implies that 	$ 1 - \dfrac{\theta}{2} \leq 1 -\gamma \mu$ and $$\dfrac{1}{2 \gamma} - \dfrac{L}{2} - \dfrac{\beta \LAMsq \gamma}{\theta} \geq 0.$$ The last inequality follows from the bound $$\gamma^2\dfrac{2\beta \LAMsq}{\theta} + \gamma L \leq 1.$$
	Thus,
	\begin{eqnarray*}
		\delta^{t+1}+\dfrac{\gamma}{ \theta} s^{t+1} \leq (1-\gamma \mu) \left(\delta^{t}+\dfrac{\gamma}{ \theta} s^{t} \right).
	\end{eqnarray*}
	It remains to unroll the recurrence.
\end{proof}

\clearpage
\addtocounter{adjsection}{1}
\section{Proof of Theorem~\ref{ch3:thm:ef21_new_result} (Improved Theory for EF21)}
We commence by redefining gradient distortion as follows:
\begin{equation}\label{ch3:eq:new_def_distortion}
	G^{t} \eqdef \dfrac{1}{n^2} \sumin \dfrac{1}{w_i} \|\nabla f_i(x^t) - g^t_i\|^2.
\end{equation}

We recall that the gradient update step for standard \algname{EF21} (Algorithm~\ref{ch3:alg:EF21}) takes the following form:
\begin{align}
	&g_i^{t+1} = g_i^t + \cC_i^t(\nabla f_i(x^{t+1}) - g_i^t), \label{ch3:eq:standard_ind_grad_upd} \\
	&g^{t+1} = \avein g_i^{t+1} \label{ch3:eq:standard_average_grad}.
\end{align}

\subsection{Two lemmas}

Once more, we start our proof with the contraction Lemma.
\begin{lemma}
	Let $\cC_i^t \in \mathbb{C}(\alpha)$ for all $i \in [n]$ and $t\geq 0$.\\
	Define $$W^t \eqdef \{g_1^t, g_2^t, \dots, g_n^t, x^t, x^{t+1}\}.$$\\
	Let Assumption~\ref{ch3:as:L_i} hold. Then
	\begin{equation}\label{ch3:eq:grad_distortion_evolution}
		\ExpBr{G^{t+1}\; | \;  W^t} \leq (1 - \theta(\alpha,s)) G^t + \beta(\alpha,s) \LAMsq \|x^{t+1} - x^t \|^2,
	\end{equation}
	where  for any $s > 0$ 
	\begin{equation*}
	\begin{aligned}
		\theta(\alpha, s) &\eqdef 1 - (1 - \alpha)(1 + s), \\
		\beta(\alpha, s) &\eqdef (1 - \alpha)(1 + s^{-1}).
	\end{aligned}
	\end{equation*}
\end{lemma}

\begin{proof}
	The proof of this Lemma starts as the similar Lemma in the standard analysis of \algname{EF21}:
	\begin{eqnarray}
		\ExpBr{G^{t+1}\; | \;  W^t} & \overset{\eqref{ch3:eq:new_def_distortion}}{=}  & \dfrac{1}{n^2}\sumin \dfrac{1}{w_i}\ExpBr{ \|\nabla f_i(x^{t+1}) - g^{t+1}_i\|^2\; | \;  W^t}  \notag \\
		& \overset{\eqref{ch3:eq:standard_ind_grad_upd}}{=} & \dfrac{1}{n^2} \sumin \dfrac{1}{w_i} \ExpBr{\|g_i^{t} + \cC_i^t(\nabla f_i(x^{t+1}) - g_i^t) -\nabla f_i(x^{t+1}) \|^2 \;|\; W^t}  \notag \\
		& \overset{\eqref{ch3:eq:compressor_contraction}}{\leq} & \dfrac{1}{n^2} \sumin \dfrac{1 - \alpha}{w_i}  \|\nabla f_i(x^{t+1}) - g_i^t) \|^2  \notag \\
		& = & \dfrac{1}{n^2} \sumin \dfrac{1 - \alpha}{w_i}  \|\nabla f_i(x^{t+1}) - \nabla f_i(x^t) + \nabla f_i(x^t) - g_i^t) \|^2 \notag \\
		& \overset{\eqref{ch3:eq:young}}{\leq} & \dfrac{1}{n^2} \sumin \dfrac{1 - \alpha}{w_i}  \Big((1 + s^{-1}) \|\nabla f_i(x^{t+1}) - \nabla f_i(x^t)) \|^2 \notag \\
		&&\qquad \qquad \qquad + (1 + s) \|g_i^t - \nabla f_i(x^t) \|^2 \Big) \notag \\
		&& 			 \label{ch3:eq:aux1}
	\end{eqnarray}
	For all  $s > 0$ we proceed with the proof as follows:
	\begin{eqnarray}
		\ExpBr{G^{t+1} \;|\; W^t}  &\overset{\eqref{ch3:eq:aux1}}{\leq} & \dfrac{1}{n^2} \sumin \dfrac{1 - \alpha}{w_i}  \Big((1 + s^{-1}) \|\nabla f_i(x^{t+1}) - \nabla f_i(x^t)) \|^2 \notag \\
		&& \qquad \qquad \qquad + (1 + s) \|g_i^t - \nabla f_i(x^t) \|^2 \Big) \notag \\
		& =  & (1 - \theta(\alpha,s)) \dfrac{1}{n^2} \sumin \dfrac{1}{w_i} \|g_i^t - \nabla f_i(x^t)\|^2 \notag \\
		&& + \dfrac{\beta(\alpha,s)}{n^2} \sumin \dfrac{1}{w_i} \|\nabla f_i(x^{t+1}) - \nabla f_i(x^t)) \|^2  \notag \\
		& \overset{\eqref{ch3:eq:new_def_distortion}}{=} & (1 - \theta(\alpha,s)) G^t \notag \\
		&& + \dfrac{\beta(\alpha,s)}{n^2} \sumin \dfrac{1}{w_i } \|\nabla f_i(x^{t+1}) - \nabla f_i(x^t) \|^2  \notag \\
		& \overset{\eqref{ch3:eq:L_i}}{\leq} & (1 - \theta(\alpha,s)) G^t \notag \\
		&& + \dfrac{\beta(\alpha,s)}{n^2} \sumin \dfrac{L_i^2}{w_i} \|x^{t+1} - x^t \|^2.\label{ch3:eq:aux2}
	\end{eqnarray}
	Note that this is the exact place where the current analysis differs from the standard one. It fully coincides with it when $w_i = \dfrac{1}{n}$, i.e., when we assign the same weight for each individual gradient distortion $\|g_i^t - \nabla f_i(x^t) \|^2$. However, applying weights according to ``importance'' of each function, we proceed as follows:
	\begin{eqnarray*}
		\ExpBr{G^{t+1}\; | \; W^t} &\overset{\eqref{ch3:eq:aux2}}{\leq}  & (1 - \theta(\alpha,s)) G^t + \dfrac{\beta(\alpha,s)}{n^2} \sumin \dfrac{L_i^2}{w_i} \|x^{t+1} - x^t \|^2 \\
		& \overset{\eqref{ch3:eq:weight_definition}}{=}  & (1 - \theta(\alpha,s)) G^t + \dfrac{\beta(\alpha,s)}{n^2} \sumin \dfrac{L_i^2}{L_i} \left(\sum\limits_{i=1}^n L_i\right) \|x^{t+1} - x^t \|^2 \\
		& =  & (1 - \theta(\alpha,s)) G^t + \dfrac{\beta(\alpha,s)}{n^2} \sum_j L_j \left(\sum\limits_{i=1}^n L_i\right) \|x^{t+1} - x^t \|^2 \\
		& = & (1 - \theta(\alpha,s)) G^t + \beta(\alpha,s) \LAMsq \|x^{t+1} - x^t \|^2,
	\end{eqnarray*}
	what finishes the proof.
\end{proof}
To prove the main convergence theorem, we also need the following Lemma.
\begin{lemma}
	For the variable $g^t$ from Algorithm~\ref{ch3:alg:EF21}, the following inequality holds:
	\begin{equation}\label{ch3:eq:distortion_connection}
		\|g^t - \nabla f(x^t) \|^2 \leq G^t.
	\end{equation}
\end{lemma}
\begin{proof}
	The proof is straightforward:
	\begin{eqnarray*}
		\|g^t - \nabla f(x^t) \|^2 &\overset{\eqref{ch3:eq:standard_average_grad}}{=} & \left\| \sumin  \dfrac{1}{n}\left( g_i^t - \nabla f_i(x^t) \right) \right\|^2 	\\	&=& \left\| \sumin w_i \dfrac{1}{w_i n} \left( g_i^t - \nabla f_i(x^t) \right) \right\|^2 \\ 
		&\leq & \sumin w_i \left\| \dfrac{1}{w_i n} \left( g_i^t - \nabla f_i(x^t) \right) \right\|^2  \\
		&=& \sumin \dfrac{1}{w_i n^2} \|g^t - \nabla f_i(x^t)\|^2 \quad \overset{\eqref{ch3:eq:new_def_distortion}}{=} \quad G^t,
	\end{eqnarray*}
	where the only inequality in this series of equations is derived using Jensen's inequality.
\end{proof}

\subsection{Main result}

We are now equipped with all the necessary tools to establish the convergence theorem.
\begin{proof}
	Let us define the Lyapunov function $$\Phi^t \eqdef f(x^t) - f^\ast + \dfrac{\gamma}{2 \theta(\alpha,s)} G^t .$$ Let us also define $W^t \eqdef \{g_1^t, g_2^t, \dots, g_n^t, x^t, x^{t+1}\}$. We start as follows:
	\begin{eqnarray*}
		& \ExpBr{\Phi^{t+1}\; | \;  W^t}  \\
		& =  & \ExpBr{f(x^{t+1}) - f^\ast\; | \;  W^t}  + \dfrac{\gamma}{2\theta(\alpha,s)} \ExpBr{G^{t+1}\; | \;  W^t}\\
		& \overset{\eqref{ch3:eq:descent_lemma}}{\leq}  & f(x^t) - f^\ast - \dfrac{\gamma}{2} \|\nabla f(x^t)\|^2 - \left(\dfrac{1}{2\gamma} - \dfrac{L}{2} \right) \|x^{t+1} - x^t \|^2 + \\
		&& \qquad \dfrac{\gamma}{2}\|g^t - \nabla f(x^t) \|^2 + \dfrac{\gamma}{2\theta(\alpha,s)} \ExpBr{G^{t+1}\; | \;  W^t}\\
		& \overset{\eqref{ch3:eq:distortion_connection}}{\leq}  &  f(x^t) - f^\ast - \dfrac{\gamma}{2} \|\nabla f(x^t)\|^2 - \left(\dfrac{1}{2\gamma} - \dfrac{L}{2} \right) \|x^{t+1} - x^t \|^2 + \dfrac{\gamma}{2} G^t \\
		&& \qquad + \dfrac{\gamma}{2\theta(\alpha,s)} \ExpBr{G^{t+1}\; | \;  W^t}\\
		&\overset{\eqref{ch3:eq:grad_distortion_evolution}}{\leq}  &  f(x^t) - f^\ast - \dfrac{\gamma}{2} \|\nabla f(x^t)\|^2 - \left(\dfrac{1}{2\gamma} - \dfrac{L}{2} \right) \|x^{t+1} - x^t \|^2 + \dfrac{\gamma}{2} G^t\\
		& =   & f(x^t) - f^\ast + \dfrac{\gamma}{2\theta(\alpha,s)} G^t - \dfrac{\gamma}{2} \|\nabla f(x^t)\|^2 \\
		&& \qquad - \underbrace{\left(\dfrac{1}{2\gamma} - \dfrac{L}{2} - \dfrac{\gamma \beta(\alpha,s)}{2\theta(\alpha,s)} \LAMsq \right)}_{\geq 0} \|x^{t+1} - x^t \|^2 \\
		& \leq  & f(x^t) - f^\ast + \dfrac{\gamma}{2\theta(\alpha,s)} G^t - \dfrac{\gamma}{2} \|\nabla f(x^t)\|^2 \\
		& = & \Phi^t  - \dfrac{\gamma}{2} \|\nabla f(x^t)\|^2.
	\end{eqnarray*}
	The inequality in the last but one line is valid if $$\gamma \leq \dfrac{1}{ L + \LAM \sqrt{\dfrac{\beta(\alpha,s)}{\theta(\alpha,s)}}},$$ according to Lemma~\ref{ch3:lm:stepsize_bound}. By optimizing the step size bound through the selection of $s$ in accordance with Lemma~\ref{ch3:lm:best_theta_beta_choice}, we derive the final step size and establish the optimal value for $\theta$ in defining the Lyapunov function. Applying the tower property and unrolling the recurrence, we finish the proof.
\end{proof}

\subsection{Main result for Polyak-\L ojasiewicz functions}

For completeness, we also provide a convergence result under Polyak-\L ojasiewicz condition (Assumption~\ref{ch3:as:PL}). The main result is presented next.

\begin{theorem}
	Let Assumptions~\ref{ch3:as:smooth},~\ref{ch3:as:L_i}, and~\ref{ch3:as:PL} hold. Assume that $\cC_i^t \in \mathbb{C}(\alpha)$ for all $i\in [n]$ and $t\geq 0$. Let the step size in Algorithm~\ref{ch3:alg:EF21-W} be set as 
	$$
	0 < \gamma \leq \min\left\{ \dfrac{1}{L + \sqrt{2} \LAM  \xi(\alpha)}, \dfrac{\theta(\alpha,s)}{2\mu}\right\}.
	$$
	Let $$\Psi^t \eqdef f(x^t) - f(x^\ast) + \dfrac{\gamma}{\theta(\alpha,s)} G^t .$$ Then, for any $T>0$ the following inequality holds:
	\begin{equation}
		\ExpBr{\Psi^T} \leq (1 - \gamma \mu)^T \Psi^0.
	\end{equation}
\end{theorem}

\begin{proof}
	We proceed as in the previous proof, starting from the descent Lemma with the same vector but using the PL inequality~\eqref{ch3:eq:PL} and            subtracting $f(x^\star)$ from both sides:
	\begin{eqnarray}
		\ExpBr{f(x^{t+1})- f(x^\star)}  &\overset{\eqref{ch3:eq:descent_lemma}}{\leq} &  \ExpBr{ f(x^{t})- f(x^\star) }-\dfrac{\gamma}{2}\sqnorm{\nabla f(x^{t})} \notag \\
		&& - \left(\dfrac{1}{2 \gamma}-\dfrac{L}{2}\right)\sqnorm{x^{t+1}-x^{t}}+\dfrac{\gamma}{2} G^t \notag \\ 
		& \overset{\eqref{ch3:eq:PL}}{\leq} & (1-\gamma \mu) \ExpBr{ f(x^{t})- f(x^\star) } \notag \\
		&& - \left(\dfrac{1}{2 \gamma}-\dfrac{L}{2}\right)\sqnorm{x^{t+1}-x^{t}}+\dfrac{\gamma}{2} G^t. \label{ch3:eq:weighted_aux_4}
	\end{eqnarray}
	
	Let
	\begin{align*}
		\delta^{t} &\eqdef \ExpBr{f(x^{t})-f(x^\star)},\\
		s^{t} &\eqdef \ExpBr{G^t}, \\
		r^{t} &\eqdef \ExpBr{\sqnorm{x^{t+1}-x^{t}}}.
	\end{align*}	
	Thus, 
	\begin{eqnarray*}
		\delta^{t+1}+\dfrac{\gamma}{ \theta(\alpha,s)} s^{t+1} &\overset{\eqref{ch3:eq:weighted_aux_4}}{\leq}& (1-\gamma \mu)\delta^{t} -\left(\dfrac{1}{2 \gamma} - \dfrac{L}{2}\right) r^{t} + \dfrac{\gamma}{2} s^{t} + \dfrac{\gamma}{ \theta(\alpha,s)} s^{t+1} \\
		& \overset{\eqref{ch3:eq:grad_distortion_evolution}}{\leq}  & (1-\gamma \mu)\delta^{t} -\left(\dfrac{1}{2 \gamma} - \dfrac{L}{2}\right) r^{t} + \dfrac{\gamma}{2} s^{t}  \\
		&& \qquad + \dfrac{\gamma}{ \theta (\alpha,s)}\left( (1-\theta(\alpha,s)) s^t + \beta \left(\avein L_i \right)^2 r^t       \right) \\
		&=&  (1-\gamma \mu) \delta^{t} + \dfrac{\gamma}{\theta(\alpha,s)} \left(1-\dfrac{\theta(\alpha,s)}{2}\right) s^t  \\
		&& \qquad -\left(\dfrac{1}{2 \gamma} - \dfrac{L}{2} - \dfrac{\beta \LAMsq \gamma}{\theta (\alpha,s)}\right) r^{t}.
	\end{eqnarray*}
	
	Note that our extra assumption on the step size implies that 	$$1 - \dfrac{\theta(\alpha,s)}{2} \leq 1 -\gamma \mu$$ and $$\dfrac{1}{2 \gamma} - \dfrac{L}{2} - \dfrac{\beta \LAMsq \gamma}{\theta(\alpha,s)} \geq 0.$$ The last inequality follows from the bound $$\gamma^2\dfrac{2\beta \LAMsq}{\theta(\alpha,s)} + \gamma L \leq 1.$$ Thus,
	\begin{eqnarray*}
		\delta^{t+1}+\dfrac{\gamma}{ \theta (\alpha,s)} s^{t+1} \leq (1-\gamma \mu) \left(\delta^{t}+\dfrac{\gamma}{ \theta (\alpha,s)} s^{t} \right).
	\end{eqnarray*}
	It remains to unroll the recurrence which finishes the proof.
\end{proof}


\clearpage
\addtocounter{adjsection}{1}
\section{EF21-W-SGD: Weighted Error Feedback 2021 with Stochastic Subsampled Gradients} \label{ch3:sec:EF21-W-SGD}

The \algname{EF21-W} algorithm assumes that all clients can compute the exact gradient in each round. In some scenarios, the exact gradients may be unavailable or too costly to compute, and only approximate gradient estimators can be obtained. In this section, we present the convergence result for \algname{EF21-W} in the setting where the gradient computation on the clients, $\nabla f_i(x^{t+1})$, is replaced by a specific stochastic gradient estimator. For a variation of  \algname{EF21-W-SGD} which is working under a more general setting please see Appendix~\ref{ch3:sec:EF21-W-SGD-ABC}.


\subsection{Algorithm}




In this section, we extend \algname{EF21-W} to handle stochastic gradients, and we call the resulting algorithm \algname{EF21-W-SGD} (Algorithm~\ref{ch3:alg:weighted_ef21_sgd}). Our analysis of this extension follows a similar approach as the one used by \citet{fatkhullin2021ef21} for studying the stochastic gradient version of the vanilla \algname{EF21} algorithm, which they called \algname{EF21-SGD}. Analysis of \algname{EF21-W-SGD} has two important differences with vanilla \algname{EF21-SGD}:

\begin{enumerate}
	\item Vanilla \algname{EF21-SGD} provides maximum theoretically possible $$\gamma = \left( L + \LQM \sqrt{\dfrac{\beta_1}{\theta}} \right)^{-1},$$ where \algname{EF21-W-SGD} has $$\gamma = \left( L + \LAM \sqrt{\dfrac{\beta_1}{\theta}} \right)^{-1}.$$
	\item Vanilla \algname{EF21-SGD} and \algname{EF21-W-SGD} formally differs in a way how it reports iterate $x^T$ which minimizes $\ExpBr{\sqnorm{\nabla f(x^{T})}}$ due to a slightly different definition of $\widetilde{A}$. The \algname{EF21-W-SGD} (Algorithm \ref{ch3:alg:weighted_ef21_sgd}) requires output iterate $\hat{x}^T$ randomly according to the probability mass function described by \eqref{ch3:eq:09u09fd-0ff}.
\end{enumerate}


\begin{algorithm}
	\begin{algorithmic}[1]
		\STATE {\bfseries Input:} initial model $x^0 \in \RR^d$; initial gradient estimates $g_1^0, g_2^0, \dots,g_n^0 \in \R^d$ stored at the server and the clients; step size $\gamma>0$; number of iterations $T > 0$; weights ${\color{ForestGreen}w_i}= \dfrac{L_i}{\sum_j L_j}, i\in [n]$
		\STATE {\bfseries Initialize:} $g^0 = \sumin {\color{ForestGreen}w_i} g_i^0$ on the server
		\FOR{$t = 0, 1, 2, \dots, T - 1 $}
		\STATE Server computes $x^{t+1} = x^t - \gamma g^t$ and  broadcasts  $x^{t+1}$ to all $n$ clients		
		\FOR{$i = 1, \dots, n$ {\bf on the clients in parallel}}
		\STATE Compute a stochastic estimator  $\hat{g_i} (x^{t+1})$ of the gradient $\nabla f_i(x^{t+1})$
		\STATE  Compute $u_i^t=\cC_i^t\left(\dfrac{1}{n {\color{ForestGreen}w_i}} \hat{g_i} (x^{t+1}) - g_i^t\right)$ and update $g_i^{t+1} = g_i^t +u_i^t$ \label{ch3:line:weighted_ef21_sgd_grad_update}
		\STATE Send the compressed message $u_i^{t}$ to the server		
		\ENDFOR
		\STATE Server updates $g_i^{t+1} = g_i^t +u_i^t$ for all $i\in [n]$, and computes $g^{t+1} = \sum_{i=1}^n {\color{ForestGreen}w_i} g_i^{t+1}$		
		\ENDFOR
		\STATE {\bfseries Output:} Point $\hat{x}^T$ chosen from the set $\{x^0, \dots, x^{T-1}\}$ randomly according to the law \eqref{ch3:eq:09u09fd-0ff}
	\end{algorithmic}
	\caption{\algname{EF21-W-SGD}: Weighted Error Feedback 2021 with Stochastic Gradients.}
	\label{ch3:alg:weighted_ef21_sgd}
\end{algorithm}






\begin{assumption}[General assumption for stochastic gradient estimators]\label{ch3:as:general_as_for_stoch_gradients}
	We assume that for all $i \in [n]$ there exist parameters $A_i, C_i \ge 0$, $B_i \ge 1$ such that
	\begin{equation}
		\ExpBr{\|\nabla f_{\xi_{i j}^{t}}(x)\|^2} \le 2A_i\left(f_i(x) - f_i^{\inf}\right) + B_i\|\nabla f_i(x)\|^2 + C_i, \label{ch3:eq:general_second_mom_upp_bound}
	\end{equation}
	holds for all $x\in \RR^d$, 
	where\footnote{When $A_i = 0$ one can ignore the first term in the right-hand side of \eqref{ch3:eq:general_second_mom_upp_bound}, i.e., assumption $\inf_{x\in\R^d}f_i(x) > -\infty$ is not required in this case.} $f_i^{\inf} = \inf_{x\in\R^d}f_i(x) > -\infty$.
\end{assumption}


We study \algname{EF21-W-SGD} under the same assumption as was used for analyzing \algname{Vanilla EF21-SGD}, which we denote as Assumption~\ref{ch3:as:general_as_for_stoch_gradients}. To the best of our knowledge, this assumption, which was originally presented as Assumption 2 by \citet{khaled2020better}, is the most general assumption for a stochastic gradient estimator in a non-convex setting.

Next, to be aligned with original \algname{Vanilla EF21-SGD} \citep{fatkhullin2021ef21} we have considered a specific form of gradient estimator. This specific form of gradient estimator from \algname{Vanilla EF21-SGD} is presented in Section 4.1.2. of \citet{fatkhullin2021ef21} where the stochastic gradient $\hat{g_i}$ has been computed as follows:
\begin{eqnarray*}
	\hat{g_i} (x^{t+1}) = \dfrac{1}{\tau_i} \sum_{j=1}^{\tau_i}\nabla f_{\xi_{ij}^t}(x^{t+1}),
\end{eqnarray*}

Here $\tau_i$ is a minibatch size of sampled datapoint indexed by $\xi_{ij}^t$ of client $i$ in iteration $t$. And  $\xi_{ij}^t$ are independent random variables. For a version of  \algname{EF21-W-SGD} which is working under a more general setting please see Appendix~\ref{ch3:sec:EF21-W-SGD-ABC}.


\subsection{A lemma}

The contraction Lemma in this case gets the following form:
\begin{lemma} 
	Let $\cC_i^t \in \mathbb{C}(\alpha)$ for all $i\in [n]$ and $t\geq 0$. Define
	$$G_i^t \eqdef  \sqnorm{ g_i^t - \dfrac{\nabla f_i(x^{t})}{n w_i} } , \qquad G^t \eqdef \sumin w_i G_i^t.$$ Let Assumptions~\ref{ch3:as:L_i} and \ref{ch3:as:general_as_for_stoch_gradients} hold. Then, for any $s, \nu >0$ we have
	\begin{eqnarray}
		\label{ch3:eq:weighted_ef21_sgd_grad_full_contraction}
		\ExpBr{G^{t+1}} &\leq& (1-\hat{\theta}) \ExpBr{G^t} \notag \\
		&& + \hat{\beta_1} \LAMsq  \ExpBr{\sqnorm{ x^{t+1} - x^t}}  + { \widetilde{A} {\hat{\beta_2}}} \ExpBr{f(x^{t+1}) - f^{\inf}} \notag \\
		&& + {\widetilde{C} {\hat{\beta_2}}},
	\end{eqnarray}
	where 
	\begin{eqnarray*}
		w_i &\eqdef & \dfrac{L_i}{\sum_j L_j}, \\
		\hat{\theta} & \eqdef  & 1 - \rb{1-\alpha} (1+s) (1+\nu),\\
		\hat{\beta_1} &\eqdef  & 2(1- \alpha ) \left(1+ s \right)\left(s+\nu^{-1}\right), \\
		\hat{\beta_2} & \eqdef  & 2(1 - \alpha) (1 + s) (1+\nu^{-1}) + (1 + s^{-1}),\\
		\widetilde{A} &\eqdef  & \max_{i=1,\ldots,n} \left( \dfrac{2(A_i+L_i(B_i-1))}{\tau_i}  \dfrac{1}{n w_i} \right), \\
		\widetilde{C}  & \eqdef  &  \max_{i=1,\ldots,n} \left( \dfrac{C_i}{\tau_i}  \dfrac{1}{n w_i} \right).
	\end{eqnarray*}
	
\end{lemma}

\begin{proof}
	Define  $W^t \eqdef \{g_1^t, \dots,    g_n^t, x^t, x^{t+1}\}$. The proof starts as follows:
	\begin{eqnarray*}
		\ExpBr{ G_i^{t+1} \;|\; W^t} &\overset{\eqref{ch3:eq:weighted_def_grad_distortion}}{=}& \ExpBr{  \sqnorm{g_i^{t+1} 
				- \dfrac{\nabla f_i(x^{t+1})}{n w_i}}  \;|\; W^t}	 \\
		&\overset{\text{line}~\ref{ch3:line:weighted_ef21_sgd_grad_update}}{=}& \ExpBr{  \sqnorm{g_i^t + \cC_i^t \left( \dfrac{\hat{g_i}(x^{t+1})}{n w_i} - g_i^t \right) 
				- \dfrac{\nabla f_i(x^{t+1})}{n w_i}}  \;|\; W^t}	 \\
		&=& \ExpBrBig{\sqnorm{\cC_i^t \left( \dfrac{\hat{g_i}(x^{t+1})}{n w_i} - g_i^t \right) - \left(\dfrac{\hat{g_i}(x^{t+1})}{n w_i} - g_i^t\right) \right.\notag \\&& \qquad \left.+ \dfrac{\hat{g_i}(x^{t+1})}{n w_i}
				- \dfrac{\nabla f_i(x^{t+1})}{n w_i}}  \;|\; W^t } \\
		&\overset{\eqref{ch3:eq:young}}{\leq}& (1+s) \ExpBr{ \sqnorm{\cC_i^t \left( \dfrac{\hat{g_i}(x^{t+1})}{n w_i} - g_i^t \right) - \left(\dfrac{\hat{g_i}(x^{t+1})}{n w_i} - g_i^t\right)}\;|\; W^t}  \\ 
		& & \qquad + (1+s^{-1})  \ExpBr{\sqnorm{\dfrac{\hat{g_i}(x^{t+1})}{n w_i}
				- \dfrac{\nabla f_i(x^{t+1})}{n w_i}}\;|\; W^t} \\
		&\overset{\eqref{ch3:eq:compressor_contraction}}{\leq}&  (1-\alpha) (1+s) \ExpBr{\sqnorm{\dfrac{\hat{g_i}(x^{t+1})}{n w_i} - \dfrac{\nabla f_i(x^t)}{n w_i} +\dfrac{\nabla f_i(x^t)}{n w_i} - g_i^t}\;|\; W^t}  \\ 
		& & \qquad + (1+s^{-1})  \ExpBr{\sqnorm{\dfrac{\hat{g_i}(x^{t+1})}{n w_i}
				- \dfrac{\nabla f_i(x^{t+1})}{n w_i}}\;|\; W^t} \\
		&\overset{\eqref{ch3:eq:young}}{\leq}& (1 - \alpha) (1 + s) (1+\nu) \ExpBr{\sqnorm{{g_i}^{t} - \dfrac{\nabla f_i(x^t)}{n w_i}}} \\
		& & \quad + (1 - \alpha) (1 + s) (1+\nu^{-1}) \ExpBr{\sqnorm{\dfrac{\nabla f_i(x^t)}{n w_i} - \dfrac{\hat{g_i}(x^{t+1})}{n w_i}}\;|\; W^t} \\
		& & \quad + (1+s^{-1}) \ExpBr{\sqnorm{\dfrac{\hat{g_i}(x^{t+1})}{n w_i}
				- \dfrac{\nabla f_i(x^{t+1})}{n w_i}}\;|\; W^t} \\
		&\overset{\eqref{ch3:eq:young}}{\leq}& (1 - \alpha) (1 + s) (1+\nu) \ExpBr{\sqnorm{ {g_i}^{t} - \dfrac{\nabla f_i(x^t)}{n w_i}}\;|\; W^t} \\
		& & + {2(1 - \alpha) (1 + s) (1+\nu^{-1})} \ExpBr{\sqnorm{\dfrac{\nabla f_i(x^{t+1})}{n w_i} - \dfrac{\hat{g_i}(x^{t+1})}{n w_i}}\;|\; W^t} \\
		& & + 2(1 - \alpha) (1 + s) (1+\nu^{-1}) {\sqnorm{\dfrac{\nabla f_i(x^{t+1})}{n w_i} - \dfrac{\nabla f_i(x^{t})}{n w_i}}} \\
		& & + {(1+s^{-1})} \ExpBr{\sqnorm{\dfrac{\hat{g_i}(x^{t+1})}{n w_i}
				- \dfrac{\nabla f_i(x^{t+1})}{n w_i}}\;|\; W^t}.
	\end{eqnarray*}
	
	To further bound the last term, which contains multiple ${(1+s^{-1})}$ factors, we leverage the property that $\hat g_i(x^{t+1})$ is a random variable serving as an unbiased estimator of $\nabla f_i(x^{t+1})$, taking the form $\hat{g_i} (x^{t+1}) = \dfrac{1}{\tau_i} \sum_{j=1}^{\tau_i}\nabla f_{\xi_{ij}^t}(x^{t+1}),$ where $\xi_{ij}^t$ are independent random variables. Next, we can continue as follows:
	\begin{eqnarray*}
		\ExpBr{ G_i^{t+1} \;|\; W^t} &\leq& (1-\hat{\theta}) \ExpBr{G_i^t \;|\; W^t} + \hat{\beta_1} \dfrac{1}{n^2 w_i^2} \sqnorm{\nabla f_i(x^{t+1}) - \nabla f_i(x^t)} \\
		&& + \dfrac{{\hat{\beta_2}}}{(n w_i)^2} \left( \ExpBr{\sqnorm{\dfrac{1}{\tau_i} \sum_{j=1}^{\tau_i}\nabla f_{\xi_{ij}^t}(x^{t+1}) - \dfrac{1}{\tau_i} \sum_{j=1}^{\tau_i} {\nabla f_i}(x^{t+1})}\;|\; W^t} \right) \\
		& = & (1-\hat{\theta}) \ExpBr{G_i^t \;|\; W^t} + \hat{\beta_1} \dfrac{1}{n^2 w_i^2} \sqnorm{\nabla f_i(x^{t+1}) - \nabla f_i(x^t)} \\
		&& + \dfrac{{\hat{\beta_2}}}{(n w_i)^2 \tau^2} \left( \ExpBr{\sqnorm{ \sum_{j=1}^{\tau_i} \left( \nabla f_{\xi_{ij}^t}(x^{t+1}) - {\nabla f_i}(x^{t+1})\right) }\;|\; W^t} \right)  \\
		& = & (1-\hat{\theta}) \ExpBr{G_i^t \;|\; W^t} + \hat{\beta_1} \dfrac{1}{n^2 w_i^2} \sqnorm{\nabla f_i(x^{t+1}) - \nabla f_i(x^t)} \\
		&& + \dfrac{{\hat{\beta_2}}}{(n w_i)^2 {\tau_i}^2} \sum_{j=1}^{\tau_i} \Big( \ExpBr{\sqnorm{ \nabla f_{\xi_{ij}^t}(x^{t+1})}\;|\; W^t} \\
		&& \qquad \qquad \qquad \qquad - \sqnorm{\ExpBr{ \nabla f_{\xi_{ij}^t}(x^{t+1})\;|\; W^t}} \Big) \\
		&\le& (1-\hat{\theta}) \ExpBr{G_i^t \;|\; W^t} + \hat{\beta_1} \dfrac{1}{n^2 w_i^2} \sqnorm{\nabla f_i(x^{t+1}) - \nabla f_i(x^t)} \\
		&&  + \dfrac{{\hat{\beta_2}}}{(n w_i)^2 {\tau_i}^2} \sum_{j=1}^{\tau_i} \Big(
		2A_i\left(f_i(x^{t+1}) - f_i^{\inf}\right) \\
		&& \qquad \qquad \qquad \qquad + B_i\|\nabla f_i(x^{t+1})\|^2 + C_i - \sqnorm{\nabla f_i(x^{t+1}}) \Big) \\
		&=& (1-\hat{\theta}) \ExpBr{G_i^t \;|\; W^t} + \hat{\beta_1} \dfrac{1}{n^2 w_i^2} \sqnorm{\nabla f_i(x^{t+1}) - \sqnorm{\nabla f_i(x^t)}} \\
		&& + \dfrac{ 2A_i {\hat{\beta_2}}}{(n w_i)^2 \tau_i} \left(f_i(x^{t+1}) - f_i^{\inf}\right) + \dfrac{2(B_i-1){\hat{\beta_2}}}{(n w_i)^2 \tau_i} \left(\dfrac{1}{2} \|\nabla f_i(x^{t+1})\|^2 \right) \\
		&& + \dfrac{C_i {\hat{\beta_2}}}{(n w_i)^2 \tau_i} \\
		&\le& (1-\hat{\theta}) \ExpBr{G_i^t \;|\; W^t} + \hat{\beta_1} \dfrac{1}{n^2 w_i^2} \sqnorm{\nabla f_i(x^{t+1}) - {\nabla f_i(x^t)}} \\
		&& + \dfrac{ 2A_i {\hat{\beta_2}}}{(n w_i)^2 \tau_i} \left(f_i(x^{t+1}) - f_i^{\inf}\right)
		+  \dfrac{2(B_i-1){\hat{\beta_2}}}{(n w_i)^2 \tau_i} L_i \left(f_i(x^{t+1}) - f_i^{\inf} \right) \\
		&& + \dfrac{C_i {\hat{\beta_2}}}{(n w_i)^2 \tau_i} \\
		&=& (1-\hat{\theta}) \ExpBr{G_i^t \;|\; W^t} + \hat{\beta_1} \dfrac{1}{n^2 w_i^2} \sqnorm{\nabla f_i(x^{t+1}) - {\nabla f_i(x^t)}} \\
		&& + \dfrac{ 2(A_i + L_i(B_i - 1)) {\hat{\beta_2}}}{(n w_i)^2 \tau_i} \left(f_i(x^{t+1}) - f_i^{\inf}\right)
		+  \dfrac{C_i {\hat{\beta_2}}}{(n w_i)^2 \tau_i}.
	\end{eqnarray*}
	
	Furthermore, as a result of leveraging Assumption~\ref{ch3:as:L_i}, we can derive the subsequent bound:
	\begin{eqnarray*}
		\ExpBr{ G_i^{t+1} \;|\; W^t} & \leq & (1-\hat{\theta})   G_i^t +  \dfrac{\hat{\beta_1} L_i^2}{n^2 w_i^2} \sqnorm{ x^{t+1} - x^t}  \\	
		&& \qquad 
		+ \dfrac{ 2(A_i + L_i(B_i - 1)) {\hat{\beta_2}}}{(n w_i)^2 \tau_i} \left(f_i(x^{t+1}) - f_i^{\inf}\right)
		+  \dfrac{C_i {\hat{\beta_2}}}{(n w_i)^2 \tau_i}.
	\end{eqnarray*}
	Applying the tower property and subsequently taking the expectation, we obtain:
	\begin{equation}\label{ch3:eq:weighted_ef21_sgd_aux_1}
		\begin{aligned}
			\ExpBr{ G_i^{t+1}} &\leq (1-\hat{\theta}) \ExpBr{G_i^t} + \hat{\beta_1} \dfrac{1}{n^2 w_i^2} L_i^2 \ExpBr{\sqnorm{ x^{t+1} - x^t}} 
			\\
			& \qquad + \dfrac{ 2(A_i + L_i(B_i - 1)) {\hat{\beta_2}}}{(n w_i)^2 \tau_i} \ExpBr{f_i(x^{t+1}) - f_i^{\inf}}
			+  \dfrac{C_i {\hat{\beta_2}}}{(n w_i)^2 \tau_i}.
		\end{aligned}
	\end{equation}
	Regarding the expectation of $G^{t+1}$, we derive the subsequent bound:
	\begin{eqnarray*}
		\ExpBr{G^{t+1}} &=& \ExpBr{\sumin w_i G_i^{t+1}} \\
		& =& \sumin w_i \ExpBr{G_i^{t+1}} \\
		&\overset{\eqref{ch3:eq:weighted_ef21_sgd_aux_1}}{\leq} & (1-\hat{\theta}) \sumin w_i \ExpBr{G_i^t} + \sumin w_i \hat{\beta_1} \dfrac{1}{n^2 w_i^2} L_i^2 \cdot \ExpBr{\sqnorm{ x^{t+1} - x^t}} \\
		&& \qquad + \sumin w_i \dfrac{ 2(A_i + L_i(B_i - 1)) {\hat{\beta_2}}}{(n w_i)^2 \tau_i} \cdot \ExpBr{f_i(x^{t+1}) - f_i^{\inf}} \\
		&& \qquad +  \sumin w_i \dfrac{C_i {\hat{\beta_2}}}{(n w_i)^2 \tau_i}	\\
		&= &(1-\hat{\theta}) \ExpBr{G^t} + \sumin \hat{\beta_1} \dfrac{1}{n^2 w_i} L_i^2 \cdot \ExpBr{\sqnorm{ x^{t+1} - x^t}} \\
		&& \qquad + \sumin \dfrac{ 2(A_i + L_i(B_i - 1)) {\hat{\beta_2}}}{n^2 w_i \tau_i} \cdot \ExpBr{f_i(x^{t+1}) - f_i^{\inf}} \\
		&& \qquad + \sumin \dfrac{C_i {\hat{\beta_2}}}{n^2 w_i \tau_i}.
	\end{eqnarray*}
	
	
	Employing quantities $\tilde{A}$ and $\tilde{C}$, the final bound can be reformulated as follows:
	\begin{eqnarray*}
		\ExpBr{G^{t+1}} & \leq & (1-\hat{\theta}) \ExpBr{G^t} + \sumin \hat{\beta_1} \dfrac{1}{n^2 w_i} L_i^2 \cdot \ExpBr{\sqnorm{ x^{t+1} - x^t}} \notag \\
		&& \qquad 	+ \dfrac{1}{n} \sumin { \widetilde{A} {\hat{\beta_2}}} \cdot \ExpBr{f_i(x^{t+1}) - f_i^{\inf}} +   {\widetilde{C} {\hat{\beta_2}}} \notag \\
		& \leq & (1-\hat{\theta}) \ExpBr{G^t} + \sumin \hat{\beta_1} \dfrac{1}{n^2 w_i} L_i^2 \cdot \ExpBr{\sqnorm{ x^{t+1} - x^t}} \notag \\
		&& \qquad 	+ \dfrac{1}{n} \sumin { \widetilde{A} {\hat{\beta_2}}} \cdot \ExpBr{f_i(x^{t+1}) - f^{\inf}} +   {\widetilde{C} {\hat{\beta_2}}} \notag \\
		& \leq & (1-\hat{\theta}) \ExpBr{G^t} + \sumin \hat{\beta_1} \dfrac{1}{n^2 w_i} L_i^2 \cdot \ExpBr{\sqnorm{ x^{t+1} - x^t}} \\
		&& \qquad 	+ { \widetilde{A} {\hat{\beta_2}}} \notag \ExpBr{f(x^{t+1}) - f^{\inf}}
		+  {\widetilde{C} {\hat{\beta_2}}}.
	\end{eqnarray*}

	
	Given that $w_i=\dfrac{L_i}{\sum_j L_j}$, we have:
	\begin{eqnarray*}
		\ExpBr{G^{t+1}} &\leq & (1-\hat{\theta}) \ExpBr{G^t} + \dfrac{1}{n}\sumin \hat{\beta_1} \dfrac{\sum_j L_j}{n} L_i \cdot \ExpBr{\sqnorm{ x^{t+1} - x^t}} \notag \\
		&&  \qquad + { \widetilde{A} {\hat{\beta_2}}} \ExpBr{f(x^{t+1}) - f^{\inf}}
		+  {\widetilde{C} {\hat{\beta_2}}} \notag \\
		& = & (1-\hat{\theta}) \ExpBr{G^t} + \hat{\beta_1} \left(\avein L_i\right)^2 \cdot \ExpBr{\sqnorm{ x^{t+1} - x^t}} \notag \\
		&& \qquad + { \widetilde{A} {\hat{\beta_2}}} \ExpBr{f(x^{t+1}) - f^{\inf}}
		+  {\widetilde{C} {\hat{\beta_2}}},
	\end{eqnarray*}
	what completes the proof.
\end{proof}

\subsection{Main result}

Now we are ready to prove the main convergence theorem.
\begin{theorem} Let $\cC_i^t \in\mathbb{C}(\alpha)$ for all $\in [n]$ and $t\geq 0$ in Algorithm~\ref{ch3:alg:weighted_ef21_sgd}. Set the following quantities:
	\begin{eqnarray*} 
		\hat{\theta} &\eqdef & 1 - \rb{1-\alpha} (1+s) (1+\nu),\\
		\hat{\beta_1} &\eqdef  & 2(1- \alpha ) \left(1+ s \right)\left(s+\nu^{-1}\right),\\
		\hat{\beta_2} &\eqdef  & 2(1 - \alpha) (1 + s) (1+\nu^{-1}) + (1 + s^{-1}),\\
		w_i &\eqdef &  \dfrac{L_i}{\sum_j L_j},\\
		\widetilde{A} & \eqdef  & \max_{i=1,\ldots,n} \left( \dfrac{2(A_i+L_i(B_i-1))}{\tau_i}  \dfrac{1}{n w_i} \right), \\
		\widetilde{C} & \eqdef  & \max_{i=1,\ldots,n} \left( \dfrac{C_i}{\tau_i}  \dfrac{1}{n w_i} \right).
	\end{eqnarray*} 
	Under Assumptions~\ref{ch3:as:smooth},~\ref{ch3:as:L_i}, and~\ref{ch3:as:general_as_for_stoch_gradients}, and selection of $s>0$, $\mu>0$ such that $$(1+s)(1+\mu) < \dfrac{1}{1-\alpha}$$ set the step size in the following way:
	\begin{equation}
		\gamma \leq \dfrac{1}{L + \LAM \sqrt{\dfrac{\hat{\beta}_1}{\hat{\theta}}}}.
	\end{equation}
	Choose an iterate $\hat{x}^T$ from $\{x^0, x^1, \dots, x^{T-1}\}$ with probability 
	\begin{equation} \label{ch3:eq:09u09fd-0ff}
		\Prob(\hat{x}^T = x^t) = \dfrac{v_t}{V_T},\end{equation} where
	$$ v_t \eqdef \left(1 - \dfrac{\gamma \tilde{A} \tilde{\beta}_2}{2\theta}\right)^t; \qquad V_T \eqdef \sum\limits_{t=0}^{T-1} v_t.
	$$
	Then,
	\begin{equation}
		\ExpBr{\sqnorm{\nabla f(\hat{x}^{T})}} \leq \dfrac{2 (f(x^0) - f^\text{inf})}{\gamma T  \left(1 - \dfrac{\gamma \widetilde{A} \hat{\beta_2}}{2 \theta}\right)^T } + \dfrac{G^0}{ \hat{\theta}T  \left(1 - \dfrac{\gamma \widetilde{A} \hat{\beta_2}}{2 \theta}\right)^T} + \dfrac{\widetilde{C}\beta_2}{\hat{\theta}},
	\end{equation}
	where $$G^0 \eqdef \sumin w_i \|g_i^0 - \dfrac{1}{nw_i}\nabla f_i(x^0)\|^2.$$
\end{theorem}
\begin{proof}
	In the derivation below, we  use Lemma~\ref{ch3:lm:descent_lemma} for 
	\begin{equation}\label{ch3:eq:sgd_weighted_grad_estimate_def}
		g^t=\sum\limits_{i=1}^{n} w_i g_i^t.
	\end{equation}
	We start as follows:
	\begin{eqnarray}
		f(x^{t+1}) &\overset{\eqref{ch3:eq:descent_lemma}}{\leq} &
		f(x^{t})-\dfrac{\gamma}{2}\sqnorm{\nabla f(x^{t})}-\left(\dfrac{1}{2 \gamma}-\dfrac{L}{2}\right)\sqnorm{x^{t+1}-x^{t}} \notag \\
		&& +\dfrac{\gamma}{2}\sqnorm{
			g^t- \sumin \nabla f_i(x^{t}) } \notag \\
		& \overset{\eqref{ch3:eq:sgd_weighted_grad_estimate_def}}{=} &
		f(x^{t})-\dfrac{\gamma}{2}\sqnorm{\nabla f(x^{t})}-\left(\dfrac{1}{2 \gamma}-\dfrac{L} 
		{2}\right)\sqnorm{x^{t+1}-x^{t}} \notag \\
		&& +\dfrac{\gamma}{2}\sqnorm{
			\sumin w_i \left(g_i^t-  \dfrac{\nabla f_i(x^{t})}{n w_i} \right) }  \notag  \\
		& \leq &
		f(x^{t})-\dfrac{\gamma}{2}\sqnorm{\nabla f(x^{t})}-\left(\dfrac{1}{2 \gamma}-\dfrac{L}       
		{2}\right)\sqnorm{x^{t+1}-x^{t}} \notag \\
		&& +\dfrac{\gamma}{2}
		\sumin w_i \sqnorm{g_i^t-  \dfrac{\nabla f_i(x^{t})}{n w_i} }  \notag \\
		& = & f(x^{t})-\dfrac{\gamma}{2}\sqnorm{\nabla f(x^{t})}-\left(\dfrac{1}{2 \gamma}-\dfrac{L}       
		{2}\right)\sqnorm{x^{t+1}-x^{t}} \notag \\
		&& +\dfrac{\gamma}{2} G^t.
	\end{eqnarray}
	
	Subtracting $f^\ast$ from both sides and taking expectation, we get
	\begin{eqnarray}
		\ExpBr{f(x^{t+1})-f^\ast} & \leq &  \ExpBr{f(x^{t})-f^\ast}
		-\dfrac{\gamma}{2} \ExpBr{\sqnorm{\nabla f(x^{t})}} \notag \\
		&& \qquad  -\left(\dfrac{1}{2 \gamma}
		-\dfrac{L}{2}\right) \ExpBr{\sqnorm{x^{t+1}-x^{t}}} \notag \\
		&& \qquad +\dfrac{\gamma}{2}\ExpBr{G^t}.\label{ch3:eq:weighted_function_descent_sgd}
	\end{eqnarray}

	Let 
	\begin{eqnarray*}
		\delta^{t} &\eqdef& \ExpBr{f(x^{t})-f^\ast}, \\
		s^{t} &\eqdef& \ExpBr{G^t},\\
		r^{t} &\eqdef& \ExpBr{\sqnorm{x^{t+1}-x^{t}}}.
	\end{eqnarray*}
	
	Then by adding $\dfrac{\gamma}{2\theta} s^{t+1}$ and employing \eqref{ch3:eq:weighted_ef21_sgd_grad_full_contraction}, we obtain:
	\begin{align*}
		\delta^{t+1}+\dfrac{\gamma}{2 \hat{\theta}} s^{t+1} 
		&\leq \delta^{t}-\dfrac{\gamma}{2}\ExpBr{\sqnorm{\nabla f(x^{t})}} -          \left(\dfrac{1}{2 \gamma}-\dfrac{L}{2}\right) r^{t}+\dfrac{\gamma}{2} s^{t} \\
		& \qquad +\dfrac{\gamma}{2 \hat{\theta}}\left( \hat{\beta_1} \LAMsq  r^t + (1-\hat{\theta}) s^t + { \widetilde{A} {\hat{\beta_2}}} \delta^{t+1} +  {\widetilde{C} {\hat{\beta_2}}} \right) \\
		&=\delta^{t}+\dfrac{\gamma}{2\hat{\theta}} s^{t}-\dfrac{\gamma}{2}\ExpBr{\sqnorm{\nabla f(x^{t})}}  -\left(\dfrac{1}{2\gamma} -\dfrac{L}{2} -  \dfrac{\gamma}{2\hat{\theta}} \hat{\beta_1}  \LAMsq  \right) r^{t} \\
		& \qquad + \dfrac{\gamma \widetilde{A} \beta_2}{2\hat{\theta}} \delta^{t+1} + \dfrac{\gamma \widetilde{C}}{2 \hat{\theta}} \beta_2\\
		& \leq \delta^{t}+\dfrac{\gamma}{2\hat{\theta}} s^{t} -\dfrac{\gamma}{2}\ExpBr{\sqnorm{\nabla f(x^{t})}} + \dfrac{\gamma \widetilde{A} \beta_2}{2\hat{\theta}} \delta^{t+1} + \dfrac{\gamma \widetilde{C}}{2 \hat{\theta}} \beta_2.
	\end{align*}
	
	
	The last inequality follows from the bound $$\gamma^2\dfrac{\hat{\beta_1} \LAMsq }{\hat{\theta}} + L\gamma \leq 1,$$ 
	which holds due to Lemma~\ref{ch3:lm:stepsize_bound} for $$\gamma \leq \dfrac{1}{L +  \LAM \sqrt{\dfrac{\hat{\beta}_1}{\hat{\theta}}}}.$$
	Subsequently, we will reconfigure the final inequality and perform algebraic manipulations, taking into account that $\dfrac{2}{\gamma} > 0.$ In the final step of these algebraic transformations, we will leverage the fact that $s^t \ge 0$:
	\begin{eqnarray*}
		\delta^{t+1}+\dfrac{\gamma}{2 \hat{\theta}} s^{t+1} &\leq& \delta^{t}+\dfrac{\gamma}{2\hat{\theta}} s^{t} -\dfrac{\gamma}{2}\ExpBr{\sqnorm{\nabla f(x^{t})}} + \dfrac{\gamma \widetilde{A} \beta_2}{2\hat{\theta}} \delta^{t+1} + \dfrac{\gamma \widetilde{C}}{2 \hat{\theta}} \beta_2 .
	\end{eqnarray*}	
	
	Therefore,
	\begin{eqnarray*}
		\dfrac{2}{\gamma} \delta^{t+1}+\dfrac{2}{\gamma} \dfrac{\gamma}{2 \hat{\theta}} s^{t+1} &\leq& \dfrac{2}{\gamma} \delta^{t} + \dfrac{2}{\gamma} \dfrac{\gamma}{2\hat{\theta}} s^{t} -\ExpBr{\sqnorm{\nabla f(x^{t})}} + \dfrac{2}{\gamma} \dfrac{\gamma \widetilde{A} \beta_2}{2\hat{\theta}} \delta^{t+1} + \dfrac{2}{\gamma} \dfrac{\gamma \widetilde{C}}{2 \hat{\theta}} \beta_2.
	\end{eqnarray*}	
	
	Further,
	\begin{eqnarray*}	
		\ExpBr{\sqnorm{\nabla f(x^{t})}} &\leq& -\dfrac{2}{\gamma} \delta^{t+1} - \dfrac{2}{\gamma} \dfrac{\gamma}{2 \hat{\theta}} s^{t+1} + \dfrac{2}{\gamma} \delta^{t} + \dfrac{2}{\gamma} \dfrac{\gamma}{2\hat{\theta}} s^{t} + \dfrac{2}{\gamma} \dfrac{\gamma \widetilde{A} \beta_2}{2\hat{\theta}} \delta^{t+1} + \dfrac{2}{\gamma} \dfrac{\gamma \widetilde{C}}{2 \hat{\theta}} \beta_2  \\
		&\leq&
		-\dfrac{2}{\gamma} \delta^{t+1} - \dfrac{2}{\gamma} \dfrac{\gamma}{2\hat{\theta}} s^{t+1} + \dfrac{2}{\gamma} \left( \delta^{t} + \dfrac{\gamma}{2\hat{\theta}} s^{t} \right) + \dfrac{2}{\gamma} \dfrac{\gamma \widetilde{A} \beta_2}{2 \hat{\theta}} \delta^{t+1} + \dfrac{\widetilde{C}\beta_2}{ \hat{\theta}}  \\
		&\leq&
		\dfrac{2}{\gamma} \left( \left( \delta^{t} + \dfrac{\gamma}{2\hat{\theta}} s^{t} \right) -1\left(1 - \dfrac{\gamma \widetilde{A} \beta_2}{2 \hat{\theta}} \right) \delta^{t+1} -\left(\dfrac{\gamma}{2\hat{\theta}} s^{t+1} \right) \right) + \dfrac{\widetilde{C}\beta_2}{ \hat{\theta}} \\
		&\leq& \dfrac{2}{\gamma} \left( \left( \delta^{t} + \dfrac{\gamma}{2\hat{\theta}} s^{t} \right) -\left(1 - \dfrac{\gamma \widetilde{A} \beta_2}{2 \hat{\theta}} \right) \left(\delta^{t+1} + \dfrac{\gamma}{2\hat{\theta}} s^{t+1} \right) \right) + \dfrac{\widetilde{C}\beta_2}{ \hat{\theta}}.
	\end{eqnarray*}
	
	Next, we sum up inequalities above with weights $v_t/V_T$, where 
	$$v_t \eqdef (1 - \dfrac{\gamma \widetilde{A} \hat{\beta_2}}{2 \theta})^t \quad \mathrm{and} \quad V_T \eqdef \sum_{i=1}^{T} v_i.$$
	\begin{eqnarray*}
		\ExpBr{\sqnorm{\nabla f(\hat{x}^{T})}} &=& \sum_{t=0}^{T} \dfrac{v_t}{V_T} \ExpBr{\sqnorm{\nabla f(x^{t})}} \\
		&=& \dfrac{1}{V_T} \sum_{t=0}^{T} v_t \ExpBr{\sqnorm{\nabla f(x^{t})}} \\
		&\leq& \dfrac{1}{V_T} \sum_{t=0}^{T} v_t \Big(\dfrac{2}{\gamma} \left( \left( \delta^{t} + \dfrac{\gamma}{2\hat{\theta}} s^{t} \right) -\left(1 - \dfrac{\gamma \widetilde{A} \beta_2}{2 \hat{\theta}} \right) \left(\delta^{t+1} + \dfrac{\gamma}{2\hat{\theta}} s^{t+1} \right) \right) \\
		&& \qquad \qquad + \dfrac{\widetilde{C}\beta_2}{ \hat{\theta}} \Big) \\
		&=& \dfrac{2}{\gamma V_T} \sum_{t=0}^{T} w_t \left( \left( \delta^{t} + \dfrac{\gamma}{2\hat{\theta}} s^{t} \right) -\left(1 - \dfrac{\gamma \widetilde{A} \beta_2}{2 \hat{\theta}} \right) \left(\delta^{t+1} + \dfrac{\gamma}{2\hat{\theta}} s^{t+1} \right) \right)	\\
		&& + \sum_{t=0}^{T} \dfrac{w_t}{W_T} \cdot \dfrac{\widetilde{C}\beta_2}{ \hat{\theta}} \\
		&=& \dfrac{2}{\gamma V_T} \sum_{t=0}^{T} w_t \left( \left( \delta^{t} + \dfrac{\gamma}{2\hat{\theta}} s^{t} \right) -\left(1 - \dfrac{\gamma \widetilde{A} \beta_2}{2 \hat{\theta}} \right) \left(\delta^{t+1} + \dfrac{\gamma}{2\hat{\theta}} s^{t+1} \right) \right)\\
		&& + \dfrac{\widetilde{C}\beta_2}{ \hat{\theta}} \\
		&=& \dfrac{2}{\gamma V_T} \sum_{t=0}^{T} \left(w_t \left( \delta^{t} + \dfrac{\gamma}{2\hat{\theta}} s^{t} \right) -w_{t+1} \left(\delta^{t+1} + \dfrac{\gamma}{2\hat{\theta}} s^{t+1} \right) \right) \\
		&& + \dfrac{\widetilde{C}\beta_2}{ \hat{\theta}} \\
		&\leq& \dfrac{2 \delta^0}{\gamma V_T} + \dfrac{s^0}{ \hat{\theta}V_T} + \dfrac{\widetilde{C}\beta_2}{\hat{\theta}}.
	\end{eqnarray*}
	Finally, we notice that $$V_T = \sum\limits_{t=1}^T (1 - \dfrac{\gamma \widetilde{A} \hat{\beta_2}}{2 \theta})^t \geq T \cdot (1 - \dfrac{\gamma \widetilde{A} \hat{\beta_2}}{2 \theta})^T,$$ what concludes the proof.
\end{proof}

\clearpage
\addtocounter{adjsection}{1}
\section{EF21-W-SGD: Weighted Error Feedback 2021 with Stochastic Gradients under the ABC Assumption} \label{ch3:sec:EF21-W-SGD-ABC}

In this section, we present the convergence result for \algname{Weighted EF21} in the setting where the gradient computation on the clients is replaced with a pretty general unbiased stochastic gradient estimator.

\subsection{Algorithm}


The \algname{EF21-W} algorithm assumes that all clients can compute the exact gradient in each round. In some scenarios, the exact gradients may be unavailable or too costly to compute, and only approximate gradient estimators can be obtained. To have the ability for \algname{EF21-W} to work in such circumstances we extended \algname{EF21-W} to handle stochastic gradients. We called the resulting algorithm \algname{EF21-W-SGD} (Algorithm~\ref{ch3:alg:weighted_ef21_sgd_abc}).

\begin{algorithm}
	\begin{algorithmic}[1]
		\STATE {\bfseries Input:} initial model $x^0 \in \RR^d$; initial gradient estimates $g_1^0, \dots, g_n^0 \in \R^d$ stored at the server and the clients; step size $\gamma>0$; number of iterations $T > 0$; weights ${\color{ForestGreen}w_i}= \dfrac{L_i}{\sum_j L_j}$ for $i\in [n]$
		\STATE {\bfseries Initialize:} $g^0 = \sumin {\color{ForestGreen}w_i} g_i^0$ on the server
		\FOR{$t = 0, 1, 2, \dots, T - 1 $}
		\STATE Server computes $x^{t+1} = x^t - \gamma g^t$ and  broadcasts  $x^{t+1}$ to all $n$ clients		
		\FOR{$i = 1, \dots, n$ {\bf on the clients in parallel}}
		\STATE Compute a stochastic gradient $\hat{g_i} (x^{t+1})$ estimator of the gradient $\nabla f_i(x^{t+1})$
		\STATE  Compute $u_i^t=\cC_i^t\left(\dfrac{1}{n {\color{ForestGreen}w_i}} \hat{g_i} (x^{t+1}) - g_i^t\right)$ and update $g_i^{t+1} = g_i^t +u_i^t$ \label{ch3:line:weighted_ef21_sgd_grad_update_abc}
		\STATE Send the compressed message $u_i^{t}$ to the server		
		\ENDFOR
		\STATE Server updates $g_i^{t+1} = g_i^t +u_i^t$ for all $i\in [n]$, and computes $g^{t+1} = \sum_{i=1}^n {\color{ForestGreen}w_i} g_i^{t+1}$		
		\ENDFOR
		\STATE {\bfseries Output:} Point $\hat{x}^T$ chosen from the set $\{x^0, \dots, x^{T-1}\}$ randomly according to the law \eqref{ch3:eq:09u09fd-0ff-abc}
	\end{algorithmic}
	\caption{\algname{EF21-W-SGD}: Weighted EF-21 with Stochastic Gradients under ABC assumption.}
	\label{ch3:alg:weighted_ef21_sgd_abc}
\end{algorithm}

Our analysis of this extension follows a similar approach as the one used by \citet{fatkhullin2021ef21} for studying the stochastic gradient version under the name \algname{EF21-SGD}. However, \algname{EF21-W-SGD} has four important differences with vanilla \algname{EF21-SGD}:

\begin{enumerate}
	\item Vanilla \algname{EF21-SGD} algorithm analyzed by \citet{fatkhullin2021ef21} worked under a specific sampling schema for a stochastic gradient estimator. Our analysis works under a more general ABC Assumption~\ref{ch3:as:general_as_for_stoch_gradients_abc}.
	\item  Vanilla \algname{EF21-SGD} provides maximum theoretically possible $$\gamma = \left( L + \LQM \sqrt{\dfrac{\beta_1}{\theta}} \right)^{-1},$$ where \algname{EF21-W-SGD} has $$\gamma = \left( L + \LAM \sqrt{\dfrac{\beta_1}{\theta}} \right)^{-1}.$$
	\item In contrast to the original analysis \algname{Vanilla EF21-SGD} our analysis provides a more aggressive $\beta_1$ parameter which is smaller by a factor of $2$.	
	\item Vanilla \algname{EF21-SGD} and \algname{EF21-W-SGD} formally differs in a way how it reports iterate $x^T$ which minimizes $\ExpBr{\sqnorm{\nabla f(x^{T})}}$ due to a slightly different definition of $\widetilde{A}$. The \algname{EF21-W-SGD} (Algorithm \ref{ch3:alg:weighted_ef21_sgd_abc}) requires output iterate $\hat{x}^T$ randomly according to the probability mass function described by Equation~\eqref{ch3:eq:09u09fd-0ff-abc}.
\end{enumerate}

\begin{assumption}[General assumption for stochastic gradient estimators]
	\label{ch3:as:general_as_for_stoch_gradients_abc}
	We assume that for all $i \in [n]$ there exist parameters $A_i, C_i \ge 0$, $B_i \ge 1$ such that
	\begin{equation}
		\ExpBr{\|\nabla \hat{g_i} (x)\|^2} \le 2A_i\left(f_i(x) - f_i^{\inf}\right) + B_i\|\nabla f_i(x)\|^2 + C_i, \label{ch3:eq:general_second_mom_upp_bound_abc}
	\end{equation}
	holds for all $x\in \RR^d$, 
	where\footnote{When $A_i = 0$ one can ignore the first term in the right-hand side of \eqref{ch3:eq:general_second_mom_upp_bound_abc}, i.e., assumption $\inf_{x\in\R^d}f_i(x) > -\infty$ is not required in this case.} $f_i^{\inf} = \inf_{x\in\R^d}f_i(x) > -\infty$.
\end{assumption}

\begin{assumption}[Unbiased assumption for stochastic gradient estimators]\label{ch3:as:general_as_unbiased_for_stoch_gradients_abc}
	We assume that for all $i \in [n]$ there following holds for all $x\in \RR^d$:
	\begin{equation*}
		\ExpBr{\hat{g_i} (x)} = \nabla f_{{i}}(x) .
	\end{equation*}
\end{assumption}

We study \algname{EF21-W-SGD} under Assumption~\ref{ch3:as:general_as_for_stoch_gradients_abc} and Assumption~\ref{ch3:as:general_as_unbiased_for_stoch_gradients_abc}.To the best of our knowledge, this Assumption~\ref{ch3:as:general_as_for_stoch_gradients_abc}, which was originally presented as Assumption 2 by \citet{khaled2020better}, is the most general assumption for a stochastic gradient estimator in a non-convex setting. For a detailed explanation of the generality of this assumption see Figure 1 of \citet{khaled2020better}.

\subsection{A lemma}

The contraction Lemma in this case gets the following form:
\begin{lemma} 
	Let $\cC_i^t \in \mathbb{C}(\alpha)$ for all $i\in [n]$ and $t\geq 0$. Define
	$$G_i^t \eqdef  \sqnorm{ g_i^t - \dfrac{\nabla f_i(x^{t})}{n w_i} } , \qquad G^t \eqdef \sumin w_i G_i^t.$$ Let Assumptions~\ref{ch3:as:L_i}, \ref{ch3:as:general_as_for_stoch_gradients_abc}, \ref{ch3:as:general_as_unbiased_for_stoch_gradients_abc} hold. Then, for any $s >0, \nu >0$ during execution of the Algorithm~\ref{ch3:alg:weighted_ef21_sgd_abc} the following holds:
	\begin{eqnarray}
		\label{ch3:eq:weighted_ef21_sgd_grad_full_contraction_abc}
		\ExpBr{G^{t+1}} &\leq& (1-\hat{\theta}) \ExpBr{G^t} \notag \\
		&& + \hat{\beta_1} \LAMsq  \ExpBr{\sqnorm{ x^{t+1} - x^t}}  + { \widetilde{A} {\hat{\beta_2}}} \ExpBr{f(x^{t+1}) - f^{\inf}}  \notag \\
		&& +  {\widetilde{C} {\hat{\beta_2}}},
	\end{eqnarray}
	where
	\begin{eqnarray*}
		w_i &\eqdef & \dfrac{L_i}{\sum_j L_j}, \\
		\hat{\theta} & \eqdef & 1 - \rb{1-\alpha} (1+s) (1+\nu) \\
		\hat{\beta_1} &\eqdef  & (1- \alpha ) \left(1+ s \right)\left(s+\nu^{-1}\right), \\
		\hat{\beta_2}  & \eqdef  & (1 - \alpha) (1 + s) + (1 + s^{-1}),\\
		\widetilde{A} & \eqdef &  \max_{i=1,\ldots,n} \left( {2(A_i+L_i(B_i-1))}  \dfrac{1}{n w_i} \right), \\  
		\widetilde{C}  &  \eqdef  & \max_{i=1,\ldots,n} \left( {C_i}  \dfrac{1}{n w_i} \right).
	\end{eqnarray*} 
\end{lemma}

\begin{proof}
	Define  $W^t \eqdef \{g_1^t, \dots,    g_n^t, x^t, x^{t+1}\}$. The proof starts as follows:
	\begin{eqnarray*}
		\ExpBr{ G_i^{t+1} \;|\; W^t} &\overset{\eqref{ch3:eq:weighted_def_grad_distortion}}{=}& \ExpBr{  \sqnorm{g_i^{t+1} 
				- \dfrac{\nabla f_i(x^{t+1})}{n w_i}}  \;|\; W^t}	 \\
		&\overset{\text{line}~\ref{ch3:line:weighted_ef21_sgd_grad_update_abc}}{=}& \ExpBr{  \sqnorm{g_i^t + \cC_i^t \left( \dfrac{\hat{g_i}(x^{t+1})}{n w_i} - g_i^t \right) 
				- \dfrac{\nabla f_i(x^{t+1})}{n w_i}}  \;|\; W^t}	 \\
		&=& \ExpBrBig{  \sqnorm{\cC_i^t \left( \dfrac{\hat{g_i}(x^{t+1})}{n w_i} - g_i^t \right) - \left(\dfrac{\hat{g_i}(x^{t+1})}{n w_i} - g_i^t\right) + \dfrac{\hat{g_i}(x^{t+1})}{n w_i} \right. \\
		&& \left. \qquad - \dfrac{\nabla f_i(x^{t+1})}{n w_i}}  \;|\; W^t}	 \\
		&\overset{\eqref{ch3:eq:young}}{\leq}& (1+s) \ExpBr{ \sqnorm{\cC_i^t \left( \dfrac{\hat{g_i}(x^{t+1})}{n w_i} - g_i^t \right) - \left(\dfrac{\hat{g_i}(x^{t+1})}{n w_i} - g_i^t\right)}\;|\; W^t}  \\ 
		& & \qquad + (1+s^{-1})  \ExpBr{\sqnorm{\dfrac{\hat{g_i}(x^{t+1})}{n w_i}
				- \dfrac{\nabla f_i(x^{t+1})}{n w_i}}\;|\; W^t} \\
		&\overset{\eqref{ch3:eq:compressor_contraction}}{\leq}&  (1-\alpha) (1+s) \ExpBrBig{\sqnorm{ \left( \dfrac{\hat{g_i}(x^{t+1})}{n w_i} - \dfrac{\nabla f_i(x^{t+1})}{n w_i} \right) \right.\\
				&& \qquad \qquad \qquad \qquad \left. + \left( \dfrac{\nabla f_i(x^{t+1})}{n w_i} - g_i^t\right) } \;|\; W^t}  \\ 
		& & \qquad + (1+s^{-1})  \ExpBr{\sqnorm{\dfrac{\hat{g_i}(x^{t+1})}{n w_i}
				- \dfrac{\nabla f_i(x^{t+1})}{n w_i}}\;|\; W^t}\\
		&=& (1 - \alpha) (1 + s) \ExpBr{\sqnorm{{g_i}^{t} - \dfrac{\nabla f_i(x^{t+1})}{n w_i}}|\; W^t} \\
		& & \qquad + (1 - \alpha) (1 + s) \ExpBr{\sqnorm{\dfrac{\nabla f_i(x^{t+1})}{n w_i} - \dfrac{\hat{g_i}(x^{t+1})}{n w_i}}\;|\; W^t} \\
		& & \qquad + (1+s^{-1}) \ExpBr{\sqnorm{\dfrac{\hat{g_i}(x^{t+1})}{n w_i}
				- \dfrac{\nabla f_i(x^{t+1})}{n w_i}}\;|\; W^t} \\
		&=& (1 - \alpha) (1 + s) \ExpBrBig{\sqnorm{{g_i}^{t} - \dfrac{\nabla f_i(x^{t})}{n w_i} \right. \\
				&& \left. \qquad \qquad \qquad \qquad +
				\dfrac{\nabla f_i(x^{t})}{n w_i} - \dfrac{\nabla f_i(x^{t+1})}{n w_i}}|\; W^t} \\
		& & \qquad + (1 - \alpha) (1 + s) \ExpBr{\sqnorm{\dfrac{\nabla f_i(x^{t+1})}{n w_i} - \dfrac{\hat{g_i}(x^{t+1})}{n w_i}}\;|\; W^t} \\
		& & \qquad + (1+s^{-1}) \ExpBr{\sqnorm{\dfrac{\hat{g_i}(x^{t+1})}{n w_i}
				- \dfrac{\nabla f_i(x^{t+1})}{n w_i}}\;|\; W^t} .			
	\end{eqnarray*}				
	
	Further, we continue as follows				
	\begin{eqnarray*}
		\ExpBr{ G_i^{t+1} \;|\; W^t} 							
		&\overset{\eqref{ch3:eq:young}}{\leq}& (1 - \alpha) (1 + s) (1+\nu) \ExpBr{\sqnorm{ {g_i}^{t} - \dfrac{\nabla f_i(x^t)}{n w_i}}\;|\; W^t} \\
		&& + (1 - \alpha) (1 + s) (1+\nu^{-1}) {\sqnorm{\dfrac{\nabla f_i(x^{t+1})}{n w_i} - \dfrac{\nabla f_i(x^{t})}{n w_i}}} \\
		&&  + {\left((1+s^{-1})+ (1-\alpha)(1 + s) \right)} \ExpBr{\sqnorm{\dfrac{\hat{g_i}(x^{t+1})}{n w_i}
				- \dfrac{\nabla f_i(x^{t+1})}{n w_i}}\;|\; W^t}.
	\end{eqnarray*}
	
	To further bound the last term, which contains multiple ${(1+s^{-1})}$ factors, we leverage the property that $\hat g_i(x^{t+1})$ is a random variable serving as an unbiased estimator of $\nabla f_i(x^{t+1})$. Our approach is as follows:
	\begin{eqnarray*}
		\ExpBr{ G_i^{t+1} \;|\; W^t} &\leq& (1-\hat{\theta}) \ExpBr{G_i^t \;|\; W^t} + \hat{\beta_1} \dfrac{1}{n^2 w_i^2} \sqnorm{\nabla f_i(x^{t+1}) - \nabla f_i(x^t)} \\
		& & \qquad + \dfrac{{\hat{\beta_2}}}{(n w_i)^2} \ExpBr{\sqnorm{ \hat{g_i} (x^{t+1}) - {\nabla f_i}(x^{t+1})}\;|\; W^t}.
	\end{eqnarray*}
	
	Now due to the requirement of unbiasedness of gradient estimators expressed in the form of Assumption~\ref{ch3:as:general_as_unbiased_for_stoch_gradients_abc} we have the following:
	\begin{eqnarray}
		\label{ch3:eq:var-decompoposition-abc}
		\ExpBr{\sqnorm{ \hat{g_i} (x^{t+1}) - {\nabla f_i}(x^{t+1})}\;|\; W^t} &=& \ExpBr{\sqnorm{ \hat{g_i} (x^{t+1})}\;|\; W^t} \notag \\
		& & \qquad- {\sqnorm{ \nabla f_{{i}} (x^{t+1})}}
	\end{eqnarray}
	
	Using this variance decomposition, we can proceed as follows.
	
	\begin{eqnarray*}
		\ExpBr{ G_i^{t+1} \;|\; W^t} &\overset{\eqref{ch3:eq:var-decompoposition-abc}}{\leq}& (1-\hat{\theta}) \ExpBr{G_i^t \;|\; W^t} + \hat{\beta_1} \dfrac{1}{n^2 w_i^2} \sqnorm{\nabla f_i(x^{t+1}) - \nabla f_i(x^t)} \\
		& & \qquad + \dfrac{{\hat{\beta_2}}}{(n w_i)^2 } \left( \ExpBr{\sqnorm{ \hat{g_i} (x^{t+1})}\;|\; W^t} - {\sqnorm{ \nabla f_{i}(x^{t+1})}} \right) \\
		&\overset{\eqref{ch3:eq:general_second_mom_upp_bound_abc}}{\leq}&	(1-\hat{\theta}) \ExpBr{G_i^t \;|\; W^t} + \hat{\beta_1} \dfrac{1}{n^2 w_i^2} \sqnorm{\nabla f_i(x^{t+1}) - \nabla f_i(x^t)} \\
		& &  + \dfrac{{\hat{\beta_2}}}{(n w_i)^2} \Big(
		2A_i\left(f_i(x^{t+1}) - f_i^{\inf}\right) \\
		&& \qquad \qquad + B_i\|\nabla f_i(x^{t+1})\|^2 \\
		&& \qquad \qquad + C_i - \sqnorm{\nabla f_i(x^{t+1}}) \Big) \\
		&=& (1-\hat{\theta}) \ExpBr{G_i^t \;|\; W^t} + \hat{\beta_1} \dfrac{1}{n^2 w_i^2} \sqnorm{\nabla f_i(x^{t+1}) - {\nabla f_i(x^t)}} \\
		& & \qquad + \dfrac{ 2A_i {\hat{\beta_2}}}{(n w_i)^2} \left(f_i(x^{t+1}) - f_i^{\inf}\right) \\
		& & \qquad + \dfrac{2(B_i-1){\hat{\beta_2}}}{(n w_i)^2} \left(\dfrac{1}{2} \|\nabla f_i(x^{t+1})\|^2 \right)
		+  \dfrac{ C_i {\hat{\beta_2}}}{(n w_i)^2} \\
		&\le& (1-\hat{\theta}) \ExpBr{G_i^t \;|\; W^t} + \hat{\beta_1} \dfrac{1}{n^2 w_i^2} \sqnorm{\nabla f_i(x^{t+1}) - {\nabla f_i(x^t)}} \\
		& & \qquad + \dfrac{ 2A_i {\hat{\beta_2}}}{(n w_i)^2} \left(f_i(x^{t+1}) - f_i^{\inf}\right) \\
		& & \qquad + \dfrac{2(B_i-1){\hat{\beta_2}}}{(n w_i)^2} L_i \left(f_i(x^{t+1}) - f_i^{\inf} \right)
		+  \dfrac{C_i {\hat{\beta_2}}}{(n w_i)^2} \\
		&=& (1-\hat{\theta}) \ExpBr{G_i^t \;|\; W^t} + \hat{\beta_1} \dfrac{1}{n^2 w_i^2} \sqnorm{\nabla f_i(x^{t+1}) - {\nabla f_i(x^t)}} \\
		& & \qquad + \dfrac{ 2(A_i + L_i(B_i - 1)) {\hat{\beta_2}}}{(n w_i)^2} \left(f_i(x^{t+1}) - f_i^{\inf}\right)
		+  \dfrac{C_i {\hat{\beta_2}}}{(n w_i)^2 }.
	\end{eqnarray*}
	
	Next leveraging Assumption~\ref{ch3:as:L_i} we replace the second term in the last expression, and we can derive the subsequent bound:
	\begin{eqnarray*}
		\ExpBr{ G_i^{t+1} \;|\; W^t} &\overset{\eqref{ch3:eq:L_i}}{\leq}& (1-\hat{\theta})   G_i^t +  \dfrac{\hat{\beta_1} L_i^2}{n^2 w_i^2} \sqnorm{ x^{t+1} - x^t}  \\
		&& \qquad 
		+ \dfrac{ 2(A_i + L_i(B_i - 1)) {\hat{\beta_2}}}{(n w_i)^2} \left(f_i(x^{t+1}) - f_i^{\inf}\right)
		+  \dfrac{C_i {\hat{\beta_2}}}{(n w_i)^2}.
	\end{eqnarray*}
	Applying the tower property and subsequently taking the expectation, we obtain:
	\begin{equation}\label{ch3:eq:weighted_ef21_sgd_aux_1_abc}
		\begin{aligned}
			\ExpBr{ G_i^{t+1}} &\leq (1-\hat{\theta}) \ExpBr{G_i^t} + \hat{\beta_1} \dfrac{1}{n^2 w_i^2} L_i^2 \ExpBr{\sqnorm{ x^{t+1} - x^t}} 
			\\
			& \qquad + \dfrac{ 2(A_i + L_i(B_i - 1)) {\hat{\beta_2}}}{(n w_i)^2} \ExpBr{f_i(x^{t+1}) - f_i^{\inf}}
			+  \dfrac{C_i {\hat{\beta_2}}}{(n w_i)^2}.
		\end{aligned}
	\end{equation}
	
	Next for the expectation of the main quantity of our interest $G^{t+1}$, we derive the subsequent bound:
	\begin{eqnarray*}
		\ExpBr{G^{t+1}} &= & \ExpBr{\sumin w_i G_i^{t+1}} \\
		& = &  \sumin w_i \ExpBr{G_i^{t+1}} \\
		&\overset{\eqref{ch3:eq:weighted_ef21_sgd_aux_1_abc}}{\leq}  & (1-\hat{\theta}) \sumin w_i \ExpBr{G_i^t} + \sumin w_i \hat{\beta_1} \dfrac{1}{n^2 w_i^2} L_i^2 \cdot \ExpBr{\sqnorm{ x^{t+1} - x^t}} \\
		& &  \qquad + \sumin w_i \dfrac{ 2(A_i + L_i(B_i - 1)) {\hat{\beta_2}}}{(n w_i)^2} \cdot \ExpBr{f_i(x^{t+1}) - f_i^{\inf}} \\
		& & \qquad + \sumin w_i \dfrac{C_i {\hat{\beta_2}}}{(n w_i)^2}	\\
		&=  & (1-\hat{\theta}) \ExpBr{G^t} + \sumin \hat{\beta_1} \dfrac{1}{n^2 w_i} L_i^2 \cdot \ExpBr{\sqnorm{ x^{t+1} - x^t}} \\
		& &  \qquad + \sumin \dfrac{ 2(A_i + L_i(B_i - 1)) {\hat{\beta_2}}}{(n)^2 w_i} \cdot \ExpBr{f_i(x^{t+1}) - f_i^{\inf}} \\
		& & \qquad + \sumin \dfrac{C_i {\hat{\beta_2}}}{n^2 w_i}
	\end{eqnarray*}
	
	
	Employing quantities $\tilde{A}$ and $\tilde{C}$, the final bound can be reformulated as follows:
	\begin{eqnarray}
		\ExpBr{G^{t+1}} \leq (1-\hat{\theta}) \ExpBr{G^t} + \sumin \hat{\beta_1} \dfrac{1}{n^2 w_i} L_i^2 \cdot \ExpBr{\sqnorm{ x^{t+1} - x^t}} \notag \\
		+ \dfrac{1}{n} \sumin { \widetilde{A} {\hat{\beta_2}}} \cdot \ExpBr{f_i(x^{t+1}) - f_i^{\inf}}
		+   {\widetilde{C} {\hat{\beta_2}}} \notag \\
		\leq (1-\hat{\theta}) \ExpBr{G^t} + \sumin \hat{\beta_1} \dfrac{1}{n^2 w_i} L_i^2 \cdot \ExpBr{\sqnorm{ x^{t+1} - x^t}} \notag \\
		+ \dfrac{1}{n} \sumin { \widetilde{A} {\hat{\beta_2}}} \cdot \ExpBr{f_i(x^{t+1}) - f^{\inf}}
		+   {\widetilde{C} {\hat{\beta_2}}} \notag \\
		\le (1-\hat{\theta}) \ExpBr{G^t} + \sumin \hat{\beta_1} \dfrac{1}{n^2 w_i} L_i^2 \cdot \ExpBr{\sqnorm{ x^{t+1} - x^t}} \notag \\
		+ { \widetilde{A} {\hat{\beta_2}}} \notag \ExpBr{f(x^{t+1}) - f^{\inf}}
		+  {\widetilde{C} {\hat{\beta_2}}}. \notag
	\end{eqnarray}

	
	Given that $w_i=\dfrac{L_i}{\sum_j L_j}$, we have:
	\begin{equation}
		\begin{aligned}
			\ExpBr{G^{t+1}} &\leq (1-\hat{\theta}) \ExpBr{G^t} + \dfrac{1}{n}\sumin \hat{\beta_1} \dfrac{\sum_j L_j}{n} L_i \ExpBr{\sqnorm{ x^{t+1} - x^t}} \notag \\
			& \qquad + { \widetilde{A} {\hat{\beta_2}}} \ExpBr{f(x^{t+1}) - f^{\inf}}
			+  {\widetilde{C} {\hat{\beta_2}}} \notag \\
			& = (1-\hat{\theta}) \ExpBr{G^t} + \hat{\beta_1} \left(\avein L_i\right)^2 \cdot \ExpBr{\sqnorm{ x^{t+1} - x^t}} \notag \\
			& \qquad + { \widetilde{A} {\hat{\beta_2}}} \ExpBr{f(x^{t+1}) - f^{\inf}}
			+  {\widetilde{C} {\hat{\beta_2}}},
		\end{aligned}
	\end{equation}
	what completes the proof.
\end{proof}

\subsection{Main result}

Now we are ready to prove the main convergence theorem.
\begin{theorem} Let $\cC_i^t \in\mathbb{C}(\alpha)$ for all $\in [n]$ and $t\geq 0$ in Algorithm~\ref{ch3:alg:weighted_ef21_sgd_abc}. set the following quantities:
	\begin{eqnarray*}
		\hat{\theta} &\eqdef & 1 - \rb{1-\alpha} (1+s) (1+\nu),\\
		\hat{\beta_1} &\eqdef  & (1- \alpha ) \left(1+ s \right)\left(s+\nu^{-1}\right),\\
		\hat{\beta_2} &\eqdef  & (1 - \alpha) (1 + s) + (1 + s^{-1}),\\
		w_i &\eqdef  & \dfrac{L_i}{\sum_{j=1}^n L_j},\\		
		\widetilde{A} & \eqdef  & \max_{i=1,\ldots,n}  \dfrac{2(A_i+L_i(B_i-1))}{n w_i}, \\
		\widetilde{C} & \eqdef  & \max_{i=1,\ldots,n}  \dfrac{C_i}{n w_i} .
	\end{eqnarray*}
	Under Assumptions~\ref{ch3:as:smooth}, \ref{ch3:as:L_i}, \ref{ch3:as:general_as_for_stoch_gradients_abc}, \ref{ch3:as:general_as_unbiased_for_stoch_gradients_abc}, and  selection of $s >0, \nu >0$ small enough such that
	$(1+s)(1+\nu) < \dfrac{1}{1-\alpha}$ holds, set the step size in the following way:
	\begin{equation}
		\gamma \leq \dfrac{1}{L + \LAM \sqrt{\dfrac{\hat{\beta}_1}{\hat{\theta}}}}.
	\end{equation}
	Choose an iterate $\hat{x}^T$ from $\{x^0, x^1, \dots, x^{T-1}\}$ with probability 
	\begin{equation} 
		\label{ch3:eq:09u09fd-0ff-abc}
		\Prob(\hat{x}^T = x^t) = \dfrac{v_t}{V_T},
	\end{equation} 
	where
	$$ v_t \eqdef \left(1 - \dfrac{\gamma \tilde{A} \tilde{\beta}_2}{2\theta}\right)^t; \qquad V_T \eqdef \sum\limits_{t=0}^{T-1} v_t.
	$$
	Then,
	\begin{equation}
		\ExpBr{\sqnorm{\nabla f(\hat{x}^{T})}} \leq \dfrac{2 (f(x^0) - f^\text{inf})}{\gamma T  \left(1 - \dfrac{\gamma \widetilde{A} \hat{\beta_2}}{2 \theta}\right)^T } + \dfrac{G^0}{ \hat{\theta}T  \left(1 - \dfrac{\gamma \widetilde{A} \hat{\beta_2}}{2 \theta}\right)^T} + \dfrac{\widetilde{C}\beta_2}{\hat{\theta}},
	\end{equation}
	where $$G^0 \eqdef \sumin w_i \norm{g_i^0 - \dfrac{1}{nw_i}\nabla f_i(x^0)}^2.$$
\end{theorem}
\begin{proof}
	In the derivation below, we  use Lemma~\ref{ch3:lm:descent_lemma} for 
	\begin{equation}
		\label{ch3:eq:sgd_weighted_grad_estimate_def_abc}
		g^t=\sum\limits_{i=1}^{n} w_i g_i^t.
	\end{equation}
	We start as follows:
	\begin{eqnarray*}
		f(x^{t+1}) &\overset{\eqref{ch3:eq:descent_lemma}}{\leq} & 
		f(x^{t})-\dfrac{\gamma}{2}\sqnorm{\nabla f(x^{t})}-\left(\dfrac{1}{2 \gamma}-\dfrac{L}{2}\right)\sqnorm{x^{t+1}-x^{t}} \\
		&& \qquad +\dfrac{\gamma}{2}\sqnorm{
			g^t- \dfrac{1}{n} \sumin \nabla f_i(x^{t}) } \\
		& \overset{\eqref{ch3:eq:sgd_weighted_grad_estimate_def}}{=} & 
		f(x^{t})-\dfrac{\gamma}{2}\sqnorm{\nabla f(x^{t})}-\left(\dfrac{1}{2 \gamma}-\dfrac{L} 
		{2}\right)\sqnorm{x^{t+1}-x^{t}} \\
		&& \qquad +\dfrac{\gamma}{2}\sqnorm{
			\sumin w_i \left(g_i^t-  \dfrac{\nabla f_i(x^{t})}{n w_i} \right) }  \\
		& \leq & 
		f(x^{t})-\dfrac{\gamma}{2}\sqnorm{\nabla f(x^{t})}-\left(\dfrac{1}{2 \gamma}-\dfrac{L}       
		{2}\right)\sqnorm{x^{t+1}-x^{t}} \\
		&& \qquad +\dfrac{\gamma}{2}
		\sumin w_i \sqnorm{ g_i^t-  \dfrac{\nabla f_i(x^{t})}{n w_i}  } \\
		& = & f(x^{t})-\dfrac{\gamma}{2}\sqnorm{\nabla f(x^{t})}-\left(\dfrac{1}{2 \gamma}-\dfrac{L}       
		{2}\right)\sqnorm{x^{t+1}-x^{t}} \\
		&& \qquad +\dfrac{\gamma}{2} G^t.
	\end{eqnarray*}
	
	Subtracting $f^\ast$ from both sides and taking expectation, we get
	\begin{eqnarray*}
		\ExpBr{f(x^{t+1})-f^\ast} &\leq& \ExpBr{f(x^{t})-f^\ast}
		-\dfrac{\gamma}{2} \ExpBr{\sqnorm{\nabla f(x^{t})}} \\
		&& -\left(\dfrac{1}{2 \gamma} - \dfrac{L}{2}\right) \ExpBr{\sqnorm{x^{t+1}-x^{t}}} + \dfrac{\gamma}{2}\ExpBr{G^t}.
	\end{eqnarray*}

	Let
	\begin{eqnarray*}
		\delta^{t} &\eqdef& \ExpBr{f(x^{t})-f^\ast},\\
		s^{t} &\eqdef& \ExpBr{G^t },\\ 
		r^{t} &\eqdef& \ExpBr{\sqnorm{x^{t+1}-x^{t}}}.
	\end{eqnarray*}
	
	Then by adding $\dfrac{\gamma}{2\theta} s^{t+1}$ and employing inequality \eqref{ch3:eq:weighted_ef21_sgd_grad_full_contraction}, we obtain:
	\begin{align*}
		\delta^{t+1}+\dfrac{\gamma}{2 \hat{\theta}} s^{t+1} 
		&\leq \delta^{t}-\dfrac{\gamma}{2}\ExpBr{\sqnorm{\nabla f(x^{t})}} -          \left(\dfrac{1}{2 \gamma}-\dfrac{L}{2}\right) r^{t}+\dfrac{\gamma}{2} s^{t} \\
		& \qquad +\dfrac{\gamma}{2 \hat{\theta}}\left( \hat{\beta_1} \LAMsq  r^t + (1-\hat{\theta}) s^t + { \widetilde{A} {\hat{\beta_2}}} \delta^{t+1} +  {\widetilde{C} {\hat{\beta_2}}} \right) \\
		&=\delta^{t}+\dfrac{\gamma}{2\hat{\theta}} s^{t}-\dfrac{\gamma}{2}\ExpBr{\sqnorm{\nabla f(x^{t})}}  -\left(\dfrac{1}{2\gamma} -\dfrac{L}{2} -  \dfrac{\gamma}{2\hat{\theta}} \hat{\beta_1}  \LAMsq  \right) r^{t} \\
		& \qquad + \dfrac{\gamma \widetilde{A} \beta_2}{2\hat{\theta}} \delta^{t+1} + \dfrac{\gamma \widetilde{C}}{2 \hat{\theta}} \beta_2\\
		& \leq \delta^{t}+\dfrac{\gamma}{2\hat{\theta}} s^{t} -\dfrac{\gamma}{2}\ExpBr{\sqnorm{\nabla f(x^{t})}} + \dfrac{\gamma \widetilde{A} \beta_2}{2\hat{\theta}} \delta^{t+1} + \dfrac{\gamma \widetilde{C}}{2 \hat{\theta}} \beta_2.
	\end{align*}
	
	
	The last inequality follows from the bound $$\gamma^2\dfrac{\hat{\beta_1} \LAMsq }{\hat{\theta}} + L\gamma \leq 1,$$ which holds due to Lemma~\ref{ch3:lm:stepsize_bound} for $$\gamma \leq \dfrac{1}{L +  \LAM \sqrt{\dfrac{\hat{\beta}_1}{\hat{\theta}}}}.$$ Subsequently, we will reconfigure the final inequality and perform algebraic manipulations, taking into account that $\dfrac{2}{\gamma} > 0$. In the final step of these algebraic transformations, we will leverage the fact that $s^t \ge 0$:
	\begin{eqnarray*}
		\delta^{t+1}+\dfrac{\gamma}{2 \hat{\theta}} s^{t+1} &\leq& \delta^{t}+\dfrac{\gamma}{2\hat{\theta}} s^{t} -\dfrac{\gamma}{2}\ExpBr{\sqnorm{\nabla f(x^{t})}} + \dfrac{\gamma \widetilde{A} \beta_2}{2\hat{\theta}} \delta^{t+1} + \dfrac{\gamma \widetilde{C}}{2 \hat{\theta}} \beta_2.
	\end{eqnarray*}		
	
	Therefore,		
	\begin{eqnarray*}		
		\dfrac{2}{\gamma} \delta^{t+1}+\dfrac{2}{\gamma} \dfrac{\gamma}{2 \hat{\theta}} s^{t+1} &\leq& \dfrac{2}{\gamma} \delta^{t} + \dfrac{2}{\gamma} \dfrac{\gamma}{2\hat{\theta}} s^{t} -\ExpBr{\sqnorm{\nabla f(x^{t})}} + \dfrac{2}{\gamma} \dfrac{\gamma \widetilde{A} \beta_2}{2\hat{\theta}} \delta^{t+1} + \dfrac{2}{\gamma} \dfrac{\gamma \widetilde{C}}{2 \hat{\theta}} \beta_2 .
	\end{eqnarray*}				
	
	Further,	
	\begin{eqnarray*}	
		\ExpBr{\sqnorm{\nabla f(x^{t})}} &\leq& -\dfrac{2}{\gamma} \delta^{t+1} - \dfrac{2}{\gamma} \dfrac{\gamma}{2 \hat{\theta}} s^{t+1} + \dfrac{2}{\gamma} \delta^{t} + \dfrac{2}{\gamma} \dfrac{\gamma}{2\hat{\theta}} s^{t} + \dfrac{2}{\gamma} \dfrac{\gamma \widetilde{A} \beta_2}{2\hat{\theta}} \delta^{t+1} + \dfrac{2}{\gamma} \dfrac{\gamma \widetilde{C}}{2 \hat{\theta}} \beta_2  \\
		&\leq&
		-\dfrac{2}{\gamma} \delta^{t+1} - \dfrac{2}{\gamma} \dfrac{\gamma}{2\hat{\theta}} s^{t+1} + \dfrac{2}{\gamma} \left( \delta^{t} + \dfrac{\gamma}{2\hat{\theta}} s^{t} \right) + \dfrac{2}{\gamma} \dfrac{\gamma \widetilde{A} \beta_2}{2 \hat{\theta}} \delta^{t+1} + \dfrac{\widetilde{C}\beta_2}{ \hat{\theta}}  \\
		&\leq&
		\dfrac{2}{\gamma} \left( \left( \delta^{t} + \dfrac{\gamma}{2\hat{\theta}} s^{t} \right) -1\left(1 - \dfrac{\gamma \widetilde{A} \beta_2}{2 \hat{\theta}} \right) \delta^{t+1} -\left(\dfrac{\gamma}{2\hat{\theta}} s^{t+1} \right) \right) + \dfrac{\widetilde{C}\beta_2}{ \hat{\theta}} \\
		&\leq& \dfrac{2}{\gamma} \left( \left( \delta^{t} + \dfrac{\gamma}{2\hat{\theta}} s^{t} \right) -\left(1 - \dfrac{\gamma \widetilde{A} \beta_2}{2 \hat{\theta}} \right) \left(\delta^{t+1} + \dfrac{\gamma}{2\hat{\theta}} s^{t+1} \right) \right) + \dfrac{\widetilde{C}\beta_2}{ \hat{\theta}}.
	\end{eqnarray*}
	
	We sum up inequalities above with weights $v_t/V_T$, where 
	$$v_t \eqdef (1 - \dfrac{\gamma \widetilde{A} \hat{\beta_2}}{2 \theta})^t, \qquad V_T \eqdef \sum_{i=1}^{T} v_i.$$
	\begin{eqnarray*}
		\ExpBr{\sqnorm{\nabla f(\hat{x}^{T})}} &=& \sum_{t=0}^{T} \dfrac{v_t}{V_T} \ExpBr{\sqnorm{\nabla f(x^{t})}} \\
		&=& \dfrac{1}{V_T} \sum_{t=0}^{T} v_t \ExpBr{\sqnorm{\nabla f(x^{t})}} \\
		&\leq& \dfrac{1}{V_T} \sum_{t=0}^{T} v_t \Big(\dfrac{2}{\gamma} \left( \left( \delta^{t} + \dfrac{\gamma}{2\hat{\theta}} s^{t} \right) -\left(1 - \dfrac{\gamma \widetilde{A} \beta_2}{2 \hat{\theta}} \right) \left(\delta^{t+1} + \dfrac{\gamma}{2\hat{\theta}} s^{t+1} \right) \right) \\
		&& \qquad \qquad + \dfrac{\widetilde{C}\beta_2}{ \hat{\theta}} \Big) \\
		&=& \dfrac{2}{\gamma V_T} \sum_{t=0}^{T} w_t \left( \left( \delta^{t} + \dfrac{\gamma}{2\hat{\theta}} s^{t} \right) -\left(1 - \dfrac{\gamma \widetilde{A} \beta_2}{2 \hat{\theta}} \right) \left(\delta^{t+1} + \dfrac{\gamma}{2\hat{\theta}} s^{t+1} \right) \right)	\\
		&& \qquad + \sum_{t=0}^{T} \dfrac{w_t}{W_T} \cdot \dfrac{\widetilde{C}\beta_2}{ \hat{\theta}} \\
		&=& \dfrac{2}{\gamma V_T} \sum_{t=0}^{T} w_t \left( \left( \delta^{t} + \dfrac{\gamma}{2\hat{\theta}} s^{t} \right) -\left(1 - \dfrac{\gamma \widetilde{A} \beta_2}{2 \hat{\theta}} \right) \left(\delta^{t+1} + \dfrac{\gamma}{2\hat{\theta}} s^{t+1} \right) \right)	\\
		&& \qquad + \dfrac{\widetilde{C}\beta_2}{ \hat{\theta}} \\
		&=& \dfrac{2}{\gamma V_T} \sum_{t=0}^{T} \left(w_t \left( \delta^{t} + \dfrac{\gamma}{2\hat{\theta}} s^{t} \right) -w_{t+1} \left(\delta^{t+1} + \dfrac{\gamma}{2\hat{\theta}} s^{t+1} \right) \right) \\ 
		&& \qquad +\dfrac{\widetilde{C}\beta_2}{ \hat{\theta}} \\
		&\leq& \dfrac{2 \delta^0}{\gamma V_T} + \dfrac{s^0}{ \hat{\theta}V_T} + \dfrac{\widetilde{C}\beta_2}{\hat{\theta}}.
	\end{eqnarray*}
	Finally, we notice that $$V_T = \sum\limits_{t=1}^T (1 - \dfrac{\gamma \widetilde{A} \hat{\beta_2}}{2 \theta})^t \geq T \cdot (1 - \dfrac{\gamma \widetilde{A} \hat{\beta_2}}{2 \theta})^T,$$ what concludes the proof.
\end{proof}

\clearpage
\addtocounter{adjsection}{1}
\section{EF21-W-PP: Weighted Error Feedback 2021 with Partial Participation} \label{ch3:sec:EF21-W-PP}

In this section, we present another extension of error feedback. Again, to maintain brevity, we show our results for~\algname{EF21-W}, however, we believe getting an enhanced rate for standard \algname{EF21} should be straightforward. 

\subsection{Algorithm}
Building upon the delineation of \algname{EF21-W} in Algorithm~\ref{ch3:alg:EF21-W}, we turn our attention to its partial participation variant, \algname{EF21-W-PP}, and seek to highlight the primary distinctions between them. One salient difference is the introduction of a distribution, denoted as $\cD$, across the clients. For clarity, consider the power set ${\cal P}$ of the set $[n] \eqdef \{1, 2, \dots, n\}$, representing all possible subsets of $[n]$. Then, the distribution ${\cal D}$ serves as a discrete distribution over ${\cal P}$.

While \algname{EF21-W-PP} runs, at the start of each communication round $t$, the master, having computed a descent step as $x^{t+1} = x^t - \gamma g^t$, samples a client subset $S^t$ from the distribution $\cD$. Contrasting with Algorithm~\ref{ch3:alg:EF21-W} where the new iteration $x^{t+1}$ is sent to all clients, in this variant, it is sent exclusively to those in $S^t$.

Any client $i \in S^t$ adheres to procedures akin to \algname{EF21-W}: it compresses the quantity $\dfrac{1}{nw_i} \nabla f_i(x^t) - g_i^t$ and transmits this to the master. Conversely, client $j$ omitted in $S^t$, i.e., $j \notin S^t$, is excluded from the training for that iteration. Concluding the round, the master updates $g^{t+1}$ by integrating the averaged compressed variances received from clients in the set $S^t$.

\begin{algorithm}
	\begin{algorithmic}[1]
		\STATE {\bfseries Input:} initial model parameters $x^0 \in \RR^d$; initial gradient estimates $g_1^0, g_2^0, \dots,g_n^0 \in \R^d$ stored at the clients; weights ${\color{ForestGreen}w_i} = \nicefrac{L_i}{\sum_j L_j}$; step size $\gamma>0$; number of iterations $T > 0$; distribution $\cD$ over clients
		\STATE {\bfseries Initialize:} $g^0 = \sum_{i=1}^n {\color{ForestGreen}w_i} g_i^0 $ on the server	
		\FOR{$t = 0, 1, 2, \dots, T - 1 $}
		\STATE Server computes $x^{t+1} = x^t - \gamma g^t$		
		\STATE Server samples a subset $S^t \sim \cD$ of clients
		\STATE Server broadcasts  $x^{t+1}$ to clients in $S^t$
		\FOR{$i = 1, \dots, n$ {\bf on the clients in parallel}} 		
		\IF{$i\in S^t$}		
		\STATE Compute $u_i^t=\cC_i^t (\dfrac{1}{n {\color{ForestGreen}w_i}}\nabla f_i(x^{t+1}) - g_i^t)$ and update $g_i^{t+1} = g_i^t +u_i^t$ \label{ch3:line:weighted_ef21_pp_grad_ind_update_step}	
		\STATE Send the compressed message $u_i^{t}$ to the server	
		\ENDIF
		\IF{$i\notin S^t$}
		\STATE Set $u_i^t = 0$		for the client and the server
		\STATE Do not change local state $g_i^{t+1} = g_i^t$
		\ENDIF
		\ENDFOR
		\STATE Server updates $g_i^{t+1} = g_i^t +u_i^t$ for all $i\in [n]$, and computes $g^{t+1} = \sum_{i=1}^n {\color{ForestGreen}w_i} g_i^{t+1}$		\label{ch3:line:averaging_ef21_pp_weighting} 
		\ENDFOR
		\STATE {\bfseries Output:} Point $\hat{x}^T$ chosen from the set $\{x^0, \dots, x^{T-1}\}$ uniformly at random		
	\end{algorithmic}
	\caption{\algname{EF21-W-PP}: Weighted Error Feedback 2021 with Partial Participation.}
	\label{ch3:alg:weighted_ef21_pp}
\end{algorithm}

Assume $S$ is drawn from the distribution $\cD$. Let us denote 
\begin{equation}\label{ch3:eq:p_i_def}
	p_i \eqdef \Prob(i \in S^t).
\end{equation}
In other words, $p_i$ represents the probability of client $i$ being selected in any iteration.  For given parameters $p_i$ such that $p_i \in (0,1]$ for  $i\in [n]$, we introduce the notations $p_{\min} \eqdef \min_i p_i$ and $p_{\max} \eqdef \max_i p_i$, respectively.

\subsection{A lemma}

Having established the necessary definitions, we can now proceed to formulate the Lemma.
\begin{lemma} \label{ch3:lemma:contraction_ef21_pp_weighted}
	Let $\cC_i^t\in \mathbb{C}(\alpha)$ for all $i\in [n]$ and $t\geq 0$. Let Assumption~\ref{ch3:as:L_i} hold. Define
	\begin{equation}\label{ch3:eq:grad_distortion_def_pp}
		G_i^t \eqdef  \sqnorm{ g_i^t - \dfrac{\nabla f_i(x^{t})}{n w_i} } , \qquad G^t \eqdef \sumin w_i G_i^t.
	\end{equation}
	For any $s>0$ and $\rho>0$, let us define the following quantities: 
	\begin{eqnarray*}
		\theta(\alpha, s) &\eqdef& 1 - (1 - \alpha)(1 + s)\\
		\beta(\alpha, s) &\eqdef& \beta(\alpha, s) = (1 - \alpha)(1+s^{-1})\\
		\theta_p &\eqdef& p_{\min}\rho + \theta(\alpha,s) p_{\max} - \rho - (p_{\max}-p_{\min}) \\
		\tilde{B} &\eqdef& \left({\beta(\alpha,s) p_{\max}} + (1-p_{\min}){(1+\rho^{-1})} \right) \LAMsq.
	\end{eqnarray*}
	Additionally, assume that
	$$
	\dfrac{1 + \rho(1 - p_{\min}) + (p_{\max} - p_{\min}) }{p_{\max}}\geq \theta(\alpha, s) > \dfrac{\rho(1 - p_{\min}) + (p_{\max} - p_{\min}) }{p_{\max}}.
	$$
	Then, we have
	\begin{equation} 
		\ExpBr{G^{t+1}} \le (1-\theta_p) \ExpBr{G^t} +
		\tilde{B} \ExpBr{\sqnorm{x^{t+1} - x^{t}}}.
	\end{equation}
\end{lemma}

\begin{proof}
	Let us define $W^t \eqdef \{g_1^t, \dots,    g_n^t, x^t, x^{t+1}\}$. If client $i$ participates in the training at iteration $t$, then
	
	\begin{equation*}
		\begin{aligned}
			&\ExpBr{ G_i^{t+1} \;|\; W^t, i \in S^t} \\
			&\qquad \overset{\eqref{ch3:eq:grad_distortion_def_pp}}{=} \ExpBr{  \sqnorm{g_i^{t+1} 
					- \dfrac{\nabla f_i(x^{t+1})}{n w_i}}  \;|\; W^t, i \in S^t} \\ 
			&\qquad \overset{\text{line}~\ref{ch3:line:weighted_ef21_pp_grad_ind_update_step}\text{ of Algorithm~\ref{ch3:alg:weighted_ef21_pp}}}{=} \ExpBr{  \sqnorm{g_i^t + \cC_i^t \left( \dfrac{\nabla f_i(x^{t+1})}{n w_i} - g_i^t \right) 
					- \dfrac{\nabla f_i(x^{t+1})}{n w_i}}  \;|\; W^t, i \in S^t} \\ 
			&\qquad \overset{\eqref{ch3:eq:compressor_contraction}}{\leq} (1-\alpha) 
			\sqnorm{\dfrac{\nabla f_i(x^{t+1})}{n w_i} - g_i^t} \\ 
			&\qquad = (1-\alpha) 
			\sqnorm{\dfrac{\nabla f_i(x^{t+1})}{n w_i} - \dfrac{\nabla f_i(x^{t})}{n w_i} + \dfrac{\nabla f_i(x^{t})}{n w_i} - g_i^t} \\ 
			&\qquad \overset{\eqref{ch3:eq:young}}{\leq} (1-\alpha) (1+ s) \sqnorm{ \dfrac{\nabla f_i(x^{t})}{n w_i} - g_i^t} \\
			&\qquad \quad + (1-\alpha)  \left(1+s^{-1}\right) \dfrac{1}{n^2 w_i^2} 
			\sqnorm{\nabla f_i(x^{t+1}) - \nabla f_i(x^t)} \\ 
			&\qquad \overset{\eqref{ch3:eq:L_i}}{\leq} (1-\alpha) (1+ s) \sqnorm{ \dfrac{\nabla f_i(x^{t})}{n w_i} - g_i^t} \\
			&\qquad \quad + (1-\alpha)  \left(1+s^{-1}\right) \dfrac{L_i^2}{n^2 w_i^2} 
			\sqnorm{x^{t+1} - x^t}.
		\end{aligned}
	\end{equation*}
	
	Utilizing the tower property and taking the expectation with respect to $W^t$, we derive:
	\begin{eqnarray} \label{ch3:eq:g_i_for_sampled_ef21_weighted}
		\ExpBr{ G_i^{t+1} \;|\; i \in S^t} \leq (1-\theta(\alpha,s)) \ExpBr{G_i^{t}} + \beta(\alpha,s) \dfrac{L_i^2}{n^2 w_i^2} 
		\ExpBr{\sqnorm{x^{t+1} - x^t}},
	\end{eqnarray}
	where $\theta(\alpha, s) = 1 - (1 - \alpha)(1 + s)$, and $\beta(\alpha, s) = (1 - \alpha)(1+s^{-1})$. We now aim to bound the quantity $\ExpBr{G_i^{t+1} \;|\; i \notin S^t}$, starting with  an  application of the tower property:
	\begin{eqnarray*}
		\ExpBr{ G_i^{t+1} \;|\; i \notin S^t} &=&
		\ExpBr{\ExpBr{G_i^{t+1} \;|\; W^t, i \notin S^t}} \\
		&\overset{\eqref{ch3:eq:grad_distortion_def_pp}}{=}& \ExpBr{\ExpBr{\sqnorm{ g_i^{t+1} - \dfrac{\nabla f_i(x^{t+1})}{n w_i} }| \; W^t, i \notin S^t}} \\
		&=&
		\ExpBrBig{\ExpBrBig{\sqnorm{ g_i^{t} - \dfrac{\nabla f_i(x^{t+1})}{n w_i} + \right.\\
					&& \left. \qquad \dfrac{\nabla f_i(x^{t})}{n w_i} - \dfrac{\nabla f_i(x^{t})}{n w_i} }| \; W^t, i \notin S^t}} \\
		&\overset{\eqref{ch3:eq:young}}{\le}& \ExpBrBig{\ExpBrBig{(1+\rho)\sqnorm{ g_i^{t} - \dfrac{\nabla f_i(x^{t})}{n w_i} } \\
		& & \qquad + (1+\rho^{-1})\sqnorm{\dfrac{\nabla f_i(x^{t})}{n w_i} - \dfrac{\nabla f_i(x^{t+1})}{n w_i}}| \; W^t, i \notin S^t}} \\
		&=&
		(1+\rho) \ExpBr{G_i^t} + \dfrac{(1+\rho^{-1})}{n^2 w_i^2} \ExpBr{\sqnorm{\nabla f_i(x^{t+1}) - \nabla f_i(x^{t})}\; }.
	\end{eqnarray*}
	
	Given that Assumption~\ref{ch3:as:L_i} is satisfied, by applying \eqref{ch3:eq:L_i} to the second term, we obtain:
	\begin{equation}\label{ch3:eq:g_i_for_non_sampled_ef21_weighted}
		\ExpBr{ G_i^{t+1} \;|\; i \notin S^t} \leq (1+\rho) \ExpBr{G_i^t} + \dfrac{L_i^2(1+\rho^{-1})}{n^2 w_i^2} \ExpBr{\sqnorm{x^{t+1} - x^{t}}}.
	\end{equation}
	
	We combine the two preceding bounds:
	\begin{eqnarray*}
		\ExpBr{ G_i^{t+1} } & = &  \Prob(i \in S^t) \ExpBr{G_i^{t+1} \;|\; i \in S^t} \\
		&&\quad + \Prob(i \notin S^t) \ExpBr{G_i^{t+1}\;|\; i \notin S^t}\\
		&\overset{\eqref{ch3:eq:p_i_def}}{=}& p_i \ExpBr{ G_i^{t+1}|\; i \in S^t} \\
		&&\quad +(1-p_i)\ExpBr{ G_i^{t+1}|\; i \notin S^t}\\
		&\overset{\eqref{ch3:eq:g_i_for_sampled_ef21_weighted}+\eqref{ch3:eq:g_i_for_non_sampled_ef21_weighted}}{\le}& p_i \left[(1-\theta(\alpha,s)) \ExpBr{G_i^{t}} +\beta(\alpha,s) \dfrac{L_i^2}{n^2 w_i^2} \ExpBr{\sqnorm{x^{t+1} - x^t}}\right] \\
		&&\quad + (1-p_i)\left[(1+\rho) \ExpBr{G_i^t} +  \dfrac{L_i^2(1+\rho^{-1})}{n^2 w_i^2} \ExpBr{\sqnorm{x^{t+1} - x^{t}}}\right] \\
		&=& \left( (1-\theta(\alpha,s))p_i + (1 - p_i)(1+\rho) \right) \ExpBr{G_i^{t}}\\ 
		& & \quad + \left({\beta(\alpha,s) p_i} + (1-p_i){(1+\rho^{-1})} \right) \dfrac{L_i^2}{n^2 w_i^2} \ExpBr{\sqnorm{x^{t+1} - x^{t}}}.
	\end{eqnarray*}
	
	Consequently, for $\ExpBr{G^{t+1}}$, we derive the subsequent bound:
	\begin{eqnarray*}
		\ExpBr{G^{t+1}} &\overset{\eqref{ch3:eq:grad_distortion_def_pp}}{=}& \ExpBr{\sumin w_i G_i^{t+1}} \\
		&=& \sumin w_i \ExpBr{G_i^{t+1}} \\
		&\le& \sumin w_i \left( (1-\theta(\alpha,s))p_i + (1 - p_i)(1+\rho) \right) \ExpBr{G_i^{t}} \\
		&& \quad + \sumin w_i \left({\beta(\alpha,s) p_i} + (1-p_i){(1+\rho^{-1})} \right) \dfrac{L_i^2}{n^2 w_i^2} \ExpBr{\sqnorm{x^{t+1} - x^{t}}},
	\end{eqnarray*}
	where we applied the preceding inequality. Remembering the definitions
	\[
	p_{\min} \eqdef \min_i p_i, \quad 
	p_{\max} \eqdef \max_i p_i,
	\]		
	we subsequently obtain:
	\begin{eqnarray*}
		\ExpBr{G^{t+1}} &\le& \sumin w_i \left( (1-\theta(\alpha,s))p_{\max} + (1 - p_{\min})(1+\rho) \right) \ExpBr{G_i^{t}} \\
		&& \qquad +	\sumin \left({\beta(\alpha,s) p_{\max}} + (1-p_{\min}){(1+\rho^{-1})} \right)\dfrac{L_i^2}{n^2 w_i} \ExpBr{\sqnorm{x^{t+1} - x^{t}}} \\
		&=& \left( (1-\theta(\alpha,s))p_{\max} + (1 - p_{\min})(1+\rho) \right) \sumin w_i \ExpBr{G_i^t} \\
		&& \qquad +	\left({\beta(\alpha,s) p_{\max}} + (1-p_{\min}){(1+\rho^{-1})} \right) \sumin \dfrac{L_i^2}{n^2 w_i} \ExpBr{\sqnorm{x^{t+1} - x^{t}}}.
	\end{eqnarray*}
	
	Applying~\eqref{ch3:eq:grad_distortion_def_pp} and~\eqref{ch3:eq:weight_definition}, we obtain:
	\begin{eqnarray*}
		\ExpBr{G^{t+1}} &=& \left( (1-\theta(\alpha,s))p_{\max} + (1 - p_{\min})(1+\rho) \right) \ExpBr{G^t} \\
		&& + \left({\beta(\alpha,s) p_{\max}} + (1-p_{\min}){(1+\rho^{-1})} \right) \sumin \dfrac{L_i^2}{n^2 \dfrac{L_i}{\sum_{j=1}^{n} L_j}} \ExpBr{\sqnorm{x^{t+1} - x^{t}}} \\
		&=& \left( (1-\theta(\alpha,s))p_{\max} + (1 - p_{\min})(1+\rho) \right) \ExpBr{G^t} \\
		&& +	\left({\beta(\alpha,s) p_{\max}} + (1-p_{\min}){(1+\rho^{-1})} \right) \sum_{j=1}^{n} \dfrac{L_j}{n} \sumin \dfrac{L_i}{n} \ExpBr{\sqnorm{x^{t+1} - x^{t}}} \\
		&=&	\left( (1-\theta(\alpha,s))p_{\max} + (1 - p_{\min})(1+\rho) \right) \ExpBr{G^t} \\
		&& +	\left({\beta(\alpha,s) p_{\max}} + (1-p_{\min}){(1+\rho^{-1})} \right)  \LAMsq  \ExpBr{\sqnorm{x^{t+1} - x^{t}}}.
	\end{eqnarray*}
	Subsequently, in order to simplify the last inequality, we introduce the variables $1 - \theta_p$ and $\tilde{B}$:
	\begin{eqnarray*}
		1 - \theta_p &\eqdef& (1-\theta(\alpha,s))p_{\max} + (1 - p_{\min})(1+\rho) \\
		&=&	p_{\max} - p_{\max} \theta(\alpha,s) + 1 - p_{\min} + \rho - p_{\min}\rho \\
		&=&	1 - \left(-p_{\max} + p_{\max} \theta(\alpha,s) + p_{\min} - \rho + p_{\min}\rho\right) \\
		&=&	1 - \left(p_{\min}\rho + p_{\max} \theta(\alpha,s) - \rho - (p_{\max}-p_{\min})\right) .
	\end{eqnarray*}	
	
	Therefore,
	\begin{eqnarray*}	
		\theta_p &=& \left(p_{\max} \theta(\alpha,s) - \rho(1 - p_{\min}) - (p_{\max}-p_{\min})\right) \\
		\tilde{B} &\eqdef& \left({\beta(\alpha,s) p_{\max}} + (1-p_{\min}){(1+\rho^{-1})} \right) \LAMsq.
	\end{eqnarray*}
	
	Expressed in terms of these variables, the final inequality can be reformulated as:
	\begin{eqnarray*}
		\ExpBr{G^{t+1}} \le (1-\theta_p) \ExpBr{G^t} +
		\tilde{B} \ExpBr{\sqnorm{x^{t+1} - x^{t}}}.
	\end{eqnarray*}
	
	Since we need the contraction property over the gradient distortion $\ExpBr{G^{t+1}}$, we require $0 < \theta_p \leq 1$. We rewrite these conditions as follows:
	$$
	\dfrac{1 + \rho(1 - p_{\min}) + (p_{\max} - p_{\min}) }{p_{\max}}\geq \theta(\alpha, s) > \dfrac{\rho(1 - p_{\min}) + (p_{\max} - p_{\min}) }{p_{\max}}.
	$$
\end{proof}

\subsection{Main result}

We are ready to prove the main convergence theorem.
\begin{theorem} Consider Algorithm~\ref{ch3:alg:weighted_ef21_pp} (\algname{EF21-W-PP}) applied to the distributed optimization Problem~\eqref{ch3:eq:main_problem}. Let Assumptions~\ref{ch3:as:smooth},~\ref{ch3:as:L_i},~\ref{ch3:as:lower_bound} hold, assume that $\cC_i^t \in \mathbb{C}(\alpha)$ for all $i \in [n]$ and $t > 0$, set 
	\begin{equation*}
		G^t \eqdef \sumin w_i \left\|g_i^t - \dfrac{1}{n w_i} \nabla f_i(x^t)\right\|^2,
	\end{equation*}	
	where 
	$$w_i = \dfrac{L_i}{\sum_j L_j}, \forall i \in [n],$$
	and let the step size satisfy
	\begin{equation}
		\label{ch3:eq:ef21-w-pp-gamma}
		0 < \gamma \leq \left(L + \sqrt{\dfrac{\tilde{B}}{\theta_p}} \right)^{-1},
	\end{equation}
	where $s>0$, $\rho > 0$, and
	\begin{eqnarray*}
		\theta(\alpha, s) &\eqdef& 1 - (1 - \alpha)(1 + s)\\
		\beta(\alpha, s) &\eqdef&  (1 - \alpha)(1+s^{-1})\\
		\theta_p &\eqdef& p_{\min}\rho + \theta(\alpha,s) p_{\max} - \rho - (p_{\max}-p_{\min}) \\
		\tilde{B} &\eqdef& \left({\beta(\alpha,s) p_{\max}} + (1-p_{\min}){(1+\rho^{-1})} \right) \LAMsq.
	\end{eqnarray*}
	Additionally, assume that
	$$
	\dfrac{1 + \rho(1 - p_{\min}) + (p_{\max} - p_{\min}) }{p_{\max}}\geq \theta(\alpha, s) > \dfrac{\rho(1 - p_{\min}) + (p_{\max} - p_{\min}) }{p_{\max}}.
	$$
	If for $T > 1$ we define $\hat{x}^T$ as an element of the set $\{x^0, x^1, \dots, x^{T-1}\}$ chosen uniformly at random, then
	\begin{equation}
		\ExpBr{\|\nabla f(\hat{x}^T) \|^2} \leq \dfrac{2 (f(x^0) - f^\ast)}{\gamma T} + \dfrac{G^0}{\theta_p T}.
	\end{equation}
\end{theorem}

\begin{proof}
	Following the same approach employed in the proof for the \algname{SGD} case, we obtain
	\begin{eqnarray*}
		\ExpBr{f(x^{t+1})-f^\ast} & \leq &  \ExpBr{f(x^{t})-f^\ast}
		-\dfrac{\gamma}{2} \ExpBr{\sqnorm{\nabla f(x^{t})}}   \\
		&& \qquad 	
		-\left(\dfrac{1}{2 \gamma}-\dfrac{L}{2}\right) \ExpBr{\sqnorm{x^{t+1}-x^{t}}}+ \dfrac{\gamma}{2}\ExpBr{G^t}.
	\end{eqnarray*}
	Let 
	\begin{eqnarray*}
		\delta^{t} &\eqdef& \ExpBr{f(x^{t})-f^\ast}, \\
		s^{t} &\eqdef& \ExpBr{G^t}, \\
		r^{t} &\eqdef& \ExpBr{\sqnorm{x^{t+1}-x^{t}}}.
	\end{eqnarray*}
		
	Applying the previous Lemma, we obtain:
	\begin{eqnarray*}
		\delta^{t+1}+\dfrac{\gamma}{2 \theta_p} s^{t+1} &\leq& \delta^{t}-\dfrac{\gamma}{2}\ExpBr{\sqnorm{\nabla f(x^{t})}} \\
		&& \qquad - \left(\dfrac{1}{2 \gamma}-\dfrac{L}{2}\right) r^{t}+\dfrac{\gamma}{2} s^{t}+\dfrac{\gamma}{2 \theta_p}\left(\tilde{B} r^t + (1 - \theta_p) s^{t}\right) \\
		&=&\delta^{t}+\dfrac{\gamma}{2\theta} s^{t}-\dfrac{\gamma}{2}\ExpBr{\sqnorm{\nabla f(x^{t})}}-\underbrace{\left(\dfrac{1}{2\gamma} -\dfrac{L}{2} - \dfrac{\gamma}{2\theta_p} \tilde{B} \right)}_{\geq 0} r^{t} \\
		& \leq& \delta^{t}+\dfrac{\gamma}{2\theta_p} s^{t} -\dfrac{\gamma}{2}\ExpBr{\sqnorm{\nabla f(x^{t})}}.
	\end{eqnarray*}
	By summing up inequalities for $t =0, \ldots, T-1,$ we get
	$$
	0 \leq \delta^{T}+\dfrac{\gamma}{2 \theta_p} s^{T} \leq \delta^{0}+\dfrac{\gamma}{2 \theta_p} s^{0}-\dfrac{\gamma}       {2} \sum_{t=0}^{T-1} \ExpBr{\sqnorm{\nabla f(x^{t})}}.
	$$
	
	Finally, via multiplying both sides by $\dfrac{2}{\gamma T}$, after rearranging we get:
	\begin{eqnarray*}
		\sum_{t=0}^{T-1} \dfrac{1}{T} \ExpBr{\sqnorm{\nabla f (x^{t})}} \leq \dfrac{2 \delta^{0}}{\gamma T} + \dfrac{s^0}{\theta_p T}.
	\end{eqnarray*}
	
	It remains to notice that the left-hand side can be interpreted as $\ExpBr{\sqnorm{\nabla f(\hat{x}^{T})} }$, where $\hat{x}^{T}$ is chosen from the set $\{x^{0}, x^{1}, \ldots, x^{T-1}\}$ uniformly at random.
\end{proof}

Our analysis of this extension follows a similar approach as the one used by \citet{fatkhullin2021ef21}, for an algorithm they called \algname{EF21-PP}. Presented analysis of \algname{EF21-W-PP} has a better provides a better multiplicative factor in the definition of $\widetilde{B} \propto \LAMsq$, where in vanilla \algname{EF21-PP} had $\widetilde{B} \propto \LQMsq$. This fact  improves the upper bound on allowable step size in \eqref{ch3:eq:ef21-w-pp-gamma}.

\clearpage
\addtocounter{adjsection}{1}
\section{Improved Theory for EF21 in the Rare Features Regime} \label{ch3:sec:RF}

In this section, we adapt our new results to the {\em rare features} regime proposed and studied by~\citet{richtarik2023error}.

\subsection{Algorithm}

In this section, we focus on Algorithm~\ref{ch3:alg:EF21_RF}, which is an adaptation of \algname{EF21} (as delineated in Algorithm~\ref{ch3:alg:EF21}) that specifically employs \compname{TopK} operators. This variant is tailored for the rare features scenario, enhancing the convergence rate by shifting from the average of squared Lipschitz constants to the square of their average. The modifications introduced in Algorithm~\ref{ch3:alg:EF21_RF}, compared to the standard \algname{EF21}, are twofold and significant.

Primarily, the algorithm exclusively engages \compname{TopK} compressors, leveraging the inherent sparsity present in the data. Additionally, the initial gradient estimates $ g_i^0$ are confined to the respective subspaces $\RR^d_i$, as characterized by Equation~\eqref{ch3:eq:r_d_i_def}. With the exception of these distinct aspects, the algorithm's execution parallels that of the original \algname{EF21}.

\begin{algorithm}
	\begin{algorithmic}[1]
		\STATE {\bfseries Input:} initial model $x^0 \in \RR^d$; initial gradient estimates $g_1^0\in \R^d_1,\dots,g_n^0 \in \R^d_n$ (as defined in Equation~\eqref{ch3:eq:r_d_i_def}) stored at the server and the clients; step size $\gamma>0$; sparsification levels $K_1,\dots,K_n \in [d]$; number of iterations $T > 0$	
		\STATE {\bfseries Initialize:} $g^0 = \avein g_i^0 $ on the server
		\FOR{$t = 0, 1, 2, \dots, T - 1 $}
		\STATE Server computes $x^{t+1} = x^t - \gamma g^t$ and  broadcasts  $x^{t+1}$ to all $n$ clients
		\FOR{$i = 1, \dots, n$ {\bf on the clients in parallel}} 
		\STATE Compute $u_i^t=\text{Top}K_i (\nabla f_i(x^{t+1}) - g_i^t)$ and update $g_i^{t+1} = g_i^t +u_i^t$ \label{ch3:line:sparse_g_update_step}		
		\STATE Send the compressed message $u_i^{t}$ to the server
		\ENDFOR 
		\STATE Server updates $g_i^{t+1} = g_i^t +u_i^t$ for all $i\in [n]$, and computes $g^{t+1} = \avein g_i^{t+1}$		 \label{ch3:line:sparse_averaging_line}
		\ENDFOR
		\STATE {\bfseries Output:} Point $\hat{x}^T$ chosen from the set $\{x^0, \dots, x^{T-1}\}$ uniformly at random
	\end{algorithmic}
	\caption{\algname{EF21}: Error Feedback 2021 with \compname{TopK} Compressors.}
	\label{ch3:alg:EF21_RF}
\end{algorithm}

\subsection{New sparsity measure}

To extend our results to the {\em rare features} regime, we need to slightly change the definition of the parameter $c$ in the original paper. The way we do it is unrolled as follows. First, we recall the following definitions from~\citep{richtarik2023error}:
\begin{equation}
	\cZ \eqdef \{ (i, j)\in [n] \times [d]\; | \;  [\nabla f_i(x)]_j = 0 \ \forall x \in \RR^d \},
\end{equation}
and
\begin{equation}
	\cI_j \eqdef \{ i \in [n]\; | \;   (i, j) \notin \cZ\}, \qquad \cJ_i \eqdef \{ j \in [d]\; | \;   (i, j) \notin \cZ\}.
\end{equation}

We also need the following definition of $\RR^d_i$:
\begin{equation}\label{ch3:eq:r_d_i_def}
	\RR^d_i \eqdef \{u = (u_1, \dots, u_d) \in \RR^d : u_j = 0 \text{ whenever } (i,j) \in \cZ \}.
\end{equation}

Now we are ready for a new definition of the sparsity parameter $c$:
\begin{equation}\label{ch3:eq:sparse_c_definiton}
	c \eqdef n \cdot \max\limits_{j \in [d]} \sum\limits_{i \in \cI_j} w_i,
\end{equation}
where $w_i$ is defined as in~\eqref{ch3:eq:weight_definition}. We note that $c$ recovers the standard definition from \citep{richtarik2023error} when $w_i = \dfrac{1}{n}$ for all $i\in [n]$.

\subsection{Lemmas}

We will proceed through several Lemmas.

\begin{lemma}\label{ch3:lm:sparse_tighter_bound}
	Let $u_i \in \RR^d_i$ for all $i \in [n]$. Then, the following inequality holds:
	\begin{equation}\label{ch3:eq:sparse_tighter_bound}
		\left\|\sumin w_i u_i \right\|^2 \leq \dfrac{c}{n} \sumin w_i \|u_i\|^2.
	\end{equation}
\end{lemma}
\begin{proof}
	Initially, we observe that
	\begin{align}\label{ch3:eq:sparse_aux_1}
		\left\|\sumin w_i u_i \right\|^2 = \sum\limits_{j=1}^d \left(\sumin w_i u_{ij} \right)^2.
	\end{align}
	We note that for any $j \in [d]$ it holds that
	\begin{equation}\label{ch3:eq:sparse_aux_2}
		\begin{aligned}
			\left(\sumin w_i u_{ij} \right)^2 &= \left(\sum\limits_{i \in \cI_j} w_i u_{ij} \right)^2 \\
			&= \left(\sum\limits_{i \in \cI_j} w_i \right)^2 \left(\sum\limits_{i \in \cI_j} \dfrac{w_i}{\sum\limits_{i' \in \cI_j} w_{i'}} u_{ij} \right)^2  \leq \left(\sum\limits_{i \in \cI_j} w_i \right)^2  \sum\limits_{i \in \cI_j} \dfrac{w_i}{\sum\limits_{i' \in \cI_j} w_{i'}} u_{ij}^2,
		\end{aligned}
	\end{equation}
	where on the last line we used Jensen's inequality. Subsequent arithmetic manipulations and the incorporation of definition~\eqref{ch3:eq:sparse_c_definiton} yield:
	\begin{eqnarray}
		\left(\sumin w_i u_{ij} \right)^2 & \overset{\eqref{ch3:eq:sparse_aux_2}}{\leq}  & \left(\sum\limits_{i \in \cI_j} w_i \right) \cdot \sum\limits_{i \in \cI_j} w_i u_{ij}^2  \notag \\
		& = & \left(\sum\limits_{i \in \cI_j} w_i \right) \cdot \sumin w_i u_{ij}^2  \notag \\
		& \leq & \left[\max_{j \in [d]}\left(\sum\limits_{i \in \cI_j} w_i \right)\right] \cdot \sumin w_i u_{ij}^2 \notag  \\
		& \overset{\eqref{ch3:eq:sparse_c_definiton}}{=} & \dfrac{c}{n} \sumin w_i u_{ij}^2.
		\label{ch3:eq:sparse_aux_3}
	\end{eqnarray}
	
	Substituting Equation~\eqref{ch3:eq:sparse_aux_3} into Equation~\eqref{ch3:eq:sparse_aux_1} completes the proof.
\end{proof}

\begin{lemma}
	Assume that $g_i^0 \in \RR^d_i$ for all $i \in [n]$. Then, it holds for all $t>0$ that
	\begin{equation}\label{ch3:eq:grad_est_distortion}
		\| g^t - \nabla f(x^t)\|^2 \leq \dfrac{c}{n} G^t.
	\end{equation}
\end{lemma}

\begin{proof}
	By Lemma 8 in~\citet{richtarik2023error}, for \algname{EF21} the update $g_i^t$ stays in $\RR^d_i$ if $g_i^0 \in \RR^d_i$. We then proceed as follows:
	\begin{equation}\label{ch3:eq:sparse_aux_4}
		\begin{aligned}
			\|g^t - \nabla f(x^t) \|^2 \overset{\eqref{ch3:line:sparse_averaging_line}}{=}\left\|\avein g_i^t - \nabla f_i(x^t) \right\|^2 = \left\| \sumin w_i \left[\dfrac{1}{nw_i} (g_i^t - \nabla f_i(x^t)) \right] \right\|^2.
		\end{aligned}
	\end{equation}
	Since $g_i^t \in \RR^d_i$, as was noted, and $\nabla f_i(x^t) \in \RR^d_i$, by the definition of $\RR^d_i$, then $\dfrac{1}{nw_i} (g_i^t - \nabla f_i(x^t))$ also belongs to $\RR^d_i$. By Lemma~\ref{ch3:lm:sparse_tighter_bound}, we further proceed:
	\begin{eqnarray*}
		\|g^t - \nabla f(x^t) \|^2 &\overset{\eqref{ch3:eq:sparse_aux_4}} {=} &  \left\| \sumin w_i \left[\dfrac{1}{nw_i} (g_i^t - \nabla f_i(x^t)) \right] \right\|^2 \\
		& \overset{\eqref{ch3:eq:sparse_tighter_bound}}{\leq} & \dfrac{c}{n} \sumin w_i \left\| \dfrac{1}{nw_i} (g_i^t - \nabla f_i(x^t))\right\|^2\\
		& = & \dfrac{c}{n} \sumin \dfrac{1}{n^2 w_i} \| g_i^t - \nabla f_i(x^t)\|^2\\
		& = & \dfrac{c}{n} G^t,
	\end{eqnarray*}
	which completes the proof.
\end{proof}

For the convenience of the reader, we briefly revisit Lemma 6 from~\citep{richtarik2023error}.
\begin{lemma}[Lemma 6 from~\citet{richtarik2023error}]
	If Assumption~\ref{ch3:as:L_i} holds, then for $i \in [n]$, we have
	\begin{equation}\label{ch3:eq:sparse_smoothness}
		\sum\limits_{j: (i, j) \notin \cZ} ((\nabla f_i(x))_j - (\nabla f_i(y))_j )^2 \leq L_i^2 \sum\limits_{j: (i, j) \notin \cZ}  (x_j - y_j)^2 \quad \forall x, y \in \RR^d.
	\end{equation}
\end{lemma}

Now, we proceed to the following Lemma, which aims to provide a tighter bound for the quantity $$\sumin \dfrac{1}{L_i} \|\nabla f_i(x) - \nabla f_i(y)\|^2.$$

\begin{lemma}\label{ch3:lm:sparse_L+}
	If Assumption~\ref{ch3:as:L_i} holds, then
	\begin{equation}\label{ch3:eq:sparse_L_+}
		\sumin \dfrac{1}{L_i} \|\nabla f_i(x) - \nabla f_i(y) \|^2 \leq c \LAM \|x - y\|^2.
	\end{equation}
\end{lemma}
\begin{proof}
	The proof commences as follows:
	\begin{eqnarray}
		\sumin \dfrac{1}{L_i} \|\nabla f_i(x) - \nabla f_i(y) \|^2 &= & \sumin \dfrac{1}{L_i} \sum\limits_{j: (i, j) \notin \cZ} ((\nabla f_i(x))_j - (\nabla f_i(y))_j )^2 \notag \\
		& \overset{\eqref{ch3:eq:sparse_smoothness}}{\leq}  & \sumin \dfrac{1}{L_i}  L_i^2 \sum\limits_{j: (i, j) \notin \cZ}  (x_j - y_j)^2 \notag  \\
		& =  & \sumin \sum\limits_{j: (i, j) \notin \cZ}  L_i (x_j - y_j)^2 \notag  \\
		& = & \sum\limits_{j=1}^d  \sum\limits_{i: (i, j) \notin \cZ} L_i (x_j - y_j)^2  \notag \\
		& = & \sum\limits_{j=1}^d \left[(x_j - y_j)^2 \sum\limits_{i: (i, j) \notin \cZ} L_i \right]. \label{ch3:eq:sparse_aux_5}
	\end{eqnarray}
	To advance our derivations, we consider the maximum value over $\sum\limits_{i: (i, j) \notin \cZ} L_i$:
	\begin{eqnarray*}
		\sumin \dfrac{1}{L_i} \|\nabla f_i(x) - \nabla f_i(y) \|^2 &\overset{\eqref{ch3:eq:sparse_aux_5}}{\leq} & \sum\limits_{j=1}^d \left[(x_j - y_j)^2 \sum\limits_{i: (i, j) \notin \cZ} L_i \right] \notag \\
		& \leq  & \sum\limits_{j=1}^d \left[(x_j - y_j)^2 \max\limits_{j \in [d]} \sum\limits_{i: (i, j) \notin \cZ} L_i \right] \notag \\
		& = & \left[\max\limits_{j \in [d]} \sum\limits_{i: (i, j) \notin \cZ} L_i \right] \sum\limits_{j=1}^d (x_j - y_j)^2 \notag \\
		& =   & \left[\max\limits_{j \in [d]} \sum\limits_{i: (i, j) \notin \cZ} L_i \right] \|x - y\|^2 \notag \\
		& =  & \left[\max\limits_{j \in [d]} \sum\limits_{i \in \cI_j} L_i \right] \|x - y\|^2 \\
		& \overset{\eqref{ch3:eq:sparse_c_definiton}}{=}& c \LAM \|x - y\|^2,
	\end{eqnarray*}
	what completes the proof.
\end{proof}

For clarity and easy reference, we recapitulate Lemma 10 from~\citet{richtarik2023error}.
\begin{lemma}[Lemma 10 from~\citet{richtarik2023error}]\label{ch3:lm:sparse_grad_est_evolution}
	The iterates of Algorithm~\ref{ch3:alg:EF21_RF}  method satisfy 
	\begin{equation}\label{ch3:eq:sparse_grad_est_evolution}
		\left\|g_i^{t+1} - \nabla f_i(x^{t+1})\right\|^2 \leq (1 - \theta(\alpha)) \left\|g_i^t - \nabla f_i(x^t)\right\|^2 + \beta(\alpha) \|\nabla f_i(x^{t+1}) - \nabla f_i(x^t)\|^2,
	\end{equation}
	where  $$\alpha = \min\left\{\min\limits_{i \in [n]} \dfrac{K_i}{|\cJ_i|}, 1 \right\}.$$
\end{lemma}

\begin{lemma}
	Under Assumption~\ref{ch3:as:L_i}, iterates of Algorithm~\ref{ch3:alg:EF21_RF} satisfies 
	\begin{equation}
		G^{t+1} \leq (1 - \theta(\alpha)) G^t + \beta(\alpha) \dfrac{c}{n} \LAMsq \|x - y\|^2.
	\end{equation}
\end{lemma}
\begin{proof}
	The proof is a combination of Lemmas~\ref{ch3:lm:sparse_L+} and~\ref{ch3:lm:sparse_grad_est_evolution}:
	\begin{eqnarray*}
		G^{t+1} &\overset{\eqref{ch3:eq:new_def_distortion}}{=}  & \dfrac{1}{n^2} \sumin \dfrac{1}{w_i} \left\|g_i^{t+1} - \nabla f_i(x^{t+1})\right\|^2 \\
		& \overset{\eqref{ch3:eq:sparse_grad_est_evolution}}{\leq}  & \dfrac{1}{n^2} \sumin \dfrac{1}{w_i} \left[ (1 - \theta(\alpha)) \|g_i^t - \nabla f_i(x^t)\|^2 + \beta(\alpha) \|\nabla f_i(x^{t+1}) - \nabla f_i(x^t) \|^2 \right] \\
		& =  & (1 - \theta(\alpha)) G^t + \dfrac{\beta(\alpha)}{n^2} \sumin \dfrac{1}{w_i} \|\nabla f_i(x^{t+1}) - \nabla f_i(x^t) \|^2 \\
		& \overset{\eqref{ch3:eq:weight_definition}}{=}  & (1 - \theta(\alpha)) G^t + \dfrac{\beta(\alpha)}{n^2}  \left(\sum_{j=1}^n L_j\right)\sumin \dfrac{1}{L_i} \|\nabla f_i(x^{t+1}) - \nabla f_i(x^t) \|^2 \\
		& \overset{\eqref{ch3:eq:sparse_L_+}}{\leq}  & (1 - \theta(\alpha)) G^t + \dfrac{\beta(\alpha)}{n^2} \dfrac{c}{n} \left( \sum_{j=1}^n L_j \right)^2\|x - y\|^2\\
		& = & (1 - \theta (\alpha) ) G^t + \beta(\alpha) \dfrac{c}{n} \LAMsq \|x - y\|^2.
	\end{eqnarray*}
\end{proof}

\subsection{Main result}

And now we are ready to formulate the main result.
\begin{theorem}
	Let Assumptions~\ref{ch3:as:smooth},~\ref{ch3:as:L_i} and~\ref{ch3:as:lower_bound} hold. Let $g_i^0 \in \RR^d_i$ for all $i \in [n]$, $$\alpha = \min\left\{\min\limits_{i \in [n]} \dfrac{K_i}{|\cJ_i|}, 1 \right\}, \qquad 0<\gamma \leq \dfrac{1}{L + \dfrac{c}{n} \LAM \xi(\alpha) }.$$ Under these conditions, the iterates of Algorithm~\ref{ch3:alg:EF21_RF} satisfy
	\begin{equation}
		\dfrac{1}{T} \sum\limits_{t=0}^{T-1} \|\nabla f(x^t)\|^2 \leq \dfrac{2 (f(x^0) - f^\ast) }{\gamma T} + \dfrac{c}{n} \dfrac{G^0}{\theta(\alpha) T}.
	\end{equation}
\end{theorem}
\begin{proof}
	Let us define the Lyapunov function: 
	\begin{equation}\label{ch3:eq:sparse_Lyapunov}
		\Psi^t \eqdef f(x^t) - f^\ast + \dfrac{\gamma c}{2\theta n} G^t.
	\end{equation}
	We start the proof as follows:
	\begin{eqnarray*}
		\Psi^{t+1} &\overset{\eqref{ch3:eq:sparse_Lyapunov}}{=}  & f(x^{t+1}) - f^\ast + \dfrac{\gamma c}{2\theta n} G^{t+1} \\
		& \overset{\eqref{ch3:eq:descent_lemma}}{\leq}  & f(x^t) - f^\ast - \dfrac{\gamma}{2} \|\nabla f(x^t)\|^2 - \left(\dfrac{1}{2\gamma} - \dfrac{L}{2}\right) \|x^{t+1} - x^t\|^2 \\
		&& \qquad + \dfrac{\gamma}{2} \|g^t - \nabla f(x^t)\|^2 +  \dfrac{\gamma c}{2\theta n} G^{t+1} \\
		& \overset{\eqref{ch3:eq:grad_est_distortion}}{\leq} &  f(x^t) - f^\ast - \dfrac{\gamma}{2} \|\nabla f(x^t)\|^2 - \left(\dfrac{1}{2\gamma} - \dfrac{L}{2}\right) \|x^{t+1} - x^t\|^2 \\
		&& \qquad + \dfrac{\gamma}{2}\dfrac{c}{n} G^t +  \dfrac{\gamma c}{2\theta n} G^{t+1} \\
		& \overset{\eqref{ch3:eq:sparse_grad_est_evolution}}{\leq}  & f(x^t) - f^\ast - \dfrac{\gamma}{2} \|\nabla f(x^t)\|^2 - \left(\dfrac{1}{2\gamma} - \dfrac{L}{2}\right) \|x^{t+1} - x^t\|^2 \\
		& & \qquad + \dfrac{\gamma}{2}\dfrac{c}{n} G^t +  \dfrac{\gamma c}{2\theta n} \left((1 - \theta) G^t  + \beta\dfrac{c}{n} \LAMsq \|x^{t+1} - x^t\|^2 \right) \\
		& =  & f(x^t) - f^\ast + \dfrac{\gamma c}{2\theta n} G^t - \dfrac{\gamma}{2} \|\nabla f(x^t)\|^2 \\
		&& \qquad - \left(\dfrac{1}{2\gamma} - \dfrac{L}{2} - \dfrac{\gamma}{2} \dfrac{\beta}{\theta} \dfrac{c^2}{n^2} \cdot \LAMsq \right) \|x^{t+1} - x^t\|^2\\
		& =  & \Psi^t - \dfrac{\gamma}{2} \|\nabla f(x^t)\|^2 \\
		& & \qquad- \underbrace{\left(\dfrac{1}{2\gamma} - \dfrac{L}{2} - \dfrac{\gamma}{2} \dfrac{\beta}{\theta} \dfrac{c^2}{n^2} \cdot \LAMsq \right)}_{\geq 0} \|x^{t+1} - x^t\|^2 \\
		& \leq & \Psi^t - \dfrac{\gamma}{2} \|\nabla f(x^t)\|^2.
	\end{eqnarray*}
	
	Unrolling the inequality above, we get
	\begin{align*}
		0 \leq \Psi^T \leq \Psi^{T-1} - \dfrac{\gamma}{2} \|\nabla f(x^{T-1})\|^2 \leq \Psi^0 - \dfrac{\gamma}{2}\sum\limits_{t=0}^{T-1} \|\nabla f(x^t) \|^2,
	\end{align*}
	from what the main result follows.
\end{proof}

\clearpage
\addtocounter{adjsection}{1}
\section{Experiments: Further Details}
\label{ch3:app:exp-additional-details-main-part}

\begin{center}	
	\begin{figure*}[h]
		\centering
		\captionsetup[sub]{font=normalsize,labelfont={}}	
		\captionsetup[subfigure]{labelformat=empty}		
		
		\begin{subfigure}[ht]{0.85\textwidth}
			\includegraphics[width=\textwidth]{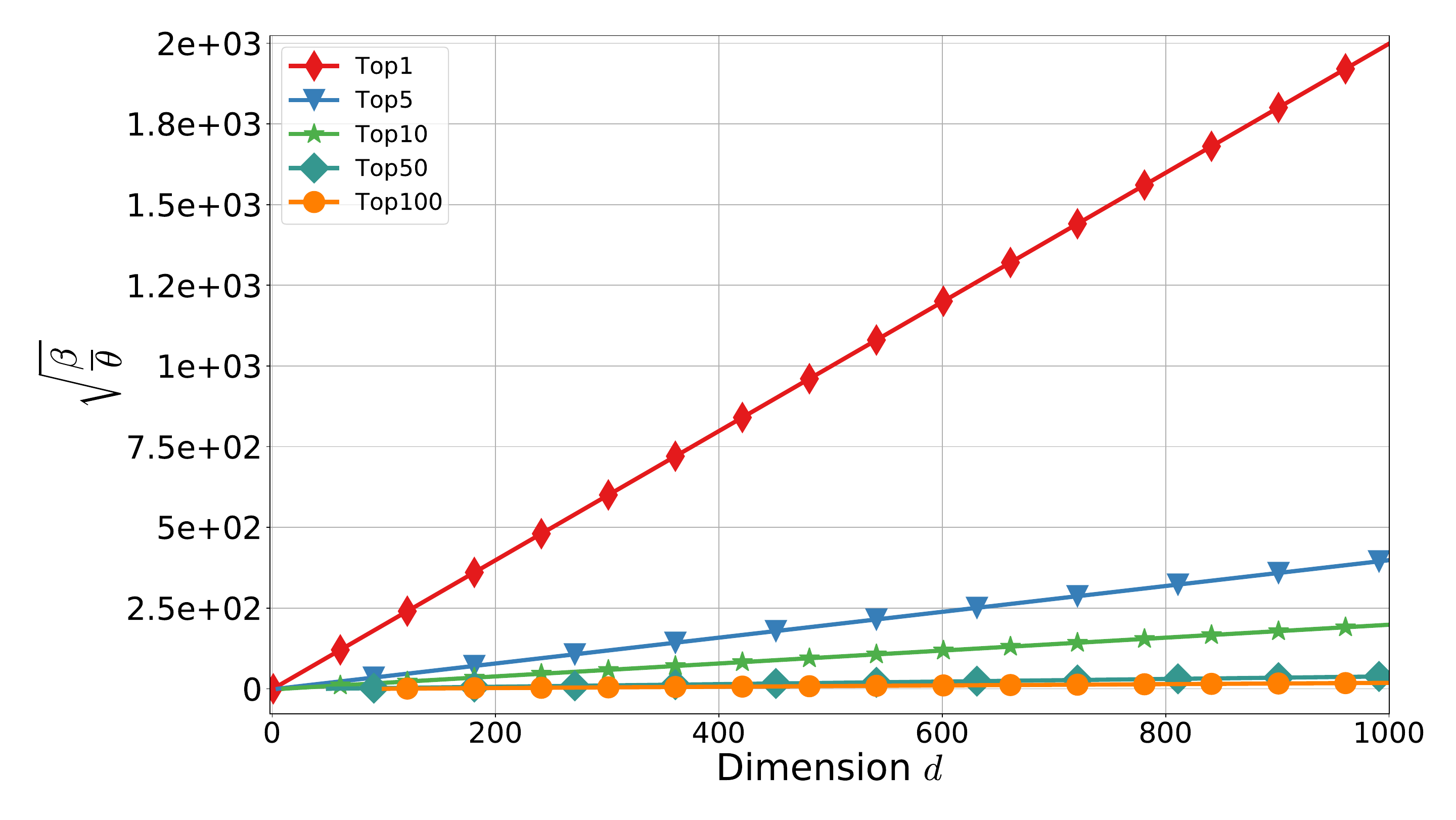} \caption{}
		\end{subfigure}
		
		\caption{{The factor $\xi=\sqrt{{\beta}/{\theta}}$ as a function of optimization variable dimension $d$ for several \compname{TopK} compressors. The behavior is independent of properties of $\{f_1(x),\dots,f_n(x)\}$ and $f(x)$.}}
		\label{ch3:fig:beta-over-theta-theoretical}
	\end{figure*}
\end{center}

\subsection{Computing and software environment}
\label{ch3:app:compute-env}

We used the Python software suite \fl \citep{burlachenko2021fl_pytorch} to simulate the distributed environment for training. We carried out experiments on a compute node with {Ubuntu 18.04 LTS}, $256$ GBytes of DRAM DDR4 memory at $2.9$ GHz, and $48$ cores ($2$ sockets with $24$ cores per socket) of {Intel(R) Xeon(R) Gold 6246 CPU at $3.3$ GHz}. We used double-precision arithmetic (FP64) during computing gradient oracles. All our computations were carried on {CPU}.

\subsection{Comments on the improvement}
\label{ch3:app:practical-role}

The standard \algname{EF21} analysis \citep{EF21} allows to utilize \algname{EF21} with maximum allowable step size $\gamma$ equal to:
\begin{eqnarray*}
	\label{ch3:frm:ef21}
	\gamma = \left( L + \LQM \sqrt{\dfrac{\beta(\alpha)}{\theta(\alpha)}} \right)^{-1}, \qquad  \theta(\alpha) = 1 - \sqrt{1-\alpha}, \qquad \beta(\alpha)=\dfrac{1-\alpha}{1 - \sqrt{1-\alpha}}.
\end{eqnarray*}

Our analysis allows us to replace the quantity $\LQM$ with $\LAM$. This improvement has an important consequence. The replaced quantity affects the step size by a factor of $\xi(\alpha)=\sqrt{\dfrac{\beta(\alpha)}{\theta(\alpha)}}$. This factor can be arbitrarily large as $d$ increases, as shown in Figure~\ref{ch3:fig:beta-over-theta-theoretical}. If $d$ is increasing and the parameter $k$ of \compname{TopK} compressor is fixed, then even a small improvement in the constant term can have a significant impact in an absolute sense on the computed step size if $\xi(\alpha) \gg L$.

\subsection{When improved analysis leads to more aggressive steps}
\label{ch3:app:when-improved-analysis-better}

The quantity $\color{red} \LQM \eqdef \sqrt{\dfrac{1}{n} \sum_{i=1}^n L_i ^2}$ plays essential role in \algname{EF21} analysis. As we saw with special consideration this quantity for \algname{EF21} and its extensions is improvable. The improved analysis allows us to replace it with $\color{blue}  \LAM \eqdef {\dfrac{1}{n}\sum_{i=1}^n L_i}.$ Clearly, by the arithmetic-quadratic mean inequality,  $$\Lvar \eqdef \LQMsq - \LAMsq  \ge 0.$$ The difference $\LQM - \LAM$ can be expressed as follows:

\begin{eqnarray*}
	\LQM - \LAM &=&  (\LQM - \LAM)\left(\dfrac{\LQM + \LAM}{\LQM + \LAM}\right) \\
	&=& \dfrac{\LQMsq - \LAMsq}{\LQM + \LAM} \quad = \quad \dfrac{1}{\LQM + \LAM} \cdot {\dfrac{1}{n} \sum_{i=1}^{n} \left(L_i - \dfrac{1}{n} \sum_{i=1}^{n} L_i\right) ^2}.
\end{eqnarray*}

The coefficient $\dfrac{1}{\LQM+\LAM}$ in the last equation can be bounded from below and above as follows:
$$\dfrac{1}{2 \LQM} = \dfrac{1}{2 \cdot \max(\LQM, \LAM)} \le \dfrac{1}{\LQM+\LAM} \le \dfrac{1}{2 \cdot \min(\LQM, \LAM)} \le \dfrac{1}{2\LAM}.$$

As a consequence, the difference $\LQM - \LAM$ is bound above by the estimated variance of $L_i$ divided by the mean of $L_i$, also known as \textit{Index of Dispersion} in statistics. From this consideration, we can more easily observe that \algname{EF21-W} can have an arbitrarily better step size than vanilla \algname{EF21} if the variance of $L_i$ is increasing faster than the mean of $L_i$.

\subsection{Dataset generation for synthetic experiment}
\label{ch3:app:dataset-gen-synthetic}

First, we assume that the user provides two parameters: $\mu \in \mathbb{R}_+$ and $L \in \mathbb{R}_+$. These parameters define the construction of strongly convex function $f_i(x)$, which are modified by meta-parameters $q \in [-1,1]$ and $z > 0$, described next.
\begin{enumerate}
	\item  Each client initially has
	$$f_i(x) \eqdef \dfrac{1}{n_i} \norm{\mA_i x - {b_i}}^2,$$ where $\mA_i$ is initialized in such way that $f_i$ is $L_{i}$ smooth and $\mu_{f_i}$ strongly convex. Parameters are defined in the following way: 
	\[L_{i}=\dfrac{i}{n} \cdot (L-\mu) + \mu, \qquad \mu_{f_i} = \mu.\]
	
	\item The scalar value $q \in [-1,+1]$ informally plays the role of meta parameter to change the distribution of $L_{i}$ and make values of $L_{i}$ close to one of the following: (i) $\mu$; (ii) $L$; (iii) $(L+\mu)/2$. The exact modification of $L_{i}$ depends on the sign of meta parameter $q$.
	
	\begin{itemize}
		\item Case $q \in [0,1]$. In this case for first $n/2$ (i.e., $i \in [0, n/2]$) compute the value $L_{i,q} = \mathrm{lerp}(L_{i}, \mu, q)$, where $\mathrm{lerp}(a,b,t):\RD \times \RD \times [0,1] \to \RD$ is standard linear interpolation	
		\[
		\mathrm{lerp}(a, b, t) = a (1-t) + bt.
		\]
		
		The last $n/2$ ($i \in [n/2+1, n]$) compute the value $L_{i,q} = \mathrm{lerp}(L_{i}, L, q)$. For example, if $q=0$ then $L_{i,q}=L_{i}, \forall i \in [n]$, and if $q=1$ then $L_{i,q} = \mu$ for first ${n}/{2}$ clients and $L_{i,q} = L$ for last ${n}/{2}$ clients.

		\item  Case $q \in [-1,0]$. In this for all $n$ clients the new value $L_{i,q} = \mathrm{lerp}(L_{i}, ({L+\mu})/{2}, -q)$. In this case for example if $q=0$ then $L_{i,q}=L_{i}$ and if $q=-1$ then $L_{i,q} = ({L+\mu})/{2}$.

	\end{itemize}
	
	The process firstly fills the $\mA_i$ in such form that $L_{i}$ forms a uniform spectrum in $[\mu, L]$ with the center of this spectrum equal to $a=\dfrac{L+\mu}{2}$. And then as $q \to 1$, the variance of $L_{i,q}$ is increasing. 
	
	\item  We use these new values  $L_{i,q}$ for all $i \in [n]$ clients as a final target $L_{i}^{\mathrm{new}}$ values. Due to numerical issues, we found that it is worthwhile for the first and last client to add extra scaling. First client scales $L_{1,q}$ by 
	factor $1/z$, and last $n$-th client scales $L_{n,q}$ by factor $z$. Here $z \ge 0$ is an additional meta-parameter.
	
	\item Next obtained values are used to generate $\mA_i$ in such way that $\nabla^2 f_i(x)$ has uniform spectrum in $[\mu_{f_i,q,z}, L_{i,q,z}]$.
	
	\item  As a last step the objective function $f(x)$ is scaled in such a way that it is $L$ smooth with constant value $L$. The $b_i$ for each client is initialized as $b_i \eqdef \mA_i \cdot x_{\mathrm{solution}}$, where $x_{\mathrm{solution}}$ is fixed solution.
\end{enumerate}

\subsection{Dataset shuffling strategy for LIBSVM dataset}
\label{ch3:app:dataset-shuffling-for-libsvm}

Our dataset shuffling strategy heuristically splits data points so that $\Lvar$ is maximized. It consists of the following steps:
\begin{enumerate}
	\item  {\bf Sort data points from the whole dataset according to $L$ constants.} Sort all data points according to the smoothness constants of the loss function for each single data point.
	\item  {\bf Assign a single data point to each client} Assume that there are total $m$ data points in the datasets, and the total number of clients is $n$. At the beginning each client $i$ holds a single data point $\lfloor (i-1 + 1/2) \cdot \dfrac{m}{n}\rfloor$.
	\item  {\bf Data assignment.} First, initialize set $F=\{\}$. Next, iterate through all points except those assigned from the previous step. For each point we find the best client $i'\in [n] \backslash F$ to assign the point in a way that assignment of point to client $i'$ maximizes $\Lvar \eqdef \LQMsq - \LAMsq$. Once the client already has $\lceil \dfrac{m}{n} \rceil$ data points assigned to it, the client is added to the set $F$. 
\end{enumerate}

The set $F$ in the last step guarantees each client will have $\dfrac{m}{n}$ data points. In general, this is a heuristic greedy strategy that approximately maximizes $\Lvar$ under the constraint that each client has the same amount of data points equal to $\lfloor \dfrac{m}{n} \rfloor$. Due to its heuristic nature, the Algorithm does not provide deep guarantees, but it was good enough for our experiments.

\clearpage
\addtocounter{adjsection}{1}
\section{Additional Experiments}
\label{ch3:app:exp-additional-experiments}

In this section, we present additional experiments for comparison \algname{EF21-W}, \algname{EF21-W-PP}, \algname{EF21-W-SGD} with their vanilla versions. We applied these algorithms in a series of synthetically generated convex and non-convex optimization problems and for training \modelname{logistic regression} with non-convex regularized with using several \dataname{LIBSVM} datasets \citep{chang2011libsvm}. While carrying out additional experiments we will use three quantities. These quantities have already been mentioned in the main part, but we will repeat them here:
\begin{equation*}
	\begin{split}
		\LQM &\eqdef {\color{red} \sqrt{\avein L_i^2}}, \\
		\LAM &\eqdef {\color{blue} \avein L_i}, \\
		\Lvar &\eqdef \LQMsq - \LAMsq = \dfrac{1}{n} \sum_{i=1}^{n} \left(L_i - \dfrac{1}{n} \sum_{i=1}^{n} L_i\right) ^2.
	\end{split}
\end{equation*}

The relationship between these quantities was discussed in Appendix~\ref{ch3:app:when-improved-analysis-better}. In our experiments, we used \compname{TopK} compressor. The \compname{TopK} compressor returns sparse vectors filled with zeros, except $K$ positions, which correspond to $K$ maximum values in absolute value and which are unchanged by the compressor. Even if this compressor breaks ties arbitrarily, it is possible to show that $\alpha = \nicefrac{K}{d}$.
The compressor parameter $\alpha$ is defined without considering properties of $f_i$. The quantities $\beta$, $\theta$, $\dfrac{\beta}{\theta}$ are derived from $\alpha$, and they do not depend on $L_{i}$.

\subsection{Additional experiments for {EF21}}

\begin{center}	
	\begin{figure*}[t]
		\centering
		\captionsetup[sub]{font=normalsize,labelfont={}}	
		\captionsetup[subfigure]{labelformat=empty}
		
		\begin{subfigure}[ht]{0.49\textwidth}
			\includegraphics[width=\textwidth]{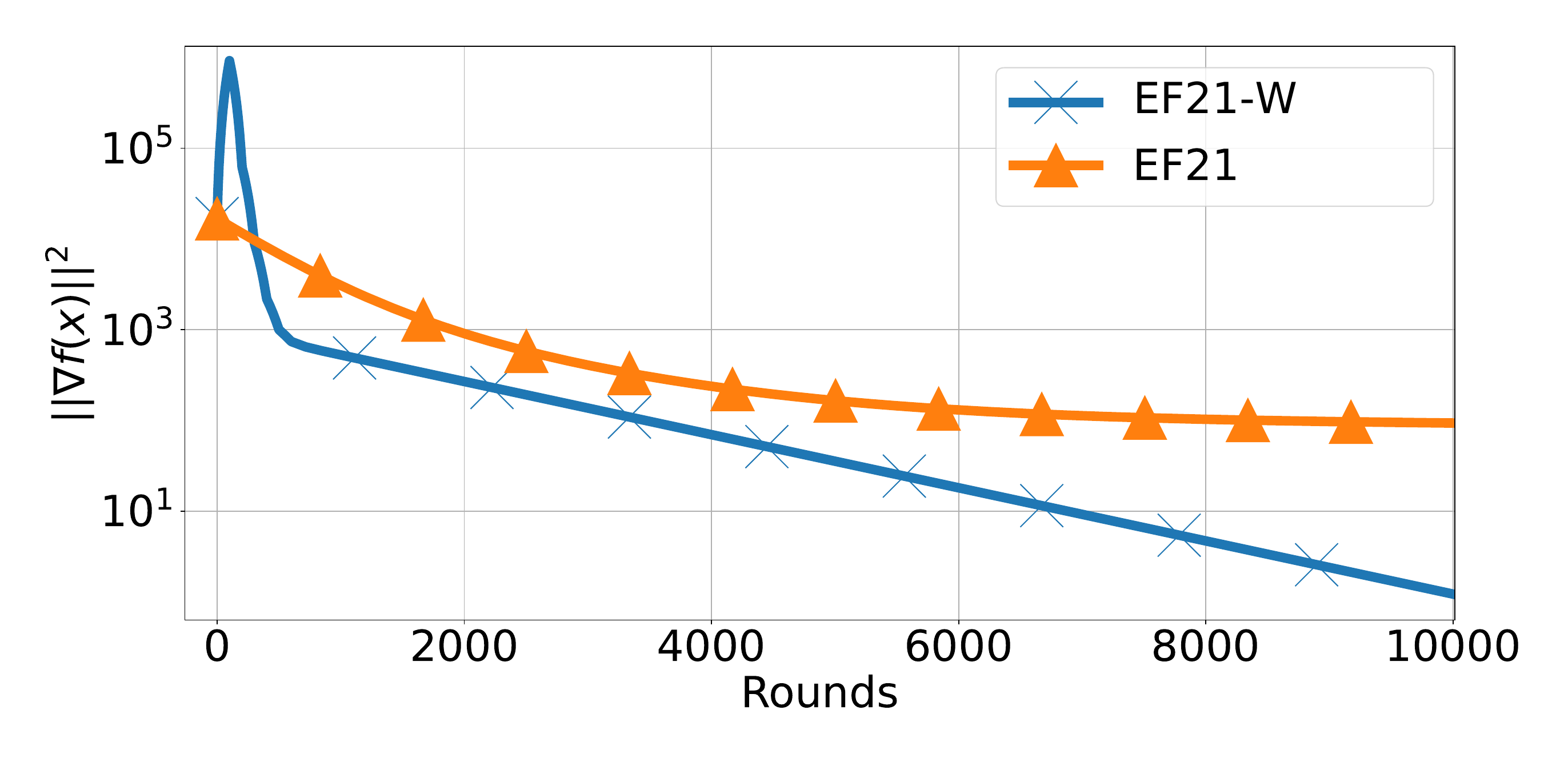} 
			\caption{(a) $\Lvar\approx 4.45 \cdot 10^6$}
		\end{subfigure}		
		\begin{subfigure}[ht]{0.49\textwidth}
			\includegraphics[width=\textwidth]{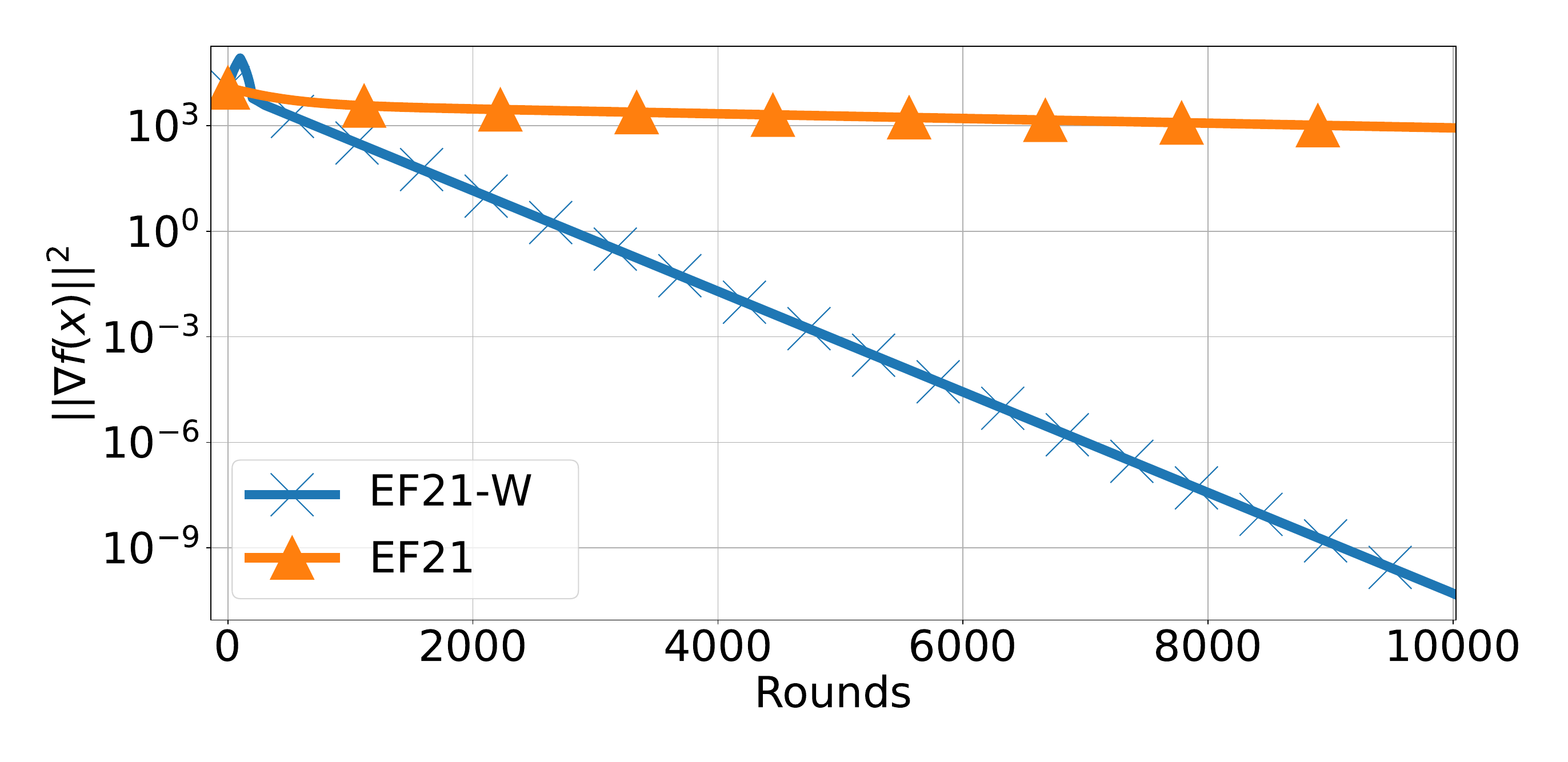}
			\caption{(b) $\Lvar\approx 1.97 \cdot 10^6$}
		\end{subfigure}
		
		\begin{subfigure}[ht]{0.49\textwidth}
			\includegraphics[width=\textwidth]{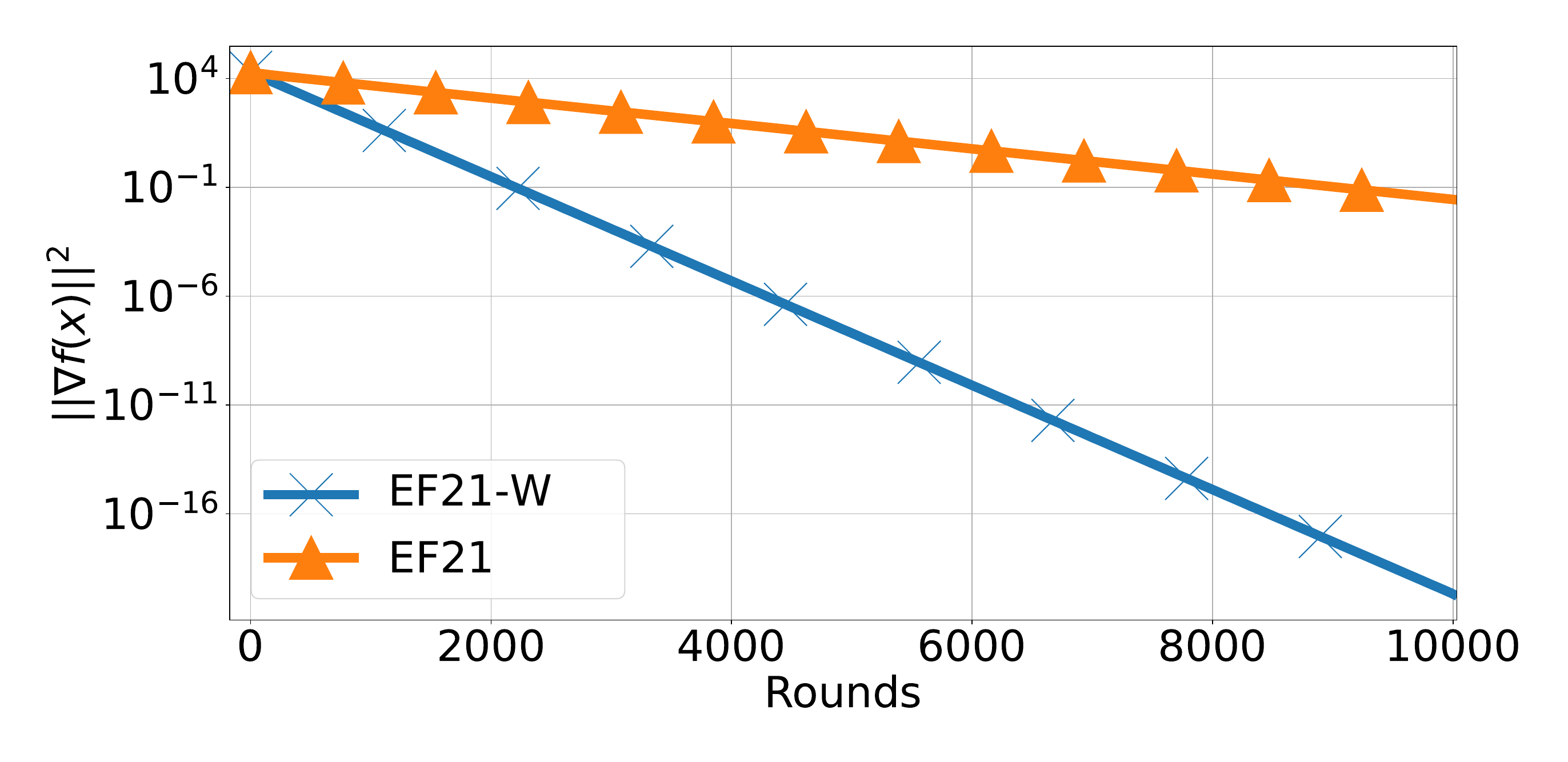}
			\caption{(c) $\Lvar\approx 1.08 \cdot 10^5$}
		\end{subfigure}		
		\begin{subfigure}[ht]{0.49\textwidth}
			\includegraphics[width=\textwidth]{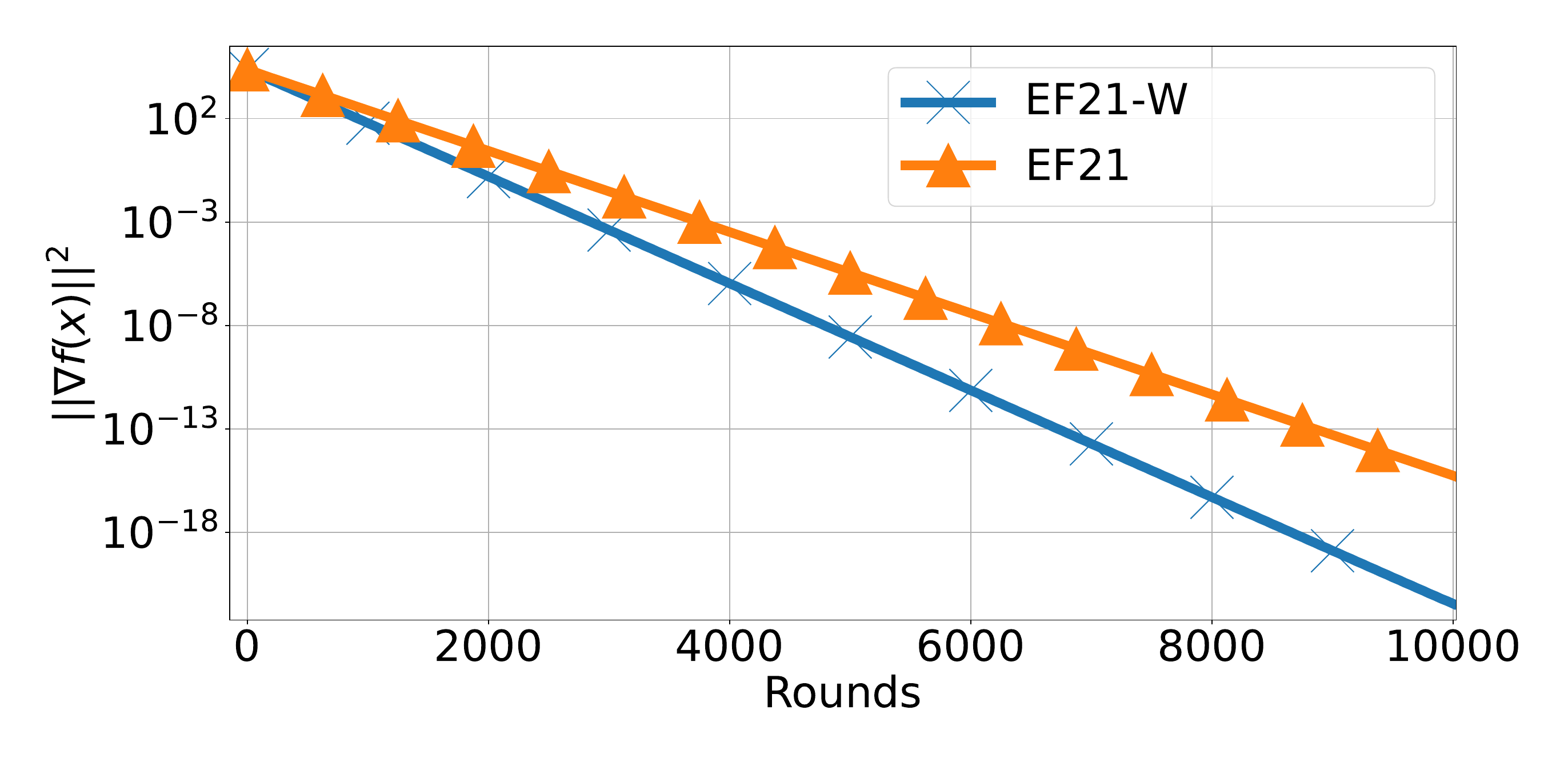}
			\caption{(d) $\Lvar\approx 5.42 \cdot 10^3$}
		\end{subfigure}
		
		\caption{{Convex smooth optimization. \algname{EF21} and \algname{EF21-W} with \compname{Top1} client compressor, $n=2\,000$, $d=10$. The objective function is constituted of $f_i(x)$ defined in Equation~\eqref{ch3:eq:linreg-cvx}. Regularization term $\lambda \nicefrac{\|x\|^2}{2}$, where $\lambda=0.01$. Theoretical step size. Full participation. Extra details are in Table~\ref{ch3:tbl:app-syn-ef21-cvx}.}}
		\label{ch3:fig:app-syn-ef21-cvx}
	\end{figure*}
\end{center}

\paragraph{Convex case with synthetic datasets.} 
We aim to solve optimization Problem~\eqref{ch3:eq:main_problem} in the case when  the functions $f_1,\dots,f_n$  are strongly convex. In particular, we work with
\begin{eqnarray}
	\label{ch3:eq:linreg-cvx}
	f_i(x) \eqdef \dfrac{1}{n_i} \norm{\mA_i x - {b_i}}^2 + \dfrac{\lambda}{2} \|x\|^2,
\end{eqnarray}
where $\lambda =0.01$. It can be shown that $L_{i} = \dfrac{2}{n_i} \lambda_{\max}({\mA_i}^\top \mA_i) + \lambda$. The result of experiments for training linear regression model with a convex regularized is presented in Figure~\ref{ch3:fig:app-syn-ef21-cvx}. The total number of rounds for simulation is $r=10,000$.  Instances of optimization problems were generated for values $L=50$, $\mu = 1$ with several values of $q,z$ with using the dataset generation schema described in Appendix~\ref{ch3:app:dataset-gen-synthetic}. The summary of derived quantities is presented in Table~\ref{ch3:tbl:app-syn-ef21-cvx}. We present several optimization problems to demonstrate the possible different relationships between $\LQM$ and $\LAM$. As we see from experiments, the \algname{EF21-W} is superior as the variance of $L_{i}$ tends to increase. The plots in Figure~\ref{ch3:fig:app-syn-ef21-cvx} (a--d) correspond to decreasing variance of $L_{i}$. As we see, as the variance of $L_i$ decreases, the difference between \algname{EF21-W} and \algname{EF21} also tends to decrease. Finally, \algname{EF21-W} is always at least as best as \algname{EF21}.

\begin{table}[h]
	\footnotesize
	\begin{center}
		\caption{{Convex Optimization experiment in Figure~\ref{ch3:fig:app-syn-ef21-cvx}. Quantities that define theoretical step size.}}
		\label{ch3:tbl:app-syn-ef21-cvx}
		\begin{tabular}{|c||c|c|c|c|c|c|c|c|c|c|}
			\hline
			Tag & $L$ & $q$ & $z$ & $\Lvar$ & $\xi=\sqrt{\dfrac{\beta}{\theta}}$ & $\LQM$ & $\LAM$ & $\gamma_{\algnametiny{EF21}}$ & $\gamma_{\algnametiny{EF21-W}}$ \\
			\hline
			\hline
			(a) & $50$ & $1$ & $10^4$ & $4.45  \cdot  10^6$ & $18.486$ & $2111.90$ & $52.04$ & $2.55 \cdot 10^{-5}$ & $9.87 \cdot 10^{-4}$ \\
			\hline
			(b) & $50$ & $1$ & $10^3$ & $1.97  \cdot  10^6$ & $18.486$ & $1408.49$ & $63.56$ & $3.83 \cdot 10^{-5}$ & $8.16  \cdot 10^{-4}$ \\
			\hline
			(c) & $50$ & $1$ & $10^2$ & $1.08  \cdot 10^5$ & $18.486$ & $339.34$ & $80.97$ & $1.58 \cdot 10^{-4}$ & $6.46  \cdot 10^{-4}$ \\
			\hline
			(d) & $50$ & $0.8$ & $1$ & $5.42  \cdot  10^3$ & $18.486$ & $112.51$ & $85.03$ & $4.69 \cdot 10^{-4}$ & $6.16  \cdot  10^{-4}$\\
			\hline
		\end{tabular}
	\end{center}
\end{table}

\paragraph{Non-convex case with synthetic datasets.} 
We aim to solve optimization Problem~\eqref{ch3:eq:main_problem} in the case when  the functions $f_1,\dots,f_n$  are non-convex. In particular, we work with \begin{eqnarray}
	\label{ch3:eq:ncvx-linear-reg}
	f_i(x) \eqdef \dfrac{1}{n_i} \norm{\mA_i x - {b_i}}^2 + \lambda \cdot \sum_{j=1}^{d} \dfrac{x_j^2}{x_j^2 + 1}.
\end{eqnarray}

\begin{center}	
	\begin{figure*}[t]
		\centering
		\captionsetup[sub]{font=normalsize,labelfont={}}	
		\captionsetup[subfigure]{labelformat=empty}
		
		\begin{subfigure}[ht]{0.49\textwidth}
			\includegraphics[width=\textwidth]{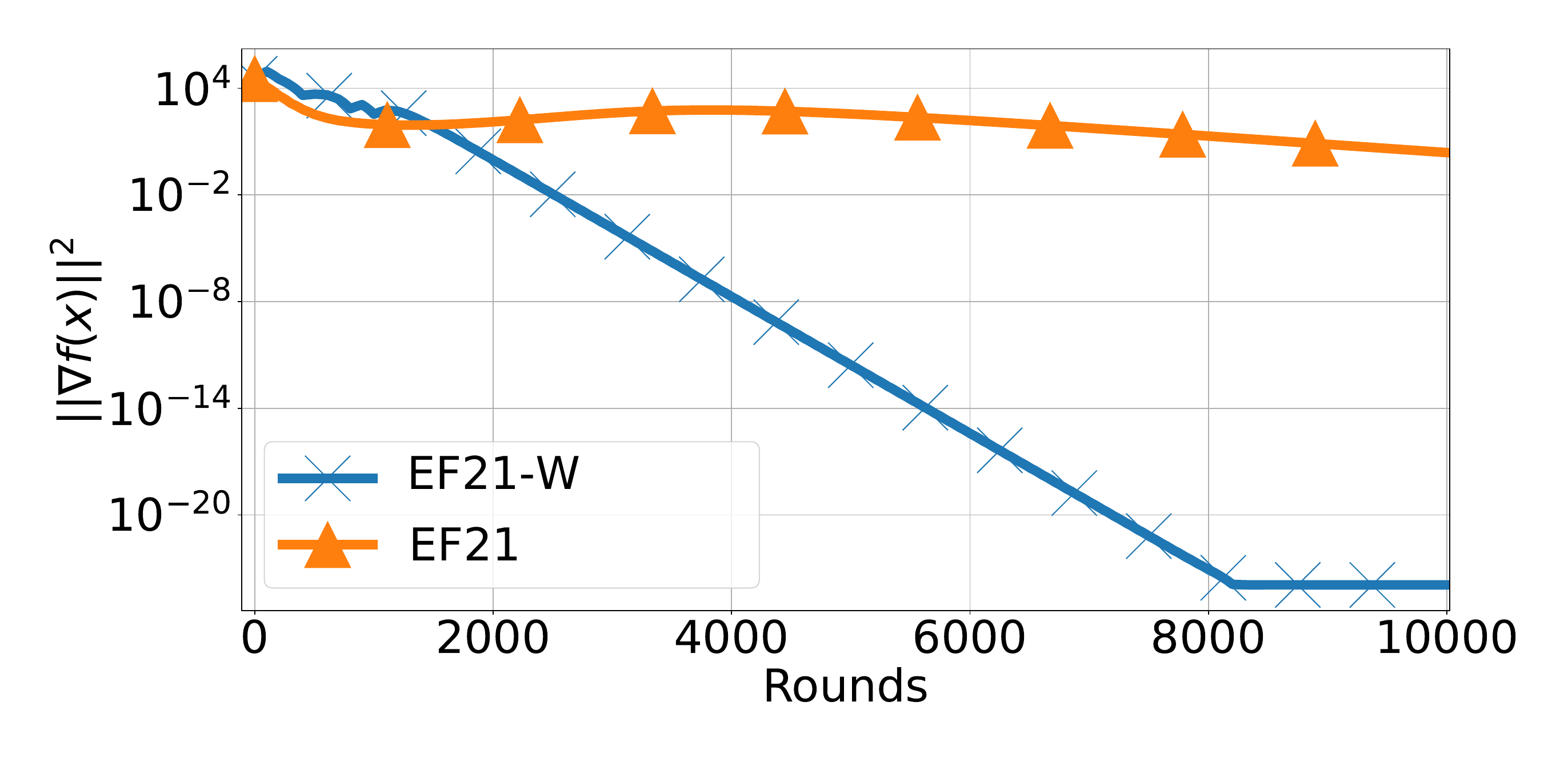} 
			\caption{(a) $\Lvar \approx 4.45 \cdot 10^6$}
		\end{subfigure}		
		\begin{subfigure}[ht]{0.49\textwidth}
			\includegraphics[width=\textwidth]{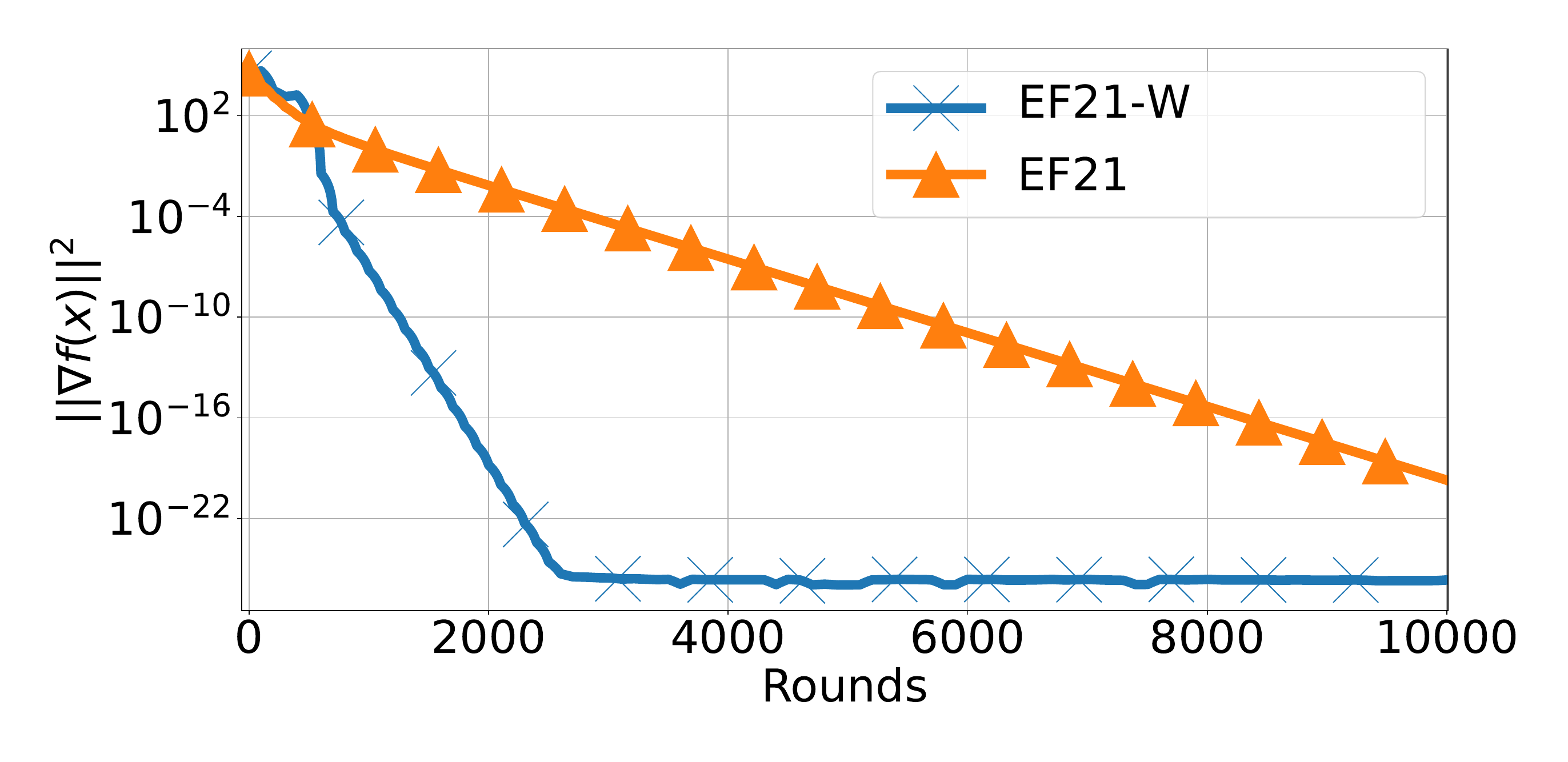} 
			\caption{(b) $\Lvar \approx 1.97  \cdot 10^6$}
		\end{subfigure}
		
		\begin{subfigure}[ht]{0.49\textwidth}
			\includegraphics[width=\textwidth]{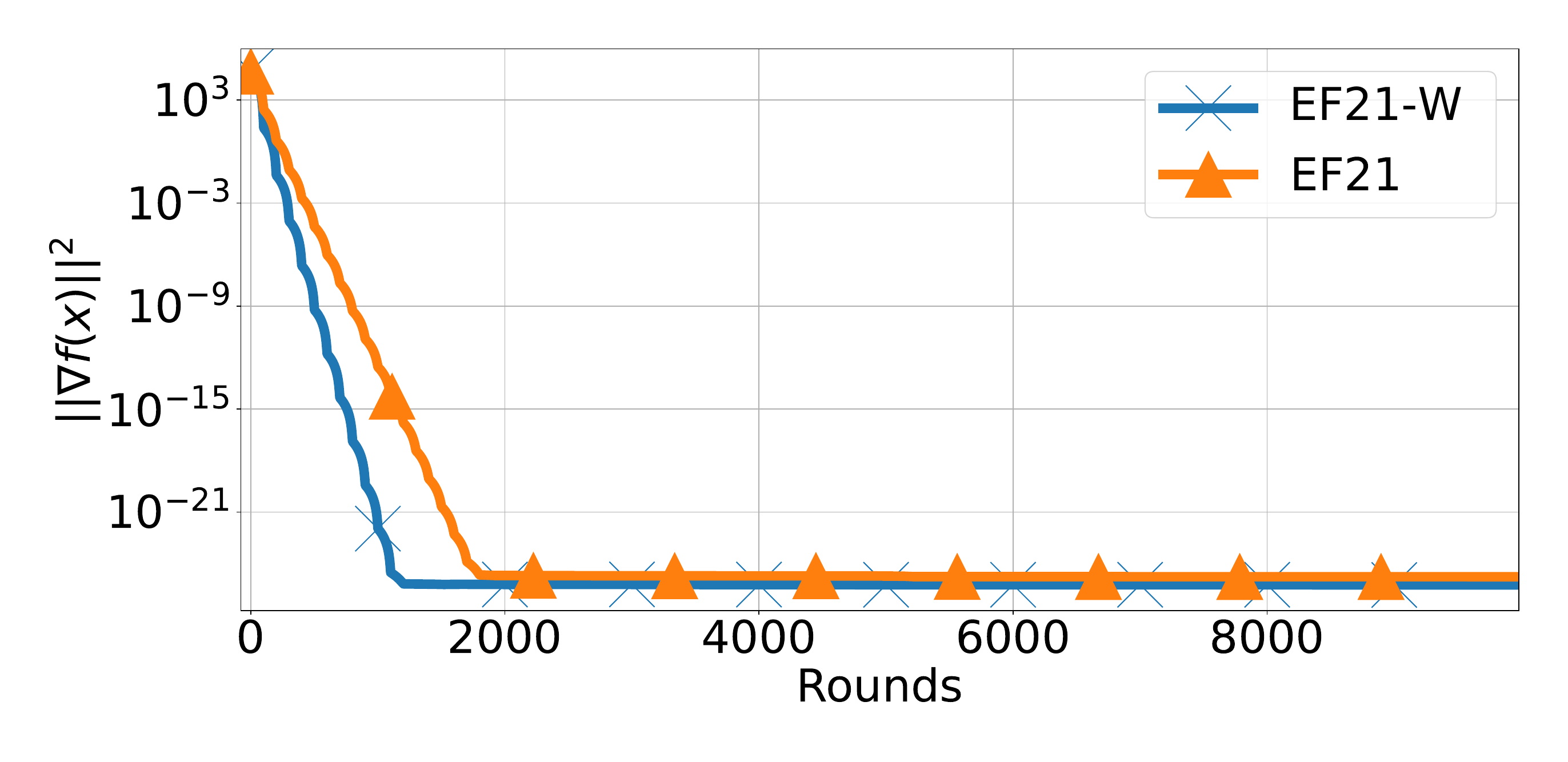} 
			\caption{(c) $\Lvar \approx 1.08  \cdot  10^5$}
		\end{subfigure}		
		\begin{subfigure}[ht]{0.49\textwidth}
			\includegraphics[width=\textwidth]{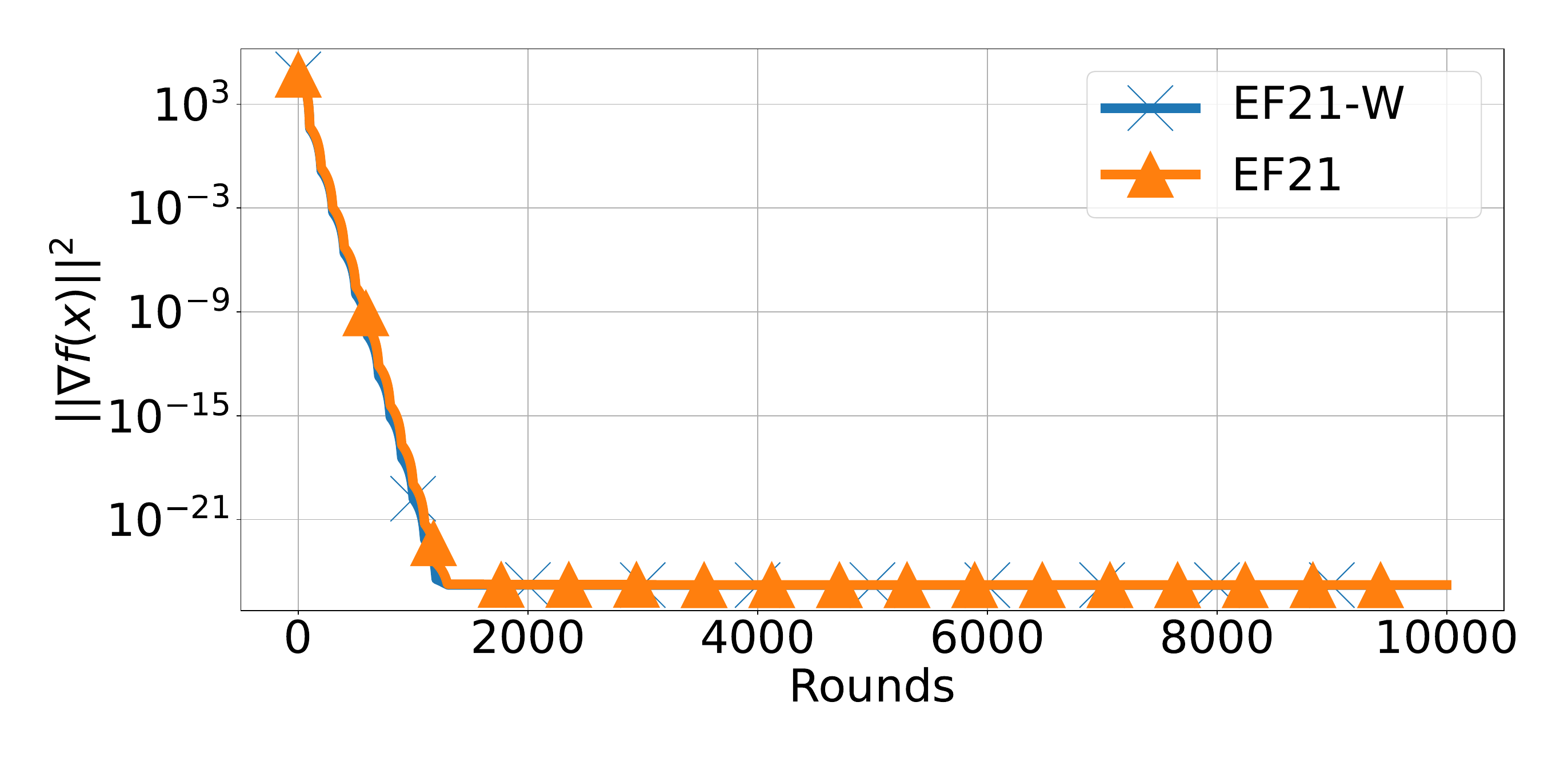} 
			\caption{(d) $\Lvar \approx 5.42  \cdot  10^3$}
		\end{subfigure}
		
		\caption{{Non-Convex smooth optimization. \algname{EF21} and \algname{EF21-W} with \compname{Top1} client compressor, $n=2,000$, $d=10$. The objective function is constituted of $f_i(x)$ defined in Equation~\eqref{ch3:eq:ncvx-linear-reg}.
				Regularization term $\lambda \sum_{j=1}^{d} \dfrac{x_j^2}{x_j^2 + 1}$, with $\lambda = 100$. Theoretical step size. Full client participation. Extra details are in Table~\ref{ch3:tbl:app-syn-ef21-ncvx}.}}
		\label{ch3:fig:app-syn-ef21-ncvx}
	\end{figure*}
\end{center}

The result of experiments for training linear regression model with a non-convex regularization is presented in Figure~\ref{ch3:fig:app-syn-ef21-ncvx}. The regularization coefficient $\lambda=100$. Instances of optimization problems were generated for values $L=50, \mu = 1$ and several values of $q,z$ for employed 
dataset generation schema from Appendix~\ref{ch3:app:dataset-gen-synthetic}.

The summary of derived quantities is presented in Table~\ref{ch3:tbl:app-syn-ef21-ncvx}.  We present various instances of optimization problems to demonstrate the different relationships between $\LQM$ and $\LAM$. As we see in the case of small variance of $L_{i}$ algorithm \algname{EF21-W} is at least as best as standard \algname{EF21}.

\begin{table}[h]
	\footnotesize
	\begin{center}
		\caption{{Non-Convex optimization experiment in Figure~\ref{ch3:fig:app-syn-ef21-ncvx}. Quantities that define theoretical step size.}}				
		\label{ch3:tbl:app-syn-ef21-ncvx}
		\begin{tabular}{|c||c|c|c|c|c|c|c|c|c|c|}
			\hline
			Tag & $L$ & $q$ & $z$ & $\Lvar$ & $\xi=\sqrt{\dfrac{\beta}{\theta}}$ & $\LQM$ & $\LAM$ & $\gamma_{\algnametiny{EF21}}$ & $\gamma_{\algnametiny{EF21-W}}$ \\
			\hline
			\hline
			(a) & $50$ & $1$ & $10^4$ & $4.45  \cdot 10^6$ & $18.486$ & $2126.25$ & $252.035$ & $2.52 \cdot 10^{-5}$ & $2.03  \cdot  10^{-4}$ \\
			\hline
			(b) & $50$ & $1$ & $10^3$ & $1.97 \cdot 10^6$ & $18.486$ & $1431.53$ & $263.55$ & $3.74 \cdot 10^{-5}$ & $1.95 \cdot 10^{-4}$ \\
			\hline
			(c) & $50$ & $1$ & $10^2$ & $1.08 \cdot 10^5$ & $18.486$ & $433.05$ & $280.958$ & $1.21 \cdot 10^{-4}$ & $1.83  \cdot  10^{-4}$ \\
			\hline
			(d) & $50$ & $0.8$ & $1$ & $5.42  \cdot 10^3$ & $18.486$ & $294.39$ & $285.022$ & $1.17 \cdot 10^{-4}$ & $1.81  \cdot 10^{-4}$\\
			\hline
		\end{tabular}
	\end{center}
\end{table}

\paragraph{Non-convex \modelname{logistic regression} on benchmark datasets.} We aim to solve optimization Problem~\eqref{ch3:eq:main_problem} in the case when the functions $f_1,\dots,f_n$ are non-convex. In particular, we work with \modelname{logistic regression} with a non-convex robustification regularization term:
\begin{eqnarray}
	\label{ch3:eq:ncvx-log-reg}
	f_i(x) \eqdef \dfrac{1}{n_i} \sum_{j=1}^{n_i} \log \left(1+\exp({-y_{ij} \cdot a_{ij}^{\top} x})\right) + \lambda \cdot \sum_{j=1}^{d} \dfrac{x_j^2}{x_j^2 + 1},
\end{eqnarray}
where $ (a_{ij},  y_{ij}) \in \mathbb{R}^{d} \times \{-1,1\}$.

\begin{center}	
	\begin{figure*}[t]
		\centering
		\captionsetup[sub]{font=normalsize,labelfont={}}	
		\captionsetup[subfigure]{labelformat=empty}
		
		\begin{subfigure}[ht]{0.49\textwidth}
			\includegraphics[width=\textwidth]{ch3imgs/ef21-vc-figs/expreal/fig1-09JULY23-w1a.pdf}
			\caption{{(a) \dataname{W1A}}}
		\end{subfigure}
		\begin{subfigure}[ht]{0.49\textwidth}
			\includegraphics[width=\textwidth]{ch3imgs/ef21-vc-figs/expreal/fig2-09JULY23-w2a.pdf}
			\caption{{(b) \dataname{W2A}}}
		\end{subfigure}
		
		\begin{subfigure}[ht]{0.49\textwidth}
			\includegraphics[width=\textwidth]{ch3imgs/ef21-vc-figs/expreal/fig3-09JULY23-w3a.pdf} 
			\caption{{(c) \dataname{W3A}}}
		\end{subfigure}
		\begin{subfigure}[ht]{0.49\textwidth}
			\includegraphics[width=\textwidth]{ch3imgs/ef21-vc-figs/expreal/fig4-09JULY23-mushrooms}
			\caption{{(d) \dataname{MUSHROOMS}}} 
		\end{subfigure}
		
		\begin{subfigure}[ht]{0.49\textwidth}
			\includegraphics[width=\textwidth]{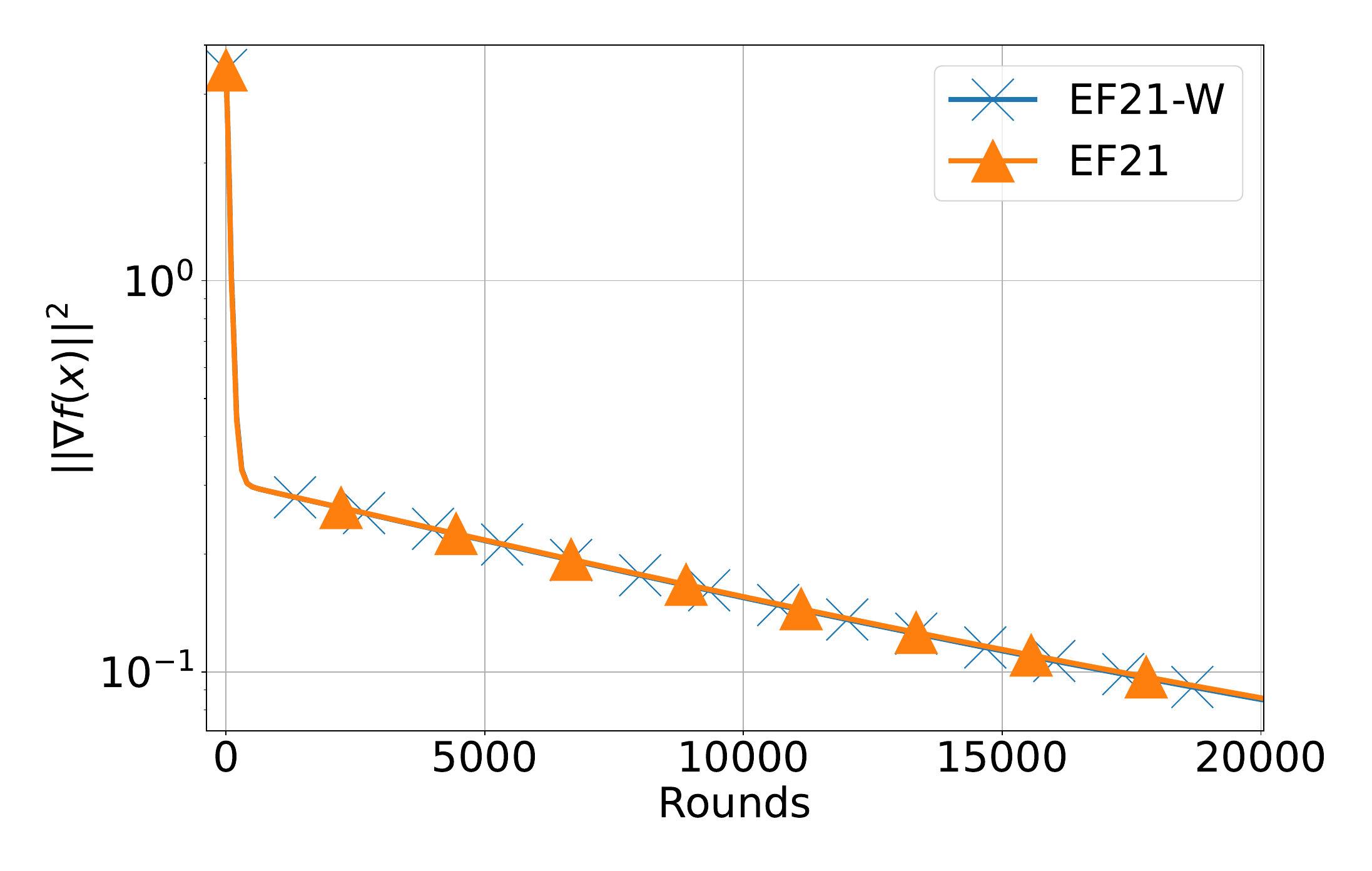} 
			\caption{{(e) \dataname{SPLICE}}}
		\end{subfigure}
		\begin{subfigure}[ht]{0.49\textwidth}
			\includegraphics[width=\textwidth]{ch3imgs/ef21-vc-figs/expreal/fig6-09JULY23-phishing.pdf} 
			\caption{{(f) \dataname{PHISHING}}} 
		\end{subfigure}

		\caption{{Non-Convex \modelname{logistic regression}: comparison of \algname{EF21} and \algname{EF21-W}. The used compressor is \compname{Top1}. The number of clients $n=1,000$. Regularization term $\lambda \sum_{j=1}^{d} \dfrac{x_j^2}{x_j^2 + 1}$, with $\lambda=0.001$. Theoretical step size. Full client participation. The objective function is constituted of $f_i(x)$ defined in Equation~\eqref{ch3:eq:ncvx-log-reg}. Extra details are in Table~\ref{ch3:tbl:app-real-ef21-ncvx}.}}
	\label{ch3:fig:app-real-ef21-ncvx}
	\end{figure*}
\end{center}

\begin{center}	
\begin{figure*}[t]
\centering
\captionsetup[sub]{font=normalsize,labelfont={}}	
\captionsetup[subfigure]{labelformat=empty}

\begin{subfigure}[ht]{0.8\textwidth}
\includegraphics[width=\textwidth]{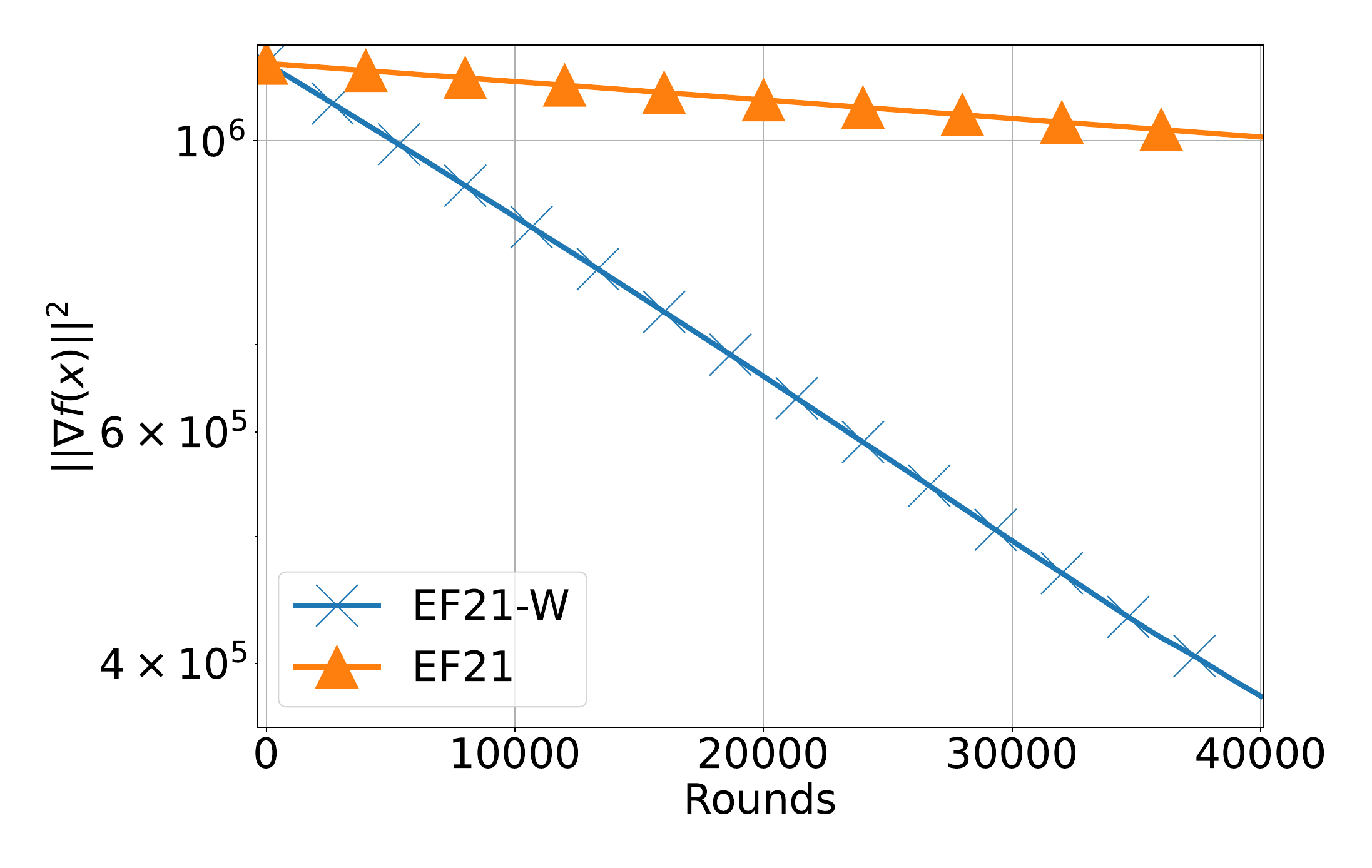} 
\caption{{(g) \dataname{AUSTRALIAN}}}
\end{subfigure}

\caption{{Non-Convex \modelname{logistic regression}: comparison of the performance of standard \algname{EF21} and \algname{EF21-W}. The used compressor is \compname{Top1}. The number of clients $n=200$. Regularization term $\lambda \sum_{j=1}^{d} \dfrac{x_j^2}{x_j^2 + 1}$, with $\lambda=1,000$. Theoretical step size. The objective function is constituted of $f_i(x)$ defined in Equation~\eqref{ch3:eq:ncvx-log-reg}. Extra details are in Table \ref{ch3:tbl:app-real-ef21-ncvx}.}}
\label{ch3:fig:app-real-ef21-ncvx-aus}
\end{figure*}
\end{center}

We used several \dataname{LIBSVM} datasets \citep{chang2011libsvm} for our benchmarking purposes. The results are presented in Figures~\ref{ch3:fig:app-real-ef21-ncvx},~\ref{ch3:fig:app-real-ef21-ncvx-aus}. The important quantities for these instances of optimization problems are summarized in Table~\ref{ch3:tbl:app-real-ef21-ncvx}. From Figure \ref{ch3:fig:app-real-ef21-ncvx} (a, b, c), we can observe that for these datasets, the \algname{EF21-W} is better, and this effect is observable in practice. From these examples, we can observe that $12.5$K rounds of \algname{EF21-W} corresponds to only $10$K rounds of \algname{EF21}. This improvement is essential for Federated Learning, in which both communication rounds and communication information during a round represent the main bottlenecks and are the subject of optimization. Figure \ref{ch3:fig:app-real-ef21-ncvx} (d, e, f) demonstrate that sometimes the \algname{EF21-W} can have practical behavior close to \algname{EF21}, even if there is an improvement in step size (For exact values of step size see Table~\ref{ch3:tbl:app-real-ef21-ncvx}). The experiment on \dataname{AUSTRALIAN} datasets is presented in Figure~\ref{ch3:fig:app-real-ef21-ncvx-aus}. This example demonstrates that in this \dataname{LIBSVM} benchmark datasets, the relative improvement in the number of rounds for \algname{EF21-W} compared to \algname{EF21} is considerable. For example $40$K rounds of \algname{EF21} corresponds to $5$K rounds of \algname{EF21-W} in terms of attainable $\| \nabla f(x^t)\|^2$.

\begin{table}[h]
	\footnotesize
	\begin{center}
		\tiny
		\caption{{Non-Convex optimization experiments in Figures \ref{ch3:fig:app-real-ef21-ncvx}, \ref{ch3:fig:app-real-ef21-ncvx-aus}. Derived quantities that define theoretical step size.}}
		\label{ch3:tbl:app-real-ef21-ncvx}
		\begin{tabular}{|c||c|c|c|c|c|c|c|c|c|c|}
			\hline
			Tag & $L$ & $\Lvar$ & $\xi=\sqrt{\dfrac{\beta}{\theta}}$ & $\LQM$ & $\LAM$ & $\gamma_{\algnametiny{EF21}}$ & $\gamma_{\algnametiny{EF21-W}}$ \\
			\hline
			\hline
			\tiny{(a) W1A} & $0.781$ & $3.283$ & $602.49$ & $2.921$ & $2.291$ & $5.678  \cdot 10^{-4}$ & $7.237   \cdot  10^{-4}$ \\
			\hline
			\tiny{(b) W2A} & $0.784$ & $2.041$ & $602.49$ & $2.402$ & $1.931$ & $6.905  \cdot 10^{-4}$ & $8.589  \cdot 10^{-4}$ \\
			\hline
			\tiny{(c) W3A} & $0.801$ & $1.579$ & $602.49$ & $2.147$ & $1.741$ & $7.772  \cdot 10^{-4}$ & $9.523   \cdot  10^{-4}$ \\
			\hline
			\tiny{(d) MUSHROOMS} & $2.913$ & $5.05 \times10^{-1}$ & $226.498$ & $3.771$ & $3.704$ & $1.166  \cdot 10^{-3}$ & $1.187  \cdot 10^{-3}$\\
			\hline
			\tiny{(e) SPLICE} & $96.082$ & $2.23   \cdot  10^{2}$ & $122.497$ & $114.43$ & $113.45$ & $7.084  \cdot 10^{-5}$ & $7.14  \cdot 10^{-5}$\\
			\hline
			\tiny{(f) PHISHING} & $0.412$ & $9.2   \cdot  10^{-4}$ & $138.498$ & $0.429$ & $0.428$ & $1.670  \cdot 10^{-2}$ & $1.674   \cdot  10^{-2}$\\			
			\hline				
			\tiny{(g) AUSTRALIAN} & $3.96   \cdot  10^6 $ & $1.1   \cdot  10^{16}$ & $18.486$ & $3.35   \cdot  10^{7}$ & $3.96   \cdot  10^{6}$ & $9.733  \cdot 10^{-10}$ & $8.007   \cdot  10^{-9}$\\
			\hline				
		\end{tabular}
	\end{center}
\end{table}

\paragraph{Non-convex \modelname{logistic regression} with non-homogeneous compressor.}

\begin{center}	
\begin{figure*}[t]
\centering
\captionsetup[sub]{font=normalsize,labelfont={}}	
\captionsetup[subfigure]{labelformat=empty}

\begin{subfigure}[ht]{0.49\textwidth}
\includegraphics[width=\textwidth]{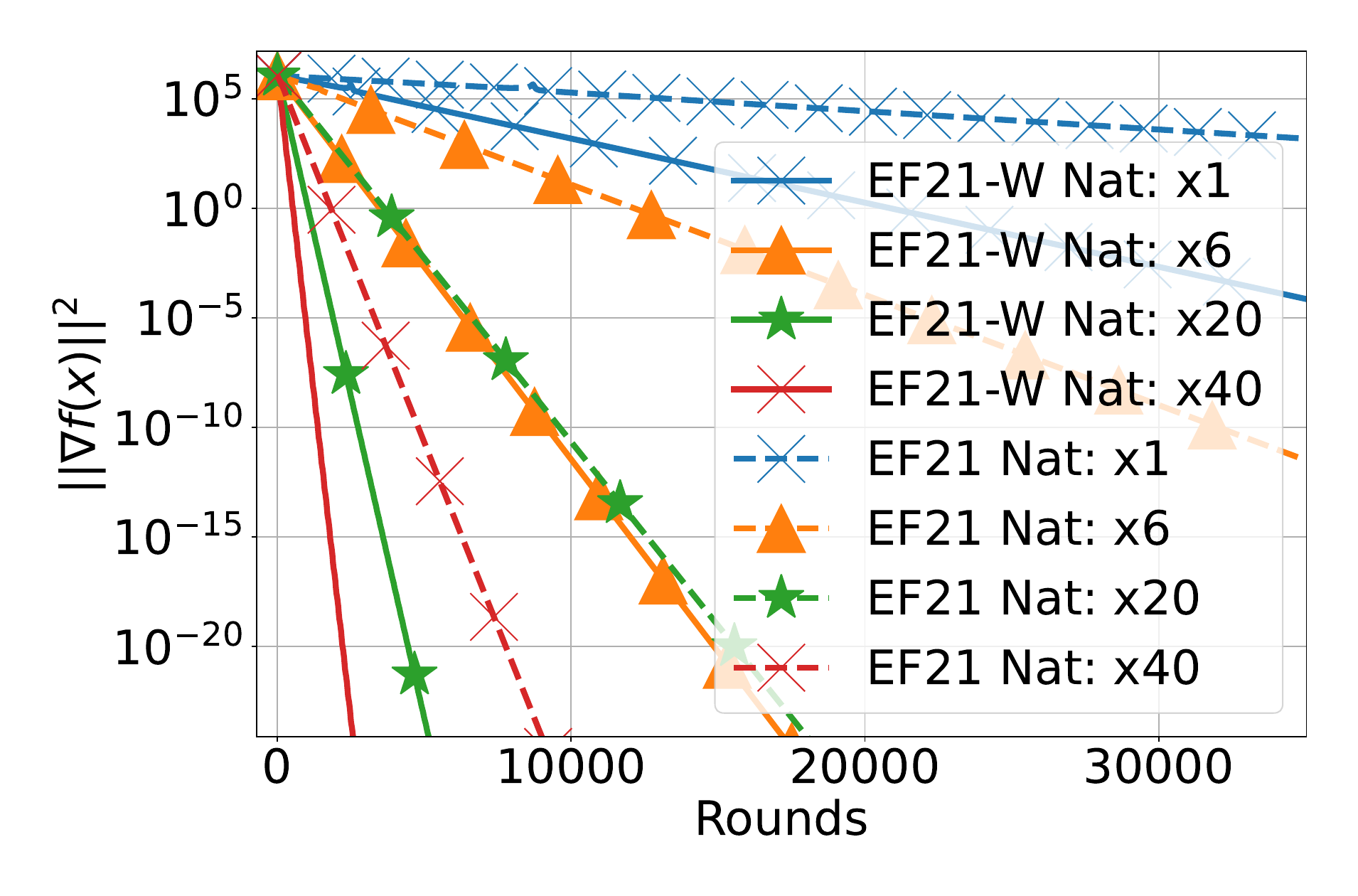} \caption{{(a)}}
\end{subfigure}
\begin{subfigure}[ht]{0.49\textwidth}
\includegraphics[width=\textwidth]{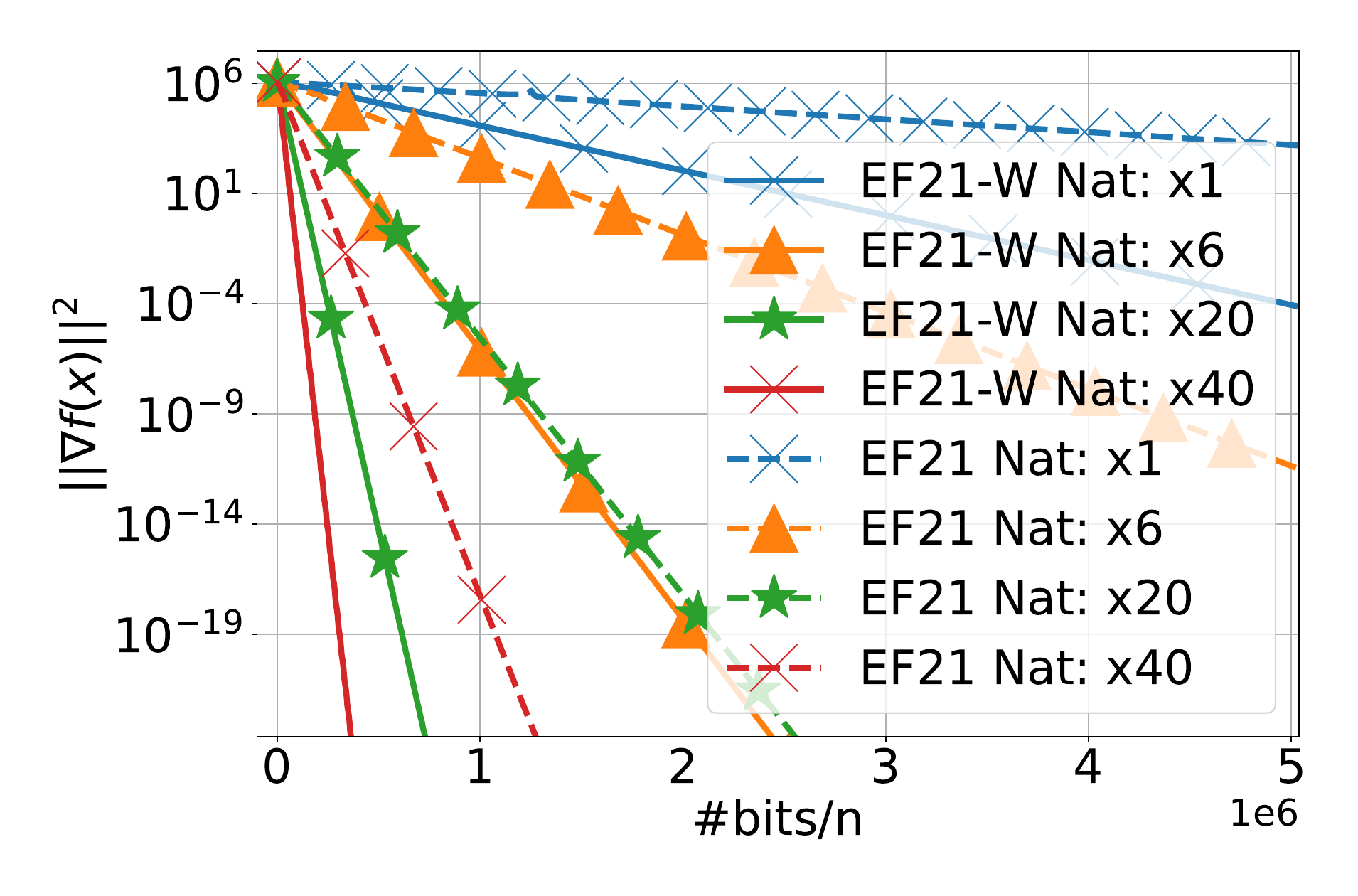} \caption{{(b)}}
\end{subfigure}

\caption{{Non-Convex \modelname{logistic regression}: comparison of the performance of standard \algname{EF21} and \algname{EF21-W}. The used compressor for \algname{EF21} and \algname{EF21-W} is \compname{Natural} compressor \citet{horvath2019natural}. The number of clients $n=200$. The objective function is constituted of $f_i(x)$ defined in Equation~\eqref{ch3:eq:ncvx-log-reg}. Regularization term $\lambda \sum_{j=1}^{d} \dfrac{x_j^2}{x_j^2 + 1}$, with $\lambda=1,000$. Multipliers of theoretical step size. Full participation. Computation format single precision (FP32). Dataset: \dataname{AUSTRALIAN}.}}
\label{ch3:fig:app-natual-ef21-ncvx}
\end{figure*}
\end{center}

In this supplementary experiment, we used \dataname{AUSTRALIAN} dataset \citep{chang2011libsvm} to train a \modelname{logistic regression} model with a non-convex sparsity-enhanced regularization term, as defined in Equation~\eqref{ch3:eq:ncvx-log-reg}. We employed the non-homogeneous \algname{Natural} compressor \citep{horvath2019natural}, an unbiased compressor satisfying Definition~\ref{ch3:eq:unbiased} with \( w = 1/8 \). This compressor randomly rounds the exponential part and discards the mantissa when using the IEEE 754 Standard~\citep{IEEE754-2008} floating-point representation. In FP32 format, only 9 bits per scalar are transmitted to the master, while the remaining 23 mantissa bits are entirely dropped. Figure~\ref{ch3:fig:app-natual-ef21-ncvx} presents the experimental results. We fine-tuned the theoretical step size by multiplying it by a constant and observed that \algname{EF21-W} consistently outperforms \algname{EF21} across all considered step size multipliers. \algname{EF21-W} effectively leverages unbiased non-homogeneous compressors, extending its advantages beyond applications to homogeneous compressors. Finally, the theoretically increased step size in \algname{EF21-W} in this experiment does not fully reflect the potential for significantly larger practical step size increases such as a $\times 40$ multiplier, which remains an open question for future research.

\clearpage

\subsection{Additional experiments for {EF21-W-PP}}

\paragraph{Convex case with synthetic datasets.} 

\begin{center}	
\begin{figure*}[t]
\centering
\captionsetup[sub]{font=normalsize,labelfont={}}	
\captionsetup[subfigure]{labelformat=empty}

\begin{subfigure}[ht]{0.49\textwidth}
\includegraphics[width=\textwidth]{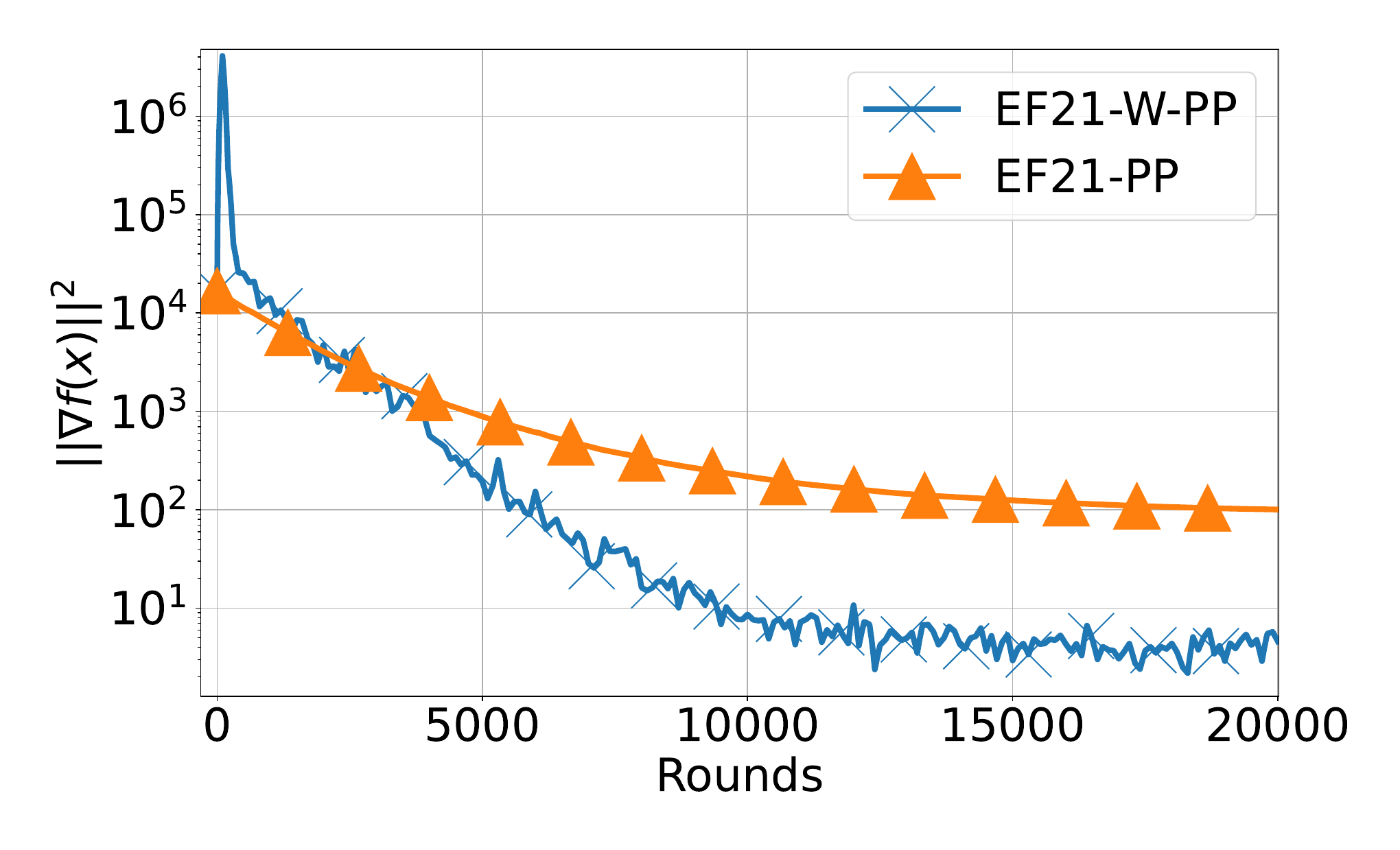} \caption{ (a) $\Lvar = 4.45   \cdot  10^6$ }
\end{subfigure}
\begin{subfigure}[ht]{0.49\textwidth}
\includegraphics[width=\textwidth]{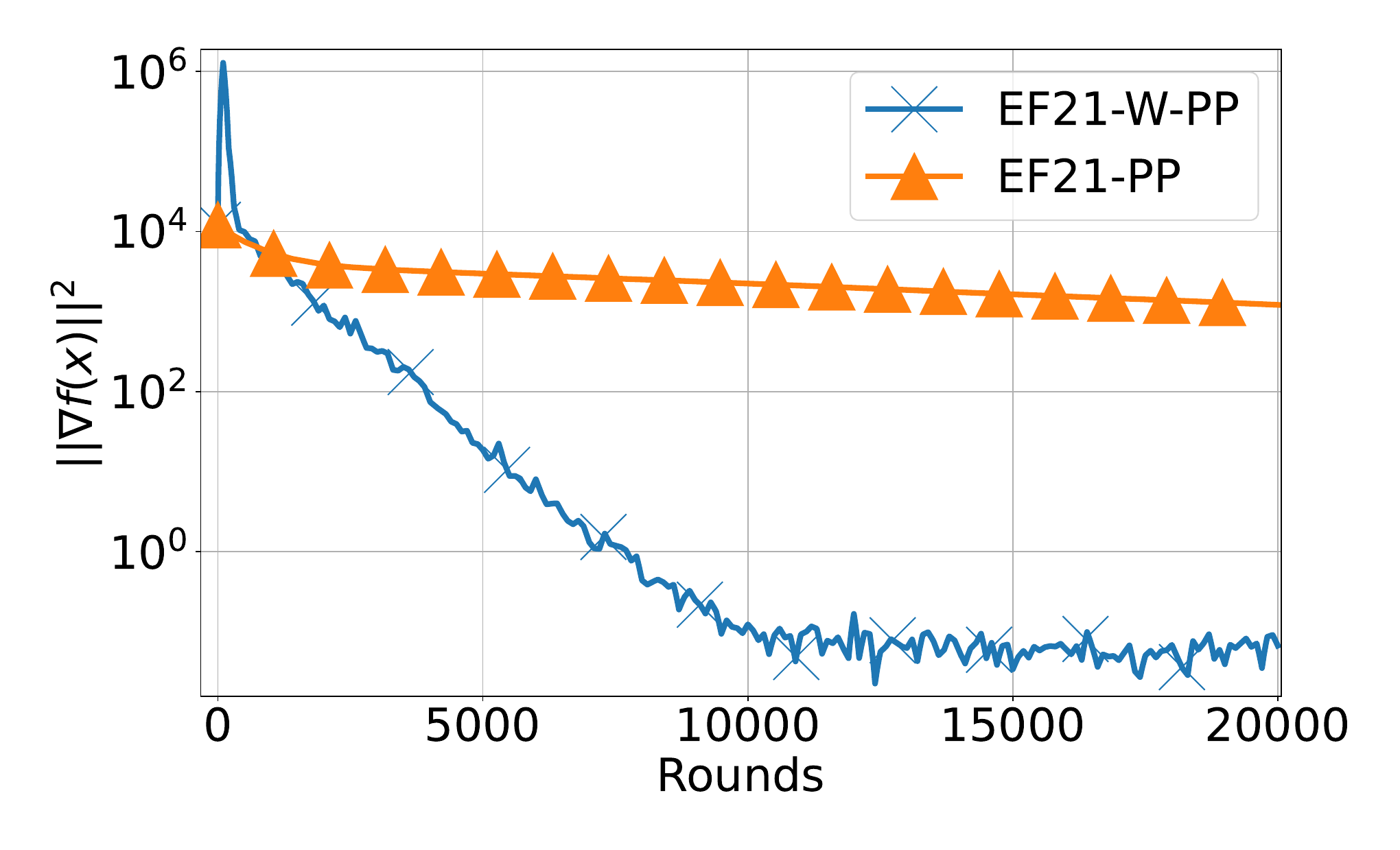} \caption{ (b) $\Lvar = 1.97   \cdot  10^6$ }
\end{subfigure}

\begin{subfigure}[ht]{0.49\textwidth}
\includegraphics[width=\textwidth]{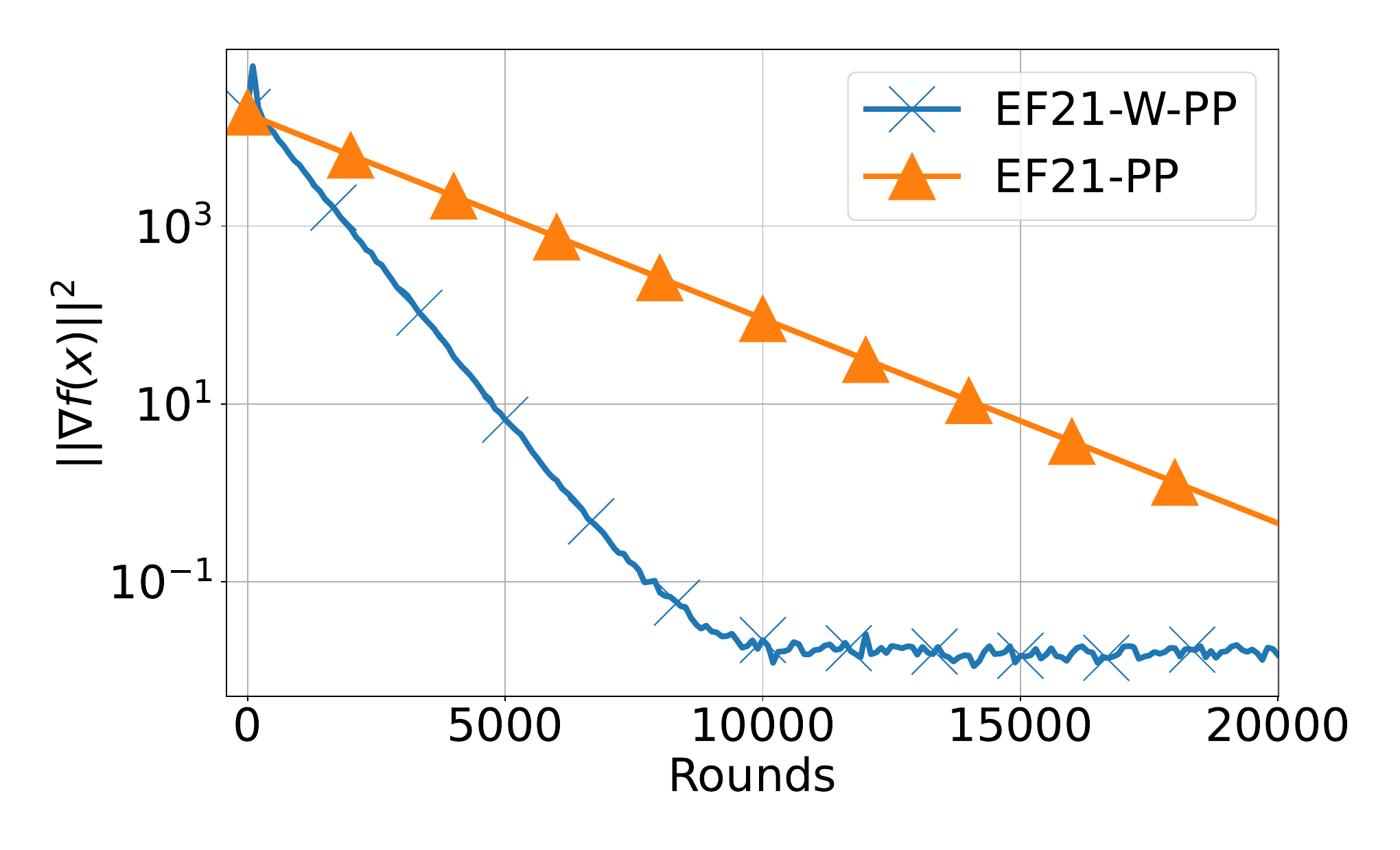} \caption{ (c) $\Lvar = 1.08   \cdot  10^5$ }
\end{subfigure}

\caption{{Convex smooth optimization. \algname{EF21-PP} and \algname{EF21-W-PP} with \compname{Top1} client compressor, $n=2\,000$, $d=10$. The objective function is constituted of $f_i(x)$ defined in Equation~\eqref{ch3:eq:cvx-linear-reg}. Regularization term $\lambda \dfrac{\|x\|^2}{2}$, $\lambda=0.01$. Theoretical step size. The objective function is constitute of $f_i(x)$ defined in Equation~\eqref{ch3:eq:cvx-linear-reg}.	Each client participates in each round with probability $p_i=0.5$. Extra details are in Table~\ref{ch3:tbl:app-syn-ef21-pp-cvx}.}}
\label{ch3:fig:app-syn-ef21-pp-cvx}
\end{figure*}
\end{center}

\begin{table}[h]
\footnotesize
\begin{center}
\caption{{Convex optimization experiment in Figure~\ref{ch3:fig:app-syn-ef21-pp-cvx}. Derived quantities that define theoretical step size.}}
\label{ch3:tbl:app-syn-ef21-pp-cvx}
\begin{tabular}{|c||c|c|c|c|c|c|c|c|c|c|}
\hline
Tag & $L$ & $q$ & $z$ & $\Lvar$ & $\sqrt{\dfrac{\beta}{\theta}}$ & $\LQM$ & $\LAM$ & $\gamma_{\algnametiny{EF21-PP}}$ & $\gamma_{\algnametiny{EF21-W-PP}}$ \\
\hline
\hline
(a) & $50$ & $1$ & $10^4$ & $4.45   \cdot  10^6$ & $18.486$ & $2111.90$ & $52.04$ & $2.55  \cdot 10^{-5}$ & $9.87   \cdot  10^{-4}$ \\
\hline
(b) & $50$ & $1$ & $10^3$ & $1.97   \cdot  10^6$ & $18.486$ & $1408.49$ & $63.56$ & $3.83  \cdot 10^{-5}$ & $8.16   \cdot  10^{-4}$ \\
\hline
(c) & $50$ & $1$ & $10^2$ & $1.08   \cdot  10^5$ & $18.486$ & $339.34$ & $80.97$ & $1.58  \cdot 10^{-4}$ & $6.46   \cdot  10^{-4}$ \\
\hline
\end{tabular}
\end{center}
\end{table}

We aim to solve optimization Problem~\eqref{ch3:eq:main_problem} in the case when 
the functions $f_1,\dots,f_n$ are strongly convex. In particular, we choose: \begin{eqnarray}
\label{ch3:eq:cvx-linear-reg}
f_i(x) \eqdef \dfrac{1}{n_i} \norm{\mA_i x - {b_i}}^2 + \dfrac{\lambda}{2} \|x\|^2.
\end{eqnarray}

In this synthetic experiment, we have used the maximum allowable step size suggested by the theory of \algname{EF21-PP} and for the proposed \algname{EF21-W-PP} algorithm. The initial  gradient estimators have been initialized as $g_i^0 = \nabla f_i (x^0)$ for all $i$. The number of clients in simulation $n=2000$, dimension of optimization problem $d=10$, number of samples per client $n_i=10$, and number of communication rounds is $r=10,000$. For both \algname{EF21-PP} and \algname{EF21-W-PP} clients we used \compname{Top1} biased contractile compressor. In our experiment, each client's participation in each communication round is governed by an independent Bernoulli trial which takes $p_i=0.5$. The result of experiments for training linear regression model with a convex regularizer is presented in Figure~\ref{ch3:fig:app-syn-ef21-pp-cvx}. The regularization constant was chosen to be $\lambda=0.01$. Instances of optimization problems were generated for values $L=50$, $\mu = 1$ with several values of $q$ and $z$. The summary of derived quantities is presented in Table~\ref{ch3:tbl:app-syn-ef21-pp-cvx}. We present several optimization problems to demonstrate the possible different relationships between $\LQM$ and $\LAM$. As we see from experiments, the \algname{EF21-W-PP} is superior as the variance of $L_{i}$ tends to increase. As we can observe \algname{EF21-W-PP} is always at least as best as \algname{EF21-PP}.

\paragraph{Non-convex \modelname{logistic regression} on benchmark datasets.}  We provide additional numerical experiments in which we compare \algname{EF21-PP} and \algname{EF21-W-PP} for solving \eqref{ch3:eq:main_problem}. We address the problem of training a binary classifier via a logistic model on several \dataname{LIBSVM} datasets \citep{chang2011libsvm} with non-convex regularization. We consider the case when the functions $f_1,\dots,f_n$ are non-convex; in particular, we set
$f_i(x)$ as follows:
\begin{eqnarray}
\label{ch3:eq:ncvx-log-reg-2}
f_i(x) \eqdef \dfrac{1}{n_i} \sum_{j=1}^{n_i} \log \left(1+\exp({-y_{ij} \cdot a_{ij}^{\top} x})\right) + \lambda \cdot \sum_{j=1}^{d} \dfrac{x_j^2}{x_j^2 + 1}, 
\end{eqnarray}
where $(a_{ij},  y_{ij}) \in \mathbb{R}^{d} \times \{-1,1\}$.

\begin{center}	
\begin{figure*}[t]
\centering
\captionsetup[sub]{font=normalsize,labelfont={}}	
\captionsetup[subfigure]{labelformat=empty}

\begin{subfigure}[ht]{0.49\textwidth}
\includegraphics[width=\textwidth]{ch3imgs/ef21-vc-figs-pp/expreal/w1a.pdf} \caption{ {(a) \dataname{W1A} }}
\end{subfigure}		
\begin{subfigure}[ht]{0.49\textwidth}
\includegraphics[width=\textwidth]{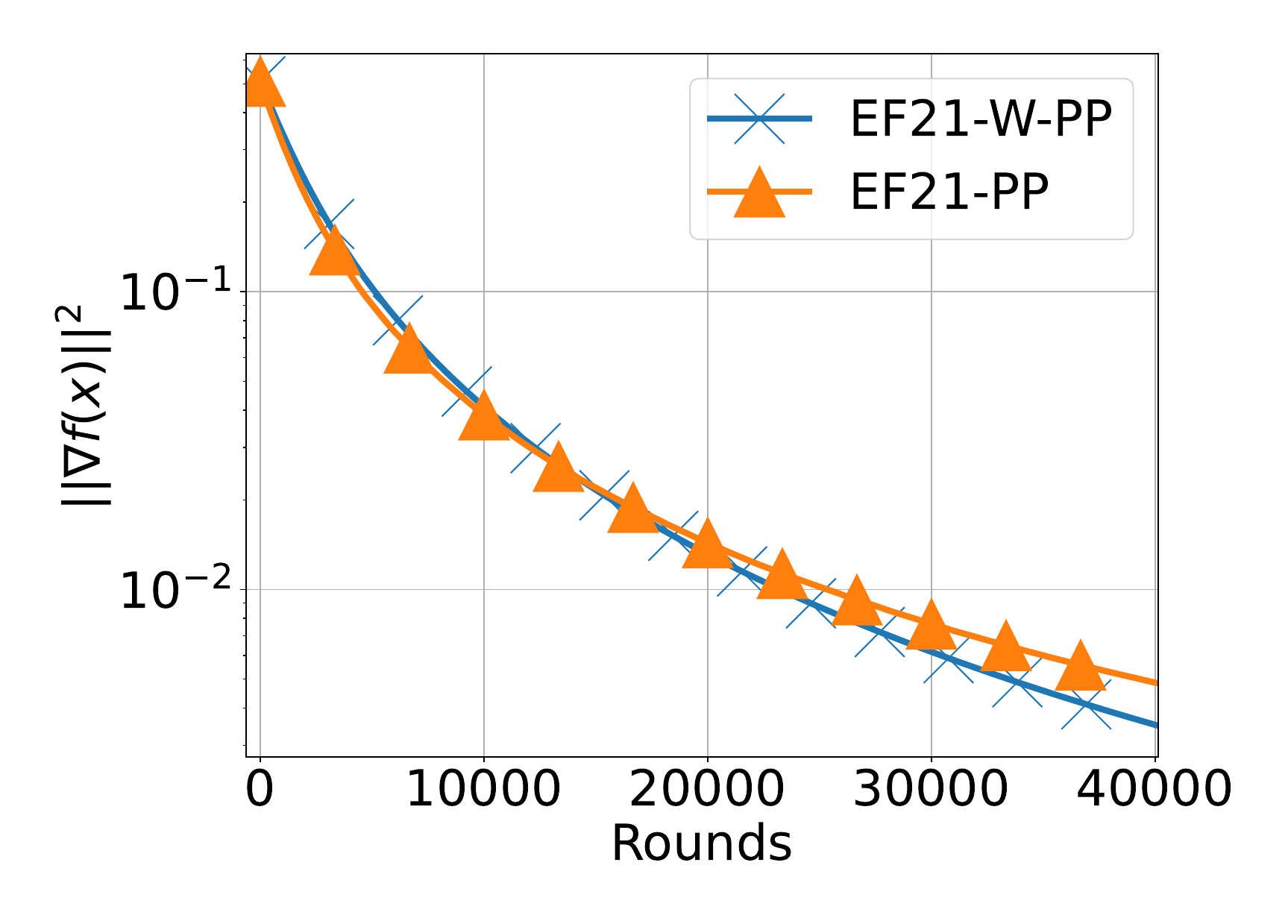} \caption{ { (b) \dataname{W2A} }}
\end{subfigure} \\
\begin{subfigure}[ht]{0.49\textwidth}
\includegraphics[width=\textwidth]{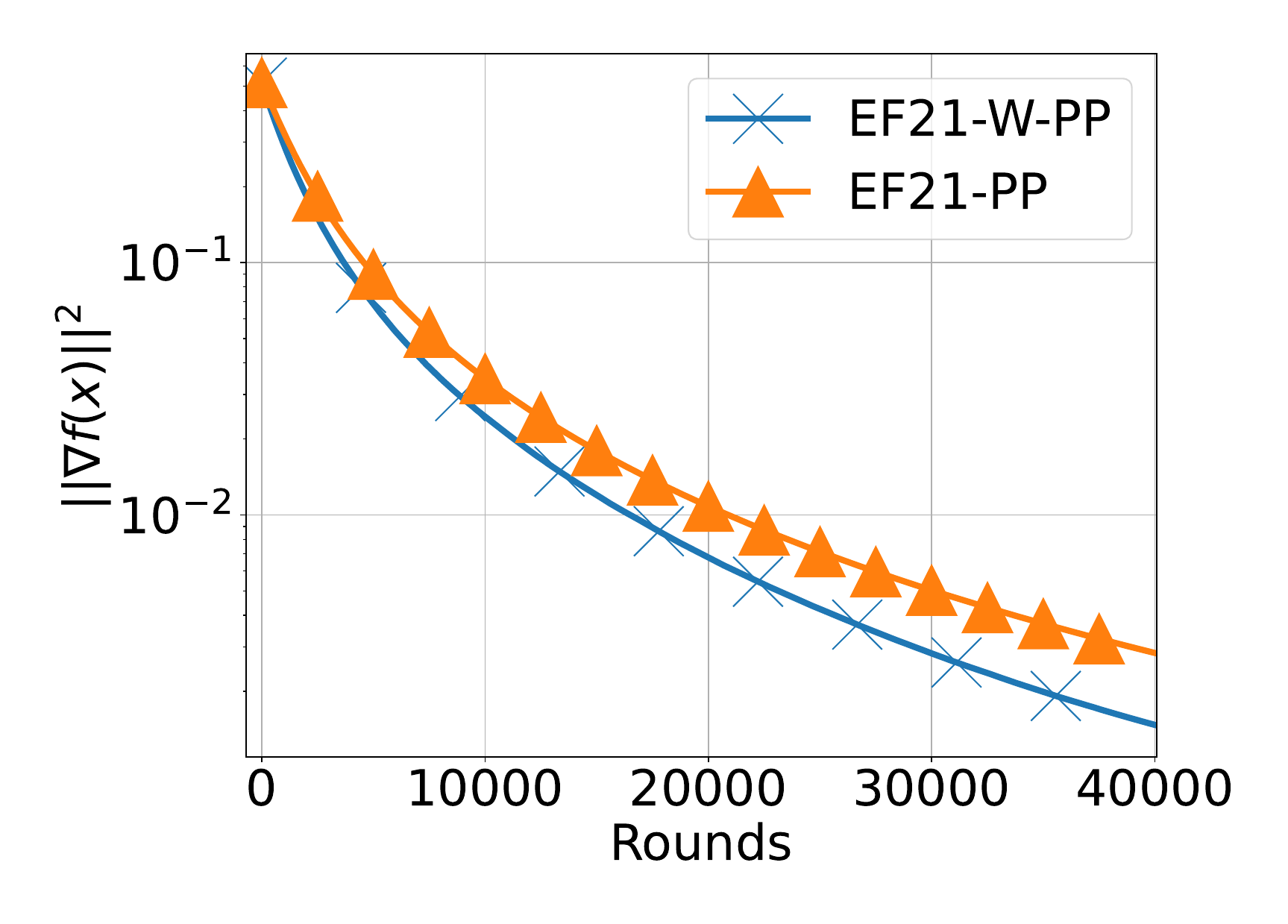} \caption{ {(c) \dataname{W3A} }}
\end{subfigure}		
\begin{subfigure}[ht]{0.49\textwidth}
\includegraphics[width=\textwidth]{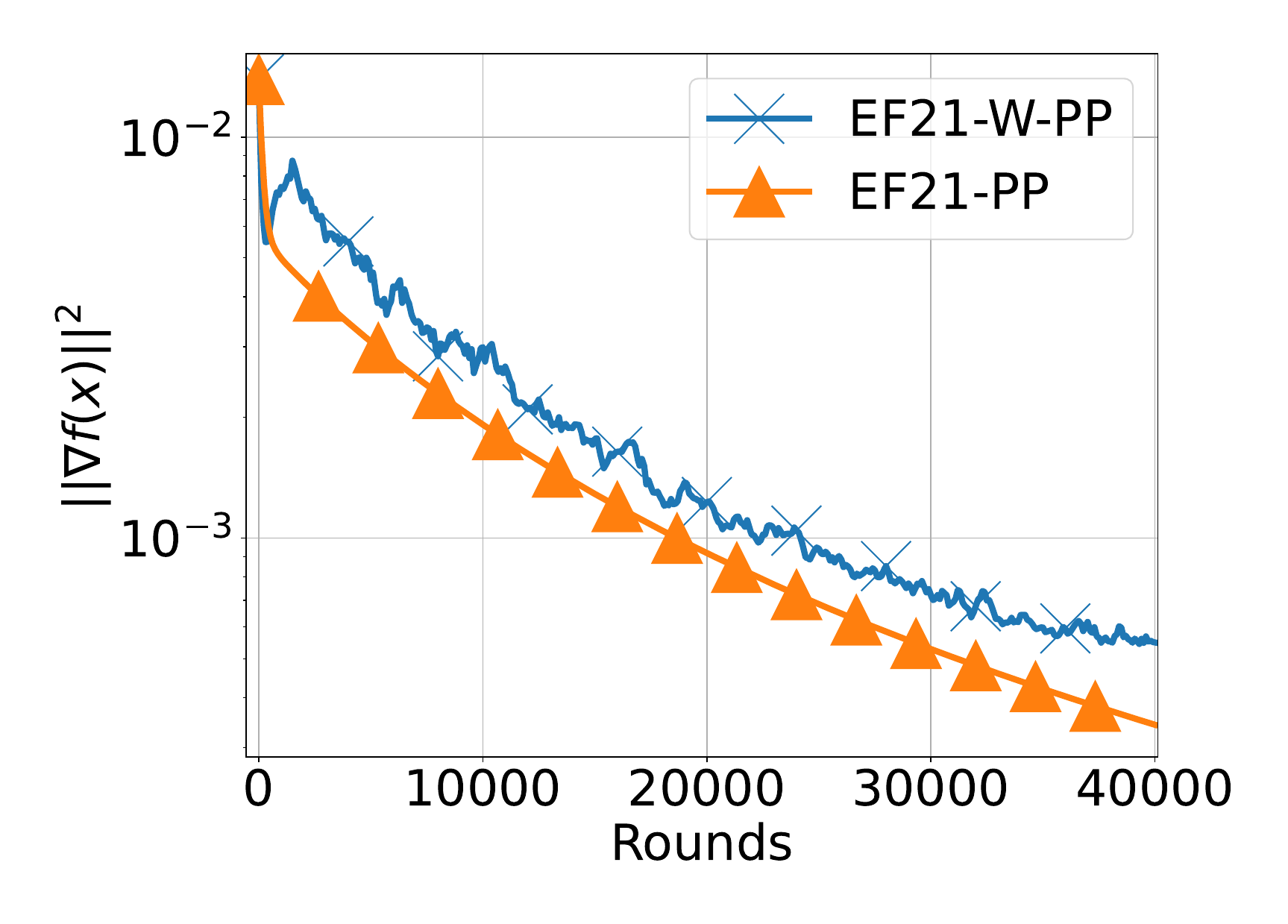} \caption{ {(d) \dataname{PHISHING}}}
\end{subfigure}

\caption{{Non-Convex \modelname{logistic regression}: comparison of \algname{EF21-PP} and \algname{EF21-W-PP}. The used compressor is \compname{Top1}. The number of clients $n=1,000$. Regularization term $\lambda \sum_{j=1}^{d} \dfrac{x_j^2}{x_j^2 + 1}$, $\lambda=0.001$.  Theoretical step size. Each client participates in each round with probability $p_i=0.5$. The objective function is constitute of $f_i(x)$ defined in Equation~\eqref{ch3:eq:ncvx-log-reg-2}. Extra details are in Table~\ref{ch3:tbl:app-real-ef21-pp-ncvx}.}}
\label{ch3:fig:app-real-ef21-pp-ncvx}
\end{figure*}
\end{center}

\begin{center}	
\begin{figure*}[t]
\centering
\captionsetup[sub]{font=normalsize,labelfont={}}	
\captionsetup[subfigure]{labelformat=empty}

\begin{subfigure}[ht]{0.8\textwidth}
\includegraphics[width=\textwidth]{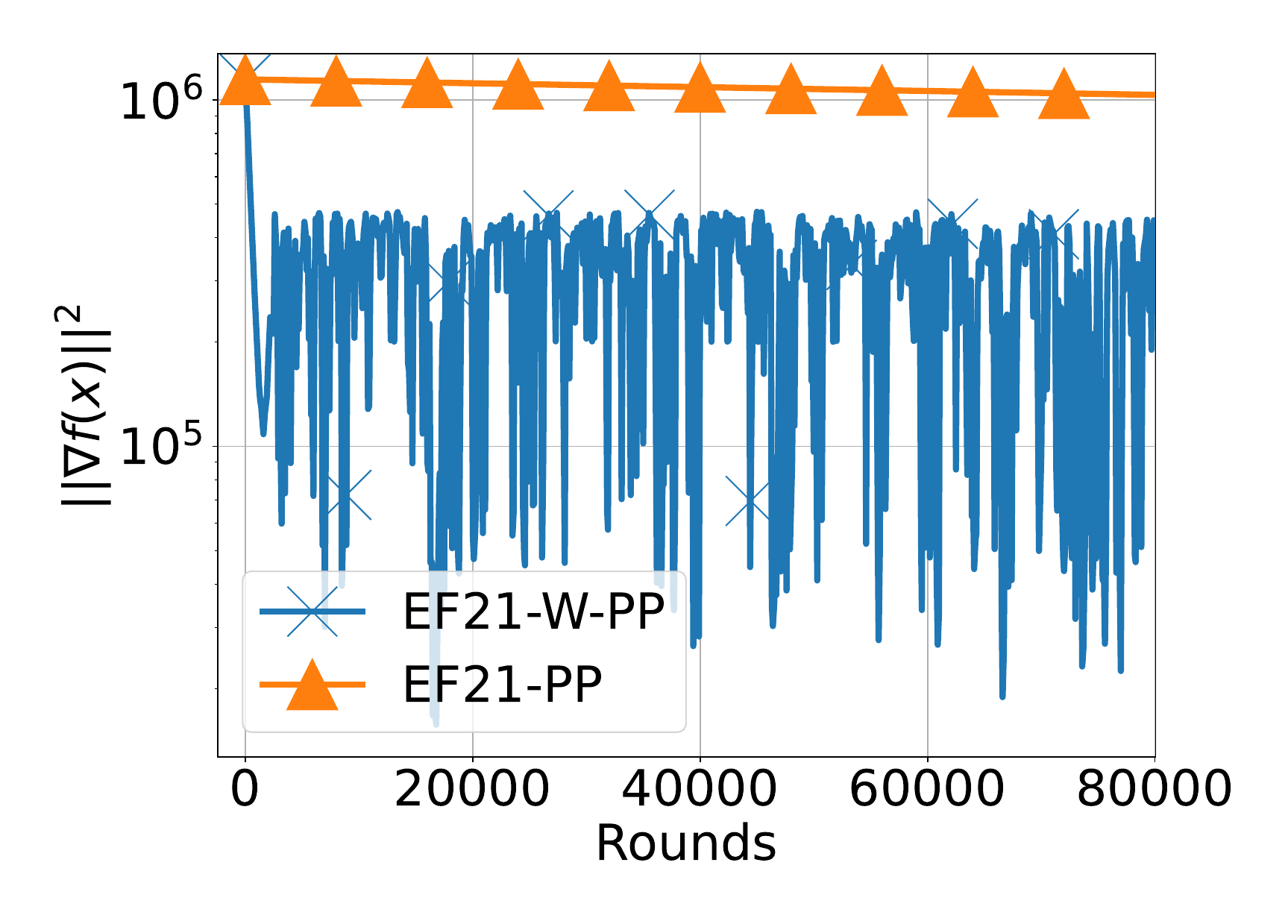} \caption{{(e) \dataname{AUSTRALIAN}}}
\end{subfigure}

\caption{{Non-Convex \modelname{logistic regression}: comparison of \algname{EF21-PP} and \algname{EF21-W-PP}. The used compressor is \compname{Top1}. The number of clients $n=200$. Regularization term $\lambda \sum_{j=1}^{d} \dfrac{x_j^2}{x_j^2 + 1}$, with $\lambda=1,000$.  Theoretical step size. Each client participates in each round with probability $p_i=0.5$. The objective function is constitute of $f_i(x)$ defined in Equation~\eqref{ch3:eq:ncvx-log-reg-2}. Extra details are in Table~\ref{ch3:tbl:app-real-ef21-pp-ncvx}.}}
\label{ch3:fig:app-real-ef21-pp-ncvx-aus}
\end{figure*}
\end{center}

We conducted distributed training of a \modelname{logistic regression} model on \dataname{W1A}, \dataname{W2A}, \dataname{W3A}, \dataname{PHISHING}, and \dataname{AUSTRALIAN} datasets with non-convex regularization. The initial gradient estimators are set $g_i^0 = \nabla f_i(x^0)$ for all $ i \in [n]$. For comparison of \algname{EF21-PP} and \algname{EF21-W-PP}, we used the largest step size allowed by theory. We used the dataset shuffling strategy described in Appendix~\ref{ch3:app:dataset-shuffling-for-libsvm}. The results are presented in Figure~\ref{ch3:fig:app-real-ef21-pp-ncvx} and Figure~\ref{ch3:fig:app-real-ef21-pp-ncvx-aus}. The important quantities for these instances of optimization problems are summarized in Table~\ref{ch3:tbl:app-real-ef21-pp-ncvx}.

\begin{table}[h]
\footnotesize
\begin{center}
\caption{{Non-Convex optimization experiments in Figures \ref{ch3:fig:app-real-ef21-pp-ncvx}, \ref{ch3:fig:app-real-ef21-pp-ncvx-aus}. Quantities that define theoretical step size.}}
\label{ch3:tbl:app-real-ef21-pp-ncvx}
\begin{tabular}{|c||c|c|c|c|c|c|c|c|c|}
\hline
Tag & $L$ & $\Lvar$ & $\LQM$ & $\LAM$ & $\gamma_{\algnametiny{EF21-PP}}$ & $\gamma_{\algnametiny{EF21-W-PP}}$ \\
\hline
\hline
\tiny{(a) W1A} & $0.781$ & $3.283$ & $2.921$ & $2.291$ & $2.315  \cdot 10^{-4}$ & $2.95   \cdot  10^{-4}$ \\
\hline
\tiny{(b) W2A} & $0.784$ & $2.041$ & $2.402$ & $1.931$ & $2.816  \cdot 10^{-4}$ & $3.503  \cdot 10^{-4}$ \\
\hline
\tiny{(c) W3A} & $0.801$ & $1.579$ & $2.147$ & $1.741$ & $3.149   \cdot  10^{-4}$ & $3.884   \cdot  10^{-4}$ \\
\hline
\tiny{(d) PHISHING} & $0.412$ & $9.2   \cdot  10^{-4}$ & $0.429$ & $0.428$ & $6.806  \cdot 10^{-3}$ & $6.823   \cdot  10^{-3}$\\
\hline				
\tiny{(e) AUSTRALIAN} & $3.96   \cdot  10^6 $ & $1.1   \cdot  10^{16}$ & $3.35   \cdot  10^{7}$ & $3.96   \cdot  10^{6}$ & $3.876  \cdot 10^{-10}$ & $3.243   \cdot  10^{-9}$\\
\hline				
\end{tabular}
\end{center}
\end{table}

From Figure~\ref{ch3:fig:app-real-ef21-pp-ncvx} (a, b, c), we can observe that for these datasets, the \algname{EF21-W-PP} is better, and this effect is observable in practice and is not negligible. Figure \ref{ch3:fig:app-real-ef21-pp-ncvx} (d), demonstrate that sometimes \algname{EF21-W-PP} in terms of the full gradient at last iterate can have slightly worse behavior compared to \algname{EF21-PP}, even though theory allow more aggressive step size (For exact values of step size see Table~\ref{ch3:tbl:app-real-ef21-pp-ncvx}. The experiment on \dataname{AUSTRALIAN} dataset is presented in Figure~\ref{ch3:fig:app-real-ef21-pp-ncvx-aus}. This example demonstrates that in this \dataname{LIBSVM} benchmark datasets, the relative improvement in the number of rounds for \algname{EF21-W-PP} compared to \algname{EF21-PP} is considerable. The \algname{EF21-W-PP} exhibits more oscillation behavior in terms of $\|\nabla f(x^t)\|^2$ for \dataname{AUSTRALIAN} dataset, however as we can see observe in expectation $\|\nabla f(x^t)\|^2$ tends to decrease faster compare to \algname{EF21-PP}.

\subsection{Additional experiments for {EF21-W-SGD}}

The standard \algname{EF21-SGD} with the analysis described in Corollary 4 \citep{fatkhullin2021ef21} allows performing the optimization procedure with maximum allowable step size up to the factor of $2$ equal to:  
$$\gamma_{\text{\algnametiny{EF21-SGD}}} = \left(L + \sqrt{\dfrac{\hat{\beta_1}}{\hat{\theta}}} {\color{red}{\LQM}} \right)^{-1}.$$

In last expression quantities $\hat{\theta} = 1 - (1-\alpha)(1+s) (1+\nu)$, and $\hat{\beta_1} = 2(1- \alpha ) \left(1+ s \right)\left(s+\nu^{-1}\right)$. Improved analysis for \algname{EF21-W-SGD} allows to apply step size:
$$\gamma_{\text{\algnametiny{EF21-W-SGD}}} = \left(L + \sqrt{\dfrac{\hat{\beta_1}}{\hat{\theta}}}{\color{blue}{\LAM}} \right)^{-1}.$$

Therefore in terms of step size $$\gamma_{\text{\algnametiny{EF21-W-SGD}}} \ge \gamma_{\text{\algnametiny{EF21-SGD}}}$$ and  \algname{EF21-W-SGD} exhibits a more aggressive step size.

We conducted distributed training of a \modelname{logistic regression} model on \dataname{W1A}, \dataname{W2A}, \dataname{W3A}, \dataname{PHISHING}, \dataname{AUSTRALIAN} datasets with non-convex regularization. For all datasets, we consider the  optimization Problem~\eqref{ch3:eq:main_problem}, 
where
\begin{eqnarray}
\label{ch3:eq:ncvx-log-reg-3}
f_i(x) \eqdef \dfrac{1}{n_i} \sum_{j=1}^{n_i} \log \left(1+\exp({-y_{ij} \cdot a_{ij}^{\top} x})\right) + \lambda  \sum_{j=1}^{d} \dfrac{x_j^2}{x_j^2 + 1},
\end{eqnarray}
and $(a_{ij},  y_{ij}) \in \mathbb{R}^{d} \times \{-1,1\}$.

The initial gradient estimators are set to $g_i^0 = \nabla f_i(x^0)$ for all  $i \in [n]$. For comparison of \algname{EF21-SGD} and \algname{EF21-W-SGD}, we used the largest step size allowed by theory. The dataset shuffling strategy repeats the strategy that we have used for \algname{EF21-W-PP} and \algname{EF21-W} and it is described in Appendix~\ref{ch3:app:dataset-shuffling-for-libsvm}. The algorithms \algname{EF21-SGD} and \algname{/EF21-W-SGD} employed an unbiased gradient estimator, which was estimated by sampling a single training point uniformly at random and independently at each client.

\begin{table}[h]
\footnotesize
\begin{center}
\caption{{Non-Convex optimization experiments in Figures \ref{ch3:fig:app-real-ef21-sgd-ncvx}, \ref{ch3:fig:app-real-ef21-sgd-ncvx-aus}. Quantities that define theoretical step size.}}
\label{ch3:tbl:app-real-ef21-sgd-ncvx}
\begin{tabular}{|c||c|c|c|c|c|c|c|c|c|}
\hline
Tag & $L$ & $\Lvar$ & $\LQM$ & $\LAM$ & $\gamma_{\text{\algnametiny{EF21-SGD}}}$ & $\gamma_{\text{\algnametiny{EF21-W-SGD}}}$ \\
\hline
\hline
\tiny{(a) W1A} & $0.781$ & $3.283$ & $2.921$ & $2.291$ & $4.014  \cdot 10^{-4}$ & $5.118   \cdot  10^{-4}$ \\
\hline
\tiny{(b) W2A} & $0.784$ & $2.041$ & $2.402$ & $1.931$ & $4.882  \cdot 10^{-4}$ & $6.072  \cdot 10^{-4}$ \\
\hline
\tiny{(c) W3A} & $0.801$ & $1.579$ & $2.147$ & $1.741$ & $5.460   \cdot  10^{-4}$ & $6.733   \cdot  10^{-4}$ \\
\hline
\tiny{(f) PHISHING} & $0.412$ & $9.2   \cdot  10^{-4}$ & $0.429$ & $0.428$ & $1.183  \cdot 10^{-2}$ & $1.186   \cdot  10^{-2}$\\
\hline				
\tiny{(g) AUSTRALIAN} & $3.96   \cdot  10^6 $ & $1.1   \cdot  10^{16}$ & $3.35   \cdot  10^{7}$ & $3.96   \cdot  10^{6}$ & $3.876  \cdot 10^{-10}$ & $3.243   \cdot  10^{-9}$\\
\hline				
\end{tabular}
\end{center}
\end{table}

\begin{center}	
\begin{figure*}[t]
\centering
\captionsetup[sub]{font=normalsize,labelfont={}}	
\captionsetup[subfigure]{labelformat=empty}

\begin{subfigure}[ht]{0.49\textwidth}
\includegraphics[width=\textwidth]{ch3imgs/ef21-vc-figs-sgd/expreal/w1a-sgd.pdf} \caption{{ (a) \dataname{W1A}}}
\end{subfigure}		
\begin{subfigure}[ht]{0.49\textwidth}
\includegraphics[width=\textwidth]{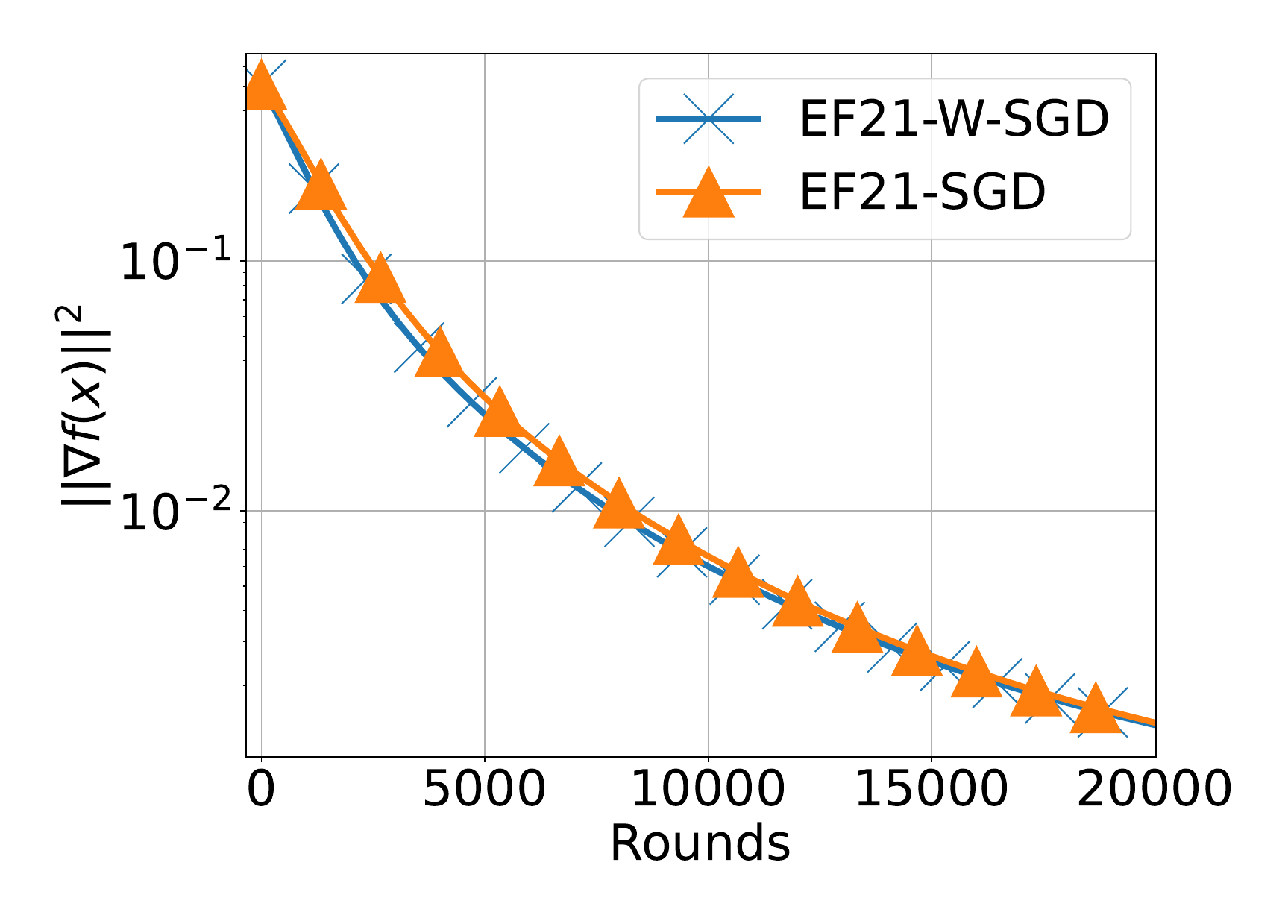} \caption{{ (b) \dataname{W2A}}}
\end{subfigure} \\
\begin{subfigure}[ht]{0.49\textwidth}
\includegraphics[width=\textwidth]{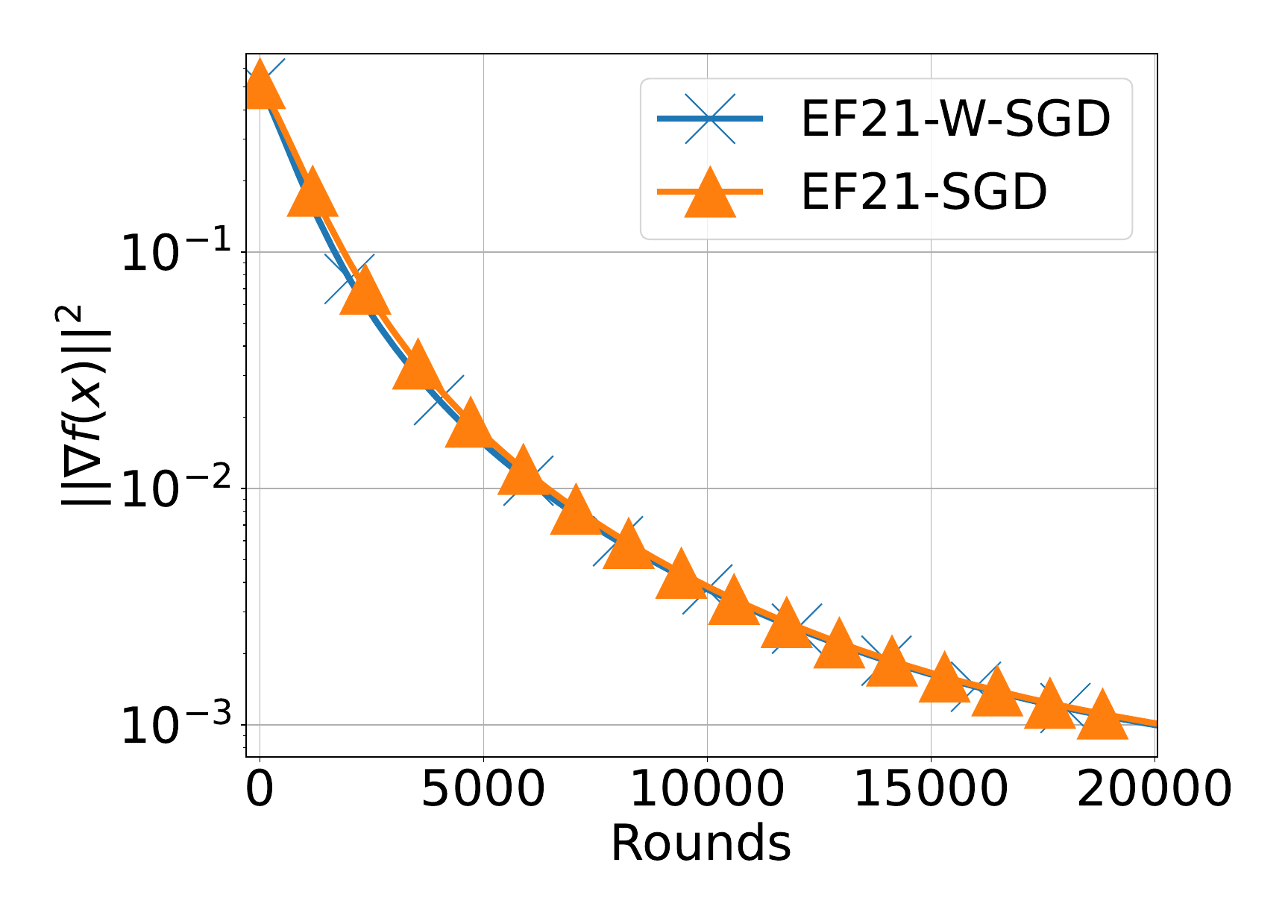} \caption{{ (c) \dataname{W3A}}}
\end{subfigure}		
\begin{subfigure}[ht]{0.49\textwidth}
\includegraphics[width=\textwidth]{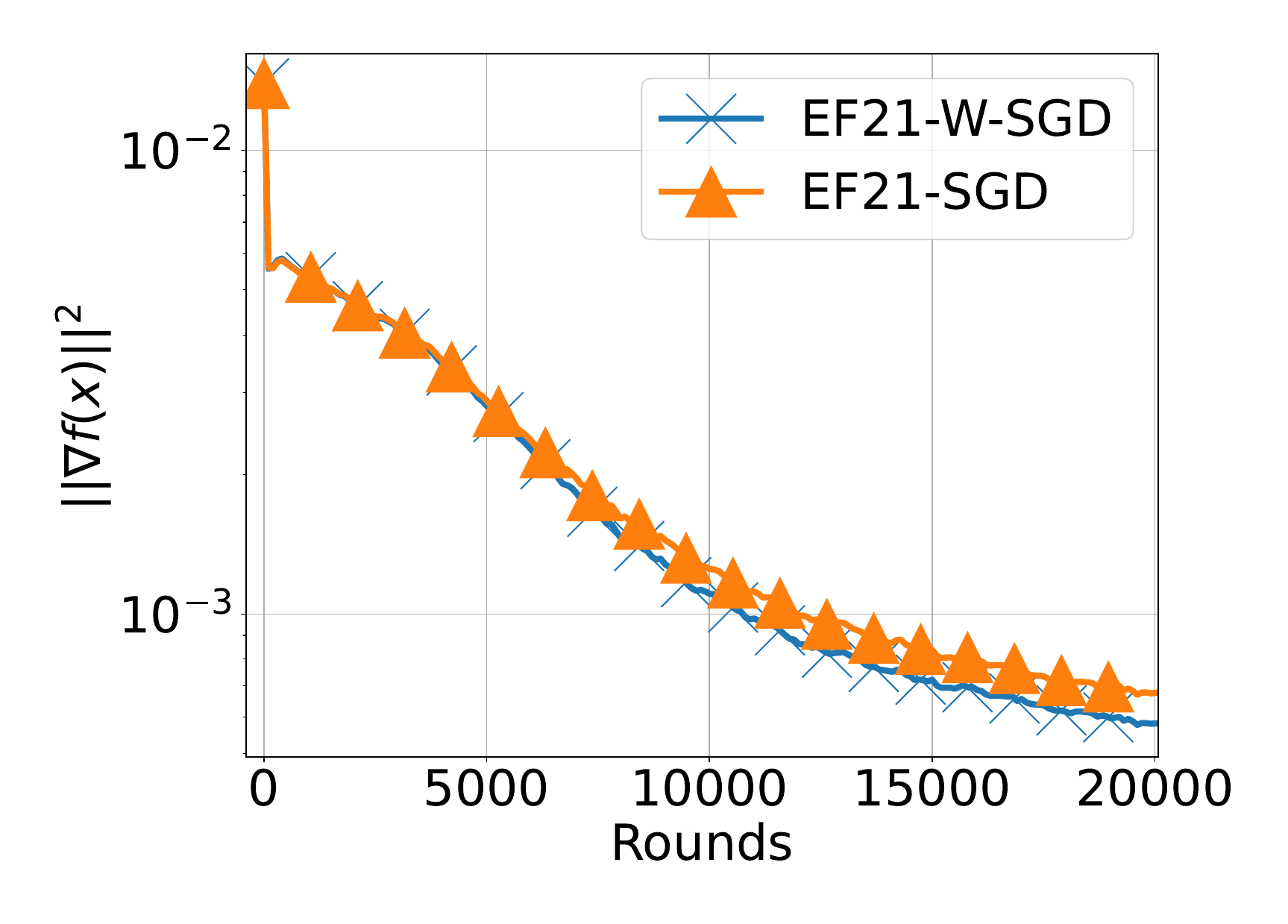} \caption{{ (f) \dataname{PHISHING}}}
\end{subfigure}

\caption{{Non-Convex \modelname{logistic regression}: comparison of \algname{EF21-SGD} and \algname{EF21-W-SGD}. The used compressor is \compname{Top1}. The \algname{SGD} gradient estimator is \algname{SGD-US}, $\tau=1$. The number of clients $n=1,000$. The objective function is constitute of $f_i(x)$ defined in Equation~\eqref{ch3:eq:ncvx-log-reg-3}.	Regularization term $\lambda \sum_{j=1}^{d} \dfrac{x_j^2}{x_j^2 + 1}$, $\lambda=0.001$. Theoretical step size. See also  Table~\ref{ch3:tbl:app-real-ef21-sgd-ncvx}.}
\label{ch3:fig:app-real-ef21-sgd-ncvx}}
\end{figure*}
\end{center}

\begin{center}	
\begin{figure*}[t]
\centering
\captionsetup[sub]{font=normalsize,labelfont={}}	
\captionsetup[subfigure]{labelformat=empty}

\begin{subfigure}[ht]{0.8\textwidth}
\includegraphics[width=\textwidth]{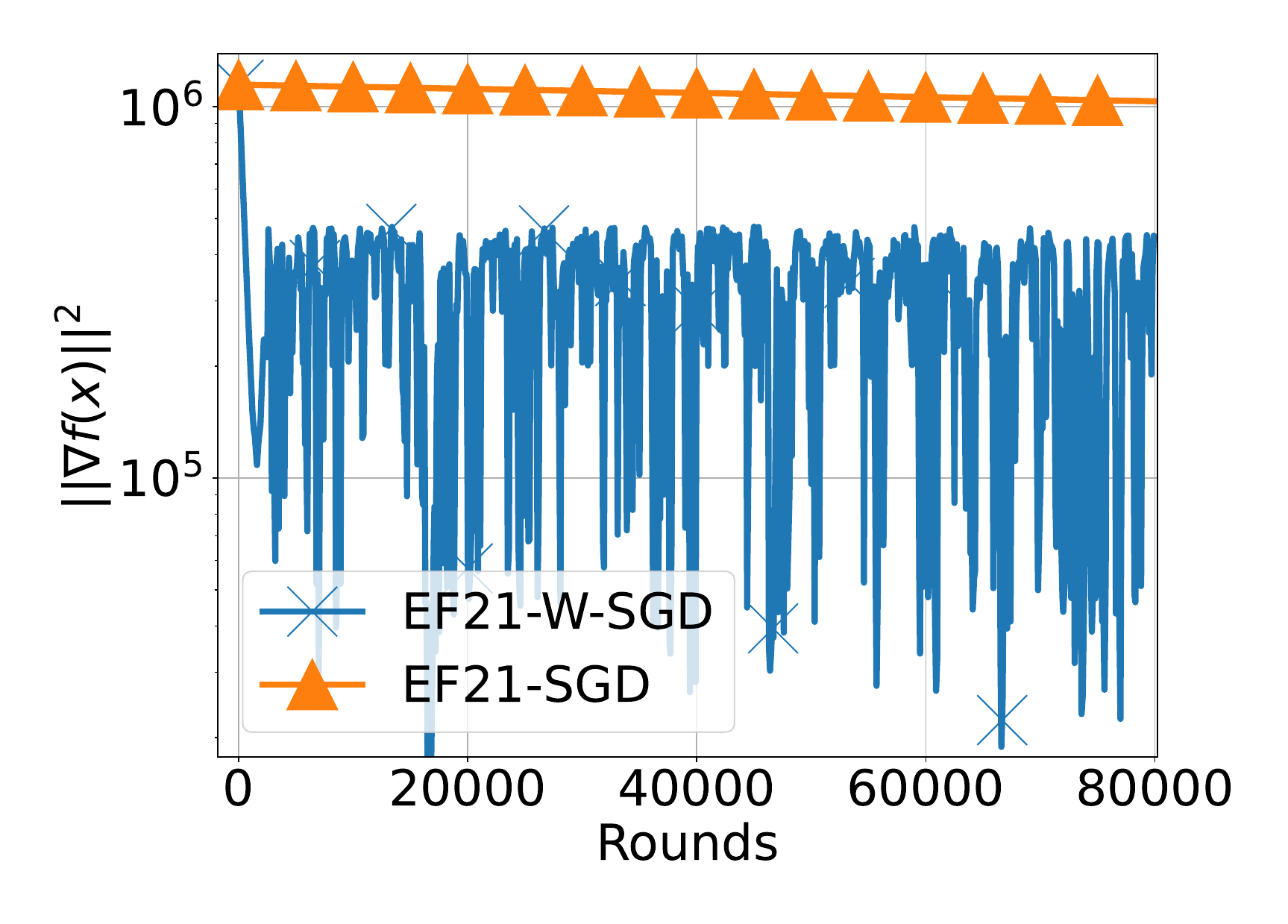} \caption{{ \dataname{AUSTRALIAN}}}
\end{subfigure}

\caption{{Non-Convex \modelname{logistic regression}: comparison of \algname{EF21-SGD} and \algname{EF21-W-SGD}. The used compressor is \compname{Top1}. The \algname{SGD} gradient estimator is \algname{SGD-US}, $\tau=1$. The number of clients $n=200$. The objective function is constitute of $f_i(x)$ defined in Equation~\eqref{ch3:eq:ncvx-log-reg-3}. Regularization term $\lambda \sum_{j=1}^{d} \dfrac{x_j^2}{x_j^2 + 1}$, with $\lambda=1,000$. Theoretical step size. Full participation. Extra details are in Table~\ref{ch3:tbl:app-real-ef21-sgd-ncvx}.}}
\label{ch3:fig:app-real-ef21-sgd-ncvx-aus}
\end{figure*}
\end{center}

The results are presented in Figure~\ref{ch3:fig:app-real-ef21-sgd-ncvx} and Figure~\ref{ch3:fig:app-real-ef21-sgd-ncvx-aus}. The important quantities for these instances of optimization problems are summarized in Table~\ref{ch3:tbl:app-real-ef21-sgd-ncvx}. In Figure \ref{ch3:fig:app-real-ef21-sgd-ncvx} (a, b, c, d), we can observe that for these datasets, the \algname{EF21-W-SGD} is better, and this effect is observable in practice. The experiment on the \dataname{AUSTRALIAN} datasets is presented in Figure~\ref{ch3:fig:app-real-ef21-sgd-ncvx-aus}. This example demonstrates that in this \dataname{LIBSVM} benchmark datasets, the relative improvement in the number of rounds for \algname{EF21-W-SGD} compared to \algname{EF21-SGD} is considerable. Finally, we address oscillation behavior to the fact that the employed step size for \algname{EF21-SGD} is too pessimistic, and its employed step size removes oscillation of $\|\nabla f(x^t)\|^2$.


\addtocounter{adjsection}{1}
\section{Reproducibility}

To ensure reproducibility, we use the following \fl \citep{burlachenko2021fl_pytorch} simulator features: (i) random seeds were fixed for data synthesis; (ii) random seeds were fixed for the runtime pseudo-random generators involved in \algname{EF21-PP} and \algname{EF21-SGD} across clients and the server; (iii)  the thread pool size was turned off to avoid the non-deterministic order of client updates in the server. The source code for the experiments, along with a description for reproducing the experiment, can be downloaded from:
\begin{center}
	\href{https://openreview.net/forum?id=Ch7WqGcGmb}{https://openreview.net/forum?id=Ch7WqGcGmb}
\end{center}


\unappendix

\chapter{DCGD/PermK/AES: Classical Cryptography in FL}
\label{chapter4}


The goals and summaries of this chapter are outlined in Table \ref{ch1:tbl:algorithms} and Section~\ref{ch1:sec:overview-4}.

\section{Introduction}
Effective machine learning models necessitate vast training data from diverse sources \citep{zhang2020batchcrypt}. However, such data, often dispersed across various entities, faces sharing restrictions due to privacy concerns \citep{jiang2021flashe,liu2022privacy}. Federated Learning ({FL}) provides a viable solution, enabling collaborative global model training without exposing sensitive information \citep{mcmahan17fedavg,kairouz2019advances}. {FL} algorithms bifurcate into \textit{cross-device} {FL}, involving big amount of devices, and \textit{cross-silo} {FL}, typically engaging several distinct organizations \citep{liu2022privacy}. In both cases, preserving the privacy of clients' datasets is significant. {FL} strives to ensure confidentiality by retaining private data on the client side, yet it falls short of providing strong privacy assurance if relying only on that. Model parameters and derived quantities from it, transmitted from clients to a server, may embed sensitive data (see Appendix \ref{ch4:app:reconstruction}). Detailed information about privacy methods for {FL} is presented in Appendix \ref{ch4:app:overview_of_privacy_mechanisms}. We briefly discuss two of them.

Differential Privacy (\abr{DP}) is an approach to assess the privacy of a specific Algorithm provided by \citet{dwork2006our,dwork2006calibrating}. \abr{DP} algorithms ensure that, for any set of training examples, no attacker, no matter how powerful, can learn much more information about a single training example than they could if this example had been excluded from the training data. Despite offering privacy protection for individual users, integrating \abr{DP} with {FL} could amplify communication overhead, and diminish accuracy. When striving for strong privacy guarantees in scenarios where the user count is small (which is possible in \textit{cross-silo}) the noise impact from \abr{DP} techniques makes training challenging. Understanding the convergence of \algname{SGD} with \abr{DP} is an active research topic \citep{chen2020understanding}. 

While \abr{DP} statistically hides the training data, Homomorphic Encryption (\abr{HE}) provides means to perform computation operations under encrypted numbers from $\mathbb{Z}^d$, $\mathbb{R}^d$, $\mathbb{B}^d$ without decryption, protecting input and output from execution part. Therefore \abr{HE} can be used for aggregating encrypted gradients \citep{liu2019secure,aono2017privacy} in a server that clients do not trust, given that the master has a public key which is required for execution of arithmetic operations.

\paragraph{Classical Cryptography in FL.} Classical Cryptography operates on the binary representation of data. It does not preserve even linear relations in a homomorphic way. The fact of exhibiting poor algebraic properties is an underlying reason why using Classical Cryptography was considered challenging by previous research papers \citep{kaissis2020secure,lauter2022private,jain2023revisiting,DBLP:journals/cybersec/PanCWYLW24}. In these works, authors stated that Advanced Encryption Standard (\aesname{AES}) \citep{daemen2001reijndael} is not suitable for {FL} or is too challenging.

\paragraph{Communication compression in FL.}
Numerous compression methods such as quantization \citep{DBLP:conf/nips/WenXYWWCL17,safaryan2019stochastic}, sparsification \citep{wangni2018gradient,Alistarh-EF-NIPS2018}, and dithering \citep{alistarh2017qsgd,horvath2019natural} have been explored to mitigate communication cost during {FL} training. However, these techniques necessitate secure server-side aggregation, and there are no guarantees that communication reduction techniques and techniques aimed to preserve privacy or security are combinable.

\subsection{Problem formulation}
\label{ch4:problem-formulation}

In this work, we develop a practical communication efficient privacy and secure aware framework for {FL} training. 

\paragraph{Requirements for optimization objective.}
From the perspective of Machine Learning ({ML}), our objective is to select a function from a parameterized function class $\mathcal{F}$ indexed by $x \in \mathbb{R}^d$, by solving the following optimization problem:
\begin{equation}\label{ch4:eq:main}
	\min \limits_{x\in \R^d} \left\{f(x)\eqdef \dfrac{1}{n}\sum \limits_{i=1}^n f_i(x) \right\}, 
\end{equation}

Here, $n\in \mathbb{N}$ represents the number of clients, and $x\in \mathbb{R}^d$ denotes the $d$ parameters or weights of a model $\hat{F}(\cdot;x) \in \mathcal{X}\to \mathcal{Y}$ that need to be learned across all $n$ clients. The function $f_i:\mathbb{R}^d \to \mathbb{R}$ provides the score criteria for using the model $\hat{F}(\cdot;x)$ on the client's $i$ data. In the context of {FL} the functions $f_i$ typically represented as:

\begin{equation}\label{ch4:eq:fi_for_erm} 
	f_i(x) = \dfrac{w_i}{n_{i}} \sum \limits_{j=1}^{n_{i}} \left(\mathcal{L}_{ij}(b_{ij}, \hat{F}(a_{ij};x)) + R_{i}(x) \right),
\end{equation}

Here, $n_i \in \mathbb{R}$ denotes the number of data points at client $i \in [n]$, and $(a_{ij}, b_{ij}) \in \mathcal{X} \times \mathcal{Y}$ represent the input-output pairs at client $i$. The function $\mathcal{L}_{ij}(y_{\mathrm{real}}, y_{\mathrm{pred}} ): \mathcal{Y} \times \mathcal{Y} \to \mathbb{R}$ is a loss function that scores prediction, $R_{i}: \mathbb{R}^d \to \mathbb{R}$ is the regularization function used for parameter $x$ at client $i$. The weight $w_i \in \mathbb{R}$ encodes knowledge about the role of client $i$. 
For example: (i) $w_i \eqdef {\left(n_i \cdot n\right)}/{\left( \sum_{i=1}^{n}n_i \right)}$ corresponds to a case when \textit{all data point}s are equally important; (ii) $w_i \eqdef 1$ corresponds to a case when \textit{all devices} are equally important. We aim to solve Problem~\eqref{ch4:eq:main} under the next requirements:



\subsection{Assumptions}

\paragraph{Assumptions for applying our framework.} 

\begin{enumerate}[label={}]
	\item i) $f(x)$ is differentiable in training variable $x\in \mathbb{R}^d$.
	\item ii) Clients trust each other through the established key.
	\item iii) $\mathrm{dom}(f)=\RD$.
\end{enumerate}

\paragraph{Security and system requirements for the {FL} training procedure.}

\begin{enumerate}[label=(\Alph*)]
	\item Clients never transfer the training data to the master.
	\item Clients do not trust communication devices.
	\item System should detect attempts of message tampering by adversaries.
	\item Clients do not trust server.
	\item Preventing competitor worker interference.
	\item Limit memory traffic from clients to the master. 
	\item System should allow overlapping communication and computing in clients because separate physical devices implement it. 
\end{enumerate}


\subsection{Real-world scenarios for applying our framework.}

The potential benefits of our framework can be demonstrated through at least three possible scenarios.

(a) An individual using multiple \abr{IoT} devices wants to train an {ML} model via third-party servers. In practice, engineers at these companies often have unrestricted access to server software, which raises privacy concerns. Individuals may seek guarantees that third-party providers only offer computation and storage services without being able to access the transmitted data.

(b) In cross-device settings, multiple mutually trusted \abr{IoT} clients collaborate on {FL} training. However, a central coordinator may still be needed. For instance, if a client is temporarily unavailable, updates must be buffered. Additionally, applied optimization steps should be stored to reconstruct the training trajectory for statistical analysis, but \abr{IoT} devices may lack sufficient storage. Another reason for requiring a coordinator is to facilitate training restarts in case of system failures.

(c) We aim to train an {FL} model without a physical master in scenarios where the communication topology naturally supports broadcasting. However, communication must remain secure against tampering and eavesdropping. Our framework can be instantiated in this setting (see Appendix \ref{ch4:app:comm_networks}).

\subsection{Contributions}

We discovered that recently proposed permutated correlated compressors \citep{szlendak2021permutation} \compname{PermK} exhibit properties essential for using \textit{Classical Cryptography in FL} while preserving the ability of \textit{Communication Compression}. In our work, we introduce a framework that provides privacy and secure preserving training process to {FL} applications in which previously \abr{HE} methods have been used. Summary of our contributions:

\begin{enumerate}
	\item We addressed the challenge highlighted in \citep{kaissis2020secure, lauter2022private, jain2023revisiting, DBLP:journals/cybersec/PanCWYLW24}, which states that using \aesname{AES} in {FL} is challenging, by constructing an {FL} system based on \aesname{AES}.
	\item We demonstrated the operational advantages of the proposed privacy and secure aware optimization Algorithm~\ref{ch4:alg:dcgd_permk_aes} over \abr{HE} in the setting in which \abr{HE} is typically applied in {FL}. 
	\item We demonstrated the framework's ability to train a \modelname{ResNet-18} \citep{resnet} {DL} model on \dataname{CIFAR-10} and discussed its potential benefits for {DL} more broadly in Appendix~\ref{ch4:app:flexibility_for_dl_training}.
	\item We demonstrated a possibility of computation communication overlap and handling compute heterogeneity in Appendix~\ref{ch4:app:simulation_experiment}. Deployment flexibility in communication topologies is discussed in Appendix~\ref{ch4:app:comm_networks}.
\end{enumerate}

To support readers with various backgrounds we provide:
\begin{enumerate}[label={}]
\item i) Glossary in Appendix~\ref{ch4:app:glossary}.
\item ii) Details about \aesname{AES} in Appendix~\ref{ch4:app:aes_details}.
\item iii) Overview of privacy mechanisms in {FL} in Appendix~\ref{ch4:app:overview_of_privacy_mechanisms}, discussion about the difference between privacy and security in Appendix~\ref{ch4:app:privacy_vs_security}.
\item iv) Overview of \ecryptname{CKKS} in Appendix~\ref{ch4:app:ckks_details}.
\end{enumerate}



\section{Framework of Security Aware {FL} with Permuted Compressors}

The Distributed Compressed Gradient Descent 
(\algname{DCGD (Baseline)}) \citep{khirirat2018distributed}, presented as (Algorithm \ref{ch4:alg:dcgd}, \myred{Option B}), enables the use of independent unbiased compressors $\mathcal{C}_i$ if $\Exp{\|\mathcal{C}_i(x) - x\|^2} \le w\|x\|^2$. If $w\ne0$ this algorithm does not induce variance for a $\mu$-strongly convex objective $f$ only in a specific (\textit{overparameterized}) mode: $\nabla f_i(x)=0, \forall i \in [n]$. If $\mathcal{C}_i (\nabla f_i(x)) \eqdef \nabla f_i(x)$, it reconstitutes distributed Gradient Descent (\algname{GD}). One line of research involves removing variance from the use of compressors. For instance, \algname{MARINA} \citep{gorbunov2021marina} and \algname{COFIG/FRECON} \citep{zhao2021faster} do not induce variance when using unbiased independent compressors, while \algname{EF21} \citep{richtarik2021ef21} avoids inducing variance from any independent contractive compressors. 

In all these algorithms, the logic in the master starts to include extra state updates based on obtaining messages from clients. But this is what we aim to avoid. Thus, the development of our framework was based on stateless \algname{DCGD (Baseline)}. For plain \algname{DCGD (Baseline)} we have two issues: (a) In Line 6, the algorithm sends sparsified information about $\nabla f_i (x)$ to the master, which is insecure; (b) In Line 8, the master computes the average of $g_i^k \in \mathbb{R}^d$, but the master should not obtain $g_i^k=\nabla f_i(x^k)$ or quantities that are subject to reconstruction attacks (see Appendix \ref{ch4:app:reconstruction}). One way to address these challenges is through \abr{HE} schemes. The current practical state-of-the-art \abr{HE} scheme that operates on elements of $\mathbb{R}^d$ is \ecryptname{CKKS}, designed by \citet{cheon2017homomorphic}. \ecryptname{CKKS} is the most widely used scheme for {ML} applications and is considered a highly practical choice \citep{lauter2022protecting}. As we demonstrate in our experiments, \ecryptname{CKKS} introduces an approximation error due to its inherently lossy design. In our experiments, we used \ecryptname{CKKS} as a black-box with an appropriate configuration. For more background and details on \ecryptname{CKKS}, see Appendix~\ref{ch4:app:ckks_details}, and for a broader overview of \abr{HE}, see Appendix~\ref{ch4:app:overview_of_privacy_mechanisms}.

\begin{algorithm}[H]
	\footnotesize
	\algsetup{linenosize=\footnotesize}
	\captionsetup{position=top}
	\begin{algorithmic}[1]
		\label{ch4:algo:1}
		\STATE  \textbf{Input and Initialization:} step size $\gamma>0$, iterate $x^0\in\R^d$, clients' compressors $\cC_i$ \newline
		\myblue{Option A}: Clients negotiate a secret key ${\color{blue}sk}$ 
		\FOR {$k=0,1,2, \ldots$}
		\FOR {{\bf all workers $i \in \{1,2,\dots,n\}$ in parallel}}
		\STATE {Compute and compress local gradient $g_i^k = \cC_i^k(\nabla f_i(x^k))$} 
		\STATE \myblue{Option A}: {Send tuple $m_i^k = ({nonce}_i, {mac}_i, \tilde{g}_i^k) = {\color{blue}Enc}(g_i^k, {\color{blue}sk})$ to master, $\tilde{g}_i^k$ is encryption of $g_i^k$}
		\STATE \myred{Option B}: {Send $g_i^k$ to master as message $m_i^k$}
		\ENDFOR      
		\STATE {Master collects the messages from clients $G^k = (m_1^k,\dots,m_n^k)$}
		\STATE \myblue{Option A:} {Master broadcasts $G^k \in \mathbb{R}^{dn + |nonce|n + |mac|n}$ to workers}
		\STATE \myred{Option B:} {Master computes the aggregate $\hat{g}^k = \dfrac{1}{n}\sum_{i=1}^n  g_i^k$, $g_i^k \eqdef m_i^k$}
		\STATE \myred{Option B:} {Master broadcasts $\hat{g}^k \in \mathbb{R}^d$ to all $n$ workers}
		\FOR {{\bf all workers $i \in \{1,2,\dots,n\}$ in parallel}} 
		
		\STATE \myblue{Option A:} {From ${G}^k$ unpack ${mac}_j$ and $\tilde{g}_j^k \in \mathbb{R}^d, \forall j \in [n]$; \newline ${g}_j^k = {\color{blue}Decrypt}( \tilde{g}_j^k, {\color{blue}sk})$; $Verify({\color{blue}sk}, {g}_j^k, mac_j)$}
		\STATE \myblue{Option A:} {Workers computes the aggregate $\hat{g}^k = \dfrac{1}{n} \sum_{i=1}^n  g_i^k$}
		\STATE \myred{Option B:} {Workers obtain $\hat{g}^k$ from master during round $k$.}
		\STATE Compute the next iterate $x^{k+1} = {x^k - \gamma \hat{g}^k}$
		\ENDFOR
		\ENDFOR
	\end{algorithmic}
	\caption{{\algname{DCGD} with \myblue{Naive Usage of AES  (A)} and \myred{Baseline (B)}.}}
	\label{ch4:alg:dcgd}    
\end{algorithm}


\begin{algorithm}[H]
	\footnotesize
	\algsetup{linenosize=\footnotesize}
	\begin{algorithmic}[1]
		\label{ch4:algo:3}
		\STATE  \textbf{Input and Initialization:}  Dimension of opt. problem $d>0$, number of clients $n>0$.
		\STATE Uniformly at random generate permutation $z$ of sequence $[d]=\{1,2,\dots,d\}$
		\STATE Split z into $n$ buckets, where each bucket has a size at least $B=\left\lfloor \nicefrac{d}{n}\right\rfloor$
		\STATE Each bucket $b_i$ is initialized with $\{z_{ (i\cdot B) - B + 1}, \dots, z_{ (i\cdot B)}\}, 1 \le i \le n$
		\STATE Compute the residual $t = d - n \cdot \left\lfloor \nicefrac{d}{n}\right\rfloor$
		\STATE Sample without replacement $t$ clients from $n$ as a set $S$, $|S| = t$
		\STATE Scan the set $S=\{s_1,\dots,s_k, \dots, s_t\}$ and update $b_{s_k} = b_{s_k} \cup z_{d - t + k}$
		\STATE {Setup compression $\cC=(\cC_1,\dots, \cC_n)$: $[\cC_i(x)]_j = n \cdot x_j \cdot I( j \in b_i)$}
	\end{algorithmic}
	\caption{{Sampling of Correlated Permutation Compressors (\compname{PermK}) ($d>n$).}}
	\label{ch4:alg:perm_k_gen}    
\end{algorithm}
\begin{algorithm}[H]
	\footnotesize
	\algsetup{linenosize=\footnotesize}
	\begin{algorithmic}[1]
		\label{ch4:algo:4}
		\STATE  \textbf{Input and Initialization:} learning rate $\gamma>0$, start iterate $x^0\in\R^d$ \newline
		All clients negotiate a secret key ${\color{blue}sk}$.	All clients and master negotiate a ${seed}$ for pseudo-random number generator (\algnamesmall{PRG}).
		\FOR {$k=0,1,2, \ldots$}
		\FOR {{\bf all workers $i \in \{1,2,\dots,n\}$ in parallel}}
		\STATE {Generate permutation compressor $\cC_i^k(\cdot; {seed})$ for round $k$ at worker $i$ with Algorithm~\ref{ch4:alg:perm_k_gen} using the known $seed$.}
		\STATE {Compute and compress local gradient $g_i^k = \cC_i^k(\nabla f_i(x^k))$}
		\STATE {Represent $g_i^k \in \mathbb{R}^d$ in sparse form.}
		\STATE {Send $m_i^k = ({nonce}_i, {mac}_i, \hat{g}_i^k)={\color{blue}Enc}(g_i^k,{\color{blue}sk})$ to the master.}
		\ENDFOR      
		\STATE {Master concatenates the message ${G}^k = concat(m_1^k, \dots, m_n^k)$}
		\STATE {Master broadcasts the compressed aggregate ${G}^k$ to all workers}
		\FOR {{\bf all workers $i \in \{1,2,\dots,n\}$ in parallel}}      		
		\STATE {Reconstruct indices from (\algnamesmall{PRG}) for all compressors $\mathcal{C}_1^k, \dots, \mathcal{C}_n^k$ with Algorithm~\ref{ch4:alg:perm_k_gen} using the known ${seed}$.}
		\FOR {{\bf all block of coordinates $b \in \{1,2,\dots,n\}$ in parallel}}
		\STATE Obtain part of $G^k$ corresponds to $m_b^k$
		\STATE Unpack ${nonce}_b$, ${mac}_b$, $\hat{g}_b^k$, and ${g}_b^k = {\color{blue}Decrypt}(\tilde{g}_b^k, {\color{blue}sk})$
		\STATE $Verify({\color{blue}sk}, \hat{g}_b^k, {mac}_b)$. If no - halt the training process.
		\STATE Compute the next iterate $x_b^{k+1} = {x_b^k - \dfrac{\gamma}{n} \cdot \hat{g}_b^k}$
		\ENDFOR
		\ENDFOR
		\ENDFOR
	\end{algorithmic}
	\caption{{\algnamewithaes{DCGD/PermK/AES}, $d \ge n$.}}
	\label{ch4:alg:dcgd_permk_aes}    
\end{algorithm}

The clients need to send in parallel $n$ \textit{encrypted} messages to the master. Since clients trust each other due to Assumption (ii) from Section~\ref{ch4:problem-formulation}, symmetric key encryption is a natural choice. In the world of symmetric ciphers, we have decided to use a block cipher, specifically the industry-standard \aesname{AES} \citep{daemen1999aes}, instead of stream ciphers like \algname{SALSA} \citep{bernstein2008salsa20}. The choice was made due to the advantages of block ciphers, as: 
\begin{enumerate}[label={}]
\item i) Hardware support within CPU \footnote{AES is supported by Intel x86 Westmere, AMD x86 Bulldozer, ARM Cortex-A53.}.
\item ii) The provision of multiple security levels ($128,192,256$ bits).
\item iii) The necessary flexibility to work with available parts of the vector $G^k$ in coming Algorithm \ref{ch4:alg:dcgd_permk_aes}.
\end{enumerate}

The \aesname{AES} block-cipher provides security only for a single 16-byte block. The \textit{Modes of Operation} provide a way to use \aesname{AES} for more than one block. \textit{Message Authentication Code} (\abr{MAC}) provides guarantees from tampering. In our work, we used \aesname{AES/EAX} mode of operation. For details about \aesname{AES} see Appendix \ref{ch4:app:aes_details}.

The naive way to use \aesname{AES} is to use Algorithm \ref{ch4:alg:dcgd}, \myblue{(Option A)}. It solves problems with an untrusted server and channels. Without knowing $sk$, the untrusted party cannot join the training procedure. Due to the use of \abr{MAC} (Lines 5, 13), the training process verifies the integrity and protects against malicious attacks. To provide semantic security against a chosen-plaintext attack (\attackname{CPA}) and have the ability to use $sk$ in a distributed way, we employ random $nonce$ based selection. It eliminates the need for client coordination in a $nonce$ selection if $nonce$ is big enough such as $128$ bits. Details about the role of $nonce$ in block ciphers are presented in Appendix~\ref{ch4:app:aes_details}.

With this Algorithm~\ref{ch4:alg:dcgd} \myblue{Option-A}, solves (A)--(E) from Section~\ref{ch4:problem-formulation}. However, this strategy has two big downsides: (i) Computation at Lines 13 and 14 is repeated by all clients with complexity per client $\mathcal{O}(dn)$; (ii) Amount of information from the master in Line 9 is $\mathcal{O}(dn)$, not $\mathcal{O}(d)$. 

We pay this price due to the use of \aesname{AES}, which does not allow us to perform aggregation in the server. From one point of view, the natural way is to perform averaging for vectors in $\mathbb{R}^d$ in the master (Line 10), but on the other hand, we perform bit-wise operations (Line 5) inside \aesname{AES} cipher. There is no way to connect it from the first point of view, but we have discovered how to do it!

The \compname{PermK} correlated compressor from work \citep{szlendak2021permutation} possesses a compelling property that has yet to be recognized. The correlated compressors operate interdependently for clients and the schema for \compname{PermK} was outlined in Algorithm \ref{ch4:alg:perm_k_gen}. Let $\cC_1,\dots,\cC_n:\R^d\to \R^d$ be randomized \compname{PermK} compressors, then $$\Exp{\dfrac{1}{n}\sum_{i=1}^n \cC_i(v_i)} = \dfrac{1}{n}\sum_{i=1}^n v_i, \forall v_i \in \R^d.$$

If $d \ge n$, then $\dfrac{1}{n}\sum_{i=1}^n \cC_i(v_i)$ fulfills the variance bound: \newline $$\Exp{\norm{ \dfrac{1}{n}\sum_{i=1}^n \cC_i(v_i) - \dfrac{1}{n}\sum_{i=1}^n v_i}^2} \leq \dfrac{1}{n}\sum_{i=1}^n \norm{v_i}^2 - \norm{\dfrac{1}{n}\sum_{i=1}^n v_i}^2$$. 

This compressor is used in our \algname{DCGD/PermK} Algorithm \ref{ch4:alg:dcgd_permk_aes}. Its unbiasedness property ensures the absence of systematic errors and the variance bound guarantees that if we are in an overparameterized setting and $x^k \to x^*$, then the estimator's variance will decay to zero, which should remove any oscillation behavior near the solution. The crucial property of \algname{DCGD/PermK} is that the aggregation of compressed gradients can be replaced by concatenation, which is only an algebraic monoid for $(\mathbb{R}^*, \mathrm{concat})$, and not the algebraic group. Scaling an encrypted vector by $\nicefrac{1}{n}$ is unfeasible in the master, but this can be delegated to the iterate update Algorithm \ref{ch4:alg:dcgd_permk_aes}, Line 17. We can use non-secure pseudo-random generators as \citep{matsumoto1998mersenne} for sampling indices because they are independent of the input. 
\section{The Resilience for Attacks}
\label{ch4:sed:resilience}

In chosen-plaintext-attack (\attackname{CPA}), an attacker may adaptively ask for the encryption $(e_1, \dots)$ of arbitrary messages $(m_1,\dots)$ to obtain the ability to correctly guess from the set of two encrypted cipher-texts $\{E_1, E_2\}$ which encryption belongs to which message in the set $\{M_1, M_2\}$, given that attacker knows the set $\{M_1, M_2\}$. \ecryptname{CKKS} and \aesname{AES} in \aesname{EAX} mode of operation \citep{bellare2004eax} are \attackname{CPA}-secure. 

In a chosen-ciphertext attack (\attackname{CCA}), the attacker has unlimited access to a decryption oracle and aims to achieve the same goal as in \attackname{CPA}. No \abr{HE} schemes can achieve \attackname{CCA} security \citep{fauzi2022ind}. However, it has been proven that \aesname{AES}/\aesname{EAX} is secure against \attackname{CCA} \citep{bellare2004eax}.

Algorithms are often executed on general-purpose microprocessors, which process a stream of instructions during execution. However, one subtle aspect lies in the fact that the energy consumption and latency for different operations are different \citep{horowitz20141}. This discrepancy leads to the possibility of side-channel attacks, where the attacker can use information about measurable physical information to detect when, how much, and the type of operations have been executed. To have this ability the executable operations should be some partially invertible function of the secret key or plaintext. The \aesname{AES} block cipher and \textit{Operation Modes} perform the same stream of operation types (see Appendix~\ref{ch4:app:aes_details}). However, in \ecryptname{CKKS} polynomial multiplication, encoding, decoding, and randomization can lead to a trace to plain message \citep{aydin2022reveal} using side-channel attacks. 

Since the standardization of \aesname{AES}, significant efforts have been made to protect it from side-channel attacks \citep{rahaman2008side,gross2017efficient}. However, protecting \abr{HE} libraries from such attacks remains an open challenge. For instance, \citep{aydin2022reveal} highlights that the \abr{HE} implementation in the reference \libname{SEAL} library \citep{chen2017simple}, as well as in derived libraries like \libname{TenSEAL} \citep{benaissa2021tenseal}, needs to be revised to improve protection against side-channel attacks. The work by \citet{aydin2022reveal} demonstrates how an asymmetry in ciphertext generation can be exploited, reducing the security level of the Brakerski/Fan-Vercauteren (\abr{BFV}) \citep{fan2012somewhat} Ring Learning With Error (\abr{RLWE}) scheme from $2^{128}$ to $2^{4.4}$ (see Appendix~\ref{ch4:app:lwe}). As a result, the protection against side-channel attacks for \aesname{AES} can be considered more developed.


\section{Experiments}

We conducted an experimental comparison of several optimization algorithms with different compression and security methods. In Table~\ref{ch4:tbl:list_of_optimization_algos}, presented in Appendix~\ref{ch4:app:list_of_optimization_algos}, we summarize their qualitative aspects. Since there is a chain of modifications to \algname{DCGD} baseline, this summary can help trace the steps in our algorithm design.

\subsection{Synthetic experiments}
\label{ch4:sec:syntetic_exp}

The ultimate goal of the forthcoming chain of the algorithm design process is to illustrate the advantages of using the block cipher \aesname{AES} (see Appendix \ref{ch4:app:aes_details}) over \ecryptname{CKKS} (see Appendix \ref{ch4:app:ckks_details}). The experiments were conducted in a \fl simulator \citep{burlachenko2021fl_pytorch}. For details on the computing environment see Appendix~\ref{ch4:app:reconstruction}. We configured the \ecryptname{CKKS} to offer security guarantees as \aesname{AES} with $128$ bit ($16$ bytes) key. In this case, the size of public and private keys for \ecryptname{CKKS} has a lower bound $420\,000$ bytes (see Appendix~\ref{ch4:app:ckks_details} for underlying reason). During usage of \ecryptname{CKKS}, the public key should be reported somehow to the master once. As we will see while using \algnamewithaes{DCGD/PermK/AES} no key at all should be reported to the master for the master to operate. Therefore, the key size can be a problem for \ecryptname{CKKS} already when the volume of communicated information is far smaller than $0.42 \cdot 10^6$ Bytes. In our experiments, we neglect the overhead from \textit{key agreement}.

\paragraph{Optimization problem and experimental setup.}
In our synthetically controlled experiments, we consider a specific smooth convex optimization problem which is obtained from Equation~\eqref{ch4:eq:main} via $$f_i(x) = \dfrac{1}{n_i} \|\mA_i x - b_i\|^2, \mathrm{where\,} \mA_i \in \mathbb{R}^{n_i \times d}, b_i \in \mathbb{R}^{d}.$$

\begin{figure*}[ht!]
	\centering             
	
	\captionsetup[sub]{font=footnotesize,labelfont={},labelformat=empty}		
	\captionsetup[subfigure]{font=footnotesize,labelfont={},labelformat=empty}
	\captionsetup[figure]{font=footnotesize,labelfont={},labelformat=empty}
	
	
	\begin{subfigure}[ht]{0.8\textwidth}
		\includegraphics[width=\textwidth]{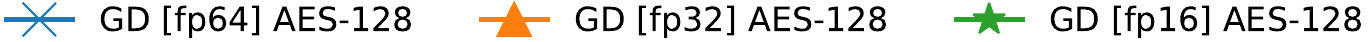} 
	\end{subfigure}
	\begin{subfigure}[ht]{0.55\textwidth}
		\includegraphics[width=\textwidth]{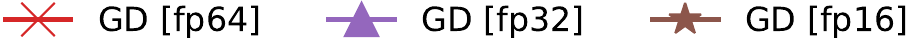} 
	\end{subfigure}
	
	\begin{subfigure}[ht]{0.49\textwidth}
		\includegraphics[width=\textwidth]{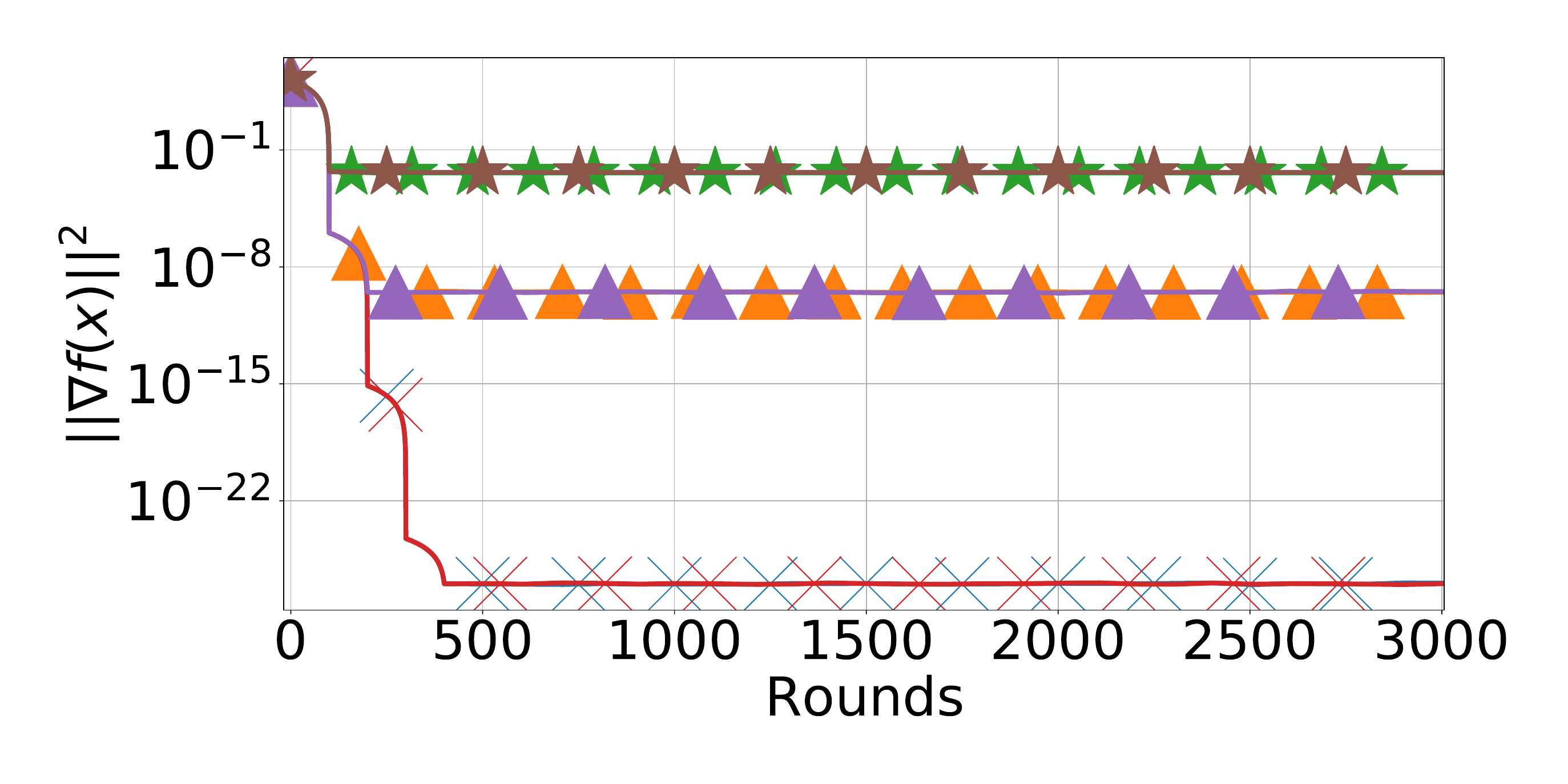} 
		\caption{{ (a) }}
	\end{subfigure}
	\begin{subfigure}[ht]{0.49\textwidth}
		\includegraphics[width=\textwidth]{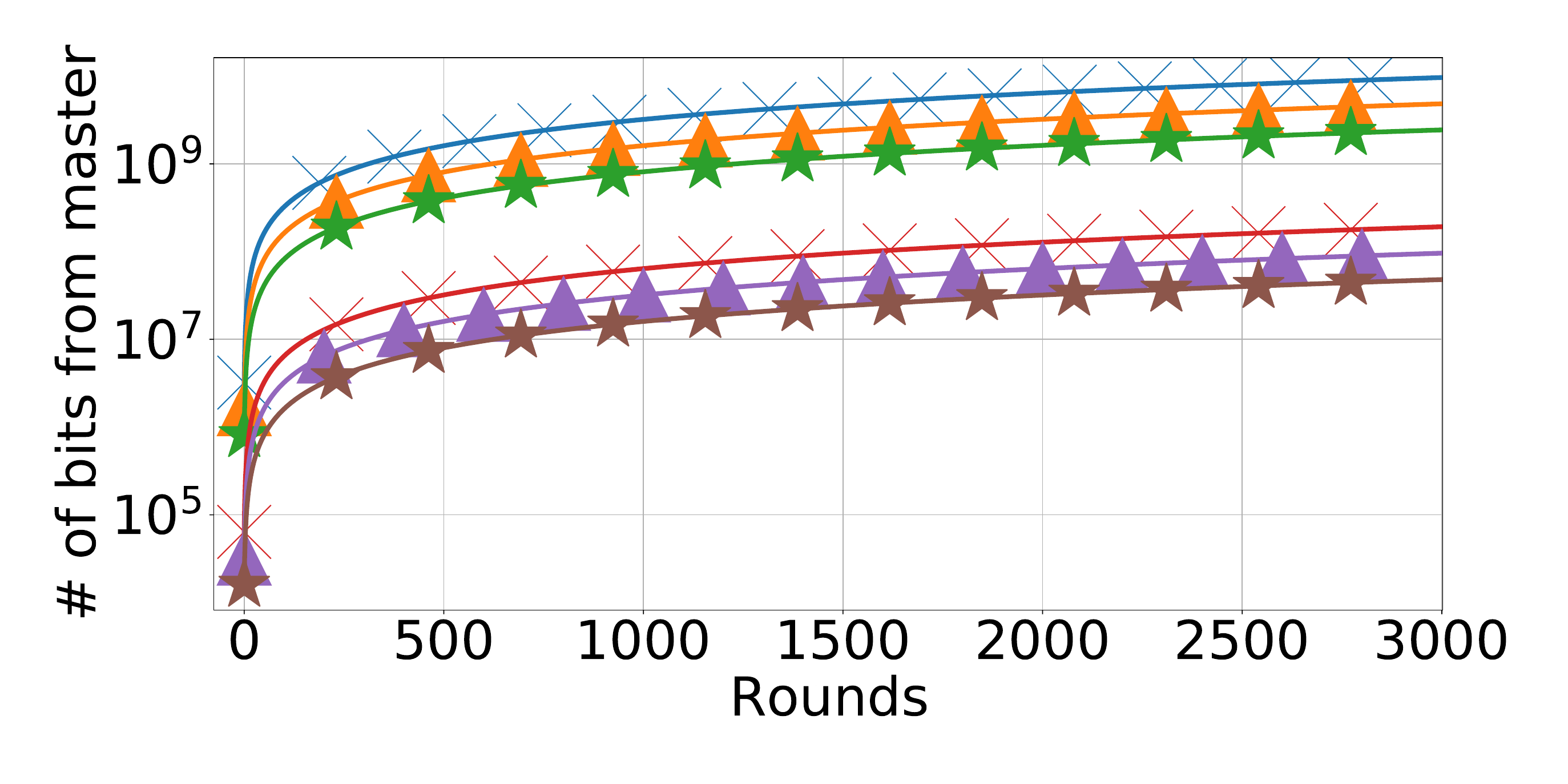} 
		\caption{{ (b) }}
	\end{subfigure}
	
	\begin{subfigure}[ht]{0.49\textwidth}
		\includegraphics[width=\textwidth]{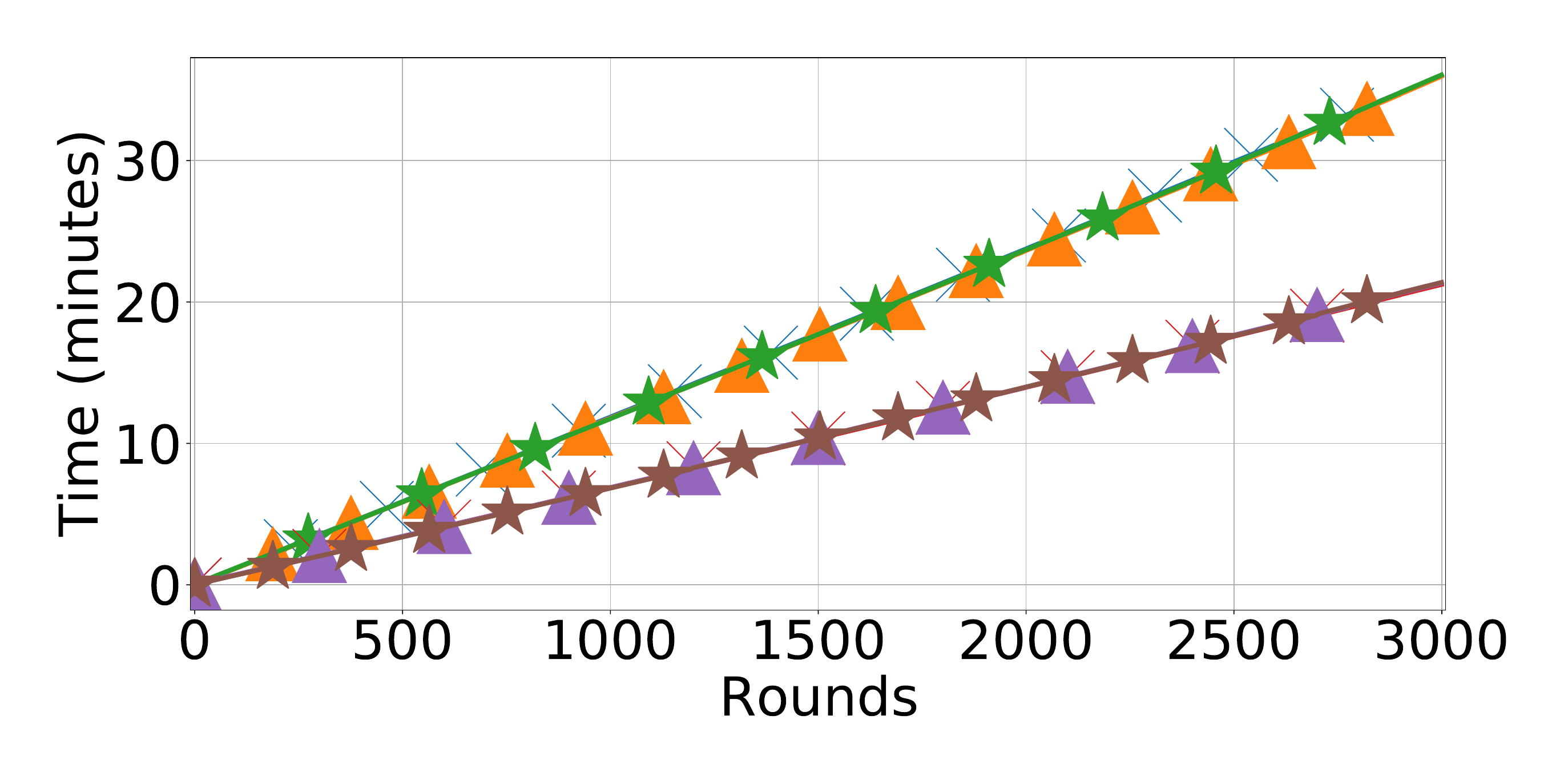} 
		\caption{{ (c) }}
	\end{subfigure}
	
	\caption{{Synthesized \modelname{linear regression} in interpolation mode, $n_i=12$, $n=50$, $d=1000$. No compression. Theoretical step sizes.}}
	\label{ch4:fig:exp_syn_1}
\end{figure*}

\begin{figure*}[ht]
	\centering
	\captionsetup[sub]{font=footnotesize,labelfont={},labelformat=empty}		
	\captionsetup[subfigure]{font=footnotesize,labelfont={},labelformat=empty}
	\captionsetup[figure]{font=footnotesize,labelfont={},labelformat=empty}
	
	\begin{subfigure}[ht]{0.80\textwidth}
		\includegraphics[width=\textwidth]{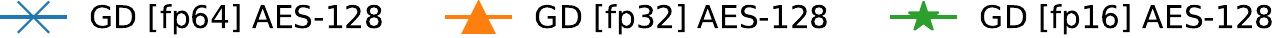} 
	\end{subfigure}
	\begin{subfigure}[ht]{0.70\textwidth}
		\includegraphics[width=\textwidth]{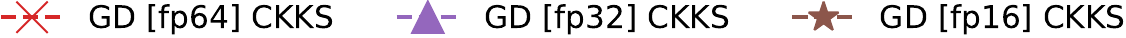} 
	\end{subfigure}

	\begin{subfigure}[ht]{0.49\textwidth}
		\includegraphics[width=\textwidth]{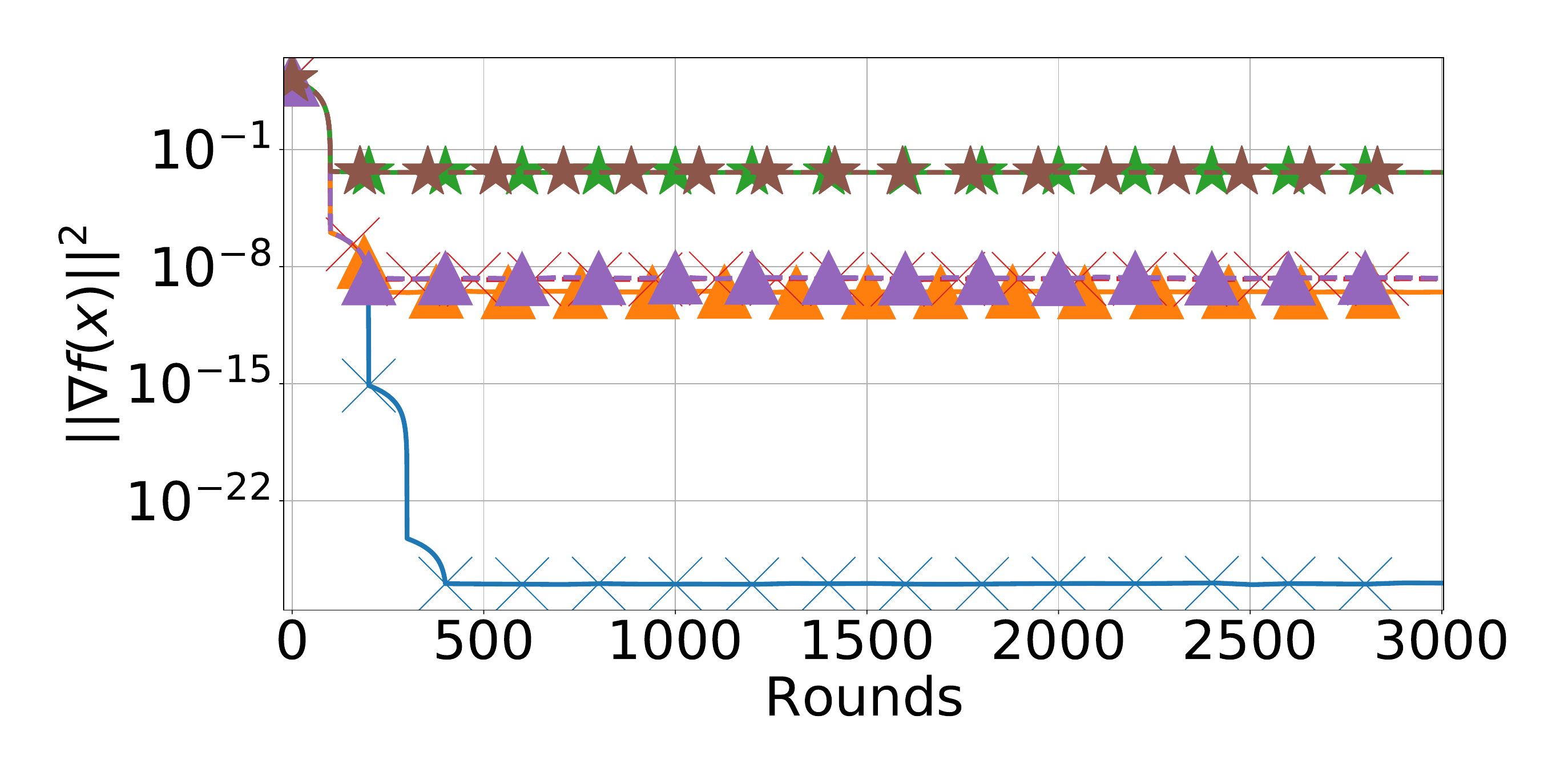} 
		\caption{{ (a) }}
	\end{subfigure}
	\begin{subfigure}[ht]{0.49\textwidth}
		\includegraphics[width=\textwidth]{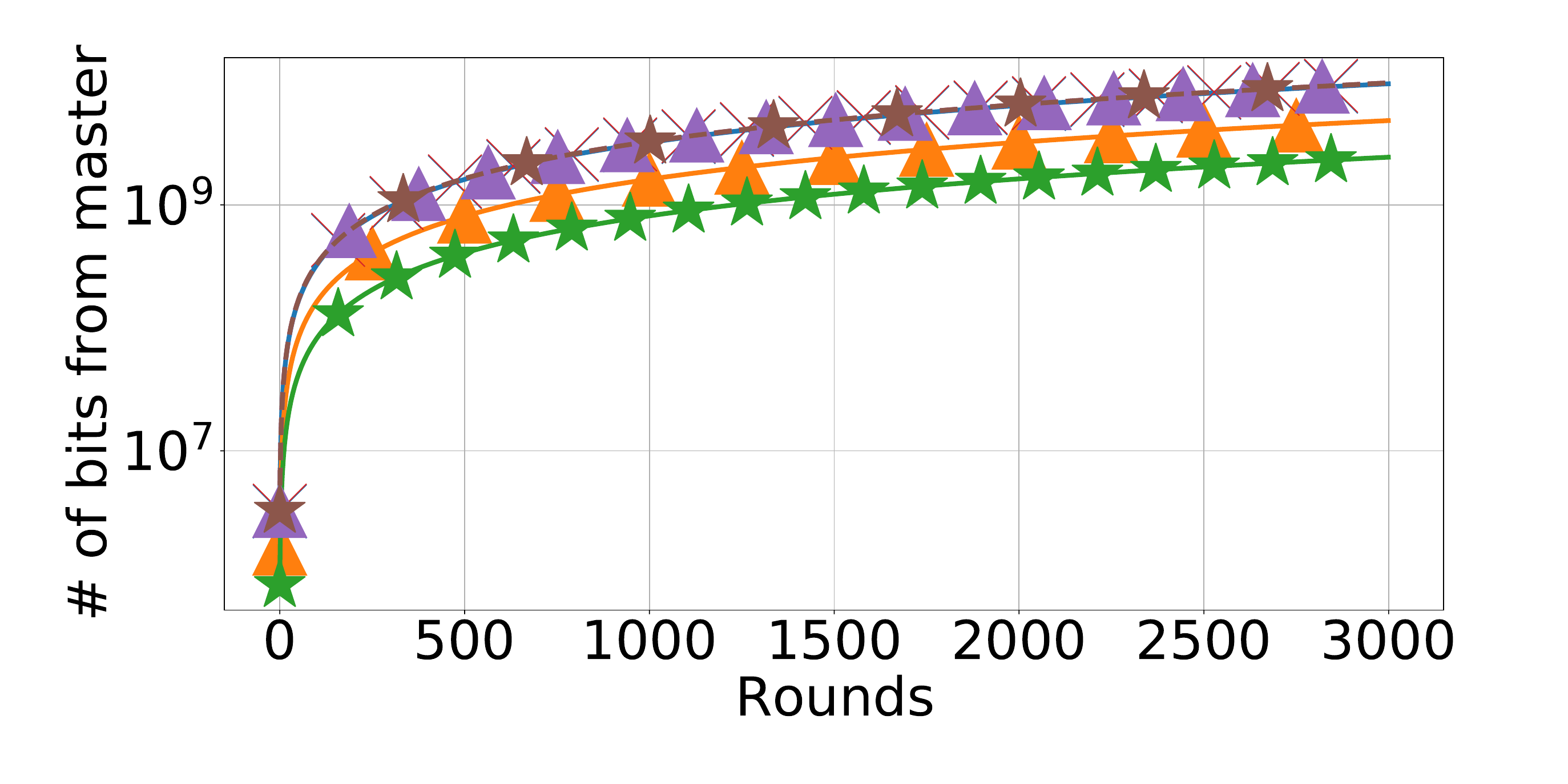} 
		\caption{{ (b) }}
	\end{subfigure}
	
	\begin{subfigure}[ht]{0.49\textwidth}
		\includegraphics[width=\textwidth]{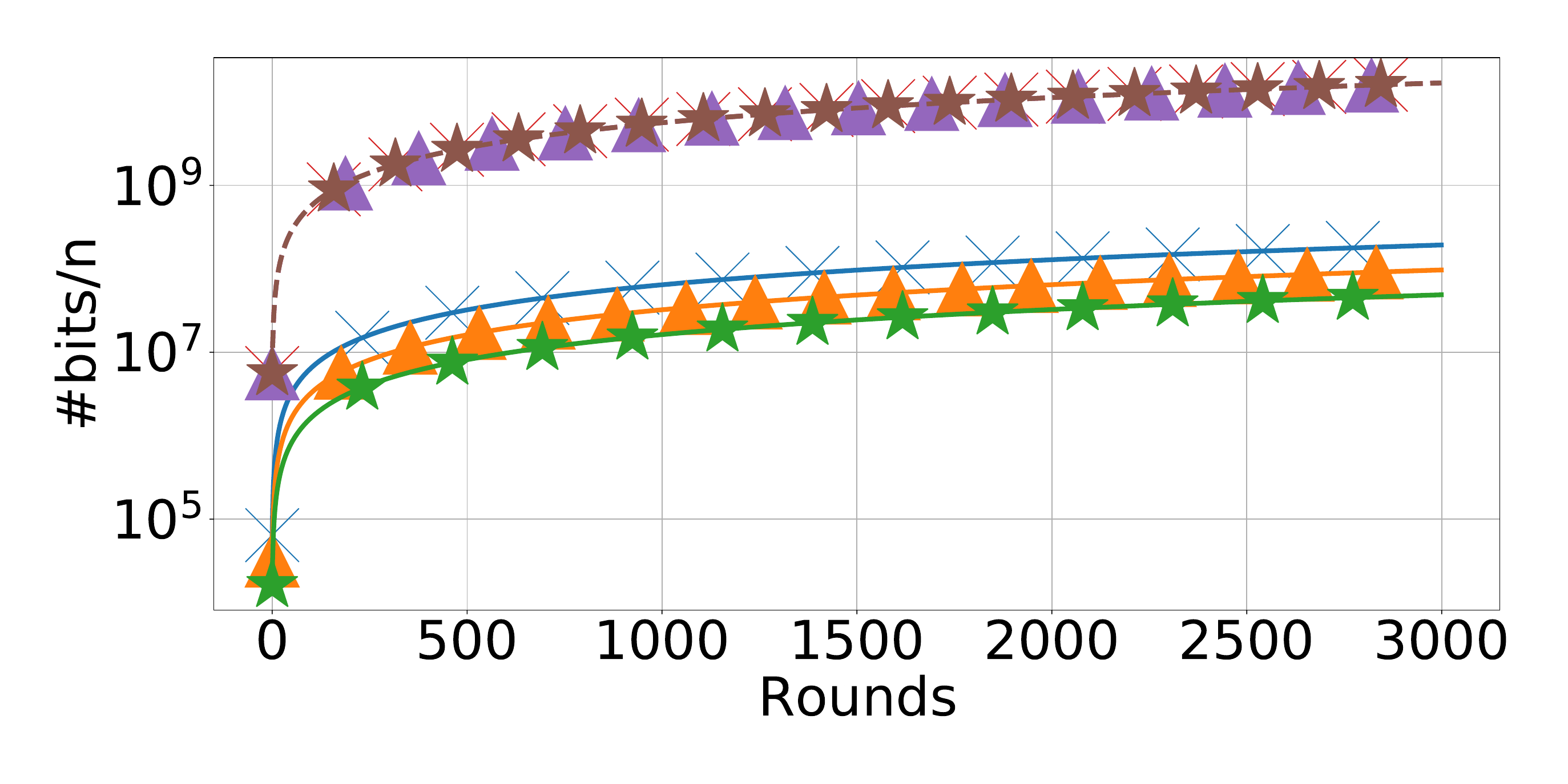} 
		\caption{{ (c) }}
	\end{subfigure}
	\begin{subfigure}[ht]{0.49\textwidth}
		\includegraphics[width=\textwidth]{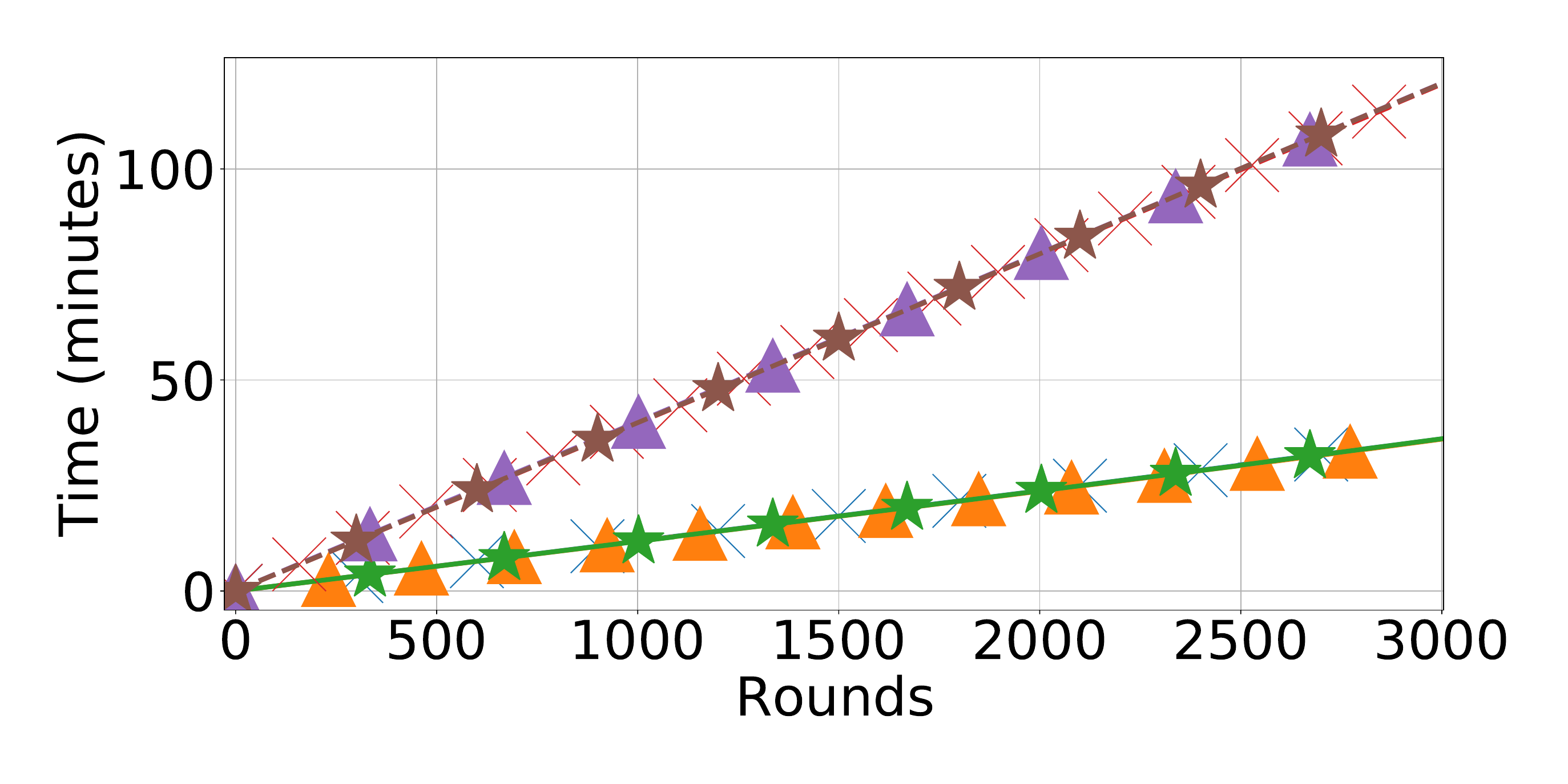} 
		\caption{{ (d) }}
	\end{subfigure}
	
	\caption{{Synthesized \modelname{linear regression} in interpolation mode, $n_i=12$, $n=50$, $d=1000$. No compression. Theoretical step sizes.}}
	\label{ch4:fig:exp_syn_2}
\end{figure*}

\paragraph{Case 1: Distributed GD with and without AES/CKKS.} We conducted experiments with $d=1000$, $n=50$, and $n_i=12$. In our designed experimental setup, we filled the Hessian $\nabla ^2 f(x)$ such that its nonzero eigenvalues lie uniformly in $[1.0, 10.0]$, therefore $L_f=10$. We used the maximum theoretical constant step size $\gamma=1/L_f$ for \algname{GD}. Figure~\ref{ch4:fig:exp_syn_1} (a) shows the impact of {IEEE-754 FP16}, {FP32}, and {FP64} formats. It shows that \algnamewithaes{GD/AES} does not hurt float arithmetic. However, \algnamewithaes{GD/AES} increases traffic between the master and clients by a factor of $n$ as we see from  Figure~\ref{ch4:fig:exp_syn_1} (b) and increases wall clock time by a factor of 1.7 as in Figure~\ref{ch4:fig:exp_syn_1} (c). Next, Figure~\ref{ch4:fig:exp_syn_2} compares \algname{GD} with \aesname{AES} and \ecryptname{CKKS}. According to Figure~\ref{ch4:fig:exp_syn_2} (a), \algname{GD/CKKS} and FP64 arithmetic, it is possible to obtain $| \nabla f(x^k)|^2 \approx 10^{-9}$, when \algnamewithaes{GD/AES} attains $| \nabla f(x^k)|^2 \approx 10^{-23}$. From Figure~\ref{ch4:fig:exp_syn_2} (b, c) we see that \algname{GD/CKKS} increases load from master  slightly compared to \algnamewithaes{GD/AES}, but increases the load from clients by $\times 10^4$. As per Figure~\ref{ch4:fig:exp_syn_2} (d), \ecryptname{CKKS} is approximately $\times 3$ slower compared to \aesname{AES}. Using \algnamewithaes{GD/AES} is reasonable only if  $d$ and $n$ are small.

\begin{figure*}[ht!]
	\centering
	\captionsetup[sub]{font=footnotesize,labelfont={},labelformat=empty}		
	\captionsetup[subfigure]{font=footnotesize,labelfont={},labelformat=empty}
	\captionsetup[figure]{font=footnotesize,labelfont={},labelformat=empty}
	
	\begin{subfigure}[ht]{1.0\textwidth}
		\includegraphics[width=\textwidth]{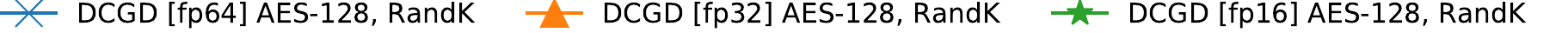}
	\end{subfigure}
	\begin{subfigure}[ht]{0.75\textwidth}
		\includegraphics[width=\textwidth]{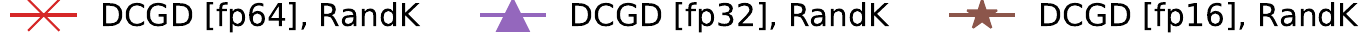}
	\end{subfigure}
	
	\begin{subfigure}[ht]{0.49\textwidth}
		\includegraphics[width=\textwidth]{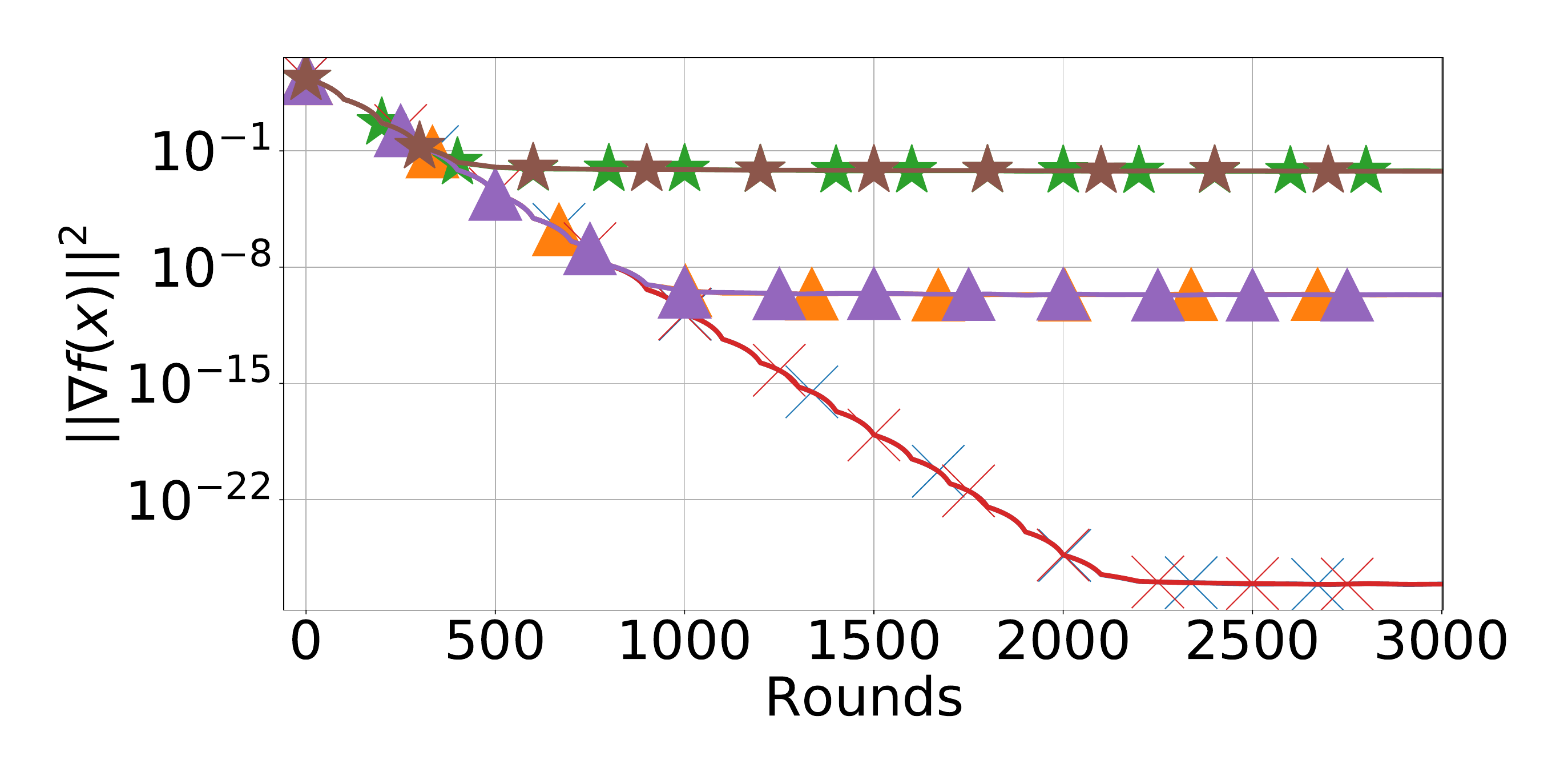} 
		\caption{{ (a) }}
	\end{subfigure}
	\begin{subfigure}[ht]{0.49\textwidth}
		\includegraphics[width=\textwidth]{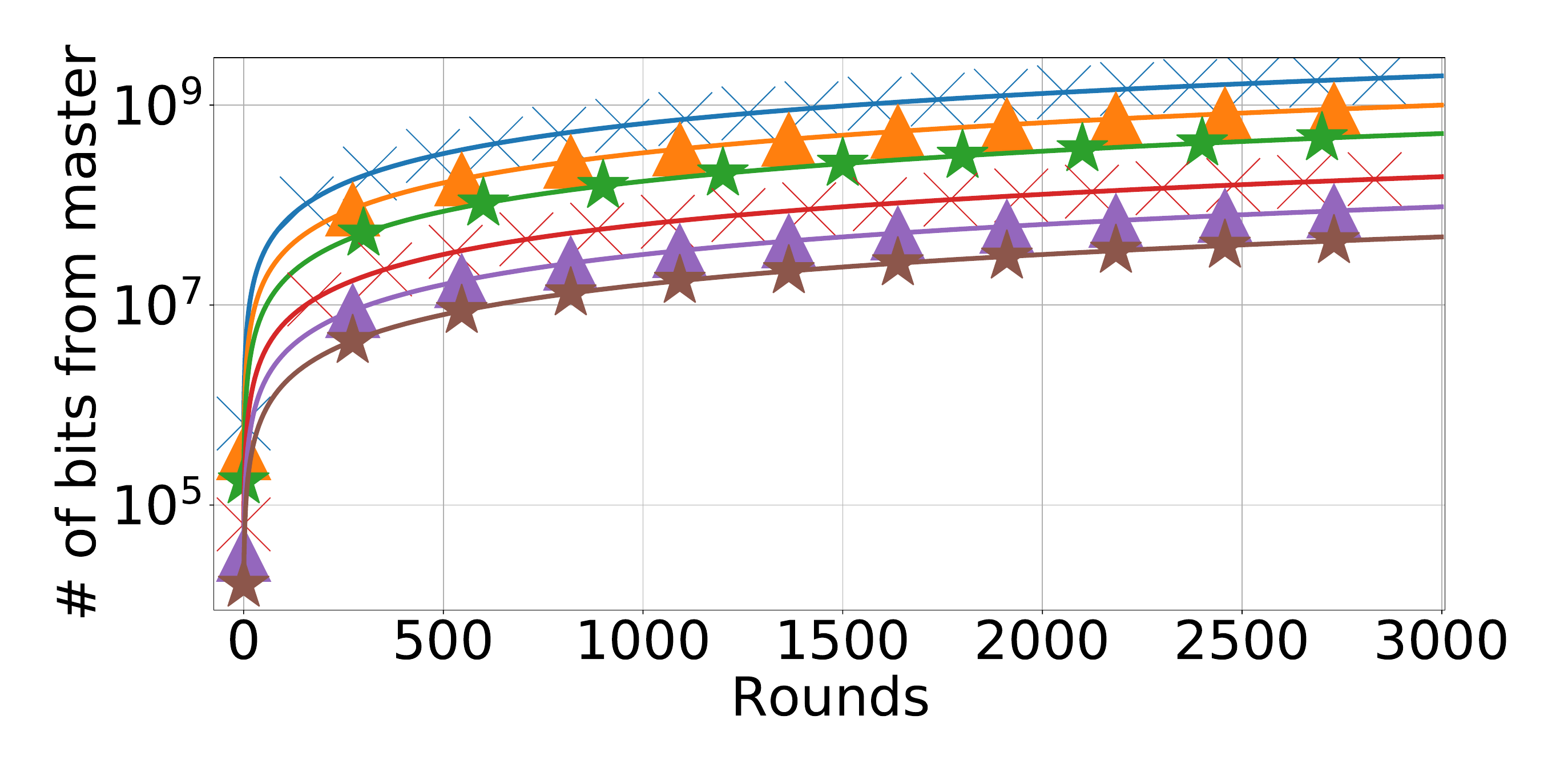} 
		\caption{{ (b) }}
	\end{subfigure}
	\begin{subfigure}[ht]{0.49\textwidth}
		\includegraphics[width=\textwidth]{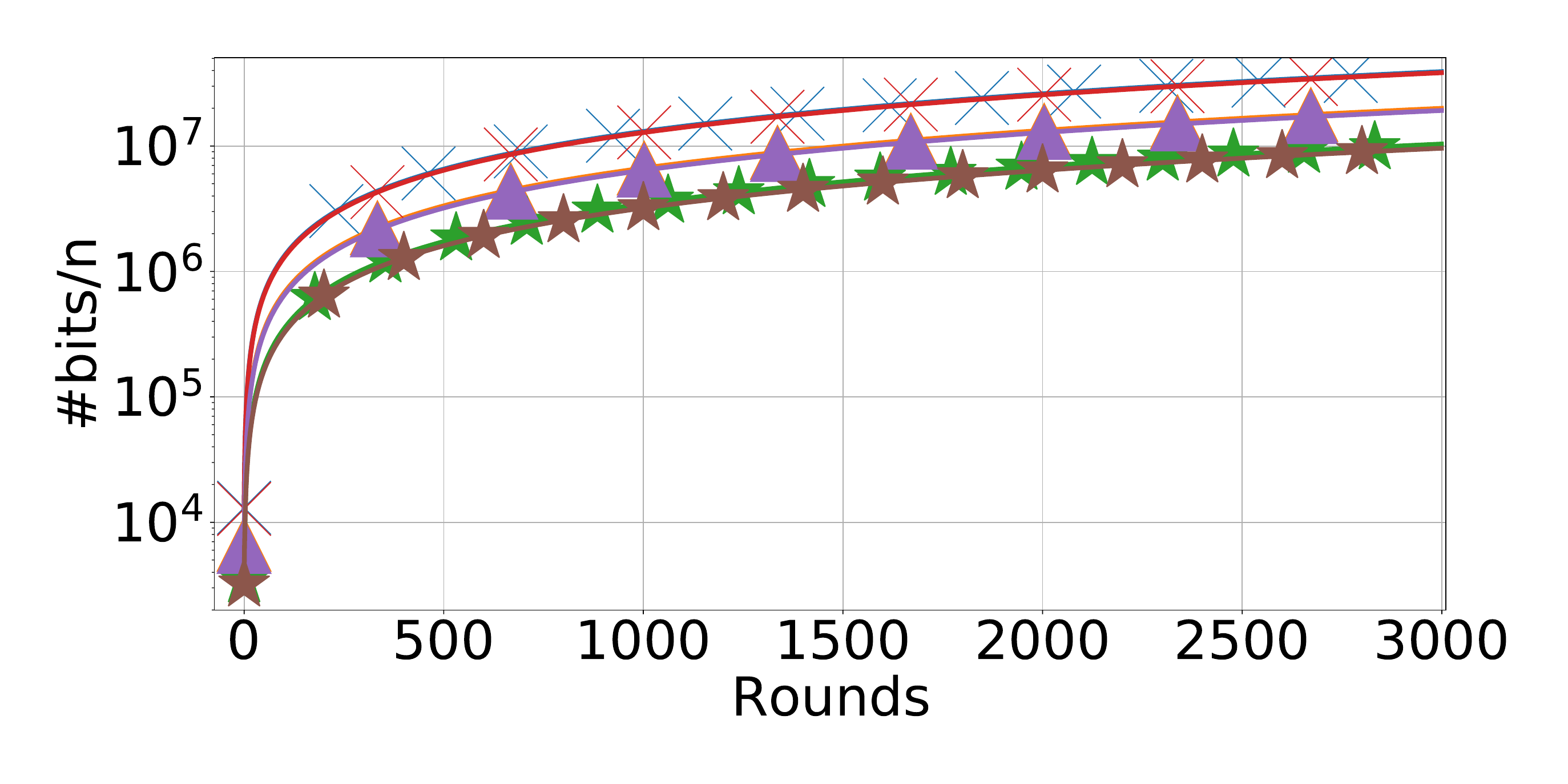} 
		\caption{{ (c) }}
	\end{subfigure}
	
	\caption{{Synthesized \modelname{linear regression} in interpolation, $n_i=12$, $n=50$, $d=1000$. Compressors: \compname{RandK}[$K=0.2d$]. Theoretical step sizes.}}
	\label{ch4:fig:exp_syn_3}
\end{figure*}

\begin{figure*}[ht]
	\centering
	\captionsetup[sub]{font=footnotesize,labelfont={},labelformat=empty}		
	\captionsetup[subfigure]{font=footnotesize,labelfont={},labelformat=empty}
	\captionsetup[figure]{font=footnotesize,labelfont={},labelformat=empty}
	
	\begin{subfigure}[ht]{\textwidth}
		\includegraphics[width=\textwidth]{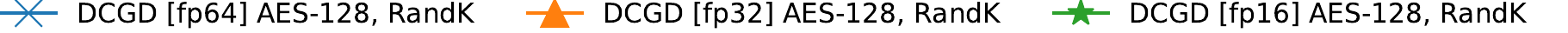}
	\end{subfigure}
	\begin{subfigure}[ht]{0.86\textwidth}
		\includegraphics[width=\textwidth]{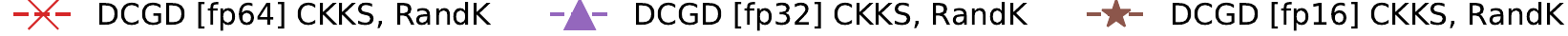}
	\end{subfigure}
	
	\begin{subfigure}[ht]{0.49\textwidth}
		\includegraphics[width=\textwidth]{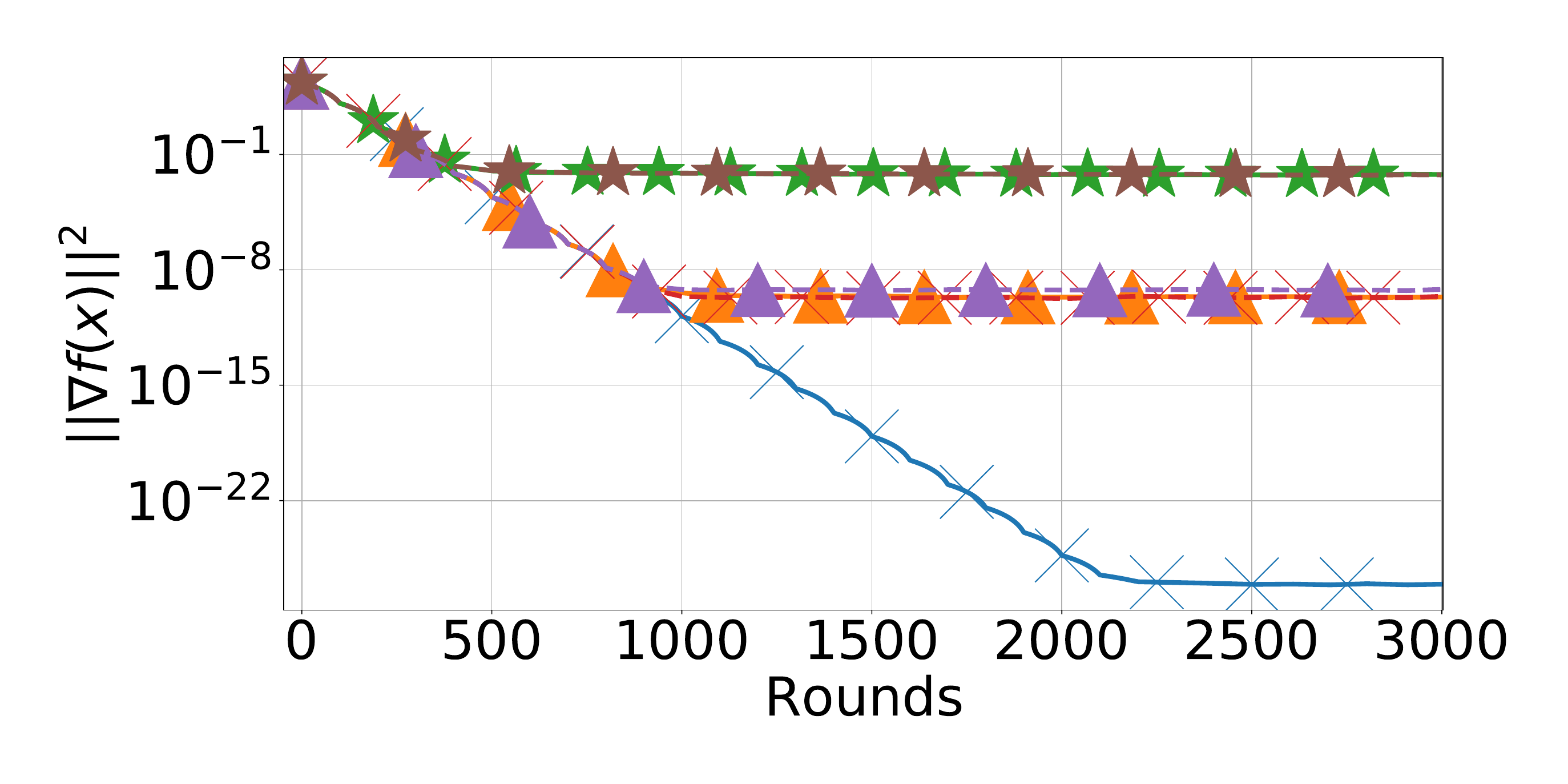} 
		\caption{{ (a) }}
	\end{subfigure}
	\begin{subfigure}[ht]{0.49\textwidth}
		\includegraphics[width=\textwidth]{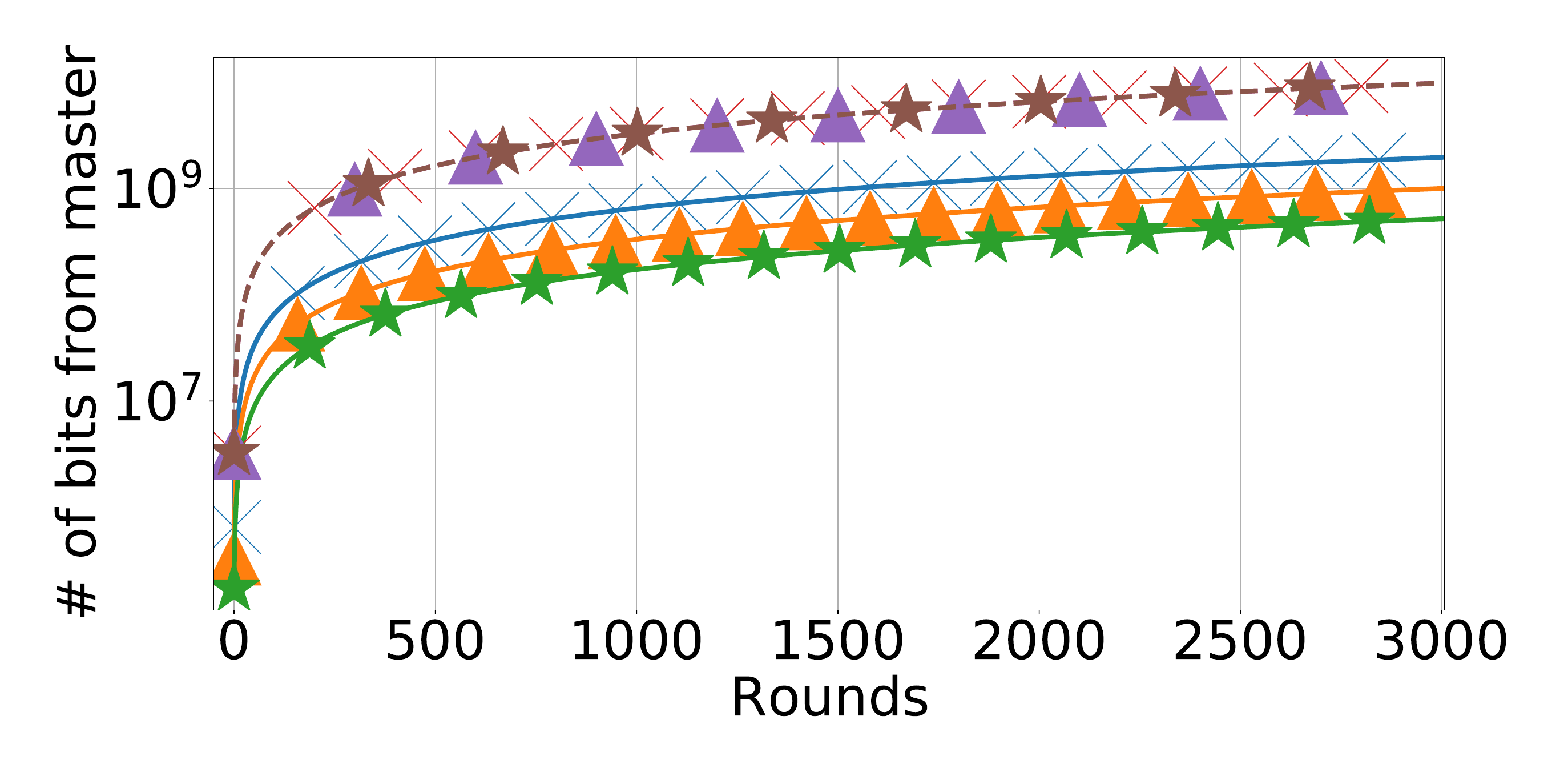} 
		\caption{{ (b) }}
	\end{subfigure}
	\begin{subfigure}[ht]{0.49\textwidth}
		\includegraphics[width=\textwidth]{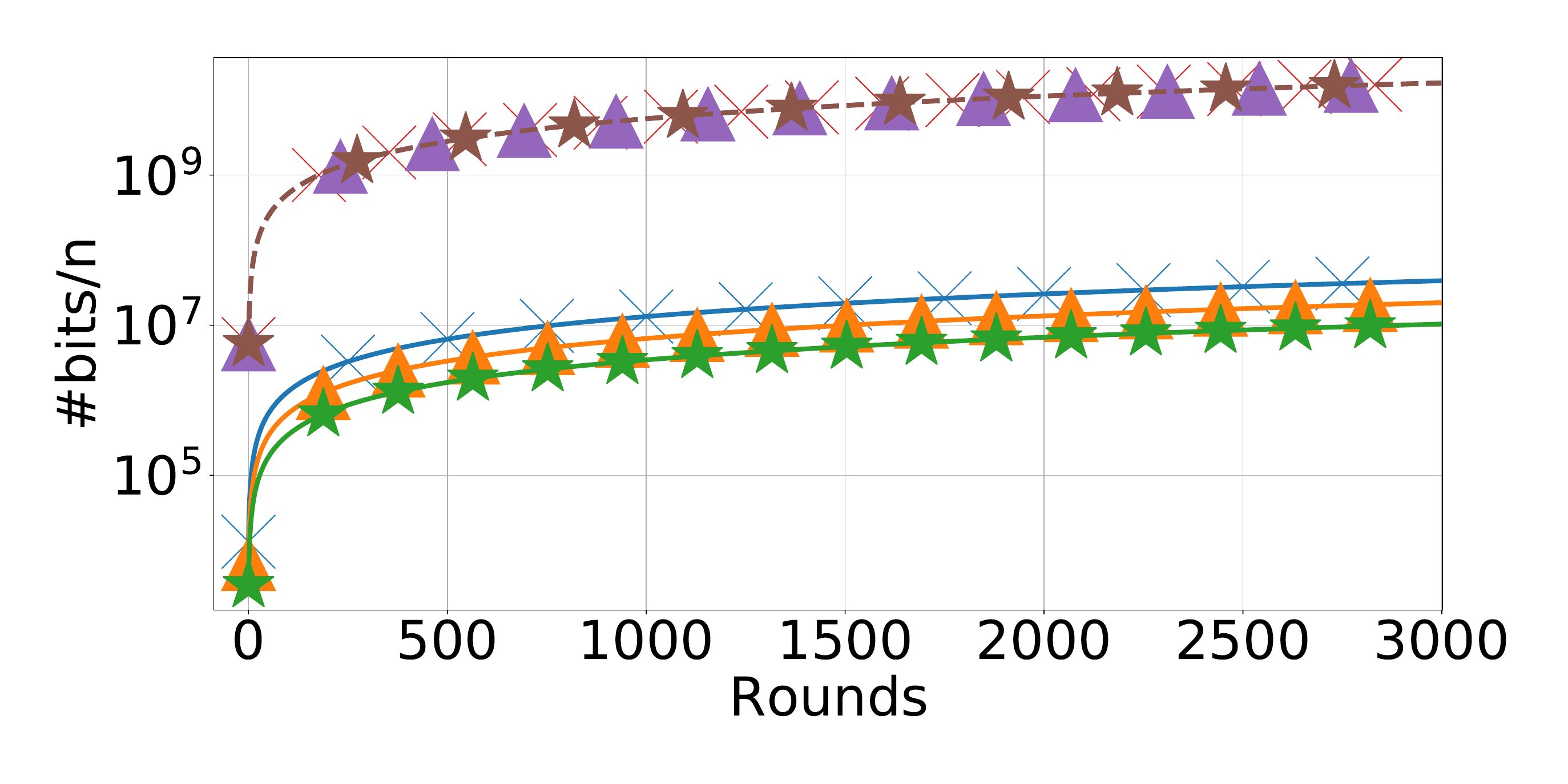} 
		\caption{{ (c) }}
	\end{subfigure}
	\begin{subfigure}[ht]{0.49\textwidth}
		\includegraphics[width=\textwidth]{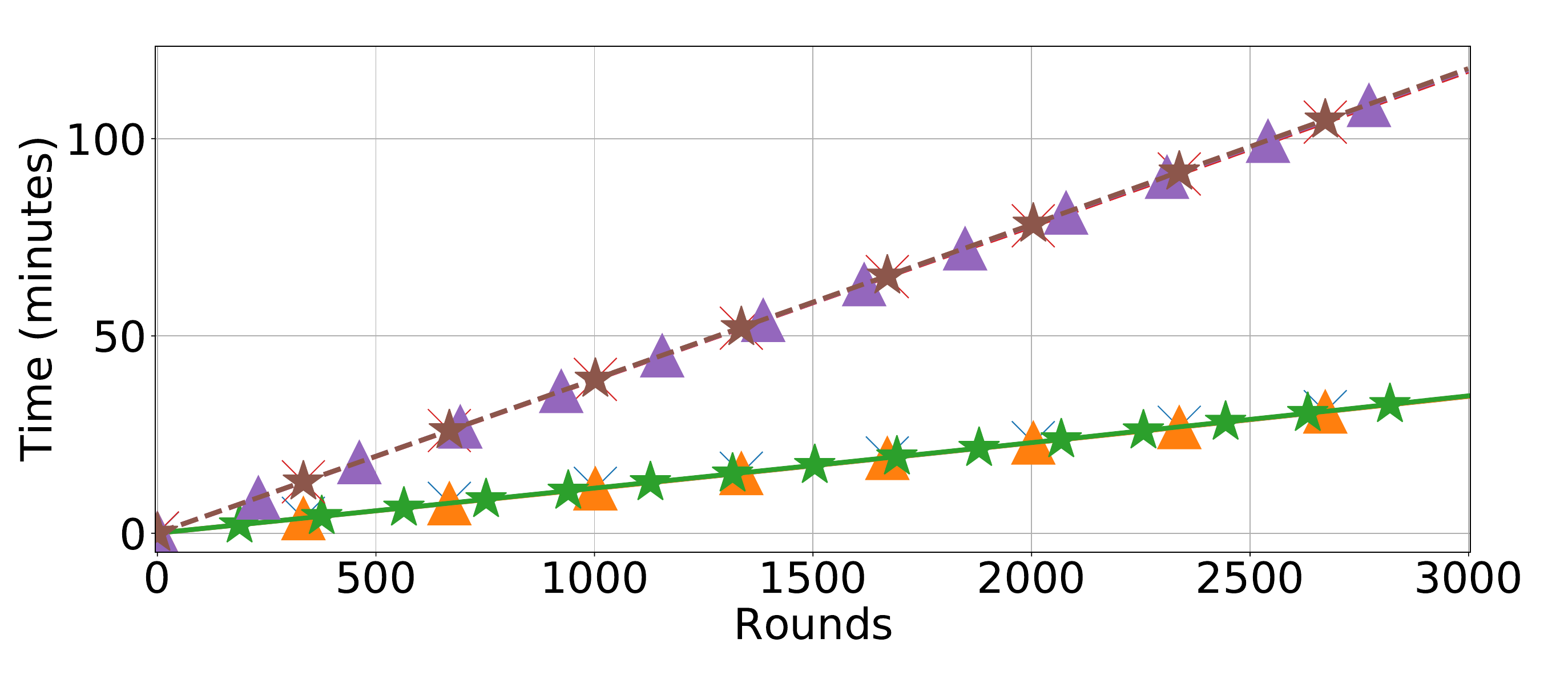} 
		\caption{{ (d) }}
	\end{subfigure}
	
	\caption{{Synthesized \modelname{linear regression} in interpolation, $n_i=12$, $n=50$, $d=1000$. Compressors: \compname{RandK} [$K=0.2d$]. Theoretical step sizes.}}
	\label{ch4:fig:exp_syn_4}
\end{figure*}

\paragraph{Case 2: DCGD with AES/CKKS.} In this experiment, we employ \algname{DCGD} with \compname{RandK} sparsification compressor to analyze the possibility of gradient sparsification while preserving the client's privacy from the master. To compress $\nabla f_i(x)$ each client creates a set $S_i\subset\{1,2,\dots,d\}$ of size $K$ chosen uniformly at random, and compute $C(\nabla f_i(x)) \eqdef \dfrac{d}{K} \sum_{j \in S_i} [\nabla f_i(x)]_j \cdot e_j$, where $e_j$ are unit vectors of standard basis of $\mathbb{R}^d$. Results are presented in Figure~\ref{ch4:fig:exp_syn_3}, and Figure~\ref{ch4:fig:exp_syn_4}. Figure~\ref{ch4:fig:exp_syn_3} (a) and Figure~\ref{ch4:fig:exp_syn_4} (a) demonstrates that using \aesname{AES} does not lead to numerical issues, whereas \ecryptname{CKKS} for FP64 does. In \algname{GD}, the master broadcasts a vector from $\mathbb{R}^d$ in each round. For \algnamewithaes{DCGD/RandK/AES}, if clients can reconstruct sparsified indices, then the master has to broadcast $d = K \cdot n = \dfrac{1}{5} \cdot n$ encrypted scalars and $32n$ bytes from employing \textit{nonce}, and \textit{mac} for privacy and integrity. This process reduces the number of bits transmitted from the master to the clients by a factor $\times 5$, compared to standard \algnamewithaes{GD/AES}, as depicted in Figure~\ref{ch4:fig:exp_syn_3} (b) and Figure~\ref{ch4:fig:exp_syn_1} (b). Figure~\ref{ch4:fig:exp_syn_4} (c) shows that \ecryptname{CKKS} does not leverage the sparsity. For \ecryptname{HE} schemas, encoding of any two vectors (for example, sparse and dense) should be indistinguishable due to semantic security requirements. However, from a computational perspective, ignoring sparsity is sometimes highly impractical, and this gap presents an open research question for \ecryptname{HE}. Figure~\ref{ch4:fig:exp_syn_4} (b) highlights that \aesname{AES} adapts to any bit representation of scalars. In contrast, \libname{TenSEAL} \citep{benaissa2021tenseal} implementation of \ecryptname{CKKS} does not exhibit this property.

\begin{figure*}[ht!]
	\centering
	\captionsetup[sub]{font=footnotesize,labelfont={},labelformat=empty}		
	\captionsetup[subfigure]{font=footnotesize,labelfont={},labelformat=empty}
	\captionsetup[figure]{font=footnotesize,labelfont={},labelformat=empty}
	
	\begin{subfigure}[ht]{\textwidth}
		\includegraphics[width=\textwidth]{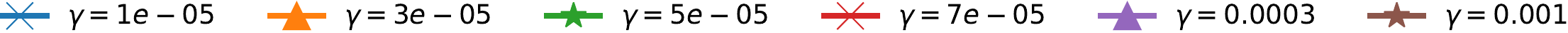} 
	\end{subfigure}
	
	\begin{subfigure}[ht]{0.73\textwidth}
		\includegraphics[width=\textwidth]{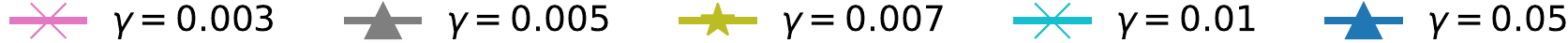} 
	\end{subfigure}
	
	\begin{subfigure}[ht]{0.49\textwidth}
		\includegraphics[width=\textwidth]{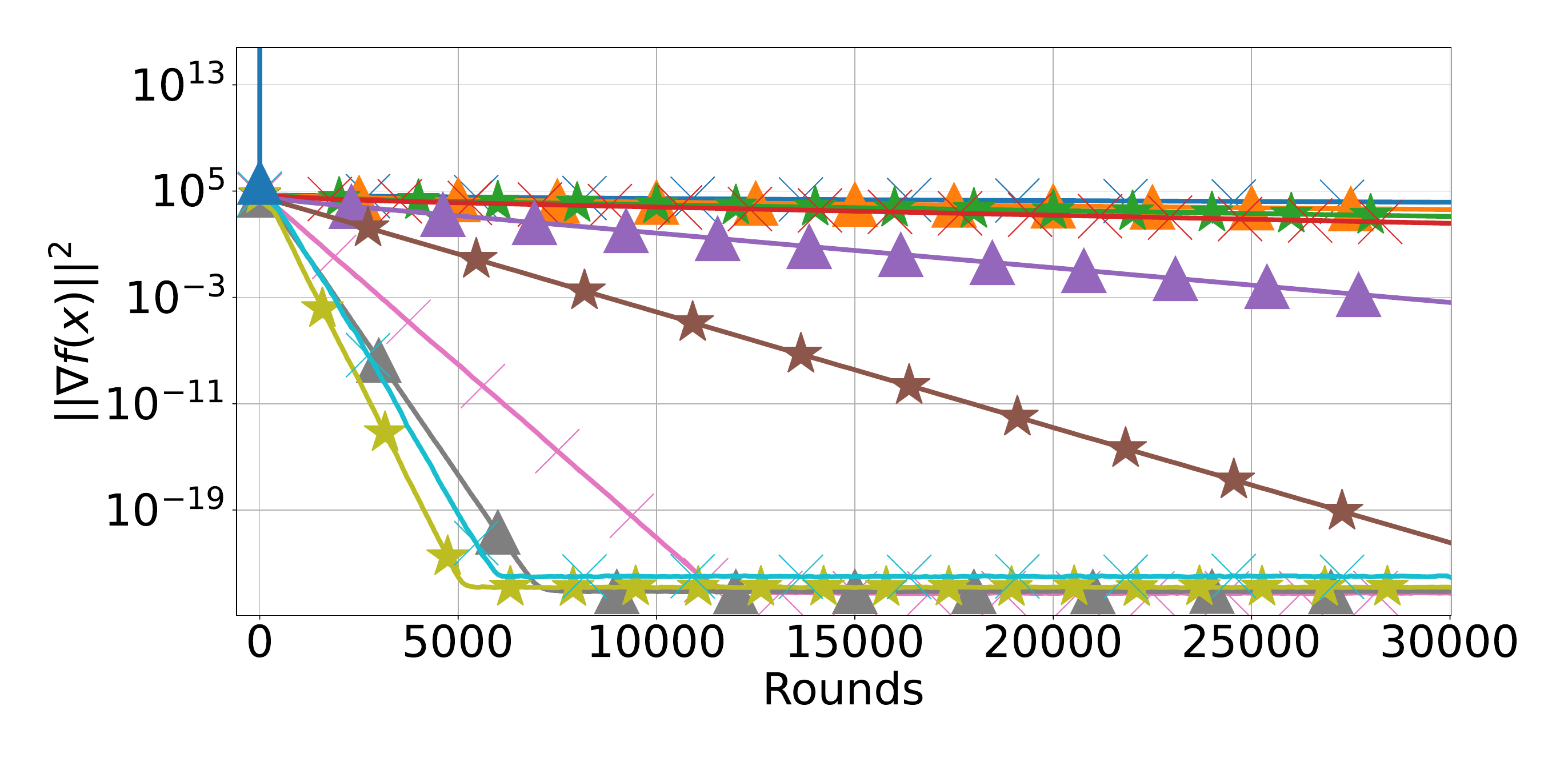}
		\caption{{ (a) }}
	\end{subfigure}
	\begin{subfigure}[ht]{0.49\textwidth}
		\includegraphics[width=\textwidth]{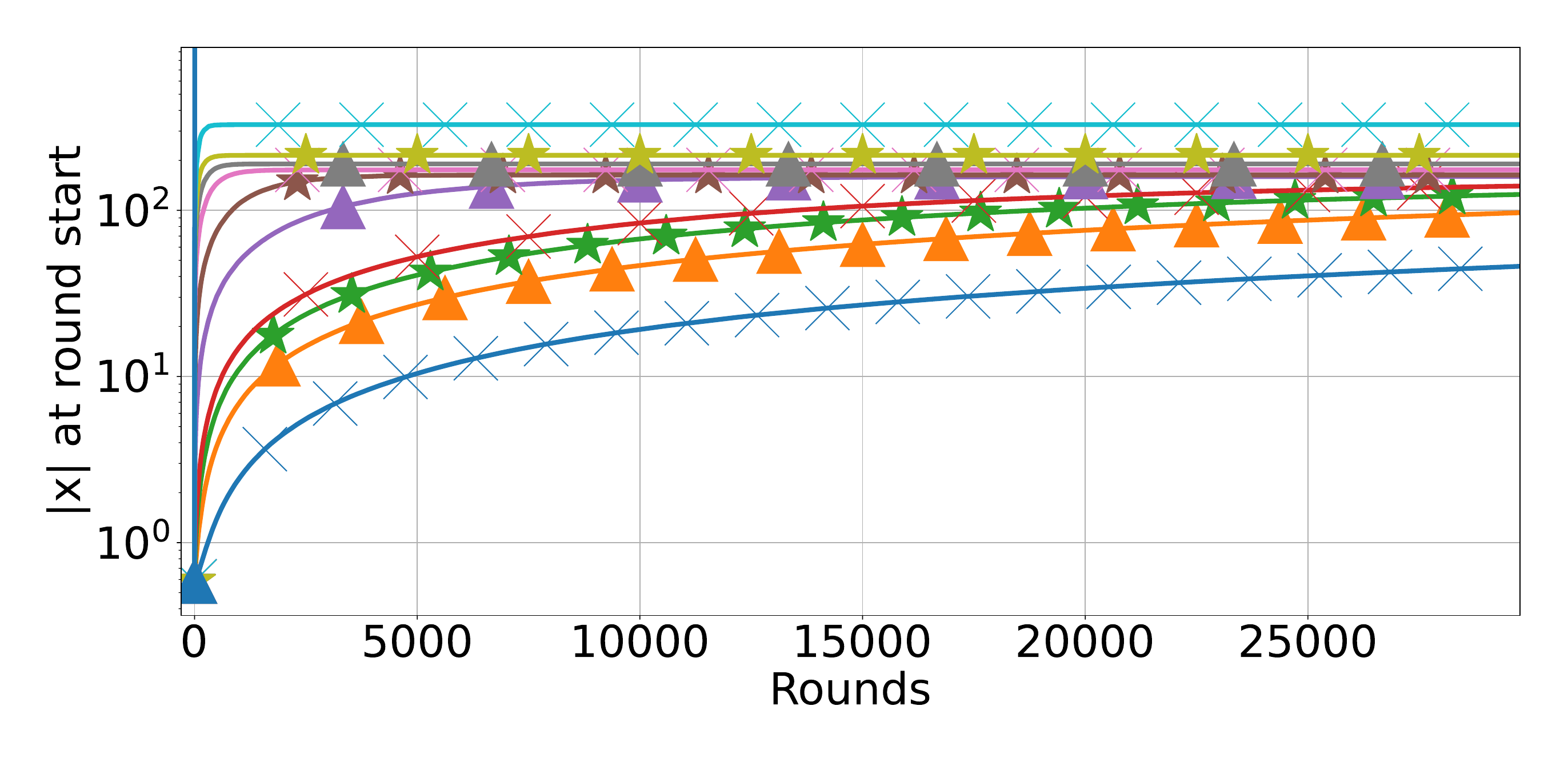}
		\caption{{ (b) }}
	\end{subfigure}
	\begin{subfigure}[ht]{0.49\textwidth}
		\includegraphics[width=\textwidth]{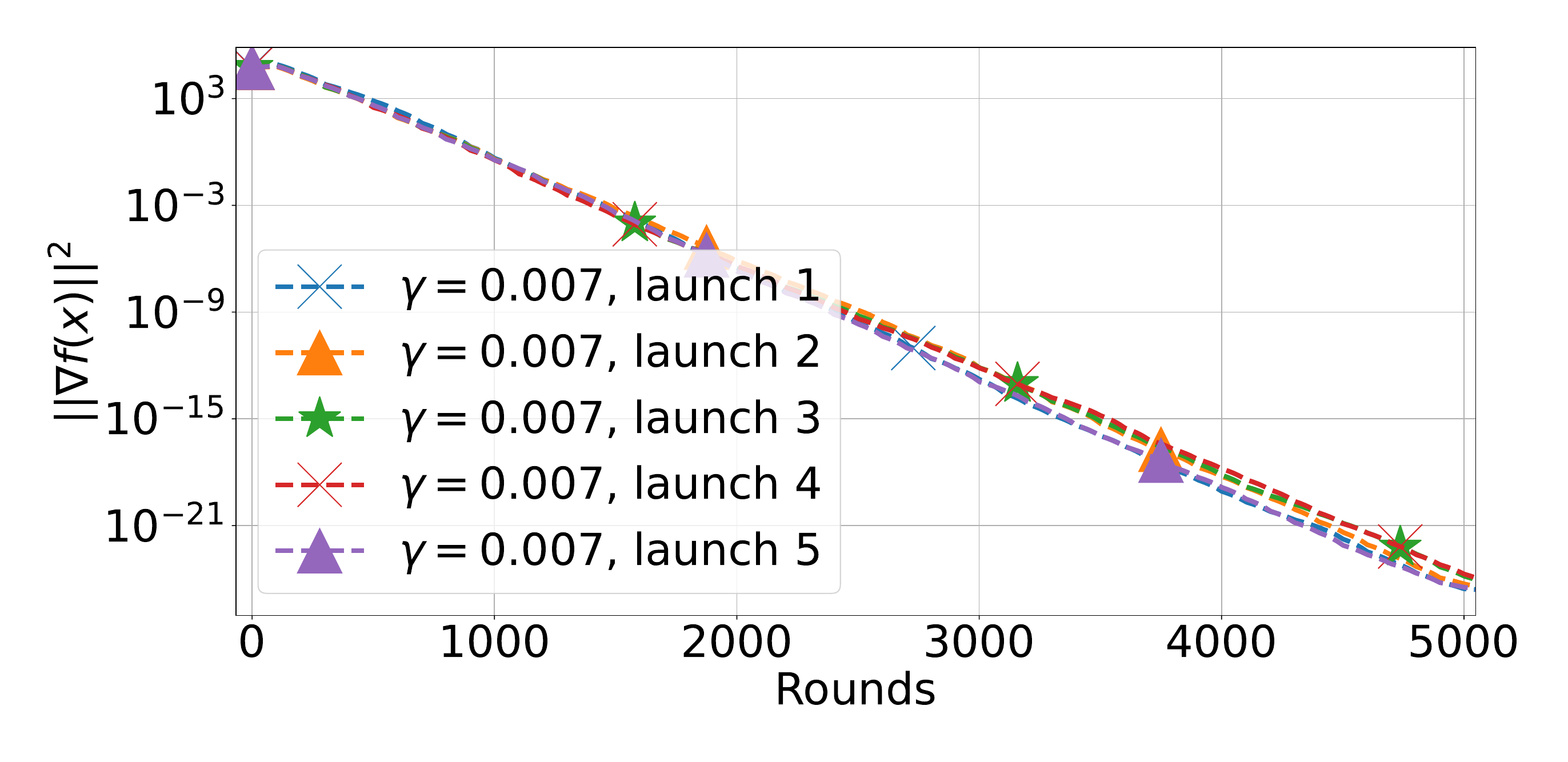}
		\caption{{ (c) }}
	\end{subfigure}
	
	\caption{{Tuning step size $\gamma$ without decay for \algname{DCGD/PermK}. Synthesized \modelname{linear regression} in interpolation, $5$ launches, FP64.}}
	\label{ch4:fig:exp_syn_5}
\end{figure*}

\paragraph{Case 3: DCGD with PermK.} Both \algnamewithaes{GD/AES} and \algnamewithaes{DCGD/AES} require a significant amount of data to be sent from the master to the clients. In the case of using correlated compressors as \compname{PermK}, all clients do not intersect in supports of sparsified gradients by design. The encryption of the global direction can be obtained by concatenating encrypted messages from clients. When using \compname{PermK}, clients do not need to perform any aggregation on their side, and decryption of the whole global direction obtained from master can be done in $\mathcal{O}(d)$ independent of $n$. We aim to find an approximate $\gamma$ for \algname{DCGD/PermK}. We generated $5$ problems with matrices $\mA_i \sim U[0,1)^{n_i \times d}$, projected $\mA=[\mA_1, \dots, \mA_n]^\top$ to have $L_{f} = 10$, and computed $b_i \eqdef \mA_i x_{\mathrm{fixed}}$. We tested various step sizes demonstrated in Figure~\ref{ch4:fig:exp_syn_5}. We found that using a step size $\gamma \ge \dfrac{1}{2L_{f}}=0.05$ led to divergence, as shown in Figure~\ref{ch4:fig:exp_syn_5} (a, b). From Figure~\ref{ch4:fig:exp_syn_5} (a), we see that the method exhibits linear convergence without oscillation near the solution, similar to \algname{DCGD} with \compname{RandK} (Here $\nabla f_i (x^*)=0, \forall i \in [n]$, because $d > n_i \cdot n$). From Figure~\ref{ch4:fig:exp_syn_5} (c), we see that the variance of the optimization path using fixed step size $\gamma=0.007$ and fixed $d, n_i, n, L_f$ is negligible.

\begin{figure*}[ht!]
	\centering
	\captionsetup[sub]{font=footnotesize,labelfont={},labelformat=empty}		
	\captionsetup[subfigure]{font=footnotesize,labelfont={},labelformat=empty}
	\captionsetup[figure]{font=footnotesize,labelfont={},labelformat=empty}
	
	\begin{subfigure}[ht]{\textwidth}
		\includegraphics[width=\textwidth]{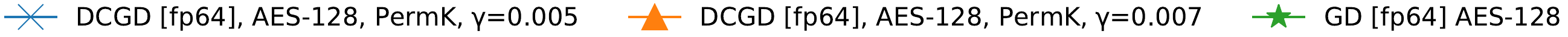}
	\end{subfigure}
	\begin{subfigure}[ht]{\textwidth}
		\includegraphics[width=\textwidth]{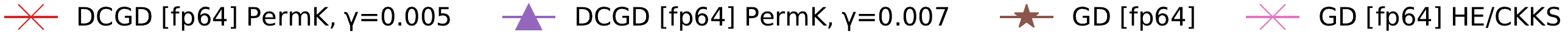} 
	\end{subfigure}
	
	\begin{subfigure}[ht]{0.49\textwidth}
		\includegraphics[width=\textwidth]{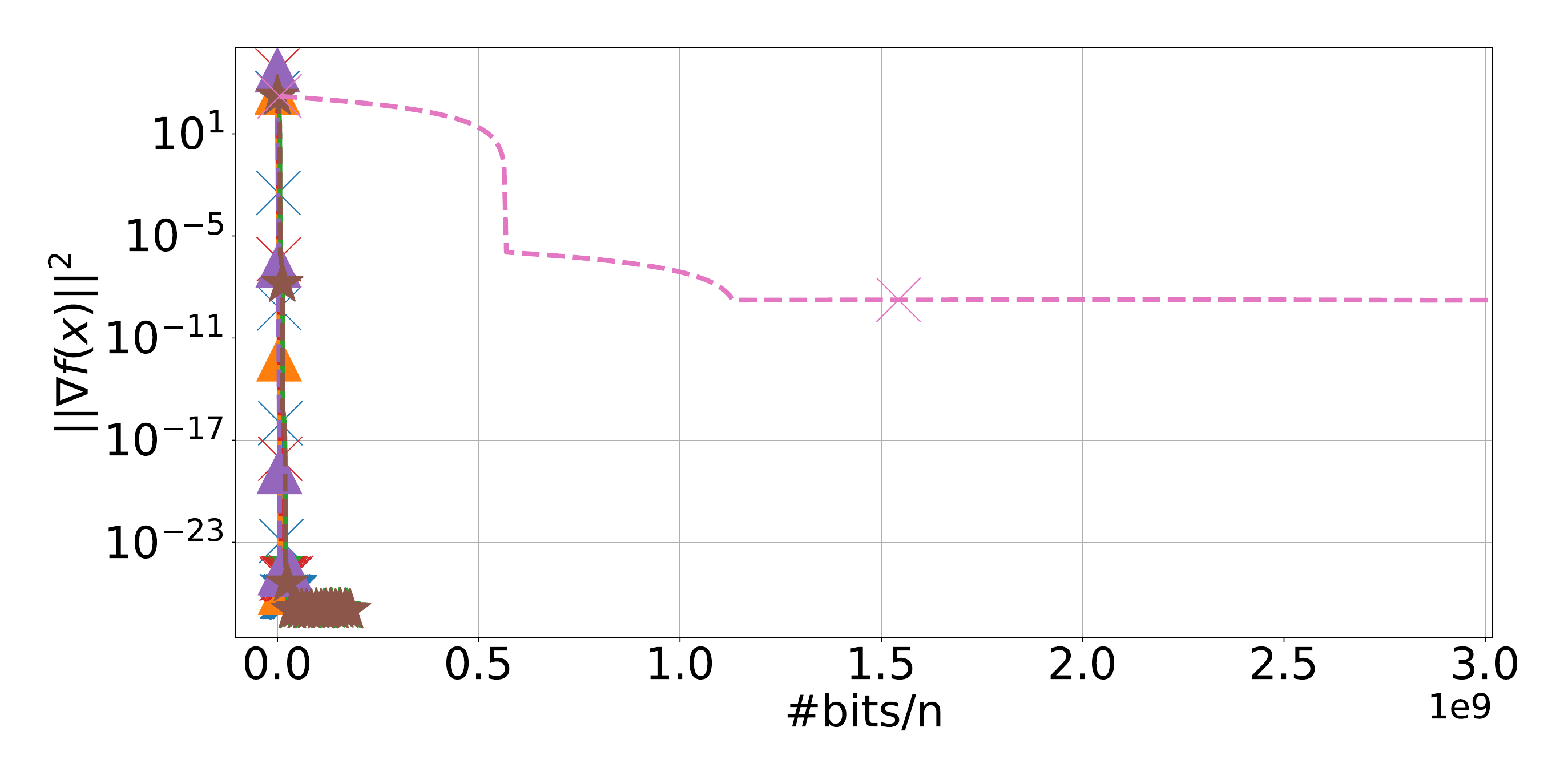} 
		\caption{{ (a) }}
	\end{subfigure}
	\begin{subfigure}[ht]{0.49\textwidth}
		\includegraphics[width=\textwidth]{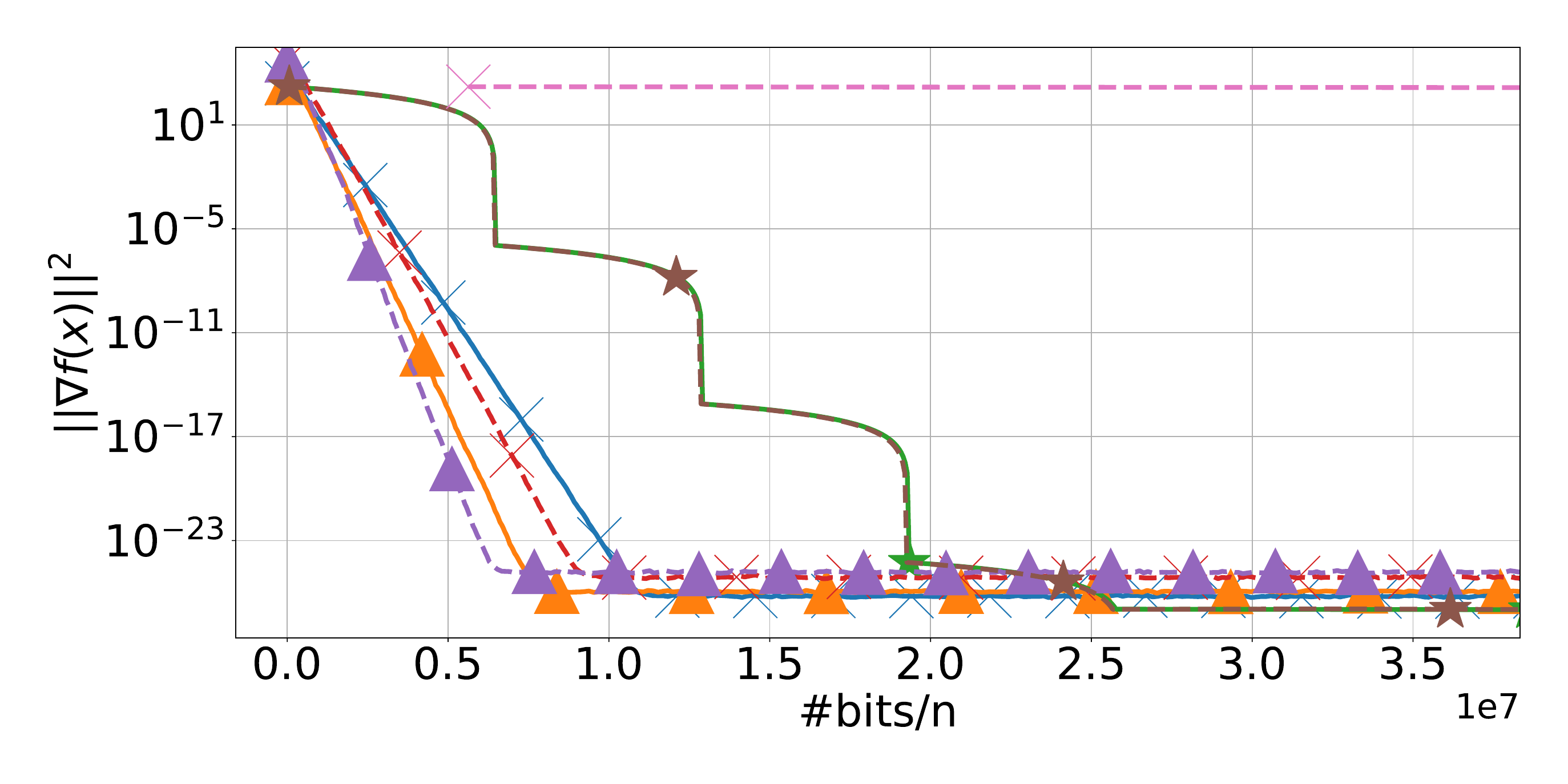} 
		\caption{{ (b) }}
	\end{subfigure}
	\begin{subfigure}[ht]{0.49\textwidth}
		\includegraphics[width=\textwidth]{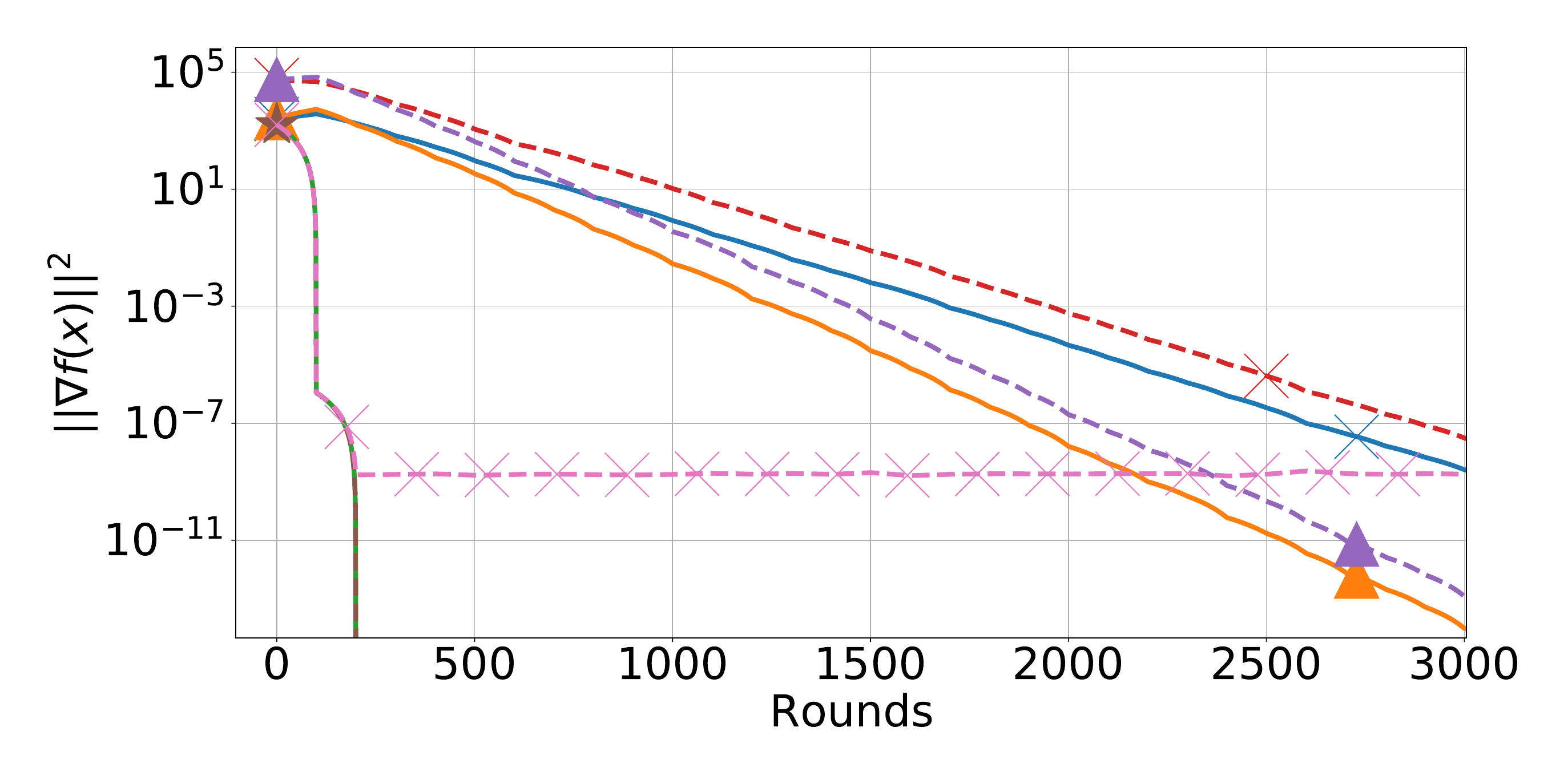} 
		\caption{{ (c) }}
	\end{subfigure}
	\begin{subfigure}[ht]{0.49\textwidth}
		\includegraphics[width=\textwidth]{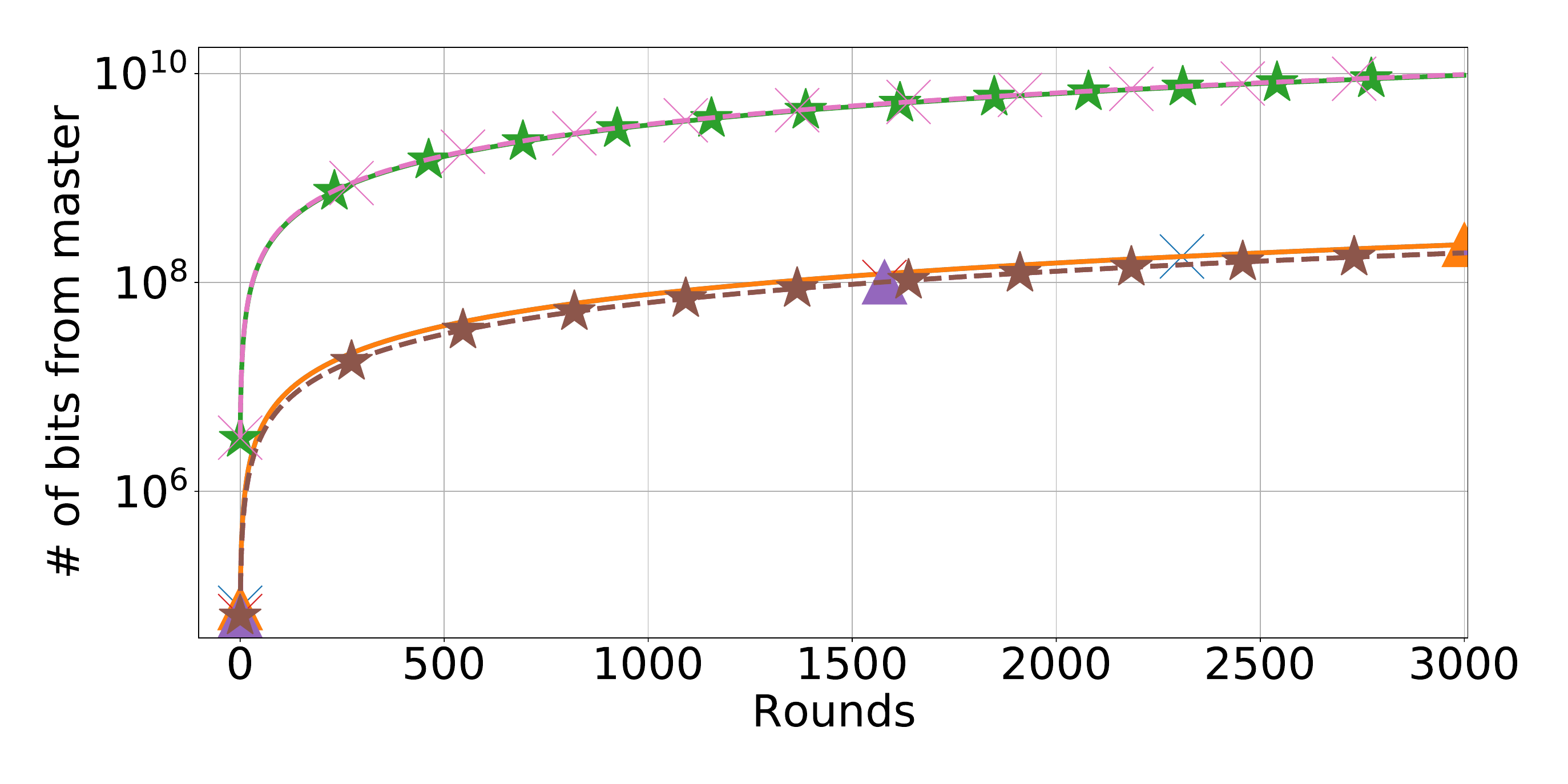} 
		\caption{{ (d) }}
	\end{subfigure}	
	\caption{{\modelname{Linear regression} in an interpolation. \algname{DCGD} uses tuned step size. \algname{GD}, \algname{GD/CKKS}, \algnamewithaes{GD/AES} use theoretical.}}
	\label{ch4:fig:exp_syn_6}
\end{figure*}

\paragraph{Comparison of {DCGD/PermK/AES} and {GD/CKKS}.} Figure~\ref{ch4:fig:exp_syn_6} compares \algname{GD}, \algname{GD/CKKS}, \algname{GD/PermK} with \algnamewithaes{GD/AES}, \algnamewithaes{GD/PermK/AES}. We see that the \ecryptname{CKKS} schema does not leverage the sparsity of vectors, making sparsification-based compression ineffective in reducing communication during privacy-preserving training. Therefore, there is no benefit from using \compname{RandK} for \algname{DCGD/CKKS}, and it is better to use vanilla \algname{GD}. For \algname{GD}, we used theoretical step size, which in practice is extremely tight. If communication isn't free, Figure~\ref{ch4:fig:exp_syn_6} (a) and Figure~\ref{ch4:fig:exp_syn_6} (d) suggest that \ecryptname{CKKS} is impractical in settings where client-master communication is a bottleneck. Figure~\ref{ch4:fig:exp_syn_6} (b) shows that \algnamewithaes{GD/AES} does not increase client-to-master traffic but does significantly increase master-to-client traffic, as seen in Figure~\ref{ch4:fig:exp_syn_6} (d). Figure~\ref{ch4:fig:exp_syn_6} (c) shows that the best convergence in terms of rounds is attained for \algname{GD} or \algnamewithaes{GD/AES} with preserving security. If communication is free, \algname{GD/CKKS} remains suboptimal due to the approximate nature of floating-point operations in \ecryptname{CKKS}. Next, suppose the key metric is the convergence of $\|{\nabla f(x^k)}\|$ relative to the number of bits transmitted from client to master. Figure~\ref{ch4:fig:exp_syn_6} (b) shows that \algname{GD/CKKS} uses approximately $0.6 \cdot 10^7$ bits per client after the first round of optimization. \algnamewithaes{DCGD/PermK/AES} with this transfers can attain $\|{\nabla f(x)}\|^2 \approx 10^{-20}$.

\begin{figure*}[ht]
	\centering
	\captionsetup[subfigure]{labelformat=empty}
	\begin{subfigure}[ht]{0.65\textwidth}
		\includegraphics[width=\textwidth]{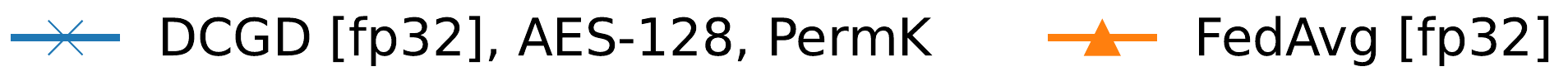} \caption{{  }}
	\end{subfigure}
	
	
	\begin{subfigure}[ht]{0.495\textwidth}
		\includegraphics[width=\textwidth]{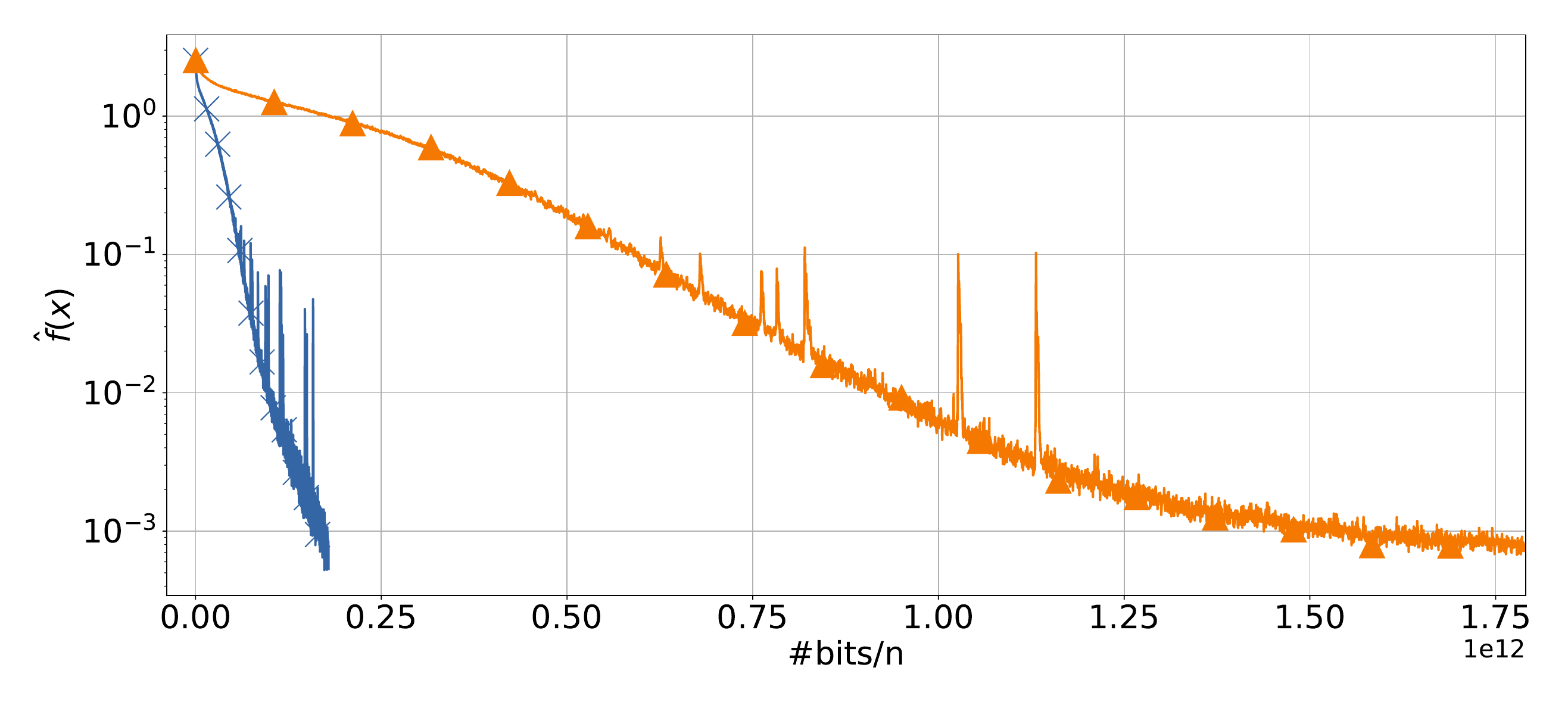} \caption{{  }}
	\end{subfigure}
	\begin{subfigure}[ht]{0.495\textwidth}
		\includegraphics[width=\textwidth]{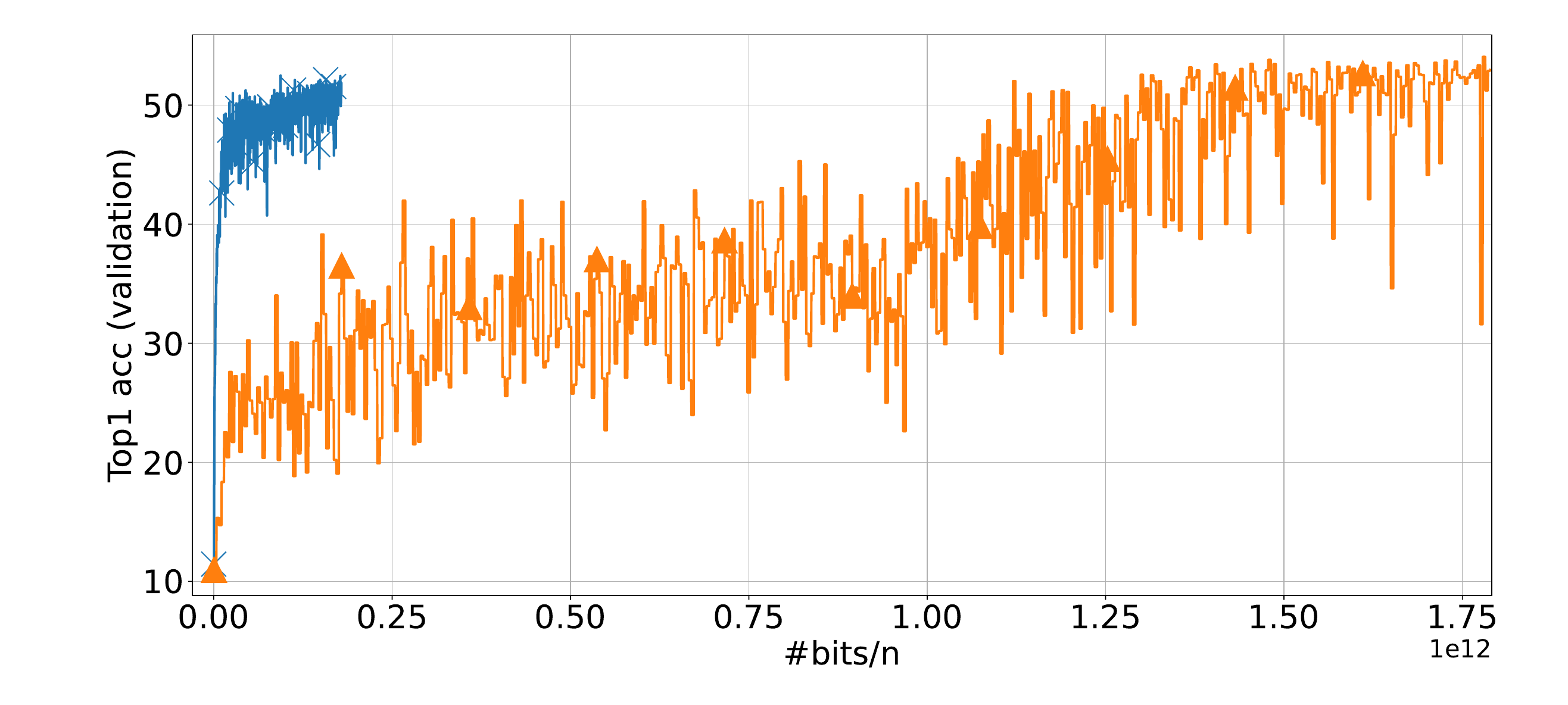} \caption{{  }}
	\end{subfigure}
	\begin{subfigure}[ht]{0.495\textwidth}
		\includegraphics[width=\textwidth]{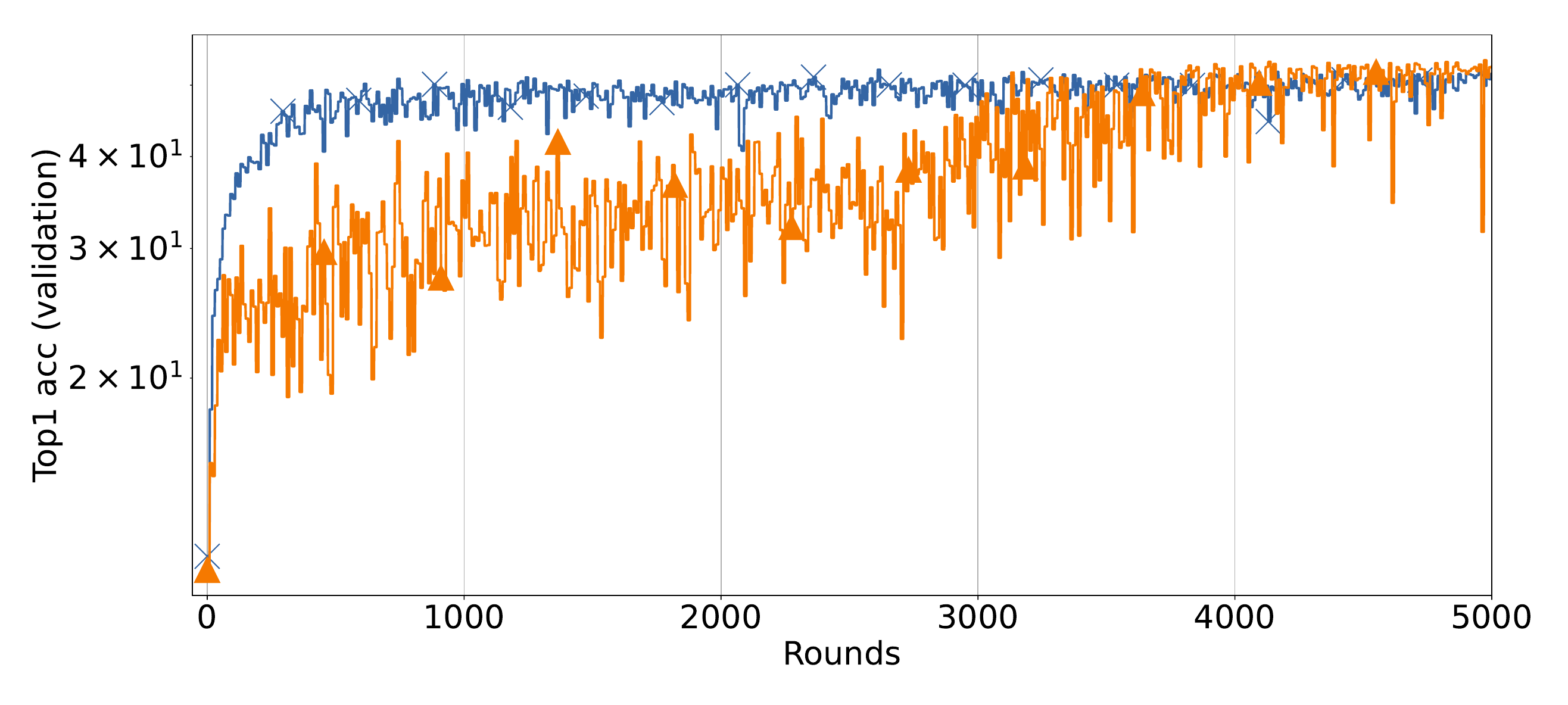} \caption{{  }}
	\end{subfigure}
	
	
	\caption{{\modelname{ResNet-18} in \dataname{CIFAR-10}, $n=10$, $d = 11\,181\,642$. \algnamewithaes{DCGD/PermK/AES} and \algname{FedAVG}.}}
	\label{ch4:fig:resnet_exp}
\end{figure*}

\subsection{Image classification application}
\label{ch4:sec:image_classification}

We evaluated the applicability of using \algnamewithaes{DCGD/PermK/AES} on Deep Neural Networks training. We used the \modelname{ResNet-18} architecture \citep{resnet} and trained it on \dataname{CIFAR-10} dataset \citep{krizhevsky2009learning}, which consists of $60\,000$ images across $10$ classes with a resolution of $32\times32$ pixels. We used \modelname{ResNet-18} implementation from TorchVision library, part of \libname{PyTorch} \citep{paszke2019pytorch}. The model size $d = 11,181,642$. For Optimization Problem~\eqref{ch4:eq:main}, we used a standard cross-entropy loss for $\mathcal{L}_{ij}$ terms in Equation~\eqref{ch4:eq:fi_for_erm}. Next, we distributed the dataset uniformly across $n = 10$ clients, with each client participating in every round. Global learning rate $0.1$, local learning rate $0.1$, local weight decay $5\cdot 10^{-4}$, number of rounds $5\,000$. During each round, clients evaluate $\nabla f_i(x)$ using a fixed 10\% subset of each client's local data points. In {FL}, there are cases where clients cannot store a large number of data points (for example, high-resolution images, recorded voice in \abr{IoT} devices) due to storage limitations. Our experiment simulates this scenario. 
We compared our implementation with \algname{FedAVG} \citep{mcmahan17fedavg}, where no compression or encryption is performed. We tracked during training communicated message size,  accuracy, and convergence. Our reported metrics in Figure~\ref{ch4:fig:resnet_exp} are computation framework and communication topology independent. In these experiments, we did not carry out an analysis of the spent time for communication, compression, and \aesname{AES} encryption/decryption, because the actual numbers are highly implementation-dependent quantities.

In \algnamewithaes{DCGD/PermK/AES}, the message size from the client to the master contains $\left\lceil \nicefrac{d}{10} \right\rceil = 1,118,165$ parameters with $4$ bytes each one. The overhead from \aesname{AES-128/EAX} Operation Mode is $32$ bytes per single message, which is negligible. Figure~\ref{ch4:fig:resnet_exp} shows that the message size was reduced from $42.65$ (for \algname{FedAVG}) {MBytes} to $4.26$ {MBytes} (for \algname{DCGD/PermK}). We cannot launch \ecryptname{CKKS} with security guarantees as \aesname{AES-128} in our environment due to memory overhead. In Appendix \ref{ch4:sec:syntetic_exp}, we demonstrated that problems start appearing in \ecryptname{CKKS} with \aesname{AES-128} security guarantees already for $d=10^6$. For a discussion about extra flexibility in Training DL models see Appendix~\ref{ch4:app:flexibility_for_dl_training}. 

\section{Deployment Flexibility}

The \textit{physical network topologies} describe the  arrangement of the computation and routing devices. In a mesh topology, every pair of nodes is connected with a dedicated link. It has high bandwidth and fault tolerance but requires a lot of cables. In this setting, \algname{DCGD/PermK} is the natural choice. The  Algorithm~\ref{ch4:alg:dcgd_permk_aes} can be observed as a series of broadcast operations to reconstitute the optimization step. See Appendix~\ref{ch4:app:comm_networks} for the benefits in other topologies.

\section{Exploring Computation-Communication Overlap}

In Appendix~\ref{ch4:app:simulation_experiment}, we present the refined scheduled communication and computation plan that utilizes the possibility of computation communication overlap and handling computation heterogeneity during training \modelname{linear regression}. The modeled situation that we studied contained $5$ clients (one of them is a straggler) which all are connected to the master with a shared channel. We showed that the execution plan can be refined, achieving a speedup of $\times 1.31$ for \algname{GD} and $\times 3.33$ for \algname{DCGD/PermK}. \algname{DCGD/PermK} offers the flexibility to start multiple compute operations while waiting for the straggler, a feature not possible in \algname{GD}. Therefore, \algname{DCGD/PermK}, with knowledge of timings, can be rescheduled more effectively.

\section{Conclusions}
\label{ch4:conclusions}

We proposed a novel secure {FL} framework that employs symmetric-key encryption \aesname{AES} \citep{daemen1999aes} with permutation compressors \compname{PermK} \citep{szlendak2021permutation} to protect and compress gradients during training. We conducted experiments on both real and synthetic data. Our work introduces a new path for applying classical cryptography to {FL} and challenges claims about its limitations.

The appendices for this work contain the following:

\begin{itemize} 
	\item {Appendices \ref{ch4:app:glossary}, \ref{ch4:app:overview_of_privacy_mechanisms}, \ref{ch4:app:discussions}, \ref{ch4:app:aes_details}, and \ref{ch4:app:ckks_details}.} Because our work is multidisciplinary, we provide a series of appendices to support readers from various backgrounds.
		
	\item {Appendix~\ref{ch4:app:extra_experiment}.} Experiments studying the effect of problem dimension, overlapping communication, and computation.
	
	\item {Appendix~\ref{ch4:app:flexibility_for_dl_training}.} Flexibility in training DL models and auxiliary research opportunities for parallelization within Backpropagation.
	
	\item {Appendix~\ref{ch4:app:comm_networks}.} Deployment flexibility across various network topologies.
	

\end{itemize}

\clearpage
\appendix

\part*{Appendices to Chapter \ref{chapter4}}
\label{ch4:app:toc_1}
\newpage

\phantomsection
\addcontentsline{toc}{chapter}{Appendices to Chapter 4}


\addtocounter{adjsection}{1}
\section{Glossary}
\label{ch4:app:glossary}

\iftrue
\begin{longtable}{|c|p{0.68\textwidth}|}
	\caption{General terminology in the context of Federated Learning.} \\
	\cline{1-2}
	\parbox{0.25\textwidth}{\centering \textbf{Term}} & \textbf{Meaning} \\
	\cline{1-2}
	\endhead
	\cline{1-2}
	\endfoot
	\cline{1-2}
	FL & Federated Learning. \\
	GD & Gradient Descent. \\
	DCGD & Distributed Compressed Gradient Descent.\\
	TEE & Trusted Execution Environments. \\
	DP & Differential Privacy. \\
	MPC & The term is overloaded. In the context of Federated Learning literature, typically means Multi-Party Computation. \\
	HE & Homomorphic Encryption. \\
	IoT & The collective network of connected devices or sensors.
\end{longtable}
\fi

\iftrue
\begin{longtable}{|c|p{0.68\textwidth}|}
	\caption{Terminology from Mathematical Optimization.} \\
	\cline{1-2}
	\parbox{0.25\textwidth}{\centering \textbf{Term}} & \textbf{Meaning} \\
	\cline{1-2}
	\endhead 
	\cline{1-2}
	\endfoot
	\cline{1-2}
	$d$ & Dimension of optimization variable. \\
	$n$ & Number of clients/agents/devices. \\
	$f_i$ & Local loss function on client number $i$. \\
	$f$ & Objective function with we want to minimize with image $\mathbb{R}$ and domain $\mathbb{R}^d$. \\
	$\gamma$ & Step size of learning rate. \\
	$x$ & Trainable or Optimization variable from $\mathbb{R}^d$. \\
	$x^k$ & Trainable or Optimization variable at most outer loop of optimization algorithm number $k$. \\
	round & The iteration in the outermost loop of the optimization algorithm. \\
	$e_i$ & The vector from $\mathbb{R}^d$ whose i-th component is equal $1$ while all the others are zeros. \\
	u.a.r. & uniformly at random. \\
	r.v. & random variable. \\
	p.d.f. & probability distribution function. \\
	$\mathbb{E}[.]$ & Expectation of some Random Variable. \\
	\compname{PermK} & Permutated correlated compressors. \\
	Master & Entity in {FL} which performs aggregation and other forms of reductions.\\
\end{longtable}
\fi

\iftrue

\begin{longtable}{|c|p{0.68\textwidth}|}
	\caption{Terminology from Discrete Mathematics.} \\
	\cline{1-2}
	\parbox{0.25\textwidth}{\centering \textbf{Term}} & \textbf{Meaning} \\
	\cline{1-2}
	\endhead 
	\cline{1-2}
	\endfoot
	\cline{1-2}
	Algebra & Set of elements and defined operations on set which lead to elements of the same set.\\
	Monoid & Any algebra with the binary, associative operation, which also has a neutral
	element. Example - concatenation.\;\\
	Group & Any algebra with the binary, associative operation, with a neutral element and each element has an inverse.\\
	Ring & Algebra with summation, and multiplication. A summation is a commutative group, multiplication is a monoid. Multiplication is distributive with respect to summation. In a Ring without extra assumptions, it can have a situation such that $a\ne0$, $b\ne0$, but $ab=0$.
	\\
	Polynomial & $p(x) = p_0 + p_1 \cdot x + p_2 \cdot x^2 + \dots$. \\
	$K[x]$ & Let K be some field. The $K[x]$ is the set of all polynomials with coefficients in some field $K$. In the Ring of polynomials $K[X]$ there is no division in general. However, it is possible to perform division with residual. One basis for $K[x]$ is $\{1, x, x^2, \dots\}$.\\
	$Z[x]$ & Integer polynomial rings over commutative ring $Z$ is denoted as $Z[x]$. This is the set of polynomials whose coefficients are integers and polynomials depend only on one variable.\\
	$P[x]/(x^2 + 1)$ & This notation means polynomials which are obtained in the following way. We take the polynomial ring $P[x]$ and perform modulus division by the polynomial $x^2 + 1$. This restricts obtained polynomials to have a power less than $2$. Such a ring is an example of a quotient ring.\\
	Unity Roots & Roots of unity are roots of the following equation $Z^n=1$. If $Z\in \mathbb{C}$ the roots are: $z_k=\exp(2 \pi k i / n )$ for $k=0,1,\dots,n-1$.\\
	GCD & The greatest common divisor (GCD) of two or more integers, that are not zero, is the largest positive integer that divides each of the integers. For example $\gcd(8, 12) = 4$. \\
	Homomorphism & Homomorphism of two groups $G_1$ and $G_2$ is a mapping $f:G_1 \to G_2$ between two groups, such that $\forall x,y \in G_1$ the following holds: $f(x*y)=f(x)*f(y)$\\	
	Isomomorphism & Isomomorphism of groups $G_1$ and $G_2$ is a mapping $f:G_1 \to G_2$ between two groups, such $\forall x,y \in G_1$ the following holds: $f(x*y)=f(x)*f(y)$ and $f$ is bijection mapping. From the point of view of Algebraic structures, two isomorphic groups are the same, even if they have different nature.\\	
\end{longtable}

\fi

\iftrue
\begin{longtable}{|c|p{0.68\textwidth}|}
	\caption{Terminology from Cryptography} \\
	\cline{1-2}
	\parbox{0.25\textwidth}{\centering \textbf{Term}} & \textbf{Meaning} \\
	\cline{1-2}
	\endhead 
	\cline{1-2}
	\endfoot
	\cline{1-2}
	FHE & Fully Homomorphic Encryption. \\
	SWHE & Somewhat Homomorphic Encryption. \\
	LFHE & Leveled Fully Homomorphic Encryption. \\
	CKKS & Cheon-Kim-Kim-Song Homomorphic Encryption schema, which is formally only SWHE. \\
	AES & Advanced Encryption Standard.\\
	sk & Secret Key.\\
	pk & Public Key.\\
	\makecell[c]{Symmetric\\Encryption} & Encryption schema in which the secret and public key are the same.\\
	PRP & Pseudo Random Permutation.\\
	PRF & Pseudo Random Function. \\
	MAC & Message Authentication Code. \\
	CRC & Cyclic Redundancy Check.\\
	CPA & Chosen plaintext attack.\\
	CCA & Chosen ciphertext attack.\\
	CTR & Counter Mode Randomized.\\
	CBC & Cipher Block Chaining.\\
	EAX & Encrypt then Authenticate Mode.\\
	LWE & The Learning With Errors (LWE) search problem to find a solution to a noisy system of linear equations. \\
	RLWE & The Ring Learning With Errors (RLWE) problem involves finding a solution to a noisy system of linear equations, where the underlying algebra is a polynomial over a ring. \\
	Integer Lattice& It is a set of points that correspond to all possible linear combinations with integer coefficients of the $n$ vectors $b_i \in \mathbb{R}^d$: $\{x:x=\sum_{i=1}^{n} {\alpha}_i b_i,{\alpha}_i \in \ZS\}$.\\
\end{longtable}
\fi 

\iftrue
\begin{longtable}{|c|p{0.68\textwidth}|}
	\caption{Terminology from Compute Architecture and Systems.} \\
	\cline{1-2}
	\parbox{0.25\textwidth}{\centering \textbf{Term}} & \textbf{Meaning} \\
	\cline{1-2}
	\endhead 
	\cline{1-2}
	\endfoot
	\cline{1-2}
	DRAM & Dynamic Random Access Memory.\\
	GPU & Graphics Processing Unit.\\
	FP & Floating-point arithmetic format.\\
	FP16, FP32, FP64 & Floating-point format with a specified number of bits per single scalar.\\
	CUDA & Compute Unified Device Architecture.\\
	CPU & Central Processing Unit.\\
	AVX & Advanced Vector Extensions.\\
	NIC & Network Interface Controller.\\
\end{longtable}
\fi
\clearpage
\addtocounter{adjsection}{1}
\section{Compared Optimization Algorithms}
\label{ch4:app:list_of_optimization_algos}

In our work, we made a comparison of several optimization algorithms with different compression and security methods. In Table~\ref{ch4:tbl:list_of_optimization_algos} below we summarize their qualitative and qualitative features.
\iftrue

\begin{longtable}{|>{\raggedright\arraybackslash}p{0.20\textwidth}|p{0.73\textwidth}|}		
	\caption{Compared optimization algorithms with compression and privacy mechanisms.} \\
	\cline{1-2}
	\textbf{Name} & \textbf{Description} \\
	\cline{1-2}
	\endhead
	\cline{1-2}
	\endfoot
	\cline{1-2}
	1. \algname{GD [FP16|FP32|FP64]}	
	& Algorithm~\ref{ch4:alg:dcgd}, Baseline (\myred{B}), $\mathcal{C}_i(x) \eqdef x$. 
	\newline
	\newline
	Gradient Descent. In this algorithm, clients in a distributed way compute the gradient for the function $f_i$ defined in Equation~\eqref{ch4:eq:fi_for_erm} in the current iterate. Master obtains gradients from the clients and averages them. The master broadcasts the result directly to all clients. The specified Floating-Point format is used to represent both data and trainable variables in all clients and during aggregation in the master. 
	\newline
	\newline
	\textbf{Quantitative and qualitative characteristics:}
	\begin{enumerate}
		\item It's completely insecure (see Appendix~\ref{ch4:app:reconstruction}).
		\item Volume information from clients to master is $\mathcal{O}(dn)$ bytes per round.
		\item Broadcasted information from master to clients is $\mathcal{O}(d)$ bytes per round.
	\end{enumerate}
	For details on why information about $\nabla f_i(x; D_i)$ can potentially reveal information about dataset $D_i$ see discussion in Appendix~\ref{ch4:app:reconstruction}.	
	\\
	\cline{1-2}
	2. \algnamewithaes{GD [FP16|FP32|FP64] /AES-128} & 
	Algorithm~\ref{ch4:alg:dcgd}, Naive Usage of \aesname{AES} (\myblue{A}), $\mathcal{C}_i(x) \eqdef x$. 
	\newline
	\newline
	Gradient Descent with symmetric \aesname{AES} encryption which uses a key size of \myblue{$128$} bits ($16$ bytes). In this algorithm, clients in a distributed way compute the gradient for the function $f_i$ defined in Equation~\eqref{ch4:eq:fi_for_erm} in the current iterate. Then clients encrypt computed $\nabla f_i(x)$ in parallel. After that, the Master collects the encoded gradients as a communication hub. Unfortunately, the master can not perform any algebraic operations, because $\nabla f_i(x)$ is encrypted, and the master needs to broadcast all directions to the clients. The specified Floating-Point format is used to represent both data and trainable variables in all clients and during aggregation in the master. 
	
	During using \aesname{AES/EAX} there exists a fixed negligible overhead from using EAX Mode Operation Mode of $32$ bytes in each client to master communication information transfer. Specifically, this Mode Operation includes the following overheads: \textit{Nonce} - a random $16$-byte value, and the \textit{Tag} - $16$-byte message authentication code.	
	\newline
	\newline
	\textbf{Quantitative and qualitative characteristics:}
	\begin{enumerate}
		\item Master cannot average obtained encrypted clients' local gradient.
		\item Volume information from clients to master is $\mathcal{O}(dn + 32n)$ bytes per round.
		\item Broadcasted information from master to clients is $\mathcal{O}(dn + 32n)$ bytes per round.
	\end{enumerate}
	For details about \aesname{AES} see Appendix~\ref{ch4:app:aes_details}.
	\\
	\cline{1-2}
	3. \algnamewithaes{DCGD [FP16|FP32|FP64] /AES-128/RandK} &
	
	Algorithm~\ref{ch4:alg:dcgd}, Naive Usage of \aesname{AES} (\myblue{A}), $$\mathcal{C}_i(x) \eqdef \dfrac{d}{k} \sum_{i \in S} x_i \cdot e_i, S \sim_{u.a.r} \{s: s\in 2^{[d]}, |s|=k \}.$$
	\newline	
	\newline
	Distributed Compressed Gradient Descent with symmetric \aesname{AES} encryption which uses a key size of \myblue{$128$} bits ($16$ bytes), and with using \compname{RandK} sparsifier.
	
	In this algorithm, clients in a distributed way compute the gradient for the function $f_i$ defined in Equation~\eqref{ch4:eq:fi_for_erm} in the current iterate. After this, the clients compress them by selecting $k$ components from $d$ u.a.r. Then clients use a sparse bitwise representation of the sparsified gradient and encode non-zero values with \aesname{AES/EAX} mode of operation. Then clients encrypt computed $\mathcal{C}(\nabla f_i(x))$ in parallel. After that, the Master collects the encoded gradients as a communication hub. Unfortunately, the master can not perform any algebraic operations, because $\nabla f_i(x)$ is encrypted, and the master needs to broadcast all directions to the clients. The specified Floating-Point format is used to represent both data and trainable variables in all clients and during aggregation in the master. 
	
	During using \aesname{AES/EAX} there exists a fixed negligible overhead from using EAX Mode Operation Mode of $32$ bytes in each client to master communication information transfer. Specifically, this Mode Operation includes the following overheads: \textit{Nonce} - a random $16$-byte value, and the \textit{Tag} - $16$-byte message authentication code.	
	The implementation of \compname{RandK} sparsifier in the case of using a pseudo-random generator can be implemented in a way that no indices should be transferred during training because they can be reconstituted. 
	\newline
	\newline
	\textbf{Quantitative and qualitative characteristics:}
	\begin{enumerate}
		\item Master cannot average obtained encrypted and compressed clients' local gradient.
		\item Volume information from clients to master is $\mathcal{O}(kn + 32n)$ bytes per round. 
		\item Broadcasted information from master to clients is $\mathcal{O}(kn + 32n)$ bytes per round.
		\item Fundamentally, there is some redundancy induced by the inability to perform aggregation in the master.
	\end{enumerate}
	\\
	\cline{1-2}
	4. \algname{GD [FP16|FP32|FP64] /CKKS} & 
	Algorithm~\ref{ch4:alg:dcgd}, Baseline (\myred{B}), $\mathcal{C}_i(x) \eqdef x$.
	Extra encryption is carried via employing \ecryptname{CKKS}.
	\newline
	\newline
	In this algorithm, clients in a distributed way compute the gradient for the function $f_i$ defined in Equation~\eqref{ch4:eq:fi_for_erm} in the current iterate. After this, each client using the public key in \ecryptname{CKKS} schema performs encryption of gradient vectors. Master obtains the encoded gradients from the clients. Then the master using a public key performs aggregation and relinearization to reduce the size of a ciphertext after arithmetic operations. After this master broadcast aggregated gradient estimator. Clients using private (secret) keys perform the decryption of the encrypted direction from the master and then make a step.
	\newline
	\newline
	\textbf{Quantitative and qualitative characteristics:} 
	\begin{enumerate}
		\item \ecryptname{CKKS} encryption considerably increases ciphertext space. The volume of information from clients to master $\mathcal{O}(C \cdot dn)$ bytes, where $C$ is a constant which depends on $d$, security level, degree of the used polynomial, and cardinality of space $\mathbb{Z}_q$ to represent integer coefficient of two transferred polynomials after encoding.
		\item Size of public and private key to guarantee \aesname{AES-128} for \ecryptname{CKKS} is approximately $420\,000$ bytes, while for \aesname{AES-128} the key size is $16$ bytes. In some tasks, this key size is negligible, in some tasks, it is not. For details see Appendix~\ref{ch4:app:ckks_details}.
		\item \ecryptname{CKKS} does not operate on the level of bits only and can operate only with FP64 floating-point format in \libname{TenSeal} implementation. Therefore, there is no possibility of using less precision in combination with \ecryptname{CKKS}. 		
	\end{enumerate}
	For details about \ecryptname{CKKS} see Appendix~\ref{ch4:app:ckks_details}.
	\\
	\cline{1-2}
	5. \algname{DCGD [FP16|FP32|FP64] /RandK/CKKS} & 
	Algorithm~\ref{ch4:alg:dcgd}, Baseline (\myred{B}), $$\mathcal{C}_i(x) \eqdef \dfrac{d}{k} \sum_{i \in S} x_i \cdot e_i, S \sim_{u.a.r} \{s: s\in 2^{[d]}, |s|=k \}.$$ Extra encryption is carried via employing \ecryptname{CKKS}.
	\newline
	\newline
	Distributed Compressed Gradient Descent with \ecryptname{CKKS}. In this algorithm, clients in a distributed way compute the gradient for the function $f_i$ defined in Equation~\eqref{ch4:eq:fi_for_erm} in the current iterate. After this, the clients compress them by selecting $k$ components from $d$ u.a.r. After this, each client using the public key in \ecryptname{CKKS} schema performs encryption of sparsified gradient vectors. Master obtains the encoded gradients from the clients. Then the master using a public key performs aggregation and relinearization to reduce the size of a ciphertext after arithmetic operations. After this master broadcast aggregated gradient estimator. Clients using private (secret) keys perform the decryption of the encrypted direction from the master and then make a step.
	\newline
	\newline
	\textbf{Quantitative and qualitative characteristics:} 
	\begin{enumerate}
		\item \ecryptname{CKKS} encryption considerably increases ciphertext space. The volume of information from clients to master $\mathcal{O}(C \cdot dn)$, where $C$ is a big constant which depends on $d$, security level, degree of the used polynomial, and cardinality of space $\mathbb{Z}_q$ to represent integer coefficient of two transferred polynomials after encoding. For details see Appendix~\ref{ch4:app:ckks_details}.
		\item Size of public and private key to guarantee \aesname{AES-128} for \ecryptname{CKKS} is approximately $420\,000$ bytes, while for \aesname{AES-128} the key size is $16$ bytes. In some tasks, this key size is negligible, in some tasks, it is not. For details see Appendix~\ref{ch4:app:ckks_details}.
		\item \ecryptname{CKKS} does not operate on the level of bits only and can operate only with FP64 floating-point format in \libname{TenSeal} implementation. Therefore, there is no possibility of using less precision in combination with \ecryptname{CKKS}.
		\item The \ecryptname{CKKS} does not support linear operations over sparse vectors from $\RD$. For \ecryptname{HE} schemas, encoding of any two vectors (for example, sparse and dense) should be indistinguishable due to semantic security requirements. However, from a computational perspective, ignoring sparsity is sometimes highly impractical. This gap presents an open research question for \ecryptname{HE}.
	\end{enumerate}
	\\
	\cline{1-2}
	6. \algnamewithaes{DCGD [FP16|FP32|FP64] /PermK/AES} & 
	Algorithm~\ref{ch4:alg:dcgd_permk_aes}.
	\newline
	\newline
	Distributed Compressed Gradient Descent with using \compname{PermK} sparsifier. In this algorithm, clients in a distributed way compute gradients in the current iterate. After this, the clients compress them by selecting $k$ components from $d$ jointly	via using Algorithm~\ref{ch4:alg:perm_k_gen}. 
	
	Then clients send a sparse bitwise representation of the sparsified gradient. Master collects the encoded gradients as a communication hub. 
	
	The \compname{PermK} compressor guarantees that what is needed to perform at master is only concatenation. Therefore, the algorithm can be implemented in practice in situations when the master can store, but cannot compute. 
	
	The specified floating-point format is used to represent both data and trainable variables. The implementation of \compname{PermK} in the case of using a pseudo-random generator can be implemented in a way, that no indices should be transferred during training because they can be reconstituted, and negotiation between clients should be carried in runtime if clients have negotiated an initial seed. 
	\newline
	\newline
	\textbf{Quantitative and qualitative characteristics:}
	\begin{enumerate}
		\item There is $32$ byte overhead during message transfers compared to \algname{DCGD/RandK}.
		\item The current implementation works only in case $d \ge n$, when \algname{DCGD/RandK} and \algname{GD/CKKS} do not require this
		\item Volume information from clients to master is $\mathcal{O}(d + 32n)$ bytes per round. 
		\item Broadcasted information from master to clients is $\mathcal{O}(d + 32n)$ bytes per round.
	\end{enumerate}
	\\
	\cline{1-2}
	7. \algnamewithaes{DCGD [FP16|FP32|FP64] /PermK} & 
	Algorithm~\ref{ch4:alg:dcgd_permk_aes} in which $\myblue{Encrypt}(x) \eqdef x$, $\myblue{Decrypt}(x) \eqdef x$.
	\newline
	\newline
	Distributed Compressed Gradient Descent with using \compname{PermK} sparsifier. In this algorithm, clients in a distributed way compute gradients in the current iterate. After this, the clients compress them by selecting $k$ components from $d$ jointly	via using Algorithm~\ref{ch4:alg:perm_k_gen}. Then clients send a sparse bitwise representation of the sparsified gradient. Master collects the encoded gradients as a communication hub. The \compname{PermK} compressor guarantees that what is needed to perform at master is only concatenation. Therefore, the algorithm can be implemented in practice in situations when the master can store, but cannot compute. The selected floating point format is used to represent both data and trainable variables. The implementation of \compname{PermK} in case of using a pseudo-random generator can be implemented in a way, that no indices should be transferred during training because they can be reconstituted and negotiation between clients should be carried in runtime if clients have negotiatedan initial seed. 
	\newline	
	\newline
	\textbf{Quantitative and qualitative characteristics:}
	\begin{enumerate}
		\item The current implementation works only in case $d \ge n$ only, when \algname{DCGD/RandK} and \algname{GD/CKKS} does not require this.
		\item This algorithm ignores privacy and security aspects.
		\item Volume information from clients to master is $\mathcal{O}(d + 32n)$ bytes per round. 
		\item Broadcasted information from master to clients is $\mathcal{O}(d + 32n)$ bytes per round.
	\end{enumerate}
	\label{ch4:tbl:list_of_optimization_algos}
\end{longtable}

\fi

\clearpage
\addtocounter{adjsection}{1}
\section{Computing and Software Environment}
\label{ch4:app:environment}

We have conducted numerical experiments using the Python software suite \fl~\citep{burlachenko2021fl_pytorch}, running on Python 3.9. The target machine is a server-grade system operating on Ubuntu 18.04 and Linux Kernel v5.4.0-148. It is equipped with a 48-core Intel(R) Xeon(R) Gold 6246 CPU (2 sockets with 24 cores per socket) running at $3.3$ GHz. The machine is equipped with 256 GBytes of DDR4 DRAM system memory operating at $2.9$ GHz. The installed CPU does not support both the AVX512FP16 Instruction Set Architecture
\footnote{\href{https://www.intel.com/content/www/us/en/content-details/669773/intel-avx-512-fp16-instruction-set-for-intel-xeon-processor-based-products-technology-guide.html}{https://www.intel.com/content/www/us/en/content-details/669773/intel-avx-512-fp16-instruction-set-for-intel-xeon-processor-based-products-technology-guide.html} - {Intel AVX-512 FP16 Instruction Set}} and it does not support {FP16} arithmetic. The machine also has an NVIDIA GeForce RTX 3090 GPU built on Ampere microarchitecture with 24 GBytes of DRAM $9.7$ GHz GPU memory. This GPU supports the CUDA Compute Capability 8.6. and as a consequence, supports the {FP16} arithmetic, which is supported for devices with Compute Capability 5.3. and higher. \footnote{\href{https://docs.nvidia.com/cuda/cuda-c-programming-guide/index.html\#features-and-technical-specifications}{https://docs.nvidia.com/cuda/cuda-c-programming-guide/index.html\#features-and-technical-specifications} - {NVIDIA Specification for Compute Capabilities}}. Numerical experiments with FP16 arithmetic were carried out in this GPU.

We have patched \fl to include necessary functionality from \libname{TenSeal} \citep{benaissa2021tenseal} library version 0.3.14 and PyCryptodome \citep{pycryptodome} version 3.17. The experiment's scheduler for computation and communication from Appendix \ref{ch4:app:simulation_experiment} generates text descriptions in dot graphviz format. Finally, we used Graphviz 8.0.5 \citep{ellson2002graphviz} to generate a directed graph from this description.

\clearpage
\addtocounter{adjsection}{1}
\section{Overview of Existing Privacy Mechanisms in FL}
\label{ch4:app:overview_of_privacy_mechanisms}

\paragraph{Trusted Execution Environments (\abr{TEE}).} The \abr{TEE} brings the idea of handling self-isolation of processes more seriously than it has been done in Operating Systems before. An example of the particular implementation of this idea is ARM TrustZone \citep{pinto2019demystifying} and Intel SGX \citep{costan2016intel}, where both technologies have been implemented at the hardware level. Such technologies allow the protection of memory from reading and writing operations not only from another process in the Operating System (OS) but also from the kernel of the OS by itself. The goal of \abr{TEE}: 
\begin{center}
	\textit{TEE protects the execution environment from illegal intervention, which will defect training}.
\end{center}

\paragraph{Differential Privacy ({DP}).} The goal of \abr{DP} is to ensure statistical analysis does not hurt the privacy aspect. The intuitive definition can be as follows: "An algorithm A is differentially private if an observer seeing its output cannot tell if a particular individual's data was used in the computation." \abr{DP} criteria assess the private property of the algorithm, but it does not dictate how this functionality should be implemented. In general, as more noise is appended to the data or derived quantities from the data that are released publicly, the algorithm starts to be more \abr{DP}, but on the other hand, the statistical properties of the data to solve the original task start to be lost.

\begin{definition} [Approximate Differential Privacy] \label{ch4:def:approx_dp}
	An algorithm $M:\mathcal{X}^n \to \mathcal{Y}$ satisfies approximate $(\varepsilon, \delta)$~-~DP if $\forall X, X' \in \mathcal{X}^n$ such that datasets are different in one point (neighboring datasets denoted as $X \sim X'$) and $\forall T \subseteq \mathcal{Y}$ the following holds:
	\[
	\Prob(M(X) \in T) \le \exp(\varepsilon) \Prob(M(X') \in T) + \delta.
	\]
\end{definition}

The Definition~\ref{ch4:def:approx_dp} is a relaxation of pure DP, first proposed by \citet{dwork2006our}. The pure DP definition, introduced by \citet{dwork2006calibrating}, requires $\delta = 0$. The Approximate DP possesses weaker privacy guarantees but allows the addition of less noise.

DP-ERM aims to output $\hat{\theta}$, which is $(\varepsilon, \delta)$ - DP with respect to the training dataset $D$. To do that, there are three types of approaches: (i) Output perturbation; (ii) Objective perturbation; (iii) Gradient perturbation. Algorithms that solves that DP-ERM problem are quantified by \textit{expected excess empirical risk} $\E[\mathcal{L}_{erm}(\hat{\theta}, D) - \mathcal{L}_{erm}(\theta_{erm}^*, D)]$, and for DP-RM problem by \textit{expected population risk} $\E[\mathcal{L}_{rm}(\hat{\theta}, D) - \mathcal{L}_{rm}(\theta_{rm}^*, D)]$. These quantities are sometimes named as \textit{utility}. \\
The goal of \abr{DP}:
\begin{center}
	\textit{DP protects output of algorithms so that users' data are not leaking from Algorithm execution}.
\end{center}

One important classification in \abr{DP} algorithms targeted to distributed environments is their separation into two classes: \textit{Centralized} and \textit{Local} Differential Private settings. In the centralized model, a client trusts the curator (or master), and DP protection mechanisms are applied in the master. In the local model, each individual applies a differentially private mechanism to their own data before sending it to an untrusted curator (or master). Centralized models have lower privacy loss since the noise is added only once at the end of the process, such as aggregation at the master. However, the centralized model also requires more trust in the aggregator.

\paragraph{Aggregation with Multi-Party Computation ({MPC}).} The MPC is a sub-field of Cryptography concerned with the problem of having a set of parties that compute an agreed function of their private inputs. The goal of secure multi-party computation (\abr{MPC}) is to enable independent data owners who do not trust each other or any common third party to jointly compute a function that depends on all of their private inputs. \abr{MPC} protocols are typically implemented with (a) \textit{Secret sharing} - in this case, it requires a lot of total communication rounds to compute the average across $n$ clients; (b) \textit{Garbled circuits} - in this case there both communication and computation overhead is added to the training \citep{zhao2019secure}. The goal of \abr{MPC}: 

\begin{center} 
	\textit{MPC allows for protecting inputs for the algorithm at the cost of communication.}
\end{center}

\paragraph{Homomorphic Encryption (\ecryptname{HE}).}

Homomorphic Encryption (\ecryptname{HE}) enables numerical computation on encrypted data, for example aggregating vectors from $\mathbb{R}^d$. However, the result can only be decrypted with the private key. The concept dates back to 1978, with the work of \citet{rivest1978data}, who proposed the idea but did not provide a complete solution. \ecryptname{HE} allows any device party to compute functions on encrypted data using only a public key and encrypt the result. \ecryptname{HE} aims to hide the plaintext input and output from the executor, who only sees the encrypted versions of it. This is possible with  \ecryptname{HE}. On the other hand, \textit{obfuscation} is the process of encrypting the program (function $f$), not the input or output from the caller. The \textit{obfuscation} is impossible under weak technical conditions \citep{barak2001possibility}.

Fully Homomorphic Encryption (\ecryptname{FHE}) allows any computable algorithm to be executed on encrypted data without any restrictions on the binary operations. The first \ecryptname{FHE} scheme was proposed by \citet{gentry2009fully}, based on lattices and a novel bootstrapping technique. Let $c_i$ be the encryption of message $m_i$, and $c$ be the result of evaluating a function $\hat{f}(c_1,\dots c_n)$, which should decrypt to $m=f(m_1,\dots m_n)$. In his Ph.D. thesis, Craig Gentry listed some requirements that any \ecryptname{FHE} scheme should satisfy, which we summarize below:

\begin{enumerate}
	\item \textit{Correctness.} Decryption should always recover the correct evaluation of the function, in other words, $\Prob( {Decrypt(c) = f(m_1, m_2,\dots)}) = 1.$
	
	\item \textit{Semantic security.} The encryption of any two messages should be computationally indistinguishable.
	
	\item \textit{Efficiency.} Decryption should not be more expensive than evaluating the function itself.
	
	\item \textit{Compactness.} Ciphertexts should have a polynomial size in the security parameter, independent of the size of the function evaluated.
	
	\item \textit{Security.} The best-known attack should have exponential complexity in the security parameter, in other words, $\Omega({2^k})$ ($k$ is a security parameter).
	
	\item \textit{Feasibility.} Key generation, encryption, and decryption should have polynomial complexity in the security parameter.

\end{enumerate}

By C.Gentry, only algorithms that allow executing any computable algorithm on encrypted data that satisfies properties (1)--(6) can be called \ecryptname{FHE}. The \ecryptname{FHE} with this requirement can firstly be hard to construct, and secondly, in practice, they may not be computationally efficient. To mitigate these issues, two main strategies are the following:

\begin{itemize}
	\item \textit{Somewhat Homomorphic Encryption} (\ecryptname{SWHE}). One way to make \ecryptname{FHE} more efficient in practice is to restrict the class of functions that can be evaluated. This leads to Somewhat Homomorphic Encryption, which can handle functions from restricted classes (for example, low-degree polynomials). 
	
	\item \textit{Leveled FHE} (\ecryptname{LFHE}). Another way is to limit the depth of the binary circuit that represents the function. This leads to a notion of Leveled FHE, which can handle arbitrary functions represented by boolean circuits but with a fixed bound on the circuit depth.
\end{itemize}

Challenges of applying \ecryptname{HE} for training machine learning models and scientific computation:

\begin{enumerate}
	\item The \ecryptname{HE} adds noise to the plaintext to ensure security. The challenge is to manage it with error-correction techniques, as it typically grows after each arithmetic operation.
	\item The \ecryptname{HE} cannot perform random access on encrypted data without revealing information.
	\item The \ecryptname{HE} cannot exploit the advantages of Random Access, which can compute some algorithms faster than binary circuits. Example - Binary Search.
	\item The \ecryptname{HE} does not support multiple keys natively. Originally, it was designed as a single-private-key system.
	\item The \ecryptname{HE} does not obfuscate the function itself, only the input and output. Obfuscation is impossible under weak conditions \citep{barak2001possibility}.
	\item The \ecryptname{HE} works on binary circuits or functional schemas. This class of representation is Turing Complete, assuming we can create circuits of different levels for different inputs. The pure \ecryptname{FHE} operations are computationally intensive and currently for practical purposes the \ecryptname{SWHE} and \ecryptname{LFHE} schemas should be considered instead.	
	\item The \ecryptname{HE} methods require large ciphertext sizes.
	\item Choosing the right \ecryptname{HE} scheme for a given machine learning task is not trivial.
\end{enumerate}

The goal of \abr{HE}: 
\begin{center}
	\textit{HE allows meaningful manipulation under encrypted data without revealing it.}
\end{center}

\clearpage
\addtocounter{adjsection}{1}
\section{Discussions}
\label{ch4:app:discussions}

\subsection{The Imperative of safeguarding against eavesdropping}
\label{ch4:app:reconstruction}

Assume that during training with first-order optimization method, each client $i$ discloses the following information at each iteration $k \in \{1, \dots, K\}$: $$\dfrac{\partial f_i(x^k)}{\partial x_j} \eqdef \lim_{dx_j \to 0} \dfrac{f_i(x^k + e_k \cdot {dx}_j; {\color{red}D_i}) - f(x^k; {\color{red}D_i})}{{dx}_j}.$$ If $f_i$ is twice differentiable and has bounded Hessians near point $x^k$, the knowledge of partial derivative provide the following approximate equality: $$\dfrac{\partial f_i (x^k)}{\partial x_j} \cdot dx_j \approx {f_i(x^k + e_j \cdot {dx}_j; {\color{red}D_i}) - f(x^j; {\color{red}D_i})}.$$ In the last approximate equality, $e_j$ is a unit norm vector of the standard basis of $\mathbb{R}^d$. Let's assume that:
\begin{enumerate}
	\item The $f_i(x)$ is linear with respect to ${\color{red}D_i}$ in original form, or there exists a bijective change of variable ${\color{red}D_i \to {D_i}'}$ such that $f_i(x)$ is linear in ${\color{red}D_i'}$.	
	\item Iterates $x^k$ are uniformly distributed in $\mathbb{R}^d$. 
\end{enumerate}

Under these assumptions, a training process might reveal sensitive information. Indeed, if an adversary obtains this information, then information about partial derivative provides noisy response linear function in ${\color{red}D_i}$. This general setting has been studied by \citep{dinur2003revealing}. Authors demonstrated that there exist linear attacks that could, with high probability, expose ${\color{red}D_i}$.

Specifically, let's denote the length of ${\color{red}D_i}$ is equal to $N$ bits. Let's assume adversarial obtains $K$ noisy answers (for a single partial derivative) with an additive error of $E=o(\sqrt{N})$ to each answer. Having $K \ge \theta(N^2/E^2)$, such adversarial queries can reveal information about ${\color{red}D_i}$ with high probability. The hidden constant is around $256$ but can be improved. A single full gradient $\nabla f_i(x; {\color{red}D_i})$ provides $d$ equations instead of $1$ for each response. This demonstrates that in specific modes the pretty big amount of partial derivative already reveals enough information to reconstruct the client's dataset. Such attacks are not only in theoretical interest but can be practically mounted \citep{cohen2018linear}, \citep{kasiviswanathan2013power}.

\subsection{Privacy and security}
\label{ch4:app:privacy_vs_security}

As we have described in Appendix~\ref{ch4:app:overview_of_privacy_mechanisms}, different mechanisms can be used for {FL} to provide privacy for different aspects training process. Each mechanism has a specific target of protection (such as input, output, execution, or communication channel) and a specific adversary model. The precise meaning of protection is defined by rigorous formalization and the details are crucial. For example, \abr{DP} protects the output of an algorithm so that it does not reveal sensitive information about the input data, while \abr{MPC} protects the input data from being exposed to other parties during computation. Recent research papers on {FL} deployments have mostly used Local DP \citep{bhowmick2018protection}, or a combination of \abr{MPC} and centralized \abr{DP}.

In contrast, the proposed design in our work can be understood as a mechanism for providing \textit{security} for the training process rather than \textit{privacy}. The response obtained from the server by the client is computed securely and accurately with protection against attacks on the server or the communication channels to the server. The concept of \textit{privacy} ensures that the entity being protected can engage in an algorithm without being observed by unauthorized parties. In contrast, \textit{security} is a more comprehensive concept that defines the algorithm or protocol, how personal information and derived data are safeguarded, the types and strengths of attacks it can withstand, and the requirement of a secret key to access the data. In some cases, a complex security protocol may not be practical or suitable for certain situations, and a privacy mechanism that is less exhaustive but more flexible may be a more appropriate choice.

\clearpage
\addtocounter{adjsection}{1}
\section{Usage of {AES} Cipher During Distributed Training}
\label{ch4:app:aes_details}

This section overviews the current state-of-the-art block cipher \aesname{AES} and explains how it can be used for symmetric key encryption in scenarios where multiple client messages must be encrypted and decrypted securely.

\subsection{{AES} block cipher} 
A block cipher is a fundamental cryptographic primitive that transforms a fixed-length block of bits into another block of the same length using a secret key. The \aesname{AES} is today's most widely used secure block cipher. Some CPUs have hardware support for it. Examples of CPUs with x86 instruction set architecture that support \aesname{AES} in the hardware level are Intel Westmere, AMD Bulldozer. Examples of CPU with AArch64 instruction set that support \aesname{AES} in the hardware level is ARM Cortex-A53.

\aesname{AES} block cipher supports three key sizes: $128$, $192$, or $256$. The key size determines the level of security and the computational cost of the encryption and decryption operations. The key space $K$ for \aesname{AES-128} has size of $|K|=2^{128}$. The input message space $M$ and the output cipher text space $C$ of the block cipher \aesname{AES} for all three types of keys have the same size equal to $128$ bits. Therefore cardinality of $M$ and $C$ is equal to $|M|=|C|=2^{128}$.

For each key $k\in K$, the \aesname{AES} block cipher maps $M \to C$ with a bijective function, with $M = C$, and $|M|=|C| \le \infty$. Essentially, the key $k$ is a selector of bijective mapping or permutation. Each key realizes the permutation of the input message. If we assume that each of two distinct keys $k_1, k_2 \in K$ implements different bijective mappings $M \to C$, then the number of permutations that \aesname{AES} can realize is $|K|=2^{128}$. Even though it is a big number, this is much smaller than $2^{128}!$, i.e. the size of all possible permutations if input and output are $128$ bits in length. In the Cryptography community, the block cipher \aesname{AES} is sometimes observed as a primitive that implements a Pseudo Random Permutation (\abr{PRP}). This means that for a fixed key, it defines a permutation. The discrepancy between the number of all possible permutations and the number of \abr{PRPs}  that \aesname{AES} can realize is elegantly resolved in the Cryptography community. It is resolved by introducing the notion of a \abr{Secure PRP}. The algorithm which implements \abr{PRP} implements a \abr{Secure PRP}  if an adversary from observing the realization of permutation $f$ cannot distinguish by using an arbitrarily tractable algorithm between two events:
\begin{enumerate}
	\item The permutation function $f:M \to M$ is chosen uniformly at random from all possible $M!$ permutations.
	\item The permutation function $f:M \to M$ is chosen as one of the permutations that block cipher $\mathrm{BlockCipher}(\cdot,k):M \to M$ with $k \sim_{u.a.r} K$ can realize.
\end{enumerate}

By the current status in Cryptography, the \aesname{AES} is believed to be a secure \abr{PRP}. Having two input and output pairs, adversarial may wish to derive the secret key. The brute force search on \aesname{AES-128} is computationally infeasible because $2^{128}$ is too large to enumerate. The best-known attack on the full version of \aesname{AES-128} that can recover the complete key has a complexity of $2^{126}$ \citep{bogdanov2011biclique}. The \aesname{AES-128} is secure against brute force search and also against linear and quantum attacks \citep{jang2022quantum}.

\subsection{Internals of {AES} block cipher} The \aesname{AES} block cipher has $10$ rounds for \aesname{AES-128} and $14$ rounds for \aesname{AES-256}. The key is expanded into $11$ or $15$ subkeys of $128$ bits in length each. Subkeys are used in each round. Each round consists of four invertible steps: key addition (in the sense of "exclusive boolean or", which we will denote as XOR), byte substitution, row shift, and column mix. These steps transform a $4 \times 4$ matrix of bytes representing each round's input. Byte substitution adds non-linearity, row shift rotates each row cyclically, and column mix applies a linear transformation to each column. The last subkey is used for a final key addition (XOR) to mask the output. The \aesname{AES} decryption reverses the encryption steps. The \aesname{AES} has different implementations for different devices and code size requirements. All occurred transformations are stateless; consequently, the \aesname{AES} block cipher is stateless.

\subsection{Apply {AES} block cipher for more than one input block}
\label{ch4:app:internals-of-aes}

The \aesname{AES} is a secure block cipher, but using the same key $k$ to encrypt multiple blocks deterministically is not recommended. This undermines the notion of Semantic Security, which ensures that an adversary cannot deduce any useful information by analyzing sequences of ciphertexts. For instance, if an adversary observes $c_1 = c_2$, they could infer that $m_1 = m_2$, even though they do not know the exact values of $m_1$ and $m_2$. This is because encryption is fully deterministic, based on the input message. To preserve Semantic Security, encryption must be resistant to chosen-plaintext attacks, which refers to the adversary's ability to gain information by choosing plaintexts and analyzing the resulting ciphertexts.

Formally, in chosen-plaintext-attack (\attackname{CPA}) an attacker may adaptively ask for the encryption $(e_1, e_2, \dots)$ of arbitrary messages $(m_1,m_2,\dots)$ of his choice. The attacker's goal is to obtain the ability to correctly guess from two obtained encryption $\{e_a,e_b\}$ from experiment $A$ and experiment $B$ which encryption belongs to which message from the set $\{m_a, m_b\}$. The attacker does not know encryption of $\{m_a, m_b\}$ in advance, and the attacker does not know which plain message has been used in which experiment. In the context of this attack, the advantage of adversarial is defined as: $${ADV}_{cpa} = |\Prob(\{ \mathrm{ExperimentA\, uses\,} m_a \}) - \Prob(\{ \mathrm{ExperimentB\, uses\,} m_b\})|.$$ Fundamentally, there are two ways to provide security against \attackname{CPA} attacks:

\begin{enumerate}
	\item Increase ciphertext space and allow encryption to work randomly. This is the underlying reason why \ecryptname{CKKS} is \attackname{CPA} secure. The downside of this is that ciphertext space is increasing substantially.
	\item Augment key space $K$ with extra counter, named as a \textit{nonce} from a nonce space $N$. The clients who perform encryption guarantee that the pair $(key, nonce)$ is unique during the life of the $key$. In modern cryptography, the $nonce$ is considered public and accessible to everyone (conceptually).
\end{enumerate}

Next, there are two different ways to select \textit{nonce} in nonce-based protection against \attackname{CPA} attacks:
\begin{enumerate}
	\item \textit{Deterministic counter.} In this mode, a single deterministic integer counter is used. Sometimes there is no need to send nonce itself with ciphertext during communication if the receiver of ciphertext can recover counter from other available information.
	\item \textit{Randomized counter.} If encryption happens on several devices, the coordination of \textit{nonce} complicates the process. One way around this is to select $nonce \sim_{u.a.r.} N$, where $N$ is sufficiently large. For example, a nonce space size of $|N| > 2^{128}$ ensures that we expect to encounter at least one collision after sampling $ \sqrt{2|N|} + 1 = 2^{64}$ values, as predicted by the famous Birthday Paradox \citep{cormen2022introduction}.
\end{enumerate}

The two popular ways to use nonce-based encryption for \aesname{AES}, also known as \textit{operation modes}, are as follows:

\begin{enumerate}
	\item \textbf{Nonce-based Cipher Block Chaining Mode} (\ecryptname{CBC}): In this scheme, the input message is divided into blocks. Each block is encrypted using the \aesname{AES} block cipher, but the plaintext is first masked before encryption. The first block is masked (via XOR) with the \textit{nonce}, while subsequent blocks are masked with the ciphertext of the previous block. The output consists of the public nonce and the sequence of encrypted blocks.
	
	It can be proven that \aesname{CBC} achieves \aesname{CPA} security with an advantage of at most ${ADV}_{cpa} \le 1/2^{32}$ when used for up to $2^{48}$ \aesname{AES} ($128$-bit) blocks. If the \textit{nonce} is generated using a non-secure PRG, it should first be encrypted with a secure \aesname{AES} block to ensure its security.
	
	\item \textbf{Randomized Counter Mode} (\ecryptname{CTR}): The input message is divided into blocks, but in this mode, each block $i$ is masked using a value derived from the \aesname{AES} block cipher: $\aesname{AES}(k, \textit{nonce} + i)$. The plaintext is then XORed with this value.
	
	This mode is \attackname{CPA}-secure with an advantage of at most $1/2^{32}$ as long as the total number of encrypted blocks does not exceed $2^{64}$. Compared to \ecryptname{CBC}, \ecryptname{CTR} is more flexible, as it does not require padding and allows for truncation of unnecessary bits in the final block.
\end{enumerate}

The \ecryptname{CKKS} and \aesname{AES} in \aesname{EAX} mode of operation \citep{bellare2004eax} are secure against \attackname{CPA} attacks. or further details see \citet{boneh2020graduate}.

\subsection{Apply {AES} with integrity guarantees}

The \attackname{CPA} security and operations mode solve security against eavesdropping, but they don't provide integrity of the delivered messages. Message Authentication Code (\ecryptname{MAC}) is Cryptography protected analogous to (non-secure) Cyclic Redundancy Check (\abr{CRC}), which is used in Networking communication to protect from random (not adversarial) noise in communication channels. The goal of \abr{MAC} is that the value of \abr{MAC} coupled with a ciphertext guarantees that the message has not been modified during the transfer by the adversary. The \textit{signing algorithm} for generating \abr{MAC} tag takes a plain message $m$ and secret key $k$ as input. The signature is added to the transferred message with the public nonce. 

Adversarial without a secret key $k$ cannot generate valid $(Message, Mac)$ pairs in such a way that it will pass the \textit{verification algorithm}. The \textit{verification algorithm} takes as input the secret key $k$, decrypted message $m$, and \abr{MAC} value $tag$. In practice, nonce and \abr{MAC} are typically $16$ bytes long. If the optimization algorithm already requires sending far more than $2$ elements of $\mathbb{R}$ encoded in {IEEE-754 FP64} format then this overhead is negligible during the training.

Two popular constructions to provide integrity in the context of \aesname{AES} are the following schemes:

\begin{enumerate}
	\item \textbf{Encrypted CBC-MAC}. The input message is split into $16$ bytes buckets. After this, the first bucket is plugged into \aesname{AES} as input. Output from this block is masked with XOR with the next input $16$ byte bucket. After the XOR operation is finished, the output of XOR is plugged into the next \aesname{AES} block as input, and the process repeats in a chaining fashion. It is important to notice that all \aesname{AES} blocks use the same key $k$. Finally, the last output is plugged into one more \aesname{AES} block as input, which uses another key $k'$ and schema releases \abr{MAC} tag.
	
	\item \textbf{Nested MAC}. In this approach, the message is split into $16$ bytes buckets. After this, the first \aesname{AES} block obtains as input this first bucket and $k$ is used as a secret key. The output of this \aesname{AES} block is used as a key for the next block, and data for this \aesname{AES} block is the next input bucket. After processing all messages, the last output is plugged into the last \aesname{AES} block as input, but for the last \aesname{AES} block, another secret key $k'$ should be used similarly as in \textbf{CBC-MAC}.
\end{enumerate}

For using \textbf{CBC-MAC} and \textbf{Nested MAC} securely, the number of signed messages (of arbitrary length) should be on the order of $2^{64}$ \aesname{AES} $128$-bits blocks. For details, please check the analysis of these schemes.

\subsection{Chosen Ciphertext Attack {(CCA)} and authentication}

In a chosen ciphertext attack (\attackname{CCA}), the attacker has access to a decryption oracle and can decrypt any ciphertext of his choice. Also, the attacker can access encryption oracles and obtain encryption of arbitrary messages of his choice. Adversarial aims to break semantic security in the \attackname{CPA} sense. To provide the \attackname{CCA} security, the two processes \algname{Encrypt} and {\algname{MAC}} should be performed sequentially, and typically it is preferable to perform them in this order. 

In our work, we have used \aesname{EAX}, which is \algname{CTR} mode for encryption combined with \algname{CMAC} (standardized version of \textbf{CBC-MAC}) for integrity purposes.

\subsection{Using {AES} with {EAX} operation mode in our work}
We use \aesname{AES} encryption with \ecryptname{EAX} mode, which produces a triple as output: 
\[\langle m=C_i(\nabla f(x^t) \in \mathbb{R}^d (16/32/64 \cdot d\,\mathrm{bits}), nonce \in N (128\, \mathrm{bits}), Tag (128\,\mathrm{bits}, ciphertext)\rangle\].

The ciphertext has the same bit length as the original message. Because \ecryptname{EAX} mode employs \ecryptname{CTR} for stream encryption, it does not require any padding. The $nonce$ is a random 16-byte value, and the $Tag$ is a $16$ byte message authentication code. As it has been mentioned in Section \ref{ch4:sed:resilience}, the \ecryptname{CKKS} and all \abr{HE} schemas cannot achieve \attackname{CCA} security \citep{fauzi2022ind}. For \aesname{AES}/\aesname{EAX}, it has been proved in \citep{bellare2004eax} that it is secure against \attackname{CCA} attacks. Cryptography is an area of science by itself. Readers can gain more information about Classical Cryptography from \citep{boneh2020graduate}.

\clearpage
\addtocounter{adjsection}{1}
\section{Homomorphic Encryption with {CKKS}}
\label{ch4:app:ckks_details}

\subsection{Learning with errors problem and REGEV09 algorithm}
\label{ch4:app:lwe}

The security of the \ecryptname{CKKS} scheme depends mainly on the hardness of the Learning With Errors (\abr{LWE}) problem which we will overview next. The \textit{"Learning with errors"} search problem represents was analyzed by \citet{regev2009lattices}. For this work Oded Regev has obtained a G{\"o}del prize in 2018:
\begin{equation*}
	\begin{aligned}
		\mathrm{find}\,s\in \ZS_q^n \\
		\mathrm{such\,that:}\, s^T \bA + e^T &= b \,\mathrm{,where}\,e\,\mathrm{is\,random\,variable.} \\
	\end{aligned}
\end{equation*}

Parameters have the following properties: 

\begin{itemize}
	\item The $q$ is a prime number.
	\item $\bA\in \ZS_q^{n \times m}$ is a random matrix drawn u.a.r. from $\ZS_q^{n \times m}$.	
	\item $m > n$.
	\item $e\in \ZS_q^m$ is small (in terms of $L_2$ norm) random variable. For $e$ there is no assumption about its distribution. The only one constraint is that $e \ne 0$.
	\item $b \in \ZS_q^{m}$.
\end{itemize}

As can be observed from the description LWE problem operates on a system of over-determined sets of equations. When $e=0$ the system can be solved via \textit{Gaussian elimination} in variable $s$, but when $e \ne 0$ we believe it is a hard search problem. Specifically, Theorem 1 from \citet{regev2009lattices} makes a connection between LWE and the search problem in integer lattices via the following theorem:

\begin{center}
	\textit{Instance of LWE problems with parameter $n$ is as hard as the Short Integer Solution Problem}.
\end{center}

The best well-known solution for solving the LWE problem works in the following time: ${q}^{\mathcal{O}\left({n}/{\log(n)}\right)}$ \citep{blum2003noise}.

In usual LWE we can not find a solution for a noisy set of linear equations effectively based on knowledge of $A$ and $b$.

There exists a variation of the LWE problem named as Decision Learning With Errors problem. In theory, it has been shown that Decisional LWE is as hard as LWE. In \textit{Decisional LWE} the adversaries can not from observing tuple $(\bA,b)$ distinguish the following two scenarios:
\begin{enumerate}
	\item The tuple $(\bA,b)$ has been generated completely uniformly at random. In some sense, there are no real hidden linear dependence structures between columns of $A$ and $b$.
	\item The tuple $(\bA,b)$ in fact has a specific structure $(\bA=\bA,b=\bA^T s + e)$.
\end{enumerate}

The LWE problem provides a way to publish a lot of perturbated linear equations in variable $s$ in the form of $s^T \bA + e^T = b$, and essentially hide $s\in\ZS^n$ from parties who do not know exactly $e$.

Next, we will describe concrete examples of using this idea. The method described next represents a symmetric Somewhat Homomorphic Encryption (\abr{SWHE}) Learning With Errors (LWE) based scheme known as \algname{{REGEV09}}. The method was proposed by \citet{regev2009lattices}. Its security properties are based on hardness to solve the decisional LWE.

\paragraph{KeyGen.} $s\in \ZS_q^n \sim U$ is a secret key, $q$ is a prime, $n$ is a secure parameter.

\paragraph{Encrypt.} Encryption is working bit by bit. The message that we encrypt without loss of generality can be considered as $m\in\{0,1\}$. The algorithm for encryption produces ciphertext as $c=(a,b) \in \ZS_q^n \times \ZS_q$ via following rules:

\begin{enumerate}
	\item $e \in \ZS$ is a "short" noise generated from some distribution, satisfied constraint $|e|<q/4$).
	\item $a \in \ZS_q^n$ is a random vector sampled uniformly from its domain.
	\item $b \eqdef \langle a,s \rangle + e + (m \cdot \lfk q/2 \rfk)|\mod q$.
	\item The released ciphertext $c=(a,b) \in \ZS_q^n \times \ZS_q$.
\end{enumerate}

\paragraph{Decrypt.} Decryption happens via using the following formula which involves function ${round}_a(x)$ which round to $0$ or $a$ depends on what is more close to $x$.
\begin{eqnarray*}
	\hat{m} &=& \dfrac{round_{q/2}\left(b- (\langle a,s \rangle \pmod q)
		\right) }{q/2} \\
	&=& \dfrac{round_{q/2}({\color{red}{(\langle a,s \rangle + e + m \cdot \lfk q/2 \rfk)}} - \langle a, s \rangle)}{q/2} = \dfrac{{round_{q/2}(e + {\color{red}{m}} \cdot \lfk q/2 \rfk))}}{q/2} .
\end{eqnarray*}

For decryption to work correctly, we need to have $$e \in (-\lfk q/4 \rfk, \lfk q/4 \rfk ).$$ 
Now let's verify that this formula works correctly:
\begin{enumerate}
\item If ${\color{red}{m=0}}$, and because $|e| < \lfk q/4 \rfk$ $\implies \hat{m} = 0$. 
\item If ${\color{red}{m=1}}$, and because $|e| < \lfk q/4 \rfk \implies e > -\lfk q/4 \rfk$ $\implies \hat{m} = 1$.
\end{enumerate}

\paragraph{HE Add.} We add two ciphertext messages $c_{i}=(a_{i},b_{i}),i \in \{0,1\}$ by components:
\begin{equation*}
	(a_1 + a_2, b_1 + b_2) = (a_1+a_2, \langle a_1 + a_2, s \rangle + (e_1 + e_2) + (m_1 + m_2) \lfk q/2 \rfk \pmod q .
\end{equation*}

This schema is SWHE and homomorphic additive. In the worst case error is doubled in each addition operation of each bit. If the adversary listens to the channel, he can collect $(a,b)$ for each transferred bit and construct $\bA=[a_1, a_2,\dots]$ and the right-hand side $b$. But encoding of ciphertext directly follows LWE problem description with additive error correction code $e + m \cdot \lfk q/2 \rfk$. By properties of the \textit{Decision LWE} Problem, adversaries can not get any information from it because data distribution $(\bA,b)$ is indistinguishable from a uniform.

\subsection{Introduction to CKKS}
There are different variants of \abr{FHE}, \abr{LFHE}, and \abr{SWHE}, but most of them operate on boolean or integer arithmetic. In our work, we compare \algnamewithaes{DCGD/PermK/AES} against \algname{GD} with Cheon-Kim-Kim-Song (\ecryptname{CKKS}) \citep{cheon2017homomorphic} schema. This schema allows approximate arithmetic on encrypted real and complex numbers and dense linear vectors. 

The \ecryptname{CKKS} schema violates property (1) of \abr{FHE} from Appendix \ref{ch4:app:overview_of_privacy_mechanisms}, and it is specialized to work with these two linear spaces $\mathbb{R}^d$ and $\mathbb{C}^d$. Therefore, \ecryptname{CKKS} is an \abr{SWHE} scheme. Still, it is sufficient for most machine learning problems that operate on real spaces during the training phase, and it is the most popular schema in machine learning applications that require \abr{HE}. Supporting multiple keys (MK) with privacy guarantees is an active research topic \citep{kluczniak2023circuit}. None of \ecryptname{MK-CKKS} possible schemas are currently implemented 
in \libname{TenSEAL} \citep{benaissa2021tenseal} or \libname{SEAL} \citep{seal} libraries. In our work, we used the classical single-key \ecryptname{CKKS} scheme for comparison because it is suitable for our setting. 

As we mentioned, the \ecryptname{CKKS} scheme allows us to perform \abr{HE} computations on vectors of complex and real values. \ecryptname{CKKS} provides ways to perform element-wise addition, multiplication, and rotation of elements in encrypted form. The \ecryptname{CKKS} uses three different keys:

\begin{itemize}
	\item \textit{Public key.} The public key in \ecryptname{CKKS} schema is used for the encryption of plain messages.
	
	\item \textit{Relinearization key}. Relinearization key reduces the size of a ciphertext after arithmetic operations on ciphertexts.
	
	\item \textit{Private (secret) key}. The private (secret) key is used for decryption of encrypted messages.
	
\end{itemize}

To perform arithmetic operations on encrypted vectors on the master device, the master should know the public key and the master should know the relinearization keys to reduce the size of a ciphertext after arithmetic operations. The public key in \ecryptname{CKKS} schema is used for encryption and can be shared with the master. However, a private (secret) key is used for decryption and must be kept confidential from the master if clients do not trust the master. The security of the scheme depends on the hardness of the Ring Learning With Errors (\abr{RLWE}) problem. The \abr{RLWE} is a generalization of \abr{LWE} problem, but instead of using $\mathbb{Z}_q^n$ field, it is based on underlying algebra which is polynomials over a ring. \abr{RLWE} inherits useful properties of hardness from \abr{LWE}, but it is more space efficient. For an overview of \abr{LWE} see Appendix~\ref{ch4:app:lwe}.

\subsection{Description of processes inside CKKS}

The schematic process of how \ecryptname{CKKS} operates is depicted in Figure~\ref{ch4:fig:ckks_enc}. The general schema of \ecryptname{CKKS} consists of several steps.

\paragraph{First step.} Firstly, a vector of values $m \in \mathbb{R}^{N/2}$ on which we want to perform certain computations is encoded into a plaintext polynomial $p(x)$. This is necessary because encryption, decryption, and other mechanisms work on polynomial commutative rings. In the case of encoding into polynomials from $\mathbb{Z}[X]/(X^N+1)$, the encoding process is reversible. In the original paper, \citet{cheon2017homomorphic} this process is carried out using canonical embedding. This embedding is carried in such a way that if evaluate the obtained polynomial from $\mathbb{Z}[X]/(X^N+1)$ at the roots of cyclotomic polynomial $X^N+1$ we will recover entries of the original message of $n \in \mathbb{R}^{N/2}$.

\paragraph{Second step.} Next, the actual encryption part encrypted the plain polynomial with integer coefficients via a public key into polynomial $(c0, c1) \in (Z_q[x]/(X^N+1))^2$. The $q$ is a modulus number in the \ecryptname{CKKS} scheme. Here $q=\prod_{i=1}^{L}q_i$, where $q_i$ are prime numbers. So $q$ is not necessarily a prime number, but it is chosen to be a product of several prime numbers. To carry encryption the public key in the form of 
$Z_q[X]/(X^N+1))$ after specific transformation is used. It means that the size of the public key in bits is approximately equal to $N \cdot q$. To have \aesname{AES-128} security level for \ecryptname{CKKS} the key size is at least equal to $420$ KBytes.

\paragraph{Third step.} Next, the algebraic operations are carried on encrypted messages $(c_0, c_1)$ by specific rules which we will not go deep into.

\paragraph{Fourth step.} After carrying out the arithmetic operation, the result will be from the same space $(c0, c1) \in (Z_q(x)/(x^N+1))^2$. To carry decryption the private key in the form of $Z_q[X]/(X^N+1))$ after a specific transformation is used. Therefore it means that the size of the private key in bits is approximately equal to $N \cdot q$. To have \aesname{AES-128} security level the key size at least equal should be equal to $420$ KBytes.

\paragraph{Fifth step.} Finally during decoding, a message from space $(Z_q(x)/(x^N+1))^2$ to plain text in $Z[x]/(x^N+1)$ and finally perform decoding into $\mathbb{R}^{N/2}$.

\newpage

\subsection{CKKS configuration equivalent to AES-128}

To provide privacy grantees similar to \aesname{AES-128} encryption: $N$ should satisfy this condition $N>16384$, and $q=\prod_{i=1}^{K}q_i$ should be at least $438$ bits long. This ensures that the security parameter $N$, $q$ is large enough to provide sufficient security. These details can be found in the reference implementation of Microsoft Research \libname{SEAL} Library
\footnote{\href{https://github.com/microsoft/SEAL/blob/master/native/src/seal/util/hestdparms.h}{https://github.com/microsoft/SEAL/blob/master/native/src/seal/util/hestdparms.h} - Microsoft Research SEAL Library.}. This is an underlying reason why \ecryptname{CKKS} has a more memory requirement compared to \aesname{AES-128}. The encoding consists of $2$ polynomials with $2 \max(d, N)$ coefficients, and each coefficient is not 32 (for FP32) or 64 (for FP64) bits long, but it is essentially $q=438$ bits long. In addition, when input/output vectors do not match $N$ \ecryptname{CKKS/HE} requires performing chunking of input and output - it requires additional operations to maintain the correctness and efficiency of the computation.

In all our experiments, we used the following \ecryptname{CKKS} configuration inside \libname{TenSeal} \citep{benaissa2021tenseal} library version 0.3.14. For this library to obtain \aesname{AES-128}, we have used the recommended configuration:

\begin{enumerate}
	\item Polynomial degree: $2^{14}=16\,384$.
	\item Coefficient modulus: $q_1, q_2, q_3, q_4, q_5 = (60, 30, 30, 30, 60)$ bits, which corresponds to $q$ size of $210$ bits. This configuration is recommended, by \libname{TenSeal} for \aesname{AES-128} security level.
	\item Scale factor: $2^{30}$.
	\item Scheme type: \ecryptname{CKKS}.
\end{enumerate}

\tikzstyle{startstop} = [rectangle, rounded corners, minimum width=3cm, minimum height=1cm,text centered, draw=black, fill=red!30]
\tikzstyle{process} = [rectangle, minimum width=3cm, minimum height=1cm, text centered, draw=black, fill=orange!30]
\tikzstyle{arrow} = [thick,->;,>;=stealth]

\begin{center}
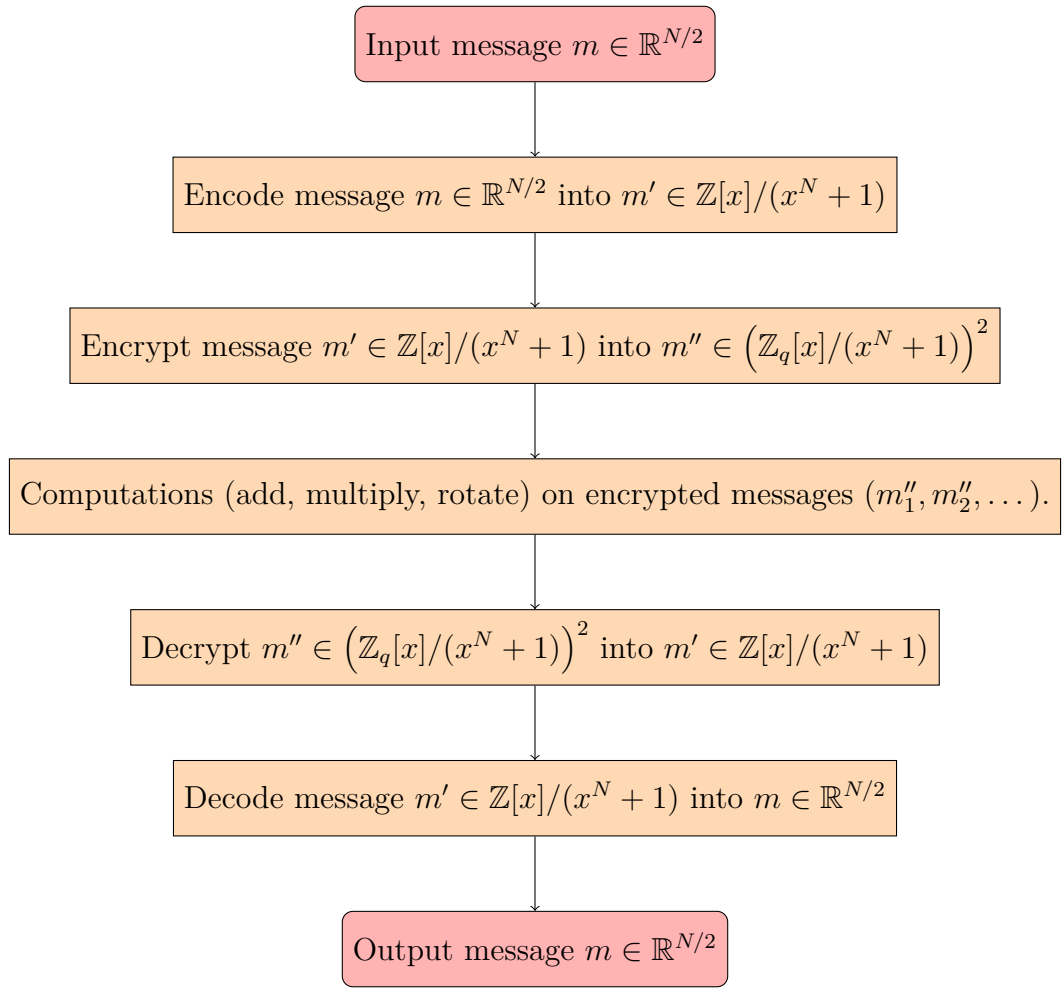
\begin{figure}
	\centering
	\begin{tikzpicture}[node distance=2cm]
		\label{ch4:fig:ckks_enc}
		\node (input) [startstop] {Input message $m \in \mathbb{R}^{N/2}$ };
		\node (encode) [process, below of=input] {Encode message $m \in \mathbb{R}^{N/2}$ into $m' \in \mathbb{Z}[x]/(x^N+1)$ };
		\node (encrypt) [process, below of=encode] {Encrypt message $m' \in \mathbb{Z}[x]/(x^N+1)$ into $m'' \in \left(\mathbb{Z}_q[x]/(x^N+1)\right)^2$ };	
		\node (compute) [process, below of=encrypt] {Computations (add, multiply, rotate) on encrypted messages $\left(m_1'', m_2'', \dots\right)$.};
		\node (decrypt) [process, below of=compute] {Decrypt $m'' \in \left(\mathbb{Z}_q[x]/(x^N+1)\right)^2$ into $m' \in \mathbb{Z}[x]/(x^N+1)$};
		\node (decode) [process, below of=decrypt] {Decode message $m' \in \mathbb{Z}[x]/(x^N+1)$ into $m \in \mathbb{R}^{N/2}$ };	
		\node (output) [startstop, below of=decode] {Output message $m \in \mathbb{R}^{N/2}$ };	
		
		\draw [->] (input) -- (encode);
		\draw [->] (encode) -- (encrypt);
		\draw [->] (encrypt) -- (compute);
		\draw [->] (compute) -- (decrypt);
		\draw [->] (decrypt) -- (decode);
		\draw [->] (decode) -- (output);
	\end{tikzpicture}
	\caption{A high-level view of operations inside the \ecryptname{CKKS} schema.} \label{ch4:fig:ckks_enc}
\end{figure}
\end{center}

\clearpage
\addtocounter{adjsection}{1}
\section{Extra Experiments}
\label{ch4:app:extra_experiment}

\subsection{Exploring problem dimension}
\label{ch4:app:simulation_experiment}

\begin{figure*}[h]
	\centering
	\captionsetup[subfigure]{labelformat=empty}
	
	\begin{subfigure}[ht]{0.325\textwidth}
		\includegraphics[width=\textwidth]{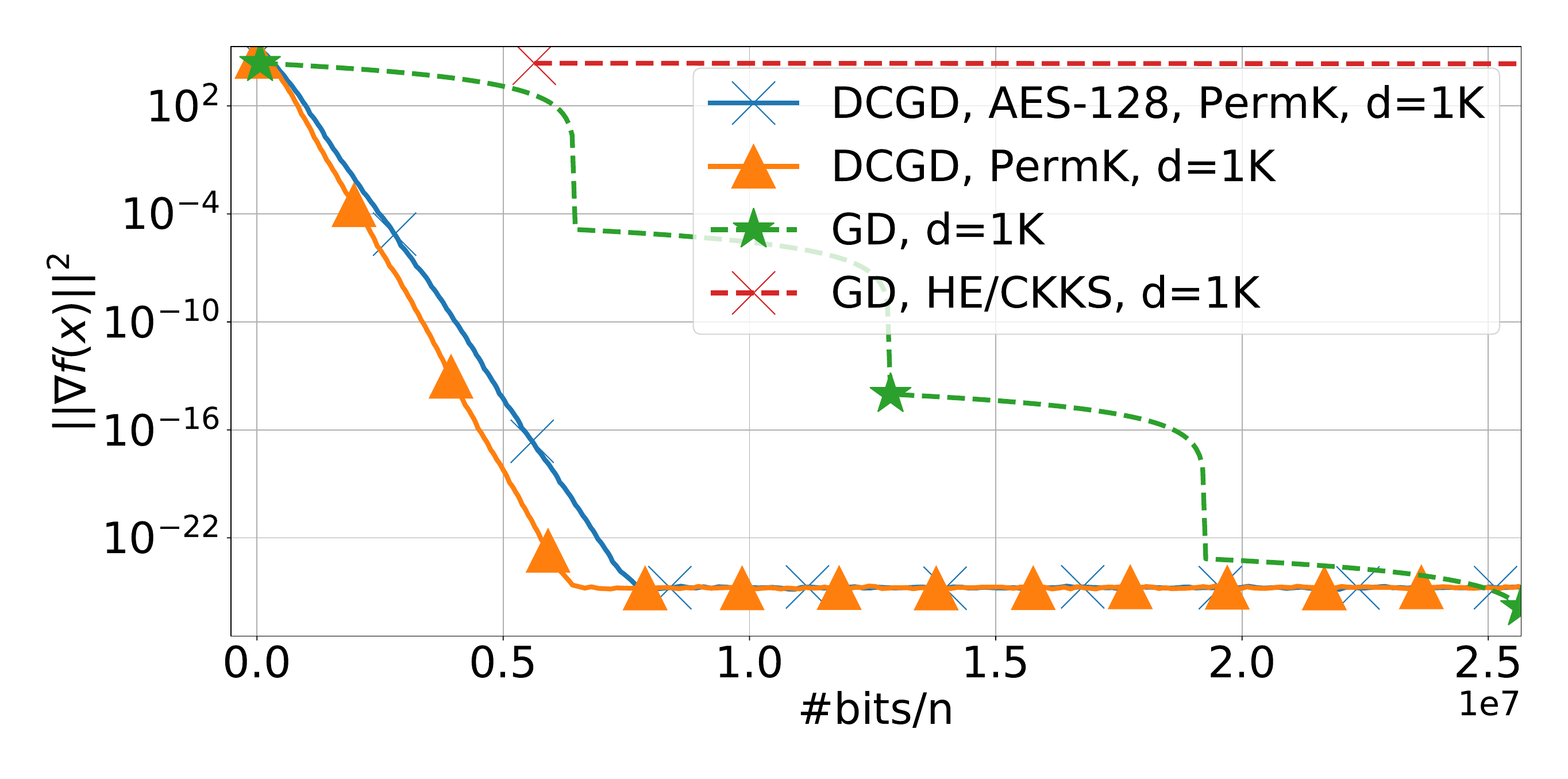} 
		\caption{{  }}
	\end{subfigure}
	\begin{subfigure}[ht]{0.325\textwidth}
		\includegraphics[width=\textwidth]{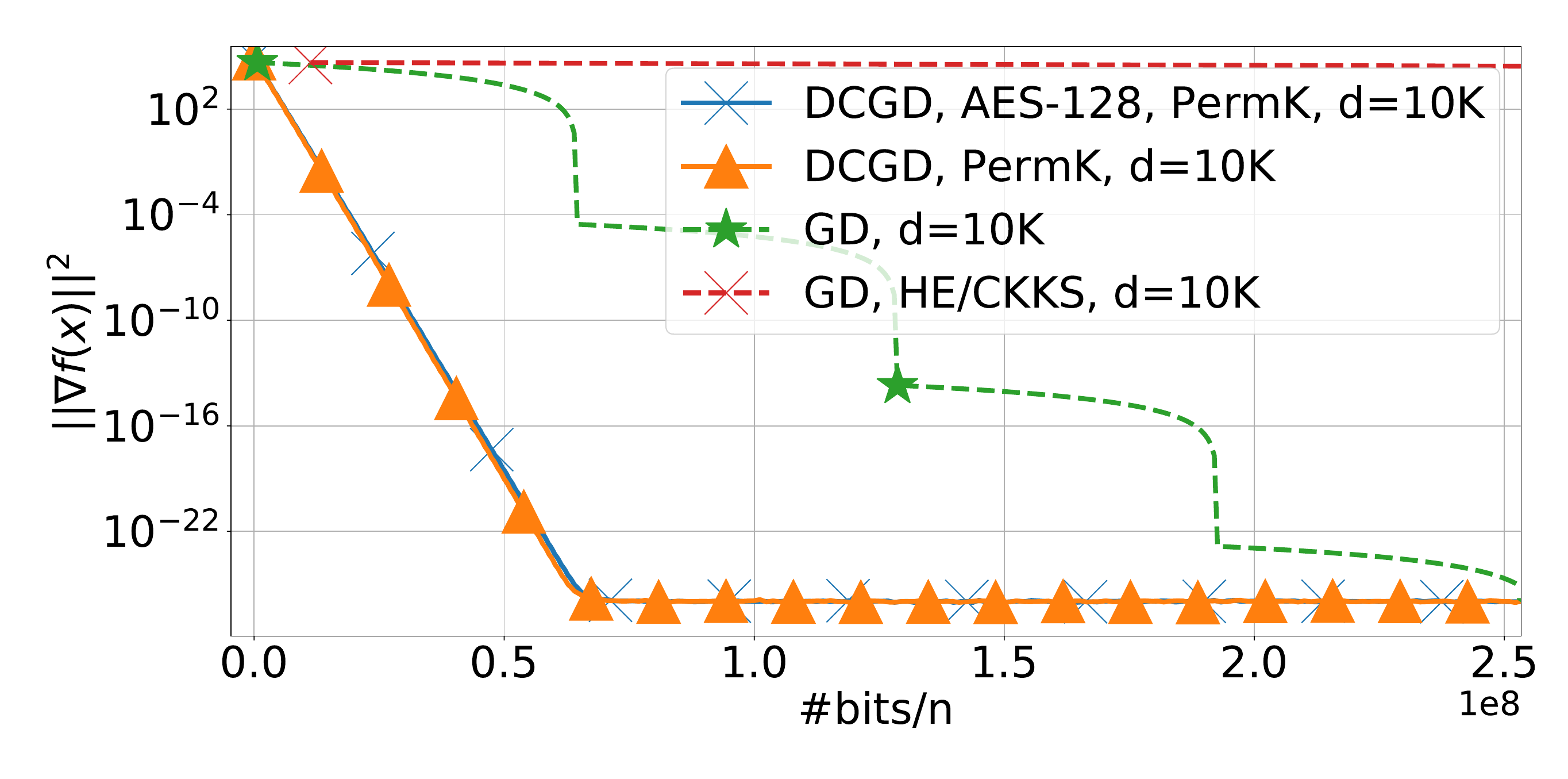} 
		\caption{{ }}
	\end{subfigure}
	\begin{subfigure}[ht]{0.325\textwidth}
		\includegraphics[width=\textwidth]{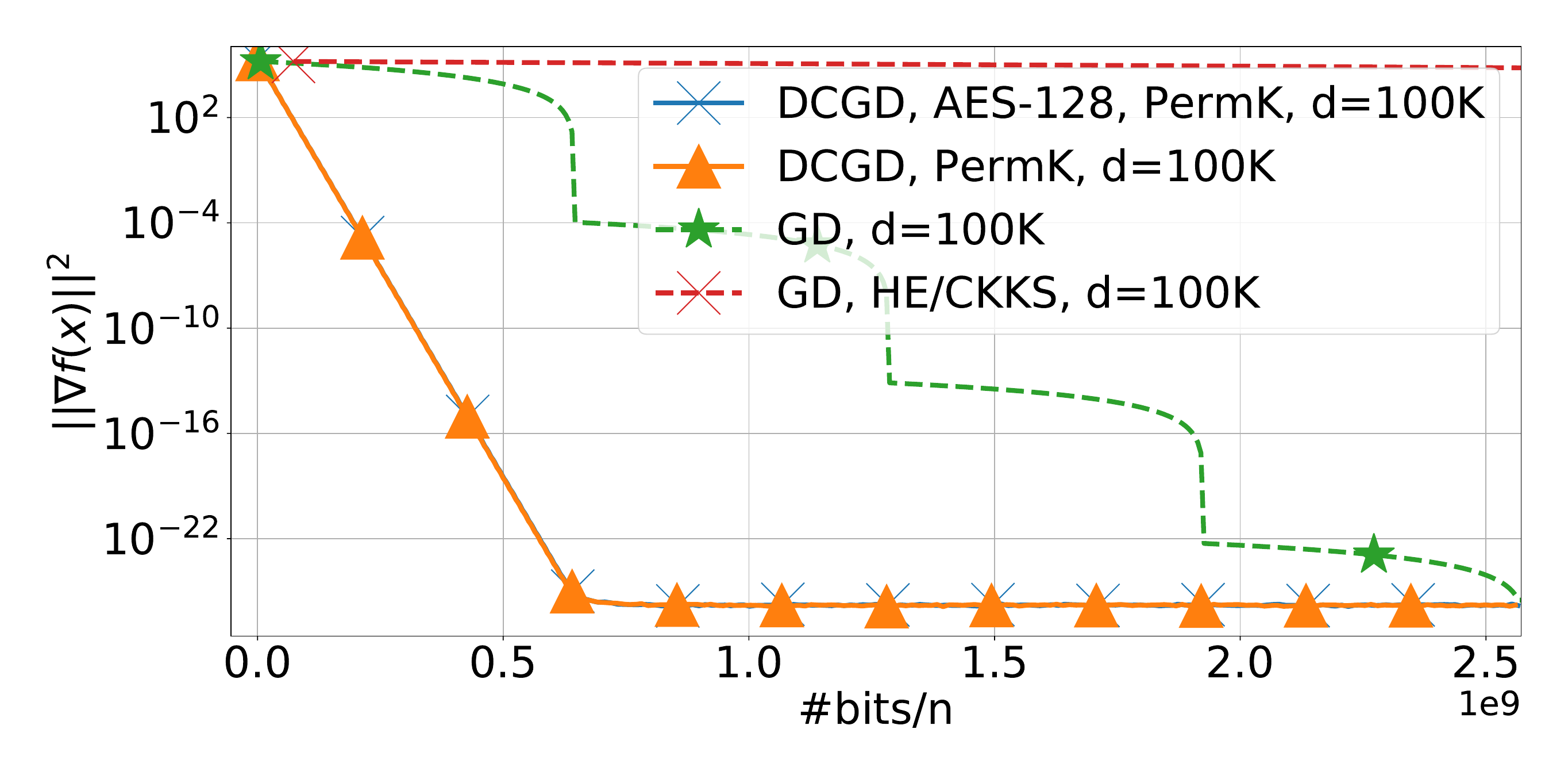}
		\caption{{  }}
	\end{subfigure}
	
	\begin{subfigure}[ht]{0.325\textwidth}
		\includegraphics[width=\textwidth]{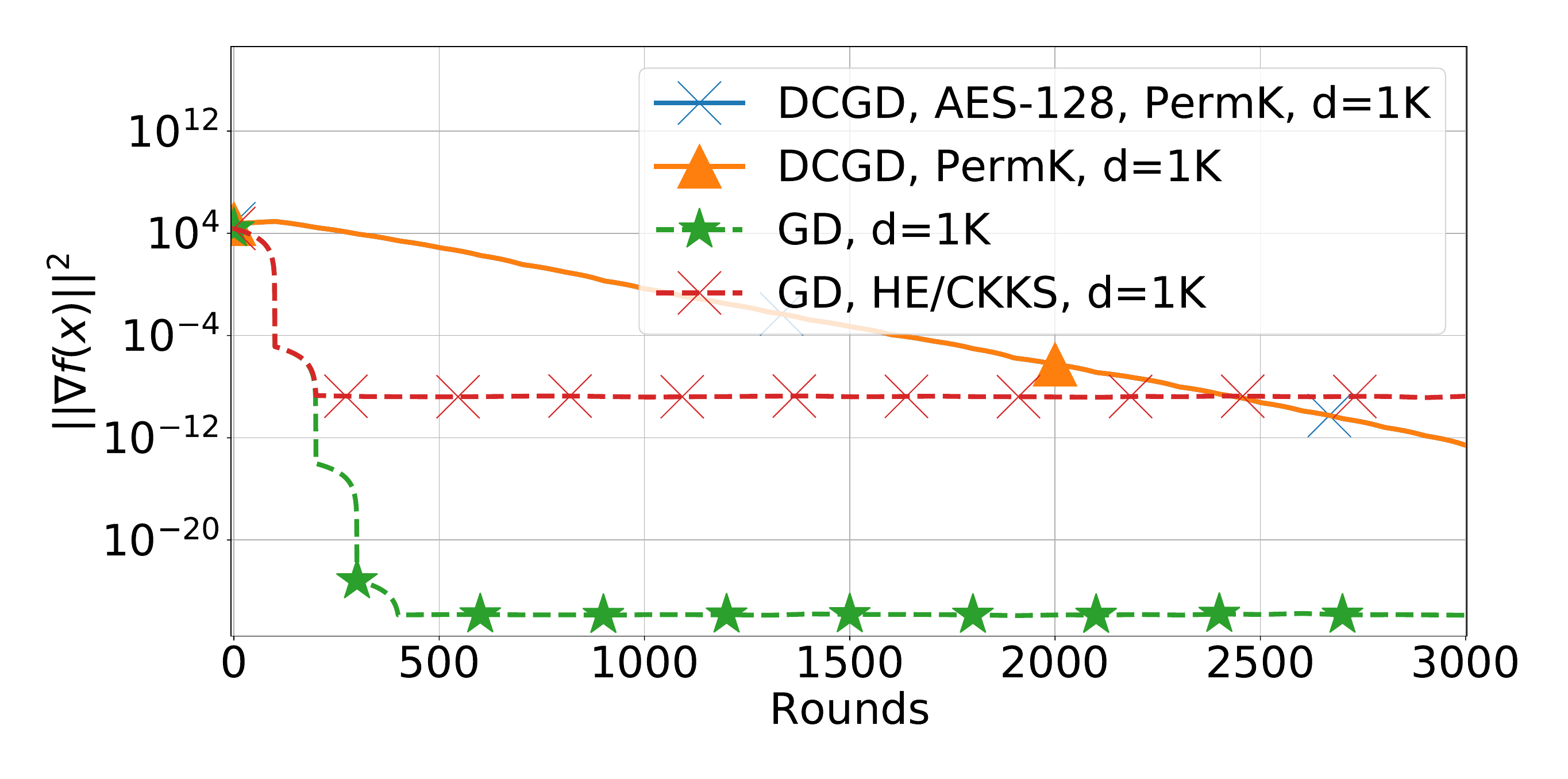} 
		\caption{{  }}
	\end{subfigure}	
	\begin{subfigure}[ht]{0.325\textwidth}
		\includegraphics[width=\textwidth]{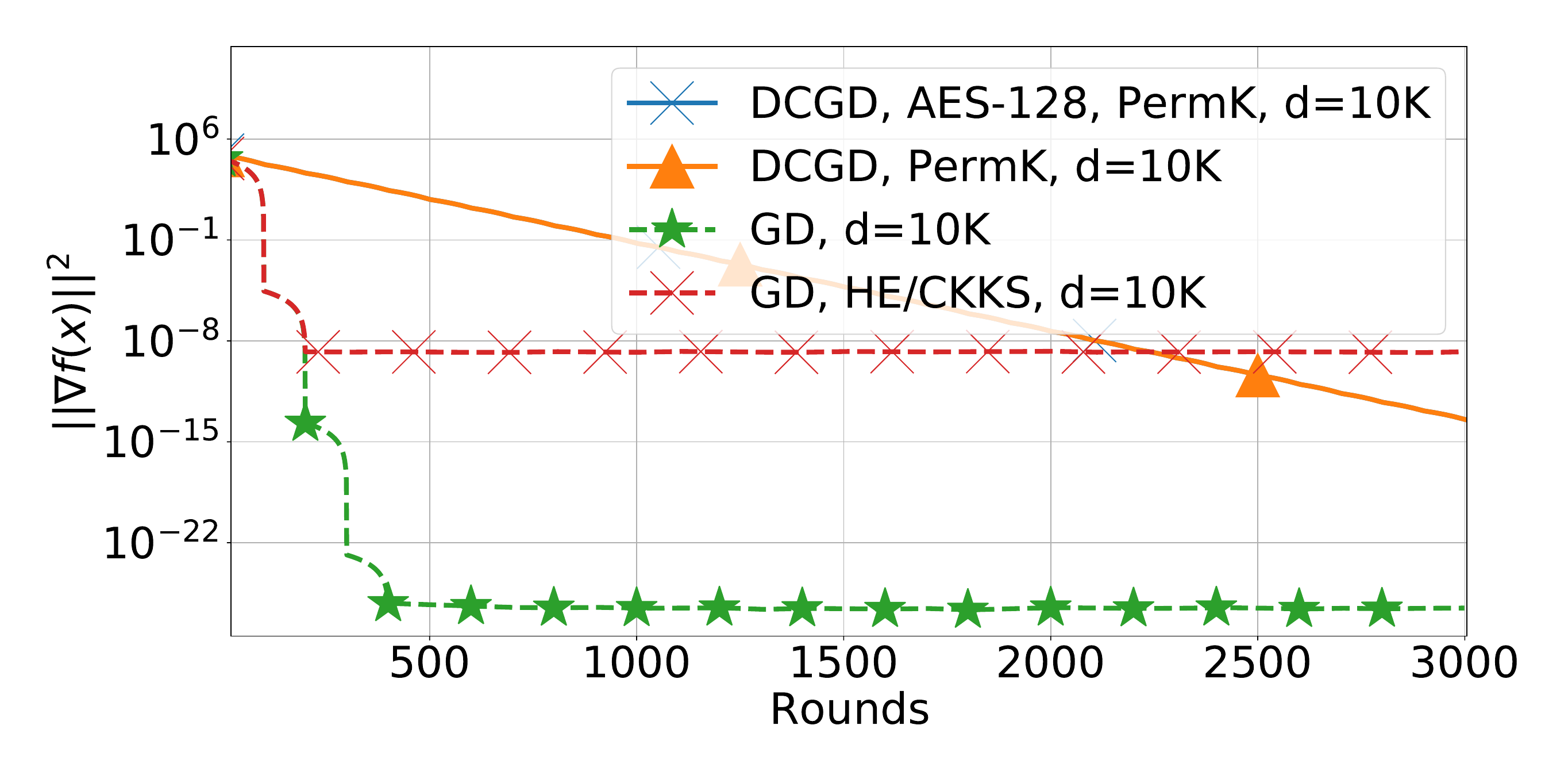} 
		\caption{{  }}
	\end{subfigure}
	\begin{subfigure}[ht]{0.325\textwidth}
		\includegraphics[width=\textwidth]{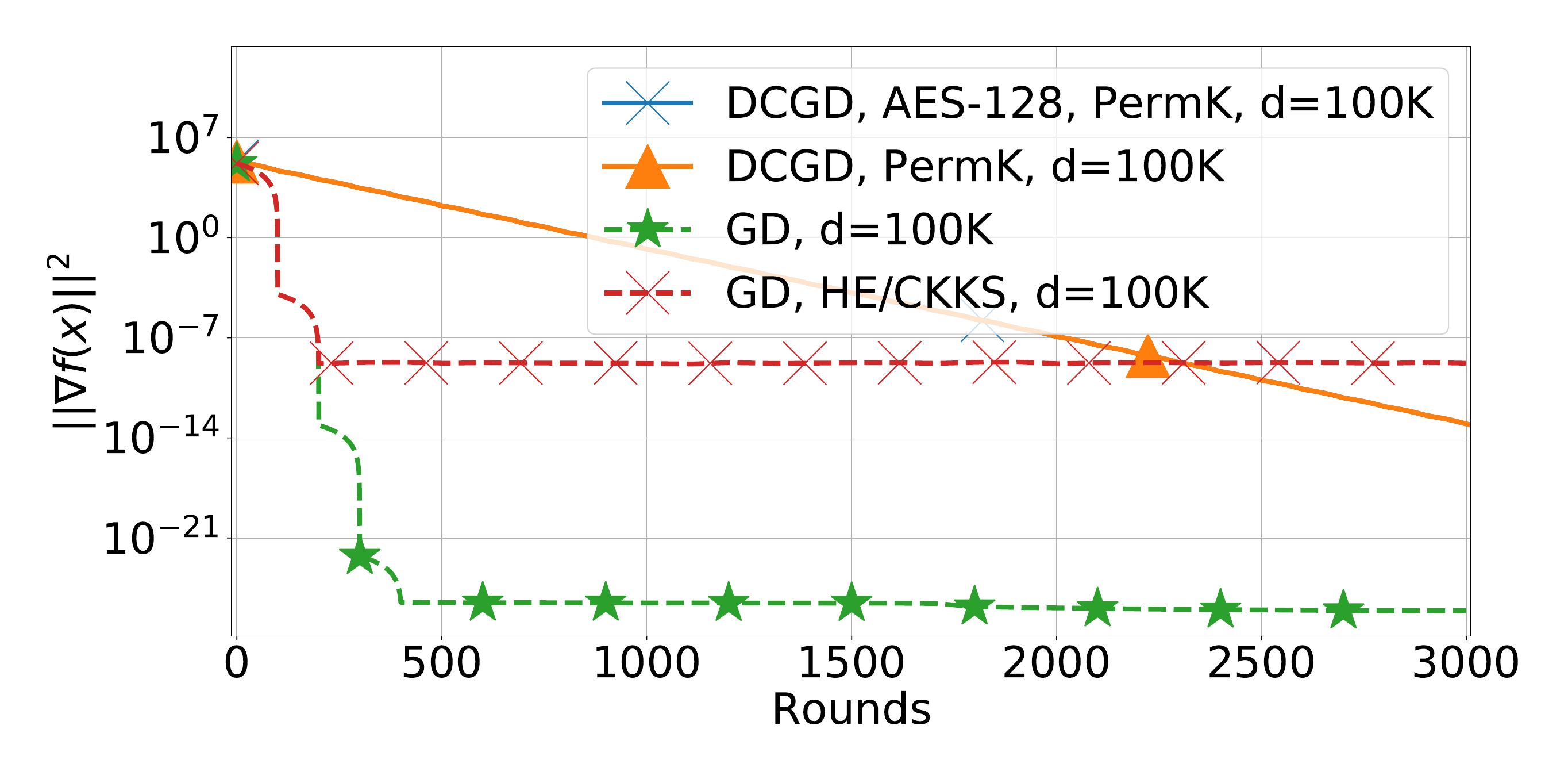} 
		\caption{{ }}
	\end{subfigure}
	
	\begin{subfigure}[ht]{0.325\textwidth}
		\includegraphics[width=\textwidth]{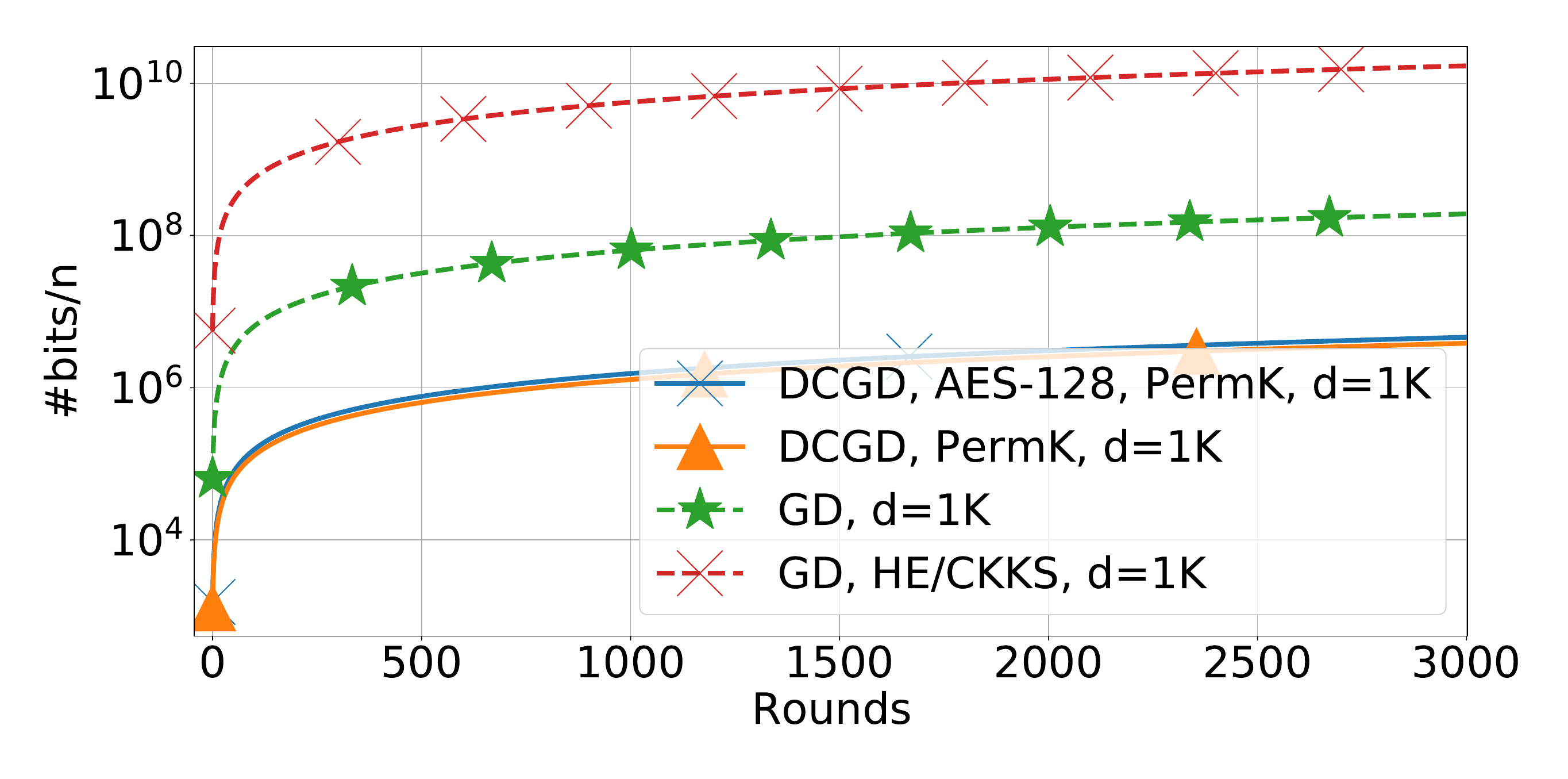} 
		\caption{{  }}
	\end{subfigure}
	\begin{subfigure}[ht]{0.325\textwidth}
		\includegraphics[width=\textwidth]{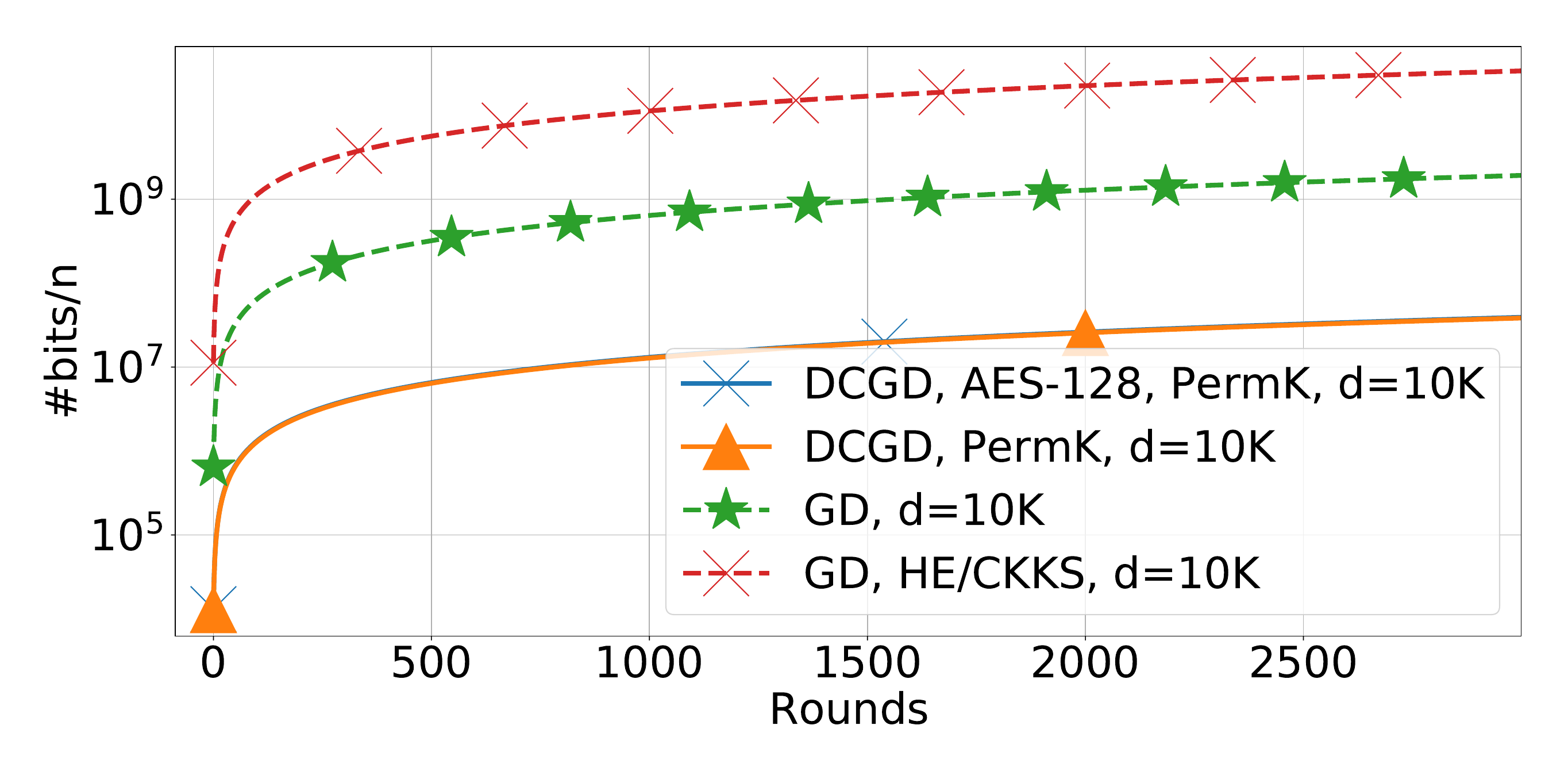} 
		\caption{{  }}
	\end{subfigure}
	\begin{subfigure}[ht]{0.325\textwidth}
		\includegraphics[width=\textwidth]{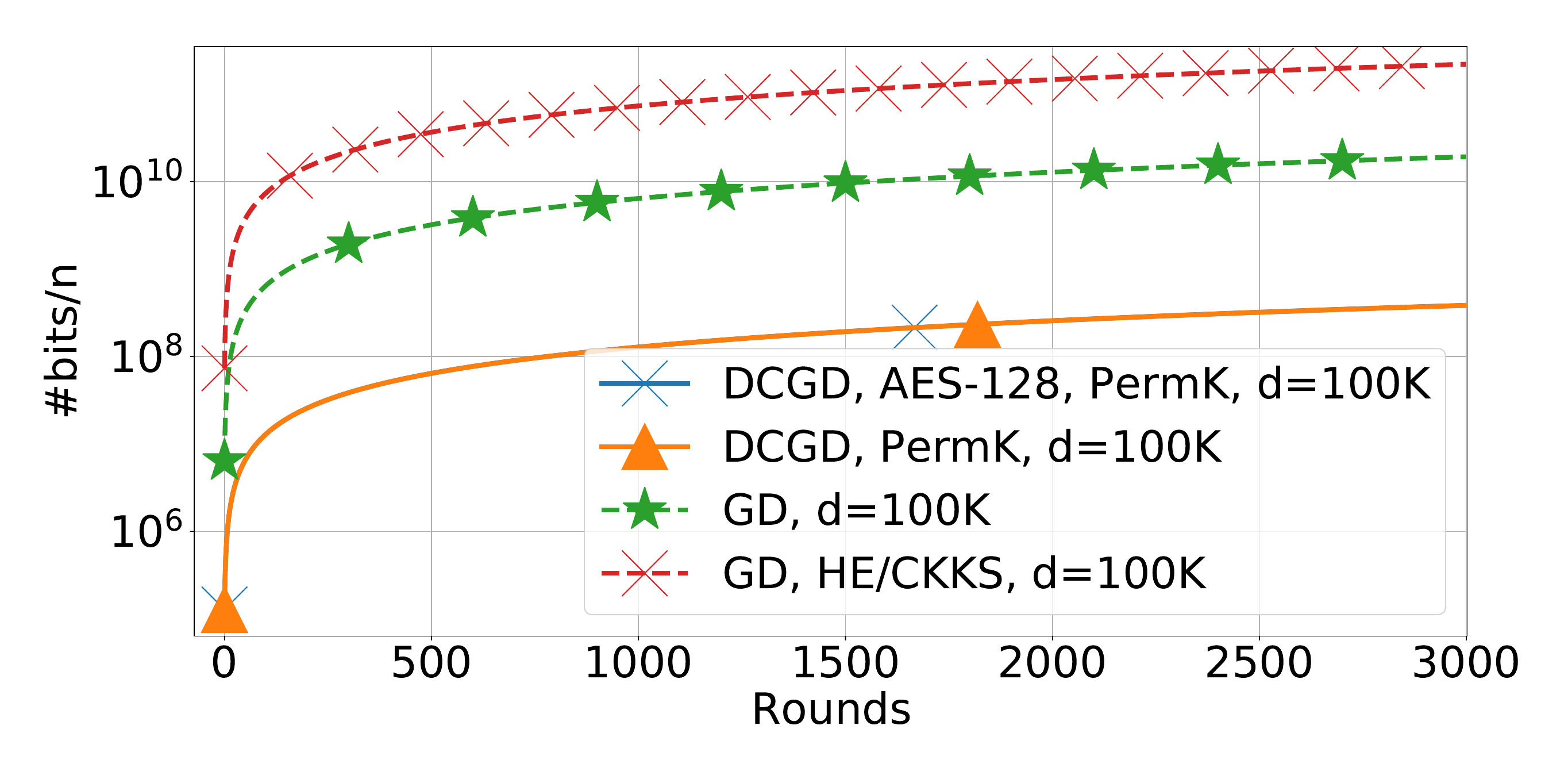} 
		\caption{{  }}
	\end{subfigure}
	
	\begin{subfigure}[ht]{0.325\textwidth}
		\includegraphics[width=\textwidth]{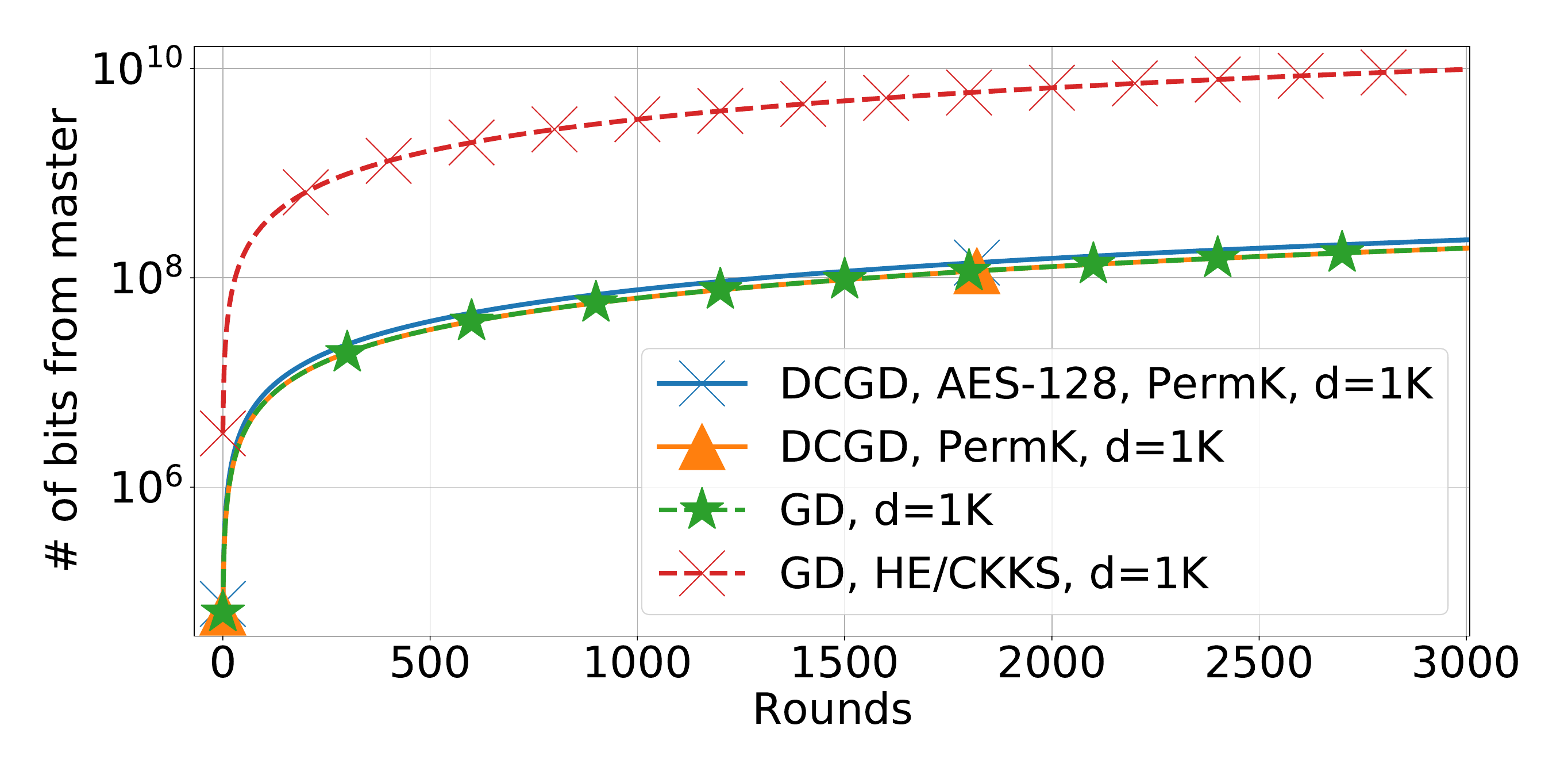} 
		\caption{{ (a) $d=1\,000$ }}
	\end{subfigure}
	\begin{subfigure}[ht]{0.325\textwidth}
		\includegraphics[width=\textwidth]{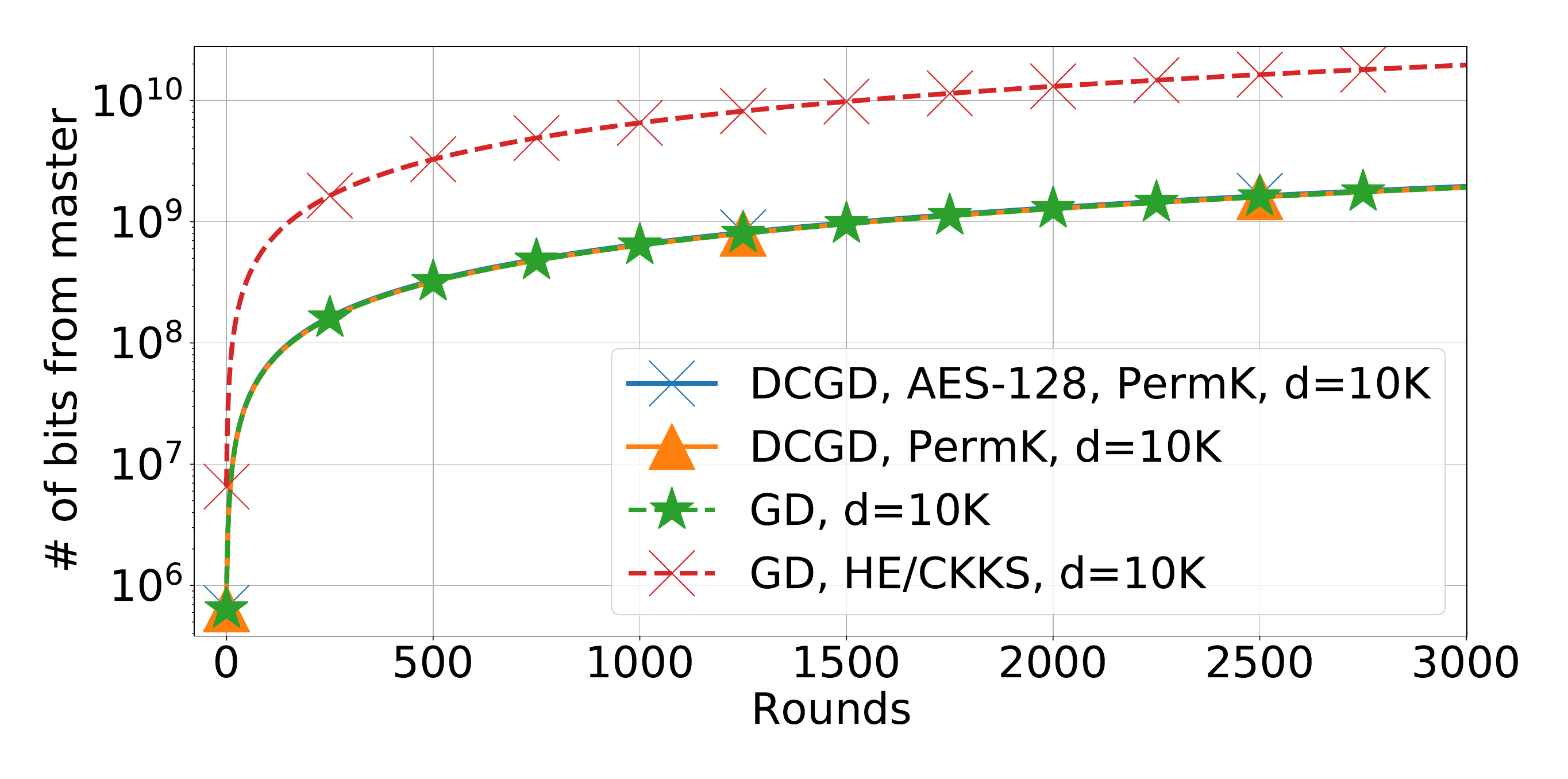} 
		\caption{{ (b) $d=10\,000$ }}
	\end{subfigure}
	\begin{subfigure}[ht]{0.325\textwidth}
		\includegraphics[width=\textwidth]{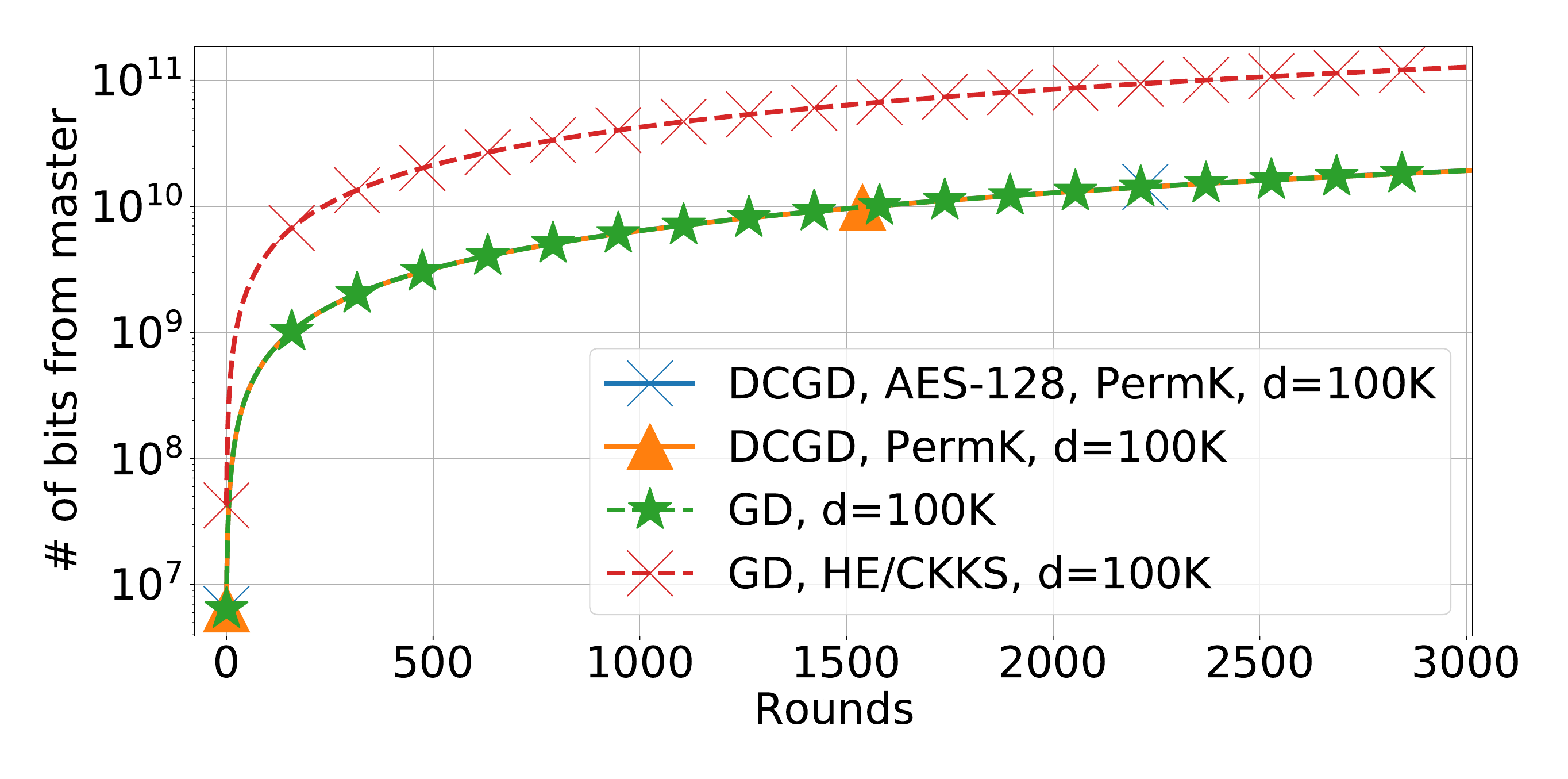} 
		\caption{{ (c) $d=100\,000$ }}
	\end{subfigure}
	
	\caption{{\modelname{Linear regression} in interpolation, $n=50$, $n_i=12$, compute FP64. Tuned $\gamma=0.007$ for \algname{DCGD}. Theoretical $\gamma$ for \algname{GD}.}}
	\label{ch4:fig:exp_syn_7}
\end{figure*}

This experiment investigates the impact of problem dimension $d\in \{10^3,10^4,10^5\}$. Figure~\ref{ch4:fig:exp_syn_7} shows that the \ecryptname{CKKS} overhead from encryption is $\times 10^3$ more both in master to the client, and the client to master communication direction compared to \algnamewithaes{DCGD/PermK/AES}. With $d=10^6$ the memory footprint for \ecryptname{CKSS} configured to guarantee the same guarantees as \aesname{AES-128} in the master to store $n=50$ encrypted gradients is $46$ GBytes, rendering storage of such information in the master challenging. The best convergence relative to the volume of information sent to the master is achieved with \algname{DCGD/PermK}. The behavior of \algname{DCGD/PermK} and \algnamewithaes{DCGD/PermK/AES} is indistinguishable for $d>10K$. Despite the ciphertext size being the same as the input when using \aesname{AES}, proper use of \aesname{AES} block ciphers for communication requires the addition of a Message Authentication Code (\abr{MAC}) for protection against malicious errors and a unique pseudo-random identifier (nonce). Each of these adds an overhead of $16$ bytes. It explains different behavior observed for \algname{DCGD/PermK} and \aesname{DCGD/PermK/AES} at $d=1K$ in Figure~\ref{ch4:fig:exp_syn_7}. 

Given that, \algnamewithaes{DCGD/PermK/AES} emerges as a more viable alternative to \ecryptname{CKKS} in {FL} context, in the setting when \abr{HE} previously has been applied.

\clearpage

\subsection{Compute and communication overlap for DCGD/PermK}
\label{ch4:app:simulation_experiment}

In this experiment, we simulated a compute and communication environment with realistic assumptions and evaluated the performance of \algname{DCGD/PermK}. We ignored overheads from the Operating System (OS) and its components (drivers, kernel, services). Firstly, we provide background on Network Communication and System Architecture to justify our modeling choices.

\subsubsection{Background in network communication}
\label{ch4:app:systems_background}



\paragraph{Communication bandwidth during master client communication.}
We consider a scenario where clients are connected to the master via the Internet. These devices are called \textit{end systems} or \textit{hosts} in the context of a communication network. The communication between end systems is facilitated by intermediate \textit{routing devices}, such as routers and switches, connected by \textit{links}. Internet communication is based on \textit{packet switching}, which means that exchange messages $m$ are divided into smaller units called \textit{packets}. Then, they are delivered with the best effort through the communication network without reserving a path (circuit) from sender to receiver in advance. 

The \textit{bandwidth} of a link is the maximum rate at which data can be transferred over it and is measured in bits/second. This rate depends on various factors and generally can vary over time. It depends on:

\begin{enumerate}
	\item The distance between the linked devices.
	\item The type of physical medium for carrying the signal.
	\item The number of packets in the queue to transfer in routing devices.
	\item The probability that clients on the Internet will use the same link simultaneously.
	\item The probability that a packet will be dropped in a routing device.
	\item The number of clients attached to the link.
\end{enumerate}

We can distinguish between two types of bandwidth:
\begin{enumerate}
	\item \textit{Instantaneous bandwidth} - bandwidth at a specific moment of time.
	\item \textit{Average bandwidth} - average bandwidth over a complete message transfer.
\end{enumerate}

If the information flows from client to master via a directed path across $N$ links, then the effective bandwidth of the path is determined by the minimum bandwidth among all links:
\[B_{\mathrm{transfer}}=\min_{i\in [N]} B_{i}\].

The link with this minimum bandwidth is called the \textit{bottleneck link}.

\paragraph{Communication delays during master client communication.}

The delay of a single link device is the actual time it takes for a specific bit of data to travel from one link device to the next. The delay consists of several components:

\begin{enumerate}
	\item \textit{Processing delay} is the time required to examine the packet header and determine where to forward the it in a link device with several output ports.
	\item \textit{Transmission delay} is the time required for the router to push out the packet into the link, without considering the propagation of signal by itself.
	\item \textit{Propagation delay} is the time required for the signal to travel along the physical medium.
	\item \textit{Queuing delay} is the time required for a packet to wait in the output buffer if the link output port is busy.
	\item \textit{Loss delay} is the time required for a packet to be retransmitted if it is dropped due to buffer overflow or errors due to medium.
\end{enumerate}

Suppose a client and a server are connected by $N$ links sequentially, and the information flows through them in a directed path. In that case, the total delay for transferring a single package in which a single concrete bit is located over this path is given by the sum of all delays along each link over the path:
\[\tau_{\mathrm{delay}}=\sum_{i=1}^{N} \tau_{\mathrm{delay},i}.\]

\subsubsection{Network communication model}

As we have seen, the total delay and bandwidth vary, even for the fixed topology of connecting the master and clients. According to \footnote{\href{https://www.speedtest.net/global-index}{Speedtest Global Index}} in the average download bandwidth across the globe is $B=41.54\,\mathrm{MBps}$ and communication latency is $RTT=28\,\mathrm{ms}$. The delay time for communication with messages $m \in \mathbb{R}^d$ where each component is represented by $bpp$ bits per component from master to client is modeled in the following way:
\[\tau_{delay} = \dfrac{RTT}{2} + \dfrac{d}{B} \cdot bpp.\]

Here, $B$ represents the bandwidth (or throughput) of the communication channel between the master and the client, which we assume to be constant. Next, $RTT$ is the round-trip time for a small packet to travel from the client to the server and back. This is an approximation and does not account for factors such as packet loss delay. The worst-case maximum delay is particularly difficult to estimate. We assume that both $\frac{RTT}{2}$ and $B$ remain unchanged during training. However, in real communication networks, the exact path from the client to the master can already vary over time.
 
In reality, a Network Interface Controller (\abr{NIC}) implements network communication in a local computing device, and it is typically connected with a PCI-Express bus as an external input/output device to the whole local computation system. For modern PCI-Express buses such as PCI-E v5, the bandwidth, even for a single physical lane x1, is in the order of $4000.0$ MBps, and latency for this bus is on the order of $35\cdot 10^{-3}$ ms. Increasing the number of PCI-E lanes increases throughput (bandwidth) but does not reduce latency.

Therefore in context when devices are communicated via the Internet, the effect of delays from PCI-Express is negligible, or at least it does not represent the bottleneck both in terms of bandwidth and latency. The same holds for involving communication time to transfer data from CPU to DRAM memory. Modern DDR5 memory has a bandwidth of $51 200$ MBps, and latency is measured in the order of nanoseconds. It means that what is represented as a bottleneck from a communication point of view (both in terms of latency and bandwidth) with the server is the connection to another \abr{NIC} installed in another side. This means that what is considered a bottleneck from a communication perspective, both in terms of latency and bandwidth, is the connection between two \abr{NIC}s installed in the client and the server.

\subsubsection{CPU-based computation in clients} 

To evaluate the compressed gradient $C_i(\nabla f_i(x))$ in a client, different algorithms can be used, such as analytical, numerical, symbolic, and methods that leverage Automatic Differentiation. These algorithms need to run on some device that can execute them efficiently. We will focus on the execution aspects of modern Central Processing Units (CPUs), which are the most flexible devices from a programming perspective.

\subsubsection{Background in modern Central Processing Unit}

CPUs have multiple cores that contain various components that work together to execute algorithms.

\paragraph{Instruction decode.} In this stage, the CPU decodes the instructions and obtains information about the input, output, and operation type. Then, it splits the instructions into \textit{micro operations} and puts them into the Operation Issue Queue. This stage may introduce some complexities. One complexity is that some CPUs can decode multiple instructions simultaneously and issue them in parallel. This is called a multi-issue (or \textit{superscalar}) design, which aims to exploit instruction-level parallelism by executing independent instructions concurrently. Another complexity is that the CPU with \textit{out of order} issue capability can execute instructions whenever they are ready, regardless of the original order.

\paragraph{Operation issue queue.} This is a hardware queue where decoded instructions, in the form of \textit{micro-operations}, are stored. The queue has a minimum length that can accommodate the longest sequence of \textit{micro-operations} for any instruction in the CPU's instruction set.

\paragraph{Control unit (CU).} The CU operates at the level of \textit{micro-operations} and performs the following functions: (a) selecting the way to connect electrical components using multiplexers and demultiplexers; (b) turning on/off different electronic components; and (c) controlling the control lines of electronic components.

\paragraph{Multiplexers and demultiplexers.} During pipeline execution, intermediate inputs and results are stored in the latches (small intermediate buffers) of electrical components. To route signals within the CPU, multiplexers, and demultiplexers are utilized. Signals propagate through the component once all the components are connected, and data is applied to the input ports. The results are then produced when the enabled control signal reaches the electrical component.

\paragraph{Adders and multipliers.} The adders and multipliers are electrical circuits that perform addition or multiplication when turned on. The input for these devices is read from the intermediate latches. The typical input source (after intermediate routing with multiplexers and demultiplexers) is the Register File.

\paragraph{Register file.} The Register File is a storage unit that holds all the registers in the CPU. It may have multiple ports that allow parallel access to it. The Load and Store Units may have direct access to the Register File.

\paragraph{Load and store units (LS).} The Execution pipeline sends requests to the LS units for memory access. The LS unit can access the Register File, the Translation Lookaside Buffer (TLB) for address translation, and the Memory Cache.	

\paragraph{Translation lookaside Buffer (TLB).} To read code or data from memory in the user space or kernel space of the OS, the first step is to find the actual physical address of the specific memory location. This operation occurs for every instruction of a program. The TLB is a cache that stores the mapping between virtual page numbers and physical frame numbers, speeding up the address translation process for memory access. Without the TLB, the virtual addressing mechanism would require several accesses to different page tables, significantly increasing the time. The TLB relies on the locality of code and data in most algorithms.

\paragraph{CPU memory cache.} The CPU memory cache is a fast storage unit that holds frequently accessed data and instructions, operating on fixed blocks of bytes known as \textit{cache lines}. It is used to reduce the latency of accessing the DRAM memory. The Load and Store Units have access to the Cache. Modern high-end systems support three levels of Cache. The L1 cache is typically split between data and instructions, and it is the closest to the CPU. The L2 and L3 caches are larger and slower, and they can be shared by multiple CPU cores. The cache implementation varies across different CPUs, depending on the trade-offs between speed, potential conflicts, hardware complexity, cache replacement policy, and power consumption. If the data is not available in the Cache, then the CPU Cache requests a block of memory from the Memory Controller (MU), which accesses the DRAM memory. 

When data from DRAM is stored in multiple CPU caches, it fundamentally means that data may be stored in several places. In this situation, another aspect becomes important: (a) \textit{cache consistency}, which essentially means that all copies of DRAM cache lines should be the same in all caches in the system; (b) \textit{cache coherence} which essentially means that any read of memory returns the most recent update anywhere in the system.

\paragraph{DRAM memory controller (MC).} The DRAM Memory Controller is a device that manages access to the main DRAM memory Chips. It is used to fetch data from the Main Memory when it is not available in the CPU cache. The Memory Controller returns the data to the caches in blocks of a fixed size, called cache lines, and typically it is $64$ bytes. The Memory Controller is also responsible for running the memory bus transactions, which are the transfers of data between the MC and DRAM memory chips. The Memory Controller is typically implemented in the hardware as a device that is shared by multiple cores.

\paragraph{DRAM memory chips.} Memory chips are devices that store data in binary form. They are usually specified by the number of bits stored and the number of bits accessed in one read or write operation. For example, a common DRAM chip \textit{4Gbx1} means that it can store 4G bits and access one bit at a time. To protect data from corruption, extra logic may store bits for Error Correction Codes.

\paragraph{Input and output buses.} Input and Output buses serve as pathways connecting external devices to the CPU. Examples of I/O buses include SATA, USB, and PCI-Express. For example, PCI-Express is commonly used to communicate with devices such as graphics cards and network cards. Communication with such external devices can be achieved using Direct Memory Access (DMA) or Programmed Input-Output (PIO). DMA enables devices to transfer data directly to or from memory without involving the CPU, while PIO requires the CPU to issue commands and wait for data. DMA is more efficient and faster than PIO but necessitates additional hardware support.


\subsubsection{Modeling of computation}

We model clients' compute capability by assuming that they have a computation device similar to Intel-Xeon-E5-2666-v3 CPU \footnote{\href{https://www.intel.com/content/www/us/en/products/sku/81706/intel-xeon-processor-e52660-v3-25m-cache-2-60-ghz/specifications.html}{Intel Xeon Processor E5-2660 v3 Specification}}. We assume that the computation device of the client and the master is represented by a CPU with $10$ CPU cores, working at frequency $3.2$ GHz, CPU support hyper-threading with executing $2$ computation works per core, we assume that Multiply - Add (MAD) operation is possible which effectively doubles compute throughput. There are $2$ functional units (FU) or compute ports per core for Floating Point arithmetic in such a device. We assume that add, subtract, and multiply operations for float numbers require a throughput of $1$ operation/clock and latency of $1$ clock per execution unit for FP32 arithmetic, which is realistic. With these assumptions, this device has a peak computation throughput of $238.41$ GFLOPS with FP32 arithmetic. And we suppose all $n=4$ clients and masters are equipped with it.

In our model, we will assume that, on average, the train data is located in the L2 cache, and all memory operations in the client can be executed via accessing the L2 cache by utilizing one of $3$ Load/Store Units per Core. We assume access latency to read a cache line of size $64$ bytes requires $10$ clocks. Next, we assume that the Network Interface Controller (NIC) in the master and clients have data-direct I/O access to the L3 cache. Access latency to it is $40$ CPU cycles for $64$ byte cache line size. The effect of DRAM and caches can be ignored during inter-node communication but not during memory operation during gradient oracle computation.

\subsubsection{The optimization problem}

For modeling purposes, we consider solving a \modelname{linear regression} in the form:
\begin{eqnarray*}
	f(x) \eqdef \dfrac{1}{n} \sum_{i=1}^{n} f_i(x),\\
	f_i(x)=\dfrac{1}{n_i} \norm{\mA_i x - b_i}^2.
\end{eqnarray*}

For this problem the value of $\nabla f_i(x)=\dfrac{2}{n_i} \mA_i^\top(\mA_i x - b_i)$. In implementing gradient oracles, we utilize dense matrix and matrix-vector operations. To add two vectors, we execute $d$ scalar additions and $2d$ memory access operations for reading. In modern computing, hardware loads are more expensive because writes essentially can be queued. The need time of inner product operation of two vectors of dimension $d$ is equal to $$(d-1) \cdot \mathrm{add_{cost}} + d\cdot \mathrm{mult_{cost}} + 2d \cdot \mathrm{memaccess_{cost}}.$$

To estimate computing time for the Matrix-Vector and Matrix-Matrix Operations, we have assumed that their calculation is a sequence of inner products. In reality, not all data may fit into the Cache. If we go one step further with modeling, then Cache misses effects should be modeled as well.

\subsubsection{Implementation benefits of DCGD/PermK}

We will leverage several flexibility aspects of \algname{DCGD/PermK} in our experiment.

\begin{enumerate}
	\item The \compname{PermK} operator compressor behaviorism is independent of the input. This means that clients can use a prior sample to need coordinates for sparsification and increase the speed of $\nabla f_i(x)$ in clients.
	\item Next, we will exploit the fact that after the master receives the message, it can immediately broadcast it to all clients, regardless of whether the client has finished the work for the current round. This is possible in the context of using \algnamewithaes{DCGD/PermK/AES}, but it is impossible when using \ecryptname{CKKS}.
	\item If during the {FL} process, there is a slow client with a slow CPU or with a big amount of samples, then \algname{DCGD/PermK} allows other clients to start several operations which are impossible for \algname{GD}:
	\begin{itemize}
		\item [i)] obtain (partial) results from the master by using a communication network for current round $\left[\nabla f(x^{k})\right]_{part}$,
		\item [ii)] apply partial update for current $x^k$ and obtain partially new $x_{part}^{k+1}$,
		\item [iii)] and start performing partial computations $\left[\nabla f(x_{part}^{k+1})\right]_{part}$.
	\end{itemize}
\end{enumerate}

\subsubsection{Scheduling of clients and master computation and communication with Critical Path Method}

We modeled \algname{GD} and \algname{DCGD} Algorithms with unrolling $r=4$ rounds of computation and communication over $n=4$ clients and $1$ master. We represented the optimization process as the directed graph of different elementary tasks.

Task dependencies are represented in the form of weighted, orientated, directed, acyclic graph:
$$G=(V, E, W), \qquad E=V \times V, \qquad W: {E} \to \mathbb{R}_{+}.$$ 

The graph is constructed following the next rules:

\begin{enumerate}
	\item Tasks are represented as vertices in graph $G$.
	\item Task $s$ is connected to task $e$ with weight $0$, if and only if task $e$ can not start before task $s$.
	\item If the task $e$ obtains as input from task $s$, then the task $s$ is connected to task $e$ with the weight of the duration of work (in seconds) that $s$ should perform to produce input for task $e$.
\end{enumerate}

The algorithm for the scheduling process involves the following steps:

\begin{enumerate}
	\item Introduce fake source and sink vertices.
	\item Add zero weight directed edge from a fake source vertex to all $v \in V$.
	\item From all original vertices $v \in V$ add zero weight edge to fake sink vertex.
	\item Add fake source and sink to set $V$.
	\item Compute topological order of $G$ starting from fake source $s$.
	\item Compute the longest path in the directed acyclic graph of tasks. It is achieved via temporally changing weights of $G=(V, E)$ to negative, computing the topological order of $G$, and relaxing all vertices in topological order.
	\item The longest path to all vertices $v \in V$ from fake source $s$ creates a schedule for executing operation $v$.
\end{enumerate} 

Essentially, this is a critical path method (CPM) to solve the parallel precedence-constrained scheduling problem. The running time of this algorithm is $\mathcal{O}\left(V+E\right)$. 

\paragraph{Correctness proof.}

Let us examine the longest path $s {\leadsto} v$. The previously scheduled jobs before $v$ are vertices $x \in V$, such that $x {\leadsto} v$. The construction of the longest path implies that $w(s {\leadsto} v) \ge w(s {\leadsto} x)+w(x {\leadsto} v) \ge w(s {\leadsto} x), \forall s {\leadsto} x$.

From this, we observe that the start times obtained by the longest paths are feasible because jobs $x$ that need to be executed before $v$ will be scheduled before $v$. Furthermore, the length of any path $s {\leadsto} v$ is a lower bound on the actual time to start $v$ because $v$ cannot begin earlier than previous tasks due to dependency constraints. This proves that the longest path from $s \in V$ to $v \in V$ determines the start time for task $v$.

Once task $v$ can begin execution in the timeline, it can potentially activate the execution of all tasks $(v, z) \in E$, where $z \in adj(v)$ and $w(v, z) > 0$.

\subsubsection{Iterative refinement of Critical Path Method}

Parallel precedence-constrained scheduling using the CPM method determines the execution schedule for a graph of jobs. However, there may be cases where the duration of a task (which is the input for the CPM Algorithm) is defined by the number of other tasks during specific time intervals. This creates a circular dependency between the input and output of the CPM Algorithm. In our scenarios, we encounter this situation due to the following reasons:

\begin{enumerate}
	\item If clients make a partial computation and the CPU is not busy with other work, parallelizable operations (such as Matrix-Vector multiply) can be parallelized across several CPU cores. Consequently, it increases clients' computational throughput for this partial update.
	\item If clients share the same bottleneck link to the master and some are still busy with compute $\nabla f_i(x^k)$, then other clients can transmit data at a fast bandwidth because the bottleneck link is shared across a smaller number of clients. This observation leads to a situation in which effective bandwidth can be increased.
\end{enumerate}

\begin{figure*}[ht!]
	\centering
	\captionsetup[subfigure]{labelformat=empty}
	
	\begin{subfigure}[ht]{0.49\textwidth}
		\includegraphics[width=\textwidth]{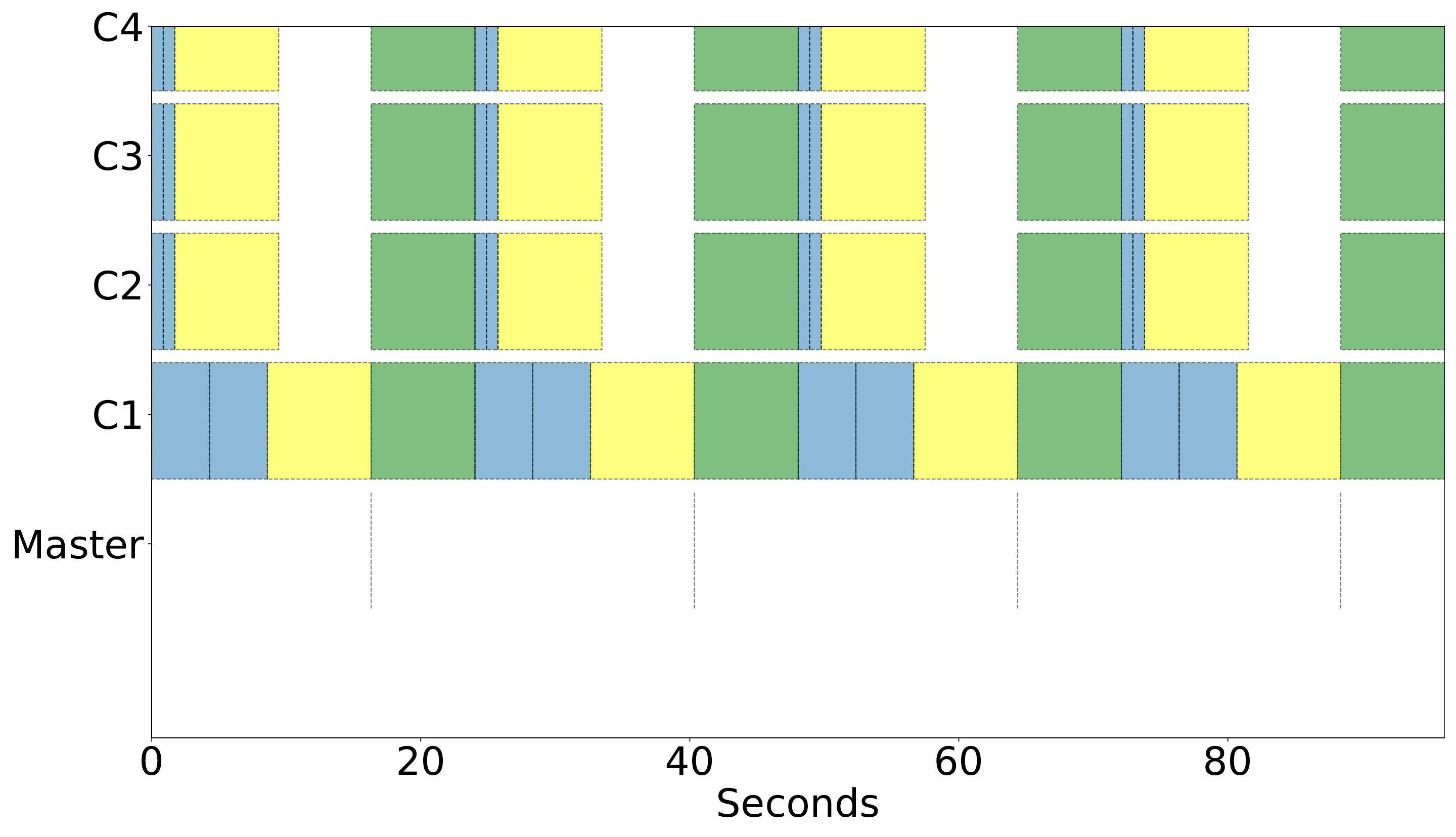} 
		\caption{{(a) \algname{GD}: $96.11$s.}}
	\end{subfigure}
	\begin{subfigure}[ht]{0.49\textwidth}
		\includegraphics[width=\textwidth]{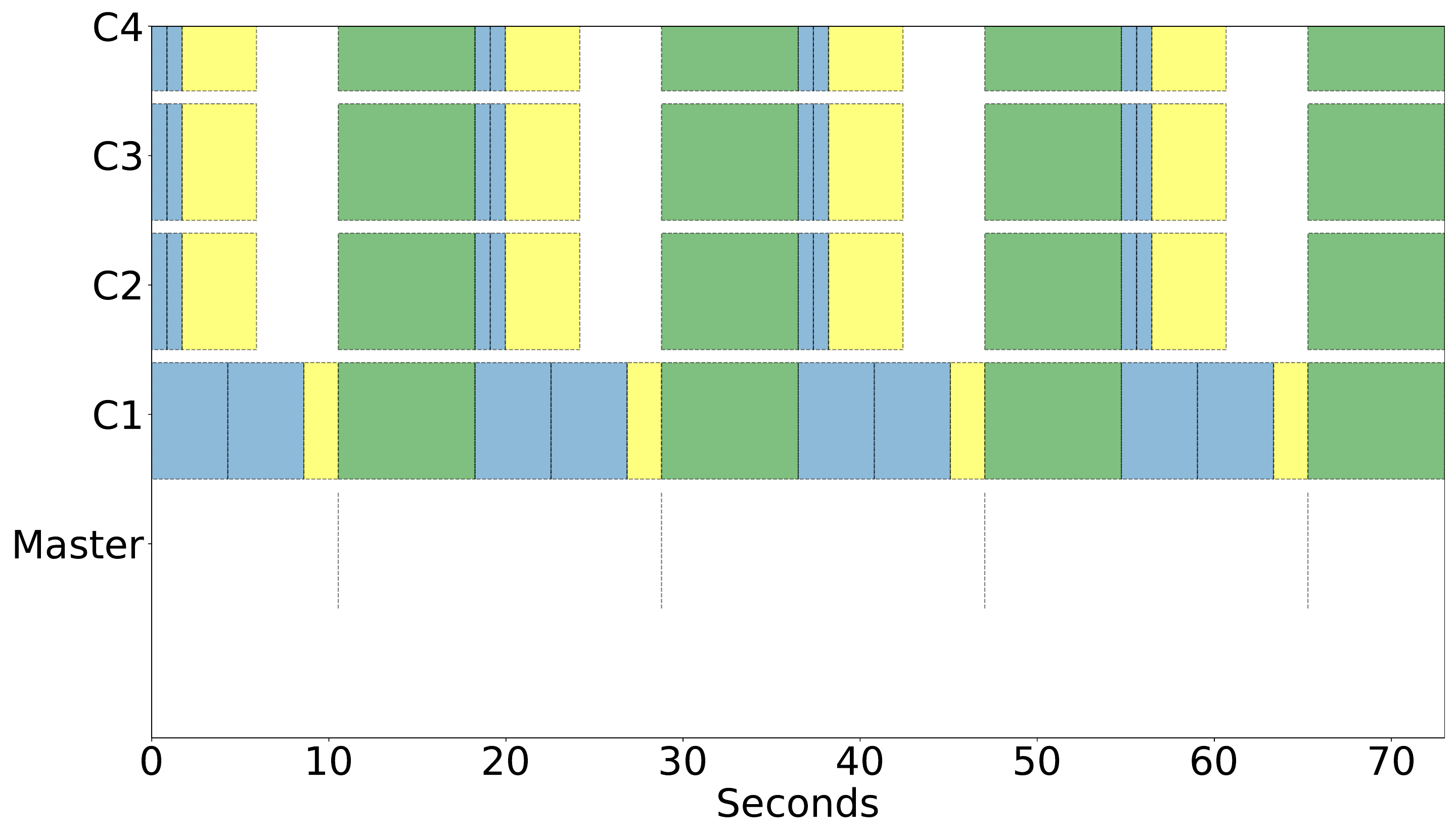} 
		\caption{{(b) Refined \algname{GD}: $73.00$s.}}
	\end{subfigure}
	
	\begin{subfigure}[ht]{0.49\textwidth}
		\includegraphics[width=\textwidth]{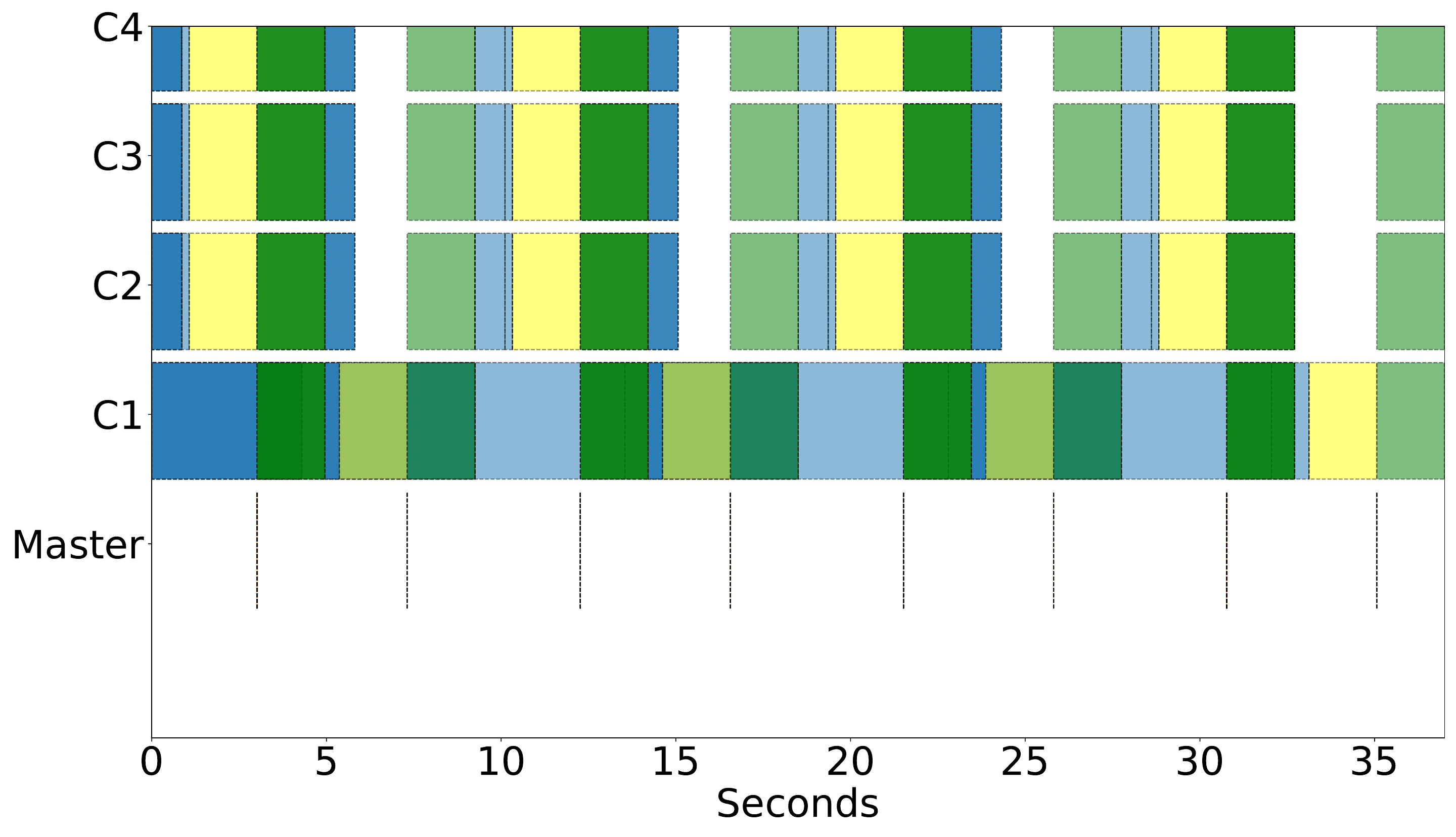} 
		\caption{{(c) \algnamewithaes{DCGD/PermK/AES}: $37.003$ s.}}
	\end{subfigure}
	\begin{subfigure}[ht]{0.49\textwidth}
		\includegraphics[width=\textwidth]{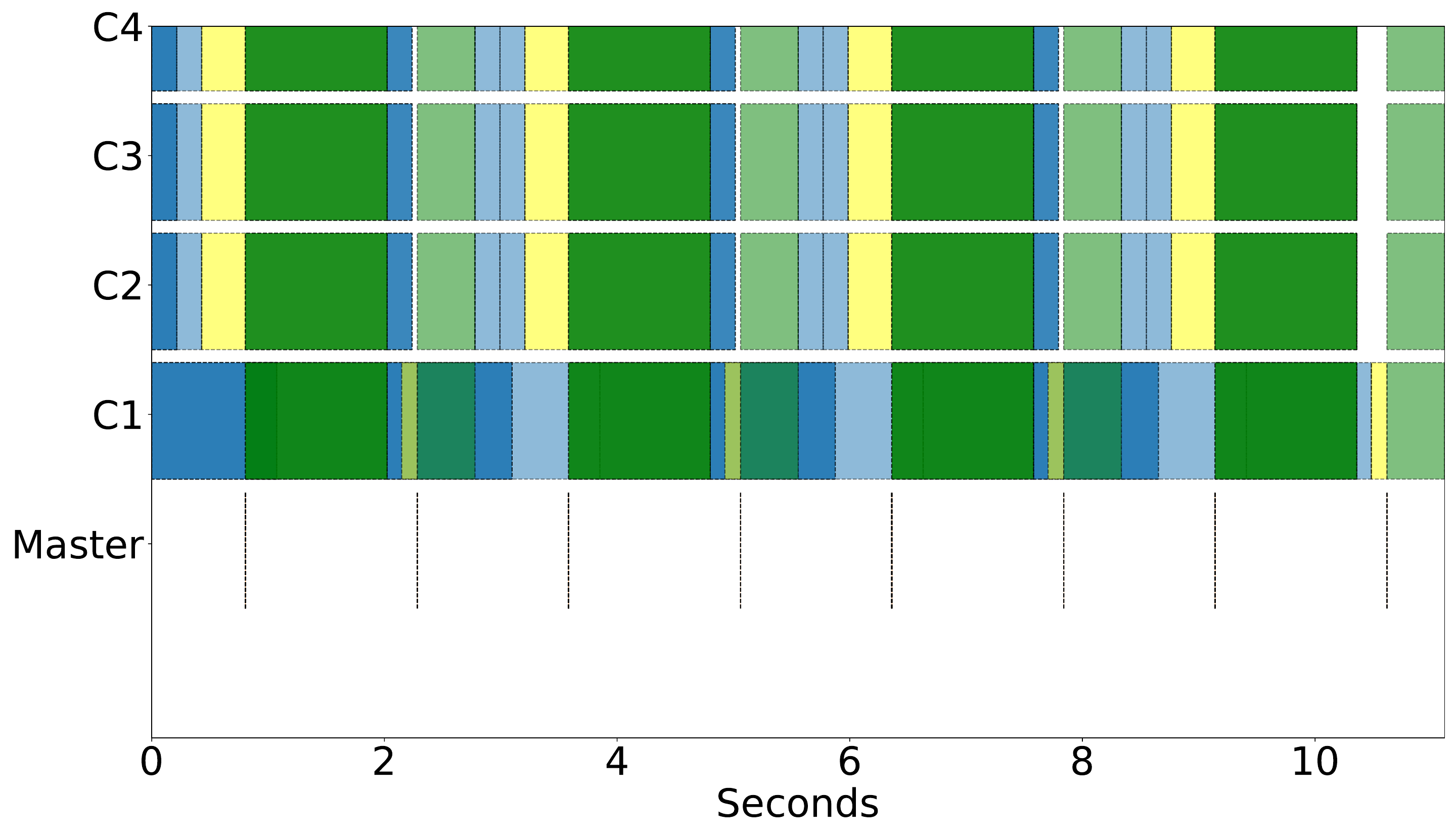} 
		\caption{{(d) Refined \algnamewithaes{DCGD/PermK/AES}: $11.113$s.}}
	\end{subfigure}
	
	\caption{
		Event-based modeling for training \modelname{linear regression} across $n=4$ clients, $d=10\cdot 10^6$, $n_1=55000, n_2=n_3=n_4=11000$ during $4$ rounds. The uplink and downlink bandwidth is $41.54$ MBps, latency $28$ms, and computational throughput of modeled CPUs is $238.41$ GFLOPS. Legend:\\
		\phantom{x} \colorbox{blue}{\phantom{x}} - Computation and local memory access using all available CPU cores, \\
		\phantom{x} \colorbox{yellow}{\phantom{x}} - Client to master communication (Clients share the same bottleneck link),\\ \phantom{x} \colorbox{green}{\phantom{x}} - Master to client Communication (Clients share the same bottleneck link).
	}
	\label{ch4:fig:exp_syn_8}
\end{figure*}

\subsubsection{Results of refined scheduling for GD and DCGD/PermK/AES}

Scheduling results are presented in Figure~\ref{ch4:fig:exp_syn_8}. In our experiment, all clients have the same compute power equal to $238.41$ GFLOPS. Communication bandwidth and latency from all clients to the master and vice versa is the same, namely $B_i=41.54$ Mbps, and latency (or delay) for transmission is $28\cdot10^{-3}$ seconds. The assumption for computation is that each memory operation is carried out with an L2 cache. In case there is a need to perform read and write, we treat it as two memory operations. We assume all clients during communication to the master share the same bottleneck communication link. Also, we assume that the target CPU supports \aesname{AES}. The detailed generated execution plans can not be represented in this work due to their size. However, they can be found in the source code attached to the work. 

As illustrated in Figure~\ref{ch4:fig:exp_syn_8}, overlapping communication, computation, and the refined scheduling due to time-varying into shared communication bus and time-varying load into computing devices in clients leads to different speedups in execution plans (Figure~\ref{ch4:fig:exp_syn_8}, (b, d)) for \algname{GD} and \algname{DCGD/PermK}. The total computation speedup per round when compared to refined \algname{GD} is $6.578$. We hope this simulation will prove valuable for those seeking to adopt \algnamewithaes{DCGD/PermK/AES} to align with hardware requirements closely.




\clearpage
\addtocounter{adjsection}{1}
\section{Flexibility in Training Deep Learning Models}

\label{ch4:app:flexibility_for_dl_training}

\subsection{Introduction to backpropagation}
In Appendix \ref{ch4:app:simulation_experiment} we have explored that \algname{DCGD/PermK} exhibits practical flexibility for training linear models in detail. Described communication and computation overlap can be important not only for linear models but also for {DL} models. 

The evaluation of the gradient of the score function in modern computational frameworks such as \libname{TensorFlow} \citep{abadi2016tensorflow} or \libname{PyTorch} \citep{paszke2019pytorch} for complex computational graph is automated via leveraging algorithms for the numeric evaluation of derivatives. In most cases, this is achieved using \textit{Automatic Differentiation (AD) in Reverse Accumulation mode} \citep{linnainmaa1970representation} named as a \textit{Backpropagation Algorithm} \citep{rumelhart1986learning} in {ML} literature. A composed loss function for DL models typically has the following structure: 
\begin{eqnarray}
	\label{ch4:eq:loss_for_dl}
	f_i(x) = \sum_{j=1}^{n_i} \mathcal{L}(b_{real,j}, \textcolor{red}{g_1}([x]_{Q_1}, a_j, \textcolor{red}{g_2}([x]_{Q_2}, a_j, \dots))) + R(x)\\
	Q_1 \cup Q_2 \dots = [d], Q_i \cap Q_j \dots = 0, \forall i,j. \notag
\end{eqnarray}

The score function $f_i(x)$ is represented by a computation graph and has a nested structure where prediction is driven by a nested composition of functions $\textcolor{red}{g_i}$ and the predicted label is scored with a true label for training input-output pair $(a_{j}, b_{real,j})$ with a score loss function $\mathcal{L}$. Each function $\textcolor{red}{g_i}$ has the following source of inputs: output from the prevision function (or layer) $g_{i-1}$, and input (trainable) scalar parameters $x_p$ where $p \in \{Q_1, Q_2, \dots\}$, and $Q_i \subseteq [d]$.

Assume that $f_i(x)$ is differentiable in the points where derivatives according to the chain rule are evaluated. In this case, the needed partial derivatives $$[\nabla f_i(x)]_k = \dfrac{\partial f_i}{\partial x_k}$$
can be computed with chaining Jacobians calculated with intermediate variables. The \textit{Backpropagation Algorithm} allows for fixed input samples $D_i={(a_1,b_1), \dots, (a_{n_i},b_{n_i})}$ stored in client number $i$ and for fixed computational graph $$f_i(x; D_i)$$
compute the required gradient $\nabla f_i(x)$ in two \textit{passes} or two \textit{phases}:
\begin{enumerate}
	\item \textit{Forward Pass}. Compute all intermediate variables, namely values of $g_i()$ for every training sample, for the whole compute graph. All these variables are stored in memory for the next pass.
	\item \textit{Backward Pass}. Compute the intermediate Jacobians and produce final partial derivatives of a full gradient. 
\end{enumerate}

There is no need to store all intermediate Jacobians simultaneously. They are stored implicitly in special variables typically denoted as $\delta$ in the implementation of {backpropagation}. However, there is a need to store all intermediate outputs explicitly (named as activation in {ML} literature) after \textit{Forward Pass}. 

\subsection{Examples of parallelization inside backpropagation}
\label{ch4:app:parallforbackprop}

Computationally {backpropagation} is often a preferred strategy when the entire gradient needs to be computed. The compute scheduling and parallelism strategies for effective computation of $\nabla f_i(x)$ is an active line of research in DL system literature \citep{jia2019beyond, krizhevsky2014one}. The standard strategies for performing parallelism inside computation of $\nabla f_i(x)$ involve the following strategies:

\begin{enumerate}
	\item \textbf{Data parallelism.} Assign train samples $(a_j, b_j)$ to different computation devices in client $i$.
	\item \textbf{Model parallelism.} Assign different functions $g_i$ to different computation devices available in client $i$.
	\item \textbf{Attributes parallelism.} Split the image of functions $g_{i-1}$ into parts and process it by parts if function $g_{i}$ allows to do it.
	\item \textbf{Parameters parallelism.} Partition trainable variables for function $g_i(x)$, into smaller chunks $[x]_{Q_i}$. After partitioning $[x]_{Q_i}$ if $g_i(x)$ can be computed in parallel, then compute it in parallel in different devices.
\end{enumerate}

\paragraph{Remark on terminology.} Some researchers adopt this terminology, while others use alternative terms. In fact, the object from Equation~\eqref{ch4:eq:loss_for_dl} is already sufficiently complex and remains an active area of research. Therefore, constructing a universal terminology is likely to be challenging.	

\subsection{Research opportunities for parallelization in backpropagation from DCGD/PermK}

\label{ch4:app:flexibility_for_dl_training_from_permk}

Assume that clients know the current iterate $x^{k}$, however, if there is even one straggler that still did not send ${g_s}^k$ to the Master, then no clients can proceed in training. Master has to wait for gradient estimator ${g_s}^k$ from straggler, and what can be done for \algname{DCGD/RandK} is at least challenging in these circumstances. 

However, \algname{DCGD/PermK} exhibits useful properties that can be partially helpful in dealing with this situation, which we previously discussed in Appendix \ref{ch4:app:simulation_experiment}:

\begin{itemize}
	\item Clients can obtain (partial) results from the master $\left[\nabla f(x^{k})\right]_{part}$.
	\item Clients can apply partial update for $x^k$ and obtain partially new model $x_{part}^{k+1}$.
\end{itemize}

It depends on the specific situation, but during the waiting of a straggler, the forward potentially can be started using the available $x_{part}^{k+1}$ (\textit{forward pass} in practice takes at most $50\%$ of $\nabla f_i(x)$ computation). Assume that we have statistical information about stragglers, then potentially this information can be provided to DL schedulers and parallelization strategies described in the previous section Appendix~\ref{ch4:app:parallforbackprop} to optimize $\nabla f_i(x^{k+1})$. 

This opens new opportunities to refine parallelization strategies for training in general, and \textit{{parameter parallelism}} in particular, due to the relaxed synchronization requirements of \algname{DCGD/PermK}.

\clearpage
\addtocounter{adjsection}{1}
\section{Flexibility of DCGD/PermK Across Communication Topologies}
\label{ch4:app:comm_networks}

In High-Performance Computation and Network Communication, the \textit{physical network topologies} describe the physical arrangement of the devices and routing devices for organizing communication. Next, we will explore what benefits and flexibility \algname{DCGD/PermK} can bring if this algorithm runs in the mentioned popular physical network topologies.

\paragraph{Potential benefits in point-to-point topology}
In this type of topology, some clients are directly connected with a single link pairwise. It is possible to instantiate \algname{DCGD/PermK} so that clients do not coordinate during runtime for aggregation computation. It is possible because there is no need to perform reduction in the sense of averaging $\nabla f_i(x)$. If some pair clients are connected with the fast link, this pair of clients can utilize this channel and deliver $\left[\nabla f(x_{part}^{k+1})\right]_{part}$ for his neighbor with a fast point-to-point link without obtaining it from a master.

\paragraph{Potential benefits in bus topology}
A single cable or bus connects all the nodes in this type of topology. In this type of topology, if some clients have slow computing, it removes them from competition for a shared bus and decreases communication contention. In a bus topology, the benefit of \algname{DCGD/PermK} is that messages from each client can be effectively broadcast. In fact, \algname{DCGD/PermK} can be implemented without a centralized server, and a shared bus is enough for organized communication. Employing \aesname{AES} based encryption guarantees protection from eavesdropping on the shared bus.

\paragraph{Potential benefits in star topology}
In this type, all the nodes are connected to a central master device with individual links. In the case of using \algname{DCGD/PermK}, the master in this topology only plays the role of communication hub. Interestingly, while obtaining parts of the gradients from clients, computation is unnecessary, and synchronization is far more relaxed, as we have observed in Appendix~\ref{ch4:app:simulation_experiment}. While using \algname{DCGD/PermK}, the server can potentially utilize communication links with Full Duplex mode. Examples of Full Duplex communication buses include PCI-Express, InfiniBand\footnote{\url{https://network.nvidia.com/pdf/whitepapers/HPC_Clustering_130.pdf}}, and some forms of Ethernet (see \citep{spurgeon2000ethernet} Table 4.1 in Chapter 4). In our setting, we consider the situation when clients are connected to the Internet, and as it has been described in Appendix~\ref{ch4:app:simulation_experiment}, PCI-Express is not a bottleneck. However, in data centers, PCI-Express can also represent a bottleneck \citep{li2019priority}. With \algname{DCGD/PermK}, during some period, a single link can be used simultaneously to obtain information from clients and deliver information to the clients. This is not the case for \algname{DCGD/RandK} or \algname{GD} because they require explicit synchronization. But it is the case for \algname{DCGD/PermK}. This doubles the maximum bandwidth during master client communication and decreases latency by a factor of two.

\paragraph{Potential benefits in ring topology}
This type of topology connects all the nodes circularly. The ring topology is sometimes preferable because it reduces the number of \abr{NIC} and cables required for connecting multiple devices. Specifically, each device only needs one \abr{NIC} to participate in all-reduce. Also, ring topology avoids congestion and collisions at the central hub as messages are distributed evenly along the ring. The time delay during performing aggregation for \algname{GD} or \algname{DCGD/RandK} in a ring topology represents a sum of delays along the circle and is equal to $$\tau_{\mathrm{ring\,delay}} = \sum_{i=1}^{n} \dfrac{RTT_i}{2} + \dfrac{d}{B_i} \cdot bpp.$$
Fundamentally, this is because when using \algname{GD} or \algname{DCGD/RandK}, clients cannot begin computation for the next iteration until they have computed the entire gradient. For \algname{DCGD/PermK} clients, after obtaining $\left[\nabla f(x_{part}^{k})\right]_{part}$ from the neighbor, can partially start computation for the next iteration.

\paragraph{Potential benefits in mesh topology}
This topology connects every node to every other node with dedicated links. It has high bandwidth and fault tolerance, but requires a lot of cables and ports and is complex to install. In this form of topology, \algname{DCGD/PermK} is the most natural choice because underlying communication topology naturally maps to communication pattern in \algname{DCGD/PermK}, which can observed if view Algorithm~\ref{ch4:alg:dcgd_permk_aes} as a series of broadcast operations to reconstitute the global direction for optimization step.

\paragraph{Potential benefits in tree topology}
This topology combines multiple star topologies into a hierarchical structure with a root node. In this configuration, broadcasting can be implemented very efficiently. One potential benefit of \algname{DCGD/PermK} is that it can operate without a central server; instead, each client broadcasts information to other clients in Line 6 of Algorithm \ref{ch4:alg:dcgd_permk_aes}.

\addtocounter{adjsection}{1}
\section{Reproducibility}

To ensure reproducibility, we use the following \fl simulator \citep{burlachenko2021fl_pytorch} features: random seeds were fixed for data reshuffling, and random seeds were fixed for the runtime pseudo-random generators involved in variants of \algname{DCGD} for randomized compressors.

The source code of our experiments is a part of our submission to the \textit{4th International Workshop on Distributed Machine Learning, co-located with CoNEXT 2023}. The source code, along with the implementation and documentation, is publicly available at the following link:

\begin{center} 
	\href{https://github.com/burlachenkok/fl-with-nonhe}{https://github.com/burlachenkok/fl-with-nonhe}
\end{center}

\unappendix

\chapter{PAGE Extensions: Refining the Optimal Single-Node SGD Optimizer}
\label{chapter5}

The goals and summaries of this chapter are outlined in Table \ref{ch1:tbl:algorithms} and Section~\ref{ch1:sec:overview-5}.

\section{Introduction}
In this work, we consider the minimization of the average of $n$ smooth functions \eqref{ch5:eq:main_problem} in the non-convex setting of the regime when the number of functions $n$ is very large. In this regime, calculation of the exact gradient can be infeasible and the classical gradient descent method (\algname{GD}) \citep{nesterov2018lectures} can not be applied. The structure of the problem is generic, and such problems arise in many applications, including Machine Learning \citep{bishop2006pattern} and Computer Vision \citep{goodfellow2016deep}. Problems of this form are the basis of empirical risk minimization (ERM), which is the prevalent paradigm for training supervised machine learning models.

\subsection{Finite-sum optimization in the smooth non-convex regime}
We consider the finite-sum optimization problem
\begin{align}
	\label{ch5:eq:main_problem} 
	  \min \limits_{x \in \R^d}\left\{f(x) \eqdef \dfrac{1}{n}\sum \limits_{i=1}^n f_i(x)\right\},
\end{align}
where $f_i\,:\,\R^d \rightarrow \R$ is a smooth (and possibly non-convex) function for all $i \in [n] \eqdef \{1, \dots, n\}.$ We are interested in randomized algorithms that find an $\varepsilon$-stationary point of \eqref{ch5:eq:main_problem} by returning a random point $\widehat{x}$ such that $$\Exp{\norm{\nabla f(\widehat{x})}^2} \leq \varepsilon.$$ The main efficiency metric of gradient-based algorithms for finding such a point is the (expected) number of gradient computations $\nabla f_i$; we will refer to it as the \textit{complexity} of an algorithm. 

\subsection{Related work}

The area of algorithmic research devoted to designing methods for solving the Problem~\eqref{ch5:eq:main_problem} in the smooth non-convex regime is one of the most highly developed and most competitive in optimization.

{\bf The path to optimality.} Let us provide a lightning-speed overview of recent progress. 
The complexity of \algname{GD} for solving \eqref{ch5:eq:main_problem} is $\cO\left(n\varepsilon^{-1}\right)$, but this was subsequently improved by more elaborate stochastic methods, including \algname{SAGA}, \algname{SVRG} and \algname{SCSG} \citep{SAGA, johnson2013accelerating, lei2017non, horvath2019nonconvex}, which enjoy the better complexity $\cO\left(n + n^{2/3} \varepsilon^{-1}\right)$. Further progress was obtained by methods such as \algname{SNVRG} and \algname{Geom-SARAH} \citep{zhou2018stochastic, horvath2020adaptivity},  improving the complexity to $\widetilde{\cO}\left(n + n^{1/2}\varepsilon^{-1}\right)$. Finally, the methods \algname{SPIDER}, \algname{SpiderBoost}, \algname{SARAH} and \algname{PAGE} \citep{fang2018spider,wang2019spiderboost,SARAH,li2021page}, among others, shaved-off certain logarithmic factors and obtained the {\em optimal} complexity $\cO\left(n + n^{1/2}\varepsilon^{-1}\right)$, matching lower bounds \citep{li2021page}.

{\bf Optimal, but hiding a secret.} While it may look like this is the end of the road, the starting point of our work is the following observation.
\begin{center}
{\em The big-$\cO$ notation in the above results hides important and typically very large data-dependent constants.}
\end{center}
For instance, it is rarely noted that the more precise complexity of \algname{GD} is $\cO\left(L_{-} n\varepsilon^{-1}\right)$ and where $L_{-}$ is L-smooth constant of function $f(x)$ (see Assumption~\ref{ch5:ass:lipschitz_constant}), while the complexity of the optimal methods, for instance \algname{PAGE}, is $\cO\left(n + (L_{-}+L_{+}n^{1/2})\varepsilon^{-1}\right),$ where $L_{-}\leq L_{+}$ are {\em different} and often {\em very large} smoothness constants. Moreover, it is easy to generate examples of problems (see Example~\ref{ch5:ex:lipt}) in which the ratio $L_{+}/ L_{-}$ is {\em as large as one desires}. 

{\bf Client and data sampling in Federated Learning.} Furthermore, several modern applications, notably Federated Learning~\citep{FEDLEARN,mcmahan17fedavg}, depend on elaborate {\em client} and {\em data sampling} mechanisms, which are not properly understood. However, optimal \algname{SGD} methods were considered in combination with very simple mechanisms only, such as sampling a random function $f_i$ several times independently with replacement \citep{li2021page}. We thus believe that an in-depth study of sampling mechanisms for optimal methods will be of interest to the Federated Learning community. 

There exists prior work on analyzing non-optimal \algname{SGD} variants with flexible mechanisms.
For example, using the ``arbitrary sampling'' paradigm, originally proposed by \citet{richtarik2016parallel} in the study of randomized coordinate descent methods, \citet{horvath2019nonconvex} and \citet{qian2021svrg} analyzed \algname{SVRG}, \algname{SAGA}, and \algname{SARAH} methods, and showed that it is possible to improve the dependence of these methods on the smoothness constants via carefully crafted sampling strategies. Further, \citet{zhao2014accelerating} investigated the stratified sampling, but only  provided the analysis for vanilla \algname{SGD}, and in the convex case.

{\bf Our goal.} Motivated by the above considerations, in this chapter, we propose a {\em general framework} for the study of a broad family of sampling mechanisms for and provide a {\em refined analysis} of the optimal method \algname{PAGE} for solving problems \eqref{ch5:eq:main_problem} in the smooth non-convex regime. 

\subsection{Assumptions}
\label{ch5:sec:assumptions}
In our analysis, we rely on the following weak assumptions on the functions $f_i.$ 
\begin{assumption}
	\label{ch5:ass:lower_bound}
	There exists $f^* \in \R$ such that $f(x) \geq f^*$ for all $x \in \R^d$.
\end{assumption}
\begin{assumption}
	\label{ch5:ass:lipschitz_constant}
	There exists $L_{-}\geq 0$ such that $\norm{\nabla f(x) - \nabla f(y)} \leq L_{-} \norm{x - y}$ for all $x, y \in \R^d.$
\end{assumption}
\begin{assumption}
	\label{ch5:ass:local_lipschitz_constant}
	For all $i \in [n],$ there exists constants $L_{i}>0$ such that $\norm{\nabla f_i(x) - \nabla f_i(y)} \leq L_{i} \norm{x - y}$ for all $x, y \in \R^d.$
\end{assumption}

\subsection{Contributions}

The summary of our contributions is as follows.

\begin{enumerate}

\item In the original paper \citet{li2021page}, the optimal (with respect to $n, \varepsilon$) optimization method \algname{PAGE} was analyzed with a simple uniform mini-batch sampling with replacement. We analyze \algname{PAGE} with virtually any (unbiased) sampling mechanism using a novel Assumption~\ref{ch5:ass:sampling}. Moreover, we show that some samplings can improve the convergence of \algname{PAGE} (see Table~\ref{ch5:tbl:compl}).

\item We improve the analysis of \algname{PAGE} using a new quantity, the weighted Hessian Variance $L_{\pm}$ (or $L_{\pm,w}$), that is well-defined if the functions $f_i$ are $L_i$--smooth. We show that, when the functions $f_i$ are ``similar'' in the sense of the weighted Hessian Variance, \algname{PAGE} enjoys faster convergence rates (see Table~\ref{ch5:tbl:compl}). Also, unlike \citep{szlendak2021permutation}, we introduce weights $w_i$ that can play a crucial role in some samplings. Moreover, the experiments in Section~\ref{ch5:sec:experiments} agree with our theoretical results.

\item Our framework is flexible and can be generalized to \textit{the composition of samplings}. These samplings naturally emerge in Federated Learning \citep{FEDLEARN, mcmahan17fedavg}, and we show that our framework can be helpful in the analysis of problems from Federated Learning.

\end{enumerate}

\renewcommand{\arraystretch}{1.5}
\begin{table}
	\centering
	\footnotesize
	\caption{The constants $A$, $B,$ $w_i$ and $|\samplefunc|$ that characterize the samplings in Assumption~\ref{ch5:ass:sampling}.}
	\label{ch5:tbl:AB}
	\begin{threeparttable}
		\begin{tabular}{|cccccc|}
			\hline
			\makecell[c]{\textbf{Sampling}\\ \textbf{Scheme}} & \textbf{$A$} &\textbf{$w_i$}& \textbf{$B$} & \textbf{$|\samplefunc|$} & \makecell[c]{\textbf{Reference}\\ \textbf{Section}}  \\
			\hline
			\hline
			\makecell{\samplingname{\scriptsize Uniform} \\ \samplingname{\scriptsize With Replacement}} & $\nicefrac{1}{\tau}$& $\nicefrac{1}{n}$& $\nicefrac{1}{\tau}$ & $\leq \tau$ &\ref{ch5:subsec:3}\\
			\hline			
			\samplingname{\scriptsize Importance} & $\nicefrac{1}{\tau}$& $q_i$& $\nicefrac{1}{\tau}$ & $\leq \tau$ &\ref{ch5:subsec:3} \\
			\hline			
			\samplingname{\scriptsize Nice} &$\dfrac{n-\tau}{\tau(n-1)}$ &$\nicefrac{1}{n}$ &$\dfrac{n-\tau}{\tau(n-1)}$ & $\tau$ &\ref{ch5:subsec:0} \\
			\hline			
			\samplingname{\scriptsize Independent} &$\dfrac{1}{\sum_{i=1}^n\dfrac{p_i}{1-p_i}}$ & $\dfrac{\dfrac{p_i}{1-p_i}}{\sum_{i=1}^n\dfrac{p_i}{1-p_i}}$& $0$& $\sum_{i=1}^n p_i$ &\ref{ch5:subsec:2} \\
			\hline
			\samplingname{\scriptsize Extended Nice} & $\dfrac{n-\tau}{\tau(n-1)}$ &$\dfrac{l_i}{\sum_{i=1}^n l_i}$ &$\dfrac{n-\tau}{\tau(n-1)}$ & $\leq \tau$ &\ref{ch5:subsec:1} \\
			\hline
		\end{tabular}
		\scriptsize Notation: $n = \#$ of data points; $\tau = $ batch size; $q_i = $ probability to sample $i$\textsuperscript{th} data point in the multinomial distribution; $p_i = $ probability to sample $i$\textsuperscript{th} data point in the Bernoulli distribution; $l_i = \#$ of times to repeat $i$\textsuperscript{th} data point before apply the \samplingname{Nice} sampling.
	\end{threeparttable}
\end{table}

{\section{Tight Variance Control of General Sampling Estimators} In Algorithm~\ref{ch5:alg:page} (a generalization of \algname{PAGE}), we form an estimator of the gradient $\nabla f$ via subsampling. In our search for achieving the combined goal of providing a {\em general} (in terms of the range of sampling techniques we cater for) and {\em refined} (in terms of the sharpness of our results, even when compared to known results using the {\em same} sampling technique) analysis of \algname{PAGE}, we have identified several  powerful tools, the first of which is Assumption~\ref{ch5:ass:sampling}.
	
	Let $\mathcal{S}^n \eqdef \left\{(w_1, \dots, w_n) \in \R^n\,|\, w_1, \dots, w_n \geq 0,\,\sum_{i=1}^n w_i = 1\right\}$ be the \textit{standard simplex} and tuple $(\Omega, \mathcal{F},\mathbf{P})$ be a probability space.
	
	\begin{assumption}[Weighted $AB$ Inequality]
		\label{ch5:ass:sampling}
		Consider the random mapping $$\samplefunc\,:\, \R^d \times \dots \times \R^d \times\,\,\Omega \rightarrow \R^d,$$ which we will call ``sampling'', such that for all $a_1, \dots, a_n \in \R^d$ 
		$$\Exp{\samplefunc(a_1, \dots, a_n;\omega)} = \dfrac{1}{n} \sum_{i=1}^n a_i.$$
		
		Assume that there exist $A, B \geq 0$ and weights $(w_1, \dots, w_n) \in \mathcal{S}^n$ such that for all $a_1,\dots,a_n\in \R^d$ the following holds
		\begin{eqnarray}
			\label{ch5:eq:sampling}
			   \Exp{\norm{\samplefunc(a_1, \dots, a_n;\omega) - \dfrac{1}{n}\sum \limits_{i = 1}^n a_i}^2} \leq \dfrac{A}{n}\sum \limits_{i = 1}^n \dfrac{1}{n w_i}\norm{a_i}^2 - B \norm{\dfrac{1}
				{n}\sum \limits_{i = 1}^n a_i}^2.
		\end{eqnarray}
	\end{assumption}
	
	For simplicity, we denote $$\samplefunc\left(\{a_i\}_{i=1}^n\right) \eqdef \samplefunc(a_1, \dots, a_n) \eqdef \samplefunc(a_1, \dots, a_n;\omega).$$ Further, the collection  of samplings satisfying Assumption~\ref{ch5:ass:sampling} will be denotes as $\mathbb{S}(A, B, \{w_i\}_{i=1}^n).$ The main purpose of a sampling $\samplefunc \in \mathbb{S}(A, B, \{w_i\}_{i=1}^n)$ is to estimate the mean $\dfrac{1}{n} \sum_{i=1}^n a_i$ using some random subsets (possibly containing some elements more than once) of the set $\{a_1, \dots, a_n\}.$ 
	
	We now define the cardinality $|\samplefunc|$ of a sampling $\samplefunc \in \mathbb{S}(A, B, \{w_i\}_{i=1}^n)$.
	\begin{definition}[Cardinality of a Sampling]
		Let us take $\samplefunc \in \mathbb{S}(A, B, \{w_i\}_{i=1}^n),$ and define the function $\samplefunc_{\omega}(a_1, \dots, a_n)\,:\,\R^d \times \dots \times \R^d \rightarrow \R^d$ such that $$\samplefunc_{\omega}(a_1, \dots, a_n) \eqdef \samplefunc(a_1, \dots, a_n; \omega).$$ 
		If the function $\samplefunc_{\omega}(a_1, \dots, a_n)$ depends only on a subset $\mathcal{A}(\omega)$ of the arguments $(a_1, \dots, a_n),$ where $\mathcal{A}(\omega)\,:\,\Omega \rightarrow 2^{\{a_1, \dots, a_n\}},$ we define $|\samplefunc| \eqdef \Exp{|\mathcal{A}(\omega)|}.$
	\end{definition}
	
	Assumption~\ref{ch5:ass:sampling} is most closely related to two independent works: \citep{horvath2019nonconvex} and \citep{szlendak2021permutation}. \citet{horvath2019nonconvex} analyzed several non-optimal \algname{SGD} methods for ``arbitrary samplings''; these are random set-valued mappings $S$ with values being the subsets of $[n]$. The distribution of such a sampling is uniquely determined by assigning probabilities to all $2^{n}$ subsets  of $[n]$. In particular, they show that Assumption~\ref{ch5:ass:sampling} holds with 
	
	\begin{eqnarray*}
			\samplefunc(a_1, \dots, a_n) &=& \dfrac{1}{n} \sum_{i \in S} \dfrac{a_i}{p_i},\\
			p_i &\eqdef& \Prob(i \in S),\\
			|\samplefunc| &=& |S|,\\
			\mathrm{some\,} A &\geq& 0,\\
			w_1, \dots, w_n &\geq& 0,\\
			B &=& 0.
	\end{eqnarray*}
		
	Recently, \citet{szlendak2021permutation} studied a similar inequality, but in the context of communication-efficient distributed training with randomized gradient compression operators. They explicitly set out to study {\em correlated} compressors, and for this reason, introduced the second term in the right-hand side; in other words, they considered the possibility of $B$ being nonzero, as in this way they obtain a tighter inequality, which they can use in their analysis. However, their inequality only involves uniform weights $\{w_i\}$.
	Our Assumption~\ref{ch5:ass:sampling} offers the tightest known way to control the variance of the sampling estimator, and our analysis can take advantage of it. See Table~\ref{ch5:tbl:AB} for an overview of several samplings and the values $A,B$ and $\{w_i\}$ for which Assumption~\ref{ch5:ass:sampling} is satisfied.

	\newcommand{\degeneration}[1]{{\color{red}#1}}
	
	\iftrue
	\begin{table}
		\centering
		\footnotesize
		\caption{The complexity of methods and samplings from Table~\ref{ch5:tbl:AB} and Section~\ref{ch5:sec:sampling_in_sampling}.}
		\label{ch5:tbl:compl}
		\begin{threeparttable}
			\begin{tabular}{|ccc|}
				\hline
				\textbf{Sampling Scheme} & \textbf{Complexity} & \textbf{Comment}\\
				\hline
				\hline
				\makecell{\samplingname{\scriptsize Independent} \\ \scriptsize \citep{horvath2019nonconvex}}
				 & \scriptsize $\Theta\left(n + \dfrac{\degeneration{n^{2/3}} \left(\dfrac{1}{n} \sum_{i=1}^n L_i\right)}{\varepsilon}\right)$ & \scriptsize \shortstack{\algname{\scriptsize SVRG} method \\ $p_i \propto L_i$}\\
				\hline
				\makecell{\samplingname{\scriptsize Uniform-With-Replacement} \\ \scriptsize \citep{li2021page}} & \scriptsize $\Theta\left(n + \dfrac{\sqrt{n} \degeneration{L_{+}}}{\varepsilon}\right)$ & ---\\
				\hline
				\makecell{\samplingname{\scriptsize Uniform-With-Replacement} \\ \scriptsize \samplingname{(new)}}
				& \scriptsize $\Theta\left(n + \dfrac{\max\{\sqrt{n} L_{\pm}, L_{-}\}}{\varepsilon}\right)$ & ---\\
				\hline
				\samplingname{\scriptsize Importance} & \scriptsize $\Theta\left(n + \dfrac{\sqrt{n} \left(\dfrac{1}{n} \sum_{i=1}^n L_i\right)}{\varepsilon}\right)$ & \scriptsize $q_i = \dfrac{L_i}{\sum_{i=1} L_i}$\\
				\hline
				\samplingname{\scriptsize Stratified} & \scriptsize $\Theta\left(n + \dfrac{\max\left\{\sqrt{n} \sqrt{\dfrac{1}{g}\sum_{i = 1}^g L_{i,\pm}^2}, g L_{-}\right\}}{\varepsilon}\right)$ & \scriptsize \shortstack{The functions $f_i$ \\ are split into $g$ groups} \\
				\hline
			\end{tabular}
			\scriptsize Notation: $n = \#$ of data points; $\varepsilon = $ error tolerance; $L_{-}, L_i, L_{\pm}, L_{+}$ and $L_{i,\pm}$ are smoothness constants such that $L_{-} \leq \dfrac{1}{n}\sum_{i=1}^n L_i,$ $L_{-} \leq L_{+}$ and $L_{\pm} \leq L_{+};$ $g = \#$ of groups in the \samplingname{Stratified} sampling.
		\end{threeparttable}
	\end{table}
	\fi
	
	\section{Sampling-dependent Smoothness Constants}
	We now define two smoothness constants that depend on the weights $\{w_i\}_{i=1}^n$ of a sampling $\samplefunc$ and on the functions $f_i.$ 
	\begin{definition}
		\label{ch5:def:weighted_local_lipschitz_constant}
		Given a sampling $\samplefunc \in \mathbb{S}(A, B, \{w_i\}_{i=1}^n)$, let $L_{+,w}$ be a constant such that for all $x, y \in \R^d$ the following holds:
		$$ \dfrac{1}{n} \sum \limits_{i=1}^n \dfrac{1}{n w_i}\norm{\nabla f_i(x) - \nabla f_i(y)}^2 \leq L_{+,w}^2 \norm{x - y}^2.$$
		For $(w_1, \dots, w_n) = (\nicefrac{1}{n}, \dots, \nicefrac{1}{n}),$ we define $L_{+} \eqdef L_{+,w}.$
	\end{definition}
	\begin{definition}
		\label{ch5:def:weighted_hessian_varaince}
		Given a sampling $\samplefunc \in \mathbb{S}(A, B, \{w_i\}_{i=1}^n)$, let $L_{\pm,w}$ be a constant such that for all $x, y \in \R^d$ the following holds:
		\begin{equation*}
		 \dfrac{1}{n} \sum \limits_{i=1}^n \dfrac{1}{n w_i}\norm{\nabla f_i(x) - \nabla f_i(y)}^2 - \norm{\nabla f(x) - \nabla f(y)}^2 \leq L_{\pm,w}^2 \norm{x - y}^2.
		\end{equation*}
		
		For $(w_1, \dots, w_n) = (\nicefrac{1}{n}, \dots, \nicefrac{1}{n}),$ we define $L_{\pm} \eqdef L_{\pm,w}.$
	\end{definition}
	One can interpret Definition~\ref{ch5:def:weighted_local_lipschitz_constant} as \textit{weighted} mean-squared smoothness property \citep{arjevani2019lower}, and Definition~\ref{ch5:def:weighted_hessian_varaince} as \textit{weighted} Hessian variance~\citep{szlendak2021permutation} that captures the similarity between the functions $f_i$.
	The constants $L_{+,w}$ and $L_{\pm,w}$ help us better to understand the structure of the optimization Problem~\eqref{ch5:eq:main_problem} {\em in connection} with a particular choice of a sampling scheme. The next result states that $L_{+,w}^2$ and $L_{\pm,w}^2$ are finite provided the functions $f_i$ are $L_i$ smooth for all $i \in [n].$
	\begin{theorem}
		\label{ch5:theorem:l_estimation}
		If Assumption~\ref{ch5:ass:local_lipschitz_constant} holds, then $$L_{+,w}^2 = L_{\pm,w}^2 = \dfrac{1}{n} \sum_{i=1}^n \dfrac{1}{n w_i} L_i^2$$
		satisfy Definition~\ref{ch5:def:weighted_hessian_varaince} and Definition~\ref{ch5:def:weighted_local_lipschitz_constant}.
	\end{theorem}
	Indeed, from Assumption~\ref{ch5:ass:local_lipschitz_constant} and the inequality $\norm{\nabla f(x) - \nabla f(y)}^2 \geq 0$ we get
	$$ \dfrac{1}{n} \sum \limits_{i=1}^n \dfrac{1}{n w_i}\norm{\nabla f_i(x) - \nabla f_i(y)}^2 - \norm{\nabla f(x) - \nabla f(y)}^2 \leq \left(\dfrac{1}{n} \sum \limits_{i=1}^n \dfrac{1}{n w_i}L_i^2\right)\norm{x - y}^2,$$
	thus we can take for Definition~\ref{ch5:def:weighted_hessian_varaince} the value $$L_{\pm,w}^2 = \dfrac{1}{n} \sum_{i=1}^n \dfrac{1}{n w_i}L_i^2.$$
	
	The proof for $L_{+,w}^2$ for Definition~\ref{ch5:def:weighted_local_lipschitz_constant} is the same.\qed
	
	From the proof, one can see that we ignore $\norm{\nabla f(x) - \nabla f(y)}^2$ when estimating $L_{\pm,w}^2.$ However, by doing that, the obtained result is not tight.

	\section{A General and Refined Theoretical Analysis of {PAGE}}
	\label{ch5:sec:theorems}
	In the Algorithm~\ref{ch5:alg:page}, we provide the description of \algname{PAGE}. The choice of \algname{PAGE} as the base method is driven by the simplicity of the proof in the original paper. However, we believe that other methods, including \algname{SPIDER} and \algname{SARAH}, can also admit samplings from Assumption~\ref{ch5:ass:sampling}.
	
	\begin{algorithm}[!t]
		\caption{\algname{PAGE}.}
		\label{ch5:alg:page}
		\begin{algorithmic}[1]
			\STATE \textbf{Input and Initialization:} initial point $x^0\in \R^d$, step size $\gamma>0$, probability ${p} \in (0, 1]$
			\STATE $g^0 = \nabla f(x^0)$
			\FOR{$t =0,1,\dots$}
			\STATE $x^{t+1} = x^t - \gamma g^t$
			\STATE Generate a random sampling function $\samplefunc^t$
			\STATE $
			g^{t+1}=
			\begin{cases}
				\nabla f(x^{t+1}) & \text{with probability}\ p \\
				g^t + \samplefunc^t\left(\{\nabla f_i(x^{t+1}) - \nabla f_i(x^{t})\}_{i=1}^n\right) & \text{with probability}\ 1 - p \\
			\end{cases}
			$
			\ENDFOR
		\end{algorithmic}
	\end{algorithm}
	
	In this section, we provide theoretical results for Algorithm~\ref{ch5:alg:page}. Let us define $$\Delta_0 \eqdef f(x^0) - f^*.$$
	
	\begin{restatable}{theorem}{THEOREMCONVERGENCEPAGE}
		\label{ch5:theorem:pageab}
		Suppose that Assumptions~\ref{ch5:ass:lower_bound}, \ref{ch5:ass:lipschitz_constant}, \ref{ch5:ass:local_lipschitz_constant} hold and the samplings $$\samplefunc^t \in \mathbb{S}(A, B, \{w_i\}_{i=1}^n).$$
		Then Algorithm~\ref{ch5:alg:page} \algname{(PAGE)} has the convergence rate
		$$\Exp{\norm{\nabla f(\widehat{x}^T)}^2} \leq \dfrac{2\Delta_0}{\gamma T},$$
		where
		$$\gamma \leq \left(L_- + \sqrt{\dfrac{1 - p}{p} \left(\left(A - B\right)L_{+,w}^2 + B L_{\pm,w}^2\right)}\right)^{-1}.$$
	\end{restatable}
	
	To reach an $\varepsilon$-stationary point, it is enough to do 
	\begin{align}
		\label{ch5:eq:rate}
		  T \eqdef \dfrac{2\Delta_0}{\varepsilon}\left(L_- + \sqrt{\dfrac{1 - p}{p} \left(\left(A - B\right)L_{+,w}^2 + B L_{\pm,w}^2\right)}\right)
	\end{align}
	iterations of Algorithm~\ref{ch5:alg:page}. To deduce the gradient complexity, we provide the following corollary.
	
	\begin{restatable}{corollary}{COROLLARYCONVERGENCEPAGE}
		\label{ch5:corollary:pageab}
		Suppose that the assumptions of Theorem~\ref{ch5:theorem:pageab} hold. Let us take $$p = \dfrac{|\samplefunc|}{|\samplefunc| + n}.$$
		Then the complexity (the expected number of gradient calculations $\nabla f_i$) of Algorithm~\ref{ch5:alg:page} equals
		\begin{eqnarray*}
			 		N \eqdef \Theta\left(n + |\samplefunc| T\right) = \Theta\left(n + \dfrac{\Delta_0}{\varepsilon} |\samplefunc| \left(L_- + \sqrt{\dfrac{n}{|\samplefunc|} \left(\left(A - B\right)L_{+,w}^2 + B L_{\pm,w}^2\right)}\right)\right).
		\end{eqnarray*}
	\end{restatable}
	\begin{proof}
		At each iteration, the expected \# gradient calculations equals $$p n + (1 - p) |\samplefunc| \leq 2 |\samplefunc|.$$ Thus the total expected number of gradient calculations of Algorithm~\ref{ch5:alg:page} is bounded above by $n + 2 |\samplefunc| T$ to get an $\varepsilon$-stationary point.
	\end{proof}
	
	The original result from \citep{li2021page} states that the complexity of \algname{PAGE} with batch size $\tau$ is
	\begin{eqnarray}
		\label{ch5:eq:compl_page}
		N_{\textnormal{orig}} \eqdef \Theta\left(n + \dfrac{\Delta_0}{\varepsilon} \tau \left(L_- + \dfrac{\sqrt{n}}{\tau} L_{+}\right)\right).
	\end{eqnarray}

	\subsection{{Uniform-with-replacement} sampling}
	Let us do a sanity check and substitute the parameters of the sampling that the original paper uses. We take the \samplingname{Uniform-With-Replacement} sampling (see Section~\ref{ch5:subsec:3}) with batch size $\tau$ (note that $\tau \geq |\samplefunc|$), $A = B = \nicefrac{1}{\tau}$ and $w_i = \nicefrac{1}{n}$ for all $i \in [n]$ (see Table~\ref{ch5:tbl:AB}) and get the complexity $N_{\textnormal{uniform}} = \Theta\left(n + \dfrac{\Delta_0}{\varepsilon} \tau \left(L_- + \dfrac{\sqrt{n}}{\tau} L_{\pm}\right)\right).$ Next, let us fix $$\tau \leq \max\left\{\dfrac{\sqrt{n} L_{\pm}}{L_{-}},1\right\},$$ and, finally, obtain that $$N_{\textnormal{uniform}} = \Theta\left(n + \dfrac{\Delta_0\max\{\sqrt{n} L_{\pm}, L_{-}\}}{\varepsilon}\right).$$ Let us compare it with \eqref{ch5:eq:compl_page}. With the \textbf{same sampling}, our analysis provides better complexity; indeed, note that $\max\{\sqrt{n} L_{\pm}, L_{-}\} \leq \sqrt{n} L_{+}$ (see Lemma~2 in \citet{szlendak2021permutation}). Moreover, \citet{szlendak2021permutation} provides examples of the optimization problems when $L_{\pm}$ is small and $L_{+}$ is large, so the difference can be arbitrarily large.
	
	\subsection{{Nice} sampling}
	Next, we consider the \samplingname{Nice} sampling (see Section~\ref{ch5:subsec:0}) and get that the complexity $$N_{\textnormal{nice}} = \Theta\left(n + \dfrac{\Delta_0}{\varepsilon} \tau \left(L_- + \dfrac{1}{\tau}\sqrt{\dfrac{n (n-\tau)}{(n-1)}}L_{\pm}\right)\right).$$
	Unlike the \samplingname{Uniform-With-Replacement} sampling, the \samplingname{Nice} sampling recovers the complexity of \algname{GD} for $\tau = n.$
	
	\subsection{{Importance} sampling}
	Let us consider the \samplingname{Importance} sampling (see Section~\ref{ch5:subsec:3}) that justifies the introduction of the weights $w_i.$ For $$\tau \leq \max\left\{\dfrac{\sqrt{n} L_{\pm,w}}{L_{-}},1\right\}$$ we can get the complexity $$N_{\textnormal{importance}} = \Theta\left(n + \dfrac{\Delta_0}{\varepsilon} \tau \left(L_- + \dfrac{\sqrt{n}}{\tau} L_{\pm,w}\right)\right) \leq \Theta\left(n + \dfrac{\Delta_0\max\{\sqrt{n} L_{\pm,w}, L_{-}\}}{\varepsilon}\right).$$ Now, we take $$q_i = w_i = \dfrac{L_i}{\sum_{i=1} L_i}$$ and use the results from Section~\ref{ch5:sec:optimal_weights} to obtain $$N_{\textnormal{importance}} = \Theta\left(n + \dfrac{\Delta_0\sqrt{n} \left(\dfrac{1}{n} \sum_{i=1}^n L_i\right)}{\varepsilon}\right) \leq N_{\textnormal{orig}}.$$ In Example~\ref{ch5:ex:lipt_sum}, we consider the optimization task where $\dfrac{1}{n} \sum_{i=1}^n L_i$ is $\sqrt{n}$ times smaller than $L_{+}.$ Thus the complexity $N_{\textnormal{importance}}$ can be at least $\sqrt{n}$ times smaller that the complexity $N_{\textnormal{orig}}.$
	
	\subsection{The power of $B > 0$}
	In all previous examples, the constant $A = B > 0.$ If $A = B,$ then the complexity from Corollary~\ref{ch5:corollary:pageab} is simplified to:
	$$N  = \Theta\left(n + \dfrac{\Delta_0}{\varepsilon} |\samplefunc| \left(L_- + \sqrt{\dfrac{n}{|\samplefunc|} B L_{\pm,w}^2}\right)\right).$$ 
	Thus the complexity $N$ does not depend on $L_{+,w}^2,$ which greater of equal to $L_{\pm,w}^2.$ This is the first analysis of optimal SGD, which uses $B > 0.$
	
	\subsection{Analysis under Polyak-\L ojasiewicz condition}
	
	The previous results can be extended to the optimization problems that satisfy the Polyak-\L ojasiewicz condition. Under this assumption, Algorithm~\ref{ch5:alg:page} enjoys a linear convergence rate.
	
	\begin{assumption}
		\label{ch5:ass:pl}
		There exists $\mu > 0$ such that the function $f$ satisfy (Polyak-\L ojasiewicz) P\L-condition:
		\begin{align*}
			\norm{\nabla f(x)}^2 \geq 2\mu(f(x) - f^*) \quad \forall x \in \R,
		\end{align*}
		where $f^* = \inf_{x \in \R^d} f(x) > -\infty.$
	\end{assumption}
	
	Using Assumption~\ref{ch5:ass:pl}, we can improve the convergence rate of \algname{PAGE}.
	
	\begin{restatable}{theorem}{THEOREMCONVERGENCEPAGEPL}
		\label{ch5:theorem:pageabpl}
		Suppose that Assumptions~\ref{ch5:ass:lower_bound}, \ref{ch5:ass:lipschitz_constant}, \ref{ch5:ass:local_lipschitz_constant}, \ref{ch5:ass:pl} and the samplings $$\samplefunc^t \in \mathbb{S}(A, B, \{w_i\}_{i=1}^n).$$ Then Algorithm~\ref{ch5:alg:page} \algname{(PAGE)} has the convergence rate
		$$\Exp{f(x^T)} - f^* \leq (1 - \gamma \mu)^T \Delta_0,$$
		where
		$$\gamma \leq \min\left\{\left(L_- + \sqrt{\dfrac{2(1 - p)}{p} \left(\left(A - B\right)L_{+,w}^2 + B L_{\pm,w}^2\right)}\right)^{-1}, \dfrac{p}{2\mu}\right\}.$$
	\end{restatable}

	\section{Composition of Samplings: Application to Federated Learning}
	\label{ch5:sec:sampling_in_sampling}
	In Section~\ref{ch5:sec:theorems}, we analyze \algname{PAGE} method with samplings that satisfy Assumption~\ref{ch5:ass:sampling}. Now, let us assume that the functions $f_i$ have the finite-sum form, in other words, $$f_i(x) \eqdef \dfrac{1}{m_i} \sum_{j=1}^{m_i} f_{ij}(x),$$ thus we an optimization problem
	\begin{align}
		\label{ch5:eq:main_problem:group} 
		   \min \limits \limits_{x \in \R^d}\left\{f(x) \eqdef \dfrac{1}{n} \sum \limits_{i=1}^n \left( \dfrac{1}{m_i} \sum \limits_{j=1}^{m_i} f_{ij}(x)\right) \right\},
	\end{align}
	Another way to get the problem is to assume that we split the functions $f_i$ into groups of sizes $m_i.$ In summary, let us consider \eqref{ch5:eq:main_problem:group} instead of \eqref{ch5:eq:main_problem}.
	
	The Problem~\eqref{ch5:eq:main_problem:group} occurs in many applications, including distributed optimization and Federated Learning \citep{FEDLEARN, mcmahan17fedavg}. In Federated Learning, many devices and machines (nodes) store local datasets that they do not share with other nodes. The local datasets are represented by functions $f_i,$ and all nodes solve the common optimization Problem~\eqref{ch5:eq:main_problem:group}. Due to privacy reasons and communication bottlenecks \citep{kairouz2019advances}, 
	it is infeasible to store and compute the functions $f_i$ locally in one machine.
	
	In general, when we solve \eqref{ch5:eq:main_problem} in one machine, we have the freedom of choosing a sampling $\samplefunc$ for the functions $f_i,$ which we have shown in Section~\ref{ch5:sec:theorems}. However, in Federated Learning, a sampling of nodes or the functions $f_i$ is dictated by hardware limits or network quality \citep{kairouz2019advances}. Still, each $i$\textsuperscript{th} node can choose sampling $\samplefunc_i$ to sample the functions $f_{ij}.$ As a result, we have a composition of the sampling $\samplefunc$ and the samplings $\samplefunc_i$ (see Algorithm~\ref{ch5:alg:page:composition}).
	
	\begin{assumption}
		\label{ch5:ass:local_local_lipschitz_constant}
		For all $j \in [m_i], i \in [n],$ there exists a Lipschitz constant $L_{ij}$ such that $\norm{\nabla f_{ij}(x) - \nabla f_{ij}(y)} \leq L_{ij} \norm{x - y}$ for all $x, y \in \R^d.$
	\end{assumption}
	
	We now introduce the counterpart of Definitions~\ref{ch5:def:weighted_local_lipschitz_constant} and \ref{ch5:def:weighted_hessian_varaince}.
	\begin{definition}
		\label{ch5:def:weighted_local_lipschitz_constant:local}
		For all $i \in [n]$ and any sampling $\samplefunc_i \in \mathbb{S}(A_i, B_i, \{w_{ij}\}_{j=1}^{m_i})$, define constant $L_{i,+,w_i}$ such that 
		$$ \dfrac{1}{m_i} \sum \limits_{j=1}^{m_i} \dfrac{1}{m_i w_{ij}}\norm{\nabla f_{ij}(x) - \nabla f_{ij}(y)}^2 \leq L_{i,+,w_i}^2 \norm{x - y}^2 \quad \forall x, y \in \R^d.$$
	\end{definition}
	\begin{definition}
		\label{ch5:def:weighted_hessian_varaince:local}
		For all $i \in [n]$ and any sampling $\samplefunc_i \in \mathbb{S}(A_i, B_i, \{w_{ij}\}_{j=1}^{m_i})$, define constant $L_{i,\pm,w_i}$ such that $\forall x, y \in \R^d$:
		$$ \dfrac{1}{m_i} \sum \limits_{j=1}^{m_i} \dfrac{1}{m_i w_{ij}}\norm{\nabla f_{ij}(x) - \nabla f_{ij}(y)}^2 - \norm{\nabla f_i(x) - \nabla f_i(y)}^2 \leq L_{i,\pm,w_i}^2 \norm{x - y}^2.$$
	\end{definition}
	
	\begin{algorithm}[!t]
		\caption{\algname{PAGE} with Composition of Samplings.}
		\label{ch5:alg:page:composition}
		\begin{algorithmic}[1] 
			\STATE \textbf{Input and Initialization:} initial point $x^0\in \R^d$, step size $\gamma>0$, probability ${p} \in (0, 1]$, $g^0 = \nabla f(x^0)$
			\FOR{$t =0,1,\dots$}
			\STATE $x^{t+1} = x^t - \gamma g^t$
			\STATE $
			c^{t+1}=
			\begin{cases}
				1 & \text{with probability}\ p \\
				0 & \text{with probability}\ 1 - p \\
			\end{cases}
			$
			\IF{$c^{t+1} = 1$}
			\STATE $g^{t+1} = \nabla f(x^{t+1})$ 
			\ELSE
			\STATE Generate samplings $\samplefunc^t_i$ for all $i \in [n]$
			\STATE $h^{t+1}_i = \samplefunc^t_i\left(\{\nabla f_{ij}(x^{t+1}) - \nabla f_{ij}(x^{t})\}_{j=1}^{m_i}\right)$ for all $i \in [n]$
			\STATE Generate a sampling $\samplefunc^t$ and set $g^{t+1} = g^t + \samplefunc^t\left(\{h^{t+1}_i\}_{i=1}^n\right)$
			\ENDIF
			\ENDFOR
		\end{algorithmic}
	\end{algorithm}
	
	Let us provide the counterpart of Theorem~\ref{ch5:theorem:pageab} for Algorithm~\ref{ch5:alg:page:composition}.
	
	\begin{restatable}{theorem}{THEOREMCONVERGENCEPAGECOMPOSITION}
		\label{ch5:theorem:pageab:composition}
		Suppose that Assumptions~\ref{ch5:ass:lower_bound}, \ref{ch5:ass:lipschitz_constant}, \ref{ch5:ass:local_lipschitz_constant}, \ref{ch5:ass:local_local_lipschitz_constant} hold and the samplings $$\samplefunc^t \in \mathbb{S}(A, B, \{w_{i}\}_{i=1}^{n})$$ and the samplings $$\samplefunc^t_i \in \mathbb{S}(A_i, B_i, \{w_{ij}\}_{j=1}^{m_i}), \forall i \in [n].$$ Moreover, $B \leq 1.$ Then Algorithm~\ref{ch5:alg:page:composition} has the convergence rate
		$$\Exp{\norm{\nabla f(\widehat{x}^T)}^2} \leq \dfrac{2\Delta_0}{\gamma T},$$
		where
		\\
		\resizebox{1.0\linewidth}{!}{
			\begin{minipage}{\linewidth}
				\begin{align*}
					\gamma \leq \left(L_- + \sqrt{\dfrac{1 - p}{p} \left(\dfrac{1}{n}\sum_{i = 1}^n \left(\dfrac{A}{n w_i} + \dfrac{(1 - B)}{n}\right)\left((A_i - B_i) L_{i,+,w_i}^2 + B_i L_{i,\pm,w_i}^2\right) + (A - B)L_{+,w}^2 + B L_{\pm,w}^2\right)}\right)^{-1}.
				\end{align*}
			\end{minipage}
		}
	\end{restatable}
	
	The obtained theorem provides a general framework that helps analyze the convergence rates of the composition of samplings that satisfy Assumption~\ref{ch5:ass:sampling}. We discuss the obtained result in different contexts.
	
	\subsection{Federated learning}
	For simplicity, let us assume that the samplings $\samplefunc^t$ and $\samplefunc_i^t$ are \textit{Uniform-With-Replacement} samplings with batch sizes $\tau$ and $\tau_i$ for all $i \in n$, accordingly, then to get $\varepsilon$-stationary point, it is enough to do
	$$
	T \eqdef \Theta\left(\dfrac{\Delta_0}{\varepsilon} \left(L_- + \sqrt{\dfrac{1 - p}{p \tau} \left(\dfrac{1}{n}\sum_{i = 1}^n \dfrac{1}{\tau_i} L_{i,\pm}^2 + L_{\pm}^2\right)}\right)\right)
	$$
	iterations. Note that for all $\tau_i \geq 1$ for all $i \in [n]$ the following holds:
	$$
	T \geq \Theta\left(\dfrac{\Delta_0}{\varepsilon} \left(L_- + \sqrt{\dfrac{1 - p}{p \tau} L_{\pm}^2}\right)\right).
	$$
	It means that after some point, there is no benefit in increasing batch sizes $\tau_i.$ In order to balance
	\begin{equation*}
 		\dfrac{1}{n}\sum_{i = 1}^n \dfrac{1}{\tau_i} L_{i,\pm}^2 \qquad \mathrm{and} \qquad L_{\pm}^2,
	\end{equation*}
	 one can take $$\tau_i = \Theta\left(\nicefrac{L_{i,\pm}^2}{L_{\pm}^2}\right).$$ The constant $L_{i,\pm}^2$ captures the \textit{intra-variance} inside $i$\textsuperscript{th} node, while $L_{\pm}^2$ captures the \textit{inter-variance} between nodes. 
	
	If the \textit{intra-variance} is small with respect to the \textit{inter-variance}, then our theory suggests taking small batch sizes and vice versa. 
	
	\subsection{{Stratified} sampling}
	
	Let us provide another example that is closely related to \citep{zhao2014accelerating}. Let us consider \eqref{ch5:eq:main_problem} and use a variation of the \textit{Stratified} sampling \citep{zhao2014accelerating}: we split the functions $f_i$ into $g = \nicefrac{n}{m}$ groups, where $m$ is the number of functions in each group. Thus we get the Problem~\eqref{ch5:eq:main_problem:group} with $$f(x) = \dfrac{1}{g} \sum_{i=1}^g \dfrac{1}{m} \sum_{j=1}^{m} f_{ij}(x).$$
	Let us assume that we always sample \textit{all} groups, thus $A = B = 0,$ and the sampling $\samplefunc_i^t$ are \textit{Nice} samplings with batch sizes $\tau_1$ for all $i \in [n].$ Applying Theorem~\ref{ch5:theorem:pageab:composition}, we get the convergence rate
	$$
	T_{\textnormal{group}} \eqdef \Theta\left(\dfrac{\Delta_0}{\varepsilon} \left(L_- + \sqrt{\dfrac{1 - p}{p g \tau_1}\left(\dfrac{1}{g}\sum_{i = 1}^g L_{i,\pm}^2\right)}\right)\right).
	$$
	At each iteration, the algorithm calculates $g \tau_1$ gradients, thus we should take $$p = \dfrac{g \tau_1}{g \tau_1 + n}$$ to get the complexity
	$$
	N_{\textnormal{group}} \eqdef \Theta\left(n + g \tau_1 T\right) = \Theta\left(n + \dfrac{\Delta_0}{\varepsilon} \left(g \tau_1 L_- + \sqrt{n}\sqrt{\dfrac{1}{g}\sum_{i = 1}^g L_{i,\pm}^2}\right)\right).
	$$
	Let us take $\tau_1 \leq \max\left\{\dfrac{\sqrt{n} \sqrt{\dfrac{1}{g}\sum_{i = 1}^g L_{i,\pm}^2}}{g L_{-}}, 1\right\}$ to obtain the complexity $$N_{\textnormal{group}} = \Theta\left(n + \dfrac{\Delta_0 \max\left\{\sqrt{n} \sqrt{\dfrac{1}{g}\sum_{i = 1}^g L_{i,\pm}^2}, g L_{-}\right\}}{\varepsilon}\right).$$ Comparing the complexity $N_{\textnormal{group}}$ with the complexity $N_{\textnormal{uniform}}$ from Section~\ref{ch5:sec:theorems}, one can see that if split the functions $f_i$ in a ``right way'', such that $L_{i,\pm}$ is small for $i \in [n]$ (see Example~\ref{ch5:ex:groups}), then we can get at least $\nicefrac{\sqrt{n}}{\sqrt{g}}$ times improvement with the \textit{Stratified} sampling.
	
	\section{Experiments}
	\label{ch5:sec:experiments}
	We now provide experiments with synthetic quadratic optimization tasks, where the functions $f_i$, in general, are non-convex quadratic functions. Note that our goal here is to check whether the dependencies that our theory predicts are correct for the Problem~\eqref{ch5:eq:main_problem}. The procedures that generate synthetic quadratic optimization tasks give us control over the choice of smoothness constants. 
	
	All parameters of Algorithm~\ref{ch5:alg:page} are chosen as suggested in Theorem~\ref{ch5:theorem:pageab} and Cor~\ref{ch5:corollary:pageab}. In the plots, we represent the relation between the norm of gradients and the number of gradient calculations $\nabla f_i$.
	
	\subsection{Quadratic optimization tasks with various Hessian variances $L_{\pm}$}
	
	\label{ch5:sec:experiment:quadratic_l_pm}
	
	Using Algorithm~\ref{ch5:algorithm:matrix_generation} (see Appendix), we generated various quadratic optimization tasks with different smoothness constants $L_{\pm} \in [0, 1.0]$ and fixed $L_{-} \approx 1.0$ (see Figure~\ref{ch5:fig:page_ab_arxiv_2022_nodes_1000_nl_0_01_05_1_10_dim_10_wc_with_compl_more_info}). We choose $d = 10,$  $n = 1000,$ regularization $\lambda = 0.001,$ and the noise scale $s \in \{0, 0.1, 0.5, 1\}.$ According to Section~\ref{ch5:sec:theorems} and Table~\ref{ch5:tbl:compl}, the gradient complexity of original \algname{PAGE} method (``Vanilla PAGE'' in Figure~\ref{ch5:fig:page_ab_arxiv_2022_nodes_1000_nl_0_01_05_1_10_dim_10_wc_with_compl_more_info}) is proportional to $L_{+}.$ While the gradient complexity of the new analysis with the \samplingname{Uniform-With-Replacement} 
	sampling (``Uniform-With-Replacement'' in Figure~\ref{ch5:fig:page_ab_arxiv_2022_nodes_1000_nl_0_01_05_1_10_dim_10_wc_with_compl_more_info}) is proportional to $L_{\pm},$ which is always less or equal $L_{+}.$ In Figure~\ref{ch5:fig:page_ab_arxiv_2022_nodes_1000_nl_0_01_05_1_10_dim_10_wc_with_compl_more_info}, one can see that the smaller $L_{\pm}$ with respect to $L_{+},$ the better the performance of ``Uniform-With-Replacement.'' Moreover, we provide experiments with the \samplingname{Importance} sampling (``Importance'' in Figure~\ref{ch5:fig:page_ab_arxiv_2022_nodes_1000_nl_0_01_05_1_10_dim_10_wc_with_compl_more_info}) with $q_i = \nicefrac{L_{i}}{\sum_{i=1}^n L_i}$ for all $i \in [n].$ This sampling has the best performance in all regimes. 

	\iftrue
	\begin{figure*}[h]
		\centering	
		\captionsetup[sub]{font=footnotesize,labelfont={},labelformat=empty}		
		\captionsetup[subfigure]{font=footnotesize,labelfont={},labelformat=empty}
		\captionsetup[figure]{font=footnotesize,labelfont={},labelformat=empty}
					
		\begin{subfigure}[ht]{0.38\textwidth}
			\includegraphics[width=\textwidth]{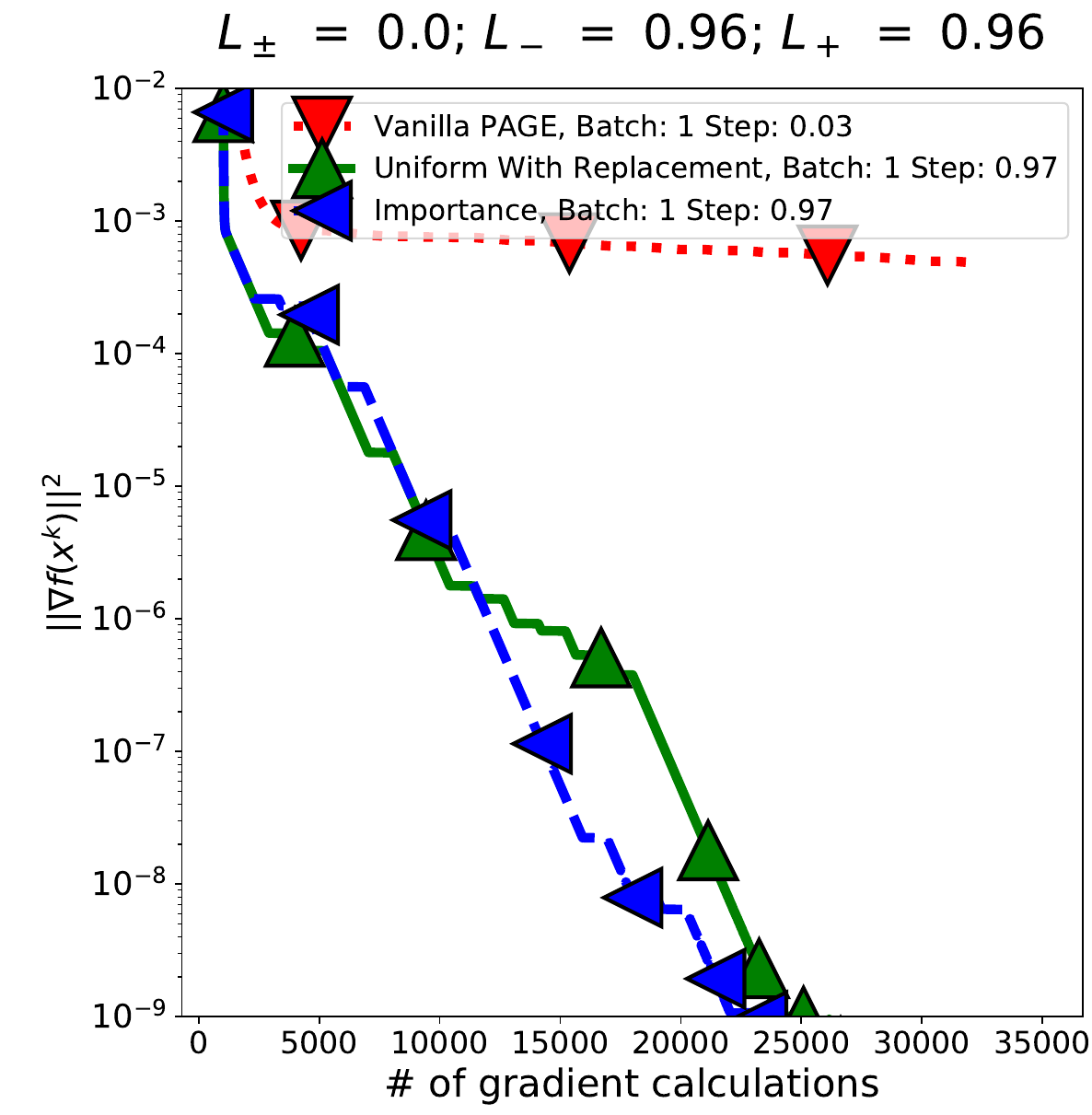} 
			\caption{}
		\end{subfigure}
		\begin{subfigure}[ht]{0.38\textwidth}
			\includegraphics[width=\textwidth]{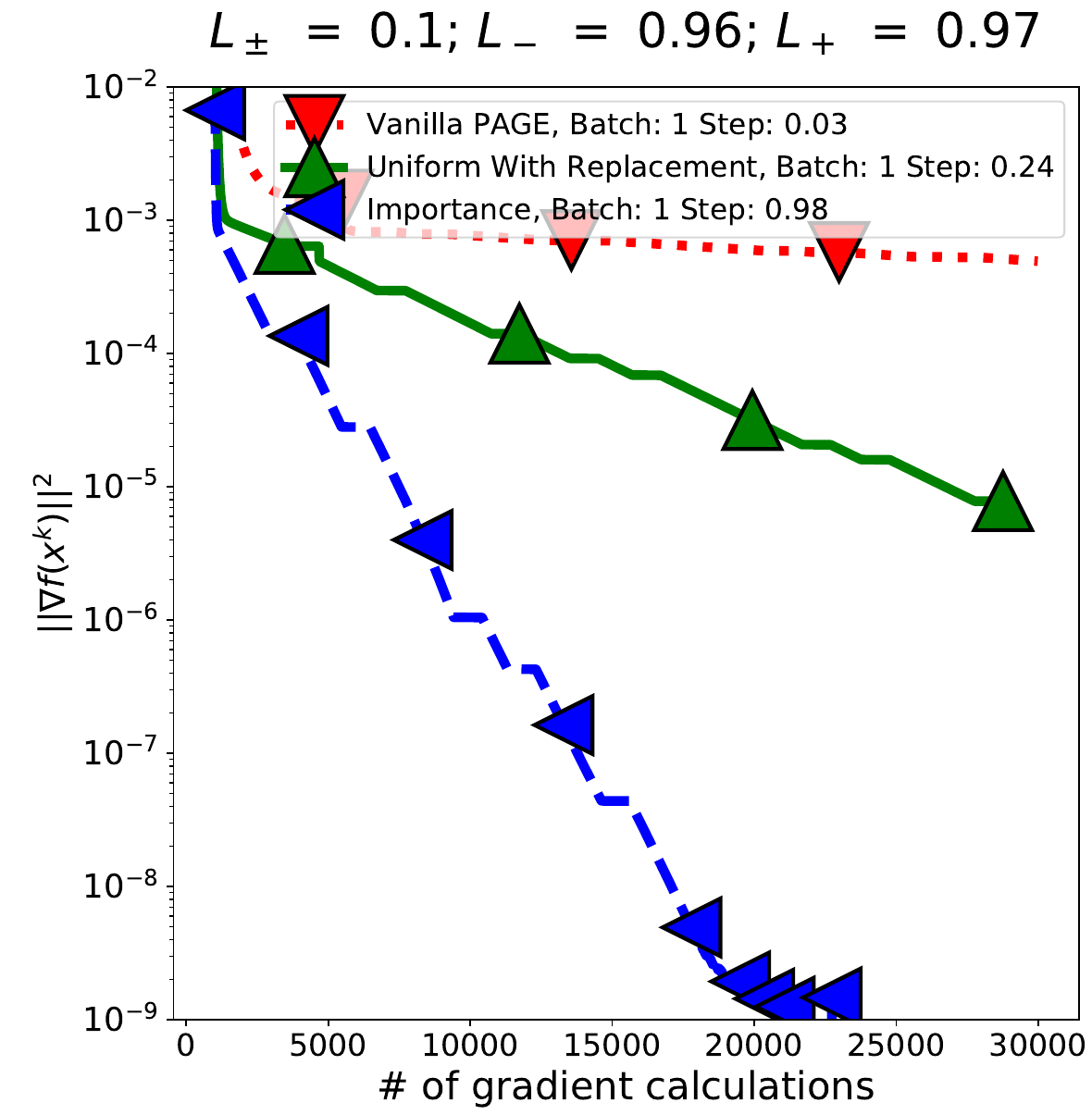}
			\caption{}
		\end{subfigure}
		
		\begin{subfigure}[ht]{0.38\textwidth}
			\includegraphics[width=\textwidth]{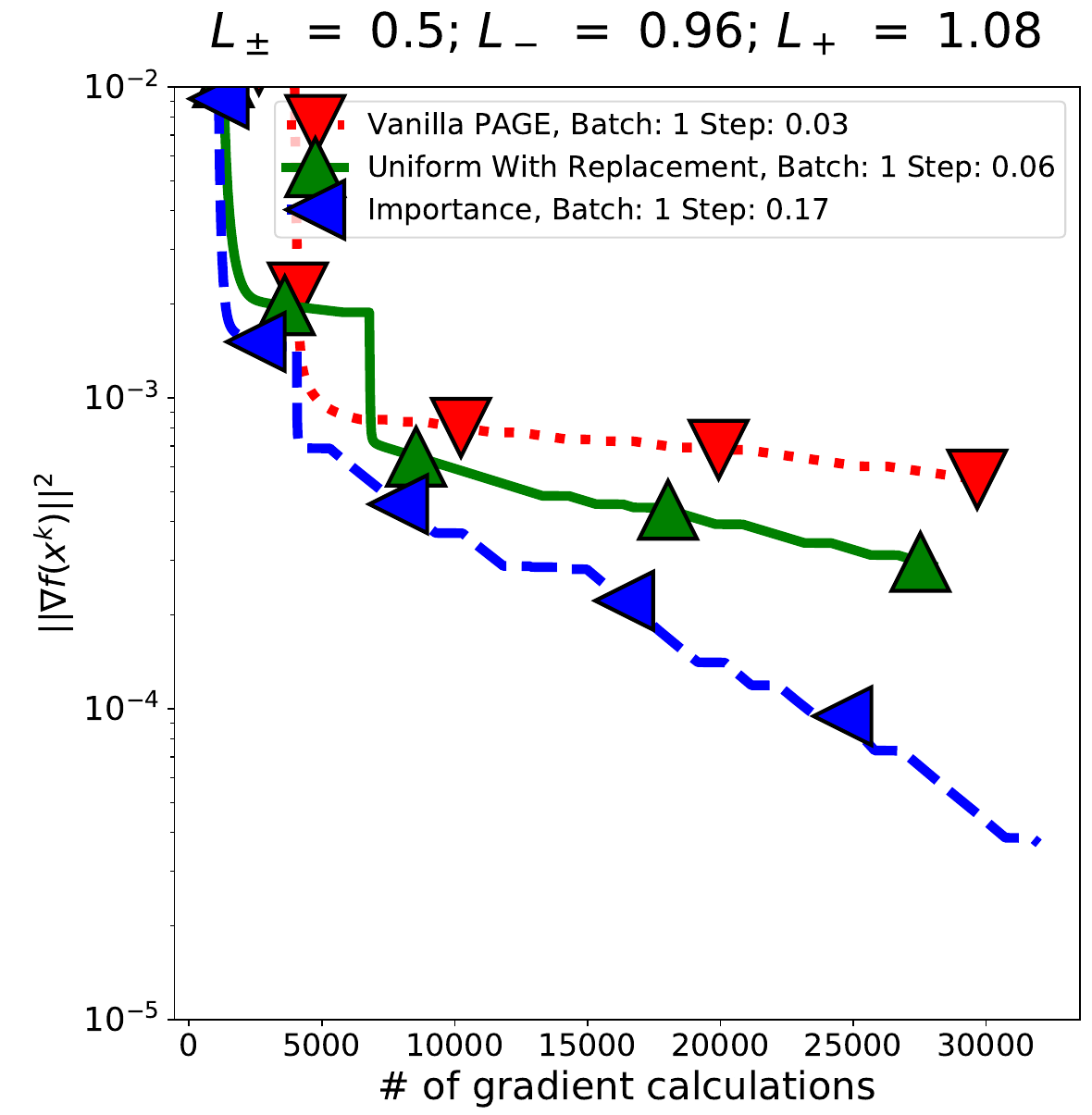} 
			\caption{}
		\end{subfigure}
		\begin{subfigure}[ht]{0.38\textwidth}
			\includegraphics[width=\textwidth]{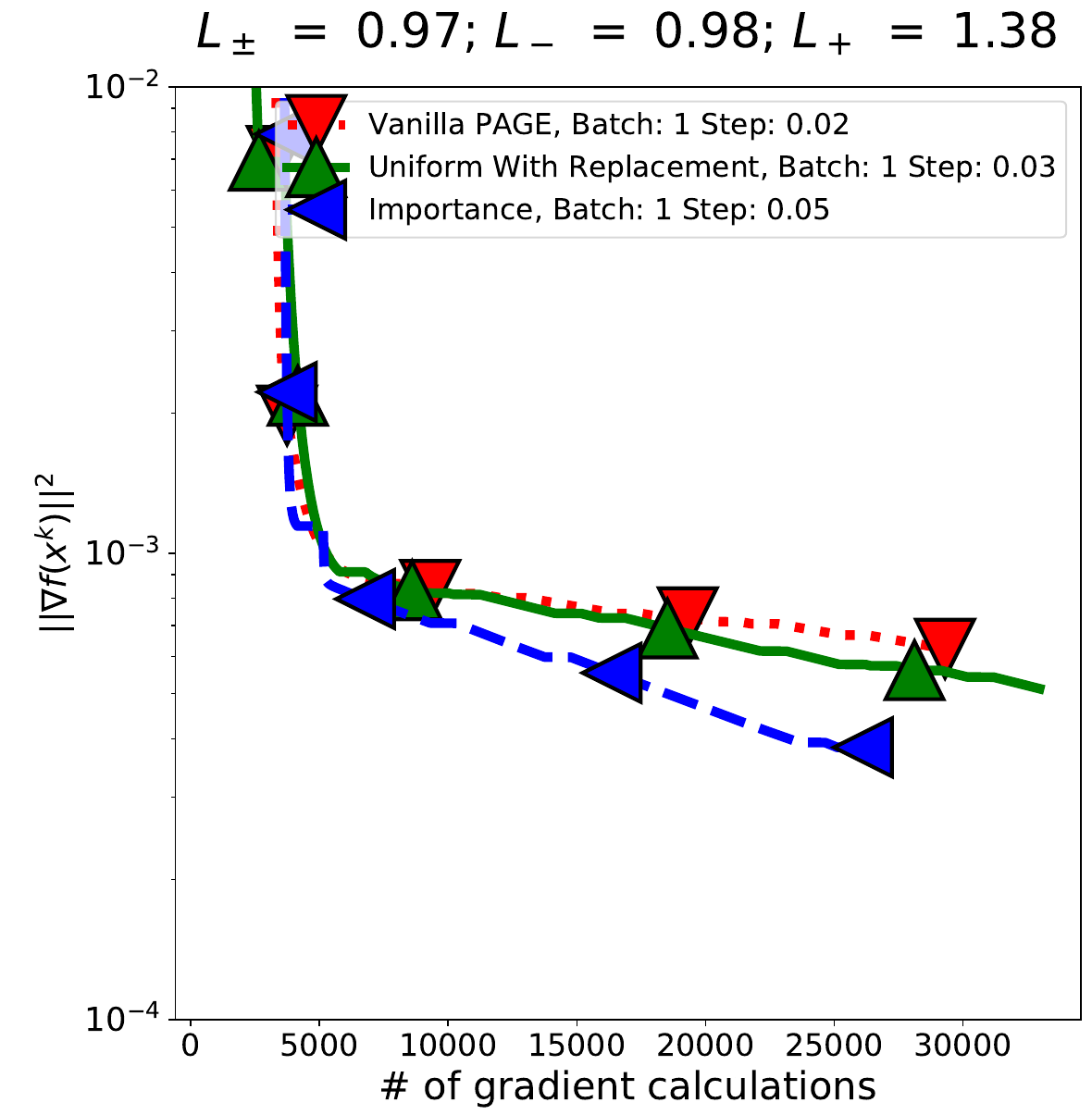} 
			\caption{}
		\end{subfigure}

		\caption{Comparison of samplings and methods on quadratic optimization tasks with various $L_{\pm}$. The number of functions is $1000$.}
		\label{ch5:fig:page_ab_arxiv_2022_nodes_1000_nl_0_01_05_1_10_dim_10_wc_with_compl_more_info}		
	\end{figure*}
	\fi
	
	\subsection{Quadratic optimization tasks with various local Lipschitz constatns $L_{i}$}
	\label{ch5:sec:experiment:quadratic_l_i}
	
	Using Algorithm~\ref{ch5:algorithm:matrix_generation_l_i} (see Appendix), we synthesized various quadratic optimization tasks with different smoothness constants $L_{i}$ (see Figure~\ref{ch5:fig:page_ab_arxiv_2022_nodes_1000_nl_0_01_05_1_10_dim_10_with_compl_more_info}). We choose $d = 10,$ $n = 1000,$ the regularization $\lambda = 0.001,$ and the noise scale $s \in \{0, 0.1, 0.5, 10.0\}.$
	We generated tasks in such a way that the difference between $\max_i L_i$ and $\min_i L_i$ increases. First, one can see that the \samplingname{Uniform-With-Replacement} sampling with the new analysis (``Uniform-With-Replacement'' in Figure~\ref{ch5:fig:page_ab_arxiv_2022_nodes_1000_nl_0_01_05_1_10_dim_10_with_compl_more_info}) has better performance even in the cases of significant variations of $L_i$. Next, we see the stability of the \samplingname{Importance} sampling (``Importance'' in Figure~\ref{ch5:fig:page_ab_arxiv_2022_nodes_1000_nl_0_01_05_1_10_dim_10_with_compl_more_info}) with respect to these variations.

	\iftrue
	\begin{figure*}[h]
		\centering	
		\captionsetup[sub]{font=footnotesize,labelfont={},labelformat=empty}		
		\captionsetup[subfigure]{font=footnotesize,labelfont={},labelformat=empty}
		\captionsetup[figure]{font=footnotesize,labelfont={},labelformat=empty}
		
		\begin{subfigure}[ht]{0.4\textwidth}
			\includegraphics[width=\textwidth]{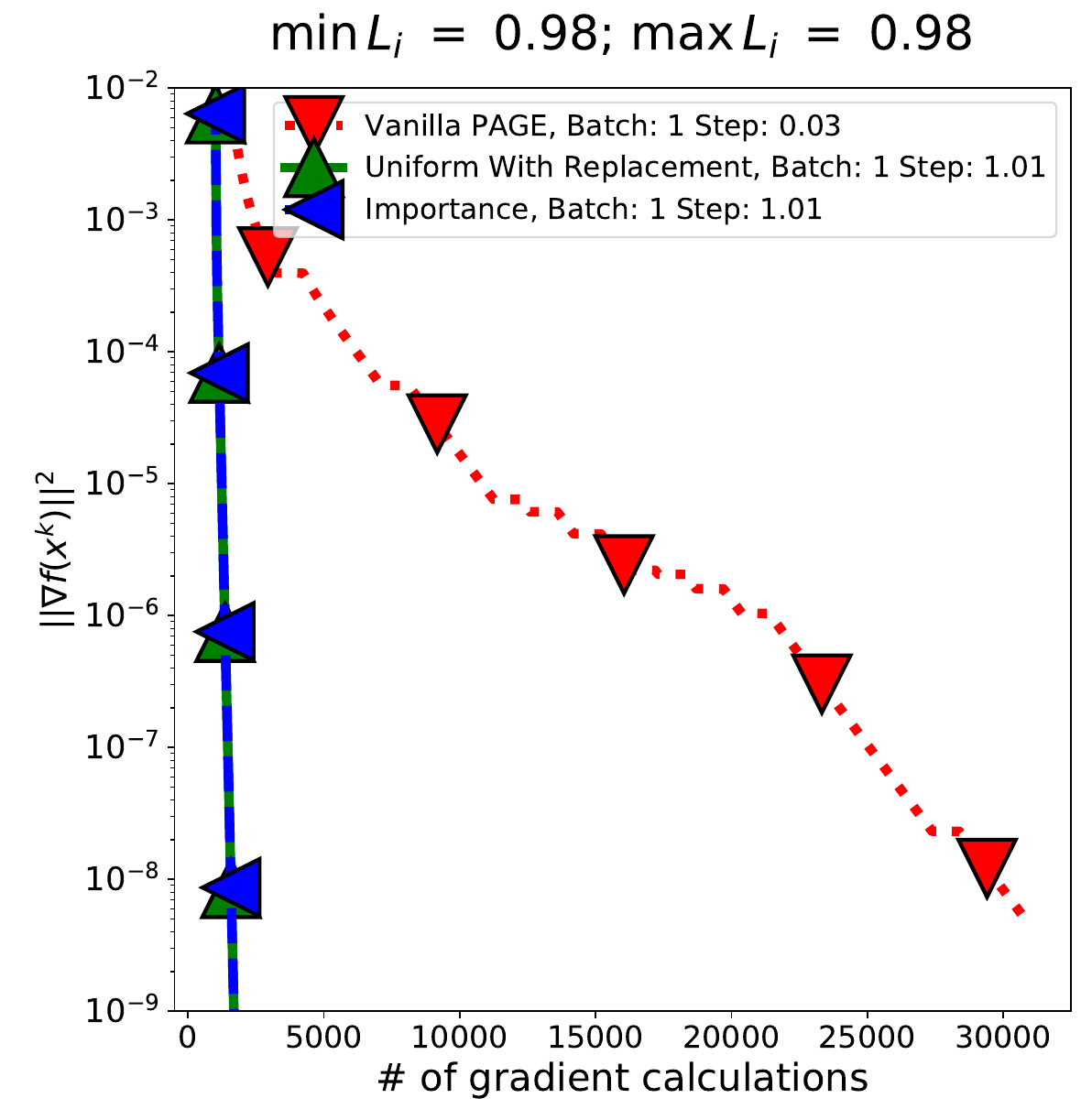} 
			\caption{}
		\end{subfigure}
		\begin{subfigure}[ht]{0.4\textwidth}
			\includegraphics[width=\textwidth]{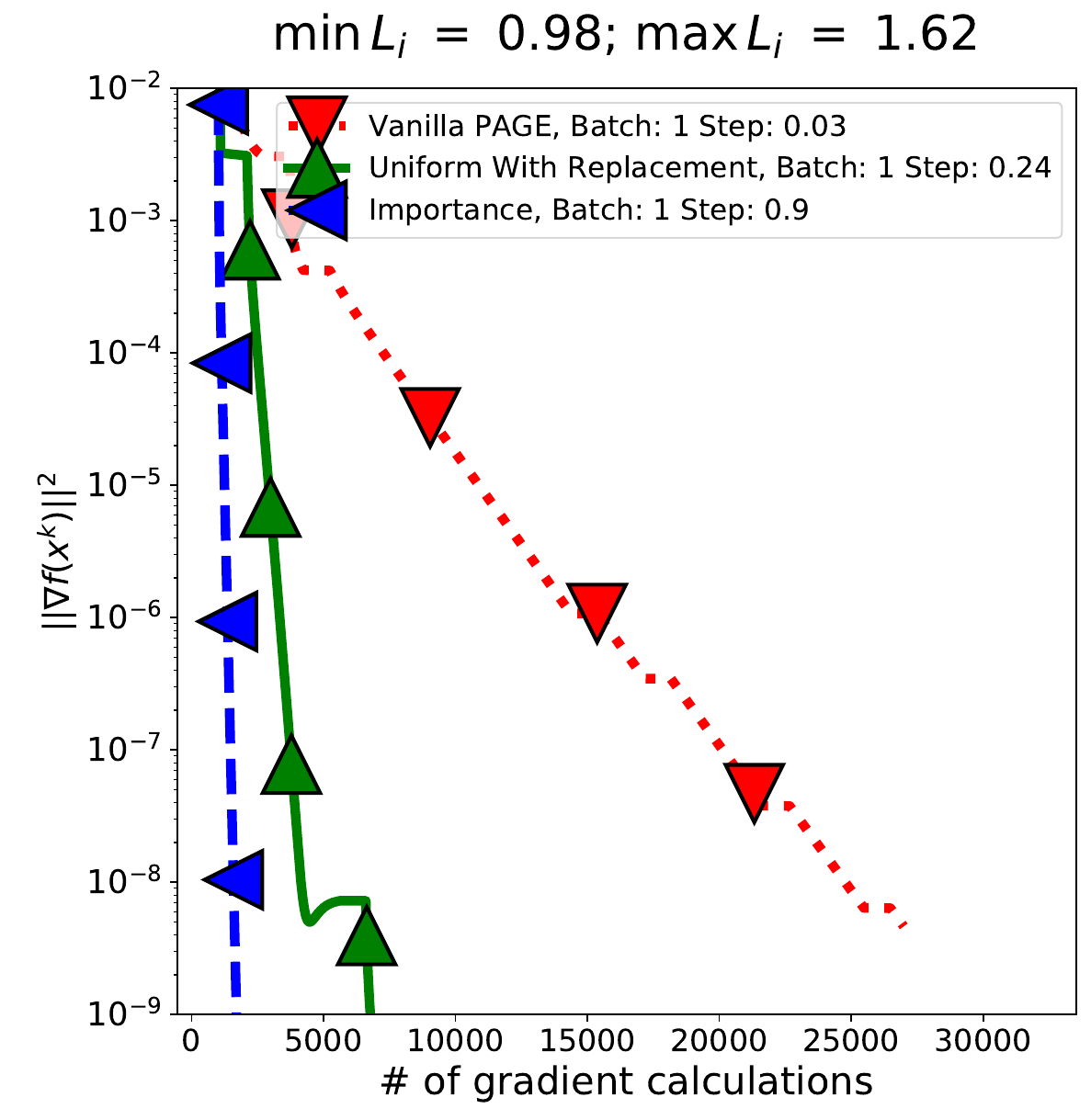}
			\caption{}
		\end{subfigure}
		
		\begin{subfigure}[ht]{0.4\textwidth}
			\includegraphics[width=\textwidth]{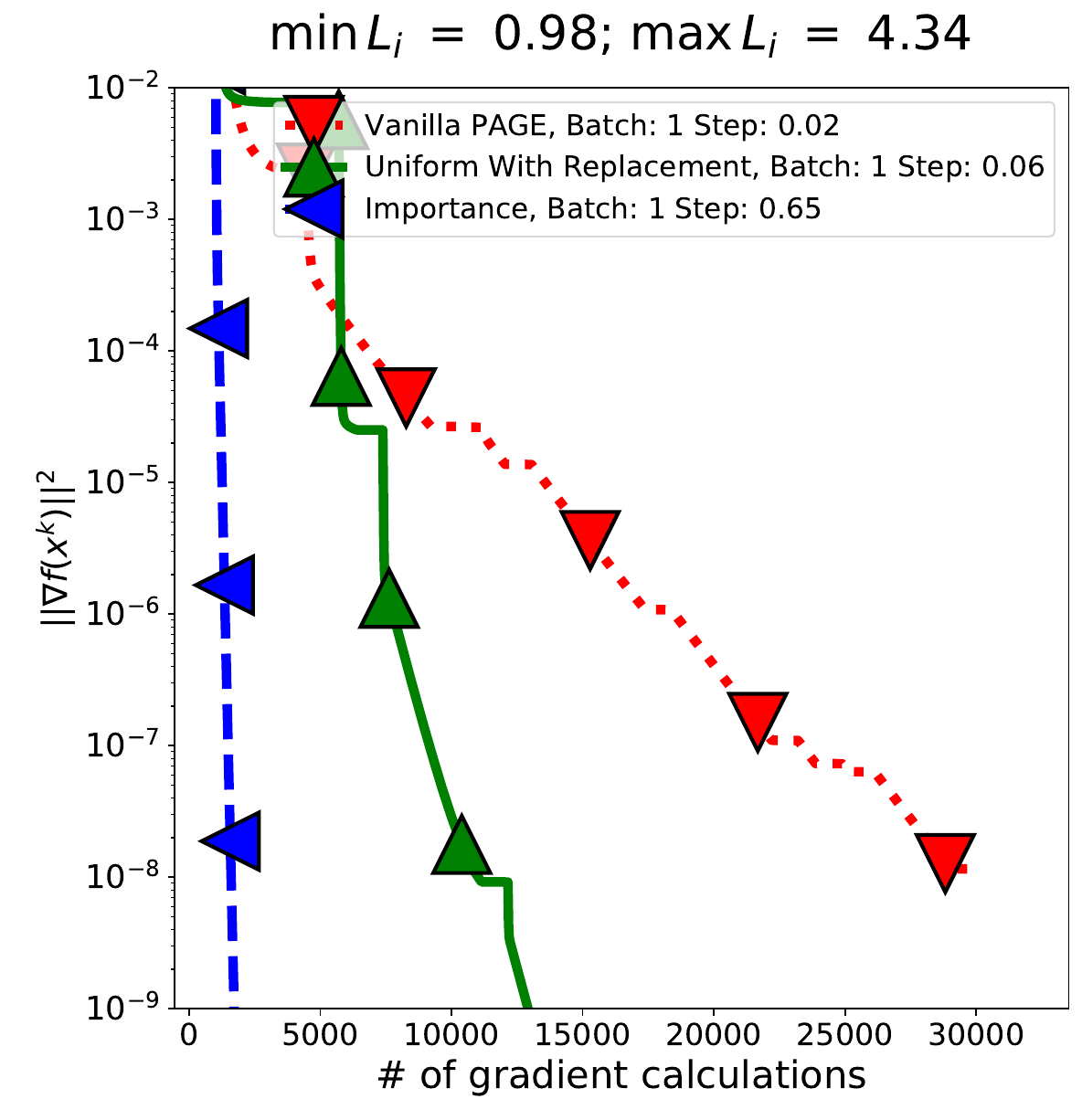} 
			\caption{}
		\end{subfigure}
		\begin{subfigure}[ht]{0.4\textwidth}
			\includegraphics[width=\textwidth]{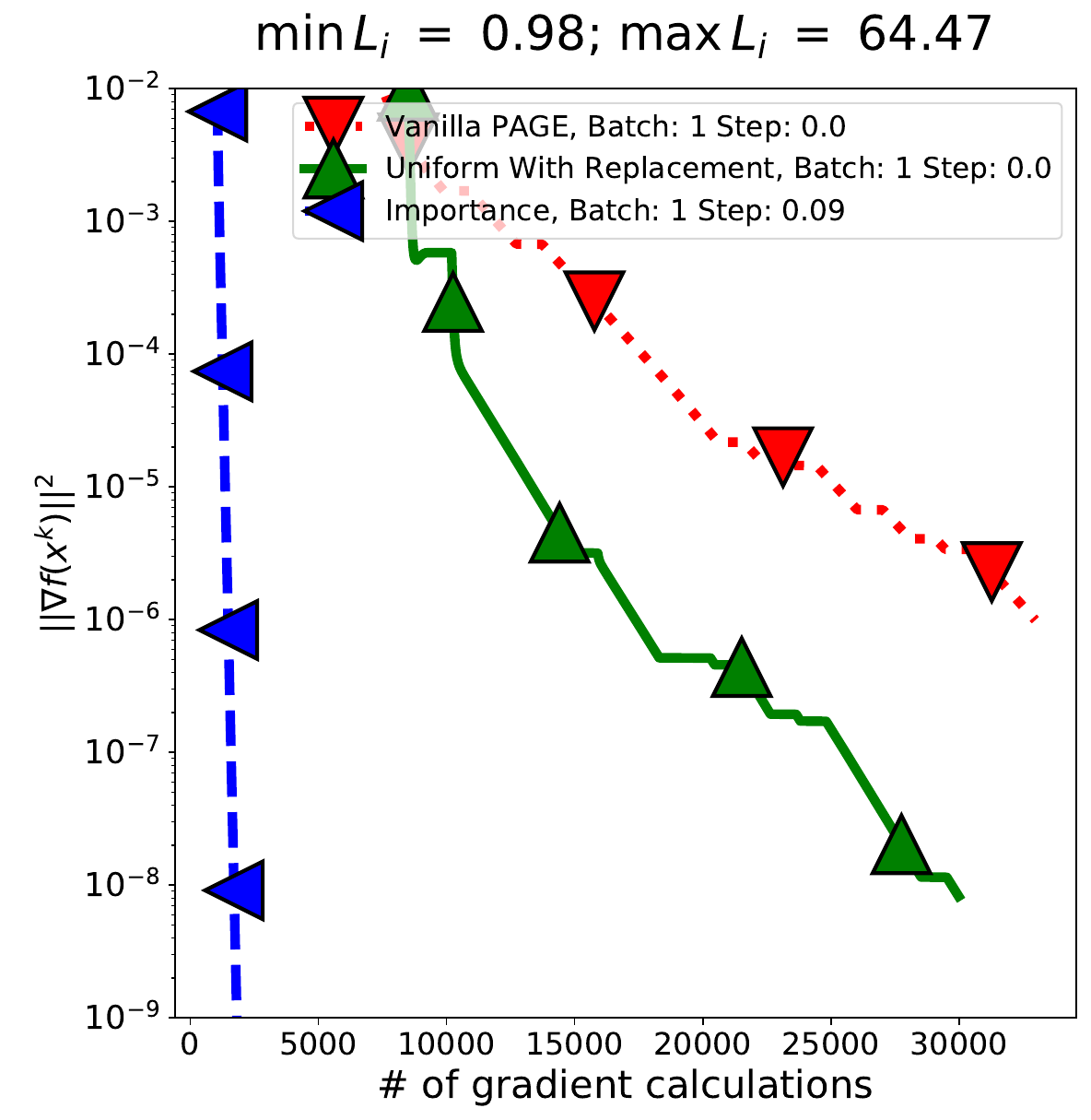} 
			\caption{}
		\end{subfigure}
		
		\caption{Comparison of samplings and methods on quadratic optimization tasks with various $L_{i}$. The number of functions is $1000$.}
		\label{ch5:fig:page_ab_arxiv_2022_nodes_1000_nl_0_01_05_1_10_dim_10_with_compl_more_info}	
	\end{figure*}
	\fi
	
	\subsection{Non-convex classification problem with LIBSVM datasets}
	We now solve non-convex machine learning tasks and compare samplings on \dataname{LIBSVM} datasets \citep{chang2011libsvm} (see details in Section~\ref{ch5:sec:experiment:libsvm:details}). As in previous sections, \algname{PAGE} with the \samplingname{Importance} sampling performs better, especially in \dataname{AUSTRALIAN} dataset where the variation of $L_i$ is large.
	
	\begin{figure}[H]
		\centering
		\begin{subfigure}{.495\textwidth}
			\centering
			\includegraphics[width=1.0\linewidth]{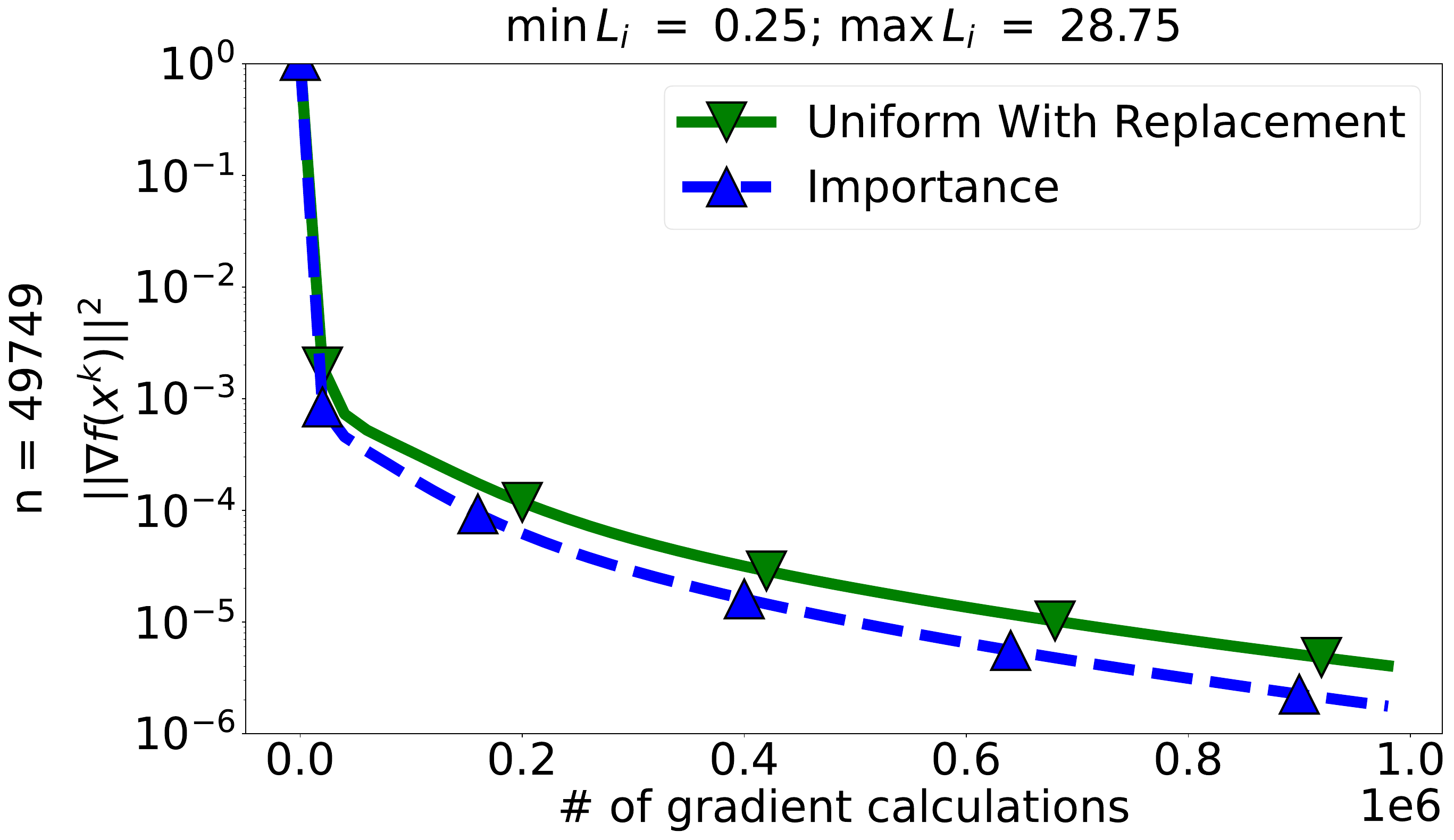}
			\caption*{\dataname{W8A} dataset.}
		\end{subfigure}
		\begin{subfigure}{.495\textwidth}
			\centering
			\includegraphics[width=1.0\linewidth]{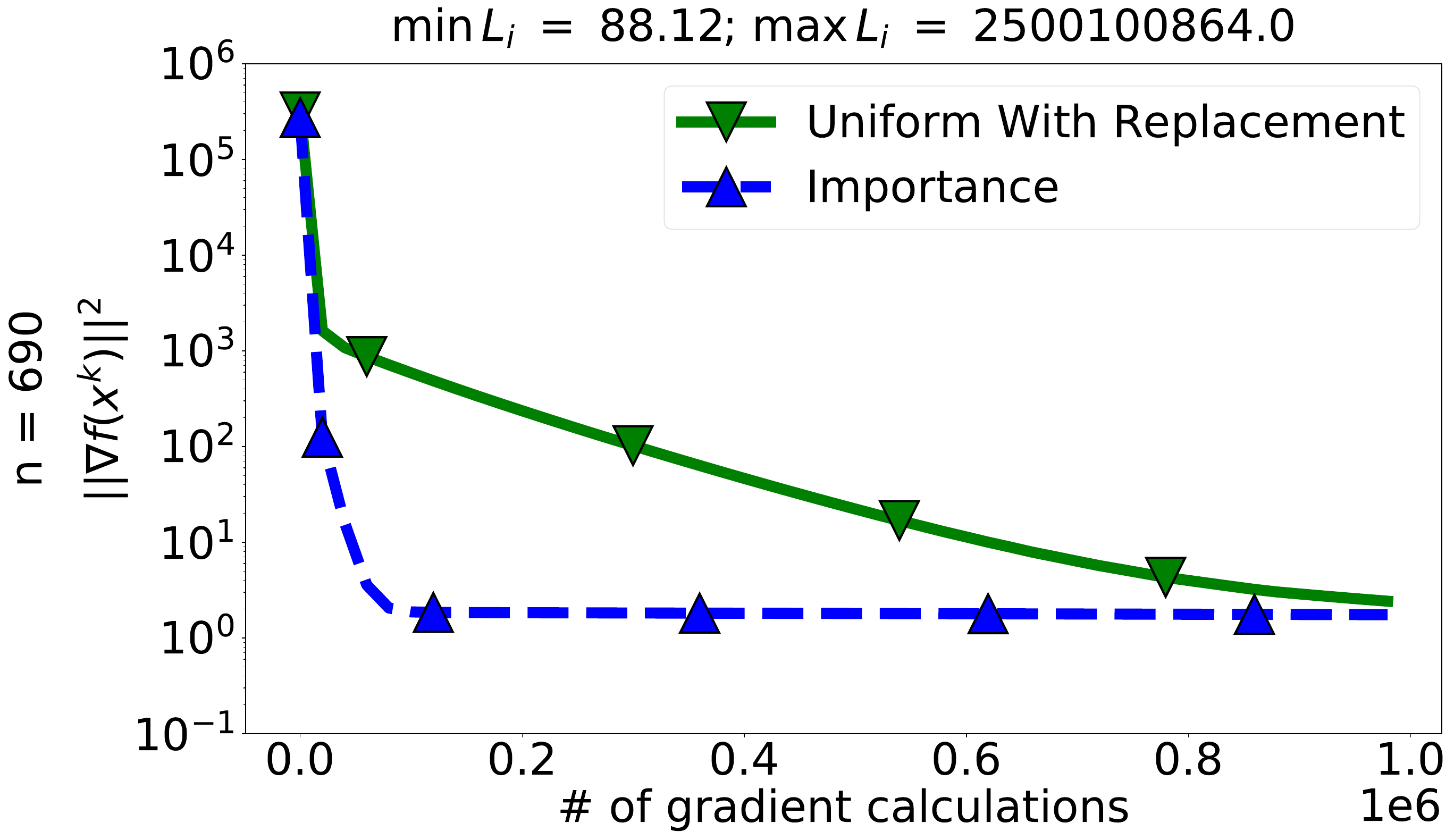}
			\caption*{\dataname{AUSTRALIAN} dataset.}
		\end{subfigure}
		\caption{Comparison of samplings on non-convex machine learning tasks with \dataname{LIBSVM} datasets.}
		\label{ch5:fig:test}
	\end{figure}
	
	\section{Conclusions}
	In this work, we revisit the classical problem of finding an approximately stationary point of the average of $n$ smooth and possibly non-convex functions. 
	The optimal complexity of stochastic first-order methods in terms of the number of gradient computations of individual functions is $\mathcal{O}\left(n + n^{1/2}\varepsilon^{-1}\right)$, attained by the optimal \algname{SGD} methods \algname{SPIDER} \citep{fang2018spider} and \algname{PAGE} \citep{li2021page}, for example, where $\varepsilon$ is the error tolerance. However, i) the big-$\mathcal{O}$ notation hides crucial dependencies on the smoothness constants associated with the functions, and ii) the rates and theory in these methods assume simplistic sampling mechanisms that do not offer any flexibility. In this work, we remedy the situation. First, we generalize \algname{PAGE} algorithm so that it can provably work with virtually any (unbiased) sampling mechanism. This is particularly useful in Federated Learning, as it allows us to construct and better understand the impact of various combinations of client and data sampling strategies. Second, our analysis is sharper as we make explicit use of certain novel inequalities that capture the intricate interplay between the smoothness constants and the sampling procedure. Indeed, our analysis is better even for the simple sampling procedure analyzed in \algname{PAGE} paper. However, this already improved bound can be further sharpened by a different sampling scheme that we propose. In summary, we provide the most general and most accurate analysis of optimal \algname{SGD} in the smooth non-convex regime. Our theoretical findings are supposed with carefully designed experiments.
	
	\clearpage
	\appendix
		
	\part*{Appendices to Chapter \ref{chapter5}}
	\label{ch5:app:toc_1}
	\newpage
	
	\phantomsection
	\addcontentsline{toc}{chapter}{Appendices to Chapter 5}
	
	\addtocounter{adjsection}{1}
	\section{Extra Experiments and Details}
	
	\subsection{Quadratic optimization tasks with various batch sizes}
	\label{ch5:sec:experiment:sythetic}
	
	In this section, we consider the same setup as in Section~\ref{ch5:sec:experiment:quadratic_l_pm}. In Figure~\ref{ch5:fig:page_ab_arxiv_2022_nodes_1000_nl_0_01_05_1_10_dim_10_wc_bz_more_info}, we fix $L_{\pm},$ and show that the \samplingname{Importance} sampling has better convergence rates with different batch sizes. Note that with large batches, the competitors reduce to \algname{GD} method, and the difference is not significant.
	
	\begin{figure}[h]
		\centering
		\includegraphics[width=0.6\textwidth]{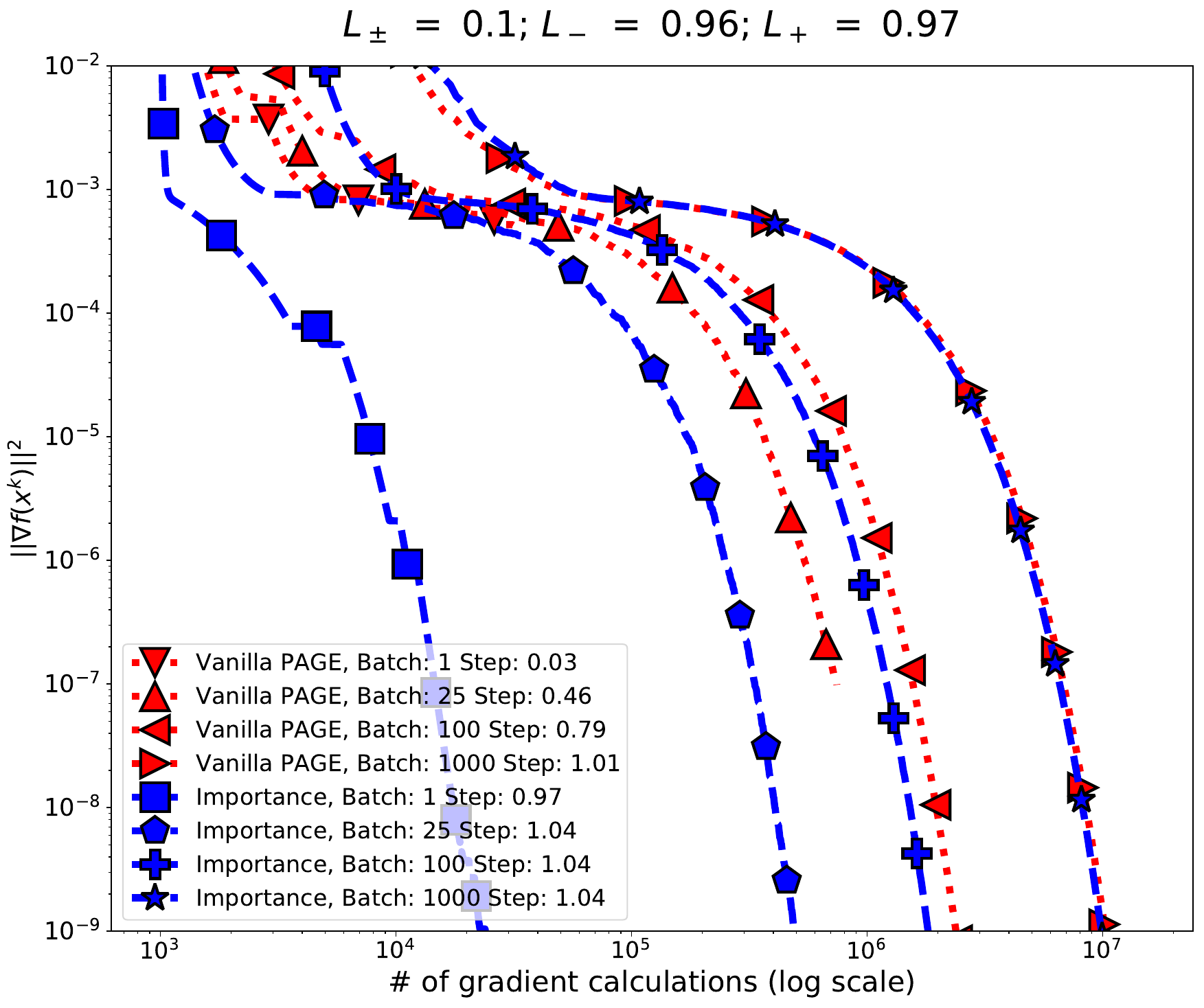}
		\caption{Comparison of samplings and methods with various batch sizes. The number of functions is $1000$.}
		\label{ch5:fig:page_ab_arxiv_2022_nodes_1000_nl_0_01_05_1_10_dim_10_wc_bz_more_info}
	\end{figure}
	
	\subsection{Details on experiments with LIBSVM datasets}
	\label{ch5:sec:experiment:libsvm:details}
	
	\begin{figure}
		\centering
		\begin{subfigure}{.49\textwidth}
			\centering
			\includegraphics[width=1.0\linewidth]{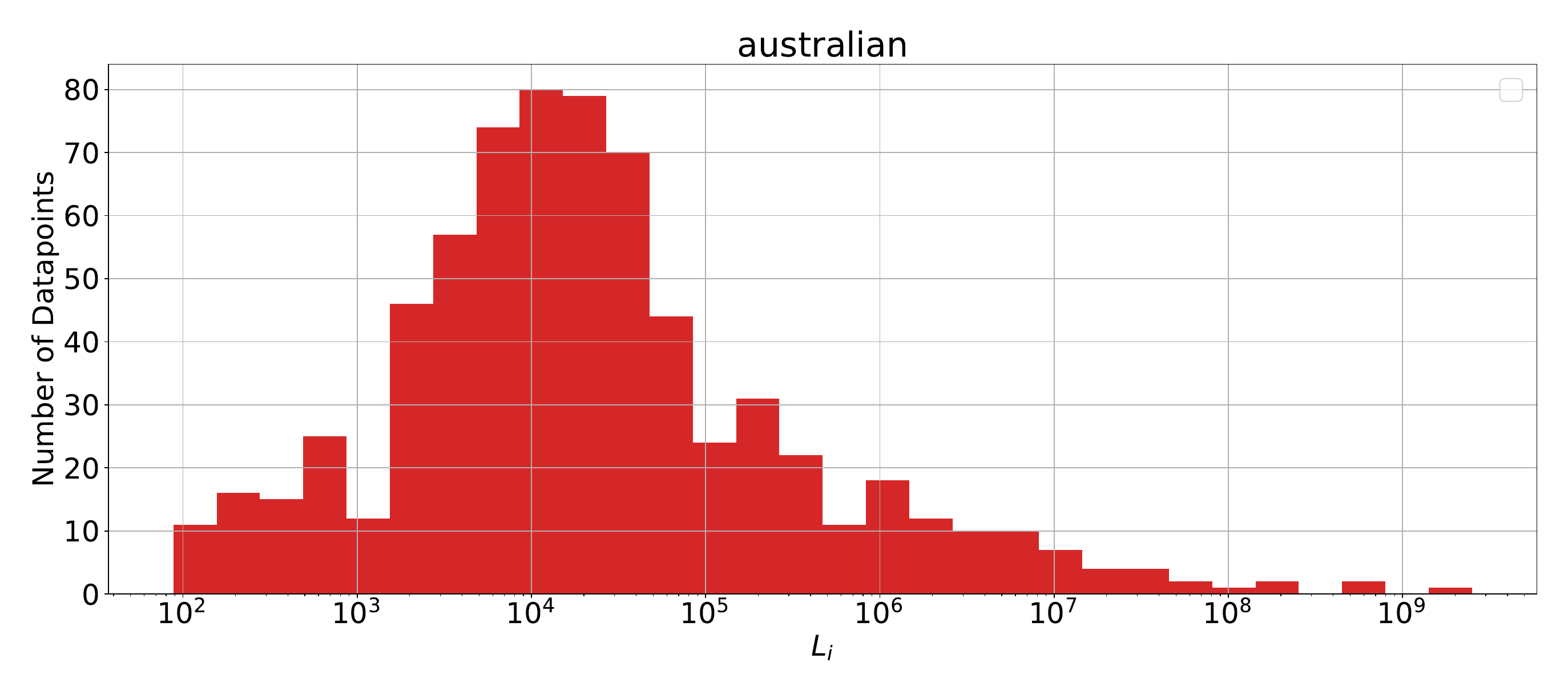}
			\caption*{}
		\end{subfigure}
		\begin{subfigure}{.49\textwidth}
			\centering
			\includegraphics[width=1.0\linewidth]{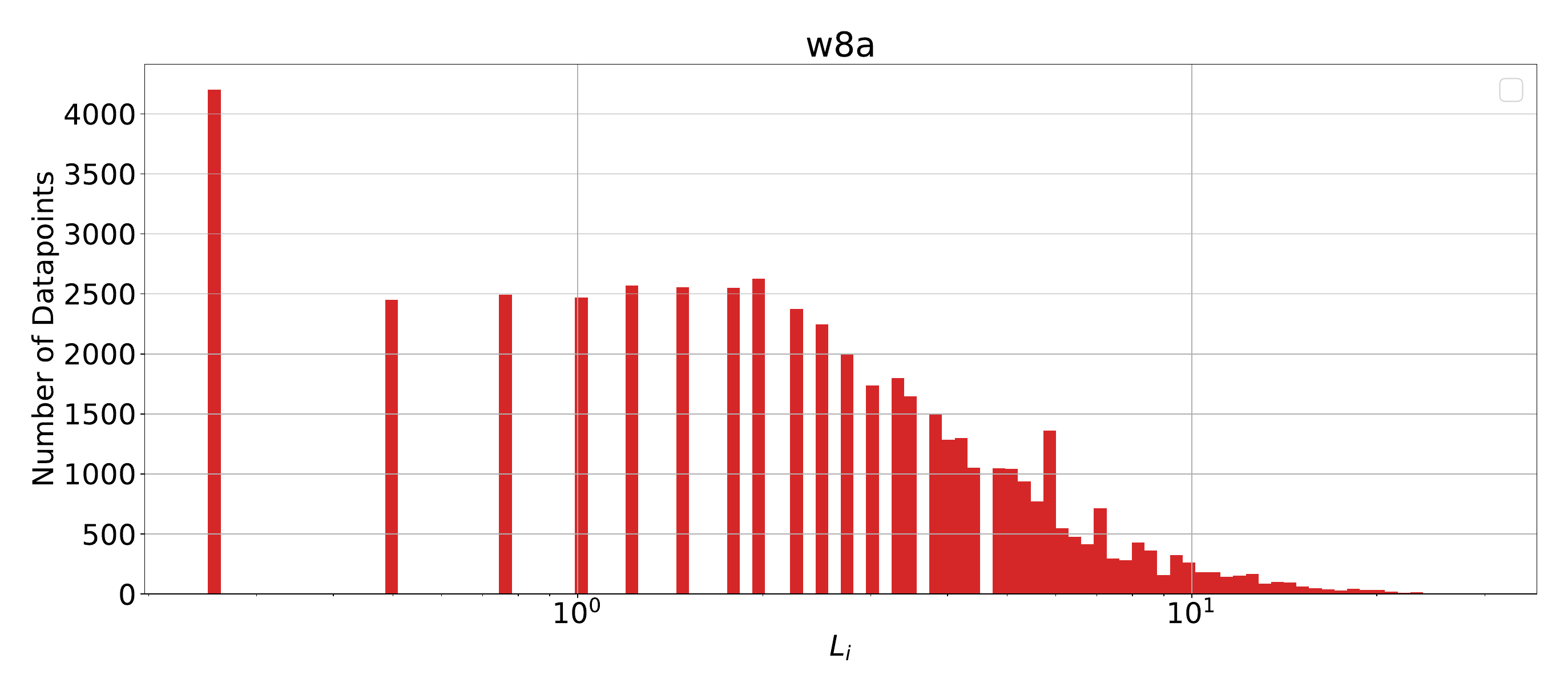}
			\caption*{}
		\end{subfigure}
		\caption{The distribution of Lipschitz constants $L_{i}$.}
		\label{ch5:fig:Li_smooth_for_log_regression_w8a_and_australian}
	\end{figure}

	We compare the samplings on practical machine learning with \dataname{LIBSVM} datasets \citep{chang2011libsvm} (under the 3-clause BSD license). Parameters of Algorithm~\ref{ch5:alg:page} are chosen as suggested in Theorem~\ref{ch5:theorem:pageab} and Corollary\ref{ch5:corollary:pageab}. 
	
	We take the parameters for \samplingname{Uniform-With-Replacement} and \samplingname{Importance} samplings from Table~\ref{ch5:tbl:AB} with $q_i = \nicefrac{L_i}{\sum_{i=1}^n L_i}.$ We consider the \modelname{logistic regression} task with a non-convex regularization \citep{wang2019spiderboost} in the following form:
	$$\min_{x_1, x_2 \in \R^d} f(x_1, x_2),$$
	where
	$$f(x_1, x_2) \eqdef \dfrac{1}{n}\sum_{i=1}^n \left[-\log\left(\dfrac{\exp\left(a_{i}^\top x_{y_{i}}\right)}{\sum_{y \in \{1, 2\}} \exp\left(a_{i}^\top x_{y}\right)}\right) + \lambda \sum_{y \in \{1, 2\}} \sum_{k = 1}^d \dfrac{\{x_{y}\}_k^2}{1 + \{x_{y}\}_k^2}\right]$$
	where $\{\cdot\}_k$ is an indexing operation,
	$a_{i} \in \R^{d}$ is the feature of a $i$\textsuperscript{th} sample, $y_{i} \in \{1, 2\}$ is the label of a $i$\textsuperscript{th} sample, constant $\lambda = 0.001.$ We fix batch size $\tau = 1$ and take \dataname{W8A} dataset (dimension $d = 300$, number of samples $n = \num[group-separator={,}]{49749}$) and \dataname{AUSTRALIAN} dataset (dimension $d = 14$, number of samples $n = \num[group-separator={,}]{690}$) from \dataname{LIBSVM}. For the \modelname{logistic regression}, the Lipschitz constants $L_i$ can be estimated. The distribution of Lipschitz constants $L_i$ across data points for those two datasets is presented in Figure~\ref{ch5:fig:Li_smooth_for_log_regression_w8a_and_australian}. We use Theorem~\ref{ch5:theorem:l_estimation} to obtain $L_{+,w}^2$ and $L_{\pm,w}^2$.
	
	\subsection{Federated learning experiments with LIBSVM dataset}
	\label{ch5:sec:experiment:libsvm_multi_node:details}
	
	In this experiment, we compare the \textit{Uniform-With-Replacement} sampling and the \textit{Importance} sampling on the \modelname{logistic regression} task from Section~\ref{ch5:sec:experiment:libsvm:details} in a distributed environment. 
	The training of the models is carried out on \dataname{AUSTRALIAN} dataset from \dataname{LIBSVM}. The dataset is reshuffled with uniform distribution, and then it is split across $n=10$ clients. 
	In all experiments, we use Algorithm~\ref{ch5:alg:page:composition} with theoretical step sizes according to Theorem~\ref{ch5:theorem:pageab:composition}.
	
	We take the parameters of the \samplingname{Uniform-With-Replacement} and \samplingname{Importance} samplings from Table~\ref{ch5:tbl:AB} with $q_i = \nicefrac{L_i}{\sum_{i=1}^n L_i}.$ 
	
	According to Algorithm~\ref{ch5:alg:page:composition}, we have the samplings $\samplefunc^t$ that sample clients, and the samplings $\samplefunc_i^t$ that sample data from the local datasets of clients.
	Algorithm~\ref{ch5:alg:page:composition} allows mixed sampling strategies that satisfy Assumption~\ref{ch5:ass:sampling}. For simplicity, we consider that the samplings $\samplefunc^t$ and $\samplefunc_i^t$ are of the same type.
	
	For the \modelname{logistic regression}, the Lipschitz constants $L_i$ and $L_{ij}$ of the gradients of functions $f_i(x)$ and $f_{ij}(x)$ can be estimated. As in Section~\ref{ch5:sec:experiment:libsvm:details}, we use Theorem~\ref{ch5:theorem:l_estimation} to obtain the constants $L_{i,+,w}^2,$ $L_{i,\pm,w}^2,$ $L_{+,w}^2$ and $L_{\pm,w}^2.$ 
	
	The results of experiments are provided in Figure~\ref{ch5:fig:test_multinode_australian}. 
	We denote by $\tau_{points}$ the batch size of the samplings $\samplefunc_i^t$ for all $i \in [n],$ and by $\tau_{clients}$ the batch size of the sampling $\samplefunc^t.$ The number of gradient calculations in Figure~\ref{ch5:fig:test_multinode_australian} stands for the total number of gradient calculations in all clients.
	
	We demonstrate results for different values of the batch sizes $\tau_{clients}$ and $\tau_{points}$. 
	As in previous experiments, the \textit{Importance} sampling has better empirical performance than the \samplingname{Uniform-With-Replacement} sampling. In addition to it, we observe that plots with small batch sizes $\tau_{points}$ converge faster.
	
	\begin{figure}
		\centering
		\begin{subfigure}{.75\textwidth}
			\centering
			\includegraphics[width=1.0\linewidth]{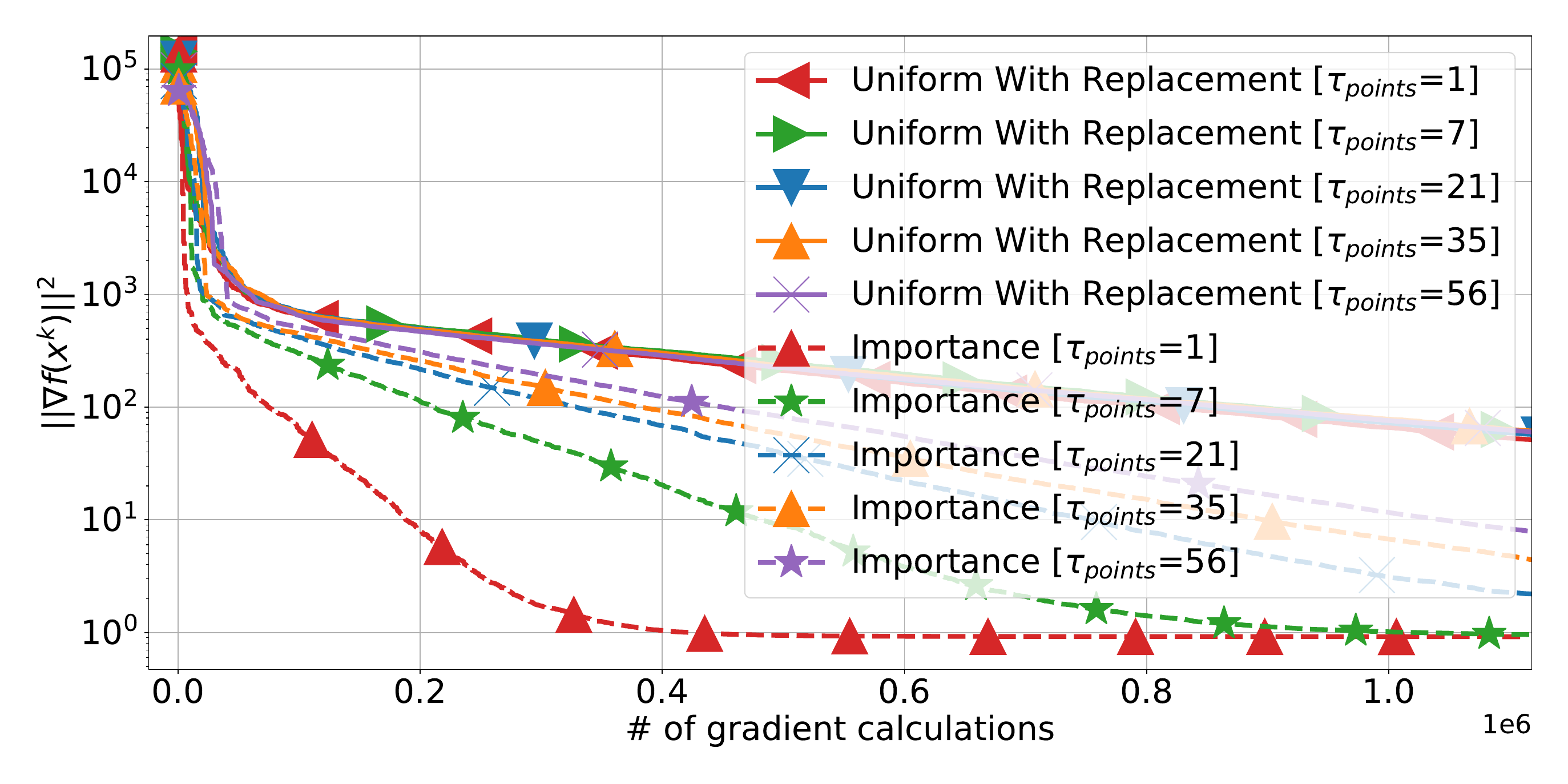}
			\caption{$\tau_{clients}=1$, \#clients $n=10$.}
		\end{subfigure}
		
		\begin{subfigure}{.75\textwidth}
			\centering
			\includegraphics[width=1.0\linewidth]{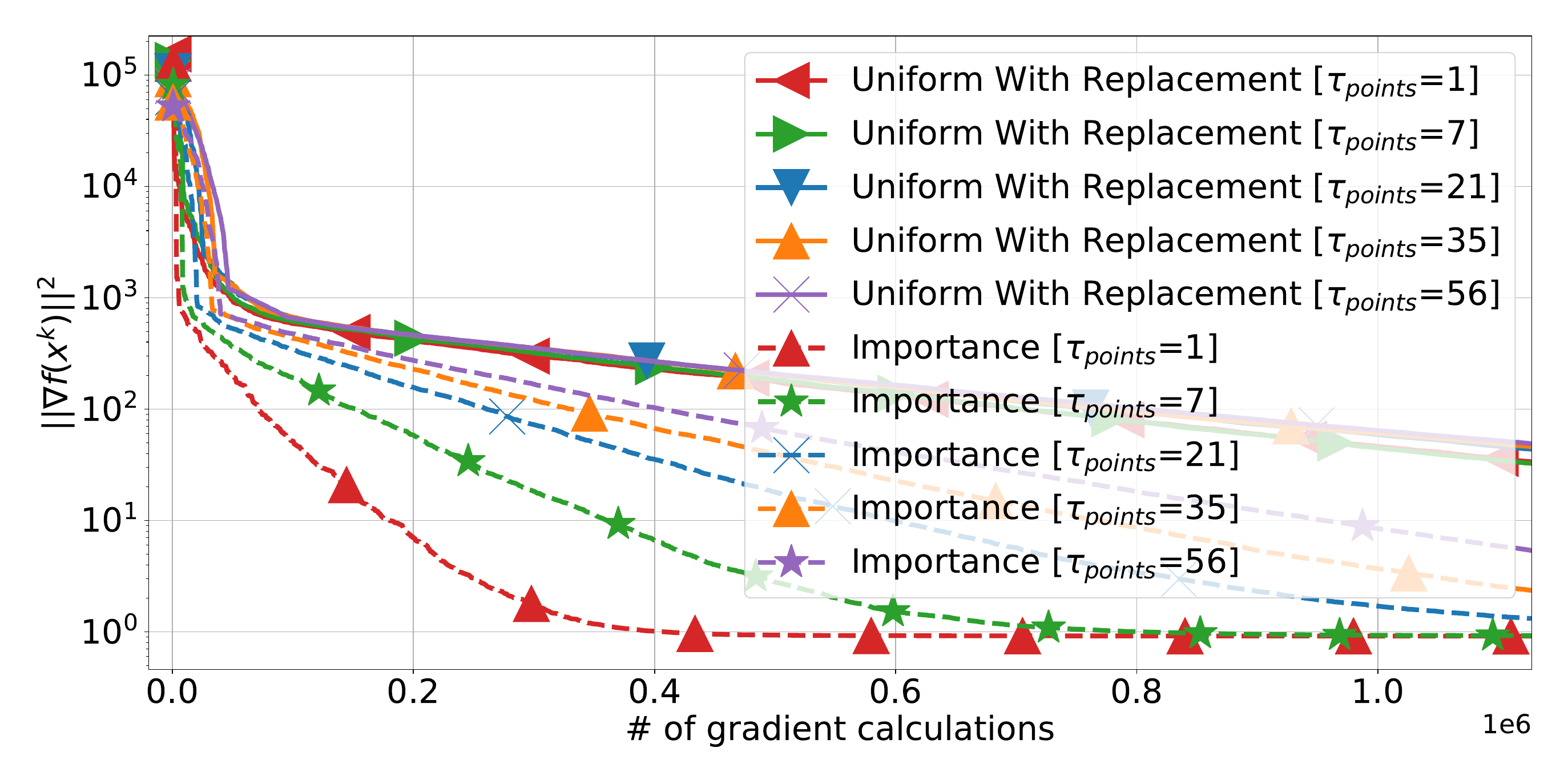}
			\caption{$\tau_{clients}=3$, \#clients $n=10$.}
		\end{subfigure}
		
		\begin{subfigure}{.75\textwidth}
			\centering
			\includegraphics[width=1.0\linewidth]{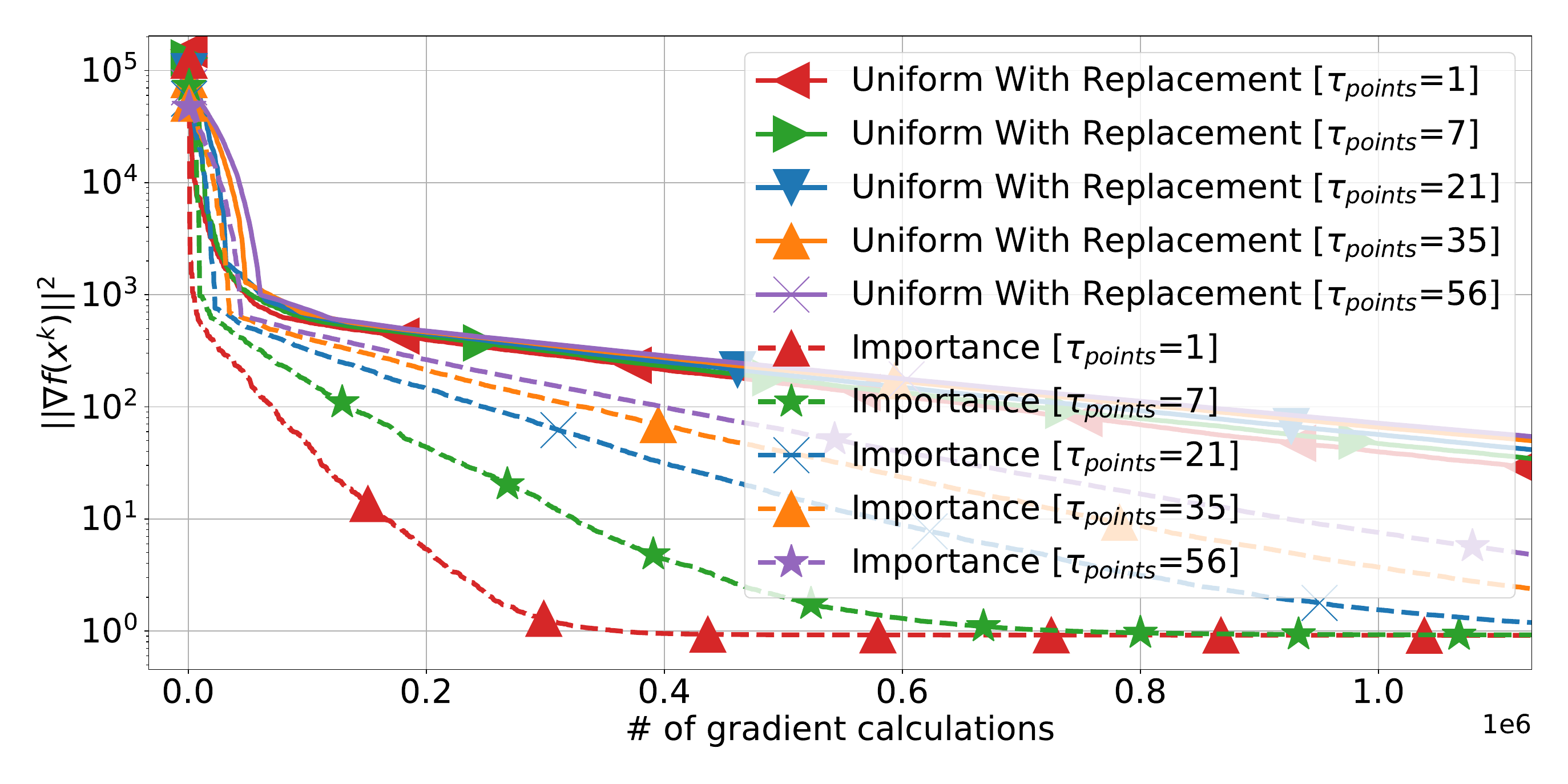}
			\caption{$\tau_{clients}=6$, \#clients $n=10$.}
		\end{subfigure}
		
		\begin{subfigure}{.75\textwidth}
			\centering
			\includegraphics[width=1.0\linewidth]{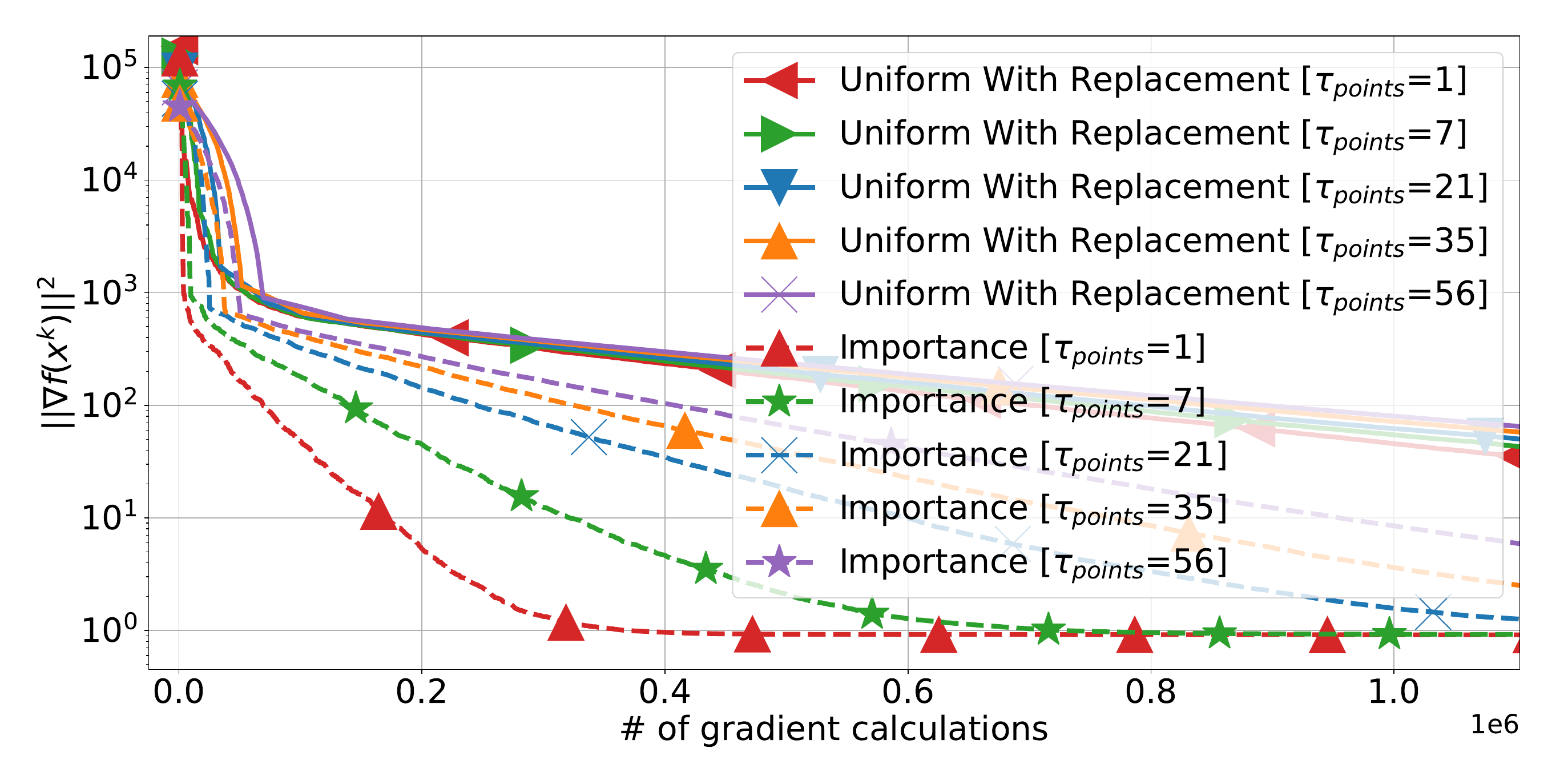}
			\caption{$\tau_{clients}=9$, \#clients $n=10$.}
		\end{subfigure}
		
		\caption{Comparison of methods on \dataname{AUSTRALIAN} dataset from \dataname{LIBSVM}.}
		\label{ch5:fig:test_multinode_australian}
	\end{figure}
	
	\subsection{Computing environment}
	\label{ch5:sec:experiment:details}
	The code was written in Python 3.6.8 using \libname{PyTorch} 1.9 \citep{paszke2019pytorch} and optimization research simulator \libname{FL\_PyTorch} \citep{burlachenko2021fl_pytorch}. The distributed environment was emulated on a machine with Intel(R) Xeon(R) Gold 6226R CPU $2.90$ GHz and 64 cores.

	\clearpage
	\addtocounter{adjsection}{1}
	\section{Auxiliary Facts}
	We use the following auxiliary fact in our proofs.
		Let us take a \textit{random vector} $\xi \in \R^d$, then
		\begin{align}
			\Exp{\norm{\xi}^2} = \Exp{\norm{\xi - \Exp{\xi}}^2} + \norm{\Exp{\xi}}^2.
			\label{ch5:auxiliary:variance_decomposition}
		\end{align}
	
	\addtocounter{adjsection}{1}
	\section{Examples of Optimization Problems}
	
	\begin{example}
		\normalfont
		\label{ch5:ex:lipt}
		For simplicity, let us assume that $n$ is even. Let us consider the optimization Problem~\eqref{ch5:eq:main_problem} with $f_i(x):\R \to \R$:
		\begin{itemize}
			\item $f_i(x) = \dfrac{a}{2} x^2 + \dfrac{b}{2} x^2$ for $i \in \{1, \cdots, n / 2\}$.
			\item $f_i(x) = -\dfrac{a}{2} x^2 + \dfrac{b}{2} x^2$ for $i \in \{n / 2 + 1, \cdots, n\}$.
			\item The $b \geq 0$.
		 \end{itemize}
		 
		 Then $f(x) = \dfrac{b}{2} x^2$ and $$L_{-}^2 = \sup_{x \neq y} \dfrac{\norm{\nabla f(x) - \nabla f(y)}^2}{\norm{x - y}^2} = b^2,$$ 
		$$L_{+}^2 = \sup_{x \neq y} \dfrac{\dfrac{1}{n}\sum_{i=1}^n\norm{\nabla f_i(x) - \nabla f_i(y)}^2}{\norm{x - y}^2} = \dfrac{1}{2}\left((a + b)^2 + (a - b)^2\right),$$
		and we can take $a$ arbitrary large.
	\end{example}
	
	\begin{example}
		\normalfont
		\label{ch5:ex:lipt_sum}
		Let us assume that $n \geq 2$ and consider the optimization Problem~\eqref{ch5:eq:main_problem} with $f_1(x) = \dfrac{b}{2} x^2$ and $f_i(x) = 0$ for $i \in \{2, \cdots, n\},$ where $x \in \R$ and $b \geq 0.$ Then $f(x) = \dfrac{b}{2 n} x^2,$$$L_{-} = \sup_{x \neq y} \dfrac{\norm{\nabla f(x) - \nabla f(y)}}{\norm{x - y}} = \dfrac{b}{n},$$
		$$\dfrac{1}{n}\sum_{i=1}^n L_i = \dfrac{1}{n}\sup_{x \neq y} \dfrac{\norm{\nabla f_1(x) - \nabla f_1(y)}^2}{\norm{x - y}^2} = \dfrac{b}{n},$$
		and
		$$L_{+} = \sqrt{\sup_{x \neq y} \dfrac{\dfrac{1}{n}\sum_{i=1}^n\norm{\nabla f_i(x) - \nabla f_i(y)}^2}{\norm{x - y}^2}} = \dfrac{b}{\sqrt{n}}.$$
	\end{example}
	
	\begin{example}
		\normalfont
		\label{ch5:ex:groups}
		Let us consider the optimization Problem~\eqref{ch5:eq:main_problem:group} with $$f(x) = \dfrac{1}{g} \sum_{i=1}^g \dfrac{1}{m} \sum_{j=1}^{m} f_{ij}(x)$$ and $f_{ij}(x) = \dfrac{b_i}{2} x^2$ for all $i \in [g]$ and $j \in [m]$, where $x \in \R$ and $b_1 \geq 0$ and $b_i = 0$ for all $i \in \{2, \dots, g\}.$
		Then $f(x) = \dfrac{b_1}{2 g} x^2,$$$L_{-} = \sup_{x \neq y} \dfrac{\norm{\nabla f(x) - \nabla f(y)}}{\norm{x - y}} = \dfrac{b_1}{g},$$		
		\begin{eqnarray*}
		L_{\pm}^2 &=& \sup_{x \neq y} \dfrac{\dfrac{1}{g m}\sum_{i=1}^g \sum_{j=1}^m \norm{\nabla f_{ij}(x) - \nabla f_{ij}(y)}^2 - \norm{\nabla f(x) - \nabla f(y)}^2}{\norm{x - y}^2} \\
		&=& \left(\dfrac{1}{g} - \dfrac{1}{g^2}\right) b_1^2,
		\end{eqnarray*}
		and 
		$$L_{i,\pm}^2 = \sup_{x \neq y} \dfrac{\dfrac{1}{m}\sum_{j=m}^n\norm{\nabla f_{ij}(x) - \nabla f_{ij}(y)}^2 - \norm{\nabla f_i(x) - \nabla f_i(y)}^2}{\norm{x - y}^2} = 0 \quad \forall i \in [n].$$
		
		Substituting the smoothness constants to the complexity $N_{\textnormal{uniform}}$ from Section~\ref{ch5:sec:theorems} and $N_{\textnormal{uniform}}$ from Section~\ref{ch5:sec:sampling_in_sampling}, one can show that
		$$N_{\textnormal{uniform}} = \Theta\left(n + \dfrac{\Delta_0\max\{\sqrt{n} L_{\pm}, L_{-}\}}{\varepsilon}\right) = \Theta\left(n + \dfrac{\Delta_0\sqrt{n} b_1}{\varepsilon \sqrt{g}}\right)$$
		and
		$$N_{\textnormal{group}} = \Theta\left(n + \dfrac{\Delta_0 \max\left\{\sqrt{n} \sqrt{\dfrac{1}{g}\sum_{i = 1}^g L_{i,\pm}^2}, g L_{-}\right\}}{\varepsilon}\right) = \Theta\left(n + \dfrac{\Delta_0 b_1}{\varepsilon}\right).$$ The complexity $N_{\textnormal{group}}$ is $\dfrac{\sqrt{n}}{\sqrt{g}}$ times better than the complexity $N_{\textnormal{uniform}}.$
	\end{example}
	
	\addtocounter{adjsection}{1}
	\section{Missing Proofs}
	
	\begin{lemma}\label{ch5:descentlemma}
		Suppose that Assumption~\ref{ch5:ass:lipschitz_constant} holds and let $x^{t+1} = x^{t}-\gamma g^{t} .$ Then for any $g^{t} \in \mathbb{R}^{d}$ and $\gamma>0$, we have
		\begin{eqnarray}
			\label{ch5:eq:14frompage}
			f(x^{t+1}) &\leq& f(x^{t})-\dfrac{\gamma}{2}\norm{\nabla f(x^{t})}^{2}-\left(\dfrac{1}{2 \gamma}-\dfrac{L_{-}}{2}\right)\norm{x^{t+1}-x^{t}}^{2} \notag \\
			&& \qquad +\dfrac{\gamma}{2}\norm{g^{t}-\nabla f(x^{t})}^{2}.
		\end{eqnarray}
	\end{lemma}
	\begin{proof}
		Using Assumption~\ref{ch5:ass:lipschitz_constant}, we have 
		\begin{align*}
			f(x^{t+1}) &\leq f(x^t) + \inp{\nabla f(x^t)}{x^{t+1} - x^{t}} + \dfrac{L_{-}}{2} \norm{x^{t+1} - x^{t}}^2 \\
			&= f(x^t) - \gamma \inp{\nabla f(x^t)}{g^t} + \dfrac{L_{-}}{2} \norm{x^{t+1} - x^{t}}^2.
		\end{align*}
		Next, due to $-\inp{x}{y} = \dfrac{1}{2}\norm{x - y}^2 - \dfrac{1}{2}\norm{x}^2 - \dfrac{1}{2}\norm{y}^2,$ we obtain 
		\begin{align*}
			f(x^{t+1}) \leq f(x^t) -\dfrac{\gamma}{2} \norm{\nabla f(x^t)}^2 - \left(\dfrac{1}{2\gamma} - \dfrac{L_{-}}{2}\right) \norm{x^{t+1} - x^{t}}^2 + \dfrac{\gamma}{2}\norm{g^t - \nabla f(x^t)}^2.
		\end{align*}
	\end{proof}
	
	\THEOREMCONVERGENCEPAGE*
	
	\begin{proof}
		We start with the estimation of the variance of the noise:
		\begin{align*}
			&\Exp{\norm{g^{t+1} - \nabla f(x^{t+1})}^2} \\
			&= (1 - p) \Exp{\norm{g^t + \samplefunc^t\left(\{\nabla f_i(x^{t+1}) - \nabla f_i(x^{t})\}_{i=1}^n\right) - \nabla f(x^{t+1})}^2} \\
			&= (1 - p) \norm{\samplefunc^t\left(\{\nabla f_i(x^{t+1}) - \nabla f_i(x^{t})\}_{i=1}^n\right) - \left(\nabla f(x^{t+1}) - \nabla f(x^{t})\right)}^2 \\
			& \qquad + (1 - p) \norm{g^t - \nabla f(x^{t})}^2,
		\end{align*}
		where we used the unbiasedness of the sampling.
		Using Assumption~\ref{ch5:ass:sampling}, we have
		\begin{align*}
			&\Exp{\norm{g^{t+1} - \nabla f(x^{t+1})}^2} \\
			&\leq (1 - p) \left(A \sum_{i = 1}^n \dfrac{1}{n^2w_i}\norm{\nabla f_i(x^{t+1}) - \nabla f_i(x^{t})}^2 - B \norm{\nabla f(x^{t+1}) - \nabla f(x^{t})}^2\right)\\
			&\quad + (1 - p) \norm{g^t - \nabla f(x^{t})}^2.
		\end{align*}
		Using the definition of $L_{+,w}$ and $L_{\pm,w}$, we get
		\begin{equation}\label{ch5:eq:15frompage}
			\begin{aligned}
				&\Exp{\norm{g^{t+1} - \nabla f(x^{t+1})}^2} \\
				&\leq (1 - p) \left(A \sum_{i = 1}^n\dfrac{1}{n^2w_i} \norm{\nabla f_i(x^{t+1}) - \nabla f_i(x^{t})}^2 - B \norm{\nabla f(x^{t+1}) - \nabla f(x^{t})}^2\right)\\
				&\quad + (1 - p) \norm{g^t - \nabla f(x^{t})}^2 \\
				&= (1 - p) \Bigg(\left(A - B\right) \left(\sum_{i = 1}^n \dfrac{1}{n^2w_i}\norm{\nabla f_i(x^{t+1}) - \nabla f_i(x^{t})}^2\right) \\
				&\quad + B \left(\sum_{i = 1}^n \dfrac{1}{n^2w_i}\norm{\nabla f_i(x^{t+1}) - \nabla f_i(x^{t})}^2 - \norm{\nabla f(x^{t+1}) - \nabla f(x^{t})}^2\right)\Bigg)\\
				&\quad + (1 - p) \norm{g^t - \nabla f(x^{t})}^2 \\
				&\leq (1 - p) \left(\left(A - B\right) L_{+,w}^2 + B L_{\pm,w}^2 \right)\norm{x^{t+1} - x^t}^2 \\
				& \quad + (1 - p) \norm{g^t - \nabla f(x^{t})}^2.
			\end{aligned}
		\end{equation}
		
		We now continue the proof using Lemma \ref{ch5:descentlemma}. 
		
		We add \eqref{ch5:eq:14frompage} with $\dfrac{\gamma}{2p}\times$ \eqref{ch5:eq:15frompage}, and take expectation to get 
		\begin{equation}\label{ch5:eq:19frompage}
			\begin{aligned}
				&\Exp{f(x^{t+1})-f^{*}+\dfrac{\gamma}{2p}\normsq{g^{t+1}-\nabla f(x^{t+1})}}\\
				&\leq \Exp{f\left(x^{t}\right)-f^{*}-\dfrac{\gamma}{2}\left\|\nabla f\left(x^{t}\right)\right\|^{2}-\left(\dfrac{1}{2 \gamma}-\dfrac{L_{-}}{2}\right)\left\|x^{t+1}-x^{t}\right\|^{2}+\dfrac{\gamma}{2}\left\|g^{t}-\nabla f\left(x^{t}\right)\right\|^{2}} \\
				&\quad+\dfrac{\gamma}{2 p} \Exp{(1-p)\left\|g^{t}-\nabla f\left(x^{t}\right)\right\|^{2}+(1 - p) \left(\left(A - B\right) L_{+,w}^2 + B L_{\pm,w}^2 \right)\left\|x^{t+1}-x^{t}\right\|^{2}} \\
				&=\Exp{f\left(x^{t}\right)-f^{*}+\dfrac{\gamma}{2 p}\left\|g^{t}-\nabla f\left(x^{t}\right)\right\|^{2}-\dfrac{\gamma}{2}\left\|\nabla f\left(x^{t}\right)\right\|^{2}\right.\\
					&\left.\quad-\left(\dfrac{1}{2 \gamma}-\dfrac{L_{-}}{2}-\dfrac{(1-p)\gamma}{2 p}\left(\left(A - B\right) L_{+,w}^2 + B L_{\pm,w}^2 \right)\right)\left\|x^{t+1}-x^{t}\right\|^{2}}\\
				&\leq \Exp{f\left(x^{t}\right)-f^{*}+\dfrac{\gamma}{2 p}\left\|g^{t}-\nabla f\left(x^{t}\right)\right\|^{2}-\dfrac{\gamma}{2}\left\|\nabla f\left(x^{t}\right)\right\|^{2}},
			\end{aligned}
		\end{equation}
		where the last inequality holds due to $$\dfrac{1}{2 \gamma}-\dfrac{L_{-}}{2}-\dfrac{(1-p)\gamma}{2 p}\left(\left(A - B\right) L_{+,w}^2 + B L_{\pm,w}^2 \right) \geq 0$$ by choosing step size
		$$
		\gamma \leq \left(L_- + \sqrt{\dfrac{1 - p}{p} \left(\left(A - B\right)L_{+,w}^2 + B L_{\pm,w}^2\right)}\right)^{-1}.
		$$
		Now, if we define $$\Phi_{t} \eqdef f\left(x^{t}\right)-f^{*}+\dfrac{\gamma}{2 p}\left\|g^{t}-\nabla f\left(x^{t}\right)\right\|^{2},$$ then \eqref{ch5:eq:19frompage} can be written in the form
		$$
		\Exp{\Phi_{t+1}}\leq \Exp{\Phi_{t}}-\dfrac{\gamma}{2} \Exp{\left\|\nabla f\left(x^{t}\right)\right\|^{2}}.
		$$
		Summing up from $t=0$ to $T-1$, we get
		$$
		\Exp{\Phi_{T}} \leq \Exp{\Phi_{0}}-\dfrac{\gamma}{2} \sum_{t=0}^{T-1} \Exp{\left\|\nabla f\left(x^{t}\right)\right\|^{2}}.
		$$
		Then according to the output of the algorithm, $\widehat{x}_{T}$ is randomly chosen from $\left\{x^{t}\right\}_{t \in[T]}$ and $$\Phi_{0}=f\left(x^{0}\right)-f^{*}+\dfrac{\gamma}{2 p} \| g^{0}-\nabla f\left(x^{0}\right) \|^{2},$$
		$$\Delta_{0} \stackrel{\text { def }}{=} f\left(x^{0}\right)-f^{*},$$ 
		we have
		$$
		\Exp{\left\|\nabla f\left(\widehat{x}_{T}\right)\right\|^{2}}\leq \dfrac{2 \Delta_{0}}{\gamma T}.
		$$
	\end{proof}
	
	\THEOREMCONVERGENCEPAGEPL*
	
	\begin{proof}
		From the proof of Theorem~\ref{ch5:theorem:pageab}, we know that
		\begin{align}
			\Exp{\norm{g^{t+1} - \nabla f(x^{t+1})}^2} &\leq (1 - p) \left(\left(A - B\right) L_{+,w}^2 + B L_{\pm,w}^2 \right)\norm{x^{t+1} - x^t}^2 \nonumber \\
			& \qquad + (1 - p) \norm{g^t - \nabla f(x^{t})}^2.
			\label{ch5:eq:47frompage}
		\end{align}
		
		Using Lemma \ref{ch5:descentlemma}, we add \eqref{ch5:eq:14frompage} with $\dfrac{\gamma}{p}\times$ \eqref{ch5:eq:47frompage}, and take expectation to get 
		\begin{equation*}
			\begin{aligned}
				&\Exp{f(x^{t+1})-f^{*}+\dfrac{\gamma}{p}\normsq{g^{t+1}-\nabla f(x^{t+1})}}\\
				&\leq \Exp{f\left(x^{t}\right)-f^{*}-\dfrac{\gamma}{2}\left\|\nabla f\left(x^{t}\right)\right\|^{2}-\left(\dfrac{1}{2 \gamma}-\dfrac{L_{-}}{2}\right)\left\|x^{t+1}-x^{t}\right\|^{2}+\dfrac{\gamma}{2}\left\|g^{t}-\nabla f\left(x^{t}\right)\right\|^{2}} \\
				&\quad+\dfrac{\gamma}{p} \Exp{(1-p)\left\|g^{t}-\nabla f\left(x^{t}\right)\right\|^{2}+(1 - p) \left(\left(A - B\right) L_{+,w}^2 + B L_{\pm,w}^2 \right)\left\|x^{t+1}-x^{t}\right\|^{2}} \\
				&=\Exp{f\left(x^{t}\right)-f^{*}+\left(1 - \dfrac{p}{2}\right)\dfrac{\gamma}{p}\left\|g^{t}-\nabla f\left(x^{t}\right)\right\|^{2}-\dfrac{\gamma}{2}\left\|\nabla f\left(x^{t}\right)\right\|^{2}\right.\\
					&\left.\quad-\left(\dfrac{1}{2 \gamma}-\dfrac{L_{-}}{2}-\dfrac{(1-p)\gamma}{p}\left(\left(A - B\right) L_{+,w}^2 + B L_{\pm,w}^2 \right)\right)\left\|x^{t+1}-x^{t}\right\|^{2}}\\
				&\leq \Exp{f\left(x^{t}\right)-f^{*}+\left(1 - \dfrac{p}{2}\right)\dfrac{\gamma}{p}\left\|g^{t}-\nabla f\left(x^{t}\right)\right\|^{2}-\dfrac{\gamma}{2}\left\|\nabla f\left(x^{t}\right)\right\|^{2}},
			\end{aligned}
		\end{equation*}
		where the last inequality holds due to $$\dfrac{1}{2 \gamma}-\dfrac{L_{-}}{2}-\dfrac{(1-p)\gamma}{p}\left(\left(A - B\right) L_{+,w}^2 + B L_{\pm,w}^2 \right) \geq 0$$ by choosing step size
		$$
		\gamma \leq \left(L_- + \sqrt{\dfrac{2(1 - p)}{p} \left(\left(A - B\right)L_{+,w}^2 + B L_{\pm,w}^2\right)}\right)^{-1}.
		$$
		Next, using Assumption~\ref{ch5:ass:pl} and $$\gamma \leq \dfrac{p}{2\mu},$$
		we have
		\begin{equation*}
			\begin{aligned}
				&\Exp{f(x^{t+1})-f^{*}+\dfrac{\gamma}{p}\normsq{g^{t+1}-\nabla f(x^{t+1})}}\\
				&\leq \left(1 - \gamma \mu\right) \Exp{ f\left(x^{t}\right)-f^{*} + \dfrac{\gamma}{p}\left\|g^{t}-\nabla f\left(x^{t}\right)\right\|^{2}}.
			\end{aligned}
		\end{equation*}
		Unrolling the recursion and considering that $g^0 = \nabla f(x^0),$ we can complete the proof of the theorem.
	\end{proof}
	
	\addtocounter{adjsection}{1}
	\section{Derivations of the Parameters for the Samplings}
	
	\subsection{{Nice} sampling}
	\label{ch5:subsec:0}
	
	Let $S$ be a random subset uniformly chosen from $[n]$ with a fixed cardinality $\tau$. 
	Let us fix $a_1, \dots, a_n \in \R^d.$ A sampling $$\samplefunc(a_1, \dots, a_n) \eqdef \dfrac{1}{n}\sum_{i \in S} \dfrac{a_i}{p_i}$$ is called the \samplingname{Nice} sampling, where $p_i \eqdef \Prob(i \in S).$ 
	
	Let us bound $\Exp{\norm{\samplefunc(a_1, \dots, a_n) - \dfrac{1}{n}\sum_{i = 1}^n a_i}^2}$ and find parameters from Assumption~\ref{ch5:ass:sampling}. Note that $|\samplefunc| = |S| = \tau.$
	We introduce auxiliary random variables
	\begin{equation*}
		\chi_i\eqdef\begin{cases}
			1&i\in S\\
			0&\text{otherwise.}
		\end{cases}.
	\end{equation*}
	Due to $p_i = \Prob\left(i \in S\right) = \dfrac{\tau}{n},$ we have
	\begin{equation*}
		\begin{aligned}
			\Exp{\norm{\dfrac{1}{n}\sum_{i \in S} \dfrac{a_i}{p_i}}^2}&=\Exp{\normsq{\dfrac{1}{\tau}\sum_{i=1}^n\chi_i a_i}}\\
			&=\dfrac{1}{\tau^2}\sum_{i=1}^n\Exp{\normsq{\chi_ia_i}}+\dfrac{1}{\tau^2}\sum_{i\neq j}\Exp{\inner{\chi_ia_i}{\chi_ja_j}}\\
			&=\dfrac{1}{\tau^2}\sum_{i=1}^n\Exp{\chi_i}\normsq{a_i}+\dfrac{1}{\tau^2}\sum_{i\neq j}\Exp{\inner{\chi_i}{\chi_j}}\inner{a_i}{a_j}\\
			&=\dfrac{1}{n\tau}\sum_{i=1}^n\normsq{a_i}+\dfrac{\tau-1}{n(n-1)\tau}\sum_{i\neq j}\inner{a_i}{a_j}\\
			&=\dfrac{1}{n\tau}\sum_{i=1}^n\normsq{a_i}+\dfrac{\tau-1}{n(n-1)\tau}\left(\normsq{\sum_{i=1}^na_i}-\sum_{i=1}^n\normsq{a_i}\right)\\
			&=\dfrac{n-\tau}{\tau(n-1)}\dfrac{1}{n}\sum_{i=1}^n\normsq{a_i}+\dfrac{\tau-1}{n(n-1)\tau}\normsq{\sum_{i=1}^na_i},
		\end{aligned}
	\end{equation*}
	where we use $\Exp{\chi_i^2}=\Exp{\chi_i}=\dfrac{\tau}{n}$ and $\Exp{\chi_i\chi_j}=\dfrac{\tau(\tau-1)}{n(n-1)},$ when $i\neq j$.
	
	Finally, we have 
	\begin{equation*}
		\begin{aligned}
			\Exp{\norm{\dfrac{1}{n}\sum_{i \in S} \dfrac{a_i}{p_i} - \dfrac{1}{n}\sum_{i = 1}^n a_i}^2}&=\Exp{\norm{\dfrac{1}{n}\sum_{i \in S} \dfrac{a_i}{p_i}}^2} - \norm{\dfrac{1}{n}\sum_{i = 1}^n a_i}^2\\
			&=\dfrac{n-\tau}{\tau(n-1)}\dfrac{1}{n}\sum_{i=1}^n\normsq{a_i}+\dfrac{\tau-1}{n(n-1)\tau}\normsq{\sum_{i=1}^na_i} \\
			&\qquad -\normsq{\dfrac{1}{n}\sum_{i=1}^na_i}\\
			&=\dfrac{n-\tau}{\tau(n-1)}\left(\dfrac{1}{n}\sum_{i=1}^n\normsq{a_i}-\normsq{\dfrac{1}{n}\sum_{i=1}^na_i}\right).
		\end{aligned}
	\end{equation*}
	Thus we have $A=B=\dfrac{n-\tau}{\tau(n-1)}$ and $w_i=\dfrac{1}{n}$ for all $i \in [n].$
	
	\subsection{{Independent} sampling}
	\label{ch5:subsec:2}
	
	Let us define independent and identically distributed (i.i.d.) random variables
	\begin{equation*}
		\chi_i=\begin{cases}
			1&\text{with probability} \ p_i\\
			0&\text{with probability} \ 1-p_i,\\
		\end{cases}.
	\end{equation*}
	for all $i \in [n]$ and take $S\eqdef\{i \in [n] \,|\, \chi_i = 1\}.$ We now fix $a_1, \dots, a_n \in \R^d.$ A sampling $$\samplefunc(a_1, \dots, a_n) \eqdef \dfrac{1}{n}\sum_{i \in S} \dfrac{a_i}{p_i}$$ is called the \samplingname{Independent} sampling, where $p_i \eqdef \Prob(i \in S).$ We get
	\begin{equation*}\label{ch5:eq:ss}
		\begin{aligned}
			\Exp{\norm{\dfrac{1}{n}\sum_{i \in S} \dfrac{a_i}{p_i} - \dfrac{1}{n}\sum_{i = 1}^n a_i}^2}&=\Exp{\normsq{\dfrac{1}{n}\sum_{i=1}^n\dfrac{1}{p_i}\chi_ia_i}}-\normsq{\dfrac{1}{n}\sum_{i=1}^na_i}\\
			&=\sum_{i=1}^n\dfrac{\Exp{\chi_i}}{n^2p_i^2}\normsq{a_i}+ \sum_{i \neq j} \dfrac{\Exp{\chi_i} \Exp{\chi_j}}{n^2 p_i p_j} \inp{a_i}{a_j} \\
			& \qquad -\normsq{\dfrac{1}{n}\sum_{i=1}^na_i}\\
			&=\sum_{i=1}^n\dfrac{1}{n^2p_i}\normsq{a_i}+\dfrac{1}{n^2}\left(\normsq{\sum_{i=1}^na_i}-\sum_{i=1}^n\normsq{a_i}\right) \\
			& \qquad -\normsq{\dfrac{1}{n}\sum_{i=1}^na_i}\\
			&=\dfrac{1}{n^2}\sum_{i=1}^n\left(\dfrac{1}{p_i}-1\right)\normsq{a_i}.
		\end{aligned}
	\end{equation*}
	Thus we have $A=\dfrac{1}{\sum_{i=1}^n\dfrac{p_i}{1-p_i}},$ $B=0$ and $w_i=\dfrac{\dfrac{p_i}{1-p_i}}{\sum_{i=1}^n\dfrac{p_i}{1-p_i}}$ for all $i \in [n].$
	
	\subsection{{Importance} and {uniform-with-replacement} sampling}
	\label{ch5:subsec:3}
	
	Let us fix $\tau > 0.$ For all $k \in [\tau],$ we define i.i.d. random variables 
	\begin{equation*}
		\chi_{k}=\begin{cases}
			1& \text{with probability} \ q_1\\
			2&\text{with probability} \ q_2\\
			\quad &\vdots\\
			n&\text{with probability}\ q_n,
		\end{cases} 
	\end{equation*}
	where $(q_1, \dots, q_n) \in \mathcal{S}^n$ (simple simplex). 
	A sampling 
	\begin{equation*}
		\samplefunc(a_1, \dots, a_n) \eqdef \dfrac{1}{\tau} \sum_{k=1}^{\tau} \dfrac{a_{\chi_{k}}}{n q_{\chi_{k}}}
	\end{equation*} is called the \samplingname{Importance} sampling.
	The \samplingname{Importance} sampling reduces to the \samplingname{Uniform-With-Replacement} sampling when $q_i = \nicefrac{1}{n}$ for all $i \in [n].$ Note that $|\samplefunc| \leq \tau.$ 
	
	Let us bound the variance
	\begin{align*}
		\Exp{\norm{\dfrac{1}{\tau} \sum_{k=1}^{\tau} \dfrac{a_{\chi_{k}}}{n q_{\chi_{k}}} - \dfrac{1}{n}\sum_{i=1}^n a_i}^2} &= \dfrac{1}{\tau^2}\sum_{k=1}^{\tau} \Exp{\norm{\dfrac{a_{\chi_{k}}}{n q_{\chi_{k}}} - \dfrac{1}{n}\sum_{i=1}^n a_i}^2} \\
		&\quad+\dfrac{1}{\tau^2}\sum_{k\neq k'} \Exp{\inp{\dfrac{a_{\chi_{k}}}{n q_{\chi_{k}}} - \dfrac{1}{n}\sum_{i=1}^n a_i}{\dfrac{a_{\chi_{k'}}}{n q_{\chi_{k'}}} - \dfrac{1}{n}\sum_{i=1}^n a_i}}.
	\end{align*}
	Using the independence and unbiasedness of the random variables, the last term vanishes, and we get
	\begin{align*}
		\Exp{\norm{\dfrac{1}{\tau} \sum_{k=1}^{\tau} \dfrac{a_{\chi_{k}}}{n q_{\chi_{k}}} - \dfrac{1}{n}\sum_{i=1}^n a_i}^2} &= \dfrac{1}{\tau^2}\sum_{k=1}^{\tau} \Exp{\norm{\dfrac{a_{\chi_{k}}}{n q_{\chi_{k}}} - \dfrac{1}{n}\sum_{i=1}^n a_i}^2} \\
		&\overset{\eqref{ch5:auxiliary:variance_decomposition}}{=} \dfrac{1}{\tau^2}\sum_{k=1}^{\tau} \Exp{\norm{\dfrac{a_{\chi_{k}}}{n q_{\chi_{k}}}}^2} - \dfrac{1}{\tau} \norm{\dfrac{1}{n}\sum_{i=1}^n a_i}^2 \\
		&= \dfrac{1}{\tau} \sum_{i=1}^n q_i \norm{\dfrac{a_{i}}{n q_{i}}}^2 - \dfrac{1}{\tau} \norm{\dfrac{1}{n}\sum_{i=1}^n a_i}^2\\
		&= \dfrac{1}{\tau} \left(\dfrac{1}{n}\sum_{i=1}^n \dfrac{1}{n q_i} \norm{a_{i}}^2 - \norm{\dfrac{1}{n}\sum_{i=1}^n a_i}^2\right).
	\end{align*}
	Thus we have $A=B=\dfrac{1}{\tau},$ and $w_i=q_i$ for all $i \in [n].$
	
	\subsection{{Extended nice} sampling}
	\label{ch5:subsec:1}
	
	In this section, we analyze the extension of \samplingname{Nice} sampling. First, we $l_i$ times repeat each vector $a_i$; then, we use the \samplingname{Nice} sampling. We define
	
	\begin{equation*}
		\tilde{a}_i\eqdef\begin{cases}
			\dfrac{\sum_{j=1}^nl_j}{nl_1}a_1&1\leq i\leq l_1\\
			\dfrac{\sum_{j=1}^nl_j}{nl_2}a_2&l_1+1\leq i\leq l_1+l_2\\
			\quad&\vdots\\
			\dfrac{\sum_{j=1}^nl_j}{nl_n}a_n&\sum_{j=1}^{n-1}l_j\leq i\leq \sum_{j=1}^nl_j,\\
		\end{cases},
	\end{equation*}
	where $a_i \in \R^d$ and $l_i \geq 1$ for all $i \in [n].$
	Then we have
	\begin{equation*}
		\dfrac{1}{n}\sum_{i=1}^n a_i(x)=\dfrac{1}{N}\sum_{i=1}^{N}\tilde{a}_i(x),
	\end{equation*}
	where $N\eqdef\sum_{j=1}^nl_j$. Also, we denote $N_k \eqdef \sum_{j=1}^kl_j.$
	
	For some $\tau > 0,$ we apply the \samplingname{Nice} sampling method:
	\begin{equation*}
		\samplefunc(a_1, \dots, a_n) \eqdef \dfrac{1}{N}\sum_{i \in S} \dfrac{\tilde{a}_i}{p_i} = \sum_{i=1}^N\dfrac{1}{\tau}\chi_i\tilde{a}_i,
	\end{equation*}
	where \begin{equation*}
		\chi_i=\begin{cases}
			1&i\in S\\
			0&\text{otherwise}
		\end{cases},\quad p_i = \Prob\left(i \in S\right),
	\end{equation*}
	and $S$ is a random set with cardinality $\tau$ from $[N]$. The sampling $\samplefunc(a_1, \dots, a_n)$ is called the \samplingname{Extended Nice} sampling.
	
	We are now ready to bound the variance. Using the results for the \samplingname{Nice} sampling, we obtain
	\begin{align*}
		&\Exp{\norm{\samplefunc(a_1, \dots, a_n) - \dfrac{1}{n}\sum_{i=1}^n a_i(x)}^2}\\
		&=\Exp{\norm{\samplefunc(a_1, \dots, a_n) - \dfrac{1}{N}\sum_{i=1}^{N}\tilde{a}_i(x)}^2}\\
		&=\dfrac{n-\tau}{\tau(n-1)}\dfrac{1}{N}\sum_{i=1}^N\normsq{\tilde{a}_i}-\dfrac{n-\tau}{\tau(n-1)}\normsq{\dfrac{1}{N}\sum_{i=1}^N\tilde{a}_i}\\
		&=\dfrac{n-\tau}{\tau(n-1)}\left(\dfrac{1}{N}\left(\dfrac{N}{nl_1}\right)^2\sum_{i=1}^{N_1}\normsq{{a}_1}+\dfrac{1}{N}\left(\dfrac{N}{nl_2}\right)^2\sum_{i=N_1+1}^{N_2}\normsq{{a}_2}\right.\\
		&\left.\quad+\cdots+\dfrac{1}{N}\left(\dfrac{N}{nl_n}\right)^2\sum_{i=N_{n-1}+1}^{N}\normsq{{a}_n}\right)-\dfrac{n-\tau}{\tau(n-1)}\normsq{\dfrac{1}{n}\sum_{i=1}^na_i}\\
		&={\dfrac{n-\tau}{\tau(n-1)}{\left(\dfrac{N}{nl_1}\dfrac{1}{n}\normsq{a_1}+\dfrac{N}{nl_2}\dfrac{1}{n}\normsq{a_2}+\cdots+\dfrac{N}{nl_n}\dfrac{1}{n}\normsq{a_n}\right)}-\dfrac{n-\tau}{\tau(n-1)}\normsq{\dfrac{1}{n}\sum_{i=1}^na_i}}\\
		&{=\dfrac{n-\tau}{\tau(n-1)}\left(\sum_{i=1}^n\dfrac{1}{n^2w_i}\normsq{a_i}\right)-\dfrac{n-\tau}{\tau(n-1)}\normsq{\dfrac{1}{n}\sum_{i=1}^na_i}}
	\end{align*}
	where $w_i=\dfrac{l_i}{N}$. Thus we have $A=B = \dfrac{n-\tau}{\tau(n-1)}$ and $w_i=\dfrac{l_i}{N}$ for $i \in [n].$
	
	\addtocounter{adjsection}{1}
	\section{The Optimal Choice of $w_i$}
	\label{ch5:sec:optimal_weights}
	Let us consider $L_{+,w}^2$ and $L_{\pm,w}^2.$ In Section~\ref{ch5:sec:assumptions}, we show that one can take $$L_{+,w}^2 = L_{\pm,w}^2 = \dfrac{1}{n} \sum_{i=1}^n \dfrac{1}{n w_i} L_i^2.$$ Let us minimize $\dfrac{1}{n} \sum_{i=1}^n \dfrac{1}{n w_i} L_i^2$ with respect to the weights $w_i$ such that $$w_1, \dots, w_n \geq 0,\quad \sum_{i=1}^n w_i = 1.$$ Using the method of Lagrange multipliers, we can construct a Lagrangian
	\begin{equation*}
		\cL(w,\lambda)\eqdef\dfrac{1}{n} \sum_{i=1}^n \dfrac{1}{n w_i} L_i^2-\lambda \left(\sum_{i=1}^nw_i-1\right).
	\end{equation*}
	Next, we calculate partial derivatives
	\begin{align*}
		&\dfrac{\partial\cL}{\partial w_i}=-\dfrac{1}{n^2w_i^2}L_i^2-\lambda=0 \forall i \in [n]
	\end{align*}
	and get
	\begin{equation*}
		w_i^2=-\dfrac{L_i^2}{n^2\lambda}.
	\end{equation*}
	Using $\sum_{i=1}^n w_i = 1,$ we can show that the weights $w_i^*=\dfrac{L_i}{\sum_{i=1}^nL_i}$ are the solutions of the minimization problem. Moreover, 
	\begin{equation*}
		\dfrac{1}{n} \sum_{i=1}^n \dfrac{1}{n w_i^*} L_i^2 = \left(\dfrac{1}{n}\sum_{i=1}^nL_i\right)^2.
	\end{equation*}
	
	\addtocounter{adjsection}{1}
	\section{Missing Proofs: The Composition of Samplings}
	
	\begin{restatable}{lemma}{LEMMASAMPLINGINSAMPLING}
		\label{ch5:lemma:sampling}
		Let us assume that a random sampling function $\samplefunc$ satisfies Assumption~\ref{ch5:ass:sampling} with some $A, B$ and weights $w_i,$ and a random sampling function $\samplefunc_i$ satisfy Assumption~\ref{ch5:ass:sampling} with some $A_i, B_i$ and weights $w_{ij}$ for all $i \in [n].$ Moreover, $B \leq 1.$ Then
		\begin{align*}
			&\Exp{\norm{\samplefunc\left(\samplefunc_1\left(a_{11}, \dots, a_{1 m_1}\right), \dots, \samplefunc_n\left(a_{n1}, \dots, a_{n m_n}\right)\right) - \dfrac{1}{n}\sum_{i = 1}^n \left(\dfrac{1}{m_i} \sum_{j=1}^{m_i} a_{ij}\right)}^2} \\
			&\leq \dfrac{1}{n}\sum_{i = 1}^n \left(\dfrac{A}{n w_i} + \dfrac{(1 - B)}{n}\right)\left(\dfrac{A_i}{m_i}\sum_{j = 1}^{m_i} \dfrac{1}{m_i w_{ij}}\norm{a_{ij}}^2 - B_i \norm{\dfrac{1}{m_i} \sum_{j=1}^{m_i} a_{ij}}^2\right) \\
			&\quad + \dfrac{A}{n}\sum_{i = 1}^n \dfrac{1}{n w_i}\norm{\dfrac{1}{m_i} \sum_{j=1}^{m_i} a_{ij}}^2 - B \norm{\dfrac{1}{n}\sum_{i = 1}^n \left(\dfrac{1}{m_i} \sum_{j=1}^{m_i} a_{ij}\right)}^2,
		\end{align*}
		where $a_{ij} \in \R^d$ for all $j \in [m_i]$ and $i \in [n].$
	\end{restatable}
	
	\begin{proof}
		We denote $\widehat{a}_i \eqdef \samplefunc_i\left(a_{i1}, \dots, a_{i m_i}\right)$ and $a_i \eqdef \dfrac{1}{m_i}\sum_{j=1}^{m_i} a_{ij}.$ Using \eqref{ch5:auxiliary:variance_decomposition}, we have
		\begin{align*}
			&\Exp{\norm{\samplefunc\left(\widehat{a}_1, \dots, \widehat{a}_n\right) - \dfrac{1}{n}\sum_{i = 1}^n a_i}^2} \\
			&=\Exp{\ExpSub{S}{\norm{\samplefunc\left(\widehat{a}_1, \dots, \widehat{a}_n\right) - \dfrac{1}{n}\sum_{i = 1}^n a_i}^2}} \\
			&=\Exp{\ExpSub{S}{\norm{\samplefunc\left(\widehat{a}_1, \dots, \widehat{a}_n\right) - \dfrac{1}{n}\sum_{i = 1}^n \widehat{a}_i}^2}} + \Exp{\norm{\dfrac{1}{n}\sum_{i = 1}^n \widehat{a}_i - \dfrac{1}{n}\sum_{i = 1}^n a_i}^2}.
		\end{align*}
		Next, using Assumption~\ref{ch5:ass:sampling} for the sampling $\samplefunc,$ we get
		\begin{align*}
			\Exp{\norm{\samplefunc\left(\widehat{a}_1, \dots, \widehat{a}_n\right) - \dfrac{1}{n}\sum_{i = 1}^n a_i}^2} &\leq A \dfrac{1}{n}\sum_{i = 1}^n \dfrac{1}{n w_i}\Exp{\norm{\widehat{a}_i}^2} - B \Exp{\norm{\dfrac{1}{n}\sum_{i = 1}^n \widehat{a}_i}^2} \\
			&\quad + \Exp{\norm{\dfrac{1}{n}\sum_{i = 1}^n \widehat{a}_i - \dfrac{1}{n}\sum_{i = 1}^n a_i}^2}.
		\end{align*}
		Due to \eqref{ch5:auxiliary:variance_decomposition}, we obtain
		\begin{align*}
			&\Exp{\norm{\samplefunc\left(\widehat{a}_1, \dots, \widehat{a}_n\right) - \dfrac{1}{n}\sum_{i = 1}^n a_i}^2} \\
			&\leq A \dfrac{1}{n}\sum_{i = 1}^n \dfrac{1}{n w_i}\Exp{\norm{\widehat{a}_i - a_i}^2} + A \dfrac{1}{n}\sum_{i = 1}^n \dfrac{1}{n w_i}\norm{a_i}^2 \\
			&\quad - B \Exp{\norm{\dfrac{1}{n}\sum_{i = 1}^n \widehat{a}_i - \dfrac{1}{n}\sum_{i = 1}^n a_i}^2} - B \norm{\dfrac{1}{n}\sum_{i = 1}^n a_i}^2\\
			&\quad + \Exp{\norm{\dfrac{1}{n}\sum_{i = 1}^n \widehat{a}_i - \dfrac{1}{n}\sum_{i = 1}^n a_i}^2} \\
			&= A \dfrac{1}{n}\sum_{i = 1}^n \dfrac{1}{n w_i}\Exp{\norm{\widehat{a}_i - a_i}^2} + A \dfrac{1}{n}\sum_{i = 1}^n \dfrac{1}{n w_i}\norm{a_i}^2 \\
			&\quad + (1 - B) \Exp{\norm{\dfrac{1}{n}\sum_{i = 1}^n \widehat{a}_i - \dfrac{1}{n}\sum_{i = 1}^n a_i}^2} - B \norm{\dfrac{1}{n}\sum_{i = 1}^n a_i}^2 \\
			&= A \dfrac{1}{n}\sum_{i = 1}^n \dfrac{1}{n w_i}\Exp{\norm{\widehat{a}_i - a_i}^2} + A \dfrac{1}{n}\sum_{i = 1}^n \dfrac{1}{n w_i}\norm{a_i}^2 \\
			&\quad + \dfrac{(1 - B)}{n^2} \sum_{i=1}^n \Exp{\norm{\widehat{a}_i - a_i}^2} - B \norm{\dfrac{1}{n}\sum_{i = 1}^n a_i}^2 \\
			&= \dfrac{1}{n}\sum_{i = 1}^n \left(\dfrac{A}{n w_i} + \dfrac{(1 - B)}{n}\right)\Exp{\norm{\widehat{a}_i - a_i}^2} + A \dfrac{1}{n}\sum_{i = 1}^n \dfrac{1}{n w_i}\norm{a_i}^2 - B \norm{\dfrac{1}{n}\sum_{i = 1}^n a_i}^2. 
		\end{align*}
		Using Assumption~\ref{ch5:ass:sampling} for the samplings $\samplefunc_i,$ we have
		\begin{align*}
			&\Exp{\norm{\samplefunc\left(\widehat{a}_1, \dots, \widehat{a}_n\right) - \dfrac{1}{n}\sum_{i = 1}^n a_i}^2} \\
			&\leq \dfrac{1}{n}\sum_{i = 1}^n \left(\dfrac{A}{n w_i} + \dfrac{(1 - B)}{n}\right)\left(A_i \dfrac{1}{m_i}\sum_{j = 1}^{m_i} \dfrac{1}{m_i w_{ij}}\norm{a_{ij}}^2 - B_i \norm{a_i}^2\right) \\
			&\quad + A \dfrac{1}{n}\sum_{i = 1}^n \dfrac{1}{n w_i}\norm{a_i}^2 - B \norm{\dfrac{1}{n}\sum_{i = 1}^n a_i}^2.
		\end{align*}
	\end{proof}
	
	\THEOREMCONVERGENCEPAGECOMPOSITION*
	
	\begin{proof}
		We start with the estimation of the variance of the noise:
		\begin{align*}
			&\Exp{\norm{g^{t+1} - \nabla f(x^{t+1})}^2} \\
			&= (1 - p) \Exp{\norm{g^t + \samplefunc\left(\left\{\samplefunc_i\left(\{\nabla f_{ij}(x^{t+1}) - \nabla f_{ij}(x^{t})\}_{j=1}^{m_i}\right)\right\}_{i=1}^n\right) - \nabla f(x^{t+1})}^2} \\
			&= (1 - p) \norm{\samplefunc\left(\left\{\samplefunc_i\left(\{\nabla f_{ij}(x^{t+1}) - \nabla f_{ij}(x^{t})\}_{j=1}^{m_i}\right)\right\}_{i=1}^n\right) - \left(\nabla f(x^{t+1}) - \nabla f(x^{t})\right)}^2 \\
			& \qquad + (1 - p) \norm{g^t - \nabla f(x^{t})}^2,
		\end{align*}
		where we used the unbiasedness of the composition of samplings.
		Using Lemma~\ref{ch5:lemma:sampling}, we have
		\begin{align*}
			&\Exp{\norm{g^{t+1} - \nabla f(x^{t+1})}^2} \\
			&\leq (1 - p) \left(\dfrac{1}{n}\sum_{i = 1}^n \left(\dfrac{A}{n w_i} + \dfrac{(1 - B)}{n}\right) \right.\\
			& \qquad \qquad \left. \left(\dfrac{A_i}{m_i}\sum_{j = 1}^{m_i} \dfrac{1}{m_i w_{ij}}\norm{\nabla f_{ij}(x^{t+1}) - \nabla f_{ij}(x^{t})}^2 - B_i \norm{\nabla f_{i}(x^{t+1}) - \nabla f_{i}(x^{t})}^2\right) \right.\\
			&\left.\qquad \qquad + \dfrac{A}{n}\sum_{i = 1}^n \dfrac{1}{n w_i}\norm{\nabla f_{i}(x^{t+1}) - \nabla f_{i}(x^{t})}^2 - B \norm{\nabla f(x^{t+1}) - \nabla f(x^{t})}^2\right)\\
			&\quad + (1 - p) \norm{g^t - \nabla f(x^{t})}^2.
		\end{align*}
		Using Definitions~\ref{ch5:def:weighted_local_lipschitz_constant}, \ref{ch5:def:weighted_hessian_varaince}, \ref{ch5:def:weighted_local_lipschitz_constant:local} and \ref{ch5:def:weighted_hessian_varaince:local}, we get
		\begin{align*}
			&\Exp{\norm{g^{t+1} - \nabla f(x^{t+1})}^2} \\
			&\leq (1 - p) \left(\dfrac{1}{n}\sum_{i = 1}^n \left(\dfrac{A}{n w_i} + \dfrac{(1 - B)}{n}\right)\left(\dfrac{A_i - B_i}{m_i}\sum_{j = 1}^{m_i} \dfrac{1}{m_i w_{ij}}\norm{\nabla f_{ij}(x^{t+1}) - \nabla f_{ij}(x^{t})}^2 \right.\right.\\
			&\left.\left.+\quad B_i \left(\dfrac{1}{m_i}\sum_{j = 1}^{m_i} \dfrac{1}{m_i w_{ij}}\norm{\nabla f_{ij}(x^{t+1}) - \nabla f_{ij}(x^{t})}^2 - \norm{\nabla f_{i}(x^{t+1}) - \nabla f_{i}(x^{t})}^2\right)\right) \right.\\
			&\left.\quad + \dfrac{A - B}{n}\sum_{i = 1}^n \dfrac{1}{n w_i}\norm{\nabla f_{i}(x^{t+1}) - \nabla f_{i}(x^{t})}^2 \right.\\
			&\left.\quad+ B\left(\dfrac{1}{n}\sum_{i = 1}^n \dfrac{1}{n w_i}\norm{\nabla f_{i}(x^{t+1}) - \nabla f_{i}(x^{t})}^2 - \norm{\nabla f(x^{t+1}) - \nabla f(x^{t})}^2\right)\right) \\
			&\quad + (1 - p) \norm{g^t - \nabla f(x^{t})}^2 \\
			&\leq (1 - p) \left(\dfrac{1}{n}\sum_{i = 1}^n \left(\dfrac{A}{n w_i} + \dfrac{(1 - B)}{n}\right)\left((A_i - B_i) L_{i,+,w_i}^2 + B_i L_{i,\pm,w_i}^2\right) \right. \\
			& \left. \qquad \qquad \qquad + (A - B)L_{+,w}^2 + B L_{\pm,w}^2\right) \norm{x^{t+1} - x^t}^2 \\
			&\quad + (1 - p) \norm{g^t - \nabla f(x^{t})}^2.
		\end{align*}
		From this point the proof of theorem repeats the proof of Theorem~\ref{ch5:theorem:pageab} with $$\dfrac{1}{n}\sum_{i = 1}^n \left(\dfrac{A}{n w_i} + \dfrac{(1 - B)}{n}\right)\left((A_i - B_i) L_{i,+,w_i}^2 + B_i L_{i,\pm,w_i}^2\right) + (A - B)L_{+,w}^2 + B L_{\pm,w}^2$$ instead of $$\left(A - B\right) L_{+,w}^2 + B L_{\pm,w}^2.$$
	\end{proof}

	\clearpage
	\addtocounter{adjsection}{1}
	\section{Artificial Quadratic Optimization Tasks}
	
	In this section, we provide algorithms that we use to generate artificial optimization tasks for experiments. Algorithm~\ref{ch5:algorithm:matrix_generation} and Algorithm~\ref{ch5:algorithm:matrix_generation_l_i} allow us to control the smoothness constants $L_{\pm}$ and $L_{i},$ accordingly, via the noise scales.
	\begin{algorithm}
		\caption{Generate Quadratic Optimization Task with Controlled $L_{\pm}$ (homogeneity).}
		\begin{algorithmic}[1]
			\label{ch5:algorithm:matrix_generation}
			\STATE \textbf{Input and Initialization:} number nodes $n$, dimension $d$, regularizer $\lambda$, and noise scale $s$.
			\FOR{$i = 1, \dots, n$}
			\STATE Generate random noises $\nu_i^s = 1 + s \xi_i^s$ and $\nu_i^b = s \xi_i^b,$ i.i.d. $\xi_i^s, \xi_i^b \sim \textnormal{NormalDistribution}(0, 1)$
			\STATE Take vector $b_i = \dfrac{\nu_i^s}{4}(-1 + \nu_i^b, 0, \cdots, 0) \in \R^{d}$
			\STATE Take the initial tridiagonal matrix
			\[\mA_i = \dfrac{\nu_i^s}{4}\left( \begin{array}{cccc}
				2 & -1 & & 0\\
				-1 & \ddots & \ddots & \\
				& \ddots & \ddots & -1 \\
				0 & & -1 & 2 \end{array} \right) \in \R^{d \times d}\]
			\ENDFOR
			\STATE Take the mean of matrices $\mA = \dfrac{1}{n}\sum_{i=1}^n \mA_i$
			\STATE Find the minimum eigenvalue $\lambda_{\min}(\mA)$
			\FOR{$i = 1, \dots, n$}
			\STATE Update matrix $\mA_i = \mA_i + (\lambda - \lambda_{\min}(\mA)) \mI$
			\ENDFOR
			\STATE Take starting point $x^0 = (\sqrt{d}, 0, \cdots, 0)$
			\STATE \textbf{Output:} matrices $\mA_1, \cdots, \mA_n$, vectors $b_1, \cdots, b_n$, starting point $x^0$
		\end{algorithmic}
	\end{algorithm}
	
	\begin{algorithm}
		\caption{Generate Quadratic Optimization Task with Controlled $L_i$.}
		\begin{algorithmic}[1]
			\label{ch5:algorithm:matrix_generation_l_i}
			\STATE \textbf{Input and Initialization:} number nodes $n$, dimension $d$, regularizer $\lambda$, and noise scale $s$.
			\FOR{$i = 1, \dots, n$}
			\STATE Generate random noises $\nu_i^s = 1 + s \xi_i^s,$ where i.i.d. $\xi_i^s \sim \textnormal{ExponentialDistribution}(1.0)$
			\STATE Generate random noises $\nu_i^b = s \xi_i^b,$ i.i.d. $\xi_i^b \sim \textnormal{NormalDistribution}(0, 1)$
			\STATE Take vector $b_i = (-\dfrac{1}{4} + \nu_i^b, 0, \cdots, 0) \in \R^{d}$
			\STATE Take the initial tridiagonal matrix
			\[\mA_i = \dfrac{\nu_i^s}{4}\left( \begin{array}{cccc}
				2 & -1 & & 0\\
				-1 & \ddots & \ddots & \\
				& \ddots & \ddots & -1 \\
				0 & & -1 & 2 \end{array} \right) \in \R^{d \times d}\]
			\ENDFOR
			\STATE Take starting point $x^0 = (\sqrt{d}, 0, \cdots, 0)$
			\STATE \textbf{Output:} matrices $\mA_1, \cdots, \mA_n$, vectors $b_1, \cdots, b_n$, starting point $x^0$
		\end{algorithmic}
	\end{algorithm}

\addtocounter{adjsection}{1}
\section{Reproducibility}

The source code for the experiments, along with a description for reproducing the experiment, can be downloaded from:
\begin{center}
	\href{https://github.com/mysteryresearcher/sampling-in-optimal-sgd}{https://github.com/mysteryresearcher/sampling-in-optimal-sgd}
	\href{https://github.com/mysteryresearcher/page_ab_fl_experiment_a3}{https://github.com/mysteryresearcher/page\_ab\_fl\_experiment\_a3}
\end{center}

\unappendix

\chapter{Compressed L2GD: Personalized FL with Compression}
\label{chapter6}

The goals and summaries of this chapter are outlined in Table \ref{ch1:tbl:algorithms} and Section~\ref{ch1:sec:overview-6}.

\section{Introduction}

We live in the era of big data, and mobile devices have become a part of our daily lives. While the training of ML models using the diverse data stored on these devices is becoming increasingly popular, the traditional data center-based approach to training them faces serious {\em privacy issues} and has to deal with {\em high communication and energy cost} associated with the transfer of data from users to the data center~\citep{dean2012large}.
{\em Federated Learning}~(FL) provides an attractive alternative to the traditional approach as it aims to train the models directly on {\em resource constrained} heterogeneous devices without any need for the data to leave them \citep{FEDLEARN,kairouz2019advances}. 

The prevalent paradigm for training FL models is empirical risk minimization, where the aim is to train a {\em single global model} using the aggregate of all the training data stored across all participating devices. Among the popular algorithms for training FL models for this formulation belong \algname{FedAVG} \citep{mcmahan17fedavg}, \algname{Local GD} \citep{FirstLocalGDHeter, khaled_lgd}, \algname{Local SGD} \citep{stich2018local, khaled_lgd, LSGDunified2020} and \algname{Shifted Local SVRG} \citep{LSGDunified2020}. All these methods require the participating devices to perform a local training procedure (for example, by taking multiple steps of some optimization algorithm) and subsequently communicate the resulting model to an orchestrating server infrastructure for aggregation; see Figure~\ref{ch6:fig:l2gd} (a). This process is repeated until a model of suitable qualities is found. For more variants of local methods and further pointers to the literature, we refer the reader to \citet{LSGDunified2020}.

\begin{figure}
	\centering
	\includegraphics[width=0.65\textwidth]{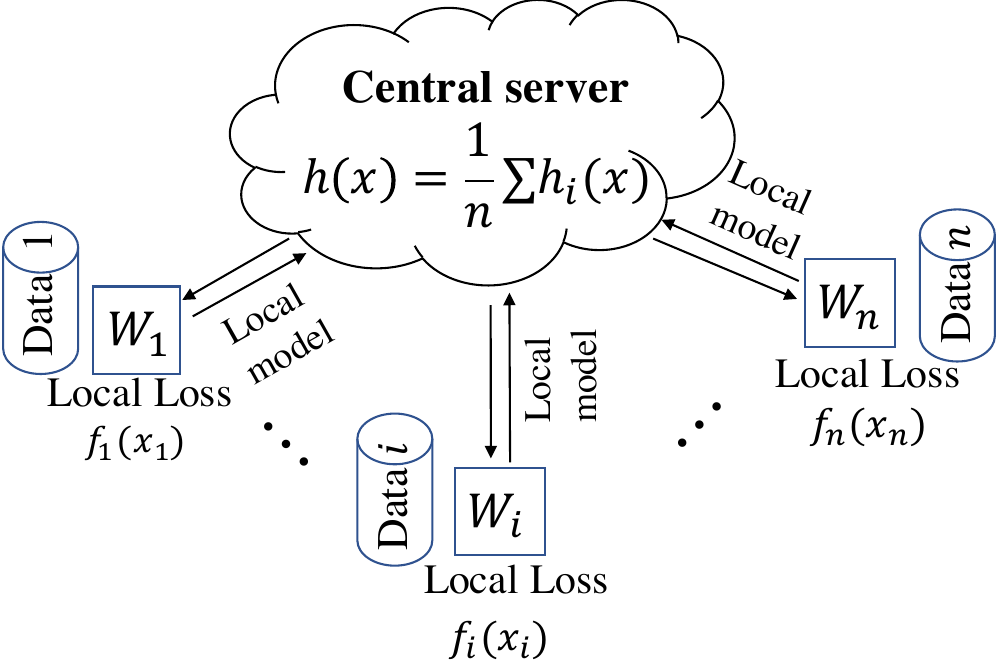}
	\caption{{Training $n$ local devices, $\{W_i\}$ on the loss, $f_i$ of their local model, $x_i$ with a central server/master node, where $h_i$ penalizes for dissimilarity between the local model, $x_i$ and the average of all local models, $\Bar{x}.$}}
	\label{ch6:fig:fl}
\end{figure}

\begin{figure}
	\centering
	
	\begin{subfigure}[ht]{0.49\textwidth}
		\includegraphics[width=0.94\textwidth]{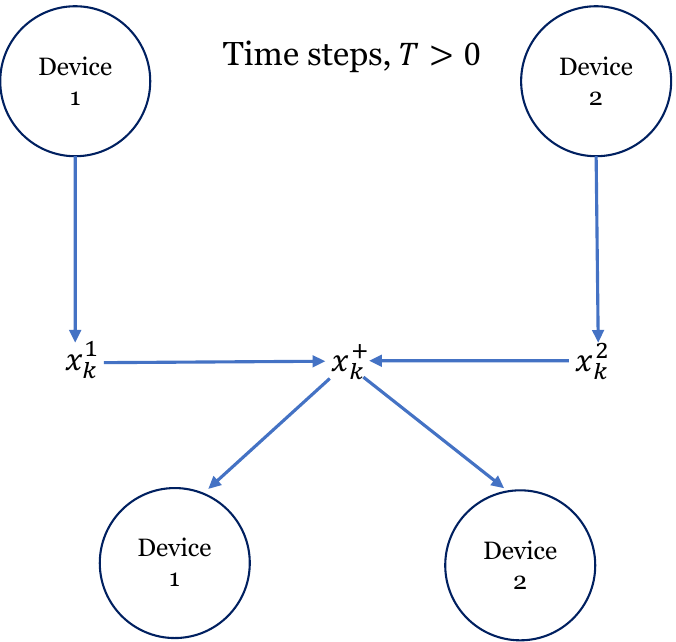}
		\caption{{ \algname{FedAVG} }}
	\end{subfigure}
	\begin{subfigure}[ht]{0.49\textwidth}
		\includegraphics[width=\textwidth]{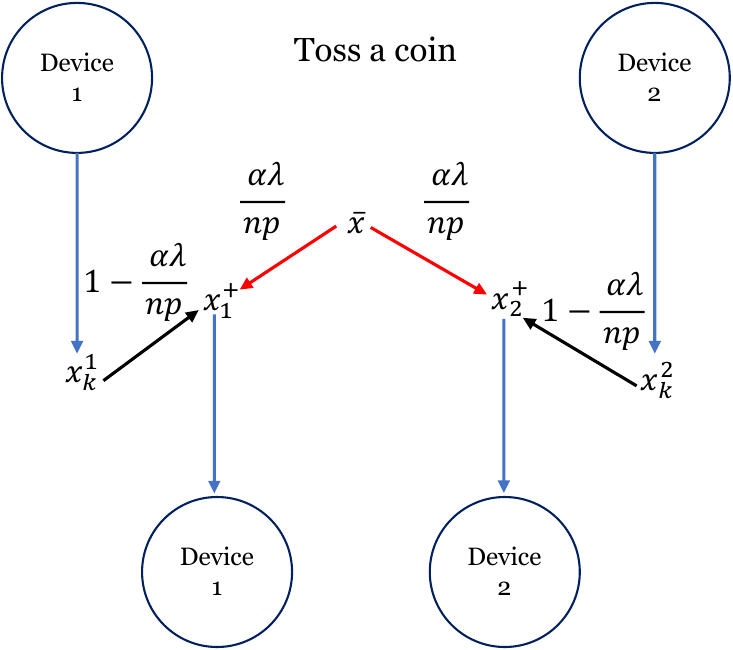}
		\caption{{ \algname{L2GD} }}
	\end{subfigure}	
	\caption{{The \algname{FedAVG} \citep{mcmahan17fedavg} and \algname{L2GD} \citep{Hanzely2020} algorithm on 2 devices. Unlike \algname{FedAVG}, \algname{L2GD} does not communicate after a fixed $T$ local steps, it communicates based on a probabilistic protocol.}}\label{ch6:fig:l2gd}
\end{figure}

\subsection{Personalized FL}
In contrast, \citet{Hanzely2020} recently introduced a new formulation of FL as an alternative to the existing ``single-model-suits-all'' approach embodied by empirical risk minimization. Their formulation explicitly aims to find a {\em personalized} model for every device; see Figure~\ref{ch6:fig:fl}. In particular, \citet{Hanzely2020} considered the formulation\footnote{\citet{esgd} considered a similar model in a different context and with different motivations.}
\begin{equation} 
	\label{ch6:eq:problem}
	\min \limits_{x\in\R^{nd}} \left[F(x) \eqdef f(x) + h(x)\right]
\end{equation}
for simultaneous training of $n$ personalized FL models $x_1,\dots,x_n\in \R^d$ for $n$ participating devices. They chose
$${f(x) \eqdef \dfrac{1}{n}\sum \limits_{i=1}^n f_i(x_i), \quad \text{and}\quad h(x) \eqdef \dfrac{1}{n}\sum\limits_{i=1}^n h_i(x),}$$ where $f_i$ represents the loss of model $x_i$ over the local data stored on device $i$. Function $h_i$ penalizes for dissimilarity between the local model $x_i$ and the average of all local models $$\Bar{x} \eqdef \dfrac{1}{n} \sum_{i=1}^n x_i,$$ and is defined to be
$$h_i(x) = \dfrac{\lambda}{2} \|x_i -\Bar{x}\|_2^2,$$ where ${\lambda>0}$ controls for the strength of this penalization. At one extreme, ${\lambda\rightarrow\infty}$ forces the local models to be equal to their average and hence mutually identical. Therefore, \eqref{ch6:eq:problem} reduces to the classical empirical risk minimization formulation
\begin{equation*} 
	{\min \limits_{z\in\R^{d}} \dfrac{1}{n} \sum \limits_{i=1}^n f_i(z).}
\end{equation*}
On the other hand, for $\lambda=0$ Problem~\eqref{ch6:eq:problem} is equivalent to each client (node) training independently using their own data only. And the $i^{\rm th}$ client solves 
\begin{equation*} 
	\min_{x_i \in\R^{d}} f_i(x_i).
\end{equation*}
By choosing $\lambda$ to a value in between these two extremes, $${0<\lambda<\infty},$$ 
we control the level of similarity among the personalized models $\{x_i\}_{i=1}^n$.


We remark that, local methods such as \algname{FedAVG} by \citet{mcmahan17fedavg} (also see similar methods in \citet{haddadpour2019local, stich2018local, wang2018adaptive, zhou2018convergence, lin2018don}), are popular for training FL models. Their main drawback in the heterogeneous setting with data and device heterogeneity is inefficient communication. \citet{Hanzely2020} solved this, and we are using their model to build our compressed, personalized FL. 

To solve Problem~\eqref{ch6:eq:problem}, \citet{Hanzely2020} proposed a {\em probabilistic} gradient descent algorithm for which they coined the name loopless local gradient descent (\algname{L2GD}). \citet{Hanzely2020} shows how \algname{L2GD} can be interpreted as a simple variant of \algname{FedAVG}, typically presented as a method for solving the standard empirical risk minimization (ERM) formulation of FL. However, alongside \citet{Hanzely2020} argue, \algname{L2GD} is better seen as an algorithm for solving the personalized FL formulation \eqref{ch6:eq:problem}. By doing so, they interpret the nature of local steps in classical FL: the role of local steps in classical FL methods is to provide personalization and not communication efficiency as was widely believed---\algname{FedAVG} can diverge on highly non-identical data partitions \citep{mcmahan17fedavg}. Instead, communication efficiency in local methods comes from their tendency to gear towards personalization, and personalized models are easier to train.

\smartparagraph{Communication compression.} We observe that the {\em \algname{L2GD} algorithm does not support any compression mechanism} for the master-worker and worker-master communication that needs to happen. This is the starting point of our work. 
\begin{center}
	{\em We believe that equipping personalized FL with fast and theoretically tractable communication compression mechanisms is an important open problem.} 
\end{center}
In distributed training of deep neural network (DNN) models, synchronous data-parallelism~\citep{dean2012large} is most widely used and adopted by mainstream Deep Learning toolkits (such as \libname{PyTorch}, \libname{TensorFlow}). However, exchanging the stochastic gradients for aggregation creates a communication bottleneck, and this results in slower training \citep{xu2021grace}. One way to save on communication costs is to use compression operators \citep{alistarh2017qsgd, DIANA2, xu2021grace}. Gradient compression techniques, such as quantization \citep{alistarh2017qsgd, signsgd, horvath2019natural, beznosikov2020biased, safaryan2021}, sparsification \citep{Suresh2017, konevcny2018randomized, aji_sparse, sahu2021rethinking, Stich-EF-NIPS2018, layer-wise, safaryan2021}, hybrid compressors \citep{Strom15}, and low-rank methods \citep{PowerSGD} have been proposed to overcome this issue~\footnote{Model compression~\citep{guo2018, DFPMCI2019} is orthogonal to gradient compression and not in the scope of this work.}.

Although recent works have introduced compression in traditional FL formulation \citep{FEDLEARN, fedpaq, artemis,9054168,amiri2020federated,kostopoulou2021deepreduce}; except \citep{horvath2019natural, artemis, Lin_EC_SGD, amiri2020federated}, others use compression only for the {\em throughput limited uplink} channel, that is, to upload the local models from the devices to the central server. But limited bandwidth in the downlink channel may pose a communication latency between the server and the devices and consequently, slow down the training; see detailed discussion in Section~\ref{ch6:sec:related_work}. As of now, no study combines {\em bidirectional} compression techniques with a probabilistic communication protocol in the FL set-up by using a mixture of a local and global model as in Problem~\eqref{ch6:eq:problem}. In this work, we combine these aspects and make subsequent contributions. 

\subsection{Related work}
\label{ch6:sec:related_work}

Numerous studies are proposed to reduce communication but not all of them are in the FL setting. In this scope, for completeness, we quote a few representatives from each class of communication-efficient \algname{SGD}.  

\citet{smith2017federated} proposed a communication-efficient primal-dual optimization that learns separate but related models for each participating device. The \algname{FedAVG} by \citet{mcmahan17fedavg} performs local steps on a subset of participating devices in an FL setting. Similar to \algname{FedAVG}, but without data and device heterogeneity, \citet{haddadpour2019local, stich2018local, wang2018adaptive, zhou2018convergence, lin2018don} independently proposed local SGD, where several local steps are taken on the participating devices before periodic communication and averaging the local models. While \algname{FedProx} \citep{li2018federated} generalizes \algname{FedAVG}, \algname{SCAFFOLD} employs variance reduction to correct local updates arising from non-independent and identically distributed (non-i.i.d.) data in \algname{FedAVG}. From the system's perspective, on \libname{TensorFlow}, \citet{bonawitz2019towards} built a FL system on mobile devices. 

Compression has also been introduced in the FL. \citet{9054168} combined universal vector quantization with FL for throughput limited uplink channel. In \algname{FedPAQ}~\citep{fedpaq}, each local device sends a compressed difference between its input and output model to the central server, after computing the local updates for a fixed number of iterations. While \citet{amiri2020federated} used a bidirectional compression in FL, \citet{artemis} combined it with a memory mechanism or error feedback \citep{Stich-EF-NIPS2018}.  

Among other communication-efficient \algname{SGDs}, parallel restarted \algname{SGD} \citep{hao2018b} reduces the number of communication rounds compared to the baseline \algname{SGD}. \citet{Farzin2018} showed that redundancy reduces residual error as compared with the baseline \algname{SGD} where all nodes can sample from the complete data and this leads to lower communication overheads. \algname{CoCoA} \citep{NIPS2014_5599}, \algname{Dane} \citep{dane} perform several local steps and hence fewer communication rounds before communicating with the other workers. Lazily Aggregated Gradient (\algname{LAG}) \citep{Chen2018LAGLA}, selects a subgroup of workers and uses their gradients, rather than obtaining a fresh gradient from each one.

In decentralized training, where the nodes only communicate with their neighbors, \citet{Koloskova2019chocosgd} implemented an {\em average consensus} where the nodes can communicate to their neighbors via a fixed communication graph. \citet{pipe_sgd} proposed \algname{Pipe-SGD}---a framework with decentralized pipelined training. 

Personalization in FL is a growing research area. \citet{arivazhagan2019federated} proposed \algname{FedPer} to mitigate statistical heterogeneity of data; also see adaptive personalized FL algorithm in \citet{deng2020adaptive}. \citet{layerwise_personalizedFL} proposed to obtain personalization in FL by using layer-wise parameters, and two-stage training; also see \citet{ma2022layer} and model personalization in \citet{shen2022cd2}. \citet{shamsian2021personalized} trained a central hypernetwork model to generate a set of personalized models for the local devices. \citet{li2021hermes} proposed \algname{Hermes}---a communication-efficient personalized FL, where each local device identifies a small subnetwork by applying the structured pruning, communicates these subnetworks to the server and the devices, the server performs the aggregation on only overlapped parameters across each subnetwork; also see \citet{pillutla2022federated} for partial model personalization in FL. \algname{DispFL} is another communication-efficient personalized FL algorithm proposed by \citet{dai2022dispfl}. In recent work, \citet{zhang2021personalized} introduce personalization by calculating optimal weighted model combinations for each client without assuming any data distribution. For a connection between personalization in FL and model-agnostic-meta-learning (\algname{MAML}), see \citet{fallah2020personalized}. Additionally, we refer to the surveys \citet{kulkarni2020survey,tan2022towards} for an overview of personalization in FL. 

\subsection{Contributions}

\begin{enumerate}
	\item \textbf{{L2GD} algorithm with bidirectional compression.} Communication compression is prevalent in recent local FL training algorithms, but these algorithms are not robust to data and device heterogeneity. \algname{L2GD} by \citet{Hanzely2020} remedies this issue by introducing personalization in FL. Integrating compression with \algname{L2GD} algorithm is a nontrivial task---unlike other FL algorithms, \algname{L2GD} does not communicate after fixed local steps, it communicates based on a probabilistic protocol; see Section~\ref{ch6:sec:L2GD} and Figure~\ref{ch6:fig:l2gd}. Additionally, due to this probabilistic protocol, the communication involves local model updates, as well as gradients; see Section~\ref{ch6:sec:L2GD}. To reduce the communication bottleneck in \algname{L2GD}, we use compression techniques on top of its probabilistic communication at both master and the participating local devices; see Section~\ref{ch6:sec:L2GDComp}. To the best of our knowledge, we are the first to integrate {\em bidirectional compression techniques} with a probabilistic communication in the FL set-up, and we call our algorithm \algname{Compressed L2GD}; see Algorithm \ref{ch6:alg:ComL2GD}.

	\item \textbf{Convergence analysis.} In Section~\ref{ch6:sec:convergence}, we prove the convergence of our \algname{Compressed L2GD} algorithm based on the most recent theoretical development, such as expected smoothness as in \citet{Gower2019}. Admittedly, convergence analysis of first-order optimization algorithms with bidirectional compression exists in the literature, see works \citet{DoubleSqueeze,horvath2019natural,amiri2020federated, layer-wise}, integrating arbitrary unbiased compressors with a probabilistic communication protocol into personalized FL, and showing convergence are nontrivial and the first one in its class. Our compressed \algname{L2GD} algorithm maintains a similar asymptotic convergence rate as the baseline vanilla \algname{SGD} without compression in both strongly convex and smooth non-convex cases; see Theorem~\ref{ch6:thm:mainconvergenceresult} and Theorem~\ref{ch6:thm:nncc} in Section~\ref{ch6:sec:convergence}. 

	\item \textbf{Optimal rate and communication.} We optimized the complexity bounds of our algorithm as a function of the parameters involved in the algorithm. This leads us to the ``optimal" setting of our algorithm. Mainly, we derived the optimal expected number of local steps to get the optimal iteration complexity and communication rounds; see Section~\ref{ch6:sec:optimalrate}. Although our analysis is based on some hard-to-compute constants in real life, for example, the Lipchitz constant, this may help the practitioners to get an insight into the iteration complexity and communication trade-off; see Theorem~\ref{ch6:thm:optimalrate} and Theorem~\ref{ch6:thm:optimalcommunication} in Section~\ref{ch6:sec:optimalrate}.

	\item \textbf{Empirical study.} We perform diverse numerical experiments on {synthetic and real datasets by using both convex and non-convex problems (using 4 DNN models) and} invoking various compression techniques; see details in Section~\ref{ch6:sec:empirical}, Table~\ref{ch6:tab:summary}. In training larger DNN models, to obtain the same global \compname{Top1} test accuracy, \algname{Compressed L2GD} reduces the communicated data volume (bits normalized by the number of local devices or clients, $\mathrm{\#bits/n}$), from $10^{15}$ to $10^{11}$, rendering approximately $10^4$ times improvement compared to \algname{FedAVG}; see Section~\ref{ch6:sec:dnn}. Moreover, \algname{L2GD} with \compname{Natural} compressor (that by design has smaller variance) empirically behaves the best and converges approximately $5$ times faster, and reaches the best accuracy on both train and the test sets compared to no-compression \algname{FedOpt} \citep{reddi2020adaptive} baseline; see  Section~\ref{ch6:sec:dnn} and Section~\ref{ch6:sec:app_nn}. These experiments validate the effect of the parameters used, and the effect of compressors, and show the efficiency of our algorithm in practice.
\end{enumerate}}

\section{Background and Preliminaries}
\label{ch6:sec:L2GD}

\smartparagraph{Notation.} For a given vector, $${x\in\R^{nd}},$$ by $x_i$ we denote the $i^{\rm th}$ subvector of $x$, and write $${x=\left(x_1^\top,\ldots,x_n^\top \right)^\top}, \mathrm{where\,}{x_i\in\R^d}.$$ 
We denote the $i^{\rm th}$ component of $x$ by $x_{(i)}$ and $\|x\|$ represents its Euclidean norm. 
By ${[n]}$ we denote the set of indexes, ${\{1,\ldots,n\}}.$
By $\E_{\xi}(\cdot)$ we define the expectation over the randomness of $\xi$ conditional to all the other potential random variables. 
The operator, $${\mathcal{C}(\cdot)\eqdef\left(\mathcal{C}_1(\cdot)^\top,\ldots,\mathcal{C}_n(\cdot)^\top \right)^\top:\R^{nd}\to\R^{nd}}$$ 
denotes a compression operator with each $\mathcal{C}_i(\cdot)$ being compatible with $x_i$. Denote ${\mQ\eqdef[\mI, \mI,\ldots,\mI]^\top\in\R^{nd \times d}}$, where $\mI$ denotes the identity matrix of $\R^{d \times d}$. With our Assumptions that we will introduce later in the chapter, the Problem~\eqref{ch6:eq:problem} has a
unique solution, which we denote by  $x^*$ and we define $\Bar{x}^*$ as $${\Bar{x}^* \eqdef \dfrac{1}{n} \sum_{i=1}^n x_i^*.}$$ 
Finally, we denote the cardinality of a set $S$ by $|S|$.

\smartparagraph{Loopless local gradient descent~(\algname{L2GD}).} We give a brief overview of the loopless local gradient descent (\algname{L2GD}) algorithm by \citet{Hanzely2020} to solve \eqref{ch6:eq:problem} as a two sum problem. At each iteration, to estimate the gradient of $F$, \algname{L2GD} samples either the gradient of $f$ or the gradient of $h$ and updates the local models via:
$${x_i^{k+1} = x_i^k - \alpha G_i(x^k), ~i=1,\ldots,n,}$$
where $G_i(x^k)$ for $~i=1,\ldots,n,$ is the $i^{\rm th}$ block of the vector $${G(x^k)=
	\left\{
	\begin{array}{ll}
		\dfrac{\nabla f(x^k)}{1-p} & \text{ with probability } 1-p, \\
		& \textbf{(Local gradient step)}
		\\
		\dfrac{ \nabla h(x^k)}{p} & \text{ with probability } p, \\
		& \textbf{(Aggregation step)}
	\end{array}
	\right.}$$
where $${0< p <1},$$ and $\nabla f(x^k)$ is the gradient of $f$ at $x^k,$ and $${\nabla_i h(x^k)=\dfrac{  \lambda}{n} \left( x_i^k - \Bar{x}^k \right)}$$
is  the $i^{\rm th}$ block of the gradient of $h$ at $x^k$.  

In this approach, there  is a hidden communication between the local devices because in aggregation steps they need the average of the local models. That is, the communication occurs when the devices switch from a local gradient step to an aggregation step. Note that there is no need for communication between the local devices when they switch from an aggregation step to a local gradient step. There is also no need for communication after two consecutive aggregation steps since the average of the local models does not change in this case. If $k$ and $k+1$ are both aggregation steps, we have 
$$\bar{x}^{k+1} = \dfrac{1}{n}\sum_{i=1}^n x_i^{k+1} = \dfrac{1}{n}\sum_{i=1}^n x_i^{k} - \dfrac{  \alpha \lambda}{n} \dfrac{1}{n}\sum_{i=1}^n \left( x_i^k - \Bar{x}^k \right)  = \bar{x}^{k} -  \dfrac{\alpha \lambda}{n} \cdot 0 = \bar{x}^{k}.$$

\section{Compressed L2GD}
\label{ch6:sec:L2GDComp}

\begin{algorithm}
	\caption{\algname{Compressed L2GD}.}
	\label{ch6:alg:ComL2GD}
	\begin{algorithmic}
		\STATE {\bfseries Input:} step size $\eta>0$,  probability $p$
		\STATE {\bfseries Initialize:} $\{x_i^0\}_{i=1,\ldots,n}$, $\xi_{-1}=1$, $\Bar{x}^{-1}=\dfrac{1}{n} \sum_{i=1}^n x_i^0$
		\FOR{$k=0,1,2,\ldots$}
		\STATE \textbf{Draw:} $\xi_{k}=1$ with probability $p$\;
		\IF{$\xi_k = 0 $}
		\STATE \textbf{on all devices:} $x_i^{k+1}=x_i^k - \dfrac{\eta}{n(1-p)}\nabla f_i(x_i^k)$ for $i\in[n]$\;
		\ELSE 
		\IF {$\xi_{k-1}=0$}
		\STATE  \textbf{on all devices:} Compress $x_i^k$ to $\mathcal{C}_i(x_i^k)$ and communicate $\mathcal{C}_i(x_i^k)$ to the master\;
		\STATE		 \textbf{ on master:}\;
		\begin{enumerate}[label={}]
			\item 1. Receive $\mathcal{C}_i(x_i^k)$ from all devices $i\in[n]$\;
			\item 2. Compute $\Bar{y}^k \eqdef \dfrac{1}{n}\sum_{j=1}^n \mathcal{C}_j(x_j^k)$\;
			\item 3. Compress $\Bar{y}^k$ to $\mathcal{C}_M(\Bar{y}^k)$\;
			\item 4. Communicate  $\mathcal{C}_M(\Bar{y}^k)$ to all devices\;
		\end{enumerate}		
		\STATE      \textbf{ on all devices:} Perform aggregation step $x_i^{k+1}=x_i^k  - \dfrac{\eta \lambda }{np} \left( x_i^k - \mathcal{C}_M(\Bar{y}^k)\right)$\;    
		\ELSE
		\STATE  	 \textbf{ on all devices:} 
		\begin{enumerate}[label={}]
			\item a. $\Bar{x}^k = \Bar{x}^{k-1}$\;
			\item b. Perform aggregation step $x_i^{k+1}=x_i^k  - \dfrac{\eta \lambda}{n p} \left( x_i^k - \Bar{x}^k\right)$\;
		\end{enumerate}
		\ENDIF
		\ENDIF
		\ENDFOR
	\end{algorithmic}
\end{algorithm}

Now, we are all set to describe \algname{Compressed L2GD} Algorithm~\ref{ch6:alg:ComL2GD} for solving Problem~\eqref{ch6:eq:problem}. We start by defining how the compression operates in our set-up. 

\subsection{Compressed communication}
Recall that the original \algname{L2GD} algorithm has a probabilistic communication protocol---the devices do not communicate after every fixed number of local steps. The communication occurs when the devices switch from a local gradient step to an aggregation step. Therefore, instead of using the compressors in a fixed time stamp (after every ${T>0}$ iteration, say), each device $i$, requires to compress its local model $x_i$ when it needs to communicate it to the master, based on the probabilistic protocol. We assume that device $i$ uses the compression operator, $${\mathcal{C}_i(\cdot): \R^d \to \R^d}.$$ Moreover, another compression happens when the master needs to communicate with the devices. We assume that the master uses the compression operator, $${\mathcal{C}_M(\cdot): \R^d \to \R^d}.$$
Therefore, the compression is used in uplink and downlink channels similar to \citet{layer-wise,horvath2019natural}, but occurs in a probabilistic fashion. There exists another subtlety---although the model parameters (either from the local devices or the global aggregated model) are communicated in the network for training the FL model via \algname{Compressed L2GD}, the compressors that we use in this work are the compressors used for gradient compression in distributed DNN training; see \citet{xu2021grace}. 

\subsection{The algorithm}
Note that, in each iteration ${k\ge0}$, there exists a random variable, ${\xi_k\in\{0,1\}}$ with $${\Prob(\xi_k=1)=p}, \quad {\Prob(\xi_k=0)=1-p}.$$
If ${\xi_k=0}$, all local devices at iteration $k$, perform one local gradient step. Otherwise (if ${\xi_k=1}$), all local devices perform an aggregation step. However, to perform an aggregation step, the local devices need to know the average of the local models. If the previous iteration (i.e, ${{k-1}^{\rm th}}$ iteration) was an aggregation step (i.e, ${\xi_{k-1}=1}$), then at the current iteration the local devices can use the same average as the one at iteration ${k-1}$ (recall, the average of the local models does not change after two consecutive aggregation steps). Otherwise, a communication happens with the master to compute the average. In this case, each local device $i$ compresses its local model $x_i^k$ to $\mathcal{C}_i(x_i^k)$ and communicates the result to the master. The master computes the average based on the compressed values of local models:
$${\Bar{y}^k \eqdef \dfrac{1}{n}\sum \limits_{j=1}^n \mathcal{C}_j(x_j^k),}$$
then it compresses $\Bar{y}^k$ to $\mathcal{C}_M(\Bar{y}^k)$ by using a compression operator at the master's end and communicates it back to the local devices. The local devices further perform an aggregation step by using $\mathcal{C}_M(\Bar{y}^k)$ instead of the exact average. This process continues until convergence.  
From Algorithm \ref{ch6:alg:ComL2GD}, 
we have, for $i=1,\ldots,n: $
$$x_i^{k+1}= x_i^k - \eta G_i(x^k),$$ 
where 
$${G_i(x^{k})=
	\left\{
	\begin{array}{lll}
		\dfrac{\nabla f_i\left(x_i^k\right)}{n(1-p)} & \text{ if } \xi_k=0
		\\
		\dfrac{  \lambda }{np}\left( x_i^k - \mathcal{C}_M(\Bar{y}^k)\right) 
		& \text{ if } \xi_k=1 ~\& ~\xi_{k-1}=0, \\
		\dfrac{  \lambda}{n p} \left( x_i^k - \Bar{x}^k \right) & \text{ if } \xi_k=1 ~\& ~\xi_{k-1}=1.\\
	\end{array}
	\right.}$$
	
For the pseudocode of \algname{Compressed L2GD}, see Algorithm \ref{ch6:alg:ComL2GD}.

\section{Convergence Analysis}\label{ch6:sec:convergence}
With the above setup, we now present the convergence results of Algorithm \ref{ch6:alg:ComL2GD}. Detailed proofs can be found in Section~\ref{ch6:sec:Proofs}.

\subsection{Assumptions}
We make the following general assumptions in this chapter. 
\begin{assumption} \label{ch6:ass:compression}
	For $i=1,\ldots,n$:
	\begin{itemize}
		\item The compression operator, ${\mathcal{C}_i(\cdot):\R^d \to \R^d}$ is unbiased, 
		$$ \E_{\mathcal{C}_i}\left[\mathcal{C}_i(x)\right] = x, \quad \forall x\in \R^d.$$
		\item There exists 
		constant, $\omega_i>0$ such that the variance of $\mathcal{C}_i$ is bounded as follows:
		$${\E_{\mathcal{C}_i} \left[\|\mathcal{C}_i(x) -  x\|^2\right] \le \omega_i \|  x\|^2, \forall x \in \R^d.}$$
		\item The operators, $\{\mathcal{C}_i(\cdot)\}_{i=1}^n$ are independent from each other, and independent from $\xi_k$, for all $k\ge 0$.
		\item The compression operator, $\mathcal{C}_M(\cdot)$ is unbiased, independent from $\{\mathcal{C}_i\}_{i=1}^n$ and has compression factor, $\omega_M$.
	\end{itemize}
\end{assumption}
From the above assumption we conclude that for all $x \in \R^d$, we have $${\E_{\mathcal{C}_i}
	\left[\|\mathcal{C}_i(x)\|^2\right] \le (1+\omega_i) \|  x\|^2.}$$
The following Lemma characterizes the compression factor, $\omega$ of the joint compression operator, ${\mathcal{C}(\cdot)= \left(\mathcal{C}_1(\cdot)^\top,\ldots,\mathcal{C}_n(\cdot)^\top \right)^\top}$ as a function of $\omega_1,\ldots,\omega_n$.
\begin{lemma}\label{ch6:lem:lm1}
	Let $x \in R^{nd}$, then
	$${\E_{\mathcal{C}} \left[\|\mathcal{C}(x)\|^2\right]\le (1+\omega) \|x\|^2,}$$
	where ${\omega = \max_{i=1,\ldots,n} \{\omega_i\}.}$
\end{lemma}
\begin{assumption}\label{ch6:ass:smoothBound}
	We assume that $f$ is $L_f$-smooth and $\mu$-strongly convex. 
\end{assumption}

\subsection{Auxiliary results}

Before we state our main convergence theorem, we state several intermediate results needed for the convergence. In the following two Lemmas, we show that based on the randomness of the compression operators, in expectation, we recover the exact average of the local models and the exact gradients for all iterations. 
\begin{lemma}\label{ch6:lem:compmean}
	Let Assumption \ref{ch6:ass:compression} hold, then for all ${k\ge 0}$, 
	${\E_{\mathcal{C},\mathcal{C}_M} \left[\mathcal{C}_M(\Bar{y}^k)\right] = \Bar{x}^k.}$
\end{lemma}

\begin{lemma}\label{ch6:lem:unbiasGrad}
	Let Assumptions \ref{ch6:ass:compression} hold. Then for all $k\ge 0$, knowing $x^k$,     
	$G(x^k)$ is an unbiased estimator of the gradient of function $F$ at $x^k$. 
\end{lemma}

Our next Lemma \ref{ch6:lem:xrelF} gives an upper bound on the iterate at each iteration. This bound is composed of two terms---the optimality gap, $F(x^k) - F(x^*)$, and the norm at the optimal point, $x^*$.  

\begin{lemma}
	\label{ch6:lem:xrelF}
	Let Assumption \ref{ch6:ass:smoothBound} hold, then 
	$${\left\| x^k\right\|^2 \le \dfrac{4}{\mu} \left( F(x^k) - F(x^*) \right) + 2 \left\| x^*\right\|^2.}$$
\end{lemma}

Our next Lemma \ref{ch6:lem:mathcalA} helps us to prove the expected smoothness property \citep{Gower2019}. 
The bound in Lemma \ref{ch6:lem:mathcalA} is composed of---the optimality gap, the difference between the gradients of $h$ at $x^k$ and $x^*$, and an extra constant, $\beta$ that depends on the used compressors. 

\begin{figure*}[t]
	\centering
	\begin{subfigure}[ht]{0.495\textwidth}
		\includegraphics[width=\textwidth]{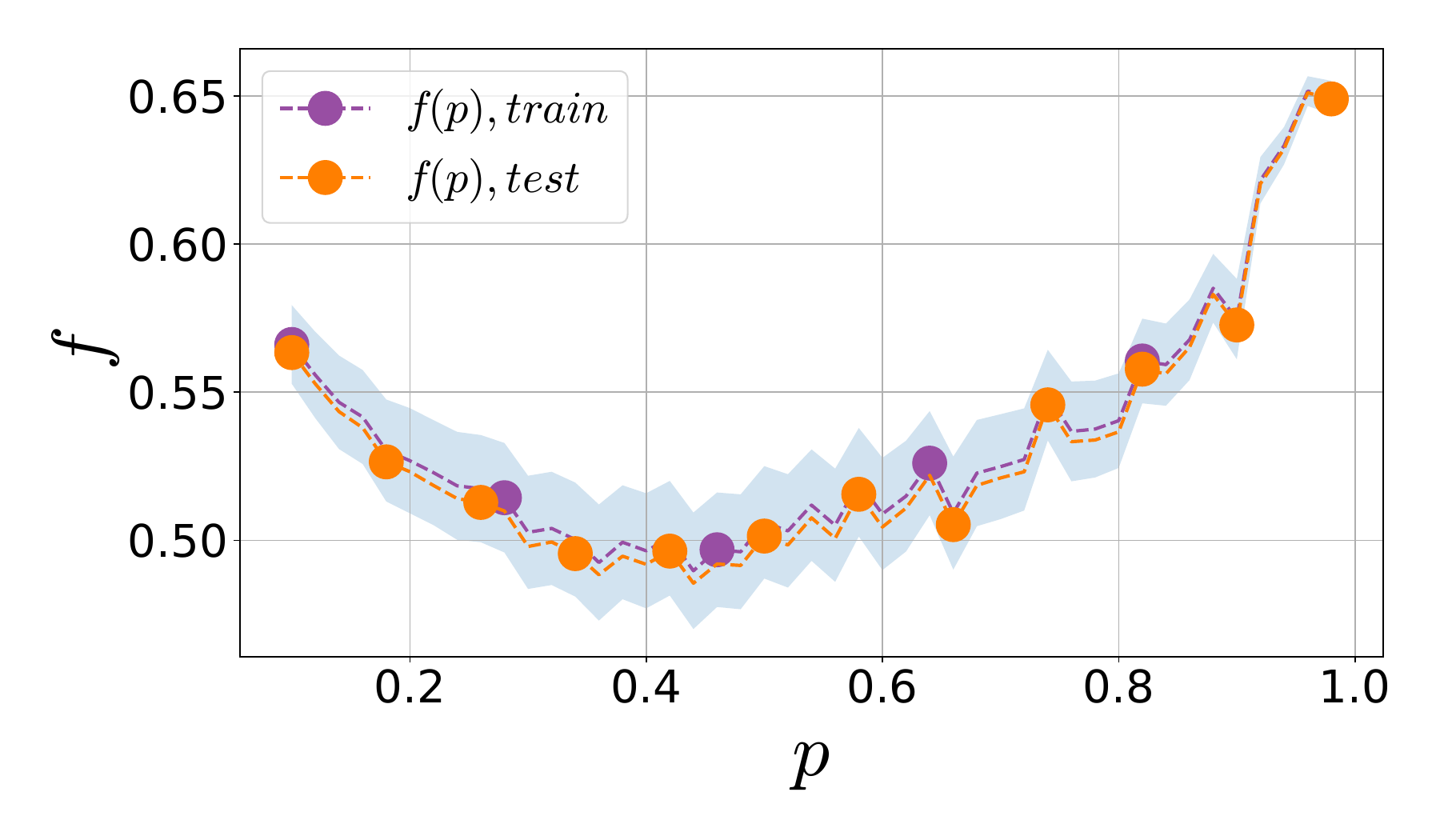} \caption{}
	\end{subfigure}
	\begin{subfigure}[ht]{0.495\textwidth}
		\includegraphics[width=\textwidth]{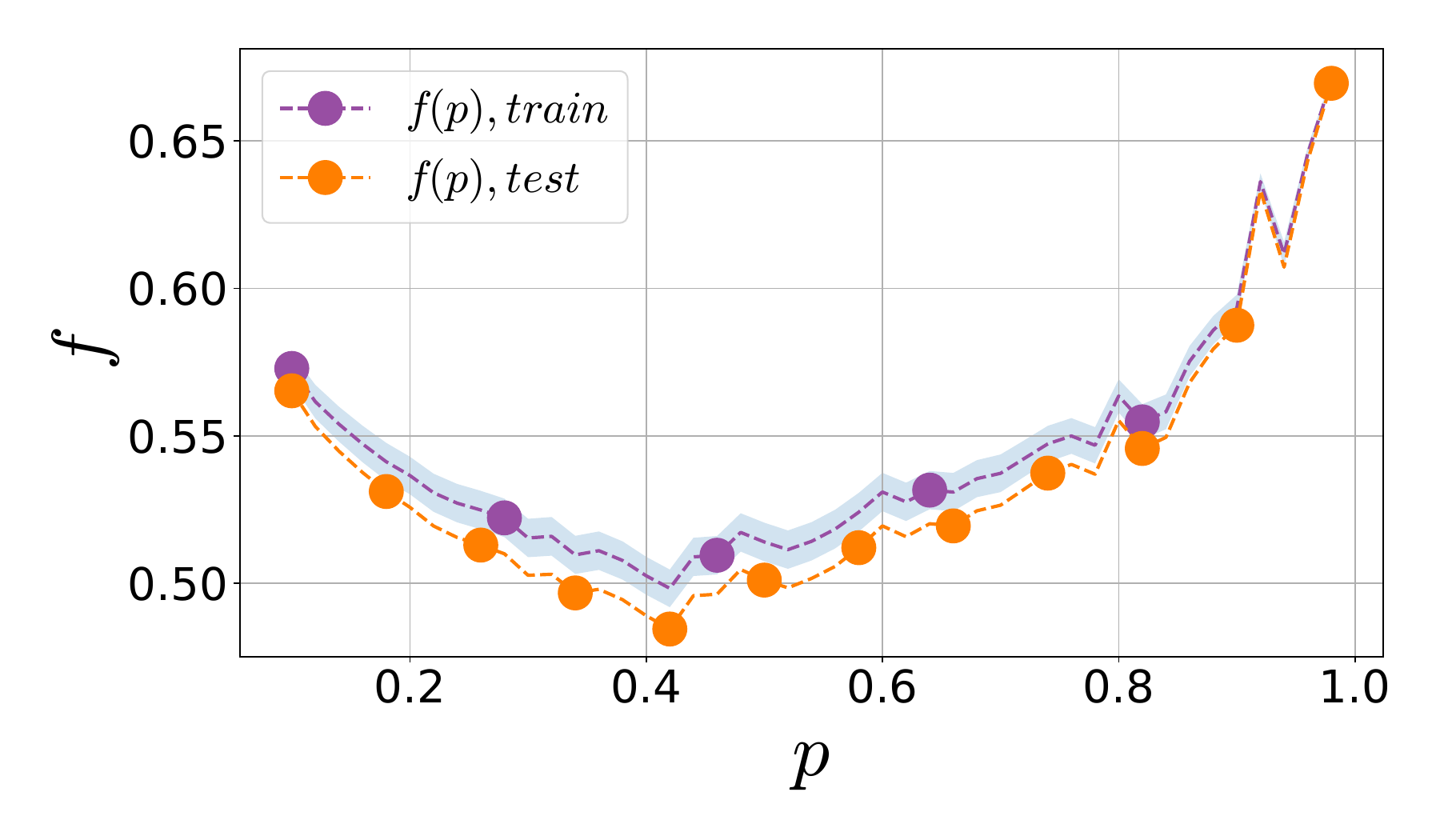} \caption{}
	\end{subfigure}
	
	\begin{subfigure}[ht]{0.495\textwidth}
		\includegraphics[width=\textwidth]{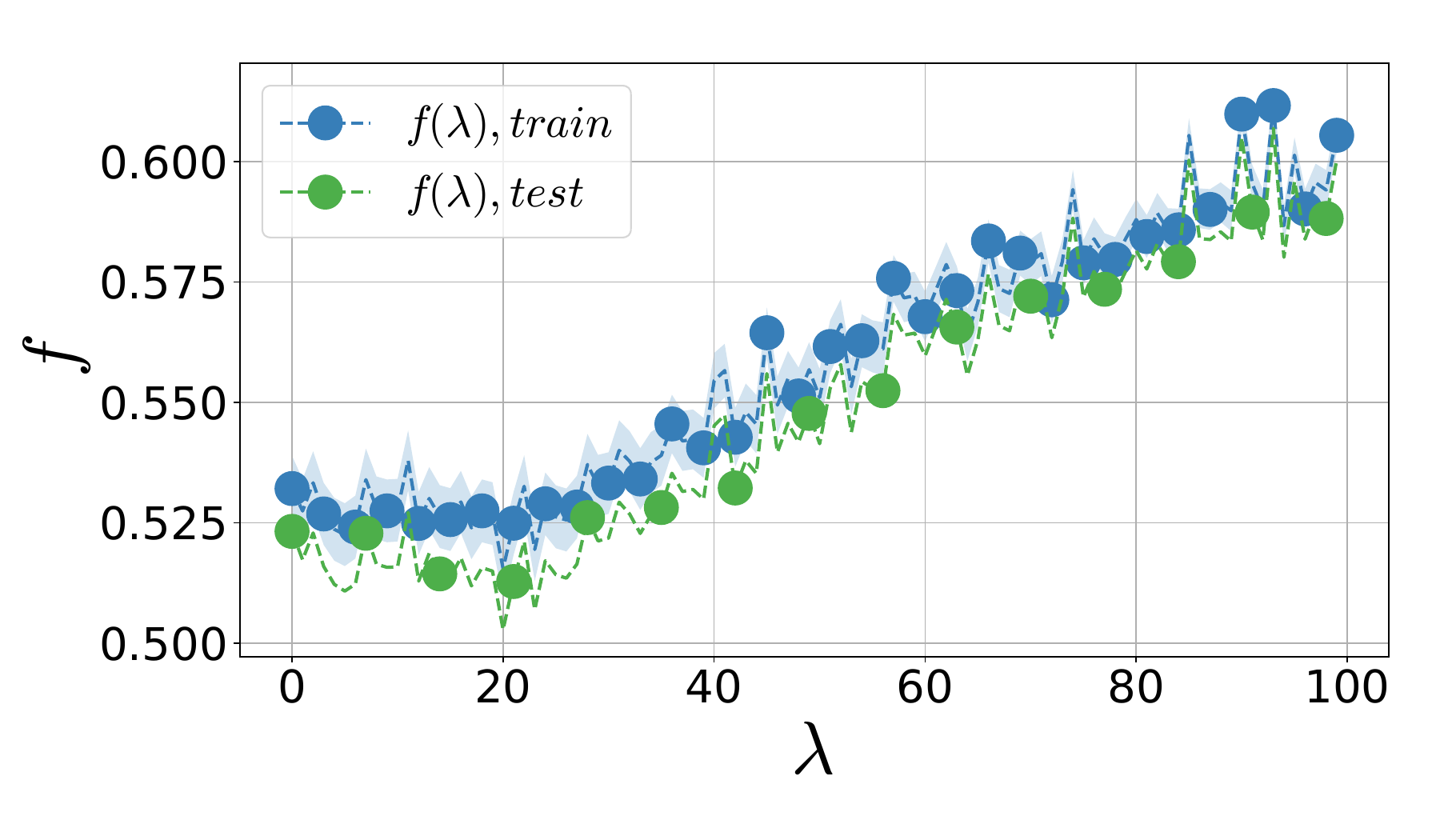}\caption{}
	\end{subfigure}
	\begin{subfigure}[ht]{0.495\textwidth}
		\includegraphics[width=\textwidth]{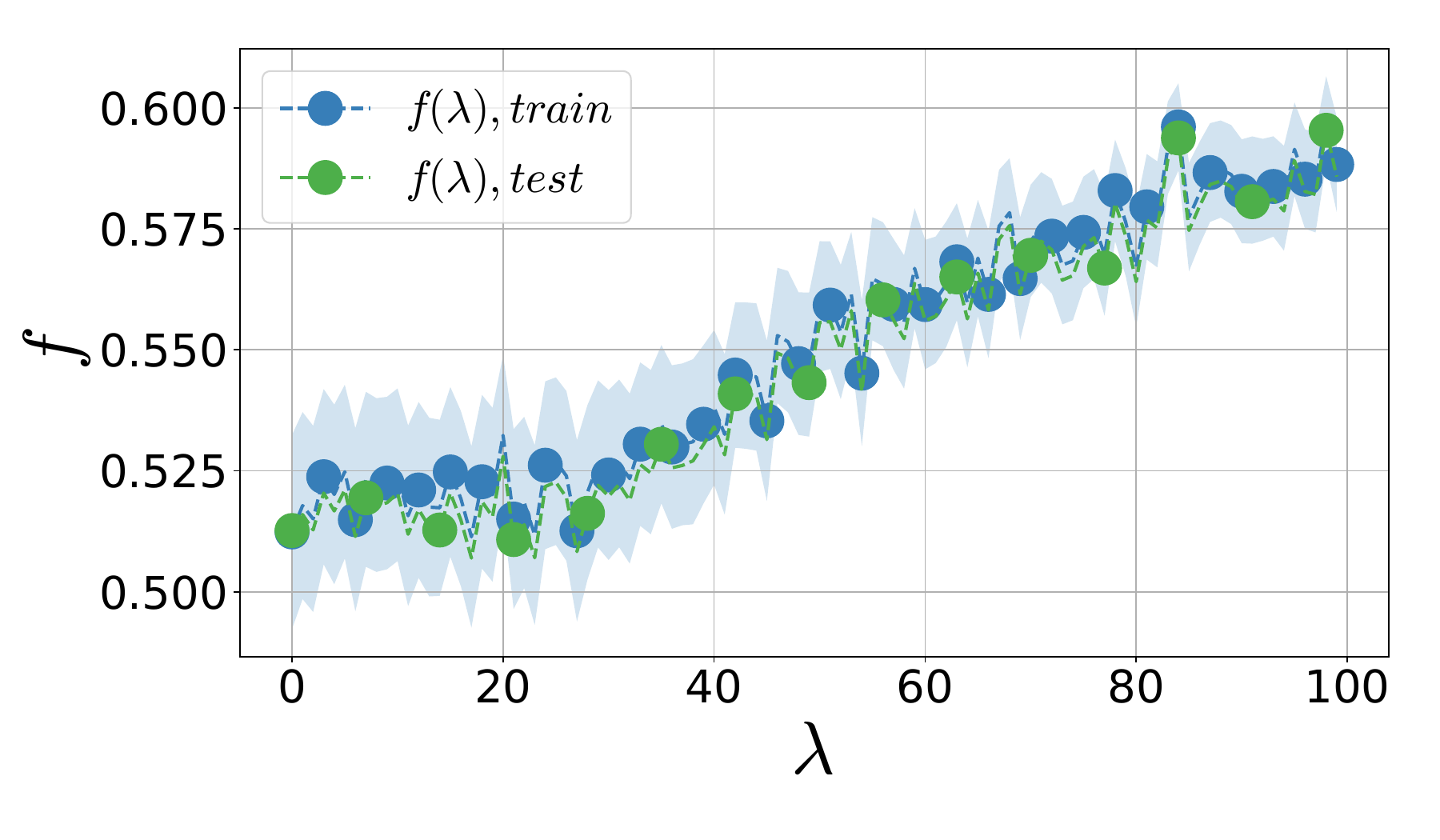} \caption{}
	\end{subfigure}
	\caption{{Uncompressed \algname{L2GD} on $n=5$ workers. We show the  loss, $f$ as a function of $p$ and $\lambda$ obtained after $K=100$ iterations of Algorithm \ref{ch6:alg:ComL2GD} with $\cC$ an identity compressor. (a) \dataname{A1A} dataset, $d=124,\lambda=10$, (b) \dataname{A2A} dataset, $d=124,\lambda=10$, (c) \dataname{A1A} dataset, $d=124,p=0.65$ (d) \dataname{A2A} dataset, $d=124,p=0.65$.}}
	\label{ch6:fig:lambda_and_p_selection}
\end{figure*}

\begin{lemma}\label{ch6:lem:mathcalA}
	Let Assumptions \ref{ch6:ass:compression} and \ref{ch6:ass:smoothBound} hold. Then
	\begin{eqnarray*}
		\mathcal{A} &\eqdef& \E_{\mathcal{C}_M,\mathcal{C}} \left\| x^k - \mQ\mathcal{C}_M(\Bar{y}^k) - x^* + \mQ\mathcal{C}_M(\Bar{y}^*) \right\|^2\\
		&\le& \dfrac{4n^2}{\lambda^2}\left\|\nabla h(x^k) - \nabla h(x^*)\right\|^2+ \alpha \left(F(x^k) - F(x^*)\right) + \beta,    
	\end{eqnarray*}
	where 
	and 
	\begin{eqnarray*}
		\Bar{y}^* &\eqdef& \dfrac{1}{n}\sum_{j=1}^n \mathcal{C}_j(x_j^*),\\
		\alpha &\eqdef& \dfrac{4\left (4\omega + 4\omega_M (1+\omega) \right)}{\mu},\\
		\beta &\eqdef& 2 \left (4\omega + 4\omega_M (1+\omega) \right) \left\|{x}^*\right\|^2
		+ 4\E_{\mathcal{C}_M,\mathcal{C}}\left\|\mQ\mathcal{C}_M(\Bar{y}^*)-  \mQ\Bar{x}^*\right\|^2.
	\end{eqnarray*} 
\end{lemma}

Next Lemma \ref{ch6:lem:ES} is the final result that (together with Lemma \ref{ch6:lem:mathcalA}) shows the expected smoothness property and gives an upper bound on the stochastic gradient. This bound is composed of three terms:
\begin{enumerate}
	\item The optimality gap, $F(x^k) - F(x^*)$.
	\item The expected norm of the stochastic gradient at the optimal point, $\E \| G(x^{*})\|^2$.
	\item Some other quantity that involves interplay between the parameters used in Algorithm \ref{ch6:alg:ComL2GD} and the used compressor.
\end{enumerate}
\begin{lemma}[Expected Smoothness]\label{ch6:lem:ES}
	Let Assumptions \ref{ch6:ass:compression} and \ref{ch6:ass:smoothBound} hold, then
	\begin{equation}
		\E\left[\|G(x^{k})\|^2|x^k\right] \le 4 \gamma \left(F(x^k) - F(x^*)\right) + \delta,
	\end{equation}  
	where 
	\begin{eqnarray*} 
	\gamma &\eqdef& \dfrac{\alpha \lambda^2 (1-p)}{2 n^2 p} + \max \left\{ \dfrac{ L_f}{(1-p)}, \dfrac{\lambda}{n} \left(1+\dfrac{4(1-p)}{p}\right)
		\right\},\\
	\delta &\eqdef& \dfrac{2 \beta \lambda^2 (1-p)}{n^2 p}+2\E \| G(x^{*})\|^2.
	\end{eqnarray*} 
\end{lemma}

\begin{remark} 
	If there is no compression, the operators, $\mathcal{C}_i(\cdot)$, for $i\in[n]$, and $\mathcal{C}_M(\cdot)$ are equal to identity. The compression constants, $\omega_i$, for $i\in[n]$, and $\omega_M$ are equal to zero. Therefore, ${\alpha = \beta=0}$ (Lemma~\ref{ch6:lem:mathcalA}), and the factor $4$ in the formula of $\gamma$ (Lemma~\ref{ch6:lem:ES}) can be replaced by $1$ and thus  
	\begin{eqnarray*}
		\gamma &=& \max \left\{ \dfrac{ L_f}{(1-p)}, \dfrac{\lambda}{n} \left(1+\dfrac{(1-p)}{p}\right)
		\right\}=\max \left\{ \dfrac{ L}{n(1-p)}, \dfrac{\lambda}{np}
		\right\},\\
		\delta &=& \dfrac{2 \beta \lambda^2 (1-p)}{n^2 p}+2\E \| G(x^{*})\|^2 = 2\E \| G(x^{*})\|^2.
	\end{eqnarray*}
	where $L = n L_f$. Same constants arise in the expected smoothness property in \citet{Hanzely2020}.
\end{remark}

{For nonconvex convergence of our algorithm, we cannot use the expected smoothness of Lemma \ref{ch6:lem:ES}, as it requires $\mu$-strong convexity of the loss function. Therefore, similar to \citep{sahu2021rethinking} we require a new assumption to bound the stochastic gradient, $G(x^k)$. Let $G(x^k)$ is of the form: $$G(x^k)=\nabla F(x^k)+\zeta_k,$$ where $\zeta_k$ is the stochastic noise on the gradient.} 

\begin{assumption}
	\label{ch6:assumption:bounded_g}
	\label{ch6:remark:m-sigma2-bounded_noise}
	{There exist $M,\sigma^2\ge 0$, such that $$\E[\|G(x^k)\|^2\mid x^k]\leq M\|\nabla F(x^k)\|^2+\sigma^2.$$}
\end{assumption}

\subsection{Main result}

We now state the convergence result for Algorithm \ref{ch6:alg:ComL2GD} for both strongly convex and non-convex cases.
\begin{theorem}(Strongly convex case)\label{ch6:thm:mainconvergenceresult} Let Assumptions \ref{ch6:ass:compression} and \ref{ch6:ass:smoothBound} hold. 
	If ${\eta \le \dfrac{1}{2\gamma}}$, then 
	$${\E\left\| x^k - x^*\right\|^2 \le \left( 1-\dfrac{\eta \mu}{n}\right)^k \left\| x^0 - x^*\right\|^2 + \dfrac{n \eta \delta}{\mu}.}$$
\end{theorem}
\begin{proof}
	The proof follows directly from Lemma \ref{ch6:lem:unbiasGrad}, \ref{ch6:lem:ES}, and Theorem 3.1 from \citet{Gower2019}.
\end{proof}

\begin{theorem}(Non convex case)\label{ch6:thm:nncc} {Let Assumptions \ref{ch6:ass:compression} and \ref{ch6:assumption:bounded_g} hold. Assume also that $F$ is $L_f$-smooth, bounded from below by $F(x^*)$. Then to reach a precision, $\textstyle{\epsilon>0}$, set the step size, 
		$$\textstyle{\eta=\min\{\frac{1}{{L_f M}},\frac{\epsilon^2}{2L_f\sigma^2}\}},$$ 
		such that for $$\textstyle{K\ge \frac{4L_f M(F(x^0)-F(x^*))}{\epsilon^2},}$$
		we have 
		$$\min_{k=0,1,\dots,K}  \E \|\nabla F(x^k)\|_2\le \epsilon.$$}
\end{theorem}


\begin{remark}\label{ch6:rem:nncc}
	For smooth non-convex problems, we recover the optimal $O(\epsilon^4)$ classical rate as vanilla \algname{SGD} in a non-convex smooth case.
\end{remark}

\section{Optimal Rate and Communication} 
\label{ch6:sec:optimalrate} 
In this section, we provide the ``optimal" setting of our algorithm that is obtained by optimizing the complexity bounds of our algorithm as a function of the parameters involved.  The analysis in this section is based on the following upper bound of $\gamma$. We recall from Lemma~\ref{ch6:lem:ES} that 
\begin{eqnarray*}
	\gamma &=& \dfrac{\alpha \lambda^2 (1-p)}{2 n^2 p} + \max \left\{ \dfrac{ L_f}{(1-p)}, \dfrac{\lambda}{n} \left(1+\dfrac{4(1-p)}{p}\right)
	\right\} 
	\\  
	&\le&\dfrac{\alpha \lambda^2 (1-p)}{2 n^2 p} + \max \left\{ \dfrac{ L_f}{(1-p)}, \dfrac{4\lambda}{np} 
	\right\} \eqdef \gamma_{u}.
\end{eqnarray*}
Note that the number of iterations is linearly dependent on $\gamma$. Therefore, to minimize the total number of iterations, it suffices to minimize $\gamma$. Define $L \eqdef n L_f$.

\begin{theorem}[Optimal rate] \label{ch6:thm:optimalrate}
The probability $p^*$ minimizing $\gamma$ is equal to 
$\max\{p_e,p_A\}$, where $p_e = \dfrac{7 \lambda + L - \sqrt{\lambda^2 + 14 \lambda L + L^2}}{6 \lambda}$ and $p_A$ is the optimizer of the function $A(p) = \dfrac{\alpha \lambda^2}{2 n^2 p } + \dfrac{ L}{n(1-p)}$ in $(0,1)$.
\end{theorem}

\begin{remark}\label{ch6:rem:max_upp_bound}
If we maximize the upper bound, $\gamma_u$ instead of $\gamma$ then $p_e$ simplifies to $\dfrac{4 \lambda}{L + 4 \lambda}$.
\end{remark}

\begin{lemma}\label{ch6:lem:pa} The optimizer probability $p_A$   of the function $A(p) = \dfrac{\alpha \lambda^2}{2 n^2 p } + \dfrac{ L}{n(1-p)}$ in $(0,1)$ is equal to 
$$p_A=
\left\{
\begin{array}{lll}
	\dfrac{1}{2}  & \text{ if } 2nL =\alpha \lambda^2
	\\
	\dfrac{-2 \alpha \lambda^2 + 2\lambda \sqrt{2\alpha n L}}{2(2nL -\alpha \lambda^2)}   & \text{ if } 2nL > \alpha \lambda^2
	\\
	\dfrac{-2 \alpha \lambda^2 - 2\lambda \sqrt{2\alpha n L}}{2(2nL -\alpha \lambda^2)}   & \text{otherwise.} 
\end{array}
\right.$$
\end{lemma}

Note that the number of communication rounds is linearly proportional to $C \eqdef p(1-p)\gamma$. Therefore, minimizing the total number of communication rounds suffices to minimize $C$ or $nC$.

\begin{theorem}[Optimal communication] \label{ch6:thm:optimalcommunication}
The probability $p^*$ optimizing $C$ is equal to 
$\max\{p_e,p_A\}$, where $p_e = \dfrac{7 \lambda + L - \sqrt{\lambda^2 + 14 \lambda L + L^2}}{6 \lambda}$ and $p_A = 1 - \dfrac{Ln}{\alpha \lambda^2}$.
\end{theorem}

\begin{remark}
As in Remark \ref{ch6:rem:max_upp_bound}, we note that, if we use the upper bound, $\gamma_u$ instead of $\gamma$ then $p_e$ simplifies to $\dfrac{4 \lambda}{L + 4 \lambda}$.
\end{remark}  

We note that $\lambda\to0$ implies $p^*\to0$. This means that the optimal strategy, in this case, is {\em no communication at all}. This result is intuitive since for $\lambda=0$, we deal with pure local models which can be computed without any communication. As $\lambda\to\infty$ implies $p^*\to 1$ denoting that the optimal strategy is to communicate often to find the global model. 

\begin{table*}[t]
\caption{Gradient compression methods used in this work. Note that $\|\tilde{g}\|_0$ and $\|g\|_0$ are the number of elements in the compressed and uncompressed gradient, respectively; the nature of operator $\mQ$ is random or deterministic. We implement those mechanisms for \libname{FedML.ai} framework.}

\label{ch6:tab:summary}
\small
\centering
\begin{tabular}{lllcc}
	\toprule
	\textbf{Method}                                & \textbf{Similar Methods} & \textbf{$\|\tilde{g}\|_0$} & \makecell[c]{\textbf{Nature}\\\textbf{of $\cC$}} & \makecell[c]{\textbf{Implemen-}\\\textbf{tation}} \\ \midrule
	\multicolumn{5}{c}{\textbf{Quantization}} \\
	\midrule \\
	\makecell[l]{\compname{QSGD} \\ \citet{alistarh2017qsgd}}                           & 
	\makecell[l]{\citet{horvath2019natural}, \\
		 \citet{ATOMO},\\
		 \citet{DBLP:conf/nips/WenXYWWCL17} \\ \citet{errorSGD} \\
		 \citet{Yu2019ExploringFA},\\ \citet{zipml}}                         &                          $\|g\|_0$        &  
		 \makecell[l]{Rand,\\unbiased}             
	& \makecell[l]{FedML.ai\\standalone / \\ distributed}                      
	\\
	\\
	\makecell[l]{\compname{Natural}\\ \citet{horvath2019natural}}                           &  \makecell[l]{\citet{alistarh2017qsgd}\\ \citet{Yu2019ExploringFA} \\ \citet{zipml}}                            &                        $\|g\|_0$          &   \makecell[l]{Rand,\\unbiased}                                               & \makecell[l]{FedML.ai\\standalone / \\ distributed}                   
	\\
	\\
	\makecell[l]{\compname{TernGrad} \\ \citet{DBLP:conf/nips/WenXYWWCL17}}                             & \makecell[l]{\citet{alistarh2017qsgd},\\ \citet{ATOMO}, \\ \citet{Yu2019ExploringFA}}                             &                          $\|g\|_0$        & \makecell[l]{Rand,\\unbiased}                                    & 
	\makecell[l]{FedML.ai\\standalone / \\distributed}      
	\\
	\\
	\makecell[l]{\compname{Bernoulli} \\ \citet{khirirat2018distributed}}                             & ---                             &                          ---       & \makecell[l]{Rand,\\unbiased}                                    & \makecell[l]{FedML.ai\\standalone / \\distributed}                   
	\\\midrule
	\multicolumn{5}{c}{\textbf{Sparsification}} \\
	\midrule \\
	\makecell[l]{\compname{TopK} \\ \citet{aji_sparse}}                              & \makecell[l]{\citet{Alistarh-EF-NIPS2018}, \\ \citet{Stich-EF-NIPS2018}}                             &         $k$                           &   \makecell[l]{Det,\\biased}                                         & \makecell[l]{FedML.ai\\standalone / \\ distributed}                   \\
	\bottomrule
\end{tabular}
\end{table*}


\begin{figure*}[t]
\centering

\hspace{-1.04cm}\begin{subfigure}[ht]{0.28\textwidth}
	\includegraphics[width=\textwidth]{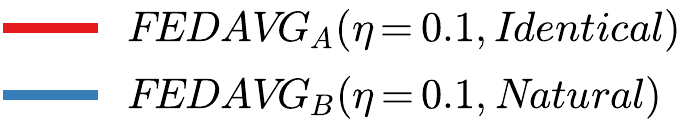}
\end{subfigure}\hspace{2.90cm}
\begin{subfigure}[ht]{0.34\textwidth}
	\includegraphics[width=\textwidth]{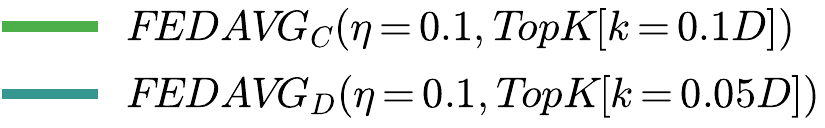}
\end{subfigure}

\begin{subfigure}[ht]{0.9\textwidth}
	\vspace{0.1cm}
	\includegraphics[width=\textwidth]{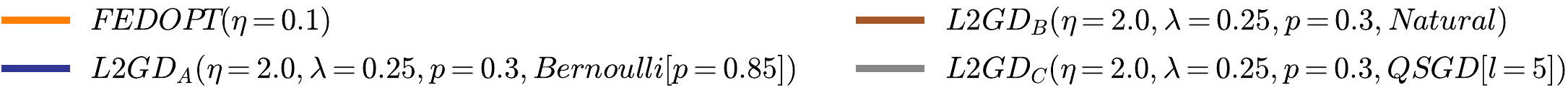}
\end{subfigure}

\begin{subfigure}[ht]{0.495\textwidth}
	\includegraphics[width=\textwidth]{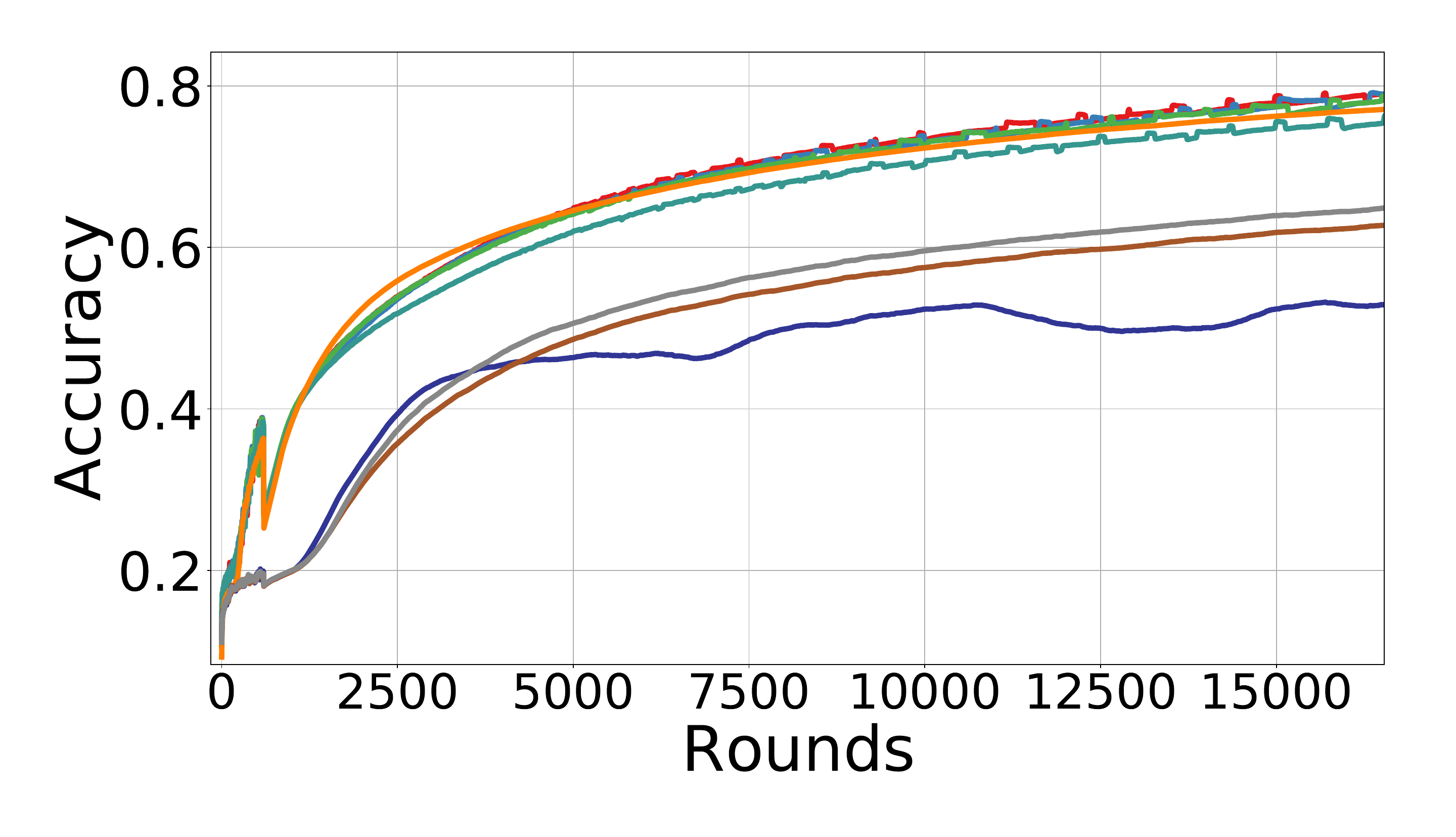} \caption{\textbf{Train:} Top-1 accuracy}
\end{subfigure}
\begin{subfigure}[ht]{0.495\textwidth}
	\includegraphics[width=\textwidth]{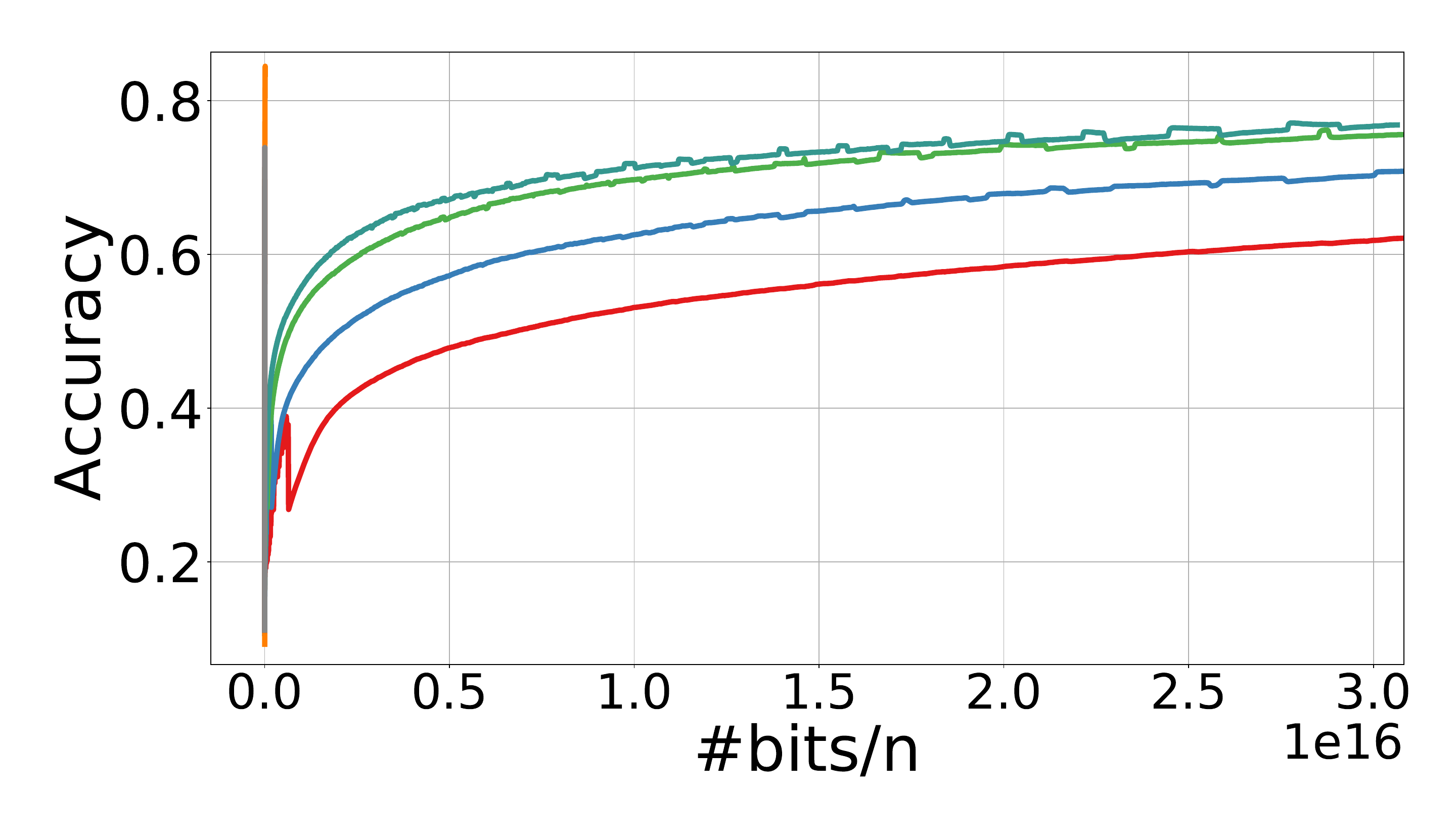}
	\caption{\textbf{Train:} Top-1 accuracy}
\end{subfigure}

\begin{subfigure}[ht]{0.495\textwidth}
	\includegraphics[width=\textwidth]{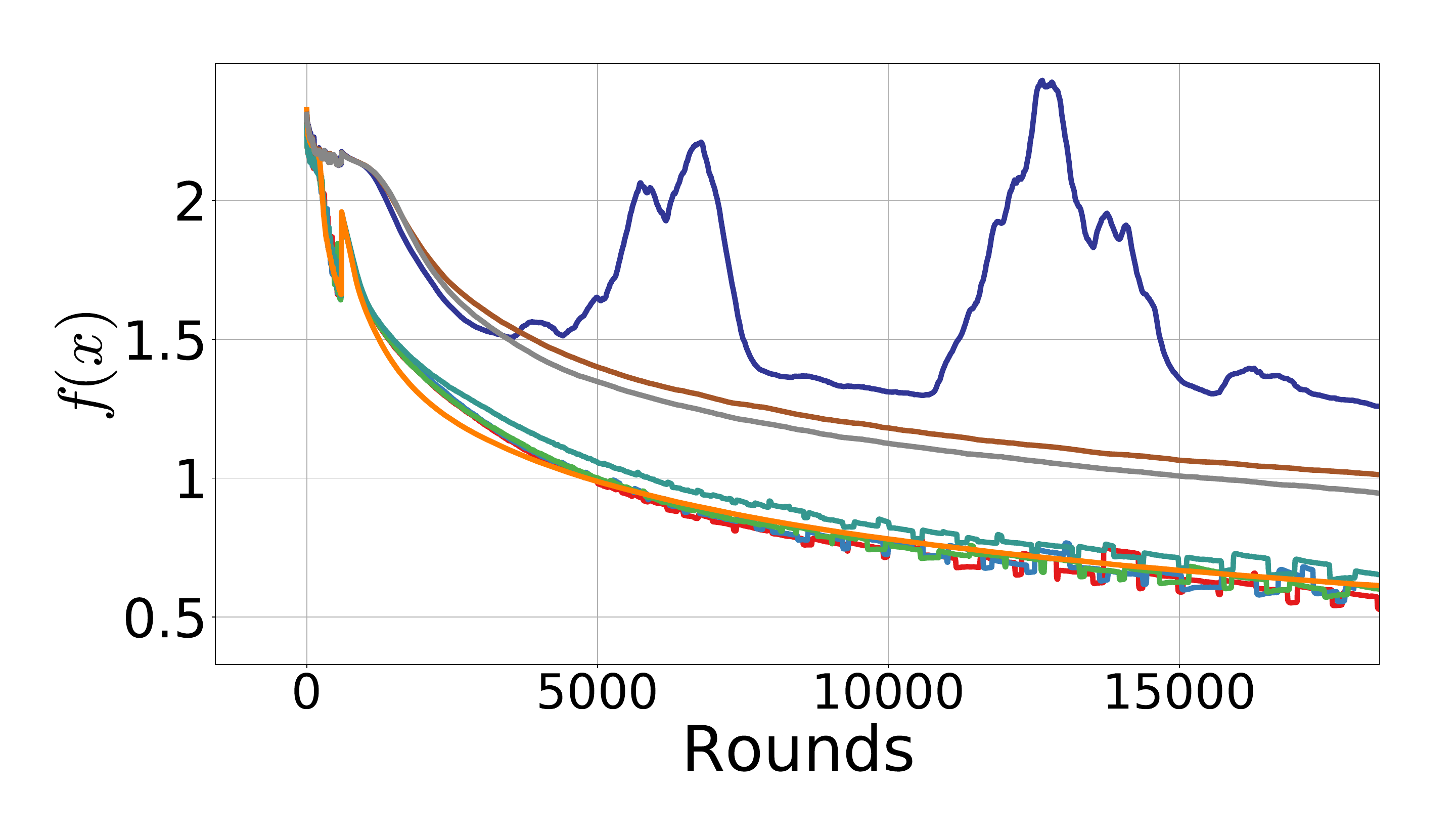} \caption{\textbf{Train:} Loss functional value}
\end{subfigure}
\begin{subfigure}[ht]{0.495\textwidth}
	\includegraphics[width=\textwidth]{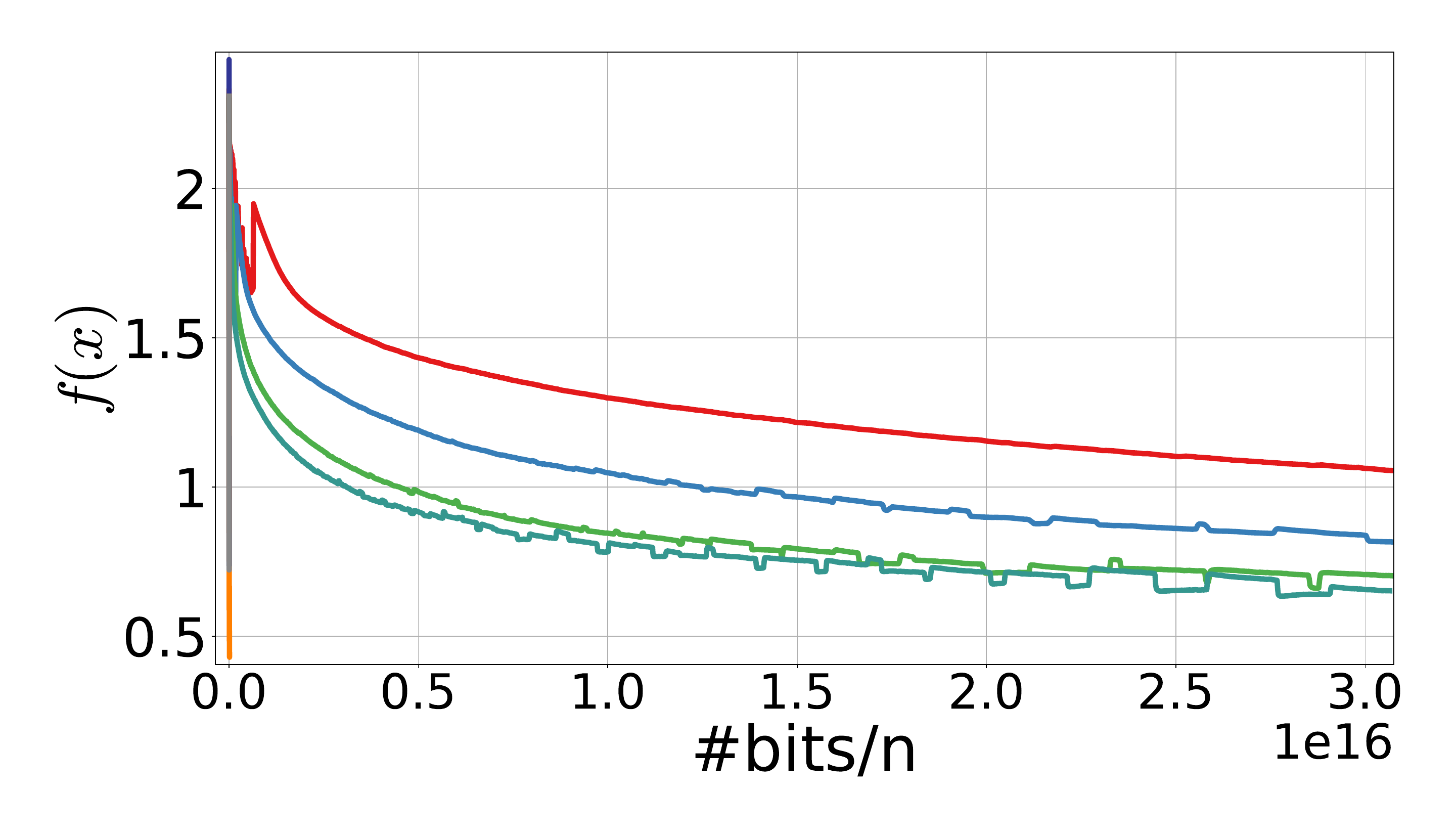} \caption{\textbf{Train:} Loss functional value}
\end{subfigure}

\vspace{0.25cm} 
\rule{\textwidth}{0.4pt} 
\vspace{0.25cm} 

\begin{subfigure}[ht]{0.495\textwidth}
	\includegraphics[width=\textwidth]{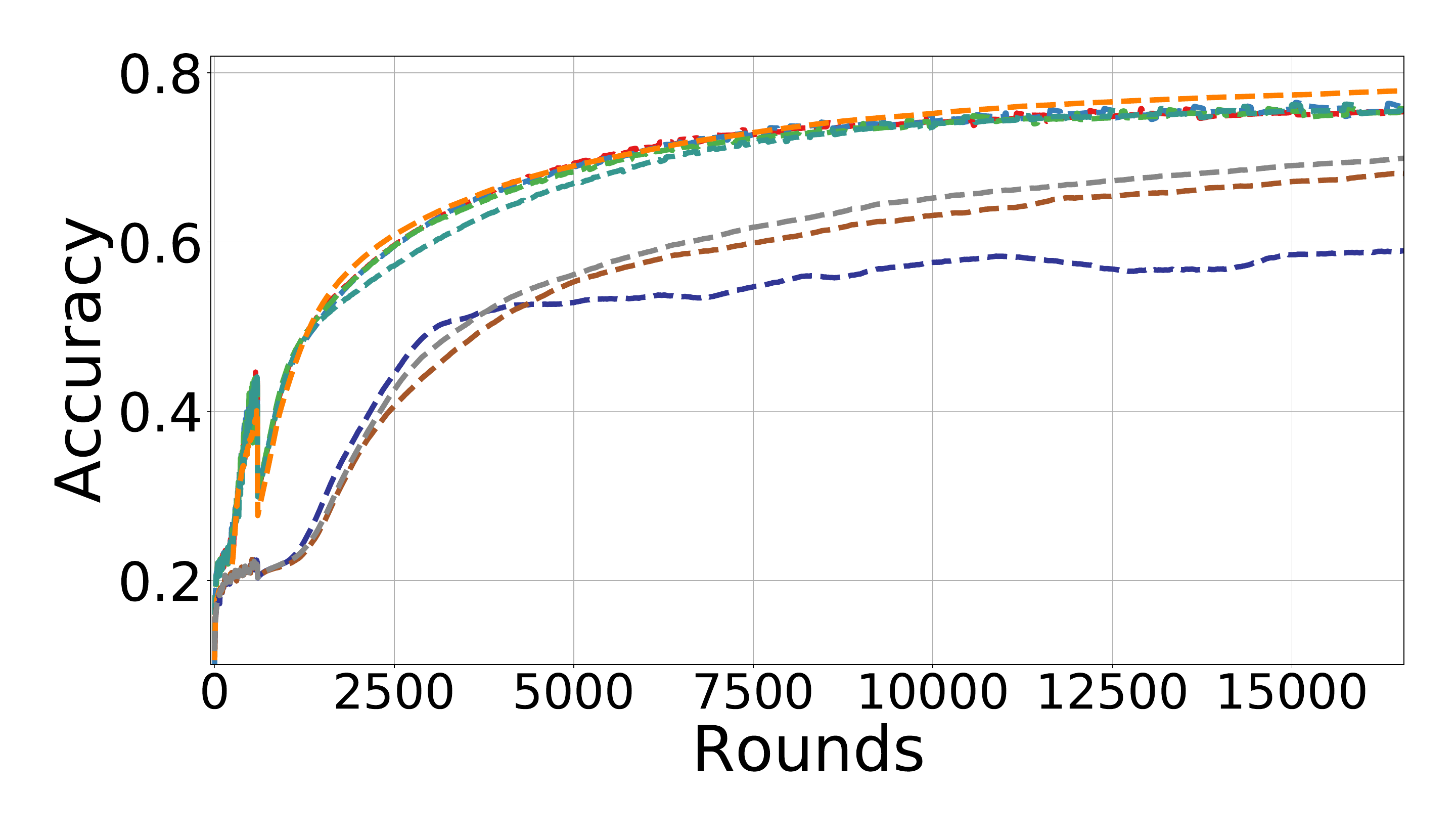} \caption{\textbf{Test:} Top-1 accuracy}
\end{subfigure}
\begin{subfigure}[ht]{0.495\textwidth}
	\includegraphics[width=\textwidth]{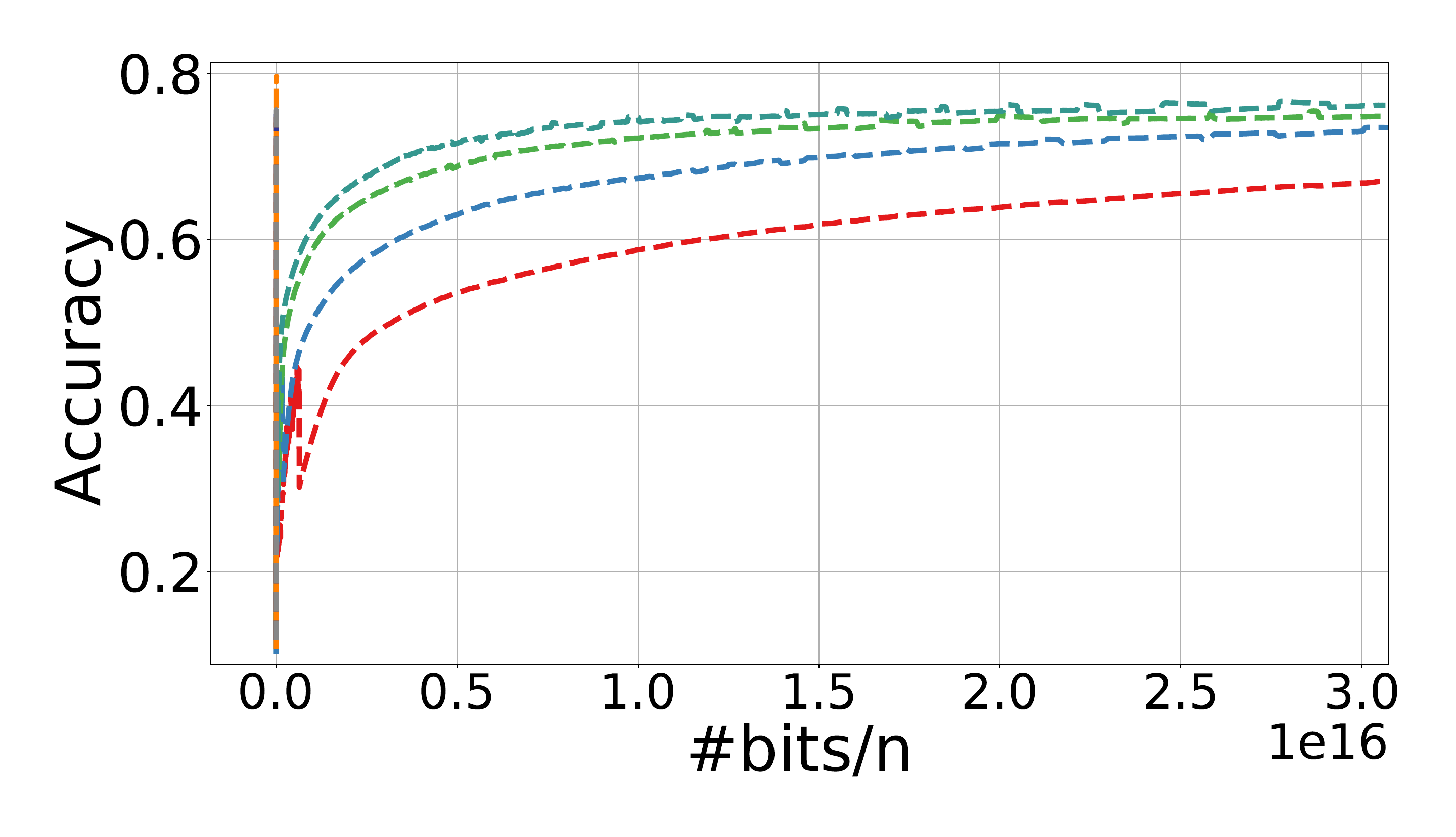}
	\caption{\textbf{Test:} Top-1 accuracy}
\end{subfigure}

\begin{subfigure}[ht]{0.495\textwidth}
	\includegraphics[width=\textwidth]{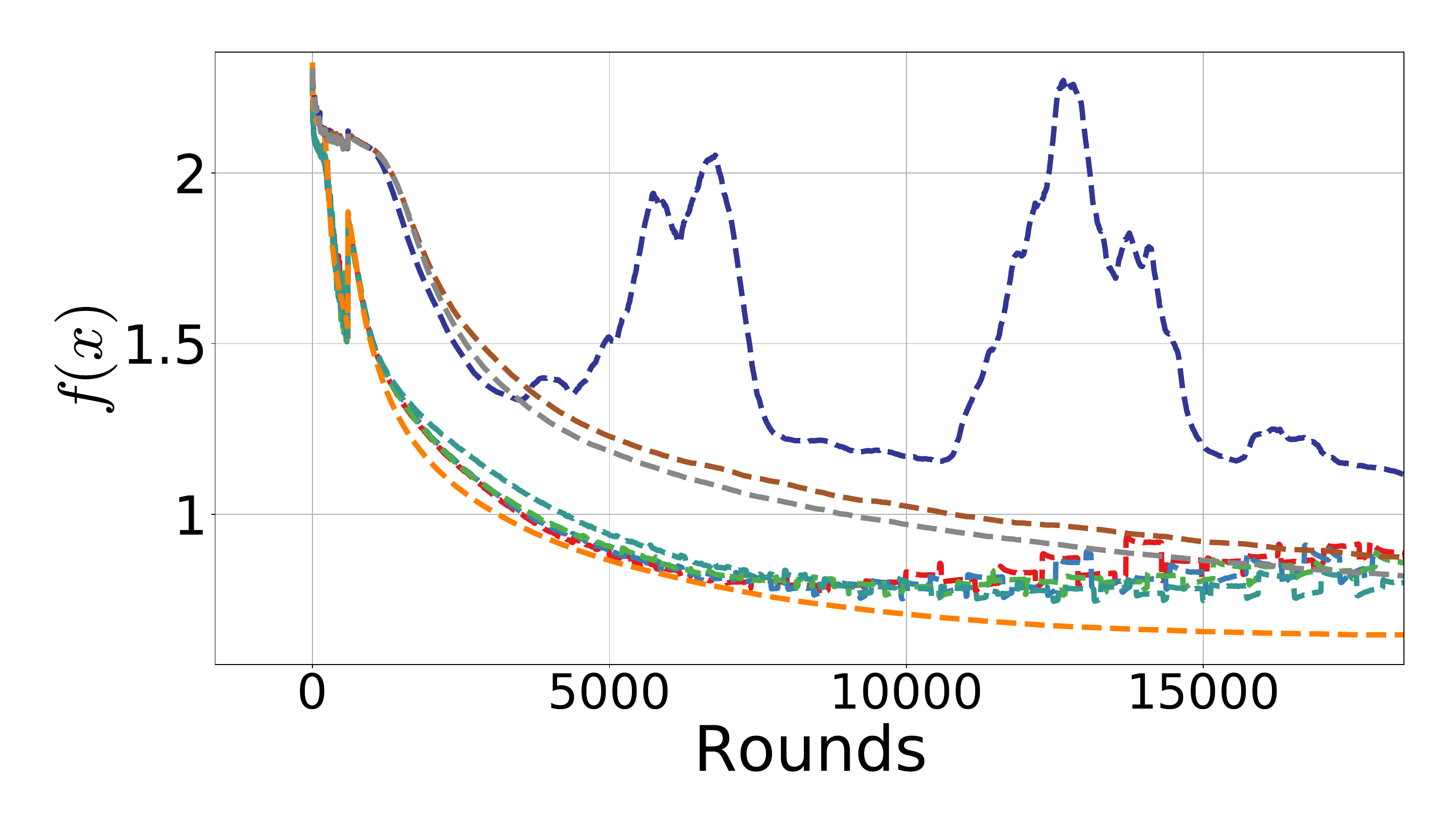} \caption{\textbf{Test:} Loss functional value}
\end{subfigure}
\begin{subfigure}[ht]{0.495\textwidth}
	\includegraphics[width=\textwidth]{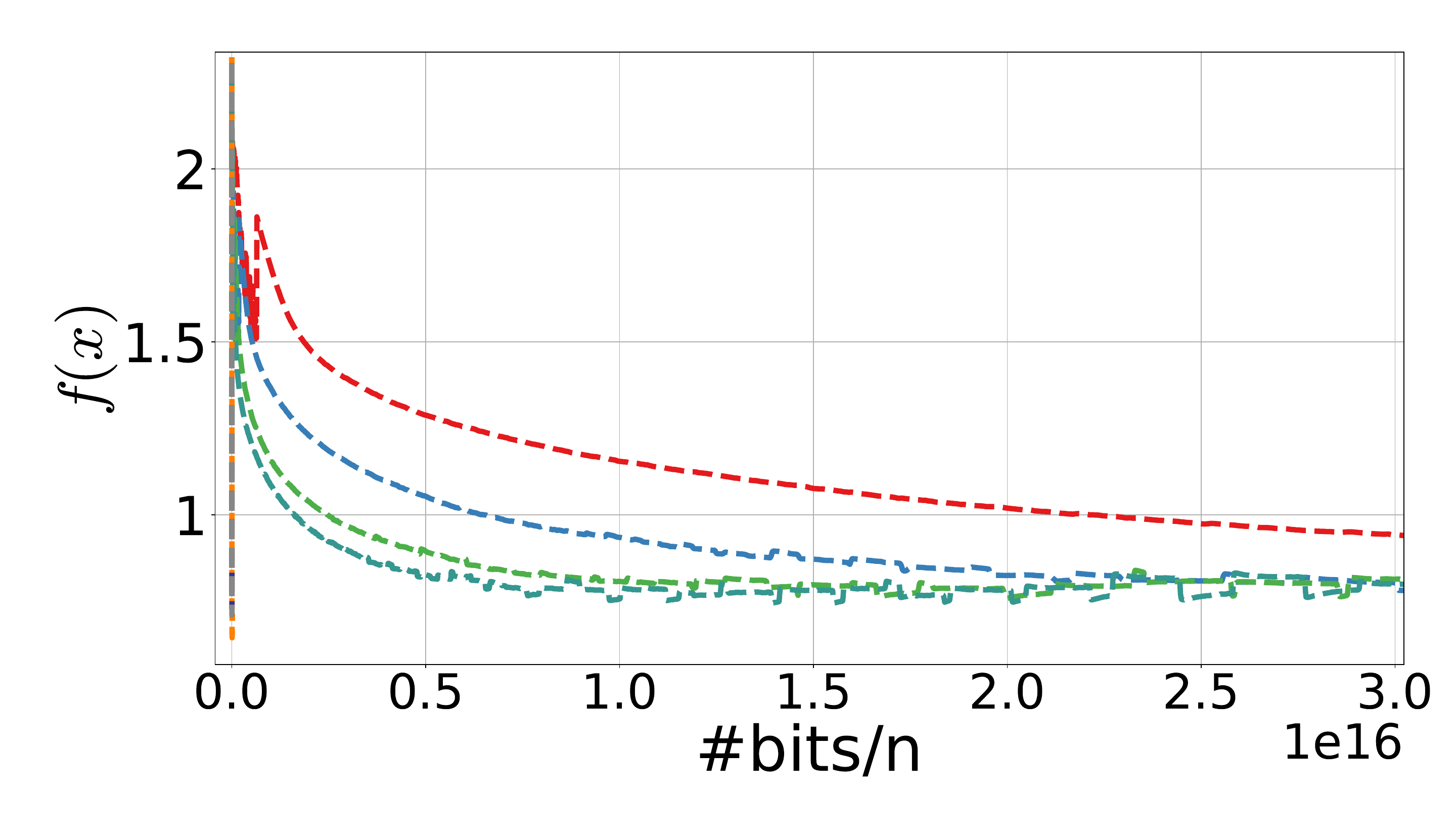} \caption{\textbf{Test:} Loss functional value}
\end{subfigure}

\caption{{Training \modelname{ResNet-18} on \dataname{CIFAR-10} with $n=10$ workers. Plots (a)--(d) show behavior on the \textit{train set}, while (e)--(h) show behavior on the \textit{test set}.}}
\label{ch6:fig:training_resnet}
\end{figure*}

\begin{figure*}[t]
\centering

\hspace{-1.3cm}\begin{subfigure}[ht]{0.32\textwidth}
	\includegraphics[width=\textwidth]{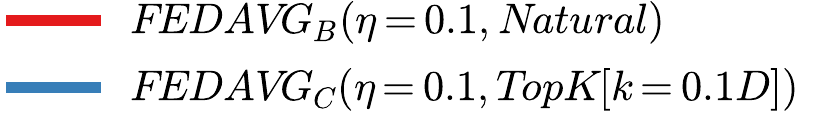}
\end{subfigure}\hspace{2.3cm}
\begin{subfigure}[ht]{0.32\textwidth}
	\includegraphics[width=\textwidth]{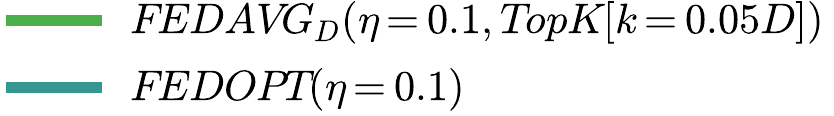}
\end{subfigure}

\begin{subfigure}[ht]{0.9\textwidth}
	\vspace{0.1cm}
	\includegraphics[width=\textwidth]{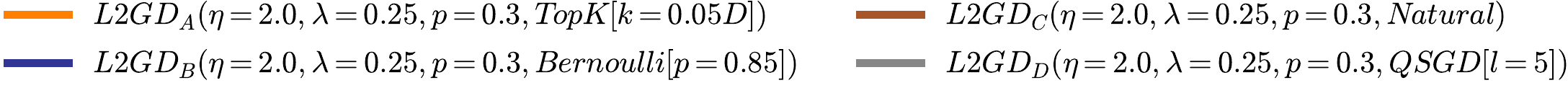}
\end{subfigure}

\begin{subfigure}[ht]{0.495\textwidth}
	\includegraphics[width=\textwidth]{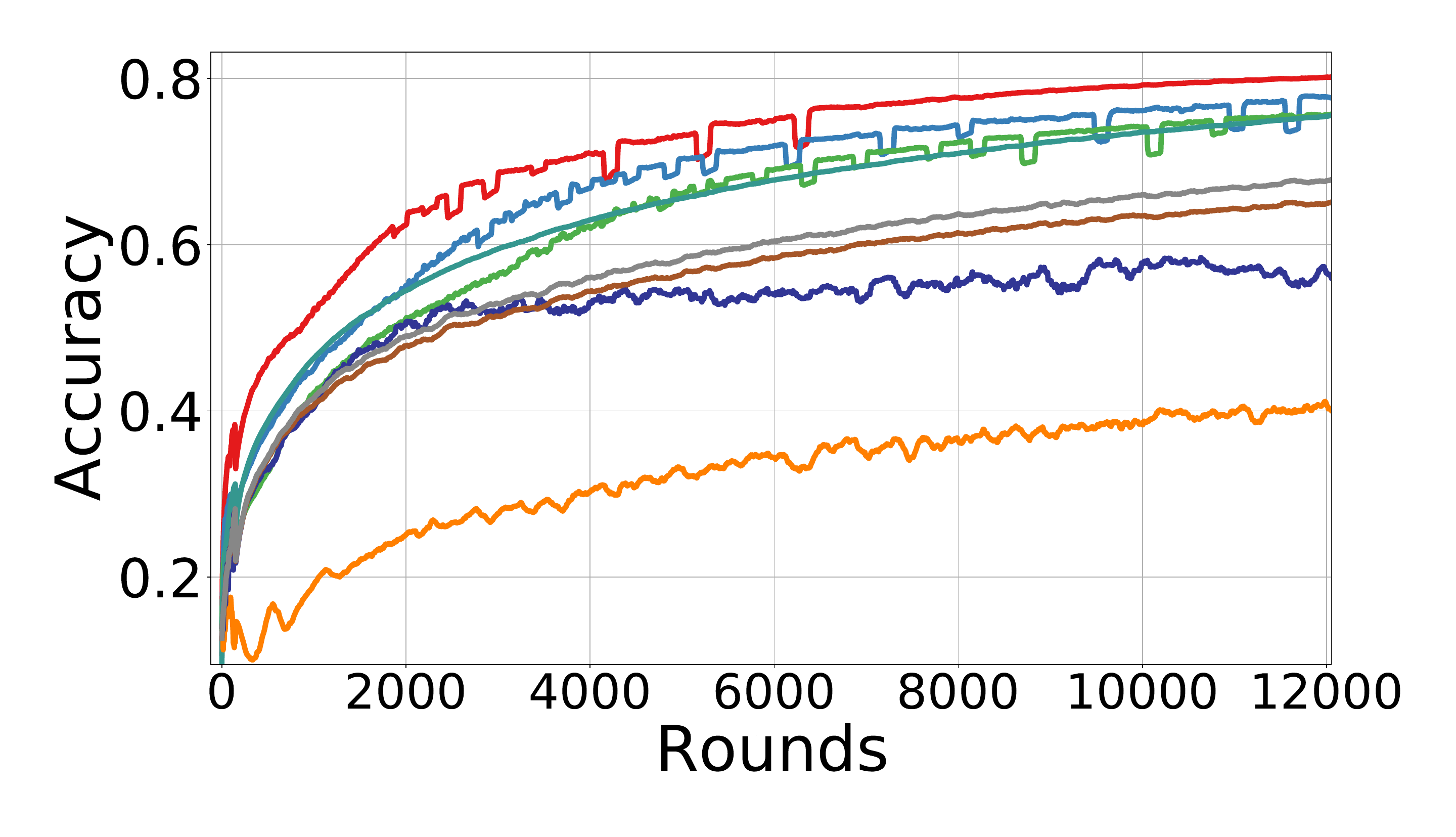} 
	\caption{\textbf{Train:} Top-1 accuracy}
\end{subfigure}
\begin{subfigure}[ht]{0.495\textwidth}
	\includegraphics[width=\textwidth]{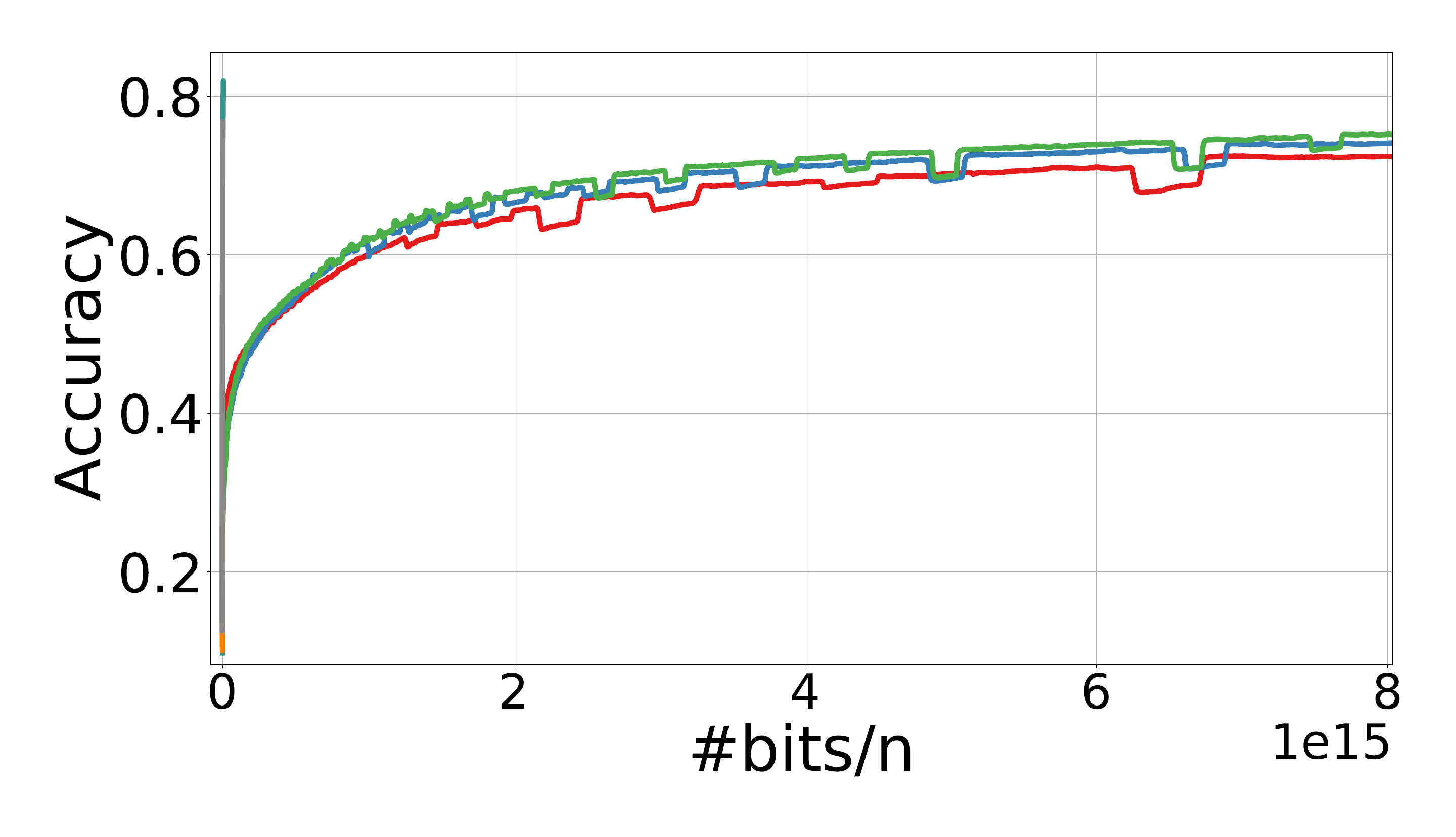}
	\caption{\textbf{Train:} Top-1 accuracy}
\end{subfigure}

\begin{subfigure}[ht]{0.495\textwidth}
	\includegraphics[width=\textwidth]{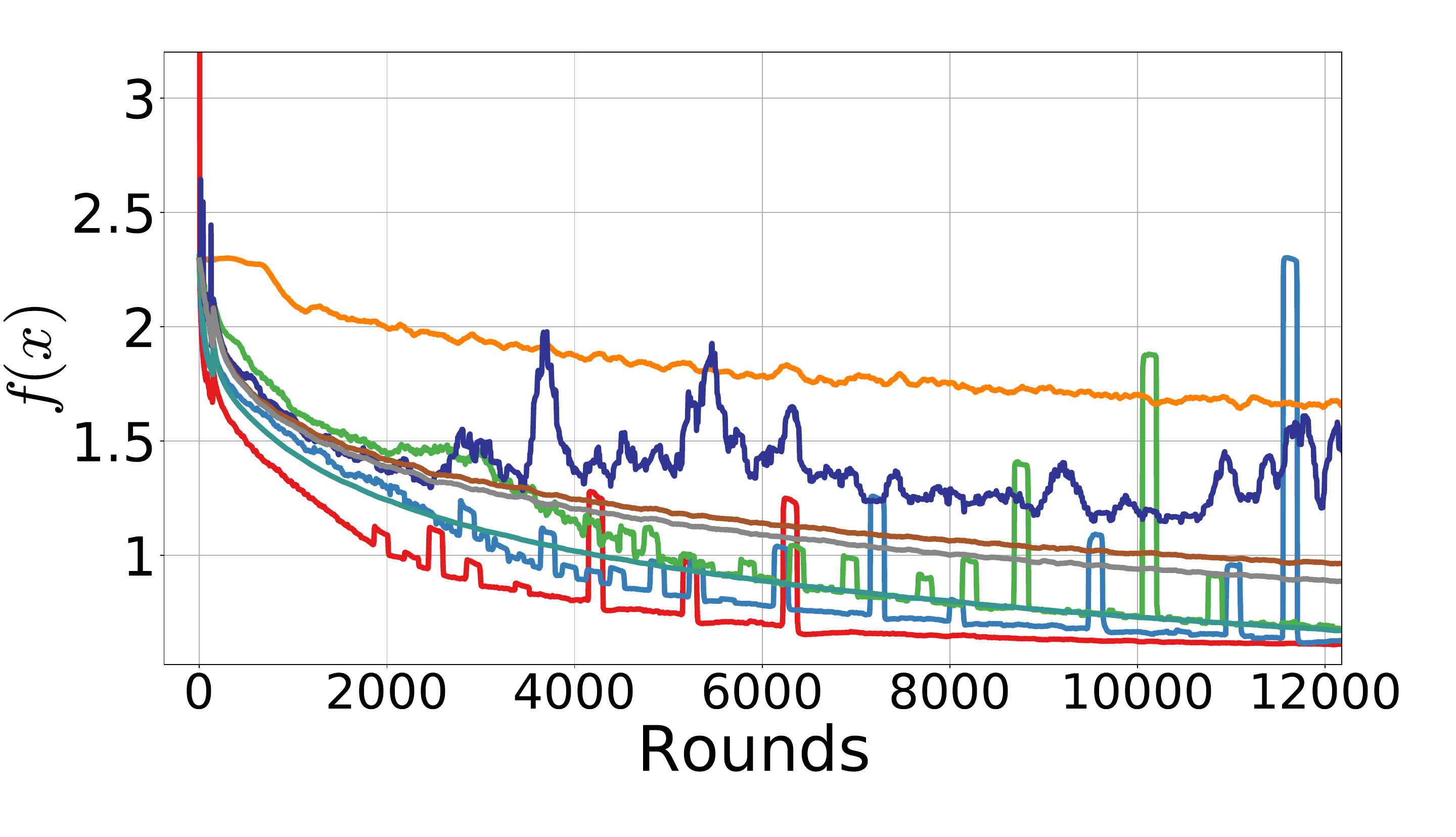}
	\caption{\textbf{Train:} Loss functional value}
\end{subfigure}
\begin{subfigure}[ht]{0.495\textwidth}
	\includegraphics[width=\textwidth]{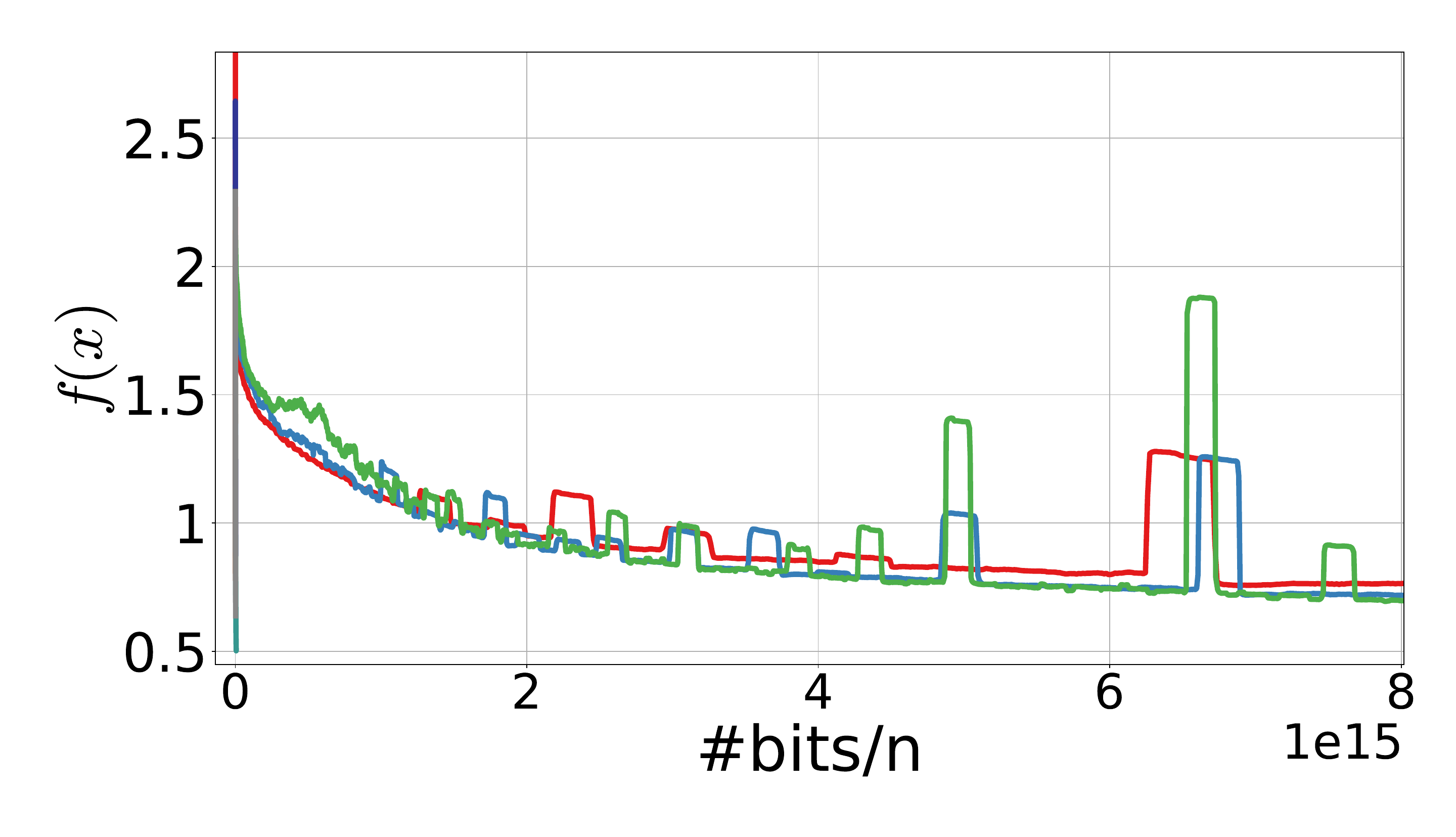}
	\caption{\textbf{Train:} Loss functional value}
\end{subfigure}

\vspace{0.25cm} 
\rule{\textwidth}{0.4pt} 
\vspace{0.25cm} 

\begin{subfigure}[ht]{0.495\textwidth}
	\includegraphics[width=\textwidth]{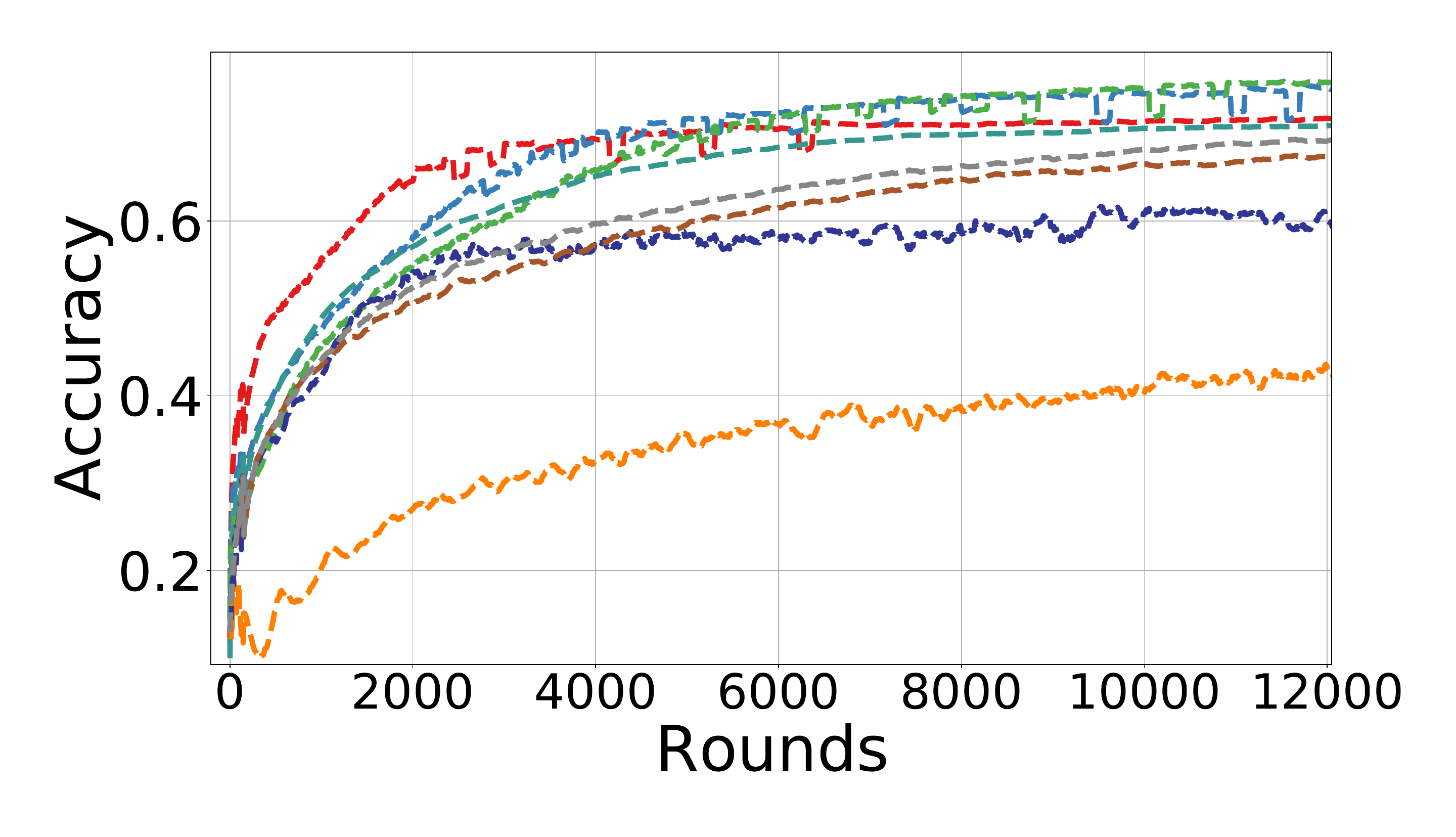} 
	\caption{\textbf{Test:} Top-1 accuracy}
\end{subfigure}
\begin{subfigure}[ht]{0.495\textwidth}
	\includegraphics[width=\textwidth]{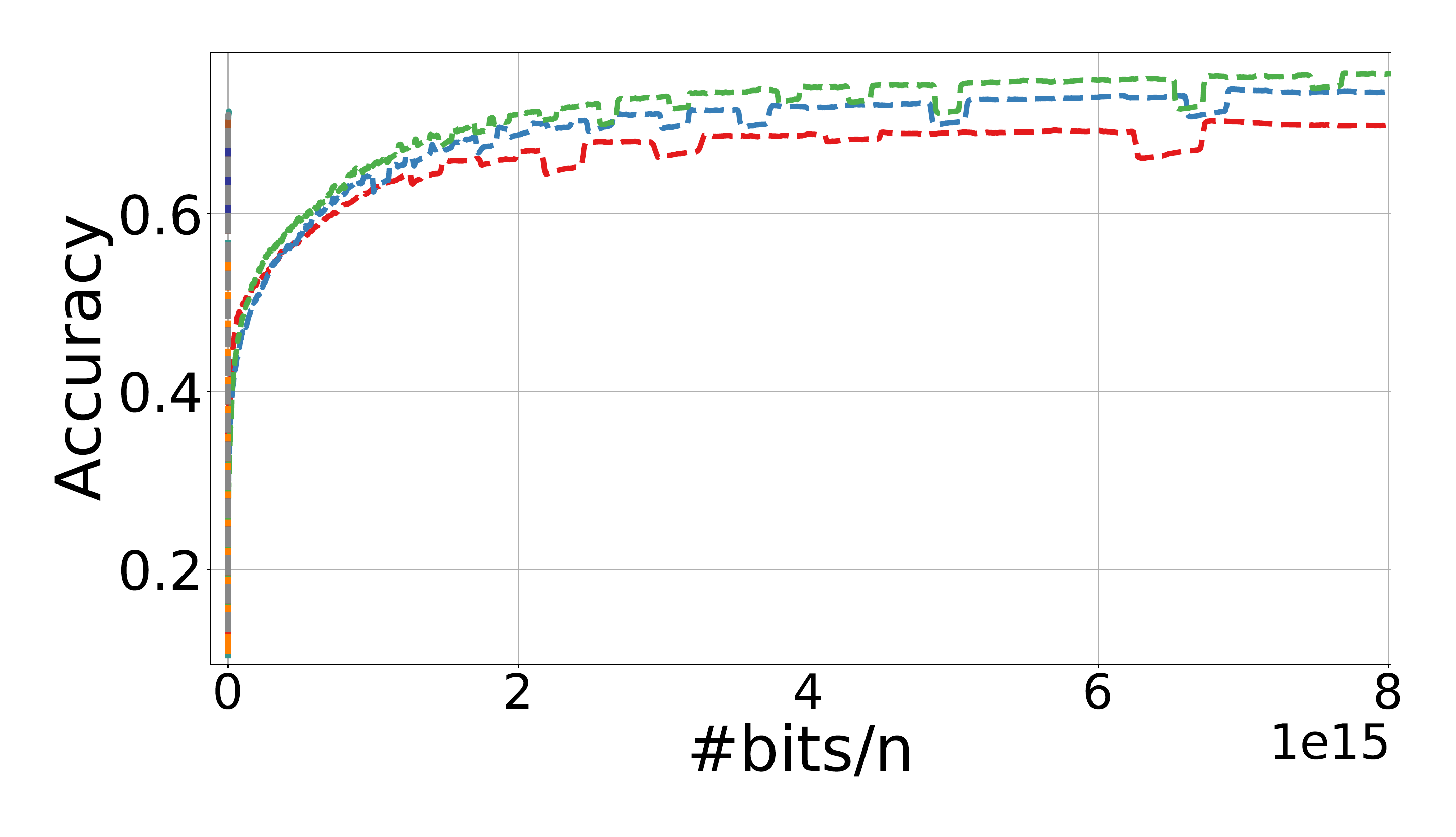}
	\caption{\textbf{Test:} Top-1 accuracy}
\end{subfigure}

\begin{subfigure}[ht]{0.495\textwidth}
	\includegraphics[width=\textwidth]{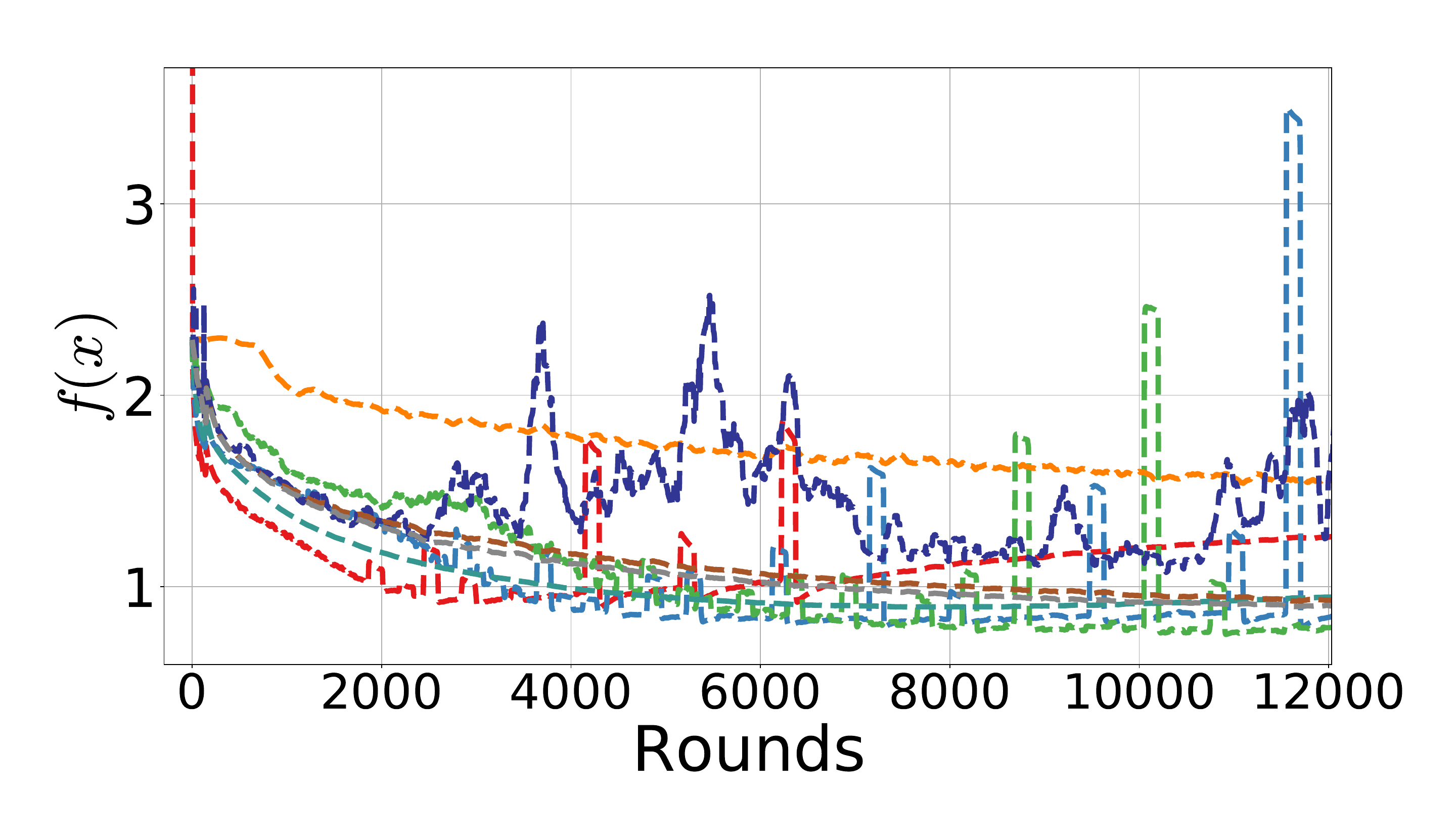} 
	\caption{\textbf{Test:} Loss functional value}
\end{subfigure}
\begin{subfigure}[ht]{0.495\textwidth}
	\includegraphics[width=\textwidth]{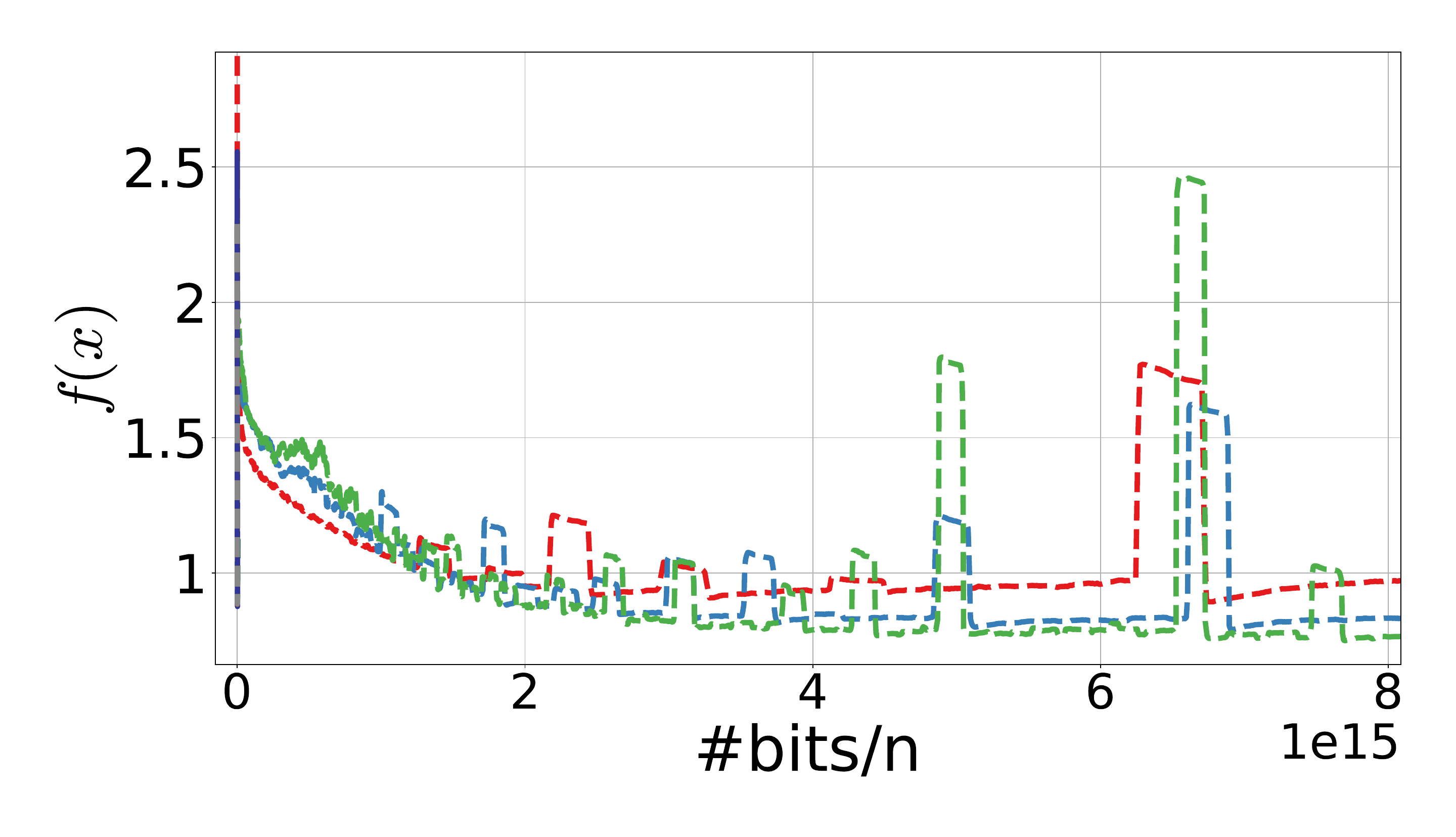} 
	\caption{\textbf{Test:} Loss functional value}
\end{subfigure}

\caption{{Training \modelname{DenseNet-121} on \dataname{CIFAR-10} with $n=10$ workers. Plots (a)--(d) show behavior on the \textit{train set}, while (e)--(h) show behavior on the \textit{test set}.}}
\label{ch6:fig:training_densenet}
\end{figure*}


\begin{figure*}[t]
\centering

\begin{subfigure}[ht]{0.5\textwidth}
	\includegraphics[width=\textwidth]{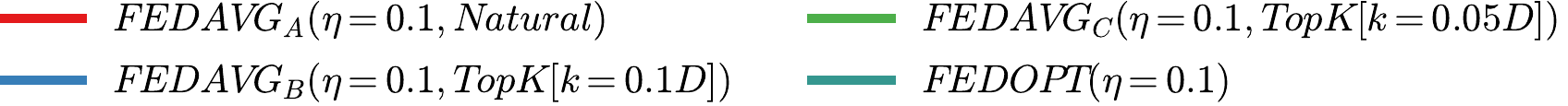}
\end{subfigure}
\begin{subfigure}[ht]{1.0\textwidth}
	\vspace{0.1cm}
	\includegraphics[width=\textwidth]{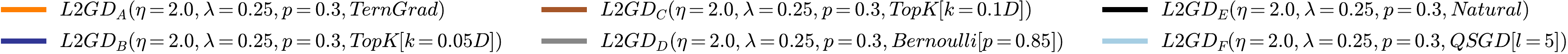}
\end{subfigure}

\begin{subfigure}[ht]{0.495\textwidth}
	\includegraphics[width=\textwidth]{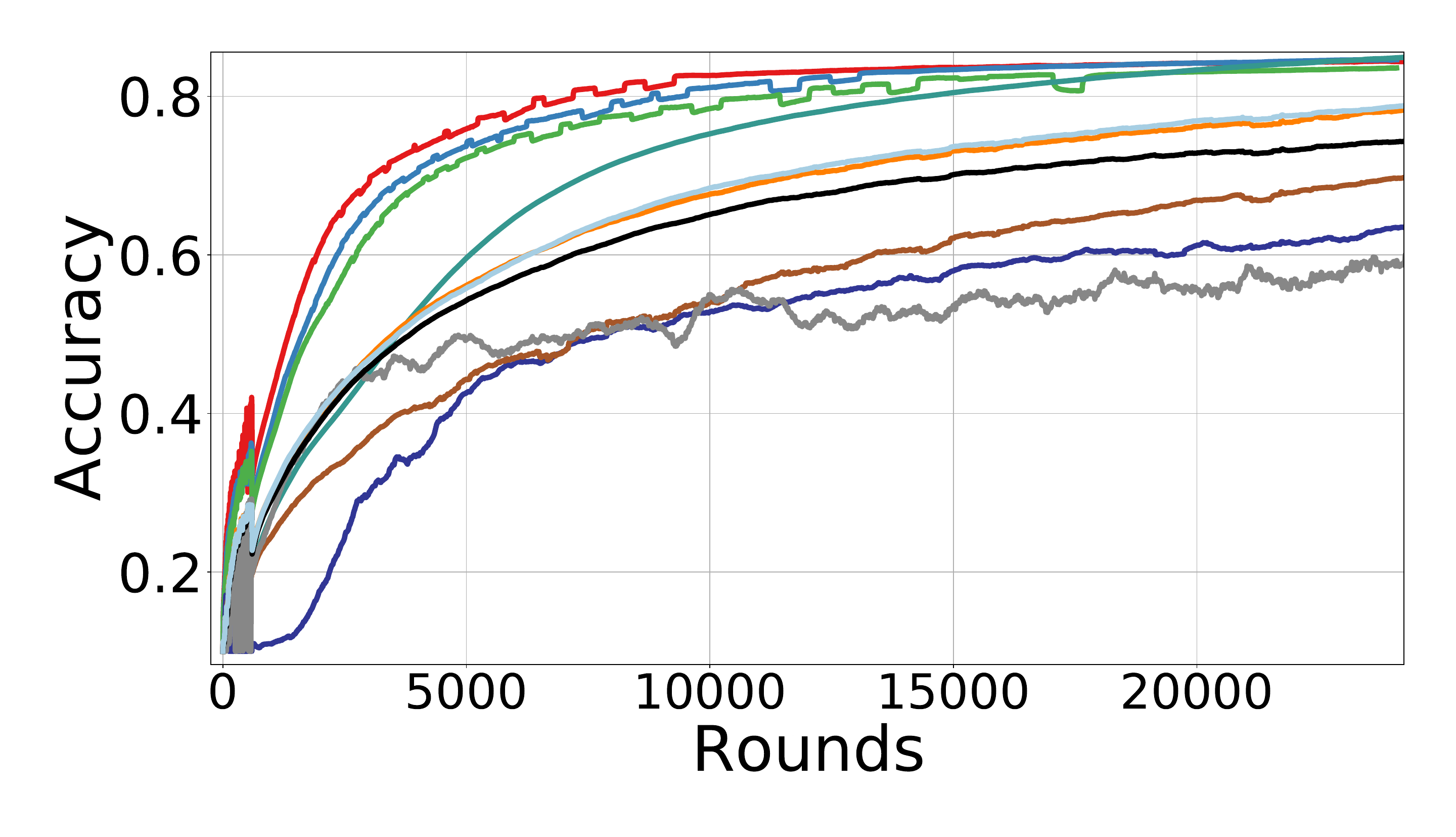}
	\caption{\textbf{Train:} Top-1 accuracy}
\end{subfigure}
\begin{subfigure}[ht]{0.495\textwidth}
	\includegraphics[width=\textwidth]{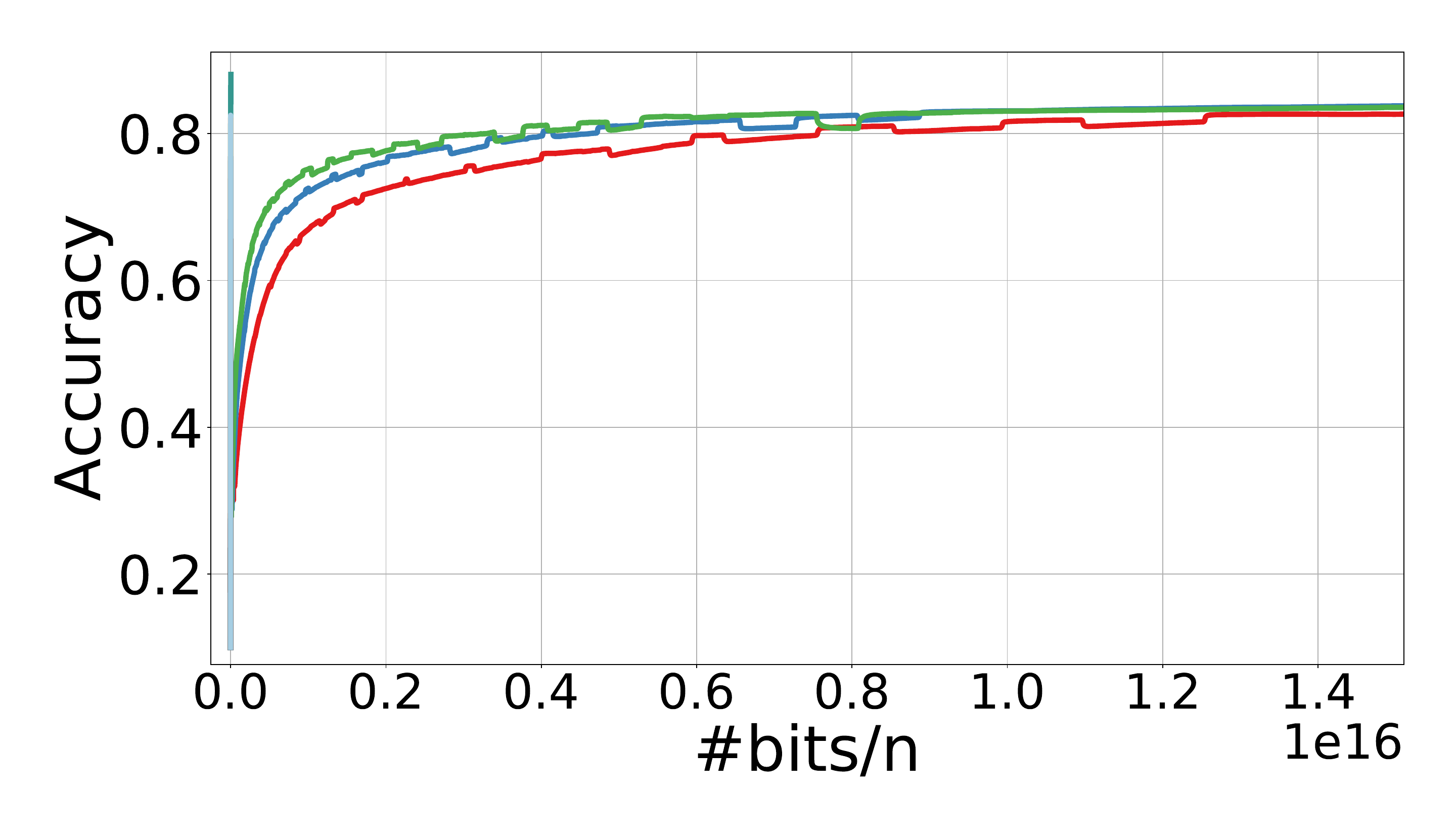}
	\caption{\textbf{Train:} Top-1 accuracy}
\end{subfigure}

\begin{subfigure}[ht]{0.495\textwidth}
	\includegraphics[width=\textwidth]{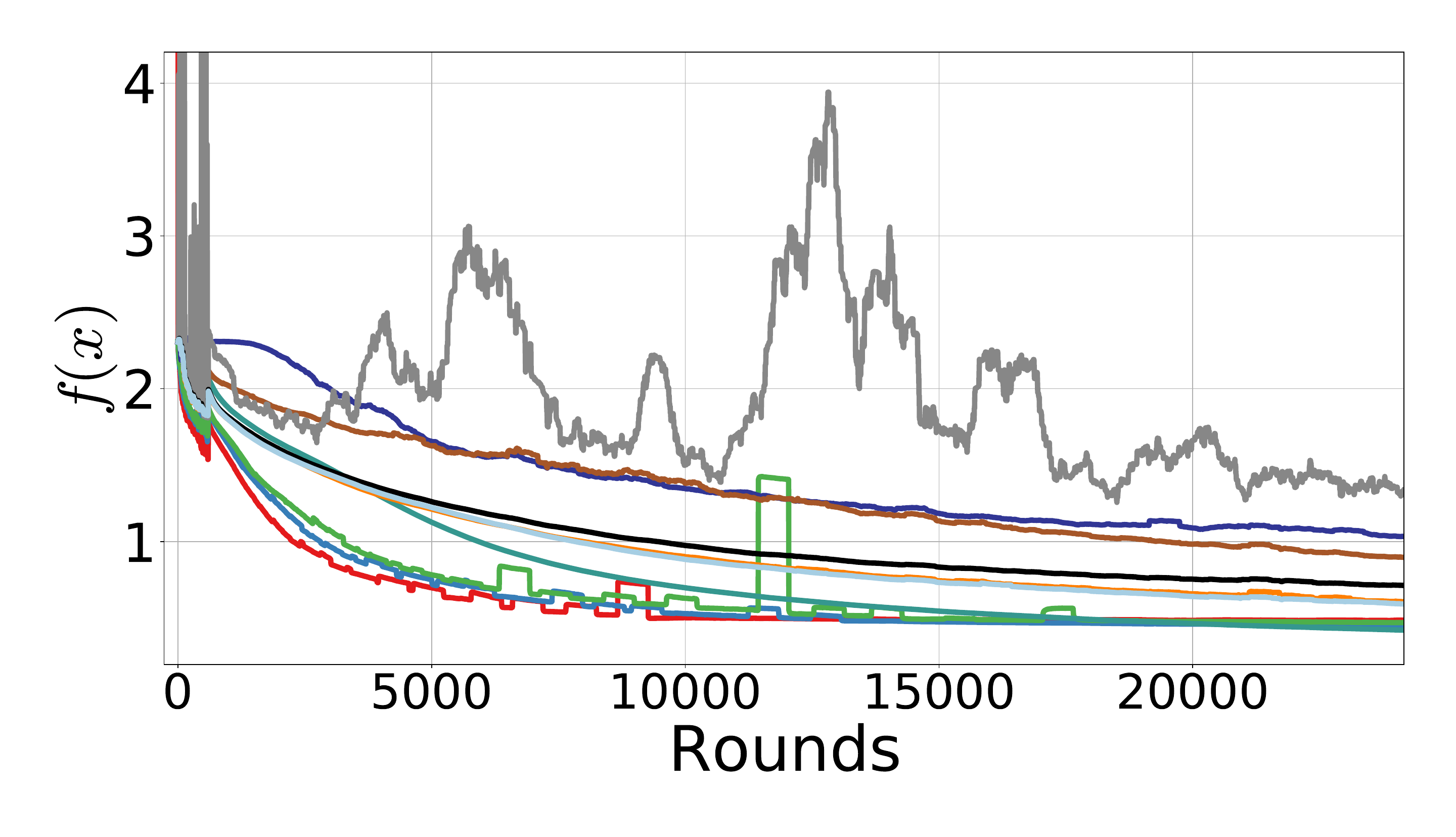}
	\caption{\textbf{Train:} Loss functional value}
\end{subfigure}
\begin{subfigure}[ht]{0.495\textwidth}
	\includegraphics[width=\textwidth]{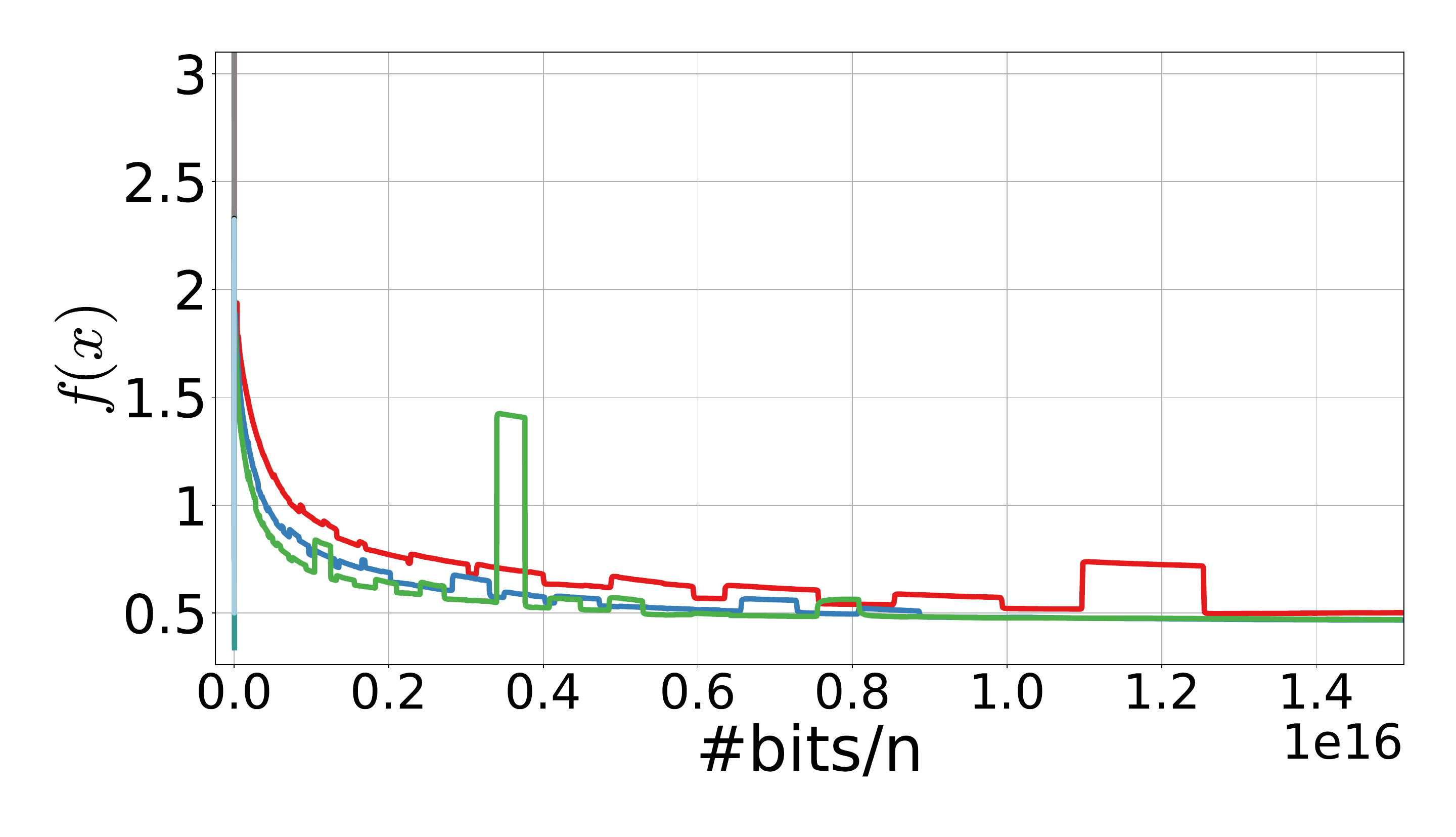}
	\caption{\textbf{Train:} Loss functional value}
\end{subfigure}

\vspace{0.25cm} 
\rule{\textwidth}{0.4pt} 
\vspace{0.25cm} 

\begin{subfigure}[ht]{0.495\textwidth}
	\includegraphics[width=\textwidth]{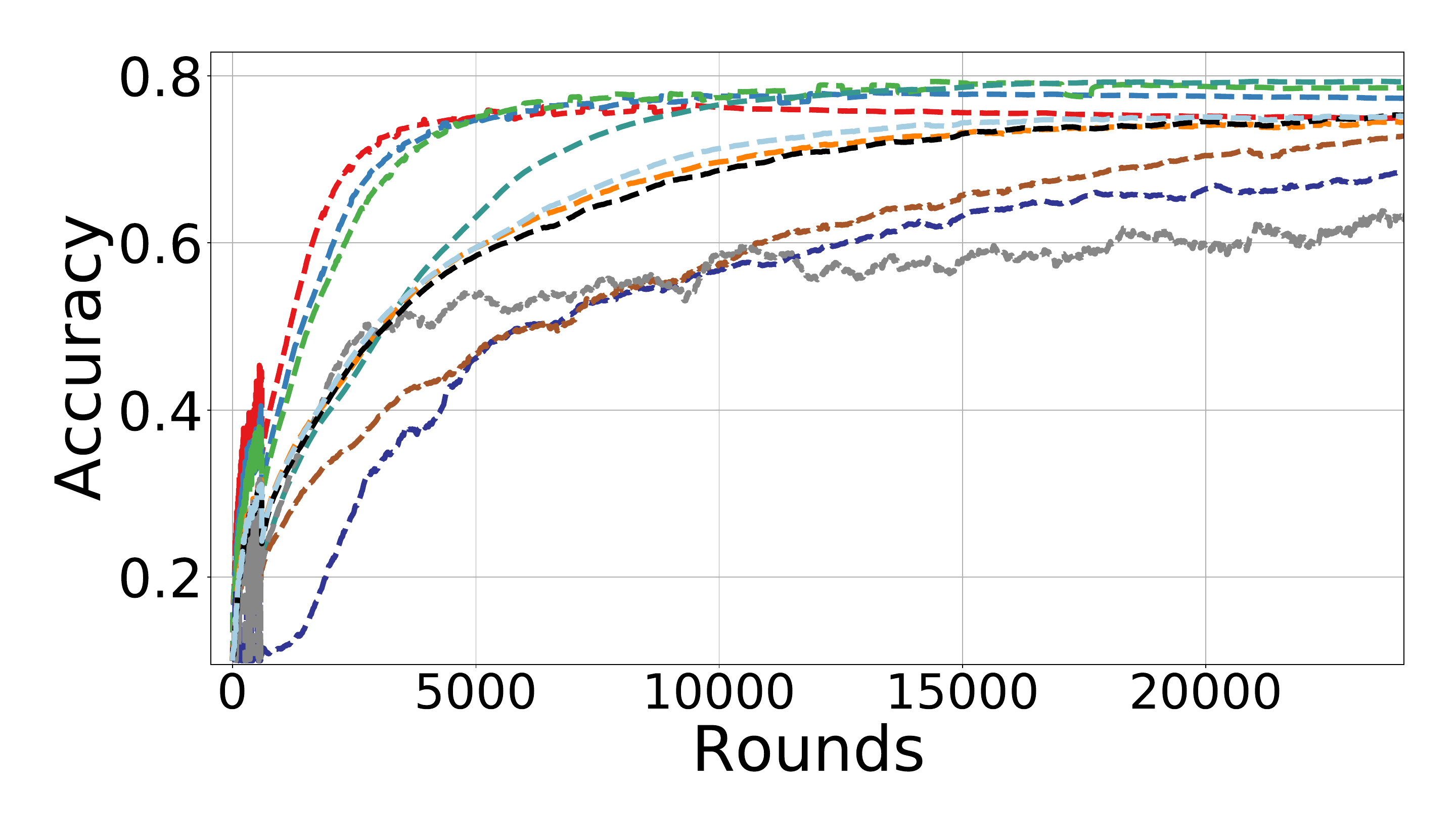} 
	\caption{\textbf{Test:} Top-1 accuracy}
\end{subfigure}
\begin{subfigure}[ht]{0.495\textwidth}
	\includegraphics[width=\textwidth]{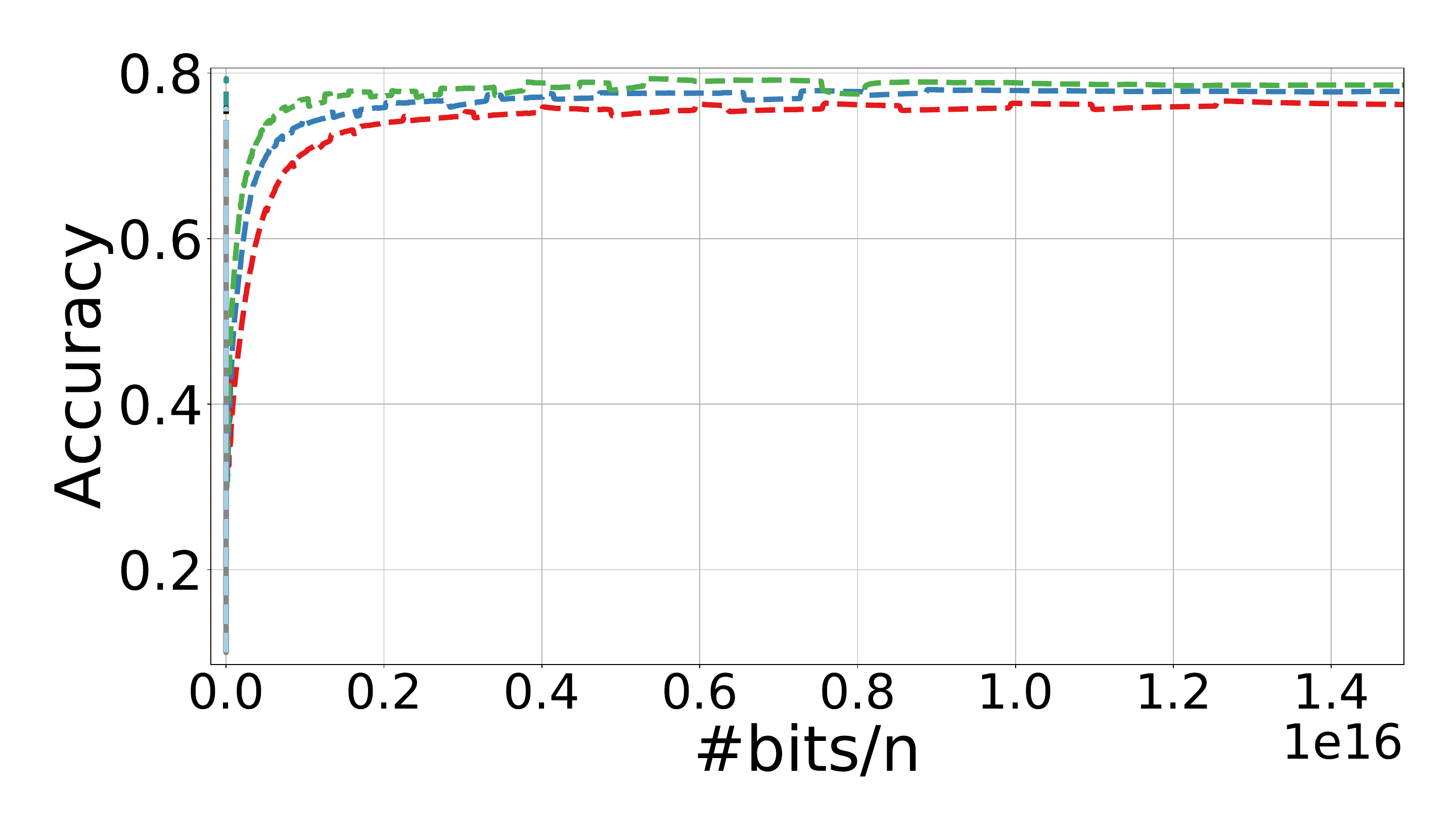}
	\caption{\textbf{Test:} Top-1 accuracy}
\end{subfigure}

\begin{subfigure}[ht]{0.495\textwidth}
	\includegraphics[width=\textwidth]{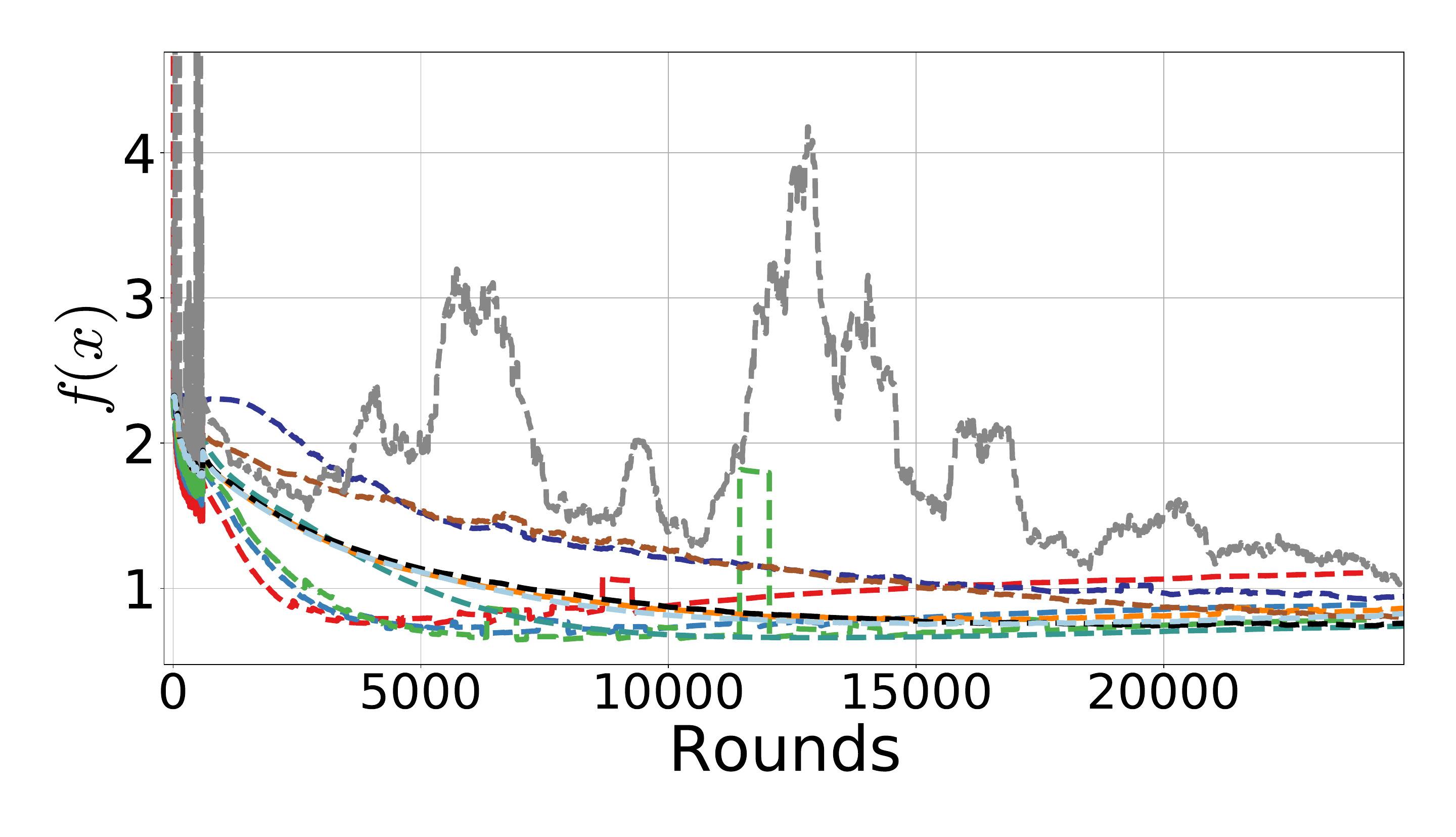} 
	\caption{\textbf{Test:} Loss functional value}
\end{subfigure}
\begin{subfigure}[ht]{0.495\textwidth}
	\includegraphics[width=\textwidth]{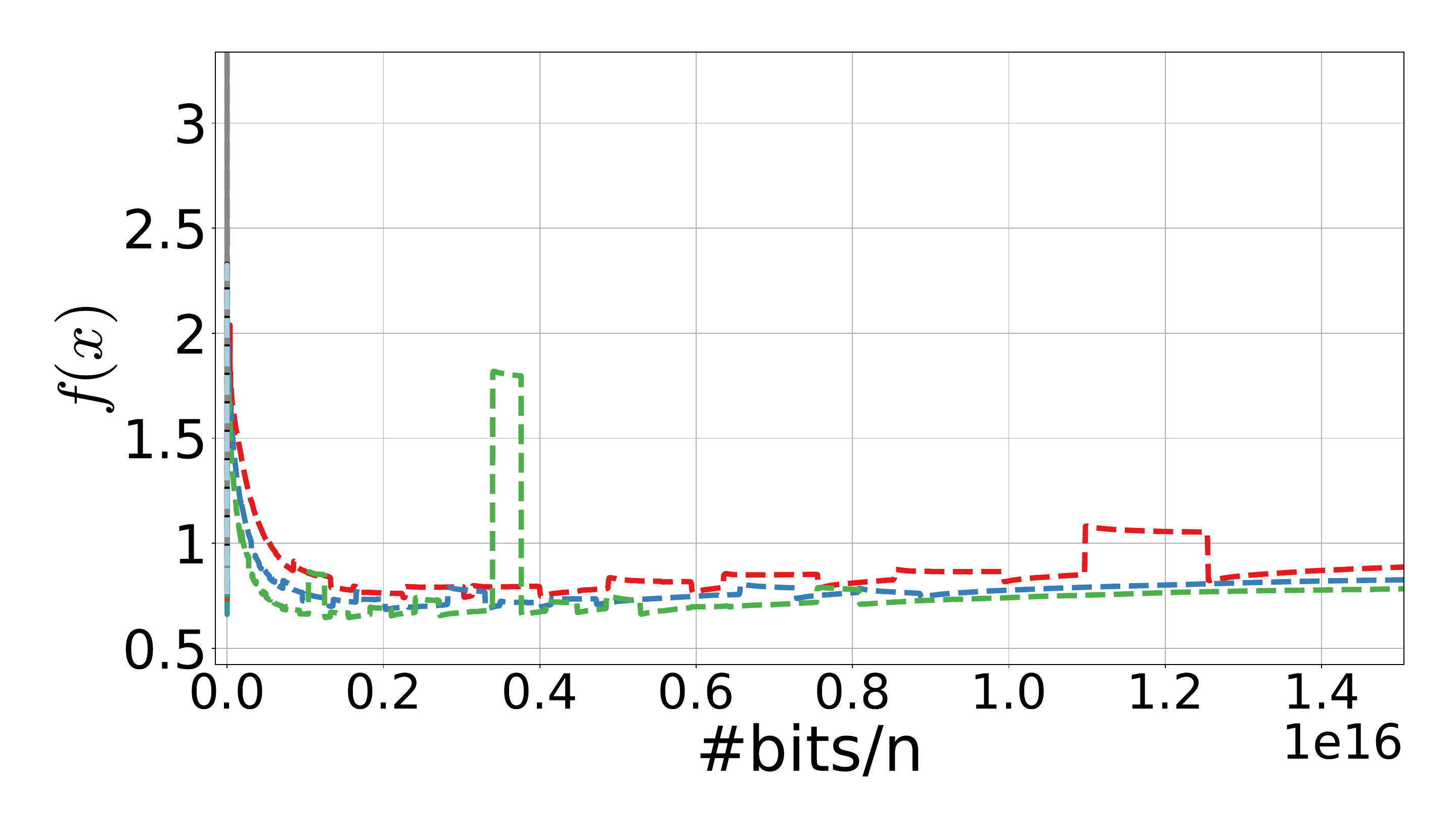} 
	\caption{\textbf{Test:} Loss functional value}
\end{subfigure}

\caption{{Training \modelname{MobileNet} on \dataname{CIFAR-10} with $n=10$ workers. Plots (a)--(d) show behavior on the \textit{train set}, while (e)--(h) show behavior on the \textit{test set}.}}
\label{ch6:fig:training_mobilenet}
\end{figure*}

\section{Experiments}
\label{ch6:sec:empirical}

We conducted diverse numerical experiments with \algname{L2GD} algorithm that includes: \myNum{i}~Analysis of algorithm meta-parameters for \modelname{logistic regression} in strongly convex setting; see Section~\ref{ch6:sec:meta};~\myNum{ii}~analysis of \algname{Compressed L2GD} algorithm on image classification with DNNs; see Section~\ref{ch6:sec:dnn}. 

\smartparagraph{Computing environment.} We performed experiments on server-grade machines running Ubuntu 18.04 and Linux Kernel v5.4.0, equipped with 8-cores $3.3$ GHz Intel Xeon and a single NVIDIA GeForce RTX 2080 Ti.Tesla-V100-SXM2 GPU with 32GB of GPU memory. 

The computation backend for \modelname{logistic regression} experiments was \libname{NumPy} library with leveraging \libname{MPI4PY} for inter-node communication. For DNNs, we used a recent version of \libname{FedML} \citep{he2020fedml} benchmark\footnote{{FedML.AI}} and patched it with: \myNum{i} distributed and standalone version of Algorithm \ref{ch6:alg:ComL2GD}; \myNum{ii} serializing and plotting mechanism; \myNum{iii} modifications in standalone, distributed version of \algname{FedAVG}~\citep{mcmahan17fedavg} and \algname{FedOpt}~\citep{reddi2020adaptive} to be consistent with \eqref{ch6:eq:problem}; \myNum{iv} not to drop the last batch while processing the dataset.

\subsection{Meta-parameter study}\label{ch6:sec:meta}
The purpose of these experiments is to study the meta-parameters involved in the uncompressed \algname{L2GD} algorithm. We used \algname{L2GD} algorithm without compression for solving $\ell_2$ regularized \modelname{logistic regression} on LIBSVM \dataname{A1A} and \dataname{A2A} datasets \citep{libsvm}. Both datasets contain shuffled examples in the train set, and we did not perform any extra shuffling. To simulate the FL settings, we divided both datasets into $5$ parts. After splitting, each worker has $321$ and $453$ records for \dataname{A1A}, and \dataname{A2A}, respectively.

\smartparagraph{Setup and results.}
We define $f_i(x)$ to be local empirical risk minimization for logistic loss with additive regularization term for local data $D_i$ and of the form:
\[f_i(x)=\dfrac{1}{n_i} \sum_{j=1}^{n_i}\log(1+\exp(-b^{(j)} x^{\top} a^{(j)})) +  \dfrac{L_2}{2}\|x\|^2,\]
where $a^{(j)} \in \mathbb{R}^{124}, b^{(j)} \in \{+1,-1\},n_i=|D_i|$. 
We set $L_2=0.01$, and varied meta-parameters  $p$ and $\lambda$. For each parameter, we performed $100$ iterations of Algorithm \ref{ch6:alg:ComL2GD}. Note that, as meta-parameter $\lambda$ decreases the models will fit more to its local data, while $p$ provides a stochastic balance between local gradient steps with probability, $1-p$ and aggregation with probability, $p$.

\iftrue
\begin{table*}[h!]
	\small
	\centering
	\caption{Summary of the benchmarks. The measured quantity is ${\rm bits}/n$ to achieve $0.7$ \compname{Top1} test accuracy, with $n=10$ clients. For \modelname{DenseNet-121}, \modelname{MobileNet}, \modelname{ResNet-18} the baseline is \algname{FedAVG} with\compname{Natural} compressor with $1$ local epoch; \algname{L2GD} also uses \compname{Natural} compressor.}
	
	\begin{tabular}{cccc}
		\toprule
		\textbf{Model} & \makecell{\textbf{Training} \\ \textbf{Parameters}} & \makecell{\textbf{L2GD} \\ \textbf{${\rm bits}/n$}} & \makecell{\textbf{Baseline} \\ \textbf{${\rm bits}/n$}}   \\\hline
		\\
		\modelname{DenseNet-121} &	$~79\times 10^5$ & $8\times 10^{11}$ & $4\cdot10^{15}$ \\ 
		
		\modelname{MobileNet} &	$~32\times 10^5$ & $1.7\times 10^{11}$ & $1\times 10^{15}$ \\ 
		\modelname{ResNet-18} & $~11\times 10^6$ & $1.1 \times 10^{12}$  & $1.5 \times 10^{16}$\\
		\hline
	\end{tabular}
	\label{ch6:tab:modelsize}
\end{table*}

\begin{figure}[h!]
	\centering
	\includegraphics[width=.48\linewidth]{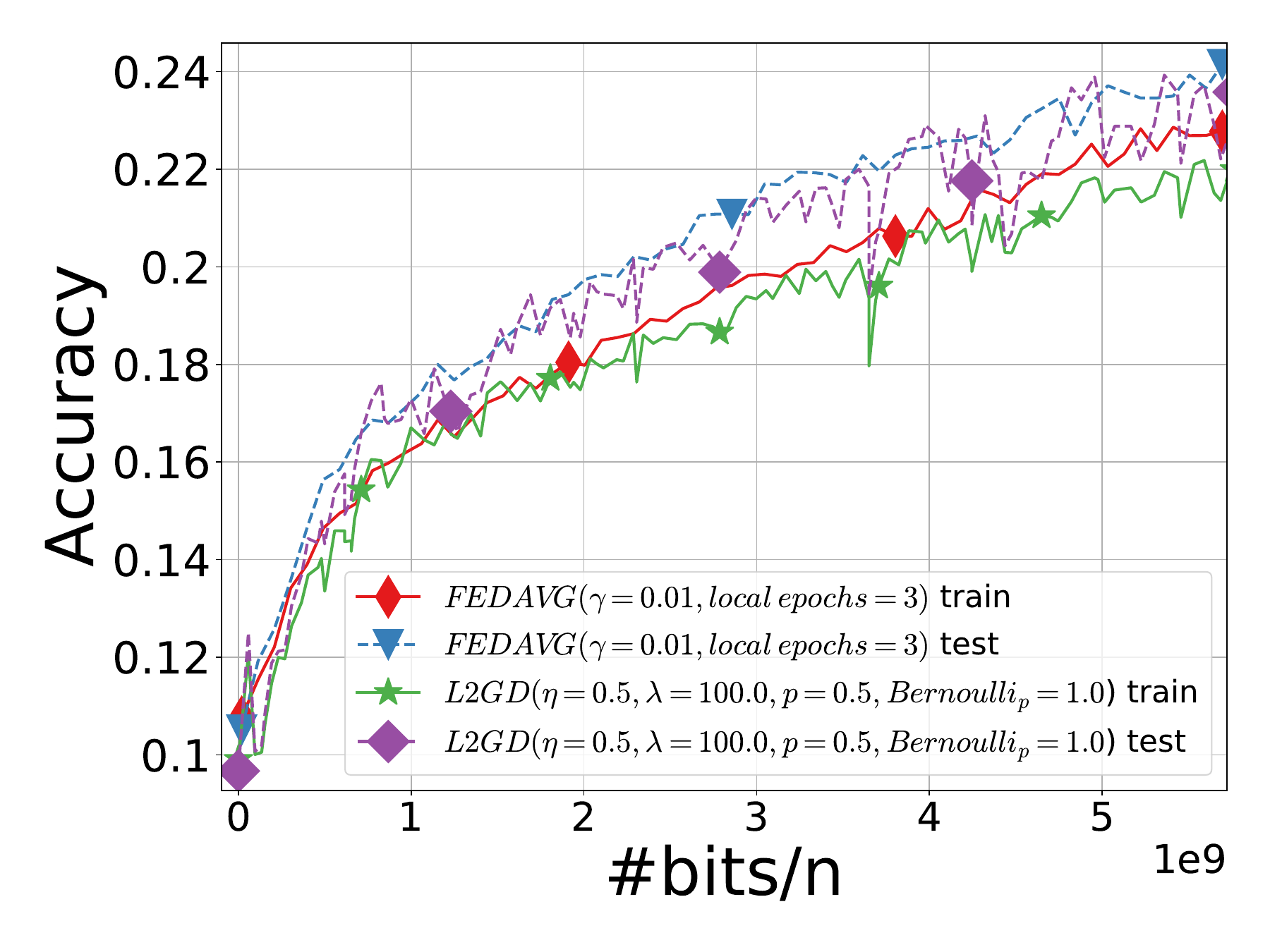}
	\caption{The \algname{FedAVG} as a particular case of \algname{L2GD}: Test and train accuracy for \modelname{ResNet-56} on \dataname{CIFAR-10}.}
	\label{ch6:fig:nn_experiments_fedavg_like_acc}
\end{figure}
\begin{figure}[h!]
	\centering
	\includegraphics[width=.48\linewidth]{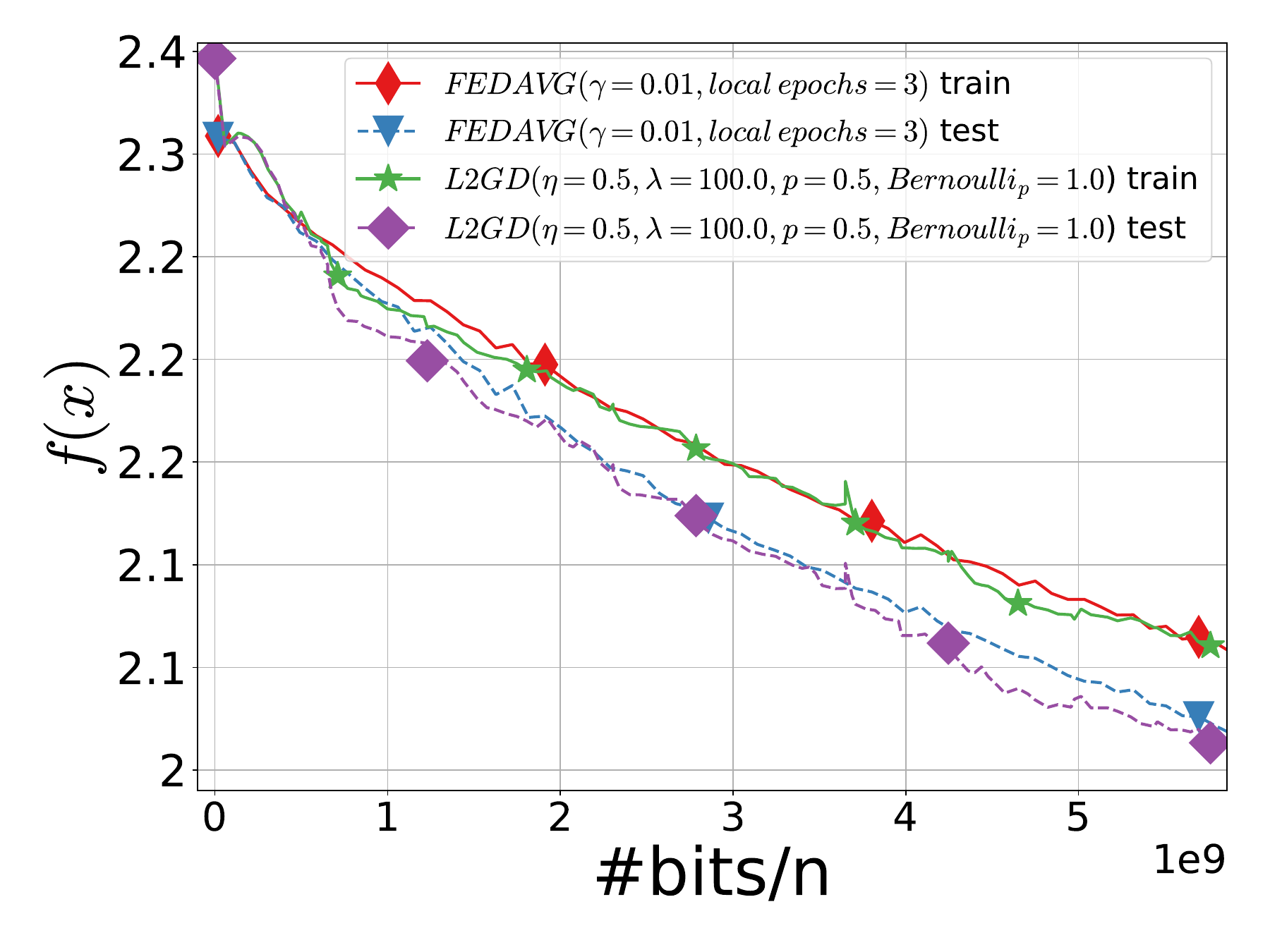}
	\caption{The \algname{FedAVG} as a particular case of \algname{L2GD}: Test and train loss for \modelname{ResNet-56} on \dataname{CIFAR-10}.}
	\label{ch6:fig:nn_experiments_fedavg_like_f}
\end{figure}
\fi


\smartparagraph{Takeaway message.} The results in Figure~\ref{ch6:fig:lambda_and_p_selection} support the theoretical finding---there exists an optimal choice of $(p, \lambda$), where the loss function, $f$ achieves the least value. Nevertheless, this choice is problem-dependent. Additionally, we find small $p$ is not good due to the lack of samples in a single node compared to samples available at other nodes. There is a trade-off for each node in learning from the other nodes' data and spending time to learn from its own data. In the experiments, the ``optimal" setting of our algorithm is attained for $p=0.4$ and $\lambda$ in $[0,25]$.
Finally, we observe that \emph{to get the smallest errors on the training and validation sets, it is better not to  perform the {averaging step} too often}. 

\subsection{Training DNN models}\label{ch6:sec:dnn}
We choose three practically important DNN models used for image classification, and other down-streaming tasks, such as feature extractions for image segmentation, object detection, image embedding, and image captioning.
\begin{itemize}
\item \modelname{ResNet-18}~\citep{resnet}. The overwhelmingly popular \modelname{ResNet} architecture exploits residual connections to remedy vanishing gradients. The network supported the trend toward smaller filters and deeper architectures. Additionally, we use \modelname{ResNet-56}~\citep{resnet}. 

\item \modelname{DenseNet}~\citep{densenet} contains a short connection between layers via connecting each layer to every other layer in a feed-forward fashion. A dense connection allows propagating information to the final classifier via concatenating all feature maps. Each layer in \modelname{DenseNet} is narrow and contains only 12 filters---another practical model for FL training. 
\item \modelname{MobileNet}~\citep{howard2017mobilenets}. DNN architecture has a trade-off between computational complexity and accuracy \cite[p.3, Figure 1]{bianco2018benchmark}. For mobile devices that appear in cross-device FL, the computation cost and energy consumption are both important. The energy consumption is mostly driven by memory movement \citep{chen2018understanding, horowitz20141}. In \modelname{MobileNet} architecture standard convolution blocks perform depth-wise convolution followed by $1\times 1$ convolution. This is computationally less expensive in flops during inference time (see \cite[Figure 1, p.3]{bianco2018benchmark}) and is $\sim3.5\times$ more power efficient compared to \modelname{DenseNet} \cite[p.85, Table 7]{garcia2019estimation}. This makes \modelname{MobileNet} an attractive model for FL training. 
\end{itemize}

\smartparagraph{Dataset and Setup.}  
We consider \dataname{CIFAR-10} dataset \citep{krizhevsky2009learning} for image classification. It contains color images of resolution $28\times28$ from 10 classes. The training and the test set are of size, $5\times 10^4$ and $10^4$, respectively. The training set is partitioned heterogeneously across $10$ clients. The proportion of samples of each class stored at each local node is drawn by using the Dirichlet distribution ($\alpha=0.5$). In our experiments, all clients are involved in each communication round. Additionally, we added a linear head in all CNN models for \dataname{CIFAR-10}, as they were originally designed for classification tasks with $1000$ output classes. 

\smartparagraph{Loss function.} Denote $${f_i(x)=w_i \cdot \dfrac{1}{|D_i|} \sum_{ (a_i, b_i) \in D_i }^{} l(a_i, b_i, x)}$$ 
to be a weighted local empirical risk associated with the local data, $\mathcal{D}_i$ stored in node, $i$. We note that $l(a_i,b_i,x)$ is a standard unweighted cross-entropy loss, $a_i \in \mathbb{R}^{28 \times 28 \times 3}$, $b_i \in \{0,1\}^{10}$ with only one component equal to $1$, the ground truth value, and the weight is set to $w_i={|D_i|}/{|D_1 \cup \dots \cup D_n|}$.

\smartparagraph{Metrics.} To measure the performance, we examine the loss function value, $f(x)$, and the Top-1 accuracy of the global model on both the train and the test set. Additionally, we measure the number of rounds and ${\rm bits}/n$---communicated bits normalized by the number of local clients, $n$. The intuition behind using the last metric is to measure the communicated data volume; it is widely {\em hypothesized} that the reduced data volume translates to faster training in a constant speed network in distributed setup \citep{gajjala2020huffman, xu2021grace}. 

\smartparagraph{Compressors used.} The theoretical results of compressed \algname{L2GD} are attributed to unbiased compressors. We used 4 different unbiased compressors at the clients: \compname{Bernoulli} \citep{khirirat2018distributed}, \compname{Natural} compressor \citep{horvath2019natural}, random dithering a.k.a. \compname{QSGD} \citep{alistarh2017qsgd}, and \compname{Terngrad} \citep{DBLP:conf/nips/WenXYWWCL17}; see Table~\ref{ch6:tab:summary} for  details. Additionally, we note that biased compressors (mostly sparsifiers) are popular in DNN training. 
Therefore, out of scientific curiosity, we used a popular sparsifier: \compname{TopK} \citep{aji_sparse, sahu2021rethinking} as a proof of concept. We note that extending \algname{Compressed L2GD} theory for biased compressors (with or without error-feedback \citep{xu2021grace}) is nontrivial and mathematically involved, and left for future work.

\smartparagraph{Algorithms used for comparison.} We used state-of-the-art FL training algorithms, \algname{FedAVG} \citep{mcmahan17fedavg} and \algname{FedOpt} \citep{reddi2020adaptive} as no compression baseline to compare against our \algname{Compressed L2GD}. However, the performance of \algname{FedAVG} is not stable but improves with the compression mechanism. The original \algname{FedAVG} algorithm does not contain any compression mechanism, but for comparison, we incorporated compressors into \algname{FedAVG} via the following schema which is similar to the classic error feedback \citet{xu2021grace}: \myNum{i} After local steps, client estimates change of current iterate from the previous round and formulates direction, ${g_{c,\mathrm{computed}}}^{i}$; \myNum{ii} client sends compressed difference between previous gradient estimator from previous round and currently computed gradient estimator, $\mathcal{C}({g_{c,\mathrm{computed}}}^{i} - {g_c}^{i-1})$ to the master; \myNum{iii} both master and client updating ${g_c}^{i}$ via the following schema: ${g_c}^{i} = {g_c}^{i-1} + \mathcal{C}({g_{c,\mathrm{computed}}}^{i} - {g_c}^{i-1})$. We provide the details about step size and batch size selection in the Appendix~\ref{ch6:sec:app_nn}.

\subsubsection{Results}
We show the results for training \modelname{ResNet-18}, \modelname{DenseNet-121}, and \modelname{MobileNet} with \algname{Compressed L2GD} and other state-of-the-art FL algorithms in Figures~\ref{ch6:fig:training_resnet},~\ref{ch6:fig:training_densenet},~\ref{ch6:fig:training_mobilenet}. For these experiments, the communication rounds are set to $12\times 10^3$, $25\times 10^3$, and $20\times 10^3$, respectively. For \algname{FedAVG} algorithm, each client performs one epoch over the local data. We empirically tried $1,2,3,$ and $4$ epochs over the local data as local steps, but one epoch is empirically provided the best results. 

For training \modelname{ResNet-18}, from Figure~\ref{ch6:fig:training_resnet} we observe that \algname{FedAVG} with compression has albeit better convergence than no compression \algname{FedAVG} \footnote{ We have observed that batch normalization \citep{ioffe2015batch} in \modelname{ResNet} is sensitive for aggregation; see our discussion in Section~\ref{ch6:sec:app_nn}.}. At the same time, compressed \algname{FedAVG} affects the convergence as a function of communicated rounds only negligibly (see Figure~\ref{ch6:fig:training_resnet}~(b,d)). Therefore, for training other DNN models we use \algname{FedAVG} with compression and \algname{FedOpt} without any compressors to enjoy the best of both baselines.

\smartparagraph{Take away message.} The \algname{Compressed L2GD} with \compname{Natural} compressor sends the least data and drives the loss down the most in these experiments. At the same time, \algname{L2GD} with \compname{Natural} compressor (by design it has a smaller variance) reaches the best accuracy for both train and test sets. \algname{Compressed L2GD} outperforms \algname{FedAVG} by a huge margin---For all DNN experiments, to reach the desired \compname{Top1} test accuracy, \algname{Compressed L2GD} reduces the communicated data-volume, $\mathrm{\#bits/n}$, from $10^{15}$ to $10^{11}$, rendering approximately a $10^4$ times improvement compared to \algname{FedAVG}; see Table~\ref{ch6:tab:modelsize}.

Interestingly, in training \modelname{MobileNet}, the performance of biased \compname{TopK} compressor degrades only about 10\% compared to \compname{Natural} compressor, while approximately degrades 35\% in training \modelname{DenseNet}. See  discussion in Section~\ref{ch6:sec:app_nn}, Figures~\ref{ch6:fig:training_resnet_l2g_only}, \ref{ch6:fig:training_densenet_l2g_only}, \ref{ch6:fig:training_mobilenet_l2g_only}. These phenomena may lead the researchers to design unbiased compressors with smaller variance to empirically harvest the best behavior of \algname{Compressed L2GD}.

Nevertheless, we also observe that \algname{Compressed L2GD} converges slower compared to other FL algorithms without compression in all cases. What follows, it can be argued, is that when we compare the communicated data volume for all DNN models, the convergence of \algname{Compressed L2GD} is much better. Additionally, the gain in terms of lowering the loss function value is significant.
\begin{center}
	\emph{By sending the same amount of data, \algname{L2GD} lowers the loss the most compared to the other no-compression FL baseline algorithms.} 
\end{center}

These experiments also demonstrate that when communication is a bottleneck, \algname{FedAVG} is not comparable with \algname{L2GD}. The only comparable baseline for \algname{L2GD} is \algname{FedOpt}. Also, see  discussion in Section~\ref{ch6:sec:app_nn}, and Figures~\ref{ch6:fig:training_resnet_l2g_only}, \ref{ch6:fig:training_densenet_l2g_only}, \ref{ch6:fig:training_mobilenet_l2g_only}. A similar observation holds for the \compname{Top1} test and train accuracy. Taken together, these indicate that for training larger DNN models in a personalized FL setting, with resource-constrained and geographically remote devices, \algname{Compressed L2GD} could be the preferred algorithm because its probabilistic communication protocol sends less data but obtains better test accuracy than no compression \algname{FedAVG} and \algname{FedOpt}.

Additionally, we observe that when $\nicefrac{\eta \lambda}{n p} \in [0.5, 0.95],$ \algname{Compressed L2GD} incurs a significant variance in objective function during training. Empirically, the best behavior was observed for $\nicefrac{\eta \lambda}{n p} \approx 1$ or $\nicefrac{\eta \lambda}{n p} \in (0, 0.17]$.

\smartparagraph{FedAVG as a particular case of L2GD.} 
We note that if $$\dfrac{\eta \lambda}{n p}=1,$$
then the aggregation step of Algorithm \ref{ch6:alg:ComL2GD} reduces to $x_i^{k+1}=\bar{x}^k$, for all devices. Thus, in this regime \algname{L2GD} works similarly as \algname{FedAVG} with a random number of local steps. E.g., if $p=0.5$, then Algorithm \ref{ch6:alg:ComL2GD} reduces to the randomized version of \algname{FedAVG} with an average of $3$ local steps.  Figures~\ref{ch6:fig:nn_experiments_fedavg_like_acc} and \ref{ch6:fig:nn_experiments_fedavg_like_f}, confirm this observation numerically, where we see that both algorithms exhibit similar performance. In that experiment we trained \modelname{ResNet-56} on \dataname{CIFAR-10} with $n=100$ workers, and  $600$ rounds. For \algname{L2GD} in this experiment, we set $\nicefrac{\eta \lambda}{p n}=1$.

\section{Conclusions} 

In this work, we equipped the loopless gradient descent (\algname{L2GD}) algorithm with a compression mechanism to reduce the communication bottleneck between local devices and the server in an FL context. We showed that the new algorithm enjoys similar convergence properties as the uncompressed \algname{L2GD} with a natural increase in the stochastic gradient variance due to compression. This phenomenon is similar to classical convergence bounds for compressed \algname{SGD} algorithms. We also show that in a personalized FL setting, there is a trade-off that must be considered by devices between learning from other devices' data and spending time learning from their own data. However, a particular parameterization of our algorithm recovers the well-known \algname{FedAVG} Algorithm. We assessed the performance of the new algorithm compared to the state-of-the-art and validated our theoretical insights through a large set of experiments.

%



\clearpage
\appendix

\part*{Appendices to Chapter \ref{chapter6}}
\label{ch6:app:toc_1}
\newpage

\phantomsection
\addcontentsline{toc}{chapter}{Appendices to Chapter 6}

\addtocounter{adjsection}{1}
\section{Overview of Convergence Analysis Results}\label{ch6:sec:Proofs}

In the Appendix, we provide the proofs of both the convex and non-convex convergence results for \algname{Compressed L2GD} algorithm. In Section~\ref{ch6:sec:app convergence_technical_results}, we provide the technical Lemmas necessary for the analyses. The Section~\ref{ch6:sec:app_main conv} contains the auxiliary results pertaining to both convex and non-convex convergence. In Section~\ref{ch6:sec:app_nnc} we provide the non-convex convergence results, and Section~\ref{ch6:sec:app optimal} provides the proofs for optimal rate and communication.   

\addtocounter{adjsection}{1}
\section{Technical Results Used for Convergence}\label{ch6:sec:app convergence_technical_results}
The following two Lemmas are instrumental in proving other compression-related results.

\begin{lemma*}
Let $x \in R^{nd}$, then
$${\E_{\mathcal{C}} \left[\|\mathcal{C}(x)\|^2\right]\le (1+\omega) \|x\|^2,}$$
where ${\omega = \max_{i=1,\ldots,n} \{\omega_i\}.}$
\end{lemma*}
\begin{proof}
By using Assumption 1, we have
\begin{eqnarray*}
\E_{\mathcal{C}} \left[\|\mathcal{C}(x)\|^2\right] 
&=& \E_{\mathcal{C}} \left[\sum_{i=1}^n \|\mathcal{C}_i(x_i)\|^2\right] \\
&=& \sum_{i=1}^n \E_{\mathcal{C}_i} \|\mathcal{C}_i(x_i)\|^2 \\
&\le&  \sum_{i=1}^n (1+\omega_i) \|x_i\|^2 \le (1+\omega) \|x\|^2.
\end{eqnarray*}
Hence the result. 
\end{proof}

\begin{lemma*}
Let Assumption \ref{ch6:ass:compression} hold, then for all ${k\ge 0}$, 
${\E_{\mathcal{C},\mathcal{C}_M} \left[\mathcal{C}_M(\Bar{y}^k)\right] = \Bar{x}^k.}$
\end{lemma*}

\begin{proof}
We have
\begin{eqnarray*}
\E_{\mathcal{C},\mathcal{C}_M} \left[\mathcal{C}_M(\Bar{y}^k)\right] &=&  \E_\mathcal{C} \left[\E_{\mathcal{C}_M}\left[\mathcal{C}_M(\Bar{y}^k)\right]\right] \\
&=& \E_\mathcal{C} \left[\dfrac{1}{n}\sum_{j=1}^n \mathcal{C}_j(x_j^k)\right] \\
&=& \dfrac{1}{n}  \sum_{j=1}^n \E_{\mathcal{C}_j} \left[\mathcal{C}_j(x_j^k)\right] \\
&=& \Bar{x}^k.
\end{eqnarray*}
Hence the result. 

\end{proof}

In the following Lemma, we show that based on the randomness of the compression operators, in expectation, we recover the exact average of the local models and the exact gradient for all iterations.  

\begin{lemma*}
Let Assumptions \ref{ch6:ass:compression} hold. Then for all $k\ge 0$, knowing $x^k$,     
$G(x^k)$ is an unbiased estimator of the gradient of function $F$ at $x^k$. 
\end{lemma*}

\begin{proof}
We have 
\begin{eqnarray*}
\E_{\mathcal{C},\mathcal{C}_M} \left[G_i(x^{k})\right]&=&
\left\{
\begin{array}{lll}
	\dfrac{\nabla f_i\left(x_i^k\right)}{n(1-p)} & \text{ if } \xi_k=0
	\\
	\dfrac{  \lambda }{np}\left( x_i^k - \E_{C,\mathcal{C}_M} \left[\mathcal{C}_M(\Bar{y}^k)\right]\right) & \text{ if } \xi_k=1 ~\& ~\xi_{k-1}=0, \\
	\dfrac{  \lambda}{n p} \left( x_i^k - \Bar{x}^k \right)& \text{ if } \xi_k=1 ~\& ~\xi_{k-1}=1, \\
\end{array}
\right.\\
&\overset{{\rm By~Lemma}~\ref{ch6:lem:compmean}}{=}&
\left\{
\begin{array}{ll}
	\dfrac{\nabla f_i(x_i^k)}{n(1-p)} & \text{ if } \xi_k=0,
	\\
	\dfrac{  \lambda}{n p} \left( x_i^k - \Bar{x}^k \right) & \text{ if } \xi_k=1.
\end{array}
\right.
\end{eqnarray*}
Therefore, 
\begin{eqnarray*}
\E[G_i(x^k) | x^k] &=&
\E_{\xi_k} \left[ \E_{C,\mathcal{C}_M}\left[G_i(x^k) \right]\right]\\
&=& (1-p) \dfrac{ \nabla f_i(x_i^k)}{n(1-p)} + p \dfrac{  \lambda}{n p} \left( x_i^k - \Bar{x}^k \right) \\
&=& \nabla_{x_i} f(x^k) +  \nabla_{x_i} h\left(x^k\right) = \nabla_{x_i} F(x^k).
\end{eqnarray*}
Hence the result. 
\end{proof}

\addtocounter{adjsection}{1}
\section{Main Convergence Results}\label{ch6:sec:app_main conv}
Based on the results given in the previous section, we are now set to quote our key convergence results. Our next Lemma gives an upper bound on the  iterate at each iteration. This bound is composed of two terms---the optimality gap, $F(x^k) - F(x^*)$,  and the norm of the optimal point, $\| x^*\|$.  

\begin{lemma*}
Let Assumption \ref{ch6:ass:smoothBound} hold, then 
$${\left\| x^k\right\|^2 \le \dfrac{4}{\mu} \left( F(x^k) - F(x^*) \right) + 2 \left\| x^*\right\|^2.}$$
\end{lemma*}
\begin{proof}
We have
\begin{eqnarray*}
\left\| x^k\right\|^2 &\overset{\|a+b\|^2\le 2\|a\|^2+2\|b\|^2}{\le}& 2 \left\| x^k - x^*\right\|^2 + 2 \left\| x^*\right\|^2 \\
&\le& \dfrac{4}{\mu} \left( F(x^k) - F(x^*) \right) + 2 \left\| x^*\right\|^2.
\end{eqnarray*}
Hence the result. 
\end{proof}

Recall that, inspired by the expected smoothness property \citet{Gower2019}, we use a similar idea in our convergence proofs. The next Lemma is a technical Lemma that helps us to prove the expected smoothness property \citep{Gower2019}. 
The bound in Lemma \ref{ch6:lem:mathcalA} is composed of the:
\begin{enumerate}
\item Optimality gap, $F(x^k) - F(x^*)$.
\item The difference between the gradients of $h$ at $x^k$ and $x^*$, $\left\|\nabla h(x^k) - \nabla h(x^*)\right\|$.
\item An extra constant, $\beta$, which depends on the used compressors. 
\end{enumerate}

\begin{lemma*}
Let Assumptions \ref{ch6:ass:compression} and \ref{ch6:ass:smoothBound} hold, then
\begin{eqnarray*}
 \mathcal{A}&\eqdef&\E_{\mathcal{C}_M,\mathcal{C}} \left\| x^k - \mQ\mathcal{C}_M(\Bar{y}^k) - x^* + \mQ\mathcal{C}_M(\Bar{y}^*) \right\|^2 \\
 &\le& \dfrac{4n^2}{\lambda^2}\left\|\nabla h(x^k) - \nabla h(x^*)\right\|^2 \\
 && \qquad + \alpha \left(F(x^k) - F(x^*)\right) \\
 && \qquad + \beta,    
\end{eqnarray*}
where 
\begin{eqnarray*}
\Bar{y}^* &\eqdef& \dfrac{1}{n}\sum_{j=1}^n \mathcal{C}_j(x_j^*),\\
\alpha &\eqdef& \dfrac{4\left (4\omega + 4\omega_M (1+\omega) \right)}{\mu},\\
\beta &\eqdef& 2 \left (4\omega + 4\omega_M (1+\omega) \right) \left\|{x}^*\right\|^2 \\
&&\qquad +4\E_{\mathcal{C}_M,\mathcal{C}}\left\|\mQ\mathcal{C}_M(\Bar{y}^*)-  \mQ\Bar{x}^*\right\|^2.
\end{eqnarray*}	
\end{lemma*}

\begin{proof}
We have
\begin{eqnarray*}
\lefteqn{\mathcal{A}}&& \\
&=& \E_{\mathcal{C}_M,\mathcal{C}} \left\| x^k - \mQ\Bar{x}^k + \mQ\Bar{x}^k - \mQ\mathcal{C}_M(\Bar{y}^k) - x^* + \mQ\Bar{x}^*  -\mQ\Bar{x}^* +\mQ\mathcal{C}_M(\Bar{y}^*) \right\|^2\\
&=& \E_{\mathcal{C}_M,\mathcal{C}} \left\| \left(x^k - \mQ\Bar{x}^k - x^* + \mQ\Bar{x}^*\right) \right. \\
&& \left. \qquad \qquad + \left( \mQ\Bar{x}^k - \mQ \Bar{y}^k\right) + \left( \mQ \Bar{y}^k- \mQ\mathcal{C}_M(\Bar{y}^k)\right) \right. \\
&& \left. \qquad \qquad +\left(\mQ\mathcal{C}_M(\Bar{y}^*)-  \mQ\Bar{x}^*\right) \right\|^2\\
&\le& 4\left\|x^k - \mQ\Bar{x}^k - x^* + \mQ\Bar{x}^*\right\|^2 \\ 
&& \qquad + 4\E_{\mathcal{C}}\left\| \mQ\Bar{x}^k - \mQ \Bar{y}^k\right\|^2 \\
&& \qquad + 4\E_{\mathcal{C}_M,\mathcal{C}}\left\| \mQ \Bar{y}^k- \mQ\mathcal{C}_M(\Bar{y}^k)\right\|^2\\
&& \qquad + 4\E_{\mathcal{C}_M,\mathcal{C}}\left\|\mQ\mathcal{C}_M(\Bar{y}^*)-  \mQ\Bar{x}^*\right\|^2\\
&=& 4\left\|x^k - \mQ\Bar{x}^k - x^* + \mQ\Bar{x}^*\right\|^2 \\
&&\qquad +4n\E_{\mathcal{C}}\left\| \Bar{x}^k -  \Bar{y}^k\right\|^2+4n\E_{\mathcal{C}_M,\mathcal{C}}\left\|  \Bar{y}^k- \mathcal{C}_M(\Bar{y}^k)\right\|^2\\
&& \qquad +4\E_{\mathcal{C}_M,\mathcal{C}}\left\|\mQ\mathcal{C}_M(\Bar{y}^*)-  \mQ\Bar{x}^*\right\|^2\\
&\le& 4\left\|x^k - \mQ\Bar{x}^k - x^* + \mQ\Bar{x}^*\right\|^2 \\ 
&& \qquad +4 \sum_{i=1}^n \E_{\mathcal{C}}\left\| x_i^k -  \mathcal{C}_i(x_i^k)\right\|^2 \\
&& \qquad +4 n \omega_M \E_{\mathcal{C}_M} \left\|  \Bar{y}^k\right\|^2\\
&& \qquad +4\E_{\mathcal{C}_M,\mathcal{C}}\left\|\mQ\mathcal{C}_M(\Bar{y}^*)-  \mQ\Bar{x}^*\right\|^2\\
&\le& 4\left\|x^k - \mQ\Bar{x}^k - x^* + \mQ\Bar{x}^*\right\|^2 \\
&& \qquad +4 \sum_{i=1}^n \omega_i\left\| x_i^k\right\|^2 \\ 
&& \qquad +4 \omega_M \sum_{i=1}^n (1+\omega_i)\left\|  x_i^k \right\|^2 \\
&& \qquad +4\E_{\mathcal{C}_M,\mathcal{C}}\left\|\mQ\mathcal{C}_M(\Bar{y}^*)-  \mQ\Bar{x}^*\right\|^2\\
&\le & 4\dfrac{n^2}{\lambda^2}\left\|\nabla h(x^k) - \nabla h(x^*)\right\|^2 \\
&& \qquad + \left (4\omega + 4\omega_M (1+\omega) \right) \left\| x^k\right\|^2 \\ 
&& \qquad +4\E_{\mathcal{C}_M,\mathcal{C}}\left\|\mQ\mathcal{C}_M(\Bar{y}^*)-  \mQ\Bar{x}^*\right\|^2\\
&\overset{{\rm By~Lemma}~\ref{ch6:lem:xrelF}}{\le}& 4\dfrac{n^2}{\lambda^2}\left\|\nabla h(x^k) - \nabla h(x^*)\right\|^2 \\
&& \qquad +\left (4\omega + 4\omega_M (1+\omega) \right) \left( \dfrac{4}{\mu} \left( F(x^k) - F(x^*) \right) + 2 \left\| x^*\right\|^2 \right) \\
&&\qquad +4\E_{\mathcal{C}_M,\mathcal{C}}\left\|\mQ\mathcal{C}_M(\Bar{y}^*)-  \mQ\Bar{x}^*\right\|^2\\
&\le & 4\dfrac{n^2}{\lambda^2}\left\|\nabla h(x^k) - \nabla h(x^*)\right\|^2+ \alpha \left(F(x^k) - F(x^*)\right) + \beta.  
\end{eqnarray*}
Hence the result. 
\end{proof}

Now we are all set to prove the expected smoothness property in our setup. 
\begin{lemma*}[Expected Smoothness]
Let Assumptions \ref{ch6:ass:compression} and \ref{ch6:ass:smoothBound} hold, then
\begin{equation}
	\E\left[\|G(x^{k})\|^2|x^k\right] \le 4 \gamma \left(F(x^k) - F(x^*)\right) + \delta,
\end{equation}  
where $${\gamma \eqdef \dfrac{\alpha \lambda^2 (1-p)}{2 n^2 p} + \max \left\{ \dfrac{ L_f}{(1-p)}, \dfrac{\lambda}{n} \left(1+\dfrac{4(1-p)}{p}\right)
\right\}}$$ and $$\delta \eqdef \dfrac{2 \beta \lambda^2 (1-p)}{n^2 p}+2\E \| G(x^{*})\|^2.$$
\end{lemma*}
\begin{proof}
We have 
$$\|G(x^{k}) - G(x^{*})\|^2=
\left\{
\begin{array}{lll}
\dfrac{\|\nabla f\left(x^k\right) - \nabla f\left(x^*\right)\|^2}{(1-p)^2} & \text{ if } \xi_k=0
\\
\dfrac{  \lambda^2 }{n^2 p^2}\left\| x^k - \mQ\mathcal{C}_M(\Bar{y}^k) - x^* + \mQ\mathcal{C}_M(\Bar{y}^*) \right\|^2 
& \text{ if } \xi_k=1 ~\& ~\xi_{k-1}=0, \\
\dfrac{  1}{ p^2} \|\nabla h(x^k) - \nabla h(x^*) \|^2 & \text{ if } \xi_k=1 ~\& ~\xi_{k-1}=1.\\
\end{array}
\right.$$
Finally,  
\begin{eqnarray*}
\E_{\xi_k,\xi_{k-1}} \|G(x^{k}) - G(x^{*})\|^2 &=& (1-p) \dfrac{\|\nabla f\left(x^k\right) - \nabla f \left(x^*\right)\|^2}{(1-p)^2} \\
&& \quad + p^2 \dfrac{  1}{ p^2} \|\nabla h(x^k) - \nabla h(x^*) \|^2 \\
&&\quad +p(1-p) \dfrac{  \lambda^2 }{n^2 p^2}\left\| x^k - \mQ\mathcal{C}_M(\Bar{y}^k) - x^* + \mQ\mathcal{C}_M(\Bar{y}^*) \right\|^2 \\
&=&  \dfrac{\|\nabla f\left(x^k\right) - \nabla f \left(x^*\right)\|^2}{(1-p)} \\
&&\quad +\|\nabla h(x^k) - \nabla h(x^*) \|^2 \\
&&\quad +\dfrac{  \lambda^2 (1-p)}{n^2 p}\left\| x^k - \mQ\mathcal{C}_M(\Bar{y}^k) - x^* + \mQ\mathcal{C}_M(\Bar{y}^*) \right\|^2.
\end{eqnarray*}
Therefore, by using Lemma \ref{ch6:lem:mathcalA} we get
\begin{eqnarray*}
\lefteqn{\E \|G(x^{k}) - G(x^{*})|x^k\|^2} && \\
&= &  \dfrac{\|\nabla f\left(x^k\right) - \nabla f \left(x^*\right)\|^2}{(1-p)} \\
&&\quad +\|\nabla h(x^k) - \nabla h(x^*) \|^2 \\
&&\quad +\dfrac{  \lambda^2 (1-p)}{n^2 p} \mathcal{A}\\
&\le&  \dfrac{\|\nabla f\left(x^k\right) - \nabla f \left(x^*\right)\|^2}{(1-p)} \\
&&\quad +\|\nabla h(x^k) - \nabla h(x^*) \|^2\\
&&\quad +\dfrac{\lambda^2 (1-p)}{n^2 p} \left(4\dfrac{n^2}{\lambda^2}\left\|\nabla h(x^k) - \nabla h(x^*)\right\|^2+ \alpha \left(F(x^k) - F(x^*)\right) + \beta\right)\\
&=& \dfrac{\|\nabla f\left(x^k\right) - \nabla f \left(x^*\right)\|^2}{(1-p)} \\
&&\quad +\left(1+\dfrac{4(1-p)}{p}\right) \|\nabla h(x^k) - \nabla h(x^*) \|^2 \\
&&\quad +\dfrac{\alpha \lambda^2 (1-p)}{n^2 p}  \left(F(x^k) - F(x^*)\right) + \dfrac{\beta \lambda^2 (1-p)}{n^2 p} \\
&\le& \dfrac{2 L_f}{(1-p)}\left(f(x^k) - f(x^*)\right) + \dfrac{2\lambda}{n} \left(1+\dfrac{4(1-p)}{p}\right)\left(h(x^k) - h(x^*)\right)\\
&&\quad +\dfrac{\alpha \lambda^2 (1-p)}{n^2 p}  \left(F(x^k) - F(x^*)\right) + \dfrac{\beta \lambda^2 (1-p)}{n^2 p} \\
&\le& 2 \gamma \left(F(x^k) - F(x^*)\right) + \dfrac{\beta \lambda^2 (1-p)}{n^2 p}.
\end{eqnarray*}
Finally, we obtain
\begin{eqnarray*}
\E \|G(x^{k})|x^k\|^2 &\le& 2\E \|G(x^{k}) - G(x^{*})|x^k\|^2+ 2\E \| G(x^{*})\|^2\\
&\le &4 \gamma \left(F(x^k) - F(x^*)\right) +\dfrac{2 \beta \lambda^2 (1-p)}{n^2 p}+2 \E \| G(x^{*})\|^2\\
&\le& 4\gamma \left(F(x^k) - F(x^*)\right) + \delta.
\end{eqnarray*}
Hence the result. 
\end{proof}

Based on the above results, the convergence of Algorithm \ref{ch6:alg:ComL2GD} for strongly convex functions follows directly from Lemmas \ref{ch6:lem:unbiasGrad}, \ref{ch6:lem:ES} and Theorem 3.1 from \citet{Gower2019}.

\subsection{Nonconvex convergence}\label{ch6:sec:app_nnc}
\begin{theorem}[Non convex case]{Let Assumptions \ref{ch6:ass:compression} and \ref{ch6:assumption:bounded_g} hold. Assume also that $F$ is $L_f$-smooth, bounded from below by $F(x^*)$. Then to reach a precision, $\textstyle{\epsilon>0}$, set the stepsize, 
		$$\textstyle{\eta=\min\{\frac{1}{{L_f M}},\frac{\epsilon^2}{2L_f\sigma^2}\}},$$
		such that for 
		$$\textstyle{K\ge \frac{4L_f M(F(x^0)-F(x^*))}{\epsilon^2},}$$ we have 
		$$\min_{k=0,1,\dots,K}  \E \|\nabla F(x^k)\|_2\le \epsilon.$$}
\end{theorem}
\begin{proof}
	{
		From $L_f$-smoothness of $F$ we have
		\begin{eqnarray*}
			F(x^{k+1}) &\leq&  F(x^k) - \eta_k{\nabla F(x^k)}^\top{G(x^k)} + \frac{L_f}{2}\eta_k^2\|G(x_k)\|^2.
		\end{eqnarray*} 
		By taking the expectation in the above inequality, conditional on $x^k$, we get 
		\begin{eqnarray*}
			\E{ \left[ F(x^{k+1}) \;|\; x^k \right]} &\overset{{\rm By\;Lemma \; \ref{ch6:lem:unbiasGrad}}}{\leq}& F(x^k)- \eta_k  \|\nabla F(x^k)\|_2^{2} + \frac{L_f\eta_k^2}{2}\E\left(\|G(x_k)\|^2|x_k\right),
		\end{eqnarray*}
		which by using Assumption \ref{ch6:assumption:bounded_g} reduces to
		\begin{eqnarray*}
			\E{ \left[ F(x^{k+1}) \;|\; x^k \right]}&\leq & F(x^k)- \eta_k  \|\nabla F(x^k)\|_2^{2} + \frac{L_f\eta_k^2}{2}\left(M\|\nabla F(x^k)\|^2+\sigma^2\right)\\
			&\leq & F(x^k)- \eta_k \left(1-\frac{L_f M\eta_k}{2}\right) \|\nabla F(x^k)\|_2^{2} +  \frac{L_f\eta_k^2\sigma^2}{2}. 
		\end{eqnarray*} 
		After rearranging, we have
		\begin{eqnarray*}
			\eta_k \left(1-\frac{L_f M\eta_k}{2}\right) \|\nabla F(x^k)\|_2^{2} &\leq & F(x^k)- \E{ \left[ F(x^{k+1}) \;|\; x^k \right]}  +  \frac{L_f\eta_k^2\sigma^2}{2}. 
		\end{eqnarray*} 
		Setting $\eta_k=\eta> 0$ in the above, taking expectation, using the tower property of expectation, and finally summing over the iterates $k=0,1,\cdots K-1$ we have
		\begin{eqnarray*}
			\eta\left(1-\frac{L_f M\eta}{2}\right) \sum_{k=0}^{K-1}\E\left[\|\nabla F(x^k)\|_2^{2}\right] &\leq & \left(F(x^0)-F(x^{*})\right)  +  \frac{KL_f\eta\sigma^2}{2}. 
		\end{eqnarray*} 
		If $\eta \le \frac{1}{L_fM}$ then
		\begin{eqnarray*}
			\sum_{k=0}^{K-1}\E\left[\|\nabla F(x^k)\|_2^{2}\right] &\leq & \frac{2}{\eta}\left(F(x^0)-F(x^{*})\right) + {L_fK\eta\sigma^2}. 
		\end{eqnarray*} 
		Dividing throughout by $K$, we get
		\begin{eqnarray*}
			\frac{1}{K}\sum_{k=0}^{K-1}\E\left[\|\nabla F(x^k)\|_2^{2}\right] &\leq & \frac{2}{\eta K}\left(F(x^0)-F(x^{*})\right) +{L_f\eta\sigma^2}. 
		\end{eqnarray*}
		Finally, setting $\eta=\frac{1}{L_fM}$ we have
		\begin{eqnarray}\label{eq:descent-con-3}
			\min_{k=0,1,\cdots K-1}\E\|\nabla F(x^{k})\|^2&\le& \frac{2L_fM}{K}\left(F(x^0)-F(x^{*})\right) + \frac{\sigma^2}{M}.
		\end{eqnarray}
		For a given precision, $\epsilon>0$, to make $\min_{k=0,1,\cdots K-1}\E\|\nabla F(x^{k})\|^2\le \epsilon^2$, we require $\frac{2L_fM\left(F(x^0)-F(x^*)\right)}{K}\le \frac{\epsilon^2}{2}$ and $L_f\eta\sigma^2\le \frac{\epsilon^2}{2}$, resulting in
		$$
		K\ge \frac{4L_fM(F(x^0)-F(x^*))}{\epsilon^2}\;{\rm and\;} \eta\le \frac{\epsilon^2}{2L_f\sigma^2}.
		$$
		Hence the result.}
\end{proof}

\subsection{Optimal rate and communication}\label{ch6:sec:app optimal}

The following proofs are related to optimal rate and communication as given in Section~\ref{ch6:sec:optimalrate}. 

\begin{theorem*}[Optimal rate] 
The probability $p^*$ minimizing $\gamma$ is equal to 
$\max\{p_e,p_A\}$, where $$p_e = \dfrac{7 \lambda + L - \sqrt{\lambda^2 + 14 \lambda L + L^2}}{6 \lambda}$$ and $p_A$ is the optimizer of the function $A(p) = \dfrac{\alpha \lambda^2}{2 n^2 p } + \dfrac{ L}{n(1-p)}$ in $(0,1)$.
\end{theorem*}
\begin{proof}
We can rewrite $\gamma$ as follows
\begin{eqnarray*}
\gamma &=&  -\dfrac{\alpha \lambda^2}{2 n^2} + \max \left\{ A(p), B(p)
\right\},
\end{eqnarray*}
where $$A(p) = \dfrac{\alpha \lambda^2}{2 n^2 p } + \dfrac{ L}{n(1-p)}$$ and $$B(p) = \dfrac{\alpha \lambda^2}{2 n^2 p }+  \dfrac{4\lambda}{np} -\dfrac{3 \lambda}{n}.$$

The function $B$ is monotonically decreasing as a function of $p$. The function $A$ goes to $\infty$ as $p$ goes to zero or one, and it has one stationary point between zero and one hence it is convex in the interval $(0,1)$. Thus it admits an optimizer $p_A$ in $(0,1)$. Note that $$p_e = \dfrac{7 \lambda + L - \sqrt{\lambda^2 + 14 \lambda L + L^2}}{6 \lambda}$$ is the point for which $A(p)$ is equal to $B(p)$. Note also that near to zero $$B(p) \ge A(p).$$
Therefore if $p_e \le p_A$ then the optimizer of $\gamma$ is $p_A$ otherwise it is equal to $p_e$.  
Thus the probability $p^*$ optimizing $\gamma$ is equal to $\max\{p_e,p_A\}$.
\end{proof}

\begin{lemma*}The optimizer probability $p_A$   of the function $$A(p) = \dfrac{\alpha \lambda^2}{2 n^2 p } + \dfrac{ L}{n(1-p)}$$ in $(0,1)$ is equal to 
$$p_A=
\left\{
\begin{array}{lll}
\dfrac{1}{2}  & \text{ if } 2nL =\alpha \lambda^2
\\
\dfrac{-2 \alpha \lambda^2 + 2\lambda \sqrt{2\alpha n L}}{2(2nL -\alpha \lambda^2)}   & \text{ if } 2nL > \alpha \lambda^2
\\
\dfrac{-2 \alpha \lambda^2 - 2\lambda \sqrt{2\alpha n L}}{2(2nL -\alpha \lambda^2)}   & \text{ otherwise } 
\end{array}
\right.$$
\end{lemma*}

\begin{proof}
If $2nL \neq \alpha \lambda^2$, then the function $A$ has the following two stationary points
$$\dfrac{-2 \alpha \lambda^2 + 2\lambda \sqrt{2\alpha n L}}{2(2nL -\alpha \lambda^2)}   $$ and 
$$\dfrac{-2 \alpha \lambda^2 - 2\lambda \sqrt{2\alpha n L}}{2(2nL -\alpha \lambda^2)}.$$ 
If  $2nL =\alpha \lambda^2$, then  the function $A$ has one stationary point equal to $\dfrac{1}{2}$.
\end{proof}

\begin{theorem*}[Optimal communication] 
The probability $p^*$ optimizing $C$ is equal to 
$\max\{p_e,p_A\}$, where $p_e = \dfrac{7 \lambda + L - \sqrt{\lambda^2 + 14 \lambda L + L^2}}{6 \lambda}$ and $p_A = 1 - \dfrac{Ln}{\alpha \lambda^2}$.
\end{theorem*}

\begin{proof}
We can rewrite $nC$ as follows
\begin{eqnarray*}
nC &=&   \max \left\{ A(p), B(p)
\right\},
\end{eqnarray*}
where $$A(p) = \dfrac{\alpha \lambda^2 p(1-p)}{2n} + \dfrac{\alpha \lambda^2(1-p)}{2 n} + Lp$$ and $$B(p) =\dfrac{\alpha \lambda^2 p(1-p)}{2n}  + \dfrac{\alpha \lambda^2(1-p)}{2 n}+  4\lambda(1-p) - 3 \lambda p (1-p).$$

The function $B$ is monotonically decreasing as a function of $p$ in $[0,1]$. Note that 
$$B(0)  = \dfrac{\alpha \lambda^2}{2 n}+  4\lambda\quad\mathrm{and}\quad B(1)=0.$$
The function $A$  admits a minimizer equal to $$p_A = 1 - \dfrac{Ln}{\alpha \lambda^2}.$$ Of course $p_A$ is a  probability   under the condition that $Ln \le \alpha \lambda^2$. Thus we consider the following  2 scenarios
\begin{enumerate}
\item If $Ln > \alpha \lambda^2$ ($p_A<0$) then $p^* = p_e$.
\item Otherwise, $p^* = \max\{p_e,p_A\}$.
\end{enumerate}
We conclude in both cases that $p^* = \max\{p_e,p_A\}.$
\end{proof}

\addtocounter{adjsection}{1}
\section{Addendum to the Experimental Results}
\label{ch6:sec:app_nn}

\smartparagraph{Batch Normalization.} Beside the trainable parameters, the \modelname{ResNet} models contain batch normalization~\citep{ioffe2015batch} layers that are crucial for training. The logic of batch normalization depends on the estimation of running mean and variance, and these statistics can be pretty personalized for each client in a heterogeneous data regime. The implementation of \algname{FedAVG} and \algname{FedOpt} in \libname{FedML} considers the averaging of these statistics during the aggregation phase. In our implementation, the batch normalization statistics are included in aggregation. 

\smartparagraph{Step size.} The step sizes for \algname{FedAVG} and \algname{FedOpt} tuned via selecting step sizes from the following set $$\{0.01, 0.1, 0.2, 0.5, 1.0, 2.0, 4.0\}.$$ 
We consider the step size for both algorithms to be $0.1$. Starting with step size $0.2$ algorithms diverge; we also did not use step size schedulers. Additionally, we have tuned a number of local epochs for \algname{FedAVG} from the following set $\{1,2,3,4\}$. The used batch size is set to $256$.

\begin{figure*}[t]
\centering
\captionsetup[sub]{font=small,labelfont={}}	


\begin{subfigure}[ht]{0.82\textwidth}
	\includegraphics[width=\textwidth]{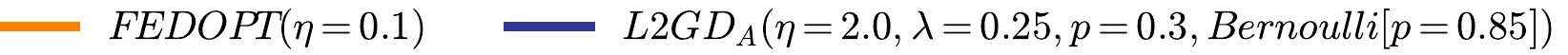}
\end{subfigure}

\begin{subfigure}[ht]{\textwidth}
	\includegraphics[width=\textwidth]{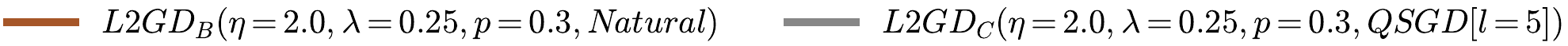}
\end{subfigure}

\begin{subfigure}[ht]{0.48\textwidth}
\includegraphics[width=\textwidth]{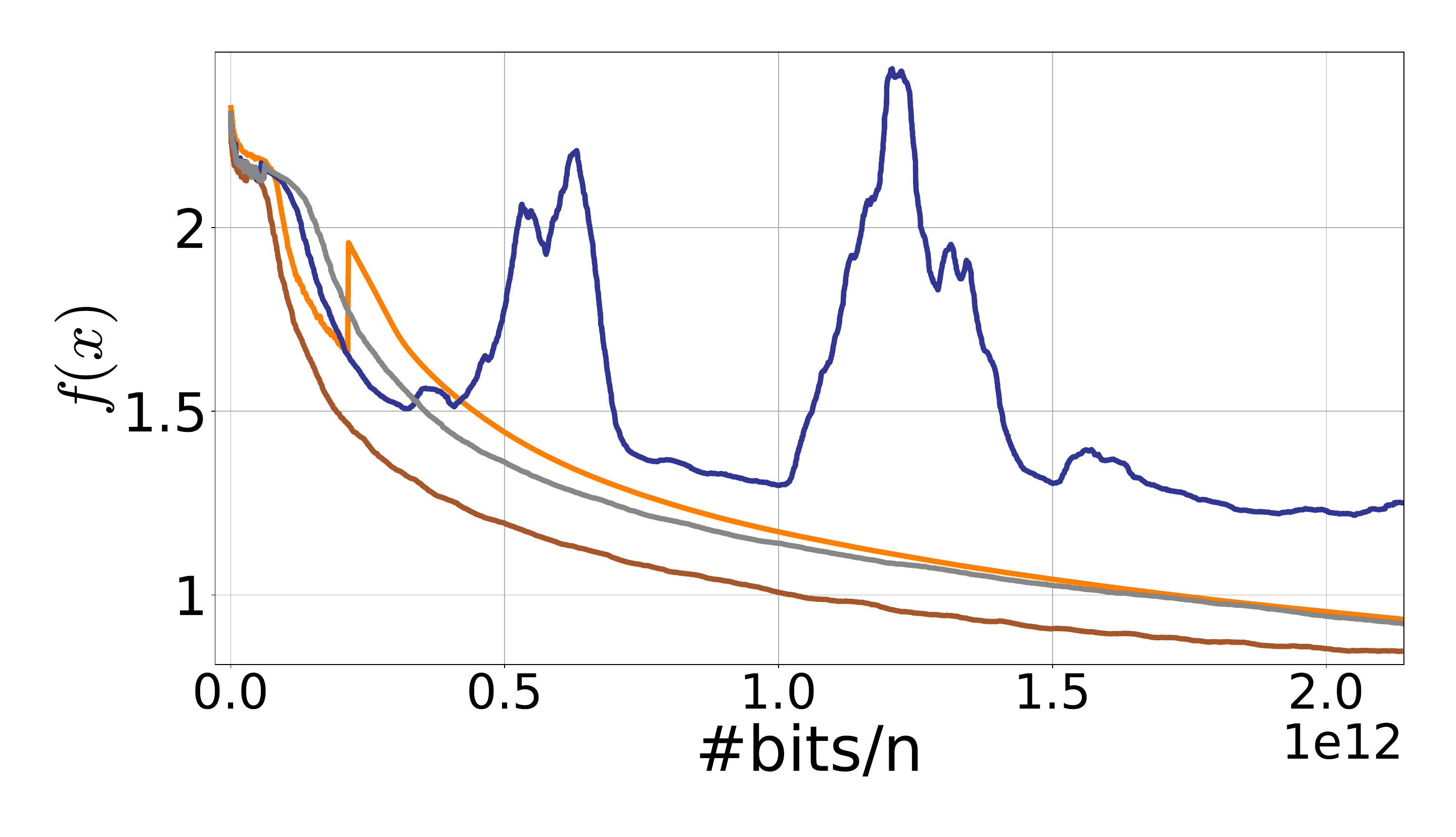} 
\caption{\textbf{Train:} Loss functional value}
\end{subfigure}
\begin{subfigure}[ht]{0.48\textwidth}
\includegraphics[width=\textwidth]{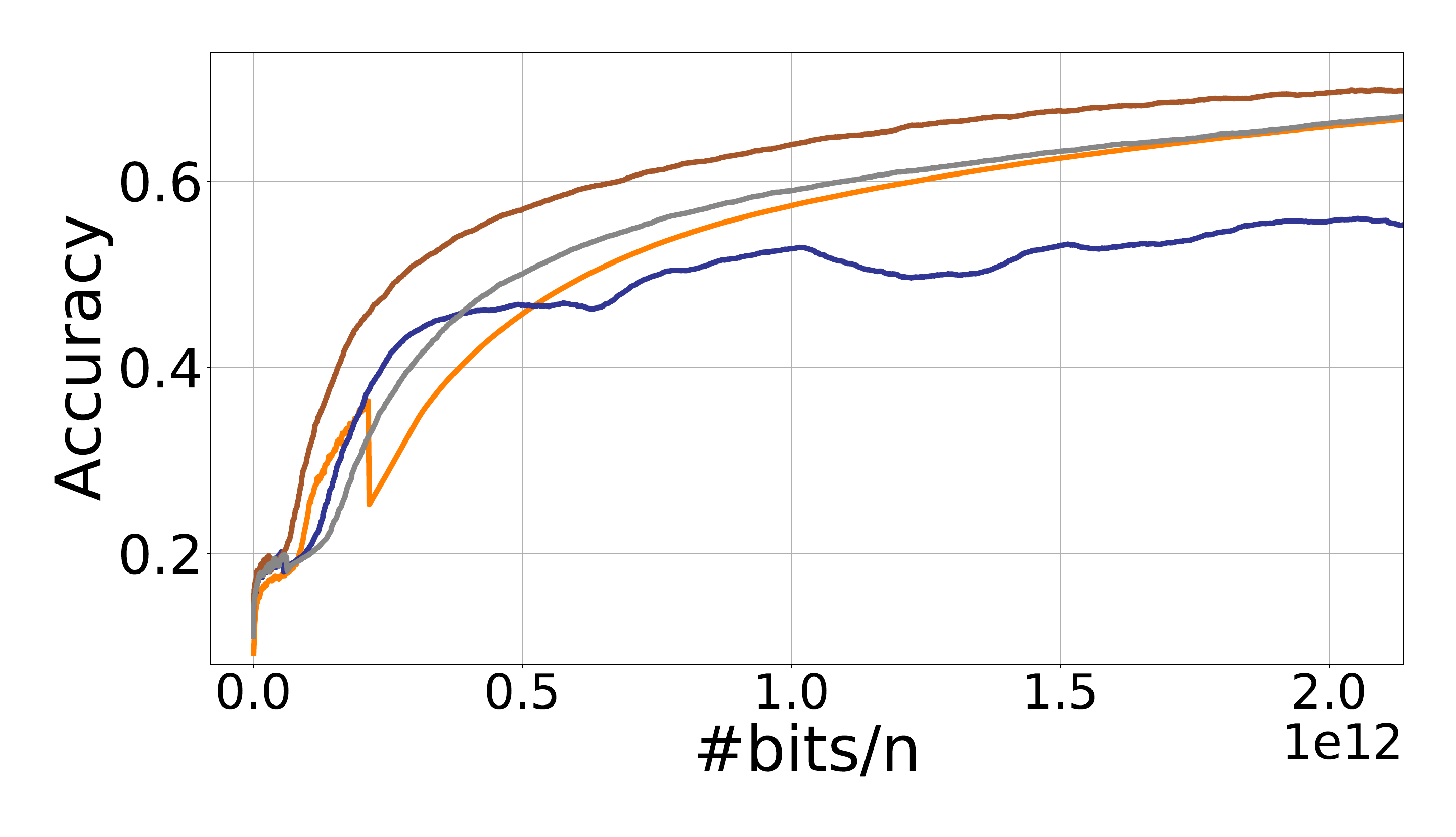} 
\caption{\textbf{Train:} Top-1 accuracy}
\end{subfigure}

\begin{subfigure}[ht]{0.48\textwidth}
\includegraphics[width=\textwidth]{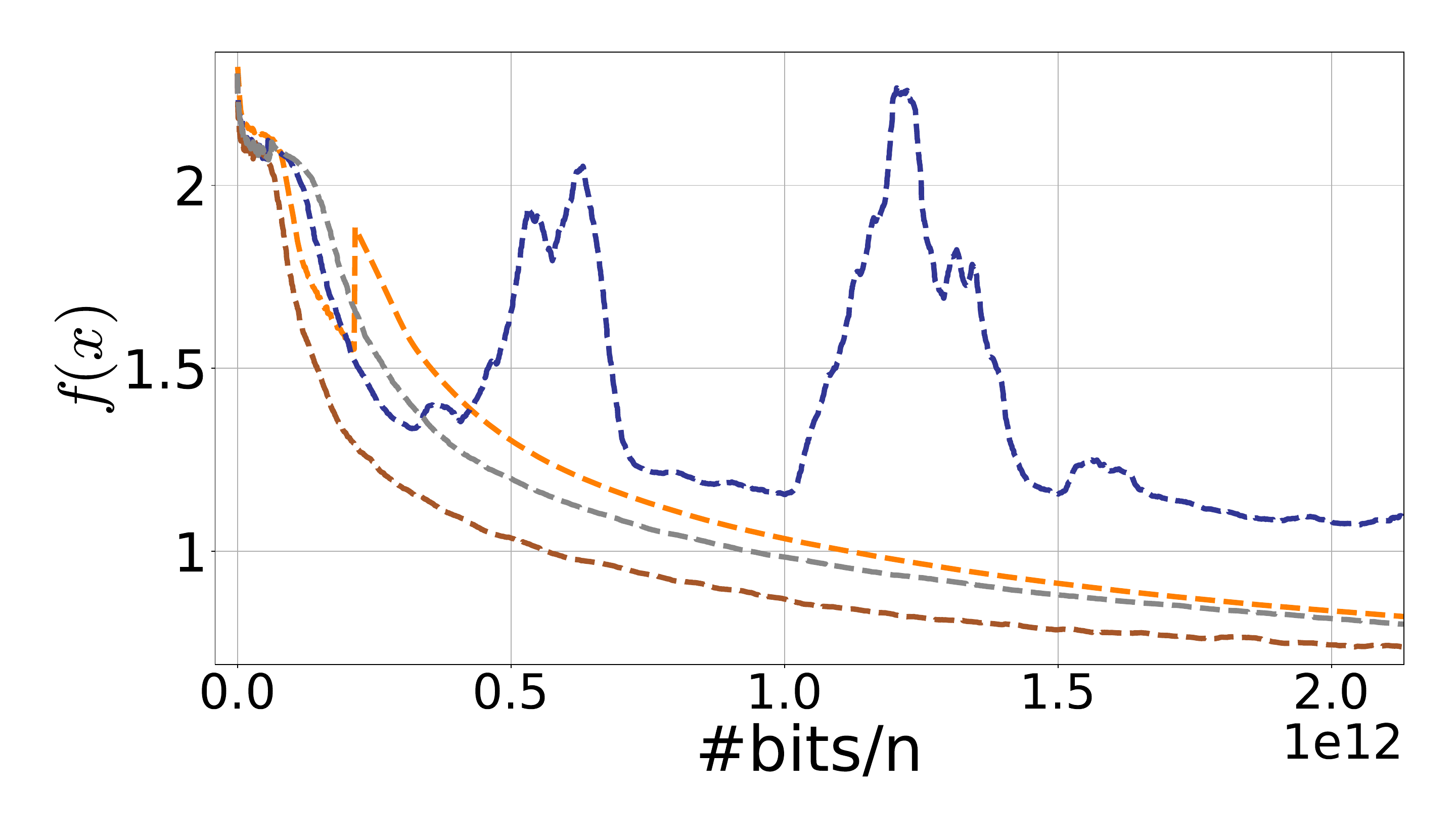}
\caption{\textbf{Test:} Loss functional value}
\end{subfigure}
\begin{subfigure}[ht]{0.48\textwidth}
\includegraphics[width=\textwidth]{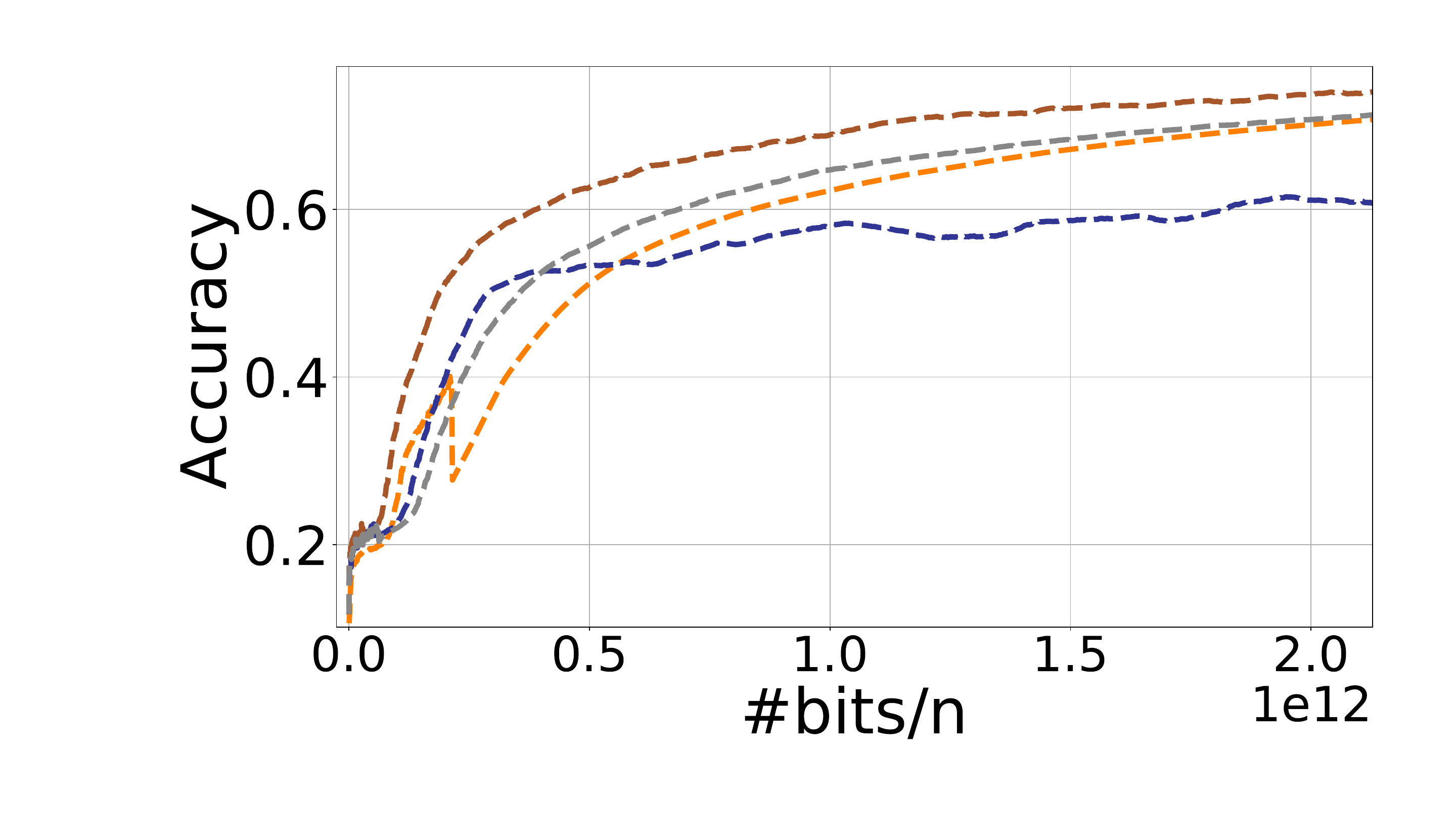} 
\caption{\textbf{Test:} Top-1 accuracy}
\end{subfigure}
\caption{{Training \modelname{ResNet-18} on \dataname{CIFAR-10}, with $n=10$ workers. Loss and Top-1 accuracy on the train (a)--(b) and test data (c)--(d).}}
\label{ch6:fig:training_resnet_l2g_only}
\end{figure*}

\begin{figure*}[t]
\centering
\captionsetup[sub]{font=small,labelfont={}}	


\begin{subfigure}[ht]{0.8\textwidth}
	\includegraphics[width=\textwidth]{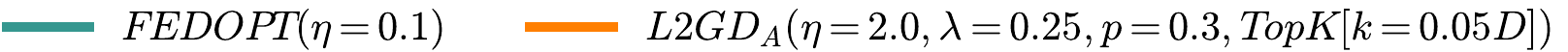}
\end{subfigure}

\begin{subfigure}[ht]{\textwidth}
	\includegraphics[width=\textwidth]{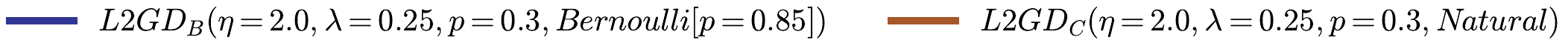}
\end{subfigure}

\begin{subfigure}[ht]{0.46\textwidth}
	\includegraphics[width=\textwidth]{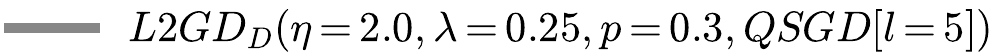}
\end{subfigure}

\begin{subfigure}[ht]{0.495\textwidth}
	\includegraphics[width=\textwidth]{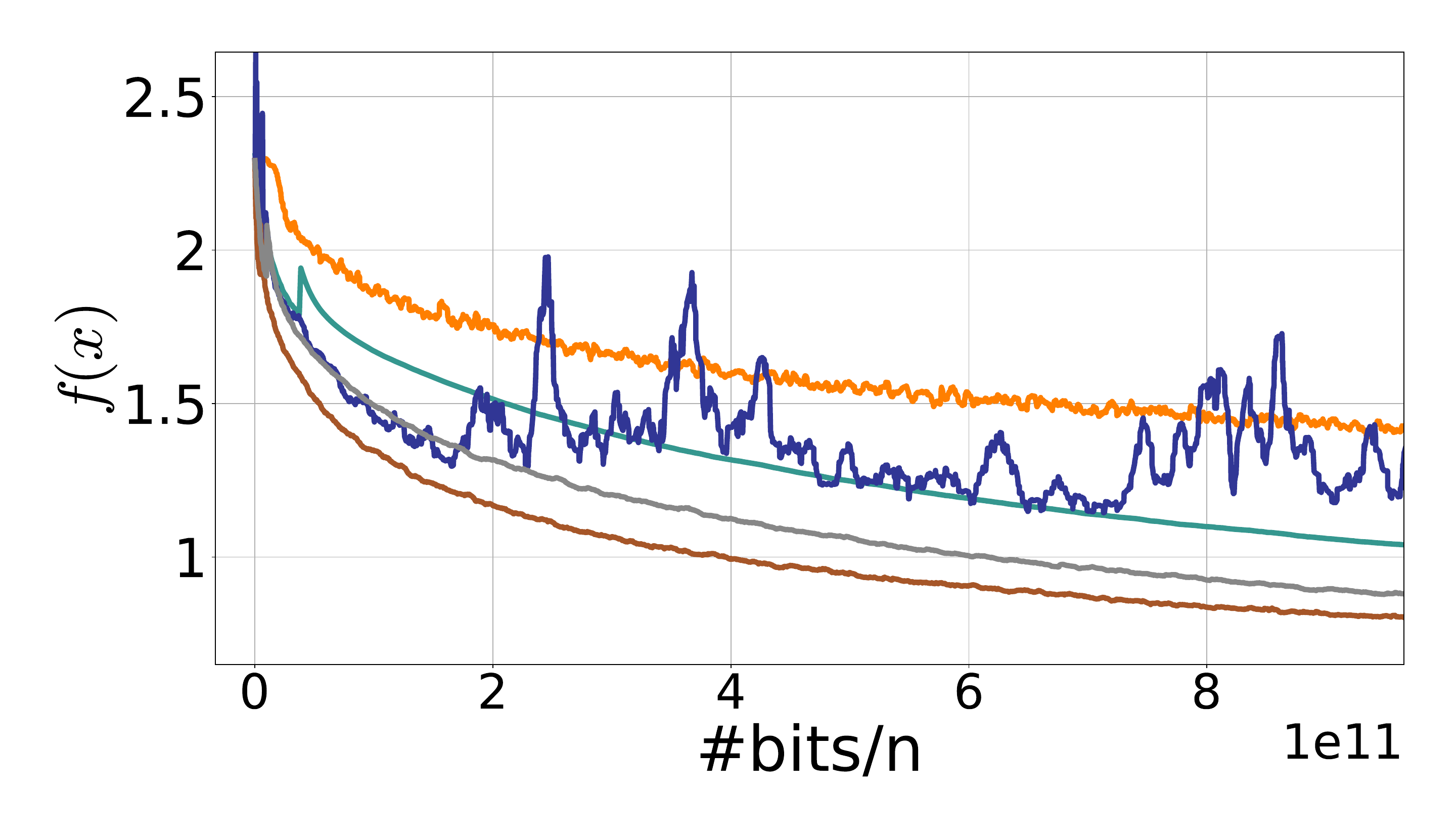}
	\caption{\textbf{Train:} Loss functional value}
\end{subfigure}
\begin{subfigure}[ht]{0.495\textwidth}
	\includegraphics[width=\textwidth]{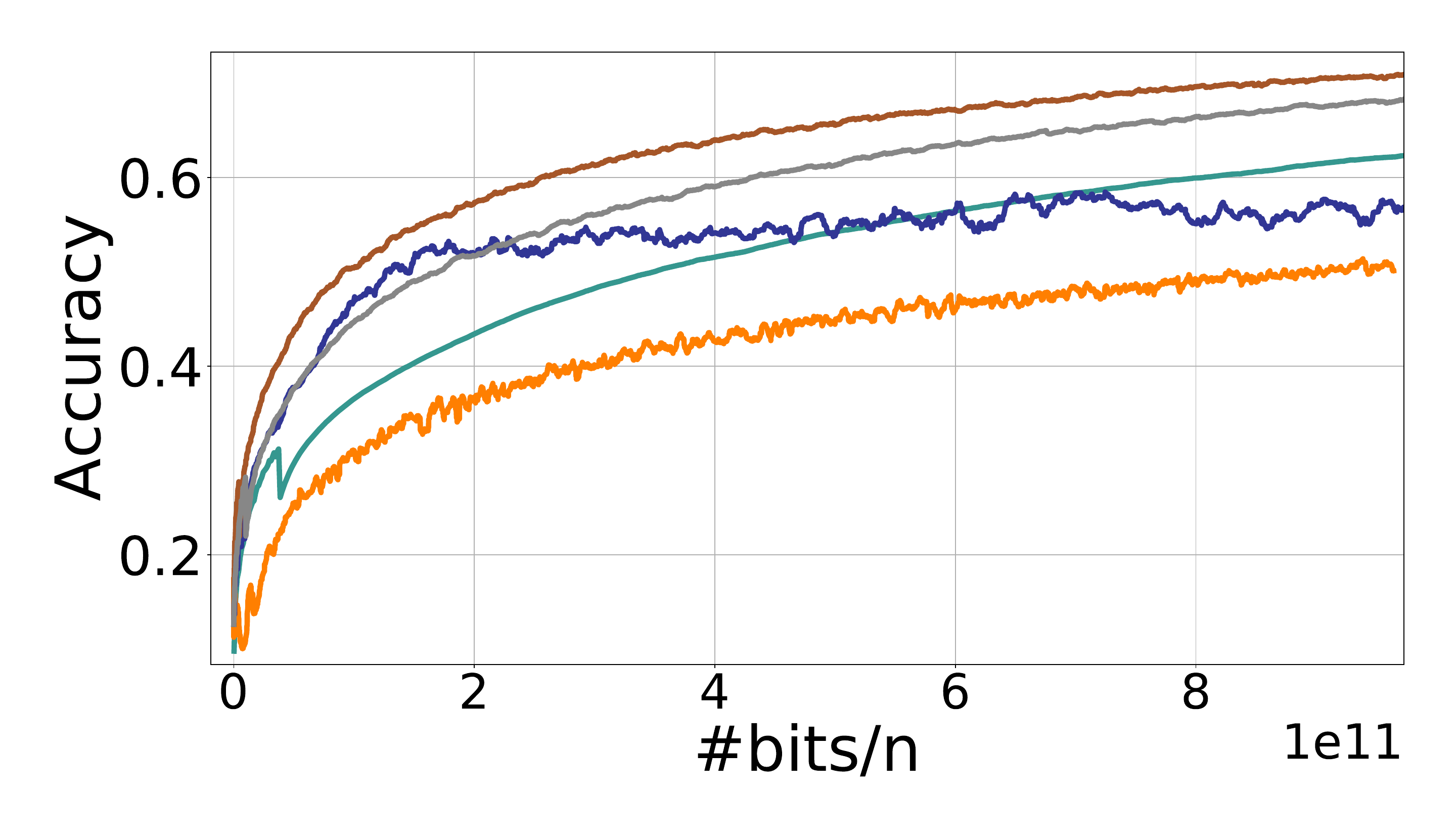} 
	\caption{\textbf{Train:} Top-1 accuracy}
\end{subfigure}

\begin{subfigure}[ht]{0.495\textwidth}
\includegraphics[width=\textwidth]{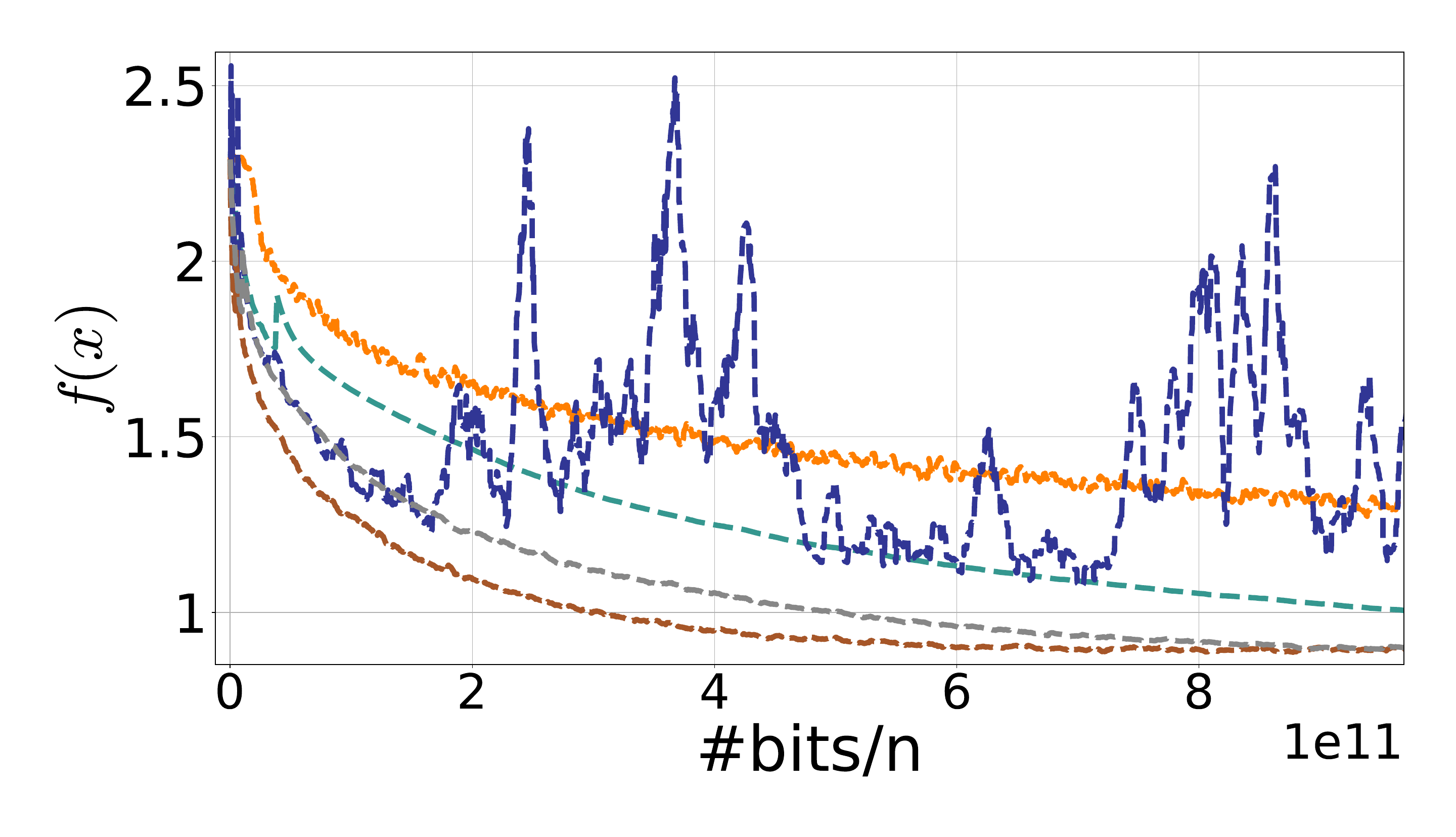} 
\caption{\textbf{Test:} Loss functional value}
\end{subfigure}
\begin{subfigure}[ht]{0.495\textwidth}
\includegraphics[width=\textwidth]{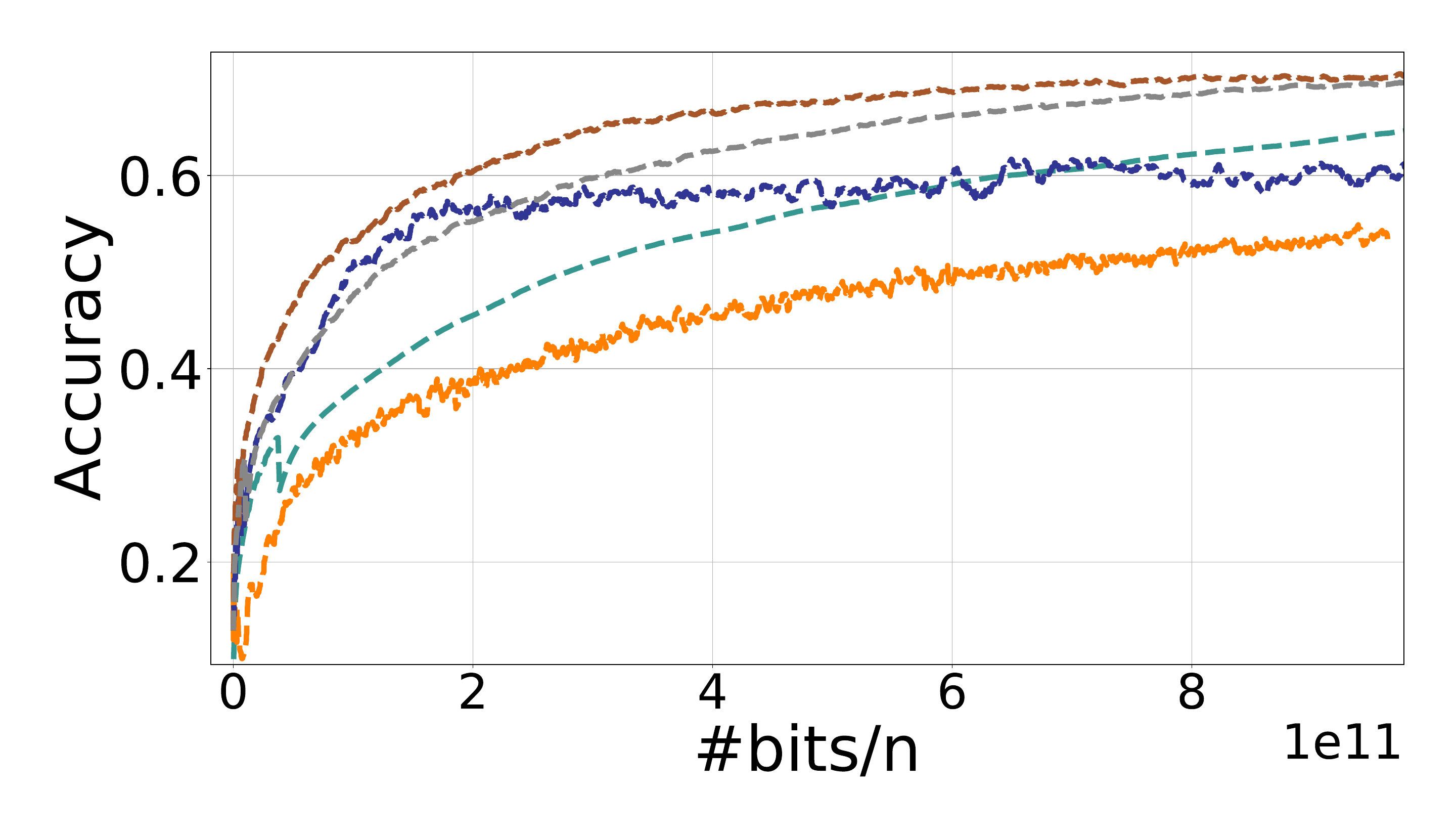}
\caption{\textbf{Test:} Top-1 accuracy}
\end{subfigure}
\caption{{Training \modelname{DenseNet-121} on \dataname{CIFAR-10}, with $n=10$ workers. Loss and Top-1 accuracy on the train (a)--(b), and test data (c)--(d).}}
\label{ch6:fig:training_densenet_l2g_only}
\end{figure*}

\begin{figure*}[t]
\centering
\captionsetup[sub]{font=small,labelfont={}}	


\begin{subfigure}[ht]{0.98\textwidth}
	\includegraphics[width=\textwidth]{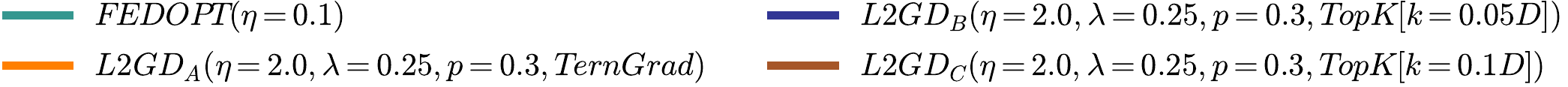}
\end{subfigure}	

\begin{subfigure}[ht]{\textwidth}
	\vspace{0.2cm}
	\hspace{0.05cm}
	\includegraphics[width=\textwidth]{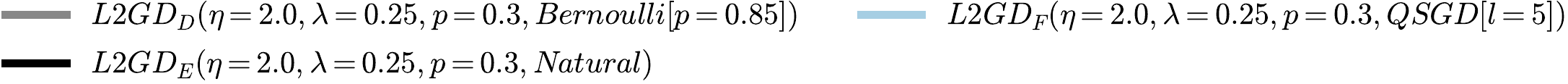}
\end{subfigure}	

\begin{subfigure}[ht]{0.5\textwidth}
\includegraphics[width=\textwidth]{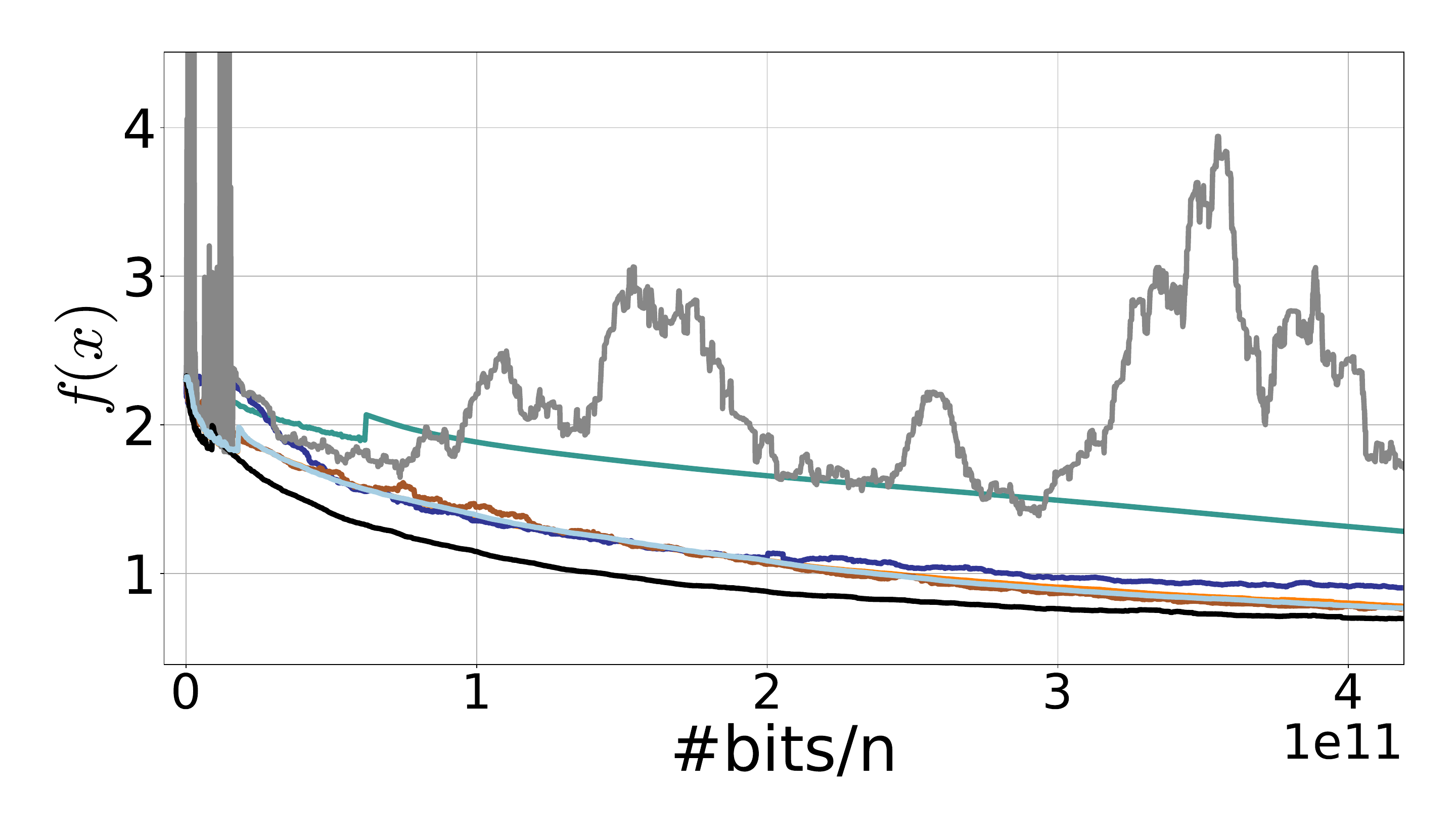} 
\caption{\textbf{Train:} Loss functional value}
\end{subfigure}
\begin{subfigure}[ht]{0.48\textwidth}
\includegraphics[width=\textwidth]{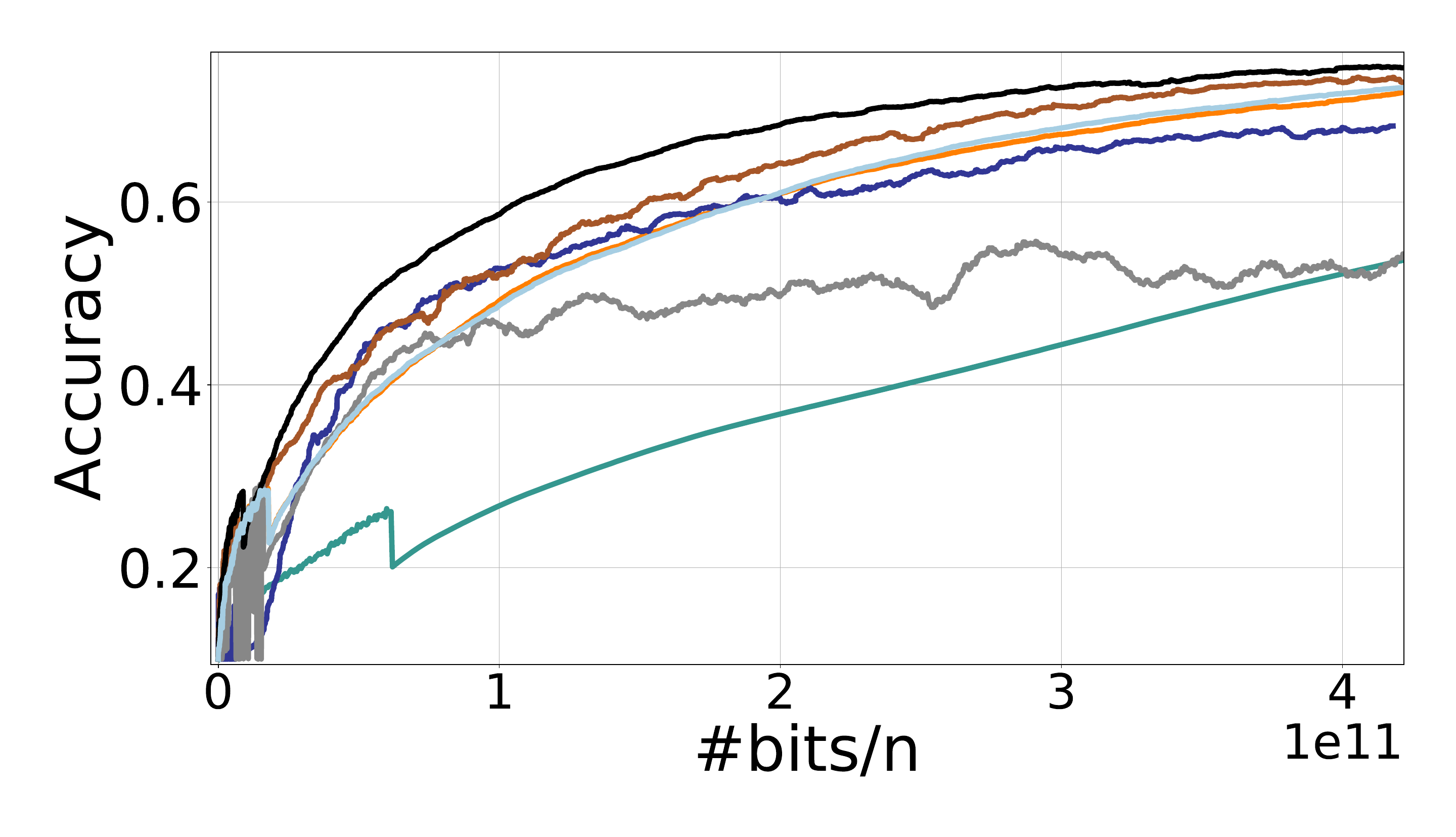} 
\caption{\textbf{Train:} Top-1 accuracy}
\end{subfigure}

\begin{subfigure}[ht]{0.48\textwidth}
\includegraphics[width=\textwidth]{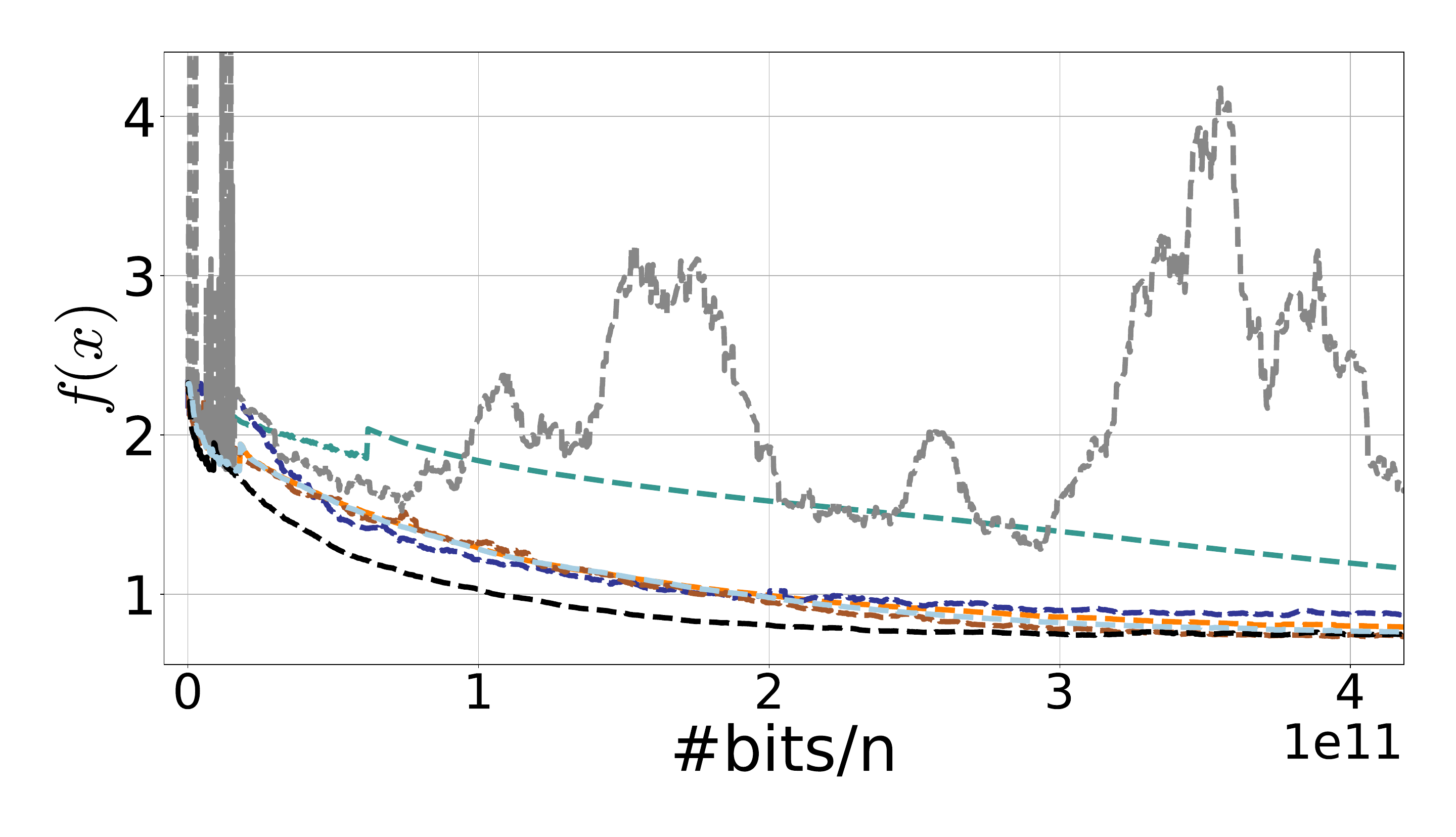} 
\caption{\textbf{Test:} Loss functional value}
\end{subfigure}
\begin{subfigure}[ht]{0.48\textwidth}
\includegraphics[width=\textwidth]{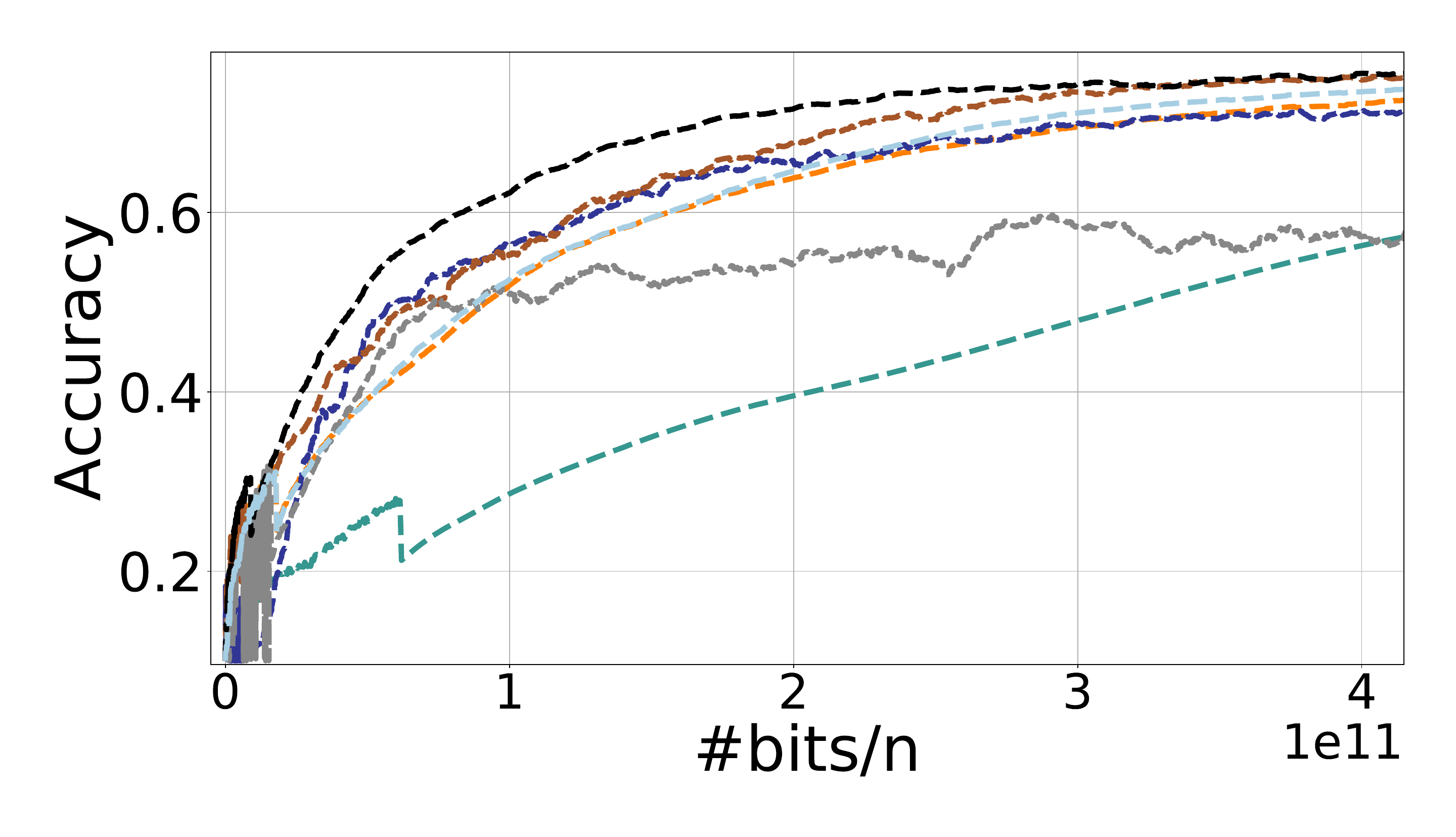}
\caption{\textbf{Test:} Top-1 accuracy}
\end{subfigure}
\caption{{Training \modelname{MobileNet} on \dataname{CIFAR-10}, with $n=10$ workers. Loss and Top-1 accuracy on the train (a)--(b), and test data (c)--(d).}}
\label{ch6:fig:training_mobilenet_l2g_only}
\end{figure*}

\smartparagraph{Compressed L2GD versus FedOpt.} From the experiments in Section \ref{ch6:sec:dnn}, Figures~\ref{ch6:fig:training_resnet},~\ref{ch6:fig:training_densenet},~\ref{ch6:fig:training_mobilenet} we realized that \algname{FedAVG} is not a competitive no-compression baseline for \algname{L2GD}; see Table~\ref{ch6:tab:modelsize}. \algname{FedOpt}, on the other hand, remains a competitive no-compression baseline comparable to \algname{Compressed L2GD}. Therefore, we separately measure the performance of \algname{Compressed L2GD} and non-compression \algname{FedOpt} for training \modelname{ResNet-18}, \modelname{DenseNet-121}, and  \modelname{MobileNet}. Figures~\ref{ch6:fig:training_resnet_l2g_only},~\ref{ch6:fig:training_densenet_l2g_only},~\ref{ch6:fig:training_mobilenet_l2g_only} demonstrate that \algname{L2GD} with \compname{Natural} compressor (that by design has small variance) empirically behaves the best and converges approximately $5$ times faster compare to \algname{FedOpt}. They also show that \algname{Compressed L2GD} with \compname{Natural} compressor sends the least data and drives the loss down the most. At the same time, \algname{L2GD} with \compname{Natural} compressor reaches the best accuracy for both train and test sets.

\addtocounter{adjsection}{1}
\section{Reproducibility}
The source code for the experiments, along with a description for reproducing the experiment, can be downloaded from:
\begin{center}
\href{https://github.com/burlachenkok/compressed-fl-l2gd-code}{https://github.com/burlachenkok/compressed-fl-l2gd-code}
\end{center}

Our source codes have been constructed on top of the following version of FedML.ai: 
\begin{center}
\href{https://github.com/FedML-AI/FedML/commit/3b9b68764d922ce239e0b84aceda986cfa977f96}{https://github.com/FedML-AI/FedML/commit/3b9b68764d922ce239e0b84aceda986cfa977f96}
\end{center}

\unappendix

\chapter{Unlocking FedNL: Practical Implementation of FedNL}
\label{chapter7}


The goals and summaries of this chapter are outlined in Table \ref{ch1:tbl:algorithms} and Section~\ref{ch1:sec:overview-7}.

\section{Introduction}
\label{ch7:intro}

Convex optimization finds applications in science and engineering and additionally serves as a tool for tackling non-tractable global optimization and combinatorial optimization problems. Examples can be found in \citet{bertsekas2003convex}, \citet{boyd2004convex}. The second-order methods represent a category of continuous optimization techniques that go beyond utilizing gradient and function value information by incorporating details about the Hessians of the minimization objective. These methods offer advantages, such as invariance to affine changes in optimization variable coordinates and rapid local convergence. For instance, the pure Newton method exhibits a quadratic convergence rate near the solution. Existing ready-to-use systems within Mathematical Optimization and Machine Learning lack good support for large-scale distributed second-order optimization. \citet{wytock2016new} attribute this issue to the following factors: (a) Distributing optimization using Newton and Quasi-Newton methods demands high bandwidth to transmit Hessian-like quantities during communication; (b) The memory requirement for forming and storing second-order information is substantial; (c) Practical low-level linear algebra libraries, have limited interfaces. However, these concerns are highly correlated with questions under study in Federated Learning.

Federated Learning ({FL}), introduced by \citet{FEDLEARN}, is a specialized multidisciplinary subfield within {ML}. It facilitates intelligent clients to collaboratively train {ML} models in a distributed way, eliminating the need for centralized data collection. The 	{FL} optimization algorithms have internal mechanisms to balance memory transfers and computation. Importantly in {FL}, they are an integral part of the optimization algorithms. One of them is \textit{communication compression}. Examples of first-order optimization algorithms that was co-designed with {communication compression} includes  \algname{MARINA} \citep{gorbunov2021marina}, \algname{COFIG/FRECON} \citep{zhao2021faster}, \algname{EF-21} \citep{richtarik2021ef21, richtarik2023error}. These algorithms and also recently started to appear algorithms for second-order optimization methods with mechanisms for \textit{communication compression} are still not spread enough, despite the robust theory. The first reason is that despite the rapid growth, {FL} is still a highly specialized subfield within {ML}. The second reason is that effectively applying methods in a multidisciplinary field demands proficiency in several disciplines. Ready-to-use implementations play a crucial role in simplifying this challenge.

The recent work by \citet{safaryan2021fednl} introduces the family of Federated Newton Learn (\algname{FedNL}) algorithms, marking a significant step in incorporating second-order methods into {FL} and large-scale optimization. The optimization problem addressed by \algname{FedNL} \citep{safaryan2021fednl} has a finite sum structure:
\begin{equation}
	\label{ch7:eq:main}
	\min \limits_{x\in \R^d} \left\{f(x)\eqdef \dfrac{1}{n}\sum \limits_{i=1}^n f_i(x) \right\}.
\end{equation}

In Equation~\eqref{ch7:eq:main}, functions $f(x), f_1(x),\dots, f_n(x) \in \mathbb{R}^d \to \mathbb{R}$ satisfy Assumptions~\ref{ch7:asm:1} and \ref{ch7:asm:2} from the following section.

\subsection{Assumptions}

\begin{assumption}\label{ch7:asm:1}
	The $f(x)$ is $\mu_f$ strongly convex. If $f(x)$ is twice continuously differentiable, it is equivalent to the condition that $\nabla^2 f(x)  - \mu I$ be positive semi-definite $\forall x \in \RD$.
\end{assumption}

\begin{assumption}\label{ch7:asm:2}
	There exist Lipschitz constants $\HS$, $\HF$, and $\HM$ such that $\forall i\in[n]$, $\forall x,y\in\R^d$:
	\begin{eqnarray*}
		\|\nabla^2 f_i(x) - \nabla^2 f_i(y)\|_2 &\leq & \HS \|x-y\|, \\
		\|\nabla^2 f_i(x) - \nabla^2 f_i(y)\|_{\rm F} & \leq & \HF \|x-y\|, \\
		\max_{j,l}| (\nabla^2 f_i(x) - \nabla^2 f_i(y))_{jl}| & \leq & \HM \|x-y\|.
	\end{eqnarray*}	
\end{assumption}

\clr{
	To the best of our knowledge, \algname{FedNL} is the state-of-the-art second-order optimization method in terms of \textit{theoretical} convergence guarantees in a class of algorithms striving to solve Problem~\eqref{ch7:eq:main} under Assumptions \ref{ch7:asm:1}, \ref{ch7:asm:2}. We will not reiterate the comparisons of \algname{FedNL} with prior methods, as this has been comprehensively covered in Appendix A \citet{safaryan2021fednl}.
}

\algname{FedNL} supports communication compression for transferring Hessian information from clients. Its extensions include \algname{FedNL-PP}, designed for partial participation among clients, and \algname{FedNL-LS}, which ensures global convergence. The algorithm exhibits local superlinear convergence rates in terms of the squared Euclidean distance to the solution, independent of the Hessian's condition number. Practically, this is indistinguishable from the local quadratic rate achieved by the Newton method. Notably, executing \algname{FedNL} does not require knowledge of any problem-specific constants. \clr{These factors make the \algname{FedNL} algorithm family a promising choice for a range of practical applications in which machine learning problems fulfill Equation~\eqref{ch7:eq:main} under Assumptions \ref{ch7:asm:1}, \ref{ch7:asm:2}.}



\subsection{Practical considerations to avoid catastrophic outcomes}
\label{ch7:sec:extra-motivation}
In 1943, Kurt Lewin, a pioneer in social science, stated that \textit{"There is nothing so practical as a good theory" \footnote{The phrase is commonly linked to Kurt Lewin rather than his specific publication.}}, which sets one connection to how theory influences practice. It is well recognized that practical applications validate theories, and prompt theorists to refine their models. However, there are \textit{two lesser-known facts} that highlight the need to balance theory and practice, discovered in communities adjacent to but not strictly within Machine Learning.

The first insight comes from the realm of Data-intensive Computing and Operating Systems. The constructive criticism by \citet{mcsherry2015scalability}, argues that an excessive focus on scalability in big-data systems hides an essential evaluation of absolute performance. The concern of the authors of this work was about whether (a) state-of-the-art distributed processing systems genuinely enhance overall performance or (b) the overheads introduced by them lead in fact to performance degradation. The authors demonstrated that in experiments with popular graph algorithms, unfortunately, the latter is true. This phenomenon has to be considered in the context of any multidisciplinary field.

The second is from the Computer Architecture. Before $2004$, there was no motivation to focus on focused performance optimization of algorithms. Up to this year, the transistor density experienced exponential growth \citep{10.1145/2133806.2133822}, and \textit{``just wait''} for new hardware was a valid strategy to improve wall clock time. From $2004$, the clock rate plateaued at $1.7-3.4$ GHz, and microprocessor vendors began incorporating multiple cores, cache hierarchies, and specialized units. After the paradigm shift, the compute architecture research  discovered that there are two ways to progress:

\paragraph{Design of domain-specific compute architectures.} The key lies in adapting algorithms to the architecture's features. This process involves closely intertwining algorithms with the presented functional compute units and the intricate rules governing the utilization.

\paragraph{Refining software implementation.} Modern scripting languages offer the advantage of democratizing algorithm implementations. But their eco-system clashes too much with the principles of the real hardware and systems (\clrshort{see arguments in Appendix~\ref{ch7:app:nopython}}). \citet{leiserson2020there} underscore this fundamental conflict by demonstrating a staggering $\times 62\,806$ speedup for dense matrix multiply.

\subsection{Contributions}
\label{ch7:contributions}

Inspired by \citet{safaryan2021fednl}, we have explored the landscape of enhancements to make \algname{FedNL}  more practical. One line of possible research involves modifying optimization algorithms where step size computation explicitly avoids dependence on problem-specific constants. Notably, \algname{FedNL} already exhibits this property. Next, during our experimentation with a reference implementation of \algname{FedNL}, a significant challenge arose in launching numerical experiments. Launching \algname{FedNL} experiments using provided prototypes took $4.8$ hours for a single optimization process. Next, we noticed that the referenced prototype only simulates a distributed environment.  The \algname{FedNL} has a super-linear local convergence rate. With this level of theory development, the gain from further theoretical improvements might not be as substantial as those derived from a highly optimized implementation. These aspects motivated us to create a more well-developed \algname{FedNL} implementation on top of the original work. We acknowledge the considerable challenges of simultaneously developing both a comprehensive theoretical framework and a proficient implementation. Therefore in the present work, we focused our efforts on enhancing its \textit{practical implementation} and \textit{practical applicability} of \algname{FedNL} algorithm family. \clr{The practical implementation also serves other needs. 
	
First, the theoretical cost model may not be entirely accurate and needs to be adjusted (for examples of nuances in modern compute systems see Section~$  $\ref{ch7:sec:int-div}, Appendix~\ref{ch7:app:memory-hierachy-latencies}). Second, without real implementation, scientific methods often remain confined to theory. As S. Boyd noted in \citet{interviewboyd}, this is one reason why Control Theory has experienced limited adoption in the broad sense.
}


Summary of our contributions:

\begin{enumerate}
	\item \textbf{Compute effectiveness.} We addressed the challenge of executing computationally intensive multi-node simulations for \algname{FedNL} on a single workstation, achieving a remarkable $\times 1000$ improvement in wall-clock time. 
	
	\item \textbf{Self-contained design.} We introduced a self-contained single-node and multi-node implementation of \algname{FedNL} (Algorithm \ref{ch7:alg:FedNL}), \algname{FedNL-PP} (Algorithm \ref{ch7:alg:FedNL-PP}), \algname{FedNL-LS} (Algorithm \ref{ch7:alg:FedNL-LS}). Our design facilitates seamless integration into resource-constrained systems and eliminates the need for library dependency management. Our solution relies only on {OS} interfaces. It is compatible with several OS, Compilers (Appendix \ref{ch7:app:supported-os-and-compilers}),
	CPUs (Appendix~\ref{ch7:app:cpus}). We provide native OS executable applications, dynamic libraries, and static libraries. In addition, we provide an easy way to generate extension modules for other programming languages (Appendix \ref{ch7:app:usability}).

	
	\item \textbf{Adaptive TopLEK compressor.} We introduced an extension of the \compname{TopK} compression mechanism, termed \compname{TopLEK}. The core idea is to perform compression for \compname{TopK} adaptively and to compress as much as theory allows, but not more (for details see Appendix~\ref{ch7:app:toplek}).
	
	\item \textbf{Cache-aware RandSeqK compressor.} We proposed a cache-aware version of \compname{RandK} compressor, named as \compname{RandSeqK}. The theory of \compname{RandK} provides the need \textit{''null space''}, which we have exploited to make the algorithm \textit{cache-aware} (for details see Appendix~\ref{ch7:app:seqk}).
	
	\item \textbf{Logistic regression with FedNL outperforms best-practice solutions.} In a single-node setup, our implementation practically outperforms solvers encapsulated within \libname{CVXPY} \citep{diamond2016cvxpy}, including the commercial \libname{MOSEK} \citep{aps2022mosek} for solving \modelname{logistic regression}. In a multi-node setup, our implementation surpasses the performance of \libname{Apache Spark MLlib} \citep{meng2016mllib} and \libname{Ray/Scikit-Learn} \citep{moritz2018ray}. Our implementation has an initialization time smaller by $\times 25 - \times 50$, and exhibits faster solving by a factor $\times 7 - \times 82$ in wall clock time (see Section~\ref{ch7:sec:experiments}).
	
	\item \clr{\textbf{First robust practical implementation.} To the best of our knowledge, our implementation is the first \textit{robust}, \textit{practical} implementation for training (strongly) convex objectives Equation~\eqref{ch7:eq:main} in FL settings (see Section~\ref{ch7:sec:existing-fl-sota-system}).
	}
	
\end{enumerate}

Contributions \clrshort{(3)}--(6) are more focused toward {FL}. The principles from contributions (1)--(\clrshort{2}) are valuable in scenarios when a theoretical  compelling\footnote{A robust theory is necessary. The good practical implementation, especially in C++, demands a significant amount of time.} ML algorithm requires a strong realization. We believe some of our findings and improvements raise interesting questions for designing training algorithms because the achievement of improvements on the order of $\times 1000$ (if the underlying compute and storage hardware is fixed) indicates underlying fundamental issues. The broader impact of our work is elaborated in Appendix~\ref{ch7:app:discussions}.

\section{Background on FedNL}
\label{ch7:preliminaries}

{
\begin{algorithm}[H]
	\caption{\algname{FedNL}: {Baseline} from \citet{safaryan2021fednl} \clrshort{\textbf{[Existing]}}.}
	\label{ch7:alg:FedNL}
	\begin{algorithmic}[1]
		\STATE \textbf{Input:} Hessian learning rate $\alpha\ge0$; compression operators $\{\cC_1^k, \dots,\cC_n^k\}$
		\STATE \textbf{Initialize:} $x^0\in\R^d$; $\mH_1^0, \dots, \mH_n^0 \in \R^{d\times d}$ and $\mH^0 \eqdef \dfrac{1}{n}\sum_{i=1}^n \mH_i^0$
		\FOR{each device $i = 1, \dots, n$ in parallel} 
		\STATE Get $x^k$ from the server and compute local gradient $\nabla f_i(x^k)$ and local Hessian $\nabla^2 f_i(x^k)$
		\STATE Send $\nabla f_i(x^k)$,\; $\mS_i^k \eqdef \cC_i^k(\nabla^2 f_i(x^k) - \mH_i^k)$ and $l_i^k \eqdef \|\mH_i^k - \nabla^2 f_i(x^k)\|_{\rm F}$ to the server
		\STATE Update local Hessian shift to $\mH_i^{k+1} = \mH_i^k + \alpha\mS_i^k$
		\ENDFOR
		\STATE On Server:
		\STATE \quad Get $\nabla f_i(x^k),\; \mS_i^k$ and $l_i^k$ from each node $i\in [n]$
		\STATE \quad $\mS^k = \dfrac{1}{n}\sum\limits_{i=1}^n \mS_i^k,\; l^k = \dfrac{1}{n}\sum\limits_{i=1}^n l_i^k,\; \mH^{k+1} = \mH^k + \alpha\mS^k$
		\STATE \quad {\textbf{Option 1 (a)}:} $x^{k+1} = x^k - [\mH^{k}]_{\mu}^{-1} \nabla f(x^k)$ \quad {\textbf{Option 2 (b)}:} $x^{k+1} = x^k - [\mH^{k} + l^k\mI]^{-1} \nabla f(x^k)$
	\end{algorithmic}
\end{algorithm}
}

The optimization problem which \algname{FedNL} solves has a finite sum structure described by \eqref{ch7:eq:main}. Here, $n \in \mathbb{N}$ is the number of clients in a distributed system, $d \in \mathbb{N}$ denotes the dimension of the problem, and $x^k \in \mathbb{R}^d$ represents the model parameters at iteration $k$. When using \algname{FedNL} for {FL} problems in {ML}, $f_i(x)$ provides the score criteria for using the model $x$ on client $i$ data. The optimization formulation in Problem~\eqref{ch7:eq:main} encodes the final goal of the training process as a selection of a function from a parameterized function class $\mathcal{F}$ indexed by $x \in \mathbb{R}^d$. In this context, $f_i(x)$ often takes the form $$f_i(x) \eqdef \dfrac{w_i}{n_{i}} \sum_{1 \le j \le n_i} (\mathcal{L}_{ij}(b_{ij}, \hat{F}(a_{ij};x)) + R_{i}(x)).$$ Here $f_i(x)$ assesses a predictive model $\hat{F}(\cdot;x) \in \mathcal{X}\to \mathcal{Y}$ common to all $n$ clients, $n_i \in \mathbb{N}$ denotes the number of input-output pairs at client $i \in [n]$, and $(a_{ij}, b_{ij}) \in \mathcal{A} \times \mathcal{B}$ represent the single input-output pair with the number $j$ at client $i$. The function $\mathcal{L}_{ij}(y_{\mathrm{real}}, y_{\mathrm{pred}} ): \mathcal{B} \times \mathcal{B} \to \mathbb{R}$ is a loss function that scores prediction, and $R_{i}(x): \mathbb{R}^d \to \mathbb{R}$ is the regularization function. The weight $w_i \ge 0$ encodes the role of client $i$.

The convergence guarantees of \algname{FedNL} hold if $f(x)$ satisfies Assumption \ref{ch7:asm:1} and $f_i(x)$ satisfies Assumption \ref{ch7:asm:2}. While \algname{FedNL} does not require strong convexity of functions $f_i$, it is required for $f(x)$. In \algname{FedNL} (Algorithm~\ref{ch7:alg:FedNL} from Section 3.4 \citet{safaryan2021fednl}), the only quantity not evaluated in runtime is $\alpha \in \mathbb{R}$. It is derived from the global characteristics of the used compressor.

\clr{

To integrate user-defined optimization problems, users must explicitly define oracles $\nabla^2 f_i(x)$, $\nabla f_i(x)$, and $f_i(x)$. To facilitate this process, we provide a comprehensive collection of mathematical primitives for computations on both CPU and NVIDIA GPUs, as well as system primitives for efficient input/output management and memory utilization. Also, we offer tools for numerically verifying the correctness of the $\nabla^2 f_i(x)$ and $\nabla f_i(x)$ oracles (see Appendix~\ref{ch7:app:usability}, \ref{ch7:app:futresearch}). Our implementation follows a modular design. Due to the principles of the selected language (C++ ISO/IEC 14882:2020), modern compilation tools, and static linkage, there is no performance penalty from the modularity. 
}

\section{Compute and Storage Demands of a Single Worker}
\label{ch7:problem_single_node}

Algorithm \ref{ch7:alg:FedNL} encapsulates the essential procedures employed when multiple \textit{clients} collaboratively solve the optimization problem \eqref{ch7:eq:main}. The algorithm's versatility is independent of specific computing hardware, but in realistic scenarios, computing devices should be fixed. In our implementation, we target modern general-purpose central processing units. 

\clr{This type of computing device is ubiquitous, spanning from server-grade machines to portable devices. Even when electronic components are integrated into a single System-on-Chip, the Central Processing Unit (CPU) remains a fundamental element. The architectural details at the assembly language level, have evolved sustainably over the decades, facilitating effective decoupling between Electrical Engineering and Computer Science and Engineering. In contrast, the situation with creating low-level implementations for Graphics Processing Units (GPUs) is entirely different. The Application Programming Interfaces\footnote{OpenCL for ARM Mali GPU, CUDA for NVIDIA GPU, Metal API for Apple GPU, and ROCm for AMD GPU}, memory management rules, and computational organization principles evolve rapidly, even within a single vendor.
}

\clrshort{To evaluate our implementation of the \algname{FedNL}} a specific class of optimization problems must be chosen and we chose $L_2$ regularized \modelname{logistic regression}. This selection is guided by the existence of experiments with this objective in the original \algname{FedNL} paper, and the fact that it ensures strong convexity. 

The \modelname{logistic regression} can be obtained from Problem~\eqref{ch7:eq:main} through the following specialization $f_i(x)$ for client $i \in \{1\dots,n\}$ which contains $n_i$ samples:

\begin{equation}
\label{ch7:eq:fi_log_reg_structure}
f_i(x) = \dfrac{1}{n_i} \sum_{j=1}^{n_i} \log (1 + \exp(-b_{ij} \cdot a_{ij}^\top x)) + \dfrac{\lambda}{2} \|x\|_2^2, \quad b \in \mathbb{R}^{n \times n_i}, a \in \mathbb{R}^{n \times n_i \times d}
\end{equation}
From Equation~\eqref{ch7:eq:fi_log_reg_structure} we can analytically compute $\nabla f_i(x)$, $\nabla^2 f_i(x)$ by:

\begin{eqnarray}
\label{ch7:eq:grad_fi_logreg_structure}
\nabla f_i(x) = \mA_i \times \begin{bmatrix} \dfrac{-1/n_i}{\exp (x^\top (b_{i,1} a_{i,1})) + 1} \\ \dots \\ \dfrac{-1/n_i}{\exp (x^\top (b_{i,n_i} a_{i,n_i})) + 1} \end{bmatrix} + \lambda \cdot x,
\end{eqnarray}

\begin{eqnarray}
\label{ch7:eq:hessian_fi_logreg_structure}
\nabla^2 f_i(x) = \mA_i \times {\textcolor{blue}{H_i}} \times \mA_i^{\top} + \lambda \cdot \mI_{d \times d}.
\end{eqnarray}
Where $\mI_{d,d} \in \mathbb{R}^{d \times d}$ is the identity matrix, $A_i$ matrix contains samples by columns and $H_i \in \mathbb{R}^{n_i \times n_i}$ is diagonal matrix consisting of:
\begin{eqnarray}
\label{ch7:eq:hessian_structure_h_k}
[{\textcolor{blue}{H_i}}]_{j,j} &=& \dfrac{1}{n_i} \dfrac{\exp(x^\top \cdot b_{ij} a_{ij})}{ \left( 1 + \exp(x^\top \cdot b_{ij} a_{ij}) \right)^2 }, \\
\mA_i &\eqdef& \begin{bmatrix}	b_{i,1} \cdot a_{i,1}, b_{i,2}\cdot a_{i,1} , \dots, b_{i,n_i} \cdot a_{i,n_i} \end{bmatrix}^{d \times n_i} \notag.
\end{eqnarray}

These analytical formulas have been verified numerically. Regarding memory, our implementation allocates virtual memory in the executing processes without utilizing specific mechanisms in the operating system to force dedicated physical memory (via \textit{page-locking}, also known as \textit{pinning} mechanisms). 

The virtual memory is used to store $x \in \mathbb{R}^d$, $\nabla f_i(x) \in \mathbb{R}^d$, $\nabla^2 f_i(x) \in \mathbb{R}^{d \times d}$, design matrix $\mA_i \in \mathbb{R}^{d \times n_i}$ and other intermediate buffers. The selected dataset in LIBSVM \citep{chang2011libsvm} format is read from disk storage twice, following a schema similar to Scikit-Learn~\citep{pedregosa2011scikit} and \libname{Apache Spark MLlib}~\citep{meng2016mllib}.

\section{Compute Issues in Original Implementation}
\label{ch7:sec:issues}

The reference implementation of \algname{FedNL} was developed in Python \citep{van1995python} with \libname{NumPy} \citep{van2011numpy} employed as the computational backbone. A single execution of the provided implementation for \modelname{logistic regression} utilized parameters $d=301$, $n=142$, derived by splitting the LIBSVM \dataname{W8A} dataset into chunks $n_i=348$, takes $19,770$ seconds for \compname{TopK[$k=8d$]} and $17,510$ seconds for \compname{RandK[$K=8d$]}, with $r=1000$ rounds of repeating Lines 3-11 of Algorithm~\ref{ch7:alg:FedNL}. This measurements were taken on a machine equipped with an \textit{Intel(R) Xeon(R) Gold 6246 CPU} \footnote{\href{https://ark.intel.com/content/www/us/en/ark/products/193969/intel-xeon-gold-6246-processor-24-75m-cache-3-30-ghz.html}{Intel Xeon Gold 6246 Processor Technical Specification}.} with 12 physical computation cores. The CPU clock frequency was set to $\mu=3.3$ GHz (For preparation details see Appendix~\ref{ch7:app:careful-reproduce}).

\paragraph{Back-of-the-Envelope calculation.} 
To estimate the lower bound on the execution time of the simulation for a fixed algorithm, we need to examine the internal organization of the \textit{{Intel(R) Xeon(R) Gold 6246 CPU}}. This organization is determined by the micro-architecture of \textit{{Cascade Lake}} processors, which encompasses information about the components inside the {CPU}. The adders and multipliers are electrical circuits that perform corresponding operations at the rising edge of the clock that synchronize the operations inside the CPU. In the \textit{Cascade Lake} microarchitecture, each physical CPU core contains 3 Float Functional Units (FPUs)~\footnote{\href{https://en.wikichip.org/wiki/intel/microarchitectures/cascade_lake}{Reference information about Cascade Lake Microarchitecture}.}. These functional units, responsible for float add, subtract, and multiply typically have a throughput of $1$ operation per clock cycle in modern {CPU} (See \citet{fog_instruction_tables}). Each {FPU} is a pipelined device. If the pipeline needs to be restarted, typically additional $4$ clocks are incurred per operation. The computation demands of Algorithm \ref{ch7:alg:FedNL} require each of $n$ client compute Hessian with  $\mathcal{O}(d^2\cdot n_i)$ arithmetic operations, full gradient with $\mathcal{O}(d\cdot n_i)$ arithmetic operations, function values $\mathcal{O}(d\cdot n_i)$ per round. Hessian compression takes $\mathcal{O}(d^2)$, update the Hessian shift $\mathcal{O}(d^2)$ arithmetic operations. 

Clients' compute logic happened during $r$ rounds requires: $$\mathcal{O}\left( (d^2\cdot n_i + d n_i + 2d^2) \cdot r\right)/(\mu \cdot \rm{cores} \cdot \rm{fpu}) \propto 0.26\,\rm{sec.}$$

The master has to perform $n$ additions of Hessian with $d \cdot k$ elements, and $n$ additions of gradients with dimension $d$. The master's compute time for processing:
$$\mathcal{O}\left( (d\cdot k + d) \cdot r \cdot n\right)/(\mu \cdot \rm{cores} \cdot \rm{fpu}) \propto 0.0032 \,\rm{sec.}$$

The master also needs to solve a linear system in each round. In the reference implementation, it was done using Gaussian elimination, which demands $(2/3)d^3$ arithmetic operations. Time for it:
$$\mathcal{O} (
{3}/{2} \cdot d^3 \cdot r 
) / (\mu \cdot \rm{fpu}) \propto 4.1316 \,\rm{sec.}$$

With optimal CPU core utilization, the time for \textit{float arithmetic} should be a small multiple of $4.394$ sec. However, \textit{float operations} are also required to get the actual operands. For doing this, the {CPU} control unit sends requests to the Load and Store Units (LS), which is also in charge of memory access to {CPU} registers and caches. The number of LS units in \textit{{Cascade Lake}} is $3$. If we assume that LS units work most of the time with {L1} cache, then extra penalty $\times 3$ (See Table~\ref{ch7:tbl:latencies-for-memory}) should be paid per single access. If we assume that each float operation requires $3$ memory accesses, then the memory access penalty for \textit{float operations} will be $(4.394 \cdot \rm{fpu}) / \rm{ls} \cdot 3 = 13.182$ sec.

\paragraph{Insights from rough estimates.} The rough lower bound of the estimated target execution time is $17.576$ sec. This estimate does not include memory cost for transfers in client-master communication,  controlling logic, and \textit{possible improvements} from utilizing Single Instruction Multiple Data (SIMD) instructions available in this {CPU}. The observed time for launching the \algname{FedNL} baseline is $19770\,\rm{sec.}$ ($5.5$ hours), indicating a notable discrepancy:

$$5.5\, \rm{hours} \ggg 17.576\,\rm{sec.}$$ 

We addressed the challenges related to creating a practical single-node simulation of \algname{FedNL}, ensuring its compatibility with diverse {OS}/{CPU} configurations (see Appendix~\ref{ch7:app:supported-os-and-compilers}). After providing an optimized single-node simulation, we created a practical multi-node implementation.

\subsection{Computation problems of reference implementation}

After examining Python/\libname{NumPy} reference implementations, we found a deeper issue. Major ML frameworks, burdened by extensive auxiliary management code, are suboptimal for complex system and algorithm creation with high-performance requirements. 

To tackle this, we have shifted away from general-purpose ML and FL middleware and the prevalent Python-centric design philosophy. \clrshort{For further discussions on the performance limitations of the Python ecosystems, see Appendix~\ref{ch7:app:nopython}.}

\section{Structure of x1000 Time Improvement}

We present more notable improvements from the \algname{FedNL} simulation in one machine, focusing on the wall clock time improvement for training \modelname{logistic regression} based on  Equations~\eqref{ch7:eq:main}, \eqref{ch7:eq:fi_log_reg_structure}, using \dataname{LIBSVM} \dataname{W8A} dataset. For finer granularity of improvements see Appendix~\ref{ch7:app:history-of-improvements}. We augmented each sample in a dataset with an artificial feature equal to $1$ to have an intercept term. After augmentation, \dataname{W8A} dataset contains $d=301$ features. Number of rounds $r=1000$. The dataset is reshuffled u.a.r and was split across $n=142$ clients with $n_i=350$. The $x^0=0$ and regularization coefficient $\lambda=0.001$, for this problem $\lambda\left(\nabla^2 f\right) \in [0.001, 0.0058]$. 

The overall time encompasses: (i) loading and parsing the dataset; (ii) distributing the dataset and preparing runtime; (iii) saving the outcome of the experiment to the disk; (iv) training with Algorithm \ref{ch7:alg:FedNL}. The (i)-(iii) takes $4.7\%$, and (iv) $95.3\%$ of the  simulation time.

\subsection{Naive C++ implementation: x20}
\label{ch7:sec:depart-from-python}

Targeting the {CPU} requires the \algname{FedNL} to be implemented in a formal programming language. Defining a \textit{programming language} lacks universal consensus. The definition advocated by authors of compiler-based languages suggested that a language must directly translate logic into computation devices. Scripting languages lack this feature due to the requirement of an interpreter. Studies \citet{pereira2017energy}, \citet{leiserson2020there} further highlight the superior speed and energy efficiency of compiled languages. Transitioning from Python/\libname{NumPy} to C++ \citep{cpp} yielded a $\times 22$ time improvement for \algname{FedNL/TopK} and $\times 16$ for \algname{FedNL/RandK}. Each client is represented as a user-space thread in the simulation.


\subsection{Data processing optimization: x1.077}

We optimized input data processing in \dataname{LIBSVM} format by shifting from sequential I/O to memory-mapped files, along with custom string$\to$FP64 conversion, resulting in a speedup of $\times 1.077$. \clrshort{For details on memory-mapped files, see Appendix~\ref{ch7:app:mmap}.}

\subsection{Eliminating some integer division: x1.225}
\label{ch7:sec:int-div}

Combination circuits handling integer and floating-point operations exhibit latency from $1$ to $4$ clocks. The division poses a challenge due to microarchitectural complexities, often requiring approximately $56$ clocks \citep{fog_instruction_tables}. In our dense linear algebra implementation, we optimized indexing operations by eliminating a division during indexing. These optimizations yielded a $\times 1.2$ gain. In our dense matrix/vector implementation, we leverage SIMD CPU instructions ({AVX-512}). To chunk dense arrays into 512-bit packs and the rest we utilized bit tricks. Our implementation requires $7$ x86-64 instructions, and GCC 11.4 emits $9$ (from C++ code). It gave extra $\times 1.0212$ speedup.

\subsection{Utilizing AVX512 CPU extension: x1.379}


Data alignment and utilization of the AVX-512 instruction set resulted in an additional gain $\times 1.379$.

\subsection{Compiler and linker optimization: x1.128}

By explicitly disabling exceptions and runtime type information in C++ compilers, we achieved a gain of $\times 1.07$. Leveraging whole program optimization, where code-emitting decisions are deferred to the final stages, adds an extra gain of $\times 1.047$. Employing force inlining at specific locations results in an improvement of $\times 1.007$. The combined effect of compile-time optimization is $\times 1.128$.

\subsection{Use sparsity from FedNL compressors: x1.44}

The \algname{FedNL} Lines 5,6,10 contain sparse updates. Exploiting this for the dense matrices has a drawback in that SIMD cannot be used, but the gain from not performing useless arithmetics outweighs it.

\subsection{Reuse computations from oracles structure: x1.50}


There exists a significant redundancy in the structure of $f_i(x)$, $\nabla f_i(x)$, $\nabla^2 f_i(x)$ as presented in Equation~\eqref{ch7:eq:fi_log_reg_structure}, \eqref{ch7:eq:grad_fi_logreg_structure}, \eqref{ch7:eq:hessian_fi_logreg_structure}, \eqref{ch7:eq:hessian_structure_h_k}. The classification margins $b_{ij} \langle x^\top, a_{ij} \rangle$ are reused within a single client three times. The computational cost in clock cycles is approximately $\mathcal{O}(d)$ for elementary add/multiply operations. Not only is the classification margin reused in all oracles but also the sigmoid function values $g(z)=(1 + \exp(-z))^{-1}$ and quantities derived from it, such as $g(-z)=1-g(z)$ and $g(-z) \cdot g(z)=\exp(z)/(1+\exp(z))^2$ are implicitly reused in all three oracles. We eliminated this by reusing common values, incurring the cost of storing two vectors of dimension $\mathcal{O}(n_i)$ for margins and sigmoid values.

\subsection{Basic linear algebra improvements: x1.338}

A careful implementation of adding the same scalar to the diagonal in Line 11 resulted in a gain of $\times 1.06$. Exploiting the symmetry of the input matrix when computing the Frobenius norm in Line 5 provided a gain of $\times 1.00751$. Explicitly storing information about the number of columns in a dense matrix yielded a $\times 1.0421$ gain. Implementing an efficient memory \textit{move} and \textit{copy} for matrix gave $\times 1.019$. Improving the initialization of matrices gave $\times 1.057$. To enhance instruction-level parallelism, the manual loop unrolling for vector-vector and vector-scalar operations gave $1.034$. Eliminate the aliasing effect problem in C++ code gain $\times 1.08$. Cumulative gain is $\times 1.3386$.

\subsection{Linear system solve improvements: x1.31}

We transitioned from dense Gaussian elimination to the more numerically stable Cholesky-Banachiewicz decomposition for dense matrices \citep{golub2013matrix}. We didn't explore the connections between \algname{FedNL} and \textit{indirect solvers} like Krylov methods \citep{kelley1999iterative} and Multi-grid methods \citep{hackbusch2013multi}. \textit{Dense direct} methods are nonheuristic, independent of problem instances.  The switch to Cholesky, with optimized forward-backward substitution, yielded a $\times 1.196$ gain. Further enhancements, with in-place vector arithmetic and cache-aware computation organization, contributed to an addition  $\times 1.096$ gain.

\subsection{Hessian and gradients oracles: x3.072}
\label{ch7:sec:better-hessians-for-logreg}

We began with the computation of $\nabla^2 f_i(x)$ using a naive implementation that employed straightforward matrix multiplication with three nested loops. We progressed to cache-aware matrix tiled multiplication, incorporating $9$ nested loops. The optimal tile size can be estimated based on the CPU's L1 and L2 data cache sizes, which are $32$ and $1024$ KBytes in our {CPU}. The optimal tile size to operate on three matrices (two inputs and output) are $\sqrt{L1/8}/3$, $ \sqrt{L2/8}/3$ \citep{jia1981complexity} for FP64 matrix items. Theory recommends tile sizes of $21$ and $120$, but practical experimentation favored the smaller sizes as $4$ and $32$. The gap arises from the difference between cache-aware and cache-oblivious (\abr{CO}) algorithms. The \abr{CO} algorithms adapt to the multilevel structure of caches, available cache size in a specific moment implicitly \citep{demaine2002cache} and do not require explicit knowledge of caches in advance. We tried \abr{CO} schema \footnote{\href{https://math.mit.edu/~stevenj/18.335/oblivious-matmul-handout.pdf}{Cache-Oblivious Matrix Multiplication} by \href{https://math.mit.edu/\~stevenj}{Steven G. Johnson}: \href{https://math.mit.edu/\~stevenj/18.335/oblivious-matmul-handout.pdf}{https://math.mit.edu/\~{}stevenj/18.335/oblivious-matmul-handout.pdf}} for matrix multiplication with 8-way splitting which is to best of our knowledge is standard schema for \abr{CO} matrix multiplication to date. It resulted in a $\times 3$ speedup compared to the tuned tiled version! The final possible gain is $\times 1.2$.

\paragraph{Better strategy.} Alternatively, we can compute $\nabla^2 f_i(x^k)$ from Equation~\eqref{ch7:eq:hessian_fi_logreg_structure} as the sum of symmetric rank-1 matrices. We can focus on the upper diagonal part and symmetrize the result matrix afterward (gain $\times 1.85$ overall). We did switch to processing $4$ samples with instruction-level parallelism inside the Hessian oracle (gain $\times 1.6303$). We fused operations for matrix-vector operations and added multiples of vectors required from Equation~\eqref{ch7:eq:grad_fi_logreg_structure} (gain $\times 1.0203$). The cumulative gain is $\times 3.077$.

\subsection{Better compressors implementation: x1.14}

Optimized \abr{SIMD} generation of integer sequences $(s, s+1, \dots)$, coupled with a known at compile-time $s$ value, yielded a $\times 1.006$ gain because this operation was used in data shuffling and in \compname{RandK}. Next, we computed and stored indices for the upper triangular part of $\nabla^2 f_i$ once without recomputing (gain $\times 1.0165$). For sparsification compressors, spending an additional $\mathcal{O}(k\log_2(k))$ time per client on sorting by index improved memory access in master, resulting in a $\times 1.0182$ gain.

In implementing \compname{TopK}, diverse approaches were explored, including quicksort, merge sort, radix sort, randomized order statistics, \abr{CO} Multi-way merge sort \citep{cormen2022introduction}, \abr{CO} Funnelsort \citep{frigo1999cache}. The most efficient implementation involved a 4-way Min-Heap \citep{tarjan1983data} for supporting $K$ smallest items found so far, resulting in a $\times 1.0412$ gain. In the \compname{RandK} compressor, adopting in-place memory shuffling (instead of using two separate arrays of size $\mathcal{O}(d^2)$) gained $\times 1.073$. Finally, during waiting for $x^{k+1}$ (Line 4, Algorithm~\ref{ch7:alg:FedNL}), clients start precomputation of indices for the next round (gain is $\times 1.0211$).

\subsection{Subtleties with multi-threading: \clrshort{x}1.412}

To minimize client wait times we utilized busy loops on atomic variables instead of semaphore {OS} primitives, resulting in a gain of $\times 1.0057$. 

To minimize system contention we established a thread pool with a size matching the number of physical cores. Clients were statically dispatched to this pool, avoiding unnecessary congestion for compute cores. Next, we processed messages from clients as they became available. The last two techniques result in a combined gain of $\times 1.404$. This is a common technique in parallel and concurrent programming. These optimizations aim to minimize contention and enhance overall system efficiency.

\subsection{Subtle memory optimization: \clrshort{x}1.278}

In C++ runtime, threads allocate memory from a global heap with {OS} locking mechanism. To enhance efficiency, we introduced thread-custom memory pools for vectors and matrices, resulting in a gain of $\times 1.0109$. Additionally, using aligned load/store for dense vector/matrix operations in $\nabla^2 f_i(x)$, $\nabla f(x)$ oracles provided a gain of $\times 1.11$. The client-master communication was organized via a memory buffer. Next, we observed that labels $b_{ij}$ is not needed in Equations \eqref{ch7:eq:fi_log_reg_structure}, \eqref{ch7:eq:grad_fi_logreg_structure}, \eqref{ch7:eq:hessian_fi_logreg_structure} explicitly and can absorbed into $\mA_i$ explicitly. It gave a gain of $\times 1.009$. Next, in the naive implementation, clients stored both $\mA_i$ and ${\mA_i}^\top$. By eliminating the storage of $\mA_i$ and adding the functionality for matrix operation with transposed argument, we gained $\times 1.129$. Cumulative gain is $\times 1.278$.

\section{Intermediate Conclusion on Single-Node Simulation}




The total speedup gained from the previous steps is $\times 1000$. Extra improvements are more technical and obtained with profiling tools such as Valgrind \citep{nethercote2007valgrind}, and LLVM remarks \citep{lattner2004llvm}. For details, see Appendix~\ref{ch7:app:history-of-improvements}.

\section{Network Improvements: x1.29}

\paragraph{Baseline.} We focused on a setting where communication happens via \abr{TCP/IP} for the reasons in Appendix~\ref{ch7:app:networks-why-tcp}. We took the setting $n=10$, \compname{TopK[$k=8d$]}, $r=300$, \dataname{W8A} dataset. Each client$\to$master stream of information was organized as a separate \abr{TCP} connection. The baseline took $14.92$ seconds. 

\paragraph{Improvements.} We observed that it is more effective to have a single communication channel from client to master. Next, we interleaved connection establishment with the master and dataset loading in the clients. For \compname{RandK} and \compname{RandSeqK} compressors, we enable index reconstruction using a pseudo-random generator. For \compname{TopK} and \compname{TopLEK}, it is necessary to transfer indices. We found that employing a fixed-width $32$-bit integer format surpassed the performance of a strategy involving varying sizes. Next, we intentionally disabled the Nagle algorithm \citep{nagle1984congestion} due to the explicit formulation of small buffers, as detailed in Appendix~\ref{ch7:app:networks-background}.

\clearpage
\section{Introduced Compressors}

In our pursuit of bridging the gap between \textit{theory} and \textit{practice} we asked ourselves:
\begin{center}
\textit{"Is any nullspace of the introduced compressors that can be exploited for practical purposes?"}
\end{center}

We observed that the \compname{TopK} compressor in the original work is used solely through its contractive definition proposed in \citet[p.11]{safaryan2021fednl}. This absence of an unbiasedness requirement allows for the construction of \textit{"any algorithm"} satisfying this property. We introduced \compname{TopLEK[k']} described in Appendix~\ref{ch7:app:toplek}. It adapts the \compname{TopK[$k$]} compressor, transforming it into a search process for \compname{TopK[$k' \le k$]}. The selection of $k' \in [0, k]$ ensures that contractive inequality becomes a tight equality.

Next, the \compname{RandK} compressor exhibits an intriguing degree of freedom. While \compname{RandK} selects a subset of coordinates of cardinality $k$ uniformly at random (u.a.r.) from a total of $d$ coordinates, zeroing out the rest and scaling the output to preserve unbiasedness, an alternative realization of item selection is possible. In \compname{RandSeqK} schema, the group of coordinates is sequentially chosen from the start index $s\sim_{u.a.r.} [d]$, but next $k-1$ are sequential deterministic indices $\mathbb{Z}_d$. This \compname{RandSeqK} schema is unbiased and satisfies \citet[p.11]{safaryan2021fednl} and has the same variance as \compname{RandK}. But it is more appealing for practice because (a) it is cache-aware; (b) the number of invocations of the used pseudo-random generator is $1$. For a more elaborate presentation of \compname{RandSeqK} see Appendix~\ref{ch7:app:seqk}.

\section{Experiments}
\label{ch7:sec:experiments}

We trained $L_2$ regularized \modelname{logistic regression} as defined in Equation~\eqref{ch7:eq:fi_log_reg_structure} with $\lambda=0.001$. The initial iterate $x^0=0$. When utilizing \compname{RandK} and \compname{RandSeqK}, we leveraged our implementation's ability to (optionally) reconstruct indices of sparsified information. For \compname{TopK} and \compname{TopLEK}, the indices are transferred as 32-bit integers. See Appendix~\ref{ch7:app:software-env},~\ref{ch7:app:hardware-env} for details on the software and hardware environment. Employed steps for reliable time measurements are described in Appendix~\ref{ch7:app:careful-reproduce}. We augmented implementation with \compname{Natural} compressor \citep{horvath2019natural}, which behaves remarkably well for \algname{FedNL}. The selected datasets are example datasets used in the original \algname{FedNL} paper by \citet{safaryan2021fednl}.



\subsection{Single-node: baseline vanilla FedNL improvement}
\label{ch7:sec:single-node-fednl}

In a single-node simulation, we tackled \modelname{logistic regression} with a dimensionality of $d=301$. The experiment involved a total of $n=142$ simulated clients and the number of rounds $r=1000$. The \dataname{W8A} dataset reshuffled u.a.r. and was partitioned into equal $n_i$ chunks. We utilized \algname{FedNL} without Line Search, Option-B, $\alpha$ using option-2. Results are presented in Table~\ref{ch7:tbl:compare-in-singlenode}. It can be observed that our \algname{FedNL/RandK[K=8d]} implementation achieves a significant total speedup of $\times 929.4$ in a single-node setup compared to the baseline. Similarly, for \algname{FedNL/TopK[K=8d]}, the total speedup from our implementation is $\times 1053.9$. The master aggregates the following data from all clients:
(i) $2\,937.0$ MBytes for \compname{RandK} and \compname{RandSeqK} compressors;
(ii) $49\,568.7$ MBytes for \compname{Identical} mapping $\mathbb{C}(x)\eqdef x$ compressor;
(iii) $4\,241.4$ MBytes for \compname{TopK};
(iv) $358.8$ MBytes for \compname{TopLEK} compressor.

We employed a two-level schema for $\nabla f_i(x)$ aggregation in our implementation. After obtaining $\nabla f_i(x)$, the master dispatches updates to one of the $4$ helper threads (configurable) in a round-robin fashion, incurring an extra $\mathcal{O}(d)$ memory storage per helper. Once all workers finish work, the master performs the final aggregation. For Hessian updates, the master utilizes another pool of configurable $4$ (configurable) helpers responsible for decompression and atomic updates to $H^k$.

{
\begin{table}[h!]
	\footnotesize
	\centering
	\begin{threeparttable}
		\caption{Single-node setting, $n=142$, \algname{FedNL} (B), $r=1000$, $\lambda=0.001$, $\alpha$ - option 2, FP64, $24$ cores at $3.3$ GHz.}
		\begin{tabular}{|l|p{0.43\textwidth}|c|l|}
			\hline
			\textbf{\#} & \textbf{Client Compression} & $\| \nabla f(x^{last}) \|$ & \makecell[l]{\textbf{Total Time}\\\textbf{(seconds)}} \\
			\hline
			\hline
			\cellcolor{bgcolorwe}{1} &\cellcolor{bgcolorwe}RandK[K=8d] (We) & $3 \cdot 10^{-18}$ & $18.84$ \\
			\hline
			2& RandK[$k=8d$] (Base) & $3 \cdot 10^{-18}$  & $17\,510.00$ \\
			\hline
			\cellcolor{bgcolorwe}3& \cellcolor{bgcolorwe}TopK[K=8d] (We) & $2.80 \cdot 10^{-18}$ &  $18.72$    \\
			\hline
			4&TopK[$k=8d$] (Base) & $2.80 \cdot 10^{-18}$ & $19\,770.00$   \\
			\hline
			\cellcolor{bgcolorwe}5&\cellcolor{bgcolorwe}RandSeqK[K=8d] (We) & $3.19 \cdot 10^{-18}$ & $16.70$      \\
			\hline
			\cellcolor{bgcolorwe}6&\cellcolor{bgcolorwe}TopLEK[K=8d] (We) & $3.45 \cdot 10^{-18}$ & $18.48$   \\
			\hline
			\cellcolor{bgcolorwe}7&\cellcolor{bgcolorwe}Natural (We) & $3.10 \cdot 10^{-18}$ & $27.02$   \\
			\hline
			\cellcolor{bgcolorwe}8&\cellcolor{bgcolorwe}Ident (We) & $2.46 \cdot 10^{-18}$ & $24.12$   \\
			\hline
		\end{tabular}
		\label{ch7:tbl:compare-in-singlenode}
	\end{threeparttable}
\end{table}
}

\begin{table}[h!]
\footnotesize
\centering
\begin{threeparttable}
	\caption{Single-node setting, $n=142$, \algname{FedNL-LS} (B), $\|\nabla f(x^{last}) \| \approx 9\cdot~10^{-10}$, FP64, $24$ cores at $3.3$ GHz.}
	\begin{tabular}{|l|l|c|c|c|c|c|}
		\hline
		\textbf{\#} & \makecell{\parbox{5cm}{\textbf{Solver}}}
		& \makecell{\dataname{W8A}, \\ $d=301$, \\ $n_i=350$} & \makecell{\dataname{A9A},\\ $d=124$, \\ $n_i=229$} & \makecell{\dataname{PHISHING},\\ $d=69$, \\ $n_i=77$} \\
		\hline
		\hline
		\multicolumn{5}{|c|}{\textbf{Initialization Time (seconds)}} \\
		\hline
		1 & CVXPY & +2.54 & +2.33 & +2.28      \\
		\hline
		2 & \algname{FedNL}  & +0.939 & +0.196 & +0.081 \\
		\hline
		\multicolumn{5}{|c|}{\textbf{Solving Time (seconds)}} \\				
		\hline
		3 & CLARABEL & 19.24 & 10.83 & 2.50 \\
		\hline
		4 & ECOS & 22.22 & 8.02  & 2.55 \\
		\hline
		5 & ECOS-BB & 22.00 & 8.00 &  2.12    \\
		\hline
		6 & SCS & 31.14 & 19.36 & 4.57   \\
		\hline
		7 & MOSEK & 16.90 & 9.59 &  3.55      \\
		\hline 
		8 & \makecell[l]{\algname{FedNL-LS} / RandK[$k=8d$]} & \cellcolor{bgcolorwe} 4.35 & \cellcolor{bgcolorwe} 0.34 & \cellcolor{bgcolorwe} 0.12   \\
		\hline
		9 & \makecell[l]{\algname{FedNL-LS} / RandSeqK[$k=8d$]} & \cellcolor{bgcolorwe} 3.34s & \cellcolor{bgcolorwe} 0.29 & \cellcolor{bgcolorwe} 0.06 \\
		\hline 
		10 & \makecell[l]{\algname{FedNL-LS} / TopK[$k=8d$]} & \cellcolor{bgcolorwe} 4.49 & \cellcolor{bgcolorwe} 0.46 & \cellcolor{bgcolorwe} 0.10  \\
		\hline 
		11 & \makecell[l]{\algname{FedNL-LS} / TopLEK[$k=8d$]} & \cellcolor{bgcolorwe} 4.79 & \cellcolor{bgcolorwe} 0.34 & \cellcolor{bgcolorwe} 0.61  \\
		\hline
		12 & \makecell[l]{\algname{FedNL-LS} / Natural} & \cellcolor{bgcolorwe} 3.13 & \cellcolor{bgcolorwe} 0.17 & \cellcolor{bgcolorwe} 0.08   \\
		\hline
		13 & \makecell[l]{\algname{FedNL-LS} / Identical} & \cellcolor{bgcolorwe} 0.63 & \cellcolor{bgcolorwe} 0.09 & \cellcolor{bgcolorwe} 0.06 \\
		\hline	
	\end{tabular}
	\label{ch7:tbl:compare-vs-cvxpy}
\end{threeparttable}
\end{table}

\begin{table}[h!]
\footnotesize
\centering
\begin{threeparttable}
	\caption{Multi-node setting, $n=50$ clients, $1$ master, $|\nabla f(x^{last})|\approx 10^{-9}$, FP64, $1$ {CPU} core/node.
	}
	\begin{tabular}{|l|l|c|c|c|c|c|}
		\hline	
		\textbf{\#} & \makecell{\parbox{5cm}{\textbf{Solution}}} & \makecell{\dataname{W8A} \\ $d=301$, \\ $n_i=994$} & \makecell{\dataname{A9A} \\ $d=124$, \\ $n_i=651$} & \makecell{\dataname{PHISHING}\\ $d=69$, \\ $n_i=221$} \\
		\hline
		\hline
		\multicolumn{5}{|c|}{\textbf{Initialization Time (seconds)}} \\
		\hline
		1 & Ray & \multicolumn{3}{c|}{+52.0}    \\
		\hline
		2 & Apache Spark &  \multicolumn{3}{c|}{+25.82}      \\
		\hline
		3 & \algname{FedNL} &  \multicolumn{3}{c|}{+1.1} \\
		\hline
		\multicolumn{5}{|c|}{\textbf{Solving Time (seconds)}} \\
		\hline				
		4 & Ray & 116.17 & 28.13 & 11.54 \\
		\hline
		5 & Apache Spark & 36.65 & 33.59 & 33.14   \\
		\hline	
		6 & \makecell[l]{ \algname{FedNL} / RandK[$k=8d$]} & \cellcolor{bgcolorwe} 12.6 & \cellcolor{bgcolorwe} 4.52 & \cellcolor{bgcolorwe} 0.21   \\
		\hline
		7 & \makecell[l]{\algname{FedNL} / RandSeqK[$k=8d$]} & \cellcolor{bgcolorwe} 12.56 & \cellcolor{bgcolorwe} 5.10  & \cellcolor{bgcolorwe} 0.14 \\
		\hline 
		8 & \makecell[l]{\algname{FedNL} / TopK[$k=8d$]} & \cellcolor{bgcolorwe} 12.20  & \cellcolor{bgcolorwe} 5.79  & \cellcolor{bgcolorwe} 5.23  \\
		\hline 
		9 & \makecell[l]{\algname{FedNL} / TopLEK[$k=8d$]} & \cellcolor{bgcolorwe} 15.11  & \cellcolor{bgcolorwe} 3.26 & \cellcolor{bgcolorwe} 0.82 \\
		\hline
		10 & \makecell[l]{\algname{FedNL} / Natural} & \cellcolor{bgcolorwe} 5.75 & \cellcolor{bgcolorwe} 1.56 & \cellcolor{bgcolorwe} 0.14 \\
		\hline
	\end{tabular}
	\label{ch7:tbl:compare-in-multinode}
\end{threeparttable}
\end{table}



\subsection{Single-node: extension FedNL-LS and CVXPY}
\label{ch7:sec:single-node-cmp-vs-industry}

We implemented \algname{FedNL-LS} which represents the \algname{FedNL} with global convergence guarantees (See pseudocode in Appendix \ref{ch7:app:fednl-ls-descr} or \citet{safaryan2021fednl}) both for simulation and practical distributed usage. In experiments, the line search procedure requires almost always a $1$ step. We have compared \algname{FedNL-LS} against solvers from \libname{CVXPY} \citet{diamond2016cvxpy} which can solve \modelname{logistic regression}. The summary is presented in Table~\ref{ch7:tbl:compare-vs-cvxpy}. Plots are available in  Appendix~\ref{ch7:app:single-node-cmp-vs-industry-extra}. 

We have fine-tuned tolerance for \libname{CVXPY} solvers so that $\norm{\nabla f(x^k)}$ is the same. Experiment shows that our implementation already outperforms solvers from \libname{CVXPY} \citet{diamond2016cvxpy}: \libname{CLARABEL} \citet{clarabel}, \libname{MOSEK} \citet{aps2022mosek}, \libname{SCS} \citet{ocpb:16}, \libname{ECOS} \citet{domahidi2013ecos} by $\times 20$. The solving times in Table~\ref{ch7:tbl:compare-vs-cvxpy} were obtained from \libname{CVXPY}'s internal functions. These times reflect the actual duration used by the corresponding solver, excluding any overhead introduced by \libname{CVXPY}. The initialization time refers to the period required to load all necessary Python extensions into the interpreter and to parse the data using \libname{sklearn.datasets}. As shown for \dataname{A9A} and \dataname{PHISHING} datasets, the {initialization time alone} is $\times 7$ longer than both the initialization and solving times combined for \algname{FedNL-LS} with any compressor. We did not perform a comparison with \libname{GUROBI} \citep{gurobi}. Despite its strength in solving linear and mixed-integer programming problems and accessibility to be used from \libname{CVXPY} and ability to obtain an academic license, \libname{GUROBI} solver has inherent limitations when applied to \modelname{logistic regression} or more complex convex models (that includes intermediate transformation into exponential cone).

Additionally, our experiments indicate that the benefits introduced by \compname{RandSeqK} are already noticeable when  $d$ and $n_i$  are small. In this low-dimensional regime, the internal logic of the compressor and its memory access patterns become significant.

\subsection{Multi-node: FedNL, Apache Spark, and Ray}
\label{ch7:sec:multi-node-cmp-vs-spark}

To the best of our knowledge, the best-known open-source ready-to-use solutions that can be used for distributing training \modelname{logistic regression} are \libname{Apache Spark} \citep{meng2016mllib} and \libname{Ray} \citep{moritz2018ray}. We provide the time required for \algname{FedNL}, \libname{Ray}, and \libname{Apache Spark} to achieve a tolerance of $|\nabla f(x^k)|\approx 10^{-9}$ via configuring final tolerance for solvers. For details about the cluster configuration see Appendix~\ref{ch7:app:software-env}.

From Table~\ref{ch7:tbl:compare-in-multinode}, it is evident that \libname{Ray} and \libname{Apache Spark} necessitate more initialization time. \algname{FedNL} surpasses alternative approaches in this benchmark not only in initialization time but in solving time as well. For extra multi-node experiments with \algname{FedNL}, \algname{FedNL-PP}, \algname{FedNL-LS} see Appendix~\ref{ch7:app:experiments-multi-node}.

\clr
\subsection{Practical limitations of existing FL systems}
\label{ch7:sec:existing-fl-sota-system}

In cross-device Federated Learning (FL), clients typically consist of edge devices such as IoT and mobile devices. The simplest IoT devices may lack an Operating System (OS), operating in a bare-metal environment where feasible implementations can be carried out only in Assembly, C, or C++. Mobile applications, on the other hand, function in a highly competitive and performance-sensitive market. Given the time and energy constraints of many mobile applications, Python is often the least favorable choice for mobile development. For competitive analysis of programming languages, see \citet{pereira2021ranking}. 

Consequently, FL systems that rely solely on Python runtimes, such as \libname{OpenFL}~\citep{reina2021openfl} and \libname{Flute}~\citep{garcia2022flute}, can face significant limitations in real-world applications. Some frameworks, like \libname{FL\_PyTorch} \citep{burlachenko2021fl_pytorch} and \libname{Apple PFL-Research}~\citep{granqvist2024pfl}, explicitly state that their focus is on simulation rather than practical implementation. Interestingly, in response to real-world challenges, frameworks like \libname{FedML} \citep{he2020fedml} and \libname{Flower} \citep{beutel2020flower}, initially designed in Python, are also considering support for non-scripting languages like Java and C++ to enhance their applicability and performance, despite the strong preference for Python in the AI community.

Without additional effort, existing FL frameworks from \citep{he2020fedml}, \citep{beutel2020flower}, \citep{granqvist2024pfl}, \citep{roth2022nvidia}, \citep{garcia2022flute}, and \citep{ro2021fedjax} lack mechanisms for robustly and autonomously training \modelname{logistic regression} models on arbitrary datasets without human intervention. As a result, our work does not compare with these frameworks.

For baselines comparison, we have selected to compete against solvers accessible via \libname{CVXPY} \citep{diamond2016cvxpy}, and solutions such as \libname{Apache Spark} \citep{meng2016mllib}, and \libname{Ray/Scikit-Learn} \citep{moritz2018ray}. Although these frameworks provide effective and robust industrial solutions, they do not fully adhere to FL principles, particularly regarding partial client participation or communication compression. Despite this limitation, we have conducted comparisons with them as they represent well-developed tools.

\section{Conclusions}


Our work represents a significant contribution to advancing the field of FL by addressing a crucial gap between theoretical advancements in \algname{FedNL} optimization algorithm family and their practical implementation. Drawing inspiration from \citet{safaryan2021fednl}, our work is rooted in cutting-edge optimization theory, demonstrating the way of implementing theory into resource-constrained FL settings. Our work serves as a guiding beacon for researchers navigating the intricate path of translating theoretical algorithms into efficient implementations across diverse domains of Machine Learning. Our work emphasizes the multifaceted considerations involved in aiming to improve the actual wall clock time.

Also, our work challenges the predominant Python-centric design philosophy in Machine Learning. It underscores the significance of considering alternative languages when prioritizing computational and memory efficiency.

\clearpage
\appendix

\part*{Appendices to Chapter \ref{chapter7}}
\label{ch7:app:toc_1}
\newpage

\phantomsection
\addcontentsline{toc}{chapter}{Appendices to Chapter 7}
\addtocounter{adjsection}{1}
\section{Practically Implemented FedNL Extensions}
\label{ch7:app:extra-pesudo-code-for-fednl}

{
\begin{algorithm}[H]
	\caption{\algname{FedNL-LS}: {FedNL} with Line Search. Baseline from \citet{safaryan2021fednl} \clrshort{\textbf{[Existing]}.}
	}
	\label{ch7:alg:FedNL-LS}
	\begin{algorithmic}[1]
		\STATE \textbf{Input:} Hessian learning rate $\alpha\ge0$; compression operators $\{\cC_1^k, \dots,\cC_n^k\}$; {line search parameters $c \in (0,\nicefrac{1}{2}]$ and $\gamma \in (0,1)$}
		\STATE \textbf{Initialize:} $x^0\in\R^d$; $\mH_1^0, \dots, \mH_n^0 \in \R^{d\times d}$ and $\mH^0 \eqdef \dfrac{1}{n}\sum_{i=1}^n \mH_i^0$
		\FOR{each device $i = 1, \dots, n$ in parallel} 
		\STATE Get $x^k$ from the server; compute $f_i(x^k)$,\; $\nabla f_i(x^k)$ and $\nabla^2 f_i(x^k)$
		\STATE Send $f_i(x^k)$,\; $\nabla f_i(x^k)$ and $\mS_i^k \eqdef \cC_i^k(\nabla^2 f_i(x^k) - \mH_i^k)$ to the server
		\STATE Update local Hessian shifts $\mH_i^{k+1} = \mH_i^k + \alpha\mS_i^k$
		\ENDFOR
		\STATE On Server
		\STATE \quad Get $f_i(x^k)$,\; $\nabla f_i(x^k)$ and $\mS_i^k$ from all devices $i\in[n]$
		\STATE \quad $f(x^k) = \dfrac{1}{n}\sum_{i=1}^n f_i(x^k), \; \nabla f(x^k) = \dfrac{1}{n}\sum_{i=1}^n \nabla f_i(x^k), \; \mS^k = \dfrac{1}{n}\sum_{i=1}^n \mS_i^k$
		\STATE \quad {Compute search direction $d^k = [\mH^k]_\mu^{-1}\nabla f(x^k)$}
		\STATE \quad {Find the smallest integer $s\ge0$ satisfying $f(x^k+\gamma^sd^k) \le f(x^k) + c\gamma^s \langle \nabla f(x^k), d^k\rangle$}
		\STATE \quad Update global model to $x^{k+1} = x^k + \gamma^sd^k$
		\STATE \quad Update global Hessian shift to $\mH^{k+1} = \mH^k + \alpha\mS^k$
	\end{algorithmic}
\end{algorithm}
}

\subsection{FedNL-LS extension: globalization via line search}
\label{ch7:app:fednl-ls-descr}

The first extension of \algname{FedNL} that we implemented is \algname{FedNL-LS}. We selected an algorithm in which the globalization technique remains independent of any problem-specific parameters. The pseudocode for this algorithm, presented as Algorithm \ref{ch7:alg:FedNL-LS}, has been sourced from \citet{safaryan2021fednl}.

The \algname{FedNL} family includes an extension incorporating a globalization strategy through cubic regularization \algname{FedNL-CR}. However, its practical implementation requires knowledge of problem-dependent $L_*$.

\clearpage

{
\begin{algorithm}[h!]
	\caption{\algname{FedNL-PP}: {FedNL} with Partial Participation. {Baseline from \citet{safaryan2021fednl}} \clrshort{\textbf{[Existing]}}.}
	\label{ch7:alg:FedNL-PP}
	\begin{algorithmic}[1]
		\STATE {\bfseries Input:} Hessian learning rate $\alpha>0$; compression operators $\{\cC_1^k, \dots,\cC_n^k\}$; {number of participating devices $\tau \in \{1,2,\dots,n\}$}
		\STATE {\bfseries Initialize:}
		For all $i\in [n]$: $w^0_i = x^0 \in \R^d$; $\mH_i^0 \in \R^{d\times d}$; $l_i^0 = \|\mH_i^{0} - \nabla^2 f_i(w_i^{0})\|_{\rm F}$; $g_i^0 = (\mH_i^{0} + l_i^{0} \mI)w_i^{0} - \nabla f_i(w_i^{0})$; Moreover: $\mH^0 = \dfrac{1}{n} \sum_{i=1}^n \mH_i^0$; $l^0 = \dfrac{1}{n} \sum_{i=1}^n l_i^0$; $g^0 = \dfrac{1}{n} \sum_{i=1}^n g_i^0$
		\STATE \textbf{on} server
		\STATE ~~~ $x^{k+1} = \left(  \mH^k + l^k\mI  \right)^{-1} g^k$ \hfill { \scriptsize Main step: Update the global model}
		\STATE ~~~ {Choose a subset $S^{k} \subseteq \{1,\dots, n\}$ of devices of cardinality $\tau$, uniformly at random}
		\STATE ~~~ Send $x^{k+1}$ to {the selected devices $i\in S^k$} \hfill { \scriptsize Communicate to selected clients}
		\FOR{each device $i = 1, \dots, n$ in parallel}
		\STATE {{\bf for participating devices} $i \in S^k$ {\bf do} }
		\STATE $w_i^{k+1} = x^{k+1}$ \hfill { \scriptsize Update local model}
		\STATE $\mH_i^{k+1} = \mH_i^k + \alpha \cC_i^k(\nabla^2 f_i(w_i^{k+1}) - \mH_i^k)$ \hfill { \scriptsize Update local Hessian estimate}
		\STATE $l_i^{k+1} = \|\mH_i^{k+1} - \nabla^2 f_i(w_i^{k+1})\|_{\rm F}$ \hfill { \scriptsize Compute local Hessian error}
		\STATE $g_i^{k+1} = (\mH_i^{k+1} + l_i^{k+1} \mI)w_i^{k+1} - \nabla f_i(w_i^{k+1})$ \hfill { \scriptsize Compute Hessian-corrected local gradient}
		\STATE Send $\cC_i^k(\nabla^2 f_i(w_i^{k+1}) - \mH_i^k)$,\; $l_i^{k+1} - l_i^k$ and $g_i^{k+1} - g_i^k$ to server \hfill { \scriptsize Communicate to server}
		\STATE { {\bf for non-participating devices} $i \notin S^k$ {\bf do} }
		\STATE $w_i^{k+1} = w_i^k$, $\mH_i^{k+1} = \mH_i^k$, $l_i^{k+1} = l_i^k$, $g_i^{k+1} = g_i^k$  \hfill { \scriptsize Do nothing}
		\ENDFOR
		
		\STATE On Server
		\STATE ~~~ $g^{k+1} = g^k + \dfrac{1}{n}\sum_{i\in S^k} \left(  g_i^{k+1} - g_i^k  \right)$  \hfill { \scriptsize Maintain the relationship $g^k = \dfrac{1}{n} \sum_{i=1}^n g_i^k$}
		\STATE ~~~ $\mH^{k+1} = \mH^k + \dfrac{\alpha}{n}\sum_{i\in S^k} \cC_i^k(\nabla^2 f_i(w_i^{k+1}) - \mH_i^k)$   \hfill { \scriptsize Update  the Hessian estimate on the server}
		
		\STATE ~~~ $l^{k+1} = l^k + \dfrac{1}{n}\sum_{i\in S^k} \left(  l_i^{k+1} - l_i^k  \right)$ \hfill { \scriptsize Maintain the relationship $l^k = \dfrac{1}{n} \sum_{i=1}^n l_i^k$}
	\end{algorithmic}
\end{algorithm}
}

\subsection{FedNL-PP extension: FedNL with clients partial participation}
\label{ch7:app:fednl-pp-descr}

The subsequent extension to \algname{FedNL}, which we incorporate into our practical implementation, is \algname{FedNL-PP} from the original \algname{FedNL} paper by \citet{safaryan2021fednl} (Algorithm \ref{ch7:alg:FedNL-PP}). This extension enables the handling of the situation, where only randomly selected clients participate in each iteration.





\clearpage
\addtocounter{adjsection}{1}
\section{Progression of Practical Enhancements: Chronological Overview}
\label{ch7:app:history-of-improvements}

There is an inherent complexity involved in concurrently addressing three cornerstones:

\begin{enumerate}

\item \textit{{Theoretical algorithm.}} Serves as a foundational framework.

\item \textit{{Modern computation systems.}} Serving for efficient and reliable computation.

\item \textit{{Best engineering practices.}} Serves as a bridge between the previous two.
\end{enumerate}

In our work, we transition a well-developed theoretical algorithm analyzed with worst-case guarantees, exploring possible practical improvement gains. We adhered to a set of decisions in creating a practical implementation, taking steps that do not violate theory at all on one hand, but on the other side, they target to make \algname{FedNL} practical.

In refining \algname{FedNL}, our focus was on addressing $L_2$ regularized \modelname{logistic regression} as a proxy, which is used for measuring improvements. The LIBSVM \dataname{W8A} dataset was selected for experimentation as a proxy to drive progress. Each sample in the \dataname{W8A} dataset was augmented with an artificial feature set to $1$ to introduce an intercept term. Post-augmentation, our dataset \dataname{W8A} has $d=301$ features, and encompasses $49\,749$ samples. Uniform shuffling and equitable distribution across $n=142$ clients were ensured, with each receiving a share of $350$ sample data points, while the remaining $49$ samples were excluded. The initial value of the optimization variable $x$ is initialized at zero. The regularization coefficient is $\lambda=0.001$. In our instance, the strong convexity parameter is $\mu_f \ge 0.001$ and smoothness contact is  $L_f \le 0.0058$. This lead to condition number $\nicefrac{L_f}{\mu_f} \le 5.8$. We executed $1000$ rounds.

\paragraph{Experiments setup.} All our experiments have been carried out under the Assumption that the preliminary step from Appendix~\ref{ch7:app:careful-reproduce} was done, and next, the launching has been done $4$ times with a minimum time from $4$ launches. The employed hardware for single-node multi-core simulation is described in Appendix~\ref{ch7:app:hardware-env-single-node}. The start iterate is $x^0=0$.  Our \algname{FedNL} implementation contains function value evaluation $f(x)$ that is optionally tracked and computed via obtaining this information from clients. Norm of a full gradient in last iterate $\| \nabla f(x^k)\| \approx 3 \cdot 10^{-18}$ for \algname{FedNL} with \compname{RandK[$k=8d$]} compressor, and $\| \nabla f(x^k)\| \approx 3.88 \cdot 10^{-18}$ for \algname{FedNL} with \compname{TopK[$k=8d$]} compressor. 

\paragraph{Improvements history.} The detailed history of improvements presented in the Table~\ref{ch7:tab:improvements-log} below. The total speedup of our implementation in a single-node multi-core machine {v63} compared to baseline {v0} is {$\times 951.88$} for \compname{RandK[$k=8d$]} and {$\times 1069$} for \compname{TopK[$k=8d$]}.

\iftrue
{
\footnotesize

\begin{longtable}{|p{0.4\textwidth}|p{0.10\textwidth}|p{0.13\textwidth}|p{0.10\textwidth}|p{0.13\textwidth}|}	
	\caption{Improvemens for \compname{TopK[K=8d]} and  \compname{RandK[K=8d]}, $d=301,n_i=350,n=142$, \dataname{W8A} dataset.}
	\label{ch7:tab:improvements-log} \\
	\cline{1-5}
	\textbf{Improvement Step} & \textbf{TopK. Time (sec.)} & \textbf{TopK. Relative speedup $t_{i+1}/t_{i}$} & \textbf{RandK. Time (sec.)} & \textbf{RandK. Relative speedup $t_{i+1}/t_{i}$} \\
	\cline{1-5}
	\endhead
	\cline{1-5}
	\endfoot
	\cline{1-5}
	v63. Compiler and Language relative improvements. & 18.488 & 1.002 & 18.395 & 1.002 \\
	\cline{1-5}		
	v62. Improvements inside pseudo-random generators and Knuth shuffle implementation. & 18.525 & 1.0105 & 18.438 & 1.0218 \\	
	\cline{1-5}
	v61. Optimize data alignment in dense linear algebra operations by transitioning to AVX-512 vectorization from AVX2(256 bits). & 18.72 & 1.3633 & 18.84 & 1.3305 \\	
	\cline{1-5}
	v60. Compiler improvements for local \algname{FedNL} implementation. & 25.521 & 1.0018 & 25.067 & 0.9954 \\
	\cline{1-5}		
	v59. Bit tricks for computing residual from division by a number which is a power of two. & 25.569 & 1.00406 & 24.953 & 1.021239 \\
	\cline{1-5}
	v58. Use more aligned load/store in Hessian oracle. & 25.673 & 1.016 & 25.483 & 1.019 \\ 		
	\cline{1-5}
	v57. Use aligned load/store for dense vector operations. & 26.104 & 1.10423 & 25.787 & 1.0895 \\
	\cline{1-5}
	v56. Eliminating the need for information in memory buffers, and storing a vector of labels explicitly. & 28.825 & 1.000173 & 28.097 & 1.03185 \\		
	\cline{1-5}
	v55. Precompute indices without rechecking in a hot-loop for \compname{RandK} and other minor improvements. & 29.098 & 1.014 & 28.992 & 1.0159 \\		
	\cline{1-5}		
	v54. Eliminate division during memory allocation from constructed memory pools. & 29.542 & 1.0002 & 29.455 & 1.0008 \\		
	\cline{1-5}		
	v53. Eliminate the storage of design matrix $\mA_i^\top$. Add need methods for matrix-vector multiplication with $\mA_i^\top$. & 29.549 & 1.13144 & 29.480 & 1.12869 \\
	\cline{1-5}
	v52. Additional internal vectorization across samples with reducing stores during Hessian evaluation. & 33.433 & 1.6303 & 33.274 & 1.6267 \\
	\cline{1-5}
	v51. Use symmetry during evaluating $\norm{.}_{F}$. & 54.506 & 1.00066 & 54.128 & 1.00751 \\
	\cline{1-5}
	v50. Unroll the first iteration in Hessian Oracle. & 54.542 & 1.0489 & 54.535 & 1.0351 \\
	\cline{1-5}
	v49. More space-efficient method for \compname{TopK} based on min-heaps. & 57.211 & 1.012 & 56.451 & 1 \\
	\cline{1-5}
	v48. Optimized version for generating sequences with SSE/AVX vectorization. & 57.940 & 1.014 & 56.451 & 1.006 \\
	\cline{1-5}
	v47. More Systems Optimization. Use memory-mapped files for parsing input data. & 58.788 & 1.023 & 56.802 & 1.022 \\		
	\cline{1-5}
	v46. Yield client execution while waiting for a signal from the server instead of a complete busy loop. & 60.198 & 1.0057 & 58.102 & 1.0005 \\
	\cline{1-5}
	v45. Use light vector for gradient aggregations and minor improvement of Cholesky Factorization. & 60.544 & 1.0016 & 58.393 & 1.0029 \\
	\cline{1-5}
	v44. Compute and send $L_k$ once the difference that defines $L_k$ is available. This quantity is in a critical path. & 60.644 & 1.003 & 58.567 & 1.013 \\
	\cline{1-5}		
	v43. Compute indices for the \compname{RandK} while waiting for the master. Add a RandSeqK compressor (which works $58.4$ sec.). & 60.826 & 1.0 & 59.358 & 1.0211 \\
	\cline{1-5}		
	v42. Remove not needed conditions in matrix-vector and matrix-matrix operations. Create extra conversion methods to obtain a flattened index in the matrix. Fused operation for matrix-vector operation and add multiple of vector. & 60.826 & 1.0203 & 60.618 & 1.0342 \\
	\cline{1-5}
	v41. Sorting indices for \compname{RandK} and \compname{TopK} to make computation more cache-friendly. & 62.063 & 1.0182 & 62.697 & 1.01033 \\
	\cline{1-5}
	v40. Force inlining for compression mechanisms. & 63.193 & 1.007 & 63.345 & 1.0005 \\
	\cline{1-5}
	v39. Switch from creating from the number of userspace threads equal to the number of clients. Create a pool of workers equal to the number of physical cores available in the system (in our case it is $12$). Process several clients or work items within one thread. Also, eliminate extra copying of current iterate $x_i$ inside workers. Process messages from the clients once they are available. & 63.647 & 1.404 & 63.379 & 1.819 \\
	\cline{1-5}
	v38. Improvement in parsing and analyzing input files in LIBSVM dataset format. Elimination of creating temporary strings with low-level primitives from C++17/20. & 89.368 & 1.0835 & 115.340 & 1.0533 \\
	\cline{1-5}
	v37. Improved \compname{TopK} implementation. Through various approaches - quicksort, merge sort, Multi-way merge sort, CO Funnelsort, and radix sort, the fastest approach is based on a D-way heap to support K smallest items found so far. & 96.833 & 1.0412 & 121.490 & $\approx$ 1.0 \\
	\cline{1-5}
	v36. Applying it to sparse updates of the dense matrix in the server. Cache-friendly implementation and usage of byte buffers by clients used to prepare information to send to the master. & 100.829 & 1.40216 & 121.499 & 1.44588 \\
	\cline{1-5}
	v35. Custom memory pools for memory allocations for dense vectors and matrices. The memory pool implementation does not allow moving vectors from one computation CPU thread to another. However, this makes memory allocation very fast. & 141.379 & 1.0419 & 175.673 & 1.0109 \\		
	\cline{1-5}
	v34. Store information about the number of columns in a dense matrix explicitly without recomputing it every time it is needed. & 150.207 & 1.0052 & 177.601 & 1.0421 \\
	\cline{1-5}
	v33. Add vectorized implementation for light dense vector (light read-only view of a dense vector) for $L_2$ norm and inner product. Useful in Backward/Forward substitution and Cholesky factorization implementation in Master. & 150.998 & 1.0689 & 185.093 & 1.021 \\
	\cline{1-5}
	v32. Manually unroll loops for vector and vector scalar operations in dense vector vectorized implementation.  & 161.410 & 1.034 & 189.075 & 1.000 \\
	\cline{1-5}
	v31. Compute and store indices for the upper triangular part of the matrix with shape [d,d] only once.  Reuse these indices in all rounds without recomputing them for \compname{TopK} and \compname{RandK}. & 166.899 & 1.045 & 189.134 & 1.0165 \\
	\cline{1-5}
	v30. Cache-friendly Cholesky Factorization implementation which produces both L and L transpose factors, that are subsequently used cache-friendly forward/backward substitution. & 174.495 & 1.034 & 192.271 & 0.986 \\
	\cline{1-5}
	v29. Turn on the whole program optimization with the linker. & 180.588 & 1.047 & 189.612 & 1.034 \\
	\cline{1-5}
	v28. Turn off support of exception and runtime type information from the C++ compiler. & 189.216 & 1.07 & 196.108 & 1.04 \\
	\cline{1-5}
	v27. Extra vectorization in CPU implementation of Hessian oracle. & 204.315 & 1.02 & 204.219 & 1.001 \\
	\cline{1-5} 
	v26. An alternative implementation of \modelname{logistic regression} Hessian Oracle. Evaluate as the sum of symmetric rank-1 matrices. Compute only the upper diagonal part and symmetrize once. & $208.937$ & $1.603$ & $204.440$ & $1.805$ \\
	\cline{1-5} 
	v25. Improve diagonal matrix multiplication by dense. Improve Matrix-Matrix multiplication - tune tile sizes. & $334.992$ & $1.089$ & $369.037$ & $1.025$ \\
	\cline{1-5} 
	v24. Removing one division during conversion from flat matrix index into matrix row/column index.  Use uninitialized buffers inside Hessian/function/gradient oracles. & $364.886$ & $1.14$ & $378.563$ & $1.26$ \\
	\cline{1-5}
	v21. Compute the exponent of classification margins only once during computing Hessian/gradient oracles. (I.e. compute $\exp(x^\top a_i \cdot b_i)$ only once and reuse them in all oracle types). & $406.550$ & $1.0004$ & $477.912$ & $1.04$ \\
	\cline{1-5}
	v20. More effective initialization of design matrix (by columns). Remove double default initialization of dense matrix. & $406.730$ & $1.057$ & $497.912$ & $1.02$ \\
	\cline{1-5}
	v17. Compute the classification margin once ($x^\top \cdot b_{ij} a_{ij}$) and reuse this quantity in the gradient and Hessian oracle. & $430.058$ & $1.458$ & $511.612$ & $1.44$ \\
	\cline{1-5}
	v16. Explicitly call size() functions in all places of the source code out of the loops and use it once without recalling the function again. & $627.172$ & $1.08$ & $737.735$ & $1.08$ \\
	\cline{1-5}
	v15. Move $l_i$ information into the separate data stream for sending information asynchronously to the master.  & $681.874$ & $1.01$ & $798.195$ & $1.03$ \\
	\cline{1-5}
	v14. Custom implementation of adding elements to the diagonal. & $689.630$ & $1.06$ & $824.973$ & $1.019$ \\
	\cline{1-5}
	v12. Shuffle the array in place instead of shuffling a separate array. Special copy operations for big and large objects of float/doubles. & $734.119$ & $1.005$ & $841.172$ & $1.073$ \\
	\cline{1-5}
	v10. Cholesky decomposition with forward/backward substitution instead of Gauss Elimination during linear solve. & $737.92$ & $1.146$ & $903.392$ & $1.196$ \\
	\cline{1-5}	
	v09. Sorting-based implementation for \compname{TopK} instead of using D-way heaps. Create a separate dedicated stream of sending information with information about gradients. & $845.982$ & $1.008$ & $1080.820$ & $1.012$ \\
	\cline{1-5}
	v7. Extra constant and no exception qualifiers into various places - make code nicer for compiler optimization. Simplify dataset representation. Remove information about the presented sample from the dataset. Add tile matrix multiplication instead of baseline. & $853.032$ & $1.016$ & $ 1093.900$ & $1.011$ \\
	\cline{1-5}
	v1. Baseline. Usual C++ implementation of the algorithm. & $867.227$ & $22.8027$ & $1106.52$ & $15.8318$ \\
	\cline{1-5}
	v0. Baseline implementation in Python/NumPy  \href{https://github.com/Rustem-Islamov/FedNL-Public}{https://github.com/Rustem-Islamov/FedNL-Public}. & $19770.0$ & $1$ & $17510.0$ & $1$ \\	
	\cline{1-5}
\end{longtable}	
}
\fi

\clearpage
\addtocounter{adjsection}{1}
\section{RandSeqK: A Practical  Cache-Aware Improvement of RandK}
\label{ch7:app:seqk}

\subsection{Background about RandK compressor}
\label{ch7:app:randk}

The sparsification of the input matrix $\bM \in \mathbb{R}^{d \times d}$ with \compname{RandK} happens by selecting $k$ items from possible $n=d(d+1)/2$ elements from the upper triangular part of the input matrix $\bM$ and scaling the result in a way to preserve unbiasedness. The distribution employed for selecting this subset is uniform across all possible subsets of cardinality $k$. The specific subset is selected with probability $\left(\dfrac{n!}{k! (n-k)!}\right)^{-1}$. The probability that a specific single element from $\bM$ will be included in the selection equals to:

\begin{eqnarray*}
p &=& \left(\dfrac{(n-1)!}{(k-1)!((n-1)-(k-1))!}\right)/
\left(\dfrac{n!}{k!(n-k)!}\right) \\
&=& \left(\dfrac{(n-1)!}{(k-1)!(n-k)!}\right)/
\left(\dfrac{n!}{k!(n-k)!}\right) \\
&=& \left(\dfrac{(n-1)!k}{k!(n-k)!}\right)/
\left(\dfrac{n!}{k!(n-k)!}\right) \\
&=& \dfrac{(n-1)!k}{n!} = \dfrac{k}{n}
\end{eqnarray*}

This implies that specific items will be selected during sparsification can be described as indicator random variables which take $1$ with probability $\nicefrac{k}{n}$, and $0$ with probability $1-\nicefrac{k}{n}$.

\paragraph{RandK as a collection of Bernoulli Random Variables.} The \compname{RandK}($\bM$) compressor applied for symmetric matrices from $\mathbb{R}^{d \times d}$ can be observed as an operator that selects elements from the upper triangular part and scales the results by scalar constant $C$. Let's use notation whereby with $e_{ij} \in \mathbb{R}^{d \times d}$ we denoted the matrix from $\mathbb{R}^{d \times d}$ with the only non-zero element in positions $(i,j)$ and $(j,i)$ equal to $1$.
Next we consider the collection of indicator random variables $Z_{ij} \in \{0,1\}; i,j \in [d]$. If $\delta \eqdef \Exp{Z_{mn}}, \forall m, n \in [d]$ we proceed as follows:
\begin{eqnarray*}	
\Exp{\mathrm{RandK}(\bM)} &=& \Exp{C \sum_{i=1}^d \sum_{j=i}^d  e_{ij} Z_{ij} m_{ij}} \\
&=&\sum_{i=1}^d \sum_{j=i}^d C \Exp{Z_{ij}} e_{ij} m_{ij} = C \cdot \delta \sum_{i=1}^d \sum_{j=i}^d e_{ij} m_{ij} = C \cdot \delta \cdot \bM.
\end{eqnarray*}

To ensure unbiased compression, there is only one choice for $C=\dfrac{1}{\delta}$.

\textbf{Observation 1:} \textit{Although we utilized the fact that 
$\Exp{Z_{ij}}$ does not vary across matrix elements, during the derivation, we did not explicitly rely on any specific dependence between $Z_{ij}$ random variables.
}

Next, the derivations for the theory of \compname{RandK} for bounding variance can be simplified by observing that ${Z_{ij}^2}={Z_{ij}}$. We proceed as follows:

\begin{eqnarray*}	
\Exp{\|\mathrm{RandK}(\bM)-\bM\|_F^2} &=& 
\Exp{\sum_{i=1}^d \sum_{j=1}^d \left( [\mathrm{RandK}(\bM)]_{ij} - m_{ij} \right)^2} \\
&=& 
\Exp{\sum_{i=1}^d \left( C\cdot Z_{ii}m_{ii} - m_{ii} \right)^2} \\
&& \qquad +
2 \Exp{\sum_{i=1}^d \sum_{j=i+1}^d \left( C\cdot Z_{ij}m_{ij} - m_{ij} \right)^2} \\
& = & \Exp{\sum_{i=1}^d \left( C\cdot Z_{ii} - 1 \right)^2 m_{ii}^2} \\
&& \qquad + 2 \Exp{\sum_{i=1}^d \sum_{j=i+1}^d \left(C\cdot Z_{ij} - 1\right)^2 m_{ij}^2} \\
& = & {\sum_{i=1}^d \Exp{(C^2\cdot Z_{ii}^2 - 2C Z_{ii} + 1)} m_{ii}^2} \\
&& \qquad +
2{\sum_{i=1}^d \sum_{j=1}^d \Exp{(C^2\cdot Z_{ij}^2 - 2C Z_{ij} + 1)} m_{ij}^2}
\\
& = & {\sum_{i=1}^d \Exp{(C^2\cdot Z_{ii}^2 - 1)} m_{ii}^2} \\
&& \qquad + 
2{\sum_{i=1}^d \sum_{j=i+1}^d \Exp{(C^2\cdot Z_{ij}^2 - 1)} m_{ij}^2} \\
& = & {\sum_{i=1}^d \Exp{(C^2\cdot Z_{ii} - 1)} m_{ii}^2} \\
&& \qquad + 
2{\sum_{i=1}^d \sum_{j=i+1}^d \Exp{(C^2\cdot Z_{ij} - 1)} m_{ij}^2} \\	 
&=& (C - 1) \|\bM\|_F^2.	
\end{eqnarray*}

Therefore for \compname{RandK} compressor $w=\dfrac{1}{\delta}-1=C-1$. 

\textbf{Observation 2:} \textit{During derivation to bound variance 
we utilized the fact that $\Exp{Z_{ij}}$ does not vary across matrix elements, however, we did not explicitly rely on any specific dependence between $Z_{ij}$ random variables.
}

\subsection{Looking for degrees of freedom in RandK compressor}
\label{ch7:app:dof-in-randk}

The theoretical derivation for \compname{RandK} does not prescribe the joint distribution of $Z_{ij}, i \le j$, therefore analysis of \compname{RandK} is more general and can be applied to other compression algorithms. The only requirement to maintain previous mathematical analysis is that $\Exp{Z_{ij}}$ is constant equal to $\delta$. This will imply automatically that $w=1/\delta - 1$ for a compressor that specifies input leaving non-zero (and scale by $C$) items from the upper triangular part of input matrix $\bM$ specified by $Z_{ij}$.

\subsection{Compression schema: cache-aware RandSeqK}
The proposed next modification of the \compname{RandK} compressor introduces a sampling strategy that ensures a cache-aware memory access pattern during the compression of a matrix $M \in \mathbb{R}^{d \times d}$ arranged densely in column-wise order. If we order all elements from the upper triangular part into the sequence $E$, its size becomes $w={d(d+1)}/{2}$. The sampling strategy begins by randomly selecting a start index $s \sim_{u.a.r.} E$. The subsequent set of indices is created deterministically:

$$I=\{s, (s+1)\mod w, \dots, (s+k-1) \mod w \}$$

The \compname{RandSeqK} compressor is then applied and utilizes the same logic of \compname{RandK}, with a difference that the selection of $Z_{ij}$ is modified in the following sense $Z_{ij}=1 \iff (i,j) \in I \subseteq E$.

The number of indices in set $I$ is $k$, implying that exactly $k$ indicator random variables $Z_{ij}$ will be equal to $1$. These random variables are sampled in a dependent manner. The total number of groups from which sampling occurs is $w$, and each group consists of:
$$\{i \mod w, (i+1) \mod w, \dots, (i+k-1)\mod w\}.$$

For each element in the original sequence $e$, there are $k$ groups with sequential indices that cover this element. The probability of selecting a specific element (in other words, $Z_{ij}=1$) from the sampling schema where the start index $s$ is initially sampled, and then the next $k-1$ are committed to, is equal to $k/n$. It does not depend on the indices $(i,j)$, allowing to reuse theory from Appendix~\ref{ch7:app:randk}.

\subsection{Advantages of RandSeqK over RandK}

First, the \compname{RandSeqK} requires only $1$ pseudo-random generator (PRG) call, while \compname{RandK} demands $k$ calls.

Second, let's consider the scenarios where the input matrix is located in DRAM memory. The \compname{RandK} compressors necessitate fetching $k$ items, each comprising $b$ bytes ($b=8$ for FP64). For \compname{RandK}, this results in a worst-case scenario of $k$ DRAM memory transactions. The \compname{RandSeqK} faces a worst-case scenario of ${kb}/{L}+2$ memory transactions, where $L$ denotes the cache line size. In contemporary systems, the cache line size is typically $64$ bytes. For FP64, \compname{RandSeqK} markedly amplifies memory access efficiency in both clients (Algorithm~\ref{ch7:alg:FedNL}, Lines $5$ and $6$) and the master (Algorithm~\ref{ch7:alg:FedNL}, Line $10$) by a factor of $L/b=8$. This improvement may seem negligible, but if considering data fetching from DRAM memory based on Table~\ref{ch7:tbl:latencies-for-memory} from Appendix \ref	{ch7:app:memory-hierachy-latencies}, the scaled improvement becomes $\times 2640$ if making a comparison with floating-point arithmetics. Finally, contemporary ARM and x86-64 processors often integrate hardware components for data prefetching, enhancing memory access latency from DRAM. Specific details about the number and types of prefetchers in CPUs are under Non-Disclosure Agreements. 

Therefore, while \compname{RandK} selects items for compression uniformly at random, \compname{RandSeqK} exhibits a more structured and predictable access pattern, which is more favorable for prefetchers. This structured access pattern contributes to improved efficiency in memory operations, as it helps reduce cache misses and enhances data locality, ultimately leading to better performance during compression.

\clearpage
\addtocounter{adjsection}{1}
\section{TopLEK: A Randomized and Adaptive Improvement over TopK}
\label{ch7:app:toplek}

\subsection{Background about TopK compressor}

To simplify exposition, let's initially focus on the class of contractive compressors to vectors:

\begin{equation}
\label{ch7:def:biased_compressors}
\{\mathcal{C} \in \mathbb{R}^d \to \mathbb{R}^d : \mathbb{E}\left[ \|\mathcal{C}(x) - x\|^2\right]\} \le (1-\alpha)\|x\|^2, \forall x \in \mathbb{R}^d, \alpha \in (0,1]\}.
\end{equation}

An important example of a compress operator that satisfies this property is the \compname{TopK} compress operator. This operator preserves the $k$ largest (in absolute value) entries of the input, zeroing out the rest. In this case, the compression algorithm is deterministic (if breaks ties in the same way).

The deterministic \compname{TopK} compressor satisfies condition in Equation~\eqref{ch7:def:biased_compressors} deterministically: $$\|\mathrm{TopK}(x) - x\|^2\ \le (1-\alpha)\|x\|^2, \forall x \in \mathbb{R}^d.$$

Firstly, we observe that \compname{TopK} is homogeneous in $\beta \in \mathbb{R}, \beta \ne 0$. For $\forall x \in \RD$:
\begin{eqnarray*}
&& \|\mathrm{TopK}(x) - x\|^2\ \le (1-\alpha)\|x\|^2 \\
&\iff& \|\mathrm{TopK}(\beta x) - \beta x\|^2\ \le (1-\alpha)\|\beta x\|^2 \\
&\iff& \|\mathrm{TopK}(\beta x) / \beta - x\|^2\ \le (1-\alpha)\|x\|^2 \\
&\iff& \|\mathrm{TopK}(x) - x\|^2\ \le (1-\alpha)\|x\|^2
\end{eqnarray*}

\paragraph{Comments.} The \compname{TopK} compressor returns the $k$ largest values in absolute value. Scaling by any nonzero number does not change the result of the compress operator (if ties are broken in the same way). After returning the result, $\mathrm{TopK}(\beta x) / \beta$ correctly reconstitutes the sign. Thus, we can conclude that \compname{TopK} is a homogeneous.

\subsection{Pessimism of TopK analysis} If $x \in \RD \backslash \{0\}^d$ and selected indices for \compname{TopK} compressor is the set $S(x) \in 2^{[d]}$ then we have:

\begin{eqnarray*}
	\lefteqn{\|\mathrm{TopK}(x) - x\|^2 = \|x-\mathrm{TopK}(x)\|^2 \le (1-\alpha)\|x\|^2, \quad \forall x \in \mathbb{R}^d} \\
	&\iff& \sum_{i=1}^{d} x_i^2 - \sum_{j \in S(x)} x_{j}^2 \le (1-\alpha) \sum_{i=1}^{d} x_i^2, \quad \forall x \in \mathbb{R}^d \\
	&\iff& 1 - \dfrac{\sum_{j \in S(x)} x_{j}^2}{\sum_{i=1}^{d} x_i^2} \le (1-\alpha), \quad \forall x \in \mathbb{R}^d, x \ne 0 \\
	&\iff& \alpha \le \dfrac{\sum_{j \in S(x)} x_{j}^2}{\sum_{i=1}^{d} x_i^2}, \quad \forall x \in \mathbb{R}^d, x \ne 0
\end{eqnarray*}


With utilizing the set $S(x)$ and $S'(x) \eqdef [d] \setminus S(x)$ the optimal $\alpha_{\rm{opt}}$ can be expressed as:

\begin{eqnarray}
	\label{ch7:eq:alpha-explicitly-S}
	\alpha_{\rm{opt}} &=& {\min}_{x \in \mathbb{R}^d} \left(\sum_{j \in S(x)} x_{j}^2 \right), \, \text{subject to:} \, \|x\|_2^2 = 1
\end{eqnarray}

\paragraph{Claim.} The minimizer of this optimization problem is attained when:

\[
\forall i,j \in [d]: x_i = x_j = c
\]

\paragraph{Proof of the claim.}
First, observe that in Problem~\eqref{ch7:eq:alpha-explicitly-S}, the value of \( \alpha(x) = \sum_{j \in S(x)} x_j^2 \) is invariant under the permutation of the elements \( x_i \) in the input vector \( x \in \mathbb{R}^d \). Therefore, without loss of generality, we can assume that the components of \( x \) are indexed such that their magnitudes follow a non-increasing order: \( |x_1| \ge |x_2| \ge \dots \ge |x_d| \). This indexing allows us to represent \( S(x) = \{1, 2, \dots, k\} \). Furthermore, for simplicity, we will consider only the case where \( x_i \ge 0 \), since \( \alpha(x) \) is not sensitive to the signs of \( x_i \).

Assume, for the sake of contradiction, that \( \alpha_{\rm{opt}} \) is attained at a point \( y \) where the elements of \( y \) are not equal (and $\|y\|^2=1$). Thus, by the ordering of components, it means \( y_1 > y_d \). In this case, \( y_1 \) contributes significantly to \( \alpha(y) \), while \( y_d \) for \( k \neq d \) does not. However, we can show that \( \alpha(y) \) can be further reduced by perturbing the components of \( y \) to form a new vector \( z \in \mathbb{R}^d \).

Let us define \( z \) with using two free parameters $\tau_1, \tau_d \in \mathbb{R}$ as follows:
\begin{eqnarray*}
	z_1 &\eqdef& y_1 + \tau_1,\\
	z_d &\eqdef& y_d + \tau_d,\\
	z_i &\eqdef& y_i, \forall i \in [d], i \neq 1, i \neq d.
\end{eqnarray*}

To maintain the constraint \( \sum_{i=1}^{d} z_i^2 = \sum_{i=1}^{d} y_i^2 = 1\), we need to ensure that the new vector \( z \) satisfies this condition. This leads to the equation:

\[
\sum_{i=1}^{d} z_i^2 = \sum_{i=1}^{d} y_i^2
\]

Expanding both sides and simplifies:

\begin{eqnarray*}
	&& (y_1 + \tau_1)^2 + (y_d + \tau_d)^2 = y_1^2 + y_d^2 \\
	&\iff& 2y_1 \tau_1 + \tau_1^2 + 2y_d \tau_d + \tau_d^2 = 0 \\
	&\iff& \tau_1^2 + 2y_1 \tau_1 + (2y_d \tau_d + \tau_d^2) = 0
\end{eqnarray*}

We now solve for \( \tau_1 \) and select one root from the two:
\[
\tau_1 = \frac{-2y_1 + \sqrt{4y_1^2 - 4(2y_d \tau_d + \tau_d^2)}}{2}=-y_1 + y_1\sqrt{1 - (2y_d \tau_d + \tau_d^2)/y_1^2}
\]

Therefore, for sufficiently small $ \tau_d > 0 $ and any $ y_1 > 0 $, we can select a sufficiently small $ \tau_1 < 0 $, because $ \tau_1 \in (-y_1, 0) $, and $ \tau_1 \to 0 $ as $ \tau_d \to 0 $. Choosing $ \tau_1 $ small enough such that $ y_1 + \tau_1 \geq y_{k+1} $ ensures that $ y_1 $ remains among the top $ k $ components. Similarly, choosing $ \tau_d $ small enough such that $ y_d + \tau_d \leq y_k $ ensures that $ y_d $ remains in the smallest $ d-k $ components.

However, this implies that $ \alpha(z) < \alpha(y) $, which contradicts the assumption that $ y $ was the minimizer. Therefore, the assumption that $ y $ is the minimizer must be false. Consequently, the optimal minimizer $ \alpha_{\rm{opt}} $ is achieved when all components of $ x $ are equal, i.e., $ x_1 = x_2 = \dots = x_d = c $. \qed

\paragraph{Consequences of optimal solution property.} The fact that optimal solution for Problem~\eqref{ch7:eq:alpha-explicitly-S} implies that for the entire space \( \mathbb{R}^d \), the minimum value of \( \alpha_{\rm{opt}} \) is attained only along the diagonal of \( \mathbb{R}^d \), where \( x_1 = x_2 = \dots = x_d = c \). In practical computer implementations, \( \mathbb{R}^d \) is represented as a set of finite values. If the set is discretized to \( M \) values, the diagonal elements in this finite representation are \( M \), and the number of non-diagonal elements is \( M^d - M \). The diagonal is only a small subset of the implementable elements of \( \mathbb{R}^d \). 

\begin{figure}
	\centering
	
	\begin{subfigure}[ht]{0.49\textwidth}
		\includegraphics[width=0.94\textwidth]{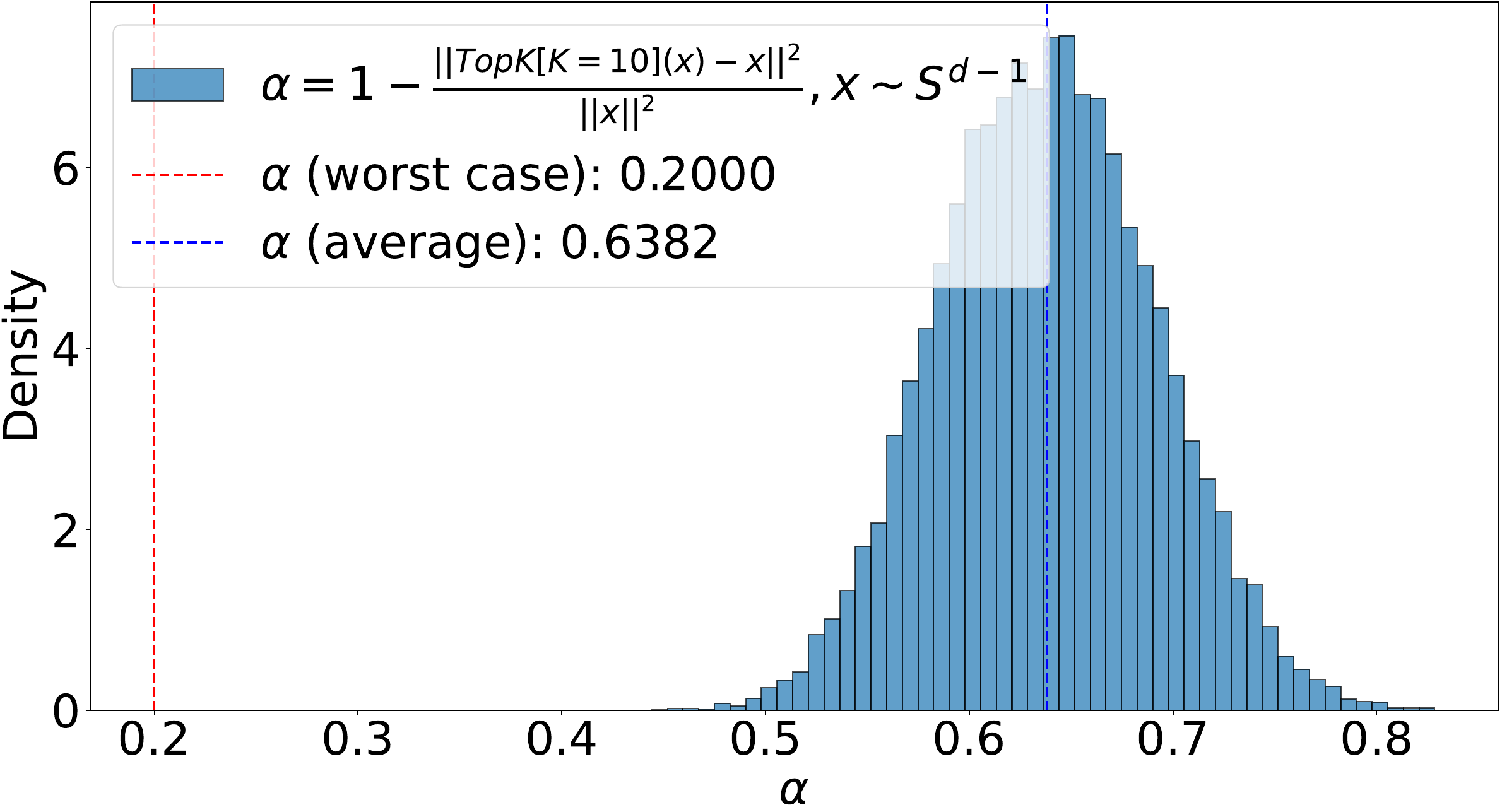}
		\caption{$d$=50, \compname{TopK} with $k=10$.}
	\end{subfigure}
	\begin{subfigure}[ht]{0.49\textwidth}
		\includegraphics[width=\textwidth]{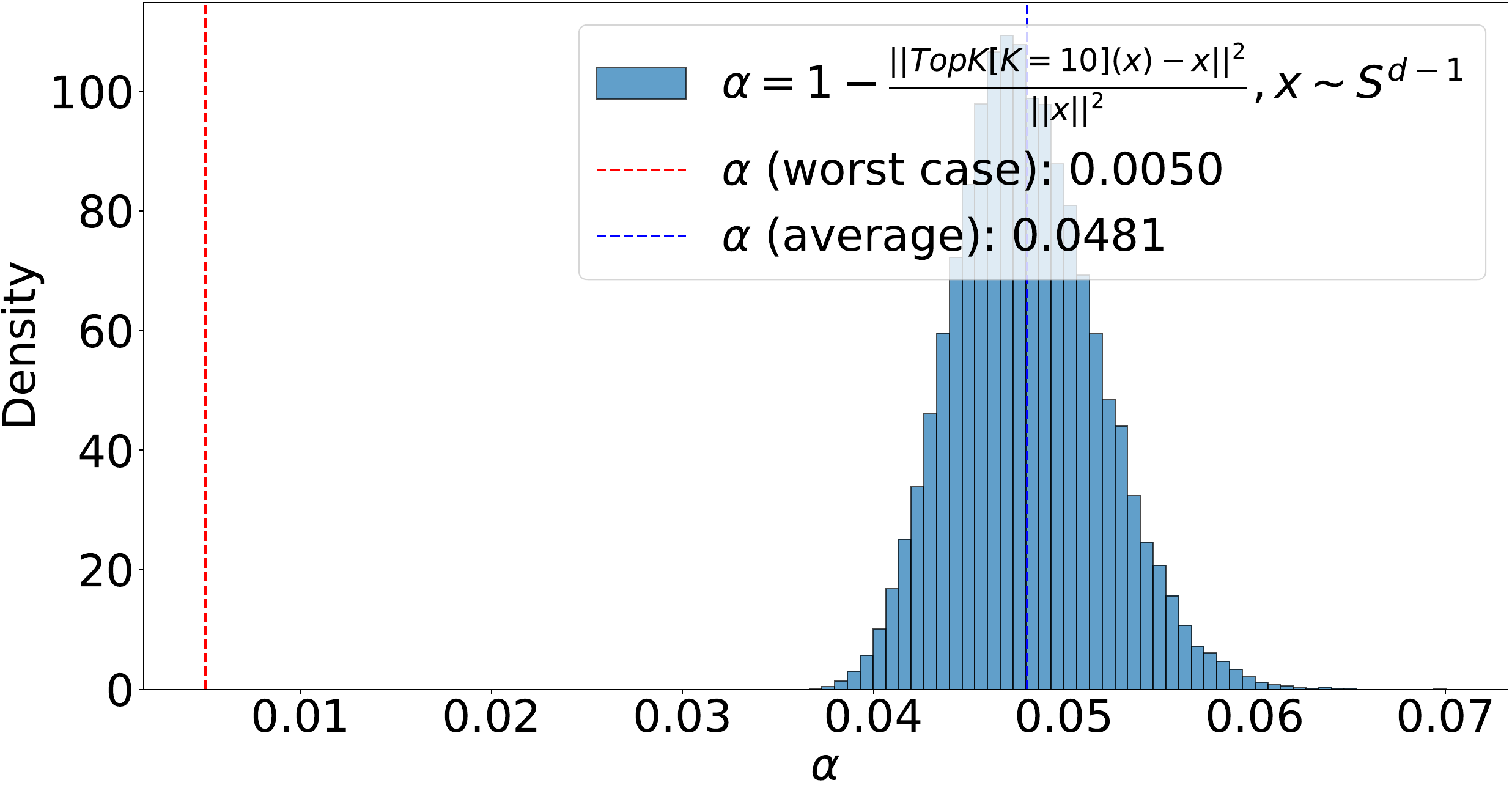}
		\caption{$d$=2000, \compname{TopK} with $k=10$.}
	\end{subfigure}	
	\caption{Discrepancy between worst-case $\alpha$ and $\alpha(x)$ when $x \sim_{\rm u.a.r.} S^{d-1}$. Number of trials $20\,000$.}
	\label{ch7:fig:alphawc}
\end{figure}

Therefore, while the worst-case $\alpha$, as defined in Equation~\ref{ch7:def:biased_compressors}, for the \compname{TopK} compressor is $\alpha_{\rm{worst\,case}} = \nicefrac{k}{d}$, this estimate is overly pessimistic and excessively conservative when the compressor’s input approximately follows $\nicefrac{x}{\|x\|} \sim_{\rm u.a.r.} S^{d-1}$. This discrepancy is illustrated in Figure~\ref{ch7:fig:alphawc}.

\subsection{Compression schema: adaptive TopLEK compressor}

The worst-case contraction factor from Definition \ref{ch7:def:biased_compressors} for \compname{TopK} compressor may be too pessimistic. One possibility is to generalize the definition of $\alpha \in \mathbb{R}$ to $\alpha(x)$ where $x$ is input to the compressor. If followed this way, then $\alpha(x)$ will vary in each client in each round. Serious reworking is required to analyze the convergence of the Lyapunov function for \algname{FedNL} in this case. 

We propose an alternative and practical approach to address this situation. Instead of introducing \textit{adaptivity} from $\alpha(x)$ directly into Optimization Algorithms, we incorporate \textit{adaptivity} into the compression. Clients during employing \algname{FedNL} obtain information about value $k$ at the beginning of the training. Then they compute $1-\alpha=1 - k/d$ also at the beginning of the optimization process to define contraction value $1-\alpha$ from Definition~\ref{ch7:def:biased_compressors}. 

Next in the compression phase clients return not $k$ values, but $k'$ items $0 \le k' \le k$ with the strategy described in Algorithm~\ref{ch7:alg:top_le_k}.

{
\begin{algorithm}
	\begin{algorithmic}[1]
		\STATE {\bfseries Input and Initialization:} Input for compressor $x \in \RD$; Integer value $0 < k \le d$;
		\STATE  Compute $1-\alpha = 1 - \dfrac{k}{d}$
		\STATE  Step $1$: Compute $\dfrac{\|\mathrm{TopK}[K=k](x) - x\|^2}{\|x\|^2}  = (1-\alpha_{k}(x))$
		\STATE $\dots$
		\STATE  Step $K+1$: Compute $\dfrac{\|\mathrm{TopK}[K=0](x) - x\|^2}{\|x\|^2}  = (1-\alpha_{0}(x)) = 1$
		\STATE Find step $i \in [K]$ such that $(1-\alpha_i(x)) \ge 1-\alpha \ge (1-\alpha_{j}(x))$ where $j=i-1$. \\ 
		Condition is equivalent to $\alpha_i(x) \le \alpha \le \alpha_{j}(x)$.
		\STATE Sample a Bernoulli random variable and:
		\begin{enumerate}
			\item Compress $x$ with \compname{TopK} and $K$ defined from step $i$ with probability ${p}$.
			\item Compress $x$ with \compname{TopK} and $K$ defined from step $j$ with probability $1-p$.
		\end{enumerate}
	\end{algorithmic}
	\caption{\compname{TopLEK}: Top Less Equal K Compressor \clrshort{\textbf{[NEW]}}.}
	\label{ch7:alg:top_le_k}
\end{algorithm}
}

In Algorithm \ref{ch7:alg:top_le_k} step $i$ and step $j \eqdef i-1$ corresponds to \compname{TopK} with $K_j > K_i$. The running value of $k$ is decreasing during the execution of Algorithm \ref{ch7:alg:top_le_k}. As steps progress leads to $1-\alpha_i \to 1$. In particular for step $K+1$ the value $1-\alpha(K=0) = 1$. Therefore we will find steps $i,j$ such that:
\begin{eqnarray*}
(1-\alpha_i(x)) \ge 1-\alpha \ge (1-\alpha_{j}(x)), j \eqdef i-1.
\end{eqnarray*}	

To make Equation~\eqref{ch7:def:biased_compressors} as tight as possible we compress in a randomized way with \compname{TopK[$k=K_i$]} compressor employed with probability ${p}$, and with \compname{TopK[$k=K_j$]} compressor with probability $1-{p}$, we ${p}$ derived as:

\begin{eqnarray*}
&&\mathbb{E}\left[ \|\mathcal{C}(x) - x\|^2\right] \le (1-\alpha) \|x\|^2 \\
&\iff& p \left[ \|\mathcal{C}_i(x) - x\|^2\right] + (1-p) \left[ \|\mathcal{C}_j(x) - x\|^2\right] \le (1-\alpha) \|x\|^2 \\
&\iff& (p (1-\alpha_{i}) + (1-p) (1-\alpha_{j})) \|x\|^2 \le (1-\alpha) \|x\|^2 \\
&\iff& p -p \alpha_{i} + 1-\alpha_{j} - p(1-\alpha_{j}) = 1-\alpha \\
&\iff& -p \alpha_{i} -\alpha_{j} + p\alpha_{j} = -\alpha \iff p (\alpha_{j} - \alpha_{i}) = \alpha_{j} - \alpha \iff p = \dfrac{\alpha_{j} - \alpha}{\alpha_{j} - \alpha_{i}}
\end{eqnarray*}

The last equation which defines ${p}$ has non-negative values 
in the numerator and denominator. Due to construction of $\alpha_{j}, \alpha, \alpha_i$ 
namely $\alpha_i(x) \le \alpha \le \alpha_{j}(x)$ the value of ${p} \le 1$. Therefore $p \in [0,1]$, represents a valid probability. The constructed \compname{TopLEK[K]} compressor contractive inequality~\eqref{ch7:def:biased_compressors} attains tight equality duing using \algname{FedNL}, that is $$\mathbb{E}\left[ \|\mathcal{C}(x) - x\|^2\right] = (1-\alpha) \|x\|^2.$$

The contraction coefficient $1-\alpha$ for clients during the execution of \algname{FedNL} remains constant and the mathematical framework of \algname{FedNL} can be seamlessly applied without any modifications. A notable advantage of \compname{TopLEK[k]} over \compname{TopK[k]} is that clients can transmit not only $k$ components but at most $k$. In fortuitous scenarios, clients may only need to send $0$ components.

\subsection{TopLEK applied for matrices}

The \algname{FedNL} operates on a family of contractive compressors applied for matrices which is defined as:
\begin{eqnarray*}
\mathcal{C}(\delta)=\{\mathcal{C} \in \mathbb{R}^{d \times d} \to \mathbb{R}^{d \times d} : &(i)& \mathbb{E}\left[ \|\mathcal{C}(\mX) - \mX\|_F ^2\right] \le (1-\delta)\|\mX\|_F^2, \forall \mX, \delta \in (0,1], \\ &(ii)& \|\mathcal{C}(\mX)\|_F \le \|\mX\|_F\}.
\end{eqnarray*}

Requirements (i) can be seen as a generalization of the class of contractive operators for vectors, defined in Equation~\eqref{ch7:def:biased_compressors}. Because both \compname{TopK} and \compname{TopLEK} only replace some matrix elements by $0$ value, and do not perform any scaling. Therefore both compressors satisfy the requirement (ii) automatically.

\clearpage
\addtocounter{adjsection}{1}
\section{Missing Plots}

\subsection{Single-node experiments with FedNL-LS}
\label{ch7:app:single-node-cmp-vs-industry-extra}

\begin{figure}[h]
\centering
\includegraphics[width=0.48\textwidth]{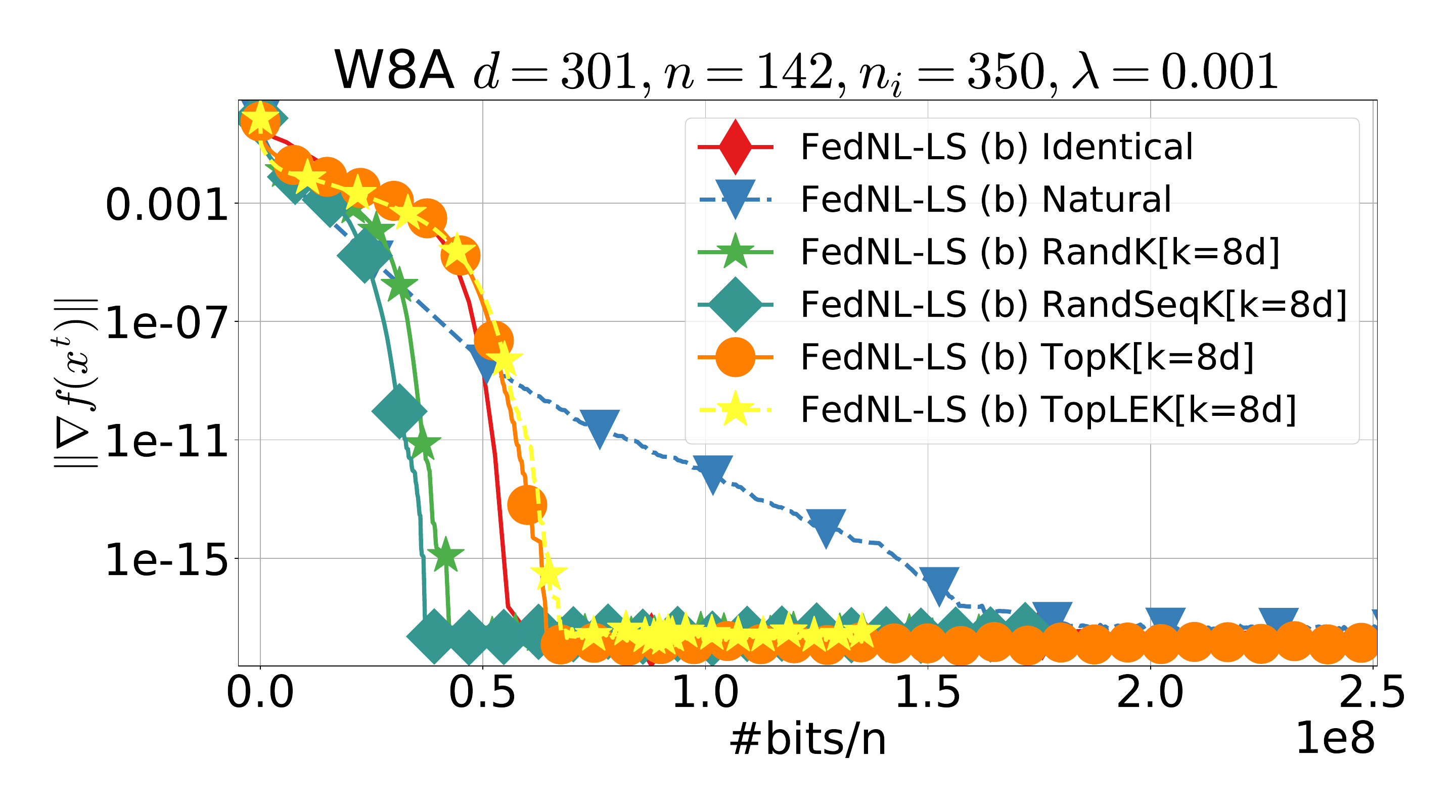}
\includegraphics[width=0.48\textwidth]{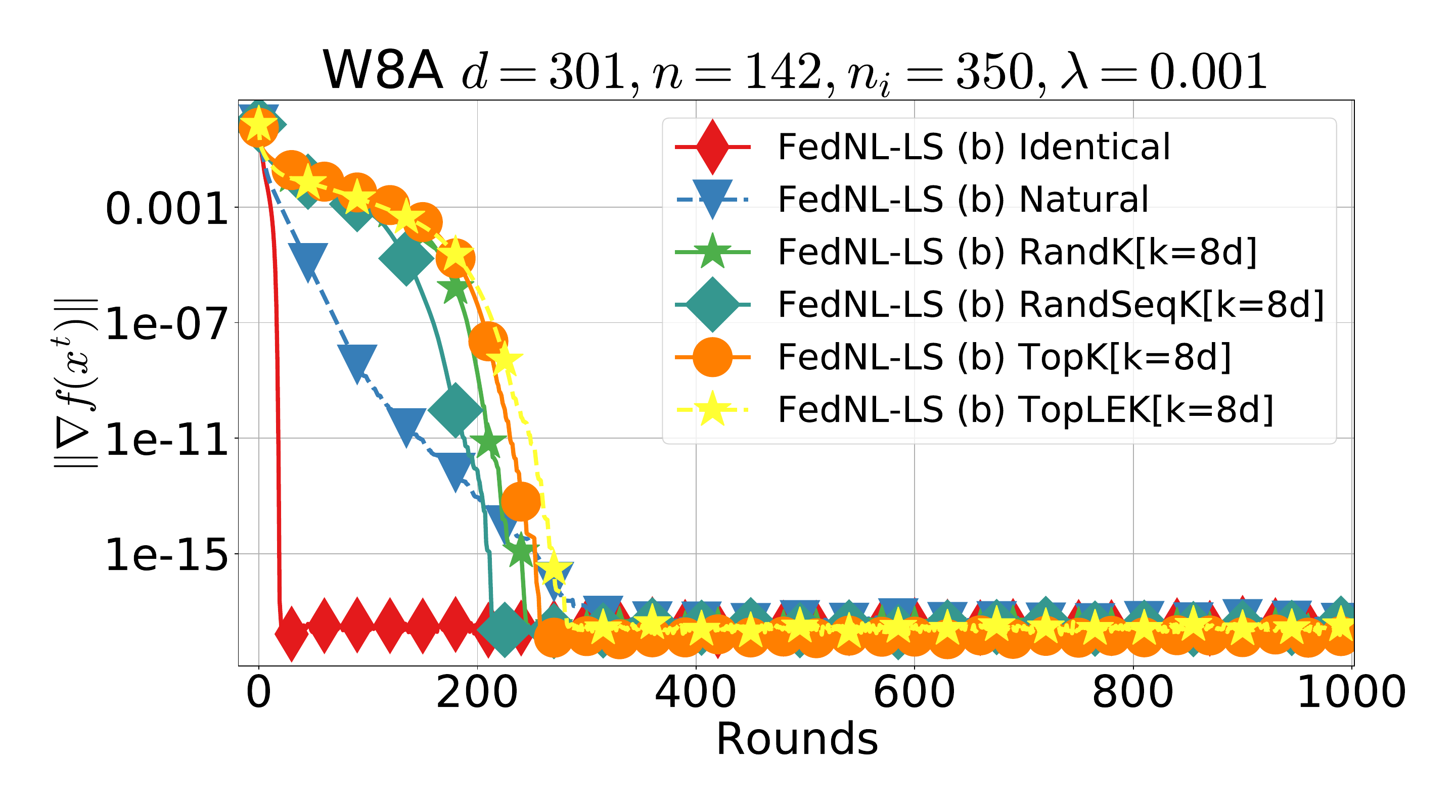}
\includegraphics[width=0.48\textwidth]{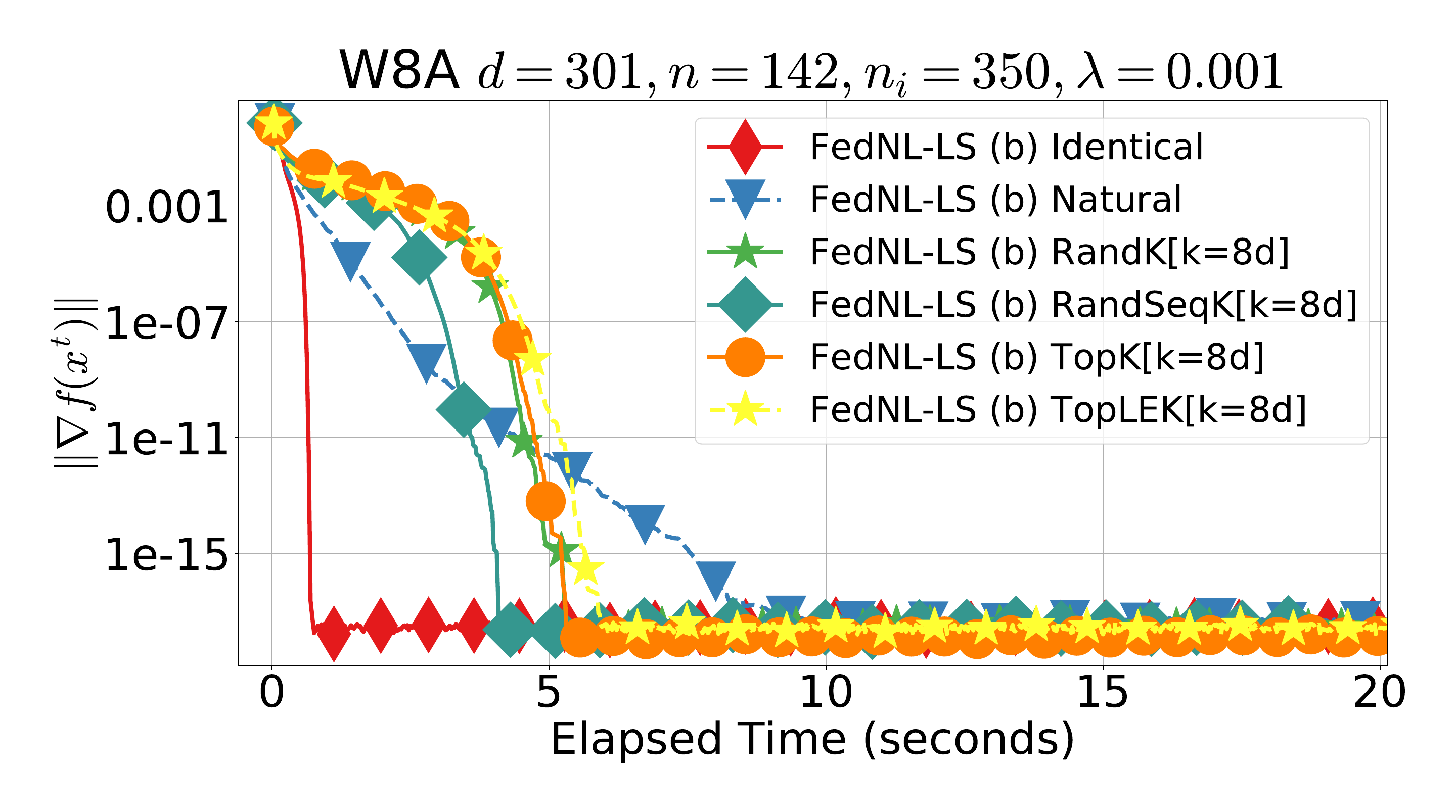}		
\caption{\algname{FedNL-LS} simulation in a single-node, $1000$ rounds, theoretical step size, FP64. Line search parameters $c=0.49,\gamma=0.5$. Dataset \dataname{W8A} ($49749$ samples) augmented with intercept	split to $n_i=350$ samples/client.}
\label{ch7:fig:fednl-ls-w8a}
\end{figure}

\begin{figure}[h]
\centering
\includegraphics[width=0.48\textwidth]{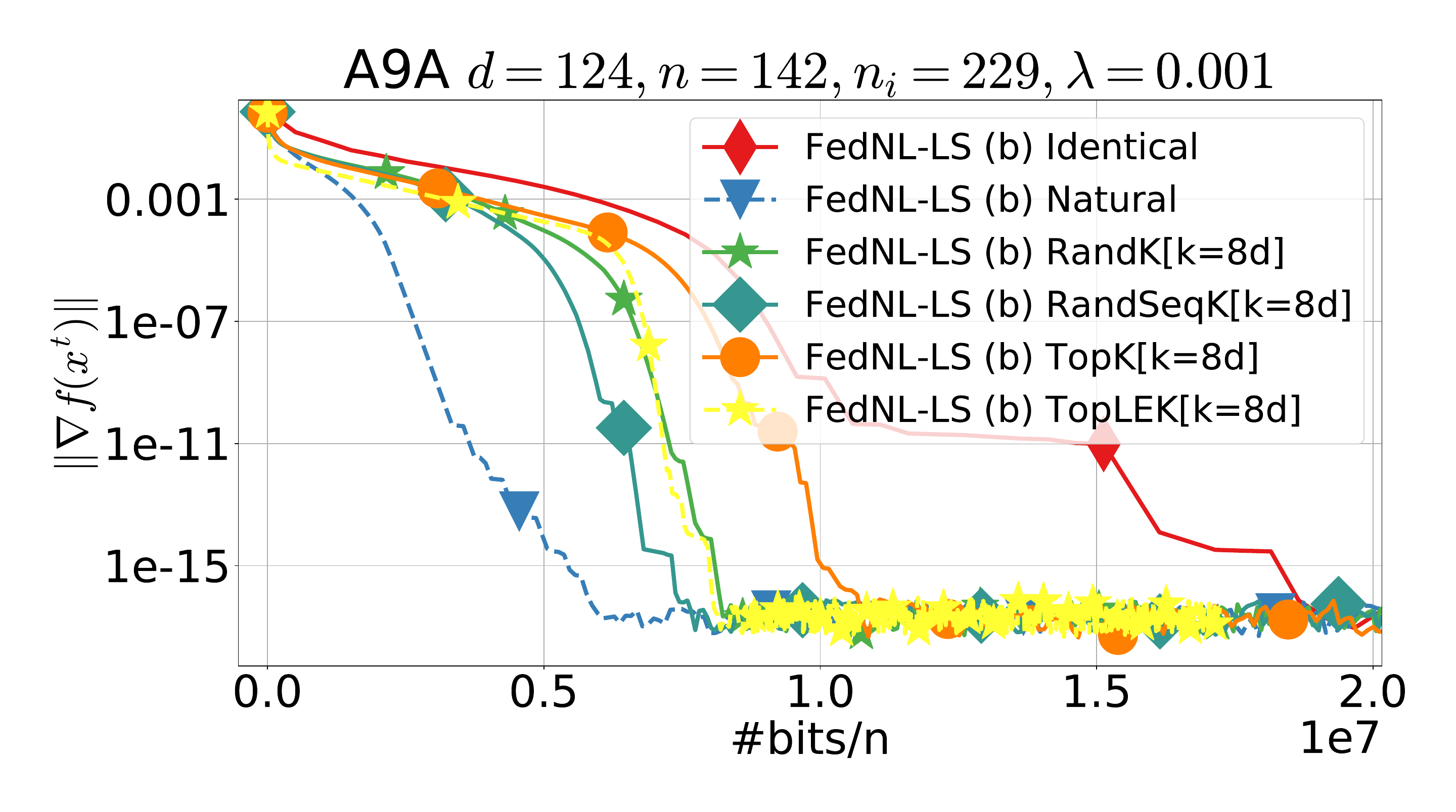}
\includegraphics[width=0.48\textwidth]{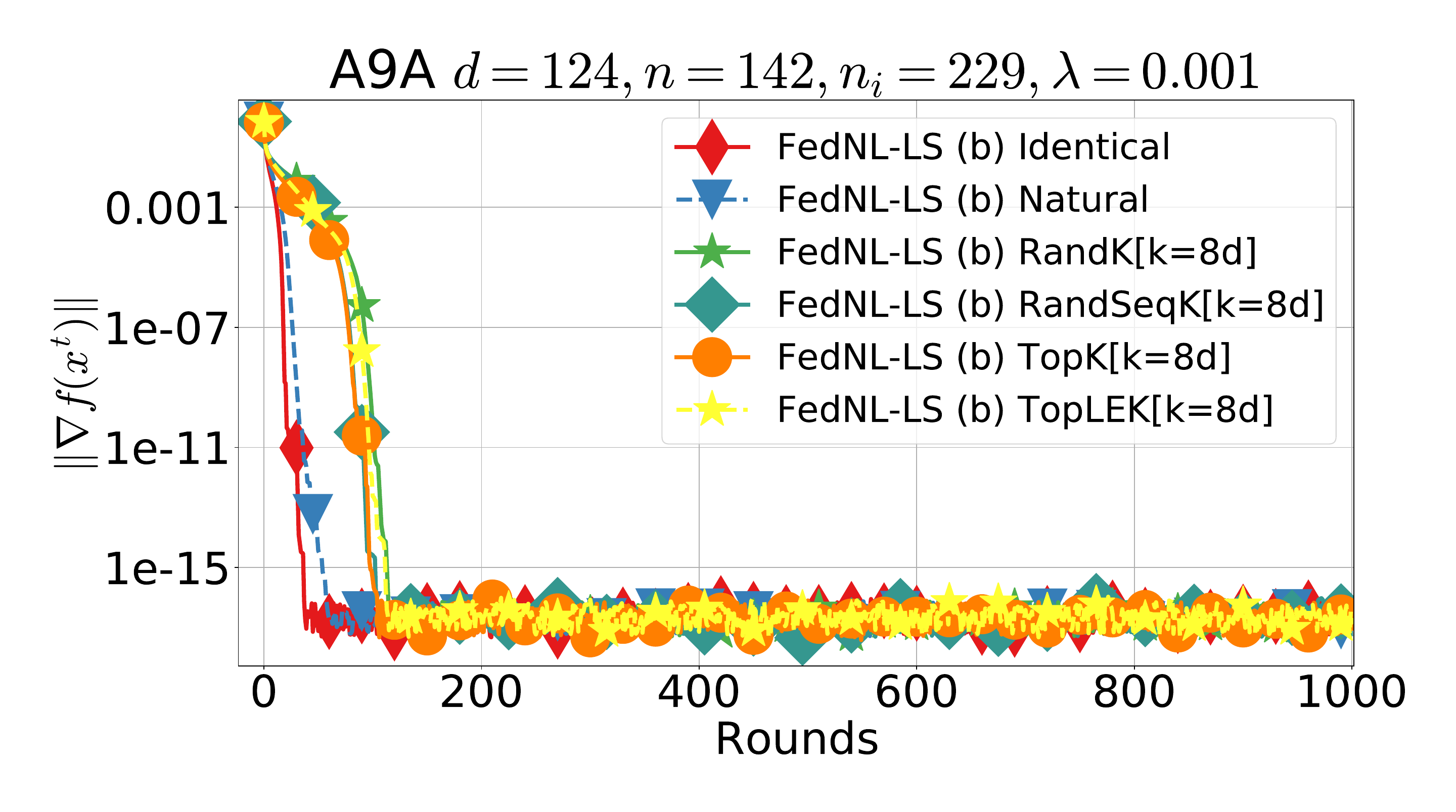}
\includegraphics[width=0.48\textwidth]{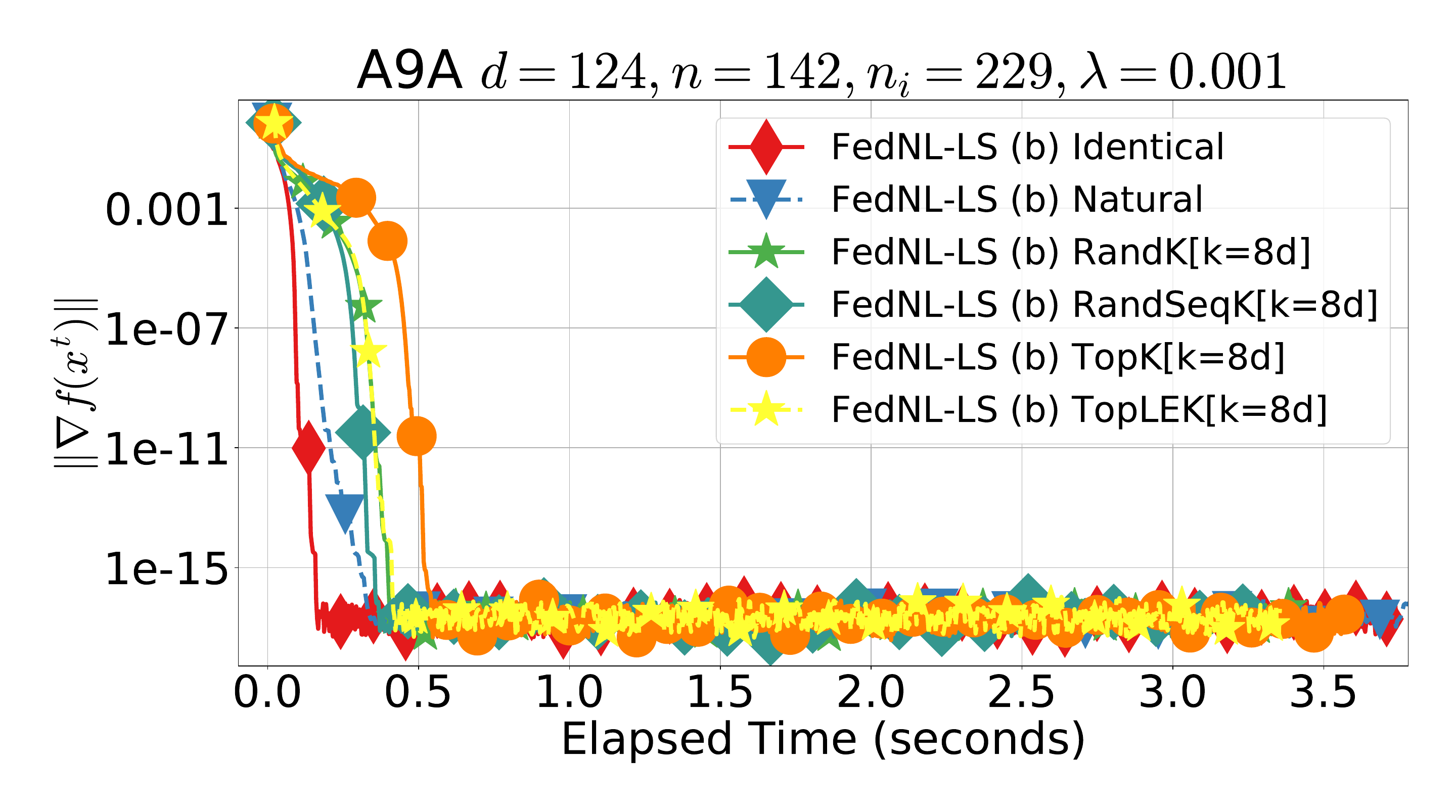}

\caption{\algname{FedNL-LS} simulation in a single-node, $1000$ rounds, theoretical step size, FP64. Line search parameters $c=0.49,\gamma=0.5$. Dataset \dataname{A9A} ($32561$ samples) augmented with intercept, split to $n_i=229$ samples/client.}

\label{ch7:fig:fednl-ls-a9a}
\end{figure}

\begin{figure}[h]
\centering
\includegraphics[width=0.48\textwidth]{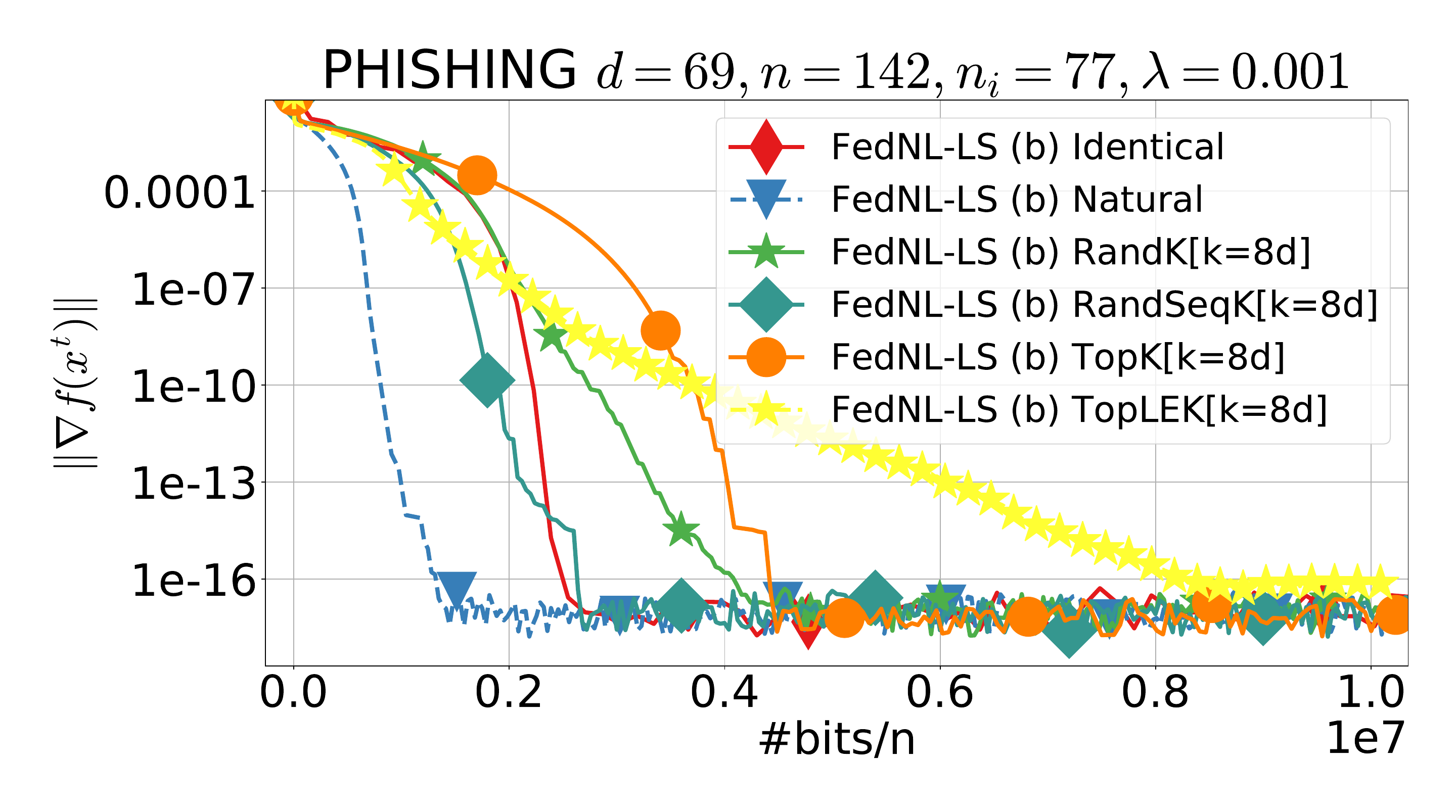}
\includegraphics[width=0.48\textwidth]{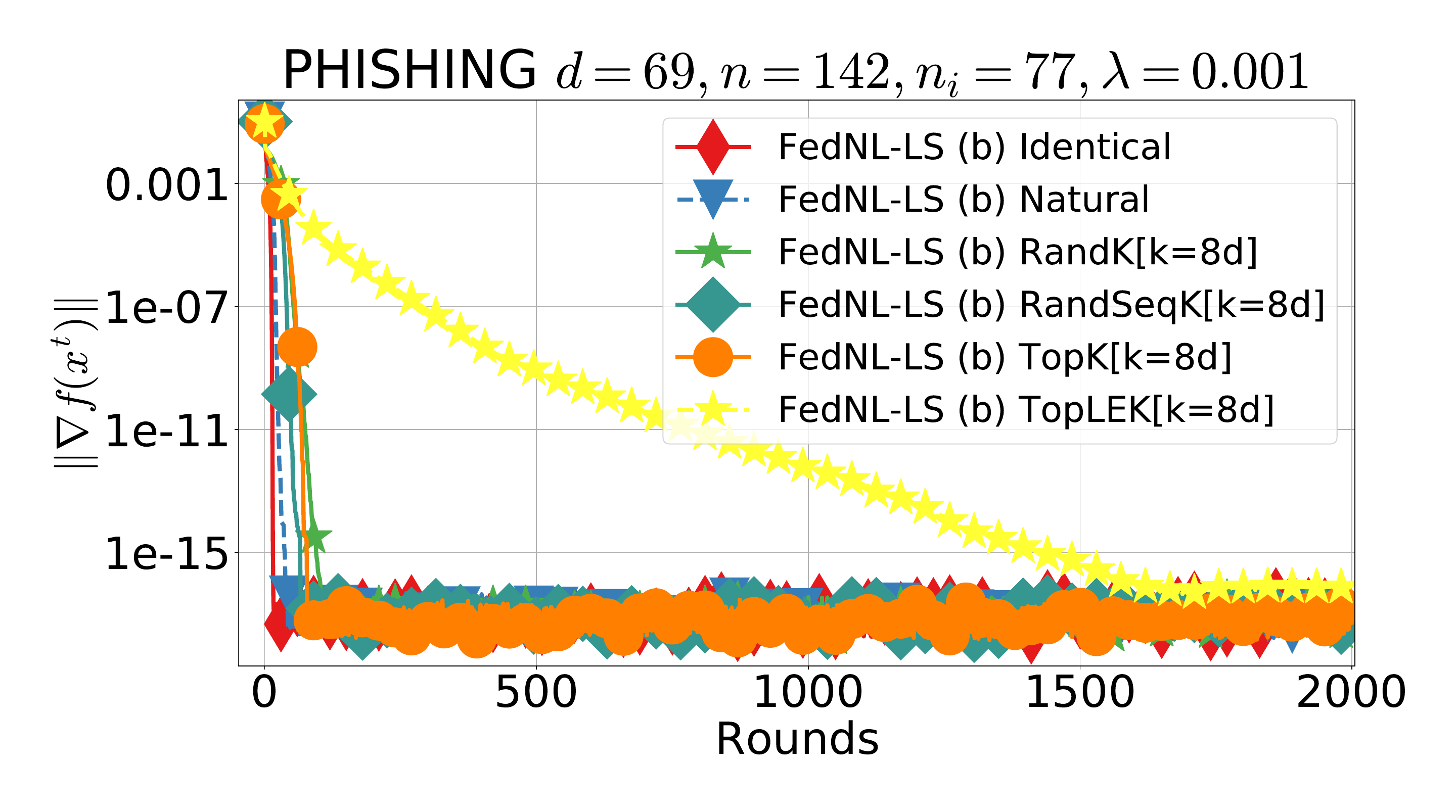}
\includegraphics[width=0.48\textwidth]{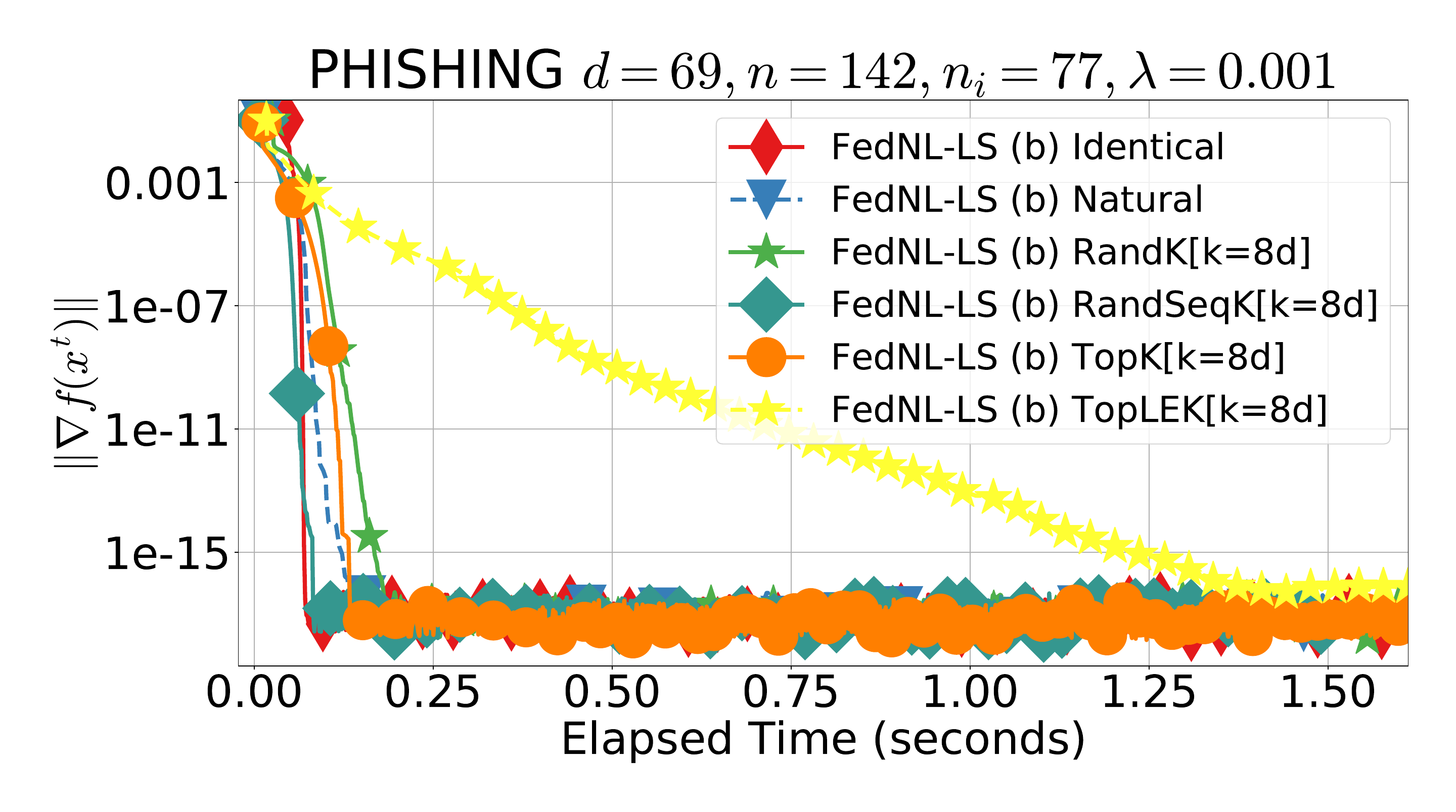}

\caption{\algname{FedNL-LS} simulation in a single-node, $2000$ rounds, theoretical step size, FP64. Line search parameters $c=0.49,\gamma=0.5$. Dataset \dataname{PHISHING} ( $11055$ samples) augmented with intercept	split to $n_i=77$ samples/client.}
\label{ch7:fig:fednl-ls-phishing}
\end{figure}

The results of an experiment using \algname{FedNL-LS} which represent a modification of \algname{FedNL} are presented in Figures \ref{ch7:fig:fednl-ls-w8a}, \ref{ch7:fig:fednl-ls-a9a}, \ref{ch7:fig:fednl-ls-phishing}. For a pseudocode of \algname{FedNL-LS} see  Appendix~\ref{ch7:app:fednl-ls-descr}. The \textit{communicated bits} include the cost for transferring the scalar values and auxiliary information from the compressor.

For \compname{TopK} and \compname{TopLEK} auxiliary information includes $32$ bits per single index transfer. For \compname{TopLEK} communicated bits include an extra $32$ bits value to encode the number of communicated components. For \compname{RandK} and \compname{RandSeqK} we support two modes: (i) indices are transferred explicitly; (ii) the master reconstructs them via using known seeds for \abr{PRG}. In all experiments (ii) strategy is used.


\clearpage
\subsection{Multi-node experiments with FedNL, FedNL-PP, and FedNL-LS}

\label{ch7:app:experiments-multi-node}


\begin{figure}[h]
\centering
\includegraphics[width=0.48\textwidth]{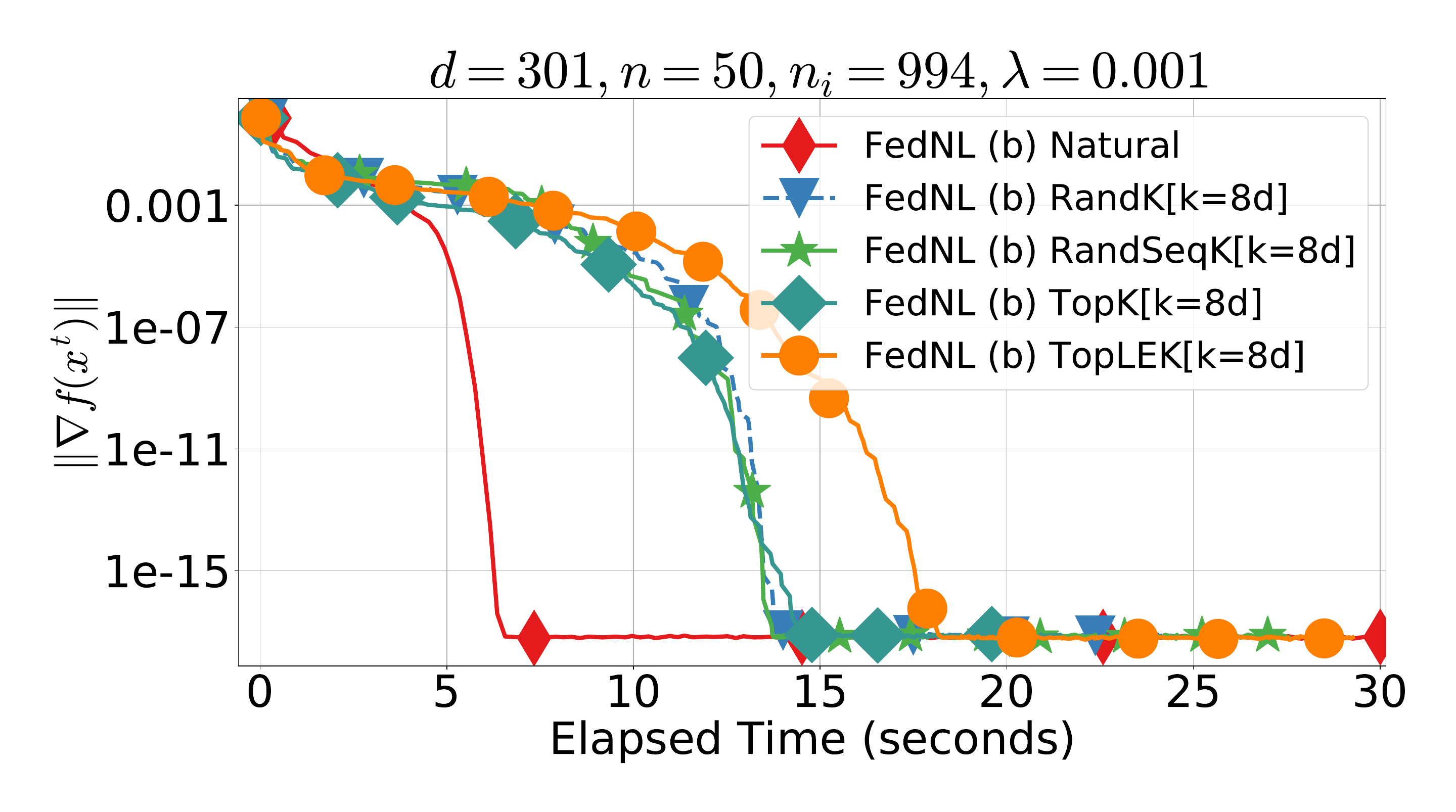}		
\includegraphics[width=0.48\textwidth]{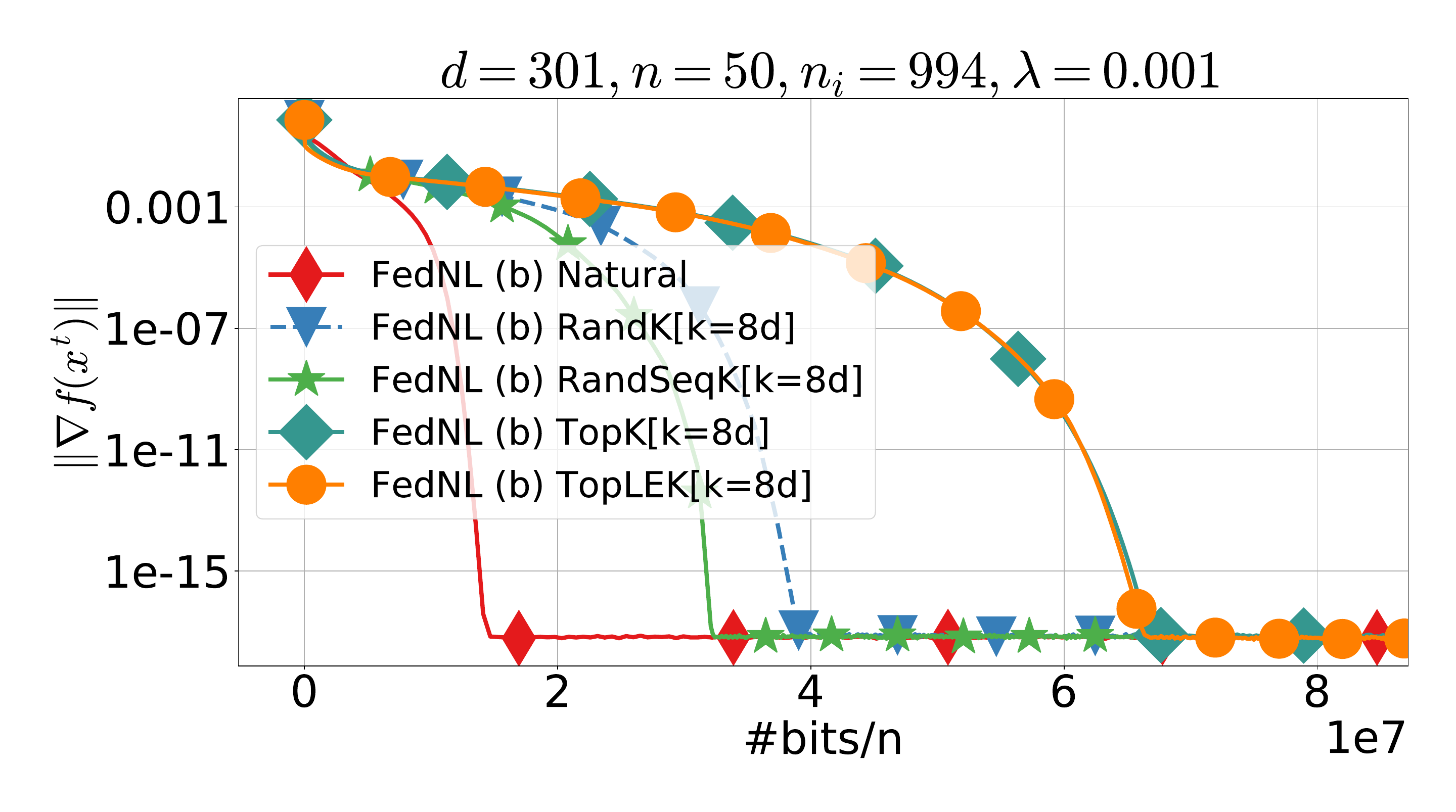}
\includegraphics[width=0.48\textwidth]{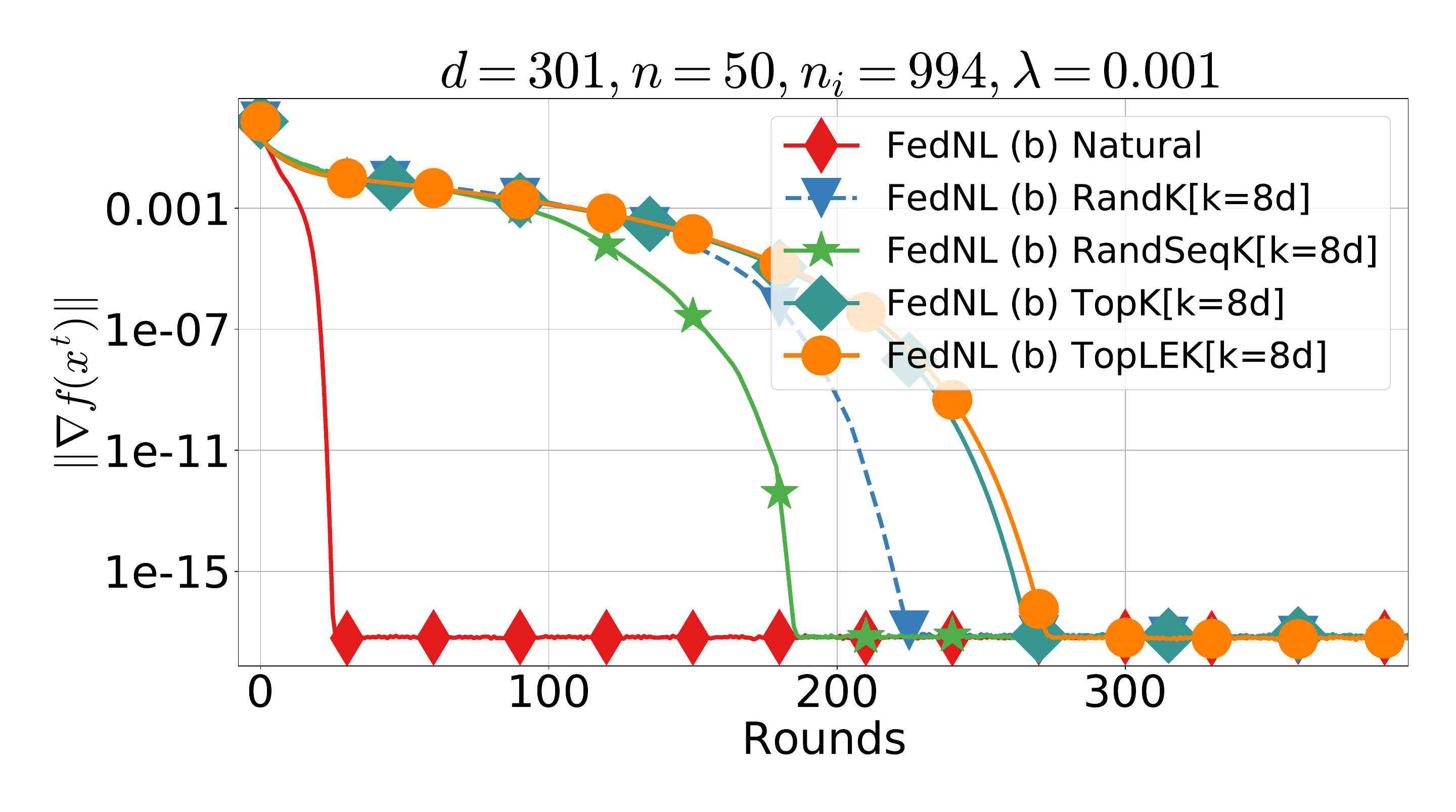}

\caption{\algname{FedNL} in multi-node setting, theoretical step size, $n=50$, FP64 arithmetic, 1 {CPU} core per node and master, \abr{TCP/IPv4}, dataset \dataname{W8A} reshuffled u.a.r. and augmented with intercept.}
\label{ch7:fig:fednl-w8a-app}
\end{figure}

\begin{figure}[h]
\centering
\includegraphics[width=0.48\textwidth]{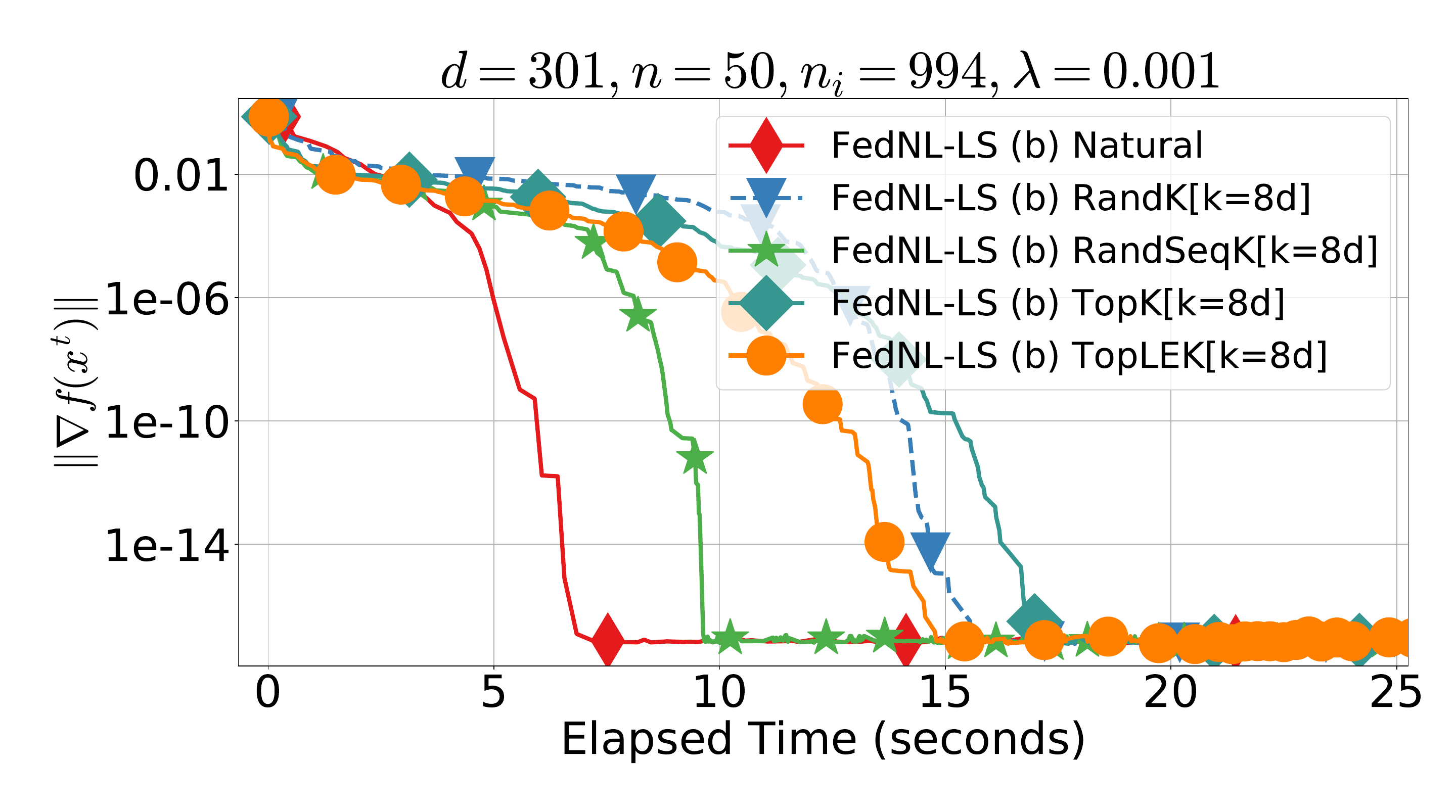}
\includegraphics[width=0.48\textwidth]{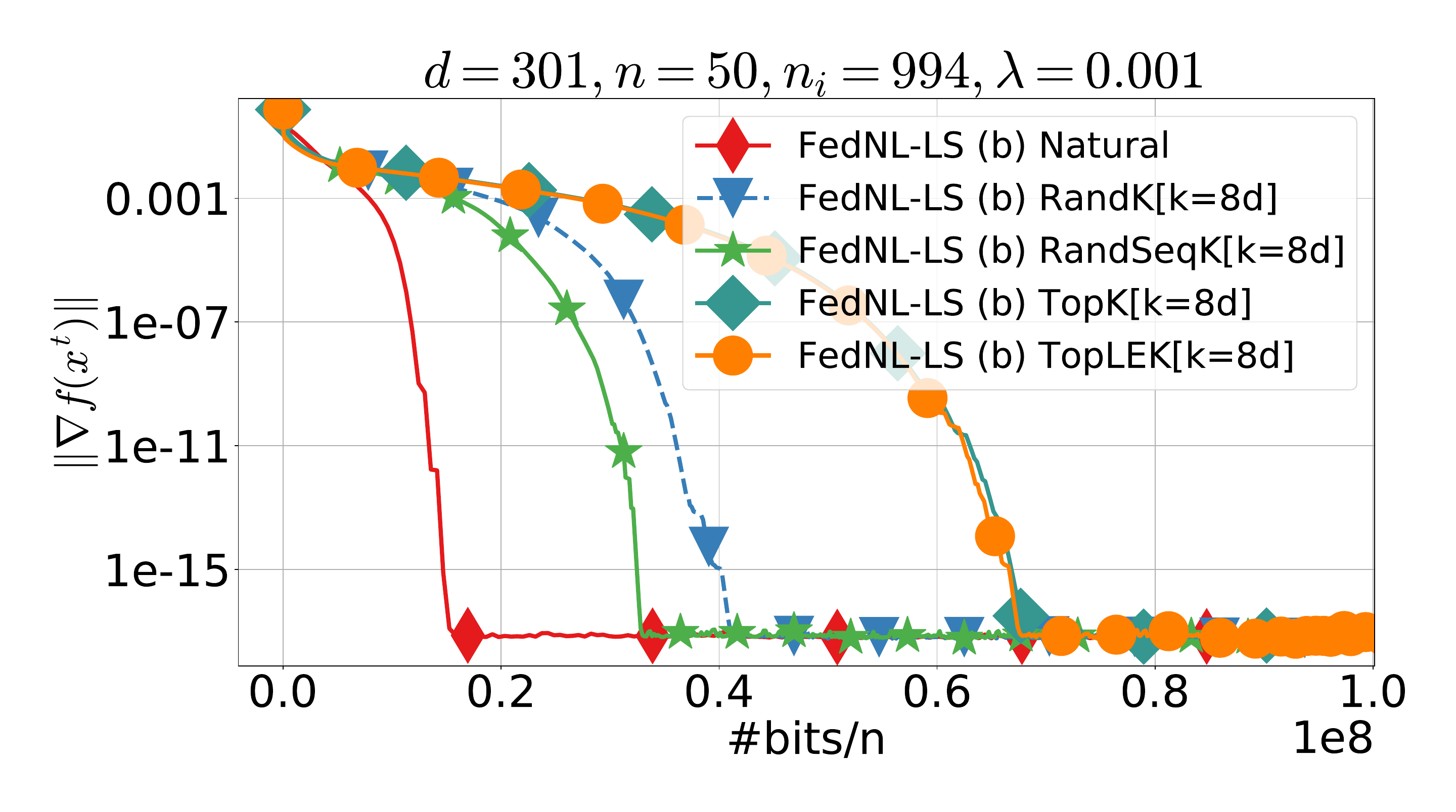}
\includegraphics[width=0.48\textwidth]{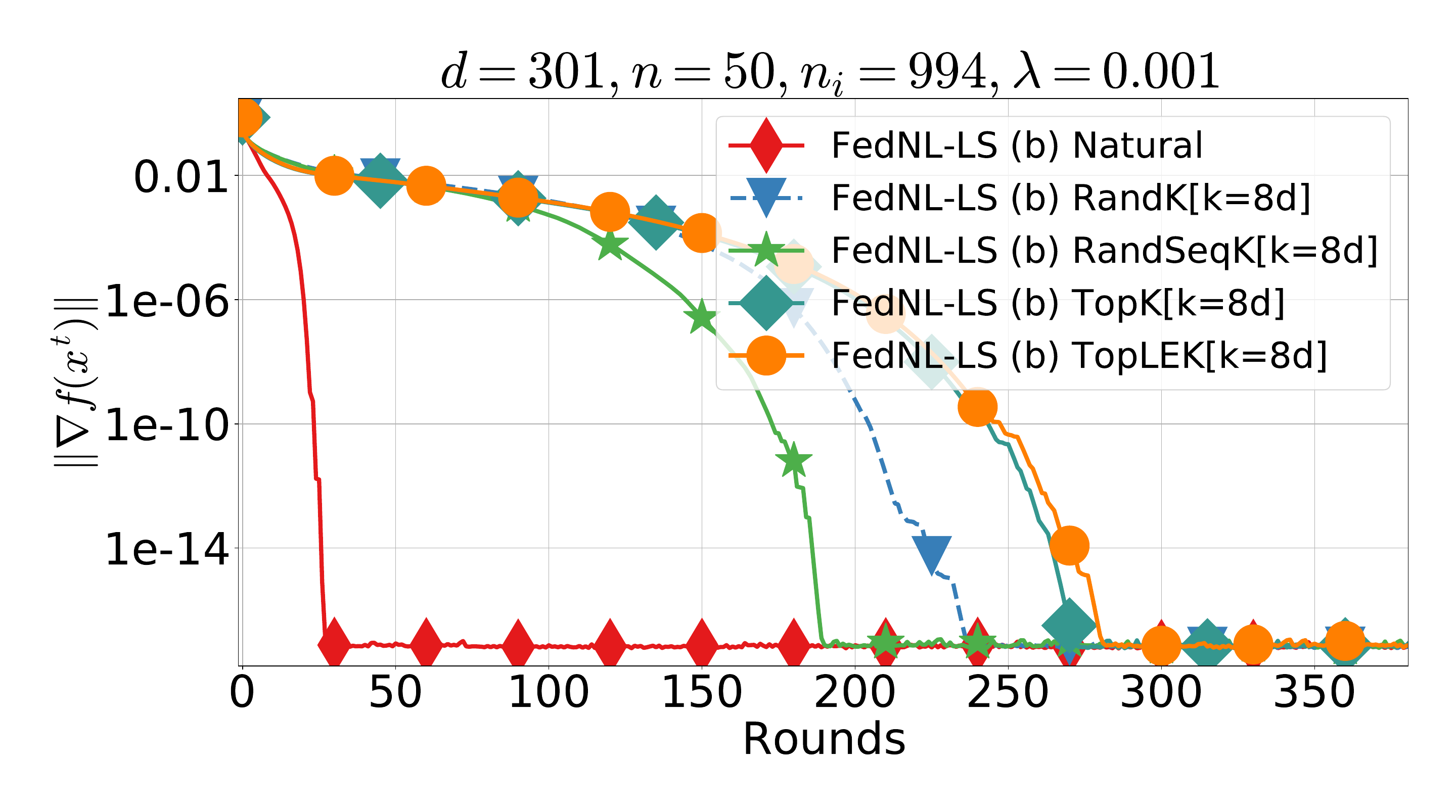}

\caption{ \algname{FedNL-LS} in multi-node setting, $n=50$, FP64 arithmetic, 1 {CPU} core per node and master, \abr{TCP/IPv4}, dataset \dataname{W8A} reshuffled u.a.r. and augmented with intercept. The line search parameters $c=0.49,\gamma=0.5$.}
\label{ch7:fig:fednl-ls-w8a-app}
\end{figure}

\begin{figure}[h]
\centering

\includegraphics[width=0.48\textwidth]{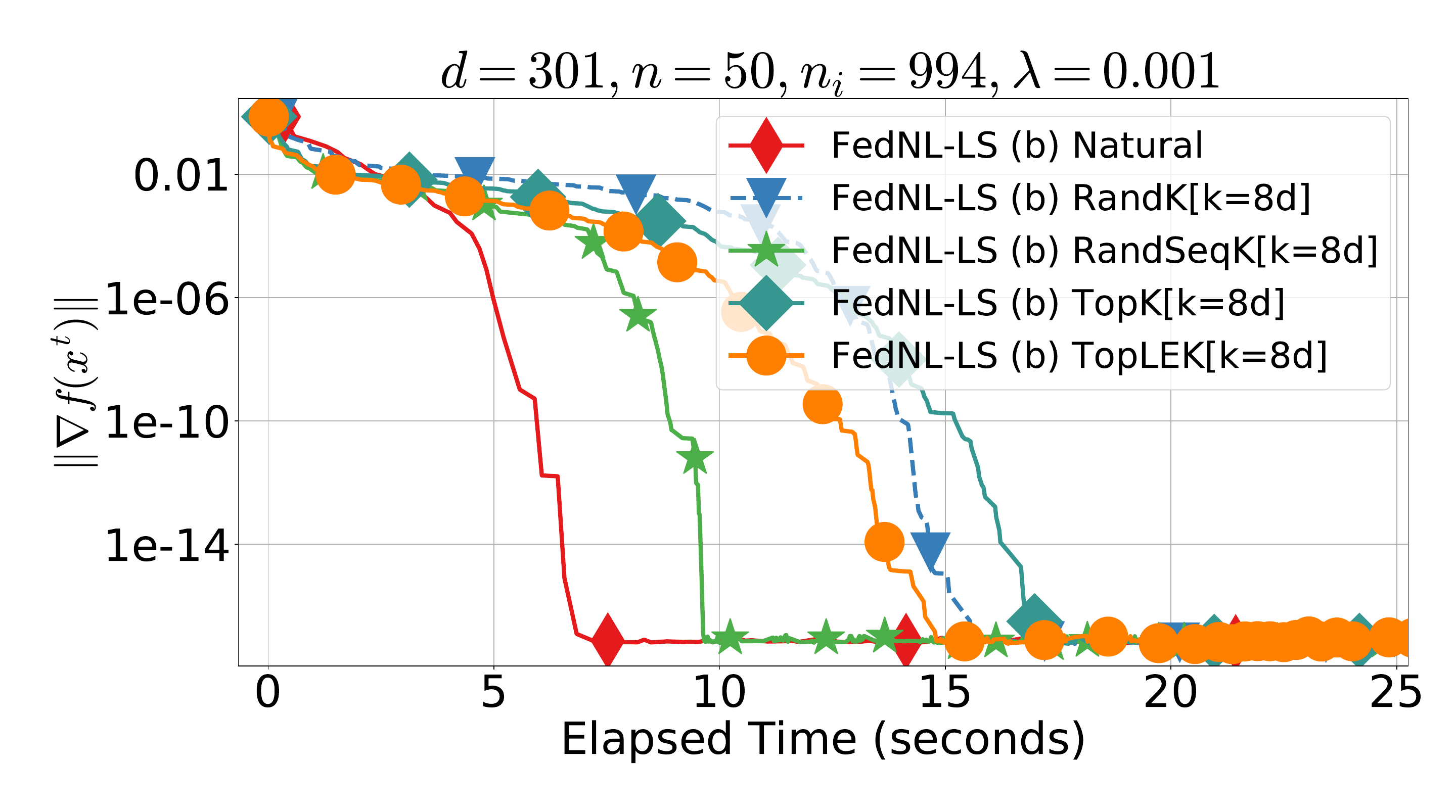}
\includegraphics[width=0.48\textwidth]{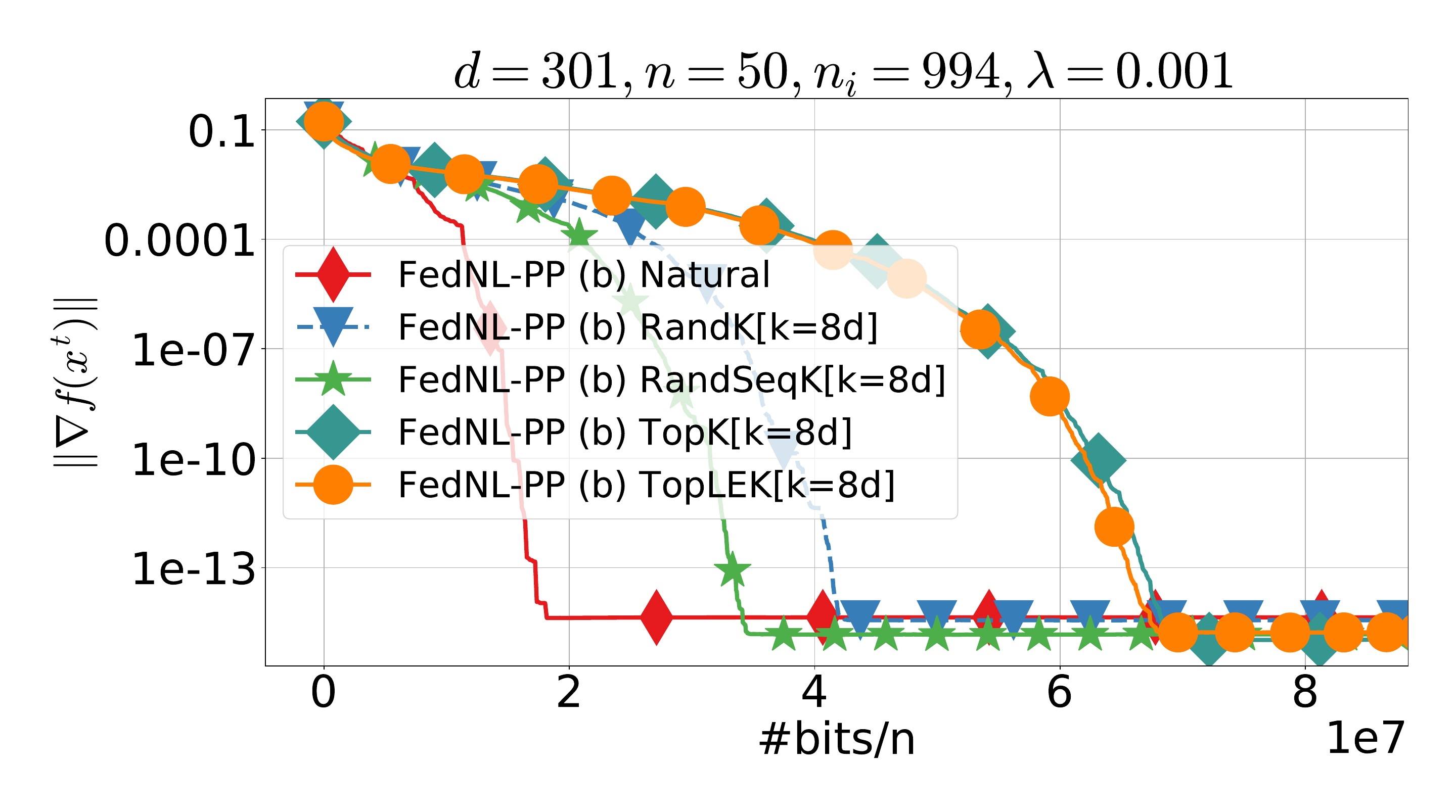}
\includegraphics[width=0.48\textwidth]{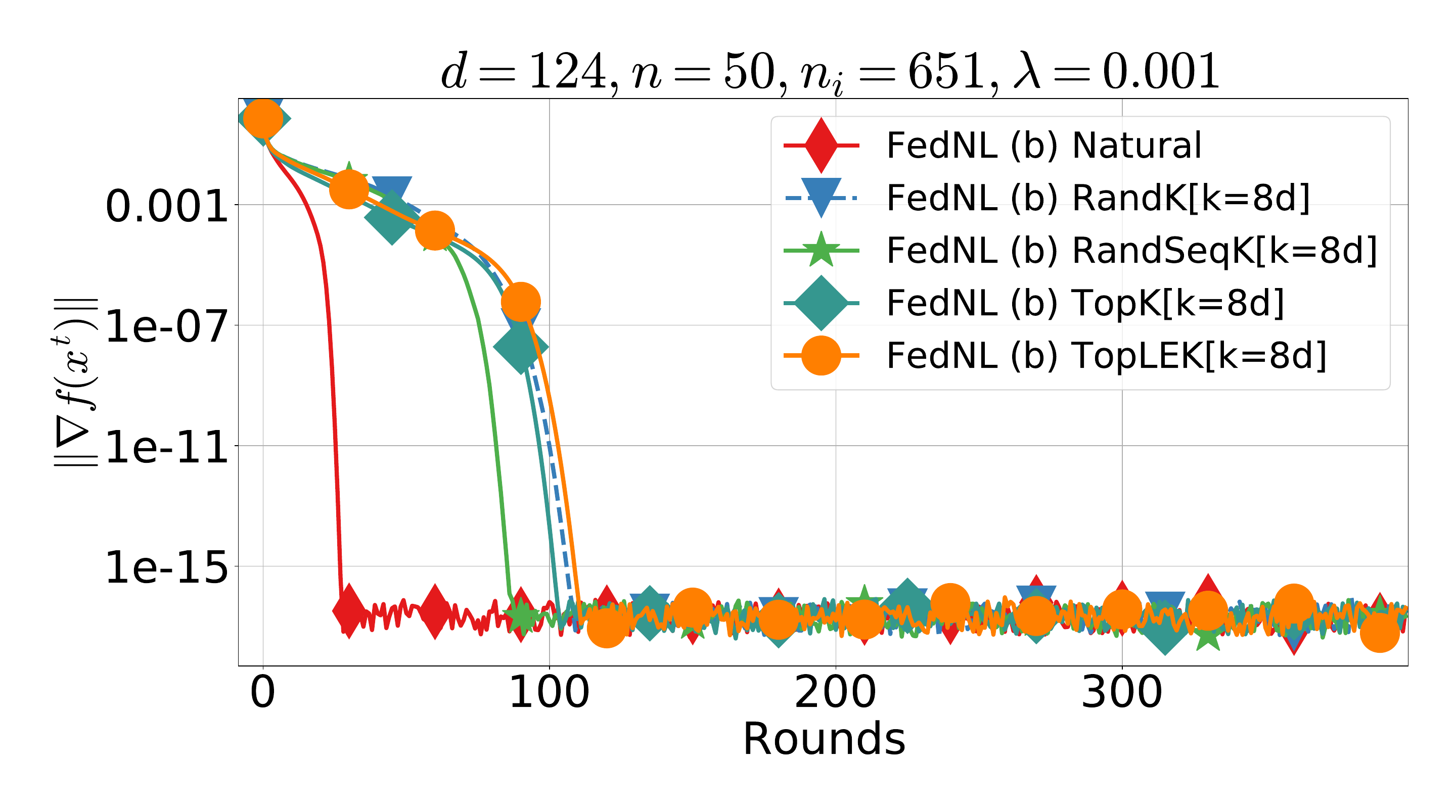}	

\caption{ \algname{FedNL-PP} in multi-node setting, $n=50$, $|S^k|=12$ clients per round, FP64 arithmetic, 1 {CPU} core per node and master, \abr{TCP/IPv4}. \dataname{W8A} dataset reshuffled u.a.r. and augmented with intercept.}
\label{ch7:fig:fednl-pp-w8a-app}
\end{figure}


\begin{figure}[h]
\centering
\includegraphics[width=0.48\textwidth]{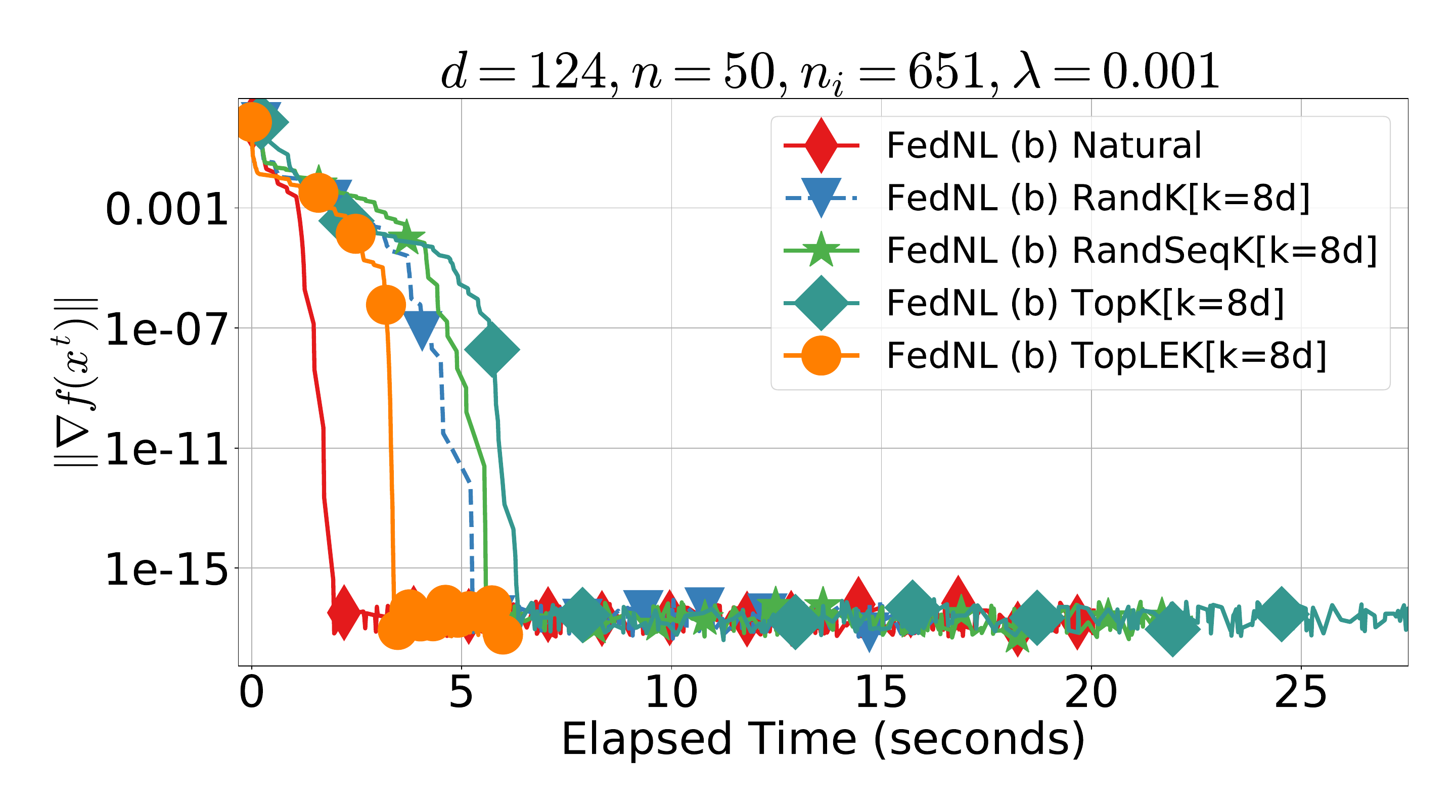}
\includegraphics[width=0.48\textwidth]{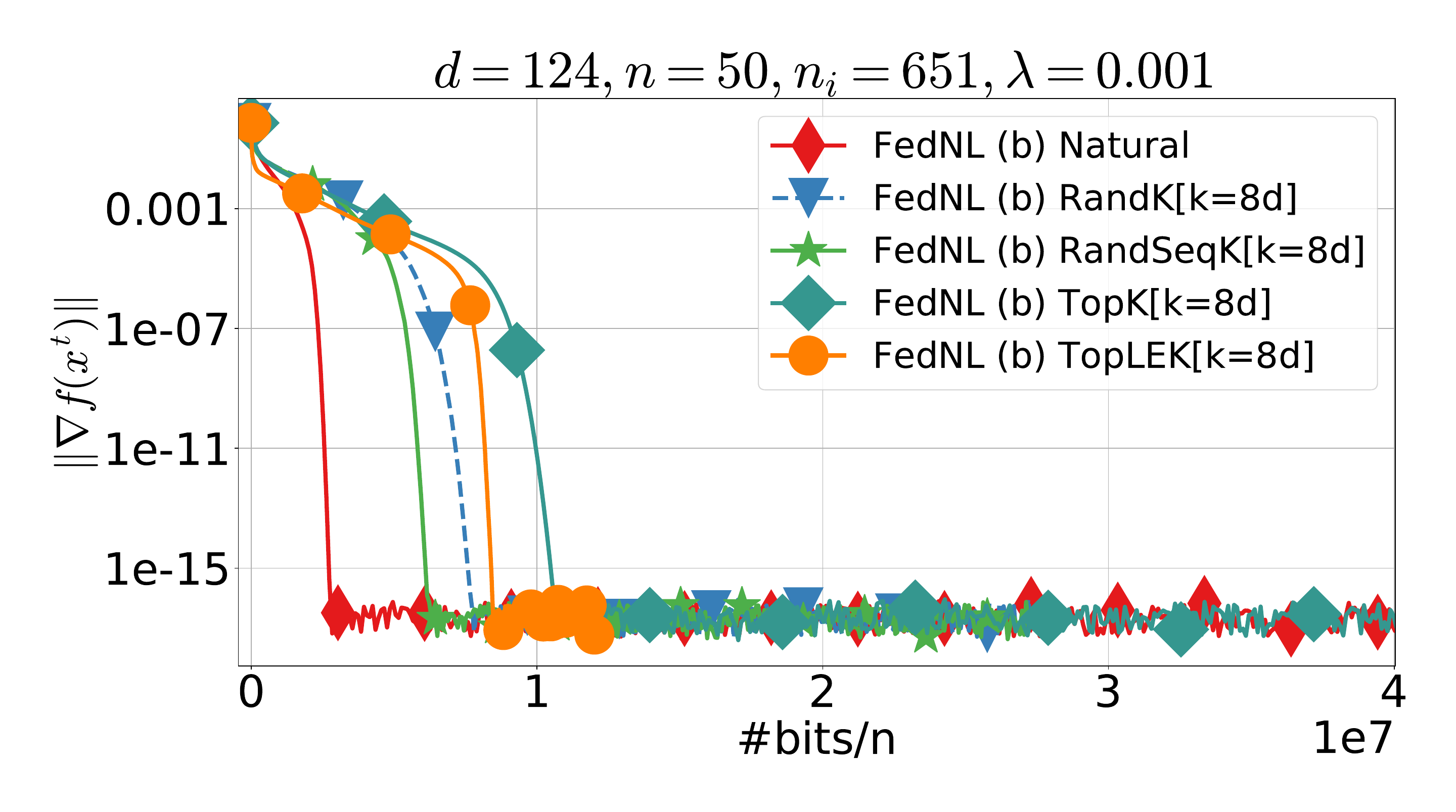}
\includegraphics[width=0.48\textwidth]{ch7imgs/figs/fednl-appendix-figs/a9a/fednl/grad-vs-rounds.pdf}

\caption{\algname{FedNL} in multi-node setting, theoretical step size, $n=50$, FP64 arithmetic, 1 {CPU} core per node and master, \abr{TCP/IPv4}, dataset \dataname{A9A} reshuffled u.a.r. and augmented with intercept.}

\label{ch7:fig:fednl-a9a-app}
\end{figure}

\begin{figure}[h]
\centering
\includegraphics[width=0.48\textwidth]{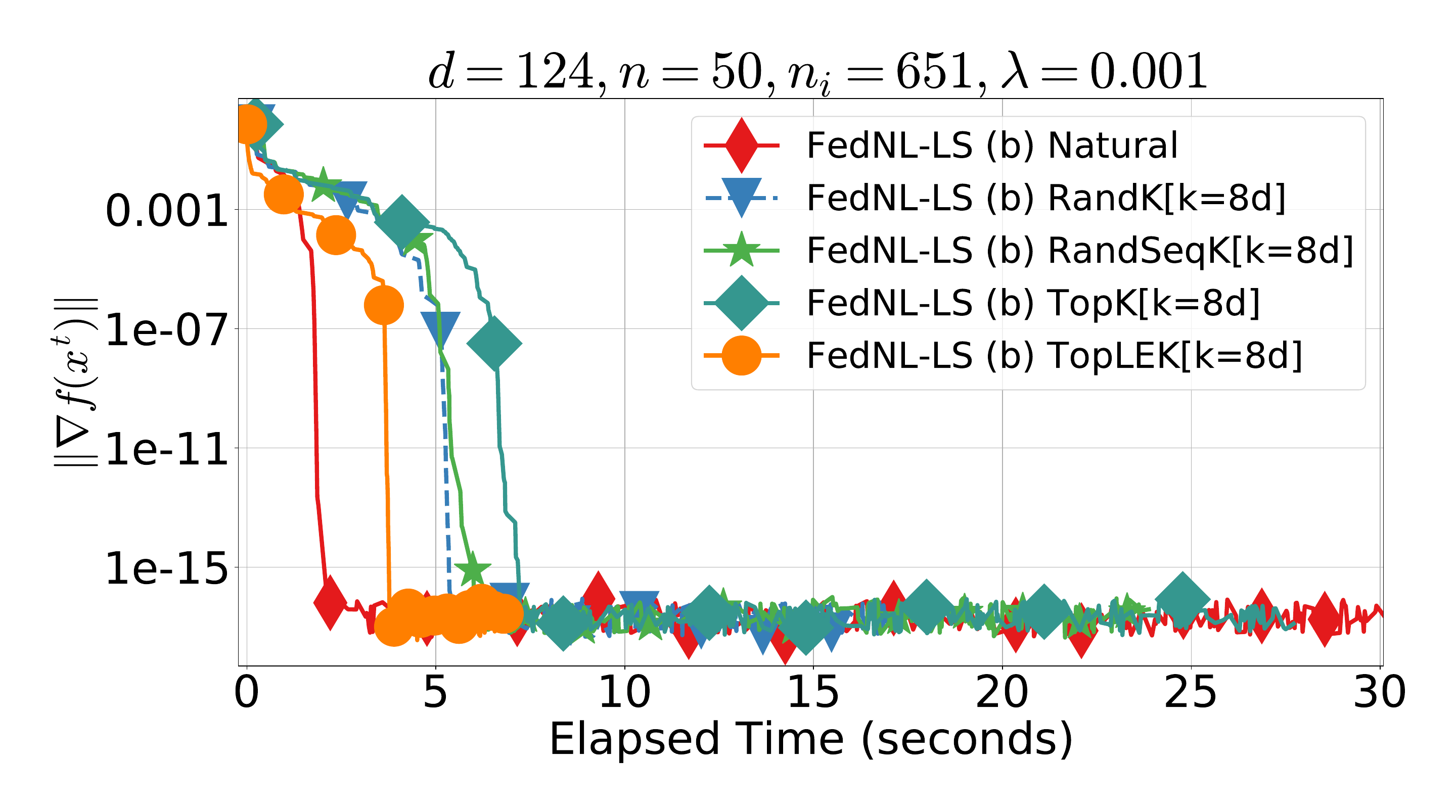}	
\includegraphics[width=0.48\textwidth]{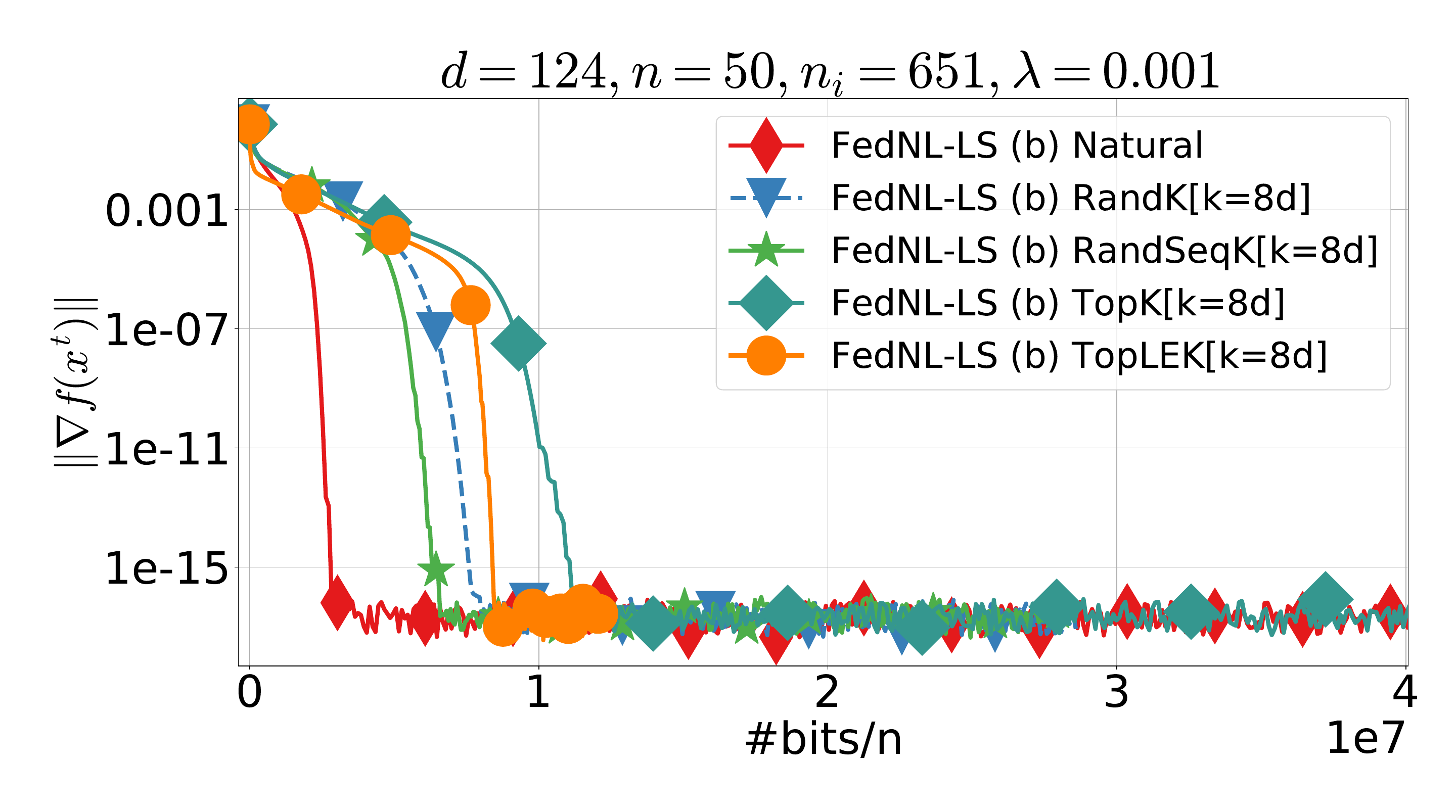}
\includegraphics[width=0.48\textwidth]{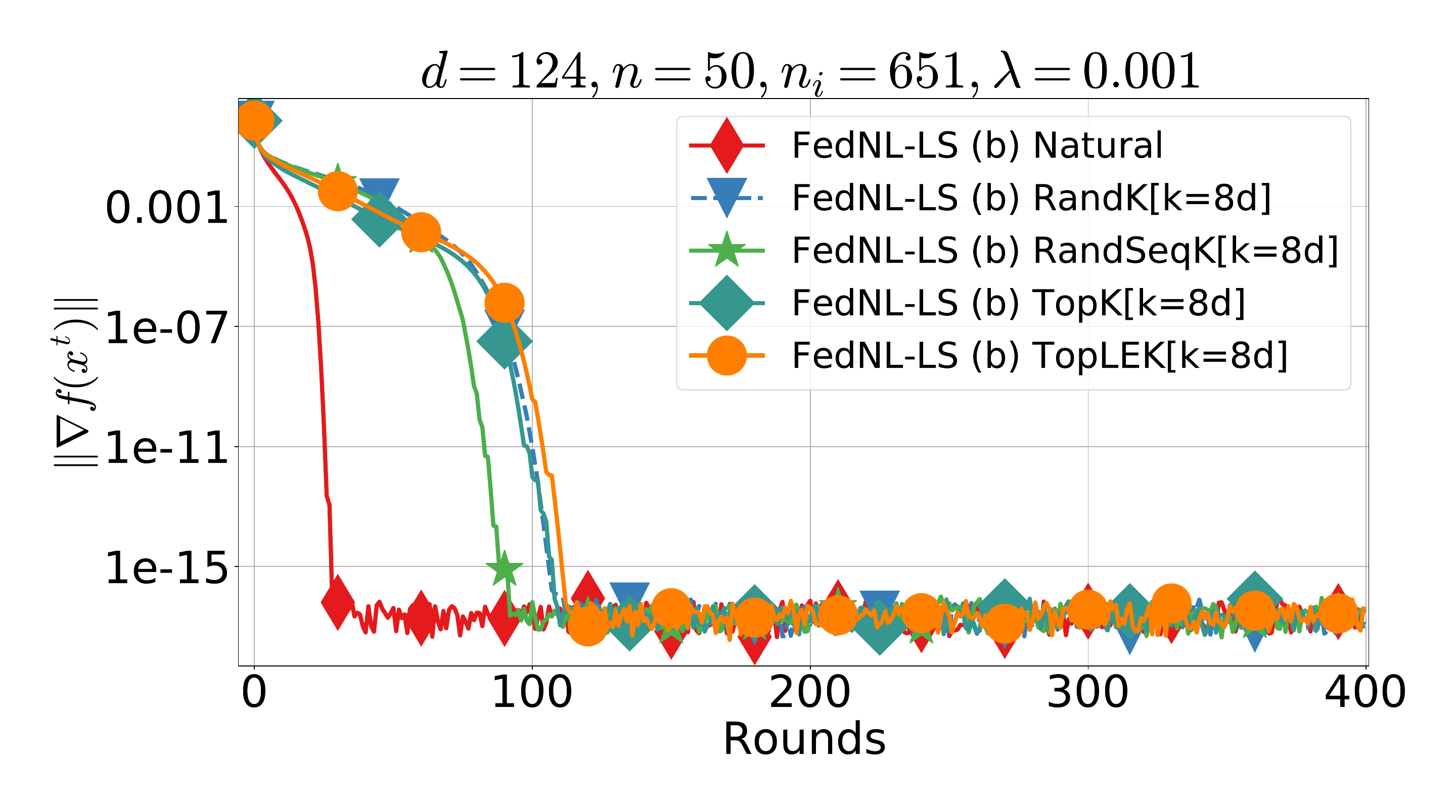}

\caption{ \algname{FedNL-LS} in multi-node setting, $n=50$, FP64 arithmetic, 1 {CPU} core per node and master, \abr{TCP/IPv4}, dataset \dataname{A9A} reshuffled u.a.r. and augmented with intercept. The line search parameters $c=0.49,\gamma=0.5$.}
\label{ch7:fig:fednl-ls-a9a-app}
\end{figure}

\begin{figure}[h]
\centering
\includegraphics[width=0.48\textwidth]{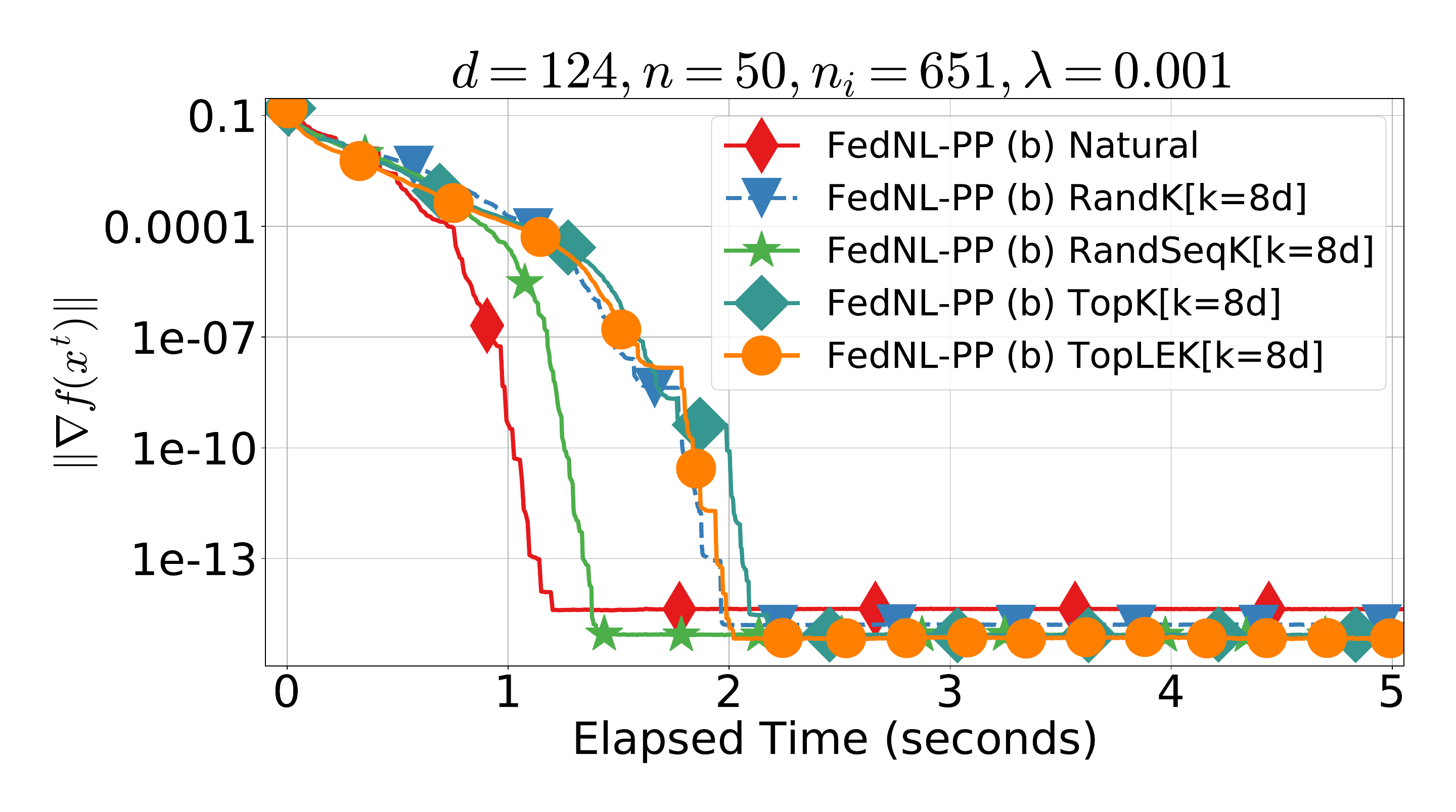}
\includegraphics[width=0.48\textwidth]{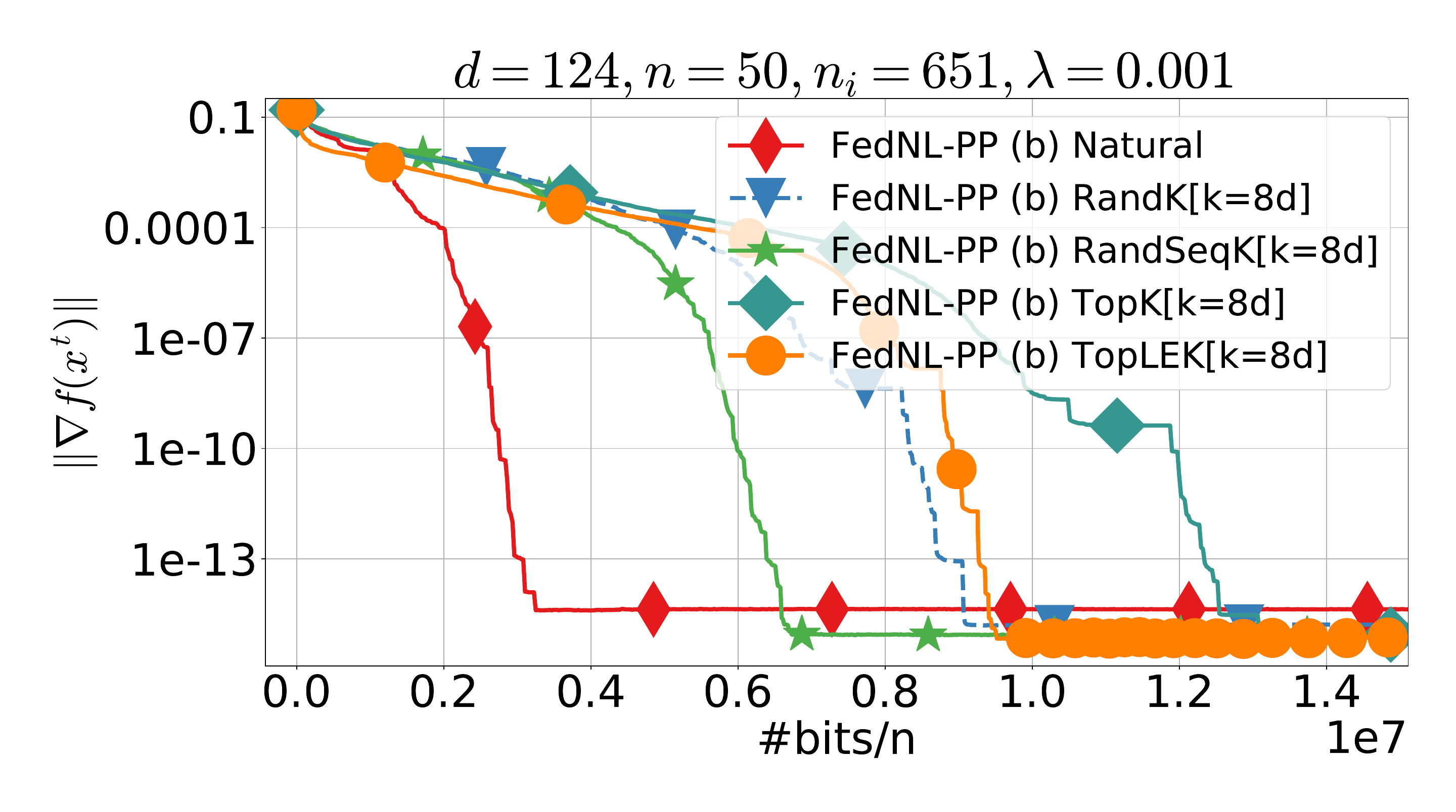}
\includegraphics[width=0.48\textwidth]{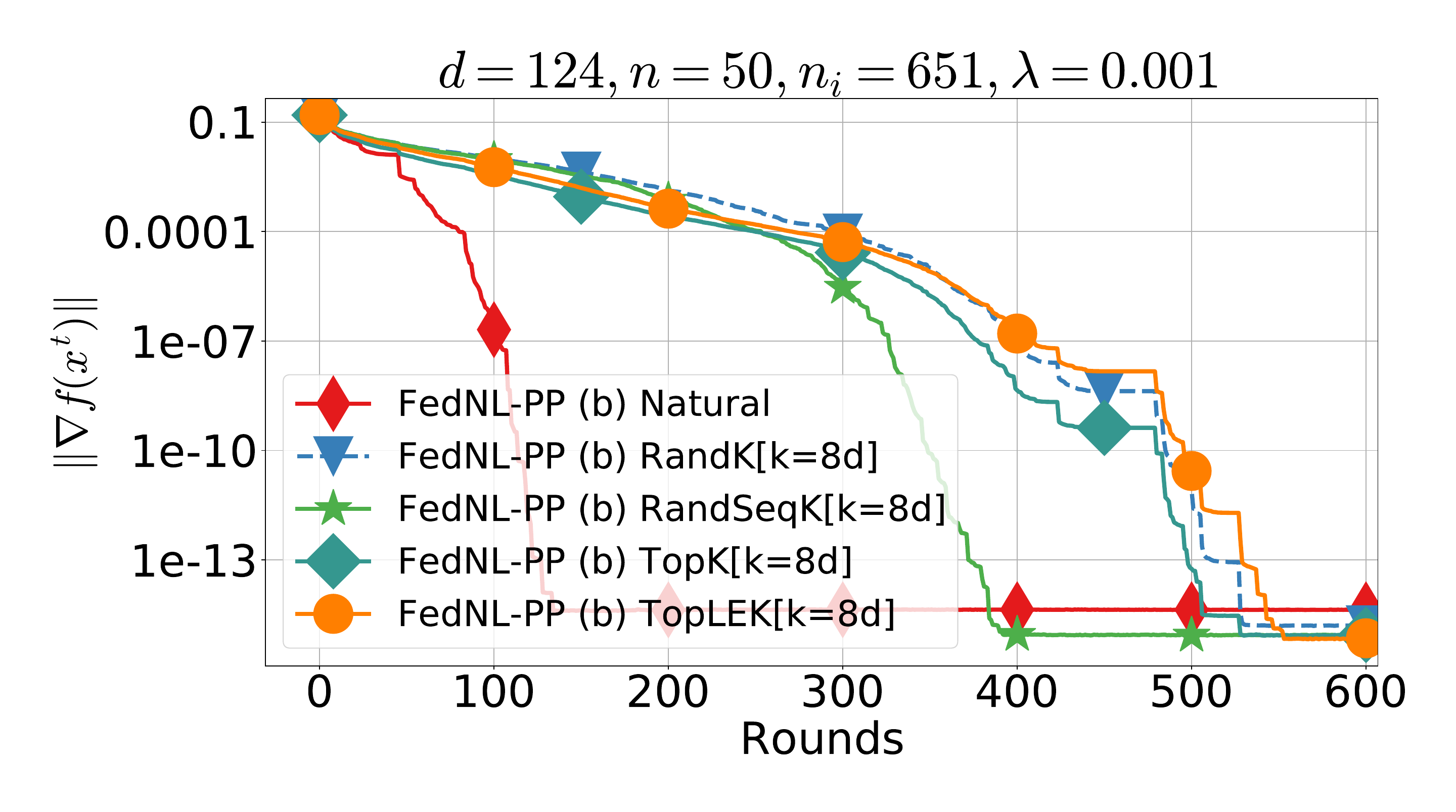}

\caption{\algname{FedNL-PP} at \dataname{A9A} in  multi-node setting, $n=50$, $|S^k|=12$ clients per round, FP64 arithmetic, 1 {CPU} core per node and master, \abr{TCP/IPv4}. \dataname{A9A} dataset reshuffled u.a.r. and augmented with intercept.}
\label{ch7:fig:fednl-pp-a9a-app}
\end{figure}


\begin{figure}[h]
\centering
\includegraphics[width=0.48\textwidth]{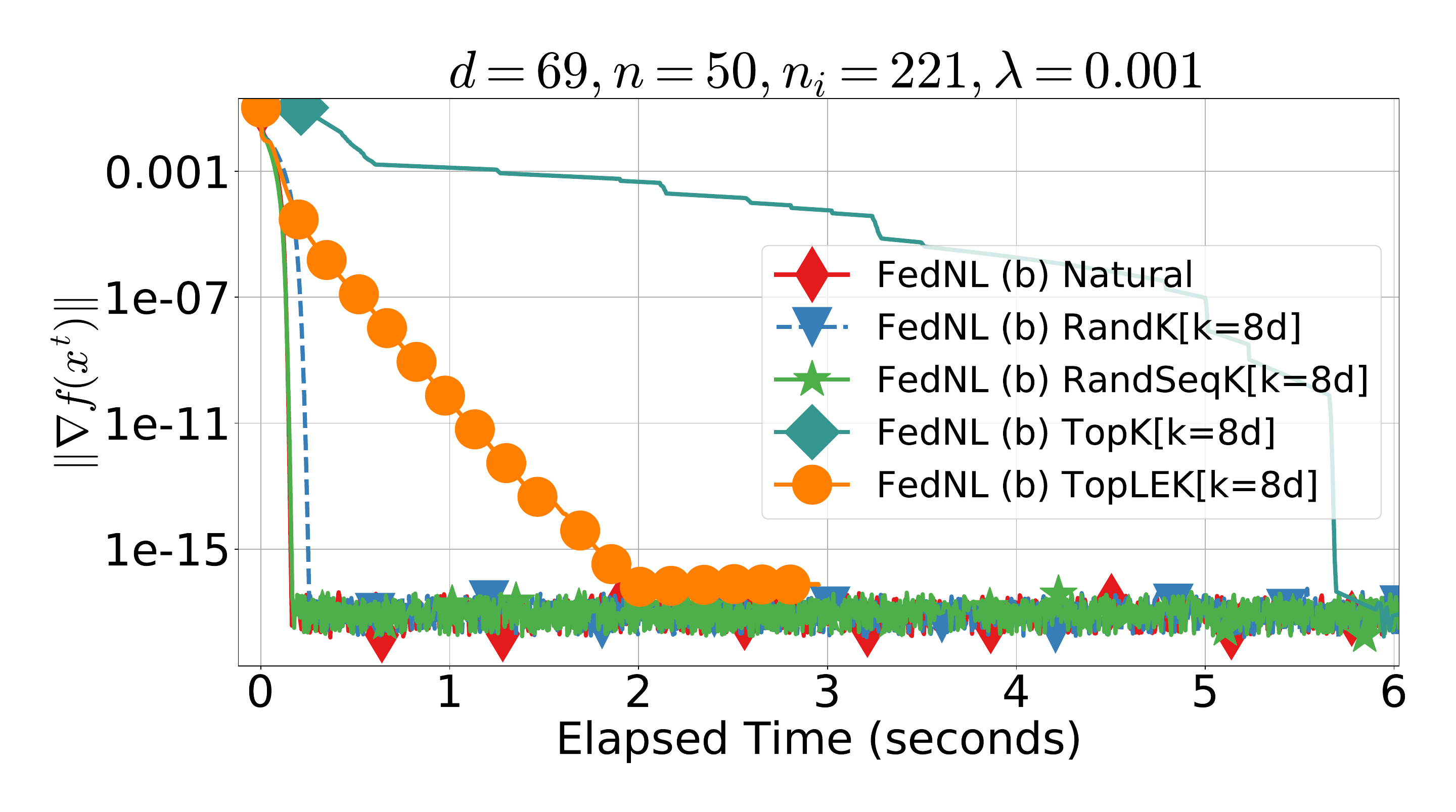}
\includegraphics[width=0.48\textwidth]{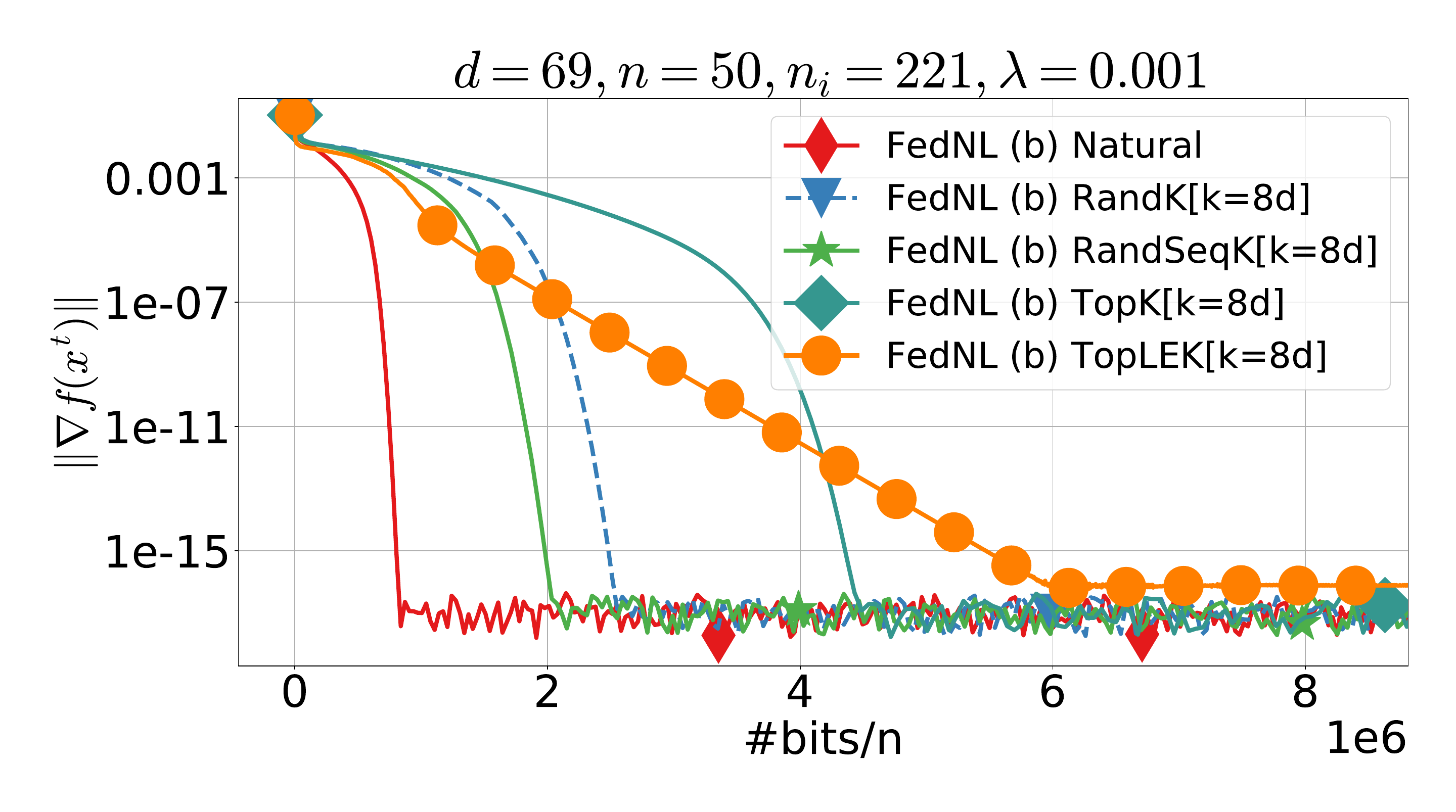}
\includegraphics[width=0.48\textwidth]{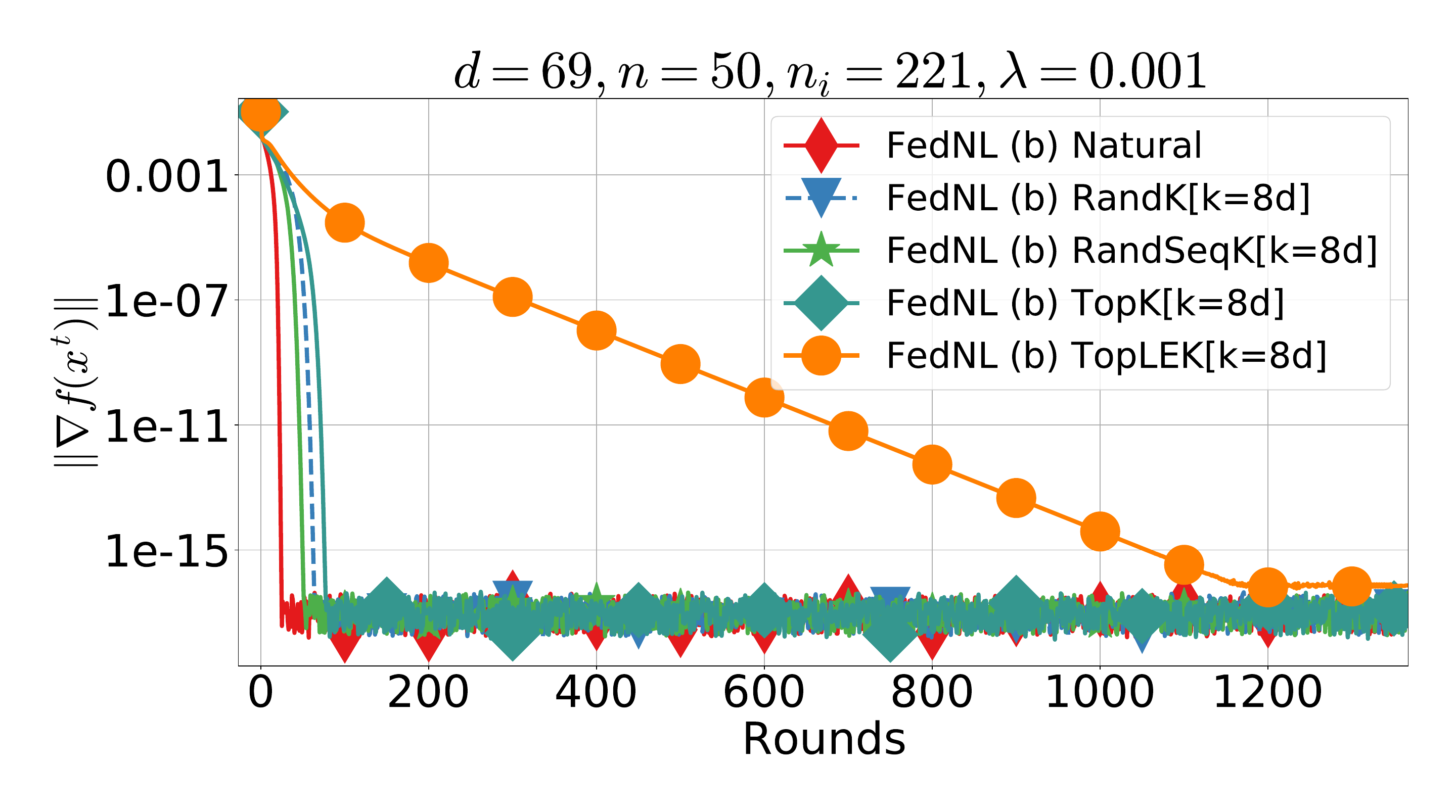}

\caption{\algname{FedNL} in multi-node setting, theoretical step size, $n=50$, FP64 arithmetic, 1 {CPU} core per node and master, \abr{TCP/IPv4}, dataset \dataname{PHISHING} reshuffled u.a.r. and augmented with intercept.}
\label{ch7:fig:fednl-phishing-app}
\end{figure}

\begin{figure}[h]
\centering
\includegraphics[width=0.48\textwidth]{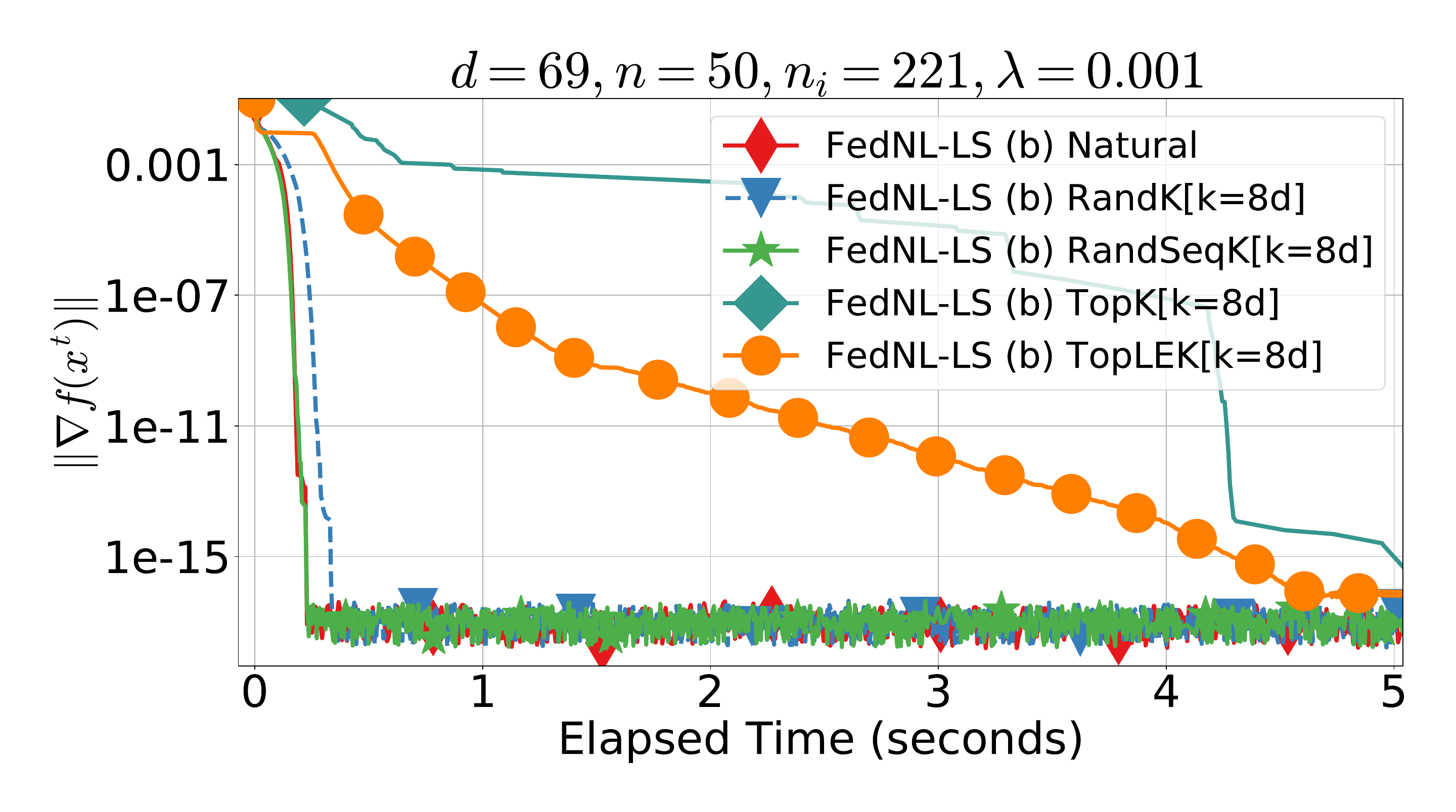}		
\includegraphics[width=0.48\textwidth]{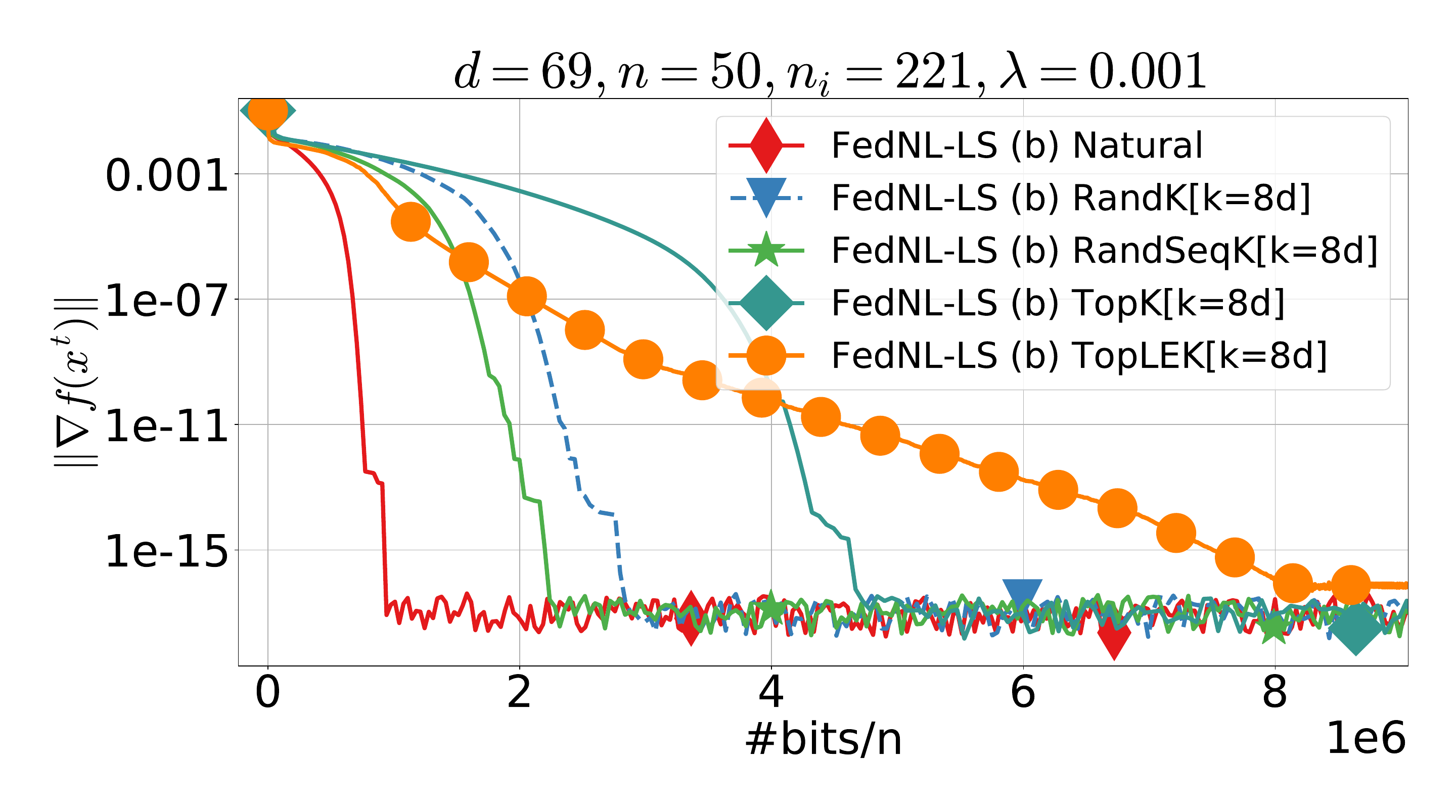}
\includegraphics[width=0.48\textwidth]{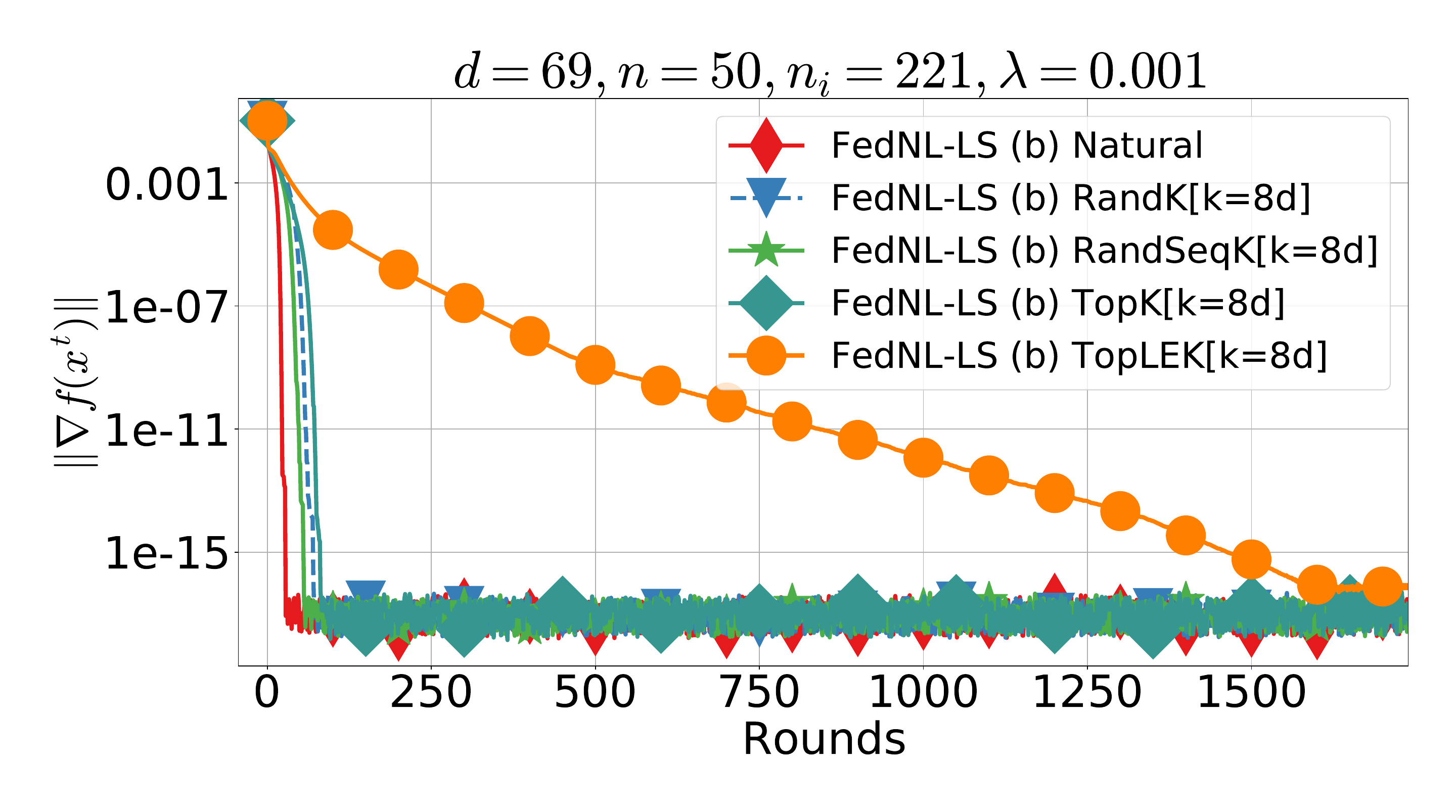}

\caption{ \algname{FedNL-LS} in multi-node setting, $n=50$, FP64 arithmetic, 1 {CPU} core per node and master, \abr{TCP/IPv4}, dataset \dataname{PHISHING} reshuffled u.a.r. and augmented with intercept. The line search parameters $c=0.49,\gamma=0.5$.}

\label{ch7:fig:fednl-ls-phishing-app}
\end{figure}

\begin{figure}[h]
\centering
\includegraphics[width=0.48\textwidth]{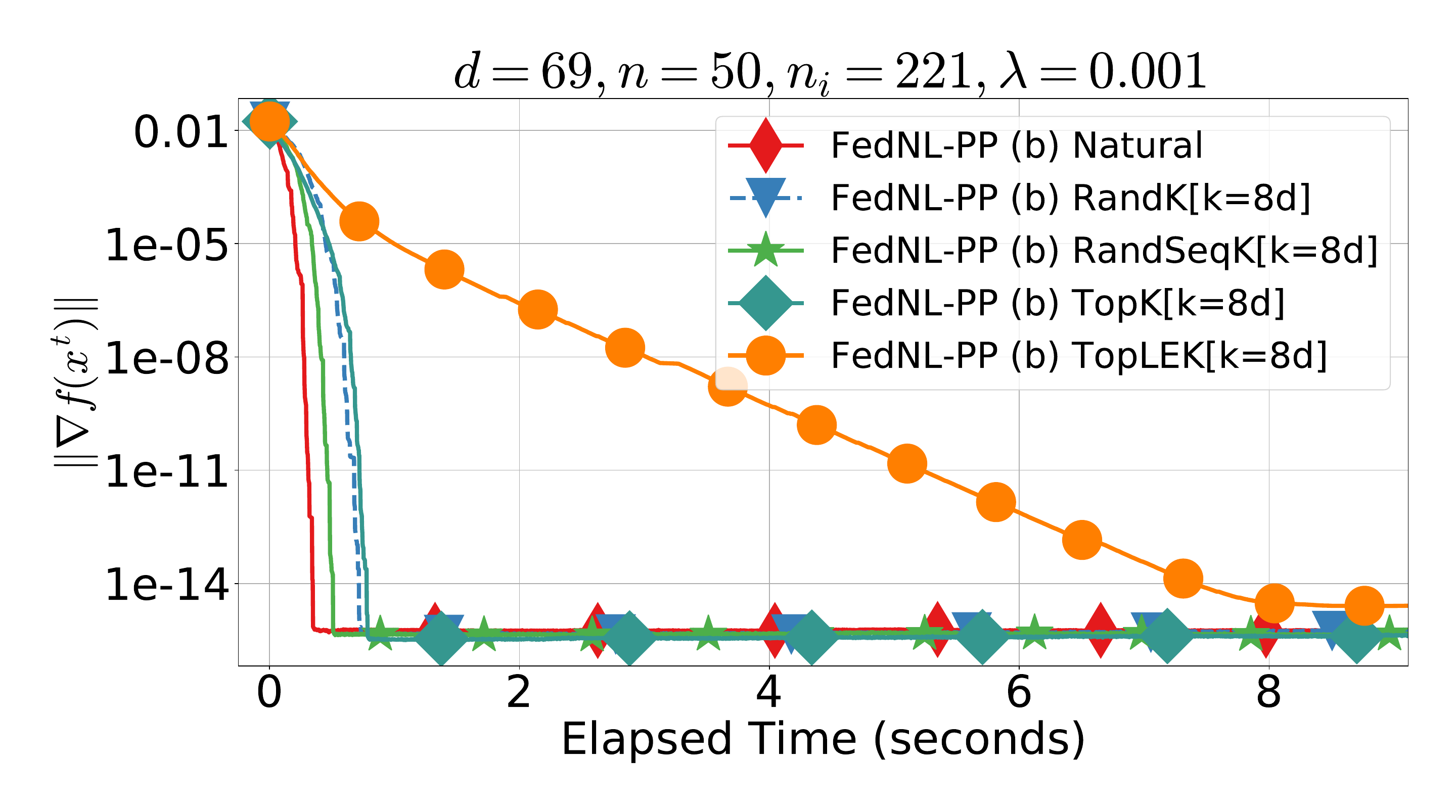}		
\includegraphics[width=0.48\textwidth]{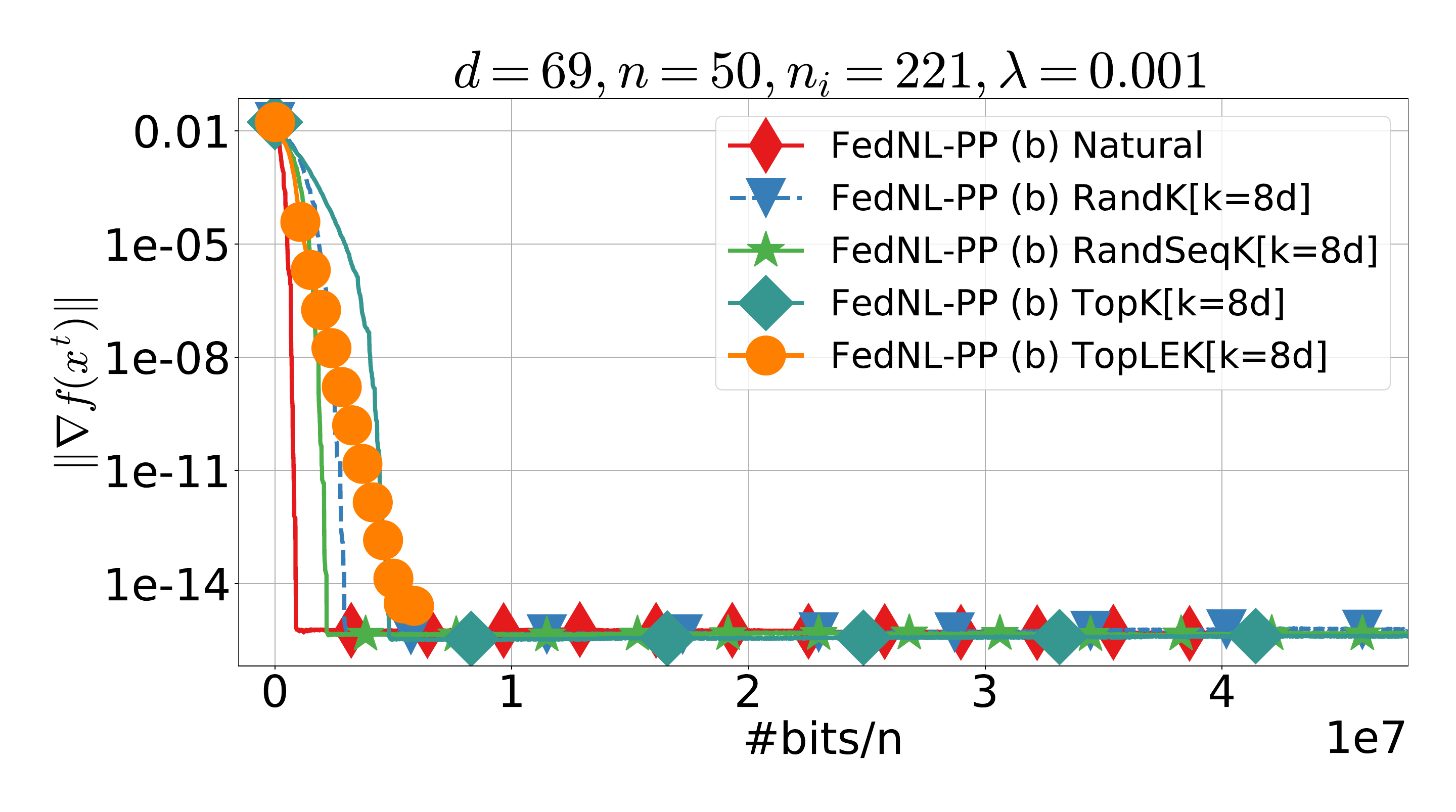}
\includegraphics[width=0.48\textwidth]{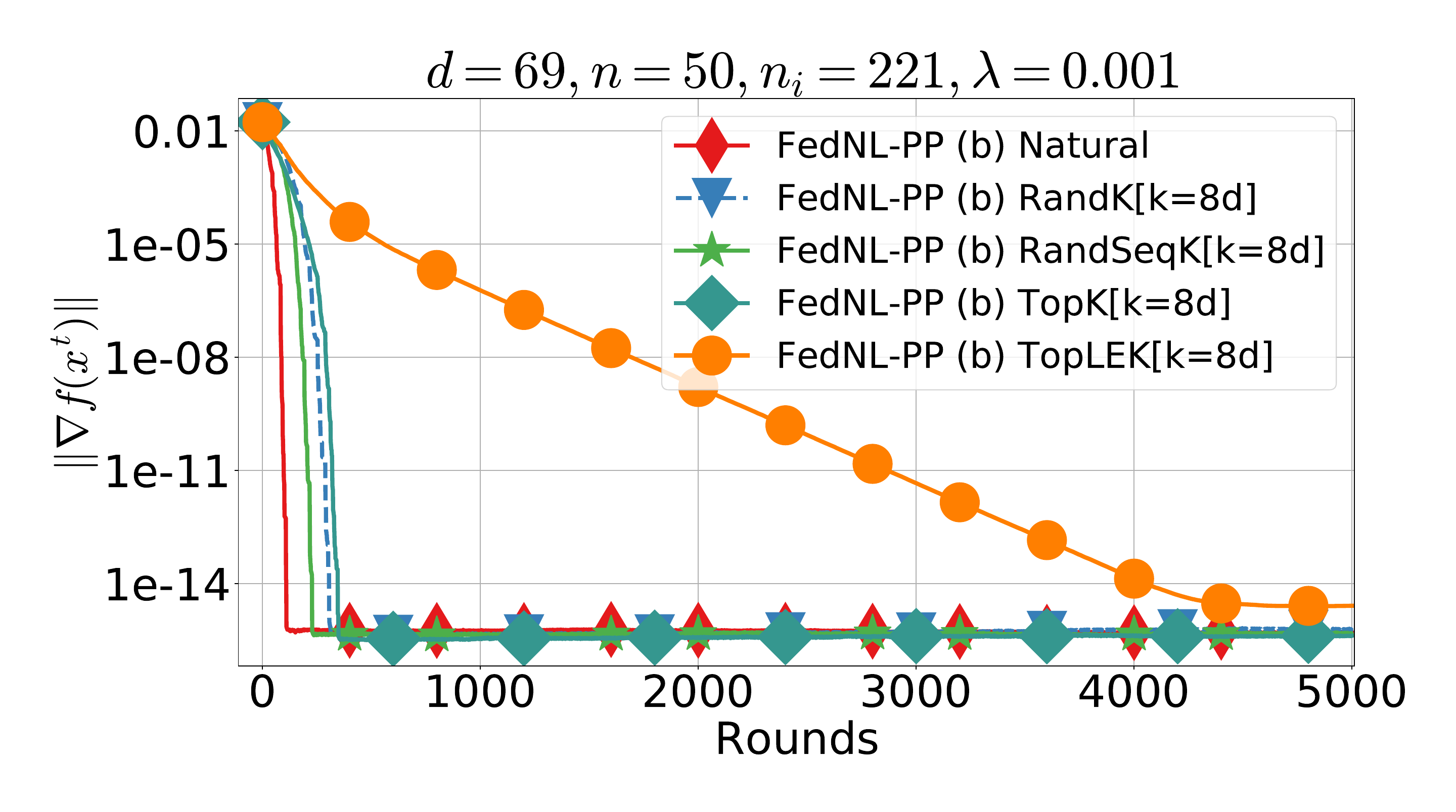} 
\caption{\algname{FedNL-PP} in multi-node setting, $n=50$, $|S^k|=12$ clients per round, FP64 arithmetic, 1 {CPU} core per node and master, \abr{TCP/IPv4}. \dataname{PHISHING} dataset reshuffled u.a.r. and augmented with intercept.}
\label{ch7:fig:fednl-pp-phishing-app}
\end{figure}


We have carried out experiments with \algname{FedNL}, \algname{FedNL-LS}, and with \algname{FedNL-PP} \algname{FedNL-PP}. The results for \dataname{W8A} are presented in Figures \ref{ch7:fig:fednl-w8a-app}, \ref{ch7:fig:fednl-ls-w8a-app}, \ref{ch7:fig:fednl-pp-w8a-app}, for \dataname{A9A} in Figures \ref{ch7:fig:fednl-a9a-app}, \ref{ch7:fig:fednl-ls-a9a-app}, \ref{ch7:fig:fednl-pp-a9a-app},  for \dataname{PHISHING} in Figures \ref{ch7:fig:fednl-phishing-app}, \ref{ch7:fig:fednl-ls-phishing-app}, \ref{ch7:fig:fednl-pp-phishing-app}.



In this experiment, we conducted the training of a \modelname{logistic regression} model, formally defined in Equation~\eqref{ch7:eq:fi_log_reg_structure}, incorporating $L_2$ regularization on three \dataname{LISVM} \citep{chang2011libsvm} benchmark datasets: \dataname{W8A}, \dataname{A9A}, \dataname{PHISHING}. The training was done within a distributed data center, utilizing the hardware detailed in Appendix~\ref{ch7:app:hardware-env-multipled-node}. Fifty clients were connected to a single master node, facilitating communication through a single \abr{TCP/IPv4} connection between client and master.

The \algname{FedNL} Algorithm~\ref{ch7:alg:FedNL}, particularly in Line 10, necessitated compute reduction for $s^k, l^k$. This reduction was implemented using various communication patterns, conceptualizing client-master communication in a star topology. The master collected information from clients in a centralized manner. Both clients and masters within the experimental setup operated with a single CPU core. Refer to Appendix~\ref{ch7:app:hardware-env-multipled-node} for a more comprehensive description of machine capabilities. In this scenario, we disabled the dedicated pool of workers for handling hierarchically gradient reduction and a pool of workers to parallelize the in-place update of the master Hessian.

Firstly, it is observed that the backtracking line search in \algname{FedNL-LS} slowed down the entire training process by a factor of $\times 1.14$, but it ensures global convergence.

The proposed \compname{RandSeqK} outperformed \compname{RandK}, showcasing better convergence as wall-clock time progressed, communication volume increased, and external iterations (rounds) progressed. This is due to two contributing factors. The  \compname{RandSeqK} is cache-aware and consistently reaches the super-linear convergence neighborhood empirically better. The proposed \compname{TopLEK} compressor emerged as the most economical method for sending information from client to master. While it did not enhance \compname{TopK} when measuring progress in terms of communicated bits, it demonstrated improvement when measuring actual time in \dataname{W8A} and \dataname{A9A} datasets. In the \dataname{PHISHING}, it is less effective compared to \compname{TopK} due to an unfortunate situation with catching the neighborhood slower.

Regarding \algname{FedNL-PP}, we caution readers against extrapolating wall-clock time beyond this experiment. The \algname{FedNL-PP} Algorithm~\ref{ch7:alg:FedNL-PP} lacks explicit support for the computation of $\nabla f(x^k)$ as part of the training process, and computation of full gradient brought the measured time overhead.

Although our primary goal initially centered on enhancing approaches from the paper \citet{safaryan2021fednl}, scientific curiosity prompted us to explore the application of the \compname{Natural} compressor. Designed to be unbiased and adapted to a real representation of $\mathbb{R}$ in FP32 and FP64 \citep{IEEE754-2008} with a small $w=1/8$ constant, this compressor was originally proposed within the realm of first-order compressed gradient descent methods. Remarkably, it exhibited favorable behavior for \algname{FedNL} as well. It is important to note that the \compname{Natural} compressor introduces technical challenges in effective implementation due to operating at the granularity of bits. Despite these, our implementation effectively supports it.


\clearpage
\addtocounter{adjsection}{1}
\section{Beyond Training Time: Relaxed Memory and System Requirements}

In this appendix, we present runtime overhead benchmarks conducted on Windows 11 operating system.

\begin{center}
\begin{table}[h!]
\footnotesize
\centering
\begin{threeparttable}
	\caption{Peak kernel handles. Single-node setting, FP64, Windows 11.}
	\begin{tabular}{|l|l|c|c|c|c|c|}
		\hline
		\textbf{\#} & \textbf{Solver} & \makecell{\textbf{\dataname{W8A}},\\\textbf{$d=301$}, \textbf{$n_i=350$}} & \makecell{\textbf{\dataname{A9A}}, \\\textbf{$d=124$}, \textbf{$n_i=229$}} & \makecell{\textbf{\dataname{PHISHING}}, \\\textbf{$d=69$}, \textbf{$n_i=77$}} \\
		\hline
		\hline
		1 & CLARABEL & 809 & 809 & 809 \\
		\hline
		2 & ECOS & 809 & 809  & 809 \\
		\hline
		3 & ECOS-BB & 811 & 811 & 811   \\
		\hline
		4 & SCS & 811 & 811 & 811   \\
		\hline
		5 & MOSEK & 1087 & 1087 & 1087   \\
		\hline 
		6 & \makecell[l]{\algname{FedNL} / RandK[$k=8d$]} & \cellcolor{bgcolorwe} 74 & \cellcolor{bgcolorwe} 74 & \cellcolor{bgcolorwe} 74   \\
		\hline
		7 & \makecell[l]{\algname{FedNL} / RandSeqK[$k=8d$]} & \cellcolor{bgcolorwe} 74 & \cellcolor{bgcolorwe} 74 & \cellcolor{bgcolorwe} 74 \\
		\hline 
		8 & \makecell[l]{\algname{FedNL} / TopK[$k=8d$]} & \cellcolor{bgcolorwe} 74 & \cellcolor{bgcolorwe} 74 & \cellcolor{bgcolorwe} 74  \\
		\hline 
		9 & \makecell[l]{\algname{FedNL} / TopLEK[$k=8d$]} & \cellcolor{bgcolorwe} 74 & \cellcolor{bgcolorwe} 74 & \cellcolor{bgcolorwe} 74  \\
		\hline
		10 & \makecell[l]{\algname{FedNL} / Natural} & \cellcolor{bgcolorwe} 74 & \cellcolor{bgcolorwe} 74 & \cellcolor{bgcolorwe} 74   \\
		\hline
		11 & \makecell[l]{\algname{FedNL} /Ident} & \cellcolor{bgcolorwe} 74 & \cellcolor{bgcolorwe} 74 & \cellcolor{bgcolorwe} 74 \\
		\hline	
	\end{tabular}
	\label{ch7:tbl:compare-vs-cvxpy-sys-handles}
\end{threeparttable}
\end{table}
\end{center}

\paragraph{Peak kernel handles.} Stacking layers of scientific software on top of each other leads to a sufficiently big amount of OS kernel objects\footnote{Examples of kernel objects include files, memory-mapped file views, processes, threads, semaphores, and mutexes. These objects belong to the operating system, but they are created and managed in response to process activity.}. Each kernel handle (in any OS) consumes system resources, such as kernel memory and system table entries. As we see from Table~\ref{ch7:tbl:compare-vs-cvxpy-sys-handles} \algname{FedNL} simulation imposes a smaller number of kernel handles (responsible for file objects, thread objects, and others). {Every kernel object occupies memory allocated within the kernel space of the OS and is managed by it. Regardless of whether the memory is paged (potentially swappable to disk) or pinned (never writable to disk under any circumstances), the object still consumes resources within the kernel.}

\begin{center}
\begin{table}[h!]
\centering
\footnotesize
\begin{threeparttable}
	\caption{Peak private bytes. Single-node setting, FP64, Windows 11.}
	\begin{tabular}{|l|l|c|c|c|c|c|}
		\hline
		\textbf{\#} & \textbf{Solver} & \makecell{\textbf{\dataname{W8A}},\\\textbf{$d=301$}, \textbf{$n_i=350$}} & \makecell{\textbf{\dataname{A9A}}, \\\textbf{$d=124$}, \textbf{$n_i=229$}} & \makecell{\textbf{\dataname{PHISHING}},\\\textbf{$d=69$}, \textbf{$n_i=77$}} \\
		\hline
		\hline
		1&CLARABEL & 5 818 936 K & 5 472 376 K & 5 248 824 K \\
		\hline
		2&ECOS & 5 681 656 K & 5 387 576 K  & 5 192 228 K \\
		\hline
		3&ECOS-BB & 5 681 936 K & 5 388 092 K & 5 192 048 K   \\
		\hline
		4&SCS & 5 859 308 K & 5 546 052 K & 5 196 608 K   \\
		\hline
		5&MOSEK & 6 685 032 K & 5 945 336 K & 5 718 484 K   \\
		\hline 
		6&\makecell[l]{\algname{FedNL} / RandK[$k=8d$]} & \cellcolor{bgcolorwe} 770 740 K & \cellcolor{bgcolorwe} 192 296 K & \cellcolor{bgcolorwe} 47 284 K   \\
		\hline
		7&\makecell[l]{\algname{FedNL} / RandSeqK[$k=8d$]} & \cellcolor{bgcolorwe} 746 072 K & \cellcolor{bgcolorwe} 191 516 K & \cellcolor{bgcolorwe} 45 236 K \\
		\hline 
		8&\makecell[l]{\algname{FedNL} / TopK[$k=8d$]} & \cellcolor{bgcolorwe} 745 072 K & \cellcolor{bgcolorwe} 192 208 K & \cellcolor{bgcolorwe} 45 484 K  \\
		\hline 
		9&\makecell[l]{\algname{FedNL} / TopLEK[$k=8d$]} & \cellcolor{bgcolorwe} 745 868 K & \cellcolor{bgcolorwe} 192 908 K & \cellcolor{bgcolorwe} 45 160 K  \\
		\hline
		10&\makecell[l]{\algname{FedNL} / Natural} & \cellcolor{bgcolorwe} 806 256 K & \cellcolor{bgcolorwe} 199 792 K & \cellcolor{bgcolorwe} 45 568  K   \\
		\hline
		11&\makecell[l]{\algname{FedNL} / Ident} & \cellcolor{bgcolorwe} 805 996 K & \cellcolor{bgcolorwe} 199 744 K & \cellcolor{bgcolorwe} 46 856 K \\
		\hline	
	\end{tabular}
	\label{ch7:tbl:compare-vs-cvxpy-vir-memory}
\end{threeparttable}
\end{table}
\end{center}
\paragraph{Used virtual memory during execution.} As we see from Table~\ref{ch7:tbl:compare-vs-cvxpy-vir-memory} there is a substantial requirement for virtual memory for launching Python scripts. Unfortunately, even a numerical library by itself such as \libname{NumPy} requires 512 MBytes of virtual address space. The reasons are: (a) \libname{NumPy} relies on numerical libraries that cumulatively have not negligible size; (b) \libname{NumPy} initializes big-size internal data structures; (c) The Python ecosystem is completely built on dynamic (shared) libraries (See also Appendix~\ref{ch7:app:nopython}).

\begin{center}
\begin{table}[h!]
\centering
\footnotesize
\begin{threeparttable}
	\centering
	\caption{Peak working set (resident set) Size. Single-node, FP64, Windows 11.}
	\begin{tabular}{|l|l|c|c|c|c|c|}
		\hline
		\textbf{\#} & \textbf{Solver} & \makecell{\textbf{\dataname{W8A}},\\\textbf{$d=301$}, \textbf{$n_i=350$}} & \makecell{\textbf{\dataname{A9A}}, \\\textbf{$d=124$}, \textbf{$n_i=229$}} & \makecell{\textbf{\dataname{PHISHING}}, \\\textbf{$d=69$}, \textbf{$n_i=77$}} \\
		\hline
		\hline
		1&CLARABEL & 816 248 K & 522 592 K & 323 364 K \\
		\hline
		2&ECOS & 668 152 K & 408 440 K  & 256 144 K \\
		\hline
		3&ECOS-BB & 668 328 K & 409 080 K & 255 916 K   \\
		\hline
		4&SCS & 813 080 K K & 504 560 K & 281 220 K   \\
		\hline
		5&MOSEK & 1 216 172 K & 706 616 K & 415 476 K   \\
		\hline 
		6&\makecell[l]{\algname{FedNL} / RandK[$k=8d$]} & \cellcolor{bgcolorwe} 750 740 K & \cellcolor{bgcolorwe} 187 888 K & \cellcolor{bgcolorwe} 47 204 K   \\
		\hline
		7&\makecell[l]{\algname{FedNL} / RandSeqK[$k=8d$]} & \cellcolor{bgcolorwe} 726 140 K & \cellcolor{bgcolorwe} 187 552 K & \cellcolor{bgcolorwe} 44 168 K \\
		\hline 
		8&\makecell[l]{\algname{FedNL} / TopK[$k=8d$]} & \cellcolor{bgcolorwe} 725 512 K & \cellcolor{bgcolorwe} 187 740 K & \cellcolor{bgcolorwe} 44 560 K  \\
		\hline                                    
		9&\makecell[l]{\algname{FedNL} / TopLEK[$k=8d$]} & \cellcolor{bgcolorwe} 725 380 K & \cellcolor{bgcolorwe} 188 416 K & \cellcolor{bgcolorwe} 44 260 K  \\
		\hline                                    
		10&\makecell[l]{\algname{FedNL} / Natural} & \cellcolor{bgcolorwe} 745 532 K & \cellcolor{bgcolorwe} 189 988 K & \cellcolor{bgcolorwe} 45 532 K   \\
		\hline                                    
		11&\makecell[l]{\algname{FedNL} / Ident} & \cellcolor{bgcolorwe} 785 960 K & \cellcolor{bgcolorwe} 195 892 K & \cellcolor{bgcolorwe} 45 732 K \\
		\hline	
	\end{tabular}
	\label{ch7:tbl:compare-vs-cvxpy-ram-memory}
\end{threeparttable}
\end{table}
\end{center}

\paragraph{Peak working set size.} In Windows OS, the \textit{"peak working set size"} refers to the maximum amount of physical memory that a process acquires. We see from Table~\ref{ch7:tbl:compare-vs-cvxpy-ram-memory} \algname{FedNL} on average exhibits a smaller peak working set by a factor $\times 1.5$ in \dataname{W8A} dataset, and by a factor $\times 6$
in \dataname{PHISHING} dataset.

\clearpage
\addtocounter{adjsection}{1}
\section{Hardware Environment for Experiments}
\label{ch7:app:hardware-env}

\subsection{Single-node experiment setup}
\label{ch7:app:hardware-env-single-node}

\begin{enumerate}
\item {CPU}: Intel(R) Xeon(R) Gold 6246 CPU $3.3$ GHz; Byte Order: Little-Endian.
\item {OS}: Ubuntu 18.04.6 LTS, bionic, x86-64; Linux kernel version: 5.4.0-150-generic.
\item Physical memory: 251 GB DDR4 Synchronous memory at $2933$ MHz.
\item Hard Drive: INTEL SSD SC2 KG03 Disk with a physical sector size of 4096 bytes.
\item Disk write speed: 347 MBytes/s; disk read speed: 2.5 GBytes/s; File System: EXT4.
\end{enumerate}

\subsection{Multi-node experiment setup}
\label{ch7:app:hardware-env-multipled-node}

\begin{enumerate}

\item {CPU}: Intel(R) Xeon(R) Gold 6148 CPU $2.50$ GHz; Byte Order: Little-Endian.
\item {OS}: Rocky Linux 9.1 (Blue Onyx), x86-64; Linux kernel version: 5.14.0
\item Physical memory: 375 GB.
\item Disk write speed: 109 MB/s; Disk read speed: 3.8 GB/s; File System: Network File System v4.
\item Network Interface: MTU Size $1500$ bytes, download 293.18 Mbit/s, upload 476.79 Mbit/s.
\item The experiments in Section \ref{ch7:sec:multi-node-cmp-vs-spark} were conducted in a compute cluster managed by the Slurm Workload Management system.
\item Masters nodes: $1$; workers nodes: $50$; number of CPU cores/node: $1$.
\end{enumerate}

\subsection{System preparation for reliable time measurements and experimental reproducibility}
\label{ch7:app:careful-reproduce}

Our experiments were conducted on a system equipped with \textit{Intel Xeon Gold 6246 CPU} \footnote{\href{https://ark.intel.com/content/www/us/en/ark/products/193969/intel-xeon-gold-6246-processor-24-75m-cache-3-30-ghz.html}{Intel Xeon Gold 6246 Processor Technical Specification}}. The CPU has a cache line size of $64$ bytes, $12$ physical cores, and $24$ logical cores, with a core frequency from $3$ GHz to $4.4$ GHz\footnote{The \textit{frequency} is the inverse of the clock \textit{period}, where the clock \textit{period} is the time required to complete one full cycle of the synchronization signal. The typical physical device that generates this signal for electrical circuits is a \textit{crystal oscillator} \citep{harris2015digital}.}. To ensure accurate time measurements, we did the following steps:

\begin{enumerate}
\item \textit{{Disabled Intel Turbo Boost.}} This technology dynamically increases the CPU clock speed above the base frequency based on workload demands. Disabling Turbo Boost ensures a more predictable and consistent measurement.

\item \textit{Disabled Simultaneous Multithreading (SMT).} SMT enables multiple execution threads to run on a single physical CPU core. We disabled SMT to ensure consistent performance measurement and minimize related fluctuations.

\item \textit{{Disabled Intel CPU Frequency scaling (CPU power management).}} This technology allows the operating system to adjust the CPU frequency dynamically.

\item \textit{{Fixed CPU frequency.}} The CPU frequency during the experiments was set at a stable $3.3$ GHz clock rate using the \texttt{cpupower frequency-set} command-line utility in Linux.

\end{enumerate}

\clearpage
\addtocounter{adjsection}{1}
\section{Software Environment for Experiments}
\label{ch7:app:software-env}

\subsection{Software environment for baselines}
\label{ch7:app:software-python-env}

Version information for used Python libraries: NumPy version: 1.26.2; Pandas version: 2.1.4; scikit-learn version: 1.3.2; Matplotlib version: 3.8.2; Ray version: 1.26.2; Cython version: 3.0.8; CVXPY version: 1.4.1;
Mosek version: 10.1.21; CVXPY solvers: CBC, CLARABEL, CVXOPT, ECOS, ECOS\_BB, GLOP,	GLPK, GLPK\_MI, GUROBI,	MOSEK, OSQP, PDLP,	SCIP, SCIPY, SCS, XPRESS.

Version information for utilized Python interpreter:
\begin{itemize}
\item Python version: 3.9.1 (default, Dec 11 2020), Linux/5.4.0-150-generic.
\item Python interpreter compiled with GCC 7.3.0, CPU architecture: x86-64.
\end{itemize}

Version information for utilized Apache Spark:
\begin{itemize}
\item Apache Spark 3.5.0; PySpark version: 3.5.0.
\item Java Runtime: OpenJDK Runtime Environment (build 11.0.19).
\end{itemize}

Cluster management system for utilized multi-node settings:
\begin{itemize}
\item Experiments from Section \ref{ch7:sec:multi-node-cmp-vs-spark} were conducted in a compute cluster managed by the Slurm Workload Management system with version: 23.02.6.
\item Linux OS Kernel: Linux 5.14.0.
\end{itemize}

\subsection{Building single-node and multi-node FedNL implementation}
\label{ch7:app:build-fednl-for-exp}

For wall-clock time measurements, our C++20 \algname{FedNL} implementation was built using the following:

\begin{enumerate}
\item {Target architecture:} AMD, Intel x86 64-bit.
\item {CMake version:} 3.27.0-rc5; {Linux GNU Make version:} 4.1.
\item {Compiler:} GNU GCC version 11.4.0 with compiling C++20.
\item Optimization flags: \textit{{-O3 -fno-rtti -fno-exceptions -flto -march=native -mtune=native -pthread}}.
\item {Floating-point format}: FP64. All computations involving matrix-matrix, matrix-vector, and vector-scalar operations are performed in FP64 (double precision)
\item {CPU extension:} The compute-intensive sections of the code leverage SIMD {CPU} extension.
\end{enumerate}

\clearpage
\addtocounter{adjsection}{1}
\section{System Requirements for Unlocking FedNL}
\label{ch7:app:system-requirements-for-ufednl}

\subsection{Supported operating systems and compilers}
\label{ch7:app:supported-os-and-compilers}

Our solution is designed to run on various desktop operating systems, including:

\begin{enumerate}
\item Microsoft Windows (Windows 10, Windows 11, and higher).
\item Linux Operating Systems (Ubuntu 18.04, 22.04, and higher).
\item macOS Monterey (12.6.1 and higher).
\end{enumerate}

The supported toolchains for building our code are the following:

\begin{enumerate}
\item Visual Studio/MSVC (minimum: VS2022, MSVC 19.30).
\item GNU GCC/G++ (minimum: 11.4).
\item LLVM CLANG (minimum: 10.0).
\item CUDA Support (minimum: CUDA Toolkit 12.4).
\end{enumerate}

Our code adheres to the C++ 2020 standard\footnote{
\href{https://www.iso.org/standard/79358.html}{ISO/IEC 14882:2020 C++ standard.}}. To facilitate the build process, we provide a helper script \texttt{project\_scripts.py} written in Python 3.9.4. This script helps with the following tasks:

\begin{enumerate}
\item Generate release/debug project files via CMake. Please use CMake (minimum: 3.12).
\item Launch the building process from the command line and parallelize it.
\item Speed up the build with Ninja (GNU Make alternative) and CMake Unity/Jumbo build.
\item Enable the usage of precompiled headers to expedite the build.
\item Enable LLVM optimization compiler remarks for study.
\item Launch unit tests, GNU code coverage, and documentation generation via Doxygen.
\end{enumerate}	

\subsection{Supported CPU architectures}
\label{ch7:app:cpus}

The code can be compiled to support the following processor supplementary instruction sets:

\begin{enumerate}
\item \textit{{SSE2 (128-bit vector registers)}} - supported by Intel (Pentium 4).
\item \textit{{AVX2 (256-bit vector registers)}} - supported by Intel (Sandy Bridge), AMD (Bulldozer).
\item \textit{{AVX-512 (512-bit vector registers)}} - supported by Intel (Knights Landing), AMD (Zen 4).
\item \textit{{ARM Neon (128-bit vector registers)}} - supported by many ARMv7 processors.
\item \textit{{No special instructions}} - explicitly turn off usage of any computing vectorization.
\end{enumerate}

\paragraph{Processors with Single Instruction Multiple Data used for testing.}

\begin{enumerate}
\item {Intel CPUs with x86-64 architecture}: Core i7-10875H, Xeon Gold 6246, Xeon Platinum 8272CL, Core i7-8700B; byte order: Little-Endian.
\item {CPUs with {ARM AArch64} architecture}: ARMv8 Neoverse N1 R3P1 Google; byte order: Little-Endian.
\end{enumerate}

\clr
{
\paragraph{List of tested processor architectures for implementation.}
We tested our implementation across the following CPU architectures on Linux/Ubuntu OS:
\begin{enumerate}
\item AMD64/x86-64: 64-bit logical address; byte order: Little-Endian.
\item ARM v7: 32-bit logical address; byte order: Little-Endian.
\item ARM64 v8: 64-bit logical address; byte order: Little-Endian.
\item PowerPC PPC64LE: 64-bit logical address; byte order: Little-Endian.
\item RISC-V 64: 64-bit logical address; byte order: Little-Endian.
\item IBM Z Series Architecture (S390X): 64-bit logical address; byte order: Big-Endian.
\end{enumerate}

We \algname{FedNL} on MacBooks with the following CPU architectures:

\begin{enumerate}
\item ARM64 v8 (Apple M1): 64-bit logical address; byte order: Little-Endian.
\item x86-64 (Intel Core-i7 1068NG7): 64-bit logical address; byte order: Little-Endian.
\end{enumerate}

This demonstrates that the \algname{FedNL} implementation is fully cross-platform, capable of running across multiple operating systems, CPU architectures, integer word sizes, and endianness of CPU configurations.
}

\paragraph{A Comparative analysis of x86-64 and ARM Instruction Sets.}

The x86-64 Instruction Set Architecture (ISA) is widely used in desktops, laptops, and servers. Processors from vendors such as Intel and AMD, including the Intel Core and AMD Ryzen series, are commonly integrated into systems based on the x86-64 architecture. The ARM AArch64 Instruction Set Architecture (also known as A64) is primarily used in mobile and embedded devices. Additionally, ARM’s energy-efficient design has contributed to the growing adoption of ARM-based servers.

Some practical differences between ARM and x86 assembly languages include the following:
\begin{enumerate}
	\item AArch64 has a dedicated register for storing return addresses (the link register $x30$), whereas x86-64 does not.
	\item AArch64 allows arithmetic instructions to avoid modifying CPU flags, whereas x86-64 does not.
	\item AArch64 includes a dedicated zero register.
	\item Both x86-64 and AArch64 support memory load/store instructions that incorporate indexing directly.
	\item Both x86-64 and AArch64 support built-in conditional execution within some instructions.
\end{enumerate}

The increasing adoption of the ARM ISA is primarily driven by its energy efficiency and simpler hardware implementation compared to x86-64, which follows a complex instruction set architecture (CISC). While x86-64 complex instructions can simplify high-level operations, they often come at the cost of increased power consumption. 

Right now, both x86-64 and ARM remain widely used processor architectures in modern computing, and our implementation supports both of them. We also tested our implementation on several other CPU architectures, as described earlier.

\subsection{Supported and tested GPU architectures}
\label{ch7:app:gpus}

Our current GPU support is limited. We primarily focused on the minimal support of creating oracles for optimization problems using NVIDIA CUDA\footnote{\href{https://developer.nvidia.com/cuda-toolkit}{https://developer.nvidia.com/cuda-toolkit} - NVIDIA CUDA Toolkit provides an environment for creating high-performance, GPU-accelerated applications for NVIDIA GPUs.}, providing minimal support for expressing computation in dense and non-structured linear algebra primitives. The used version of the NVIDIA driver is 560.35.03. Functionality has been tested on: 

\begin{enumerate} 
\item {Rocky Linux 9.1, x86-64, Kernel: Linux 5.14.0-162.23.1.el9}.
\item {Windows 11, x86-64, Versio 23H2, OS Build: 22631.4169}.
\end{enumerate}

With GPUs featuring the following microarchitectures:

\begin{enumerate}
\item NVIDIA Volta microarchitecture, CUDA Compute Capability: 7.0.
\item NVIDIA Turing microarchitecture, CUDA Compute Capability: 7.5.
\item NVIDIA Ampere microarchitecture, CUDA Compute Capability: 8.0, 8.6.
\item NVIDIA Ada Lovelace microarchitecture, CUDA Compute Capability: 8.9.
\item NVIDIA Hopper microarchitecture, CUDA Compute Capability: 9.0.
\end{enumerate}

In addition, we provide several executable applications for system introspection to detect OpenCL and CUDA-compatible devices and their characteristics (see Table~\ref{ch7:tab:auxiliary-binaries} in Appendix~\ref{ch7:app:futresearch}).

\clearpage
\addtocounter{adjsection}{1}
\section{Backgrounds}
\label{ch7:app:backgrounds}

\subsection{Memory latency variance in computing systems}
\label{ch7:app:memory-hierachy-latencies}

We present a minimum background that is enough to motivate  \compname{RandSeqK} in Appendix~\ref{ch7:app:seqk}. For more about Compute Architecture see \citet{hennessy2011computer}. Modern computing systems are equipped with various storage devices. The following table, adapted from \citet{gregg2014systems}, illustrates typical latency scales for them. More hardware-focused comparisons can be found in digital circuits design literature \citep{harris2015digital}.

\begin{table}[h!]
\centering
\footnotesize
\caption{Memory latency comparison in computing devices.}
\begin{tabular}{|c|c|c|}
\hline
\textbf{Device and Memory Level} & \makecell[c]{\textbf{Approximate} \\ \textbf{Latency} \textbf{(ns)}} & \textbf{Scale} \\
\hline
\hline
CPU cycle & $0.3$ & $x1$ \\
\hline
\clrshort{CPU register (SRAM)} & $0.3$ & $\times 1$ \\
\hline
L1 cache (SRAM) & $0.9$ & $\times 3$ \\
\hline
\clrshort{\makecell[c]{Floating Point \\addition, subtraction, and multiplication}} & $1.2$ & $\times 4$
\\
\hline
L2 cache (SRAM) & $3$ & $\times 10$ \\
\hline
L3 cache (SRAM) & $10$ & $\times 33$ \\
\hline			
\makecell[c]{Main memory or \\ Physical Memory (DRAM)} & $100$ & $\times 330$ \\
\hline
\clrshort{\makecell[c]{The OS System Call:\\ Transitioning from user to kernel space}} & $300$ & $\times 1000$ \\
\hline
Solid-State Disk (SSD) & $10\,000$ & $\times 33\,000$ \\
\hline
Rotational Hard Disk Drive (HDD) & $10\,000\,000$ & $\times 33\,000\,000$ \\
\hline
\end{tabular}
\label{ch7:tbl:latencies-for-memory}
\end{table}

\clr
{
In modern CPUs, the functional units responsible for floating-point addition, subtraction, and multiplication typically exhibit a latency of 4 clock cycles \citep{fog_instruction_tables}. While Big-O notation abstracts away many performance details, real-world execution is heavily influenced by data access times, especially when retrieval from disk or I/O is involved. As shown in Table~\ref{ch7:tbl:latencies-for-memory}, performance can vary significantly depending on the primary location of the data.	
}

\paragraph{A Hard disk drive (HDD).} HDD comprises one or more rigid disks or platters, with a head that moves to the correct location on the disk to read or write data magnetically as the disk rotates. Due to its partial mechanical nature, an HDD exhibits relatively slow access speeds. In contrast, Solid State Drives (SSDs) use flash memory, and these types of storage devices have an improved access latency compared to HDD. Only  HDD and SSD from Table~\ref{ch7:tbl:latencies-for-memory} store data persistently. 

\paragraph{The OS system call.} An OS system call is initiated by a process's thread to request the kernel to perform privileged instructions or system tasks, such as interacting with peripherals through device drivers. The transition from user space to kernel space for a specific thread is known as a \textit{mode switch}. In modern OS architectures, threads typically have both a user-level stack and a kernel stack, with the latter used to store temporary data during driver routine execution in kernel mode. While the execution time of the system calls itself can be not a major concern (see Table~\ref{ch7:tbl:latencies-for-memory}), the transition can trigger thread rescheduling, leading to cold and capacity cache misses in the CPU. One way to reduce the number of system calls is to utilize advanced OS primitives available in user space, such as memory mapping for file read operations when possible (see Section \ref{ch7:app:mmap}).

\paragraph{Dynamic random access memory (DRAM).} Dynamic Random Access Memory serves as a memory for non-persistent storage. It is structured as a series of memory chips and physically stores data typically as capacitor charge. To safeguard against data corruption, additional logic may store bits for error correction. Accessing DRAM is facilitated through the Memory Controller, which manages access to the primary Memory Chips when the data is not in the cache.

\paragraph{The CPU memory cache.} The CPU memory cache reduces latency in accessing DRAM by temporarily storing frequently accessed data. It is typically implemented using Static Random-Access Memory (SRAM), which relies on semiconductors for its electrical circuits. In SRAM, each bit of data is typically stored in cross-coupled inverters \citep{harris2015digital}, typically using six transistors. While both SRAM and DRAM lose their stored data when the supply voltage is off, they differ in the materials used and their operational mechanisms. If the requested data is not found in the cache or DRAM, the processor retrieves it from a backup stored in virtual memory on disk.
In general, low-level cache optimization at the software level is challenging due to the complexity of modern CPU architectures, where cache management is often opaque. Two contemporary strategies for optimizing memory access in algorithms designed to run on CPUs are: (i) \textit{cache-aware} algorithms, which take into account the existence and structure of caches, and (ii) \textit{cache-oblivious} algorithms, which implicitly adapt to the available multilevel cache hierarchy and its size at any given moment \citep{demaine2002cache}.

\paragraph{Connection between DRAM and CPU caches.} The DRAM Memory Controller is responsible for handling memory transactions between caches and DRAM. When interacting with caches, data is processed in blocks with a fixed size known as \textit{cache lines}. Cache line sizes can vary across microarchitectures, but the typical size is $64$ bytes in most x86-64 and AArch64 architectures.

\subsection{Background on network communication}
\label{ch7:app:networks-background}	

In the realm of the Internet, edge devices transmit packets through routers and switches. On the sender side, the long message is fragmented into smaller units called \abr{packets}. 

These packets traverse \textit{intermediate routers}, which, in turn, process them, route and enqueue to the appropriate outgoing link, and decide if a packet should be discarded.

When \textit{clients} and \textit{master} are linked via multiple routers, the delays from all intermediate routers cumulatively contribute to a \textit{single packet} delay. Sending a tiny packet from the client to master and back leads to a time metric known as Round Trip Time (\abr{RTT}), analogous to the Latency in \textit{Compute Architecture}. When multiple packets traverse the communication path, the rate at which information can be sent may become more crucial. In \textit{Compute Architecture} and \textit{Network Communication}, if several intermediate devices form a pipeline, then the speed of complete operation per time is determined by the slowest component in the pipe. The technical term denoting the speed is called \textit{bandwidth}, and in networks, it is measured in bits/second. If the link for message transfer is shared, the effective bandwidth is divided among clients. The derived quantity that describes the actual bandwidth is denoted as \textit{{throughput}}. The \textit{{bottleneck link}} is a link with the minimum throughput for communication. If the throughput is $R$ bits/s, and the  transferred message is $L$ bits, and the sender waits for acknowledgments, then the time for transferring an $L$-bit message can be estimated as:
\[
t_{\mathring{delay}} = RTT + \dfrac{L}{R_{\mathrm{bottleneck}}}.
\]

This is a classical formula commonly found in the communication network literature \citep{kurose2005computer}, but it is important to acknowledge its:

\begin{enumerate}
	\item It does not account for packet loss delays.
	\item The formula assumes that $RTT$ and $R_{\mathrm{bottleneck}}$ are constants, which may not be true.
\end{enumerate}

\paragraph{Communication network protocols stack.} The \textit{Network Communication} also serves as a bridge, linking the \textit{Electrical Engineering} with \textit{Applications} through a stack of protocols consists of:

\begin{enumerate}
\item \textit{Application layer} facilitates easy development (\abr{HTTP} [RFC 7230]).

\item \textit{Transport layer} hides defects of the underlying protocols (\abr{TCP} [RFC 793]).

\item The\textit{ Network layer} addresses routing, forwarding, and congestion control  (\abr{IPv4} [RFC 791]).

\item The \textit{Link layer} connects IP protocol with the lower levels via splitting data into frames, carrying error control, and medium access (CSMA/CD [IEEE 802.3], CSMA/CA [IEEE 802.11]).

\item The \textit{Physical layer} takes charge of the actual physics of the utilized medium.  (Bluetooth [IEEE 802.15.1], Wi-Fi [IEEE 802.11], Ethernet [IEEE 802.3]).
\end{enumerate}

This comprehensive framework driven by Institute of Electrical and Electronics Engineers (IEEE) and Request for Comments (RFC) standards\footnote{RFCs are maintained by the Internet Engineering Task Force  \href{https://www.rfc-editor.org}{https://www.rfc-editor.org}.} ensures the seamless integration and functionality of diverse elements.

\paragraph{The size of transferred messages in TCP/IP.} The theoretical maximum size of an \abr{IPv4} packet is $65535$ bytes, inclusive of a $20$-byte header. When using \abr{TCP/IP}, effective communication faces limitations if the message size exceeds the allowable Maximum Transfer Unit (\abr{MTU}) along the communication link. 

If an \abr{IP} packet surpasses the \abr{MTU}, it undergoes fragmentation, breaking into units with \abr{MTU} size. These fragments are later reassembled recreating the original packet, but \abr{IP} fragmentation is generally considered as undesirable. The \abr{IP} protocols define minimum \abr{MTU} sizes - $576$ bytes for \abr{IPv4} and $1500$ bytes for \abr{IPv6}. The \abr{TCP} header is $20$ bytes long, the \abr{IPv4} header typically spans $20$ bytes, and the \abr{IPv6} header is $40$ bytes. When operating at the \abr{TCP} level to avoid fragmentation, the payload should be less than the \abr{MTU} size minus \abr{IP} and \abr{TCP} header sizes. When Nagle Algorithm \citep{nagle1984congestion} is turned off, the \abr{TCP/IP} stack inside the {OS} will not delay sending small packages. And sending should be done with knowledge about this effect. For more see \citet{kurose2005computer}.

\paragraph{Hardware support for TCP/IP in the end systems.}

Modern communication devices supporting \abr{TCP/IP} are physically implemented as \textit{network interface cards} (NICs). These devices provide one or more ports to connect to an external communication medium, which in practice can be optical, electrical, or wireless. Internally, a NIC contains a component known as a \textit{network controller}, which is driven by its microprocessor. The network controller facilitates the efficient transfer of packets between the NIC's \textit{ports} and the Input/Output subsystem of the computing system. During increasing amounts of sent information, both the network port and hardware interface connection for the Input/Output subsystem can be a bottleneck \citep{kerrisk2010linux}.

\paragraph{OS support for TCP/IP in the end systems.}

User-level applications typically access the network interface card (NIC) using the \abr{TCP/IP} protocol through programmable endpoints, known as sockets, via the Berkeley socket API. This API, which has been in use for decades, is now ported to virtually every operating system (OS). It provides a mechanism to transfer data between the application and OS driver buffers, and OS kernel buffers. After this, the NIC reads and sends raw data from the OS kernel buffer, utilizing the Direct Memory Access (DMA) controller for packet transmission. When an incoming packet arrives at the NIC, the OS is generally notified through an asynchronous interrupt service request (IRQ). Some OS and NIC configurations slightly modify this standard behavior. For example, in Linux, if an interrupt coalescing mode is used, interrupts are not raised after arriving at each packet. Instead, an interrupt is raised either after a timeout or after a specified number of packets have been received.

The \abr{TCP} (Transmission Control Protocol) is a connection-oriented protocol that provides a reliable data transfer service on top of the unreliable IP service. The \abr{TCP} includes several unspecified characteristics, the management of which falls under the responsibility of the OS. Some of these characteristics include:

\begin{itemize}

\item The first step in a connection-oriented protocol is establishing a \abr{TCP} connection through a three-way handshake. The time to establish the initial connection is referred to as connection latency. However, \abr{TCP} at the protocol level does not specify timeouts for connection establishment.

\item During message transfer, \abr{TCP} facilitates the exchange of data between nodes in a communication graph. However, \abr{TCP} at the protocol level does not specify exactly how the OS should handle out-of-order segments.

\item To close the connection one side initiates it by sending a packet with the FIN flag. The other side acknowledges this packet and sends a symmetric FIN packet in return. After this exchange, the OS enters the TIME\_WAIT state for the specific connection (socket), which can last for a significant period (for example, 1 minute), depending on the OS configuration. The resources associated with the connection can only be released after this period. This timeout is not specified by the \abr{TCP} protocol as well.

\end{itemize}

It is also important to note that certain operating systems provide alternative ways for interacting with the NIC to use \abr{TCP/IP}. For example, the Data Plane Development Kit (DPDK)\footnote{The Data Plane Development Kit. Open source project managed by Linux Foundation: \href{https://www.dpdk.org}{https://www.dpdk.org}} and the eXpress Data Path (XDP) \footnote{eXpress Data Path (XDP). A high-performance, programmable network data path in Linux kernel: \href{https://prototype-kernel.readthedocs.io/en/latest/networking/XDP/introduction.html}{https://prototype-kernel.readthedocs.io}} in POSIX OS and Windows OS environments enable tighter coupling between user-space logic and the NIC, at the cost of added complexity in application development. Although relatively new, these interfaces open up additional opportunities for optimizing network communication.

\clearpage
\clr{
\addtocounter{adjsection}{1}
\section{{Broader Impact}}
\label{ch7:app:discussions}

\subsection{Towards more realistic scientific investigations}
\label{ch7:app:scientific-experiments}

\paragraph{Distinguishing design time from runtime.}
According to Amdahl's Law, overall performance is constrained by the weakest component. In complex, multi-layered software stacks, identifying bottlenecks becomes increasingly difficult, as multiple layers and their interactions can affect performance. The C++ language offers several distinct advantages: 
\begin{enumerate}
\item Its design facilitates systematic and modular thinking without introducing runtime overhead.
\item User-defined types are as efficient as built-in types.
\item Unused language features or libraries incur no time or memory cost.
\end{enumerate}

These properties make C++ an essential choice for our work, as it provides fine-grained control over the implementation of the components and their interactions.

\paragraph{Practical performance measurement for research ideas.}

Our implementation allows researchers to measure execution time in seconds, a capability often lacking in prototypical setups. Even theoretically optimal algorithms may underperform in practical benchmarks that measure time and memory due to large hidden implementation constants in layered designs. Our methodology enables modifications without compromising performance.
}

\clr{
\subsection{The role of theory in practical implementation}
\label{ch7:app:theory-role}

\paragraph{Convergence guarantees.}
The convergence guarantees of the \algname{FedNL} family ensure that these algorithms not only converge but also effectively address the problems outlined in Equation~\eqref{ch7:eq:main}, under Assumptions \ref{ch7:asm:1} and \ref{ch7:asm:2}. Our objective was to implement this algorithm family effectively within a Federated Learning setting, which is characterized by a less controlled training environment. The theory provides strong guarantees for the convergence of the training.

\paragraph{Problem independence in the runtime of FedNL.}
The theoretical framework of the \algname{FedNL} family \citep{safaryan2021fednl} has the advantageous property that its runtime does not explicitly depend on the characteristics of the local loss function $f_i(x)$. Thus, the algorithm can be applied to any problem in class, provided that the underlying numerical algorithms are stable.

\paragraph{Zero heuristics.}
The \algname{FedNL} Algorithm~\ref{ch7:alg:FedNL} runtime does not require any prior estimations of parameters exactly or heuristically, which provides essentially zero theoretical and practical gap. If this gap is not zero, in practice, heuristics are used to eliminate it. In this regard, several considerations must be noted: (i) their development of heuristics (or approximation) requires significant researcher time; (ii) they may unintentionally be problem-instance specific; (iii) as the number of heuristics increases, navigating their space can become exponentially complex; (iv) transferring heuristics from the design phase with a human in the loop to actual runtime (potentially without human involvement) poses challenges.

\paragraph{Leveraging FedNL theory for autonomous implementation.}
Improving core applications on edge devices requires acknowledging that mobile applications and operating systems typically cannot afford human intervention during runtime. We believe our implementation can potentially help with a rich class of applications in edge devices. In classical centralized ML, training algorithm requirements can be relaxed as long as convergence is achieved; the resulting trained model (obtained offline or during design time) can be applied in client applications with robust inference algorithms (in runtime). A key distinction in FL is that it involves not only inference on edge devices but also training itself.

\paragraph{Constructive theoretical requirements for compressors.}
Our discovery of two new compressors, presented in Appendix~\ref{ch7:app:toplek} and Appendix~\ref{ch7:app:seqk}, reinforces the idea that the compressor definitions from the original \algname{FedNL} paper \citet{safaryan2021fednl} are constructive, as they enable the development of novel compressors.

}

\clr{
\subsection{Generality and extensibility of implementation}
\label{ch7:app:extensibility-of-impl}

\paragraph{Selection of logistic regression as a benchmark.} We chose \modelname{logistic regression} as a benchmark for three main reasons. First, we used the same datasets and problem formulation as the original \algname{FedNL} implementation, as our improvements build upon that work. Second, solving \modelname{logistic regression} in a manner suitable for edge environments in FL requires no fine-tuning of the training process. Third, fully autonomous execution of \modelname{logistic regression} training on edge devices remains largely unexplored, as discussed in Section~\ref{ch7:sec:existing-fl-sota-system}.

\paragraph{Generality of the theory and our improvements to other problem classes.}

The theoretical framework of \algname{FedNL} is general and extends beyond solving the specific class of \modelname{logistic regression} problems. The algorithm is provably capable of solving any problem in the form of Equation~\eqref{ch7:eq:main} under Assumptions \ref{ch7:asm:1} and \ref{ch7:asm:2}. During improving \algname{FedNL}, only steps 17, 21, 50, and 58 from Appendix~\ref{ch7:app:history-of-improvements} were directly aimed at enhancing the Hessian Oracle for this model. Other steps contributed to performance enhancements through orthogonal improvements, independent of the specific problem class.

\paragraph{Extensibility of our implementation for future research.}

In addition to guiding how to use our solution (Appendix~\ref{ch7:app:usability}), we also detail how to incorporate new objectives and make system modifications for future research. To achieve this, we provide a robust C++ framework for implementing custom oracles, including essential tools such as linear algebra operations and linear solvers. Researchers can leverage our compute primitives, which are optimized for a wide range of CPU architectures (with and without special SIMD instructions) and NVIDIA GPU architectures via expressing computation directly in  CUDA. In addition, we provide methods for numerically verifying the correctness of computed quantities such as $f_i(x)$, $\nabla f_i(x)$, and $\nabla^2 f_i(x)$. For details see Appendix~\ref{ch7:app:futresearch}.

}

\clearpage
\addtocounter{adjsection}{1}
\section{Technical Discussions}
\label{ch7:app:technical}

\subsection{Why we selected TCP/IP for FedNL implementation}
\label{ch7:app:networks-why-tcp}

In our \algname{FedNL} implementation, clients establish connections with the master using either \abr{TCP/IPv4} or \abr{TCP/IPv6} and use this level of protocols. This level of abstraction allows us to distance ourselves from the physical principles governing information transfer, providing a high-level interface without sacrificing finer granularity optimizations. Although the UDP protocol induces less overhead, it lacks essential features such as flow control, congestion control, delivery guarantees, and ordering guarantees.

The HTTP is the application protocol employed as transport layers in various communication and remote procedure call protocols such as \libname{gRPC}, \abr{XML-RPC}, or \abr{REST}. Classical \abr{HTTP/1} introduces overheads, such as the \textit{protocol request text, request header, blank lines}, preceding the actual transferred message. If the communication session is relatively long, special treatment is required for maintaining a persistent connection. We decided \textit{not to rely on HTTP} and upper-level application protocols. Even \libname{gRPC} employs a persistent \abr{TCP} connection, we deemed \libname{gRPC} unnecessary for our purposes. In our view, \abr{gRPC} should be employed for its primary purpose only. Next, \abr{gRPC/REST/XML-RPC} relies on \abr{HTTP}, and \abr{HTTP} relies on \abr{TCP/IP}. We do not see a reason why we should work at such a high level. Any unnecessary abstractions that can not be eliminated during design time take resources and are not free in all software and hardware layers involved.


\clr{
\subsection{More about memory-mapped files}
\label{ch7:app:mmap}

Details about this mechanism can be found in specialized OS literature, such as \citet{kerrisk2010linux}, though it is rarely widely discussed. Counterintuitively, the most efficient way to read data from a file system is not through standard read operations provided by the OS. While files on disk adhere to the file system structure, conventional file-reading methods often involve an intermediate buffer managed by the OS I/O dispatcher and drivers. A memory-mapped file is an OS primitive available in modern systems such as Windows, Linux, and macOS, allowing files to be mapped directly into a process's virtual address space.

\paragraph{Key advantages of memory-mapped files from a system perspective.}

\begin{enumerate}
	
\item \textit{Direct file access.} Read and write operations bypass system calls and the I/O dispatcher, instead leveraging the CPU and the OS's built-in interrupt mechanisms for handling page faults.

\item \textit{Efficient memory usage.} Modern operating systems can use swap files to emulate larger physical memory. When utilizing memory-mapped files, even if the OS has insufficient memory, the file itself (rather than swap space) serves as backup storage for read-access operations.

\item \textit{More flexible memory management.} If physical pages in DRAM need to be evicted, the OS can release pages dedicated to memory-mapped files in read-only mode directly, without writing data back to disk. In contrast, standard buffers require data to be written back to the swap file.

\item \textit{Importance of implementation in the OS.} A well-implemented memory mapping mechanism is crucial for the OS, as it is used for loading applications and libraries and forms the foundation for inter-process communication. It is likely that this mechanism is carefully designed within the OS.

\end{enumerate}

}

\clr{
\clearpage
\subsection{Performance limitations of the Python ecosystem}
\label{ch7:app:nopython}

\paragraph{Inheriting downsides of interpretation agnostic to implementation.}

Interpreters execute scripts directly without preprocessing, while compiled languages involve multiple stages before generating executable binaries, including code generation and code optimizations. According to \citet{wexelblat2014history}, compiler research began around the 1950s with John Backus's Speedcoding project on the IBM 704 to address rising software costs in society. This project led to the creation of FORTRAN, a foundation for many later languages, including C++ \citep{stroustrup1994design}. Though compiled languages like C++ are often seen as low-level nowadays, raising the abstraction level can risk losing control over algorithm performance and its interaction with hardware and OS, potentially degrading performance despite user expertise\footnote{This assumes the algorithm is fixed. Adapting algorithms to architectural features is a viable research direction as discussed in Section~\ref{ch7:sec:extra-motivation}.}. As discussed in Appendix~\ref{ch7:app:history-of-improvements}, for our work programming language choice yielded a $\times 20$ speedup, but $\times 50$ additional speedup comes from the ability to implement needed computational and mathematical optimizations directly on the hardware.

\paragraph{The Suboptimality of design with using dynamic libraries.}

The Python interpreter leverages in addition to its core to the Python modules and extension modules to expand its capabilities. Extension modules for Python are dynamic (or shared) libraries. While they may appear to be a viable option, but for time-critical applications, heavy reliance on shared libraries can be a poor design choice per se. For performance-critical applications, static linking of isolated modules is often a far better alternative. This is because modern compilation tools can access the entire codebase during final program construction, enabling more compiler optimization techniques. Another performance drawback of dynamic libraries is the issue that arises when two or more libraries cannot be loaded at the same virtual address inside the executable process. To address this, approaches such as Code Relocation are utilized in Windows OS, and Position Independent Code (PIC) is used in macOS and POSIX operating systems. Both of them lead to performance problems. Finally, the actual application start time degrades as the number of loaded dynamic libraries increases. The overall situation can become quite problematic if these libraries (located in SSH/HDD for which access is slow based on Table~\ref{ch7:tbl:latencies-for-memory}), in a recursive manner, import additional libraries. This loading process can load a large amount of compiled code for various routines, when, in reality, only one routine is required.}

\clr
{
\paragraph{Unimposing of system thinking.}
Python's modular approach promotes the reusability of solutions constructed from distinct blocks. Sometimes, it is what is needed to focus in case of complex situations. However, this block-based thinking can contradict a system-level perspective, which emphasizes a holistic view of solutions rather than concentrating on individual components. Compiled languages offer a robust set of instrumentation tools that support a holistic and system-level mentality to isolate the problems. Such style teaches that there is a need to profile and look into the entire computation stack.
}

\clr{
\paragraph{Lack of compiler level optimization for users.} Compilers is a well-established field of Computer Science. In case using compilers, it provides ways to utilize techniques such as code inlining, global optimization, converting arithmetic operations to more efficient bit shifts, optimizing memory layout, eliminating dead code, removing loop-invariant computations, loop fusion, utilizing fast local stack storage, and profile-guided optimization. These challenging optimizations still require collaboration between the algorithm designer and the compiler. Implementing them in interpreted environments for user-provided scripts is at least more challenging.
}

\clr{

\paragraph{Challenges in developing multi-threaded implementations.} 

Creating reliable multi-threaded implementation involves several challenges, including managing memory barriers (or memory fences) to enforce the correct order of CPU instructions, ensuring proper synchronization, preventing data races, and handling atomic operations. Implementing effective multi-threading in interpreted languages is particularly complex. For example, while systems like \libname{Ray} \citep{moritz2018ray} provide distributed computing for Python, they rely on multi-processing, rather than multi-threading, for scalability, even within a single node.

}

\clr{	
\paragraph{Challenges with inefficient memory access.} As user-defined applications in the interpreter scripting language grow, the Python script execution logic may become deeply intertwined with the interpreter's operations from a CPU perspective. This intertwining can result in the interpreter consuming valuable CPU resources. For example, even when the objective is to run a training algorithm, the CPU caches may become polluted with the interpreter's logic, leaving less room for the user's algorithm to benefit from caching.
}	

\clr{
\paragraph{Inability to leverage special CPU registers and CPU instructions.}

At a fundamental level, all executable binaries, regardless of their file format, are ultimately executed using the CPU's native instruction set, which encodes operations into digital machine language. Python cannot directly interact with this hardware level, which limits performance in computation-heavy scenarios. Modern CPUs are equipped with special instructions and registers in their instruction set architectures (ISA) designed to optimize computation. Examples of such special instructions for 64-bit processors include SSE2 (16 XMM registers, each $4 \times 32$-bit), AVX2 (16 YMM registers, each $8 \times 32$-bit), AVX-512 (32 ZMM registers, each $16 \times 32$-bit), and ARM AArch64 Neon (32 V registers, each $4 \times 32$-bit). These special instructions allow highly efficient elementwise operations on small dense vectors. Additionally, special-purpose instructions such as Fused Multiply-Add (FMA) enhance performance by combining multiple operations into one, dramatically improving performance in compute-intensive tasks. Interpreted languages like Python cannot directly leverage these capabilities due to their reliance on abstraction layers that ruin latency accessing these instructions.
}

\clr
{	
\paragraph{Ineffective use of CPU registers.}
As shown in Table~\ref{ch7:tbl:latencies-for-memory}, the electrical circuits representing registers are the fastest data storage available within the CPU. Communication with the CPU from any software occurs through instructions that adhere to a specific Instruction Set Architecture (ISA). The ISA defines memory architecture, data types, input/output interfaces, and registers. The number of registers in ISA is limited and typically ranges from $32$ to $64$. This discrepancy between high-end processor vendors, who can allocate more registers in silicon and practical applications has been resolved as follows. The $32-64$ registers are the Front-End registers specified by the ISA, while at the microarchitecture level, the actual registers are stored in a common area known as the Register File, which can contain significantly more but is often limited by $1000$ registers\footnote{Microarchitecture details are typically subject to non-disclosure agreements.}. The Compiler community does not lack awareness of this discrepancy, but they do not have enough tools to operate. In reality, once the available registers are exhausted, the Compiler's code emission algorithms must resort to \textit{register spilling}, backing up registers in memory. A more effective approach to leverage CPU pipelines (and implicitly the Register File) is to express Instruction Level Parallelism (ILP) directly within the code, making the code simple and predictable enough for modern CPU's Instruction Decode and Operation Issue blocks. Interpreted languages introduce numerous wrappers that ruin the ability to use ILP.

\paragraph{The Hidden challenges of relying on vectorization in scripting languages.} In Python, vectorization often refers to processing large data blocks in parallel, typically achieved through external libraries. While this approach can significantly enhance performance, it poses challenges in certain scenarios: (i) when you need to implement such functionality yourself, relying on Python may be a strategically poor choice; (ii) when specific problems cannot be effectively represented or solved using existing library interfaces; and (iii) when the overhead of dispatching operations to external libraries exceeds the performance gains.
}

\subsection{Carried steps to ensure usability of implementation}
\label{ch7:app:usability}
\clr{\paragraph{Overview.}} Multiple optimization levels increased efficiency and performance. However, we strived to ensure that the presented work and future work building on top of the presenting implementation are practically feasible. To attain this we provide different means:

\begin{enumerate}
\item Compiled automatic code documentation in Doxygen.
\item Means to build projects and launch unit tests in macOS, Linux, and Windows.
\item Utils for observing the system configuration.
\item Generating synthetic optimization problems for \modelname{logistic regression}.
\item Code ecosystem for systems, networking, data parsing, and math routines.
\item Primitive to work with (various forms of) dense matrices and vectors.
\item Wrappers to work with CPU SIMD.
\item Means for sanity checks for $\nabla f(x)$ and $\nabla^2 f(x)$ oracles.
\item Classical iterative and dense-direct solvers for solving linear systems.
\item Means for launch automatic builds, launching unit tests, code coverage.
\end{enumerate}

\clr{
\paragraph{Step-by-Step guidelines.}
We provide step-by-step guidelines accessible through standalone Markdown files available in the code repository. These guidelines are also included in the compiled automatic code documentation, offering coherent instructions for using our work:

\begin{enumerate} 
\item \textit{Minimal environment.} Setting up the core build and runtime environment across Windows, Linux, and macOS. 
\item \textit{Extra tools.} Guidelines for setting up additional tools. 
\item \textit{Local build.} Instructions on building the project locally across Windows, Linux, and macOS.
\item \textit{Docker build.} Steps to build the project in a Docker container, ideal for testing on different CPU architectures. 
\item \textit{Experiments.} Instructions for running the experiments. 
\item \textit{GitHub integration.} Utilizing GitHub Actions for continuous integration. 
\item \textit{Project structure.} An overview of the project's structure and its components. 
\item \textit{Advanced configurations.}  Configuring the build beyond the default settings.
\end{enumerate}
}

\clr{

\paragraph{Addressing slow compilation time.}

One of the primary reasons for the limited popularity of C++ in ML is the long compile time. To mitigate this issue, we provide the following:

\begin{enumerate} 
	\item Exclude specific components from the build to reduce compile time: 
		\begin{enumerate} 
			\item Utility programs: dataset generators, system viewers, and GPU viewers. 
			\item Unit tests, which cover nearly all functionality. 
			\item CUDA support, NVIDIA's C/C++ extension for writing logic on GPUs. 
			\item OpenCL support, an API for GPUs across various vendors. 
			\item Generate bindings for other languages (for example, Python) via Simplified Wrapper and Interface Generator (SWIG) \citep{beazley1995simplified}. This process can be time-consuming. 
			\item Build shared libraries for clients and servers, including executable applications.
		\end{enumerate} 

\item The project supports invoked build processes, running unit tests, and launching binaries from the command line via created \texttt{project\_script.py}. This allows for offloading building and testing to another machine. 

\item For local builds, \texttt{project\_script.py} facilitates the use of the Ninja build system (faster than GNU Make on Linux / macOS) and allows configuration of the number of CPU cores used during compilation. 

\item Support for Unity (Jumbo) Build, a compile optimization technique that combines multiple source files into a single large file. 

\item Precompiled headers to speed up compilation times. 
\item Custom C and C++ compilers can be specified using the CXX and CC environment variables, which are supported by common build systems like CMake, GNU Make, and Microsoft Visual Studio. 

\item CPU Instruction Set Architecture-specific optimizations are automatically detected but can also be configured manually. We offer user-friendly options in a tool-independent manner to adjust build settings.

\item Extra instrumentation in the build process can be easily enabled or disabled.

\end{enumerate}
}	
\clr{

\paragraph{Handling complexity in large projects.}

As a project grows in size, making non-trivial modifications becomes more challenging. To address this, we provide the following:

\begin{enumerate} 

\item The code is organized as a collection of static libraries for easier management. 

\item We ensure near-complete test coverage, verified with the GNU Coverage. 

\item Detailed code documentation via Doxygen.

\item If your IDE has limited navigation features, we include diagrams in the compiled HTML documentation to visualize the project structure. 

\item To facilitate learning, the documentation, and the code are integrated into the compiled documentation. 
\item The project can be built using command-line utilities and exported to build and debugged into popular Integrated Development Environments such as Microsoft Visual Studio, QtCreator, and CLion. 
\end{enumerate}
}	

\clr{
\paragraph{Tools for code quality.}

To ensure code quality, we employed the following:

\begin{enumerate}

\item The compilation process ensures that there are no compile-time or linkage errors. Unlike scripting languages, where errors can be encountered at runtime, C++ requires all compile errors to be resolved before producing a valid program. In contrast, Python parses functions only at the moment of direct invocation.

\item We address compiler warnings across multiple C++ implementations. 

\item Unit tests to enhance confidence in the correctness.

\item Valgrind (memcheck) to detect memory leaks

\item Valgrind (callgrind), Intel VTune, Linux Perf Tool to detect memory cache problems

\item We used CppCheck and PVS-Studio for static code analysis.

\item We mainly used the functionality of nvprof - command line profiler from NVIDIA CUDA Toolkit to improve CUDA support implementation.

\item Algorithmic convergence is mainly supported from theory. 

\end{enumerate}
}

\clr{
\paragraph{Integration into applications.}

Our Federated Learning (FL) training runtime can be integrated with existing applications in several ways:

\begin{enumerate} 

\item \textit{Standalone executable.} The local client and server are native applications for your target OS, with no dependencies on third-party libraries at runtime.

\item \textit{Dynamic libraries.} The project can be built as distributed dynamic libraries (\texttt{.dll} for Windows, \texttt{.so} for Linux, and \texttt{.dylib} for macOS), with separate libraries for client and server, each having a single entry point. 

\item \textit{Language bindings via SWIG.} If your project uses a language supported by the Simplified Wrapper and Interface Generator (SWIG), such as Python, Perl, Java, or C$\#$, you can automatically generate interfaces to invoke the training process. We provide an example of code generation for Python. 

\end{enumerate}
}

\clr{
\clearpage
\subsection{Technical overview of project architecture}
\label{ch7:app:futresearch}
}
\begin{table}[h]
\centering
\small
\begin{tabular}{|p{0.22\textwidth}|p{0.61\textwidth}|p{0.09\textwidth}|} 
\hline
\rowcolor{gray!30}
\textcolor{black}{\textbf{Component Name}} & \textcolor{black}{\textbf{Goal}} & \textcolor{black}{\textbf{Type}} \\ 
\hline
\hline
CMakeLists.txt & Main root CMake project configuration file. & \makecell[c]{Config} \\ \hline
project\_scripts.py & Script to help launch builds, tests, documentation. & \makecell[c]{Script} \\ \hline
copylocal & Low-level utilities to work with bytes and bits and copy them effectively. & \makecell[c]{Static\\Lib} \\ \hline
cmdline & C++ cross-platform implementation of useful command line parsing mechanisms. & \makecell[c]{Static\\Lib} \\ \hline
fs & String conversion routines, operate with files and filenames, and memory map files. & \makecell[c]{Static\\Lib} \\ \hline
linalg\_linsolvers & Collection of dense and iterative specialized linear solvers in CPU. & \makecell[c]{Static\\Lib} \\ \hline
math\_routines & Special math routines - matrix sparsification, convex optimization. Also includes data structures and algorithms: sorting, tries, heaps, and indexed heaps. & \makecell[c]{Static\\Lib} \\ \hline
numerics & Number differentiation to evaluate derivatives, gradients, and Hessians numerically. & \makecell[c]{Static\\Lib} \\ \hline
timers & Various timers - systems, from C++ runtime (CPUs clocks to seconds), low-level wrappers to calculate clocks for x86-64 and AArch64. & \makecell[c]{Static\\Lib} \\ \hline
digest & Digest to check bit-bit equivalence CRC-32-IEEE 802.3, MD5 RFC1321 algorithms. & \makecell[c]{Static\\Lib} \\ \hline
network & TCP/IP[v4|v6], UDP/IP[v4|v6] wrappers for Windows and POSIX API. & \makecell[c]{Static\\Lib} \\ \hline
system & Memory pools, OS Memory Allocators, Low-level operations on Float/Double scalars. & \makecell[c]{Static\\Lib} \\ \hline
random & Calculate central statistics, uniform pseudo-random generators, shuffling with early stop. & \makecell[c]{Static\\Lib} \\ \hline
\end{tabular}
\caption{\clrshort{Components of the self-contained runtime part 1/2}.}
\label{ch7:tab:ufednl-components-a}
\end{table}

\begin{table}[h]
	\centering
	\small
	\begin{tabular}{|p{0.22\textwidth}|p{0.61\textwidth}|p{0.09\textwidth}|} 
		\hline
		\rowcolor{gray!30}
		\textcolor{black}{\textbf{Component Name}} & \textcolor{black}{\textbf{Goal}} & \textcolor{black}{\textbf{Type}} \\ \hline
		\hline
		\makecell[l]{linalg\_vectors\\(with SIMD)} & Dense vectors and light vector views with carefully designed API, including vectors with different underlying storages and custom implementation. & \makecell[c]{Static\\Lib} \\ \hline
		\makecell[l]{linalg\_matrices\\(with SIMD)} & Dense matrix implementation for BLAS operations, Cholesky, and QR factorization. & \makecell[c]{Static\\Lib} \\ \hline
		\makecell[l]{linalg\_linsolvers\\(with SIMD)} & Linear systems solvers: Jacobi, Gauss-Seidel, Conjugate-Gradient, Gauss-Elimination, backward and forward substitution. & \makecell[c]{Static\\Lib} \\ \hline		
		\makecell[l]{gpu\_compute\\\_support} & Partial supports of CUDA: dense vector and matrix operations, memory management, and wrappers over low-level GPU code invocation. & \makecell[c]{Static\\Lib} \\ \hline
		numerics & Tools for numerical differentiation to validate the correctness of Hessian and gradient computations. & \makecell[c]{Static\\Lib} \\ \hline
		optimization \_problems & Implements optimization problems, including \modelname{logistic regression} and \modelname{quadratic} minimization. & \makecell[c]{Static\\Lib} \\ \hline
		
	\end{tabular}
	\caption{\clrshort{Components of the self-contained runtime part 2/2}.}
	\label{ch7:tab:ufednl-components-b}
\end{table}

\clr{ 
\paragraph{The core components.}
The core low-level components are detailed in Table~\ref{ch7:tab:ufednl-components-a} and 
Table~\ref{ch7:tab:ufednl-components-b}. We offer out-of-the-box implementations for \modelname{logistic regression} and Symmetric Quadratic Objectives. From Tables~\ref{ch7:tab:ufednl-components-a},~\ref{ch7:tab:ufednl-components-b} it can be observed that we provide reach expressability of various optimization problems on modern CPUs beyond \modelname{logistic regression}. In terms of computation, although we support GPU computation, our implementation is pretty restricted to NVIDIA GPUs, for which we have developed functionality for dense vector and matrix operations. 

Currently, we do not support OpenCL or Apple Metal API. Additionally, our compressors operate on the CPU, with master aggregation also occurring on the CPU. Exploring robust implementations for modern GPUs, which follow a different computational model, remains a valid, practical research direction.
}

\clr{
\paragraph{Auxiliary binary tools.}

To facilitate the ongoing efforts of this project, we believe that the specifically designed binary tools listed in Table~\ref{ch7:tab:auxiliary-binaries} will  assist future researchers in advancing this work.

\paragraph{Main executable applications.}
The obtained binary applications for single node simulation are listed in Tables~\ref{ch7:tab:single-node-runners}. The obtained binary applications for multiple-node simulation are listed in Tables~\ref{ch7:tab:multiple-node-training}. To obtain extra executable binaries you should specify:
\begin{enumerate}
\item \texttt{DOPT\_SWIG\_INTERFACE\_GENERATOR} - Enables the use of SWIG to generate Python API wrappers. The generated libraries will have the prefix \texttt{"python\_"}.
\item \texttt{DOPT\_BUILD\_SHARED\_LIBRARIES} - Instructs the build system to create shared libraries, which will be prefixed with \texttt{"shared\_"}.
\end{enumerate}		

}
\begin{table}[h]
\centering
\small
\begin{tabular}{|p{0.24\textwidth}|p{0.56\textwidth}|p{0.14\textwidth}|} 
\hline
\rowcolor{bgcolorwe}

\textbf{Utility Name}            & \textbf{Goal}                                                                                                                                                                             & \textbf{Type}   \\ 
\hline
\hline
bin\_tests                       & Google Unit tests. Total number of tests is 102, but each test is pretty big. Tests cover all projects.                       & Executable \\ \hline
utils/bin\_host\_view             & Binary application to check used compiler name, flags, version, linker flags, information about OS, DRAM, and CPU extensions.                                                        & Executable \\ \hline
bin\_opencl\_view           & Observe platforms for OpenCL, devices available that support computation via OpenCL in the platform, its type, and available extensions if you will provide \texttt{show-extensions}.        & Executable \\ \hline
bin\_cuda\_view             & Observe NVIDIA CUDA compatible devices, show peak computation and memory throughput limits, available DRAM, and run simple tests to verify features. Flags: \texttt{verbose}, \texttt{benchmark}. & Executable \\ \hline
bin\_opt\_problem \_generator & Optional synthetics optimization problem generator. Can be used for debugging purposes.                                                                                              & Executable \\ \hline
bin\_split                 & Binary program to take one dataset in text, optionally reshuffle, add intercept, obtain information about clients, and split it into several files.                          & Executable \\ \hline
\end{tabular}
\caption{\clrshort{Auxiliary help binaries.}}
\label{ch7:tab:auxiliary-binaries}
\end{table}

\begin{table}[h]
\centering
\small
\begin{tabular}{|p{0.26\textwidth}|p{0.54\textwidth}|p{0.14\textwidth}|} 
\hline
\rowcolor{gray!30}
\textbf{Project} & \textbf{Goal} & \textbf{Type} \\ 
\hline
\hline
[bin|shared|python] \_fednl\_local & Local simulation on the machine for \algname{FedNL} and \algname{FedNL-LS} with mentioned compressors. & Executable \\ 
\hline
[bin|shared|python] \_fednl\_local\_pp & Local simulation on the machine for \algname{FedNL} and \algname{FedNL-LS} with compressors for Partial Participation. & Executable \\ 
\hline
\end{tabular}
\caption{\clrshort{Single-node multi-core runners.}}
\label{ch7:tab:single-node-runners}
\end{table}

\begin{table}[h]
\centering
\small
\begin{tabular}{|p{0.34\textwidth}|p{0.46\textwidth}|p{0.14\textwidth}|} 
\hline
\rowcolor{gray!30}
\textbf{Project} & \textbf{Goal} & \textbf{Type} \\ 
\hline
\hline
[bin|shared|python] \_fednl\_distr\_client & Client application to participate in \algname{FedNL} and \algname{FedNL-LS} algorithms with various compressors. & Executable \\ 
\hline
[bin|shared|python] \_fednl\_distr\_client\_pp & Client application for partial participation in \algname{FedNL} and \algname{FedNL-LS} algorithms. & Executable \\ 
\hline
[bin|shared|python] \_fednl\_distr\_master & Server application to control \algname{FedNL} and \algname{FedNL-LS} training. & Executable \\ 
\hline
[bin|shared|python] \_fednl\_distr\_client\_pp & Server application for partial participation in \algname{FedNL} and \algname{FedNL-LS} training. & Executable \\ 
\hline
\end{tabular}
\caption{\clrshort{Multiple-node FL training runners.}}
\label{ch7:tab:multiple-node-training}
\end{table}

\clearpage
\addtocounter{adjsection}{1}
\section{Reproducibility}

The source code for the experiments, implementation, and documentation is publicly available at the following link:

\begin{center} 
	\href{https://github.com/burlachenkok/ufednl}{https://github.com/burlachenkok/ufednl} 
\end{center}

\unappendix

\chapter{BurTorch: An Exquisitely Fast Backpropagation}
\label{chapter8}

The goals and summaries of this chapter are outlined in Table \ref{ch1:tbl:algorithms} and Section~\ref{ch1:sec:overview-8}.

\section{Introduction}
\label{ch8:sec:intro}

First-order optimization methods are a class of continuous optimization techniques that rely on gradient and function value information to minimize a target function, without using higher-order derivatives. For instance, the Gradient Descent (\algname{GD}) algorithm iteratively updates model parameters to approach a stationary point. Specifically, \algname{GD} updates the model parameters using the rule:
\[
x^{k+1} \eqdef x^k - \gamma \nabla f(x^k),
\]
where $\gamma \in \mathbb{R}$ is the learning rate, and $\nabla f(x^k) \in \mathbb{R}^d$ is the gradient of the objective at the current iterate $x^k$. If $\nabla f(x^k)$ does not exhibit any structure, this model update for \algname{GD} requires $d$ in-place scalar additions and multiplication by $\gamma$. The time complexities of most optimization methods, including theoretically optimal algorithms such as \algname{Nesterov Accelerated Gradient Descent} for convex objectives \citep{nesterov2018lectures} and \algname{PAGE} \citep{li2021page,tyurin2022sharper} for cases where
\begin{equation}
	\label{ch8:eq:main}
	f(x) = \frac{1}{n} \sum_{i=1}^{n} f_i(x),
\end{equation}
and $f_i(x)$ is non-convex, are typically expressed in terms of gradient oracle complexities. For this reason, efficient gradient computation is crucial, as the cost of gradient evaluations significantly impacts the overall training cost. One way to estimate $f(x)$ from Equation~\ref{ch8:eq:main} in an unbiased manner is by sampling a subset $S \subseteq [n], |S|=b$ uniformly at random from all possible subsets of cardinality $b$, and constructing:
\begin{equation}
	\label{ch8:eq:main-est}
	f_{S}(x) = \dfrac{1}{b} \sum_{i \in S} f_{i}(x).
\end{equation}

For a differentiable function $f_S(x)$ from Equation~\eqref{ch8:eq:main-est}, the following holds due to the linearity of differentiation:
\begin{equation}
	\label{ch8:eq:main-grad-est}
	\nabla f_{S}(x) = \dfrac{1}{b} \sum_{i \in S} \nabla f_i(x). 
\end{equation}

Next, assume that each $f_i(x)$ for $i \in S$ is parameterized by a compact information description, such as an input-output pair in supervised machine learning problems. In this case, it is natural to assume that the \textit{time} to compute $\nabla f_{S}(x)$ in Equation~\ref{ch8:eq:main-grad-est} is equal to the sum of the times for computing $\nabla f_i(x)$ for $i \in S$.

However, this may not hold precisely in current practice. Modern Deep Learning frameworks such as \libname{JAX} \citep{jax2018github}, \libname{PyTorch} \citep{paszke2019pytorch}, and \libname{TensorFlow} \citep{abadi2016tensorflow} are the result of collaborative efforts across various disciplines. To effectively mask internal latencies, these frameworks employ a throughput-optimized design. The relatively large batch sizes $b$ help mask latencies by batching input and intermediate data across all software and hardware layers involved in the computation. This design, while optimized for throughput, does not provide an efficient method for computing individual gradient oracles $\nabla f_i(x)$. 

Furthermore, large frameworks comprising millions of lines of code, even when open-source, are often difficult to modify and computationally optimize. A computationally optimized implementation can follow two main philosophies: (a) perfecting every detail, and (b) limiting the number of details\footnote{Jack Dorsey is an engineer who applied this approach to create systems for society.}. However, as the implementation scales and requires human intervention, both philosophies are often compromised.

To address these challenges, we introduce \libname{BurTorch}, a minimalist framework that significantly improves both the latency of computing $\nabla f_i(x)$ and overall memory efficiency. Throughput-oriented Deep Learning (DL) frameworks incur high memory overhead during $\nabla f(x)$ execution (see Appendix~\ref{ch8:sec:backprop-memory} for the memory taxonomy within backpropagation), especially for large batch sizes, $b \gg 1$. \libname{BurTorch} mitigates this issue by computing individual gradient oracles $\nabla f_i(x)$ sequentially, reducing peak memory usage from $\sum_{i=1}^{b} \mathrm{MEM}(\nabla f_i(x))$ to $\max_{i \in [b]} \mathrm{MEM}(\nabla f_i(x))$.

\subsection{Mathematical tools for gradient computation} \label{ch8:sec:background-gd-compute}

One way to view the relation between $f(x): \mathbb{R}^d \to \mathbb{R}$ and its high-order derivatives is through the Taylor expansion at the point $x\in\mathbb{R}^d$ with the Lagrange remainder:

\begin{eqnarray}
	\label{ch8:eq:taylor-exp-with-2-terms}
	\dfrac{f(x + s \cdot \varepsilon) - f(x)}{\varepsilon} = \langle s, \nabla f(x) \rangle + \dfrac{s^\top \nabla^2 f(z) s \cdot \varepsilon}{2},
\end{eqnarray}
where $\varepsilon \in \mathbb{R}$, $s \in \mathbb{R}^d$, and $z \in (x, x + s \cdot \varepsilon)$.

\paragraph{Finite difference.}

If $\varepsilon \ll \min(1, \lambda_{\text{max}}(\nabla^2 f(z)))$, neglecting the second-order term in Equation~\eqref{ch8:eq:taylor-exp-with-2-terms} gives the \textit{forward finite difference method}, a local approximation of the directional derivative of $f(x)$ at $x$ in any direction $s$, $\|s\|_2=1$. From a computational standpoint, finite-difference methods face three challenges: (i) for small $\varepsilon$, numerical instability due to floating-point \citep{IEEE754-2008} limitations is not negligible; (ii) for large $\varepsilon$, error grows unbounded if $f(x)$ has non-zero curvature in direction $s$; (iii) finite-difference schemes require repeated evaluations of $f(x + s_i \cdot \varepsilon)$ for projections of $\nabla f(x)$ in multiple directions, leading to the overhead of $\times d$ for computing $\hat{\nabla f(x)} \approx \nabla f(x)$ using the numerical scheme \eqref{ch8:eq:taylor-exp-with-2-terms}.

\paragraph{Symbolic gradient oracles.}

When the structure of $\nabla f(x)$ is simple, gradient oracles can be derived symbolically and refined numerically. For example, \citet{burlachenko2024unlocking} explores methods to improve gradient and Hessian oracles for \modelname{logistic regression} symbolically. Methods that aim to manage computational graphs symbolically generally face two main challenges: (i) already combining $x_1, \dots, x_n \in \mathcal{X}$ with a single binary operator leads to a number of all possible distinct compute graphs asymptotically bounded by the Catalan numbers $\Omega \left( \nicefrac{4^n}{n^{3/2}} \right)$ \citep{cormen2022introduction}; (ii) the number of terms in symbolic gradient expressions can increase rapidly. For example, $\nabla_x \left( \prod_{i=1}^{n} x_i \right) = \left[ \prod_{j=1, j \neq i}^{n} x_j \right]_{1 \le i \le n}$ the symbolic gradient contains $(n^2 - n)$ terms, even though the original expression having only $n$ \citep{griewank2008evaluating}. Despite these complexities, there is a line of research that operates at this level~\citep{fawzi2022discovering, jia2019taso}.

\paragraph{Automatic Differentiation.} The \textit{Automatic Differentiation (AD)}, also known as \textit{Algorithmic Differentiation}, is an exact method for computing $\nabla f(x)$, assuming the exact execution of arithmetic and explicit description of $f(x)$. AD applies the chain rule systematically, unlike symbolic differentiation, by working with numerical values instead of symbolic expressions. The AD is fundamental in ML, where gradient computation is typically automated using the \textit{{Backpropagation Algorithm}} \citep{rumelhart1986learning}, a form of AD in reverse accumulation mode \citep{linnainmaa1970representation}. The theoretical computational cost and memory access cost of backpropagation can be conservatively estimated by the $w_{\text{tang}}$ cost of computing $f(x)$ and the cost of the memory accesses. The constant $w_{\text{tang}} \in [3, 4]$ is universal \citep{griewank2008evaluating}, and this bound holds if compute operations and memory accesses for $f(x)$ are in some sense atomic.

Backpropagation reuses intermediate computations to construct the necessary Jacobians mainly in two passes. The \textit{Forward Pass} performs inference and stores intermediate variables (activations), while the \textit{Backward Pass} implicitly computes the required Jacobians and generates $\nicefrac{\partial f(x)}{\partial x_i}$. Storing activations after the \textit{Forward Pass} is crucial for {backpropagation}. While {backpropagation} is preferred for computing the entire $\nabla f(x)$, {AD in Forward Accumulation Mode} \citep{rall1981automatic} is more memory-efficient when only a single directional derivative $\langle \nabla f(x), s \rangle$ is needed, as it computes the directional derivative in a single pass, mirroring the computation of $f(x)$. The cost of computing $f(x)$ and $\langle \nabla f(x), s \rangle$ with this method is bounded by a factor in the interval $[2, \nicefrac{5}{2}]$ relative to the cost of $f(x)$ evaluation \citep{griewank2008evaluating}.

\subsection{Progression of deep learning frameworks}
\label{ch8:sec:prev-systems}

Next, we review several key DL systems, with backpropagation playing a central role in their structure. See Appendix \ref{ch8:app:exex-details-of-comp} for details on eager and graph modes.

\paragraph{{Torch (2002).}} The modern era of Deep Learning frameworks began with Torch \citep{collobert2002torch}, which laid the foundation for many subsequent systems, including \libname{PyTorch} \citep{paszke2019pytorch}. \libname{Torch} was built around the Lua scripting language, which played a key role in its flexibility.

\paragraph{\color{black}{Theano (2010).}} \libname{Theano} \citep{bergstra2010theano} pioneered efficient Automatic Differentiation (AD) and optimization, accelerating the development of complex mathematical models. It introduced the computation graph abstraction, enabling automatic gradient computation, thereby streamlining the development of DL models. The graph-based approach in \libname{Theano} (and \libname{TensorFlow 1.0}) required careful graph construction before training.

\paragraph{\color{black}{Caffe 1.0 (2014).}} In 2014, \libname{Caffe} \citep{jia2014caffe} emerged as a leading framework for computer vision, simplifying model design through its \textit{Layer} abstraction and introducing runtime memory preallocation. Unlike \libname{Theano}, \libname{Caffe} did not require the same level of manual graph construction.


\paragraph{\color{black}{TensorFlow 1.0 (2015).}} In 2015, \libname{TensorFlow 1.0} \citep{abadi2016tensorflow} became the dominant framework in academia and industry. It functioned as both a programming language and a compiler, relying on an explicit computation graph managed by a \textit{Session} object. Key optimizations included automatic memory buffer reuse, constant folding and propagation, concurrent execution of independent operations, and memory allocation for variable states.


\paragraph{\color{black}{{PyTorch} (2016).}} From its inception, \libname{PyTorch} \citep{paszke2019pytorch} gained popularity for its intuitive debugging and flexibility in experimentation. Unlike \libname{TensorFlow 1.0}, which uses a static graph, \libname{PyTorch} employs an imperative computation style that integrates implicit graph construction with the forward pass. While this approach offers greater flexibility, it complicates optimization. The design of \libname{TensorFlow 2.0} mirrored \libname{PyTorch}'s eager execution model.



\paragraph{\color{black}{Keras (2015) and TensorFlow 2.0 (2019).}} \libname{TensorFlow 2} introduced \textit{eager execution}, enabling computations without the need for explicit graph construction, bringing it closer to the dynamic graph approach of \libname{PyTorch}. Although this improved flexibility, it came at the cost of performance. The benefits of eager execution include: (i) immediate error reporting, (ii) removal of legacy constructs such as placeholders and sessions, (iii) seamless integration with language-specific data structures and control flows, and (iv) compatibility with native debugging tools. \libname{Keras} \citep{chollet2015keras} was integrated into \libname{TensorFlow 2} as its API.


\paragraph{\color{black}{JAX (2018).}} \libname{JAX} \citep{jax2018github} extends \libname{NumPy} \citep{walt2011numpy} with AD and hardware acceleration. While its eager mode lacks efficiency, its just-in-time (JIT) compilation for static graph execution focuses on highly optimized internal computations.

\paragraph{\color{black}{Standalone packages.}} \libname{Autograd} \citep{maclaurin2015autograd}, \libname{Micrograd} \citep{karpathy2020micrograd}, and \libname{Apple MLX} \citep{mlx2023} focus solely on AD implementation.


\subsection{Motivation behind our work}
\label{ch8:sec:motivation}

The foundations of the backpropagation algorithmic computational technique trace back to the seminal works of \citet{linnainmaa1970representation} and \citet{rumelhart1986learning}. Over the past two decades, systems implementing this fundamental algorithm have undergone significant evolution (see Section~\ref{ch8:sec:prev-systems}). Given this extensive history, it might seem that the computation of $\nabla f(x)$ has reached its full potential. 
\begin{center}
	\textit{However, our research challenges this assumption, revealing that there are still  opportunities for further advancements.}
\end{center}

\subsection{Contributions} 
\label{ch8:sec:contributions}


The key contributions of this work are as follows:

\begin{enumerate} 
	
	\item \textbf{Latency-Efficient backpropagation.} \libname{BurTorch} enhances memory access, computational flow of backpropagation to deliver extremely efficient performance in latency-sensitive applications, especially when batch sizes are small or high throughput is difficult to achieve. This makes it well-suited for resource-constrained environments.
	
	\item \textbf{Significant computation gains with a compact implementation.} \libname{BurTorch} delivers substantial speedups and minimizes memory usage, making it ideal for mobile and IoT devices in Federated Learning \citep{FEDLEARN}. These improvements were achieved without increasing implementation complexity, preserving the flexibility for algorithmic refinements and extensions during $\nabla f(x)$ computation.


	
	\item \textbf{Bringing compile-based languages closer to algorithms creation in ML.} High-level scripting languages, while user-friendly, often introduce excessive abstraction in DL systems, limiting system-level optimization across compilers, computing architectures, and operating systems. The runtime of these languages consumes significant system resources \citep{pereira2021ranking}, and the restricted fine-grained control over system resources hampers efforts to minimize memory movement operations. To address these challenges, \libname{BurTorch} eliminates unnecessary layers, reduces reliance on third-party libraries, and optimizes backpropagation performance. Created in modern C++20, \libname{BurTorch} maintains a compact, efficient design while adopting a \libname{PyTorch}-like syntax (see Listing~\ref{ch8:lst:exp2-listing} in Appendix~\ref{ch8:app:exp2-listing}). Its minimalistic approach avoids the common problem of prolonged compilation times in complex C++ projects. Finally, as large language models evolve, they can assist in generating the necessary system boilerplate code, which could improve the reputation of C++ in research.
	
	\item \textbf{Serialized computations to reduce memory.} Modern frameworks often parallelize batch processing to hide latency, but this increases memory consumption by storing activations for the entire batch, creating a bottleneck during backpropagation \citep{2024fwdllm, mishra2017wrpn} in both image processing and large language model tasks. \libname{BurTorch} addresses this by computing sample-gradient oracles sequentially and overwriting activations across the batch. This reduces the activation memory footprint by a factor proportional to batch size $b$, sacrificing some throughput but minimizing memory usage. In addition to a serialized computation model, \libname{BurTorch} tackles the memory problem by dramatically reducing the code size, which tends to be excessively high in modern DL frameworks.	
	
	\item \textbf{Benchmarking Against Best-Practice Solutions.} We conduct an extensive evaluation of \libname{BurTorch} by benchmarking it against several computational libraries, including \libname{JAX}, \libname{TensorFlow}, \libname{PyTorch}, \libname{Apple MLX}, \libname{Autograd}, \libname{TFLite}, and \libname{Micrograd}. Our experiments span multiple operating systems, utilize various language interfaces, explore different execution modes across these frameworks, and cover a diverse range of tasks. To provide a clear overview of the key improvements, we present a representative summary of the numerical results in Table~\ref{ch8:tab:speedup-summary}. As we will see in the main part of the paper, the performance gains of \libname{BurTorch} increase significantly as \( d \) decreases.

\end{enumerate}

\begin{table}[h!]
	\footnotesize
	\centering
	
	\caption{Significant practical performance speedups of \libname{BurTorch} over \libname{PyTorch} on CPU. A representative summary of the expected gains in computing $\nabla f_S(x)$ for small $|S|$, based on the explicit form of $f_S(x)$ in Equation~\eqref{ch8:eq:main-est}.}

	\begin{tabular}{|l|c|c|c|c|}
		\hline
		\textbf{Backpropagation Task} & \textbf{Dimension} & \textbf{\makecell[c]{Compute\\ Speedup}} & \textbf{\makecell[c]{Memory\\ Savings}} & \textbf{\makecell[c]{Initialization\\ Speedup}} \\ 
		\hline
		\hline
		1. Small compute graph $f_S(x)$ & \(d \approx 6,000\)   & \(\times 45\)    & \(\times 74\)  & \(\times 354\)  \\ \hline
		2. Medium compute graph $f_S(x)$ & \(d \approx 60,000\)  & \(\times 7.4\)   & \(\times 65\)  & \(\times 340\)  \\ \hline
		3. Large compute graph $f_S(x)$ & \(d \approx 600,000\) & \(\times 1.5\)   & \(\times 37\)  & \(\times 150\)  \\ \hline
		4. Larger compute graph $f_S(x)$ & \(d \approx 1,000,000\) & \(\times 1.2\)   & \(\times 25\)  & \(\times 100\)  \\ \hline
	\end{tabular}
	\label{ch8:tab:speedup-summary}
\end{table}

\section{Validation Through Experiments}
\label{ch8:sec:experiments}

We present evidence of \libname{BurTorch} achieving low latency and a small memory footprint during $\nabla f(x)$ computation on Windows OS across a series of tasks, starting with the simplest. Results for Linux and macOS are provided in Appendix~\ref{ch8:app:extra-experiments-linux} and \ref{ch8:app:extra-experiments-macos}, respectively. Additionally, a user-centric energy consumption measurement experiment is included in Appendix~\ref{ch8:app:energy-eff}. For detailed software and hardware setups across all experiments, see Appendix~\ref{ch8:app:exp-setup}.

\subsection{Tiny compute graph}
\label{ch8:sec:experiments-tiny}

\begin{figure}[h!]
	\centering
	\includegraphics[width=1.0\linewidth]{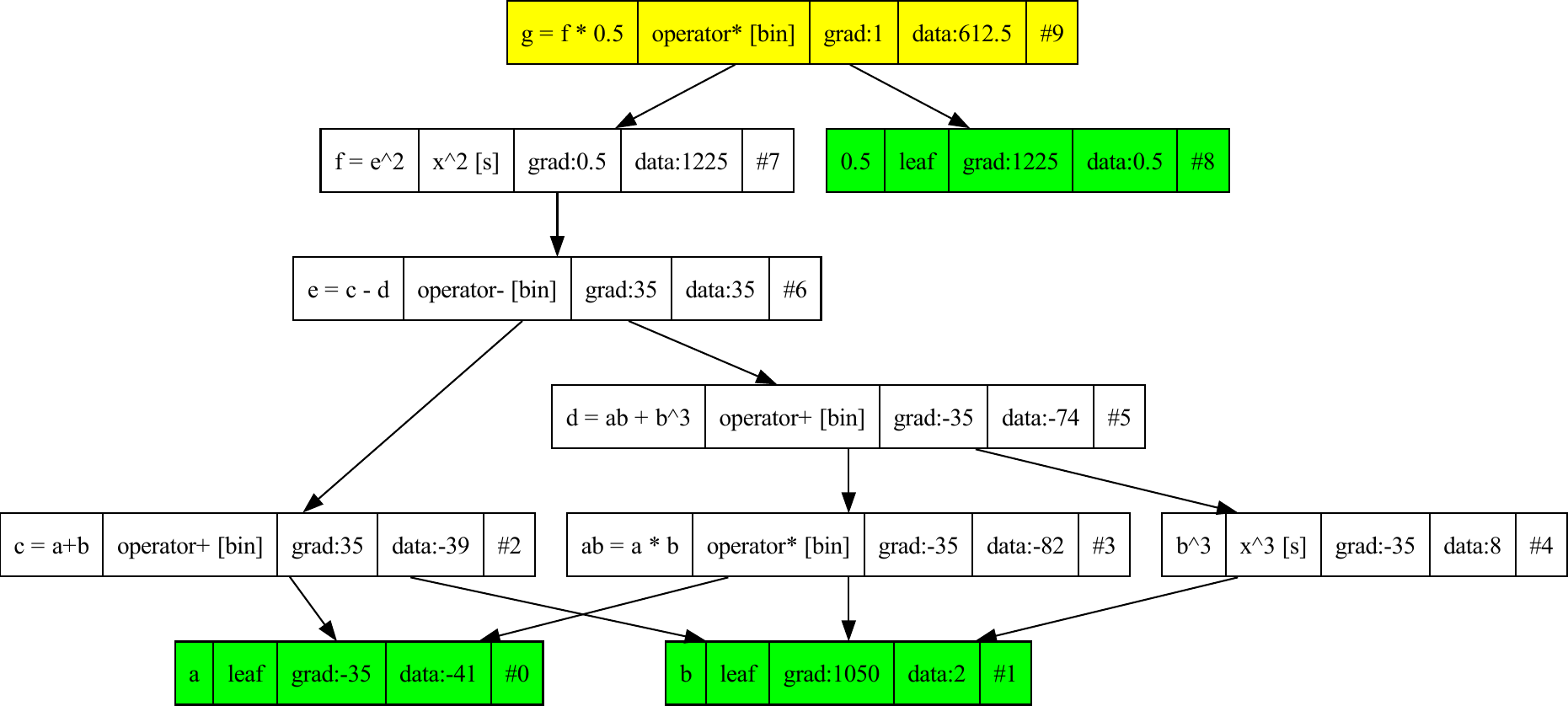}
	\caption{Tiny compute graph with $10$ nodes to evaluate $g=f/2,f=e^2,e=c-d, d=ab + b^3, c=a+b,a=-41,b=2$. Nodes contain: description, operator, $\dfrac{\partial g}{\partial [\mathrm{\textit{node}}]}$, value, raw index. The numerical results across frameworks match exactly.}
	\label{ch8:fig:tiny-compute-graph}
\end{figure}

We represent the composite function shown in Figure~\ref{ch8:fig:tiny-compute-graph} as an oriented directed tree, where the leaves correspond to constants and variables from $\mathbb{R}$, and the internal nodes represent algebraic operations. When expressing this computation across various frameworks, as detailed in Table~\ref{ch8:tab:execution_times_speedup}, we observe a substantial performance gap between \libname{BurTorch} and other state-of-the-art frameworks. \libname{BurTorch} framework consistently outperforms alternatives in computing $\nabla f(x)$ via Automatic Differentiation when handling small graphs.

Table~\ref{ch8:tab:execution_times_speedup} and Figure~\ref{ch8:fig:execution_times_speedup} present backpropagation performance over $10^5$ iterations. The reported time is the mean and standard deviation of five independent runs, with computation performed in FP64 on a single CPU core under Windows OS. The comparison highlights the compute time of different frameworks and languages, along with their relative speedup over \libname{BurTorch}. With the design philosophy outlined in Appendix~\ref{ch8:app:design-philosophy} and utilizing research-friendly eager execution, \libname{BurTorch} achieves the fastest time, taking $0.007$ seconds for $10^5$ iterations of backpropagation.

In contrast, other frameworks such as \libname{TensorFlow}, \libname{TF Lite} \citep{abadi2016tensorflow}, \libname{Autograd} \citep{maclaurin2015autograd}, \libname{PyTorch} \citep{paszke2019pytorch}, \libname{JAX} \citep{jax2018github}, and \libname{Micrograd} \citep{karpathy2020micrograd} exhibit significantly higher execution times, with some being up to $7,000$ times slower. The experiment demonstrates that as the compute graph becomes tiny, modern DL frameworks, whether accessed through high-level or optimized APIs, struggle to operate efficiently. The latency associated with transferring control to the GPU only \textit{exacerbates} the issue. These results highlight \libname{BurTorch}'s efficiency in handling tiny-scale $\nabla f(x)$ task on a CPU. Given that a $\times 4$ improvement is considered groundbreaking in Compute Architecture and Compilers, the performance gains of \libname{BurTorch} are significant.

\begin{figure*}
	\centering
	\includegraphics[width=1.0\linewidth]{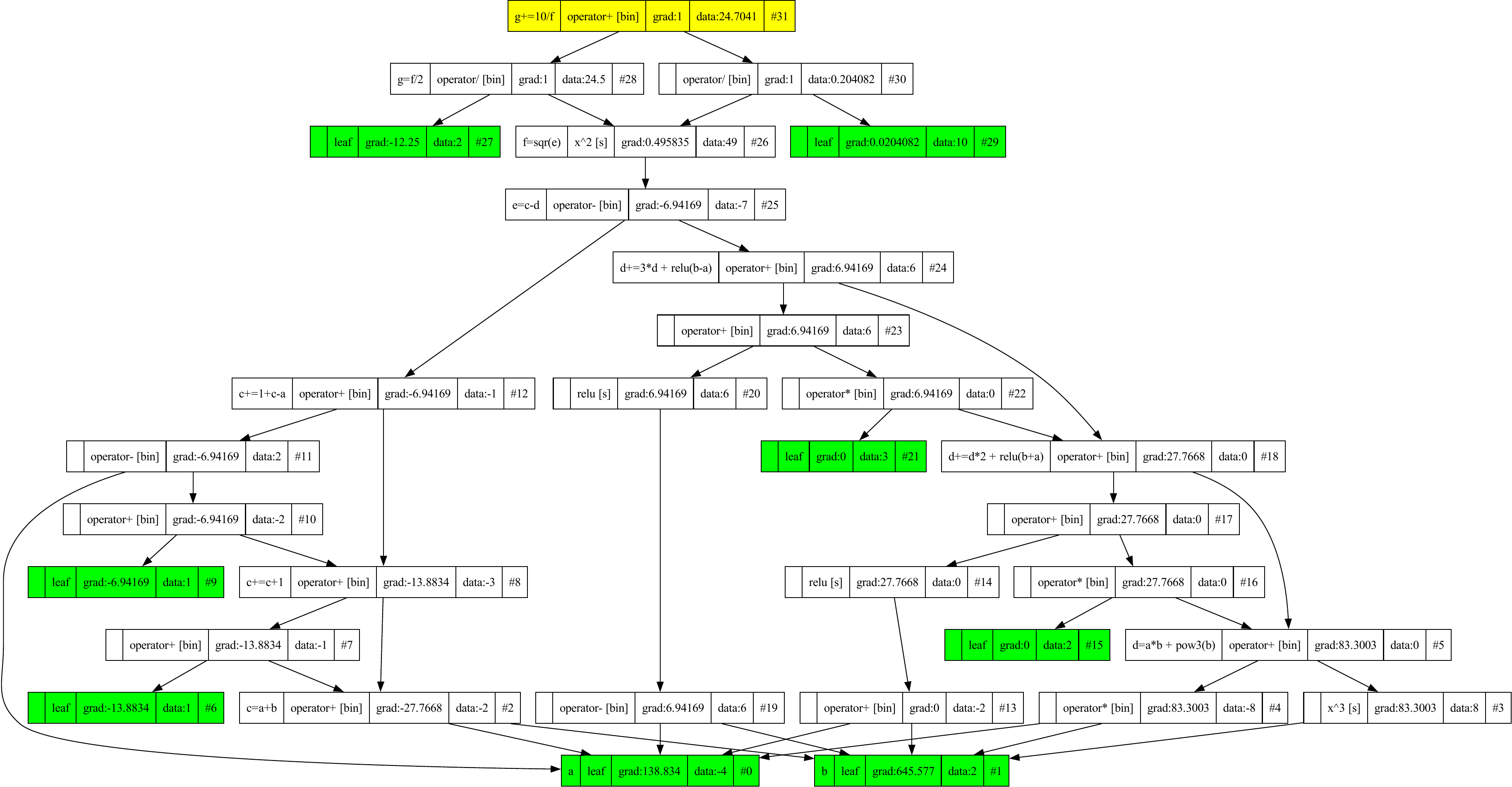}
	\caption{Small compute graph with total $V=32$ nodes and $E=44$ edges to evaluate function from \citet{karpathy2020micrograd}.}

\label{ch8:fig:exp2-small-compute-graph}
\end{figure*}

\begin{table*}[h!]
\footnotesize
\centering
\caption{Backpropagation over $100$K iterations with a {tiny} compute graph from Figure~\ref{ch8:fig:tiny-compute-graph}. Mean and std. deviation across $5$ launches, FP64, Windows OS. See also Figure~\ref{ch8:fig:execution_times_speedup}. The numerical results across frameworks match exactly.
}
\begin{tabular}{|l|l|l|l|l|}
	\hline
	\textbf{\#} & \textbf{Framework, Mode, Language} & \textbf{Device} & \textbf{\makecell[c]{Compute Time \\ (sec.)}} & \textbf{\makecell[c]{Relative \\ to \\ BurTorch}} \\ 
	\hline
	\hline
	\cellcolor{bgcolorwe}1&\cellcolor{bgcolorwe}BurTorch, Eager, C++ & \cellcolor{bgcolorwe}CPU                   & \cellcolor{bgcolorwe}$0.007 \pm 0.0004$             & \cellcolor{bgcolorwe}$\times 1.0$ (We)             \\ \hline
	2&TensorFlow 2.8.0, Eager, Python & CPU          & $55.217 \pm 0.2975$         & $\times 7\,888.1\,\,\,$                    \\ \hline
	3&TensorFlow 2.8.0, Graph, Semi-Python & CPU          & $14.469 \pm 0.0734$         & $\times 2\,067.0\,\,\,$                    \\ \hline
	4&\makecell[l]{TF Lite 2.8.0, Graph, TF Lite Interpreter} & CPU          & $0.589 \pm 0.0102$         & $\times 84\,\,\,$                    \\ \hline
	5&Autograd 1.7.0, Eager, Python & CPU            & $18.956 \pm 0.2962$        & $\times 2\,708.0\,\,\,$                    \\ \hline
	6&{PyTorch} 2.5.1, Eager, Python & \textbf{GPU}          & $51.380 \pm 0.4666$              & $\times 7\,340.0\,\,\,$                    \\ \hline
	7&{PyTorch} 2.5.1, Eager, Python & CPU          & $10.419 \pm 0.0647$              & $\times 1\,488.4\,\,\,$                    \\ \hline
	8&{PyTorch} 2.5.1, Graph, TorchScript & CPU & $9.994 \pm 0.1021$                 & $\times 1\,428.5\,\,\,$                    \\ \hline
	9&{PyTorch} 2.5.1, Eager, LibTorch, C++ & CPU & $5.300 \pm 0.0667$                 & $\times 757.14\,\,\,$                    \\ \hline
	10&{JAX} 0.4.30, Eager, Python & CPU                & $291.764 \pm 8.5373$         & $\times 41\,860.5$                      \\ \hline
	11&{JAX} 0.4.30, Graph, Semi-Python & CPU                & $5.580 \pm 0.0661$         & $\times 797.1\,\,\,\,\,\,\,$                      \\ \hline
	12&Micrograd, Eager, Python & CPU                & $1.590 \pm 0.0152$                & $\times 227.1\,\,\,\,\,\,\,$                      \\ \hline
	\cellcolor{bgcolorwe}13&\cellcolor{bgcolorwe} In Theory for this CPU (Registers Only) & \cellcolor{bgcolorwe}CPU & \cellcolor{bgcolorwe}$\Omega(0.0004)$             & \cellcolor{bgcolorwe}$\times 0.057$ \\ \hline
\end{tabular}
\label{ch8:tab:execution_times_speedup}
\end{table*}

\begin{figure*}[ht]
\centering
\includegraphics[width=1.0\linewidth]{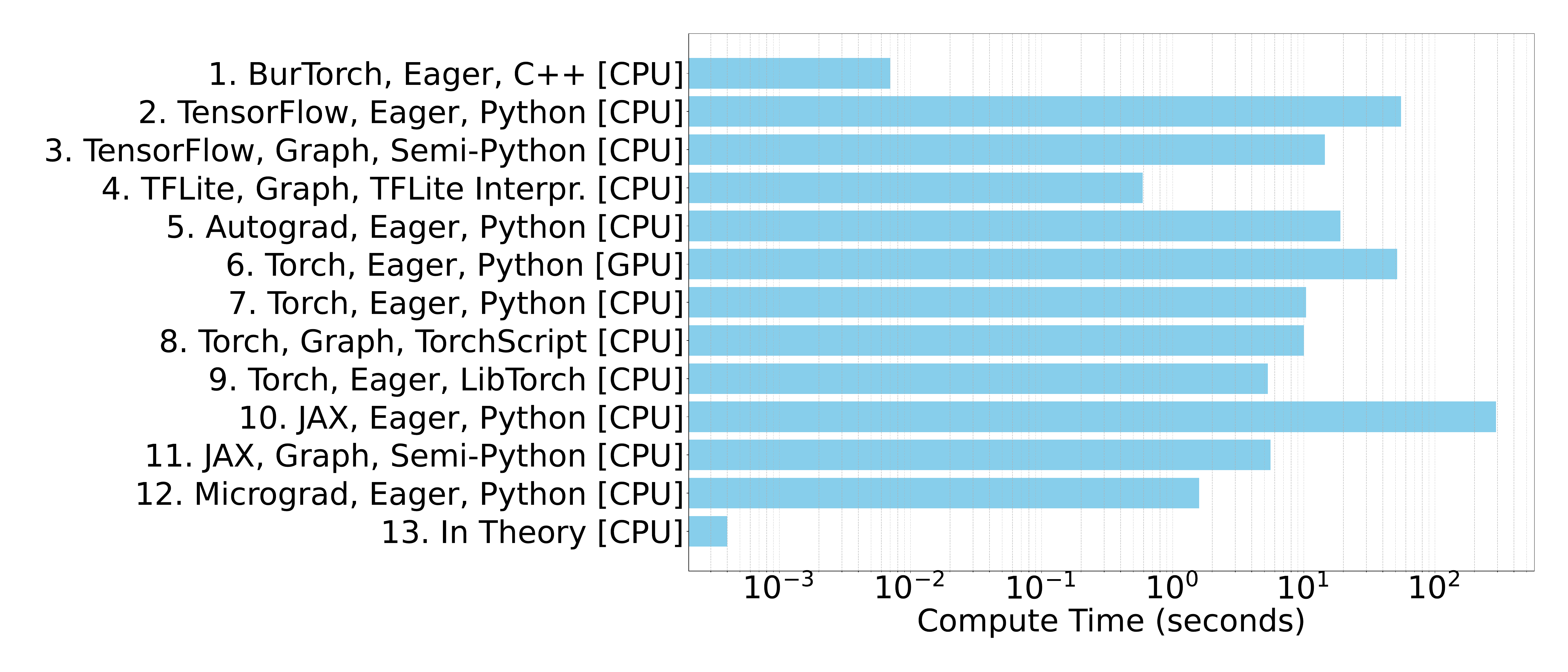}

\caption{Visualization of Table~\ref{ch8:tab:execution_times_speedup}. Backpropagation over $100$K iterations with a {tiny} dynamic compute graph from Figure~\ref{ch8:fig:tiny-compute-graph}. Computation in FP64, one CPU Core, Windows OS. The numerical results across frameworks match exactly.}

\label{ch8:fig:execution_times_speedup}
\end{figure*}

\begin{table*}[h!]
\footnotesize
\centering
\caption{Backpropagation over $20$K iterations with a \textit{small} dynamically constructed compute graph (Figure~\ref{ch8:fig:exp2-small-compute-graph}) with $32$ nodes, $5$ trials, FP64, Windows OS. The numerical results across frameworks match exactly.}

\begin{tabular}{|c|l|l|l|l|l|}
	\hline
	\textbf{\parbox{1.7cm}{\center \footnotesize{Framework,\\Mode,\\Device}}} &
	\textbf{\parbox{1.5cm}{\center \footnotesize{Compute Time\\(sec.)}}} & 
	\textbf{\parbox{1.7cm}{\center \footnotesize{Minimum Compute Time\\(sec.)}}} & 
	\textbf{\parbox{1.7cm}{\center \footnotesize{Total\\CPU Clocks \\ ($10^6$ ticks)}}} & 
	\multicolumn{2}{c|}{\makecell[c]{\textbf{Peak Memory}\\\textbf{(MB)}}} \\ 
	\cline{5-6}
	& & & & \textbf{Private} & \textbf{Resident} \\ 
	\hline
	\hline
	
	{\cellcolor{bgcolorwe} \makecell[c]{\cellcolor{bgcolorwe} BurTorch,\,\,\,\,\,\,\,\,\,\,\\ Eager, CPU}}&
	
	{\cellcolor{bgcolorwe}\begin{tabular}{@{}l@{}} \cellcolor{bgcolorwe} $0.0082$ \\ \cellcolor{bgcolorwe} $\pm 0.0003$ \end{tabular}} &
	\cellcolor{bgcolorwe}\begin{tabular}{@{}l@{}} {\color{abscolor} abs: $0.008$} \\ rel: $\times 1$ (We) \end{tabular} & 
	\cellcolor{bgcolorwe}\begin{tabular}{@{}l@{}} {\color{abscolor} abs: $39$} \\ rel: $\times 1$ \end{tabular} & 
	\cellcolor{bgcolorwe}\begin{tabular}{@{}l@{}} {\color{abscolor} abs: $0.6$} \\ rel: $\times 1$ \end{tabular} & \cellcolor{bgcolorwe}\begin{tabular}{@{}l@{}} {\color{abscolor} abs: $3.9$} \\ rel: $\times 1$ \end{tabular} \\ 
	\hline
	
	\makecell[c]{TensorFlow,\\Eager, CPU}&
	\begin{tabular}{@{}l@{}} $24.763$ \\ $\pm 0.5533$ \end{tabular} &
	\begin{tabular}{@{}l@{}} {\color{abscolor} abs: $24.158$} \\ rel: $\times 3019.7$ \end{tabular} & 
	\begin{tabular}{@{}l@{}} {\color{abscolor} abs: $75\,207$} \\ rel: $\times 1928.3$ \end{tabular} & 
	\begin{tabular}{@{}l@{}} {\color{abscolor} abs: $2120.8$} \\ rel: $\times 3534.6$ \end{tabular} & \begin{tabular}{@{}l@{}} {\color{abscolor} abs: $497.2$} \\ rel: $\times 127.5$ \end{tabular} \\ 
	\hline
	
	\makecell[c]{TensorFlow,\\Graph, CPU}&
	\begin{tabular}{@{}l@{}} $4.772$ \\ $\pm 0.4495$ \end{tabular} &
	\begin{tabular}{@{}l@{}} {\color{abscolor} abs: $3.874$} \\ rel: $\times 484.2$ \end{tabular} & 
	\begin{tabular}{@{}l@{}} {\color{abscolor} abs: $18\,065$} \\ rel: $\times 463.2$ \end{tabular} & 
	\begin{tabular}{@{}l@{}} {\color{abscolor} abs: $2121.6$} \\ rel: $\times 3536.0$ \end{tabular} & \begin{tabular}{@{}l@{}} {\color{abscolor} abs: $603.8$} \\ rel: $\times 154.82$ \end{tabular} \\ 
	\hline
	
	\makecell[c]{TF Lite,\\Graph, CPU}&
	\begin{tabular}{@{}l@{}} $2.435$ \\ $\pm 0.022$ \end{tabular} &
	\begin{tabular}{@{}l@{}} {\color{abscolor} abs: $2.400$} \\ rel: $\times 300.0$ \end{tabular} & 
	\begin{tabular}{@{}l@{}} {\color{abscolor} abs: $13\,624$} \\ rel: $\times 349.3$ \end{tabular} & 
	\begin{tabular}{@{}l@{}} {\color{abscolor} abs: $2247.1$} \\ rel: $\times 3745.1$ \end{tabular} & \begin{tabular}{@{}l@{}} {\color{abscolor} abs: $583.7$} \\ rel: $\times 149.6$ \end{tabular} \\ 
	\hline
	
	\makecell[c]{Autograd,\\Eager, CPU}&
	\begin{tabular}{@{}l@{}} $12.100$ \\ $\pm 0.0855$ \end{tabular} &
	\begin{tabular}{@{}l@{}} {\color{abscolor} abs: $11.971$} \\ rel: $\times 1496.3$ \end{tabular} & 
	\begin{tabular}{@{}l@{}} {\color{abscolor} abs: $36\,817$} \\ rel: $\times 944.0$ \end{tabular} & 
	\begin{tabular}{@{}l@{}} {\color{abscolor}  abs: $708.7$} \\ rel: $\times 1181.1$ \end{tabular} & \begin{tabular}{@{}l@{}} {\color{abscolor} abs: $29.7$} \\ rel: $\times 7.6$ \end{tabular} \\ 
	\hline
	
	\makecell[c]{PyTorch,\\Eager, \textbf{GPU}} &
	\begin{tabular}{@{}l@{}} $34.932$ \\ $\pm 0.1613$ \end{tabular} &
	\begin{tabular}{@{}l@{}} {\color{abscolor} abs: $34.784$} \\ rel: $\times 4348.0$ \end{tabular} & 
	\begin{tabular}{@{}l@{}} {\color{abscolor} abs: $83\,501$} \\ rel: $\times 2141.0$ \end{tabular} & 
	\begin{tabular}{@{}l@{}} {\color{abscolor} abs: $1639.5$} \\ rel: $\times 2732.5$ \end{tabular} & \begin{tabular}{@{}l@{}} {\color{abscolor} abs: $503.7$} \\ rel: $\times 129.15$ \end{tabular} \\ 
	\hline
	
	\makecell[c]{PyTorch,\\Eager, CPU,\\Python}&
	\begin{tabular}{@{}l@{}} $5.438$ \\ $\pm 0.0191$ \end{tabular} &
	\begin{tabular}{@{}l@{}} {\color{abscolor} abs: $5.418$} \\ rel: $\times 677.3$ \end{tabular} & 
	\begin{tabular}{@{}l@{}} {\color{abscolor} abs: $18\,977$} \\ rel: $\times 486.6$ \end{tabular} & 
	\begin{tabular}{@{}l@{}} {\color{abscolor}  abs: $1235.5$} \\ rel: $\times 2059.2$ \end{tabular} & \begin{tabular}{@{}l@{}} {\color{abscolor} abs: $334.3$} \\ rel: $\times 85.71$ \end{tabular} \\ 
	\hline
	
	\makecell[c]{PyTorch,\\ Graph, CPU, \\ TorchScript} &
	\begin{tabular}{@{}l@{}} $5.529$ \\ $\pm 0.0244$ \end{tabular} &
	\begin{tabular}{@{}l@{}} {\color{abscolor} abs: $5.505$} \\ rel: $\times 688.1$ \end{tabular} & 
	\begin{tabular}{@{}l@{}} {\color{abscolor} abs: $18\,956$} \\ rel: $\times 486.0$ \end{tabular} & 
	\begin{tabular}{@{}l@{}} {\color{abscolor} abs: $1245.7$} \\ rel: $\times 2076.2$ \end{tabular} &  \begin{tabular}{@{}l@{}} {\color{abscolor} abs: $346.8$} \\ rel: $\times 88.92$ \end{tabular} \\ 
	\hline
	
	{\centering {\begin{tabular}{@{}c@{}} PyTorch,\\Eager,CPU\\ LibTorch \end{tabular}}} &
	\begin{tabular}{@{}l@{}} $3.203$ \\ $\pm 0.0351$ \end{tabular} &
	\begin{tabular}{@{}l@{}} {\color{abscolor} abs: $3.155$} \\ rel: $\times 394.4$ \end{tabular} & 
	\begin{tabular}{@{}l@{}} {\color{abscolor} abs: $10\,283$} \\ rel: $\times 263.7$ \end{tabular} & 
	\begin{tabular}{@{}l@{}} {\color{abscolor} abs: $74.6$} \\ rel: $\times 124.3$ \end{tabular} & \begin{tabular}{@{}l@{}} {\color{abscolor} abs: $77.5$} \\ rel: $\times 19.9$ \end{tabular} \\ 
	\hline
	
	\makecell[c]{JAX,\\Eager, CPU} &
	\begin{tabular}{@{}l@{}} $185.862$ \\ $\pm 1.8866$ \end{tabular} &
	\begin{tabular}{@{}l@{}} {\color{abscolor} abs: $182.796$} \\ rel: $\times 22849.5$ \end{tabular} & 
	\begin{tabular}{@{}l@{}} {\color{abscolor} abs: $597\,088$} \\ rel: $\times 15\,309.9$ \end{tabular} & 
	\begin{tabular}{@{}l@{}} {\color{abscolor} abs: $791.3$} \\ rel: $\times 1318.8$ \end{tabular} & \begin{tabular}{@{}l@{}} {\color{abscolor} abs: $134.8$} \\ rel: $\times 34.5$ \end{tabular} \\ 
	\hline
	
	\makecell[c]{JAX,\\Graph, CPU} &
	\begin{tabular}{@{}l@{}} $1.184$ \\ $\pm 0.0189$ \end{tabular} &
	\begin{tabular}{@{}l@{}} {\color{abscolor} abs: $1.159$} \\ rel: $\times 144.9$ \end{tabular} & 
	\begin{tabular}{@{}l@{}} {\color{abscolor} abs: $4\,544$} \\ rel: $\times 116.5$ \end{tabular} & 
	\begin{tabular}{@{}l@{}} {\color{abscolor} abs: $765.2$} \\ rel: $\times 1275.3$ \end{tabular} & \begin{tabular}{@{}l@{}} {\color{abscolor} abs: $111.2$} \\ rel: $\times 28.5$ \end{tabular} \\ 
	\hline
	
	\makecell[c]{Micrograd,\\Eager, CPU}&
	\begin{tabular}{@{}l@{}} $1.071$ \\ $\pm 0.0074$ \end{tabular} &
	\begin{tabular}{@{}l@{}} {\color{abscolor} abs: $1.062$} \\ rel: $\times 132.8$ \end{tabular} & 
	\begin{tabular}{@{}l@{}} {\color{abscolor} abs: $3\,193$} \\ rel: $\times 81.9$ \end{tabular} & 
	\begin{tabular}{@{}l@{}} {\color{abscolor} abs: $6.2$} \\ rel: $\times 10.3$ \end{tabular} & \begin{tabular}{@{}l@{}} {\color{abscolor} abs: $12.2$} \\ rel: $\times 3.1$
	\end{tabular} \\
	\hline
	\cellcolor{bgcolorwe}\makecell[c]{In Theory, \\ Registers only,\,\\this CPU} & \multicolumn{2}{|c|}{\cellcolor{bgcolorwe} $\Omega(0.00033)$} &\cellcolor{bgcolorwe}{\color{abscolor}abs: $1.52$}   &
	\multicolumn{2}{|c|}{\cellcolor{bgcolorwe} \color{abscolor}abs: $0$}
	\\ \hline
	
	\hline
\end{tabular}
\label{ch8:tab:exp2-execution-times-speedup}
\end{table*}

\subsection{Small compute graph} 
\label{ch8:sec:exp2-small-compute-graphs}

In this experiment, we evaluate the performance of various frameworks on a small computation graph in Figure~\ref{ch8:fig:exp2-small-compute-graph} with 32 nodes, adapted from \citet{karpathy2020micrograd}. The comparison includes \libname{BurTorch}, \libname{TensorFlow}, \libname{TF Lite}, \libname{Autograd}, \libname{PyTorch}, \libname{Micrograd}, and \libname{JAX} running on Windows OS. The results are summarized in Table~\ref{ch8:tab:exp2-execution-times-speedup}. While popular frameworks, when utilized through C++ or Python (and Python-like languages via built-in compilers), exhibit nearly identical performance, \libname{BurTorch} stands out due to its faster computation. 

\libname{BurTorch} significantly reduces various forms of end-to-end memory allocation from the OS. We observe this in two key memory metrics. The first is the \textit{peak virtual private committed memory}. This represents the memory reserved by the OS, either in DRAM or in the swap file. While it may not be used by the application, Windows OS cannot reclaim it. This memory is non-shareable and cannot be reused by other applications. The second metric is the \textit{resident memory} (also referred to as the \textit{working set}), which is the memory allocated by the OS for applications from installed DRAM. Switching from \libname{Python} to \libname{TorchScript} and C++ with \libname{PyTorch} does lead to performance improvements, but the gains are not dramatic. This suggests that the issue with \libname{PyTorch} in this context extends beyond the language choice.


\subsection{Small compute graphs: loading and saving}
\label{ch8:sec:exp3-small-compute-graphs}

\begin{table}[h!]
\footnotesize
\small
\centering
\caption{Save and load subset of $7$ compute graph activations over $5 \cdot 10^3$ iterations to disk. The experiment uses a \textit{small}, dynamically constructed compute graph (Figure~\ref{ch8:fig:exp2-small-compute-graph}). Activations and gradients are computed in FP64 format. Raw payload size: $56$ bytes, $5$ trials.}
\begin{tabular}{|l|l|l|l|l|l|}
	\hline
	\textbf{\#} &
	\textbf{{\footnotesize{Framework, Mode, Language}}} & 
	\textbf{{\footnotesize{\parbox{1.5cm}{Save\\ (sec.)}}}} & 
	\textbf{{\footnotesize{\parbox{1.5cm}{Load\\ (sec.)}}}} & 
	\textbf{{\footnotesize\parbox{1.5cm}{{File Size\\ (bytes)}}}}
	\\ 
	\hline
	\hline
	
	\cellcolor{bgcolorwe}1 & \cellcolor{bgcolorwe}BurTorch, Eager, C++ 
	&\cellcolor{bgcolorwe}0.75 
	&\cellcolor{bgcolorwe}0.08
	&\cellcolor{bgcolorwe}56
	\\ 
	\hline

	2 & TensorFlow, Eager, Python 
	& 1.97
	& 1.66
	& 634
	\\ 
	\hline

	3 & \begin{tabular}{@{}l@{}} 
		TensorFlow, Graph, Semi-Python 
	\end{tabular}
	& 1.20
	& 0.30
	& 329
	\\ 
	\hline
	
	4 & \begin{tabular}{@{}l@{}} PyTorch, Eager, Python \\ {PyTorch} $\to$ {NumPy} $\to$ Bytes \end{tabular}
	& 1.11
	& 0.22
	& 56
	\\ 
	\hline

	5 & PyTorch, Eager, Python 
	& 2.54
	& 1.36
	& 2564
	\\ 
	\hline
	
	6 & PyTorch, Graph, TorchScript
	& 2.52
	& 1.34
	& 2564
	\\ 
	\hline
	
	7 & PyTorch, Eager, LibTorch, C++ 
	& 1.55
	& 0.95
	& 3569
	\\ 
	\hline

	\hline
\end{tabular}
\label{ch8:tab:exp2-save-load-speedup}
\end{table}

In this experiment, we assess the performance of several frameworks in terms of saving and loading activations (both leaf and intermediate nodes) from a small, dynamically constructed compute graph in Figure~\ref{ch8:fig:exp2-small-compute-graph} with 32 nodes from Section \ref{ch8:sec:exp2-small-compute-graphs}. The results are presented in Table~\ref{ch8:tab:exp2-save-load-speedup}.

The compute graph is used to simulate the saving and loading of activations over $5000$ iterations, and we have compared the performance of \libname{BurTorch} against \libname{TensorFlow}, \libname{PyTorch}, and \libname{JAX}. This experiment is conducted in the same hardware and software environment as the previous one. Activations from seven nodes, labeled (a)–(g) in Figure~\ref{ch8:fig:exp2-small-compute-graph} are saved. This results in a raw payload size of 56 bytes. Frameworks like \libname{Micrograd}, \libname{Autograd}, and \libname{JAX} are excluded from the comparison due to their lack of native support for saving computation graphs. \libname{BurTorch} not only significantly reduces save and load times but also minimizes the size of data transferred to and from disk.

\subsection{{BurTorch} on medium compute graphs}
\label{ch8:sec:exp3-medium-compute-graphs}

We evaluated \libname{BurTorch} on a character-level autoregressive prediction model, designed as a medium-complexity compute graph, based on the architecture from \citet{bengio2000neural}. This model uses English characters, each represented as a trainable embedding variable from $\mathbb{R}^{64}$. The dataset consists of 26 characters and a special token for sequence start, end, and padding, totaling 27 tokens. The context length (block size) for predicting the next-token is set to $16$. All trainable parameters, activations, and computed derivatives are represented using FP32. The training dataset, from \citet{makemore2023}, contained $n=228,146$ samples.

We compared \libname{BurTorch} with a \libname{PyTorch} implementation from \citet{makemore2023}. Tables \ref{ch8:tab:exp3-compute-and-mem-speedup-b1} and \ref{ch8:tab:exp3-compute-and-mem-speedup-b64} show the performance comparison for models with varying numbers of trainable parameters, ranging from $5,963$ to $1,079,003$.

During the experiments, we train using \algname{SGD} with the gradient estimator given by Equation~\ref{ch8:eq:main-grad-est}:
	
$$x^{k+1}=x^{k} - \gamma \dfrac{1}{b}\sum_{i\in S_k} \nabla f_i(x^k),$$
where $S_k \sim_{\mathrm{u.a.r}} \{s:s \in 2^{[n]} \land |s|=b\}$. Tables \ref{ch8:tab:exp3-compute-and-mem-speedup-b1} and \ref{ch8:tab:exp3-compute-and-mem-speedup-b64} report the results for batch sizes of $b=1$ and $b=64$, respectively. Trainable pairs for the autoregressive model were sampled uniformly at random from the training dataset. The model size was adjusted by modifying the number of hidden units in the first layer and the corresponding input units in the second layer (both layers use tanh activation).

\paragraph{Initialization time.} This time refers to the total time required for setting up the training process, including data loading, preprocessing, one backpropagation iteration (considering \libname{PyTorch’s} deferred memory allocations), and deinitialization. As shown in Table~\ref{ch8:tab:exp3-compute-and-mem-speedup-b1}, \libname{PyTorch's} initialization time for a model with $5,963$ parameters was equivalent to $3,794$ gradient oracle computations with $b=1$, whereas \libname{BurTorch} reduced this time to $488$ gradient oracles.

\paragraph{Memory efficiency.} We measured peak private virtual memory usage, representing the maximum memory allocated exclusively for the training process. For a compute graph with $1,079,003$ parameters, \libname{BurTorch} required $107$ MB of memory only, while \libname{PyTorch} required over $2.7$ GB for both $b=1$ and $b=64$. This highlights that the minimalist design of \libname{BurTorch} significantly reduces memory consumption, making it suitable for resource-constrained environments.

\paragraph{Compute time.} The reported time is the estimated mean and standard deviation over $4000$ iterations of \algname{SGD} as a computational model. \libname{BurTorch} consistently outperformed \libname{PyTorch} in terms of expected compute time for compute graphs with $d \leq 10^6$ parameters. For example, for $d=5,963$ and $b=1$, \libname{PyTorch} was approximately $45$ times slower than \libname{BurTorch}. The complex architecture of \libname{PyTorch} led to a more variable standard deviation in execution time.

\paragraph{Comparative analysis of computation time.} 

As shown in Table~\ref{ch8:tab:exp3-compute-and-mem-speedup-b1}, \libname{BurTorch} consistently outperforms \libname{PyTorch} for $b=1$ with $d$ ranging from 6K to 1M. While \libname{BurTorch} demonstrates a clear performance advantage for smaller compute graphs or small batch sizes, the gap narrows as the model size exceeds 100K parameters for $b=64$. This is because \libname{PyTorch} implementation leverages techniques such as optimized kernels and vectorized operations designed for maximum throughput.

\paragraph{Comparative analysis of memory usage.} Limited memory in edge devices poses challenges during inference \citep{laskaridis2024melting} and becomes even more critical during training. If peak memory usage exceeds available capacity and $\nabla f(x)$ computation cannot be controlled, it may \textit{seriously} disrupt the execution.

\begin{table*}[h!]
\footnotesize
\centering
\caption{Comparison of \libname{BurTorch} and \libname{PyTorch} performance for training MLP-like model. Batch: $b=1$, Compute: FP32, Single CPU core. Initialization time is end-to-end time for training with $1$ iteration. Compute time excludes batch preparation. Memory is the peak private virtual memory.}
\label{ch8:tab:exp3-compute-and-mem-speedup-b1}
\begin{tabular}{|l|l|l|l|l|l|l|l|}
	\hline
	\textbf{\#} &\textbf{Parameters (d)} & \multicolumn{3}{c|}{\textbf{\makecell[c]{PyTorch,\\ Eager, v2.5.1 [CPU]}}} & \multicolumn{3}{c|}{\cellcolor{bgcolorwe}{\textbf{{BurTorch, Eager [CPU]}}}} \\ 
	\cline{3-8}
	&\textbf{Hidden Dim.(e)} & \textbf{\makecell[c]{Init\\(ms)}} & \textbf{\makecell[c]{Compute\\(ms)}} & \textbf{\makecell[c]{Mem.\\(MB)}} & {\textbf{\makecell[c]{Init\\(ms)}}} & {\textbf{\makecell[c]{Compute\,\,\,\,\,\,\\(ms)}}} & {\textbf{\makecell[c]{Mem.\\(MB)}}} \\ 
	\hline
	\hline
	1&$5,963\ (e=4)$ & $5\,540$ & $1.46 \pm 4.63$ & $2\,651$ & $15.63$ & $0.032 \pm 0.008$ & $35.8$ \\ 
	2&$18,587\ (e=16)$ & $5\,627$ & $1.52 \pm 4.21$ & $2\,653$ & $16.51$ & $0.074 \pm 0.016$ & $36.7$ \\ 
	3&$35,419\ (e=32)$ & $5\,673$ & $1.55 \pm 5.00$ & $2\,653$ & $18.24$ & $0.124 \pm 0.019$ & $38.3$ \\ 
	4&$69,083\ (e=64)$ & $5\,537$ & $1.63 \pm 4.62$ & $2\,668$ & $18.94$ & $0.221 \pm 0.040$ & $40.8$ \\ 
	5&$136,411\ (e=128)$ & $5\,799$ & $1.79 \pm 5.19$ & $2\,660$ & $21.39$ & $0.417 \pm 0.077$ & $45.9$ \\ 
	6&$540,379\ (e=512)$ & $5\,556$ & $3.01 \pm 5.57$ & $2\,683$ & $37.09$ & $2.093 \pm 0.429$ & $71.4$ \\ 
	7&$1,079,003\ (e=1024)$ & $5\,544$ & $5.57 \pm 6.75$ & $2\,719$ & $56.57$ & $4.550 \pm 0.847$ & $107.0$ \\ 
	\hline
\end{tabular}
\end{table*}

\begin{table*}[h!]
\footnotesize
\centering
\caption{Comparison of \libname{BurTorch} and \libname{PyTorch} performance for training MLP-like model. Batch: $b=64$, Compute: FP32, Single CPU core. Initialization time is end-to-end time for training with $1$ iteration. Compute time excludes batch preparation. Memory is the peak private virtual memory.}
\label{ch8:tab:exp3-compute-and-mem-speedup-b64}
\begin{tabular}{|l|l|l|l|l|l|l|l|}
	\hline
	\textbf{\#} & \textbf{Parameters (d)} & \multicolumn{3}{c|}{\textbf{\makecell[c]{PyTorch,\\ Eager, v2.5.1 [CPU]}}} & \multicolumn{3}{c|}{\cellcolor{bgcolorwe}{\textbf{{BurTorch, Eager [CPU]}}}} \\ 
	\cline{3-8}
	&\textbf{Hidden Dim.(e)} & \textbf{\makecell[c]{Init\\(ms)}} & \textbf{\makecell[c]{Compute\\(ms)}} & \textbf{\makecell[c]{Mem.\\(MB)}} & {\textbf{\makecell[c]{Init\\(ms)}}} & {\textbf{\makecell[c]{Compute\\(ms)}}} & {\textbf{\makecell[c]{Mem.\\(MB)}}} \\ 
	\hline
	\hline
	1&$5,963\ (e=4)$ & $5\,658$ & $6.17 \pm 4.71$ & $2\,678$ & $17.33$ & $0.52 \pm 0.05$ & $35.8$ \\ 
	2&$18,587\ (e=16)$ & $5\,774$ & $6.39 \pm 4.57$ & $2\,678$ & $17.38$ & $1.70 \pm 0.12$ & $36.7$ \\ 
	3&$35,419\ (e=32)$ & $5\,941$ & $7.79 \pm 4.40$ & $2\,679$ & $18.29$ & $3.25 \pm 0.17$ & $38.3$ \\ 
	4&$69,083\ (e=64)$ & $6\,020$ & $9.43 \pm 4.81$ & $2\,686$ & $23.62$ & $6.50 \pm 0.31$ & $40.8$ \\ 
	5&$136,411\ (e=128)$ & $6\,176$ & $13.01 \pm 4.88$ & $2\,695$ & $35.62$ & $13.47 \pm 0.71$ & $45.4$ \\ 
	6&$540,379\ (e=512)$ & $6\,151$ & $34.97 \pm 4.05$ & $2\,698$ & $102.19$ & $64.17 \pm 3.22$ & $71.4$ \\ 
	7&$1,079,003\ (e=1024)$ & $5\,926$ & $64.61 \pm 6.10$ & $2\,723$ & $197.67$ & $145.87 \pm 6.04$ & $106.9$ \\ 
	\hline
\end{tabular}
\end{table*}


\subsection{{BurTorch} on a GPT-3-like computation graph} 
\label{ch8:sec:exp4-burtorch-fot-gpt3}

Recent advances in Large Language Models (LLMs) indicate significant progress toward artificial general intelligence \citep{bubeck2023sparks}. These models enable intuitive natural language interactions. Their training relies on large-scale data collection, careful tokenization strategy, and fine-tuning for question-answer alignment. First-order continuous optimization, implemented through gradient estimates as given in Equation~\ref{ch8:eq:main-grad-est}, is fundamental to their training and fine-tuning.

\paragraph{\color{black}{Experiment setup for GPT-3-like model.}}
\libname{BurTorch} offers the essential computational primitives to support gradient oracle computation while training Transformer-based models. We demonstrate this by constructing a \modelname{GPT}-like model, following the architectures of \citet{transformer} and \modelname{GPT-3} \citep{gpt}. This decoder-only Transformer is designed for the next-token character generation, with parameter counts ranging from $125 \cdot 10^6$ to $125 \cdot 10^9$ \citep{gpt}. To ensure computational feasibility, we significantly scaled down the original \modelname{GPT-3} model by modifying the configuration parameters outlined in Table 2.1 of \citet{gpt}. Although the scalar-level computation steps in the \modelname{GPT-3} architecture are complex, they can be expressed using the primitives detailed in Appendix~\ref{ch8:app:burotrch-scalar-operators} and \ref{ch8:app:burotrch-derived-operators}. For additional information on Transformer architectures, see \citet{transformer, scardapane2024alice, bishop2023deep}.

\paragraph{Input.} In this experiment, input characters from the Shakespeare dataset \citep{tinyshakespeare} are tokenized into integers, consisting of 65 unique ASCII characters (including padding), resulting in a vocabulary size of $V = 65$. The input sequence (context length) tokens are processed, and the corresponding output sequence (context length) tokens are generated. Once a specific sentence is sampled, the corresponding trainable token embeddings and positional embeddings are added elementwise at each position in the context, without additional transformations.

\paragraph{Internals.}
Next, the sequence of characters with a fixed length, known as the block size or context length, is processed. This represents a typical sequence of tokens handled by the Transformer architecture.

Internally, the generative model consists of multiple layers, which are named Transformer encoder blocks. In our experiment, we use six, but in general, it is a meta-parameter. Each Transformer encoder \citep{transformer} performs the following computations:
(a) six self-attention heads, followed by a concatenation of their outputs and an affine transformation into the embedding space.
(b) two residual connections.
(c) two layer normalization layers.
(d) one two-layer feed-forward neural network with affine transformations.

\paragraph{Output.} 
The \modelname{GPT-3} text generation process involves passing the input context through a series of encoder blocks. After this, finally, each token in the context length block is processed with an affine transformation to generate logits in a vector space of size $V$.

Logits in this space are processed with a softmax function to produce the probability distribution for the token. The loss function, $CE(p,\hat{p}) = -\sum_{i=1}^{V} p_i \log(\hat{p_i})$, compares the true probability mass function (p.m.f.) for a next-token in sequence and predicted p.m.f.

\paragraph{GPT-3-like model: configuration.}

The miniaturized \modelname{GPT-3} configuration which has been used in experiments includes: (i) \texttt{n\_layer=6} Multi-head Self-Attention encoder blocks; (ii) \texttt{k\_heads=6} heads in each block; (iii) \texttt{k\_block\_size=8} for context length (input sequence length); (iv) \texttt{d\_model=24} for the embedding dimension. Computation was performed in FP32 on a single-core CPU, with a total of $46,289$ trainable parameters. The number of \algname{SGD} iterations is $3000$.

\paragraph{GPT-3-like model: experimental results.}

The results in Table~\ref{ch8:tab:exp4-compute-and-mem-speedup-win} compare \libname{BurTorch} with a baseline \libname{Python} implementation \citet{karpathygptimplementation} of a \modelname{GPT-3}-type model, focusing on peak memory consumption and the mean and standard deviation of the computation time for a single \algname{SGD} oracle with a fixed batch size over $3$K \algname{SGD} oracles. Additionally, we evaluate its performance against \libname{PyTorch} implementation optimized with \libname{TorchScript}, a built-in compilation technique in \libname{PyTorch} (for more details see Appendix~\ref{ch8:app:torch-compile-techniques-backprop-speed-cpu}).

As shown in Table~\ref{ch8:tab:exp4-compute-and-mem-speedup-win}, \libname{BurTorch} achieves a $\times 20$ speedup with a batch size of $1$ and reduces memory requirements by $\times 100$. The memory metric represents peak virtual private memory usage. Notably, \libname{BurTorch} also exhibits lower execution time variance than the baseline. By processing samples sequentially, it avoids storing all activations simultaneously, preventing memory usage from scaling with batch size. As batch size increases to $64$, \libname{PyTorch} design outperforms \libname{BurTorch} in terms of time per batch by $\times 1.4$.

\begin{table*}
\footnotesize
\centering
\caption{\libname{BurTorch} and \libname{PyTorch} in training \modelname{GPT-3} like model, FP32, 1 CPU core, Peak private virtual memory. Trainable variables: $46$K.}
\label{ch8:tab:exp4-compute-and-mem-speedup-win}
\begin{tabular}{|l|c|c|c|c|c|c|}
	\hline
	\textbf{Batch} & \multicolumn{2}{c|}{\cellcolor{bgcolorwe}{\textbf{BurTorch, Eager, C++}}} & \multicolumn{2}{c|}{\textbf{\makecell[c]{PyTorch,\\ Graph, TorchScript}}} & \multicolumn{2}{c|}{\textbf{\makecell[c]{PyTorch,\\ Eager, Python}}} \\ 
	\cline{2-7}
	& {\textbf{\makecell[c]{Compute\\(ms)}}} & {\textbf{\makecell[c]{Mem.\\(MB)}}} & \textbf{\makecell[c]{Compute\\(ms)}} & \textbf{\makecell[c]{Mem.\\(MB)}} & \textbf{\makecell[c]{Compute\\(ms)}} & \textbf{\makecell[c]{Mem.\\(MB)}} \\ 
	\hline
	\hline
	$1$ & $0.515 \pm 0.067$ & $16.7$ & $11.119 \pm 48.118$ & $1\,624$ & $11.715 \pm 10.741$ & $1\,300$ \\ 
	$2$ & $1.027 \pm 0.091$ & $16.7$ & $11.177 \pm 37.138$ & $1\,623$ & $12.166 \pm 11.461$ & $1\,300$ \\ 
	$4$ & $2.106 \pm 0.130$ & $16.7$ & $11.762 \pm 37.171$ & $1\,624$ & $12.424 \pm 11.120$ & $1\,300$ \\ 
	$8$ & $4.222 \pm 0.238$ & $16.7$ & $12.041 \pm 36.312$ & $1\,631$ & $13.167 \pm 11.613$ & $1\,308$ \\
	$16$ & $8.358 \pm 0.644$ & $16.7$ & $13.451 \pm 37.415$ & $1\,633$ & $14.111 \pm 11.278$ & $1\,308$ \\ 
	$32$ & $16.787 \pm 1.03601$ & $16.7$ & $16.048 \pm 36.460$ & $1\,632$ & $16.661 \pm 11.122$ & $1\,308$ \\ 
	$64$ & $31.696 \pm 0.737$ & $16.8$ & $21.794 \pm 37.302$ & $1\,640$ & $22.189 \pm 11.531$ & $1\,316$ \\
	\hline
\end{tabular}
\end{table*}

\section{Designs Behind BurTorch's Low Latency}
\label{ch8:sec:design-for-low-latency}

\libname{BurTorch} excels in CPU gradient computation, especially for small batch sizes, making it ideal for memory-constrained scenarios. Next, we outline key design choices that contribute to its low latency. For further details, see Appendix \ref{ch8:app:design-philosophy}.

\paragraph{\color{black}Compile-time optimizations.}
\libname{BurTorch} maximizes compile-time optimizations usage, reducing runtime overhead by treating constants as compile-time values, thus optimizing execution. Unlike high-level frameworks, which limit such optimizations, \libname{BurTorch} leverages the full potential of modern CPUs, bypassing performance bottlenecks related to loops and memory access in all computation stacks. \libname{BurTorch} advocates for compile-time optimization across a fully integrated deep-learning training solution.

\paragraph{Eliminating unnecessary abstractions.}
Operating with no dependencies except for the OS, \libname{BurTorch} simplifies runtime execution. Through dead code elimination and whole-program optimization, compilers remove or simplify abstractions that exist only at design time. A small codebase offers several advantages. First, it results in a physically smaller compiled binary. Second, a smaller codebase allows for meticulous optimization of every detail coherently and holistically without neglecting other aspects of training or gradient computation—something that becomes practically \textit{impossible} with a large codebase.

\paragraph{\color{black}Instruction-level parallelism.} \libname{BurTorch} employs instruction-level parallelism by unrolling key operations like inner products, ensuring efficient use of CPU pipelines. By simplifying the compiler’s task, it helps maximizing performance, avoiding inefficiencies common in complex code structures. For further details, see Appendix \ref{ch8:app:compiler-in-general}.

\paragraph{\color{black}Efficient memory management.}

\libname{BurTorch} optimizes memory access by reducing latency, particularly when storing operands in DRAM. It stores partial derivatives and activations contiguously, minimizing fragmented memory access. \libname{BurTorch} introduces a "rewind" mechanism that discards unnecessary parts of the computation graph.

The ability to easily customize \libname{BurTorch} is also important. For example, in the Transformer block (Section \ref{ch8:sec:exp4-burtorch-fot-gpt3}), the concatenation of multiple self-attention heads typically involves memory copying in typical frameworks. \libname{BurTorch} avoids this by passing a sequence of memory views for linear layers, conceptually representing a split tensor without physically concatenating it. This approach eliminates unnecessary memory copies, which are approximately $\times 330$ more expensive compared to a single arithmetic operation if memory access happens to DRAM \citep{gregg2014systems}. This functionality is achievable with \libname{BurTorch}'s small codebase.

\paragraph{\color{black}Optimized backpropagation.}
\libname{BurTorch}'s backpropagation is optimized for latency. For details see Appendix~\ref{ch8:app:cap-burtorch-high-with-low}.

\section{Impact on Optimization Theory} 
\label{ch8:sec:inluence-on-theory}

\paragraph{\color{black}Practical implementations of theoretical algorithms.} \libname{BurTorch} optimizes the Backpropagation Algorithm through system-level improvements, efficient memory access, and a design that significantly reduces runtime overhead in both memory and computation.  It introduces a practically effective gradient oracle implementation  $\nabla f(x) = \nicefrac{1}{b} \sum_{i=1}^{b} \nabla f_i(x)$ which outperforms all state-of-the-art practical solutions for small batch sizes $b$ or low-dimensional settings $d$.

First, $b = 1$ is theoretically optimal for non-convex algorithms like \algname{PAGE} \citep{li2021page, tyurin2022sharper}. Before \libname{BurTorch}, achieving such low-latency gradient computation was impractical, limiting \algname{PAGE} applicability. \libname{BurTorch} removes this barrier, making \algname{PAGE} more practical. 

Second, in finite-sum convex optimization, where \algname{Proximal Stochastic Gradient Descent} with \algname{SGD-NICE} subsampling requires an optimal batch size of $\tau \approx 1$ to maintain low gradient variance \citep{Gower2019}, \libname{BurTorch} enables efficient computation of the subsampled gradient $\hat{\nabla f(x)}$.

\libname{BurTorch} demonstrates that system-level optimizations, such as refining memory access patterns, adopting non-recursive computation, and reusing memory buffers, have a significant impact on performance. These optimizations dramatically reduce hidden constants in $\mathcal{O}$ notation, offering a new perspective on optimization algorithms in the context of modern hardware constraints. Notably, DRAM access latency is approximately $330\times$ slower than register access \citep{gregg2014systems}, underscoring the importance of memory access considerations in the theoretical optimization algorithms.

\paragraph{Modifying the notion of the gradient oracle.}

Some optimization algorithms enable the interleaving of gradient computation, compression, and information transfer \citet{burlachenko2023federated}. However, implementing these techniques effectively in reality requires significant effort to seamlessly integrate communication and computation within Automatic Differentiation (AD). Research on asynchronous \algname{SGD} methods \citep{maranjyan2024mindflayer, maranjyan2025ringmasterasgdasynchronoussgd} further necessitates the incorporation of early termination—the ability to halt the computation of $\nabla f(x)$ upon request. However, even minor modifications to backpropagation often lack practical implementations in academic settings due to the inherent complexity of conventional frameworks. Maximizing throughput frequently results in overly intricate and non-maintainable implementations. \libname{BurTorch} simplifies these processes due to its compact design.

\paragraph{Randomized backpropagation.}

Early work by \citet{oktay2020randomized} investigated a randomization technique within backpropagation, but its full potential remains largely unexplored. The scalar-level granularity of \libname{BurTorch} allows these randomization techniques to be directly \textit{implemented} by modifying its codebase, rather than introducing additional layers to \textit{simulate} randomization.

\paragraph{Practical refinements for theoretical algorithms.}

A significant research direction in Federated Learning focuses on compression techniques that apply compression operations to gradient-like quantities. These methods use specialized operators, such as $\mathcal{C}(\hat{\nabla f(x)} - g)$ in \algname{EF21} by \citet{richtarik2024error} and $\mathcal{C}(\hat{\nabla f(x)} - \hat{\nabla f(y)})$ in \algname{MARINA} by \citet{gorbunov2021marina}, where $\mathcal{C}: \RD \to \RD$ is deterministic or randomized mapping. By design, these compression operators enable efficient transmission of compressed information over communication networks. For a survey on compression operators, see \citet{beznosikov2020biased}. For optimization algorithms that require efficient computation of the gradients \(\nabla f(x)\) and \(\nabla f(y)\) at two different iterates \(x, y \in \RD\), \libname{BurTorch} provides this functionality effectively out of the box. By leveraging low-level optimizations with {SIMD registers}, \libname{BurTorch} ensures high-performance execution natively.

\paragraph{Refining gradient compression to partial derivative granularity.}

Some compression methods \citep{beznosikov2020biased} operate independently of evaluated $\nabla f(x)$, meaning they do not require prior knowledge of $\nabla f(x)$ before being applied. \compname{RandK} sparsification, which selects a subset of $[d]$ coordinates independently of $\nabla f(x)$, is one such example. 

The time complexity of computing $\frac{\partial f(x)}{\partial x_i}$ depends on the location of $x_i$ within the computation graph. If $x_i$ is in the early layers of a DL model, both forward and backward passes are required. In contrast, if $x_i$ is in the later layers, only a forward pass and a single backward step are needed, avoiding traversal of the entire model during backward. When computing $\frac{\partial f}{\partial \textbf{z}}$ for $\textbf{z} \in \{x_1, \dots, x_k\}$, the computational cost depends on their positions in the computation graph. This idea can be leveraged to design more efficient compression techniques. The \compname{RandSeqK} compressor \citep{burlachenko2024unlocking} was developed to group spatially close coordinates, optimizing memory access through coalesced memory operations. A similar strategy could be employed by exploiting the internal flexibility of the \compname{RandK} compressor to create structure-aware compression methods for $f(x)$. The fast oracle in \libname{BurTorch} enables computation of $\nabla f(x)$ restricted to specific coordinate subset $S$, facilitating the development of either exact (for small $d$) or approximate cost models (for big $d$) to evaluate $[\nabla f(x)]_{i,i \in S}$.

\paragraph{Coupling gradient computation with sparsification.}

While gradient sparsification can reduce communication and storage overheads by decreasing the number of non-zero gradients, modern Deep Learning frameworks such as \libname{JAX}, \libname{TensorFlow}, and \libname{PyTorch} do not natively offer direct support for sparse gradients or gradients with specific structures. In contrast, \libname{BurTorch}, with its transparent architecture, can be conceptually adapted to support gradient sparsification, as long as the sparsification rule is based solely on individual $\frac{\partial f_i(x)}{\partial x_j}$, and not on the entire gradient $\nabla f_i(x)$ or $\nabla f(x)$.

\section{Conclusions}
\label{ch8:sec:conclusion}

We introduced \libname{BurTorch}, a CPU-based backpropagation implementation. Real-world experiments conducted across Windows (Section~\ref{ch8:sec:experiments}), Linux (Appendix~\ref{ch8:app:extra-experiments-linux}), and macOS (Appendix~\ref{ch8:app:extra-experiments-macos}) on various devices demonstrate significant improvements in computation latency, memory usage, and energy efficiency (Appendix~\ref{ch8:app:energy-eff}) during the computation of $\nabla f(x)$. \libname{BurTorch} has been compared against \libname{JAX}, \libname{TensorFlow}, \libname{PyTorch}, \libname{Apple MLX}, \libname{Autograd}, \libname{TFLite}, and \libname{Micrograd}. The compact codebase of \libname{BurTorch} lays the foundation for adapting optimization techniques to backpropagation and tailoring backpropagation to specific optimization algorithms in scenarios where $\nabla f(x)$ is computed through backpropagation.

\clearpage

\appendix

\part*{Appendices to Chapter \ref{chapter8}}
\label{ch8:app:toc_1}
\newpage

\phantomsection
\addcontentsline{toc}{chapter}{Appendices to Chapter 8}

\addtocounter{adjsection}{1}
\section{Missing Details for Experimental Setup}
\label{ch8:app:exp-setup}

\subsection{Hardware and software environment for Windows OS experiments}
\label{ch8:app:exp-setup-win}

\begin{enumerate}
\item Operating System: Microsoft Windows 11 Home.
\item CPU: Intel Core Ultra 7 155H, x86-64, Little-Endian.
\item CPU Clock Frequency: $4.48$ GHz during experiments (via \texttt{powercfg.cpl})
\item CPU Cores: 22 Logical, 16 Physical. During the experiments, only one core was used.
\item Physical Memory: 64 GBytes, DDR5, $2792$ MHz.
\item Hard Drive: KXG80 NVMe SSD, Sector size: 512 bytes, Filesystem: NTFS
\item Python Interpreter: Python: 3.9.0,
\item Python Libraries: TensorFlow: 2.8.0, Autograd 1.7.0, PyTorch: 2.5.1, JAX: 0.4.30, Micrograd: c911406, NumPy: 1.23.0, TensorFlow Lite (the same TensorFlow version): 2.8.0. C++ Libraries: LibTorch (C++ \libname{PyTorch} API): 2.5.1
\item {BurTorch} and LibTorch examples built with: Microsoft Visual Studio 2020 v17.11.4 (MSVC 14.41.34120) and \texttt{/O2,/Oi,/GL}, CMake version: 3.30.4
\end{enumerate}

\subsection{Hardware and software environment for Linux OS experiments}
\label{ch8:app:exp-setup-linux}

\begin{enumerate}
\item Operating System: Ubuntu 20.04.6 LTS.
\item CPU: Intel(R) Xeon(R) Gold 6146 CPU, x86-64, Little-Endian.
\item CPU Clock Frequency: $3.2$ GHz during experiments.
\item CPU Cores: 24 Logical, 24 Physical. During the experiments, only one core was used.
\item Physical Memory: 251 GBytes, DDR4, $2666$ MHz.
\item Hard Drive: Seagate ST4000NM0035, HDD, Sector size: 512 bytes, Filesystem: EXT4.
\item Python Interpreter: Python: 3.9.21.
\item Python Libraries: TensorFlow: 2.18.0, Autograd 1.7.0, PyTorch: 2.5.1, JAX: 0.4.30, Apple MLX: 0.22.0, Micrograd: c911406, NumPy: 2.0.2. LibTorch (C++ {PyTorch} API): 2.5.1.
\item {BurTorch} and LibTorch examples was build with g++-11 and \texttt{-O3 -flto} compilers flags, CMake version: 3.27.
\end{enumerate}

\subsection{Hardware and software environment for macOS experiments}
\label{ch8:app:exp-setup-macos}

\begin{enumerate}
\item Operating System: Sonoma 14.5.
\item CPU: Intel Quad-Core Intel Core i7, x86-64, Little Endian.
\item CPU Clock Frequency: $2.3$ GHz during experiments.
\item CPU Cores: 8 Logical, 4 Physical. During the experiments, only one core was used.
\item Physical Memory: 32 GBytes, LPDDR4X, $3733$ MHz.
\item Hard Drive: Macintosh HD APPLE SSD AP2048N, SSD, Sector size: 4096 bytes, Filesystem: APFS.
\item Python Interpreter: Python: 3.9.21.
\item Python Libraries: TensorFlow: 2.16.2, Autograd 1.7.0, PyTorch: 2.2.2, JAX: 0.4.30, Apple MLX: 0.7.0, Micrograd: c911406. LibTorch (C++ {PyTorch} API): 2.2.2.
\item {BurTorch} and LibTorch examples was build with Apple CLang 15.0.0 and \texttt{-O3 -flto} compilers flags, CMake version: 3.27.
\end{enumerate}

\clearpage
\addtocounter{adjsection}{1}
\section{Discussion on Compiler Considerations}
\label{ch8:app:compiler-in-general}

\subsection{On why a compile-based approach is fundamentally different from utilizing scripting languages}
\label{ch8:app:why-compilers}

As noted in \citet{wexelblat2014history}, compiler research dates back to the 1950s, with John Backus’s Speedcoding project addressing rising software costs. While compiled languages are often considered low-level in modern ML research, scripting languages, due to their higher levels of abstraction, can obscure essential control over interactions with hardware and the OS. Understanding why a compiler-based approach is superior requires examining the inner workings of a compiler system. A compiler operates in several distinct stages:

\begin{enumerate} 
\item \textbf{Lexical analysis.} The implementation's source code is divided into tokens which are the smallest indivisible units within a language, such as keywords, operators, separators, identifiers, and literal constants. These symbols form the foundation for all subsequent analysis.

\item \textbf{Syntax analysis.}  The compiler constructs an Abstract Syntax Tree (AST), validating that the program adheres to the language’s syntactic rules, defined typically using Backus–Naur forms for context-free grammars (CFG).

\item \textbf{Semantic analysis.} This phase ensures that the program is semantically correct, checking type compatibility, variable scope, and other areas that syntax analysis alone cannot address.

\item \textbf{Code optimization.} Code optimization improves runtime performance and reduces memory usage in various forms.

\item \textbf{Code generation.} Finally, the compiler generates final machine instructions, directly producing binary code for the target architecture or intermediate assembly code, which is then converted into the machine code.

\end{enumerate}

In contrast, interpreters execute scripts directly at runtime, without preprocessing. Therefore interpreters typically introduce significant overhead by bridging abstractions during execution. This results in inefficiency in runtime, especially in large-scale or real-time systems like ML, where performance is critical. 

To address these challenges, there are efforts from scripting-based approaches:

\begin{itemize} 
\item Establishing best practices for specific scripting languages.
\item Developing specialized libraries or frameworks for performance improvement.
\item Augmenting the scripting ecosystem with just-in-time (JIT) compilers.
\item Improving runtime efficiency through external tools or hybrid approaches.
\end{itemize}

However, in our vision, adopting a compiler-based approach in ML research presents a significantly more efficient and scalable alternative. By leveraging the power of compilers, researchers can achieve superior performance, direct hardware control, and enhanced scalability—eliminating the inefficiencies and compromises inherent in interpreted languages.

Historically, this approach has not been widely adopted, primarily because large codebases make compilation slow, creating a social barrier to its use. \libname{BurTorch}'s compact code design overcomes this limitation, making compiler-based ML development both practical and efficient.

\subsection{On performance optimizations achievable through compiler utilization}
\label{ch8:app:what-compilers-can}

At the \textbf{Optimization} stage, the Abstract Syntax Tree (AST) for the program has been constructed and augmented with information derived from the semantic analysis phase. Compiler optimizations are rooted in a broad array of innovations from various scientific and engineering disciplines, which, when applied effectively, yield substantial performance improvements. To try to optimize computation time, the compilers generally execute a sequence of transformation passes, each of which analyzes and refines the code to optimize its performance. Each transformation pass may iterate multiple times, with the steps typically following a predetermined order that has been demonstrated empirically effective in most cases. The following summarizes common optimization techniques:

\begin{itemize}
\item \textbf{Arithmetic simplification.} Optimization by converting complex arithmetic operations into less expensive ones using bitwise operations and shifts with binary numbers.

\item \textbf{Register allocation.} The substitution of stack-based storage with processor registers, improves data access speeds.

\item \textbf{Memory layout optimization.} Rearranging the layout of data structures, to improve memory access patterns.

\item \textbf{Data structures transformation.} Modifying the layout of data structures to maximize the use of registers.

\item \textbf{Dead code elimination.} Removal of unreachable or redundant code from the program’s control flow.

\item \textbf{Function inlining.} The compiler employs heuristics to decide which functions should be inlined, i.e., the invoking of a function is replaced with inserting the body of the function directly into the implementation.

\item \textbf{Hoisting.} Moving loop-invariant code outside the loop to reduce redundant computation.

\item \textbf{Loop unrolling.} Unrolling loops to reduce control overhead and potentially exploit Instruction Level Parallelism.

\item \textbf{Loop fusion or loop jamming.} Merging multiple loops that iterate over the same range of indices.

\item \textbf{Register allocation policies.} Specialized heuristics to determine optimal register usage during runtime.

\item \textbf{Constant propagation.} The propagation of constant values throughout the program’s code, enabling compile-time evaluation of expressions and eliminating computing or fetching them from memory in runtime.

\item \textbf{Global program optimizations.} The entire implementation is treated as a unified system, allowing for deeper optimization by leveraging additional knowledge such as bypassing standard calling conventions between subroutines.

\end{itemize}

\subsection{On performance optimizations beyond traditional compiler capabilities}
\label{ch8:app:what-compilers-can-not}

Compilers automate the translation of code between languages, optimize execution through iterative transformations, and ensure correctness at the programming language level. While they can handle many optimizations, several aspects fall outside their scope: (i) an algorithm's design; (ii) specific implementation choices; (iii) algebraic invariants.

These limitations have persisted for decades, though future advancements may enable compilers to incorporate more invariance-aware optimizations. Specifically, the following limitations remain:

\begin{itemize} 

\item Compilers cannot solve undecidable computational problems, such as Halting Problem~\citep{leiserson2020there}.

\item Constructing cache-oblivious algorithms (those that automatically adapt to multilevel cache hierarchies and their sizes) is a complex task that requires manual design~\citep{demaine2002cache}.

\item Automatically selecting the most efficient form of disk access for a given situation. For example, when reading datasets from disk, the optimal access method is often counterintuitive—subtle OS interfaces like \texttt{mmap} may outperform traditional read/write APIs~\citep{burlachenko2024unlocking}. 

\item Providing formal guarantees that generated code is optimal. 

\item Optimizing across multiple programming environments (for example, efficiently integrating C++ and Python). 

\item Redesigning data structures to improve spatial or temporal locality. 

\item Performing early-exit tests, such as determining whether a point lies outside a complex polyhedron using precomputed bounding box information.\footnote{See lecture slides from MIT 6.172: Performance Engineering of Software Systems by C. Leiserson and J. Shun.}

\item Implementing fundamental algorithmic improvements and eliminating semantically redundant computations beyond standard optimizations like dead code elimination and common subexpression elimination. 

\item Coarsening recursion. The concept of coarsening recursion involves constructing a base-case algorithm that effectively handles small tasks, despite poor asymptotic behavior for large input sizes, and using this algorithm within the structure of a larger computational problem.

\item Some aspects of inlining. Inline subroutine calls are useful not only for eliminating call overhead but also for opening opportunities for further improvements. However, compilers, even when forced to inline subroutines physically, sometimes face challenges in inlining functions. A notable example is recursive functions\footnote{For instance, constraints on inlining in the Microsoft MSVC Compiler can be found here: \href{https://learn.microsoft.com/en-us/cpp/cpp/inline-functions-cpp?view=msvc-170}{https://learn.microsoft.com/en-us/cpp/cpp/inline-functions-cpp?view=msvc-170}}.
\end{itemize}

Expanding compiler optimizations beyond their traditional scope is an active area of research in both the compiler community and broader fields dealing with algorithmic logic. Examples include research on extending algebraic expression replacement beyond simple transformations~\citep{jia2019taso}, automatic function generation for inference~\citep{wu2024mirage, fawzi2022discovering}, and the discovery of fundamentally new algorithms~\citep{fawzi2022discovering}.

Despite these advancements, an alternative approach with significant potential exists. By focusing on a specialized class of models (for example, Transformer-based architectures), a specific class of optimization algorithms (for example, first-order methods), and a dedicated execution strategy for gradient computations (for example, backpropagation), many applications can be expressed within these constructs. In such cases, the full generality of current compiler research may not be necessary, as many limitations can be addressed manually.

\subsection{On compilation techniques for enhancing {PyTorch} backpropagation speed on CPU for training}

\label{ch8:app:torch-compile-techniques-backprop-speed-cpu}

To the best of our knowledge, both in academia and industry, \libname{PyTorch} \citep{paszke2019pytorch} is predominantly used through its Python API. When improving backpropagation speed is critical, \libname{PyTorch} offers several optimization methods.

\paragraph{{PyTorch} graph execution mode via TorchScript.}

One of the key differences between \libname{PyTorch} and \libname{TensorFlow 1.0} lies in their execution strategies. While \libname{TensorFlow 1.0} uses a static graph-based execution model, \libname{PyTorch} adopts dynamic, eager execution, which allows for more intuitive and flexible computation. \libname{TensorFlow 2.0} later introduced eager execution (see Section~\ref{ch8:sec:prev-systems}), while maintaining backward compatibility with \libname{TensorFlow 1.0} graph mode, as static computation graphs offer advantages in execution speed and optimization opportunities. To leverage this model, \libname{PyTorch} introduced TorchScript, a tool that converts dynamic, eager code into a static, graph-based representation (with some limitations). In our experiments, we utilize TorchScript through \libname{PyTorch's} built-in tracing capabilities to achieve performance improvements.

\paragraph{{PyTorch} LibTorch C++ API.}

In addition to TorchScript, another approach for improving the performance of PyTorch-based algorithms is through \libname{PyTorch’s} native C++ interface, called LibTorch. LibTorch allows direct integration of \libname{PyTorch} into C++, offering an alternative to the Python API. At its core, LibTorch includes the ATen library, which serves as the foundation for tensor operations and Automatic Differentiation. The C++ API closely mirrors its Python counterpart, ensuring a consistent and familiar experience for developers transitioning between the two environments. Our experiments demonstrate that, unfortunately, switching from Python to C++ does not automatically unlock all the advantages of a compiler-based approach. This is partly due to the large codebase of PyTorch.

\paragraph{CPU vendor-specific extensions for PyTorch.}

The release builds of \libname{PyTorch} include functionality for specific computations (backends). However, in some cases, CPU vendors may extend \libname{PyTorch} in different ways to optimize performance on their specific hardware platforms. These extensions can further accelerate backpropagation speed by utilizing CPU-specific optimizations tailored to the vendor's architecture. To use these extensions, users typically need a specific version of \libname{PyTorch} and must define certain environment variables. By default, \libname{PyTorch} attempts to use and is compiled with the best-practice compute libraries.

\subsection{On execution modes} 
\label{ch8:app:exex-details-of-comp}

\paragraph{Eager mode.} In our experiment,  \libname{Apple MLX}, \libname{Autograd}, \libname{Micrograd}, \libname{BurTorch}, and certain configurations of \libname{PyTorch} and \libname{TensorFlow} operate with a dynamic computation graph, where forward operations are executed immediately as they are encountered in the description. This model of execution is referred to as \textit{Eager Mode}. In this mode, there is no graph compilation process, and no operation tracing for analysis or conversion takes place.

Interestingly, \libname{BurTorch} outperforms graph-based frameworks despite operating in eager mode. Our experiments show that eager execution can be faster than graph mode in certain situations, demonstrating the advantages of \libname{BurTorch} in handling such cases efficiently.

\paragraph{Graph mode.} In our experiment, we utilized \libname{TensorFlow}, \libname{TF Lite}, and \libname{PyTorch} in a Just-In-Time (JIT) manner to compile and generate a static computation graph that represents $f(x)$. This execution model is referred to as \textit{Graph Mode} and is the modern approach for constructing computation graphs, in contrast to the manual construction required in the past (see \libname{TensorFlow 1.0} in Section~\ref{ch8:sec:prev-systems}). 

In this model, the $f(x)$ computation is represented in a framework-specific manner, with the framework's runtime tracing the execution of operations. Upon the first call of $f(x)$, the internal compiler converts the function operations into a graph form. After this conversion, subsequent computations become more efficient.


\clearpage
\addtocounter{adjsection}{1}
\section{Discussion on Backpropagation}
\label{ch8:app:backprop}

\subsection{On backpropagation memory taxonomy}
\label{ch8:sec:backprop-memory}

In supervised ML tasks, the score function is typically organized according to the principle of Empirical Risk Minimization. Modern computational frameworks, such as \libname{TensorFlow} \citep{abadi2016tensorflow} and \libname{PyTorch} \citep{paszke2019pytorch}, enable the automated computation of gradients with respect to trainable weights, also known as optimization variables. These gradients are evaluated on complex and arbitrarily structured computational graphs using algorithms designed for efficient derivative computation. This process is primarily carried out through backpropagation, a form of Automatic Differentiation in Reverse Accumulation Mode~\citep{griewank2008evaluating}.

Let us assume we use {backpropagation} for a model containing $d$ trainable variables and $d' \ge d$ total variables, which include both trainable and non-trainable parameters, along with a batch of $b$ input-output pairs. The distinction between $d$ and $d'$ arises when only a subset of the model's parameters is trained, as in fine-tuning and transfer learning.

When switching from the \texttt{Forward Pass} to the \texttt{Backward Pass}, the {backpropagation} conceptually requires memory storage for the following quantities:

\begin{enumerate} 
\item \textit{Trainable parameters or variables (weights)}. Memory footprint: $d$ scalars. \newline 
Encodes the trainable part of the Deep Learning model.

\item \textit{Non-trainable parameters or non-trainable variables (frozen weights)}. Memory footprint: $d' - d$ scalars. \newline 
Represents the portion of the model not involved in training, comprising $d' - d$ scalars.

\item \textit{Activations or activation maps (response maps)}. Memory footprint: $d' \times b$ scalars. \newline
Output for all neurons (computational nodes, graph operators).

\item \textit{The $(\mathrm{Input}_i, \mathrm{Output}_i)$ pair for each trainable sample $i \in [b]$}. Memory footprint: $b \times$ \textit{"Memory for Single Input"} + $b \times$ \textit{"Memory for Single Output"}. \newline
The input part is used only in the first step of the \texttt{Forward Pass} and in the last step of the \texttt{Backward Pass}. The output part is used in the last step of the \texttt{Forward Pass} and the first step of the \texttt{Backward Pass}. If computation is not done on the CPU, transferring $(\mathrm{Input}_i, \mathrm{Output}_i)$ pairs to the computational device may become a bottleneck, especially when the descriptions of these pairs are large or when the communication bus is highly contended. For instance, transferring data to a GPU connected via a PCI-Express bus may require additional considerations. In such cases, systems like \libname{nvCOMP}~\citep{nvcomp} can optimize data transfer workflows.

\item \textit{Error signal from connected layers}. Memory footprint: $2 \times$ the maximum number of single activations in the DL model layer. \newline        
The {backpropagation} algorithm does not explicitly store Jacobians; instead, it implicitly evaluates them through the propagation of error signals, typically denoted by $\delta$. Error propagation begins as soon as the \texttt{Backward Pass} starts. The peak memory for storing $\delta$ is proportional to the maximum number of compute units in two consecutive layers.

\item \textit{Optimizer state}. Memory footprint: depends on the optimizer. \newline
The memory footprint depends on the optimizer. \algname{Gradient Descent} has a footprint of $0$. In popular optimizers like Adaptive Moment Estimation ({ADAM}), the footprint is $2d$ scalars, which track the element-wise gradient direction and the element-wise squares of the $\nabla f(x)$. Work addressing this challenge includes {MicroAdam}~\citep{modoranu2024microadam}.

\end{enumerate}

\subsection{On activation maps memory footprint discussion and solutions suggested via BurTorch} \label{ch8:sec:resolve-mem-activations-via-lat-design}

Let us now focus on the memory footprint for activations. If all operators in the compute graph are considered as single activation functions $\mathbb{R} \to \mathbb{R}$, the memory footprint for activations in throughput-oriented DL systems is $d' \times b$ scalars.

The multiplicative factor of $d'$ arises because, even if a trainable variable is located in the first layer, $d'$ can still be significantly larger than $d$. The batch size multiplier emerges due to the lack of activation sharing across samples in the subsampled batch of size $b$. Addressing this challenge remains an open research question. One promising approach is Activation Compression, which has shown potential in reducing training memory usage \citep{liu2022gact}. Another technique is gradient checkpointing~\citep{griewank2008evaluating,chen2016training}. The multiplicative factor $d' \cdot b$ often leads to memory bottlenecks in throughput-oriented designs during {backpropagation}. This issue arises both in training Convolutional Neural Networks\citep{mishra2017wrpn} and in full training and fine-tuning of Large Language Models~\citep{2024fwdllm}.

\libname{BurTorch} offers an alternative solution by addressing the root causes of this problem. The issue primarily arises in throughput-oriented designs, which focus on maximizing the parallel processing of tasks. \libname{BurTorch}, being a latency-oriented design, focuses on optimizing the processing of computations in a serialized manner. This approach can be particularly effective in situations where computations, such as the calculation of sample-gradient oracles, are performed sequentially, thus alleviating the memory bottleneck associated with activation storage in throughput-oriented systems.

\clearpage
\addtocounter{adjsection}{1}
\section{Discussion on Small Compute Graphs}
\label{ch8:app:on-small-compute-graphs}

There are several compelling reasons to emphasize the importance of small compute graphs, even though modern DL is more focused on big dimensional problems.

\paragraph{Mathematical standpoint.}

From a mathematical perspective, the classical Cantor diagonal theorem tells us that $\mathbb{R}^d \sim \mathbb{R} \sim [0,1]$, meaning that the line segment $[0,1]$ contains the same number of points as $\mathbb{R}^d$, as there exists a bijection between them. While this is theoretically interesting, it has limited practical implications. In real-world implementations with finite precision, approximations of $\mathbb{R}$ can accommodate no more points than $\mathbb{R}^d$, highlighting the constraints imposed by finite precision in practical systems.

Next, the Kolmogorov-Arnold Theorem (KAT) asserts that any continuous function $f: \mathbb{R}^d \to \mathbb{R}$ can be represented as:
\[
f(x) = \sum_{j=1}^{2d+1} g_{f,j} \left( \sum_{i=1}^d \lambda_i \gamma_i(x_i) \right),
\]
where $\gamma_i: \mathbb{R} \to \mathbb{R}$ and $\lambda_i \in \mathbb{R}$ are independent of $f$, and $g_{f,j}: \mathbb{R} \to \mathbb{R}$ is entirely determined by $f$. The KAT theorem redistributes the complexity of $f(x)$ into scalar functions $g_{f,j}$, demonstrating that the number of variables $d$ does not always serve as an adequate measure of complexity \citep{lorentz1966approximation}.

\paragraph{Discrete optimization tractability for small ML models.} In Machine Learning, particularly Deep Learning, incorporating prior knowledge often involves selecting a class of parameterized functions, applying regularization schemes for optimization, and using robust training techniques. While small compute graphs may restrict modeling flexibility, they offer advantages in tasks with a combinatorial nature. For instance, small compute graphs can be beneficial for identifying globally optimal pruned models or when certain parts of the trainable variables are constrained to small discrete sets. Discrete Optimization, when applied without heuristics, faces fundamental limitations in high-dimensional problems. However, when the number of discrete variables is small, the problem becomes more manageable.

\paragraph{Applications of on-device training on CPU.} Small compute graphs are particularly essential for resource-constrained systems, such as Internet of Things (IoT) devices, which often face limitations in computational power, memory, and energy resources. By reducing memory and computational demands, small compute graphs are well-suited for these environments. Furthermore, many IoT applications such as real-time monitoring, autonomous systems, and smart sensors—require low-latency processing. 

Smaller compute graphs enable faster inference, facilitating real-time decision-making while minimizing energy consumption. In the context of on-device training, Deep Learning models can enhance or even replace traditional processing pipelines. The engineered systems developed in recent decades present substantial opportunities for optimization across the computational stack. However, to maintain practical performance, on-device training must operate on millisecond timescales for real-time applications (for example, graphics, gaming, and video streaming) and nanosecond timescales (network protocols). 

From a hardware perspective, it is extremely challenging for CPUs to compete with GPUs in terms of processing throughput. However, it is important to note that most practical \textit{applications} are inherently implemented on CPUs. Therefore, for CPU vendors to remain competitive in the domain of practical Deep Learning systems, it may be beneficial to focus on latency-sensitive use cases in on-device training. \libname{BurTorch}, with its latency-optimized design, plays a critical role in enabling traditional applications, typically implemented as compute programs, to leverage on-device training concepts in scenarios such as Federated Learning.

\clearpage
\addtocounter{adjsection}{1}
\section{Design Philosophy Behind BurTorch}
\label{ch8:app:design-philosophy}


The design of \libname{BurTorch} prioritizes simplicity in development, debugging, and optimization while maintaining high performance. Its architecture and implementation is guided by the following principles:

\paragraph{\color{black}{1: Leveraging a compile-based Language from the ground up.}} {\libname{BurTorch}} is built using modern C++20, a widely supported language standard that aligns with the capabilities of current compilation tools. This foundation enables \libname{BurTorch} to significantly reduce the latency of gradient computation, especially for small batch sizes, by utilizing templates to generate only the necessary algorithms and code at compile time. This approach ensures maximum efficiency while minimizing runtime overhead, eliminating the need for dynamic dispatch, which is prevalent in languages like Java, C$\#$, and Python. 

While C++ is not traditionally considered the primary language for research prototyping due to its emphasis on standardization and perfection, this focus can create high barriers to entry, slowing rapid experimentation and iteration. In contrast, prototyping demands flexibility and speed, attributes that Python excels in. The philosophy behind \libname{BurTorch} is that, despite C++'s complexity, the codebase remains small and clean, and the user experience, when using compile-based techniques, is indistinguishable from that of script-based languages, as compilation time is negligible in this case.

\paragraph{2: Leverage efficient looping and function invocation.} The design philosophy behind \libname{BurTorch} is rooted in the principle that when input organization necessitates loops, we prioritize implementing them directly in the code. While Python-based frameworks offer various indexing mechanisms that provide flexibility, this flexibility comes with trade-offs. Our observations suggest that manually implementing the necessary functionality often yields better results. In manually created loops, which remain maintainable, redundant computations can be effectively removed through various optimization strategies. 

Furthermore, manually created loops—while offering comparable computational capabilities—serve as a valuable educational tool, providing users with greater control over the computational process. In contrast, users relying on interpreted environments often face limitations, as several nested loops cannot be efficiently implemented at the script level of Python. It is not only loops that cannot be implemented efficiently but also function dispatching and attribute access mechanisms. This is because, in Python, method and data attribute access are dynamically bound and resolved at runtime.\footnote{As stated in all official Python tutorials from \href{https://www.python.org}{www.python.org} up to and including Python 3.11.}

\paragraph{3: One language and compact code for debugging and optimization.}

Managing and debugging systems implemented in multiple languages is inherently challenging. Python-based systems, for example, face three key issues: (i) debugging across different languages is challenging, (ii) building extension modules with debug info as the number of used libraries grows is challenging, and (iii) understanding code intent in large codebases, especially when performance optimizations sacrifice clarity is challenging for everybody (including authors). These challenges hinder holistic optimizations. \libname{BurTorch} overcomes these issues by maintaining a small codebase and using a single language (C++20), which simplifies debugging and profiling from high-level design to scalar operations. This approach ensures streamlined debugging and is most effective when Deep Learning models are built from simple components with clear training algorithms.

\paragraph{4: Provide high-level and low-level abstractions with minimal overhead.}

A key design principle of \libname{BurTorch} is its ability to offer both high-level abstractions for convenience and low-level control for performance, with minimal overhead connecting the two. High-level abstractions may sacrifice performance, primarily due to two factors.

First, they often lack customization for critical operations. For instance, many frameworks introduce unnecessary memory copying during tensor concatenation \footnote{Concatenate tensors without memory copying \href{https://discuss.pytorch.org/t/concatenate-tensors-without-memory-copying/34609/8}{https://discuss.pytorch.org/t/concatenate-tensors-without-memory-copying/34609/8}}, which can degrade efficiency. High-level languages make it difficult to identify such low-level inefficiencies. But optimizing such things is crucial. For memory operations accessing bytes from DRAM is $\times 330$ more latency-sensitive than scalar arithmetic operations \citep{gregg2014systems}. \libname{BurTorch} delegates performance optimization to the specific implementation of the compute graph, ensuring that core computations are performed with minimal overhead.

The second factor involves the role of \textit{interfaces}. Interfaces bridge different systems, components, or fields. However, in performance-critical situations, these interfaces can introduce significant overhead. Transferring between languages and making function calls can add latency, and without measuring system performance holistically, even simple implementations can outperform complex state-of-the-art systems \citep{mcsherry2015scalability}. \libname{BurTorch} minimizes this issue by leveraging compile-time optimizations such as code inlining, C++ template generation, global program optimization, and static builds from source code. This approach decouples the interfaces used for describing algorithms from those used at runtime.

\paragraph{\color{black}{5: Self-contained design.}} \libname{BurTorch} adopts a self-contained design idea that leverages standardized operating system interfaces and fully utilizes available computational devices without relying on external runtime systems or computation libraries. This approach minimizes overhead introduced by control structures and environments, which can significantly impact computation time. Operating entirely in userspace, \libname{BurTorch} is optimized for modern OS environments. Based on our experience, implementing functionality in kernel space offers limited benefits compared to the complexity and effort involved in handling compute-intensive and I/O-bound tasks. By remaining in userspace, \libname{BurTorch} streamlines development while ensuring high performance. This self-contained design also facilitates seamless integration into resource-constrained systems and removes the need to manage external library dependencies.

\paragraph{6: Addressing control overhead in complex frameworks.}

We categorize the control overhead in system implementations into two distinct types: unavoidable and avoidable. Unavoidable overhead arises from design choices in closed-source or overly complex open-source frameworks, where users have limited control over the internal architecture. These high-level abstractions obscure crucial details of the system, leading to inefficiencies in optimizing computational resources. In contrast, avoidable overhead results from design decisions in a specific implementation of some DL framework, which prioritize ease of use and modification over peak performance.

While Deep Learning frameworks provide useful abstractions, they often sacrifice mathematical rigor and computational efficiency. For instance, many frameworks and even the Python interpreter rely on pre-built dynamic libraries that cannot be selectively excluded during runtime. This forces the OS to load unnecessary operations, resulting in performance degradation. To address these challenges, \libname{BurTorch} adopts a minimalist, system-oriented approach: (i) prioritizing compile-time optimizations via native language compilers, (ii) exposing only essential compute and OS primitives for each task, (iii) offering simplified, user-friendly views of hardware and OS details, and (iv) maintaining a compact and efficient architecture.  This design philosophy reflects \libname{BurTorch}’s commitment to minimizing overhead at all software levels, from high-level to low-level interactions.

\paragraph{\color{black}{7: Relying on the static linkage of the final executable.}}

\libname{BurTorch} philosophy avoids reliance on dynamic libraries (except those required by the OS for invocation). This approach enhances portability and reduces cascading dependencies on middleware that are often inherited when using Python. By eliminating the need to prepare environments and download dependencies, this design allows for whole-program optimization at compile time. This process constructs native CPU code that can bypass traditional calling conventions, eliminating subtle misinterpretations of compute function interfaces.

Technically, whole-program optimization is nearly infeasible when each piece of logic is implemented as a standalone distributed library. In this context, the Python interpreter serves as an ecosystem, linking its runtime with Python extension modules. Systems built on these principles are inherently challenging to optimize through whole-program optimization, as the final binaries exhibit a more compact structure compared to their original source code design. The result is a streamlined implementation with optimized performance in \libname{BurTorch}.

\paragraph{8: Recognizing the importance of visualization in model development.}

Practical ML projects generally progress through the following phases:

\begin{enumerate}
\item Selection of the problem to address.
\item Collection of data, either through statistical tests or the problem's setup.
\item Design of a mathematical model, with or without domain expertise.
\item Training of the model, with or without guarantees of convergence.
\item Evaluation of the model's performance.
\item Deployment of the model for autonomous operation.
\item Ongoing maintenance, which may involve revisiting earlier phases.
\end{enumerate}

While this classical approach persists, new methodologies are emerging with the rise of large language models (LLMs). Python's widespread use in ML stems from its low entry barrier, essential in multidisciplinary fields. Additionally, Python’s rich ecosystem, including tools like Matplotlib, excels in data exploration and model design, aiding decision-making in phases (2) and (3). In contrast, general-purpose programming languages like C++ lack comparable visualization tools and have a higher entry barrier.

To bridge this gap, \libname{BurTorch}: (a) relies on core language constructs with a minimalist design; (b) provides an API similar to PyTorch; and (c) dynamically generates Python scripts to leverage tools like Matplotlib, potentially running separate processes during code debugging to facilitate visualization. 

This hybrid approach in \libname{BurTorch} supports the generation of Python scripts during debugging, enabling real-time plotting of scalar plots through a separate Python interpreter instance. Additionally, \libname{BurTorch} allows for visualization of computation graphs, which can be exported in DOT format for easy analysis. This replicates the familiar debugging and plotting workflows found in tools like Matlab or Python, enabling quick iteration during experimentation. By leveraging Python’s strengths in debugging and visualization, \libname{BurTorch} ensures flexibility and convenience during development while benefiting from C++'s computational efficiency and scalability for fast training execution. 

This hybrid model allows researchers to harness the best of both worlds: flexibility during debugging and performance during final runtime preparation. Importantly, \libname{BurTorch} does not attempt to integrate scientific visualization directly into the C++ runtime but instead acknowledges its essential role during model development and data exploration. By seamlessly integrating Python’s visualization tools with C++’s performance, \libname{BurTorch} facilitates the visualization of scalar metrics while maintaining high performance. A common question that may still arise is:

\begin{center}
\textit{Why doesn’t C++ have such rich libraries in the first place?}    
\end{center}

This is likely a result of the language's perfectionist principles, combined with the actual time required to create such libraries. The most notable of these principles, rooted in the language's origins \citep{stroustrup1994design}, are as follows:

\begin{center}
\textit{What you don’t use, you should not pay for. If something is built into the language itself, it cannot be done better by hand.} 
\end{center}


\paragraph{9: Addressing the limitations of gradient oracle interfaces.}

In supervised ML, Reinforcement Learning, Adaptive Control, and Optimization, the gradient computation process is typically automated through Automatic Differentiation (AD). Although AD simplifies gradient computation, it often lacks the flexibility required for tasks that require greater control over the gradient computation. Achieving practical, state-of-the-art performance requires more than automation; it requires precise control over each stage of computation. \libname{BurTorch} addresses this need through the following methodologies:

\begin{enumerate} 
\item A minimalistic codebase.
\item Leveraging a language tightly integrated with OS and CPU features.
\item A transparent, simple, and contiguous memory layout for storing partial derivatives, activations, and parameters.
\end{enumerate}

The memory regions storing activations and trainable variables in \libname{BurTorch} are contiguous, meaning they are not fragmented across different locations in the process's virtual memory. This is essential for tightly coupling training implementations with the underlying operating system and communication systems.

For example, the sequential nature of these buffers enables efficient read operations, optimizing model serialization to disk. Sequential reads and writes facilitate ideal CPU cache utilization (when not already occupied), and disk transfers are executed efficiently. It's important to note that both caches and disks do not operate at the byte level. All memory transitions between DRAM and CPU caches are managed by the DRAM Memory Controller, processing these transitions in fixed-size blocks known as \textit{cache lines}, typically 64 bytes long on most x86-64 and AArch64 architectures. Disk operations, on the other hand, occur at the sector level, usually 512 bytes.

Although networks and operating systems offer straightforward interfaces, it would be a mistake to simplify them to just basic mechanisms. A deeper understanding of these systems is essential for effectively coupling backpropagation computation with communication. A concrete example of an interface that benefits from contiguous memory is sending buffers to a Network Interface Controller (NIC) via the \texttt{MSG\_ZEROCOPY} mechanism\footnote{Available from \href{https://www.kernel.org/doc/html/v4.14/networking/msg_zerocopy.html}{Linux Kernel v4.14, 2017}}. Furthermore, advanced mechanisms like the Data Plane Development Kit (DPDK)\footnote{The Data Plane Development Kit (\href{https://www.dpdk.org}{DPDK}) is an open-source project managed by Linux Foundation} offers a more direct way to bypass the kernel network stack \citep{gregg2014systems}. To fully leverage these mechanisms, a flat memory representation is crucial.

\clearpage
\addtocounter{adjsection}{1}
\section{Capabilities of BurTorch}
\label{app:cap-burtorch}

\subsection{High-level constructions}
\label{ch8:app:cap-burtorch-high}

At a higher level of abstraction, \libname{BurTorch} introduces fundamental components such as Neurons, Linear Layers, and Multi-Layer Perceptron Layers. These components encapsulate essential operations like linear (or affine) transformations and include best-practice layer parameter initialization. They also support standard activation functions, such as Sigmoid, ReLU, Tanh, or identity mapping, making it easy to model complex computations.

\subsection{Low-level constructions}
\label{ch8:app:cap-burtorch-low}

\libname{BurTorch} at the most granular level operates on simple \textit{scalars} from $\mathbb{R}$. Every scalar is indexed sequentially, maintaining simplicity and efficiency throughout the implementation. Unlike other frameworks, \libname{BurTorch} avoids the use of complex fused operations or heavy reliance on Single-Instruction Multiple-Data (SIMD) CPU registers for layer-specific optimizations or custom computation function implementation for specific cases. While SIMD can offer performance gains, its use often leads to unnecessarily complex, unmanageable code. 

Instead, \libname{BurTorch} adopts a streamlined approach, leveraging Instruction-Level Parallelism (ILP) through straightforward loop unrolling and efficient utilization of modern CPU pipelines. In architectures where floating-point functional units can handle both SIMD and ILP effectively, \libname{BurTorch} fully capitalizes on available resources without adding unnecessary implementation complexity.

\subsection{Supported scalars}
\label{ch8:app:burotrch-scalars}

\libname{BurTorch} supports the processing of computation graphs that involve operations on \textit{scalars}. Scalars, in this context, can be one of the following types:

\begin{enumerate}
\item \textbf{FP32, FP64:} These represent standard floating-point arithmetic formats, offering approximate real-number computations with single (FP32) and double (FP64) precision, as defined by the IEEE 754-2008 \citep{IEEE754-2008}.

\item \textbf{{SIMD Vector Registers}}: Operations, as listed in Table~\ref{ch8:tab:optype}, can be performed on small SIMD registers, typically 128, 256, or 512 bits in length. \libname{BurTorch} does not merely exploit SIMD registers for efficiency but can directly construct computations using these registers. Supported SIMD architectures include SSE2 (128-bit vector registers), AVX2 (256-bit vector registers), AVX-512 (512-bit vector registers), and ARM Neon (128-bit vector registers).

\item \textbf{FP16, BF16, FP128 with Switching to C++23:} While FP32 and FP64 remain the most commonly used formats, they are not the only ones available today. Modern computer hardware capable of processing 16-bit floating-point data is becoming more widely accessible. \libname{BurTorch} can be used with the following extended floating-point types from IEEE 754-2008: Half-precision floating-point format (FP16), and Quadruple-precision floating-point format (FP128). Also \libname{BurTorch} can be used with computing in brain floating-point format (BF16) \citep{wang2019bfloat16}.

\libname{BurTorch} can be compiled using any C++20 and C++23 compatible compiler. However, to use these formats, which were introduced in the C++23 standard, \libname{BurTorch} must be configured to use C++23, and a compiler that supports these formats, such as GCC 13.1\footnote{See \href{https://gcc.gnu.org/releases.html}{https://gcc.gnu.org/releases.html},
	\href{https://en.cppreference.com/w/cpp/compiler_support/23}{https://en.cppreference.com/w/cpp/compiler\_support/23}}, must be used.

\end{enumerate}

\subsection{Supported core operators at scalar level}
\label{ch8:app:burotrch-scalar-operators}

\begin{table}[h!]
\footnotesize
\centering
\caption{Supported core operations in \libname{BurTorch} at the granularity of single scalar-level computations. \textbf{Args.} = Arguments. For mnemonics see Appendix~\ref{ch8:app:burotrch-scalar-operators}.}
\begin{tabular}{@{}lllll@{}}
	\toprule
	\textbf{Mnemonics} & \textbf{Args.} & \textbf{Internal Name}  & \textbf{Description} \\ 
	\midrule
	\texttt{leaf} & \texttt{[s]} & \texttt{eLeaf} & A basic node with input \\ 
	\texttt{relu} & \texttt{[s]} & \texttt{eRelu} & Applies the $\max(0,x)$  \\
	\texttt{tanh} & \texttt{[s]} & \texttt{eTanh} & Applies the hyperbolic tangent \\ 
	\texttt{exp} & \texttt{[s]} & \texttt{eExp} & Applies the computation $\exp(x)$ \\ 
	\texttt{negativeLog} & \texttt{[s]} & \texttt{eNegLog} & The minus natural logarithm \\ 
	\texttt{sigmoid} & \texttt{[s]} & \texttt{eSigmoid} & Applies the sigmoid function  \\ 
	\texttt{inv} & \texttt{[s]} & \texttt{eInv} & Computes $\nicefrac{1}{X}$ \\ 
	\texttt{sqr} & \texttt{[s]} & \texttt{eSqr} & Computes $x^2$ \\ 
	\texttt{pow3} & \texttt{[s]} & \texttt{eCub} & Computes $x^3$ \\ 
	\texttt{logarithm} & \texttt{[s]} & \texttt{eLog} & The natural logarithm of input \\ 
	\texttt{sqrt} & \texttt{[s]} & \texttt{eSqrt} & Computes the square root $\sqrt{x}$ \\ 
	\texttt{invSqrt} & \texttt{[s]} & \texttt{eInvSqrt} & Computes  $\nicefrac{1}{\sqrt{x}}$ \\ 
	\texttt{+, add} & \texttt{[bin]} & \texttt{eBinaryAdd} & Computes for $x,y$ result $x+y$ \\ 
	\texttt{-, sub} & \texttt{[bin]} & \texttt{eBinarySub} & Computes for $x,y$ result $x-y$ \\ 
	\texttt{*, mul} & \texttt{[bin]} & \texttt{eBinaryMult} & Computes for $x,y$ result $x \times y$ \\ 
	\texttt{mulByConstant(x,c)} & \texttt{[bin]} & \texttt{eBinaryMultByConst} & Computes $x \times c$ for constant $c$ \\
	\texttt{/, div} & \texttt{[bin]} & \texttt{eBinaryDiv} & Divides two inputs $x,y$ as $x/y$ \\ 
	\texttt{mean} & \texttt{[bin]} & \texttt{eBinaryMean} & Computes the mean $\nicefrac{(x+y)}{2}$ \\   
	\texttt{addSquares} & \texttt{[bin]} & \texttt{eBinaryAddSquares} & Computes $x^2+y^2$ \\
	\texttt{meanSquares} & \texttt{[bin]} & \texttt{eBinaryMeanSquares} & Computes $\nicefrac{(x^2+y^2)}{2}$ \\ 
	\texttt{negativeMean} & \texttt{[bin]} & \texttt{eBinaryNegativeMean} & Computes $\nicefrac{-(x+y)}{2}$ \\   
	
	\texttt{reduceSum} & \texttt{[var]} & \texttt{eAddVarying} & Computes $\sum_{i=1}^{n} x_i$ \\ 
	\texttt{reduceSub} & \texttt{[var]} & \texttt{eSubVarying} & Computes $x_1 - \sum_{i=2}^{n} x_i$ \\
	\texttt{reduceMul} & \texttt{[var]} & \texttt{eMulVarying} & Computes $\prod_{i=1}^{n} x_i$ \\
	\texttt{reduceMean} & \texttt{[var]} & \texttt{eMeanVarying} & Computes $\nicefrac{1}{n} \sum_{i=1}^{n} x_i$ \\ 
	\texttt{reduceSumOfSquares} & \texttt{[var]} & \texttt{eSumOfSquaresVarying} & Computes $ \sum_{i=1}^{n} x_i^2$ \\
	\texttt{reduceMeanSquares} & \texttt{[var]} & \texttt{eMeanSquaresVarying} & Computes $ \nicefrac{1}{n}\sum_{i=1}^{n} x_i^2$ \\
	\texttt{reduceNegativeMean} & \texttt{[var]} & \texttt{eNegativeMeanVarying} & Computes $\nicefrac{-1}{n} \sum_{i=1}^{n} x_i$ \\ 
	\texttt{innerProduct} & \texttt{[var]} & \texttt{eInnerProductNoBias} & The dot product $\langle x, y \rangle$ \\ 
	\texttt{innerProductWithBias} & \texttt{[var]} & \texttt{eInnerProductWithBias} & Computes $\langle x, y \rangle + b$ \\             
	\bottomrule
	
\end{tabular}
\label{ch8:tab:optype}
\end{table}

The core computations in Table~\ref{ch8:tab:optype} are atomic operations used to construct the computational graph from simple scalars. This constructed graph, in eager mode, is suitable for both function evaluation and exact gradient computation via Automatic Differentiation. The notations \texttt{[var]}, \texttt{[bin]}, and \texttt{[s]} represent the number of arguments required for each operation:

\begin{enumerate} 
\item \texttt{[var]}: The operation accepts an arbitrary number of arguments. 
\item \texttt{[bin]}: The operation involves exactly two arguments (binary operation). 
\item \texttt{[s]}: The operation is performed on a single argument (unary operation). 
\end{enumerate}

These core computations, as outlined in Table~\ref{ch8:tab:optype}, are implemented using standard C++ constructs. All computation code for operands is written as C++ template header-only code. This design choice ensures that the code is both efficient and capable of effective dispatching. The operations are \textit{atomic} in the sense that they are implemented independently, not in the sense of atomic access for operands.

\subsection{Supported derived operators at the scalar level}
\label{ch8:app:burotrch-derived-operators}

\libname{BurTorch} supports derived operators at the scalar level, as presented in Tables \ref{ch8:tab:inplace-ops} and \ref{ch8:tab:heps-compute-ops}. The implementation of these operators requires the allocation of more than one computation node from the pool of scalar nodes. While this aspect can be disregarded from a usage perspective, it is important to note that, practically, some of the information necessary for implementing these operators may already be available within the surrounding context of the expression.

\begin{table}[h!]
\footnotesize
\centering
\caption{In-place operations supported in \libname{BurTorch} at the scalar level.}
\begin{tabular}{@{}lllll@{}}
	\toprule
	\textbf{Mnemonics} & \textbf{Arguments}  & \textbf{Description} \\ 
	\midrule
	\texttt{+=, addInplace} & \texttt{[bin]} & Adds a value in-place, i.e., $x \leftarrow x + y$ \\ 
	\texttt{-=, subInplace} & \texttt{[bin]} & Subtracts a value in-place, i.e., $x \leftarrow x - y$ \\ 
	\texttt{*=, multInplace} & \texttt{[bin]} & Multiplies a value in-place, i.e., $x \leftarrow x \times y$ \\ 
	\texttt{/=, divInplace} & \texttt{[bin]} & Divide a value in-place, i.e., $x \leftarrow \nicefrac{x}{y}$ \\ 
	\bottomrule
\end{tabular}
\label{ch8:tab:inplace-ops}
\end{table}

\begin{table}[h!]
\footnotesize
\centering
\caption{Help not-atomic scalar compute operations supported in \libname{BurTorch}.}
\begin{tabular}{@{}lllll@{}}
	\toprule
	\textbf{Mnemonics} & \textbf{Arguments}  & \textbf{Description} \\ 
	\midrule
	\texttt{varianceBiased} & \texttt{[var]} & $\sum_{i=1}^{n} \dfrac{x_i^2}{n} - \left(\sum_{j=1}^{n} \dfrac{x_j}{n} \right)^2$ \\
	\texttt{variance} & \texttt{[var]} & $\dfrac{n}{n-1} \cdot \left(\sum_{i=1}^{n} \dfrac{x_i^2}{n} - \left(\sum_{j=1}^{n} \dfrac{x_j}{n} \right)^2 \right)$ \\
	\texttt{reduceMeanAndMeanSquares} & \texttt{[var]} & $\nicefrac{1}{n}\sum_{i=1}^{n} x_i$ and  $ \nicefrac{1}{n}\sum_{i=1}^{n} x_i^2$ \\
	\bottomrule
\end{tabular}
\label{ch8:tab:heps-compute-ops}
\end{table}

\subsection{Supported Matplotlib scripts generation}
\label{ch8:app:matplotlib}

\begin{enumerate}
\item \texttt{generateHeatMapBasic:} This method generates a Python script as a string that uses Matplotlib to plot a heatmap.

\item \texttt{generateHeatMap:} This method generates a Python script as a string that uses Matplotlib to plot a heatmap with customized cell values, labels, and counters, applying specific text annotations at each cell based on the itemGetter and counterGetter functions for each matrix element.

\item \texttt{generatePlot:} This method generates a Python script as a string that plots a mathematical function over a specified range (xStart to xEnd) using Matplotlib, with the function's values computed at regular intervals and displayed with a grid and title.

\item \texttt{buildDotGraph:} This method generates a dot graph representation of a tree structure, where each node is described with relevant information, such as label, help, derivative, data references, and connections between nodes.

\item \texttt{asString:} Creates a string representation of a compute node.
\end{enumerate}

\subsection{High-level and low-level connections in BurTorch}
\label{ch8:app:cap-burtorch-high-with-low}

\paragraph{Transparency in buffer storage organization.}

At the low level, \libname{BurTorch} ensures complete transparency in the management of partial derivatives and activations, organizing them sequentially in virtual memory for optimal read/write access. The high-level constructs in \libname{BurTorch} are intentionally designed to rely exclusively on low-level implementations, promoting clarity, modularity, and adaptability across the framework. Except the optimized backpropagation internals, every aspect of \libname{BurTorch} remains highly customizable due to its compact and efficient codebase. \libname{BurTorch} is built on a minimalist philosophy, which reduces runtime complexity, dispatch overhead, and overall implementation intricacy. 
Current DL frameworks are heavily optimized for throughput and large batch sizes but are often constrained by APIs shaped by language limitations, particularly in Python. Python’s inefficiency in managing function calls and loops forces users to write mathematical computations in ways that align with the framework rather than the problem itself. This artificial constraint has become the norm. In contrast, \libname{BurTorch} addresses these inefficiencies by combining simplicity, flexibility, and computational efficiency, enabling researchers to explore straightforward models and algorithms without being hindered by complex frameworks, inefficient language ecosystems, or unnecessary abstractions.

\paragraph{Backpropagation with total memory control.}
If you need fine-grained memory control during backpropagation, use the \texttt{backwardWithScratchStorage} function. This C++ template function is designed to perform backpropagation in a computation graph, utilizing scratch storage for intermediate data. It accepts several parameters, including the root node of the graph, reverse topological order data, leaf nodes that can be computed in any order, and a recursion stack.

The \texttt{backwardWithScratchStorage} function optimizes the backpropagation process by leveraging scratch storage and efficient memory management. It ensures that the computation graph is traversed in reverse topological order, processing both internal and leaf nodes appropriately. The recursion stack enables efficient handling of nodes with multiple children. Additionally, the function relies \textit{exclusively} on scratch storage for intermediate data, guaranteeing that memory is allocated and deallocated efficiently during traversal. The function utilizes a recursion stack to maintain the state of the traversal. As each node is processed, it is pushed onto the stack. The stack dynamically grows and shrinks as nodes are processed, ensuring that the reverse topological order is followed and each node is handled in the correct sequence. This function focuses on traversing the graph in topological order, marking the nodes as processed, and managing recursive calculations on the graph nodes. It optimizes memory management by utilizing scratch storage for nodes that need to be processed in a specific order while handling leaf nodes separately, as they can be processed in any order due to the absence of children.

\paragraph{Backpropagation with simple backward.}
When using \libname{BurTorch} for early prototyping, you may prefer not to focus on memory storage concerns. If you don't need to optimize memory usage at this stage, you can invoke the simple backward function. This is standard backpropagation, and while it doesn't include the optimizations for memory allocation, recursive traversal, and efficient node processing provided by \texttt{backwardWithScratchStorage}, it may be sufficient.

\paragraph{\color{black}Saving and loading computation graph values and gradients.} \libname{BurTorch} provides efficient mechanisms for saving and loading elements of the computation graph to and from files and memory. Scalar values are indexed incrementally until an explicit collection of unused indices is invoked, or the construction of new computation nodes is reversed. Each scalar tensor is thus associated with a unique index. The framework allows for saving or loading all tensors within a specified index range (from first to last) to/from a file, or for saving and loading the entire computation graph. This approach offers compile-time flexibility in specifying which elements should be saved or loaded. These operations are highly efficient, as the memory layout for the range between two indices (retrieved using \texttt{sysGetRawNodeIndex}) is sequential.

\subsection{Comparison of {BurTorch} with explicit code snippets}
\label{ch8:app:exp2-listing}

\libname{BurTorch} provides an API that closely resembles those of \citet{karpathy2020micrograd} and \libname{PyTorch} \citep{paszke2019pytorch}. The exact code snippets for the three fully functional compute graphs (excluding the setup environment) are presented in Listing~\ref{ch8:lst:exp2-listing} below, illustrating their near-identical structure. This demonstrates that leveraging modern C++ can offer researchers a practical and efficient solution without introducing undue complexity.

\begin{figure}[h!]
\centering
\begin{tabular}{ccc}
Micrograd & BurTorch & PyTorch\\
\begin{minipage}{0.32\textwidth}
\tiny
\lstset{basicstyle=\ttfamily\tiny, breaklines=true}
\begin{lstlisting}[language=Python,caption={}]
#!/usr/bin/env python3
# Python code
from micrograd.engine import Value
			
if __name__ == "__main__":
  a = Value(-4.0)
  b = Value(2.0)
  c = a + b
  d = a * b + b**3
  c += c + 1
  c += 1 + c + (-a)
  d += d * 2 + (b+a).relu()
  d += 3 * d + (b-a).relu()
  e = c - d
  f = e**2
  g = f / 2.0
  g += 10.0 / f
  g.backward()
  print(a.grad)
\end{lstlisting}	
\end{minipage}
&
\begin{minipage}{0.32\textwidth}
\tiny
\lstset{basicstyle=\ttfamily\tiny, breaklines=true}
\begin{lstlisting}[language=C++,caption={}]
// C++ code
#include "burtorch.h"
#include <iostream>

int main() {
  auto a = Value(-4.0);
  auto b = Value(2.0);
  auto c = a + b;
  auto d = a * b + pow3(b);
  c += c + Value(1.0);
  c += Value(1.0) + c - a;
  d += d * Value(2.0) + relu(b+a);
  d += Value(3.0) * d + relu(b-a);
  auto e = c - d;
  auto f = sqr(e);
  auto g = f / Value(2.0);
  g += Value(10.0) / f;
  backward(g);
  std::cout<<a.gradCopy();
  return 0;
}
\end{lstlisting}
\end{minipage}
&
\begin{minipage}{0.32\textwidth}	
\tiny
\lstset{basicstyle=\ttfamily\tiny, breaklines=true}	
\begin{lstlisting}[language=Python,caption={}]
#!/usr/bin/env python3
# Python code
import torch

if __name__ == "__main__":
  a = torch.tensor(-4.0, requires_grad=True, dtype=torch.float64)
  b = torch.tensor(2.0, requires_grad=True,dtype=torch.float64)
  c = a + b
  d = a * b + b**3
  c += c + 1
  c += 1 + c + (-a)
  d += d * 2 + (b+a).relu()
  d += 3 * d + (b-a).relu()
  e = c - d
  f = e**2
  g = f / 2.0
  g += 10.0 / f
  g.backward()
  print(a.grad.item())
\end{lstlisting}	
\end{minipage}	
\end{tabular}

\caption{Listings for the small compute graph shown in Figure~\ref{ch8:fig:exp2-small-compute-graph}, adapted from \citet{karpathy2020micrograd}.}

\label{ch8:lst:exp2-listing}
\end{figure}

\subsection{Unique aspects of BurTorch}
\label{ch8:app:burotrch-unique}

\paragraph{\color{black}{1. Computing gradient without evaluating function value.}}

\libname{BurTorch} provides an imperative approach to constructing scalar values in a computation graph, typically culminating in a loss function. However, the final loss value itself is not required for backpropagation. Unlike traditional frameworks, \libname{BurTorch} allows gradients to be computed without explicitly evaluating the function value, providing a minor optimization. While this optimization can impact performance depending on the computation graph’s structure, it was not used in any experiments in this work.  

\paragraph{\color{black}{2. BurTorch compliance with high-reliability industry standards.}}

Motor Industry Software Reliability Association (MISRA) guidelines \citep{misra2012} are widely adopted in safety-critical industries such as automotive, aerospace, and medical devices. Most Deep Learning frameworks fail to meet MISRA’s stringent requirements, particularly its prohibitions on (i) dynamic memory allocation at runtime (Rule 4.12 \citet{misra2012}) and (ii) recursive functions (Rule 17.2 \citet{misra2012}), both of which ensure system stability and predictability. 

\libname{BurTorch} addresses these concerns by eliminating runtime memory allocation—configurable to operate solely on pre-allocated scratch buffers—and by avoiding recursion, ensuring MISRA compliance.

\paragraph{\color{black}{3. Maximizing performance with pre-allocated buffers.}}

Most Deep Learning frameworks rely on dynamic memory allocation for parameters, activations, and gradients. While they offer some control over memory management, they still depend on dynamic allocation through internal logic or external libraries. \libname{BurTorch}, in contrast, exclusively operates on pre-allocated, user-provided buffers, eliminating runtime memory allocation. This design is particularly beneficial in environments with strict memory and communication constraints, such as embedded systems or Federated Learning. By minimizing auxiliary memory usage and avoiding unnecessary copying, \libname{BurTorch} achieves optimal performance in resource-limited settings and ensures compatibility with systems where dynamic memory allocation is prohibited.

\paragraph{\color{black}{4. A small codebase enables easy customization.}}

Due to its compact codebase, \libname{BurTorch} can be modified to address research needs that cannot be easily achieved through modifications in complex frameworks. See Section~\ref{ch8:sec:inluence-on-theory}.  

\paragraph{\color{black}{5. {BurTorch} is thread-safe.}}

\libname{BurTorch} can be built with flags to ensure true thread safety, a distinguishing feature that sets it apart from many alternative frameworks. In \libname{BurTorch}, computations can be safely executed in parallel without concerns about race conditions, thanks to its carefully designed architecture and, importantly, its compact codebase. In frameworks such as \libname{TensorFlow}, \libname{PyTorch}, and \libname{JAX}, special consideration must be given to thread safety. These frameworks rely on complex chains of dependencies, where the failure of even a single component to guarantee thread safety compromises the overall framework's ability to provide strong thread safety assurances. As a consequence, the thread safety of \libname{PyTorch}, \libname{TensorFlow}, and \libname{JAX} depends on the specific use case.

\paragraph{\color{black}{6. Concurrent graph construction across multiple threads.}}

\libname{BurTorch} allows multiple real operating system threads to simultaneously contribute to the construction of a single computation graph. By enabling concurrent graph-building, \libname{BurTorch} takes full advantage of modern hardware, leading to a more efficient distribution of computational tasks. This capability to manage parallel graph construction offers a level of flexibility that is often lacking in other frameworks. It is a key advantage, enabling \libname{BurTorch} to scale more effectively in multi-threaded environments, especially when optimizing graph construction is necessary.

\paragraph{\color{black}{7. Ability to eliminate named parameters.}}

\libname{PyTorch} \citep{paszke2019pytorch}, \libname{TensorFlow} \citep{abadi2016tensorflow}, and \libname{JAX} \citep{jax2018github} with the FLAX extension \citep{flax2020github} do not provide a mechanism to remove names from trainable parameters. While named parameters are useful during research and debugging, once the model is finalized, they may no longer be necessary if the optimization algorithms do not rely on names.

\libname{BurTorch} supports names for variables, but it can be configured and built to completely remove the functionality of named parameters or variables at runtime. And it requires no changes to the source code to eliminate this redundant logic.

Although this may appear to be a minor feature if each scalar has a unique text description longer than two characters, the memory consumed by these descriptions exceeds the memory required to store the $x\in\mathbb{R}^d$ in FP16 format.

\clearpage
\addtocounter{adjsection}{1}
\section{Missing Experiment on Linux OS}
\label{ch8:app:extra-experiments-linux}

In this appendix, we present the results of comparing \libname{BurTorch} against popular Deep Learning frameworks under experimental settings similar to those described in Section \ref{ch8:sec:experiments} of the main text, with two differences.

First, the experiments were conducted on a Linux workstation running {Ubuntu 20.04.6}. Although the CPU used is also x86-based, its clock frequency is slightly lower, fixed at $3.2$ GHz during the experiments. We employed the methodology from \citet{burlachenko2024unlocking} to ensure the reproducibility of the measurement during experiments.

Second, for memory consumption, we report the peak virtual memory usage (VmSize). The memory subsystem is closely tied to the operating system, so it may be useful to revisit what constitutes this value. For more detailed explanations, we refer the reader to specialized literature, such as \citet{kerrisk2010linux}.

\paragraph{Background on the VmSize metric.} In POSIX-based OS, VmSize represents the total amount of virtual memory allocated to a process, including:

\begin{enumerate} 
\item \textbf{Program text.} Memory allocated for the executable code of the process in machine-language instructions. This memory may be shared with other applications, except for data within binary applications. In such cases, system mechanisms like copy-on-write may be involved.

\item \textbf{Initialized data.} The segment containing global and static variables that are explicitly initialized.

\item \textbf{Uninitialized data.} Global and static variables that are not explicitly initialized. Before the program starts, the system initializes such memory to zero, explaining why the size of a binary executable file does not directly reflect its runtime memory requirements.

\item \textbf{Heap memory.} Memory allocated for dynamically allocated data.

\item \textbf{Stack memory.} A dynamically growing and shrinking segment that contains stack frames. Each stack frame is allocated for the currently called function and stores local variables, function arguments, and return values.

\item \textbf{Memory-mapped files.} Additional files or resources mapped into the process’s memory space. Mapped files allow their contents to be accessed as bytes in a corresponding memory region. Pages from named memory-mapped files are loaded into memory on demand. Anonymous mappings, which are not associated with files, are backed by the swap.

\end{enumerate}

The peak \texttt{VmSize} reflects the total virtual address space of the process, encompassing both actively used memory and memory that may not be physically resident in DRAM but is backed either by swap files or another file (if the content source is a physical file mapped via the memory-mapping mechanism).

\clearpage

\begin{table}[h!]
\footnotesize
\centering
\caption{Backpropagation over $100$K iterations with a {tiny} dynamic compute graph from Figure~\ref{ch8:fig:tiny-compute-graph}. Mean and standard deviation across $5$ launches. Computation in FP64, one CPU Core $3.2$ GHz (x86-64). Physical Memory: DDR4, 251 GBytes. Linux Ubuntu 20.04. See also Figure~\ref{ch8:fig:execution_times_speedup_linux}. The numerical results across frameworks match exactly.
}
\begin{tabular}{|l|l|l|l|l|}
	\hline
	\textbf{\#} & \textbf{Framework, Mode, Language} & \textbf{Device} & \textbf{\makecell[c]{Compute Time \\ (sec.)}} & \textbf{\makecell[c]{Relative \\ to \\ BurTorch}} \\ 
	\hline
	\hline
	\cellcolor{bgcolorwe}1&\cellcolor{bgcolorwe}BurTorch, Eager, C++ & \cellcolor{bgcolorwe}CPU                   & \cellcolor{bgcolorwe}$0.011 \pm 0.00007$             & \cellcolor{bgcolorwe}$\times 1.0$ (We)             \\ \hline
	2& TensorFlow 2.18.0, Eager, Python & CPU          & $163.538 \pm 1.584$         & $\times 14\,867.0\,\,\,$                    \\ \hline
	3& TensorFlow 2.18.0, Graph, Semi-Python & CPU          & $27.882 \pm 0.280$         & $\times 2\,534.7\,\,\,\,\,$                    \\ \hline
	4& \makecell[l]{TF Lite 2.18.0, Graph,\\TF Lite Interpreter} & CPU          & $0.408 \pm 0.005$         & $\times 37$                    \\ \hline
	5& Autograd 1.7.0, Eager, Python & CPU            & $29.250 \pm 0.464$        & $\times 2659.09\,\,\,$                    \\ \hline
	6& {PyTorch} 2.5.1, Eager, Python & \textbf{GPU}          & $35.240 \pm 2.193$              & $\times 3\,203.6\,\,\,$                    \\ \hline
	7& {PyTorch} 2.5.1, Eager, Python & CPU          & $15.440 \pm 0.037$              & $\times 1\,403.6\,\,\,$                    \\ \hline
	8& {PyTorch} 2.5.1, Graph, TorchScript & CPU & $14.224 \pm 0.059$                 & $\times 1\,293.0\,\,\,$                    \\ \hline
	9& {PyTorch} 2.5.1, Eager, LibTorch, C++ & CPU & $7.611 \pm 0.079$                 & $\times 691.9\,\,\,$                    \\ \hline
	10& {JAX} 0.4.30, Eager, Python & CPU                & $506.464 \pm 1.555$         & $\times 46\,042.1$                      \\ \hline
	11& {JAX} 0.4.30, Graph, Semi-Python & CPU                & $11.156 \pm 0.0655$         & $\times 1014.1\,\,\,\,\,\,\,$                      \\ \hline
	12& {Micrograd}, Eager, Python & CPU                & $2.751 \pm 0.007$                & $\times 250.0\,\,\,\,\,\,\,$                      \\ \hline
	13& {Apple MLX} 0.22, Eager, Python & CPU                & $3.072 \pm 0.028$                & $\times 279.2\,\,\,\,\,\,\,$                      \\ \hline
\end{tabular}
\label{ch8:tab:execution_times_speedup_linux}
\end{table}

\begin{figure*}[ht]
\centering
\includegraphics[width=1.0\linewidth]{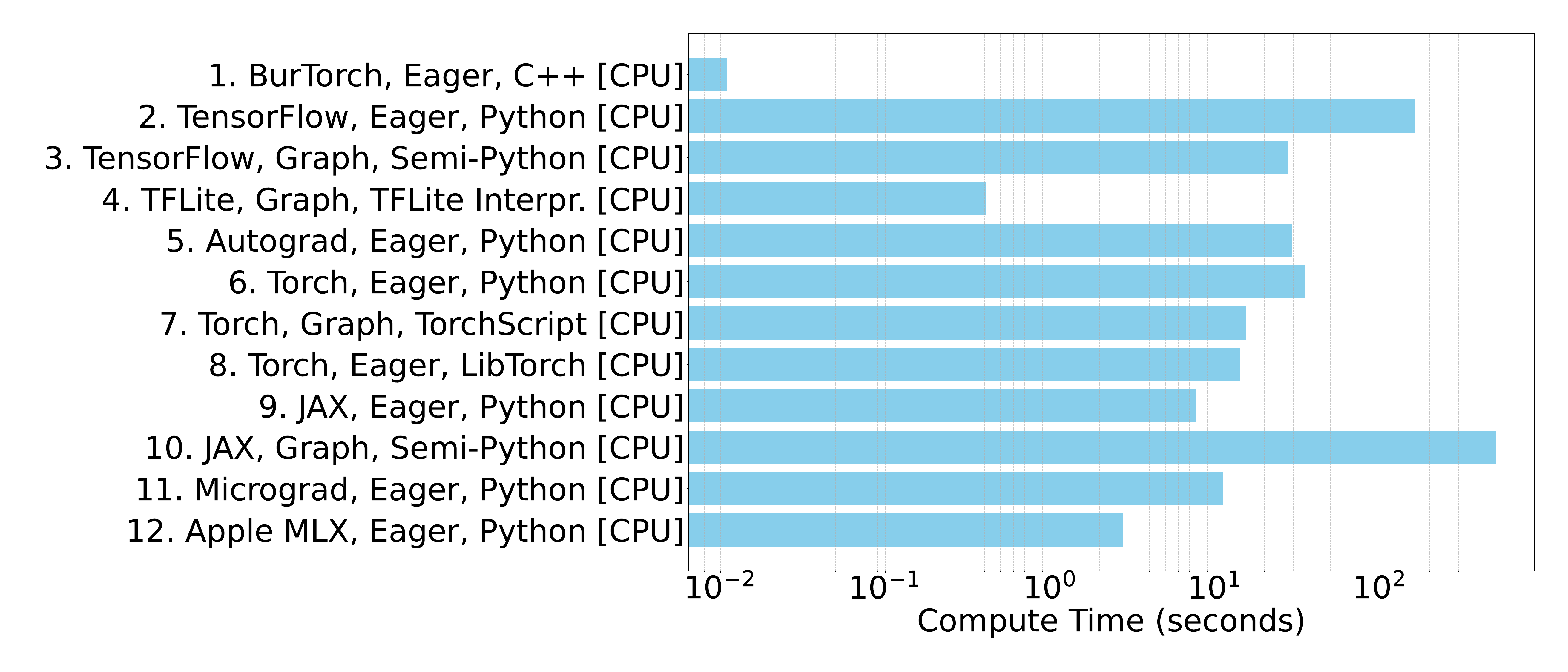}

\caption{Visualization of Table~\ref{ch8:tab:execution_times_speedup_linux}. Backpropagation over $100$K iterations with a {tiny} dynamic compute graph from Figure~\ref{ch8:fig:tiny-compute-graph}. Computation in FP64, one CPU Core $3.2$ GHz (x86-64). Linux Ubuntu 20.04.}
\label{ch8:fig:execution_times_speedup_linux}
\end{figure*}

\subsection{Experiments with tiny compute graph on Linux}
\label{ch8:app:extra-experiments-linux-tiny-graph}

This experiment replicates the setup from Section \ref{ch8:sec:experiments-tiny}. The results are presented in Table~\ref{ch8:tab:execution_times_speedup_linux} and Figure~\ref{ch8:fig:execution_times_speedup_linux}. The software and hardware environment used are detailed in Appendix \ref{ch8:app:exp-setup-linux}. In addition to the software used in Windows OS experiments, we have included a standalone Automatic Differentiation tool distributed by Apple, named \libname{Apple MLX} \citep{mlx2023}. The results in Table~\ref{ch8:tab:execution_times_speedup_linux} show that \libname{BurTorch} achieves significant speedups, thanks to its latency-optimized design.

\begin{table}[h!]
\footnotesize
\centering
\caption{Comparison of \libname{BurTorch} and \libname{PyTorch} performance for training MLP-like model. Batch: $b=1$, Compute: FP32, Single CPU core (Intel(R) Xeon(R) Gold 6146 CPU $3.20$ GHz). Initialization time is the end-to-end time for training a model with $1$ iterations. Compute time excludes batch preparation. Memory represents the peak private virtual memory per training. Linux Ubuntu 20.04.}
\label{ch8:tab:exp3-compute-and-mem-speedup-b1-linux}
\begin{tabular}{|l|l|l|l|l|l|l|l|}
	\hline
	\textbf{\#} & \textbf{Parameters (d)} & \multicolumn{3}{c|}{\textbf{\makecell[c]{PyTorch, \\Eager, v2.5.1 [CPU]}}} & \multicolumn{3}{c|}{\cellcolor{bgcolorwe}{\textbf{{BurTorch, Eager [CPU]}}}} \\ 
	\cline{3-8}
	& \textbf{Hidden Dim.(e)} & \textbf{\makecell[c]{Init\\(ms)}} & \textbf{\makecell[c]{Compute\\(ms)}} & \textbf{\makecell[c]{Mem.\\(MB)}} & {\textbf{\makecell[c]{Init\\(ms)}}} & {\textbf{\makecell[c]{Compute\\(ms)}}} & {\textbf{\makecell[c]{Mem.\\(MB)}}} \\ 
	\hline
	\hline
	1&$5,963\ (e=4)$ & $4\,649$ & $1.54 \pm 1.60$ & $13\,586$ & $8$ & $0.075 \pm 0.007$ & $41.8$ \\ 
	2&$18,587\ (e=16)$ & $4\,828$ & $1.54 \pm 1.43$ & $13\,647$ & $8$ & $0.150 \pm 0.001$ & $42.7$ \\ 
	3&$35,419\ (e=32)$ & $4\,469$ & $1.56 \pm 1.49$ & $13\,578$ & $16$ & $0.242 \pm 0.024$ & $44.1$ \\ 
	4&$69,083\ (e=64)$ & $4\,713$ & $1.67 \pm 1.62$ & $13\,643$ & $16$ & $0.378 \pm 0.038$ & $46.6$ \\ 
	5&$136,411\ (e=128)$ & $4\,879$ & $1.88 \pm 1.58$ & $13\,579$ & $27$ & $0.724 \pm 0.082$ & $51.3$ \\ 
	6&$540,379\ (e=512)$ & $4\,768$ & $2.37 \pm 1.47$ & $13\,598$ & $33$ & $2.382 \pm 0.306$ & $79.7$ \\ 
	7&$1,079,003\ (e=1024)$ & $5\,081$ & $3.23 \pm 1.52$ & $13\,661$ & $47$ & $6.192 \pm 0.948$ & $116.5$ \\ 
	\hline
\end{tabular}
\end{table}

\begin{table}[h!]
\footnotesize
\centering
\caption{Comparison of \libname{BurTorch} and \libname{PyTorch} performance for training MLP-like model. Batch: $b=64$, Compute: FP32, Single CPU core (Intel(R) Xeon(R) Gold 6146 CPU $3.20$ GHz). Initialization time is the end-to-end time for training a model with $1$ iterations. Compute time excludes batch preparation. Memory represents the peak private virtual memory. DRAM: 251 GB. Linux Ubuntu 20.04.}
\label{ch8:tab:exp3-compute-and-mem-speedup-b64-linux}
\begin{tabular}{|l|l|l|l|l|l|l|l|}
	\hline
	\textbf{\#} & \textbf{Parameters (d)} & \multicolumn{3}{c|}{\textbf{\makecell[c]{PyTorch, \\Eager, v2.5.1 [CPU]}}} & \multicolumn{3}{c|}{\cellcolor{bgcolorwe}{\textbf{{BurTorch, Eager [CPU]}}}} \\ 
	\cline{3-8}
	& \textbf{Hidden Dim.(e)} & \textbf{\makecell[c]{Init\\(ms)}} & \textbf{\makecell[c]{Compute\\(ms)}} & \textbf{\makecell[c]{Mem.\\(MB)}} & {\textbf{\makecell[c]{Init\\(ms)}}} & {\textbf{\makecell[c]{Compute\\(ms)}}} & {\textbf{\makecell[c]{Mem.\\(MB)}}} \\ 
	\hline
	\hline
	1& $5,963\ (e=4)$ & $5\,658$ & $9.98 \pm 1.68$ & $13\,602$ & $11$ & $0.976 \pm 0.023$ & $41.8$ \\
	2& $18,587\ (e=16)$ & $5\,774$ & $9.98 \pm 1.96$ & $13\,664$ & $13$ & $3.353 \pm 0.076$ & $42.7$ \\
	3& $35,419\ (e=32)$ & $5\,941$ & $10.37 \pm 2.02$ & $13\,603$ & $21$ & $6.566 \pm 0.156$ & $44.1$ \\
	4& $69,083\ (e=64)$ & $6\,020$ & $10.25 \pm 2.14$ & $13\,605$ & $28$ & $10.546 \pm 0.283$ & $46.6$ \\
	5& $136,411\ (e=128)$ & $6\,176$ & $22.49 \pm 3.55$ & $13\,625$ & $43$ & $21.557 \pm 0.496$ & $51.3$ \\
	6& $540,379\ (e=512)$ & $6\,151$ & $117.13 \pm 6.41$ & $13\,734$ & $136$ & $111.846 \pm 2.566$ & $79.7$ \\ 
	7&$1,079,003\ (e=1024)$ & $5\,926$ & $236.10 \pm 12.76$ & $13\,889$ & $272$ & $240.885 \pm 6.081$ & $116.5$ \\ 
	\hline
\end{tabular}
\end{table}

\subsection{Experiments with medium compute graph on Linux}
\label{ch8:app:extra-experiments-linux-mlp-llm}

In this experiment, we evaluated \libname{BurTorch} on a character-level autoregressive prediction model, designed as a medium-complexity compute graph, based on the architecture in \citet{bengio2000neural}, similar to the experiment in Section~\ref{ch8:sec:exp3-medium-compute-graphs}. The results, presented in Tables \ref{ch8:tab:exp3-compute-and-mem-speedup-b1-linux} and \ref{ch8:tab:exp3-compute-and-mem-speedup-b64-linux}, demonstrate \libname{BurTorch}'s efficiency in terms of initialization time and memory usage.

\subsection{Experiments with a GPT-3-like model on Linux}
\label{ch8:app:extra-experiments-linux-gpt3-like}

The experiments in this section replicate those from Section~\ref{ch8:sec:exp4-burtorch-fot-gpt3}, but were conducted on Linux OS (see Appendix~\ref{ch8:app:exp-setup-linux} for setup details). The results, presented in Table~\ref{ch8:tab:app-exp4-compute-and-mem-speedup-linux}, confirm that the findings are consistent with those in Section~\ref{ch8:sec:exp4-burtorch-fot-gpt3} across all considered OS (macOS, Linux, Windows). \libname{BurTorch} demonstrates strong potential for latency-sensitive applications, and due to its small code size, it may also be valuable for research purposes. As the batch size increases, the performance advantage of \libname{PyTorch} in terms of absolute runtime becomes more pronounced. Nevertheless, \libname{BurTorch} remains significantly more efficient in terms of real memory footprint, achieving approximately $\times 100$ smaller memory requirements.

\begin{table}[h!]
\footnotesize
\centering
\caption{Comparison of \libname{BurTorch} and \libname{PyTorch} performance for training \modelname{GPT-3} like model. FP32, Single CPU core (Intel(R) Xeon(R) Gold 6146 CPU $3.20$ GHz). Peak virtual memory per training. DRAM: 251 GB. Linux Ubuntu 20.04. Trainable variables: $46$K.}
\label{ch8:tab:app-exp4-compute-and-mem-speedup-linux}
\begin{tabular}{|l|cc|cc|cc|}
	\hline
	\textbf{Batch} & \multicolumn{2}{c|}{\cellcolor{bgcolorwe}{\textbf{BurTorch, Eager, C++}}} & \multicolumn{2}{c|}{\textbf{\makecell[c]{PyTorch,\\ Graph, TorchScript}}} & \multicolumn{2}{c|}{\textbf{\makecell[c]{PyTorch,\\ Eager, Python}}} \\ 
	\cline{2-7}
	& {\textbf{\makecell[c]{Compute\\(ms)}}} & {\textbf{\makecell[c]{Mem.\\(MB)}}} & \textbf{\makecell[c]{Compute\\(ms)}} & \textbf{\makecell[c]{Mem.\\(MB)}} & \textbf{\makecell[c]{Compute\\(ms)}} & \textbf{\makecell[c]{Mem.\\(MB)}} \\ 
	\hline
	\hline
	$1$ & $0.913 \pm 0.081$ & $22.3$ & $13.130 \pm 55.560$ & $9\,258$ & $16.144 \pm 0.653$ & $9\,111$ \\ 
	$2$ & $1.605 \pm 0.092$ & $22.3$ & $13.616 \pm 55.606$ & $9\,259$ & $16.570 \pm 0.656$ & $9\,111$ \\ 
	$4$ & $3.222 \pm 0.143$ & $22.3$ & $14.074 \pm 55.552$ & $9\,255$ & $17.034 \pm 0.713$ & $9\,111$ \\ 
	$8$ & $6.110 \pm 0.178$ & $22.3$ & $14.984 \pm 55.631$ & $9\,255$ & $17.842 \pm 0.693$ & $9\,108$ \\
	$16$ & $12.426 \pm 0.253$ & $22.3$ & $16.602 \pm 55.406$ & $9\,253$ & $19.475 \pm 0.701$ & $9\,104$ \\ 
	$32$ & $23.792 \pm 0.404$ & $22.3$ & $19.769 \pm 55.477$ & $9\,255$ & $22.481 \pm 0.724$ & $9\,105$ \\ 
	$64$ & $58.489 \pm 0.151$ & $22.3$ & $26.358 \pm 55.518$ & $9\,259$ & $28.372 \pm 0.801$ & $9\,107$ \\
	\hline
\end{tabular}
\end{table}

\clearpage
\addtocounter{adjsection}{1}
\section{Missing Experiment on macOS}
\label{ch8:app:extra-experiments-macos}

In this appendix, we present the results of comparing \libname{BurTorch} against \libname{PyTorch} under experimental settings similar to those described in the main text (Section \ref{ch8:sec:experiments}), but conducted on a macOS workstation running {macOS Sonoma 14.5}.

The CPU used in these experiments is x86-based, with a clock frequency of $2.3$ GHz during testing. For a detailed description of the hardware and software environment, please refer to Appendix~\ref{ch8:app:exp-setup-macos}.

For memory consumption, we report the peak resident memory usage (RES). RES represents the total amount of physical memory allocated to the process. The peak \texttt{VmSize} value on macOS can be somewhat ambiguous, as Apple produces a tightly integrated software-hardware ecosystem, making the exact meaning of \texttt{VmSize} less straightforward.

\subsection{Experiments with tiny compute graph on macOS}
\label{ch8:app:extra-experiments-macos-tiny-graph}

\begin{table}[h!]
\footnotesize
\centering
\caption{Backpropagation over $100$K iterations with a {tiny} dynamic compute graph from Figure~\ref{ch8:fig:tiny-compute-graph}. Mean and standard deviation across $5$ launches. Computation in FP64, one CPU Core $2.3$ GHz (x86-64). Physical Memory: 32 GB. macOS Sonoma 14.5. The numerical results across frameworks match exactly. See also Figure~\ref{ch8:fig:execution_times_speedup_macos}.
}
\begin{tabular}{|l|l|l|l|l|}
	\hline
	\textbf{\#} & \textbf{Framework, Mode, Language} & \textbf{Device} & \textbf{\makecell[c]{Compute Time \\ (sec.)}} & \textbf{\makecell[c]{Relative \\ to \\ BurTorch}} \\ 
	\hline
	\hline
	\cellcolor{bgcolorwe}1.&\cellcolor{bgcolorwe}BurTorch, Eager, C++ & \cellcolor{bgcolorwe}CPU                   & \cellcolor{bgcolorwe}$0.0118 \pm 0.00024$             & \cellcolor{bgcolorwe}$\times 1.0$ (We)             \\ 
	\hline
	2&TensorFlow 2.16.2, Eager, Python & CPU          & $145.312 \pm 2.287$         & $\times 12\,314.576\,\,\,$                    \\ \hline
	3&\makecell[l]{TensorFlow 2.16.2, Graph, \\Semi-Python} & CPU          & $33.041 \pm 0.386$         & $\times 2\,800.0847\,\,\,\,\,$                    \\ \hline
	4&\makecell[l]{TF Lite 2.16.2, Graph,\\TF Lite Interpreter} & CPU          & $0.728 \pm 0.0111$         & $\times 65$                    \\ \hline
	
	5&Autograd 1.7.0, Eager, Python & CPU            & $30.193 \pm 0.333$        & $\times 2\,558.728\,\,\,$                    \\ \hline
	6&{PyTorch} 2.2.2, Eager, Python & CPU          & $8.712 \pm 0.068$              & $\times 738.305\,\,\,$                    \\ \hline
	7&{PyTorch} 2.2.2, Graph, TorchScript & CPU & $4.978 \pm 0.050$                 & $\times 421.86\,\,\,$                    \\ \hline
	8&{PyTorch} 2.5.1, Eager, LibTorch, C++ & CPU & $5.439 \pm 0.127$                 & $\times 460.932\,\,\,$                    \\ \hline
	9&{JAX} 0.4.30, Eager, Python & CPU                & $445.015 \pm 1.921$         & $\times 37\,713.135$                      \\ \hline
	10&{JAX} 0.4.30, Graph, Semi-Python & CPU                & $12.091 \pm 0.248$         & $\times 1024.66\,\,\,\,\,\,\,$                      \\ \hline
	11&{Micrograd}, Eager, Python & CPU                & $2.399 \pm 0.039$                & $\times 203.305\,\,\,\,\,\,\,$                      \\ \hline
	12&{Apple MLX} 0.7, Eager, Python & CPU                & $3.138 \pm 0.0245$                & $\times 281.186\,\,\,\,\,\,\,$                      \\ \hline
\end{tabular}
\label{ch8:tab:execution_times_speedup_macos}
\end{table}

\begin{figure*}[ht]
\centering
\includegraphics[width=1.0\linewidth]{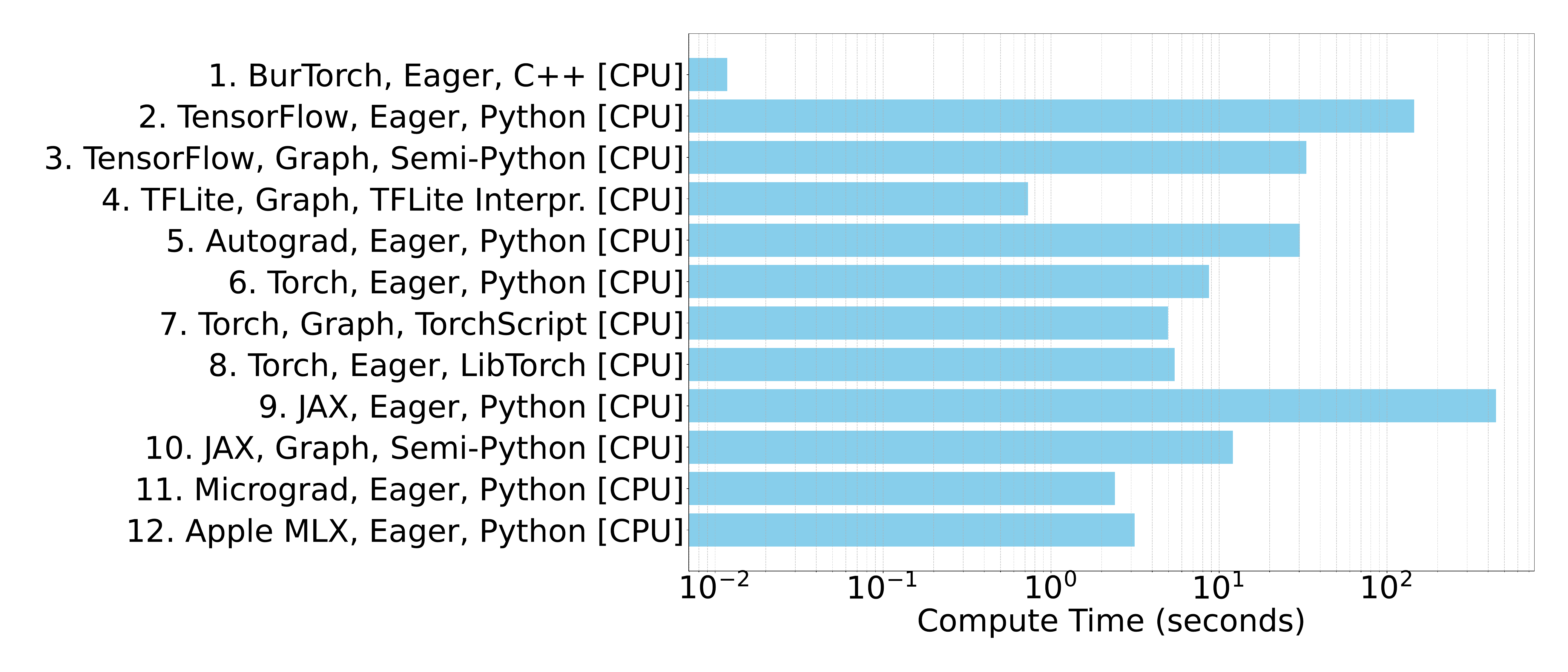}

\caption{Visualization of Table~\ref{ch8:tab:execution_times_speedup_macos}. Backpropagation over $100$K iterations with a {tiny} compute graph from Figure~\ref{ch8:fig:tiny-compute-graph}. Computation in FP64, one CPU Core $2.3$ GHz (x86-64), macOS Sonoma 14.5. The numerical results across frameworks match exactly.}
\label{ch8:fig:execution_times_speedup_macos}
\end{figure*}

This experiment replicates the setup from Section~\ref{ch8:sec:experiments-tiny}. The results are presented in Table~\ref{ch8:tab:execution_times_speedup_macos} and Figure~\ref{ch8:fig:execution_times_speedup_macos}. The software and hardware environment used is detailed in Appendix \ref{ch8:app:exp-setup-macos}. In addition to the software used in Windows OS experiments, we have included \libname{Apple MLX} \citep{mlx2023} for comparison, as done in Linux OS experiment.

\subsection{Experiments with medium compute graph on macOS}
\label{ch8:app:extra-experiments-macos-mlp-llm}

In this experiment, we evaluated \libname{BurTorch} on a character-level autoregressive prediction model, designed as a medium-complexity compute graph from \citet{bengio2000neural}, similar to the experiment in Section~\ref{ch8:sec:exp3-medium-compute-graphs}. The results, presented in Tables \ref{ch8:tab:exp3-compute-and-mem-speedup-b1-mac} and \ref{ch8:tab:exp3-compute-and-mem-speedup-b64-mac}, demonstrate \libname{BurTorch}'s efficiency in terms of initialization time and memory usage.

\begin{table}[h!]
\footnotesize
\centering
\caption{Comparison of \libname{BurTorch} and \libname{PyTorch} performance for training MLP like model. Batch: $b=1$, Compute: FP32, Single CPU core (Intel Quad-Core Intel Core i7 $2.3$ GHz). Initialization time is the end-to-end time for training a model with $1$ iterations. Compute time excludes batch preparation. Memory represents the peak private virtual memory per training. macOS Sonoma 14.5.}
\label{ch8:tab:exp3-compute-and-mem-speedup-b1-mac}
\begin{tabular}{|l|l|l|l|l|l|l|l|}
	\hline
	\textbf{\#} & \textbf{Parameters (d)} & \multicolumn{3}{c|}{\textbf{\makecell[c]{PyTorch, \\Eager, v2.5.1 [CPU]}}} & \multicolumn{3}{c|}{\cellcolor{bgcolorwe}{\textbf{{BurTorch, Eager [CPU]}}}} \\ 
	\cline{3-8}
	& \textbf{Hidden Dim.(e)} & \textbf{\makecell[c]{Init\\(ms)}} & \textbf{\makecell[c]{Compute\\(ms)}} & \textbf{\makecell[c]{Mem.\\(MB)}} & {\textbf{\makecell[c]{Init\\(ms)}}} & {\textbf{\makecell[c]{Compute\\(ms)}}} & {\textbf{\makecell[c]{Mem.\\(MB)}}} \\ 
	\hline
	\hline
	1& $5,963\ (e=4)$ & $6\,424$ & $1.08 \pm 0.35$ & $554$ & $15$ & $0.041 \pm 0.011$ & $36.2$ \\ 
	2& $18,587\ (e=16)$ & $6\,523$ & $1.08 \pm 0.36$ & $557$ & $15$ & $0.059 \pm 0.018$ & $36.8$ \\ 
	3& $35,419\ (e=32)$ & $6\,482$ & $1.12 \pm 0.38$ & $560$ & $16$ & $0.073 \pm 0.020$ & $38.2$ \\ 
	4& $69,083\ (e=64)$ & $6\,381$ & $1.23 \pm 0.37$ & $558$ & $18$ & $0.115 \pm 0.0402$ & $41.8$ \\ 
	5& $136,411\ (e=128)$ & $6\,451$ & $1.39 \pm 0.39$ & $558$ & $19$ & $0.190 \pm 0.059$ & $48.3$ \\ 
	6& $540,379\ (e=512)$ & $6\,443$ & $2.62 \pm 0.67$ & $579$ & $36$ & $0.969 \pm 0.200$ & $87.9$ \\ 
	7& $1,079,003\ (e=1024)$ & $6\,387$ & $4.40 \pm 0.88$ & $612$ & $61$ & $1.995 \pm 0.291$ & $132.0$ \\ 
	\hline
\end{tabular}
\end{table}

\begin{table}[h!]
\footnotesize
\centering
\caption{Comparison of \libname{BurTorch} and \libname{PyTorch} performance for training MLP-like model. Batch: $b=64$, Compute: FP32, Single CPU core (Intel Quad-Core Intel Core i7 $2.3$ GHz). Initialization time is the end-to-end time for training a model with $1$ iterations. Compute time excludes batch preparation. Memory represents the peak private virtual memory. DRAM: 32 GB. macOS Sonoma 14.5.}
\label{ch8:tab:exp3-compute-and-mem-speedup-b64-mac}
\begin{tabular}{|l|l|l|l|l|l|l|l|}
	\hline
	\textbf{\#} & \textbf{Parameters (d)} & \multicolumn{3}{c|}{\textbf{\makecell[c]{PyTorch, \\Eager, v2.5.1 [CPU]}}} & \multicolumn{3}{c|}{\cellcolor{bgcolorwe}{\textbf{{BurTorch, Eager [CPU]}}}} \\ 
	\cline{3-8}
	& \textbf{Hidden Dim.(e)} & \textbf{\makecell[c]{Init\\(ms)}} & \textbf{\makecell[c]{Compute\\(ms)}} & \textbf{\makecell[c]{Mem.\\(MB)}} & {\textbf{\makecell[c]{Init\\(ms)}}} & {\textbf{\makecell[c]{Compute\\(ms)}}} & {\textbf{\makecell[c]{Mem.\\(MB)}}} \\ 
	\hline
	\hline
	1& $5,963\ (e=4)$ & $6\,416$ & $7.87 \pm 1.18$ & $656$ & $16$ & $0.300 \pm 0.043$ & $36.2$ \\
	2& $18,587\ (e=16)$ & $6\,429$ & $8.21 \pm 1.04$ & $658$ & $16$ & $0.884 \pm 0.097$ & $37.3$ \\
	3& $35,419\ (e=32)$ & $6\,447$ & $9.24 \pm 1.22$ & $659$ & $18$ & $1.751 \pm 0.16$ & $38.9$ \\
	4& $69,083\ (e=64)$ & $6\,504$ & $11.43 \pm 1.08$ & $659$ & $20$ & $3.512 \pm 0.293$ & $41.7$ \\
	5& $136,411\ (e=128)$ & $6\,488$ & $16.63 \pm 1.35$ & $672$ & $29$ & $6.756 \pm 0.447$ & $47.4$ \\
	6& $540,379\ (e=512)$ & $6\,452$ & $43.02 \pm 1.66$ & $689$ & $75$ & $35.049 \pm 2.439$ & $102.3$ \\ 
	7& $1,079,003\ (e=1024)$ & $6\,512$ & $80.64 \pm 5.93$ & $715$ & $144$ & $80.363 \pm 2.885$ & $140.0$ \\ 
	\hline
\end{tabular}
\end{table}

\clearpage
\subsection{Experiments with a GPT-3-like model on macOS}
\label{ch8:app:extra-experiments-macos-gpt3-like}

The experiments in this section replicate those from Section~\ref{ch8:sec:exp4-burtorch-fot-gpt3}, but were conducted on macOS (see Appendix~\ref{ch8:app:exp-setup-macos} for setup details). The results, presented in Table~\ref{ch8:tab:app-exp4-compute-and-mem-speedup-mac}, show that the findings are consistent with those in Section~\ref{ch8:sec:exp4-burtorch-fot-gpt3}. \libname{BurTorch} demonstrates strong potential for latency-sensitive applications, and due to its small code size, it may also be valuable for research purposes. As the batch size increases, the performance advantage of \libname{PyTorch} in terms of absolute runtime becomes more noticeable. Nevertheless, as seen in Windows and Linux experiments, \libname{BurTorch} on macOS remains significantly more efficient in terms of real memory footprint.

\begin{table}[h!]
\footnotesize
\centering
\caption{Comparison of \libname{BurTorch} and \libname{PyTorch} performance for training \modelname{GPT-3} like model. FP32, Single CPU core (Intel Quad-Core Intel Core i7 $2.3$ GHz). Peak virtual memory per training. DRAM: 32 GB. macOS Sonoma 14.5. Trainable variables: $46$K.}
\label{ch8:tab:app-exp4-compute-and-mem-speedup-mac}
\begin{tabular}{|l|cc|cc|cc|}
	\hline
	\textbf{Batch} & \multicolumn{2}{c|}{\cellcolor{bgcolorwe}{\textbf{BurTorch, Eager, C++}}} & \multicolumn{2}{c|}{\textbf{\makecell[c]{PyTorch,\\ Graph, TorchScript}}} & \multicolumn{2}{c|}{\textbf{\makecell[c]{PyTorch,\\ Eager, Python}}} \\ 
	\cline{2-7}
	& {\textbf{\makecell[c]{Compute\\(ms)}}} & {\textbf{\makecell[c]{Mem.\\(MB)}}} & \textbf{\makecell[c]{Compute\\(ms)}} & \textbf{\makecell[c]{Mem.\\(MB)}} & \textbf{\makecell[c]{Compute\\(ms)}} & \textbf{\makecell[c]{Mem.\\(MB)}} \\ 
	\hline
	\hline
	$1$ & $0.739 \pm 0.157$ & $17.5$ & $14.598 \pm 61.975$ & $513$ & $12.765 \pm 2.396$ & $235$ \\ 
	$2$ & $1.429 \pm 0.227$ & $18.2$ & $13.363 \pm 53.726$ & $508$ & $13.147 \pm 2.412$ & $242$ \\ 
	$4$ & $3.001 \pm 0.495$ & $18.3$ & $14.594 \pm 56.745$ & $527$ & $13.159 \pm 2.459$ & $235$ \\ 
	$8$ & $5.617 \pm 0.542$ & $17.9$ & $15.242 \pm 58.458$ & $509$ & $14.781 \pm 3.501$ & $246$ \\
	$16$ & $11.218 \pm 0.722$ & $18.7$ & $14.466 \pm 45.581$ & $520$ & $14.765 \pm 2.480$ & $237$ \\ 
	$32$ & $22.575 \pm 1.269$ & $17.6$ & $16.819 \pm 46.574$ & $533$ & $18.216 \pm 3.069$ & $244$ \\ 
	$64$ & $45.047 \pm 2.538$ & $18.8$ & $22.288 \pm 46.398$ & $540$ & $25.403 \pm 4.393$ & $262$ \\
	\hline
\end{tabular}
\end{table}

\clearpage
\section{Experiment with Energy Drain on Windows OS}
\label{ch8:app:energy-eff}

\begin{table*}[h!]
\footnotesize
\centering
\caption{Power drain over 200K iterations with a small dynamically constructed compute graph (Figure~\ref{ch8:fig:exp2-small-compute-graph}) consisting of 32 nodes, using FP64. Voltage: 11.7V, Battery: DELL J8FK941J, Chemistry: Lithium-ion polymer, OS: Windows 11, 1 mWh = 3.6 Joules.
	See Figure~\ref{ch8:fig:exp2-power-drains-windows}. 
	The numerical results across frameworks match exactly.
}

\begin{tabular}{|l|l|l|l|l|l|}
	\hline
	\textbf{{\footnotesize{\#}}} & 
	\textbf{{\footnotesize{\makecell[l]{Framework, Mode,\\Language, Device}}}} &
	\textbf{\parbox{2.0cm}{\center \footnotesize{Time with All Initialization (sec.)}}} & 
	\textbf{\parbox{2.0cm}{\center \footnotesize{Total Energy Consumed (mWh)}}} & 
	\textbf{\parbox{2.0cm}{\center \footnotesize{Task Energy Consumed (mWh)}}} & 
	\textbf{\parbox{2.0cm}{\center \footnotesize{OS Energy Consumed (mWh)}}} \\
	\hline
	\hline
	
	\cellcolor{bgcolorwe}1& 
	\cellcolor{bgcolorwe}\makecell[l]{BurTorch, Eager, \,\,\,\,\,\,\,\,\,\,\,\,\,\,\,\,\,\,\,\,\,\\C++, CPU} & 
	\cellcolor{bgcolorwe}0.089
	&\cellcolor{bgcolorwe}0.94& \cellcolor{bgcolorwe}0.593 & \cellcolor{bgcolorwe}0.347 \\
	\hline
	
	2&
	\makecell[l]{TensorFlow, Eager,\\Python, CPU}	& 239.321
	& 1710 & 776.64 & 933.35\\
	\hline
	
	3&
	\makecell[l]{TensorFlow, Graph,\\Semi-Python, CPU}	& 36.693
	& 293 & 149.89 & 143.10 \\
	\hline
	
	4&
	\makecell[l]{TF Lite, Graph,\\TF Lite Interpreter, CPU}&
	29.736
	& 304 & 188.03 & 115.97\\
	\hline
	
	5&
	\makecell[l]{Autograd, Eager,\\Python, CPU}& 
	381.295
	& 2901 & 1413.95 & 1487.05 \\
	\hline
	
	6&
	\makecell[l]{PyTorch, Eager,\\Python, \textbf{GPU}} & 
	427.265
	& 6014 
	& 4347.67
	& 1666.33 \\
	\hline
	
	7&
	\makecell[l]{PyTorch, Eager, \\Python, CPU} & 
	53.648 &
	408
	& 198.78 & 
	209.22 \\
	\hline
	
	8&
	\makecell[l]{PyTorch, Graph, \\TorchScript, CPU} & 
	53.870 &
	432
	& 221.90 & 210.09\\
	\hline
	
	9&
	\makecell[l]{PyTorch, Eager, \\LibTorch, C++, CPU}&
	31.300 & 
	280
	& 157.93 & 122.07\\
	\hline
	
	10&
	\makecell[l]{JAX, Eager, \\Python, CPU} &
	1794.833 & 14765 & 7765.16 & 6999.84\\
	\hline
	
	11&
	\makecell[l]{JAX, Graph, \\Semi-Python, CPU} &
	11.805 & 
	82
	& 35.97 & 46.03\\
	\hline
	
	12&
	\makecell[l]{Micrograd, Eager, \\Python, CPU}&
	10.691 & 
	71
	& 29.307 & 41.694
	\\
	\hline
	
	\hline
\end{tabular}
\label{ch8:tab:exp2-power-drains-windows}
\end{table*}

\begin{figure*}[ht]
\centering
\includegraphics[width=1.0\linewidth]{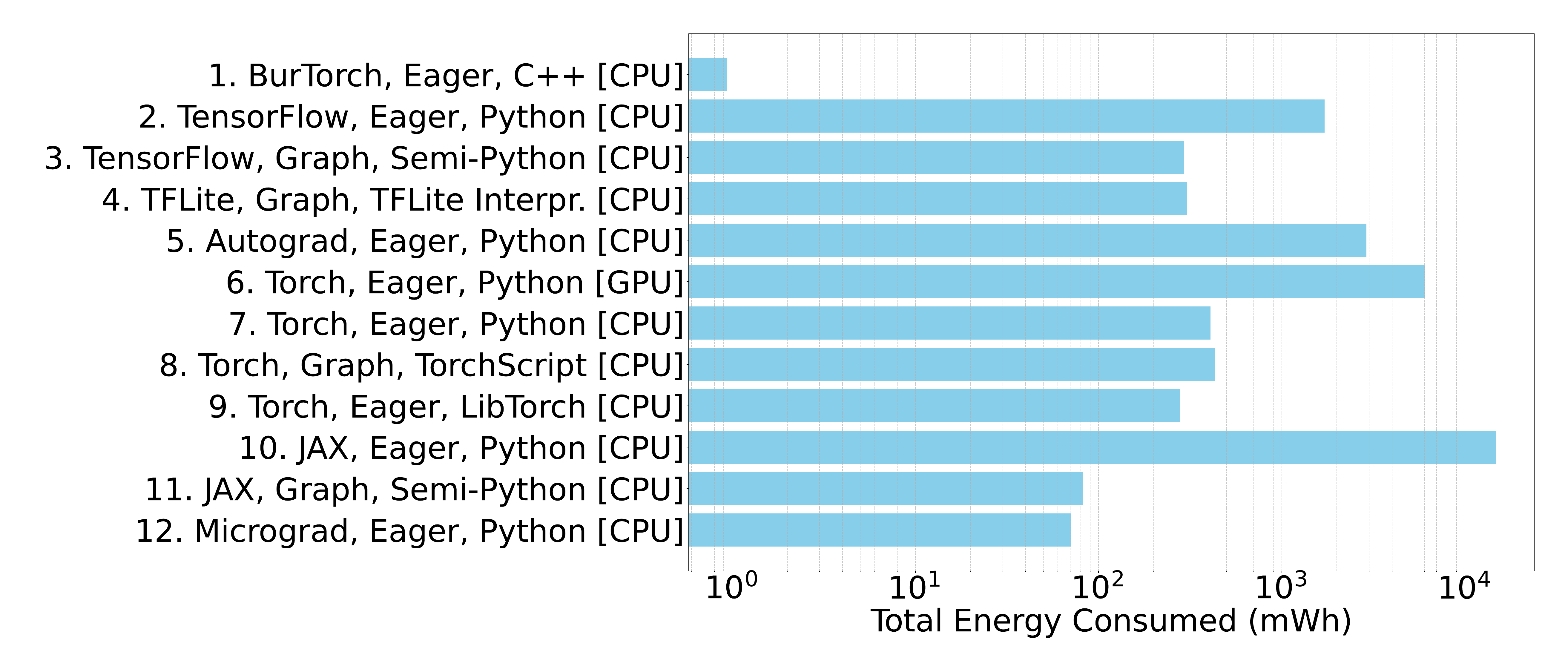}

\caption{Visualization of Table~\ref{ch8:tab:exp2-power-drains-windows}. Total power drain over 200K iterations with a \textit{small} dynamically constructed compute graph (Figure~\ref{ch8:fig:exp2-small-compute-graph}) consisting of 32 nodes, using FP64. Voltage: 11.7V, Battery: DELL J8FK941J, Chemistry: Li-poly, OS: Windows 11. The numerical results across frameworks match exactly.
}
\label{ch8:fig:exp2-power-drains-windows}
\end{figure*}

To evaluate energy consumption, we executed 200K iterations of the backpropagation algorithm across various computational frameworks for the computation graph from Figure~\ref{ch8:fig:exp2-small-compute-graph}. We compared end-to-end power consumption and end-to-end time. 

\paragraph{Preparation.} To ensure fair experimentation, we fixed all executions to run on physical CPU core number 2 (using 1-based indexing). The end-to-end computation was conducted on a Windows OS workstation, with the software and hardware environment detailed in Appendix~\ref{ch8:app:exp-setup-win}. The OS has 454 installed hardware drivers and 232 active processes. The high number of installed drivers is due to the presence of multiple hardware components. Having 232 running processes in Windows is standard practice, as many of them remain idle most of the time. Our system is equipped with the DELL J8FK941J battery. Charged Capacity: 92 828 mWh, Voltage: 11.102 volts, Chemistry: Lithium-ion polymer.

\paragraph{Cold state.}
Even when no applications are running, Windows OS core, installed hardware drivers, and system software impose a baseline load on the CPU. In this idle state, the cumulative CPU load ranges from 0.42\% to 0.76\% of its total processing capacity. During this cold phase, the system drains 3.9 mWh (milliwatt-hours) of energy per second from the battery. 
Thus, in the cold state, the hardware and OS software components consume power:
$$\mathrm{Consumed\,Power\,in\,Cold\,State} = 3.9 \cdot \dfrac{3600}{1000}=14.04 \mathrm{\,Joules\,per\,second}=14.04\mathrm{W}.$$

\paragraph{Measurements.}  
We use specific software frameworks to perform 200K iterations of backpropagation. In typical Deep Learning training, if 1000 epochs are used, with each epoch processing 200 data points, the total number of gradient oracle computations is approximately 200K. While this depends on the specific setup, the properties of \( f(x; D) \), the characteristics of the data \( D \), and the training algorithm, 200K iterations provide a reasonable (although rough) estimate.

We measure both time (in seconds) and energy (in mWh, where 1 mWh = 3.6 Joules)\footnote{To measure the battery capacity, we used the specialized tool BatteryInfoView v1.25 by NirSoft, which provides this information at one-second intervals.}. The execution includes time and energy for computation and memory transfers during end-to-end testing, covering all internal processes from initiating (for example, Python) to terminating the task in Windows 11, along with initialization, dynamic library loading, thread creation, and other overheads. The experiments run on a single $4.48$ GHz CPU core, with all hardware and software components powered solely by the installed battery. For more information about the environment, see Appendix~\ref{ch8:app:exp-setup-win}.

\paragraph{Results.}

The results are presented in Table~\ref{ch8:tab:exp2-power-drains-windows}. We observe (example 2) that, under light load, it is crucial to complete tasks quickly. As long as the OS is involved, it continues to drain energy. For example, in \libname{TensorFlow} Eager Mode experiment, the OS consumes more energy during the simulation than the application being executed. Next, we observe that using the GPU (examples 6 and 7) for small compute graphs not only increases wall-clock time but also leads to higher energy consumption. In general, non-Python solutions take less time and consume less energy (examples 2 and 3). However, relying solely on a compile-based environment (example 9) is insufficient. The choice of programming language plays a crucial role. \citet{pereira2021ranking} ranked languages based on various computational tasks, though not specifically for DL problems. Our work highlights that minimizing runtime overhead is equally critical, and this cannot be achieved solely by switching the programming language. Inefficiencies can arise in various forms in both scripting and compile-based language implementations. However, in scripting languages, when efficiency is paramount, the available tools are extremely limited.

Finally, as shown in Table~\ref{ch8:tab:exp2-power-drains-windows}, the energy consumed by \libname{BurTorch} is several orders of magnitude lower compared to current best-practice solutions. Specifically, \libname{BurTorch} (example 1) consumes only 0.94 mWh of total energy for 200K iterations, with 0.593 mWh dedicated to the task itself and 0.347 mWh for OS overhead. In contrast, \libname{TensorFlow} in Eager mode on CPU (example 2) consumes 1710 mWh, and \libname{PyTorch} in Eager mode on CPU (example 7) requires 408 mWh, highlighting a stark reduction in energy consumption by \libname{BurTorch}. These results demonstrate the significant efficiency advantages of \libname{BurTorch}, particularly when considering the energy demands of deploying this Deep Learning framework.

\addtocounter{adjsection}{1}
\section{Reproducibility}

The source code for the experiments, implementation, and documentation is publicly available at the following link:

\begin{center} 
	\href{https://github.com/burlachenkok/burtorch}{https://github.com/burlachenkok/burtorch}
\end{center}

\unappendix


\chapter{Concluding Remarks}
\label{chapater-conclusion}

\thispagestyle{empty}

\section{Summary}
\label{con:sum}

Federated Learning (FL) has emerged as a promising technique that enables edge devices to collaboratively train a shared machine learning model while keeping data localized, eliminating the need to store and access the full dataset in the cloud. This thesis explores various aspects of FL, aiming to bring it into practical use from different perspectives, ultimately broadening its future applications and uncovering intriguing insights along the way. Below, we provide a brief overview of the context for each main chapter, along with a key takeaway from each. For additional perspectives, see Table~\ref{ch1:tbl:algorithms} and Section~\ref{ch1:sec:organization}.


\paragraph{Chapter 2: FL\_PyTorch.} Chapter~\ref{chapter2} highlights the challenges of implementing, testing, and deploying Federated Learning (FL) in practice, particularly due to the heterogeneity of edge device environments. This complexity makes it fundamentally difficult for researchers to efficiently prototype and evaluate their optimization algorithms. In this chapter, we address these challenges by introducing \fl, an open-source software suite written in Python, built on top of \libname{PyTorch}, one of the most widely used Deep Learning (DL) frameworks. We designed \fl as a research simulator for FL, enabling fast development, prototyping, and experimentation with both existing and novel FL optimization algorithms. Our system provides flexible abstractions that allow researchers to explore a wide range of approaches to advance the state of the art. Additionally, \fl features a simple-to-use console interface, supporting the simultaneous execution of multiple clients on local CPUs, GPUs, or even remote compute devices—without requiring users to implement a distributed system themselves. \fl also includes a Graphical User Interface for enhanced usability. For new methods, researchers only need to provide the centralized implementation of their algorithms.

\begin{center}
	\textit{Advancing research requires tools with functionalities distinct from industrial runtimes.}
\end{center}

\paragraph{Chapter 3: EF21-W.} Chapter~\ref{chapter3} showcases our discovery of improvements to the state-of-the-art \algname{EF21} method. Error Feedback (\algname{EF}) is a highly popular and immensely effective mechanism for fixing convergence issues that arise in distributed training methods (such as distributed \algname{GD} or \algname{SGD}) when these are enhanced with greedy communication compression techniques such as \compname{TopK}. While \algname{EF} was proposed almost a decade ago~\citep{Seide2014}, and despite the concentrated effort by the community to advance the theoretical understanding of this mechanism, there is still a lot to explore. In this work, we study a modern form of error feedback called \algname{EF21}~\citep{EF21} which offers the currently best-known theoretical guarantees, under the weakest assumptions, and also works well in practice. In particular, while the theoretical communication complexity of \algname{EF21} depends on the {\em quadratic mean} of certain smoothness parameters, we improve this dependence to their {\em arithmetic mean}, which is always smaller and can be substantially smaller, especially in heterogeneous data regimes. We take the reader on a journey of our discovery process. Starting with the idea of applying \algname{EF21} to an equivalent reformulation of the underlying problem which (unfortunately) requires (often impractical) machine {\em cloning}, we continue to the discovery of a new {\em weighted} version of \algname{EF21} which can (fortunately) be executed without any cloning, and finally circle back to an improved {\em analysis} of the original \algname{EF21} method. While this development applies to the simplest form of \algname{EF21}, our approach naturally extends to more elaborate variants involving stochastic gradients and partial participation. Further, our technique improves the best-known theory of \algname{EF21} in the {\em rare features} regime~\citep{richtarik2023error}. Finally, we validate our theoretical findings with suitable experiments.

\begin{center}
	\textit{
		Discovery can also begin with a single observation, already embedded in the problem formulation, waiting to be uncovered.
	}
\end{center}

\paragraph{Chapter 4: DCGD/PermK/AES.} Chapter~\ref{chapter4} shares the way to use Classical Cryptography in some Federated Learning, even though employing Classical Cryptography was assumed to be impossible in the past in the context of {FL}. Federated Learning offers a paradigm that empowers distributed AI model training without collecting raw data. And there are different choices for providing privacy during FL training. One of the popular methodologies is employing Homomorphic Encryption (\abr{HE}) - a breakthrough in privacy-preserving computation from Cryptography. However, these methods have a price in the form of extra computation and memory footprint. To resolve these issues, we propose an innovative framework that synergizes permutation-based compressors with Classical Cryptography. Our framework offers a way to replace \abr{HE} with cheaper Classical Cryptography primitives, which provides security for the training process. It fosters asynchronous communication and provides flexible deployment options in various communication topologies.

\begin{center}
	\textit{
		The connection between two different fields can be established using tools not originally designed for this purpose.
		}
\end{center}

\paragraph{Chapter 5: PAGE Extensions.} Chapter~\ref{chapter5} revisits the classical problem of finding an approximately stationary point of the average of $n$ smooth and possibly non-convex functions. The optimal complexity of stochastic first-order methods in terms of the number of gradient computations of individual functions is $\mathcal{O}\left(n + n^{1/2}\varepsilon^{-1}\right)$, attained by the optimal \algname{SGD} methods \algname{SPIDER} \citep{fang2018spider} and \algname{PAGE} \citep{li2021page}, for example, where $\varepsilon$ is the error tolerance. However, i) the big-$\mathcal{O}$ notation hides crucial dependencies on the smoothness constants associated with the functions, and ii) the rates and theory in these methods assume simplistic sampling mechanisms that do not offer any flexibility. In this work, we remedy the situation. First, we generalize \algname{PAGE} algorithm so that it can provably work with virtually any (unbiased) sampling mechanism. This is particularly useful in Federated Learning, as it allows us to construct and better understand the impact of various combinations of client and data sampling strategies. Second, our analysis is sharper as we make explicit use of certain novel inequalities that capture the intricate interplay between the smoothness constants and the sampling procedure. Indeed, our analysis is better even for the simple sampling procedure analyzed in the \algname{PAGE} paper. However, this already improved bound can be further sharpened by a different sampling scheme that we propose. In summary, we provide the most general and most accurate analysis of optimal \algname{SGD} in the smooth non-convex regime. Finally, our theoretical findings are supported by carefully designed experiments.

\begin{center}
	\textit{
		Even theoretically optimal methods can be improved by revealing hidden details within big-$\mathcal{O}$ notation using mathematical tools, challenging the belief that improvements in big-$\mathcal{O}$ complexity come only from implementation.
	}
\end{center}

\paragraph{Chapter 6: Compressed L2GD.} Chapter~\ref{chapter6} explores the potential for combining personalization in Federated Learning (FL) with compression techniques. Existing FL algorithms aim to learn a single global model for all participating devices, which may not be helpful to all devices participating in the training due to the heterogeneity of the data across the devices. Recently, Hanzely and Richt\'{a}rik (2020) proposed a new formulation for training personalized FL models aimed at balancing the trade-off between the traditional global model and the local models that could be trained by individual devices using their private data only. They derived a new algorithm, called {\em loopless gradient descent}~(\algname{L2GD}), to solve it and showed that this algorithm leads to improved communication complexity guarantees in regimes when more personalization is required. In this work, we equip their \algname{L2GD} algorithm with a {\em bidirectional} compression mechanism to further reduce the communication bottleneck between the local devices and the server. Unlike other compression-based algorithms used in the FL-setting, our compressed \algname{L2GD} algorithm operates on a probabilistic communication protocol, where communication does not happen on a fixed schedule. Moreover, our compressed \algname{L2GD} algorithm maintains a similar convergence rate as vanilla \algname{SGD} without compression. To empirically validate the efficiency of our algorithm, we perform diverse numerical experiments on both convex and non-convex problems and use various compression techniques.

\begin{center}
	\textit{Research is not solely about creating the new; it also emerges from refining and recombining existing ideas in novel ways.}
\end{center}

\paragraph{Chapter 7: Unlocking FedNL.} Chapter~\ref{chapter7} introduces a practical improvement of Federated Newton Learn (\algname{FedNL}). The recent work \citet{safaryan2021fednl} introduces a family of Federated Newton Learn (\algname{FedNL}) algorithms, marking a significant step towards applying second-order methods to {FL} and large-scale optimization. However, the reference \algname{FedNL} prototype exhibits three serious practical drawbacks: (i) It requires $4.8$ hours to launch a single experiment in a server-grade workstation; (ii) The prototype only simulates multi-node setting; (iii) Prototype integration into resource-constrained applications is challenging. To bridge the gap between theory and practice, we present a self-contained implementation of \algname{FedNL}, \algname{FedNL-LS}, \algname{FedNL-PP} for single-node and multi-node settings. Our work resolves the aforementioned issues and reduces the wall clock time by $\times 1000$. With this \algname{FedNL} outperforms alternatives for training \modelname{logistic regression} in a single-node -- \libname{CVXPY} \citep{diamond2016cvxpy}, and in a multi-node -- \libname{Apache Spark} \citep{meng2016mllib}, \libname{Ray/Scikit-Learn} \citep{moritz2018ray}. Finally, we propose two practical-oriented compressors for \algname{FedNL} - adaptive \compname{TopLEK} and cache-aware \compname{RandSeqK}, which fulfill the theory of \algname{FedNL}.

\begin{center}
	\textit{Adapting mathematical algorithms to hardware and thinking beyond traditional research methodologies can lead to remarkable improvement gains, such as $\times 1000$.}
\end{center}

\paragraph{Chapter 8: BurTorch.} Chapter~\ref{chapter8} introduces Backpropagation Ultrafast Runtime (\libname{BurTorch}), a high-performance and compact framework designed to optimize machine learning (ML) training on single-node workstations through a highly optimized Backpropagation \citep{rumelhart1986learning, linnainmaa1970representation} implementation. While modern ML frameworks rely on compile-like stages to optimize compute graphs and improve wall-clock time, \libname{BurTorch} takes a different approach, embracing a minimalistic design that encourages the use of classical compile-based languages in ML research in the first place. Due to its small code size, the compile-based \libname{BurTorch} offers a user experience nearly indistinguishable from that of script-based programming environments. By eliminating the overhead inherent in large frameworks and leveraging careful implementation choices, \libname{BurTorch} achieves significant performance and memory efficiency gains over existing Deep Learning (DL) frameworks, including \libname{JAX} \citep{jax2018github}, \libname{PyTorch} \citep{paszke2019pytorch}, and \libname{TensorFlow} \citep{abadi2016tensorflow}, particularly during the computation of $\nabla f(x)$. With its emphasis on efficiency and compactness, \libname{BurTorch} is a valuable tool for applications requiring fine-grained control or low-latency execution. Its minimalist codebase fosters tighter integration between education, research, and industry, effectively bridging the gaps.

\begin{center}
	\textit{Progress in the field can also come from re-implementing foundational concepts from overlooked perspectives, even those concepts articulated long ago.}
\end{center}

\section{Future Research}
\label{con:future}

In this section, we outline a few promising directions for future work.
\\
\\
$\bullet$ Chapter~\ref{chapter2} introduces \fl, a suite of open-source software written in Python that builds on one of the most popular Deep Learning (DL) frameworks, \libname{PyTorch}. \fl emphasizes the importance of treating scientific experiments as assets, fostering enhanced reproducibility in FL research. For future evolution, it is crucial to provide users with more pre-implemented algorithms, datasets, and models. Additionally, expanding our visualization tools is an important goal.
\\
\\
$\bullet$ Chapter~\ref{chapter3} showcases our discovery of improvements to the state-of-the-art \algname{EF21} method. We share our discovery process with the reader, hoping that it can be useful for researchers who face difficulties in some part of the research journey. We believe that the underlying reason for the empirical good behavior of \algname{EF21} is a complex question, and further discoveries are needed in this direction.
\\
\\
$\bullet$ Chapter~\ref{chapter4} shares the way to use Classical Cryptography in some Federated Learning. One limitation of our work is the assumption that clients trust each other. However, it is also presented in \abr{HE}. The second limitation is that this work studies the case $d>n$. The work does not provide a rigorous theoretical analysis and focused on practical aspects of the proposed framework. Our method achieves strong privacy guarantees without compromising efficiency or accuracy via leveraging existing Cryptography protocols. For future research, in addition to developing rigorous theory and developing a strategy for $d>n$, we believe that our work provides a bridge to utilizing other fields of science that work only on the level of bits. In addition to Cryptography, another class of methods that operate on a bitwise representation of information is lossless data compression. This line can be investigated within \algname{DCGD/PermK} framework in future research. Next, as we have described in Appendix \ref{ch4:app:flexibility_for_dl_training}, our work potentially opens an extra degree of freedom for research in Systems and Compilers for DL that investigates different parallelism scheduling strategies for optimizing gradient oracles $\nabla f_i(x)$.
\\
\\
$\bullet$ Chapter~\ref{chapter6} introduces equipment of a new formulation for training personalized FL models aimed at balancing the trade-off between the traditional global model and the local models with a {\em bidirectional} compression mechanism to further reduce the communication bottleneck between the local devices and the server. However, several questions remain open and merit further investigation in the future. For example, one can include compression when devices calculate their local updates, especially in an FL setting, as the devices might not be powerful, and the computing energy is limited, and examine how the algorithm behaves. Additionally, we observed the efficacy of \algname{Compressed L2GD} with a biased compressor, such as \compname{TopK}. Nevertheless, extending \algname{Compressed L2GD} theory for biased compressors (with or without error-feedback \citep{xu2021grace}) is nontrivial and challenging. In the future, proving more general theory for \algname{Compressed L2GD} that includes both biased and unbiased compressors operating in a bidirectional fashion can enhance this direction further. Finally, a more detailed meta-parameter study covering different network bandwidths, diverse ML tasks with different DNN architectures, and deploying the models on real-life, geographically remote servers can also be a future empirical quest.
\\
\\
$\bullet$ Chapter~\ref{chapter7} introduces a practical improvement of Federated Newton Learn (\algname{FedNL}). Our tested operating systems encompass macOS, Linux, and Windows but exclude mobile platforms, even though our implementation supports ARM CPU with (and without) ARM Neon extensions\clrshort{, and our work was tested in a rich set of configurations (see Appendix~\ref{ch7:app:supported-os-and-compilers}, \ref{ch7:app:cpus}).} Next, the main issue that motivated our work was the extremely long duration of executing \algname{FedNL}. The memory aspect of dataset storage has not been considered in our work. Essentially, the dataset during training is read from preprocessed in-memory storage rather than from disk. Next, interesting mathematical aspects are still unexplored for \algname{FedNL}, including a more robust globalization strategy, addressing storage concerns for Hessian shifts $H_i^k$ in clients and $H^k$ in master, or integrating iterative inexact linear solvers. Finally, work has limited GPU support. GPU acceleration can significantly enhance performance, provided the overhead of executing code on the GPU and transferring input/output data is minimal or can be hidden. Our current support for GPU utilization is limited, focusing primarily on help for forming oracles using NVIDIA CUDA or exploiting linear algebra primitives. Expanding GPU support in the future presents both interesting opportunities and challenges.
\\
\\
$\bullet$ Chapter~\ref{chapter8} introduces Backpropagation Ultrafast Runtime (\libname{BurTorch}), a high-performance and compact framework designed to optimize machine learning (ML) training on single-node workstations. While \libname{BurTorch} demonstrates significant improvements in computation speed, memory, and energy efficiency, several limitations should be acknowledged: (1) \textit{Limited Scalability to Large Batch Sizes.} While \libname{BurTorch} outperforms \libname{PyTorch} in low-latency scenarios and small batch sizes, its advantages diminish as batch size increases. \libname{PyTorch}'s optimized batching strategies and extensive parallelism enable better absolute performance for large-scale training workloads; (2) \textit{CPU-Centric Design.} \libname{BurTorch} is optimized for CPU execution and is not intended for GPU-accelerated environments. The organization of latency-aware GPU computations differs significantly from CPU-based execution, and exploring this difference remains an open question for future research; (3) \textit{{Optimization for Small Compute Graphs.}} \libname{BurTorch}'s efficiency gains are particularly suited for small compute graphs and latency-sensitive applications. Scaling \libname{BurTorch} to support larger graphs remains an open research question; (4) \textit{{Community and Maintainability.}} \libname{PyTorch} and \libname{TensorFlow} benefit from large developer communities and continuous updates, ensuring ongoing improvements and bug fixes. \libname{BurTorch}, in contrast, offers unit tests, a small codebase, and example implementations, but its development is expected to be driven primarily by external researchers. Establishing an effective feedback loop with contributors potentially through a plugin system is an area for future consideration.


\begin{onehalfspacing}
	\renewcommand*\bibname{\centerline{REFERENCES}} 
    \phantomsection
	\addcontentsline{toc}{chapter}{References}
	\newcommand{\BIBdecl}{\setlength{\itemsep}{0pt}}
        \bibliographystyle{ACM-Reference-Format}
		\bibliography{References}


\begin{thebibliography}{345}


\ifx \showCODEN    \undefined \def \showCODEN     #1{\unskip}     \fi
\ifx \showDOI      \undefined \def \showDOI       #1{#1}\fi
\ifx \showISBNx    \undefined \def \showISBNx     #1{\unskip}     \fi
\ifx \showISBNxiii \undefined \def \showISBNxiii  #1{\unskip}     \fi
\ifx \showISSN     \undefined \def \showISSN      #1{\unskip}     \fi
\ifx \showLCCN     \undefined \def \showLCCN      #1{\unskip}     \fi
\ifx \shownote     \undefined \def \shownote      #1{#1}          \fi
\ifx \showarticletitle \undefined \def \showarticletitle #1{#1}   \fi
\ifx \showURL      \undefined \def \showURL       {\relax}        \fi
\providecommand\bibfield[2]{#2}
\providecommand\bibinfo[2]{#2}
\providecommand\natexlab[1]{#1}
\providecommand\showeprint[2][]{arXiv:#2}

\bibitem[Abadi et~al\mbox{.}(2016)]%
        {abadi2016tensorflow}
\bibfield{author}{\bibinfo{person}{Mart{\'{\i}}n Abadi}, \bibinfo{person}{Paul
  Barham}, \bibinfo{person}{Jianmin Chen}, \bibinfo{person}{Zhifeng Chen},
  \bibinfo{person}{Andy Davis}, \bibinfo{person}{Jeffrey Dean},
  \bibinfo{person}{Matthieu Devin}, \bibinfo{person}{Sanjay Ghemawat},
  \bibinfo{person}{Geoffrey Irving}, \bibinfo{person}{Michael Isard},
  \bibinfo{person}{Manjunath Kudlur}, \bibinfo{person}{Josh Levenberg},
  \bibinfo{person}{Rajat Monga}, \bibinfo{person}{Sherry Moore},
  \bibinfo{person}{Derek~Gordon Murray}, \bibinfo{person}{Benoit Steiner},
  \bibinfo{person}{Paul~A. Tucker}, \bibinfo{person}{Vijay Vasudevan},
  \bibinfo{person}{Pete Warden}, \bibinfo{person}{Martin Wicke},
  \bibinfo{person}{Yuan Yu}, {and} \bibinfo{person}{Xiaoqiang Zhang}.}
  \bibinfo{year}{2016}\natexlab{}.
\newblock \showarticletitle{TensorFlow: {A} system for large-scale machine
  learning}.
\newblock \bibinfo{journal}{\emph{CoRR}}  \bibinfo{volume}{abs/1605.08695}.
\newblock
\showeprint[arXiv]{1605.08695}
\urldef\tempurl%
\url{http://arxiv.org/abs/1605.08695}
\showURL{%
\tempurl}


\bibitem[Aji and Heafield(2017)]%
        {aji_sparse}
\bibfield{author}{\bibinfo{person}{Alham~Fikri Aji} {and}
  \bibinfo{person}{Kenneth Heafield}.} \bibinfo{year}{2017}\natexlab{}.
\newblock \showarticletitle{Sparse Communication for Distributed Gradient
  Descent}. In \bibinfo{booktitle}{\emph{Proceedings of the 2017 Conference on
  Empirical Methods in Natural Language Processing, {EMNLP} 2017, Copenhagen,
  Denmark, September 9-11, 2017}}, \bibfield{editor}{\bibinfo{person}{Martha
  Palmer}, \bibinfo{person}{Rebecca Hwa}, {and} \bibinfo{person}{Sebastian
  Riedel}} (Eds.). \bibinfo{publisher}{Association for Computational
  Linguistics}, \bibinfo{pages}{440--445}.
\newblock
\urldef\tempurl%
\url{https://doi.org/10.18653/V1/D17-1045}
\showDOI{\tempurl}


\bibitem[Alistarh et~al\mbox{.}(2017)]%
        {alistarh2017qsgd}
\bibfield{author}{\bibinfo{person}{Dan Alistarh}, \bibinfo{person}{Demjan
  Grubic}, \bibinfo{person}{Jerry Li}, \bibinfo{person}{Ryota Tomioka}, {and}
  \bibinfo{person}{Milan Vojnovic}.} \bibinfo{year}{2017}\natexlab{}.
\newblock \showarticletitle{{QSGD:} Communication-Efficient {SGD} via Gradient
  Quantization and Encoding}.
\newblock  (\bibinfo{year}{2017}), \bibinfo{pages}{1709--1720}.
\newblock
\urldef\tempurl%
\url{https://proceedings.neurips.cc/paper/2017/hash/6c340f25839e6acdc73414517203f5f0-Abstract.html}
\showURL{%
\tempurl}


\bibitem[Alistarh et~al\mbox{.}(2018)]%
        {Alistarh-EF-NIPS2018}
\bibfield{author}{\bibinfo{person}{Dan Alistarh}, \bibinfo{person}{Torsten
  Hoefler}, \bibinfo{person}{Mikael Johansson}, \bibinfo{person}{Nikola
  Konstantinov}, \bibinfo{person}{Sarit Khirirat}, {and}
  \bibinfo{person}{C{\'{e}}dric Renggli}.} \bibinfo{year}{2018}\natexlab{}.
\newblock \showarticletitle{The Convergence of Sparsified Gradient Methods}. In
  \bibinfo{booktitle}{\emph{Advances in Neural Information Processing Systems
  31: Annual Conference on Neural Information Processing Systems 2018, NeurIPS
  2018, December 3-8, 2018, Montr{\'{e}}al, Canada}},
  \bibfield{editor}{\bibinfo{person}{Samy Bengio}, \bibinfo{person}{Hanna~M.
  Wallach}, \bibinfo{person}{Hugo Larochelle}, \bibinfo{person}{Kristen
  Grauman}, \bibinfo{person}{Nicol{\`{o}} Cesa{-}Bianchi}, {and}
  \bibinfo{person}{Roman Garnett}} (Eds.). \bibinfo{pages}{5977--5987}.
\newblock
\urldef\tempurl%
\url{https://proceedings.neurips.cc/paper/2018/hash/314450613369e0ee72d0da7f6fee773c-Abstract.html}
\showURL{%
\tempurl}


\bibitem[Amiri et~al\mbox{.}(2020)]%
        {amiri2020federated}
\bibfield{author}{\bibinfo{person}{Mohammad~Mohammadi Amiri},
  \bibinfo{person}{Deniz G{\"{u}}nd{\"{u}}z}, \bibinfo{person}{Sanjeev~R.
  Kulkarni}, {and} \bibinfo{person}{H.~Vincent Poor}.}
  \bibinfo{year}{2020}\natexlab{}.
\newblock \showarticletitle{Federated Learning With Quantized Global Model
  Updates}.
\newblock \bibinfo{journal}{\emph{CoRR}}  \bibinfo{volume}{abs/2006.10672}
  (\bibinfo{year}{2020}).
\newblock
\showeprint[arXiv]{2006.10672}
\urldef\tempurl%
\url{https://arxiv.org/abs/2006.10672}
\showURL{%
\tempurl}


\bibitem[Apple(2017)]%
        {apple}
\bibfield{author}{\bibinfo{person}{Apple}.} \bibinfo{year}{2017}\natexlab{}.
\newblock \showarticletitle{{Learning with Privacy at Scale}}. In
  \bibinfo{booktitle}{\emph{Differential Privacy Team Technical Report}}.
\newblock


\bibitem[Apple(2019)]%
        {apple19wwdc}
\bibfield{author}{\bibinfo{person}{Apple}.} \bibinfo{year}{2019}\natexlab{}.
\newblock \bibinfo{title}{Designing for Privacy (video and slide deck)}.
\newblock \bibinfo{howpublished}{Apple WWDC,
  \url{https://developer.apple.com/videos/play/wwdc2019/708}}.
\newblock


\bibitem[ApS(2022)]%
        {aps2022mosek}
\bibfield{author}{\bibinfo{person}{Mosek ApS}.}
  \bibinfo{year}{2022}\natexlab{}.
\newblock \showarticletitle{Mosek optimizer API for python}.
\newblock \bibinfo{journal}{\emph{Version}} \bibinfo{volume}{9},
  \bibinfo{number}{17} (\bibinfo{year}{2022}), \bibinfo{pages}{6--4}.
\newblock
\urldef\tempurl%
\url{https://www.mosek.com}
\showURL{%
\tempurl}


\bibitem[Arivazhagan et~al\mbox{.}(2019)]%
        {arivazhagan2019federated}
\bibfield{author}{\bibinfo{person}{Manoj~Ghuhan Arivazhagan},
  \bibinfo{person}{Vinay Aggarwal}, \bibinfo{person}{Aaditya~Kumar Singh},
  {and} \bibinfo{person}{Sunav Choudhary}.} \bibinfo{year}{2019}\natexlab{}.
\newblock \showarticletitle{Federated Learning with Personalization Layers}.
\newblock \bibinfo{journal}{\emph{CoRR}}  \bibinfo{volume}{abs/1912.00818}
  (\bibinfo{year}{2019}).
\newblock
\showeprint[arXiv]{1912.00818}
\urldef\tempurl%
\url{http://arxiv.org/abs/1912.00818}
\showURL{%
\tempurl}


\bibitem[Arjevani et~al\mbox{.}(2023)]%
        {arjevani2019lower}
\bibfield{author}{\bibinfo{person}{Yossi Arjevani}, \bibinfo{person}{Yair
  Carmon}, \bibinfo{person}{John~C. Duchi}, \bibinfo{person}{Dylan~J. Foster},
  \bibinfo{person}{Nathan Srebro}, {and} \bibinfo{person}{Blake~E. Woodworth}.}
  \bibinfo{year}{2023}\natexlab{}.
\newblock \showarticletitle{Lower bounds for non-convex stochastic
  optimization}.
\newblock \bibinfo{journal}{\emph{Math. Program.}} \bibinfo{volume}{199},
  \bibinfo{number}{1} (\bibinfo{year}{2023}), \bibinfo{pages}{165--214}.
\newblock
\urldef\tempurl%
\url{https://doi.org/10.1007/S10107-022-01822-7}
\showDOI{\tempurl}


\bibitem[Aydin et~al\mbox{.}(2022)]%
        {aydin2022reveal}
\bibfield{author}{\bibinfo{person}{Furkan Aydin}, \bibinfo{person}{Emre
  Karabulut}, \bibinfo{person}{Seetal Potluri}, \bibinfo{person}{Erdem Alkim},
  {and} \bibinfo{person}{Aydin Aysu}.} \bibinfo{year}{2022}\natexlab{}.
\newblock \showarticletitle{RevEAL: Single-Trace Side-Channel Leakage of the
  {SEAL} Homomorphic Encryption Library}. In \bibinfo{booktitle}{\emph{2022
  Design, Automation {\&} Test in Europe Conference {\&} Exhibition, {DATE}
  2022, Antwerp, Belgium, March 14-23, 2022}},
  \bibfield{editor}{\bibinfo{person}{Cristiana Bolchini},
  \bibinfo{person}{Ingrid Verbauwhede}, {and} \bibinfo{person}{Elena{-}Ioana
  Vatajelu}} (Eds.). \bibinfo{publisher}{{IEEE}}, \bibinfo{pages}{1527--1532}.
\newblock
\urldef\tempurl%
\url{https://doi.org/10.23919/DATE54114.2022.9774724}
\showDOI{\tempurl}


\bibitem[Barak et~al\mbox{.}(2001)]%
        {barak2001possibility}
\bibfield{author}{\bibinfo{person}{Boaz Barak}, \bibinfo{person}{Oded
  Goldreich}, \bibinfo{person}{Russell Impagliazzo}, \bibinfo{person}{Steven
  Rudich}, \bibinfo{person}{Amit Sahai}, \bibinfo{person}{Salil~P. Vadhan},
  {and} \bibinfo{person}{Ke Yang}.} \bibinfo{year}{2001}\natexlab{}.
\newblock \showarticletitle{On the (Im)possibility of Obfuscating Programs}. In
  \bibinfo{booktitle}{\emph{Advances in Cryptology - {CRYPTO} 2001, 21st Annual
  International Cryptology Conference, Santa Barbara, California, USA, August
  19-23, 2001, Proceedings}} \emph{(\bibinfo{series}{Lecture Notes in Computer
  Science}, Vol.~\bibinfo{volume}{2139})},
  \bibfield{editor}{\bibinfo{person}{Joe Kilian}} (Ed.).
  \bibinfo{publisher}{Springer}, \bibinfo{pages}{1--18}.
\newblock
\urldef\tempurl%
\url{https://doi.org/10.1007/3-540-44647-8\_1}
\showDOI{\tempurl}


\bibitem[Barrett et~al\mbox{.}(2005)]%
        {barrett2005matplotlib}
\bibfield{author}{\bibinfo{person}{Paul Barrett}, \bibinfo{person}{John
  Hunter}, \bibinfo{person}{J~Todd Miller}, \bibinfo{person}{J-C Hsu}, {and}
  \bibinfo{person}{Perry Greenfield}.} \bibinfo{year}{2005}\natexlab{}.
\newblock \showarticletitle{matplotlib--A Portable Python Plotting Package}. In
  \bibinfo{booktitle}{\emph{Astronomical data analysis software and systems
  XIV}}, Vol.~\bibinfo{volume}{347}. \bibinfo{pages}{91}.
\newblock


\bibitem[Beazley and Team(1995)]%
        {beazley1995simplified}
\bibfield{author}{\bibinfo{person}{Dave Beazley} {and} \bibinfo{person}{SWIG
  Team}.} \bibinfo{year}{1995}\natexlab{}.
\newblock \bibinfo{title}{Simplified wrapper and interface generator}.
\newblock
\newblock
\urldef\tempurl%
\url{http://swig.sourceforge.net}
\showURL{%
\tempurl}
\newblock
\shownote{Software}.


\bibitem[Bell et~al\mbox{.}(2020)]%
        {bell2020secagg}
\bibfield{author}{\bibinfo{person}{James~Henry Bell},
  \bibinfo{person}{Kallista~A. Bonawitz}, \bibinfo{person}{Adri{\`{a}}
  Gasc{\'{o}}n}, \bibinfo{person}{Tancr{\`{e}}de Lepoint}, {and}
  \bibinfo{person}{Mariana Raykova}.} \bibinfo{year}{2020}\natexlab{}.
\newblock \showarticletitle{Secure Single-Server Aggregation with
  (Poly)Logarithmic Overhead}. In \bibinfo{booktitle}{\emph{{CCS} '20: 2020
  {ACM} {SIGSAC} Conference on Computer and Communications Security, Virtual
  Event, USA, November 9-13, 2020}}, \bibfield{editor}{\bibinfo{person}{Jay
  Ligatti}, \bibinfo{person}{Xinming Ou}, \bibinfo{person}{Jonathan Katz},
  {and} \bibinfo{person}{Giovanni Vigna}} (Eds.). \bibinfo{publisher}{{ACM}},
  \bibinfo{pages}{1253--1269}.
\newblock
\urldef\tempurl%
\url{https://doi.org/10.1145/3372297.3417885}
\showDOI{\tempurl}


\bibitem[Bellare et~al\mbox{.}(2004)]%
        {bellare2004eax}
\bibfield{author}{\bibinfo{person}{Mihir Bellare}, \bibinfo{person}{Phillip
  Rogaway}, {and} \bibinfo{person}{David~A. Wagner}.}
  \bibinfo{year}{2004}\natexlab{}.
\newblock \showarticletitle{The {EAX} Mode of Operation}.
\newblock   \bibinfo{volume}{3017} (\bibinfo{year}{2004}),
  \bibinfo{pages}{389--407}.
\newblock
\urldef\tempurl%
\url{https://doi.org/10.1007/978-3-540-25937-4\_25}
\showDOI{\tempurl}


\bibitem[Belson(1959)]%
        {belson1959matching}
\bibfield{author}{\bibinfo{person}{William~A Belson}.}
  \bibinfo{year}{1959}\natexlab{}.
\newblock \showarticletitle{Matching and prediction on the principle of
  biological classification}.
\newblock \bibinfo{journal}{\emph{Journal of the Royal Statistical Society:
  Series C (Applied Statistics)}} \bibinfo{volume}{8}, \bibinfo{number}{2}
  (\bibinfo{year}{1959}), \bibinfo{pages}{65--75}.
\newblock


\bibitem[Benaissa et~al\mbox{.}(2021)]%
        {benaissa2021tenseal}
\bibfield{author}{\bibinfo{person}{Ayoub Benaissa}, \bibinfo{person}{Bilal
  Retiat}, \bibinfo{person}{Bogdan Cebere}, {and} \bibinfo{person}{Alaa~Eddine
  Belfedhal}.} \bibinfo{year}{2021}\natexlab{}.
\newblock \showarticletitle{TenSEAL: {A} Library for Encrypted Tensor
  Operations Using Homomorphic Encryption}.
\newblock \bibinfo{journal}{\emph{CoRR}}  \bibinfo{volume}{abs/2104.03152}
  (\bibinfo{year}{2021}).
\newblock
\showeprint[arXiv]{2104.03152}
\urldef\tempurl%
\url{https://arxiv.org/abs/2104.03152}
\showURL{%
\tempurl}


\bibitem[Bengio et~al\mbox{.}(2000)]%
        {bengio2000neural}
\bibfield{author}{\bibinfo{person}{Yoshua Bengio},
  \bibinfo{person}{R{\'{e}}jean Ducharme}, {and} \bibinfo{person}{Pascal
  Vincent}.} \bibinfo{year}{2000}\natexlab{}.
\newblock \showarticletitle{A Neural Probabilistic Language Model}.
\newblock  (\bibinfo{year}{2000}), \bibinfo{pages}{932--938}.
\newblock
\urldef\tempurl%
\url{https://proceedings.neurips.cc/paper/2000/hash/728f206c2a01bf572b5940d7d9a8fa4c-Abstract.html}
\showURL{%
\tempurl}


\bibitem[Bergou et~al\mbox{.}(2023)]%
        {houcine2022personalized}
\bibfield{author}{\bibinfo{person}{El~Houcine Bergou},
  \bibinfo{person}{Konstantin Burlachenko}, \bibinfo{person}{Aritra Dutta},
  {and} \bibinfo{person}{Peter Richt{\'{a}}rik}.}
  \bibinfo{year}{2023}\natexlab{}.
\newblock \showarticletitle{Personalized Federated Learning with Communication
  Compression}.
\newblock \bibinfo{journal}{\emph{Trans. Mach. Learn. Res.}}
  \bibinfo{volume}{2023} (\bibinfo{year}{2023}).
\newblock
\urldef\tempurl%
\url{https://openreview.net/forum?id=dZugyhbNFY}
\showURL{%
\tempurl}


\bibitem[Bergstra et~al\mbox{.}(2010)]%
        {bergstra2010theano}
\bibfield{author}{\bibinfo{person}{James Bergstra}, \bibinfo{person}{Olivier
  Breuleux}, \bibinfo{person}{Fr{\'{e}}d{\'{e}}ric Bastien},
  \bibinfo{person}{Pascal Lamblin}, \bibinfo{person}{Razvan Pascanu},
  \bibinfo{person}{Guillaume Desjardins}, \bibinfo{person}{Joseph~P. Turian},
  \bibinfo{person}{David Warde{-}Farley}, {and} \bibinfo{person}{Yoshua
  Bengio}.} \bibinfo{year}{2010}\natexlab{}.
\newblock \showarticletitle{Theano: {A} {CPU} and {GPU} Math Compiler in
  Python}. In \bibinfo{booktitle}{\emph{Proceedings of the 9th Python in
  Science Conference 2010 (SciPy 2010), Austin, Texas, June 28 - July 3,
  2010}}, \bibfield{editor}{\bibinfo{person}{St{\'{e}}fan van~der Walt} {and}
  \bibinfo{person}{Jarrod Millman}} (Eds.). \bibinfo{publisher}{scipy.org},
  \bibinfo{pages}{18--24}.
\newblock
\urldef\tempurl%
\url{https://doi.org/10.25080/MAJORA-92BF1922-003}
\showDOI{\tempurl}


\bibitem[Berkson(1944)]%
        {berkson1944application}
\bibfield{author}{\bibinfo{person}{Joseph Berkson}.}
  \bibinfo{year}{1944}\natexlab{}.
\newblock \showarticletitle{Application of the logistic function to bio-assay}.
\newblock \bibinfo{journal}{\emph{Journal of the American statistical
  association}} \bibinfo{volume}{39}, \bibinfo{number}{227}
  (\bibinfo{year}{1944}), \bibinfo{pages}{357--365}.
\newblock


\bibitem[Bernstein(2008)]%
        {bernstein2008salsa20}
\bibfield{author}{\bibinfo{person}{Daniel~J. Bernstein}.}
  \bibinfo{year}{2008}\natexlab{}.
\newblock \showarticletitle{The Salsa20 Family of Stream Ciphers}.
\newblock   \bibinfo{volume}{4986} (\bibinfo{year}{2008}),
  \bibinfo{pages}{84--97}.
\newblock
\urldef\tempurl%
\url{https://doi.org/10.1007/978-3-540-68351-3\_8}
\showDOI{\tempurl}


\bibitem[Bernstein et~al\mbox{.}(2018)]%
        {signsgd}
\bibfield{author}{\bibinfo{person}{Jeremy Bernstein},
  \bibinfo{person}{Yu{-}Xiang Wang}, \bibinfo{person}{Kamyar Azizzadenesheli},
  {and} \bibinfo{person}{Animashree Anandkumar}.}
  \bibinfo{year}{2018}\natexlab{}.
\newblock \showarticletitle{{SIGNSGD:} Compressed Optimisation for Non-Convex
  Problems}. In \bibinfo{booktitle}{\emph{Proceedings of the 35th International
  Conference on Machine Learning, {ICML} 2018, Stockholmsm{\"{a}}ssan,
  Stockholm, Sweden, July 10-15, 2018}} \emph{(\bibinfo{series}{Proceedings of
  Machine Learning Research}, Vol.~\bibinfo{volume}{80})},
  \bibfield{editor}{\bibinfo{person}{Jennifer~G. Dy} {and}
  \bibinfo{person}{Andreas Krause}} (Eds.). \bibinfo{publisher}{{PMLR}},
  \bibinfo{pages}{559--568}.
\newblock
\urldef\tempurl%
\url{http://proceedings.mlr.press/v80/bernstein18a.html}
\showURL{%
\tempurl}


\bibitem[Bertsekas et~al\mbox{.}(2003)]%
        {bertsekas2003convex}
\bibfield{author}{\bibinfo{person}{Dimitri Bertsekas}, \bibinfo{person}{Angelia
  Nedic}, {and} \bibinfo{person}{Asuman Ozdaglar}.}
  \bibinfo{year}{2003}\natexlab{}.
\newblock \bibinfo{booktitle}{\emph{Convex analysis and optimization}}.
  Vol.~\bibinfo{volume}{1}.
\newblock \bibinfo{publisher}{Athena Scientific}.
\newblock


\bibitem[Beutel et~al\mbox{.}(2020)]%
        {beutel2020flower}
\bibfield{author}{\bibinfo{person}{Daniel~J. Beutel}, \bibinfo{person}{Taner
  Topal}, \bibinfo{person}{Akhil Mathur}, \bibinfo{person}{Xinchi Qiu},
  \bibinfo{person}{Titouan Parcollet}, {and} \bibinfo{person}{Nicholas~D.
  Lane}.} \bibinfo{year}{2020}\natexlab{}.
\newblock \showarticletitle{Flower: {A} Friendly Federated Learning Research
  Framework}.
\newblock \bibinfo{journal}{\emph{CoRR}}  \bibinfo{volume}{abs/2007.14390}
  (\bibinfo{year}{2020}).
\newblock
\showeprint[arXiv]{2007.14390}
\urldef\tempurl%
\url{https://arxiv.org/abs/2007.14390}
\showURL{%
\tempurl}


\bibitem[Beznosikov et~al\mbox{.}(2023)]%
        {beznosikov2020biased}
\bibfield{author}{\bibinfo{person}{Aleksandr Beznosikov},
  \bibinfo{person}{Samuel Horv{\'{a}}th}, \bibinfo{person}{Peter
  Richt{\'{a}}rik}, {and} \bibinfo{person}{Mher Safaryan}.}
  \bibinfo{year}{2023}\natexlab{}.
\newblock \showarticletitle{On Biased Compression for Distributed Learning}.
\newblock \bibinfo{journal}{\emph{J. Mach. Learn. Res.}}  \bibinfo{volume}{24}
  (\bibinfo{year}{2023}), \bibinfo{pages}{276:1--276:50}.
\newblock
\urldef\tempurl%
\url{http://jmlr.org/papers/v24/21-1548.html}
\showURL{%
\tempurl}


\bibitem[Bhowmick et~al\mbox{.}(2018)]%
        {bhowmick2018protection}
\bibfield{author}{\bibinfo{person}{Abhishek Bhowmick}, \bibinfo{person}{John~C.
  Duchi}, \bibinfo{person}{Julien Freudiger}, \bibinfo{person}{Gaurav Kapoor},
  {and} \bibinfo{person}{Ryan Rogers}.} \bibinfo{year}{2018}\natexlab{}.
\newblock \showarticletitle{Protection Against Reconstruction and Its
  Applications in Private Federated Learning}.
\newblock \bibinfo{journal}{\emph{CoRR}}  \bibinfo{volume}{abs/1812.00984}
  (\bibinfo{year}{2018}).
\newblock
\showeprint[arXiv]{1812.00984}
\urldef\tempurl%
\url{http://arxiv.org/abs/1812.00984}
\showURL{%
\tempurl}


\bibitem[Bianco et~al\mbox{.}(2018)]%
        {bianco2018benchmark}
\bibfield{author}{\bibinfo{person}{Simone Bianco}, \bibinfo{person}{R{\'{e}}mi
  Cad{\`{e}}ne}, \bibinfo{person}{Luigi Celona}, {and} \bibinfo{person}{Paolo
  Napoletano}.} \bibinfo{year}{2018}\natexlab{}.
\newblock \showarticletitle{Benchmark Analysis of Representative Deep Neural
  Network Architectures}.
\newblock \bibinfo{journal}{\emph{{IEEE} Access}}  \bibinfo{volume}{6}
  (\bibinfo{year}{2018}), \bibinfo{pages}{64270--64277}.
\newblock
\urldef\tempurl%
\url{https://doi.org/10.1109/ACCESS.2018.2877890}
\showDOI{\tempurl}


\bibitem[Biewald(2020)]%
        {wandb}
\bibfield{author}{\bibinfo{person}{Lukas Biewald}.}
  \bibinfo{year}{2020}\natexlab{}.
\newblock \bibinfo{title}{Experiment Tracking with Weights and Biases}.
\newblock
\newblock
\urldef\tempurl%
\url{https://www.wandb.com/}
\showURL{%
\tempurl}
\newblock
\shownote{Software available from wandb.com}.


\bibitem[Bishop(2007)]%
        {bishop2016pattern}
\bibfield{author}{\bibinfo{person}{Christopher~M. Bishop}.}
  \bibinfo{year}{2007}\natexlab{}.
\newblock \bibinfo{booktitle}{\emph{Pattern recognition and machine learning,
  5th Edition}}.
\newblock \bibinfo{publisher}{Springer}.
\newblock
\showISBNx{9780387310732}
\urldef\tempurl%
\url{https://www.worldcat.org/oclc/71008143}
\showURL{%
\tempurl}


\bibitem[Bishop and Bishop(2024)]%
        {bishop2023deep}
\bibfield{author}{\bibinfo{person}{Christopher~M. Bishop} {and}
  \bibinfo{person}{Hugh Bishop}.} \bibinfo{year}{2024}\natexlab{}.
\newblock \bibinfo{booktitle}{\emph{Deep Learning - Foundations and Concepts}}.
\newblock \bibinfo{publisher}{Springer}.
\newblock
\showISBNx{978-3-031-45467-7}
\urldef\tempurl%
\url{https://doi.org/10.1007/978-3-031-45468-4}
\showDOI{\tempurl}


\bibitem[Bishop and Nasrabadi(2006)]%
        {bishop2006pattern}
\bibfield{author}{\bibinfo{person}{Christopher~M Bishop} {and}
  \bibinfo{person}{Nasser~M Nasrabadi}.} \bibinfo{year}{2006}\natexlab{}.
\newblock \bibinfo{booktitle}{\emph{Pattern recognition and machine learning}}.
  Vol.~\bibinfo{volume}{4}.
\newblock \bibinfo{publisher}{Springer}.
\newblock


\bibitem[Blum et~al\mbox{.}(2003)]%
        {blum2003noise}
\bibfield{author}{\bibinfo{person}{Avrim Blum}, \bibinfo{person}{Adam Kalai},
  {and} \bibinfo{person}{Hal Wasserman}.} \bibinfo{year}{2003}\natexlab{}.
\newblock \showarticletitle{Noise-tolerant learning, the parity problem, and
  the statistical query model}.
\newblock \bibinfo{journal}{\emph{J. {ACM}}} \bibinfo{volume}{50},
  \bibinfo{number}{4} (\bibinfo{year}{2003}), \bibinfo{pages}{506--519}.
\newblock
\urldef\tempurl%
\url{https://doi.org/10.1145/792538.792543}
\showDOI{\tempurl}


\bibitem[Bogdanov et~al\mbox{.}(2011)]%
        {bogdanov2011biclique}
\bibfield{author}{\bibinfo{person}{Andrey Bogdanov}, \bibinfo{person}{Dmitry
  Khovratovich}, {and} \bibinfo{person}{Christian Rechberger}.}
  \bibinfo{year}{2011}\natexlab{}.
\newblock \showarticletitle{Biclique cryptanalysis of the full AES}. In
  \bibinfo{booktitle}{\emph{Advances in Cryptology--ASIACRYPT 2011: 17th
  International Conference on the Theory and Application of Cryptology and
  Information Security, Seoul, South Korea, December 4-8, 2011. Proceedings
  17}}. Springer, \bibinfo{pages}{344--371}.
\newblock


\bibitem[Bonawitz et~al\mbox{.}(2019)]%
        {bonawitz2019towards}
\bibfield{author}{\bibinfo{person}{Kallista~A. Bonawitz},
  \bibinfo{person}{Hubert Eichner}, \bibinfo{person}{Wolfgang Grieskamp},
  \bibinfo{person}{Dzmitry Huba}, \bibinfo{person}{Alex Ingerman},
  \bibinfo{person}{Vladimir Ivanov}, \bibinfo{person}{Chlo{\'{e}} Kiddon},
  \bibinfo{person}{Jakub Kone{\v{c}}n{\'y}}, \bibinfo{person}{Stefano
  Mazzocchi}, \bibinfo{person}{Brendan McMahan}, \bibinfo{person}{Timon~Van
  Overveldt}, \bibinfo{person}{David Petrou}, \bibinfo{person}{Daniel Ramage},
  {and} \bibinfo{person}{Jason Roselander}.} \bibinfo{year}{2019}\natexlab{}.
\newblock \showarticletitle{Towards Federated Learning at Scale: System
  Design}.
\newblock  (\bibinfo{year}{2019}).
\newblock
\urldef\tempurl%
\url{https://proceedings.mlsys.org/paper\_files/paper/2019/hash/7b770da633baf74895be22a8807f1a8f-Abstract.html}
\showURL{%
\tempurl}


\bibitem[Boneh and Shoup(2020)]%
        {boneh2020graduate}
\bibfield{author}{\bibinfo{person}{Dan Boneh} {and} \bibinfo{person}{Victor
  Shoup}.} \bibinfo{year}{2020}\natexlab{}.
\newblock \showarticletitle{A graduate course in applied cryptography}.
\newblock \bibinfo{journal}{\emph{Draft 0.5}} (\bibinfo{year}{2020}).
\newblock
\urldef\tempurl%
\url{https://crypto.stanford.edu/~dabo/cryptobook/BonehShoup_0_5.pdf}
\showURL{%
\tempurl}


\bibitem[Booth(2004)]%
        {booth2004research}
\bibfield{author}{\bibinfo{person}{Andrew Booth}.}
  \bibinfo{year}{2004}\natexlab{}.
\newblock \showarticletitle{What research studies do practitioners actually
  find useful?}
\newblock \bibinfo{journal}{\emph{Health information and libraries journal}}
  \bibinfo{volume}{21}, \bibinfo{number}{3} (\bibinfo{year}{2004}),
  \bibinfo{pages}{197--200}.
\newblock


\bibitem[Boyd and Vandenberghe(2014)]%
        {boyd2004convex}
\bibfield{author}{\bibinfo{person}{Stephen~P. Boyd} {and}
  \bibinfo{person}{Lieven Vandenberghe}.} \bibinfo{year}{2014}\natexlab{}.
\newblock \bibinfo{booktitle}{\emph{Convex Optimization}}.
\newblock \bibinfo{publisher}{Cambridge University Press}.
\newblock
\showISBNx{978-0-521-83378-3}
\urldef\tempurl%
\url{https://doi.org/10.1017/CBO9780511804441}
\showDOI{\tempurl}


\bibitem[Bozic et~al\mbox{.}(2024)]%
        {bovzivc2024testbed}
\bibfield{author}{\bibinfo{person}{Janez Bozic},
  \bibinfo{person}{Am{\^{a}}ndio~R. Faustino}, \bibinfo{person}{Boris Radovic},
  \bibinfo{person}{Marco Canini}, {and} \bibinfo{person}{Veljko Pejovic}.}
  \bibinfo{year}{2024}\natexlab{}.
\newblock \showarticletitle{Where is the Testbed for My Federated Learning
  Research?}. In \bibinfo{booktitle}{\emph{{IEEE/ACM} Symposium on Edge
  Computing, {SEC} 2024, Rome, Italy, December 4-7, 2024}}.
  \bibinfo{publisher}{{IEEE}}, \bibinfo{pages}{249--264}.
\newblock
\urldef\tempurl%
\url{https://doi.org/10.1109/SEC62691.2024.00027}
\showDOI{\tempurl}


\bibitem[Bradbury et~al\mbox{.}(2018)]%
        {jax2018github}
\bibfield{author}{\bibinfo{person}{James Bradbury}, \bibinfo{person}{Roy
  Frostig}, \bibinfo{person}{Peter Hawkins}, \bibinfo{person}{Matthew~James
  Johnson}, \bibinfo{person}{Chris Leary}, \bibinfo{person}{Dougal Maclaurin},
  \bibinfo{person}{George Necula}, \bibinfo{person}{Adam Paszke},
  \bibinfo{person}{Jake Vander{P}las}, \bibinfo{person}{Skye
  Wanderman-{M}ilne}, {and} \bibinfo{person}{Qiao Zhang}.}
  \bibinfo{year}{2018}\natexlab{}.
\newblock \bibinfo{title}{{JAX}: composable transformations of
  {P}ython+{N}um{P}y programs}.
\newblock
\newblock
\urldef\tempurl%
\url{http://github.com/jax-ml/jax}
\showURL{%
\tempurl}


\bibitem[Brown et~al\mbox{.}(2020)]%
        {gpt}
\bibfield{author}{\bibinfo{person}{Tom~B. Brown}, \bibinfo{person}{Benjamin
  Mann}, \bibinfo{person}{Nick Ryder}, \bibinfo{person}{Melanie Subbiah},
  \bibinfo{person}{Jared Kaplan}, \bibinfo{person}{Prafulla Dhariwal},
  \bibinfo{person}{Arvind Neelakantan}, \bibinfo{person}{Pranav Shyam},
  \bibinfo{person}{Girish Sastry}, \bibinfo{person}{Amanda Askell},
  \bibinfo{person}{Sandhini Agarwal}, \bibinfo{person}{Ariel Herbert{-}Voss},
  \bibinfo{person}{Gretchen Krueger}, \bibinfo{person}{Tom Henighan},
  \bibinfo{person}{Rewon Child}, \bibinfo{person}{Aditya Ramesh},
  \bibinfo{person}{Daniel~M. Ziegler}, \bibinfo{person}{Jeffrey Wu},
  \bibinfo{person}{Clemens Winter}, \bibinfo{person}{Christopher Hesse},
  \bibinfo{person}{Mark Chen}, \bibinfo{person}{Eric Sigler},
  \bibinfo{person}{Mateusz Litwin}, \bibinfo{person}{Scott Gray},
  \bibinfo{person}{Benjamin Chess}, \bibinfo{person}{Jack Clark},
  \bibinfo{person}{Christopher Berner}, \bibinfo{person}{Sam McCandlish},
  \bibinfo{person}{Alec Radford}, \bibinfo{person}{Ilya Sutskever}, {and}
  \bibinfo{person}{Dario Amodei}.} \bibinfo{year}{2020}\natexlab{}.
\newblock \showarticletitle{Language Models are Few-Shot Learners}.
\newblock \bibinfo{journal}{\emph{CoRR}}  \bibinfo{volume}{abs/2005.14165}
  (\bibinfo{year}{2020}).
\newblock
\showeprint[arXiv]{2005.14165}
\urldef\tempurl%
\url{https://arxiv.org/abs/2005.14165}
\showURL{%
\tempurl}


\bibitem[Bubeck et~al\mbox{.}(2023)]%
        {bubeck2023sparks}
\bibfield{author}{\bibinfo{person}{S{\'{e}}bastien Bubeck},
  \bibinfo{person}{Varun Chandrasekaran}, \bibinfo{person}{Ronen Eldan},
  \bibinfo{person}{Johannes Gehrke}, \bibinfo{person}{Eric Horvitz},
  \bibinfo{person}{Ece Kamar}, \bibinfo{person}{Peter Lee},
  \bibinfo{person}{Yin~Tat Lee}, \bibinfo{person}{Yuanzhi Li},
  \bibinfo{person}{Scott~M. Lundberg}, \bibinfo{person}{Harsha Nori},
  \bibinfo{person}{Hamid Palangi}, \bibinfo{person}{Marco~T{\'{u}}lio Ribeiro},
  {and} \bibinfo{person}{Yi Zhang}.} \bibinfo{year}{2023}\natexlab{}.
\newblock \showarticletitle{Sparks of Artificial General Intelligence: Early
  experiments with {GPT-4}}.
\newblock \bibinfo{journal}{\emph{CoRR}}  \bibinfo{volume}{abs/2303.12712}
  (\bibinfo{year}{2023}).
\newblock
\urldef\tempurl%
\url{https://doi.org/10.48550/ARXIV.2303.12712}
\showDOI{\tempurl}
\showeprint[arXiv]{2303.12712}


\bibitem[Burlachenko et~al\mbox{.}(2023)]%
        {burlachenko2023federated}
\bibfield{author}{\bibinfo{person}{Konstantin Burlachenko},
  \bibinfo{person}{Abdulmajeed Alrowithi}, \bibinfo{person}{Fahad~Ali
  Albalawi}, {and} \bibinfo{person}{Peter Richt{\'{a}}rik}.}
  \bibinfo{year}{2023}\natexlab{}.
\newblock \showarticletitle{Federated Learning is Better with Non-Homomorphic
  Encryption}. In \bibinfo{booktitle}{\emph{Proceedings of the 4th
  International Workshop on Distributed Machine Learning, DistributedML 2023,
  Paris, France, 8 December 2023}}, \bibfield{editor}{\bibinfo{person}{Stefanos
  Laskaridis}, \bibinfo{person}{Alexey Tumanov}, \bibinfo{person}{Nathalie
  Baracaldo}, {and} \bibinfo{person}{Dimitrios Vytiniotis}} (Eds.).
  \bibinfo{publisher}{{ACM}}, \bibinfo{pages}{49--84}.
\newblock
\urldef\tempurl%
\url{https://doi.org/10.1145/3630048.3630182}
\showDOI{\tempurl}


\bibitem[Burlachenko et~al\mbox{.}(2021)]%
        {burlachenko2021fl_pytorch}
\bibfield{author}{\bibinfo{person}{Konstantin Burlachenko},
  \bibinfo{person}{Samuel Horv{\'{a}}th}, {and} \bibinfo{person}{Peter
  Richt{\'{a}}rik}.} \bibinfo{year}{2021}\natexlab{}.
\newblock \showarticletitle{FL{\_}PyTorch: optimization research simulator for
  federated learning}. In \bibinfo{booktitle}{\emph{DistributedML '21:
  Proceedings of the 2nd {ACM} International Workshop on Distributed Machine
  Learning, Virtual Event / Munich, Germany, 7 December 2021}}.
  \bibinfo{publisher}{{ACM}}, \bibinfo{pages}{1--7}.
\newblock
\urldef\tempurl%
\url{https://doi.org/10.1145/3488659.3493775}
\showDOI{\tempurl}


\bibitem[Burlachenko and Richt{\'{a}}rik(2024)]%
        {burlachenko2024unlocking}
\bibfield{author}{\bibinfo{person}{Konstantin Burlachenko} {and}
  \bibinfo{person}{Peter Richt{\'{a}}rik}.} \bibinfo{year}{2024}\natexlab{}.
\newblock \showarticletitle{Unlocking FedNL: Self-Contained Compute-Optimized
  Implementation}.
\newblock \bibinfo{journal}{\emph{CoRR}}  \bibinfo{volume}{abs/2410.08760}
  (\bibinfo{year}{2024}).
\newblock
\urldef\tempurl%
\url{https://doi.org/10.48550/ARXIV.2410.08760}
\showDOI{\tempurl}
\showeprint[arXiv]{2410.08760}


\bibitem[Burlachenko and Richtárik(2025)]%
        {burlachenko2025burtorch}
\bibfield{author}{\bibinfo{person}{Konstantin Burlachenko} {and}
  \bibinfo{person}{Peter Richtárik}.} \bibinfo{year}{2025}\natexlab{}.
\newblock \bibinfo{title}{BurTorch: Revisiting Training from First Principles
  by Coupling Autodiff, Math Optimization, and Systems}.
\newblock
\newblock
\showeprint[arxiv]{2503.13795}~[cs.LG]
\urldef\tempurl%
\url{https://arxiv.org/abs/2503.13795}
\showURL{%
\tempurl}


\bibitem[Caldas et~al\mbox{.}(2018)]%
        {caldas2018leaf}
\bibfield{author}{\bibinfo{person}{Sebastian Caldas}, \bibinfo{person}{Peter
  Wu}, \bibinfo{person}{Tian Li}, \bibinfo{person}{Jakub Kone{\v{c}}n{\'y}},
  \bibinfo{person}{H.~Brendan McMahan}, \bibinfo{person}{Virginia Smith}, {and}
  \bibinfo{person}{Ameet Talwalkar}.} \bibinfo{year}{2018}\natexlab{}.
\newblock \showarticletitle{{LEAF:} {A} Benchmark for Federated Settings}.
\newblock \bibinfo{journal}{\emph{CoRR}}  \bibinfo{volume}{abs/1812.01097}
  (\bibinfo{year}{2018}).
\newblock
\showeprint[arXiv]{1812.01097}
\urldef\tempurl%
\url{http://arxiv.org/abs/1812.01097}
\showURL{%
\tempurl}


\bibitem[Carlini et~al\mbox{.}(2019)]%
        {carlini2019secret}
\bibfield{author}{\bibinfo{person}{Nicholas Carlini}, \bibinfo{person}{Chang
  Liu}, \bibinfo{person}{{\'U}lfar Erlingsson}, \bibinfo{person}{Jernej Kos},
  {and} \bibinfo{person}{Dawn Song}.} \bibinfo{year}{2019}\natexlab{}.
\newblock \showarticletitle{The secret sharer: Evaluating and testing
  unintended memorization in neural networks}. In
  \bibinfo{booktitle}{\emph{28th USENIX security symposium (USENIX security
  19)}}. \bibinfo{pages}{267--284}.
\newblock


\bibitem[Chang and Lin(2011a)]%
        {chang2011libsvm}
\bibfield{author}{\bibinfo{person}{Chih{-}Chung Chang} {and}
  \bibinfo{person}{Chih{-}Jen Lin}.} \bibinfo{year}{2011}\natexlab{a}.
\newblock \showarticletitle{{LIBSVM:} {A} library for support vector machines}.
\newblock \bibinfo{journal}{\emph{{ACM} Trans. Intell. Syst. Technol.}}
  \bibinfo{volume}{2}, \bibinfo{number}{3} (\bibinfo{year}{2011}),
  \bibinfo{pages}{27:1--27:27}.
\newblock
\urldef\tempurl%
\url{https://doi.org/10.1145/1961189.1961199}
\showDOI{\tempurl}


\bibitem[Chang and Lin(2011b)]%
        {libsvm}
\bibfield{author}{\bibinfo{person}{Chih{-}Chung Chang} {and}
  \bibinfo{person}{Chih{-}Jen Lin}.} \bibinfo{year}{2011}\natexlab{b}.
\newblock \showarticletitle{{LIBSVM:} {A} library for support vector machines}.
\newblock \bibinfo{journal}{\emph{{ACM} Trans. Intell. Syst. Technol.}}
  \bibinfo{volume}{2}, \bibinfo{number}{3} (\bibinfo{year}{2011}),
  \bibinfo{pages}{27:1--27:27}.
\newblock
\urldef\tempurl%
\url{https://doi.org/10.1145/1961189.1961199}
\showDOI{\tempurl}


\bibitem[Charles and Kone{\v{c}}n{\'y}(2021)]%
        {charles2021convergence}
\bibfield{author}{\bibinfo{person}{Zachary Charles} {and}
  \bibinfo{person}{Jakub Kone{\v{c}}n{\'y}}.} \bibinfo{year}{2021}\natexlab{}.
\newblock \showarticletitle{Convergence and Accuracy Trade-Offs in Federated
  Learning and Meta-Learning}.
\newblock   \bibinfo{volume}{130} (\bibinfo{year}{2021}),
  \bibinfo{pages}{2575--2583}.
\newblock
\urldef\tempurl%
\url{http://proceedings.mlr.press/v130/charles21a.html}
\showURL{%
\tempurl}


\bibitem[Chen et~al\mbox{.}(2017)]%
        {chen2017simple}
\bibfield{author}{\bibinfo{person}{Hao Chen}, \bibinfo{person}{Kim Laine},
  {and} \bibinfo{person}{Rachel Player}.} \bibinfo{year}{2017}\natexlab{}.
\newblock \showarticletitle{Simple Encrypted Arithmetic Library - {SEAL} v2.1}.
  In \bibinfo{booktitle}{\emph{Financial Cryptography and Data Security - {FC}
  2017 International Workshops, WAHC, BITCOIN, VOTING, WTSC, and TA, Sliema,
  Malta, April 7, 2017, Revised Selected Papers}}
  \emph{(\bibinfo{series}{Lecture Notes in Computer Science},
  Vol.~\bibinfo{volume}{10323})}, \bibfield{editor}{\bibinfo{person}{Michael
  Brenner}, \bibinfo{person}{Kurt Rohloff}, \bibinfo{person}{Joseph Bonneau},
  \bibinfo{person}{Andrew Miller}, \bibinfo{person}{Peter Y.~A. Ryan},
  \bibinfo{person}{Vanessa Teague}, \bibinfo{person}{Andrea Bracciali},
  \bibinfo{person}{Massimiliano Sala}, \bibinfo{person}{Federico Pintore},
  {and} \bibinfo{person}{Markus Jakobsson}} (Eds.).
  \bibinfo{publisher}{Springer}, \bibinfo{pages}{3--18}.
\newblock
\urldef\tempurl%
\url{https://doi.org/10.1007/978-3-319-70278-0\_1}
\showDOI{\tempurl}


\bibitem[Chen et~al\mbox{.}(2018a)]%
        {Chen2018LAGLA}
\bibfield{author}{\bibinfo{person}{Tianyi Chen}, \bibinfo{person}{Georgios~B.
  Giannakis}, \bibinfo{person}{Tao Sun}, {and} \bibinfo{person}{Wotao Yin}.}
  \bibinfo{year}{2018}\natexlab{a}.
\newblock \showarticletitle{{LAG:} Lazily Aggregated Gradient for
  Communication-Efficient Distributed Learning}. In
  \bibinfo{booktitle}{\emph{Advances in Neural Information Processing Systems
  31: Annual Conference on Neural Information Processing Systems 2018, NeurIPS
  2018, December 3-8, 2018, Montr{\'{e}}al, Canada}},
  \bibfield{editor}{\bibinfo{person}{Samy Bengio}, \bibinfo{person}{Hanna~M.
  Wallach}, \bibinfo{person}{Hugo Larochelle}, \bibinfo{person}{Kristen
  Grauman}, \bibinfo{person}{Nicol{\`{o}} Cesa{-}Bianchi}, {and}
  \bibinfo{person}{Roman Garnett}} (Eds.). \bibinfo{pages}{5055--5065}.
\newblock
\urldef\tempurl%
\url{https://proceedings.neurips.cc/paper/2018/hash/feecee9f1643651799ede2740927317a-Abstract.html}
\showURL{%
\tempurl}


\bibitem[Chen et~al\mbox{.}(2016)]%
        {chen2016training}
\bibfield{author}{\bibinfo{person}{Tianqi Chen}, \bibinfo{person}{Bing Xu},
  \bibinfo{person}{Chiyuan Zhang}, {and} \bibinfo{person}{Carlos Guestrin}.}
  \bibinfo{year}{2016}\natexlab{}.
\newblock \showarticletitle{Training Deep Nets with Sublinear Memory Cost}.
\newblock \bibinfo{journal}{\emph{CoRR}}  \bibinfo{volume}{abs/1604.06174}
  (\bibinfo{year}{2016}).
\newblock
\showeprint[arXiv]{1604.06174}
\urldef\tempurl%
\url{http://arxiv.org/abs/1604.06174}
\showURL{%
\tempurl}


\bibitem[Chen et~al\mbox{.}(2020)]%
        {chen2020understanding}
\bibfield{author}{\bibinfo{person}{Xiangyi Chen},
  \bibinfo{person}{Zhiwei~Steven Wu}, {and} \bibinfo{person}{Mingyi Hong}.}
  \bibinfo{year}{2020}\natexlab{}.
\newblock \showarticletitle{Understanding Gradient Clipping in Private {SGD:}
  {A} Geometric Perspective}.
\newblock  (\bibinfo{year}{2020}).
\newblock
\urldef\tempurl%
\url{https://proceedings.neurips.cc/paper/2020/hash/9ecff5455677b38d19f49ce658ef0608-Abstract.html}
\showURL{%
\tempurl}


\bibitem[Chen et~al\mbox{.}(2018b)]%
        {chen2018understanding}
\bibfield{author}{\bibinfo{person}{Y Chen}, \bibinfo{person}{Tien-Ju Yang},
  \bibinfo{person}{Joel Emer}, {and} \bibinfo{person}{Vivienne Sze}.}
  \bibinfo{year}{2018}\natexlab{b}.
\newblock \showarticletitle{Understanding the limitations of existing
  energy-efficient design approaches for deep neural networks}.
\newblock \bibinfo{journal}{\emph{Energy}} \bibinfo{volume}{2},
  \bibinfo{number}{L1} (\bibinfo{year}{2018}), \bibinfo{pages}{L3}.
\newblock


\bibitem[Cheon et~al\mbox{.}(2017)]%
        {cheon2017homomorphic}
\bibfield{author}{\bibinfo{person}{Jung~Hee Cheon}, \bibinfo{person}{Andrey
  Kim}, \bibinfo{person}{Miran Kim}, {and} \bibinfo{person}{Yong~Soo Song}.}
  \bibinfo{year}{2017}\natexlab{}.
\newblock \showarticletitle{Homomorphic Encryption for Arithmetic of
  Approximate Numbers}. In \bibinfo{booktitle}{\emph{Advances in Cryptology -
  {ASIACRYPT} 2017 - 23rd International Conference on the Theory and
  Applications of Cryptology and Information Security, Hong Kong, China,
  December 3-7, 2017, Proceedings, Part {I}}} \emph{(\bibinfo{series}{Lecture
  Notes in Computer Science}, Vol.~\bibinfo{volume}{10624})},
  \bibfield{editor}{\bibinfo{person}{Tsuyoshi Takagi} {and}
  \bibinfo{person}{Thomas Peyrin}} (Eds.). \bibinfo{publisher}{Springer},
  \bibinfo{pages}{409--437}.
\newblock
\urldef\tempurl%
\url{https://doi.org/10.1007/978-3-319-70694-8\_15}
\showDOI{\tempurl}


\bibitem[Chollet et~al\mbox{.}(2015)]%
        {chollet2015keras}
\bibfield{author}{\bibinfo{person}{Fran\c{c}ois Chollet} {et~al\mbox{.}}}
  \bibinfo{year}{2015}\natexlab{}.
\newblock \bibinfo{title}{Keras}.
\newblock \bibinfo{howpublished}{\url{https://keras.io}}.
\newblock


\bibitem[Chraibi et~al\mbox{.}(2019)]%
        {DFPMCI2019}
\bibfield{author}{\bibinfo{person}{S{\'{e}}lim Chraibi}, \bibinfo{person}{Ahmed
  Khaled}, \bibinfo{person}{Dmitry Kovalev}, \bibinfo{person}{Peter
  Richt{\'{a}}rik}, \bibinfo{person}{Adil Salim}, {and} \bibinfo{person}{Martin
  Tak{\'{a}}c}.} \bibinfo{year}{2019}\natexlab{}.
\newblock \showarticletitle{Distributed Fixed Point Methods with Compressed
  Iterates}.
\newblock \bibinfo{journal}{\emph{CoRR}}  \bibinfo{volume}{abs/1912.09925}
  (\bibinfo{year}{2019}).
\newblock
\showeprint[arXiv]{1912.09925}
\urldef\tempurl%
\url{http://arxiv.org/abs/1912.09925}
\showURL{%
\tempurl}


\bibitem[Cohen and Nissim(2018)]%
        {cohen2018linear}
\bibfield{author}{\bibinfo{person}{Aloni Cohen} {and} \bibinfo{person}{Kobbi
  Nissim}.} \bibinfo{year}{2018}\natexlab{}.
\newblock \showarticletitle{Linear Program Reconstruction in Practice}.
\newblock \bibinfo{journal}{\emph{CoRR}}  \bibinfo{volume}{abs/1810.05692}
  (\bibinfo{year}{2018}).
\newblock
\showeprint[arXiv]{1810.05692}
\urldef\tempurl%
\url{http://arxiv.org/abs/1810.05692}
\showURL{%
\tempurl}


\bibitem[Collobert et~al\mbox{.}(2002)]%
        {collobert2002torch}
\bibfield{author}{\bibinfo{person}{Ronan Collobert}, \bibinfo{person}{Samy
  Bengio}, {and} \bibinfo{person}{Johnny Mari{\'e}thoz}.}
  \bibinfo{year}{2002}\natexlab{}.
\newblock \showarticletitle{Torch: a modular machine learning software
  library}.
\newblock  (\bibinfo{year}{2002}).
\newblock


\bibitem[Condat et~al\mbox{.}(2023)]%
        {condat2023tamuna}
\bibfield{author}{\bibinfo{person}{Laurent Condat}, \bibinfo{person}{Grigory
  Malinovsky}, {and} \bibinfo{person}{Peter Richt{\'{a}}rik}.}
  \bibinfo{year}{2023}\natexlab{}.
\newblock \showarticletitle{{TAMUNA:} Accelerated Federated Learning with Local
  Training and Partial Participation}.
\newblock \bibinfo{journal}{\emph{CoRR}}  \bibinfo{volume}{abs/2302.09832}
  (\bibinfo{year}{2023}).
\newblock
\urldef\tempurl%
\url{https://doi.org/10.48550/ARXIV.2302.09832}
\showDOI{\tempurl}
\showeprint[arXiv]{2302.09832}


\bibitem[Condat et~al\mbox{.}(2022)]%
        {EF-BV}
\bibfield{author}{\bibinfo{person}{Laurent Condat}, \bibinfo{person}{Kai Yi},
  {and} \bibinfo{person}{Peter Richt{\'{a}}rik}.}
  \bibinfo{year}{2022}\natexlab{}.
\newblock \showarticletitle{{EF-BV:} {A} Unified Theory of Error Feedback and
  Variance Reduction Mechanisms for Biased and Unbiased Compression in
  Distributed Optimization}. In \bibinfo{booktitle}{\emph{Advances in Neural
  Information Processing Systems 35: Annual Conference on Neural Information
  Processing Systems 2022, NeurIPS 2022, New Orleans, LA, USA, November 28 -
  December 9, 2022}}, \bibfield{editor}{\bibinfo{person}{Sanmi Koyejo},
  \bibinfo{person}{S.~Mohamed}, \bibinfo{person}{A.~Agarwal},
  \bibinfo{person}{Danielle Belgrave}, \bibinfo{person}{K.~Cho}, {and}
  \bibinfo{person}{A.~Oh}} (Eds.).
\newblock
\urldef\tempurl%
\url{http://papers.nips.cc/paper\_files/paper/2022/hash/6fb9ea5197c0b8ece8a64220fb82cdfe-Abstract-Conference.html}
\showURL{%
\tempurl}


\bibitem[Cormen et~al\mbox{.}(2022)]%
        {cormen2022introduction}
\bibfield{author}{\bibinfo{person}{Thomas~H Cormen}, \bibinfo{person}{Charles~E
  Leiserson}, \bibinfo{person}{Ronald~L Rivest}, {and}
  \bibinfo{person}{Clifford Stein}.} \bibinfo{year}{2022}\natexlab{}.
\newblock \bibinfo{booktitle}{\emph{Introduction to algorithms}}.
\newblock \bibinfo{publisher}{MIT press}.
\newblock


\bibitem[Costan and Devadas(2016)]%
        {costan2016intel}
\bibfield{author}{\bibinfo{person}{Victor Costan} {and}
  \bibinfo{person}{Srinivas Devadas}.} \bibinfo{year}{2016}\natexlab{}.
\newblock \showarticletitle{Intel {SGX} Explained}.
\newblock \bibinfo{journal}{\emph{{IACR} Cryptol. ePrint Arch.}}
  (\bibinfo{year}{2016}), \bibinfo{pages}{86}.
\newblock
\urldef\tempurl%
\url{http://eprint.iacr.org/2016/086}
\showURL{%
\tempurl}


\bibitem[Daemen and Rijmen(1999)]%
        {daemen1999aes}
\bibfield{author}{\bibinfo{person}{Joan Daemen} {and} \bibinfo{person}{Vincent
  Rijmen}.} \bibinfo{year}{1999}\natexlab{}.
\newblock \showarticletitle{AES proposal: Rijndael}.
\newblock \bibinfo{journal}{\emph{NIST AES Proposal}} (\bibinfo{year}{1999}).
\newblock


\bibitem[Daemen and Rijmen(2001)]%
        {daemen2001reijndael}
\bibfield{author}{\bibinfo{person}{Joan Daemen} {and} \bibinfo{person}{Vincent
  Rijmen}.} \bibinfo{year}{2001}\natexlab{}.
\newblock \showarticletitle{Reijndael: The advanced encryption standard.}
\newblock \bibinfo{journal}{\emph{Dr. Dobb's Journal: Software Tools for the
  Professional Programmer}} \bibinfo{volume}{26}, \bibinfo{number}{3}
  (\bibinfo{year}{2001}), \bibinfo{pages}{137--139}.
\newblock


\bibitem[Dai et~al\mbox{.}(2022)]%
        {dai2022dispfl}
\bibfield{author}{\bibinfo{person}{Rong Dai}, \bibinfo{person}{Li Shen},
  \bibinfo{person}{Fengxiang He}, \bibinfo{person}{Xinmei Tian}, {and}
  \bibinfo{person}{Dacheng Tao}.} \bibinfo{year}{2022}\natexlab{}.
\newblock \showarticletitle{DisPFL: Towards Communication-Efficient
  Personalized Federated Learning via Decentralized Sparse Training}. In
  \bibinfo{booktitle}{\emph{International Conference on Machine Learning,
  {ICML} 2022, 17-23 July 2022, Baltimore, Maryland, {USA}}}
  \emph{(\bibinfo{series}{Proceedings of Machine Learning Research},
  Vol.~\bibinfo{volume}{162})}, \bibfield{editor}{\bibinfo{person}{Kamalika
  Chaudhuri}, \bibinfo{person}{Stefanie Jegelka}, \bibinfo{person}{Le~Song},
  \bibinfo{person}{Csaba Szepesv{\'{a}}ri}, \bibinfo{person}{Gang Niu}, {and}
  \bibinfo{person}{Sivan Sabato}} (Eds.). \bibinfo{publisher}{{PMLR}},
  \bibinfo{pages}{4587--4604}.
\newblock
\urldef\tempurl%
\url{https://proceedings.mlr.press/v162/dai22b.html}
\showURL{%
\tempurl}


\bibitem[Danowitz et~al\mbox{.}(2012)]%
        {10.1145/2133806.2133822}
\bibfield{author}{\bibinfo{person}{Andrew Danowitz}, \bibinfo{person}{Kyle
  Kelley}, \bibinfo{person}{James Mao}, \bibinfo{person}{John~P. Stevenson},
  {and} \bibinfo{person}{Mark Horowitz}.} \bibinfo{year}{2012}\natexlab{}.
\newblock \showarticletitle{{CPU} {DB:} recording microprocessor history}.
\newblock \bibinfo{journal}{\emph{Commun. {ACM}}} \bibinfo{volume}{55},
  \bibinfo{number}{4} (\bibinfo{year}{2012}), \bibinfo{pages}{55--63}.
\newblock
\urldef\tempurl%
\url{https://doi.org/10.1145/2133806.2133822}
\showDOI{\tempurl}


\bibitem[d'Aspremont et~al\mbox{.}(2021)]%
        {AccMethodsBook}
\bibfield{author}{\bibinfo{person}{Alexandre d'Aspremont},
  \bibinfo{person}{Damien Scieur}, {and} \bibinfo{person}{Adrien~B. Taylor}.}
  \bibinfo{year}{2021}\natexlab{}.
\newblock \showarticletitle{Acceleration Methods}.
\newblock \bibinfo{journal}{\emph{Found. Trends Optim.}} \bibinfo{volume}{5},
  \bibinfo{number}{1-2} (\bibinfo{year}{2021}), \bibinfo{pages}{1--245}.
\newblock
\urldef\tempurl%
\url{https://doi.org/10.1561/2400000036}
\showDOI{\tempurl}


\bibitem[Dean et~al\mbox{.}(2012)]%
        {dean2012large}
\bibfield{author}{\bibinfo{person}{Jeffrey Dean}, \bibinfo{person}{Greg
  Corrado}, \bibinfo{person}{Rajat Monga}, \bibinfo{person}{Kai Chen},
  \bibinfo{person}{Matthieu Devin}, \bibinfo{person}{Quoc~V. Le},
  \bibinfo{person}{Mark~Z. Mao}, \bibinfo{person}{Marc'Aurelio Ranzato},
  \bibinfo{person}{Andrew~W. Senior}, \bibinfo{person}{Paul~A. Tucker},
  \bibinfo{person}{Ke Yang}, {and} \bibinfo{person}{Andrew~Y. Ng}.}
  \bibinfo{year}{2012}\natexlab{}.
\newblock \showarticletitle{Large Scale Distributed Deep Networks}.
\newblock  (\bibinfo{year}{2012}), \bibinfo{pages}{1232--1240}.
\newblock
\urldef\tempurl%
\url{https://proceedings.neurips.cc/paper/2012/hash/6aca97005c68f1206823815f66102863-Abstract.html}
\showURL{%
\tempurl}


\bibitem[Defazio et~al\mbox{.}(2014)]%
        {SAGA}
\bibfield{author}{\bibinfo{person}{Aaron Defazio}, \bibinfo{person}{Francis~R.
  Bach}, {and} \bibinfo{person}{Simon Lacoste{-}Julien}.}
  \bibinfo{year}{2014}\natexlab{}.
\newblock \showarticletitle{{SAGA:} {A} Fast Incremental Gradient Method With
  Support for Non-Strongly Convex Composite Objectives}.
\newblock  (\bibinfo{year}{2014}), \bibinfo{pages}{1646--1654}.
\newblock
\urldef\tempurl%
\url{https://proceedings.neurips.cc/paper/2014/hash/ede7e2b6d13a41ddf9f4bdef84fdc737-Abstract.html}
\showURL{%
\tempurl}


\bibitem[Demaine(2002)]%
        {demaine2002cache}
\bibfield{author}{\bibinfo{person}{Erik~D Demaine}.}
  \bibinfo{year}{2002}\natexlab{}.
\newblock \showarticletitle{Cache-oblivious algorithms and data structures}.
\newblock \bibinfo{journal}{\emph{Lecture Notes from the EEF Summer School on
  Massive Data Sets}} \bibinfo{volume}{8}, \bibinfo{number}{4}
  (\bibinfo{year}{2002}), \bibinfo{pages}{1--249}.
\newblock


\bibitem[Deng et~al\mbox{.}(2020)]%
        {deng2020adaptive}
\bibfield{author}{\bibinfo{person}{Yuyang Deng},
  \bibinfo{person}{Mohammad~Mahdi Kamani}, {and} \bibinfo{person}{Mehrdad
  Mahdavi}.} \bibinfo{year}{2020}\natexlab{}.
\newblock \showarticletitle{Adaptive Personalized Federated Learning}.
\newblock \bibinfo{journal}{\emph{CoRR}}  \bibinfo{volume}{abs/2003.13461}
  (\bibinfo{year}{2020}).
\newblock
\showeprint[arXiv]{2003.13461}
\urldef\tempurl%
\url{https://arxiv.org/abs/2003.13461}
\showURL{%
\tempurl}


\bibitem[Diamond and Boyd(2016)]%
        {diamond2016cvxpy}
\bibfield{author}{\bibinfo{person}{Steven Diamond} {and}
  \bibinfo{person}{Stephen~P. Boyd}.} \bibinfo{year}{2016}\natexlab{}.
\newblock \showarticletitle{{CVXPY:} {A} Python-Embedded Modeling Language for
  Convex Optimization}.
\newblock \bibinfo{journal}{\emph{J. Mach. Learn. Res.}}  \bibinfo{volume}{17}
  (\bibinfo{year}{2016}), \bibinfo{pages}{83:1--83:5}.
\newblock
\urldef\tempurl%
\url{https://jmlr.org/papers/v17/15-408.html}
\showURL{%
\tempurl}


\bibitem[Dimitriadis et~al\mbox{.}(2022)]%
        {garcia2022flute}
\bibfield{author}{\bibinfo{person}{Dimitrios Dimitriadis},
  \bibinfo{person}{Mirian~Hipolito Garcia}, \bibinfo{person}{Daniel~Madrigal
  Diaz}, \bibinfo{person}{Andre Manoel}, {and} \bibinfo{person}{Robert Sim}.}
  \bibinfo{year}{2022}\natexlab{}.
\newblock \showarticletitle{{FLUTE:} {A} Scalable, Extensible Framework for
  High-Performance Federated Learning Simulations}.
\newblock \bibinfo{journal}{\emph{CoRR}}  \bibinfo{volume}{abs/2203.13789}
  (\bibinfo{year}{2022}).
\newblock
\urldef\tempurl%
\url{https://doi.org/10.48550/ARXIV.2203.13789}
\showDOI{\tempurl}
\showeprint[arXiv]{2203.13789}


\bibitem[Ding et~al\mbox{.}(2022)]%
        {ding2022federated}
\bibfield{author}{\bibinfo{person}{Jie Ding}, \bibinfo{person}{Eric~W. Tramel},
  \bibinfo{person}{Anit~Kumar Sahu}, \bibinfo{person}{Shuang Wu},
  \bibinfo{person}{Salman Avestimehr}, {and} \bibinfo{person}{Tao Zhang}.}
  \bibinfo{year}{2022}\natexlab{}.
\newblock \showarticletitle{Federated Learning Challenges and Opportunities: An
  Outlook}.
\newblock  (\bibinfo{year}{2022}), \bibinfo{pages}{8752--8756}.
\newblock
\urldef\tempurl%
\url{https://doi.org/10.1109/ICASSP43922.2022.9746925}
\showDOI{\tempurl}


\bibitem[Dinur and Nissim(2003)]%
        {dinur2003revealing}
\bibfield{author}{\bibinfo{person}{Irit Dinur} {and} \bibinfo{person}{Kobbi
  Nissim}.} \bibinfo{year}{2003}\natexlab{}.
\newblock \showarticletitle{Revealing information while preserving privacy}. In
  \bibinfo{booktitle}{\emph{Proceedings of the Twenty-Second {ACM}
  {SIGACT-SIGMOD-SIGART} Symposium on Principles of Database Systems, June
  9-12, 2003, San Diego, CA, {USA}}}, \bibfield{editor}{\bibinfo{person}{Frank
  Neven}, \bibinfo{person}{Catriel Beeri}, {and} \bibinfo{person}{Tova Milo}}
  (Eds.). \bibinfo{publisher}{{ACM}}, \bibinfo{pages}{202--210}.
\newblock
\urldef\tempurl%
\url{https://doi.org/10.1145/773153.773173}
\showDOI{\tempurl}


\bibitem[Doe(2023)]%
        {interviewboyd}
\bibfield{author}{\bibinfo{person}{John Doe}.} \bibinfo{year}{2023}\natexlab{}.
\newblock \bibinfo{title}{InControl podcast: Interview with Stephen Boyd}.
\newblock \bibinfo{howpublished}{Interview by A. Padoan}.
\newblock
\urldef\tempurl%
\url{https://stanford.edu/~boyd/papers/incontrol.html}
\showURL{%
\tempurl}


\bibitem[Domahidi et~al\mbox{.}(2013)]%
        {domahidi2013ecos}
\bibfield{author}{\bibinfo{person}{Alexander Domahidi}, \bibinfo{person}{Eric
  Chu}, {and} \bibinfo{person}{Stephen~P. Boyd}.}
  \bibinfo{year}{2013}\natexlab{}.
\newblock \showarticletitle{{ECOS:} An {SOCP} solver for embedded systems}. In
  \bibinfo{booktitle}{\emph{12th European Control Conference, {ECC} 2013,
  Zurich, Switzerland, July 17-19, 2013}}. \bibinfo{publisher}{{IEEE}},
  \bibinfo{pages}{3071--3076}.
\newblock
\urldef\tempurl%
\url{https://doi.org/10.23919/ECC.2013.6669541}
\showDOI{\tempurl}


\bibitem[Dutta et~al\mbox{.}(2020)]%
        {layer-wise}
\bibfield{author}{\bibinfo{person}{Aritra Dutta}, \bibinfo{person}{El~Houcine
  Bergou}, \bibinfo{person}{Ahmed~M. Abdelmoniem}, \bibinfo{person}{Chen{-}Yu
  Ho}, \bibinfo{person}{Atal~Narayan Sahu}, \bibinfo{person}{Marco Canini},
  {and} \bibinfo{person}{Panos Kalnis}.} \bibinfo{year}{2020}\natexlab{}.
\newblock \showarticletitle{On the Discrepancy between the Theoretical Analysis
  and Practical Implementations of Compressed Communication for Distributed
  Deep Learning}. In \bibinfo{booktitle}{\emph{The Thirty-Fourth {AAAI}
  Conference on Artificial Intelligence, {AAAI} 2020, The Thirty-Second
  Innovative Applications of Artificial Intelligence Conference, {IAAI} 2020,
  The Tenth {AAAI} Symposium on Educational Advances in Artificial
  Intelligence, {EAAI} 2020, New York, NY, USA, February 7-12, 2020}}.
  \bibinfo{publisher}{{AAAI} Press}, \bibinfo{pages}{3817--3824}.
\newblock
\urldef\tempurl%
\url{https://doi.org/10.1609/AAAI.V34I04.5793}
\showDOI{\tempurl}


\bibitem[Dwork(2008)]%
        {dwork2008differential}
\bibfield{author}{\bibinfo{person}{Cynthia Dwork}.}
  \bibinfo{year}{2008}\natexlab{}.
\newblock \showarticletitle{Differential Privacy: {A} Survey of Results}. In
  \bibinfo{booktitle}{\emph{Theory and Applications of Models of Computation,
  5th International Conference, {TAMC} 2008, Xi'an, China, April 25-29, 2008.
  Proceedings}} \emph{(\bibinfo{series}{Lecture Notes in Computer Science},
  Vol.~\bibinfo{volume}{4978})}, \bibfield{editor}{\bibinfo{person}{Manindra
  Agrawal}, \bibinfo{person}{Ding{-}Zhu Du}, \bibinfo{person}{Zhenhua Duan},
  {and} \bibinfo{person}{Angsheng Li}} (Eds.). \bibinfo{publisher}{Springer},
  \bibinfo{pages}{1--19}.
\newblock
\urldef\tempurl%
\url{https://doi.org/10.1007/978-3-540-79228-4\_1}
\showDOI{\tempurl}


\bibitem[Dwork et~al\mbox{.}(2006)]%
        {dwork2006our}
\bibfield{author}{\bibinfo{person}{Cynthia Dwork}, \bibinfo{person}{Krishnaram
  Kenthapadi}, \bibinfo{person}{Frank McSherry}, \bibinfo{person}{Ilya
  Mironov}, {and} \bibinfo{person}{Moni Naor}.}
  \bibinfo{year}{2006}\natexlab{}.
\newblock \showarticletitle{Our Data, Ourselves: Privacy Via Distributed Noise
  Generation}. In \bibinfo{booktitle}{\emph{Advances in Cryptology -
  {EUROCRYPT} 2006, 25th Annual International Conference on the Theory and
  Applications of Cryptographic Techniques, St. Petersburg, Russia, May 28 -
  June 1, 2006, Proceedings}} \emph{(\bibinfo{series}{Lecture Notes in Computer
  Science}, Vol.~\bibinfo{volume}{4004})},
  \bibfield{editor}{\bibinfo{person}{Serge Vaudenay}} (Ed.).
  \bibinfo{publisher}{Springer}, \bibinfo{pages}{486--503}.
\newblock
\urldef\tempurl%
\url{https://doi.org/10.1007/11761679\_29}
\showDOI{\tempurl}


\bibitem[Dwork et~al\mbox{.}(2016)]%
        {dwork2006calibrating}
\bibfield{author}{\bibinfo{person}{Cynthia Dwork}, \bibinfo{person}{Frank
  McSherry}, \bibinfo{person}{Kobbi Nissim}, {and} \bibinfo{person}{Adam~D.
  Smith}.} \bibinfo{year}{2016}\natexlab{}.
\newblock \showarticletitle{Calibrating Noise to Sensitivity in Private Data
  Analysis}.
\newblock \bibinfo{journal}{\emph{J. Priv. Confidentiality}}
  \bibinfo{volume}{7}, \bibinfo{number}{3}, \bibinfo{pages}{17--51}.
\newblock
\urldef\tempurl%
\url{https://doi.org/10.29012/JPC.V7I3.405}
\showDOI{\tempurl}


\bibitem[Eijs(2014)]%
        {pycryptodome}
\bibfield{author}{\bibinfo{person}{Helder Eijs}.}
  \bibinfo{year}{2014}\natexlab{}.
\newblock \bibinfo{title}{PyCryptodome}.
\newblock
\newblock
\urldef\tempurl%
\url{https://pypi.org/project/pycryptodome/}
\showURL{%
\tempurl}
\newblock
\shownote{Accessed: 2023-05-10}.


\bibitem[Ellson et~al\mbox{.}(2001)]%
        {ellson2002graphviz}
\bibfield{author}{\bibinfo{person}{John Ellson}, \bibinfo{person}{Emden~R.
  Gansner}, \bibinfo{person}{Eleftherios Koutsofios},
  \bibinfo{person}{Stephen~C. North}, {and} \bibinfo{person}{Gordon Woodhull}.}
  \bibinfo{year}{2001}\natexlab{}.
\newblock \showarticletitle{Graphviz - Open Source Graph Drawing Tools}. In
  \bibinfo{booktitle}{\emph{Graph Drawing, 9th International Symposium, {GD}
  2001 Vienna, Austria, September 23-26, 2001, Revised Papers}}
  \emph{(\bibinfo{series}{Lecture Notes in Computer Science},
  Vol.~\bibinfo{volume}{2265})}, \bibfield{editor}{\bibinfo{person}{Petra
  Mutzel}, \bibinfo{person}{Michael J{\"{u}}nger}, {and}
  \bibinfo{person}{Sebastian Leipert}} (Eds.). \bibinfo{publisher}{Springer},
  \bibinfo{pages}{483--484}.
\newblock
\urldef\tempurl%
\url{https://doi.org/10.1007/3-540-45848-4\_57}
\showDOI{\tempurl}


\bibitem[{European Commission}(2018)]%
        {gdpr}
\bibfield{author}{\bibinfo{person}{{European Commission}}.}
  \bibinfo{year}{2018}\natexlab{}.
\newblock \showarticletitle{{GDPR: 2018 Reform of EU Data Protection Rules}}.
\newblock
\urldef\tempurl%
\url{https://ec.europa.eu/commission/sites/beta-political/files/data-protection-factsheet-changes_en.pdf}
\showURL{%
\tempurl}


\bibitem[Fallah et~al\mbox{.}(2020)]%
        {fallah2020personalized}
\bibfield{author}{\bibinfo{person}{Alireza Fallah}, \bibinfo{person}{Aryan
  Mokhtari}, {and} \bibinfo{person}{Asuman~E. Ozdaglar}.}
  \bibinfo{year}{2020}\natexlab{}.
\newblock \showarticletitle{Personalized Federated Learning with Theoretical
  Guarantees: {A} Model-Agnostic Meta-Learning Approach}.
\newblock  (\bibinfo{year}{2020}).
\newblock
\urldef\tempurl%
\url{https://proceedings.neurips.cc/paper/2020/hash/24389bfe4fe2eba8bf9aa9203a44cdad-Abstract.html}
\showURL{%
\tempurl}


\bibitem[Fan and Vercauteren(2012)]%
        {fan2012somewhat}
\bibfield{author}{\bibinfo{person}{Junfeng Fan} {and} \bibinfo{person}{Frederik
  Vercauteren}.} \bibinfo{year}{2012}\natexlab{}.
\newblock \showarticletitle{Somewhat Practical Fully Homomorphic Encryption}.
\newblock \bibinfo{journal}{\emph{{IACR} Cryptol. ePrint Arch.}}
  (\bibinfo{year}{2012}), \bibinfo{pages}{144}.
\newblock
\urldef\tempurl%
\url{http://eprint.iacr.org/2012/144}
\showURL{%
\tempurl}


\bibitem[Fang et~al\mbox{.}(2018)]%
        {fang2018spider}
\bibfield{author}{\bibinfo{person}{Cong Fang}, \bibinfo{person}{Chris~Junchi
  Li}, \bibinfo{person}{Zhouchen Lin}, {and} \bibinfo{person}{Tong Zhang}.}
  \bibinfo{year}{2018}\natexlab{}.
\newblock \showarticletitle{{SPIDER:} Near-Optimal Non-Convex Optimization via
  Stochastic Path-Integrated Differential Estimator}.
\newblock  (\bibinfo{year}{2018}), \bibinfo{pages}{687--697}.
\newblock
\urldef\tempurl%
\url{https://proceedings.neurips.cc/paper/2018/hash/1543843a4723ed2ab08e18053ae6dc5b-Abstract.html}
\showURL{%
\tempurl}


\bibitem[Fatkhullin et~al\mbox{.}(2021)]%
        {fatkhullin2021ef21}
\bibfield{author}{\bibinfo{person}{Ilyas Fatkhullin}, \bibinfo{person}{Igor
  Sokolov}, \bibinfo{person}{Eduard Gorbunov}, \bibinfo{person}{Zhize Li},
  {and} \bibinfo{person}{Peter Richt{\'{a}}rik}.}
  \bibinfo{year}{2021}\natexlab{}.
\newblock \showarticletitle{{EF21} with Bells {\&} Whistles: Practical
  Algorithmic Extensions of Modern Error Feedback}.
\newblock \bibinfo{journal}{\emph{CoRR}}  \bibinfo{volume}{abs/2110.03294}
  (\bibinfo{year}{2021}).
\newblock
\showeprint[arXiv]{2110.03294}
\urldef\tempurl%
\url{https://arxiv.org/abs/2110.03294}
\showURL{%
\tempurl}


\bibitem[Fatkhullin et~al\mbox{.}(2023)]%
        {EF21M}
\bibfield{author}{\bibinfo{person}{Ilyas Fatkhullin},
  \bibinfo{person}{Alexander Tyurin}, {and} \bibinfo{person}{Peter
  Richt{\'{a}}rik}.} \bibinfo{year}{2023}\natexlab{}.
\newblock \showarticletitle{Momentum Provably Improves Error Feedback!}. In
  \bibinfo{booktitle}{\emph{Advances in Neural Information Processing Systems
  36: Annual Conference on Neural Information Processing Systems 2023, NeurIPS
  2023, New Orleans, LA, USA, December 10 - 16, 2023}},
  \bibfield{editor}{\bibinfo{person}{Alice Oh}, \bibinfo{person}{Tristan
  Naumann}, \bibinfo{person}{Amir Globerson}, \bibinfo{person}{Kate Saenko},
  \bibinfo{person}{Moritz Hardt}, {and} \bibinfo{person}{Sergey Levine}}
  (Eds.).
\newblock
\urldef\tempurl%
\url{http://papers.nips.cc/paper\_files/paper/2023/hash/f0b1515be276f6ba82b4f2b25e50bef0-Abstract-Conference.html}
\showURL{%
\tempurl}


\bibitem[Fauzi et~al\mbox{.}(2022)]%
        {fauzi2022ind}
\bibfield{author}{\bibinfo{person}{Prastudy Fauzi},
  \bibinfo{person}{Martha~Norberg Hovd}, {and} \bibinfo{person}{H{\aa}vard
  Raddum}.} \bibinfo{year}{2022}\natexlab{}.
\newblock \showarticletitle{On the {IND-CCA1} Security of {FHE} Schemes}.
\newblock \bibinfo{journal}{\emph{Cryptogr.}} \bibinfo{volume}{6},
  \bibinfo{number}{1} (\bibinfo{year}{2022}), \bibinfo{pages}{13}.
\newblock
\urldef\tempurl%
\url{https://doi.org/10.3390/CRYPTOGRAPHY6010013}
\showDOI{\tempurl}


\bibitem[Fawzi et~al\mbox{.}(2022)]%
        {fawzi2022discovering}
\bibfield{author}{\bibinfo{person}{Alhussein Fawzi}, \bibinfo{person}{Matej
  Balog}, \bibinfo{person}{Aja Huang}, \bibinfo{person}{Thomas Hubert},
  \bibinfo{person}{Bernardino Romera{-}Paredes}, \bibinfo{person}{Mohammadamin
  Barekatain}, \bibinfo{person}{Alexander Novikov}, \bibinfo{person}{Francisco
  J.~R. Ruiz}, \bibinfo{person}{Julian Schrittwieser},
  \bibinfo{person}{Grzegorz Swirszcz}, \bibinfo{person}{David Silver},
  \bibinfo{person}{Demis Hassabis}, {and} \bibinfo{person}{Pushmeet Kohli}.}
  \bibinfo{year}{2022}\natexlab{}.
\newblock \showarticletitle{Discovering faster matrix multiplication algorithms
  with reinforcement learning}.
\newblock \bibinfo{journal}{\emph{Nat.}} \bibinfo{volume}{610},
  \bibinfo{number}{7930} (\bibinfo{year}{2022}), \bibinfo{pages}{47--53}.
\newblock
\urldef\tempurl%
\url{https://doi.org/10.1038/S41586-022-05172-4}
\showDOI{\tempurl}


\bibitem[Feist(2012)]%
        {feist2012vivado}
\bibfield{author}{\bibinfo{person}{Tom Feist}.}
  \bibinfo{year}{2012}\natexlab{}.
\newblock \showarticletitle{Vivado design suite}.
\newblock \bibinfo{journal}{\emph{White Paper}} \bibinfo{volume}{5},
  \bibinfo{number}{30} (\bibinfo{year}{2012}), \bibinfo{pages}{24}.
\newblock


\bibitem[Fog(2011)]%
        {fog_instruction_tables}
\bibfield{author}{\bibinfo{person}{Agner Fog}.}
  \bibinfo{year}{2011}\natexlab{}.
\newblock \bibinfo{title}{Instruction tables: Lists of instruction latencies,
  throughputs and micro-operation breakdowns for Intel, AMD and VIA CPUs}.
\newblock
\newblock
\urldef\tempurl%
\url{https://www.agner.org/optimize/instruction\_tables.pdf}
\showURL{%
\tempurl}


\bibitem[Frigo et~al\mbox{.}(1999)]%
        {frigo1999cache}
\bibfield{author}{\bibinfo{person}{Matteo Frigo}, \bibinfo{person}{Charles~E.
  Leiserson}, \bibinfo{person}{Harald Prokop}, {and} \bibinfo{person}{Sridhar
  Ramachandran}.} \bibinfo{year}{1999}\natexlab{}.
\newblock \showarticletitle{Cache-Oblivious Algorithms}. In
  \bibinfo{booktitle}{\emph{40th Annual Symposium on Foundations of Computer
  Science, {FOCS} '99, 17-18 October, 1999, New York, NY, {USA}}}.
  \bibinfo{publisher}{{IEEE} Computer Society}, \bibinfo{pages}{285--298}.
\newblock
\urldef\tempurl%
\url{https://doi.org/10.1109/SFFCS.1999.814600}
\showDOI{\tempurl}


\bibitem[Gajjala et~al\mbox{.}(2020)]%
        {gajjala2020huffman}
\bibfield{author}{\bibinfo{person}{Rishikesh~R. Gajjala},
  \bibinfo{person}{Shashwat Banchhor}, \bibinfo{person}{Ahmed~M. Abdelmoniem},
  \bibinfo{person}{Aritra Dutta}, \bibinfo{person}{Marco Canini}, {and}
  \bibinfo{person}{Panos Kalnis}.} \bibinfo{year}{2020}\natexlab{}.
\newblock \showarticletitle{Huffman Coding Based Encoding Techniques for Fast
  Distributed Deep Learning}. In \bibinfo{booktitle}{\emph{DistributedML@CoNEXT
  2020: Proceedings of the 1st Workshop on Distributed Machine Learning,
  Barcelona, Spain, December 1, 2020}}. \bibinfo{publisher}{{ACM}},
  \bibinfo{pages}{21--27}.
\newblock
\urldef\tempurl%
\url{https://doi.org/10.1145/3426745.3431334}
\showDOI{\tempurl}


\bibitem[Garc{\'{\i}}a{-}Mart{\'{\i}}n et~al\mbox{.}(2019)]%
        {garcia2019estimation}
\bibfield{author}{\bibinfo{person}{Eva Garc{\'{\i}}a{-}Mart{\'{\i}}n},
  \bibinfo{person}{Crefeda~Faviola Rodrigues}, \bibinfo{person}{Graham~D.
  Riley}, {and} \bibinfo{person}{H{\aa}kan Grahn}.}
  \bibinfo{year}{2019}\natexlab{}.
\newblock \showarticletitle{Estimation of energy consumption in machine
  learning}.
\newblock \bibinfo{journal}{\emph{J. Parallel Distributed Comput.}}
  \bibinfo{volume}{134} (\bibinfo{year}{2019}), \bibinfo{pages}{75--88}.
\newblock
\urldef\tempurl%
\url{https://doi.org/10.1016/J.JPDC.2019.07.007}
\showDOI{\tempurl}


\bibitem[Gasanov et~al\mbox{.}(2022)]%
        {gasanov2021flix}
\bibfield{author}{\bibinfo{person}{Elnur Gasanov}, \bibinfo{person}{Ahmed
  Khaled}, \bibinfo{person}{Samuel Horv{\'{a}}th}, {and} \bibinfo{person}{Peter
  Richt{\'{a}}rik}.} \bibinfo{year}{2022}\natexlab{}.
\newblock \showarticletitle{{FLIX:} {A} Simple and Communication-Efficient
  Alternative to Local Methods in Federated Learning}.
\newblock   \bibinfo{volume}{151} (\bibinfo{year}{2022}),
  \bibinfo{pages}{11374--11421}.
\newblock
\urldef\tempurl%
\url{https://proceedings.mlr.press/v151/gasanov22a.html}
\showURL{%
\tempurl}


\bibitem[Gauss and Stewart(1995)]%
        {gauss1995theory}
\bibfield{author}{\bibinfo{person}{Carl~Friedrich Gauss} {and}
  \bibinfo{person}{GW Stewart}.} \bibinfo{year}{1995}\natexlab{}.
\newblock \bibinfo{booktitle}{\emph{Theory of the combination of observations
  least subject to errors, Part One, Part Two, Supplement}}.
\newblock \bibinfo{publisher}{SIAM}.
\newblock


\bibitem[Gavrilut et~al\mbox{.}(2009)]%
        {gavriluct2009malware}
\bibfield{author}{\bibinfo{person}{Dragos Gavrilut}, \bibinfo{person}{Mihai
  Cimpoesu}, \bibinfo{person}{Dan Anton}, {and} \bibinfo{person}{Liviu
  Ciortuz}.} \bibinfo{year}{2009}\natexlab{}.
\newblock \showarticletitle{Malware detection using machine learning}. In
  \bibinfo{booktitle}{\emph{Proceedings of the International Multiconference on
  Computer Science and Information Technology, {IMCSIT} 2009, Mragowo, Poland,
  12-14 October 2009}}. \bibinfo{publisher}{{IEEE}}, \bibinfo{pages}{735--741}.
\newblock
\urldef\tempurl%
\url{https://doi.org/10.1109/IMCSIT.2009.5352759}
\showDOI{\tempurl}


\bibitem[Gentry(2009)]%
        {gentry2009fully}
\bibfield{author}{\bibinfo{person}{Craig Gentry}.}
  \bibinfo{year}{2009}\natexlab{}.
\newblock \bibinfo{booktitle}{\emph{A fully homomorphic encryption scheme}}.
\newblock
\urldef\tempurl%
\url{https://searchworks.stanford.edu/view/8493082}
\showURL{%
\tempurl}


\bibitem[Golub and Van~Loan(2013)]%
        {golub2013matrix}
\bibfield{author}{\bibinfo{person}{Gene~H Golub} {and}
  \bibinfo{person}{Charles~F Van~Loan}.} \bibinfo{year}{2013}\natexlab{}.
\newblock \bibinfo{booktitle}{\emph{Matrix computations}}.
\newblock \bibinfo{publisher}{JHU press}.
\newblock


\bibitem[Goodfellow et~al\mbox{.}(2016)]%
        {goodfellow2016deep}
\bibfield{author}{\bibinfo{person}{Ian Goodfellow}, \bibinfo{person}{Yoshua
  Bengio}, \bibinfo{person}{Aaron Courville}, {and} \bibinfo{person}{Yoshua
  Bengio}.} \bibinfo{year}{2016}\natexlab{}.
\newblock \bibinfo{booktitle}{\emph{Deep learning}}. Vol.~\bibinfo{volume}{1}.
\newblock \bibinfo{publisher}{MIT Press}.
\newblock


\bibitem[Google(2021)]%
        {googleassistant2021}
\bibfield{author}{\bibinfo{person}{Google}.} \bibinfo{year}{2021}\natexlab{}.
\newblock \bibinfo{title}{Your voice and audio data stays private while Google
  Assistant improves}.
\newblock
  \bibinfo{howpublished}{\url{https://support.google.com/assistant/answer/10176224}}.
\newblock


\bibitem[Gorbunov et~al\mbox{.}(2021)]%
        {gorbunov2021marina}
\bibfield{author}{\bibinfo{person}{Eduard Gorbunov},
  \bibinfo{person}{Konstantin Burlachenko}, \bibinfo{person}{Zhize Li}, {and}
  \bibinfo{person}{Peter Richt{\'{a}}rik}.} \bibinfo{year}{2021}\natexlab{}.
\newblock \showarticletitle{{MARINA:} Faster Non-Convex Distributed Learning
  with Compression}. In \bibinfo{booktitle}{\emph{Proceedings of the 38th
  International Conference on Machine Learning, {ICML} 2021, 18-24 July 2021,
  Virtual Event}} \emph{(\bibinfo{series}{Proceedings of Machine Learning
  Research}, Vol.~\bibinfo{volume}{139})},
  \bibfield{editor}{\bibinfo{person}{Marina Meila} {and} \bibinfo{person}{Tong
  Zhang}} (Eds.). \bibinfo{publisher}{{PMLR}}, \bibinfo{pages}{3788--3798}.
\newblock
\urldef\tempurl%
\url{http://proceedings.mlr.press/v139/gorbunov21a.html}
\showURL{%
\tempurl}


\bibitem[Gorbunov et~al\mbox{.}(2020a)]%
        {LSGDunified2020}
\bibfield{author}{\bibinfo{person}{Eduard Gorbunov}, \bibinfo{person}{Filip
  Hanzely}, {and} \bibinfo{person}{Peter Richt{\'{a}}rik}.}
  \bibinfo{year}{2020}\natexlab{a}.
\newblock \showarticletitle{Local {SGD:} Unified Theory and New Efficient
  Methods}.
\newblock \bibinfo{journal}{\emph{CoRR}}  \bibinfo{volume}{abs/2011.02828}.
\newblock
\showeprint[arXiv]{2011.02828}
\urldef\tempurl%
\url{https://arxiv.org/abs/2011.02828}
\showURL{%
\tempurl}


\bibitem[Gorbunov et~al\mbox{.}(2020b)]%
        {sigma_k}
\bibfield{author}{\bibinfo{person}{Eduard Gorbunov}, \bibinfo{person}{Filip
  Hanzely}, {and} \bibinfo{person}{Peter Richt{\'{a}}rik}.}
  \bibinfo{year}{2020}\natexlab{b}.
\newblock \showarticletitle{A Unified Theory of {SGD:} Variance Reduction,
  Sampling, Quantization and Coordinate Descent}. In
  \bibinfo{booktitle}{\emph{The 23rd International Conference on Artificial
  Intelligence and Statistics, {AISTATS} 2020, 26-28 August 2020, Online
  [Palermo, Sicily, Italy]}} \emph{(\bibinfo{series}{Proceedings of Machine
  Learning Research}, Vol.~\bibinfo{volume}{108})},
  \bibfield{editor}{\bibinfo{person}{Silvia Chiappa} {and}
  \bibinfo{person}{Roberto Calandra}} (Eds.). \bibinfo{publisher}{{PMLR}},
  \bibinfo{pages}{680--690}.
\newblock
\urldef\tempurl%
\url{http://proceedings.mlr.press/v108/gorbunov20a.html}
\showURL{%
\tempurl}


\bibitem[Gorbunov et~al\mbox{.}(2020c)]%
        {Lin_EC_SGD}
\bibfield{author}{\bibinfo{person}{Eduard Gorbunov}, \bibinfo{person}{Dmitry
  Kovalev}, \bibinfo{person}{Dmitry Makarenko}, {and} \bibinfo{person}{Peter
  Richt{\'{a}}rik}.} \bibinfo{year}{2020}\natexlab{c}.
\newblock \showarticletitle{Linearly Converging Error Compensated {SGD}}. In
  \bibinfo{booktitle}{\emph{Advances in Neural Information Processing Systems
  33: Annual Conference on Neural Information Processing Systems 2020, NeurIPS
  2020, December 6-12, 2020, virtual}}, \bibfield{editor}{\bibinfo{person}{Hugo
  Larochelle}, \bibinfo{person}{Marc'Aurelio Ranzato}, \bibinfo{person}{Raia
  Hadsell}, \bibinfo{person}{Maria{-}Florina Balcan}, {and}
  \bibinfo{person}{Hsuan{-}Tien Lin}} (Eds.).
\newblock
\urldef\tempurl%
\url{https://proceedings.neurips.cc/paper/2020/hash/ef9280fbc5317f17d480e4d4f61b3751-Abstract.html}
\showURL{%
\tempurl}


\bibitem[Goulart and Chen(2024)]%
        {clarabel}
\bibfield{author}{\bibinfo{person}{Paul Goulart} {and} \bibinfo{person}{Yuwen
  Chen}.} \bibinfo{year}{2024}\natexlab{}.
\newblock \bibinfo{title}{Clarabel Online Documentation}.
\newblock
\newblock
\urldef\tempurl%
\url{https://clarabel.org}
\showURL{%
\tempurl}
\newblock
\shownote{Accessed: 2024-01-18}.


\bibitem[Gower et~al\mbox{.}(2019)]%
        {Gower2019}
\bibfield{author}{\bibinfo{person}{Robert~Mansel Gower},
  \bibinfo{person}{Nicolas Loizou}, \bibinfo{person}{Xun Qian},
  \bibinfo{person}{Alibek Sailanbayev}, \bibinfo{person}{Egor Shulgin}, {and}
  \bibinfo{person}{Peter Richt{\'{a}}rik}.} \bibinfo{year}{2019}\natexlab{}.
\newblock \showarticletitle{{SGD:} General Analysis and Improved Rates}.
\newblock \bibinfo{journal}{\emph{CoRR}}  \bibinfo{volume}{abs/1901.09401}.
\newblock
\showeprint[arXiv]{1901.09401}
\urldef\tempurl%
\url{http://arxiv.org/abs/1901.09401}
\showURL{%
\tempurl}


\bibitem[Granqvist et~al\mbox{.}(2024)]%
        {granqvist2024pfl}
\bibfield{author}{\bibinfo{person}{Filip Granqvist}, \bibinfo{person}{Congzheng
  Song}, \bibinfo{person}{{\'{A}}ine Cahill}, \bibinfo{person}{Rogier~C. van
  Dalen}, \bibinfo{person}{Martin Pelikan}, \bibinfo{person}{Yi~Sheng Chan},
  \bibinfo{person}{Xiaojun Feng}, \bibinfo{person}{Natarajan Krishnaswami},
  \bibinfo{person}{Vojta Jina}, {and} \bibinfo{person}{Mona Chitnis}.}
  \bibinfo{year}{2024}\natexlab{}.
\newblock \showarticletitle{pfl-research: simulation framework for accelerating
  research in Private Federated Learning}.
\newblock \bibinfo{journal}{\emph{CoRR}}  \bibinfo{volume}{abs/2404.06430}
  (\bibinfo{year}{2024}).
\newblock
\urldef\tempurl%
\url{https://doi.org/10.48550/ARXIV.2404.06430}
\showDOI{\tempurl}
\showeprint[arXiv]{2404.06430}


\bibitem[Gregg(2014)]%
        {gregg2014systems}
\bibfield{author}{\bibinfo{person}{Brendan Gregg}.}
  \bibinfo{year}{2014}\natexlab{}.
\newblock \bibinfo{booktitle}{\emph{Systems performance: enterprise and the
  cloud}}.
\newblock \bibinfo{publisher}{Pearson Education}.
\newblock


\bibitem[Griewank and Walther(2008)]%
        {griewank2008evaluating}
\bibfield{author}{\bibinfo{person}{Andreas Griewank} {and}
  \bibinfo{person}{Andrea Walther}.} \bibinfo{year}{2008}\natexlab{}.
\newblock \bibinfo{booktitle}{\emph{Evaluating derivatives - principles and
  techniques of algorithmic differentiation, Second Edition}}.
\newblock \bibinfo{publisher}{{SIAM}}.
\newblock
\showISBNx{978-0-89871-659-7}
\urldef\tempurl%
\url{https://doi.org/10.1137/1.9780898717761}
\showDOI{\tempurl}


\bibitem[Gross et~al\mbox{.}(2017)]%
        {gross2017efficient}
\bibfield{author}{\bibinfo{person}{Hannes Gross}, \bibinfo{person}{Stefan
  Mangard}, {and} \bibinfo{person}{Thomas Korak}.}
  \bibinfo{year}{2017}\natexlab{}.
\newblock \showarticletitle{An Efficient Side-Channel Protected {AES}
  Implementation with Arbitrary Protection Order}. In
  \bibinfo{booktitle}{\emph{Topics in Cryptology - {CT-RSA} 2017 - The
  Cryptographers' Track at the {RSA} Conference 2017, San Francisco, CA, USA,
  February 14-17, 2017, Proceedings}} \emph{(\bibinfo{series}{Lecture Notes in
  Computer Science}, Vol.~\bibinfo{volume}{10159})},
  \bibfield{editor}{\bibinfo{person}{Helena Handschuh}} (Ed.).
  \bibinfo{publisher}{Springer}, \bibinfo{pages}{95--112}.
\newblock
\urldef\tempurl%
\url{https://doi.org/10.1007/978-3-319-52153-4\_6}
\showDOI{\tempurl}


\bibitem[Guo(2018)]%
        {guo2018}
\bibfield{author}{\bibinfo{person}{Yunhui Guo}.}
  \bibinfo{year}{2018}\natexlab{}.
\newblock \showarticletitle{A Survey on Methods and Theories of Quantized
  Neural Networks}.
\newblock \bibinfo{journal}{\emph{CoRR}}  \bibinfo{volume}{abs/1808.04752}
  (\bibinfo{year}{2018}).
\newblock
\showeprint[arXiv]{1808.04752}
\urldef\tempurl%
\url{http://arxiv.org/abs/1808.04752}
\showURL{%
\tempurl}


\bibitem[{Gurobi Optimization, LLC}(2023)]%
        {gurobi}
\bibfield{author}{\bibinfo{person}{{Gurobi Optimization, LLC}}.}
  \bibinfo{year}{2023}\natexlab{}.
\newblock \bibinfo{title}{{Gurobi Optimizer Reference Manual}}.
\newblock
\newblock
\urldef\tempurl%
\url{http://www.gurobi.com/welcome.html}
\showURL{%
\tempurl}
\newblock
\shownote{Software}.


\bibitem[Hackbusch(2013)]%
        {hackbusch2013multi}
\bibfield{author}{\bibinfo{person}{Wolfgang Hackbusch}.}
  \bibinfo{year}{2013}\natexlab{}.
\newblock \bibinfo{booktitle}{\emph{Multi-grid methods and applications}}.
  Vol.~\bibinfo{volume}{4}.
\newblock \bibinfo{publisher}{Springer Science \& Business Media}.
\newblock


\bibitem[Haddadpour et~al\mbox{.}(2019a)]%
        {haddadpour2019local}
\bibfield{author}{\bibinfo{person}{Farzin Haddadpour},
  \bibinfo{person}{Mohammad~Mahdi Kamani}, \bibinfo{person}{Mehrdad Mahdavi},
  {and} \bibinfo{person}{Viveck~R. Cadambe}.} \bibinfo{year}{2019}\natexlab{a}.
\newblock \showarticletitle{Local {SGD} with Periodic Averaging: Tighter
  Analysis and Adaptive Synchronization}. In \bibinfo{booktitle}{\emph{Advances
  in Neural Information Processing Systems 32: Annual Conference on Neural
  Information Processing Systems 2019, NeurIPS 2019, December 8-14, 2019,
  Vancouver, BC, Canada}}, \bibfield{editor}{\bibinfo{person}{Hanna~M.
  Wallach}, \bibinfo{person}{Hugo Larochelle}, \bibinfo{person}{Alina
  Beygelzimer}, \bibinfo{person}{Florence d'Alch{\'{e}}{-}Buc},
  \bibinfo{person}{Emily~B. Fox}, {and} \bibinfo{person}{Roman Garnett}}
  (Eds.). \bibinfo{pages}{11080--11092}.
\newblock
\urldef\tempurl%
\url{https://proceedings.neurips.cc/paper/2019/hash/c17028c9b6e0c5deaad29665d582284a-Abstract.html}
\showURL{%
\tempurl}


\bibitem[Haddadpour et~al\mbox{.}(2019b)]%
        {Farzin2018}
\bibfield{author}{\bibinfo{person}{Farzin Haddadpour},
  \bibinfo{person}{Mohammad~Mahdi Kamani}, \bibinfo{person}{Mehrdad Mahdavi},
  {and} \bibinfo{person}{Viveck~R. Cadambe}.} \bibinfo{year}{2019}\natexlab{b}.
\newblock \showarticletitle{Trading Redundancy for Communication: Speeding up
  Distributed {SGD} for Non-convex Optimization}. In
  \bibinfo{booktitle}{\emph{Proceedings of the 36th International Conference on
  Machine Learning, {ICML} 2019, 9-15 June 2019, Long Beach, California,
  {USA}}} \emph{(\bibinfo{series}{Proceedings of Machine Learning Research},
  Vol.~\bibinfo{volume}{97})}, \bibfield{editor}{\bibinfo{person}{Kamalika
  Chaudhuri} {and} \bibinfo{person}{Ruslan Salakhutdinov}} (Eds.).
  \bibinfo{publisher}{{PMLR}}, \bibinfo{pages}{2545--2554}.
\newblock
\urldef\tempurl%
\url{http://proceedings.mlr.press/v97/haddadpour19a.html}
\showURL{%
\tempurl}


\bibitem[Hannun et~al\mbox{.}(2023)]%
        {mlx2023}
\bibfield{author}{\bibinfo{person}{Awni Hannun}, \bibinfo{person}{Jagrit
  Digani}, \bibinfo{person}{Angelos Katharopoulos}, {and}
  \bibinfo{person}{Ronan Collobert}.} \bibinfo{year}{2023}\natexlab{}.
\newblock \bibinfo{title}{{MLX}: Efficient and flexible machine learning on
  Apple silicon}.
\newblock
\newblock
\urldef\tempurl%
\url{https://github.com/ml-explore}
\showURL{%
\tempurl}


\bibitem[Hanzely and Richt{\'{a}}rik(2020)]%
        {Hanzely2020}
\bibfield{author}{\bibinfo{person}{Filip Hanzely} {and} \bibinfo{person}{Peter
  Richt{\'{a}}rik}.} \bibinfo{year}{2020}\natexlab{}.
\newblock \showarticletitle{Federated Learning of a Mixture of Global and Local
  Models}.
\newblock \bibinfo{journal}{\emph{CoRR}}  \bibinfo{volume}{abs/2002.05516}
  (\bibinfo{year}{2020}).
\newblock
\showeprint[arXiv]{2002.05516}
\urldef\tempurl%
\url{https://arxiv.org/abs/2002.05516}
\showURL{%
\tempurl}


\bibitem[Hard et~al\mbox{.}(2018)]%
        {hard18gboard}
\bibfield{author}{\bibinfo{person}{Andrew Hard}, \bibinfo{person}{Kanishka
  Rao}, \bibinfo{person}{Rajiv Mathews}, \bibinfo{person}{Fran{\c{c}}oise
  Beaufays}, \bibinfo{person}{Sean Augenstein}, \bibinfo{person}{Hubert
  Eichner}, \bibinfo{person}{Chlo{\'{e}} Kiddon}, {and} \bibinfo{person}{Daniel
  Ramage}.} \bibinfo{year}{2018}\natexlab{}.
\newblock \showarticletitle{Federated Learning for Mobile Keyboard Prediction}.
\newblock \bibinfo{journal}{\emph{CoRR}}  \bibinfo{volume}{abs/1811.03604}
  (\bibinfo{year}{2018}).
\newblock
\showeprint[arXiv]{1811.03604}
\urldef\tempurl%
\url{http://arxiv.org/abs/1811.03604}
\showURL{%
\tempurl}


\bibitem[Harris and Harris(2021)]%
        {harris2015digital}
\bibfield{author}{\bibinfo{person}{Sarah~L. Harris} {and}
  \bibinfo{person}{David~M. Harris}.} \bibinfo{year}{2021}\natexlab{}.
\newblock \showarticletitle{Digital Design and {RISC-V} Computer Architecture
  Textbook}. In \bibinfo{booktitle}{\emph{{ACM/IEEE} Workshop on Computer
  Architecture Education, {WCAE} 2021, Raleigh, NC, USA, June 17, 2021}}.
  \bibinfo{publisher}{{IEEE}}, \bibinfo{pages}{1--5}.
\newblock
\urldef\tempurl%
\url{https://doi.org/10.1109/WCAE53984.2021.9707615}
\showDOI{\tempurl}


\bibitem[He et~al\mbox{.}(2020)]%
        {he2020fedml}
\bibfield{author}{\bibinfo{person}{Chaoyang He}, \bibinfo{person}{Songze Li},
  \bibinfo{person}{Jinhyun So}, \bibinfo{person}{Mi Zhang},
  \bibinfo{person}{Hongyi Wang}, \bibinfo{person}{Xiaoyang Wang},
  \bibinfo{person}{Praneeth Vepakomma}, \bibinfo{person}{Abhishek Singh},
  \bibinfo{person}{Hang Qiu}, \bibinfo{person}{Li Shen},
  \bibinfo{person}{Peilin Zhao}, \bibinfo{person}{Yan Kang},
  \bibinfo{person}{Yang Liu}, \bibinfo{person}{Ramesh Raskar},
  \bibinfo{person}{Qiang Yang}, \bibinfo{person}{Murali Annavaram}, {and}
  \bibinfo{person}{Salman Avestimehr}.} \bibinfo{year}{2020}\natexlab{}.
\newblock \showarticletitle{FedML: {A} Research Library and Benchmark for
  Federated Machine Learning}.
\newblock \bibinfo{journal}{\emph{CoRR}}  \bibinfo{volume}{abs/2007.13518}
  (\bibinfo{year}{2020}).
\newblock
\showeprint[arXiv]{2007.13518}
\urldef\tempurl%
\url{https://arxiv.org/abs/2007.13518}
\showURL{%
\tempurl}


\bibitem[He et~al\mbox{.}(2016)]%
        {resnet}
\bibfield{author}{\bibinfo{person}{Kaiming He}, \bibinfo{person}{Xiangyu
  Zhang}, \bibinfo{person}{Shaoqing Ren}, {and} \bibinfo{person}{Jian Sun}.}
  \bibinfo{year}{2016}\natexlab{}.
\newblock \showarticletitle{Deep Residual Learning for Image Recognition}. In
  \bibinfo{booktitle}{\emph{2016 {IEEE} Conference on Computer Vision and
  Pattern Recognition, {CVPR} 2016, Las Vegas, NV, USA, June 27-30, 2016}}.
  \bibinfo{publisher}{{IEEE} Computer Society}, \bibinfo{pages}{770--778}.
\newblock
\urldef\tempurl%
\url{https://doi.org/10.1109/CVPR.2016.90}
\showDOI{\tempurl}


\bibitem[Heek et~al\mbox{.}(2024)]%
        {flax2020github}
\bibfield{author}{\bibinfo{person}{Jonathan Heek}, \bibinfo{person}{Anselm
  Levskaya}, \bibinfo{person}{Avital Oliver}, \bibinfo{person}{Marvin Ritter},
  \bibinfo{person}{Bertrand Rondepierre}, \bibinfo{person}{Andreas Steiner},
  {and} \bibinfo{person}{Marc van {Z}ee}.} \bibinfo{year}{2024}\natexlab{}.
\newblock \bibinfo{title}{{F}lax: A neural network library and ecosystem for
  {JAX}}.
\newblock
\newblock
\urldef\tempurl%
\url{http://github.com/google/flax}
\showURL{%
\tempurl}


\bibitem[Hennessy and Patterson(2012)]%
        {hennessy2011computer}
\bibfield{author}{\bibinfo{person}{John~L. Hennessy} {and}
  \bibinfo{person}{David~A. Patterson}.} \bibinfo{year}{2012}\natexlab{}.
\newblock \bibinfo{booktitle}{\emph{Computer Architecture - {A} Quantitative
  Approach, 5th Edition}}.
\newblock \bibinfo{publisher}{Morgan Kaufmann}.
\newblock
\showISBNx{978-0-12-383872-8}


\bibitem[Hong and Kung(1981)]%
        {jia1981complexity}
\bibfield{author}{\bibinfo{person}{Jia{-}Wei Hong} {and} \bibinfo{person}{H.~T.
  Kung}.} \bibinfo{year}{1981}\natexlab{}.
\newblock \showarticletitle{{I/O} Complexity: The Red-Blue Pebble Game}. In
  \bibinfo{booktitle}{\emph{Proceedings of the 13th Annual {ACM} Symposium on
  Theory of Computing, May 11-13, 1981, Milwaukee, Wisconsin, {USA}}}.
  \bibinfo{publisher}{{ACM}}, \bibinfo{pages}{326--333}.
\newblock
\urldef\tempurl%
\url{https://doi.org/10.1145/800076.802486}
\showDOI{\tempurl}


\bibitem[Horowitz(2014)]%
        {horowitz20141}
\bibfield{author}{\bibinfo{person}{Mark Horowitz}.}
  \bibinfo{year}{2014}\natexlab{}.
\newblock \showarticletitle{1.1 Computing's energy problem (and what we can do
  about it)}. In \bibinfo{booktitle}{\emph{2014 {IEEE} International Conference
  on Solid-State Circuits Conference, {ISSCC} 2014, Digest of Technical Papers,
  San Francisco, CA, USA, February 9-13, 2014}}. \bibinfo{publisher}{{IEEE}},
  \bibinfo{pages}{10--14}.
\newblock
\urldef\tempurl%
\url{https://doi.org/10.1109/ISSCC.2014.6757323}
\showDOI{\tempurl}


\bibitem[Horv{\'{a}}th et~al\mbox{.}(2019)]%
        {horvath2019natural}
\bibfield{author}{\bibinfo{person}{Samuel Horv{\'{a}}th},
  \bibinfo{person}{Chen{-}Yu Ho}, \bibinfo{person}{Ludovit Horvath},
  \bibinfo{person}{Atal~Narayan Sahu}, \bibinfo{person}{Marco Canini}, {and}
  \bibinfo{person}{Peter Richt{\'{a}}rik}.} \bibinfo{year}{2019}\natexlab{}.
\newblock \showarticletitle{Natural Compression for Distributed Deep Learning}.
\newblock \bibinfo{journal}{\emph{CoRR}}  \bibinfo{volume}{abs/1905.10988}
  (\bibinfo{year}{2019}).
\newblock
\showeprint[arXiv]{1905.10988}
\urldef\tempurl%
\url{http://arxiv.org/abs/1905.10988}
\showURL{%
\tempurl}


\bibitem[Horv{\'{a}}th et~al\mbox{.}(2023)]%
        {DIANA2}
\bibfield{author}{\bibinfo{person}{Samuel Horv{\'{a}}th},
  \bibinfo{person}{Dmitry Kovalev}, \bibinfo{person}{Konstantin Mishchenko},
  \bibinfo{person}{Peter Richt{\'{a}}rik}, {and} \bibinfo{person}{Sebastian~U.
  Stich}.} \bibinfo{year}{2023}\natexlab{}.
\newblock \showarticletitle{Stochastic distributed learning with gradient
  quantization and double-variance reduction}.
\newblock \bibinfo{journal}{\emph{Optim. Methods Softw.}} \bibinfo{volume}{38},
  \bibinfo{number}{1} (\bibinfo{year}{2023}), \bibinfo{pages}{91--106}.
\newblock
\urldef\tempurl%
\url{https://doi.org/10.1080/10556788.2022.2117355}
\showDOI{\tempurl}


\bibitem[Horv{\'{a}}th et~al\mbox{.}(2021)]%
        {horvath2021fjord}
\bibfield{author}{\bibinfo{person}{Samuel Horv{\'{a}}th},
  \bibinfo{person}{Stefanos Laskaridis}, \bibinfo{person}{M{\'{a}}rio Almeida},
  \bibinfo{person}{Ilias Leontiadis}, \bibinfo{person}{Stylianos~I. Venieris},
  {and} \bibinfo{person}{Nicholas~D. Lane}.} \bibinfo{year}{2021}\natexlab{}.
\newblock \showarticletitle{FjORD: Fair and Accurate Federated Learning under
  heterogeneous targets with Ordered Dropout}.
\newblock  (\bibinfo{year}{2021}), \bibinfo{pages}{12876--12889}.
\newblock
\urldef\tempurl%
\url{https://proceedings.neurips.cc/paper/2021/hash/6aed000af86a084f9cb0264161e29dd3-Abstract.html}
\showURL{%
\tempurl}


\bibitem[Horv{\'{a}}th et~al\mbox{.}(2022)]%
        {horvath2020adaptivity}
\bibfield{author}{\bibinfo{person}{Samuel Horv{\'{a}}th},
  \bibinfo{person}{Lihua Lei}, \bibinfo{person}{Peter Richt{\'{a}}rik}, {and}
  \bibinfo{person}{Michael~I. Jordan}.} \bibinfo{year}{2022}\natexlab{}.
\newblock \showarticletitle{Adaptivity of Stochastic Gradient Methods for
  Nonconvex Optimization}.
\newblock \bibinfo{journal}{\emph{{SIAM} J. Math. Data Sci.}}
  \bibinfo{volume}{4}, \bibinfo{number}{2} (\bibinfo{year}{2022}),
  \bibinfo{pages}{634--648}.
\newblock
\urldef\tempurl%
\url{https://doi.org/10.1137/21M1394308}
\showDOI{\tempurl}


\bibitem[Horv{\'{a}}th and Richt{\'{a}}rik(2019)]%
        {horvath2019nonconvex}
\bibfield{author}{\bibinfo{person}{Samuel Horv{\'{a}}th} {and}
  \bibinfo{person}{Peter Richt{\'{a}}rik}.} \bibinfo{year}{2019}\natexlab{}.
\newblock \showarticletitle{Nonconvex Variance Reduced Optimization with
  Arbitrary Sampling}. In \bibinfo{booktitle}{\emph{Proceedings of the 36th
  International Conference on Machine Learning, {ICML} 2019, 9-15 June 2019,
  Long Beach, California, {USA}}} \emph{(\bibinfo{series}{Proceedings of
  Machine Learning Research}, Vol.~\bibinfo{volume}{97})},
  \bibfield{editor}{\bibinfo{person}{Kamalika Chaudhuri} {and}
  \bibinfo{person}{Ruslan Salakhutdinov}} (Eds.). \bibinfo{publisher}{{PMLR}},
  \bibinfo{pages}{2781--2789}.
\newblock
\urldef\tempurl%
\url{http://proceedings.mlr.press/v97/horvath19a.html}
\showURL{%
\tempurl}


\bibitem[Howard et~al\mbox{.}(2017)]%
        {howard2017mobilenets}
\bibfield{author}{\bibinfo{person}{Andrew~G. Howard}, \bibinfo{person}{Menglong
  Zhu}, \bibinfo{person}{Bo Chen}, \bibinfo{person}{Dmitry Kalenichenko},
  \bibinfo{person}{Weijun Wang}, \bibinfo{person}{Tobias Weyand},
  \bibinfo{person}{Marco Andreetto}, {and} \bibinfo{person}{Hartwig Adam}.}
  \bibinfo{year}{2017}\natexlab{}.
\newblock \showarticletitle{MobileNets: Efficient Convolutional Neural Networks
  for Mobile Vision Applications}.
\newblock \bibinfo{journal}{\emph{CoRR}}  \bibinfo{volume}{abs/1704.04861}
  (\bibinfo{year}{2017}).
\newblock
\showeprint[arXiv]{1704.04861}
\urldef\tempurl%
\url{http://arxiv.org/abs/1704.04861}
\showURL{%
\tempurl}


\bibitem[Huang et~al\mbox{.}(2017)]%
        {densenet}
\bibfield{author}{\bibinfo{person}{Gao Huang}, \bibinfo{person}{Zhuang Liu},
  \bibinfo{person}{Laurens van~der Maaten}, {and} \bibinfo{person}{Kilian~Q.
  Weinberger}.} \bibinfo{year}{2017}\natexlab{}.
\newblock \showarticletitle{Densely Connected Convolutional Networks}. In
  \bibinfo{booktitle}{\emph{2017 {IEEE} Conference on Computer Vision and
  Pattern Recognition, {CVPR} 2017, Honolulu, HI, USA, July 21-26, 2017}}.
  \bibinfo{publisher}{{IEEE} Computer Society}, \bibinfo{pages}{2261--2269}.
\newblock
\urldef\tempurl%
\url{https://doi.org/10.1109/CVPR.2017.243}
\showDOI{\tempurl}


\bibitem[Huang et~al\mbox{.}(2013)]%
        {huang2013depth}
\bibfield{author}{\bibinfo{person}{Junxian Huang}, \bibinfo{person}{Feng Qian},
  \bibinfo{person}{Yihua Guo}, \bibinfo{person}{Yuanyuan Zhou},
  \bibinfo{person}{Qiang Xu}, \bibinfo{person}{Zhuoqing~Morley Mao},
  \bibinfo{person}{Subhabrata Sen}, {and} \bibinfo{person}{Oliver Spatscheck}.}
  \bibinfo{year}{2013}\natexlab{}.
\newblock \showarticletitle{An in-depth study of {LTE:} effect of network
  protocol and application behavior on performance}.
\newblock  (\bibinfo{year}{2013}), \bibinfo{pages}{363--374}.
\newblock
\urldef\tempurl%
\url{https://doi.org/10.1145/2486001.2486006}
\showDOI{\tempurl}


\bibitem[Ingerman and Ostrowski(2019)]%
        {TFF2019}
\bibfield{author}{\bibinfo{person}{Alex Ingerman} {and} \bibinfo{person}{Krzys
  Ostrowski}.} \bibinfo{year}{2019}\natexlab{}.
\newblock \bibinfo{booktitle}{\emph{TensorFlow Federated}}.
\newblock
\urldef\tempurl%
\url{https://medium.com/tensorflow/introducing-tensorflow-federated-a4147aa20041}
\showURL{%
\tempurl}


\bibitem[Intel and Consilient(2020)]%
        {intel2020}
\bibfield{author}{\bibinfo{person}{Intel} {and} \bibinfo{person}{Consilient}.}
  \bibinfo{year}{2020}\natexlab{}.
\newblock \bibinfo{title}{Intel and Consilient Join Forces to Fight Financial
  Fraud with AI}.
\newblock
  \bibinfo{howpublished}{\url{https://newsroom.intel.com/news/intel-consilient-join-forces-fight-financial-fraud-ai/}}.
\newblock


\bibitem[Intel®(2021)]%
        {OpenFLFramework}
\bibfield{author}{\bibinfo{person}{Intel®}.} \bibinfo{year}{2021}\natexlab{}.
\newblock \bibinfo{title}{Intel® Open Federated Learning}.
\newblock
\newblock
\urldef\tempurl%
\url{https://github.com/intel/openfl}
\showURL{%
\tempurl}


\bibitem[Ioffe and Szegedy(2015)]%
        {ioffe2015batch}
\bibfield{author}{\bibinfo{person}{Sergey Ioffe} {and}
  \bibinfo{person}{Christian Szegedy}.} \bibinfo{year}{2015}\natexlab{}.
\newblock \showarticletitle{Batch Normalization: Accelerating Deep Network
  Training by Reducing Internal Covariate Shift}. In
  \bibinfo{booktitle}{\emph{Proceedings of the 32nd International Conference on
  Machine Learning, {ICML} 2015, Lille, France, 6-11 July 2015}}
  \emph{(\bibinfo{series}{{JMLR} Workshop and Conference Proceedings},
  Vol.~\bibinfo{volume}{37})}, \bibfield{editor}{\bibinfo{person}{Francis~R.
  Bach} {and} \bibinfo{person}{David~M. Blei}} (Eds.).
  \bibinfo{publisher}{JMLR.org}, \bibinfo{pages}{448--456}.
\newblock
\urldef\tempurl%
\url{http://proceedings.mlr.press/v37/ioffe15.html}
\showURL{%
\tempurl}


\bibitem[Issariyakul et~al\mbox{.}(2009)]%
        {issariyakul2009introduction}
\bibfield{author}{\bibinfo{person}{Teerawat Issariyakul},
  \bibinfo{person}{Ekram Hossain}, \bibinfo{person}{Teerawat Issariyakul},
  {and} \bibinfo{person}{Ekram Hossain}.} \bibinfo{year}{2009}\natexlab{}.
\newblock \bibinfo{booktitle}{\emph{Introduction to network simulator 2
  (NS2)}}.
\newblock \bibinfo{publisher}{Springer}.
\newblock


\bibitem[Jaggi et~al\mbox{.}(2014)]%
        {NIPS2014_5599}
\bibfield{author}{\bibinfo{person}{Martin Jaggi}, \bibinfo{person}{Virginia
  Smith}, \bibinfo{person}{Martin Tak{\'{a}}c}, \bibinfo{person}{Jonathan
  Terhorst}, \bibinfo{person}{Sanjay Krishnan}, \bibinfo{person}{Thomas
  Hofmann}, {and} \bibinfo{person}{Michael~I. Jordan}.}
  \bibinfo{year}{2014}\natexlab{}.
\newblock \showarticletitle{Communication-Efficient Distributed Dual Coordinate
  Ascent}. In \bibinfo{booktitle}{\emph{Advances in Neural Information
  Processing Systems 27: Annual Conference on Neural Information Processing
  Systems 2014, December 8-13 2014, Montreal, Quebec, Canada}},
  \bibfield{editor}{\bibinfo{person}{Zoubin Ghahramani}, \bibinfo{person}{Max
  Welling}, \bibinfo{person}{Corinna Cortes}, \bibinfo{person}{Neil~D.
  Lawrence}, {and} \bibinfo{person}{Kilian~Q. Weinberger}} (Eds.).
  \bibinfo{pages}{3068--3076}.
\newblock
\urldef\tempurl%
\url{https://proceedings.neurips.cc/paper/2014/hash/894b77f805bd94d292574c38c5d628d5-Abstract.html}
\showURL{%
\tempurl}


\bibitem[Jain and Cherukuri(2023)]%
        {jain2023revisiting}
\bibfield{author}{\bibinfo{person}{Nimish Jain} {and}
  \bibinfo{person}{Aswani~Kumar Cherukuri}.} \bibinfo{year}{2023}\natexlab{}.
\newblock \showarticletitle{Revisiting Fully Homomorphic Encryption Schemes}.
\newblock \bibinfo{journal}{\emph{CoRR}}  \bibinfo{volume}{abs/2305.05904}
  (\bibinfo{year}{2023}).
\newblock
\urldef\tempurl%
\url{https://doi.org/10.48550/ARXIV.2305.05904}
\showDOI{\tempurl}
\showeprint[arXiv]{2305.05904}


\bibitem[Jang et~al\mbox{.}(2022)]%
        {jang2022quantum}
\bibfield{author}{\bibinfo{person}{Kyungbae Jang}, \bibinfo{person}{Anubhab
  Baksi}, \bibinfo{person}{Hyunji Kim}, \bibinfo{person}{Gyeongju Song},
  \bibinfo{person}{Hwajeong Seo}, {and} \bibinfo{person}{Anupam
  Chattopadhyay}.} \bibinfo{year}{2022}\natexlab{}.
\newblock \showarticletitle{Quantum Analysis of {AES}}.
\newblock \bibinfo{journal}{\emph{{IACR} Cryptol. ePrint Arch.}}
  (\bibinfo{year}{2022}), \bibinfo{pages}{683}.
\newblock
\urldef\tempurl%
\url{https://eprint.iacr.org/2022/683}
\showURL{%
\tempurl}


\bibitem[Jia et~al\mbox{.}(2014)]%
        {jia2014caffe}
\bibfield{author}{\bibinfo{person}{Yangqing Jia}, \bibinfo{person}{Evan
  Shelhamer}, \bibinfo{person}{Jeff Donahue}, \bibinfo{person}{Sergey Karayev},
  \bibinfo{person}{Jonathan Long}, \bibinfo{person}{Ross~B. Girshick},
  \bibinfo{person}{Sergio Guadarrama}, {and} \bibinfo{person}{Trevor Darrell}.}
  \bibinfo{year}{2014}\natexlab{}.
\newblock \showarticletitle{Caffe: Convolutional Architecture for Fast Feature
  Embedding}. In \bibinfo{booktitle}{\emph{Proceedings of the {ACM}
  International Conference on Multimedia, {MM} '14, Orlando, FL, USA, November
  03 - 07, 2014}}, \bibfield{editor}{\bibinfo{person}{Kien~A. Hua},
  \bibinfo{person}{Yong Rui}, \bibinfo{person}{Ralf Steinmetz},
  \bibinfo{person}{Alan Hanjalic}, \bibinfo{person}{Apostol Natsev}, {and}
  \bibinfo{person}{Wenwu Zhu}} (Eds.). \bibinfo{publisher}{{ACM}},
  \bibinfo{pages}{675--678}.
\newblock
\urldef\tempurl%
\url{https://doi.org/10.1145/2647868.2654889}
\showDOI{\tempurl}


\bibitem[Jia et~al\mbox{.}(2019a)]%
        {jia2019taso}
\bibfield{author}{\bibinfo{person}{Zhihao Jia}, \bibinfo{person}{Oded Padon},
  \bibinfo{person}{James Thomas}, \bibinfo{person}{Todd Warszawski},
  \bibinfo{person}{Matei Zaharia}, {and} \bibinfo{person}{Alex Aiken}.}
  \bibinfo{year}{2019}\natexlab{a}.
\newblock \showarticletitle{{TASO:} optimizing deep learning computation with
  automatic generation of graph substitutions}. In
  \bibinfo{booktitle}{\emph{Proceedings of the 27th {ACM} Symposium on
  Operating Systems Principles, {SOSP} 2019, Huntsville, ON, Canada, October
  27-30, 2019}}, \bibfield{editor}{\bibinfo{person}{Tim Brecht} {and}
  \bibinfo{person}{Carey Williamson}} (Eds.). \bibinfo{publisher}{{ACM}},
  \bibinfo{pages}{47--62}.
\newblock
\urldef\tempurl%
\url{https://doi.org/10.1145/3341301.3359630}
\showDOI{\tempurl}


\bibitem[Jia et~al\mbox{.}(2019b)]%
        {jia2019beyond}
\bibfield{author}{\bibinfo{person}{Zhihao Jia}, \bibinfo{person}{Matei
  Zaharia}, {and} \bibinfo{person}{Alex Aiken}.}
  \bibinfo{year}{2019}\natexlab{b}.
\newblock \showarticletitle{Beyond Data and Model Parallelism for Deep Neural
  Networks}.
\newblock  (\bibinfo{year}{2019}).
\newblock
\urldef\tempurl%
\url{https://proceedings.mlsys.org/paper\_files/paper/2019/hash/b422680f3db0986ddd7f8f126baaf0fa-Abstract.html}
\showURL{%
\tempurl}


\bibitem[Jiang et~al\mbox{.}(2021)]%
        {jiang2021flashe}
\bibfield{author}{\bibinfo{person}{Zhifeng Jiang}, \bibinfo{person}{Wei Wang},
  {and} \bibinfo{person}{Yang Liu}.} \bibinfo{year}{2021}\natexlab{}.
\newblock \showarticletitle{{FLASHE:} Additively Symmetric Homomorphic
  Encryption for Cross-Silo Federated Learning}.
\newblock \bibinfo{journal}{\emph{CoRR}}  \bibinfo{volume}{abs/2109.00675}
  (\bibinfo{year}{2021}).
\newblock
\showeprint[arXiv]{2109.00675}
\urldef\tempurl%
\url{https://arxiv.org/abs/2109.00675}
\showURL{%
\tempurl}


\bibitem[Johnson and Zhang(2013)]%
        {johnson2013accelerating}
\bibfield{author}{\bibinfo{person}{Rie Johnson} {and} \bibinfo{person}{Tong
  Zhang}.} \bibinfo{year}{2013}\natexlab{}.
\newblock \showarticletitle{Accelerating Stochastic Gradient Descent using
  Predictive Variance Reduction}.
\newblock  (\bibinfo{year}{2013}), \bibinfo{pages}{315--323}.
\newblock
\urldef\tempurl%
\url{https://proceedings.neurips.cc/paper/2013/hash/ac1dd209cbcc5e5d1c6e28598e8cbbe8-Abstract.html}
\showURL{%
\tempurl}


\bibitem[Kairouz et~al\mbox{.}(2021)]%
        {kairouz2019advances}
\bibfield{author}{\bibinfo{person}{Peter Kairouz}, \bibinfo{person}{H.~Brendan
  McMahan}, \bibinfo{person}{Brendan Avent}, \bibinfo{person}{Aur{\'{e}}lien
  Bellet}, \bibinfo{person}{Mehdi Bennis}, \bibinfo{person}{Arjun~Nitin
  Bhagoji}, \bibinfo{person}{Kallista~A. Bonawitz}, \bibinfo{person}{Zachary
  Charles}, \bibinfo{person}{Graham Cormode}, \bibinfo{person}{Rachel
  Cummings}, \bibinfo{person}{Rafael G.~L. D'Oliveira}, \bibinfo{person}{Hubert
  Eichner}, \bibinfo{person}{Salim~El Rouayheb}, \bibinfo{person}{David Evans},
  \bibinfo{person}{Josh Gardner}, \bibinfo{person}{Zachary Garrett},
  \bibinfo{person}{Adri{\`{a}} Gasc{\'{o}}n}, \bibinfo{person}{Badih Ghazi},
  \bibinfo{person}{Phillip~B. Gibbons}, \bibinfo{person}{Marco Gruteser},
  \bibinfo{person}{Za{\"{\i}}d Harchaoui}, \bibinfo{person}{Chaoyang He},
  \bibinfo{person}{Lie He}, \bibinfo{person}{Zhouyuan Huo},
  \bibinfo{person}{Ben Hutchinson}, \bibinfo{person}{Justin Hsu},
  \bibinfo{person}{Martin Jaggi}, \bibinfo{person}{Tara Javidi},
  \bibinfo{person}{Gauri Joshi}, \bibinfo{person}{Mikhail Khodak},
  \bibinfo{person}{Jakub Kone{\v{c}}n{\'y}}, \bibinfo{person}{Aleksandra
  Korolova}, \bibinfo{person}{Farinaz Koushanfar}, \bibinfo{person}{Sanmi
  Koyejo}, \bibinfo{person}{Tancr{\`{e}}de Lepoint}, \bibinfo{person}{Yang
  Liu}, \bibinfo{person}{Prateek Mittal}, \bibinfo{person}{Mehryar Mohri},
  \bibinfo{person}{Richard Nock}, \bibinfo{person}{Ayfer {\"{O}}zg{\"{u}}r},
  \bibinfo{person}{Rasmus Pagh}, \bibinfo{person}{Hang Qi},
  \bibinfo{person}{Daniel Ramage}, \bibinfo{person}{Ramesh Raskar},
  \bibinfo{person}{Mariana Raykova}, \bibinfo{person}{Dawn Song},
  \bibinfo{person}{Weikang Song}, \bibinfo{person}{Sebastian~U. Stich},
  \bibinfo{person}{Ziteng Sun}, \bibinfo{person}{Ananda~Theertha Suresh},
  \bibinfo{person}{Florian Tram{\`{e}}r}, \bibinfo{person}{Praneeth Vepakomma},
  \bibinfo{person}{Jianyu Wang}, \bibinfo{person}{Li Xiong},
  \bibinfo{person}{Zheng Xu}, \bibinfo{person}{Qiang Yang},
  \bibinfo{person}{Felix~X. Yu}, \bibinfo{person}{Han Yu}, {and}
  \bibinfo{person}{Sen Zhao}.} \bibinfo{year}{2021}\natexlab{}.
\newblock \showarticletitle{Advances and Open Problems in Federated Learning}.
\newblock \bibinfo{journal}{\emph{Found. Trends Mach. Learn.}}
  \bibinfo{volume}{14}, \bibinfo{number}{1-2} (\bibinfo{year}{2021}),
  \bibinfo{pages}{1--210}.
\newblock
\urldef\tempurl%
\url{https://doi.org/10.1561/2200000083}
\showDOI{\tempurl}


\bibitem[Kaissis et~al\mbox{.}(2020)]%
        {kaissis2020secure}
\bibfield{author}{\bibinfo{person}{Georgios Kaissis},
  \bibinfo{person}{Marcus~R. Makowski}, \bibinfo{person}{Daniel Rueckert},
  {and} \bibinfo{person}{Rickmer Braren}.} \bibinfo{year}{2020}\natexlab{}.
\newblock \showarticletitle{Secure, privacy-preserving and federated machine
  learning in medical imaging}.
\newblock \bibinfo{journal}{\emph{Nat. Mach. Intell.}} \bibinfo{volume}{2},
  \bibinfo{number}{6} (\bibinfo{year}{2020}), \bibinfo{pages}{305--311}.
\newblock
\urldef\tempurl%
\url{https://doi.org/10.1038/S42256-020-0186-1}
\showDOI{\tempurl}


\bibitem[Karimireddy et~al\mbox{.}(2020)]%
        {karimireddy2020scaffold}
\bibfield{author}{\bibinfo{person}{Sai~Praneeth Karimireddy},
  \bibinfo{person}{Satyen Kale}, \bibinfo{person}{Mehryar Mohri},
  \bibinfo{person}{Sashank~J. Reddi}, \bibinfo{person}{Sebastian~U. Stich},
  {and} \bibinfo{person}{Ananda~Theertha Suresh}.}
  \bibinfo{year}{2020}\natexlab{}.
\newblock \showarticletitle{{SCAFFOLD:} Stochastic Controlled Averaging for
  Federated Learning}. In \bibinfo{booktitle}{\emph{Proceedings of the 37th
  International Conference on Machine Learning, {ICML} 2020, 13-18 July 2020,
  Virtual Event}} \emph{(\bibinfo{series}{Proceedings of Machine Learning
  Research}, Vol.~\bibinfo{volume}{119})}. \bibinfo{publisher}{{PMLR}},
  \bibinfo{pages}{5132--5143}.
\newblock
\urldef\tempurl%
\url{http://proceedings.mlr.press/v119/karimireddy20a.html}
\showURL{%
\tempurl}


\bibitem[Karpathy(2015)]%
        {tinyshakespeare}
\bibfield{author}{\bibinfo{person}{Andrej Karpathy}.}
  \bibinfo{year}{2015}\natexlab{}.
\newblock \bibinfo{title}{char-rnn}.
\newblock \bibinfo{howpublished}{\url{https://github.com/karpathy/char-rnn}}.
\newblock


\bibitem[Karpathy(2020)]%
        {karpathy2020micrograd}
\bibfield{author}{\bibinfo{person}{Andrej Karpathy}.}
  \bibinfo{year}{2020}\natexlab{}.
\newblock \bibinfo{title}{Micrograd: A tiny autograd engine}.
\newblock \bibinfo{howpublished}{\url{https://github.com/karpathy/micrograd}}.
\newblock
\urldef\tempurl%
\url{https://github.com/karpathy/micrograd}
\showURL{%
\tempurl}
\newblock
\shownote{GitHub repository}.


\bibitem[Karpathy(2023a)]%
        {karpathygptimplementation}
\bibfield{author}{\bibinfo{person}{Andrej Karpathy}.}
  \bibinfo{year}{2023}\natexlab{a}.
\newblock \bibinfo{title}{GPT-3 Baseline Implementation}.
\newblock
\newblock
\urldef\tempurl%
\url{https://github.com/karpathy/ng-video-lecture/blob/master/gpt.py}
\showURL{%
\tempurl}
\newblock
\shownote{Accessed: 2025-01-30}.


\bibitem[Karpathy(2023b)]%
        {makemore2023}
\bibfield{author}{\bibinfo{person}{Andrej Karpathy}.}
  \bibinfo{year}{2023}\natexlab{b}.
\newblock \bibinfo{title}{makemore}.
\newblock \bibinfo{howpublished}{\url{https://github.com/karpathy/makemore}}.
\newblock
\urldef\tempurl%
\url{https://github.com/karpathy/makemore}
\showURL{%
\tempurl}
\newblock
\shownote{GitHub repository}.


\bibitem[Kasiviswanathan et~al\mbox{.}(2013)]%
        {kasiviswanathan2013power}
\bibfield{author}{\bibinfo{person}{Shiva~Prasad Kasiviswanathan},
  \bibinfo{person}{Mark Rudelson}, {and} \bibinfo{person}{Adam~D. Smith}.}
  \bibinfo{year}{2013}\natexlab{}.
\newblock \showarticletitle{The Power of Linear Reconstruction Attacks}. In
  \bibinfo{booktitle}{\emph{Proceedings of the Twenty-Fourth Annual {ACM-SIAM}
  Symposium on Discrete Algorithms, {SODA} 2013, New Orleans, Louisiana, USA,
  January 6-8, 2013}}, \bibfield{editor}{\bibinfo{person}{Sanjeev Khanna}}
  (Ed.). \bibinfo{publisher}{{SIAM}}, \bibinfo{pages}{1415--1433}.
\newblock
\urldef\tempurl%
\url{https://doi.org/10.1137/1.9781611973105.102}
\showDOI{\tempurl}


\bibitem[Kelley(1999)]%
        {kelley1999iterative}
\bibfield{author}{\bibinfo{person}{Carl~T. Kelley}.}
  \bibinfo{year}{1999}\natexlab{}.
\newblock \bibinfo{booktitle}{\emph{Iterative methods for optimization}}.
\newblock \bibinfo{publisher}{{SIAM}}.
\newblock
\showISBNx{978-0-89871-433-3}


\bibitem[Kerrisk(2010)]%
        {kerrisk2010linux}
\bibfield{author}{\bibinfo{person}{Michael Kerrisk}.}
  \bibinfo{year}{2010}\natexlab{}.
\newblock \bibinfo{booktitle}{\emph{The Linux programming interface: a Linux
  and UNIX system programming handbook}}.
\newblock \bibinfo{publisher}{No Starch Press}.
\newblock


\bibitem[Khaled et~al\mbox{.}(2019)]%
        {FirstLocalGDHeter}
\bibfield{author}{\bibinfo{person}{Ahmed Khaled}, \bibinfo{person}{Konstantin
  Mishchenko}, {and} \bibinfo{person}{Peter Richt{\'{a}}rik}.}
  \bibinfo{year}{2019}\natexlab{}.
\newblock \showarticletitle{First Analysis of Local {GD} on Heterogeneous
  Data}.
\newblock \bibinfo{journal}{\emph{CoRR}}  \bibinfo{volume}{abs/1909.04715}.
\newblock
\showeprint[arXiv]{1909.04715}
\urldef\tempurl%
\url{http://arxiv.org/abs/1909.04715}
\showURL{%
\tempurl}


\bibitem[Khaled et~al\mbox{.}(2020)]%
        {khaled_lgd}
\bibfield{author}{\bibinfo{person}{Ahmed Khaled}, \bibinfo{person}{Konstantin
  Mishchenko}, {and} \bibinfo{person}{Peter Richt{\'{a}}rik}.}
  \bibinfo{year}{2020}\natexlab{}.
\newblock \showarticletitle{Tighter Theory for Local {SGD} on Identical and
  Heterogeneous Data}. In \bibinfo{booktitle}{\emph{The 23rd International
  Conference on Artificial Intelligence and Statistics, {AISTATS} 2020, 26-28
  August 2020, Online [Palermo, Sicily, Italy]}}
  \emph{(\bibinfo{series}{Proceedings of Machine Learning Research},
  Vol.~\bibinfo{volume}{108})}, \bibfield{editor}{\bibinfo{person}{Silvia
  Chiappa} {and} \bibinfo{person}{Roberto Calandra}} (Eds.).
  \bibinfo{publisher}{{PMLR}}, \bibinfo{pages}{4519--4529}.
\newblock
\urldef\tempurl%
\url{http://proceedings.mlr.press/v108/bayoumi20a.html}
\showURL{%
\tempurl}


\bibitem[Khaled and Richt{\'{a}}rik(2023)]%
        {khaled2020better}
\bibfield{author}{\bibinfo{person}{Ahmed Khaled} {and} \bibinfo{person}{Peter
  Richt{\'{a}}rik}.} \bibinfo{year}{2023}\natexlab{}.
\newblock \showarticletitle{Better Theory for {SGD} in the Nonconvex World}.
\newblock \bibinfo{journal}{\emph{Trans. Mach. Learn. Res.}}
  \bibinfo{volume}{2023} (\bibinfo{year}{2023}).
\newblock
\urldef\tempurl%
\url{https://openreview.net/forum?id=AU4qHN2VkS}
\showURL{%
\tempurl}


\bibitem[Khaled et~al\mbox{.}(2023)]%
        {khaled2023}
\bibfield{author}{\bibinfo{person}{Ahmed Khaled}, \bibinfo{person}{Othmane
  Sebbouh}, \bibinfo{person}{Nicolas Loizou}, \bibinfo{person}{Robert~M.
  Gower}, {and} \bibinfo{person}{Peter Richt{\'{a}}rik}.}
  \bibinfo{year}{2023}\natexlab{}.
\newblock \showarticletitle{Unified Analysis of Stochastic Gradient Methods for
  Composite Convex and Smooth Optimization}.
\newblock \bibinfo{journal}{\emph{J. Optim. Theory Appl.}}
  \bibinfo{volume}{199}, \bibinfo{number}{2} (\bibinfo{year}{2023}),
  \bibinfo{pages}{499--540}.
\newblock
\urldef\tempurl%
\url{https://doi.org/10.1007/S10957-023-02297-Y}
\showDOI{\tempurl}


\bibitem[Khirirat et~al\mbox{.}(2018a)]%
        {khirirat2018distributed}
\bibfield{author}{\bibinfo{person}{Sarit Khirirat}, \bibinfo{person}{Hamid~Reza
  Feyzmahdavian}, {and} \bibinfo{person}{Mikael Johansson}.}
  \bibinfo{year}{2018}\natexlab{a}.
\newblock \showarticletitle{Distributed learning with compressed gradients}.
\newblock \bibinfo{journal}{\emph{arXiv preprint arXiv:1806.06573}}
  (\bibinfo{year}{2018}).
\newblock


\bibitem[Khirirat et~al\mbox{.}(2018b)]%
        {DCGD}
\bibfield{author}{\bibinfo{person}{Sarit Khirirat}, \bibinfo{person}{Hamid~Reza
  Feyzmahdavian}, {and} \bibinfo{person}{Mikael Johansson}.}
  \bibinfo{year}{2018}\natexlab{b}.
\newblock \showarticletitle{Distributed learning with compressed gradients}.
\newblock \bibinfo{journal}{\emph{arXiv preprint arXiv:1806.06573}}
  (\bibinfo{year}{2018}).
\newblock


\bibitem[Kluczniak and Santato(2023)]%
        {kluczniak2023circuit}
\bibfield{author}{\bibinfo{person}{Kamil Kluczniak} {and}
  \bibinfo{person}{Giacomo Santato}.} \bibinfo{year}{2023}\natexlab{}.
\newblock \showarticletitle{On Circuit Private, Multikey and Threshold
  Approximate Homomorphic Encryption}.
\newblock \bibinfo{journal}{\emph{{IACR} Cryptol. ePrint Arch.}}
  (\bibinfo{year}{2023}), \bibinfo{pages}{301}.
\newblock
\urldef\tempurl%
\url{https://eprint.iacr.org/2023/301}
\showURL{%
\tempurl}


\bibitem[Koloskova et~al\mbox{.}(2019)]%
        {Koloskova2019chocosgd}
\bibfield{author}{\bibinfo{person}{Anastasia Koloskova},
  \bibinfo{person}{Sebastian~U. Stich}, {and} \bibinfo{person}{Martin Jaggi}.}
  \bibinfo{year}{2019}\natexlab{}.
\newblock \showarticletitle{Decentralized Stochastic Optimization and Gossip
  Algorithms with Compressed Communication}. In
  \bibinfo{booktitle}{\emph{Proceedings of the 36th International Conference on
  Machine Learning, {ICML} 2019, 9-15 June 2019, Long Beach, California,
  {USA}}} \emph{(\bibinfo{series}{Proceedings of Machine Learning Research},
  Vol.~\bibinfo{volume}{97})}, \bibfield{editor}{\bibinfo{person}{Kamalika
  Chaudhuri} {and} \bibinfo{person}{Ruslan Salakhutdinov}} (Eds.).
  \bibinfo{publisher}{{PMLR}}, \bibinfo{pages}{3478--3487}.
\newblock
\urldef\tempurl%
\url{http://proceedings.mlr.press/v97/koloskova19a.html}
\showURL{%
\tempurl}


\bibitem[Kone{\v{c}}n{\'y} et~al\mbox{.}(2016a)]%
        {konevcny2016afederated}
\bibfield{author}{\bibinfo{person}{Jakub Kone{\v{c}}n{\'y}},
  \bibinfo{person}{H.~Brendan McMahan}, \bibinfo{person}{Daniel Ramage}, {and}
  \bibinfo{person}{Peter Richt{\'{a}}rik}.} \bibinfo{year}{2016}\natexlab{a}.
\newblock \showarticletitle{Federated Optimization: Distributed Machine
  Learning for On-Device Intelligence}.
\newblock \bibinfo{journal}{\emph{CoRR}}  \bibinfo{volume}{abs/1610.02527}
  (\bibinfo{year}{2016}).
\newblock
\showeprint[arXiv]{1610.02527}
\urldef\tempurl%
\url{http://arxiv.org/abs/1610.02527}
\showURL{%
\tempurl}


\bibitem[Kone{\v{c}}n{\'y} et~al\mbox{.}(2016b)]%
        {FEDLEARN}
\bibfield{author}{\bibinfo{person}{Jakub Kone{\v{c}}n{\'y}},
  \bibinfo{person}{H.~Brendan McMahan}, \bibinfo{person}{Felix~X. Yu},
  \bibinfo{person}{Peter Richt{\'{a}}rik}, \bibinfo{person}{Ananda~Theertha
  Suresh}, {and} \bibinfo{person}{Dave Bacon}.}
  \bibinfo{year}{2016}\natexlab{b}.
\newblock \showarticletitle{Federated Learning: Strategies for Improving
  Communication Efficiency}.
\newblock \bibinfo{journal}{\emph{CoRR}}  \bibinfo{volume}{abs/1610.05492}.
\newblock
\showeprint[arXiv]{1610.05492}
\urldef\tempurl%
\url{http://arxiv.org/abs/1610.05492}
\showURL{%
\tempurl}


\bibitem[Kone{\v{c}}n{\'y} and Richt{\'{a}}rik(2018)]%
        {konevcny2018randomized}
\bibfield{author}{\bibinfo{person}{Jakub Kone{\v{c}}n{\'y}} {and}
  \bibinfo{person}{Peter Richt{\'{a}}rik}.} \bibinfo{year}{2018}\natexlab{}.
\newblock \showarticletitle{Randomized Distributed Mean Estimation: Accuracy
  vs. Communication}.
\newblock \bibinfo{journal}{\emph{Frontiers Appl. Math. Stat.}}
  \bibinfo{volume}{4} (\bibinfo{year}{2018}), \bibinfo{pages}{62}.
\newblock
\urldef\tempurl%
\url{https://doi.org/10.3389/FAMS.2018.00062}
\showDOI{\tempurl}


\bibitem[Kosinski et~al\mbox{.}(2013)]%
        {kosinski2013private}
\bibfield{author}{\bibinfo{person}{Michal Kosinski}, \bibinfo{person}{David
  Stillwell}, {and} \bibinfo{person}{Thore Graepel}.}
  \bibinfo{year}{2013}\natexlab{}.
\newblock \showarticletitle{Private traits and attributes are predictable from
  digital records of human behavior}.
\newblock \bibinfo{journal}{\emph{Proceedings of the national academy of
  sciences}} \bibinfo{volume}{110}, \bibinfo{number}{15}
  (\bibinfo{year}{2013}), \bibinfo{pages}{5802--5805}.
\newblock


\bibitem[Kotha and Gupta(2018)]%
        {kotha2018iot}
\bibfield{author}{\bibinfo{person}{Harika~Devi Kotha} {and}
  \bibinfo{person}{V~Mnssvkr Gupta}.} \bibinfo{year}{2018}\natexlab{}.
\newblock \showarticletitle{IoT application: a survey}.
\newblock \bibinfo{journal}{\emph{Int. J. Eng. Technol}} \bibinfo{volume}{7},
  \bibinfo{number}{2.7} (\bibinfo{year}{2018}), \bibinfo{pages}{891--896}.
\newblock


\bibitem[Krizhevsky(2014)]%
        {krizhevsky2014one}
\bibfield{author}{\bibinfo{person}{Alex Krizhevsky}.}
  \bibinfo{year}{2014}\natexlab{}.
\newblock \showarticletitle{One weird trick for parallelizing convolutional
  neural networks}.
\newblock \bibinfo{journal}{\emph{CoRR}}  \bibinfo{volume}{abs/1404.5997}
  (\bibinfo{year}{2014}).
\newblock
\showeprint[arXiv]{1404.5997}
\urldef\tempurl%
\url{http://arxiv.org/abs/1404.5997}
\showURL{%
\tempurl}


\bibitem[Krizhevsky and Hinton(2009)]%
        {krizhevsky2009learning}
\bibfield{author}{\bibinfo{person}{Alex Krizhevsky} {and}
  \bibinfo{person}{Geoffrey Hinton}.} \bibinfo{year}{2009}\natexlab{}.
\newblock \bibinfo{booktitle}{\emph{Learning multiple layers of features from
  tiny images}}.
\newblock \bibinfo{type}{{T}echnical {R}eport}~0.
  \bibinfo{institution}{University of Toronto}, \bibinfo{address}{Toronto,
  Ontario}.
\newblock


\bibitem[Krizhevsky et~al\mbox{.}(2012)]%
        {krizhevsky2012}
\bibfield{author}{\bibinfo{person}{Alex Krizhevsky}, \bibinfo{person}{Ilya
  Sutskever}, {and} \bibinfo{person}{Geoffrey~E. Hinton}.}
  \bibinfo{year}{2012}\natexlab{}.
\newblock \showarticletitle{ImageNet Classification with Deep Convolutional
  Neural Networks}. In \bibinfo{booktitle}{\emph{Advances in Neural Information
  Processing Systems 25: 26th Annual Conference on Neural Information
  Processing Systems 2012. Proceedings of a meeting held December 3-6, 2012,
  Lake Tahoe, Nevada, United States}},
  \bibfield{editor}{\bibinfo{person}{Peter~L. Bartlett},
  \bibinfo{person}{Fernando C.~N. Pereira}, \bibinfo{person}{Christopher J.~C.
  Burges}, \bibinfo{person}{L{\'{e}}on Bottou}, {and}
  \bibinfo{person}{Kilian~Q. Weinberger}} (Eds.). \bibinfo{pages}{1106--1114}.
\newblock
\urldef\tempurl%
\url{https://proceedings.neurips.cc/paper/2012/hash/c399862d3b9d6b76c8436e924a68c45b-Abstract.html}
\showURL{%
\tempurl}


\bibitem[Kulkarni et~al\mbox{.}(2020)]%
        {kulkarni2020survey}
\bibfield{author}{\bibinfo{person}{Viraj Kulkarni}, \bibinfo{person}{Milind
  Kulkarni}, {and} \bibinfo{person}{Aniruddha Pant}.}
  \bibinfo{year}{2020}\natexlab{}.
\newblock \showarticletitle{Survey of Personalization Techniques for Federated
  Learning}.
\newblock \bibinfo{journal}{\emph{CoRR}}  \bibinfo{volume}{abs/2003.08673}.
\newblock
\showeprint[arXiv]{2003.08673}
\urldef\tempurl%
\url{https://arxiv.org/abs/2003.08673}
\showURL{%
\tempurl}


\bibitem[Kurose and Ross(2001)]%
        {kurose2005computer}
\bibfield{author}{\bibinfo{person}{James~F. Kurose} {and}
  \bibinfo{person}{Keith~W. Ross}.} \bibinfo{year}{2001}\natexlab{}.
\newblock \bibinfo{booktitle}{\emph{Computer networking - a top-down approach
  featuring the internet}}.
\newblock \bibinfo{publisher}{Addison-Wesley-Longman}.
\newblock
\showISBNx{978-0-201-47711-5}


\bibitem[Laskaridis et~al\mbox{.}(2024)]%
        {laskaridis2024melting}
\bibfield{author}{\bibinfo{person}{Stefanos Laskaridis},
  \bibinfo{person}{Kleomenis Katevas}, \bibinfo{person}{Lorenzo Minto}, {and}
  \bibinfo{person}{Hamed Haddadi}.} \bibinfo{year}{2024}\natexlab{}.
\newblock \showarticletitle{MELTing Point: Mobile Evaluation of Language
  Transformers}. In \bibinfo{booktitle}{\emph{Proceedings of the 30th Annual
  International Conference on Mobile Computing and Networking, {ACM} MobiCom
  2024, Washington D.C., DC, USA, November 18-22, 2024}},
  \bibfield{editor}{\bibinfo{person}{Weisong Shi}, \bibinfo{person}{Deepak
  Ganesan}, {and} \bibinfo{person}{Nicholas~D. Lane}} (Eds.).
  \bibinfo{publisher}{{ACM}}, \bibinfo{pages}{890--907}.
\newblock
\urldef\tempurl%
\url{https://doi.org/10.1145/3636534.3690668}
\showDOI{\tempurl}


\bibitem[Lattner and Adve(2004)]%
        {lattner2004llvm}
\bibfield{author}{\bibinfo{person}{Chris Lattner} {and}
  \bibinfo{person}{Vikram~S. Adve}.} \bibinfo{year}{2004}\natexlab{}.
\newblock \showarticletitle{{LLVM:} {A} Compilation Framework for Lifelong
  Program Analysis {\&} Transformation}. In \bibinfo{booktitle}{\emph{2nd
  {IEEE} / {ACM} International Symposium on Code Generation and Optimization
  {(CGO} 2004), 20-24 March 2004, San Jose, CA, {USA}}}.
  \bibinfo{publisher}{{IEEE} Computer Society}, \bibinfo{pages}{75--88}.
\newblock
\urldef\tempurl%
\url{https://doi.org/10.1109/CGO.2004.1281665}
\showDOI{\tempurl}


\bibitem[Lauter(2022)]%
        {lauter2022private}
\bibfield{author}{\bibinfo{person}{Kristin Lauter}.}
  \bibinfo{year}{2022}\natexlab{}.
\newblock \showarticletitle{Private Artificial Intelligence: Machine Learning
  on Encrypted Data}.
\newblock \bibinfo{journal}{\emph{Collections}} \bibinfo{volume}{55},
  \bibinfo{number}{03} (\bibinfo{year}{2022}).
\newblock


\bibitem[Lauter et~al\mbox{.}(2022)]%
        {lauter2022protecting}
\bibfield{author}{\bibinfo{person}{Kristin~Estella Lauter},
  \bibinfo{person}{Wei Dai}, {and} \bibinfo{person}{Kim Laine}.}
  \bibinfo{year}{2022}\natexlab{}.
\newblock \bibinfo{booktitle}{\emph{Protecting Privacy Through Homomorphic
  Encryption}}.
\newblock \bibinfo{publisher}{Springer}.
\newblock


\bibitem[Lei et~al\mbox{.}(2017)]%
        {lei2017non}
\bibfield{author}{\bibinfo{person}{Lihua Lei}, \bibinfo{person}{Cheng Ju},
  \bibinfo{person}{Jianbo Chen}, {and} \bibinfo{person}{Michael~I. Jordan}.}
  \bibinfo{year}{2017}\natexlab{}.
\newblock \showarticletitle{Non-convex Finite-Sum Optimization Via {SCSG}
  Methods}.
\newblock  (\bibinfo{year}{2017}), \bibinfo{pages}{2348--2358}.
\newblock
\urldef\tempurl%
\url{https://proceedings.neurips.cc/paper/2017/hash/81ca0262c82e712e50c580c032d99b60-Abstract.html}
\showURL{%
\tempurl}


\bibitem[Leiserson et~al\mbox{.}(2020)]%
        {leiserson2020there}
\bibfield{author}{\bibinfo{person}{Charles~E Leiserson},
  \bibinfo{person}{Neil~C Thompson}, \bibinfo{person}{Joel~S Emer},
  \bibinfo{person}{Bradley~C Kuszmaul}, \bibinfo{person}{Butler~W Lampson},
  \bibinfo{person}{Daniel Sanchez}, {and} \bibinfo{person}{Tao~B Schardl}.}
  \bibinfo{year}{2020}\natexlab{}.
\newblock \showarticletitle{There’s plenty of room at the Top: What will
  drive computer performance after Moore’s law?}
\newblock \bibinfo{journal}{\emph{Science}} \bibinfo{volume}{368},
  \bibinfo{number}{6495} (\bibinfo{year}{2020}).
\newblock
\urldef\tempurl%
\url{https://www.science.org/doi/10.1126/science.aam9744}
\showURL{%
\tempurl}


\bibitem[Li et~al\mbox{.}(2021b)]%
        {li2021hermes}
\bibfield{author}{\bibinfo{person}{Ang Li}, \bibinfo{person}{Jingwei Sun},
  \bibinfo{person}{Pengcheng Li}, \bibinfo{person}{Yu Pu}, \bibinfo{person}{Hai
  Li}, {and} \bibinfo{person}{Yiran Chen}.} \bibinfo{year}{2021}\natexlab{b}.
\newblock \showarticletitle{Hermes: an efficient federated learning framework
  for heterogeneous mobile clients}. In \bibinfo{booktitle}{\emph{{ACM} MobiCom
  '21: The 27th Annual International Conference on Mobile Computing and
  Networking, New Orleans, Louisiana, USA, October 25-29, 2021}}.
  \bibinfo{publisher}{{ACM}}, \bibinfo{pages}{420--437}.
\newblock
\urldef\tempurl%
\url{https://doi.org/10.1145/3447993.3483278}
\showDOI{\tempurl}


\bibitem[Li et~al\mbox{.}(2019)]%
        {li2019priority}
\bibfield{author}{\bibinfo{person}{Chen Li}, \bibinfo{person}{Jun Yang},
  \bibinfo{person}{Yifan Sun}, \bibinfo{person}{Lingling Jin},
  \bibinfo{person}{Lingjie Xu}, \bibinfo{person}{Zheng Cao},
  \bibinfo{person}{Pengfei Fan}, \bibinfo{person}{David~R. Kaeli},
  \bibinfo{person}{Sheng Ma}, {and} \bibinfo{person}{Yang Guo}.}
  \bibinfo{year}{2019}\natexlab{}.
\newblock \showarticletitle{Priority-Based PCIe Scheduling for Multi-Tenant
  Multi-GPU Systems}.
\newblock \bibinfo{journal}{\emph{{IEEE} Comput. Archit. Lett.}}
  \bibinfo{volume}{18}, \bibinfo{number}{2} (\bibinfo{year}{2019}),
  \bibinfo{pages}{157--160}.
\newblock
\urldef\tempurl%
\url{https://doi.org/10.1109/LCA.2019.2955119}
\showDOI{\tempurl}


\bibitem[Li et~al\mbox{.}(2020c)]%
        {FL_overview}
\bibfield{author}{\bibinfo{person}{Tian Li}, \bibinfo{person}{Anit~Kumar Sahu},
  \bibinfo{person}{Ameet Talwalkar}, {and} \bibinfo{person}{Virginia Smith}.}
  \bibinfo{year}{2020}\natexlab{c}.
\newblock \showarticletitle{Federated Learning: Challenges, Methods, and Future
  Directions}.
\newblock \bibinfo{journal}{\emph{{IEEE} Signal Process. Mag.}}
  \bibinfo{volume}{37}, \bibinfo{number}{3} (\bibinfo{year}{2020}),
  \bibinfo{pages}{50--60}.
\newblock
\urldef\tempurl%
\url{https://doi.org/10.1109/MSP.2020.2975749}
\showDOI{\tempurl}


\bibitem[Li et~al\mbox{.}(2020d)]%
        {li2018federated}
\bibfield{author}{\bibinfo{person}{Tian Li}, \bibinfo{person}{Anit~Kumar Sahu},
  \bibinfo{person}{Manzil Zaheer}, \bibinfo{person}{Maziar Sanjabi},
  \bibinfo{person}{Ameet Talwalkar}, {and} \bibinfo{person}{Virginia Smith}.}
  \bibinfo{year}{2020}\natexlab{d}.
\newblock \showarticletitle{Federated Optimization in Heterogeneous Networks}.
\newblock  (\bibinfo{year}{2020}).
\newblock
\urldef\tempurl%
\url{https://proceedings.mlsys.org/paper\_files/paper/2020/hash/1f5fe83998a09396ebe6477d9475ba0c-Abstract.html}
\showURL{%
\tempurl}


\bibitem[Li et~al\mbox{.}(2020a)]%
        {li2019convergence}
\bibfield{author}{\bibinfo{person}{Xiang Li}, \bibinfo{person}{Kaixuan Huang},
  \bibinfo{person}{Wenhao Yang}, \bibinfo{person}{Shusen Wang}, {and}
  \bibinfo{person}{Zhihua Zhang}.} \bibinfo{year}{2020}\natexlab{a}.
\newblock \showarticletitle{On the Convergence of FedAvg on Non-IID Data}.
\newblock  (\bibinfo{year}{2020}).
\newblock
\urldef\tempurl%
\url{https://openreview.net/forum?id=HJxNAnVtDS}
\showURL{%
\tempurl}


\bibitem[Li et~al\mbox{.}(2018)]%
        {pipe_sgd}
\bibfield{author}{\bibinfo{person}{Youjie Li}, \bibinfo{person}{Mingchao Yu},
  \bibinfo{person}{Songze Li}, \bibinfo{person}{Salman Avestimehr},
  \bibinfo{person}{Nam~Sung Kim}, {and} \bibinfo{person}{Alexander~G.
  Schwing}.} \bibinfo{year}{2018}\natexlab{}.
\newblock \showarticletitle{Pipe-SGD: {A} Decentralized Pipelined {SGD}
  Framework for Distributed Deep Net Training}. In
  \bibinfo{booktitle}{\emph{Advances in Neural Information Processing Systems
  31: Annual Conference on Neural Information Processing Systems 2018, NeurIPS
  2018, December 3-8, 2018, Montr{\'{e}}al, Canada}},
  \bibfield{editor}{\bibinfo{person}{Samy Bengio}, \bibinfo{person}{Hanna~M.
  Wallach}, \bibinfo{person}{Hugo Larochelle}, \bibinfo{person}{Kristen
  Grauman}, \bibinfo{person}{Nicol{\`{o}} Cesa{-}Bianchi}, {and}
  \bibinfo{person}{Roman Garnett}} (Eds.). \bibinfo{pages}{8056--8067}.
\newblock
\urldef\tempurl%
\url{https://proceedings.neurips.cc/paper/2018/hash/2c6a0bae0f071cbbf0bb3d5b11d90a82-Abstract.html}
\showURL{%
\tempurl}


\bibitem[Li et~al\mbox{.}(2021a)]%
        {li2021page}
\bibfield{author}{\bibinfo{person}{Zhize Li}, \bibinfo{person}{Hongyan Bao},
  \bibinfo{person}{Xiangliang Zhang}, {and} \bibinfo{person}{Peter
  Richt{\'{a}}rik}.} \bibinfo{year}{2021}\natexlab{a}.
\newblock \showarticletitle{{PAGE:} {A} Simple and Optimal Probabilistic
  Gradient Estimator for Nonconvex Optimization}. In
  \bibinfo{booktitle}{\emph{Proceedings of the 38th International Conference on
  Machine Learning, {ICML} 2021, 18-24 July 2021, Virtual Event}}
  \emph{(\bibinfo{series}{Proceedings of Machine Learning Research},
  Vol.~\bibinfo{volume}{139})}, \bibfield{editor}{\bibinfo{person}{Marina
  Meila} {and} \bibinfo{person}{Tong Zhang}} (Eds.).
  \bibinfo{publisher}{{PMLR}}, \bibinfo{pages}{6286--6295}.
\newblock
\urldef\tempurl%
\url{http://proceedings.mlr.press/v139/li21a.html}
\showURL{%
\tempurl}


\bibitem[Li et~al\mbox{.}(2020b)]%
        {ADIANA}
\bibfield{author}{\bibinfo{person}{Zhize Li}, \bibinfo{person}{Dmitry Kovalev},
  \bibinfo{person}{Xun Qian}, {and} \bibinfo{person}{Peter Richt{\'{a}}rik}.}
  \bibinfo{year}{2020}\natexlab{b}.
\newblock \showarticletitle{Acceleration for Compressed Gradient Descent in
  Distributed and Federated Optimization}. In
  \bibinfo{booktitle}{\emph{Proceedings of the 37th International Conference on
  Machine Learning, {ICML} 2020, 13-18 July 2020, Virtual Event}}
  \emph{(\bibinfo{series}{Proceedings of Machine Learning Research},
  Vol.~\bibinfo{volume}{119})}. \bibinfo{publisher}{{PMLR}},
  \bibinfo{pages}{5895--5904}.
\newblock
\urldef\tempurl%
\url{http://proceedings.mlr.press/v119/li20g.html}
\showURL{%
\tempurl}


\bibitem[Limited(2012)]%
        {misra2012}
\bibfield{author}{\bibinfo{person}{MIRA Limited}.}
  \bibinfo{year}{2012}\natexlab{}.
\newblock \bibinfo{booktitle}{\emph{MISRA C:2012: Guidelines for the Use of the
  C Language in Critical Systems}}.
\newblock
\urldef\tempurl%
\url{https://www.misra.org.uk}
\showURL{%
\tempurl}
\newblock
\shownote{Accessed: 2024-12-04}.


\bibitem[Lin et~al\mbox{.}(2022)]%
        {lin-etal-2022-truthfulqa}
\bibfield{author}{\bibinfo{person}{Stephanie Lin}, \bibinfo{person}{Jacob
  Hilton}, {and} \bibinfo{person}{Owain Evans}.}
  \bibinfo{year}{2022}\natexlab{}.
\newblock \showarticletitle{TruthfulQA: Measuring How Models Mimic Human
  Falsehoods}. In \bibinfo{booktitle}{\emph{Proceedings of the 60th Annual
  Meeting of the Association for Computational Linguistics (Volume 1: Long
  Papers), {ACL} 2022, Dublin, Ireland, May 22-27, 2022}},
  \bibfield{editor}{\bibinfo{person}{Smaranda Muresan},
  \bibinfo{person}{Preslav Nakov}, {and} \bibinfo{person}{Aline Villavicencio}}
  (Eds.). \bibinfo{publisher}{Association for Computational Linguistics},
  \bibinfo{pages}{3214--3252}.
\newblock
\urldef\tempurl%
\url{https://doi.org/10.18653/V1/2022.ACL-LONG.229}
\showDOI{\tempurl}


\bibitem[Lin et~al\mbox{.}(2020)]%
        {lin2018don}
\bibfield{author}{\bibinfo{person}{Tao Lin}, \bibinfo{person}{Sebastian~U.
  Stich}, \bibinfo{person}{Kumar~Kshitij Patel}, {and} \bibinfo{person}{Martin
  Jaggi}.} \bibinfo{year}{2020}\natexlab{}.
\newblock \showarticletitle{Don't Use Large Mini-batches, Use Local {SGD}}.
\newblock  (\bibinfo{year}{2020}).
\newblock
\urldef\tempurl%
\url{https://openreview.net/forum?id=B1eyO1BFPr}
\showURL{%
\tempurl}


\bibitem[Linnainmaa(1970)]%
        {linnainmaa1970representation}
\bibfield{author}{\bibinfo{person}{Seppo Linnainmaa}.}
  \bibinfo{year}{1970}\natexlab{}.
\newblock \emph{\bibinfo{title}{The representation of the cumulative rounding
  error of an algorithm as a Taylor expansion of the local rounding errors}}.
\newblock \bibinfo{thesistype}{Ph.\,D. Dissertation}.
  \bibinfo{school}{Master’s Thesis (in Finnish), Univ. Helsinki}.
\newblock


\bibitem[Liu et~al\mbox{.}(2018)]%
        {liu2019secure}
\bibfield{author}{\bibinfo{person}{Changchang Liu}, \bibinfo{person}{Supriyo
  Chakraborty}, {and} \bibinfo{person}{Dinesh~C. Verma}.}
  \bibinfo{year}{2018}\natexlab{}.
\newblock \showarticletitle{Secure Model Fusion for Distributed Learning Using
  Partial Homomorphic Encryption}.
\newblock   \bibinfo{volume}{11550} (\bibinfo{year}{2018}),
  \bibinfo{pages}{154--179}.
\newblock
\urldef\tempurl%
\url{https://doi.org/10.1007/978-3-030-17277-0\_9}
\showDOI{\tempurl}


\bibitem[Liu et~al\mbox{.}(2022a)]%
        {liu2022privacy}
\bibfield{author}{\bibinfo{person}{Ken~Ziyu Liu}, \bibinfo{person}{Shengyuan
  Hu}, \bibinfo{person}{Steven Wu}, {and} \bibinfo{person}{Virginia Smith}.}
  \bibinfo{year}{2022}\natexlab{a}.
\newblock \showarticletitle{On Privacy and Personalization in Cross-Silo
  Federated Learning}.
\newblock  (\bibinfo{year}{2022}).
\newblock
\urldef\tempurl%
\url{http://papers.nips.cc/paper\_files/paper/2022/hash/2788b4cdf421e03650868cc4184bfed8-Abstract-Conference.html}
\showURL{%
\tempurl}


\bibitem[Liu et~al\mbox{.}(2022b)]%
        {liu2022gact}
\bibfield{author}{\bibinfo{person}{Xiaoxuan Liu}, \bibinfo{person}{Lianmin
  Zheng}, \bibinfo{person}{Dequan Wang}, \bibinfo{person}{Yukuo Cen},
  \bibinfo{person}{Weize Chen}, \bibinfo{person}{Xu Han},
  \bibinfo{person}{Jianfei Chen}, \bibinfo{person}{Zhiyuan Liu},
  \bibinfo{person}{Jie Tang}, \bibinfo{person}{Joey Gonzalez},
  \bibinfo{person}{Michael~W. Mahoney}, {and} \bibinfo{person}{Alvin Cheung}.}
  \bibinfo{year}{2022}\natexlab{b}.
\newblock \showarticletitle{{GACT:} Activation Compressed Training for Generic
  Network Architectures}. In \bibinfo{booktitle}{\emph{International Conference
  on Machine Learning, {ICML} 2022, 17-23 July 2022, Baltimore, Maryland,
  {USA}}} \emph{(\bibinfo{series}{Proceedings of Machine Learning Research},
  Vol.~\bibinfo{volume}{162})}, \bibfield{editor}{\bibinfo{person}{Kamalika
  Chaudhuri}, \bibinfo{person}{Stefanie Jegelka}, \bibinfo{person}{Le~Song},
  \bibinfo{person}{Csaba Szepesv{\'{a}}ri}, \bibinfo{person}{Gang Niu}, {and}
  \bibinfo{person}{Sivan Sabato}} (Eds.). \bibinfo{publisher}{{PMLR}},
  \bibinfo{pages}{14139--14152}.
\newblock
\urldef\tempurl%
\url{https://proceedings.mlr.press/v162/liu22v.html}
\showURL{%
\tempurl}


\bibitem[Lorentz(1966)]%
        {lorentz1966approximation}
\bibfield{author}{\bibinfo{person}{George Lorentz}.}
  \bibinfo{year}{1966}\natexlab{}.
\newblock \showarticletitle{Approximation of Functions}.
\newblock \bibinfo{journal}{\emph{Selected Topics in Mathematics}}
  (\bibinfo{year}{1966}).
\newblock


\bibitem[Ma et~al\mbox{.}(2022)]%
        {ma2022layer}
\bibfield{author}{\bibinfo{person}{Xiaosong Ma}, \bibinfo{person}{Jie Zhang},
  \bibinfo{person}{Song Guo}, {and} \bibinfo{person}{Wenchao Xu}.}
  \bibinfo{year}{2022}\natexlab{}.
\newblock \showarticletitle{Layer-wised Model Aggregation for Personalized
  Federated Learning}. In \bibinfo{booktitle}{\emph{{IEEE/CVF} Conference on
  Computer Vision and Pattern Recognition, {CVPR} 2022, New Orleans, LA, USA,
  June 18-24, 2022}}. \bibinfo{publisher}{{IEEE}},
  \bibinfo{pages}{10082--10091}.
\newblock
\urldef\tempurl%
\url{https://doi.org/10.1109/CVPR52688.2022.00985}
\showDOI{\tempurl}


\bibitem[Ma et~al\mbox{.}(2019)]%
        {ma2019paddlepaddle}
\bibfield{author}{\bibinfo{person}{Yanjun Ma}, \bibinfo{person}{Dianhai Yu},
  \bibinfo{person}{Tian Wu}, {and} \bibinfo{person}{Haifeng Wang}.}
  \bibinfo{year}{2019}\natexlab{}.
\newblock \showarticletitle{PaddlePaddle: An Open-Source Deep Learning Platform
  from Industrial Practice}.
\newblock \bibinfo{journal}{\emph{Frontiers of Data and Domputing}}
  \bibinfo{volume}{1}, \bibinfo{number}{1} (\bibinfo{year}{2019}),
  \bibinfo{pages}{105--115}.
\newblock


\bibitem[Maclaurin et~al\mbox{.}(2015)]%
        {maclaurin2015autograd}
\bibfield{author}{\bibinfo{person}{Dougal Maclaurin}, \bibinfo{person}{David
  Duvenaud}, \bibinfo{person}{Matt Johnson}, {and} \bibinfo{person}{Jamie
  Townsend}.} \bibinfo{year}{2015}\natexlab{}.
\newblock \bibinfo{title}{Autograd}.
\newblock
\newblock
\urldef\tempurl%
\url{https://github.com/HIPS/autograd}
\showURL{%
\tempurl}
\newblock
\shownote{Accessed: 2024-12-10}.


\bibitem[Malinovskii et~al\mbox{.}(2024)]%
        {malinovskii2024pv}
\bibfield{author}{\bibinfo{person}{Vladimir Malinovskii},
  \bibinfo{person}{Denis Mazur}, \bibinfo{person}{Ivan Ilin},
  \bibinfo{person}{Denis Kuznedelev}, \bibinfo{person}{Konstantin Burlachenko},
  \bibinfo{person}{Kai Yi}, \bibinfo{person}{Dan Alistarh}, {and}
  \bibinfo{person}{Peter Richt{\'{a}}rik}.} \bibinfo{year}{2024}\natexlab{}.
\newblock \showarticletitle{PV-Tuning: Beyond Straight-Through Estimation for
  Extreme {LLM} Compression}.
\newblock  (\bibinfo{year}{2024}).
\newblock
\urldef\tempurl%
\url{http://papers.nips.cc/paper\_files/paper/2024/hash/091166620a04a289c555f411d8899049-Abstract-Conference.html}
\showURL{%
\tempurl}


\bibitem[Malinovsky et~al\mbox{.}(2023)]%
        {malinovsky2023federated}
\bibfield{author}{\bibinfo{person}{Grigory Malinovsky}, \bibinfo{person}{Samuel
  Horv{\'{a}}th}, \bibinfo{person}{Konstantin Burlachenko}, {and}
  \bibinfo{person}{Peter Richt{\'{a}}rik}.} \bibinfo{year}{2023}\natexlab{}.
\newblock \showarticletitle{Federated Learning with Regularized Client
  Participation}.
\newblock \bibinfo{journal}{\emph{CoRR}}  \bibinfo{volume}{abs/2302.03662}
  (\bibinfo{year}{2023}).
\newblock
\urldef\tempurl%
\url{https://doi.org/10.48550/ARXIV.2302.03662}
\showDOI{\tempurl}
\showeprint[arXiv]{2302.03662}


\bibitem[Maranjyan et~al\mbox{.}(2024)]%
        {maranjyan2024mindflayer}
\bibfield{author}{\bibinfo{person}{Artavazd Maranjyan},
  \bibinfo{person}{Omar~Shaikh Omar}, {and} \bibinfo{person}{Peter
  Richt{\'{a}}rik}.} \bibinfo{year}{2024}\natexlab{}.
\newblock \showarticletitle{MindFlayer: Efficient Asynchronous Parallel {SGD}
  in the Presence of Heterogeneous and Random Worker Compute Times}.
\newblock \bibinfo{journal}{\emph{CoRR}}  \bibinfo{volume}{abs/2410.04285}
  (\bibinfo{year}{2024}).
\newblock
\urldef\tempurl%
\url{https://doi.org/10.48550/ARXIV.2410.04285}
\showDOI{\tempurl}
\showeprint[arXiv]{2410.04285}


\bibitem[Maranjyan et~al\mbox{.}(2025a)]%
        {maranjyan2025ringmaster}
\bibfield{author}{\bibinfo{person}{Artavazd Maranjyan},
  \bibinfo{person}{Alexander Tyurin}, {and} \bibinfo{person}{Peter
  Richt{\'a}rik}.} \bibinfo{year}{2025}\natexlab{a}.
\newblock \showarticletitle{Ringmaster ASGD: The First Asynchronous SGD with
  Optimal Time Complexity}.
\newblock \bibinfo{journal}{\emph{arXiv preprint arXiv:2501.16168}}
  (\bibinfo{year}{2025}).
\newblock


\bibitem[Maranjyan et~al\mbox{.}(2025b)]%
        {maranjyan2025ringmasterasgdasynchronoussgd}
\bibfield{author}{\bibinfo{person}{Artavazd Maranjyan},
  \bibinfo{person}{Alexander Tyurin}, {and} \bibinfo{person}{Peter
  Richtárik}.} \bibinfo{year}{2025}\natexlab{b}.
\newblock \showarticletitle{{R}ingmaster {ASGD}: The First {A}synchronous {SGD}
  with Optimal Time Complexity}.
\newblock \bibinfo{journal}{\emph{arXiv preprint arXiv:2501.16168}}
  (\bibinfo{year}{2025}).
\newblock
\urldef\tempurl%
\url{https://arxiv.org/abs/2501.16168}
\showURL{%
\tempurl}


\bibitem[Markstein(2008)]%
        {IEEE754-2008}
\bibfield{author}{\bibinfo{person}{Peter~W. Markstein}.}
  \bibinfo{year}{2008}\natexlab{}.
\newblock \bibinfo{booktitle}{\emph{The New {IEEE-754} Standard for Floating
  Point Arithmetic}}.
\newblock \bibinfo{type}{{T}echnical {R}eport}.
\newblock
\urldef\tempurl%
\url{http://drops.dagstuhl.de/opus/volltexte/2008/1448}
\showURL{%
\tempurl}


\bibitem[Matsumoto and Nishimura(1998)]%
        {matsumoto1998mersenne}
\bibfield{author}{\bibinfo{person}{Makoto Matsumoto} {and}
  \bibinfo{person}{Takuji Nishimura}.} \bibinfo{year}{1998}\natexlab{}.
\newblock \showarticletitle{Mersenne Twister: {A} 623-Dimensionally
  Equidistributed Uniform Pseudo-Random Number Generator}.
\newblock \bibinfo{journal}{\emph{{ACM} Trans. Model. Comput. Simul.}}
  \bibinfo{volume}{8}, \bibinfo{number}{1} (\bibinfo{year}{1998}),
  \bibinfo{pages}{3--30}.
\newblock
\urldef\tempurl%
\url{https://doi.org/10.1145/272991.272995}
\showDOI{\tempurl}


\bibitem[Mattingley and Boyd(2012)]%
        {mattingley2012cvxgen}
\bibfield{author}{\bibinfo{person}{Jacob Mattingley} {and}
  \bibinfo{person}{Stephen Boyd}.} \bibinfo{year}{2012}\natexlab{}.
\newblock \showarticletitle{CVXGEN: A code generator for embedded convex
  optimization}.
\newblock \bibinfo{journal}{\emph{Optimization and Engineering}}
  \bibinfo{volume}{13} (\bibinfo{year}{2012}), \bibinfo{pages}{1--27}.
\newblock


\bibitem[McMahan et~al\mbox{.}(2017)]%
        {mcmahan17fedavg}
\bibfield{author}{\bibinfo{person}{Brendan McMahan}, \bibinfo{person}{Eider
  Moore}, \bibinfo{person}{Daniel Ramage}, \bibinfo{person}{Seth Hampson},
  {and} \bibinfo{person}{Blaise~Ag{\"{u}}era y Arcas}.}
  \bibinfo{year}{2017}\natexlab{}.
\newblock \showarticletitle{Communication-Efficient Learning of Deep Networks
  from Decentralized Data}. In \bibinfo{booktitle}{\emph{Proceedings of the
  20th International Conference on Artificial Intelligence and Statistics,
  {AISTATS} 2017, 20-22 April 2017, Fort Lauderdale, FL, {USA}}}
  \emph{(\bibinfo{series}{Proceedings of Machine Learning Research},
  Vol.~\bibinfo{volume}{54})}, \bibfield{editor}{\bibinfo{person}{Aarti Singh}
  {and} \bibinfo{person}{Xiaojin~(Jerry) Zhu}} (Eds.).
  \bibinfo{publisher}{{PMLR}}, \bibinfo{pages}{1273--1282}.
\newblock
\urldef\tempurl%
\url{http://proceedings.mlr.press/v54/mcmahan17a.html}
\showURL{%
\tempurl}


\bibitem[McSherry et~al\mbox{.}(2015)]%
        {mcsherry2015scalability}
\bibfield{author}{\bibinfo{person}{Frank McSherry}, \bibinfo{person}{Michael
  Isard}, {and} \bibinfo{person}{Derek~Gordon Murray}.}
  \bibinfo{year}{2015}\natexlab{}.
\newblock \showarticletitle{Scalability! But at what COST?}. In
  \bibinfo{booktitle}{\emph{15th Workshop on Hot Topics in Operating Systems,
  HotOS XV, Kartause Ittingen, Switzerland, May 18-20, 2015}},
  \bibfield{editor}{\bibinfo{person}{George Candea}} (Ed.).
  \bibinfo{publisher}{{USENIX} Association}.
\newblock
\urldef\tempurl%
\url{https://www.usenix.org/conference/hotos15/workshop-program/presentation/mcsherry}
\showURL{%
\tempurl}


\bibitem[Mei et~al\mbox{.}(2021)]%
        {layerwise_personalizedFL}
\bibfield{author}{\bibinfo{person}{Yuan Mei}, \bibinfo{person}{Binbin Guo},
  \bibinfo{person}{Danyang Xiao}, {and} \bibinfo{person}{Weigang Wu}.}
  \bibinfo{year}{2021}\natexlab{}.
\newblock \showarticletitle{FedVF: Personalized Federated Learning Based on
  Layer-wise Parameter Updates with Variable Frequency}. In
  \bibinfo{booktitle}{\emph{{IEEE} International Performance, Computing, and
  Communications Conference, {IPCCC} 2021, Austin, TX, USA, October 29-31,
  2021}}. \bibinfo{publisher}{{IEEE}}, \bibinfo{pages}{1--9}.
\newblock
\urldef\tempurl%
\url{https://doi.org/10.1109/IPCCC51483.2021.9679416}
\showDOI{\tempurl}


\bibitem[MELLODDY(2020)]%
        {melloddy2020}
\bibfield{author}{\bibinfo{person}{MELLODDY}.} \bibinfo{year}{2020}\natexlab{}.
\newblock \bibinfo{title}{MELLODDY Project Meets Its Year One Objective:
  Deployment Of The World’s First Secure Platform For Multi-Task Federated
  Learning In Drug Discovery Among 10 Pharmaceutical Companies}.
\newblock \bibinfo{howpublished}{\url{https://www.melloddy.eu/y1announcement}}.
\newblock


\bibitem[Meng et~al\mbox{.}(2016)]%
        {meng2016mllib}
\bibfield{author}{\bibinfo{person}{Xiangrui Meng}, \bibinfo{person}{Joseph~K.
  Bradley}, \bibinfo{person}{Burak Yavuz}, \bibinfo{person}{Evan~Randall
  Sparks}, \bibinfo{person}{Shivaram Venkataraman}, \bibinfo{person}{Davies
  Liu}, \bibinfo{person}{Jeremy Freeman}, \bibinfo{person}{D.~B. Tsai},
  \bibinfo{person}{Manish Amde}, \bibinfo{person}{Sean Owen},
  \bibinfo{person}{Doris Xin}, \bibinfo{person}{Reynold Xin},
  \bibinfo{person}{Michael~J. Franklin}, \bibinfo{person}{Reza Zadeh},
  \bibinfo{person}{Matei Zaharia}, {and} \bibinfo{person}{Ameet Talwalkar}.}
  \bibinfo{year}{2016}\natexlab{}.
\newblock \showarticletitle{MLlib: Machine Learning in Apache Spark}.
\newblock \bibinfo{journal}{\emph{J. Mach. Learn. Res.}}  \bibinfo{volume}{17}
  (\bibinfo{year}{2016}), \bibinfo{pages}{34:1--34:7}.
\newblock
\urldef\tempurl%
\url{https://jmlr.org/papers/v17/15-237.html}
\showURL{%
\tempurl}


\bibitem[Microsoft(2021)]%
        {seal}
\bibfield{author}{\bibinfo{person}{Microsoft}.}
  \bibinfo{year}{2021}\natexlab{}.
\newblock \bibinfo{title}{Microsoft SEAL}.
\newblock \bibinfo{howpublished}{\url{https://github.com/microsoft/SEAL}}.
\newblock
\newblock
\shownote{Version 4.1}.


\bibitem[Mishchenko et~al\mbox{.}(2019)]%
        {mishchenko2024distributed}
\bibfield{author}{\bibinfo{person}{Konstantin Mishchenko},
  \bibinfo{person}{Eduard Gorbunov}, \bibinfo{person}{Martin Tak{\'{a}}c},
  {and} \bibinfo{person}{Peter Richt{\'{a}}rik}.}
  \bibinfo{year}{2019}\natexlab{}.
\newblock \showarticletitle{Distributed Learning with Compressed Gradient
  Differences}.
\newblock \bibinfo{journal}{\emph{CoRR}}  \bibinfo{volume}{abs/1901.09269}
  (\bibinfo{year}{2019}).
\newblock
\showeprint[arXiv]{1901.09269}
\urldef\tempurl%
\url{http://arxiv.org/abs/1901.09269}
\showURL{%
\tempurl}


\bibitem[Mishchenko et~al\mbox{.}(2022)]%
        {mishchenko2022proxskip}
\bibfield{author}{\bibinfo{person}{Konstantin Mishchenko},
  \bibinfo{person}{Grigory Malinovsky}, \bibinfo{person}{Sebastian~U. Stich},
  {and} \bibinfo{person}{Peter Richt{\'{a}}rik}.}
  \bibinfo{year}{2022}\natexlab{}.
\newblock \showarticletitle{ProxSkip: Yes! Local Gradient Steps Provably Lead
  to Communication Acceleration! Finally!}. In
  \bibinfo{booktitle}{\emph{International Conference on Machine Learning,
  {ICML} 2022, 17-23 July 2022, Baltimore, Maryland, {USA}}}
  \emph{(\bibinfo{series}{Proceedings of Machine Learning Research},
  Vol.~\bibinfo{volume}{162})}, \bibfield{editor}{\bibinfo{person}{Kamalika
  Chaudhuri}, \bibinfo{person}{Stefanie Jegelka}, \bibinfo{person}{Le~Song},
  \bibinfo{person}{Csaba Szepesv{\'{a}}ri}, \bibinfo{person}{Gang Niu}, {and}
  \bibinfo{person}{Sivan Sabato}} (Eds.). \bibinfo{publisher}{{PMLR}},
  \bibinfo{pages}{15750--15769}.
\newblock
\urldef\tempurl%
\url{https://proceedings.mlr.press/v162/mishchenko22b.html}
\showURL{%
\tempurl}


\bibitem[Mishra et~al\mbox{.}(2018)]%
        {mishra2017wrpn}
\bibfield{author}{\bibinfo{person}{Asit~K. Mishra}, \bibinfo{person}{Eriko
  Nurvitadhi}, \bibinfo{person}{Jeffrey~J. Cook}, {and} \bibinfo{person}{Debbie
  Marr}.} \bibinfo{year}{2018}\natexlab{}.
\newblock \showarticletitle{{WRPN:} Wide Reduced-Precision Networks}.
\newblock  (\bibinfo{year}{2018}).
\newblock
\urldef\tempurl%
\url{https://openreview.net/forum?id=B1ZvaaeAZ}
\showURL{%
\tempurl}


\bibitem[Modoranu et~al\mbox{.}(2024)]%
        {modoranu2024microadam}
\bibfield{author}{\bibinfo{person}{Ionut{-}Vlad Modoranu},
  \bibinfo{person}{Mher Safaryan}, \bibinfo{person}{Grigory Malinovsky},
  \bibinfo{person}{Eldar Kurtic}, \bibinfo{person}{Thomas Robert},
  \bibinfo{person}{Peter Richt{\'{a}}rik}, {and} \bibinfo{person}{Dan
  Alistarh}.} \bibinfo{year}{2024}\natexlab{}.
\newblock \showarticletitle{MicroAdam: Accurate Adaptive Optimization with Low
  Space Overhead and Provable Convergence}.
\newblock  (\bibinfo{year}{2024}).
\newblock
\urldef\tempurl%
\url{http://papers.nips.cc/paper\_files/paper/2024/hash/000f947dcaff8fbffcc3f53a1314f358-Abstract-Conference.html}
\showURL{%
\tempurl}


\bibitem[Moore(1920)]%
        {moore1920reciprocal}
\bibfield{author}{\bibinfo{person}{Eliakim~H Moore}.}
  \bibinfo{year}{1920}\natexlab{}.
\newblock \showarticletitle{On the reciprocal of the general algebraic matrix}.
\newblock \bibinfo{journal}{\emph{Bulletin of the american mathematical
  society}}  \bibinfo{volume}{26} (\bibinfo{year}{1920}),
  \bibinfo{pages}{294--295}.
\newblock


\bibitem[Moritz et~al\mbox{.}(2018)]%
        {moritz2018ray}
\bibfield{author}{\bibinfo{person}{Philipp Moritz}, \bibinfo{person}{Robert
  Nishihara}, \bibinfo{person}{Stephanie Wang}, \bibinfo{person}{Alexey
  Tumanov}, \bibinfo{person}{Richard Liaw}, \bibinfo{person}{Eric Liang},
  \bibinfo{person}{Melih Elibol}, \bibinfo{person}{Zongheng Yang},
  \bibinfo{person}{William Paul}, \bibinfo{person}{Michael~I. Jordan}, {and}
  \bibinfo{person}{Ion Stoica}.} \bibinfo{year}{2018}\natexlab{}.
\newblock \showarticletitle{Ray: {A} Distributed Framework for Emerging {AI}
  Applications}. In \bibinfo{booktitle}{\emph{13th {USENIX} Symposium on
  Operating Systems Design and Implementation, {OSDI} 2018, Carlsbad, CA, USA,
  October 8-10, 2018}}, \bibfield{editor}{\bibinfo{person}{Andrea~C.
  Arpaci{-}Dusseau} {and} \bibinfo{person}{Geoff Voelker}} (Eds.).
  \bibinfo{publisher}{{USENIX} Association}, \bibinfo{pages}{561--577}.
\newblock
\urldef\tempurl%
\url{https://www.usenix.org/conference/osdi18/presentation/nishihara}
\showURL{%
\tempurl}


\bibitem[Nagle(1984)]%
        {nagle1984congestion}
\bibfield{author}{\bibinfo{person}{John Nagle}.}
  \bibinfo{year}{1984}\natexlab{}.
\newblock \showarticletitle{Congestion control in {IP/TCP} internetworks}.
\newblock \bibinfo{journal}{\emph{Comput. Commun. Rev.}} \bibinfo{volume}{14},
  \bibinfo{number}{4} (\bibinfo{year}{1984}), \bibinfo{pages}{11--17}.
\newblock
\urldef\tempurl%
\url{https://doi.org/10.1145/1024908.1024910}
\showDOI{\tempurl}


\bibitem[Nesterov(1983)]%
        {nesterov_accelerated}
\bibfield{author}{\bibinfo{person}{Yurii Nesterov}.}
  \bibinfo{year}{1983}\natexlab{}.
\newblock \bibinfo{title}{A method for unconstrained convex minimization
  problem with the rate of convergence O (1/k\^{} 2)}.
\newblock , \bibinfo{numpages}{543}~pages.
\newblock
\urldef\tempurl%
\url{https://cir.nii.ac.jp/crid/1370576118744597902}
\showURL{%
\tempurl}


\bibitem[Nesterov(2018)]%
        {nesterov2018lectures}
\bibfield{author}{\bibinfo{person}{Yurii Nesterov}.}
  \bibinfo{year}{2018}\natexlab{}.
\newblock \bibinfo{booktitle}{\emph{Lectures on Convex Optimization}}.
  Vol.~\bibinfo{volume}{137}.
\newblock \bibinfo{publisher}{Springer}.
\newblock


\bibitem[Nesterov(2004)]%
        {NesterovBook}
\bibfield{author}{\bibinfo{person}{Yurii~E. Nesterov}.}
  \bibinfo{year}{2004}\natexlab{}.
\newblock \bibinfo{booktitle}{\emph{Introductory Lectures on Convex
  Optimization - {A} Basic Course}}. \bibinfo{series}{Applied Optimization},
  Vol.~\bibinfo{volume}{87}.
\newblock \bibinfo{publisher}{Springer}.
\newblock
\showISBNx{978-1-4613-4691-3}
\urldef\tempurl%
\url{https://doi.org/10.1007/978-1-4419-8853-9}
\showDOI{\tempurl}


\bibitem[Nethercote and Seward(2007)]%
        {nethercote2007valgrind}
\bibfield{author}{\bibinfo{person}{Nicholas Nethercote} {and}
  \bibinfo{person}{Julian Seward}.} \bibinfo{year}{2007}\natexlab{}.
\newblock \showarticletitle{Valgrind: a framework for heavyweight dynamic
  binary instrumentation}. In \bibinfo{booktitle}{\emph{Proceedings of the
  {ACM} {SIGPLAN} 2007 Conference on Programming Language Design and
  Implementation, San Diego, California, USA, June 10-13, 2007}},
  \bibfield{editor}{\bibinfo{person}{Jeanne Ferrante} {and}
  \bibinfo{person}{Kathryn~S. McKinley}} (Eds.). \bibinfo{publisher}{{ACM}},
  \bibinfo{pages}{89--100}.
\newblock
\urldef\tempurl%
\url{https://doi.org/10.1145/1250734.1250746}
\showDOI{\tempurl}


\bibitem[Nguyen et~al\mbox{.}(2017)]%
        {SARAH}
\bibfield{author}{\bibinfo{person}{Lam~M. Nguyen}, \bibinfo{person}{Jie Liu},
  \bibinfo{person}{Katya Scheinberg}, {and} \bibinfo{person}{Martin
  Tak{\'{a}}c}.} \bibinfo{year}{2017}\natexlab{}.
\newblock \showarticletitle{{SARAH:} {A} Novel Method for Machine Learning
  Problems Using Stochastic Recursive Gradient}. In
  \bibinfo{booktitle}{\emph{Proceedings of the 34th International Conference on
  Machine Learning, {ICML} 2017, Sydney, NSW, Australia, 6-11 August 2017}}
  \emph{(\bibinfo{series}{Proceedings of Machine Learning Research},
  Vol.~\bibinfo{volume}{70})}, \bibfield{editor}{\bibinfo{person}{Doina Precup}
  {and} \bibinfo{person}{Yee~Whye Teh}} (Eds.). \bibinfo{publisher}{{PMLR}},
  \bibinfo{pages}{2613--2621}.
\newblock
\urldef\tempurl%
\url{http://proceedings.mlr.press/v70/nguyen17b.html}
\showURL{%
\tempurl}


\bibitem[NVIDIA(2019)]%
        {ClaraTraining}
\bibfield{author}{\bibinfo{person}{NVIDIA}.} \bibinfo{year}{2019}\natexlab{}.
\newblock \bibinfo{title}{NVIDIA Clara}.
\newblock
\newblock
\urldef\tempurl%
\url{https://developer.nvidia.com/clara}
\showURL{%
\tempurl}


\bibitem[NVIDIA(2020)]%
        {nvidia2020}
\bibfield{author}{\bibinfo{person}{NVIDIA}.} \bibinfo{year}{2020}\natexlab{}.
\newblock \bibinfo{title}{Triaging COVID-19 Patients: 20 Hospitals in 20 Days
  Build AI Model that Predicts Oxygen Needs}.
\newblock
  \bibinfo{howpublished}{\url{https://blogs.nvidia.com/blog/2020/10/05/federated-learning-covid-oxygen-needs/}}.
\newblock


\bibitem[{NVIDIA Corporation}(2024)]%
        {nvcomp}
\bibfield{author}{\bibinfo{person}{{NVIDIA Corporation}}.}
  \bibinfo{year}{2024}\natexlab{}.
\newblock \bibinfo{title}{nvCOMP: High-Speed Data Compression Using NVIDIA
  GPUs}.
\newblock \bibinfo{howpublished}{\url{https://developer.nvidia.com/nvcomp}}.
\newblock
\newblock
\shownote{Accessed: 2024-11-28}.


\bibitem[O'Donoghue et~al\mbox{.}(2016)]%
        {ocpb:16}
\bibfield{author}{\bibinfo{person}{Brendan O'Donoghue}, \bibinfo{person}{Eric
  Chu}, \bibinfo{person}{Neal Parikh}, {and} \bibinfo{person}{Stephen~P.
  Boyd}.} \bibinfo{year}{2016}\natexlab{}.
\newblock \showarticletitle{Conic Optimization via Operator Splitting and
  Homogeneous Self-Dual Embedding}.
\newblock \bibinfo{journal}{\emph{J. Optim. Theory Appl.}}
  \bibinfo{volume}{169}, \bibinfo{number}{3} (\bibinfo{year}{2016}),
  \bibinfo{pages}{1042--1068}.
\newblock
\urldef\tempurl%
\url{https://doi.org/10.1007/S10957-016-0892-3}
\showDOI{\tempurl}


\bibitem[Oktay et~al\mbox{.}(2021)]%
        {oktay2020randomized}
\bibfield{author}{\bibinfo{person}{Deniz Oktay}, \bibinfo{person}{Nick
  McGreivy}, \bibinfo{person}{Joshua Aduol}, \bibinfo{person}{Alex Beatson},
  {and} \bibinfo{person}{Ryan~P. Adams}.} \bibinfo{year}{2021}\natexlab{}.
\newblock \showarticletitle{Randomized Automatic Differentiation}.
\newblock  (\bibinfo{year}{2021}).
\newblock
\urldef\tempurl%
\url{https://openreview.net/forum?id=xpx9zj7CUlY}
\showURL{%
\tempurl}


\bibitem[Onay and {\"O}zt{\"u}rk(2018)]%
        {onay2018review}
\bibfield{author}{\bibinfo{person}{Ceylan Onay} {and} \bibinfo{person}{Elif
  {\"O}zt{\"u}rk}.} \bibinfo{year}{2018}\natexlab{}.
\newblock \showarticletitle{A review of credit scoring research in the age of
  Big Data}.
\newblock \bibinfo{journal}{\emph{Journal of Financial Regulation and
  Compliance}} \bibinfo{volume}{26}, \bibinfo{number}{3}
  (\bibinfo{year}{2018}), \bibinfo{pages}{382--405}.
\newblock


\bibitem[Owkin(2020)]%
        {owkin2020}
\bibfield{author}{\bibinfo{person}{Owkin}.} \bibinfo{year}{2020}\natexlab{}.
\newblock \bibinfo{title}{Story of the 1st Federated Learning Model at Owkin}.
\newblock
  \bibinfo{howpublished}{\url{https://owkin.com/federated-learning/federated-model/}}.
\newblock


\bibitem[Pan et~al\mbox{.}(2024)]%
        {DBLP:journals/cybersec/PanCWYLW24}
\bibfield{author}{\bibinfo{person}{Yao Pan}, \bibinfo{person}{Zheng Chao},
  \bibinfo{person}{He Wang}, \bibinfo{person}{Jing Yang},
  \bibinfo{person}{Hongjia Li}, {and} \bibinfo{person}{Liming Wang}.}
  \bibinfo{year}{2024}\natexlab{}.
\newblock \showarticletitle{FedSHE: privacy preserving and efficient federated
  learning with adaptive segmented {CKKS} homomorphic encryption}.
\newblock \bibinfo{journal}{\emph{Cybersecur.}} \bibinfo{volume}{7},
  \bibinfo{number}{1} (\bibinfo{year}{2024}), \bibinfo{pages}{40}.
\newblock
\urldef\tempurl%
\url{https://doi.org/10.1186/S42400-024-00232-W}
\showDOI{\tempurl}


\bibitem[Paszke et~al\mbox{.}(2019)]%
        {paszke2019pytorch}
\bibfield{author}{\bibinfo{person}{Adam Paszke}, \bibinfo{person}{Sam Gross},
  \bibinfo{person}{Francisco Massa}, \bibinfo{person}{Adam Lerer},
  \bibinfo{person}{James Bradbury}, \bibinfo{person}{Gregory Chanan},
  \bibinfo{person}{Trevor Killeen}, \bibinfo{person}{Zeming Lin},
  \bibinfo{person}{Natalia Gimelshein}, \bibinfo{person}{Luca Antiga},
  \bibinfo{person}{Alban Desmaison}, \bibinfo{person}{Andreas K{\"{o}}pf},
  \bibinfo{person}{Edward~Z. Yang}, \bibinfo{person}{Zachary DeVito},
  \bibinfo{person}{Martin Raison}, \bibinfo{person}{Alykhan Tejani},
  \bibinfo{person}{Sasank Chilamkurthy}, \bibinfo{person}{Benoit Steiner},
  \bibinfo{person}{Lu Fang}, \bibinfo{person}{Junjie Bai}, {and}
  \bibinfo{person}{Soumith Chintala}.} \bibinfo{year}{2019}\natexlab{}.
\newblock \showarticletitle{PyTorch: An Imperative Style, High-Performance Deep
  Learning Library}.
\newblock  (\bibinfo{year}{2019}), \bibinfo{pages}{8024--8035}.
\newblock
\urldef\tempurl%
\url{https://proceedings.neurips.cc/paper/2019/hash/bdbca288fee7f92f2bfa9f7012727740-Abstract.html}
\showURL{%
\tempurl}


\bibitem[Pedregosa et~al\mbox{.}(2011)]%
        {pedregosa2011scikit}
\bibfield{author}{\bibinfo{person}{Fabian Pedregosa},
  \bibinfo{person}{Ga{\"{e}}l Varoquaux}, \bibinfo{person}{Alexandre Gramfort},
  \bibinfo{person}{Vincent Michel}, \bibinfo{person}{Bertrand Thirion},
  \bibinfo{person}{Olivier Grisel}, \bibinfo{person}{Mathieu Blondel},
  \bibinfo{person}{Peter Prettenhofer}, \bibinfo{person}{Ron Weiss},
  \bibinfo{person}{Vincent Dubourg}, \bibinfo{person}{Jake VanderPlas},
  \bibinfo{person}{Alexandre Passos}, \bibinfo{person}{David Cournapeau},
  \bibinfo{person}{Matthieu Brucher}, \bibinfo{person}{Matthieu Perrot}, {and}
  \bibinfo{person}{Edouard Duchesnay}.} \bibinfo{year}{2011}\natexlab{}.
\newblock \showarticletitle{Scikit-learn: Machine Learning in Python}.
\newblock \bibinfo{journal}{\emph{J. Mach. Learn. Res.}}  \bibinfo{volume}{12}
  (\bibinfo{year}{2011}), \bibinfo{pages}{2825--2830}.
\newblock
\urldef\tempurl%
\url{https://doi.org/10.5555/1953048.2078195}
\showDOI{\tempurl}


\bibitem[Pereira et~al\mbox{.}(2017)]%
        {pereira2017energy}
\bibfield{author}{\bibinfo{person}{Rui Pereira}, \bibinfo{person}{Marco Couto},
  \bibinfo{person}{Francisco Ribeiro}, \bibinfo{person}{Rui Rua},
  \bibinfo{person}{J{\'{a}}come Cunha}, \bibinfo{person}{Jo{\~{a}}o~Paulo
  Fernandes}, {and} \bibinfo{person}{Jo{\~{a}}o Saraiva}.}
  \bibinfo{year}{2017}\natexlab{}.
\newblock \showarticletitle{Energy efficiency across programming languages: how
  do energy, time, and memory relate?}. In
  \bibinfo{booktitle}{\emph{Proceedings of the 10th {ACM} {SIGPLAN}
  International Conference on Software Language Engineering, {SLE} 2017,
  Vancouver, BC, Canada, October 23-24, 2017}},
  \bibfield{editor}{\bibinfo{person}{Beno{\^{\i}}t Combemale},
  \bibinfo{person}{Marjan Mernik}, {and} \bibinfo{person}{Bernhard Rumpe}}
  (Eds.). \bibinfo{publisher}{{ACM}}, \bibinfo{pages}{256--267}.
\newblock
\urldef\tempurl%
\url{https://doi.org/10.1145/3136014.3136031}
\showDOI{\tempurl}


\bibitem[Pereira et~al\mbox{.}(2021)]%
        {pereira2021ranking}
\bibfield{author}{\bibinfo{person}{Rui Pereira}, \bibinfo{person}{Marco Couto},
  \bibinfo{person}{Francisco Ribeiro}, \bibinfo{person}{Rui Rua},
  \bibinfo{person}{J{\'{a}}come Cunha}, \bibinfo{person}{Jo{\~{a}}o~Paulo
  Fernandes}, {and} \bibinfo{person}{Jo{\~{a}}o Saraiva}.}
  \bibinfo{year}{2021}\natexlab{}.
\newblock \showarticletitle{Ranking programming languages by energy
  efficiency}.
\newblock \bibinfo{journal}{\emph{Sci. Comput. Program.}}
  \bibinfo{volume}{205} (\bibinfo{year}{2021}), \bibinfo{pages}{102609}.
\newblock
\urldef\tempurl%
\url{https://doi.org/10.1016/J.SCICO.2021.102609}
\showDOI{\tempurl}


\bibitem[Philippenko and Dieuleveut(2020)]%
        {artemis}
\bibfield{author}{\bibinfo{person}{Constantin Philippenko} {and}
  \bibinfo{person}{Aymeric Dieuleveut}.} \bibinfo{year}{2020}\natexlab{}.
\newblock \showarticletitle{Artemis: tight convergence guarantees for
  bidirectional compression in Federated Learning}.
\newblock \bibinfo{journal}{\emph{CoRR}}  \bibinfo{volume}{abs/2006.14591}
  (\bibinfo{year}{2020}).
\newblock
\showeprint[arXiv]{2006.14591}
\urldef\tempurl%
\url{https://arxiv.org/abs/2006.14591}
\showURL{%
\tempurl}


\bibitem[Phong et~al\mbox{.}(2018)]%
        {aono2017privacy}
\bibfield{author}{\bibinfo{person}{Le~Trieu Phong}, \bibinfo{person}{Yoshinori
  Aono}, \bibinfo{person}{Takuya Hayashi}, \bibinfo{person}{Lihua Wang}, {and}
  \bibinfo{person}{Shiho Moriai}.} \bibinfo{year}{2018}\natexlab{}.
\newblock \showarticletitle{Privacy-Preserving Deep Learning via Additively
  Homomorphic Encryption}.
\newblock \bibinfo{journal}{\emph{{IEEE} Trans. Inf. Forensics Secur.}}
  \bibinfo{volume}{13}, \bibinfo{number}{5} (\bibinfo{year}{2018}),
  \bibinfo{pages}{1333--1345}.
\newblock
\urldef\tempurl%
\url{https://doi.org/10.1109/TIFS.2017.2787987}
\showDOI{\tempurl}


\bibitem[Pillutla et~al\mbox{.}(2022)]%
        {pillutla2022federated}
\bibfield{author}{\bibinfo{person}{Krishna Pillutla}, \bibinfo{person}{Kshitiz
  Malik}, \bibinfo{person}{Abdelrahman Mohamed}, \bibinfo{person}{Michael~G.
  Rabbat}, \bibinfo{person}{Maziar Sanjabi}, {and} \bibinfo{person}{Lin Xiao}.}
  \bibinfo{year}{2022}\natexlab{}.
\newblock \showarticletitle{Federated Learning with Partial Model
  Personalization}. In \bibinfo{booktitle}{\emph{International Conference on
  Machine Learning, {ICML} 2022, 17-23 July 2022, Baltimore, Maryland, {USA}}}
  \emph{(\bibinfo{series}{Proceedings of Machine Learning Research},
  Vol.~\bibinfo{volume}{162})}, \bibfield{editor}{\bibinfo{person}{Kamalika
  Chaudhuri}, \bibinfo{person}{Stefanie Jegelka}, \bibinfo{person}{Le~Song},
  \bibinfo{person}{Csaba Szepesv{\'{a}}ri}, \bibinfo{person}{Gang Niu}, {and}
  \bibinfo{person}{Sivan Sabato}} (Eds.). \bibinfo{publisher}{{PMLR}},
  \bibinfo{pages}{17716--17758}.
\newblock
\urldef\tempurl%
\url{https://proceedings.mlr.press/v162/pillutla22a.html}
\showURL{%
\tempurl}


\bibitem[Pinto and Santos(2019)]%
        {pinto2019demystifying}
\bibfield{author}{\bibinfo{person}{Sandro Pinto} {and} \bibinfo{person}{Nuno
  Santos}.} \bibinfo{year}{2019}\natexlab{}.
\newblock \showarticletitle{Demystifying Arm TrustZone: {A} Comprehensive
  Survey}.
\newblock \bibinfo{journal}{\emph{{ACM} Comput. Surv.}} \bibinfo{volume}{51},
  \bibinfo{number}{6} (\bibinfo{year}{2019}), \bibinfo{pages}{130:1--130:36}.
\newblock
\urldef\tempurl%
\url{https://doi.org/10.1145/3291047}
\showDOI{\tempurl}


\bibitem[Poplin et~al\mbox{.}(2017)]%
        {cardio_vas_DL}
\bibfield{author}{\bibinfo{person}{Ryan Poplin}, \bibinfo{person}{Avinash~V.
  Varadarajan}, \bibinfo{person}{Katy Blumer}, \bibinfo{person}{Yun Liu},
  \bibinfo{person}{Michael~V. McConnell}, \bibinfo{person}{Gregory~S. Corrado},
  \bibinfo{person}{Lily Peng}, {and} \bibinfo{person}{Dale~R. Webster}.}
  \bibinfo{year}{2017}\natexlab{}.
\newblock \showarticletitle{Predicting Cardiovascular Risk Factors from Retinal
  Fundus Photographs using Deep Learning}.
\newblock \bibinfo{journal}{\emph{CoRR}}  \bibinfo{volume}{abs/1708.09843}
  (\bibinfo{year}{2017}).
\newblock
\showeprint[arXiv]{1708.09843}
\urldef\tempurl%
\url{http://arxiv.org/abs/1708.09843}
\showURL{%
\tempurl}


\bibitem[Qian et~al\mbox{.}(2021)]%
        {qian2021svrg}
\bibfield{author}{\bibinfo{person}{Xun Qian}, \bibinfo{person}{Zheng Qu}, {and}
  \bibinfo{person}{Peter Richt{\'{a}}rik}.} \bibinfo{year}{2021}\natexlab{}.
\newblock \showarticletitle{{L-SVRG} and L-Katyusha with Arbitrary Sampling}.
\newblock \bibinfo{journal}{\emph{J. Mach. Learn. Res.}}  \bibinfo{volume}{22}
  (\bibinfo{year}{2021}), \bibinfo{pages}{112:1--112:47}.
\newblock
\urldef\tempurl%
\url{https://jmlr.org/papers/v22/20-156.html}
\showURL{%
\tempurl}


\bibitem[Qiu et~al\mbox{.}(2021)]%
        {qiu2021first}
\bibfield{author}{\bibinfo{person}{Xinchi Qiu}, \bibinfo{person}{Titouan
  Parcollet}, \bibinfo{person}{Javier Fern{\'{a}}ndez{-}Marqu{\'{e}}s},
  \bibinfo{person}{Pedro Porto~Buarque de Gusm{\~{a}}o},
  \bibinfo{person}{Daniel~J. Beutel}, \bibinfo{person}{Taner Topal},
  \bibinfo{person}{Akhil Mathur}, {and} \bibinfo{person}{Nicholas~D. Lane}.}
  \bibinfo{year}{2021}\natexlab{}.
\newblock \showarticletitle{A first look into the carbon footprint of federated
  learning}.
\newblock \bibinfo{journal}{\emph{CoRR}}  \bibinfo{volume}{abs/2102.07627}
  (\bibinfo{year}{2021}).
\newblock
\showeprint[arXiv]{2102.07627}
\urldef\tempurl%
\url{https://arxiv.org/abs/2102.07627}
\showURL{%
\tempurl}


\bibitem[Rahaman and Hossain(2008)]%
        {rahaman2008side}
\bibfield{author}{\bibinfo{person}{Mohammad~Zahidur Rahaman} {and}
  \bibinfo{person}{Mohammad~Akram Hossain}.} \bibinfo{year}{2008}\natexlab{}.
\newblock \showarticletitle{Side channel attack prevention for AES smart card}.
  In \bibinfo{booktitle}{\emph{2008 11th International Conference on Computer
  and Information Technology}}. IEEE, \bibinfo{pages}{376--380}.
\newblock


\bibitem[Rall(1981)]%
        {rall1981automatic}
\bibfield{author}{\bibinfo{person}{Louis~B. Rall}.}
  \bibinfo{year}{1981}\natexlab{}.
\newblock \bibinfo{booktitle}{\emph{Automatic Differentiation: Techniques and
  Applications}}. \bibinfo{series}{Lecture Notes in Computer Science},
  Vol.~\bibinfo{volume}{120}.
\newblock \bibinfo{publisher}{Springer}.
\newblock
\showISBNx{3-540-10861-0}
\urldef\tempurl%
\url{https://doi.org/10.1007/3-540-10861-0}
\showDOI{\tempurl}


\bibitem[Ramaswamy et~al\mbox{.}(2019)]%
        {gboard19emoji}
\bibfield{author}{\bibinfo{person}{Swaroop Ramaswamy}, \bibinfo{person}{Rajiv
  Mathews}, \bibinfo{person}{Kanishka Rao}, {and}
  \bibinfo{person}{Fran{\c{c}}oise Beaufays}.} \bibinfo{year}{2019}\natexlab{}.
\newblock \showarticletitle{Federated Learning for Emoji Prediction in a Mobile
  Keyboard}.
\newblock \bibinfo{journal}{\emph{CoRR}}  \bibinfo{volume}{abs/1906.04329}
  (\bibinfo{year}{2019}).
\newblock
\showeprint[arXiv]{1906.04329}
\urldef\tempurl%
\url{http://arxiv.org/abs/1906.04329}
\showURL{%
\tempurl}


\bibitem[Reddi et~al\mbox{.}(2020)]%
        {reddi2020adaptive}
\bibfield{author}{\bibinfo{person}{Sashank~J. Reddi}, \bibinfo{person}{Zachary
  Charles}, \bibinfo{person}{Manzil Zaheer}, \bibinfo{person}{Zachary Garrett},
  \bibinfo{person}{Keith Rush}, \bibinfo{person}{Jakub Kone{\v{c}}n{\'y}},
  \bibinfo{person}{Sanjiv Kumar}, {and} \bibinfo{person}{H.~Brendan McMahan}.}
  \bibinfo{year}{2020}\natexlab{}.
\newblock \showarticletitle{Adaptive Federated Optimization}.
\newblock \bibinfo{journal}{\emph{CoRR}}  \bibinfo{volume}{abs/2003.00295}
  (\bibinfo{year}{2020}).
\newblock
\showeprint[arXiv]{2003.00295}
\urldef\tempurl%
\url{https://arxiv.org/abs/2003.00295}
\showURL{%
\tempurl}


\bibitem[Regev(2009)]%
        {regev2009lattices}
\bibfield{author}{\bibinfo{person}{Oded Regev}.}
  \bibinfo{year}{2009}\natexlab{}.
\newblock \showarticletitle{On lattices, learning with errors, random linear
  codes, and cryptography}.
\newblock \bibinfo{journal}{\emph{J. {ACM}}} \bibinfo{volume}{56},
  \bibinfo{number}{6} (\bibinfo{year}{2009}), \bibinfo{pages}{34:1--34:40}.
\newblock
\urldef\tempurl%
\url{https://doi.org/10.1145/1568318.1568324}
\showDOI{\tempurl}


\bibitem[Reina et~al\mbox{.}(2021)]%
        {reina2021openfl}
\bibfield{author}{\bibinfo{person}{G.~Anthony Reina}, \bibinfo{person}{Alexey
  Gruzdev}, \bibinfo{person}{Patrick Foley}, \bibinfo{person}{Olga
  Perepelkina}, \bibinfo{person}{Mansi Sharma}, \bibinfo{person}{Igor
  Davidyuk}, \bibinfo{person}{Ilya Trushkin}, \bibinfo{person}{Maksim
  Radionov}, \bibinfo{person}{Aleksandr Mokrov}, \bibinfo{person}{Dmitry
  Agapov}, \bibinfo{person}{Jason Martin}, \bibinfo{person}{Brandon Edwards},
  \bibinfo{person}{Micah~J. Sheller}, \bibinfo{person}{Sarthak Pati},
  \bibinfo{person}{Prakash~Narayana Moorthy}, \bibinfo{person}{Hans~Shih{-}Han
  Wang}, \bibinfo{person}{Prashant Shah}, {and} \bibinfo{person}{Spyridon
  Bakas}.} \bibinfo{year}{2021}\natexlab{}.
\newblock \showarticletitle{OpenFL: An open-source framework for Federated
  Learning}.
\newblock \bibinfo{journal}{\emph{CoRR}}  \bibinfo{volume}{abs/2105.06413}
  (\bibinfo{year}{2021}).
\newblock
\showeprint[arXiv]{2105.06413}
\urldef\tempurl%
\url{https://arxiv.org/abs/2105.06413}
\showURL{%
\tempurl}


\bibitem[Reisizadeh et~al\mbox{.}(2020)]%
        {fedpaq}
\bibfield{author}{\bibinfo{person}{Amirhossein Reisizadeh},
  \bibinfo{person}{Aryan Mokhtari}, \bibinfo{person}{Hamed Hassani},
  \bibinfo{person}{Ali Jadbabaie}, {and} \bibinfo{person}{Ramtin Pedarsani}.}
  \bibinfo{year}{2020}\natexlab{}.
\newblock \showarticletitle{FedPAQ: {A} Communication-Efficient Federated
  Learning Method with Periodic Averaging and Quantization}. In
  \bibinfo{booktitle}{\emph{The 23rd International Conference on Artificial
  Intelligence and Statistics, {AISTATS} 2020, 26-28 August 2020, Online
  [Palermo, Sicily, Italy]}} \emph{(\bibinfo{series}{Proceedings of Machine
  Learning Research}, Vol.~\bibinfo{volume}{108})},
  \bibfield{editor}{\bibinfo{person}{Silvia Chiappa} {and}
  \bibinfo{person}{Roberto Calandra}} (Eds.). \bibinfo{publisher}{{PMLR}},
  \bibinfo{pages}{2021--2031}.
\newblock
\urldef\tempurl%
\url{http://proceedings.mlr.press/v108/reisizadeh20a.html}
\showURL{%
\tempurl}


\bibitem[Richt{\'{a}}rik et~al\mbox{.}(2023)]%
        {richtarik2023error}
\bibfield{author}{\bibinfo{person}{Peter Richt{\'{a}}rik},
  \bibinfo{person}{Elnur Gasanov}, {and} \bibinfo{person}{Konstantin
  Burlachenko}.} \bibinfo{year}{2023}\natexlab{}.
\newblock \showarticletitle{Error Feedback Shines when Features are Rare}.
\newblock \bibinfo{journal}{\emph{CoRR}}  \bibinfo{volume}{abs/2305.15264}
  (\bibinfo{year}{2023}).
\newblock
\urldef\tempurl%
\url{https://doi.org/10.48550/ARXIV.2305.15264}
\showDOI{\tempurl}
\showeprint[arXiv]{2305.15264}


\bibitem[Richt{\'{a}}rik et~al\mbox{.}(2024)]%
        {richtarik2024error}
\bibfield{author}{\bibinfo{person}{Peter Richt{\'{a}}rik},
  \bibinfo{person}{Elnur Gasanov}, {and} \bibinfo{person}{Konstantin
  Burlachenko}.} \bibinfo{year}{2024}\natexlab{}.
\newblock \showarticletitle{Error Feedback Reloaded: From Quadratic to
  Arithmetic Mean of Smoothness Constants}.
\newblock  (\bibinfo{year}{2024}).
\newblock
\urldef\tempurl%
\url{https://openreview.net/forum?id=Ch7WqGcGmb}
\showURL{%
\tempurl}


\bibitem[Richt{\'{a}}rik et~al\mbox{.}(2021a)]%
        {EF21}
\bibfield{author}{\bibinfo{person}{Peter Richt{\'{a}}rik},
  \bibinfo{person}{Igor Sokolov}, {and} \bibinfo{person}{Ilyas Fatkhullin}.}
  \bibinfo{year}{2021}\natexlab{a}.
\newblock \showarticletitle{{EF21:} {A} New, Simpler, Theoretically Better, and
  Practically Faster Error Feedback}. In \bibinfo{booktitle}{\emph{Advances in
  Neural Information Processing Systems 34: Annual Conference on Neural
  Information Processing Systems 2021, NeurIPS 2021, December 6-14, 2021,
  virtual}}, \bibfield{editor}{\bibinfo{person}{Marc'Aurelio Ranzato},
  \bibinfo{person}{Alina Beygelzimer}, \bibinfo{person}{Yann~N. Dauphin},
  \bibinfo{person}{Percy Liang}, {and} \bibinfo{person}{Jennifer~Wortman
  Vaughan}} (Eds.). \bibinfo{pages}{4384--4396}.
\newblock
\urldef\tempurl%
\url{https://proceedings.neurips.cc/paper/2021/hash/231141b34c82aa95e48810a9d1b33a79-Abstract.html}
\showURL{%
\tempurl}


\bibitem[Richt{\'{a}}rik et~al\mbox{.}(2021b)]%
        {richtarik2021ef21}
\bibfield{author}{\bibinfo{person}{Peter Richt{\'{a}}rik},
  \bibinfo{person}{Igor Sokolov}, {and} \bibinfo{person}{Ilyas Fatkhullin}.}
  \bibinfo{year}{2021}\natexlab{b}.
\newblock \showarticletitle{{EF21:} {A} New, Simpler, Theoretically Better, and
  Practically Faster Error Feedback}. In \bibinfo{booktitle}{\emph{Advances in
  Neural Information Processing Systems 34: Annual Conference on Neural
  Information Processing Systems 2021, NeurIPS 2021, December 6-14, 2021,
  virtual}}, \bibfield{editor}{\bibinfo{person}{Marc'Aurelio Ranzato},
  \bibinfo{person}{Alina Beygelzimer}, \bibinfo{person}{Yann~N. Dauphin},
  \bibinfo{person}{Percy Liang}, {and} \bibinfo{person}{Jennifer~Wortman
  Vaughan}} (Eds.). \bibinfo{pages}{4384--4396}.
\newblock
\urldef\tempurl%
\url{https://proceedings.neurips.cc/paper/2021/hash/231141b34c82aa95e48810a9d1b33a79-Abstract.html}
\showURL{%
\tempurl}


\bibitem[Richt{\'{a}}rik and Tak{\'{a}}c(2016)]%
        {richtarik2016parallel}
\bibfield{author}{\bibinfo{person}{Peter Richt{\'{a}}rik} {and}
  \bibinfo{person}{Martin Tak{\'{a}}c}.} \bibinfo{year}{2016}\natexlab{}.
\newblock \showarticletitle{Parallel coordinate descent methods for big data
  optimization}.
\newblock \bibinfo{journal}{\emph{Math. Program.}} \bibinfo{volume}{156},
  \bibinfo{number}{1-2} (\bibinfo{year}{2016}), \bibinfo{pages}{433--484}.
\newblock
\urldef\tempurl%
\url{https://doi.org/10.1007/S10107-015-0901-6}
\showDOI{\tempurl}


\bibitem[Riley and Henderson(2010)]%
        {riley2010ns}
\bibfield{author}{\bibinfo{person}{George~F. Riley} {and}
  \bibinfo{person}{Thomas~R. Henderson}.} \bibinfo{year}{2010}\natexlab{}.
\newblock \showarticletitle{The \emph{ns-3} Network Simulator}.
\newblock In \bibinfo{booktitle}{\emph{Modeling and Tools for Network
  Simulation}}, \bibfield{editor}{\bibinfo{person}{Klaus Wehrle},
  \bibinfo{person}{Mesut G{\"{u}}nes}, {and} \bibinfo{person}{James Gross}}
  (Eds.). \bibinfo{publisher}{Springer}, \bibinfo{pages}{15--34}.
\newblock
\urldef\tempurl%
\url{https://doi.org/10.1007/978-3-642-12331-3\_2}
\showDOI{\tempurl}


\bibitem[Riverbank(2016)]%
        {pyqt_docu}
\bibfield{author}{\bibinfo{person}{Riverbank}.}
  \bibinfo{year}{2016}\natexlab{}.
\newblock \bibinfo{title}{PyQt}.
\newblock
\newblock
\urldef\tempurl%
\url{https://www.riverbankcomputing.com/software/pyqt/}
\showURL{%
\tempurl}


\bibitem[Rivest et~al\mbox{.}(1978)]%
        {rivest1978data}
\bibfield{author}{\bibinfo{person}{Ronald~L Rivest}, \bibinfo{person}{Len
  Adleman}, \bibinfo{person}{Michael~L Dertouzos}, {et~al\mbox{.}}}
  \bibinfo{year}{1978}\natexlab{}.
\newblock \showarticletitle{On data banks and privacy homomorphisms}.
\newblock \bibinfo{journal}{\emph{Foundations of secure computation}}
  \bibinfo{volume}{4}, \bibinfo{number}{11} (\bibinfo{year}{1978}),
  \bibinfo{pages}{169--180}.
\newblock


\bibitem[Ro et~al\mbox{.}(2021)]%
        {ro2021fedjax}
\bibfield{author}{\bibinfo{person}{Jae~Hun Ro},
  \bibinfo{person}{Ananda~Theertha Suresh}, {and} \bibinfo{person}{Ke Wu}.}
  \bibinfo{year}{2021}\natexlab{}.
\newblock \showarticletitle{FedJAX: Federated learning simulation with {JAX}}.
\newblock \bibinfo{journal}{\emph{CoRR}}  \bibinfo{volume}{abs/2108.02117}
  (\bibinfo{year}{2021}).
\newblock
\showeprint[arXiv]{2108.02117}
\urldef\tempurl%
\url{https://arxiv.org/abs/2108.02117}
\showURL{%
\tempurl}


\bibitem[Rocher et~al\mbox{.}(2019)]%
        {rocher2019estimating}
\bibfield{author}{\bibinfo{person}{Luc Rocher}, \bibinfo{person}{Julien~M
  Hendrickx}, {and} \bibinfo{person}{Yves-Alexandre De~Montjoye}.}
  \bibinfo{year}{2019}\natexlab{}.
\newblock \showarticletitle{Estimating the success of re-identifications in
  incomplete datasets using generative models}.
\newblock \bibinfo{journal}{\emph{Nature communications}} \bibinfo{volume}{10},
  \bibinfo{number}{1} (\bibinfo{year}{2019}), \bibinfo{pages}{1--9}.
\newblock


\bibitem[Rosenblatt(1958)]%
        {rosenblatt1958perceptron}
\bibfield{author}{\bibinfo{person}{Frank Rosenblatt}.}
  \bibinfo{year}{1958}\natexlab{}.
\newblock \showarticletitle{The perceptron: a probabilistic model for
  information storage and organization in the brain.}
\newblock \bibinfo{journal}{\emph{Psychological review}} \bibinfo{volume}{65},
  \bibinfo{number}{6} (\bibinfo{year}{1958}), \bibinfo{pages}{386}.
\newblock


\bibitem[Rossum et~al\mbox{.}(1995)]%
        {van1995python}
\bibfield{author}{\bibinfo{person}{Guido Rossum} {et~al\mbox{.}}}
  \bibinfo{year}{1995}\natexlab{}.
\newblock \bibinfo{booktitle}{\emph{Python reference manual}}.
  Vol.~\bibinfo{volume}{111}.
\newblock \bibinfo{publisher}{Centrum voor Wiskunde en Informatica Amsterdam}.
\newblock


\bibitem[Roth et~al\mbox{.}(2023)]%
        {roth2022nvidia}
\bibfield{author}{\bibinfo{person}{Holger~R. Roth}, \bibinfo{person}{Yan
  Cheng}, \bibinfo{person}{Yuhong Wen}, \bibinfo{person}{Isaac Yang},
  \bibinfo{person}{Ziyue Xu}, \bibinfo{person}{Yuan{-}Ting Hsieh},
  \bibinfo{person}{Kristopher Kersten}, \bibinfo{person}{Ahmed Harouni},
  \bibinfo{person}{Can Zhao}, \bibinfo{person}{Kevin Lu},
  \bibinfo{person}{Zhihong Zhang}, \bibinfo{person}{Wenqi Li},
  \bibinfo{person}{Andriy Myronenko}, \bibinfo{person}{Dong Yang},
  \bibinfo{person}{Sean Yang}, \bibinfo{person}{Nicola Rieke},
  \bibinfo{person}{Abood Quraini}, \bibinfo{person}{Chester Chen},
  \bibinfo{person}{Daguang Xu}, \bibinfo{person}{Nic Ma},
  \bibinfo{person}{Prerna Dogra}, \bibinfo{person}{Mona Flores}, {and}
  \bibinfo{person}{Andrew Feng}.} \bibinfo{year}{2023}\natexlab{}.
\newblock \showarticletitle{{NVIDIA} {FLARE:} Federated Learning from
  Simulation to Real-World}.
\newblock \bibinfo{journal}{\emph{{IEEE} Data Eng. Bull.}}
  \bibinfo{volume}{46}, \bibinfo{number}{1} (\bibinfo{year}{2023}),
  \bibinfo{pages}{170--184}.
\newblock
\urldef\tempurl%
\url{http://sites.computer.org/debull/A23mar/p170.pdf}
\showURL{%
\tempurl}


\bibitem[Rumelhart et~al\mbox{.}(1986)]%
        {rumelhart1986learning}
\bibfield{author}{\bibinfo{person}{David~E Rumelhart},
  \bibinfo{person}{Geoffrey~E Hinton}, {and} \bibinfo{person}{Ronald~J
  Williams}.} \bibinfo{year}{1986}\natexlab{}.
\newblock \showarticletitle{Learning representations by back-propagating
  errors}.
\newblock \bibinfo{journal}{\emph{nature}} \bibinfo{volume}{323},
  \bibinfo{number}{6088} (\bibinfo{year}{1986}), \bibinfo{pages}{533--536}.
\newblock


\bibitem[Ryffel et~al\mbox{.}(2018)]%
        {ryffel2018generic}
\bibfield{author}{\bibinfo{person}{Th{\'{e}}o Ryffel}, \bibinfo{person}{Andrew
  Trask}, \bibinfo{person}{Morten Dahl}, \bibinfo{person}{Bobby Wagner},
  \bibinfo{person}{Jason Mancuso}, \bibinfo{person}{Daniel Rueckert}, {and}
  \bibinfo{person}{Jonathan Passerat{-}Palmbach}.}
  \bibinfo{year}{2018}\natexlab{}.
\newblock \showarticletitle{A generic framework for privacy preserving deep
  learning}.
\newblock \bibinfo{journal}{\emph{CoRR}}  \bibinfo{volume}{abs/1811.04017}
  (\bibinfo{year}{2018}).
\newblock
\showeprint[arXiv]{1811.04017}
\urldef\tempurl%
\url{http://arxiv.org/abs/1811.04017}
\showURL{%
\tempurl}


\bibitem[Sabt et~al\mbox{.}(2015)]%
        {sabt2015trusted}
\bibfield{author}{\bibinfo{person}{Mohamed Sabt}, \bibinfo{person}{Mohammed
  Achemlal}, {and} \bibinfo{person}{Abdelmadjid Bouabdallah}.}
  \bibinfo{year}{2015}\natexlab{}.
\newblock \showarticletitle{Trusted Execution Environment: What It is, and What
  It is Not}. In \bibinfo{booktitle}{\emph{2015 {IEEE} TrustCom/BigDataSE/ISPA,
  Helsinki, Finland, August 20-22, 2015, Volume 1}}.
  \bibinfo{publisher}{{IEEE}}, \bibinfo{pages}{57--64}.
\newblock
\urldef\tempurl%
\url{https://doi.org/10.1109/TRUSTCOM.2015.357}
\showDOI{\tempurl}


\bibitem[Sadiev et~al\mbox{.}(2022)]%
        {sadiev2022federated}
\bibfield{author}{\bibinfo{person}{Abdurakhmon Sadiev},
  \bibinfo{person}{Grigory Malinovsky}, \bibinfo{person}{Eduard Gorbunov},
  \bibinfo{person}{Igor Sokolov}, \bibinfo{person}{Ahmed Khaled},
  \bibinfo{person}{Konstantin Burlachenko}, {and} \bibinfo{person}{Peter
  Richt{\'{a}}rik}.} \bibinfo{year}{2022}\natexlab{}.
\newblock \showarticletitle{Federated Optimization Algorithms with Random
  Reshuffling and Gradient Compression}.
\newblock \bibinfo{journal}{\emph{CoRR}}  \bibinfo{volume}{abs/2206.07021}
  (\bibinfo{year}{2022}).
\newblock
\urldef\tempurl%
\url{https://doi.org/10.48550/ARXIV.2206.07021}
\showDOI{\tempurl}
\showeprint[arXiv]{2206.07021}


\bibitem[Sadiev et~al\mbox{.}(2024)]%
        {DBLP:conf/nips/SadievMG00BR24}
\bibfield{author}{\bibinfo{person}{Abdurakhmon Sadiev},
  \bibinfo{person}{Grigory Malinovsky}, \bibinfo{person}{Eduard Gorbunov},
  \bibinfo{person}{Igor Sokolov}, \bibinfo{person}{Ahmed Khaled},
  \bibinfo{person}{Konstantin Burlachenko}, {and} \bibinfo{person}{Peter
  Richt{\'{a}}rik}.} \bibinfo{year}{2024}\natexlab{}.
\newblock \showarticletitle{Don't Compress Gradients in Random Reshuffling:
  Compress Gradient Differences}. In \bibinfo{booktitle}{\emph{Advances in
  Neural Information Processing Systems 38: Annual Conference on Neural
  Information Processing Systems 2024, NeurIPS 2024, Vancouver, BC, Canada,
  December 10 - 15, 2024}}, \bibfield{editor}{\bibinfo{person}{Amir
  Globersons}, \bibinfo{person}{Lester Mackey}, \bibinfo{person}{Danielle
  Belgrave}, \bibinfo{person}{Angela Fan}, \bibinfo{person}{Ulrich Paquet},
  \bibinfo{person}{Jakub~M. Tomczak}, {and} \bibinfo{person}{Cheng Zhang}}
  (Eds.).
\newblock
\urldef\tempurl%
\url{http://papers.nips.cc/paper\_files/paper/2024/hash/99c80ceb10cb674110f03b2def6a5b76-Abstract-Conference.html}
\showURL{%
\tempurl}


\bibitem[Safaryan et~al\mbox{.}(2022)]%
        {safaryan2021fednl}
\bibfield{author}{\bibinfo{person}{Mher Safaryan}, \bibinfo{person}{Rustem
  Islamov}, \bibinfo{person}{Xun Qian}, {and} \bibinfo{person}{Peter
  Richt{\'{a}}rik}.} \bibinfo{year}{2022}\natexlab{}.
\newblock \showarticletitle{FedNL: Making Newton-Type Methods Applicable to
  Federated Learning}. In \bibinfo{booktitle}{\emph{International Conference on
  Machine Learning, {ICML} 2022, 17-23 July 2022, Baltimore, Maryland, {USA}}}
  \emph{(\bibinfo{series}{Proceedings of Machine Learning Research},
  Vol.~\bibinfo{volume}{162})}, \bibfield{editor}{\bibinfo{person}{Kamalika
  Chaudhuri}, \bibinfo{person}{Stefanie Jegelka}, \bibinfo{person}{Le~Song},
  \bibinfo{person}{Csaba Szepesv{\'{a}}ri}, \bibinfo{person}{Gang Niu}, {and}
  \bibinfo{person}{Sivan Sabato}} (Eds.). \bibinfo{publisher}{{PMLR}},
  \bibinfo{pages}{18959--19010}.
\newblock
\urldef\tempurl%
\url{https://proceedings.mlr.press/v162/safaryan22a.html}
\showURL{%
\tempurl}


\bibitem[Safaryan and Richt{\'{a}}rik(2021)]%
        {safaryan2019stochastic}
\bibfield{author}{\bibinfo{person}{Mher Safaryan} {and} \bibinfo{person}{Peter
  Richt{\'{a}}rik}.} \bibinfo{year}{2021}\natexlab{}.
\newblock \showarticletitle{Stochastic Sign Descent Methods: New Algorithms and
  Better Theory}.
\newblock   \bibinfo{volume}{139} (\bibinfo{year}{2021}),
  \bibinfo{pages}{9224--9234}.
\newblock
\urldef\tempurl%
\url{http://proceedings.mlr.press/v139/safaryan21a.html}
\showURL{%
\tempurl}


\bibitem[Safaryan et~al\mbox{.}(2020)]%
        {safaryan2021}
\bibfield{author}{\bibinfo{person}{Mher Safaryan}, \bibinfo{person}{Egor
  Shulgin}, {and} \bibinfo{person}{Peter Richt{\'{a}}rik}.}
  \bibinfo{year}{2020}\natexlab{}.
\newblock \showarticletitle{Uncertainty Principle for Communication Compression
  in Distributed and Federated Learning and the Search for an Optimal
  Compressor}.
\newblock \bibinfo{journal}{\emph{CoRR}}  \bibinfo{volume}{abs/2002.08958}
  (\bibinfo{year}{2020}).
\newblock
\showeprint[arXiv]{2002.08958}
\urldef\tempurl%
\url{https://arxiv.org/abs/2002.08958}
\showURL{%
\tempurl}


\bibitem[Sahu et~al\mbox{.}(2021)]%
        {sahu2021rethinking}
\bibfield{author}{\bibinfo{person}{Atal~Narayan Sahu}, \bibinfo{person}{Aritra
  Dutta}, \bibinfo{person}{Ahmed~M. Abdelmoniem}, \bibinfo{person}{Trambak
  Banerjee}, \bibinfo{person}{Marco Canini}, {and} \bibinfo{person}{Panos
  Kalnis}.} \bibinfo{year}{2021}\natexlab{}.
\newblock \showarticletitle{Rethinking gradient sparsification as total error
  minimization}.
\newblock  (\bibinfo{year}{2021}), \bibinfo{pages}{8133--8146}.
\newblock
\urldef\tempurl%
\url{https://proceedings.neurips.cc/paper/2021/hash/447b0408b80078338810051bb38b177f-Abstract.html}
\showURL{%
\tempurl}


\bibitem[Scardapane(2024)]%
        {scardapane2024alice}
\bibfield{author}{\bibinfo{person}{Simone Scardapane}.}
  \bibinfo{year}{2024}\natexlab{}.
\newblock \showarticletitle{Alice's Adventures in a Differentiable Wonderland -
  Volume I, {A} Tour of the Land}.
\newblock \bibinfo{journal}{\emph{CoRR}}  \bibinfo{volume}{abs/2404.17625}
  (\bibinfo{year}{2024}).
\newblock
\urldef\tempurl%
\url{https://doi.org/10.48550/ARXIV.2404.17625}
\showDOI{\tempurl}
\showeprint[arXiv]{2404.17625}


\bibitem[Science and on~Artificial~Intelligence(2023)]%
        {national2019national}
\bibfield{author}{\bibinfo{person}{National Science} {and}
  \bibinfo{person}{Technology Council (US). Select~Committee on
  Artificial~Intelligence}.} \bibinfo{year}{2023}\natexlab{}.
\newblock \bibinfo{booktitle}{\emph{National Artificial Intelligence Research
  and Development Strategic Plan 2023 Update}}.
\newblock \bibinfo{publisher}{National Science and Technology Council (US)}.
\newblock
\urldef\tempurl%
\url{https://www.nitrd.gov}
\showURL{%
\tempurl}


\bibitem[Seide et~al\mbox{.}(2014)]%
        {Seide2014}
\bibfield{author}{\bibinfo{person}{Frank Seide}, \bibinfo{person}{Hao Fu},
  \bibinfo{person}{Jasha Droppo}, \bibinfo{person}{Gang Li}, {and}
  \bibinfo{person}{Dong Yu}.} \bibinfo{year}{2014}\natexlab{}.
\newblock \showarticletitle{1-bit stochastic gradient descent and its
  application to data-parallel distributed training of speech DNNs}. In
  \bibinfo{booktitle}{\emph{15th Annual Conference of the International Speech
  Communication Association, {INTERSPEECH} 2014, Singapore, September 14-18,
  2014}}, \bibfield{editor}{\bibinfo{person}{Haizhou Li},
  \bibinfo{person}{Helen~M. Meng}, \bibinfo{person}{Bin Ma},
  \bibinfo{person}{Engsiong Chng}, {and} \bibinfo{person}{Lei Xie}} (Eds.).
  \bibinfo{publisher}{{ISCA}}, \bibinfo{pages}{1058--1062}.
\newblock
\urldef\tempurl%
\url{https://doi.org/10.21437/INTERSPEECH.2014-274}
\showDOI{\tempurl}


\bibitem[Shalev{-}Shwartz and Ben{-}David(2014)]%
        {shalev2014understanding}
\bibfield{author}{\bibinfo{person}{Shai Shalev{-}Shwartz} {and}
  \bibinfo{person}{Shai Ben{-}David}.} \bibinfo{year}{2014}\natexlab{}.
\newblock \bibinfo{booktitle}{\emph{Understanding Machine Learning - From
  Theory to Algorithms}}.
\newblock \bibinfo{publisher}{Cambridge University Press}.
\newblock
\showISBNx{978-1-10-705713-5}
\urldef\tempurl%
\url{http://www.cambridge.org/de/academic/subjects/computer-science/pattern-recognition-and-machine-learning/understanding-machine-learning-theory-algorithms}
\showURL{%
\tempurl}


\bibitem[Shamir et~al\mbox{.}(2014)]%
        {dane}
\bibfield{author}{\bibinfo{person}{Ohad Shamir}, \bibinfo{person}{Nathan
  Srebro}, {and} \bibinfo{person}{Tong Zhang}.}
  \bibinfo{year}{2014}\natexlab{}.
\newblock \showarticletitle{Communication-Efficient Distributed Optimization
  using an Approximate Newton-type Method}. In
  \bibinfo{booktitle}{\emph{Proceedings of the 31th International Conference on
  Machine Learning, {ICML} 2014, Beijing, China, 21-26 June 2014}}
  \emph{(\bibinfo{series}{{JMLR} Workshop and Conference Proceedings},
  Vol.~\bibinfo{volume}{32})}. \bibinfo{publisher}{JMLR.org},
  \bibinfo{pages}{1000--1008}.
\newblock
\urldef\tempurl%
\url{http://proceedings.mlr.press/v32/shamir14.html}
\showURL{%
\tempurl}


\bibitem[Shamsian et~al\mbox{.}(2021)]%
        {shamsian2021personalized}
\bibfield{author}{\bibinfo{person}{Aviv Shamsian}, \bibinfo{person}{Aviv
  Navon}, \bibinfo{person}{Ethan Fetaya}, {and} \bibinfo{person}{Gal Chechik}.}
  \bibinfo{year}{2021}\natexlab{}.
\newblock \showarticletitle{Personalized Federated Learning using
  Hypernetworks}. In \bibinfo{booktitle}{\emph{Proceedings of the 38th
  International Conference on Machine Learning, {ICML} 2021, 18-24 July 2021,
  Virtual Event}} \emph{(\bibinfo{series}{Proceedings of Machine Learning
  Research}, Vol.~\bibinfo{volume}{139})},
  \bibfield{editor}{\bibinfo{person}{Marina Meila} {and} \bibinfo{person}{Tong
  Zhang}} (Eds.). \bibinfo{publisher}{{PMLR}}, \bibinfo{pages}{9489--9502}.
\newblock
\urldef\tempurl%
\url{http://proceedings.mlr.press/v139/shamsian21a.html}
\showURL{%
\tempurl}


\bibitem[Shen et~al\mbox{.}(2022)]%
        {shen2022cd2}
\bibfield{author}{\bibinfo{person}{Yiqing Shen}, \bibinfo{person}{Yuyin Zhou},
  {and} \bibinfo{person}{Lequan Yu}.} \bibinfo{year}{2022}\natexlab{}.
\newblock \showarticletitle{CD\({}^{\mbox{2}}\)-pFed: Cyclic
  Distillation-guided Channel Decoupling for Model Personalization in Federated
  Learning}. In \bibinfo{booktitle}{\emph{{IEEE/CVF} Conference on Computer
  Vision and Pattern Recognition, {CVPR} 2022, New Orleans, LA, USA, June
  18-24, 2022}}. \bibinfo{publisher}{{IEEE}}, \bibinfo{pages}{10031--10040}.
\newblock
\urldef\tempurl%
\url{https://doi.org/10.1109/CVPR52688.2022.00980}
\showDOI{\tempurl}


\bibitem[Shlezinger et~al\mbox{.}(2020)]%
        {9054168}
\bibfield{author}{\bibinfo{person}{Nir Shlezinger}, \bibinfo{person}{Mingzhe
  Chen}, \bibinfo{person}{Yonina~C. Eldar}, \bibinfo{person}{H.~Vincent Poor},
  {and} \bibinfo{person}{Shuguang Cui}.} \bibinfo{year}{2020}\natexlab{}.
\newblock \showarticletitle{Federated Learning with Quantization Constraints}.
  In \bibinfo{booktitle}{\emph{2020 {IEEE} International Conference on
  Acoustics, Speech and Signal Processing, {ICASSP} 2020, Barcelona, Spain, May
  4-8, 2020}}. \bibinfo{publisher}{{IEEE}}, \bibinfo{pages}{8851--8855}.
\newblock
\urldef\tempurl%
\url{https://doi.org/10.1109/ICASSP40776.2020.9054168}
\showDOI{\tempurl}


\bibitem[Simonyan and Zisserman(2015)]%
        {vgg}
\bibfield{author}{\bibinfo{person}{Karen Simonyan} {and}
  \bibinfo{person}{Andrew Zisserman}.} \bibinfo{year}{2015}\natexlab{}.
\newblock \showarticletitle{Very Deep Convolutional Networks for Large-Scale
  Image Recognition}.
\newblock  (\bibinfo{year}{2015}).
\newblock
\urldef\tempurl%
\url{http://arxiv.org/abs/1409.1556}
\showURL{%
\tempurl}


\bibitem[Singel(2010)]%
        {singel2010netflix}
\bibfield{author}{\bibinfo{person}{Ryan Singel}.}
  \bibinfo{year}{2010}\natexlab{}.
\newblock \showarticletitle{Netflix cancels recommendation contest after
  privacy lawsuit}.
\newblock \bibinfo{journal}{\emph{Wired Magazine}} (\bibinfo{year}{2010}).
\newblock


\bibitem[Smith et~al\mbox{.}(2017)]%
        {smith2017federated}
\bibfield{author}{\bibinfo{person}{Virginia Smith}, \bibinfo{person}{Chao{-}Kai
  Chiang}, \bibinfo{person}{Maziar Sanjabi}, {and} \bibinfo{person}{Ameet
  Talwalkar}.} \bibinfo{year}{2017}\natexlab{}.
\newblock \showarticletitle{Federated Multi-Task Learning}. In
  \bibinfo{booktitle}{\emph{Advances in Neural Information Processing Systems
  30: Annual Conference on Neural Information Processing Systems 2017, December
  4-9, 2017, Long Beach, CA, {USA}}},
  \bibfield{editor}{\bibinfo{person}{Isabelle Guyon}, \bibinfo{person}{Ulrike
  von Luxburg}, \bibinfo{person}{Samy Bengio}, \bibinfo{person}{Hanna~M.
  Wallach}, \bibinfo{person}{Rob Fergus}, \bibinfo{person}{S.~V.~N.
  Vishwanathan}, {and} \bibinfo{person}{Roman Garnett}} (Eds.).
  \bibinfo{pages}{4424--4434}.
\newblock
\urldef\tempurl%
\url{https://proceedings.neurips.cc/paper/2017/hash/6211080fa89981f66b1a0c9d55c61d0f-Abstract.html}
\showURL{%
\tempurl}


\bibitem[Spurgeon(2000)]%
        {spurgeon2000ethernet}
\bibfield{author}{\bibinfo{person}{Charles~E. Spurgeon}.}
  \bibinfo{year}{2000}\natexlab{}.
\newblock \bibinfo{booktitle}{\emph{Ethernet - the definitive guide: designing
  and managing local area networks}}.
\newblock \bibinfo{publisher}{O'Reilly}.
\newblock
\showISBNx{978-1-56592-660-8}


\bibitem[Stich(2019)]%
        {stich2018local}
\bibfield{author}{\bibinfo{person}{Sebastian~U. Stich}.}
  \bibinfo{year}{2019}\natexlab{}.
\newblock \showarticletitle{Local {SGD} Converges Fast and Communicates
  Little}.
\newblock  (\bibinfo{year}{2019}).
\newblock
\urldef\tempurl%
\url{https://openreview.net/forum?id=S1g2JnRcFX}
\showURL{%
\tempurl}


\bibitem[Stich et~al\mbox{.}(2018)]%
        {Stich-EF-NIPS2018}
\bibfield{author}{\bibinfo{person}{Sebastian~U. Stich},
  \bibinfo{person}{Jean{-}Baptiste Cordonnier}, {and} \bibinfo{person}{Martin
  Jaggi}.} \bibinfo{year}{2018}\natexlab{}.
\newblock \showarticletitle{Sparsified {SGD} with Memory}. In
  \bibinfo{booktitle}{\emph{Advances in Neural Information Processing Systems
  31: Annual Conference on Neural Information Processing Systems 2018, NeurIPS
  2018, December 3-8, 2018, Montr{\'{e}}al, Canada}},
  \bibfield{editor}{\bibinfo{person}{Samy Bengio}, \bibinfo{person}{Hanna~M.
  Wallach}, \bibinfo{person}{Hugo Larochelle}, \bibinfo{person}{Kristen
  Grauman}, \bibinfo{person}{Nicol{\`{o}} Cesa{-}Bianchi}, {and}
  \bibinfo{person}{Roman Garnett}} (Eds.). \bibinfo{pages}{4452--4463}.
\newblock
\urldef\tempurl%
\url{https://proceedings.neurips.cc/paper/2018/hash/b440509a0106086a67bc2ea9df0a1dab-Abstract.html}
\showURL{%
\tempurl}


\bibitem[Strom(2015)]%
        {Strom15}
\bibfield{author}{\bibinfo{person}{Nikko Strom}.}
  \bibinfo{year}{2015}\natexlab{}.
\newblock \showarticletitle{Scalable distributed {DNN} training using commodity
  {GPU} cloud computing}. In \bibinfo{booktitle}{\emph{16th Annual Conference
  of the International Speech Communication Association, {INTERSPEECH} 2015,
  Dresden, Germany, September 6-10, 2015}}. \bibinfo{publisher}{{ISCA}},
  \bibinfo{pages}{1488--1492}.
\newblock
\urldef\tempurl%
\url{https://doi.org/10.21437/INTERSPEECH.2015-354}
\showDOI{\tempurl}


\bibitem[Stroustrup(1995)]%
        {stroustrup1994design}
\bibfield{author}{\bibinfo{person}{Bjarne Stroustrup}.}
  \bibinfo{year}{1995}\natexlab{}.
\newblock \bibinfo{booktitle}{\emph{The design and evolution of C++}}.
\newblock \bibinfo{publisher}{ACM Press/Addison-Wesley Publishing Co.},
  \bibinfo{address}{USA}.
\newblock
\showISBNx{0201543303}


\bibitem[Stroustrup(2013)]%
        {cpp}
\bibfield{author}{\bibinfo{person}{Bjarne Stroustrup}.}
  \bibinfo{year}{2013}\natexlab{}.
\newblock \bibinfo{booktitle}{\emph{The C++ Programming Language}
  (\bibinfo{edition}{4th} ed.)}.
\newblock \bibinfo{publisher}{Addison-Wesley Professional}.
\newblock
\showISBNx{0321563840}


\bibitem[Sun et~al\mbox{.}(2017)]%
        {sun2017deep}
\bibfield{author}{\bibinfo{person}{Yu Sun}, \bibinfo{person}{Yuan Liu},
  \bibinfo{person}{Guan Wang}, {and} \bibinfo{person}{Haiyan Zhang}.}
  \bibinfo{year}{2017}\natexlab{}.
\newblock \showarticletitle{Deep Learning for Plant Identification in Natural
  Environment}.
\newblock \bibinfo{journal}{\emph{Comput. Intell. Neurosci.}}
  \bibinfo{volume}{2017} (\bibinfo{year}{2017}),
  \bibinfo{pages}{7361042:1--7361042:6}.
\newblock
\urldef\tempurl%
\url{https://doi.org/10.1155/2017/7361042}
\showDOI{\tempurl}


\bibitem[Suresh et~al\mbox{.}(2017)]%
        {Suresh2017}
\bibfield{author}{\bibinfo{person}{Ananda~Theertha Suresh},
  \bibinfo{person}{Felix~X. Yu}, \bibinfo{person}{Sanjiv Kumar}, {and}
  \bibinfo{person}{H.~Brendan McMahan}.} \bibinfo{year}{2017}\natexlab{}.
\newblock \showarticletitle{Distributed Mean Estimation with Limited
  Communication}. In \bibinfo{booktitle}{\emph{Proceedings of the 34th
  International Conference on Machine Learning, {ICML} 2017, Sydney, NSW,
  Australia, 6-11 August 2017}} \emph{(\bibinfo{series}{Proceedings of Machine
  Learning Research}, Vol.~\bibinfo{volume}{70})},
  \bibfield{editor}{\bibinfo{person}{Doina Precup} {and}
  \bibinfo{person}{Yee~Whye Teh}} (Eds.). \bibinfo{publisher}{{PMLR}},
  \bibinfo{pages}{3329--3337}.
\newblock
\urldef\tempurl%
\url{http://proceedings.mlr.press/v70/suresh17a.html}
\showURL{%
\tempurl}


\bibitem[Szlendak et~al\mbox{.}(2022)]%
        {szlendak2021permutation}
\bibfield{author}{\bibinfo{person}{Rafal Szlendak}, \bibinfo{person}{Alexander
  Tyurin}, {and} \bibinfo{person}{Peter Richt{\'{a}}rik}.}
  \bibinfo{year}{2022}\natexlab{}.
\newblock \showarticletitle{Permutation Compressors for Provably Faster
  Distributed Nonconvex Optimization}. In \bibinfo{booktitle}{\emph{The Tenth
  International Conference on Learning Representations, {ICLR} 2022, Virtual
  Event, April 25-29, 2022}}. \bibinfo{publisher}{OpenReview.net}.
\newblock
\urldef\tempurl%
\url{https://openreview.net/forum?id=GugZ5DzzAu}
\showURL{%
\tempurl}


\bibitem[Tan et~al\mbox{.}(2023)]%
        {tan2022towards}
\bibfield{author}{\bibinfo{person}{Alysa~Ziying Tan}, \bibinfo{person}{Han Yu},
  \bibinfo{person}{Lizhen Cui}, {and} \bibinfo{person}{Qiang Yang}.}
  \bibinfo{year}{2023}\natexlab{}.
\newblock \showarticletitle{Towards Personalized Federated Learning}.
\newblock \bibinfo{journal}{\emph{{IEEE} Trans. Neural Networks Learn. Syst.}}
  \bibinfo{volume}{34}, \bibinfo{number}{12} (\bibinfo{year}{2023}),
  \bibinfo{pages}{9587--9603}.
\newblock
\urldef\tempurl%
\url{https://doi.org/10.1109/TNNLS.2022.3160699}
\showDOI{\tempurl}


\bibitem[Tang et~al\mbox{.}(2019)]%
        {DoubleSqueeze}
\bibfield{author}{\bibinfo{person}{Hanlin Tang}, \bibinfo{person}{Chen Yu},
  \bibinfo{person}{Xiangru Lian}, \bibinfo{person}{Tong Zhang}, {and}
  \bibinfo{person}{Ji Liu}.} \bibinfo{year}{2019}\natexlab{}.
\newblock \showarticletitle{DoubleSqueeze: Parallel Stochastic Gradient Descent
  with Double-pass Error-Compensated Compression}. In
  \bibinfo{booktitle}{\emph{Proceedings of the 36th International Conference on
  Machine Learning, {ICML} 2019, 9-15 June 2019, Long Beach, California,
  {USA}}} \emph{(\bibinfo{series}{Proceedings of Machine Learning Research},
  Vol.~\bibinfo{volume}{97})}, \bibfield{editor}{\bibinfo{person}{Kamalika
  Chaudhuri} {and} \bibinfo{person}{Ruslan Salakhutdinov}} (Eds.).
  \bibinfo{publisher}{{PMLR}}, \bibinfo{pages}{6155--6165}.
\newblock
\urldef\tempurl%
\url{http://proceedings.mlr.press/v97/tang19d.html}
\showURL{%
\tempurl}


\bibitem[Tarjan(1983)]%
        {tarjan1983data}
\bibfield{author}{\bibinfo{person}{Robert~Endre Tarjan}.}
  \bibinfo{year}{1983}\natexlab{}.
\newblock \bibinfo{booktitle}{\emph{Data structures and network algorithms}}.
  \bibinfo{series}{{CBMS-NSF} regional conference series in applied
  mathematics}, Vol.~\bibinfo{volume}{44}.
\newblock \bibinfo{publisher}{{SIAM}}.
\newblock
\showISBNx{978-0-89871-187-5}
\urldef\tempurl%
\url{https://doi.org/10.1137/1.9781611970265}
\showDOI{\tempurl}


\bibitem[Tyurin and Richt{\'{a}}rik(2023)]%
        {DASHA}
\bibfield{author}{\bibinfo{person}{Alexander Tyurin} {and}
  \bibinfo{person}{Peter Richt{\'{a}}rik}.} \bibinfo{year}{2023}\natexlab{}.
\newblock \showarticletitle{{DASHA:} Distributed Nonconvex Optimization with
  Communication Compression and Optimal Oracle Complexity}. In
  \bibinfo{booktitle}{\emph{The Eleventh International Conference on Learning
  Representations, {ICLR} 2023, Kigali, Rwanda, May 1-5, 2023}}.
  \bibinfo{publisher}{OpenReview.net}.
\newblock
\urldef\tempurl%
\url{https://openreview.net/forum?id=VA1YpcNr7ul}
\showURL{%
\tempurl}


\bibitem[Tyurin et~al\mbox{.}(2023)]%
        {tyurin2022sharper}
\bibfield{author}{\bibinfo{person}{Alexander Tyurin}, \bibinfo{person}{Lukang
  Sun}, \bibinfo{person}{Konstantin Burlachenko}, {and} \bibinfo{person}{Peter
  Richt{\'{a}}rik}.} \bibinfo{year}{2023}\natexlab{}.
\newblock \showarticletitle{Sharper Rates and Flexible Framework for Nonconvex
  {SGD} with Client and Data Sampling}.
\newblock \bibinfo{journal}{\emph{Trans. Mach. Learn. Res.}}
  \bibinfo{volume}{2023} (\bibinfo{year}{2023}).
\newblock
\urldef\tempurl%
\url{https://openreview.net/forum?id=zKgJ6TWAFE}
\showURL{%
\tempurl}


\bibitem[van Berkel(2009)]%
        {van2009multi}
\bibfield{author}{\bibinfo{person}{C.~H. van Berkel}.}
  \bibinfo{year}{2009}\natexlab{}.
\newblock \showarticletitle{Multi-core for mobile phones}. In
  \bibinfo{booktitle}{\emph{Design, Automation and Test in Europe, {DATE} 2009,
  Nice, France, April 20-24, 2009}}, \bibfield{editor}{\bibinfo{person}{Luca
  Benini}, \bibinfo{person}{Giovanni~De Micheli}, \bibinfo{person}{Bashir~M.
  Al{-}Hashimi}, {and} \bibinfo{person}{Wolfgang M{\"{u}}ller}} (Eds.).
  \bibinfo{publisher}{{IEEE}}, \bibinfo{pages}{1260--1265}.
\newblock
\urldef\tempurl%
\url{https://doi.org/10.1109/DATE.2009.5090858}
\showDOI{\tempurl}


\bibitem[van~der Walt et~al\mbox{.}(2011)]%
        {van2011numpy}
\bibfield{author}{\bibinfo{person}{St{\'{e}}fan van~der Walt},
  \bibinfo{person}{S.~Chris Colbert}, {and} \bibinfo{person}{Ga{\"{e}}l
  Varoquaux}.} \bibinfo{year}{2011}\natexlab{}.
\newblock \showarticletitle{The NumPy Array: {A} Structure for Efficient
  Numerical Computation}.
\newblock \bibinfo{journal}{\emph{Comput. Sci. Eng.}} \bibinfo{volume}{13},
  \bibinfo{number}{2} (\bibinfo{year}{2011}), \bibinfo{pages}{22--30}.
\newblock
\urldef\tempurl%
\url{https://doi.org/10.1109/MCSE.2011.37}
\showDOI{\tempurl}


\bibitem[Vaswani et~al\mbox{.}(2017)]%
        {transformer}
\bibfield{author}{\bibinfo{person}{Ashish Vaswani}, \bibinfo{person}{Noam
  Shazeer}, \bibinfo{person}{Niki Parmar}, \bibinfo{person}{Jakob Uszkoreit},
  \bibinfo{person}{Llion Jones}, \bibinfo{person}{Aidan~N. Gomez},
  \bibinfo{person}{Lukasz Kaiser}, {and} \bibinfo{person}{Illia Polosukhin}.}
  \bibinfo{year}{2017}\natexlab{}.
\newblock \showarticletitle{Attention is All you Need}. In
  \bibinfo{booktitle}{\emph{Advances in Neural Information Processing Systems
  30: Annual Conference on Neural Information Processing Systems 2017, December
  4-9, 2017, Long Beach, CA, {USA}}},
  \bibfield{editor}{\bibinfo{person}{Isabelle Guyon}, \bibinfo{person}{Ulrike
  von Luxburg}, \bibinfo{person}{Samy Bengio}, \bibinfo{person}{Hanna~M.
  Wallach}, \bibinfo{person}{Rob Fergus}, \bibinfo{person}{S.~V.~N.
  Vishwanathan}, {and} \bibinfo{person}{Roman Garnett}} (Eds.).
  \bibinfo{pages}{5998--6008}.
\newblock
\urldef\tempurl%
\url{https://proceedings.neurips.cc/paper/2017/hash/3f5ee243547dee91fbd053c1c4a845aa-Abstract.html}
\showURL{%
\tempurl}


\bibitem[Verbraeken et~al\mbox{.}(2021)]%
        {verbraeken2020survey}
\bibfield{author}{\bibinfo{person}{Joost Verbraeken}, \bibinfo{person}{Matthijs
  Wolting}, \bibinfo{person}{Jonathan Katzy}, \bibinfo{person}{Jeroen
  Kloppenburg}, \bibinfo{person}{Tim Verbelen}, {and} \bibinfo{person}{Jan~S.
  Rellermeyer}.} \bibinfo{year}{2021}\natexlab{}.
\newblock \showarticletitle{A Survey on Distributed Machine Learning}.
\newblock \bibinfo{journal}{\emph{{ACM} Comput. Surv.}} \bibinfo{volume}{53},
  \bibinfo{number}{2} (\bibinfo{year}{2021}), \bibinfo{pages}{30:1--30:33}.
\newblock
\urldef\tempurl%
\url{https://doi.org/10.1145/3377454}
\showDOI{\tempurl}


\bibitem[Vogels et~al\mbox{.}(2019)]%
        {PowerSGD}
\bibfield{author}{\bibinfo{person}{Thijs Vogels}, \bibinfo{person}{Sai~Praneeth
  Karimireddy}, {and} \bibinfo{person}{Martin Jaggi}.}
  \bibinfo{year}{2019}\natexlab{}.
\newblock \showarticletitle{PowerSGD: Practical Low-Rank Gradient Compression
  for Distributed Optimization}. In \bibinfo{booktitle}{\emph{Advances in
  Neural Information Processing Systems 32: Annual Conference on Neural
  Information Processing Systems 2019, NeurIPS 2019, December 8-14, 2019,
  Vancouver, BC, Canada}}, \bibfield{editor}{\bibinfo{person}{Hanna~M.
  Wallach}, \bibinfo{person}{Hugo Larochelle}, \bibinfo{person}{Alina
  Beygelzimer}, \bibinfo{person}{Florence d'Alch{\'{e}}{-}Buc},
  \bibinfo{person}{Emily~B. Fox}, {and} \bibinfo{person}{Roman Garnett}}
  (Eds.). \bibinfo{pages}{14236--14245}.
\newblock
\urldef\tempurl%
\url{https://proceedings.neurips.cc/paper/2019/hash/d9fbed9da256e344c1fa46bb46c34c5f-Abstract.html}
\showURL{%
\tempurl}


\bibitem[Walt et~al\mbox{.}(2011)]%
        {walt2011numpy}
\bibfield{author}{\bibinfo{person}{St{\'e}fan van~der Walt},
  \bibinfo{person}{S~Chris Colbert}, {and} \bibinfo{person}{Gael Varoquaux}.}
  \bibinfo{year}{2011}\natexlab{}.
\newblock \showarticletitle{The NumPy array: a structure for efficient
  numerical computation}.
\newblock \bibinfo{journal}{\emph{Computing in science and engineering}}
  \bibinfo{volume}{13}, \bibinfo{number}{2} (\bibinfo{year}{2011}),
  \bibinfo{pages}{22--30}.
\newblock


\bibitem[Wang et~al\mbox{.}(2018)]%
        {ATOMO}
\bibfield{author}{\bibinfo{person}{Hongyi Wang}, \bibinfo{person}{Scott
  Sievert}, \bibinfo{person}{Shengchao Liu}, \bibinfo{person}{Zachary Charles},
  \bibinfo{person}{Dimitris~S. Papailiopoulos}, {and}
  \bibinfo{person}{Stephen~J. Wright}.} \bibinfo{year}{2018}\natexlab{}.
\newblock \showarticletitle{{ATOMO:} Communication-efficient Learning via
  Atomic Sparsification}.
\newblock In \bibinfo{booktitle}{\emph{Advances in Neural Information
  Processing Systems 31: Annual Conference on Neural Information Processing
  Systems 2018, NeurIPS 2018, December 3-8, 2018, Montr{\'{e}}al, Canada}},
  \bibfield{editor}{\bibinfo{person}{Samy Bengio}, \bibinfo{person}{Hanna~M.
  Wallach}, \bibinfo{person}{Hugo Larochelle}, \bibinfo{person}{Kristen
  Grauman}, \bibinfo{person}{Nicol{\`{o}} Cesa{-}Bianchi}, {and}
  \bibinfo{person}{Roman Garnett}} (Eds.). \bibinfo{pages}{9872--9883}.
\newblock
\urldef\tempurl%
\url{https://proceedings.neurips.cc/paper/2018/hash/33b3214d792caf311e1f00fd22b392c5-Abstract.html}
\showURL{%
\tempurl}


\bibitem[Wang et~al\mbox{.}(2021)]%
        {FieldGuide2021}
\bibfield{author}{\bibinfo{person}{Jianyu Wang}, \bibinfo{person}{Zachary
  Charles}, \bibinfo{person}{Zheng Xu}, \bibinfo{person}{Gauri Joshi},
  \bibinfo{person}{H.~Brendan McMahan}, \bibinfo{person}{Blaise~Ag{\"{u}}era y
  Arcas}, \bibinfo{person}{Maruan Al{-}Shedivat}, \bibinfo{person}{Galen
  Andrew}, \bibinfo{person}{Salman Avestimehr}, \bibinfo{person}{Katharine
  Daly}, \bibinfo{person}{Deepesh Data}, \bibinfo{person}{Suhas~N. Diggavi},
  \bibinfo{person}{Hubert Eichner}, \bibinfo{person}{Advait Gadhikar},
  \bibinfo{person}{Zachary Garrett}, \bibinfo{person}{Antonious~M. Girgis},
  \bibinfo{person}{Filip Hanzely}, \bibinfo{person}{Andrew Hard},
  \bibinfo{person}{Chaoyang He}, \bibinfo{person}{Samuel Horv{\'{a}}th},
  \bibinfo{person}{Zhouyuan Huo}, \bibinfo{person}{Alex Ingerman},
  \bibinfo{person}{Martin Jaggi}, \bibinfo{person}{Tara Javidi},
  \bibinfo{person}{Peter Kairouz}, \bibinfo{person}{Satyen Kale},
  \bibinfo{person}{Sai~Praneeth Karimireddy}, \bibinfo{person}{Jakub
  Kone{\v{c}}n{\'y}}, \bibinfo{person}{Sanmi Koyejo}, \bibinfo{person}{Tian
  Li}, \bibinfo{person}{Luyang Liu}, \bibinfo{person}{Mehryar Mohri},
  \bibinfo{person}{Hang Qi}, \bibinfo{person}{Sashank~J. Reddi},
  \bibinfo{person}{Peter Richt{\'{a}}rik}, \bibinfo{person}{Karan Singhal},
  \bibinfo{person}{Virginia Smith}, \bibinfo{person}{Mahdi Soltanolkotabi},
  \bibinfo{person}{Weikang Song}, \bibinfo{person}{Ananda~Theertha Suresh},
  \bibinfo{person}{Sebastian~U. Stich}, \bibinfo{person}{Ameet Talwalkar},
  \bibinfo{person}{Hongyi Wang}, \bibinfo{person}{Blake~E. Woodworth},
  \bibinfo{person}{Shanshan Wu}, \bibinfo{person}{Felix~X. Yu},
  \bibinfo{person}{Honglin Yuan}, \bibinfo{person}{Manzil Zaheer},
  \bibinfo{person}{Mi Zhang}, \bibinfo{person}{Tong Zhang},
  \bibinfo{person}{Chunxiang Zheng}, \bibinfo{person}{Chen Zhu}, {and}
  \bibinfo{person}{Wennan Zhu}.} \bibinfo{year}{2021}\natexlab{}.
\newblock \showarticletitle{A Field Guide to Federated Optimization}.
\newblock \bibinfo{journal}{\emph{CoRR}}  \bibinfo{volume}{abs/2107.06917}
  (\bibinfo{year}{2021}).
\newblock
\showeprint[arXiv]{2107.06917}
\urldef\tempurl%
\url{https://arxiv.org/abs/2107.06917}
\showURL{%
\tempurl}


\bibitem[Wang and Joshi(2019)]%
        {wang2018adaptive}
\bibfield{author}{\bibinfo{person}{Jianyu Wang} {and} \bibinfo{person}{Gauri
  Joshi}.} \bibinfo{year}{2019}\natexlab{}.
\newblock \showarticletitle{Adaptive Communication Strategies to Achieve the
  Best Error-Runtime Trade-off in Local-Update {SGD}}. In
  \bibinfo{booktitle}{\emph{Proceedings of the Second Conference on Machine
  Learning and Systems, SysML 2019, Stanford, CA, USA, March 31 - April 2,
  2019}}, \bibfield{editor}{\bibinfo{person}{Ameet Talwalkar},
  \bibinfo{person}{Virginia Smith}, {and} \bibinfo{person}{Matei Zaharia}}
  (Eds.). \bibinfo{publisher}{mlsys.org}.
\newblock
\urldef\tempurl%
\url{https://proceedings.mlsys.org/paper\_files/paper/2019/hash/4a0151b47bd93c5de2a0b57831981a0d-Abstract.html}
\showURL{%
\tempurl}


\bibitem[Wang et~al\mbox{.}(2023b)]%
        {cocktailsgd}
\bibfield{author}{\bibinfo{person}{Jue Wang}, \bibinfo{person}{Yucheng Lu},
  \bibinfo{person}{Binhang Yuan}, \bibinfo{person}{Beidi Chen},
  \bibinfo{person}{Percy Liang}, \bibinfo{person}{Christopher~De Sa},
  \bibinfo{person}{Christopher R{\'{e}}}, {and} \bibinfo{person}{Ce Zhang}.}
  \bibinfo{year}{2023}\natexlab{b}.
\newblock \showarticletitle{CocktailSGD: Fine-tuning Foundation Models over
  500Mbps Networks}. In \bibinfo{booktitle}{\emph{International Conference on
  Machine Learning, {ICML} 2023, 23-29 July 2023, Honolulu, Hawaii, {USA}}}
  \emph{(\bibinfo{series}{Proceedings of Machine Learning Research},
  Vol.~\bibinfo{volume}{202})}, \bibfield{editor}{\bibinfo{person}{Andreas
  Krause}, \bibinfo{person}{Emma Brunskill}, \bibinfo{person}{Kyunghyun Cho},
  \bibinfo{person}{Barbara Engelhardt}, \bibinfo{person}{Sivan Sabato}, {and}
  \bibinfo{person}{Jonathan Scarlett}} (Eds.). \bibinfo{publisher}{{PMLR}},
  \bibinfo{pages}{36058--36076}.
\newblock
\urldef\tempurl%
\url{https://proceedings.mlr.press/v202/wang23t.html}
\showURL{%
\tempurl}


\bibitem[Wang and Kanwar(2019)]%
        {wang2019bfloat16}
\bibfield{author}{\bibinfo{person}{Shibo Wang} {and} \bibinfo{person}{Pankaj
  Kanwar}.} \bibinfo{year}{2019}\natexlab{}.
\newblock \showarticletitle{BFloat16: The secret to high performance on Cloud
  TPUs}.
\newblock \bibinfo{journal}{\emph{Google Cloud Blog}} \bibinfo{volume}{4},
  \bibinfo{number}{1} (\bibinfo{year}{2019}).
\newblock


\bibitem[Wang et~al\mbox{.}(2023a)]%
        {wang2023topoopt}
\bibfield{author}{\bibinfo{person}{Weiyang Wang}, \bibinfo{person}{Moein
  Khazraee}, \bibinfo{person}{Zhizhen Zhong}, \bibinfo{person}{Manya Ghobadi},
  \bibinfo{person}{Zhihao Jia}, \bibinfo{person}{Dheevatsa Mudigere},
  \bibinfo{person}{Ying Zhang}, {and} \bibinfo{person}{Anthony Kewitsch}.}
  \bibinfo{year}{2023}\natexlab{a}.
\newblock \showarticletitle{TopoOpt: Co-optimizing Network Topology and
  Parallelization Strategy for Distributed Training Jobs}. In
  \bibinfo{booktitle}{\emph{20th {USENIX} Symposium on Networked Systems Design
  and Implementation, {NSDI} 2023, Boston, MA, April 17-19, 2023}},
  \bibfield{editor}{\bibinfo{person}{Mahesh Balakrishnan} {and}
  \bibinfo{person}{Manya Ghobadi}} (Eds.). \bibinfo{publisher}{{USENIX}
  Association}, \bibinfo{pages}{739--767}.
\newblock
\urldef\tempurl%
\url{https://www.usenix.org/conference/nsdi23/presentation/wang-weiyang}
\showURL{%
\tempurl}


\bibitem[Wang et~al\mbox{.}(2019)]%
        {wang2019spiderboost}
\bibfield{author}{\bibinfo{person}{Zhe Wang}, \bibinfo{person}{Kaiyi Ji},
  \bibinfo{person}{Yi Zhou}, \bibinfo{person}{Yingbin Liang}, {and}
  \bibinfo{person}{Vahid Tarokh}.} \bibinfo{year}{2019}\natexlab{}.
\newblock \showarticletitle{SpiderBoost and Momentum: Faster Variance Reduction
  Algorithms}.
\newblock  (\bibinfo{year}{2019}), \bibinfo{pages}{2403--2413}.
\newblock
\urldef\tempurl%
\url{https://proceedings.neurips.cc/paper/2019/hash/512c5cad6c37edb98ae91c8a76c3a291-Abstract.html}
\showURL{%
\tempurl}


\bibitem[Wangni et~al\mbox{.}(2018)]%
        {wangni2018gradient}
\bibfield{author}{\bibinfo{person}{Jianqiao Wangni}, \bibinfo{person}{Jialei
  Wang}, \bibinfo{person}{Ji Liu}, {and} \bibinfo{person}{Tong Zhang}.}
  \bibinfo{year}{2018}\natexlab{}.
\newblock \showarticletitle{Gradient Sparsification for Communication-Efficient
  Distributed Optimization}.
\newblock  (\bibinfo{year}{2018}), \bibinfo{pages}{1306--1316}.
\newblock
\urldef\tempurl%
\url{https://proceedings.neurips.cc/paper/2018/hash/3328bdf9a4b9504b9398284244fe97c2-Abstract.html}
\showURL{%
\tempurl}


\bibitem[WeBank(2020)]%
        {webank2020}
\bibfield{author}{\bibinfo{person}{WeBank}.} \bibinfo{year}{2020}\natexlab{}.
\newblock \bibinfo{title}{Utilization of FATE in Anti Money Laundering Through
  Multiple Banks}.
\newblock
  \bibinfo{howpublished}{\url{https://www.fedai.org/cases/utilization-of-fate-in-anti-money-laundering-through-multiple-banks/}}.
\newblock


\bibitem[Wen et~al\mbox{.}(2017)]%
        {DBLP:conf/nips/WenXYWWCL17}
\bibfield{author}{\bibinfo{person}{Wei Wen}, \bibinfo{person}{Cong Xu},
  \bibinfo{person}{Feng Yan}, \bibinfo{person}{Chunpeng Wu},
  \bibinfo{person}{Yandan Wang}, \bibinfo{person}{Yiran Chen}, {and}
  \bibinfo{person}{Hai Li}.} \bibinfo{year}{2017}\natexlab{}.
\newblock \showarticletitle{TernGrad: Ternary Gradients to Reduce Communication
  in Distributed Deep Learning}. In \bibinfo{booktitle}{\emph{Advances in
  Neural Information Processing Systems 30: Annual Conference on Neural
  Information Processing Systems 2017, December 4-9, 2017, Long Beach, CA,
  {USA}}}, \bibfield{editor}{\bibinfo{person}{Isabelle Guyon},
  \bibinfo{person}{Ulrike von Luxburg}, \bibinfo{person}{Samy Bengio},
  \bibinfo{person}{Hanna~M. Wallach}, \bibinfo{person}{Rob Fergus},
  \bibinfo{person}{S.~V.~N. Vishwanathan}, {and} \bibinfo{person}{Roman
  Garnett}} (Eds.). \bibinfo{pages}{1509--1519}.
\newblock
\urldef\tempurl%
\url{https://proceedings.neurips.cc/paper/2017/hash/89fcd07f20b6785b92134bd6c1d0fa42-Abstract.html}
\showURL{%
\tempurl}


\bibitem[Wexelblat(2014)]%
        {wexelblat2014history}
\bibfield{author}{\bibinfo{person}{Richard~L Wexelblat}.}
  \bibinfo{year}{2014}\natexlab{}.
\newblock \bibinfo{booktitle}{\emph{History of programming languages}}.
\newblock \bibinfo{publisher}{Academic Press}.
\newblock


\bibitem[Wu et~al\mbox{.}(2018)]%
        {errorSGD}
\bibfield{author}{\bibinfo{person}{Jiaxiang Wu}, \bibinfo{person}{Weidong
  Huang}, \bibinfo{person}{Junzhou Huang}, {and} \bibinfo{person}{Tong Zhang}.}
  \bibinfo{year}{2018}\natexlab{}.
\newblock \showarticletitle{Error Compensated Quantized {SGD} and its
  Applications to Large-scale Distributed Optimization}. In
  \bibinfo{booktitle}{\emph{Proceedings of the 35th International Conference on
  Machine Learning, {ICML} 2018, Stockholmsm{\"{a}}ssan, Stockholm, Sweden,
  July 10-15, 2018}} \emph{(\bibinfo{series}{Proceedings of Machine Learning
  Research}, Vol.~\bibinfo{volume}{80})},
  \bibfield{editor}{\bibinfo{person}{Jennifer~G. Dy} {and}
  \bibinfo{person}{Andreas Krause}} (Eds.). \bibinfo{publisher}{{PMLR}},
  \bibinfo{pages}{5321--5329}.
\newblock
\urldef\tempurl%
\url{http://proceedings.mlr.press/v80/wu18d.html}
\showURL{%
\tempurl}


\bibitem[Wu et~al\mbox{.}(2024)]%
        {wu2024mirage}
\bibfield{author}{\bibinfo{person}{Mengdi Wu}, \bibinfo{person}{Xinhao Cheng},
  \bibinfo{person}{Oded Padon}, {and} \bibinfo{person}{Zhihao Jia}.}
  \bibinfo{year}{2024}\natexlab{}.
\newblock \bibinfo{title}{A Multi-Level Superoptimizer for Tensor Programs}.
\newblock
\newblock
\urldef\tempurl%
\url{https://doi.org/10.48550/ARXIV.2405.05751}
\showDOI{\tempurl}
\showeprint[arXiv]{2405.05751}


\bibitem[Wytock et~al\mbox{.}(2016)]%
        {wytock2016new}
\bibfield{author}{\bibinfo{person}{Matt Wytock}, \bibinfo{person}{Steven
  Diamond}, \bibinfo{person}{Felix Heide}, {and} \bibinfo{person}{Stephen~P.
  Boyd}.} \bibinfo{year}{2016}\natexlab{}.
\newblock \showarticletitle{A New Architecture for Optimization Modeling
  Frameworks}. In \bibinfo{booktitle}{\emph{6th Workshop on Python for
  High-Performance and Scientific Computing, PyHPC@SC 2016, Salt Lake, UT, USA,
  November 14, 2016}}, \bibfield{editor}{\bibinfo{person}{Andreas Schreiber},
  \bibinfo{person}{William Scullin}, \bibinfo{person}{William~F. Spotz}, {and}
  \bibinfo{person}{Andy~R. Terrel}} (Eds.). \bibinfo{publisher}{{IEEE}},
  \bibinfo{pages}{36--44}.
\newblock
\urldef\tempurl%
\url{https://doi.org/10.1109/PYHPC.2016.009}
\showDOI{\tempurl}


\bibitem[Xu et~al\mbox{.}(2021a)]%
        {xu2021grace}
\bibfield{author}{\bibinfo{person}{Hang Xu}, \bibinfo{person}{Chen{-}Yu Ho},
  \bibinfo{person}{Ahmed~M. Abdelmoniem}, \bibinfo{person}{Aritra Dutta},
  \bibinfo{person}{El~Houcine Bergou}, \bibinfo{person}{Konstantinos
  Karatsenidis}, \bibinfo{person}{Marco Canini}, {and} \bibinfo{person}{Panos
  Kalnis}.} \bibinfo{year}{2021}\natexlab{a}.
\newblock \showarticletitle{{GRACE:} {A} Compressed Communication Framework for
  Distributed Machine Learning}. In \bibinfo{booktitle}{\emph{41st {IEEE}
  International Conference on Distributed Computing Systems, {ICDCS} 2021,
  Washington DC, USA, July 7-10, 2021}}. \bibinfo{publisher}{{IEEE}},
  \bibinfo{pages}{561--572}.
\newblock
\urldef\tempurl%
\url{https://doi.org/10.1109/ICDCS51616.2021.00060}
\showDOI{\tempurl}


\bibitem[Xu et~al\mbox{.}(2021b)]%
        {kostopoulou2021deepreduce}
\bibfield{author}{\bibinfo{person}{Hang Xu}, \bibinfo{person}{Kelly
  Kostopoulou}, \bibinfo{person}{Aritra Dutta}, \bibinfo{person}{Xin Li},
  \bibinfo{person}{Alexandros Ntoulas}, {and} \bibinfo{person}{Panos Kalnis}.}
  \bibinfo{year}{2021}\natexlab{b}.
\newblock \showarticletitle{DeepReduce: {A} Sparse-tensor Communication
  Framework for Federated Deep Learning}.
\newblock  (\bibinfo{year}{2021}), \bibinfo{pages}{21150--21163}.
\newblock
\urldef\tempurl%
\url{https://proceedings.neurips.cc/paper/2021/hash/b0ab42fcb7133122b38521d13da7120b-Abstract.html}
\showURL{%
\tempurl}


\bibitem[Xu et~al\mbox{.}(2024)]%
        {2024fwdllm}
\bibfield{author}{\bibinfo{person}{Mengwei Xu}, \bibinfo{person}{Dongqi Cai},
  \bibinfo{person}{Yaozong Wu}, \bibinfo{person}{Xiang Li}, {and}
  \bibinfo{person}{Shangguang Wang}.} \bibinfo{year}{2024}\natexlab{}.
\newblock \showarticletitle{FwdLLM: efficient federated finetuning of large
  language models with perturbed inferences}. In
  \bibinfo{booktitle}{\emph{Proceedings of the 2024 USENIX Conference on Usenix
  Annual Technical Conference}} (Santa Clara, CA, USA)
  \emph{(\bibinfo{series}{USENIX ATC'24})}. \bibinfo{publisher}{USENIX
  Association}, \bibinfo{address}{USA}, Article \bibinfo{articleno}{36},
  \bibinfo{numpages}{18}~pages.
\newblock
\showISBNx{978-1-939133-41-0}


\bibitem[Yang et~al\mbox{.}(2019a)]%
        {yang2019federated}
\bibfield{author}{\bibinfo{person}{Qiang Yang}, \bibinfo{person}{Yang Liu},
  \bibinfo{person}{Tianjian Chen}, {and} \bibinfo{person}{Yongxin Tong}.}
  \bibinfo{year}{2019}\natexlab{a}.
\newblock \showarticletitle{Federated Machine Learning: Concept and
  Applications}.
\newblock \bibinfo{journal}{\emph{{ACM} Trans. Intell. Syst. Technol.}}
  \bibinfo{volume}{10}, \bibinfo{number}{2} (\bibinfo{year}{2019}),
  \bibinfo{pages}{12:1--12:19}.
\newblock
\urldef\tempurl%
\url{https://doi.org/10.1145/3298981}
\showDOI{\tempurl}


\bibitem[Yang et~al\mbox{.}(2019b)]%
        {YANG2019278}
\bibfield{author}{\bibinfo{person}{Tao Yang}, \bibinfo{person}{Xinlei Yi},
  \bibinfo{person}{Junfeng Wu}, \bibinfo{person}{Ye Yuan}, \bibinfo{person}{Di
  Wu}, \bibinfo{person}{Ziyang Meng}, \bibinfo{person}{Yiguang Hong},
  \bibinfo{person}{Hong Wang}, \bibinfo{person}{Zongli Lin}, {and}
  \bibinfo{person}{Karl~Henrik Johansson}.} \bibinfo{year}{2019}\natexlab{b}.
\newblock \showarticletitle{A survey of distributed optimization}.
\newblock \bibinfo{journal}{\emph{Annu. Rev. Control.}}  \bibinfo{volume}{47}
  (\bibinfo{year}{2019}), \bibinfo{pages}{278--305}.
\newblock
\urldef\tempurl%
\url{https://doi.org/10.1016/J.ARCONTROL.2019.05.006}
\showDOI{\tempurl}


\bibitem[Yu et~al\mbox{.}(2019a)]%
        {hao2018b}
\bibfield{author}{\bibinfo{person}{Hao Yu}, \bibinfo{person}{Rong Jin}, {and}
  \bibinfo{person}{Sen Yang}.} \bibinfo{year}{2019}\natexlab{a}.
\newblock \showarticletitle{On the Linear Speedup Analysis of Communication
  Efficient Momentum {SGD} for Distributed Non-Convex Optimization}. In
  \bibinfo{booktitle}{\emph{Proceedings of the 36th International Conference on
  Machine Learning, {ICML} 2019, 9-15 June 2019, Long Beach, California,
  {USA}}} \emph{(\bibinfo{series}{Proceedings of Machine Learning Research},
  Vol.~\bibinfo{volume}{97})}, \bibfield{editor}{\bibinfo{person}{Kamalika
  Chaudhuri} {and} \bibinfo{person}{Ruslan Salakhutdinov}} (Eds.).
  \bibinfo{publisher}{{PMLR}}, \bibinfo{pages}{7184--7193}.
\newblock
\urldef\tempurl%
\url{http://proceedings.mlr.press/v97/yu19d.html}
\showURL{%
\tempurl}


\bibitem[Yu et~al\mbox{.}(2019b)]%
        {Yu2019ExploringFA}
\bibfield{author}{\bibinfo{person}{Yue Yu}, \bibinfo{person}{Jiaxiang Wu},
  {and} \bibinfo{person}{Junzhou Huang}.} \bibinfo{year}{2019}\natexlab{b}.
\newblock \showarticletitle{Exploring Fast and Communication-Efficient
  Algorithms in Large-Scale Distributed Networks}. In
  \bibinfo{booktitle}{\emph{The 22nd International Conference on Artificial
  Intelligence and Statistics, {AISTATS} 2019, 16-18 April 2019, Naha, Okinawa,
  Japan}} \emph{(\bibinfo{series}{Proceedings of Machine Learning Research},
  Vol.~\bibinfo{volume}{89})}, \bibfield{editor}{\bibinfo{person}{Kamalika
  Chaudhuri} {and} \bibinfo{person}{Masashi Sugiyama}} (Eds.).
  \bibinfo{publisher}{{PMLR}}, \bibinfo{pages}{674--683}.
\newblock
\urldef\tempurl%
\url{http://proceedings.mlr.press/v89/yu19a.html}
\showURL{%
\tempurl}


\bibitem[Zagoruyko and Komodakis(2016)]%
        {wideresnet}
\bibfield{author}{\bibinfo{person}{Sergey Zagoruyko} {and}
  \bibinfo{person}{Nikos Komodakis}.} \bibinfo{year}{2016}\natexlab{}.
\newblock \showarticletitle{Wide Residual Networks}.
\newblock  (\bibinfo{year}{2016}).
\newblock
\urldef\tempurl%
\url{https://bmva-archive.org.uk/bmvc/2016/papers/paper087/index.html}
\showURL{%
\tempurl}


\bibitem[Zaharia et~al\mbox{.}(2016)]%
        {zaharia2016apache}
\bibfield{author}{\bibinfo{person}{Matei Zaharia}, \bibinfo{person}{Reynold~S.
  Xin}, \bibinfo{person}{Patrick Wendell}, \bibinfo{person}{Tathagata Das},
  \bibinfo{person}{Michael Armbrust}, \bibinfo{person}{Ankur Dave},
  \bibinfo{person}{Xiangrui Meng}, \bibinfo{person}{Josh Rosen},
  \bibinfo{person}{Shivaram Venkataraman}, \bibinfo{person}{Michael~J.
  Franklin}, \bibinfo{person}{Ali Ghodsi}, \bibinfo{person}{Joseph Gonzalez},
  \bibinfo{person}{Scott Shenker}, {and} \bibinfo{person}{Ion Stoica}.}
  \bibinfo{year}{2016}\natexlab{}.
\newblock \showarticletitle{Apache Spark: a unified engine for big data
  processing}.
\newblock \bibinfo{journal}{\emph{Commun. {ACM}}} \bibinfo{volume}{59},
  \bibinfo{number}{11} (\bibinfo{year}{2016}), \bibinfo{pages}{56--65}.
\newblock
\urldef\tempurl%
\url{https://doi.org/10.1145/2934664}
\showDOI{\tempurl}


\bibitem[Zhang et~al\mbox{.}(2020)]%
        {zhang2020batchcrypt}
\bibfield{author}{\bibinfo{person}{Chengliang Zhang}, \bibinfo{person}{Suyi
  Li}, \bibinfo{person}{Junzhe Xia}, \bibinfo{person}{Wei Wang},
  \bibinfo{person}{Feng Yan}, {and} \bibinfo{person}{Yang Liu}.}
  \bibinfo{year}{2020}\natexlab{}.
\newblock \showarticletitle{BatchCrypt: Efficient Homomorphic Encryption for
  Cross-Silo Federated Learning}. In \bibinfo{booktitle}{\emph{Proceedings of
  the 2020 {USENIX} Annual Technical Conference, {USENIX} {ATC} 2020, July
  15-17, 2020}}, \bibfield{editor}{\bibinfo{person}{Ada Gavrilovska} {and}
  \bibinfo{person}{Erez Zadok}} (Eds.). \bibinfo{publisher}{{USENIX}
  Association}, \bibinfo{pages}{493--506}.
\newblock
\urldef\tempurl%
\url{https://www.usenix.org/conference/atc20/presentation/zhang-chengliang}
\showURL{%
\tempurl}


\bibitem[Zhang et~al\mbox{.}(2017)]%
        {zipml}
\bibfield{author}{\bibinfo{person}{Hantian Zhang}, \bibinfo{person}{Jerry Li},
  \bibinfo{person}{Kaan Kara}, \bibinfo{person}{Dan Alistarh},
  \bibinfo{person}{Ji Liu}, {and} \bibinfo{person}{Ce Zhang}.}
  \bibinfo{year}{2017}\natexlab{}.
\newblock \showarticletitle{ZipML: Training Linear Models with End-to-End Low
  Precision, and a Little Bit of Deep Learning}. In
  \bibinfo{booktitle}{\emph{Proceedings of the 34th International Conference on
  Machine Learning, {ICML} 2017, Sydney, NSW, Australia, 6-11 August 2017}}
  \emph{(\bibinfo{series}{Proceedings of Machine Learning Research},
  Vol.~\bibinfo{volume}{70})}, \bibfield{editor}{\bibinfo{person}{Doina Precup}
  {and} \bibinfo{person}{Yee~Whye Teh}} (Eds.). \bibinfo{publisher}{{PMLR}},
  \bibinfo{pages}{4035--4043}.
\newblock
\urldef\tempurl%
\url{http://proceedings.mlr.press/v70/zhang17e.html}
\showURL{%
\tempurl}


\bibitem[Zhang et~al\mbox{.}(2021)]%
        {zhang2021personalized}
\bibfield{author}{\bibinfo{person}{Michael Zhang}, \bibinfo{person}{Karan
  Sapra}, \bibinfo{person}{Sanja Fidler}, \bibinfo{person}{Serena Yeung}, {and}
  \bibinfo{person}{Jos{\'{e}}~M. {\'{A}}lvarez}.}
  \bibinfo{year}{2021}\natexlab{}.
\newblock \showarticletitle{Personalized Federated Learning with First Order
  Model Optimization}. In \bibinfo{booktitle}{\emph{9th International
  Conference on Learning Representations, {ICLR} 2021, Virtual Event, Austria,
  May 3-7, 2021}}. \bibinfo{publisher}{OpenReview.net}.
\newblock
\urldef\tempurl%
\url{https://openreview.net/forum?id=ehJqJQk9cw}
\showURL{%
\tempurl}


\bibitem[Zhang et~al\mbox{.}(2015)]%
        {esgd}
\bibfield{author}{\bibinfo{person}{Sixin Zhang}, \bibinfo{person}{Anna
  Choromanska}, {and} \bibinfo{person}{Yann LeCun}.}
  \bibinfo{year}{2015}\natexlab{}.
\newblock \showarticletitle{Deep learning with Elastic Averaging {SGD}}. In
  \bibinfo{booktitle}{\emph{Advances in Neural Information Processing Systems
  28: Annual Conference on Neural Information Processing Systems 2015, December
  7-12, 2015, Montreal, Quebec, Canada}},
  \bibfield{editor}{\bibinfo{person}{Corinna Cortes}, \bibinfo{person}{Neil~D.
  Lawrence}, \bibinfo{person}{Daniel~D. Lee}, \bibinfo{person}{Masashi
  Sugiyama}, {and} \bibinfo{person}{Roman Garnett}} (Eds.).
  \bibinfo{pages}{685--693}.
\newblock
\urldef\tempurl%
\url{https://proceedings.neurips.cc/paper/2015/hash/d18f655c3fce66ca401d5f38b48c89af-Abstract.html}
\showURL{%
\tempurl}


\bibitem[Zhao et~al\mbox{.}(2019)]%
        {zhao2019secure}
\bibfield{author}{\bibinfo{person}{Chuan Zhao}, \bibinfo{person}{Shengnan
  Zhao}, \bibinfo{person}{Minghao Zhao}, \bibinfo{person}{Zhenxiang Chen},
  \bibinfo{person}{Chong{-}Zhi Gao}, \bibinfo{person}{Hongwei Li}, {and}
  \bibinfo{person}{Yu{-}an Tan}.} \bibinfo{year}{2019}\natexlab{}.
\newblock \showarticletitle{Secure Multi-Party Computation: Theory, practice
  and applications}.
\newblock \bibinfo{journal}{\emph{Inf. Sci.}}  \bibinfo{volume}{476}
  (\bibinfo{year}{2019}), \bibinfo{pages}{357--372}.
\newblock
\urldef\tempurl%
\url{https://doi.org/10.1016/J.INS.2018.10.024}
\showDOI{\tempurl}


\bibitem[Zhao et~al\mbox{.}(2024a)]%
        {zhao2024faster}
\bibfield{author}{\bibinfo{person}{Haoyu Zhao}, \bibinfo{person}{Konstantin
  Burlachenko}, \bibinfo{person}{Zhize Li}, {and} \bibinfo{person}{Peter
  Richt{\'{a}}rik}.} \bibinfo{year}{2024}\natexlab{a}.
\newblock \showarticletitle{Faster Rates for Compressed Federated Learning with
  Client-Variance Reduction}.
\newblock \bibinfo{journal}{\emph{{SIAM} J. Math. Data Sci.}}
  \bibinfo{volume}{6}, \bibinfo{number}{1} (\bibinfo{year}{2024}),
  \bibinfo{pages}{154--175}.
\newblock
\urldef\tempurl%
\url{https://doi.org/10.1137/23M1553820}
\showDOI{\tempurl}


\bibitem[Zhao et~al\mbox{.}(2024b)]%
        {zhao2021faster}
\bibfield{author}{\bibinfo{person}{Haoyu Zhao}, \bibinfo{person}{Konstantin
  Burlachenko}, \bibinfo{person}{Zhize Li}, {and} \bibinfo{person}{Peter
  Richt{\'{a}}rik}.} \bibinfo{year}{2024}\natexlab{b}.
\newblock \showarticletitle{Faster Rates for Compressed Federated Learning with
  Client-Variance Reduction}.
\newblock \bibinfo{journal}{\emph{{SIAM} J. Math. Data Sci.}}
  \bibinfo{volume}{6}, \bibinfo{number}{1} (\bibinfo{year}{2024}),
  \bibinfo{pages}{154--175}.
\newblock
\urldef\tempurl%
\url{https://doi.org/10.1137/23M1553820}
\showDOI{\tempurl}


\bibitem[Zhao and Zhang(2014)]%
        {zhao2014accelerating}
\bibfield{author}{\bibinfo{person}{Peilin Zhao} {and} \bibinfo{person}{Tong
  Zhang}.} \bibinfo{year}{2014}\natexlab{}.
\newblock \showarticletitle{Accelerating Minibatch Stochastic Gradient Descent
  using Stratified Sampling}.
\newblock \bibinfo{journal}{\emph{CoRR}}  \bibinfo{volume}{abs/1405.3080}
  (\bibinfo{year}{2014}).
\newblock
\showeprint[arXiv]{1405.3080}
\urldef\tempurl%
\url{http://arxiv.org/abs/1405.3080}
\showURL{%
\tempurl}


\bibitem[Zhao and Zhang(2015)]%
        {so_importance_sampling}
\bibfield{author}{\bibinfo{person}{Peilin Zhao} {and} \bibinfo{person}{Tong
  Zhang}.} \bibinfo{year}{2015}\natexlab{}.
\newblock \showarticletitle{Stochastic Optimization with Importance Sampling
  for Regularized Loss Minimization}. In \bibinfo{booktitle}{\emph{Proceedings
  of the 32nd International Conference on Machine Learning, {ICML} 2015, Lille,
  France, 6-11 July 2015}} \emph{(\bibinfo{series}{{JMLR} Workshop and
  Conference Proceedings}, Vol.~\bibinfo{volume}{37})},
  \bibfield{editor}{\bibinfo{person}{Francis~R. Bach} {and}
  \bibinfo{person}{David~M. Blei}} (Eds.). \bibinfo{publisher}{JMLR.org},
  \bibinfo{pages}{1--9}.
\newblock
\urldef\tempurl%
\url{http://proceedings.mlr.press/v37/zhaoa15.html}
\showURL{%
\tempurl}


\bibitem[Zhou et~al\mbox{.}(2020)]%
        {zhou2018stochastic}
\bibfield{author}{\bibinfo{person}{Dongruo Zhou}, \bibinfo{person}{Pan Xu},
  {and} \bibinfo{person}{Quanquan Gu}.} \bibinfo{year}{2020}\natexlab{}.
\newblock \showarticletitle{Stochastic Nested Variance Reduction for Nonconvex
  Optimization}.
\newblock \bibinfo{journal}{\emph{J. Mach. Learn. Res.}}  \bibinfo{volume}{21}
  (\bibinfo{year}{2020}), \bibinfo{pages}{103:1--103:63}.
\newblock
\urldef\tempurl%
\url{https://jmlr.org/papers/v21/18-447.html}
\showURL{%
\tempurl}


\bibitem[Zhou and Cong(2018)]%
        {zhou2018convergence}
\bibfield{author}{\bibinfo{person}{Fan Zhou} {and} \bibinfo{person}{Guojing
  Cong}.} \bibinfo{year}{2018}\natexlab{}.
\newblock \showarticletitle{On the Convergence Properties of a K-step Averaging
  Stochastic Gradient Descent Algorithm for Nonconvex Optimization}. In
  \bibinfo{booktitle}{\emph{Proceedings of the Twenty-Seventh International
  Joint Conference on Artificial Intelligence, {IJCAI} 2018, July 13-19, 2018,
  Stockholm, Sweden}}, \bibfield{editor}{\bibinfo{person}{J{\'{e}}r{\^{o}}me
  Lang}} (Ed.). \bibinfo{publisher}{ijcai.org}, \bibinfo{pages}{3219--3227}.
\newblock
\urldef\tempurl%
\url{https://doi.org/10.24963/IJCAI.2018/447}
\showDOI{\tempurl}


\end{thebibliography}
\end{onehalfspacing}

\phantomsection
\addcontentsline{toc}{chapter}{Thesis Appendices} 
\appendix
\renewcommand{\thesection}{\Alph{section}}


\refstepcounter{chapter}

\section{Complete List of Publications Authored During PhD}
\label{thesis:app:all-papers}


Below, we list all \textit{thirteen papers} authored by the writer of this thesis in collaboration with co-authors. This list includes both the \textit{seven research papers} included in the thesis, as detailed in Section~\ref{thesis:sec:focus}, and the \textit{six research papers} excluded from the thesis, as outlined in Section~\ref{thesis:sec:excluded-papers}. 

The inclusion of the \textit{seven research papers} in this thesis has been discussed and approved by all co-authors of the respective papers.

The complete list of publications authored during my PhD is as follows:
\\
\\
(1) \citet{houcine2022personalized} is part of this thesis \cmark.\\
El Houcine Bergou, Konstantin Burlachenko, Aritra Dutta, Peter Richt{\'a}rik.\\
Personalized Federated Learning with Communication Compression.\\
\textit{Transactions on Machine Learning Research, TMLR 2023}.\\
\\
(2) \citet{gorbunov2021marina} is not part of this thesis \xmark.\\
Eduard Gorbunov, Konstantin Burlachenko, Zhize Li, Peter Richt{\'a}rik.\\
MARINA: Faster Non-Convex Distributed Learning with Compression.\\
In \textit{Proceedings of the 38th International Conference on Machine Learning, ICML 2021.} \\
\\
(3) \citet{burlachenko2021fl_pytorch} is part of this thesis \cmark.\\
Konstantin Burlachenko, Samuel Horv{\'a}th, Peter Richt{\'a}rik.\\
Fl\_PyTorch: optimization research simulator for Federated Learning.\\
In \textit{Proceedings of the 2nd ACM International Workshop on Distributed Machine Learning, 2021}.\\
\\
(4) \citet{zhao2024faster} is not part of this thesis \xmark.\\
Haoyu Zhao, Konstantin Burlachenko, Zhize Li, Peter Richt{\'a}rik.\\
Faster Rates for Compressed Federated Learning with Client-Variance Reduction.\\
In \textit{Proceedings SIAM Journal on Mathematics of Data Science. SIMODS, 2024}.\\
\\
(5) \citet{DBLP:conf/nips/SadievMG00BR24, sadiev2022federated} is not part of this thesis \xmark.\\
Abdurakhmon Sadiev, Grigory Malinovsky, Eduard Gorbunov, Igor Sokolov, Ahmed Khaled, Konstantin Pavlovich Burlachenko, Peter Richt{\'a}rik. \\
Don't Compress Gradients in Random Reshuffling: Compress Gradient Differences.\\
In \textit{Proceedings of the Advances in Neural Information Processing Systems 37, NeurIPS 2024}.\\
\\
(6) \citet{tyurin2022sharper} is part of this thesis \cmark. \\
Alexander Tyurin, Lukang Sun, Konstantin Burlachenko, Peter Richt{\'a}rik.\\
Sharper rates and flexible framework for nonconvex SGD with client and data sampling.\\
\textit{Transactions on Machine Learning Research, TMLR 2023}.\\
\\
(7) \citet{malinovsky2023federated} is not part of this thesis \xmark. \\
Grigory Malinovsky, Samuel Horv{\'a}th, Konstantin Burlachenko, Peter Richt{\'a}rik.\\
Federated Learning with Regularized Client Participation.\\
\textit{arXiv: 2302.03662, 2023.}\\
\\
(8) \citet{richtarik2023error} is not part of this thesis \xmark. \\
Peter Richt{\'a}rik, Elnur Gasanov, Konstantin Burlachenko.\\
Error Feedback Shines when Features are Rare.\\
\textit{arXiv: 2305.15264, 2023.}\\
\\
(9) \citet{burlachenko2023federated} is part of this thesis \cmark. \\
Konstantin Burlachenko, Abdulmajeed Alrowithi, Fahad Ali Albalawi, Peter Richt{\'a}rik.\\
Federated Learning is Better with Non-Homomorphic Encryption.\\
In \textit{Proceedings of the 4th International Workshop on Distributed Machine Learning, 2023}.\\
\\
(10) \citet{richtarik2024error} is part of this thesis \cmark. \\
Peter Richt{\'a}rik, Elnur Gasanov, Konstantin Burlachenko.\\
Error Feedback Reloaded: From Quadratic to Arithmetic Mean \\
of Smoothness Constants.\\
In \textit{Proceedings of the 12th International Conference on Learning Representations, 2024}.\\
\\
(11) \citet{burlachenko2024unlocking} is part of this thesis \cmark. \\
Konstantin Burlachenko, Peter Richt{\'a}rik.\\
Unlocking FedNL: Self-Contained Compute-Optimized Implementation.\\
\textit{arXiv: 2410.08760, 2024}.\\
\\
(12) \citet{malinovskii2024pv} is not part of this thesis \xmark. \\
Vladimir Malinovskii, Denis Mazur, Ivan Ilin, Denis Kuznedelev, Konstantin Burlachenko, Kai Yi, Dan Alistarh, Peter Richt{\'a}rik.\\
PV-Tuning: Beyond Straight-Through Estimation \\
for Extreme LLM Compression.\\
In \textit{Proceedings of the Advances in Neural Information Processing Systems 37, NeurIPS 2024}.\\
\\
(13) Konstantin Burlachenko, Peter Richt{\'a}rik is part of this thesis \cmark.\\
BurTorch: Revisiting Training from First Principles by Coupling Autodiff, Math Optimization, and Systems.\\
\textit{arXiv: 2503.13795, 2015}.\\

\unappendix

\end{document}